\documentclass[twoside,11pt]{article}

\usepackage{blindtext}
\usepackage{lastpage}
\usepackage[morefloats=100]{morefloats}
\usepackage{multirow}
\usepackage{amsmath,amsfonts}
\usepackage[title]{appendix}
\usepackage{xcolor}
\usepackage{textcomp}
\usepackage{manyfoot}
\usepackage{booktabs}
\usepackage{algorithm}
\usepackage{algorithmicx}
\usepackage{algpseudocode}
\usepackage{listings}
\usepackage{pgffor}
\usepackage{caption}
\usepackage{placeins}
\usepackage{xfrac}

\usepackage{jmlr2e}

\definecolor{lightblue}{RGB}{100, 100, 255}

\hypersetup{
    colorlinks=true,
    linkcolor=blue,
    citecolor=brown,
    urlcolor=lightblue
}

\jmlrheading{}{2026}{1-\pageref{LastPage}}{20/06/2026}{\textbf{UNDER REVIEW}}{}{Riccardo Poli and Ahmet Yilmaz}
\firstpageno{1}

\begin{document}

\title{Gradient-Descent Steps to Success over Mean Accuracy:\\ A Paradigm Shift for ML}

\author{\name Riccardo Poli\email rpoli@essex.ac.uk\\
       \addr School  of Computer Science and Electronic Engineering\\
       University of Essex\\
       Wivenhoe Park, Colchester, CO4 3SQ, UK\\
       \AND
       \name Ahmet Yilmaz \email yilmazahmet@kmu.edu.tr\\
       \addr Department of Computer Engineering\\
       Karamanoglu Mehmetbey University\\
       Karaman, 70100, Turkey}

     \editor{Under Review}
     
\maketitle

\begin{abstract}
  Traditional evaluation of machine learning (ML) models typically
  focuses on achieving the maximum possible accuracy irrespective of
  the computational cost. In this article, we propose a paradigm shift
  towards evaluating performance based on computational
  effort—explicitly defined here as the total number of gradient
  descent steps required to reach an acceptable level of accuracy with
  high probability. Building upon the concept of computational effort
  originally introduced by Koza for Genetic Programming, we extend
  this metric to any ML model trained via gradient
  descent. Furthermore, we demonstrate that minimising this effort
  acts as a novel form of Automatic Machine Learning (AutoML). By evaluating it
  across 11 diverse ML models and five standard classification
  datasets, we uncover significant insights into the dynamics of
  gradient-based learning. Our findings reveal that optimal
  hyper-parameters consistently favour unusually large learning
  rates. Crucially, we demonstrate that the rapid, aggressive
  landscape traversal enabled by these large rates not only promotes
  generalisation—as seen in phenomena like superconvergence—but also
  statistically minimises the expected computational effort for
  training. Furthermore, we identify distinct phase transitions in the
  optimal search strategy: while a single training run suffices for
  lower accuracy targets, reaching a model's performance limit
  requires a dramatic shift towards conducting numerous independent,
  short restarts. Finally, we illustrate how this effort-based
  paradigm provides a robust framework for model selection, allowing
  practitioners to choose optimal algorithms based on the difficulty
  of a problem as perceived by different models for a given target
  accuracy, or to maximise the achievable accuracy for a fixed budget
  of gradient descent steps.
\end{abstract}
\vfill

\pagebreak

\section{Paradigm Shift}
\label{sec:paradigm-shift}

\begin{figure}
  \centering
  \includegraphics[width=.9\linewidth]{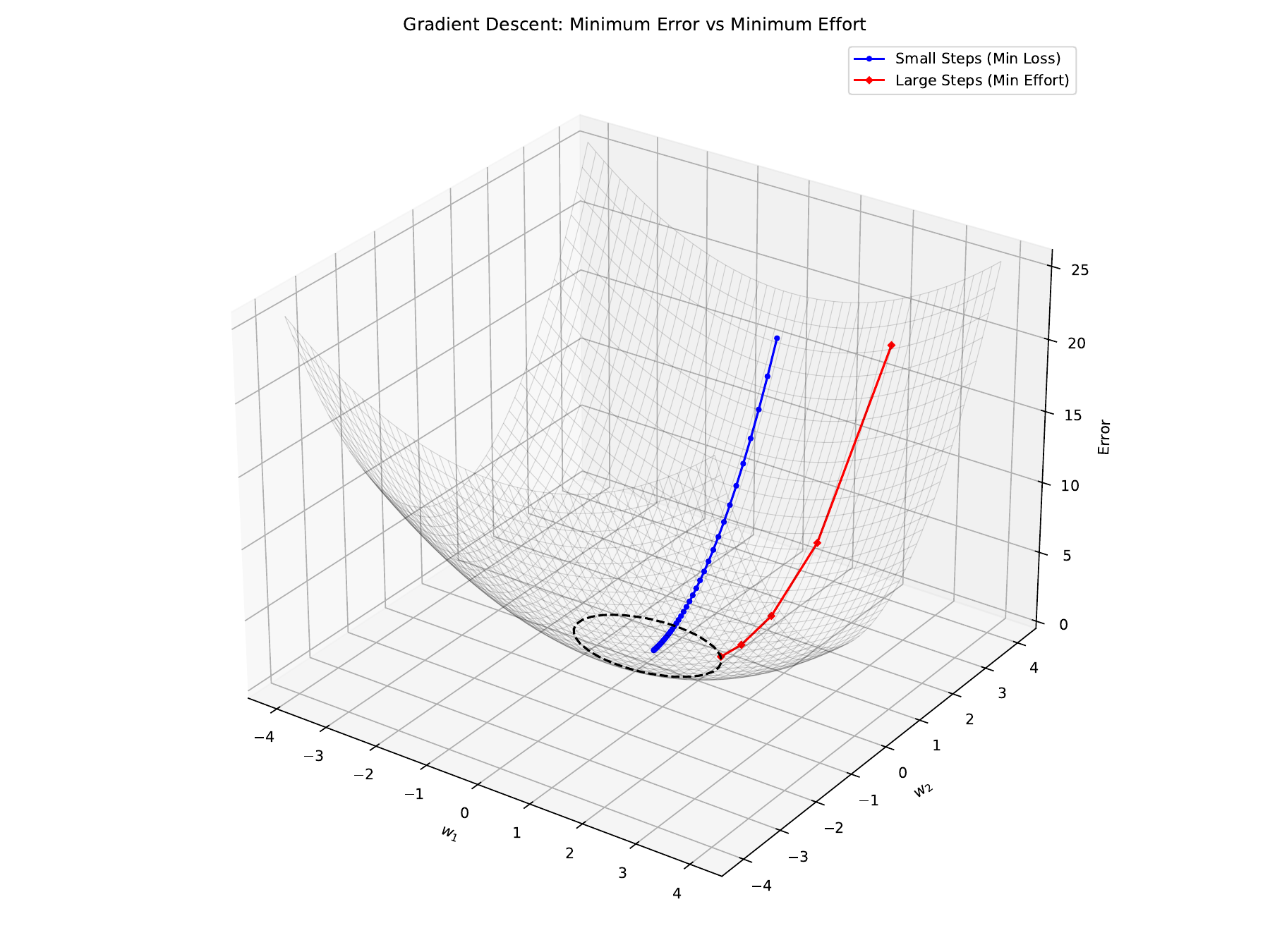}\\
  \includegraphics[width=.9\linewidth]{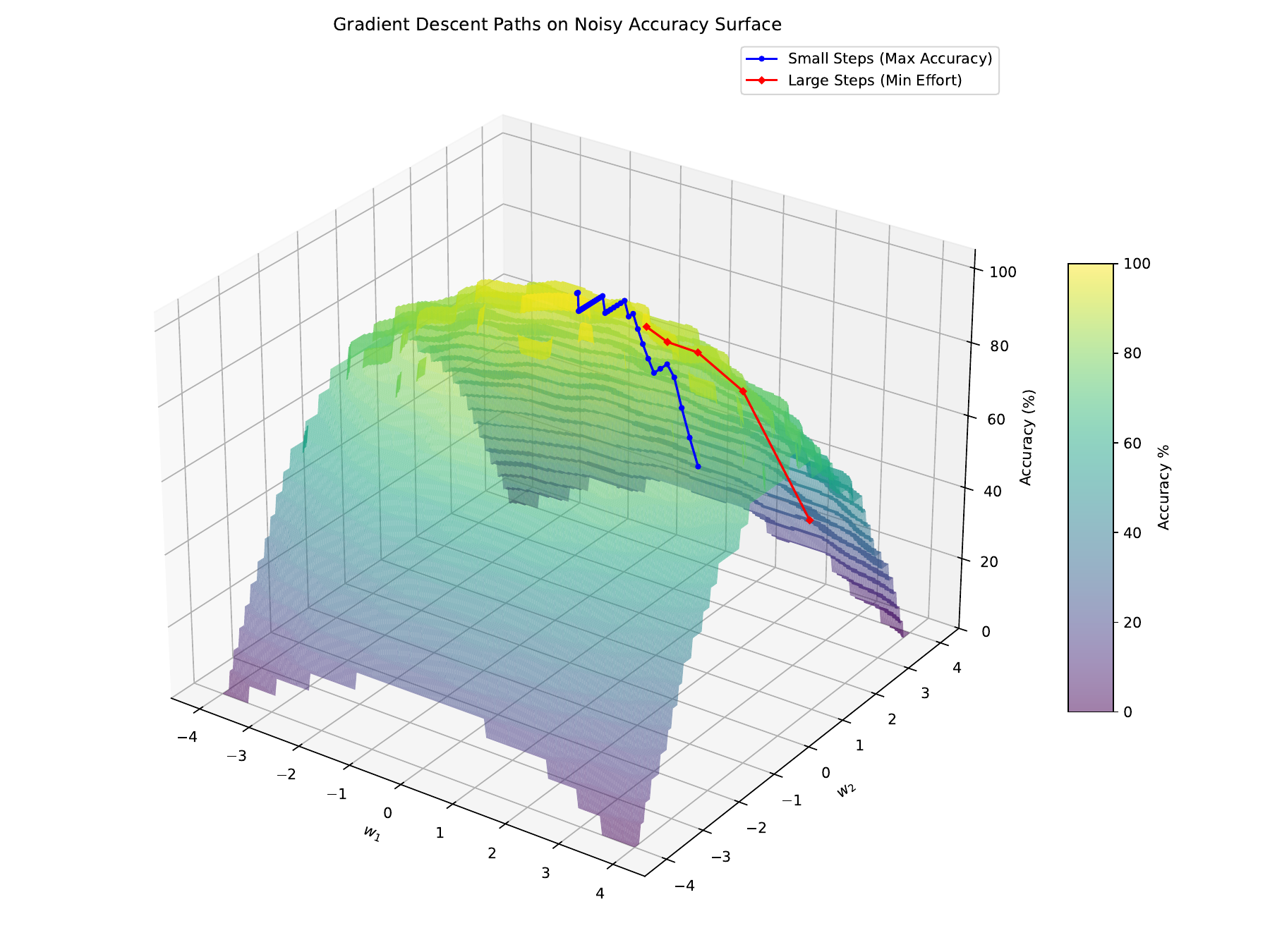}\\
  \caption{Paradigm shift proposed in this paper (see Section~\ref{sec:paradigm-shift} for explanations).}
  \label{fig:gradient_descent_comparision}
\end{figure}

This article advocates for a paradigm shift in Machine Learning (ML)
evaluation by proposing that performance be measured through the
number of gradient-descent steps required to reach an acceptable level
of performance (e.g., in terms of loss or accuracy) rather than
``whatever-the-cost'' maximum performance.  The key idea is
illustrated in Figure~\ref{fig:gradient_descent_comparision}.

In the figure the blue and red lines in the top panel represent
gradient descents on the \emph{loss surface} both starting at
approximately the same distance from the optimum. Traditionally, ML
focuses on reaching the minimum error irrespective of cost, which can
be achieved with small learning rates and many gradient-descent steps
(blue line).  In contrast, we propose that one should \emph{first
  decide on an acceptable \underline{error} level} (dashed ellipse) and
then focus on achieving such a level with the \emph{minimum number of
  gradient-descent steps} (red line), which typically require large
learning rates. Both descents reach the acceptable error level, but
the blue one requires 20 steps, while the red one only 4.

The coloured surface in the bottom panel represents the accuracy
surface associated to the error surface in the top panel, with the red
and blue lines indicating the accuracies obtained during the
corresponding gradient descents. Nothing prevents one from \emph{first
  deciding on an acceptable \underline{accuracy} level} and then
focusing on achieving such a level with the \emph{minimum number of
  gradient-descent steps}. In fact, in the article we will focus on
this second approach, prioritising
accuracies over losses, as practitioners are more interested in the
former than in the latter.
  
\section{Background}
\label{sec:background}

In  this section we provide the necessary background to understand the
work.

More specifically, in Section~\ref{sec:recent-improvements}, we
briefly review Automatic Machine Learning (AutoML), an area of ML
focusing on developing methods to automate the selection of the
elements of an ML pipeline using some form of search. We do this
because our paradigm shift allows us to introduce a new form of AutoML
based on the minimisation of the number of gradient-descent steps
required to reach an acceptable level of performance.

Then in Section~\ref{sec:comp-effort-to-success}, we review in depth
the notion of \emph{computational effort} --- the number of fitness
evaluations required to obtain a solution with 99\% probability in one
or multiple runs of Genetic Programming (GP)~\citep{koza92,langdon2002foundations,Poli2008fieldguide} --- which was introduced
by~\citet{koza92} and provided inspiration for this work. The
minimisation of such a quantity resulted in the \emph{minimum
  computational effort}, which was  used to compare problem
difficulty for a given GP system as well as the performance of
different GP systems across problems  both by Koza and other
researchers~\citep{koza92,koza1994genetic,poli1996some,poli1997evolution,koza1999gp3,miller1999empirical,koza2003gp4,miller2006redundancy,castle2012evolving,forstenlechner2018towards,rosenfeld2025survey}.

In Section~\ref{sec:koza-comp-effort-AutoML}, we clarify that Koza's
computational effort minimisation, unrecognisedly, was \emph{a new
  form of AutoML} which we will term \emph{AutoML on Computational
  Effort} (\emph{ACE}) hereafter. There we explain the unique features
of ACE, including the fact that it optimises hyper-parameters in such
a way that one is almost certain to succeed in obtaining acceptable
solutions to a problem \emph{across multiple runs}.

Following this, in
Section~\ref{sec:stat-reliability-effort}, we address historical
refinements to the computational effort metric and discuss the
statistical reliability of its empirical estimation. Finally, in
Section~\ref{sec:contr-this-article}, we list the key contributions of
the article.

\subsection{Automatic Machine Learning}
\label{sec:recent-improvements}

The research
area known as \emph{Automatic Machine Learning (AutoML)}
focuses on developing methods to automate the selection of the
elements of a machine learning
pipeline (including classification algorithms, feature extraction,
hyper-parameters,  etc.) using some form of search (Bayesian
optimisation, grid search, evolutionary algorithms, random search,
etc.)~\citep{hutter2019automated,he2021automl,karmaker2021automl,mendoza2016towards,assunccao2019denser}.

In general, AutoML has been extremely successful at both improving
over best human-tuned ML pipelines and at making ML methods
accessible to non-experts and industry. For instance, AutoML can be used to identify
optimal learning rates, number of epochs and other hyper-parameters
for neural networks (possibly including the random seed for weight
initialisation)~\citep{smith2018disciplined,yang2020hyperparameter,bethard2022we}.

There is a key characteristic of AutoML methods that we want to
highlight here: the emphasis is on finding the hyper-parameter values
such that \emph{the ML model will optimally satisfy the objective
  measure(s) in \underline{one run}}.
This is true irrespective of whether one uses a traditional
\emph{single-objective} AutoML system (where users provide a
\emph{single objective measure} that the algorithm will try to
optimise%
\footnote{Normally this is some measure of \emph{accuracy} (e.g.,
  validation accuracy) or \emph{loss} (e.g., validation loss),
  possibly measured via cross-validation.})  or one of the emerging
\emph{multi-objective} AutoML systems (see~\citep{karl2023multi} for an
extensive review).

\subsection{Computational Effort in Genetic Programming}
\label{sec:comp-effort-to-success}

Let us start with Koza's GP computational effort
measure~\citep{koza92}.

A key element of it 
is that it requires one to define a \emph{success predicate} $S$ that is
 able to recognise if an \emph{acceptable solution} to a problem has
been identified in a run. For instance, success may be defined as: (1) a
program's output being ``close enough'' to the desired value in all
fitness cases, (2) achieving a specific goal within a time limit (e.g.,
eating all the food pellets in the Artificial Ant problem or clearing
all the tiles in the Lawnmower problem), or (3) achieving a specific score
in game playing.

With this \emph{success predicate for one run} in hand, then one can
declare a \emph{set of runs successful if the performance is
  acceptable (based on $S$) in at least one run}.

Naturally, having a definition of success for a set of training runs,
we would like to impose some guarantee on \emph{how often a set of
  runs should be successful}. In other words, we need to choose a
constant $z$, which we will call \emph{confidence parameter}
hereafter, representing the desired \emph{success probability} with
which we want to be able to solve the problem (succeed) when running a
problem solver or a learning algorithm multiple times. In Koza's work
$z=99\%$. In other words, he wanted to be almost certain to succeed in
obtaining an acceptable solution in multiple runs.

Koza then defined the \emph{computational effort}, $E(M, i, z)$, as
the number of individuals one would need to process, i.e., the
\emph{number of fitness evaluations needed}, to solve a problem with
probability $z$, at generation $i$ (counting from 1) with a population
of size $M$.%
\footnote{We present the computational effort using a slightly
  different, but functionally equivalent, notation than
  in~\citep{koza92}, for easier generalisation and better adherence to
  modern ML notation.}

To compute $E$, one need to first estimate the cumulative probability
of success for a single run by generation $i$, $P(M,
i)$. In~\citep{koza92} this was
done by performing many runs (initialised using different random
seeds) up to a fixed number of generations and calculating the
fractions of runs that satisfied the success predicated.

Once $P(M, i)$ is known, the probability of \emph{not} solving the
problem by generation $i$ is $1-P(M,i)$ and, so, the \emph{probability
  of  being able to solve the problem in multiple runs} is:
\begin{equation}
  z = 1-(1-P(M,i))^R,
  \label{app:eq:Koza_z_as_func_of_p_i}
\end{equation}
where $R$ is the number of runs executed.

Solving for $R$,  one can  then determine the number
of independent runs required to achieve the target confidence
$z$:
\begin{equation}
  \label{eq:Koza_R}
  R(M,i,z) = \left\lceil \frac{\log(1 - z)}{\log(1 - P(M, i))} \right\rceil ,
\end{equation}
where $\lceil \cdot \rceil$ represents the ceiling operation, which is
applied as it does not make sense to make a fractional number of runs.

Finally,  the \emph{computational effort} (i.e., the total number of
fitness evaluations required), $E(M, i, z)$, is the product of the population
size ($M$), the number of generations completed ($i$), and the number
of runs just computed ($R$). That is:
\begin{equation}
  \label{eq:comp_effort_Koza}
  E(M, i, z) = M \times i \times R(M,i,z) .
\end{equation}

Koza then defined the   \emph{minimum computational effort} $E^*$ --- the final
test of problem difficulty for a particular GP configuration (e.g.,
with or without automatically defined functions) and population size $M$ --- as

\begin{equation}
  \label{eq:Koza_Estar}
E^* = \min_{i}  E(M, i, z) ,
\end{equation}
which is obtained after
\begin{equation}
  \label{eq:Koza_istar}
i^*=\arg\min_{i}  E(M, i, z)  
\end{equation}
generations, for a population size $M$ and success probability $z$
(typically $z=0.99$).

While Koza's traditional formulation focused on tree-based GP, the
utility of computational effort naturally extends to more complex
representations. For instance in~\citep{poli1996parallel,poli1997evolution}, we successfully employed the computational
effort to quantify the efficiency of evolving graph-like programs and
neural network topologies. These early analyses demonstrated that
effort-based evaluation is highly effective beyond computer programs,
laying the conceptual groundwork for the extension to mainstream,
gradient-based machine learning models proposed in this article.

\subsection{GP's Minimisation of Computational Effort   is AutoML}
\label{sec:koza-comp-effort-AutoML}

In this section we want to emphasise two critical innovations on
Koza's computational effort idea.

Firstly, it is important to recognise that ACE (the minimisation of the
computational effort)  was in fact \emph{one of the very first forms of
  AutoML}: using grid search (the estimation of $P(M, i)$)
it automatically yielded two optimal hyper-parameters for GP --- the
optimal number of runs and the number of generations at which to stop
them.  The connection between GP's computational effort minimisation and
hyper-parameter optimisation via AutoML could not be made back in 1992
(AutoML did not even exist as a discipline back then!) and was missed
later by AutoML researchers.

The second contribution of Koza's work is represented by a change in
perspective: \emph{the target is to achieve a satisfactory accuracy
  with the least computational effort}, not the maximum accuracy
whatever the effort. To achieve this target, effectively Koza
created \emph{the very first, and so far only, form of AutoML that
  optimises the learning effort in multiple runs}, albeit in the
specific domain of GP. The multi-run optimisation is an additional
element that distinguishes \emph{ACE} from traditional AutoML systems,
which, as mentioned in Section~\ref{sec:recent-improvements}, optimise
performance \emph{for one run}.

Why is this transition important?  The highest (validation) accuracy
possible in one run (the traditional target of AutoML/ML) is sensitive
to the random seed used for initialisation. Of course, one could use
mean or median accuracies (across training runs) for the purpose of
reducing variability~\citep{bethard2022we}.  However, when performance
varies significantly with random seeds, this method falls short, as
\emph{practitioners care more about the upper tail of the performance
  distribution}---how often the ML model achieves acceptable
results---\emph{than its central tendency}.

The reason for this is apparent if we consider a simple
example involving two learning algorithms,
\emph{HighVar} and \emph{LowVar}, with identical not-so-good average
accuracy but different accuracy distributions across multiple runs/reinitialisations.  \emph{HighVar} yields
success (i.e., acceptable accuracy) in half the runs and terrible accuracy in
the rest. \emph{LowVar}, by contrast, delivers the same accuracy every
time, never really solving the problem.
It is clear that, despite both algorithms having the same average
performance (and hence being equally good from a traditional ML standpoint),
users would prefer \emph{HighVar}, as with enough attempts it would
yield a result (e.g., a classifier) with acceptable accuracy.

Because of its dependency on the cumulative probability of success,
$P(M,i)$, the computational effort is \emph{sensitive to the upper
  tail of the distribution of accuracies recorded during training
  runs} and not only on central tendencies. So, ACE would be able to
distinguish between \emph{HighVar} and \emph{LowVar}, correctly
choosing \emph{HighVar}, the computational effort being infinite
for \emph{LowVar} but finite for \emph{HighVar}.

\subsection{Refinements and Statistical Reliability of Computational Effort}
\label{sec:stat-reliability-effort}

We should note that, approximately a decade after its definition,
Koza's computational effort measure was revisited, resulting in minor
adjustments. In \citep{christensen2002analysis}, it was found that the
ceiling operation in Equation~\eqref{eq:Koza_R} introduces a bias so
that the effort calculated with estimates of the cumulative success
probability ($P(M, i)$) differed in some cases from the actual effort
empirically calculated. Removing the ceiling operation brought the
estimate very close to the true effort. Also note that the
computational effort was originally designed for \emph{generational}
GP. In \citep{niehaus2003more}, the computational effort formulation
was adapted to consider \emph{steady-state forms of GP} (based on
tournament selection), where runs could be stopped after any number of
iterations, resulting in a better match between the estimated effort
and the theoretical value. 

Beyond these structural refinements, we must address the statistical
uncertainty inherent in empirically estimating the cumulative success
probability $P(i, a, \eta)$ from a finite number of runs. Prior
studies in evolutionary computation have demonstrated that the
relative estimation error of computational effort becomes highly
volatile when the success probability approaches the extremes of 0 or
1 \citep{barrero2011empirical, barrero2015study}.

A recent, highly complementary study by \citet{noori2026statistical}
provides a rigorous statistical analysis of this exact phenomenon in
the context of stochastic hardware and Boolean satisfiability (SAT)
solvers. They mathematically prove that the maximum relative error in
estimating the required number of restarts ($R$) increases
significantly at the boundary conditions due to the highly non-linear
formulation of the metric. To mitigate this, they propose an adaptive
sampling algorithm that dynamically adjusts the number of repeats to
guarantee a specific target error.

While \citet{noori2026statistical} focus strictly on the statistical
interval estimation and experimental design for standard heuristic
solvers, their findings highlight a critical mathematical reality for
the metric. Consequently, while computational effort provides a
powerful analytical lens for machine learning, practitioners must be
mindful that estimates derived from empirical success probabilities
near 0 or 1 are inherently susceptible to sampling noise.

However, in what follows we did not need to use these results.

\subsection{Contributions of this Article}
\label{sec:contr-this-article}

This article proposes a fundamental rethinking of how machine learning
algorithms are evaluated and parametrised. Specifically, it makes the
following key contributions:

\begin{enumerate}
\item \textbf{A New Evaluation Paradigm via Computational Effort:} We introduce a paradigm
  shift in ML evaluation by proposing that performance should be
  measured by computational effort---explicitly defined as the number
  of gradient-descent steps required to achieve a predefined,
  acceptable level of accuracy with high probability (e.g., 99\%)---rather 
  than merely pursuing maximum accuracy irrespective of cost. We formalise 
  this by redefining Koza's concept of computational effort---originally 
  designed for Genetic Programming---so that it can be seamlessly applied 
  to any mainstream machine learning model trained via full-batch gradient descent.

\item \textbf{AutoML on Computational Effort (ACE):} We operationalise
  the minimisation of this effort as a novel AutoML approach (ACE). We
  demonstrate its effectiveness by co-optimising the learning rate,
  the maximum number of epochs, and the number of independent random
  restarts across 11 widely used ML models and five standard
  classification datasets.

\item \textbf{The Dual Benefit of Rapid Landscape Traversal:} Our
  empirical analysis reveals that optimal training
  strategies consistently favour unusually large learning
  rates. Crucially, we connect this finding to the phenomenon of
  superconvergence, demonstrating that aggressive landscape traversal
  not only acts as a powerful regulariser for generalisation, but is
  also statistically optimal for minimising the expected computational
  effort during training.

\item \textbf{Identification of Phase Transitions:} We identify
  distinct phase transitions in the optimal search strategy as the
  target accuracy increases. We show that while a single training run
  is sufficient for lower accuracy thresholds, reaching a model's
  performance limit requires a dramatic shift towards conducting
  numerous independent, short restarts to explore diverse trajectories
  on the error surface.

\item \textbf{Relative Problem Difficulty and Model Selection:} We
  introduce a novel framework for model selection based on the concept
  of \emph{perceived} problem difficulty. By demonstrating that
  difficulty is highly relative to the algorithm and the target
  accuracy, we allow practitioners to diagnose whether a bottleneck
  lies in optimisability or generalisability. Ultimately, this enables
  the systematic selection of models to either achieve a target
  accuracy or to maximise the achievable accuracy for a fixed budget
  of gradient descent steps.
\end{enumerate}

\section{Extending ACE to Mainstream ML}
\label{sec:extend-kozas-automl}

In this section we extend the computational effort notion of
performance and AutoML  to compare (and do ACE
on)  forms of machine learning other than GP systems.

\subsection{Computational Effort for Machine Learning based on Gradient Descent}
\label{sec:comp-effort-mach}

As we will explain in detail in Section~\ref{sec:test-problems-and-models}, we
specifically focus on ML models which can be trained via
\emph{Gradient Descent} (GD) to learn the task at hand. These include
shallow and deep \emph{Neural Networks} (NNs), \emph{Linear
  Discriminant Analysis} (LDA) with and without regularisation,
\emph{Logistic Regression} (LR) with and without regularisation and
linear \emph{Support Vector Machines} (SVMs) with and without
regularisation.%
\footnote{By gradient descent here we mean the \emph{full-batch version of
  gradient descent} where gradient calculations are exact, i.e.,
  unaffected by noise.  While for some of the models we use,
  there are more main-stream forms of training such as the batched
  version of GD (known as Stochastic Gradient Descent), closed form
  solutions (specifically for LDA), Newton-Raphson (for LR) or
  quadratic programming (for SVM), all ML models considered can be
  trained also by GD.}

For these models, we define the \emph{computational effort to success as
the expected number of gradient descent steps required to ensure
successful training} according to a \emph{success predicate} $S$ (e.g., achieving
an accuracy of 95\%) in at least one of multiple runs with confidence $z$ (e.g.,
99\%). The minimum such effort can then be interpreted as a \emph{measure of
the intrinsic difficulty of identifying acceptable solutions for a
given model–problem pair via gradient-based optimisation}.

Importantly, this notion of difficulty is \emph{model-dependent}. It should
be distinguished from measures of \emph{dataset difficulty} that are defined
independently of any particular learning algorithm. Such measures have
been widely studied in the literature and include \emph{geometric notions}
such as linear
separability~\citep{minsky1969perceptrons,minsky1988perceptrons,vapnik1998statistical,novikoff1962convergence},
\emph{statistical notions} based on feature overlap or correlation and class
separability~\citep{ho2002complexity,lorena2019complexity,hastie2009elements},
and \emph{information-theoretic notions} such as Kolmogorov
complexity~\citep{kolmogorov1965three,li2008kolmogorov}.

We refer to our measure as ``intrinsic'' because it abstracts away from
the computational cost of individual gradient steps, which may vary
significantly across models and datasets (e.g., shallow models versus
deep neural networks, or small versus large-scale datasets). Instead,
it focuses on the number of optimisation steps required to reach
sufficiently high-quality solutions.%
\footnote{This is not dissimilar to Koza's computational effort that
  did not consider that the execution of some programs (e.g., bloated
  individuals in late generations) requires much longer than for others
  (e.g., the small individuals typical of early generations) or that
  the fitness function for some problems is generally more expensive
  (e.g., involving many more fitness cases or much more complex
  simulations) than for others.}

\subsection{Hyper-parameter Optimisation by Gradient-descent ACE}
\label{sec:hyperp-optim-ace}

We can use ACE --- with our new definition of computational effort
based on gradient decent steps --- to optimise hyper-parameters of the
machine learning algorithms mentioned above. While Koza's original
formulation focused solely on optimising the number of runs and the
iteration/epoch number where to stop the training, applying this
paradigm to gradient-based learning requires us to go
further. Specifically, it is necessary to also optimise the learning
rate $\eta$ to obtain successfully trained ML models with high
probability.  In fact, an early implementation of this explicit
co-optimisation strategy---simultaneously tuning $\eta$, the number of
runs $R$, and the stopping epoch $i$---was successfully utilised in
the PhD thesis of the second author~\citep{yilmaz2021uncovering}. There, this expanded effort metric was
used to ensure a rigorously fair performance comparison between
standard gradient descent and variants employing GP-evolved learning
rules. In this article, we formalise this methodology into the general
ACE framework and explore its dynamics across a diverse array of
mainstream ML models.%
\footnote{Naturally, adding any additional hyper-parameters to the set
  optimised by ACE extends significantly the computation required for
  the estimation of the cumulative success probability $P$ mentioned
  in Section~\ref{sec:comp-effort-to-success}. The comparatively-small
  computer power available three decades ago is the most likely reason
  why other hyper-parameters, such as the population size or the
  crossover rate, were not optimised in Koza's work.}

Unless otherwise stated, in the rest of the article we will use
\emph{accuracy} as our performance measure, so our success predicate
$S$ will be defined as \emph{the ML model achieving an accuracy of at least}
$a$. Also,  we will set the confidence parameter to $z=99\%$ as our target success probability. In
other words, we  ask users to specify what accuracy level they
consider acceptable and we want to be almost certain to succeed in
obtaining a satisfactorily trained ML model possibly in multiple
runs. However, in Section~\ref{sec:sensitivity-analysis-z} we will
study the effects on our results of adopting different values for the confidence
parameter $z$.

Note that, \emph{unlike Koza's work}, where the success criteria were
pre-fixed,%
\footnote{For instance,  in classification problems, the classes of all
  fitness cases had to be predicted correctly, i.e., an accuracy
  of $a=1.0$ (i.e., 100\%) was required.} here \emph{we also study how the computational
effort varies as the accuracy threshold $a$ is varied}.
Additionally, again unlike in Koza's work where success was defined based on
training-set performance, \emph{here we also consider (cross-validated) 
validation-set performance as a success criterion}.

There is no notion of population in traditional  ML models, so the
parameter $M$ present in the calculations of Koza's computational
effort (see Section~\ref{sec:comp-effort-to-success}) will not appear
here.  We still have epochs/iterations which play the role of
generations ($i$),  the threshold accuracy $a$ and, as indicated above,  also
the learning rate $\eta$. So, we replace the cumulative probability
of success for a single run by generation $i$, $P(M, i)$, with
 the \emph{cumulative probability of training a model to
the desired degree of accuracy $a$ by epoch $i$ and with a learning
rate $\eta$, $P(i, a, \eta)$}.

To formalise this mathematically, let $\Omega$ represent the probability 
sample space of all possible random initialisations (the initial parameter 
weights $\theta_0$), where a specific choice of random seed corresponds 
to a distinct element $\omega \in \Omega$. For a given learning rate $\eta$, 
let $\mathcal{A}_{k, \eta}$ be the random variable mapping $\Omega \to [0, 1]$ 
that represents the accuracy achieved \emph{exactly at} epoch $k$. A single 
training run executed with a specific seed $\omega$ thus represents a single, 
completely deterministic \emph{realisation} (or sample path), denoted by 
$\mathcal{A}_{k, \eta}(\omega)$. 

Because success is defined as reaching the target accuracy $a$ \emph{at any point up to} 
epoch~$i$, we must base our success probability on the running maximum of this 
stochastic process over time. Let us denote this maximum accuracy achieved 
by epoch $i$ as the random variable $\mathcal{A}'_{i, \eta} = \max_{1 \le k \le i} \mathcal{A}_{k, \eta}$, 
and let $f_{\mathcal{A}'_{i, \eta}}(x)$ be its corresponding probability density function.
Then, $P(i, a, \eta)$ is explicitly defined
as the complementary cumulative probability function over the
distribution of these maximum accuracies:
\begin{equation}
  P(i, a, \eta) = \mathbb{P}(\mathcal{A}'_{i, \eta} \ge a) = \int_{a}^{1} f_{\mathcal{A}'_{i, \eta}}(x) \, dx.
  \label{eq:prob_success_integral}
\end{equation}

Then following Koza's footsteps
(Equations~\eqref{app:eq:Koza_z_as_func_of_p_i} and~\eqref{eq:comp_effort_Koza} still apply), we
obtain the 
 \emph{expected number of runs required to
solve the problem when stopping at epoch $i$}:
\begin{equation}
  R(a,\eta,i,z) = \left\lceil \frac{\log(1-z)}{\log(1-P(i,a,\eta))} \right\rceil,
  \label{app:eq:R_as_func_of_z_and_p_i}
\end{equation}
where we made explicit the fact that $R$ depends $a$, $\eta$, $i$ and
the confidence parameter $z$.%
\footnote{We considered removing the ceiling operation as first suggested
 by  \citet{christensen2002analysis}. However, upon testing this, we found that in the cases
  where one run is sufficient to reach the accuracy threshold  $a$, the
  $R^*$ obtained when the ceiling operation is omitted can be less than 1
  (e.g., 0.8), which is not very meaningful as it hints that
  sometimes one should not  train the ML model at all. So, we kept the ceiling operation in Equation~\eqref{app:eq:R_as_func_of_z_and_p_i}.}
From this we define the 
\emph{computational effort} as  the  \emph{expected total number of
  learning steps required to solve the problem with accuracy no less
  than the threshold accuracy ($a$) with confidence $z$
  when executing training runs with learning rate $\eta$ 
  and stopping them at epoch~$i$}:
\begin{equation}
  \label{eq:comp_effort_ACE}
  E(a,\eta,i,z)  = i \times   R(a,\eta,i,z) = i \times \left\lceil \frac{\log(1-z)}{\log(1-P(i,a,\eta))} \right\rceil .
\end{equation}

The \emph{minimum computational effort} is then
\begin{equation}
  \label{eq:ACE_E_star}
  E^*(a,z) = \min_{i,\eta} E(a,\eta,i,z),  
\end{equation}
with corresponding \emph{optimal hyper-parameters}:
\begin{equation}
  \label{eq:ACE_optimal_hyperparameters}
  (i^*, \eta^*) = \arg \min_{i,\eta} E(a,\eta,i,z) \qquad
  R^*(a,z) = \left\lceil \frac{\log(1-z)}{\log(1-P(i^*,a,\eta^*))} \right\rceil . 
\end{equation}
  
 We should note that, because the computational effort
depends on the users having defined an accuracy threshold $a$, beyond
which runs are considered successful, it
intrinsically considers \emph{two objectives} which practitioners care
about: \emph{computational cost and acceptable accuracy}.

The computational effort just defined is \emph{sensitive to the upper
  tail of the distribution of maximum accuracies recorded during
  training runs} and not only to the mean
\begin{equation}
  \mathbb{E}[\mathcal{A}'_{i, \eta}] = \int_{0}^{1} x f_{\mathcal{A}'_{i, \eta}}(x) \, dx,
  \label{eq:mean_accuracy_max_expectation}
\end{equation}
at a particular
epoch.%
\footnote{\label{footnote:mean_acc}Notice that this differs from the mean
accuracy at epoch~$i$ given a learning rate $\eta$ which is given by:
\begin{equation}
  \mathbb{E}[\mathcal{A}_{i, \eta}] = \int_{0}^{1} x f_{\mathcal{A}_{i, \eta}}(x) \, dx,
  \label{eq:mean_accuracy_expectation}
\end{equation}
which is a traditional performance measure in ML, although, mean peak
training and validation accuracies (empirical estimates of
$\mathbb{E}[\mathcal{A}'_{i, \eta}]$) are sometimes used in deep
learning
research, e.g., see~\citep{baldock2021deep,pratama2024comparative,calle2025integration,rinanda2025deep}).}

Additionally, the value of $R$ provided by
Equation~\eqref{app:eq:R_as_func_of_z_and_p_i} is a \emph{very important metric},
as it signals whether a single-run approach is sufficient or a multi-run
restart strategy is needed (more on this in the next section).
The exact point when the \emph{phase transition} from a single-run approach ($R = 1$)
to a multi-run restart strategy ($R > 1$) occurs can be isolated by
relaxing the ceiling operation in the
formulation of the expected number of runs in Equation~\eqref{app:eq:R_as_func_of_z_and_p_i}
and setting $R(a, \eta, i, z) = 1$,  yielding:
\begin{equation}
  \frac{\log(1-z)}{\log(1-P(i, a, \eta))} = 1.
  \label{eq:multi_run_condition}
\end{equation}
This  results in the \emph{phase-transition condition}:
\begin{equation}
  P(i, a, \eta) = z.
  \label{eq:phase_transition_condition}
\end{equation}
In other words, \emph{an
independent, single training run remains optimal if and only if its
cumulative success probability matches or exceeds the chosen
confidence parameter}, that is $P(i, a, \eta) \ge z$.

\medskip

Finally, as will be apparent in Section~\ref{sec:valid-perf-gener}, we
should note that, for fixed
accuracy threshold $a$ and confidence $z$, \emph{the
  computational effort $E(a,\eta,i,z)$ is not necessarily a monotonic
  function of the learning rate $\eta$}. For instance, beyond a
certain value of $\eta$, gradient-descent-type of algorithms may have
trouble descending on the error surface.
Also, and this may appear counter-intuitive,
$E(a,\eta,i,z)$ is \emph{not a monotonically increasing function of}~$i$.
For instance, in some cases performing many short runs may yield a
correctly trained model with less overall effort than using a
few long runs. However, making runs too short may collapse the success
probability to zero making $E(a,\eta,i,z)\to \infty$.
\medskip

For all these reasons, we believe that \emph{the computational effort is a
  better aggregate performance measure to compare problems and ML
  models that can be trained by GD} and it is thus a better objective
measure for AutoML.

\subsection{Theoretical Sensitivity Analysis of the Confidence Parameter $z$}
\label{sec:sensitivity-analysis-z}

Throughout this article, we will make use of a fixed target success
probability at the conventional value of $z = 0.99$, aligning with the
historical choice established by \citet{koza92}. However, because the
calculation of the computational effort decouples the empirical
estimation of the cumulative success probability $P(i, a, \eta)$ from
the user-defined confidence target $z$, it is possible to analytically
predict how the computational effort  scales
under alternative values of $z$.

In any region where the multi-run restart regime dominates
($R > 1$), the denominator $\log(1-P(i, a, \eta))$ in Equation~\eqref{eq:multi_run_condition} is 
not dependent on $z$. Consequently, altering $z$ alters the required
number of runs---and by extension, the cumulative computational effort
$E$---in a completely predictable way. Both these quantities are
scaled the constant factor:
\begin{equation}
  \text{Scaling Factor} \approx \frac{\log(1-z_{\text{new}})}{\log(1-z_{\text{baseline}})},
\end{equation}
where $z_{\text{baseline}} = 0.99$. So, if we used a \emph{more
  relaxed confidence factor of $z = 0.90$}, the scaling factor is
$\frac{-2.30}{-4.61} = 0.5$. So, this would halve the computational
effort across all multi-run configurations. If instead, we
\emph{heightened the confidence factor to $z = 0.999$}, the scaling
factor would be $\frac{-6.91}{-4.61} = 1.5$. So, this would
systematically increase the required computational effort across the
board by 50\%.

\subsection{Implementation of  \emph{ACE} based on Gradient-decent Steps}
\label{sec:impl-grad-steps}

As indicated above, to optimise the performance of our learning
models with ACE,  we need to
minimise $E(a,\eta,i,z)$ over $\eta$ and $i$, obtaining the
\emph{minimum computational effort} $E^*(a,z)$, the optimisation
yielding the optimal learning rate $\eta^*$, the optimal maximum number
of epochs $i^*$ and the optimal number of runs $R^*$.

Because $E(a,\eta,i,z)$ depends solely on $P(i, a, \eta)$, the latter is the
target to be estimated for the purpose of AutoML. \emph{For each problem and model} we adopted the same \emph{grid search} method used in Koza's work
(but extended to the additional hyper-parameter $\eta$)
and estimated $P(i,a,\eta)$
using
the following simple  caching strategy to avoid recomputing values previously
obtained:
\begin{enumerate}
\item Each dataset was divided into 5 cross-validation folds.
\item For each fold, we did a set of 200 training  runs (using different
  random seeds) lasting $i_{\mathrm{max}}=500$ epochs for \emph{thirty
    nine} $\eta$ values (0.01, 0.02, \dots, 0.1, 0.2, \dots, 3.0),
  recording the \emph{full history} (training accuracy, validation
  accuracy, loss, and validation loss) \emph{for each epoch of each
    run and with each learning rate}. In each run the initial parameter 
  weights $\theta_0$ were drawn from a uniform distribution 
  $$\mathcal{U}\left(-\frac{1}{\sqrt{n}},
    +\frac{1}{\sqrt{n}}\right)$$ where $n$ is the number of input
  features $F$ for the linear models and for the input layer of the NNs,
  while it is number of incoming
  connections (\texttt{fan\_in}) for the neurons in the other layers
  of the NNs.%
  \footnote{We used this distribution because it is the default in
    \texttt{Pytorch} and is suitable to break symmetries and reduce
    the dying-neuron and vanishing/exploding-gradients problems for
    ReLU-based NNs. We use the same initialisation for all other
    models, even if they do not suffer from such problems, for
    comparability of results.}
  
\item From this information, for each $\eta$ we estimated all values
  of $P(i,a,\eta)$ for   $i=1, \cdots, i_{\mathrm{max}}$  as the
  fraction of the 200 runs/seeds where the (training or validation)
  accuracy at epoch $i$ was no less than~$a$.
\end{enumerate}

An advantage of our approach of storing the full history of runs
rather than just the accuracy is that this allows to recompute
$P(i,a,\eta)$ (at point 3 above), not only for different values of the
accuracy threshold $a$, but also for other performance measures
(validation accuracy, loss and validation loss), \emph{at no extra
  cost}.

A limitation of this approach is that it is computationally expensive:
for each problem-model pair considered, the combination of 5 folds, 39
learning rates, 200 runs and 500 epochs results in 39,000 runs and
19,500,000 gradient descent steps. A second limitation is that with
200 runs, the smallest non-zero value of $P(i,a,\eta)$ is
$\frac 1 {200}$. Substituting such a value and $z=0.99$ in
Equation~\eqref{app:eq:R_as_func_of_z_and_p_i} gives the maximum $R$
value that we can estimate: approximately 920. Substituting this and
the maximum number of iterations $i_{\mathrm{max}}=500$ used in our
experiments in Equation~\eqref{eq:comp_effort_ACE}, gives the maximum
value of $E(a,\eta,i,z)$ that we can estimate: approximately 460,000
gradient descent steps.

Naturally, if longer runs, a finer resolution, a larger range for
the learning rate $\eta$ or even lower success probabilities
$P(i,a,\eta)$ were of interest, the grid search would need to be extended
or replaced by smarter search strategies such as Bayesian Optimisation
or Evolutionary Algorithms which are commonly used in current AutoML
systems. However, out of the box, such systems are not designed to
deal with the multiple runs necessary to compute $P(i, a, \eta)$. So,
these would need to be modified and would also need to incorporate some form of smart
caching.

Another limitation is that, in this article, we consciously retain
Koza's original, unadjusted formulation, i.e., without incorporating
any of the techniques reviewed in
Section~\ref{sec:stat-reliability-effort}. Adhering to this classical
definition reduces the mathematical and experimental complexity of our
initial treatment, providing a cleaner theoretical baseline from which
to explore the novel application of computational effort to
gradient-descent dynamics. Nevertheless, we fully intend to
incorporate these structural refinements---such as the removal of the
ceiling operator and the use of finer-grained sampling criteria---in
future extensions of the ACE framework to further enhance the
precision of our effort estimates.

\section{Problems,  ML Models and Loss Functions}
\label{sec:chosen-test-problems}

\subsection{Test Problems and ML Models}
\label{sec:test-problems-and-models}

We used five well-known classification problems for experimentation:
the \emph{Iris, wine recognition (Wine), E-coli, Breast Cancer
  Wisconsin (BCW) and DNA (Promoter Gene Sequences)
  classification}. More details on these problems are provided in
Section~\ref{sec:test-problems-degree-overparam} of the Supplementary
Material (SM, hereafter).

\begin{table}[tb]
  \caption{Models used in this work.}
  \label{tab:ML_models}
  \begin{center}
    \begin{small}
      \begin{tabular}{|l|p{1.8in}|p{1.2in}|c|}
        \hline
        \emph{Model} & \emph{Description} & \emph{Loss Function} & \emph{Error Surface}  \\ 
        \hline
        LDA-inf & Linear Discriminant Analysis (no regularisation, i.e., $C=\infty$) & MSE (Eq.~\eqref{eq:MSE_loss})& Convex  \\
        \hline
        LDA-0.1 & LDA (regularised via $C=0.1$) & MSE (Eq.~\eqref{eq:MSE_loss})& Strictly Convex \\
        \hline
        LDA-0.01 & LDA (regularised via $C=0.01$) & MSE (Eq.~\eqref{eq:MSE_loss})& Strictly Convex \\
        \hline
        LR-inf & Logistic Regression   (no regularisation, i.e.,
             $C=\infty$) & Cross-Entropy (Eq.~\eqref{eq:CE})& Convex \\
        \hline
        LR-0.1 & LR (regularised via $C=0.1$) & Cross-Entropy  (Eq.~\eqref{eq:CE}) & Strictly Convex  \\
        \hline
        LR-0.01 & LR (regularised via $C=0.01$) & Cross-Entropy  (Eq.~\eqref{eq:CE}) & Strictly Convex  \\
        \hline
        SVM-inf & Linear SVM   (no regularisation, i.e., $C=\infty$) & Multi-Hinge (Eq.~\eqref{eq:hinge_loss})& Convex  \\
        \hline
        SVM-0.1 & Linear SVM (regularised via $C=0.1$) & Multi-Hinge (Eq.~\eqref{eq:hinge_loss}) & Strictly Convex  \\
        \hline
        SVM-0.01 & Linear SVM (regularised via $C=0.01$) & Multi-Hinge (Eq.~\eqref{eq:hinge_loss}) & Strictly Convex  \\
        \hline
        NN-ReLU & Shallow NN (2 hidden layers with 64 and 32 units
                  with ReLU activations and a Softmax output layer) & Cross-Entropy  (Eq.~\eqref{eq:CE}) & Non-convex  \\
        \hline
        DNN-ReLU & Deep NN (6 hidden layers with 64, 64, 64, 32, 32
                   and 32 units with ReLU activations and a Softmax output layer) & Cross-Entropy  (Eq.~\eqref{eq:CE}) & Non-convex  \\  
        \hline
      \end{tabular}
    \end{small}
  \end{center}
\end{table}

In this work, we used 11 machine learning models: one shallow and one
deep \emph{Neural Network} (NN) with ReLU activation function and
simple architectures, \emph{Linear Discriminant Analysis} (LDA) with
two levels of regularisation and without it, \emph{Logistic
  Regression} (LR) with the same levels of regularisation and without
it, and \emph{linear Support Vector Machines} (SVMs), once again, with
the same levels of regularisation and without it. More details are
provided in the first two columns of Table~\ref{tab:ML_models}.

\subsection{Loss Functions}
\label{sec:loss-functions}

By their own nature, different models require different loss functions
(as per third column of Table~\ref{tab:ML_models}). These are described
below:

\begin{itemize}
\item \emph{Mean Squared Error} (MSE): This was used for the
  LDAs. That is we  treat LDA classification as a regression toward one-hot
  encoded targets. For $K$ classes, the loss for a single observation
  is:
  \begin{equation}
    \label{eq:MSE_loss}
    L_{\mathrm{MSE}} = \sum_{j=1}^{K} (s_j - y_j)^2,
  \end{equation}
  where $s$ is the vector of logits and $y$ is the one-hot encoded
  ground truth.

\medskip
\item \emph{Cross-Entropy Loss}: This was used for LRs and Neural
  Networks. The cross-entropy loss measures the performance of a
  classification model whose output is a probability value between 0
  and 1. For a single observation with  correct class $y$:
  \begin{equation}
    \label{eq:CE}
    L_{\mathrm{CE}} = -\log\left( \frac{e^{s_y}}{\sum_{j=1}^{K} e^{s_j}}\right).
  \end{equation}

\medskip
\item \emph{Multi-Class Hinge Loss}: This was used for the SVMs. The hinge loss uses a margin-based approach to
  ensure the correct class score $s_y$ is higher than all incorrect
  class scores $s_j$ by at least a margin $\Delta$ (in our
  implementation $\Delta=1.0$):
  \begin{equation}
    \label{eq:hinge_loss}
    L_{\mathrm{Hinge}} = \sum_{j \neq y} \max(0, \Delta - s_y + s_j).
  \end{equation}
  
\end{itemize}

Naturally, such \emph{loss values need to be averaged across the $N$
training examples} for a problem to obtain the corresponding total loss,
$L(\theta)$, where we make dependency of the model parameters,
$\theta$, explicit.
  
In models where \emph{regularisation}  was used, regardless of
the loss function $L$, the \emph{objective/error function} $J$ included a
penalty term. That is:
\begin{equation}
  \label{eq:loss_with_L2_penalty}
  J(\theta) = L(\theta) + \frac{1}{2  C  N} \|\theta\|^2_2,
\end{equation}
where $C$ is the inverse regularisation strength and $\|\cdot\|_2$
is the $L_2$ norm. This term ensures the error surface remains
strictly convex for the linear models, reducing overfitting by
penalising large weights.%
\footnote{At implementation level
we applied $L_2$ regularisation via a \emph{weight decay} term $\lambda = 1/(C N)$, where the parameters $\theta$ are updated according to $\theta \leftarrow \theta - \eta (\nabla L + \lambda \theta)$.}

\subsection{Characteristics of Objective Functions and their
  Interactions with Gradient Descent}
\label{sec:char-object-funct}

The last column of Table~\ref{tab:ML_models} indicates whether a
model error surface is \emph{convex} (thereby presenting \emph{at least
  one global optimum} and no local optima), \emph{strictly
  convex} (thereby presenting \emph{exactly one global optimum} and no
local optima) or \emph{non-convex} (presenting local optima). In
principle, if $\eta$ is small, local optima can trap GD in non-convex
landscapes, making it impossible for GD to achieve the best possible
parameters $\theta$ even given a sufficient number of epochs to
converge.

The extent to which the aforementioned landscape characteristics
influence the training process is heavily modulated by the model’s
\emph{degree of over-parametrisation}, as quantified by the ratio
between the number of parameters, $P$, in a model and the number of
instances, $N$, in a training set. We report information on this for
our model-problem pairs in SM's
Table~\ref{tab:overparameterization_analysis}.

In under-parametrised
regimes, such as those occupied by LDA or LR on our test problems with
the exception of DNA, the error surface is typically constrained by
the low dimensionality of the parameter space. This often results in
simpler, convex landscapes where the global optimum is more easily
accessible but of limited quality (i.e., there is a ``high floor'').
In contrast, in the highly over-parametrised regimes of NN-ReLU and
DNN-ReLU, the non-convexity of the error surface becomes a defining
feature. However, theoretical insights suggest that in such
over-parametrised regimes, the abundance of dimensions can create a
benign landscape where local optima are less problematic than one
might expect~\citep{dauphin2014identifying,choromanska2015loss}.

Furthermore, if the number of epochs available to GD is limited (e.g.,
100 epochs), the interaction between the learning rate $\eta$ and
the random seed introduces \emph{path-dependency}, even for the convex error
surfaces. While a sufficiently small $\eta$ guarantees a monotonic
convergence toward the global optimum, it may result in slow progress
that fails to reach the minimum within the available epochs.

Conversely, a large $\eta$ may lead to an \emph{oscillatory regime} where the
model overshoots the optimum, bouncing between the walls of the convex
error surface. In such cases, the final accuracy becomes highly
sensitive to the initial parameters $\theta_0$ (the random
seed). Different seeds place the starting point at different distances
and orientations relative to the optimum; combined with a high $\eta$,
this determines the specific \emph{phase} of the oscillation at the
final epoch. The instability induced by large $\eta$ values may be
such that the region of the global optimum is too narrow for the step
size to allow for stable convergence, leading the optimiser to
oscillate within higher-error regions of the parameter space.
In
non-convex landscapes, these regions correspond to wider basins of
attraction that, while potentially having higher training error, often
provide superior validation performance, as discussed in the following
section.

Consequently, in all these
cases, one should expect \emph{a wide distribution of end-of-run accuracies
across seeds}.

As we will see in the article, this \emph{variance can be leveraged}:
multiple random restarts with a higher $\eta$ may result in a model
closer to the global optimum than a single, stable run with a small
learning rate that lacks the speed required to converge in time.

\subsection{Training and Validation Error Surfaces}
\label{sec:train-valid-error-surfaces}

Of course training by GD is only performed using the \emph{training
  set}, which is what defines the training error surface, and the
possible dynamics of GD on it discussed in the previous
section. However, of course, one is interested in how well the trained
model works on unseen data, or their surrogate: the \emph{validation set}. Because this contains
different data points from the training set, it has an associated
\emph{validation error surface} which is related to, but different
from, the error surface for the training set.

Thus, at each GD step on the training error surface, a corresponding
step is performed on the validation error surface. Most of the time
the latter is reasonably aligned with the validation error
gradient, resulting in a validation loss decrease. However, in some conditions, the alignment may be very poor
and the step may move in the wrong direction, resulting in a
validation loss \emph{increase}.

This happens because the two error surfaces are not identical, even if
the validation set is representative. In fact, it is common, even for
models where loss functions are convex or strictly convex, that the
training optima do not necessarily correspond to the validation
optima. This is more frequent where training optima are narrow (i.e.,
where to small changes of the weight vector $\theta$ correspond large
changes of the error). Instead, when optima are wider/flatter
(e.g., where to small changes of the weight vector $\theta$ correspond small
changes of the error), training performance is more aligned with
validation performance.

The use of $L_2$ regularisation increases the alignment between the
two error surfaces. Also, the trajectory effects described in the
previous section act as implicit forms of regularisation, leading GD
to settle for larger basins of attraction, i.e., places where the two
error surfaces are more similar. Indeed, in the case of NNs (with
their non-convex error surfaces), extremely large values $\eta$ have
been recently reported to have a regularising effect --- the phenomenon known as
\emph{superconvergence}~\citep{smith2019super,oymak2021provable,smith2021origin,smith2018disciplined,li2019towards,mohtashami2023special}. This allows the optimiser to ``jump'' out of sharp training minima that
do not generalise well~\citep{lewkowycz2020large}, settling instead in
flatter regions where the training and validation surfaces are more
closely aligned.

\smallskip

For these reasons, in the article, we will also focus on validation
performance obtained at each epoch, so as to ascertain what level of
alignment between error surfaces --- that is, \emph{generalisation}
--- we can expect from having trained a model with certain
hyper-parameters. Studying validation performance will also make it
possible to understand the relationship between the optimal
hyper-parameters for training performance and those for validation
performance.

\section{Experimental Results}
\label{sec:valid-perf-gener}

In this section we report on experimental results obtained with the
problems and ML models introduced in the previous section.%
\footnote{As mentioned in Section~\ref{sec:paradigm-shift}, in the
  article we will focus only on accuracies. However, we have
  equivalent results for losses which we will provide in an online
  repository and discuss in a technical report.}

As stated in Section~\ref{sec:impl-grad-steps}, the necessary data
were collected by performing $\approx 4\times 10^4$ runs and
$\approx 2\times 10^7$ gradient-descent steps per model-problem pair. So,
in total \emph{the experimentation required over two million runs and one
billion gradient descent steps}.

\subsection{Mean Accuracies}
\label{sec:mean_train_valid}

Let us start by looking at accuracies, more specifically: mean
(cross-validated) accuracies across 200 seeds. While these quantities
do not directly enter the calculations of success probability and
computational effort, they represent the central tendency of the
accuracy distribution and the standard measure of performance in ML,
and are, thus, of interest. The theoretical mean accuracy
$\mathbb{E}[\mathcal{A}_{i, \eta}]$ is reported
in Equation~\eqref{eq:mean_accuracy_expectation} (footnote~\footref{footnote:mean_acc}); the empirical
averages reported in our experimental results serve as sample
estimators of this theoretical expectation.

As examples, Figures~\ref{fig:accuracies_LR_model} and~\ref{fig:accuracies_NN_ReLU_model} report the mean training (left) and
validation (right) accuracies for the shallow neural network (NN-ReLU)
and logistic regression model without regularisation (LR),
respectively, across all problems and for all 39 learning rates
tested. Means are across the 200 reinitialisations
(seeds). SM's Section~\ref{app:mean_accuracies} reports full results.

\begin{figure*}[p]
  \centering
  \begin{tabular}{c@{}c}
    Training Accuracy & Validation Accuracy \\
    \includegraphics[clip,trim=70 30 30 30,width=.52\linewidth]{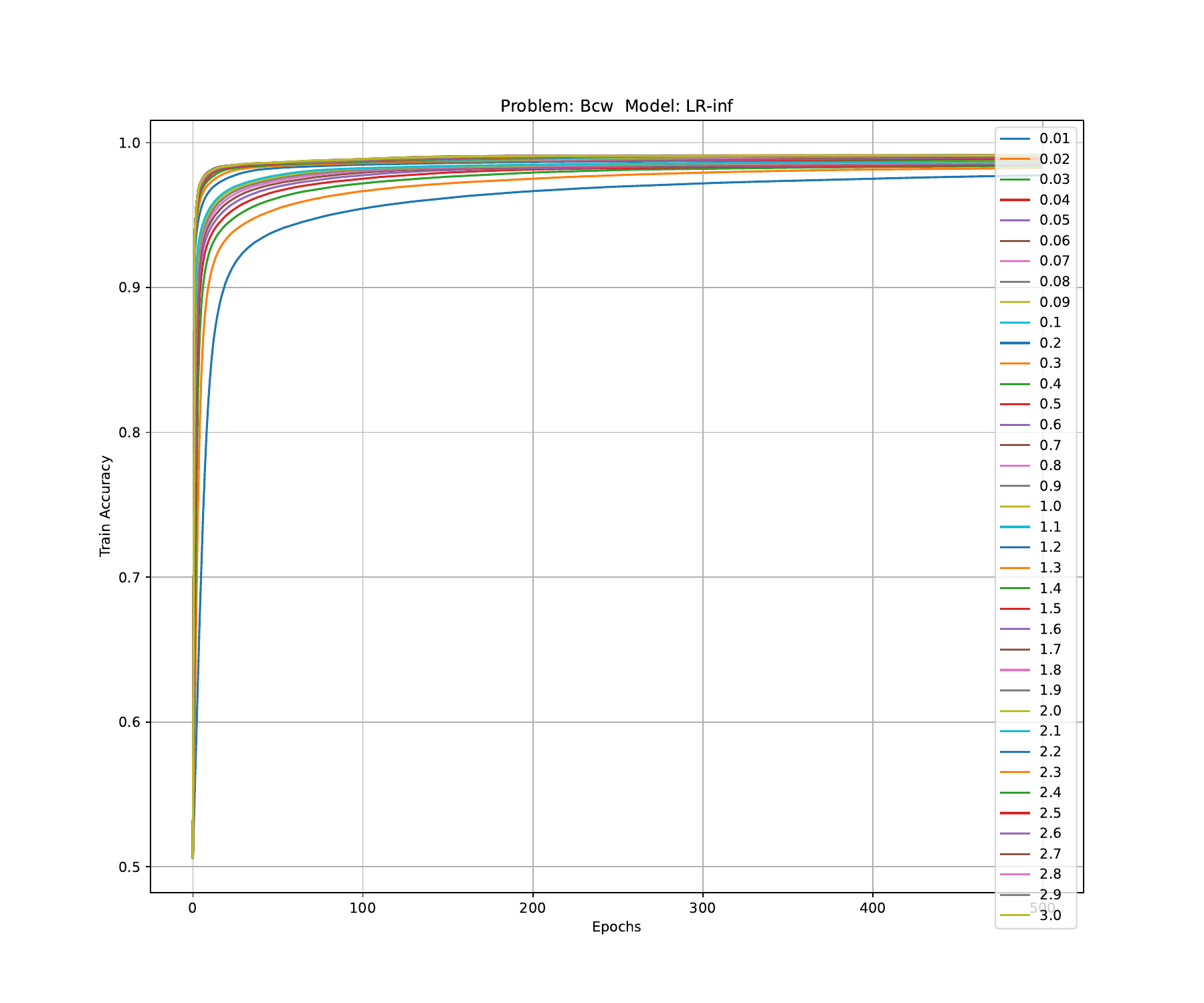} &
    \includegraphics[clip,trim=70 30 30 30,width=.52\linewidth]{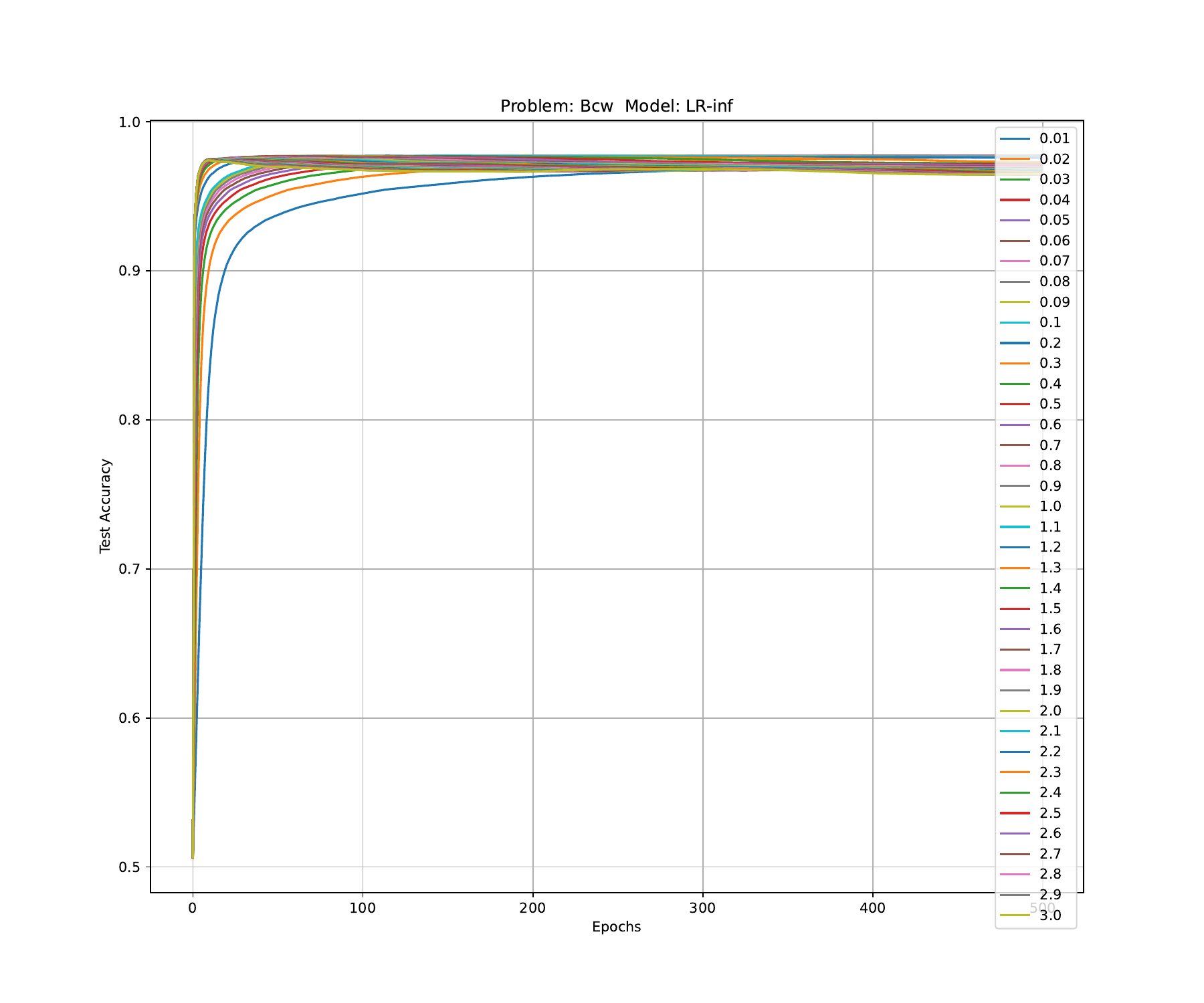} \\[-3mm]
    \includegraphics[clip,trim=70 30 30 30,width=.52\linewidth]{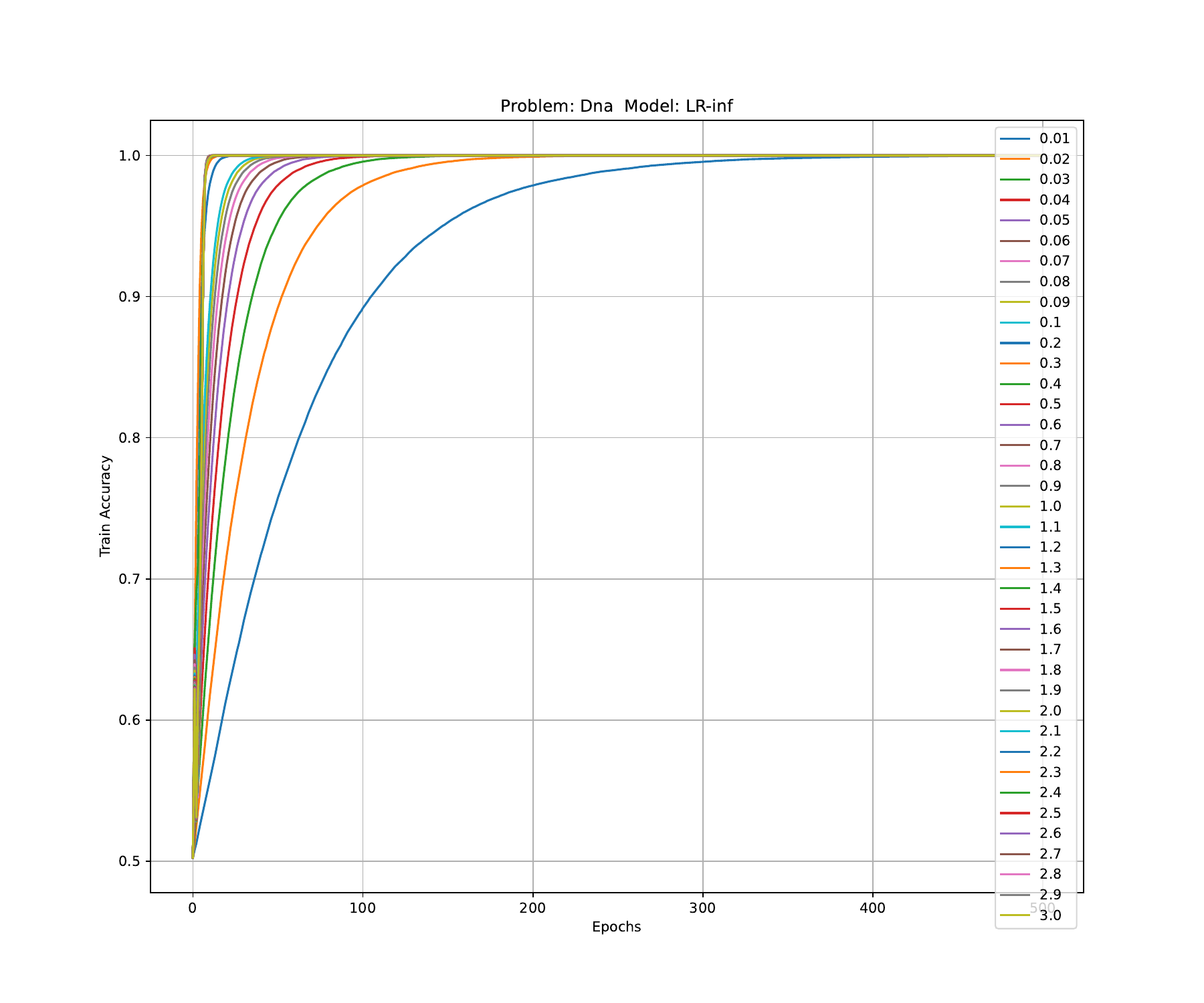} &
    \includegraphics[clip,trim=70 30 30 30,width=.52\linewidth]{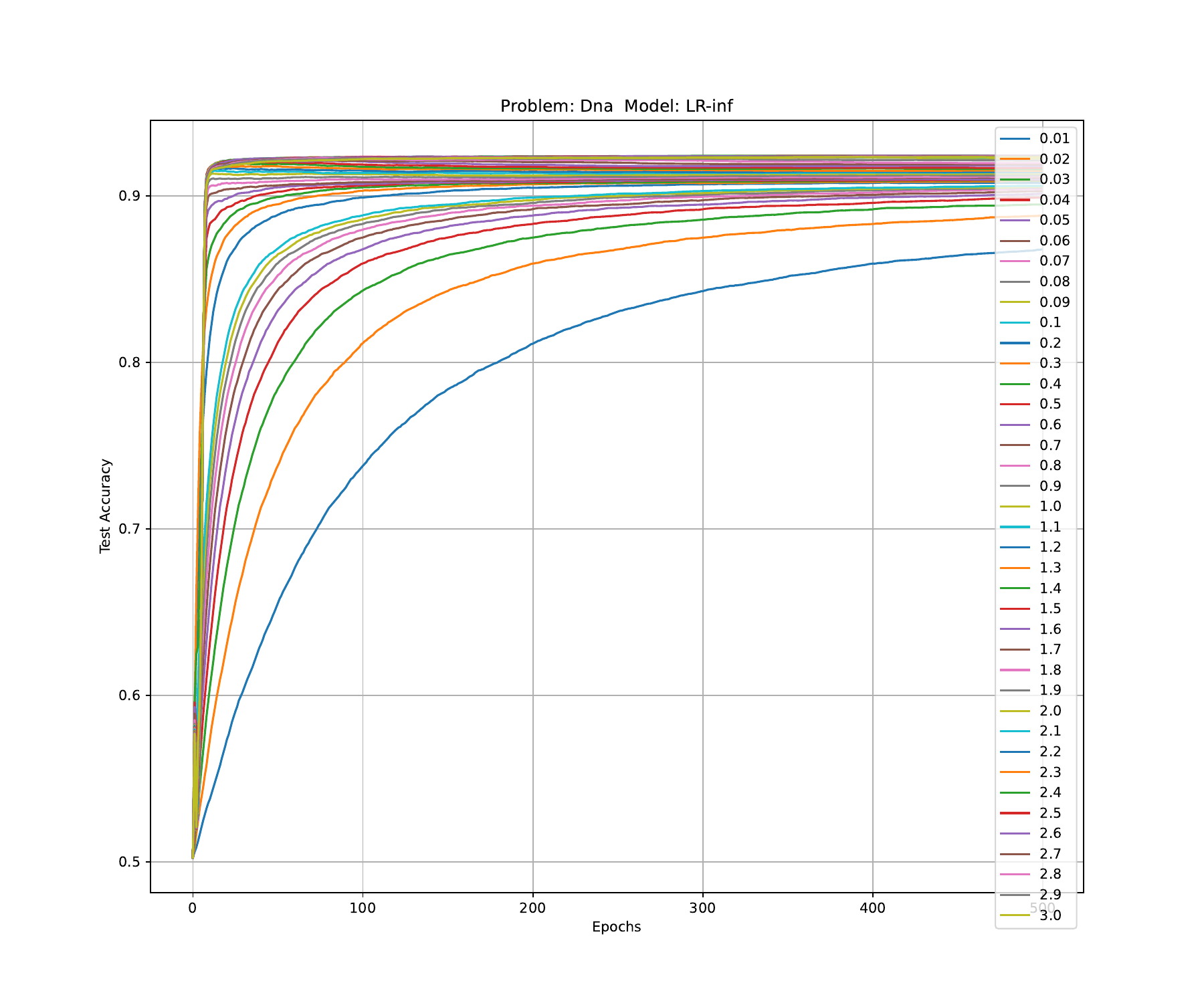} \\[-3mm]
    \includegraphics[clip,trim=70 30 30 30,width=.52\linewidth]{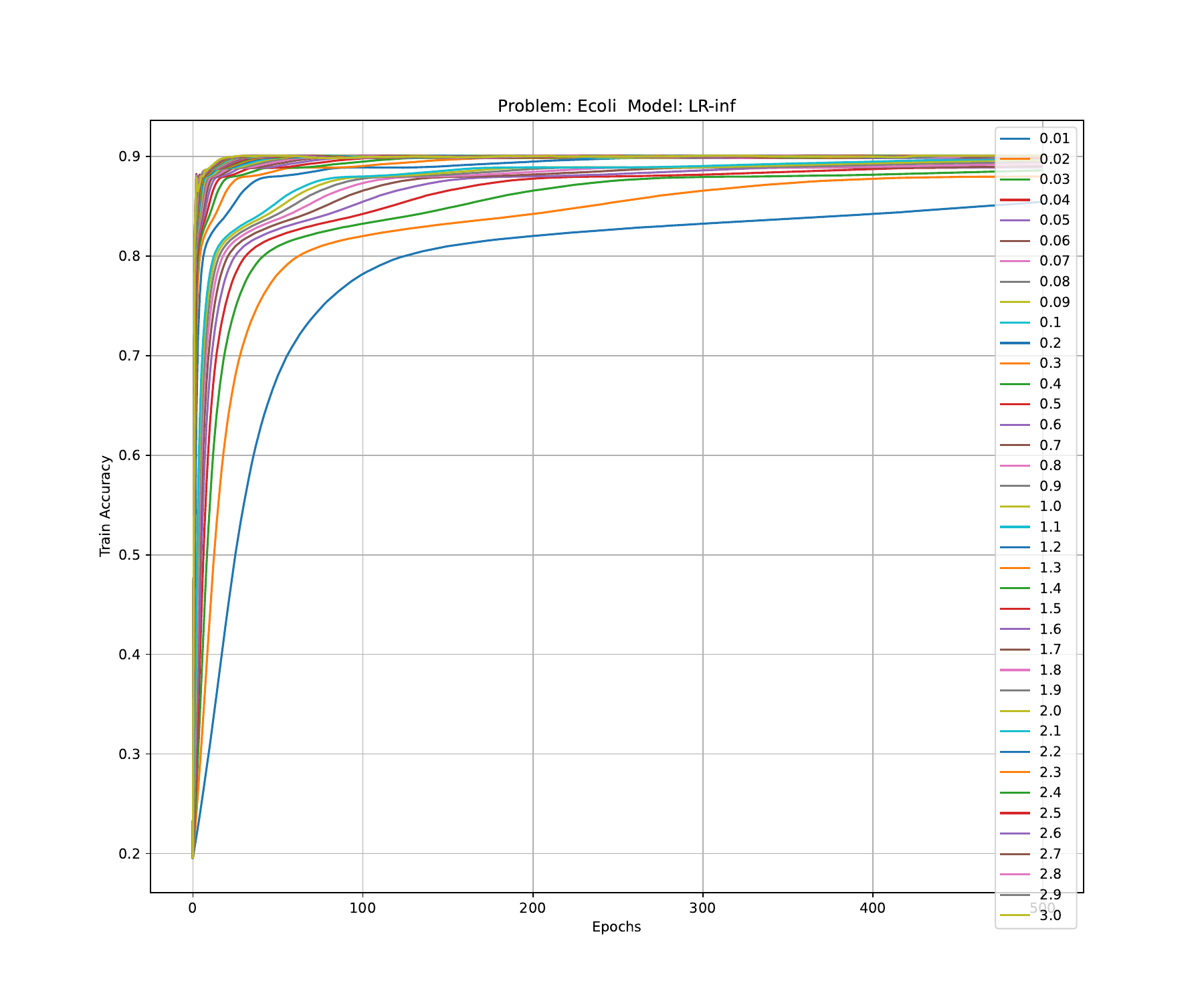} &
    \includegraphics[clip,trim=70 30 30 30,width=.52\linewidth]{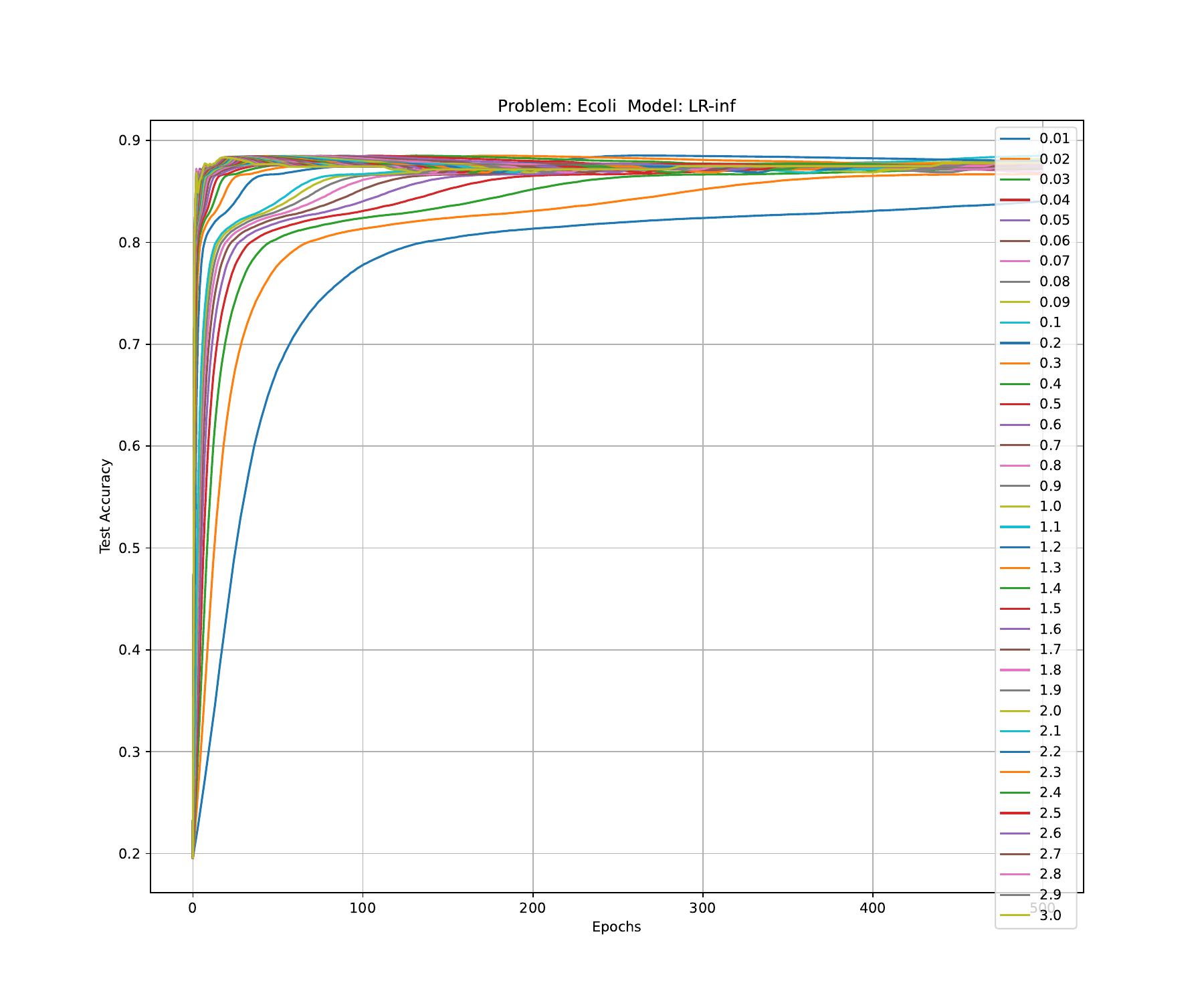} \\[-3mm]
  \end{tabular}
  \caption{Accuracies for LR model without regularisation.}
  \label{fig:accuracies_LR_model}
\end{figure*}

\begin{figure*}[t]
  \centering
  \ContinuedFloat
  \begin{tabular}{c@{}c}
    Training Accuracy & Validation Accuracy \\
    \includegraphics[clip,trim=70 30 30 30,width=.52\linewidth]{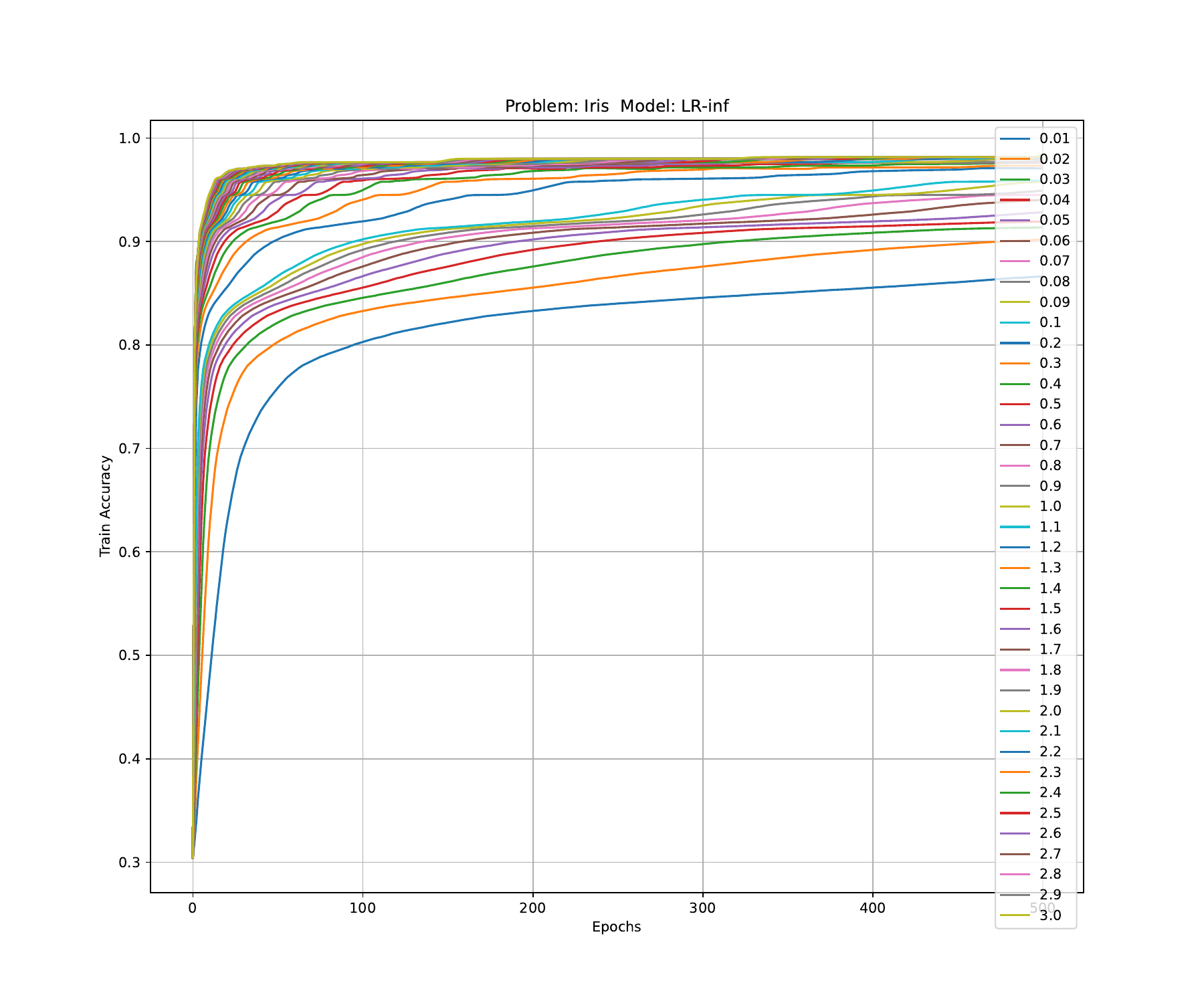} &
    \includegraphics[clip,trim=70 30 30 30,width=.52\linewidth]{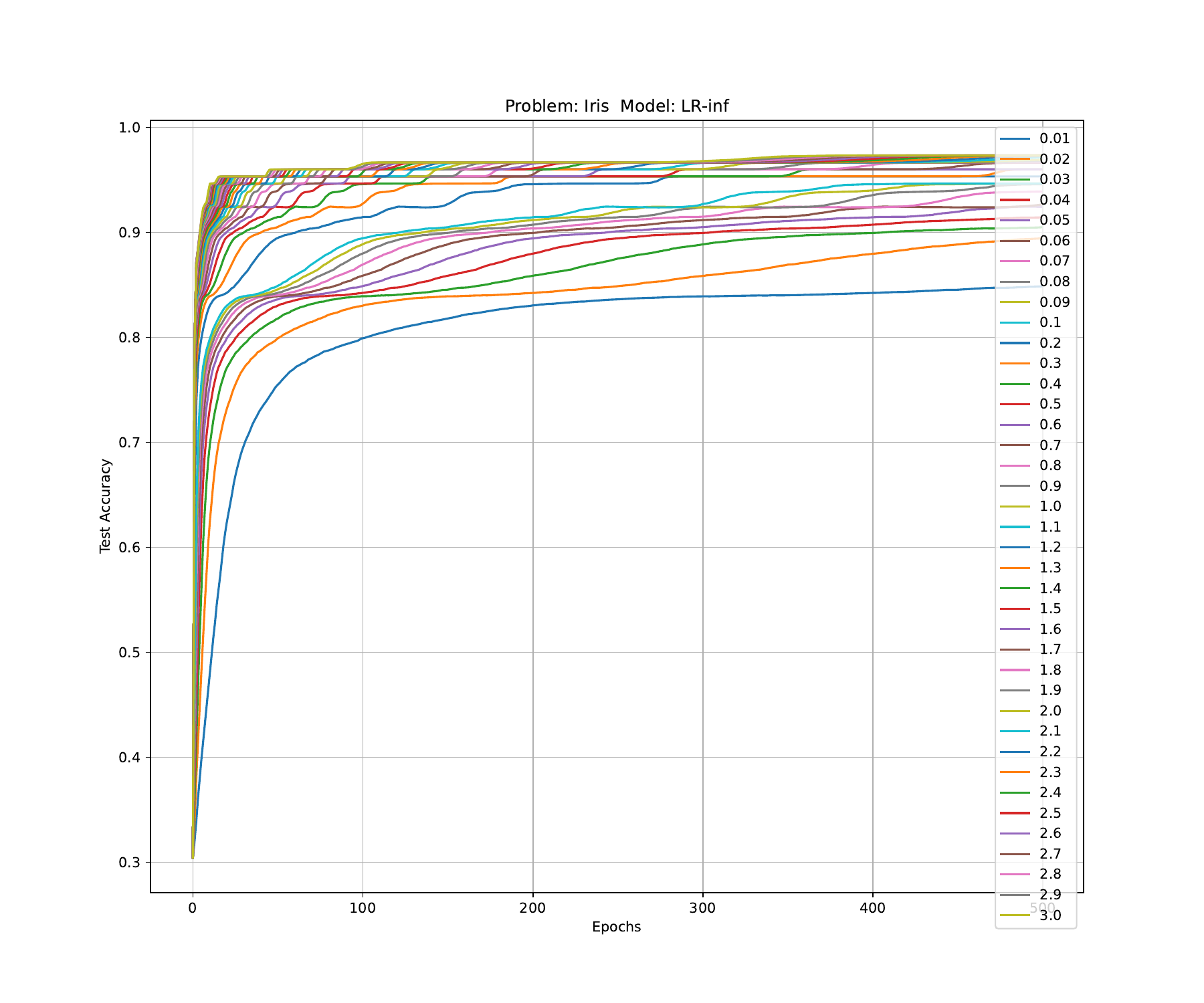} \\[-3mm]
    \includegraphics[clip,trim=70 30 30 30,width=.52\linewidth]{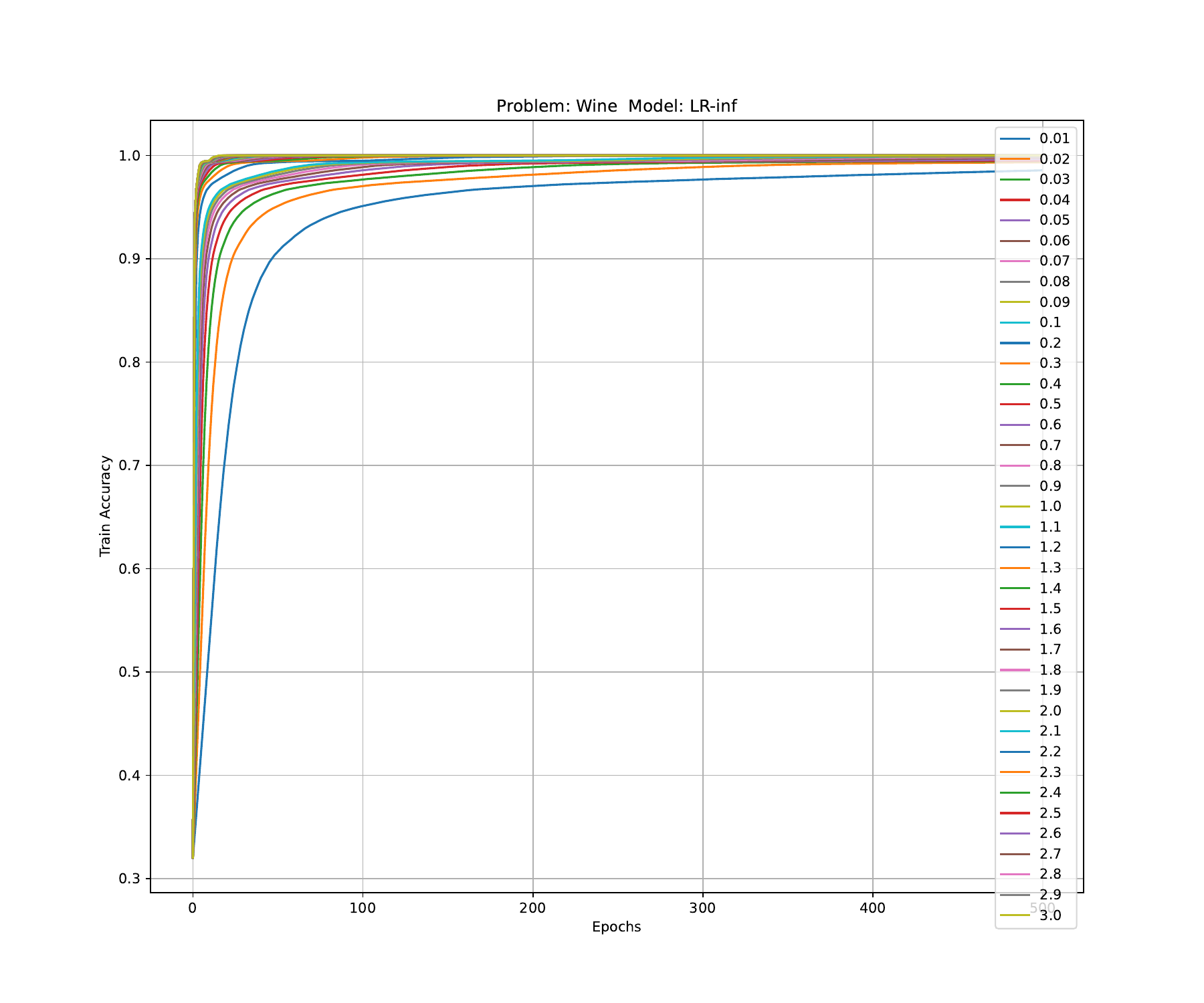} &
    \includegraphics[clip,trim=70 30 30 30,width=.52\linewidth]{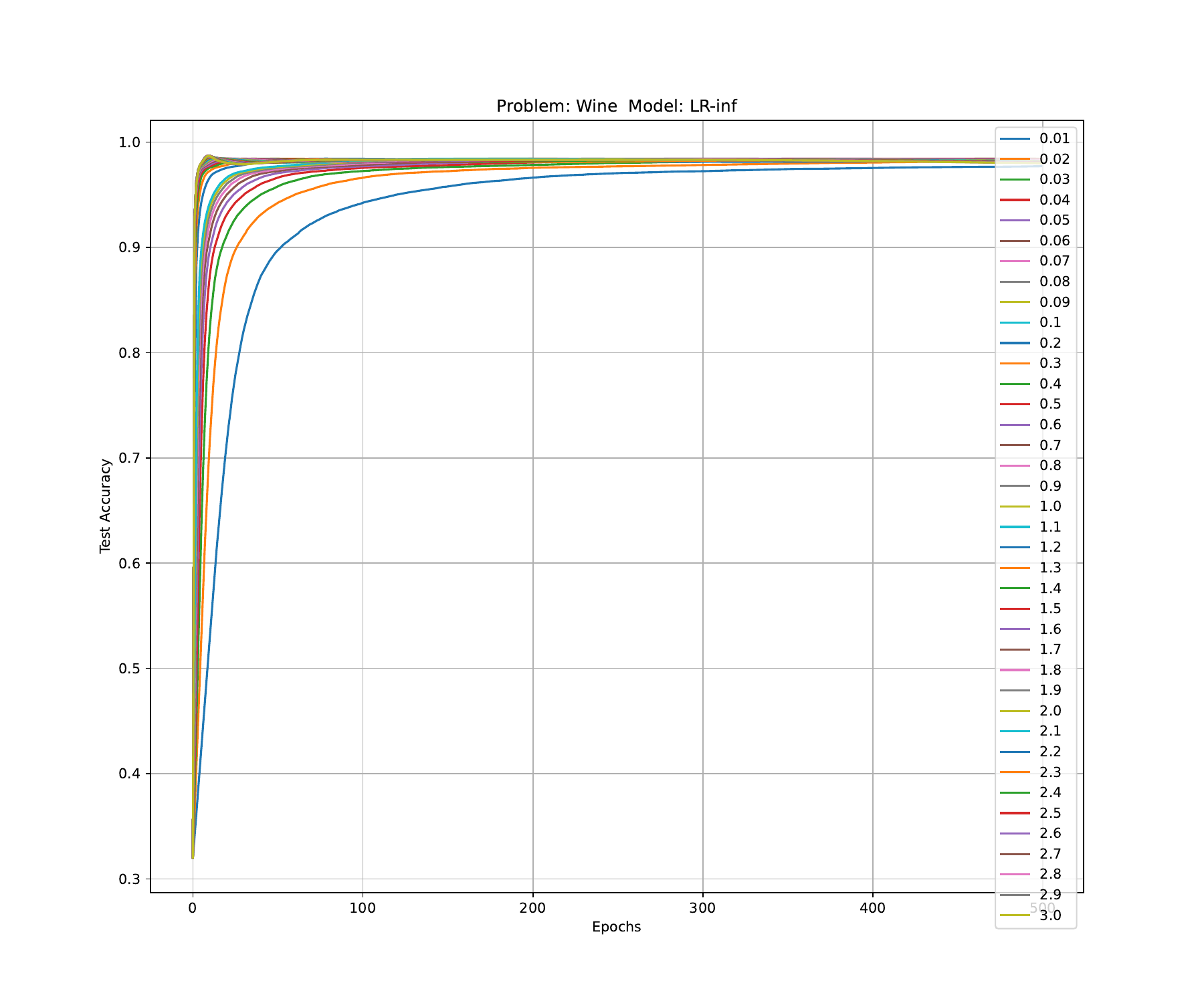} \\[-3mm]
  \end{tabular}
  \caption{Accuracies for LR model without regularisation (continued).}
\end{figure*}

\begin{figure*}[p]
  \centering
  \begin{tabular}{c@{}c}
    Training Accuracy & Validation Accuracy \\
    \includegraphics[clip,trim=70 30 30 30,width=.52\linewidth]{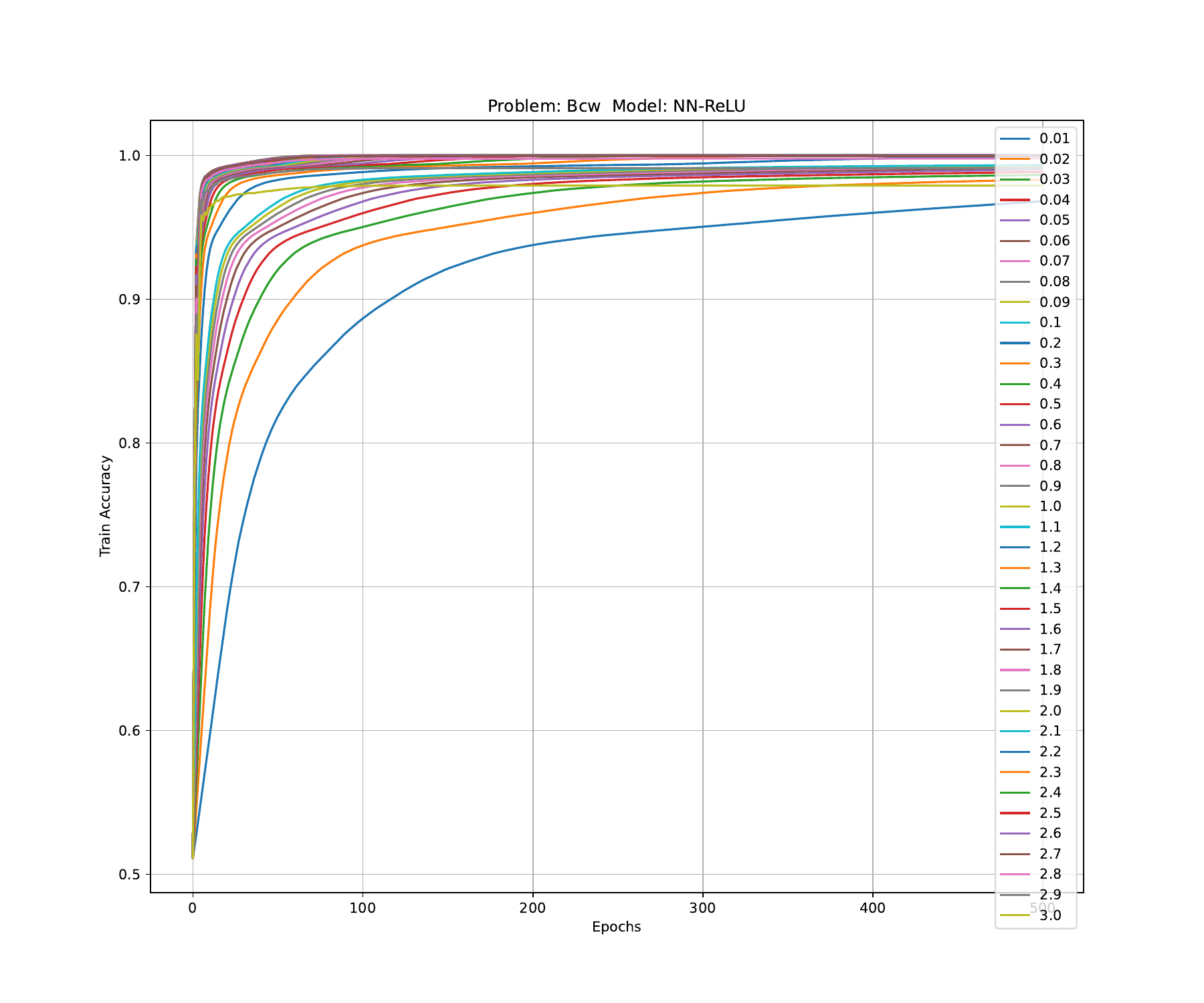} &
    \includegraphics[clip,trim=70 30 30 30,width=.52\linewidth]{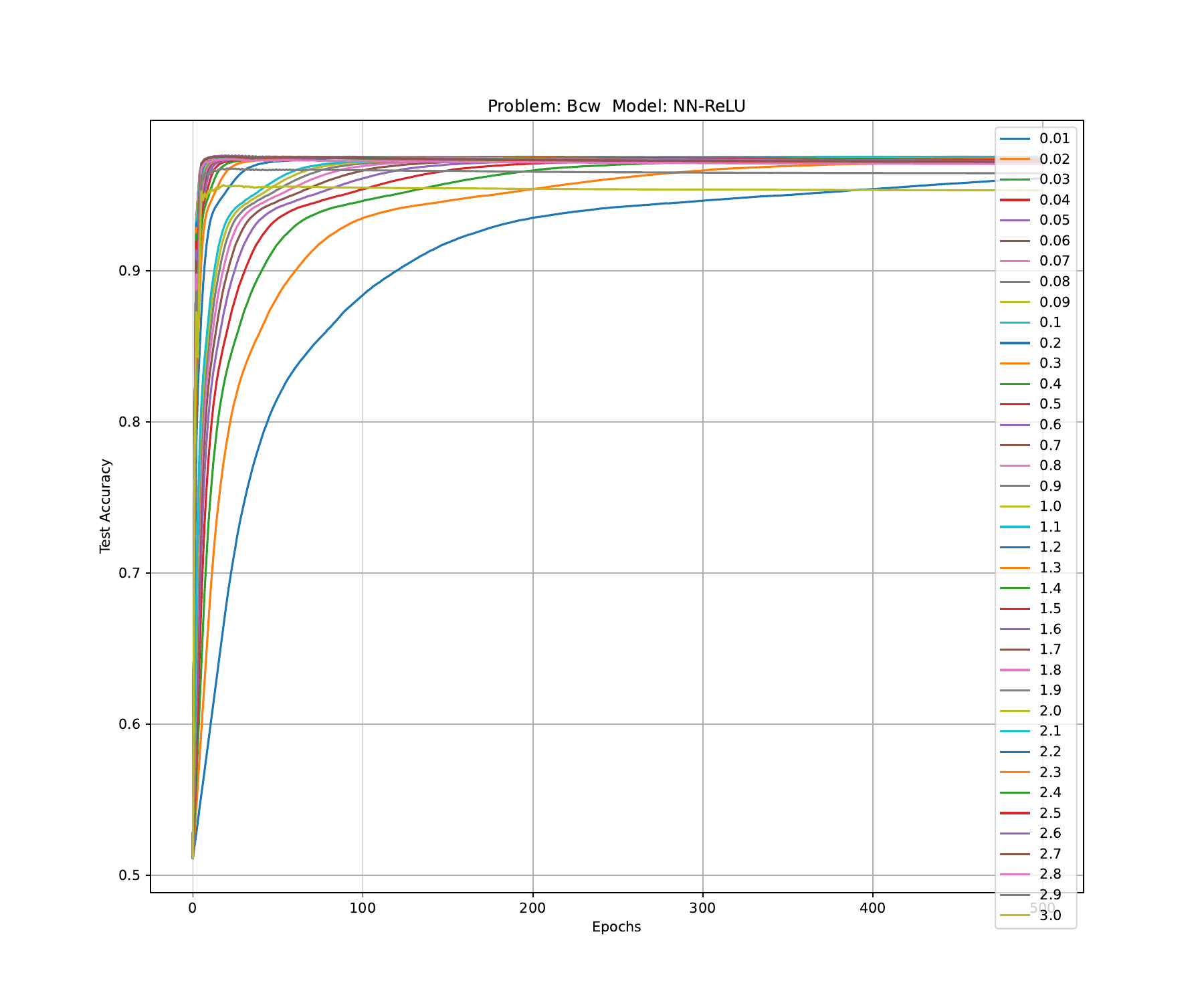} \\[-3mm]
   \includegraphics[clip,trim=70 30 30 30,width=.52\linewidth]{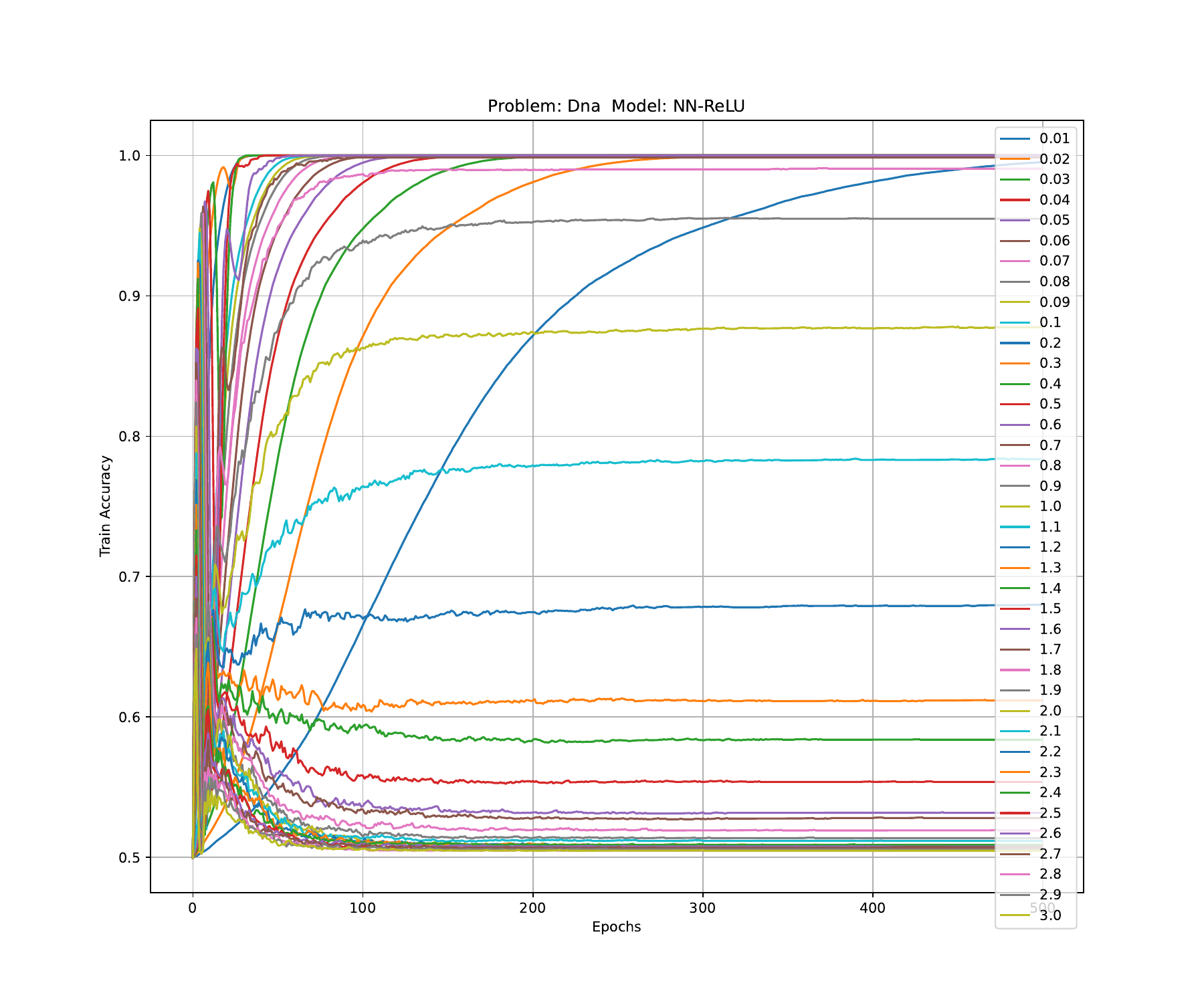} &
    \includegraphics[clip,trim=70 30 30 30,width=.52\linewidth]{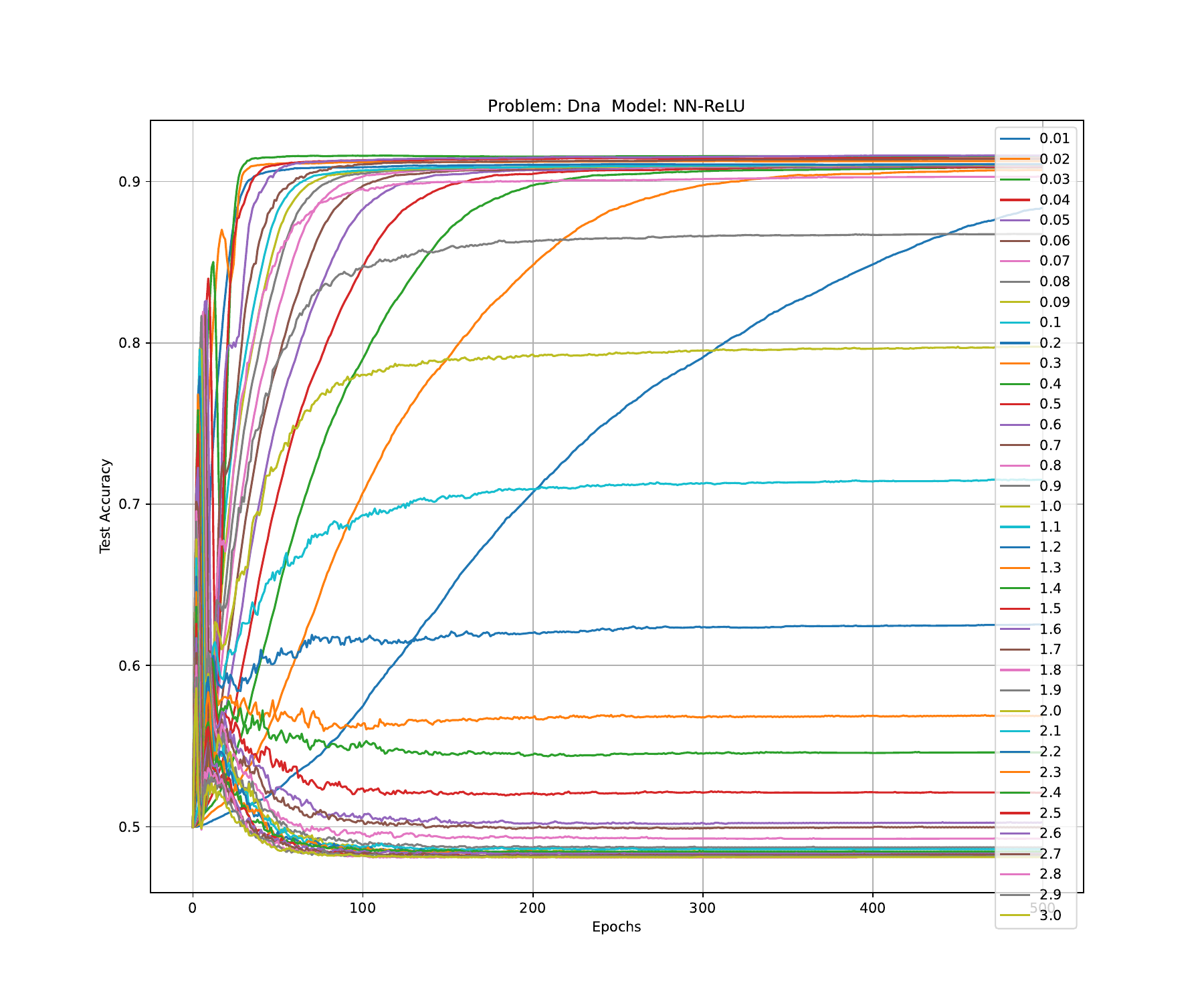} \\[-3mm]
    \includegraphics[clip,trim=70 30 30 30,width=.52\linewidth]{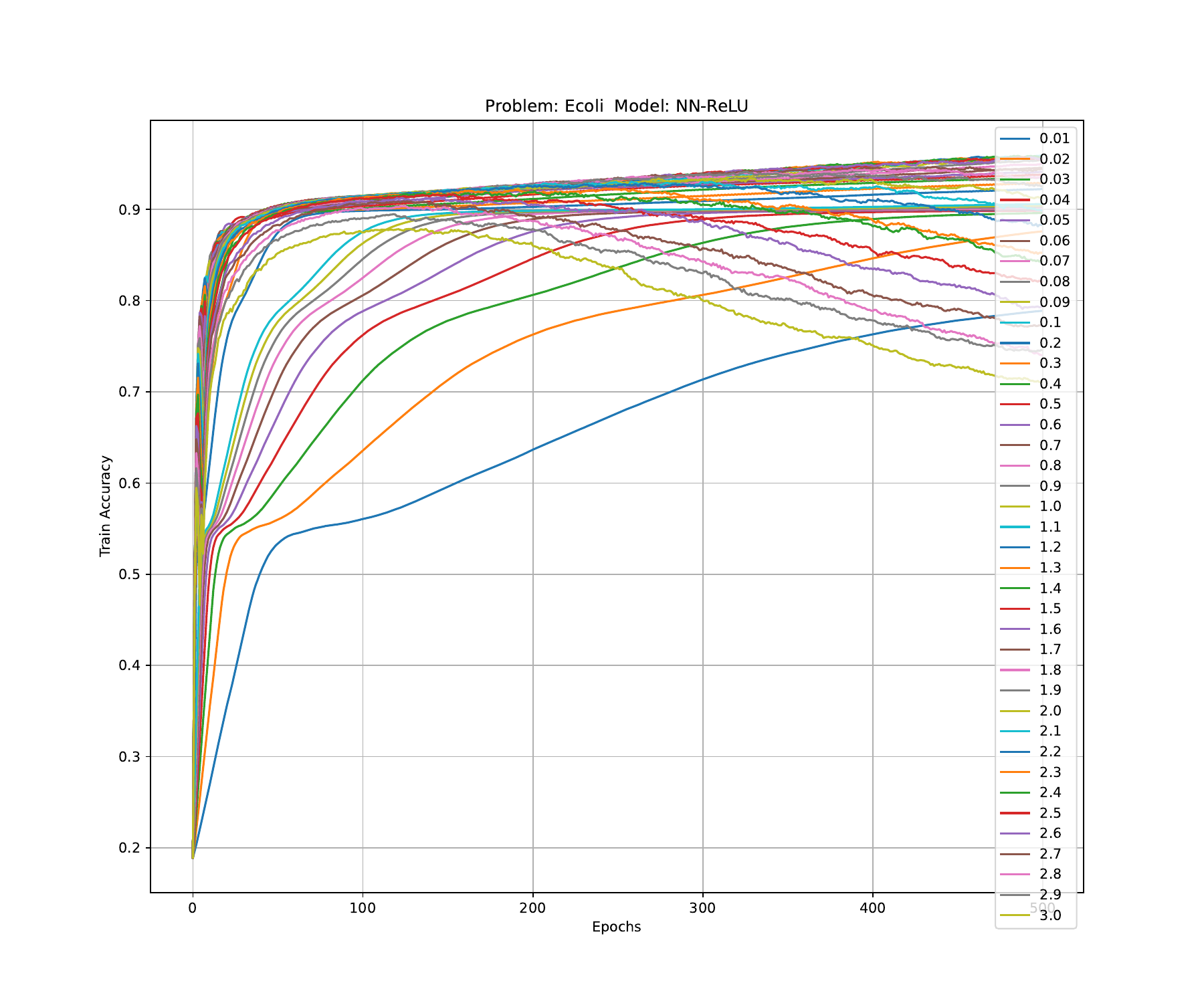} &
    \includegraphics[clip,trim=70 30 30 30,width=.52\linewidth]{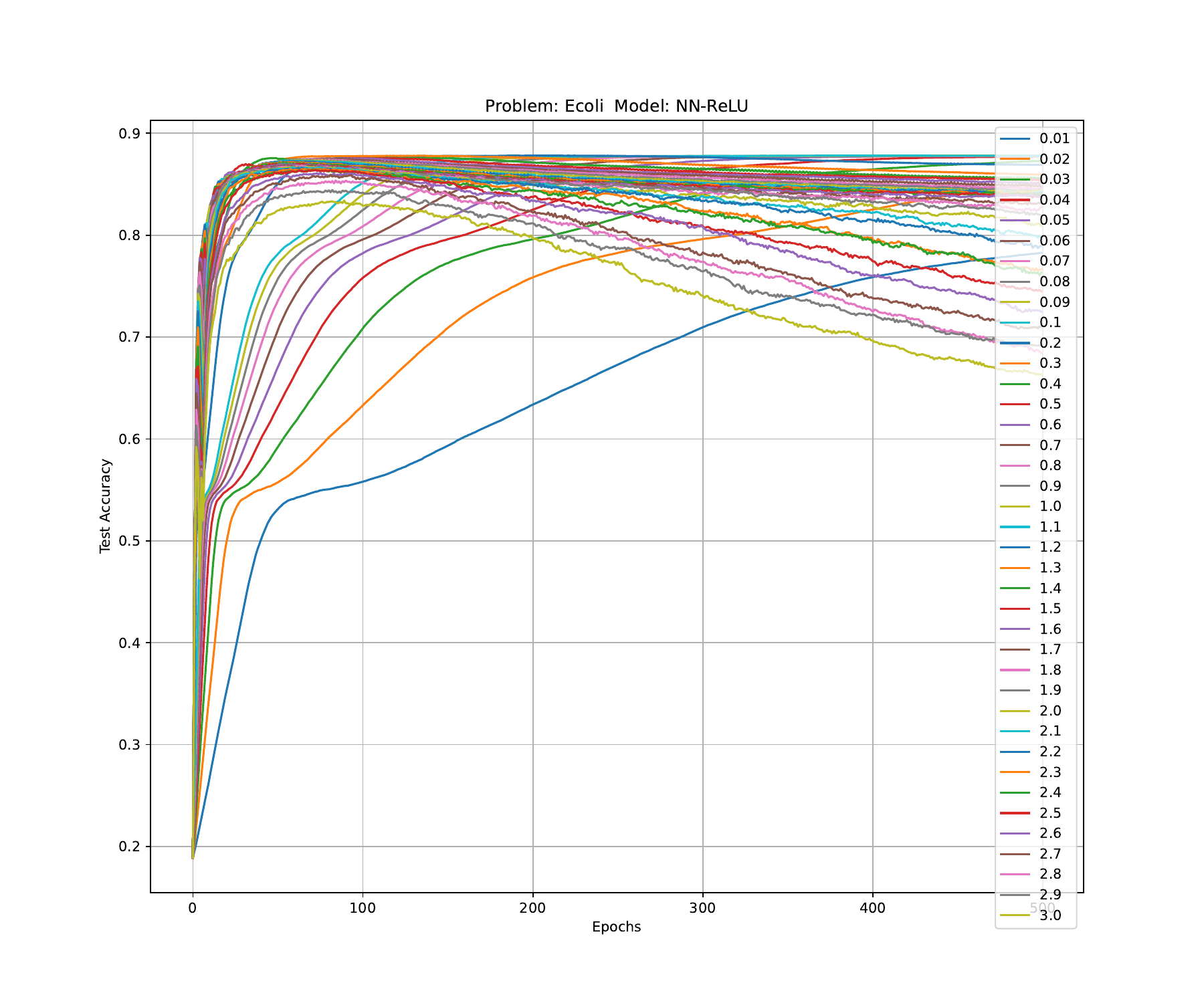} \\[-3mm]
  \end{tabular}
\caption{Accuracies for NN-ReLU model.}
  \label{fig:accuracies_NN_ReLU_model}
\end{figure*}

\begin{figure*}[t]
  \centering
  \ContinuedFloat
  \begin{tabular}{c@{}c}
    Training Accuracy & Validation Accuracy \\
    \includegraphics[clip,trim=70 30 30 30,width=.52\linewidth]{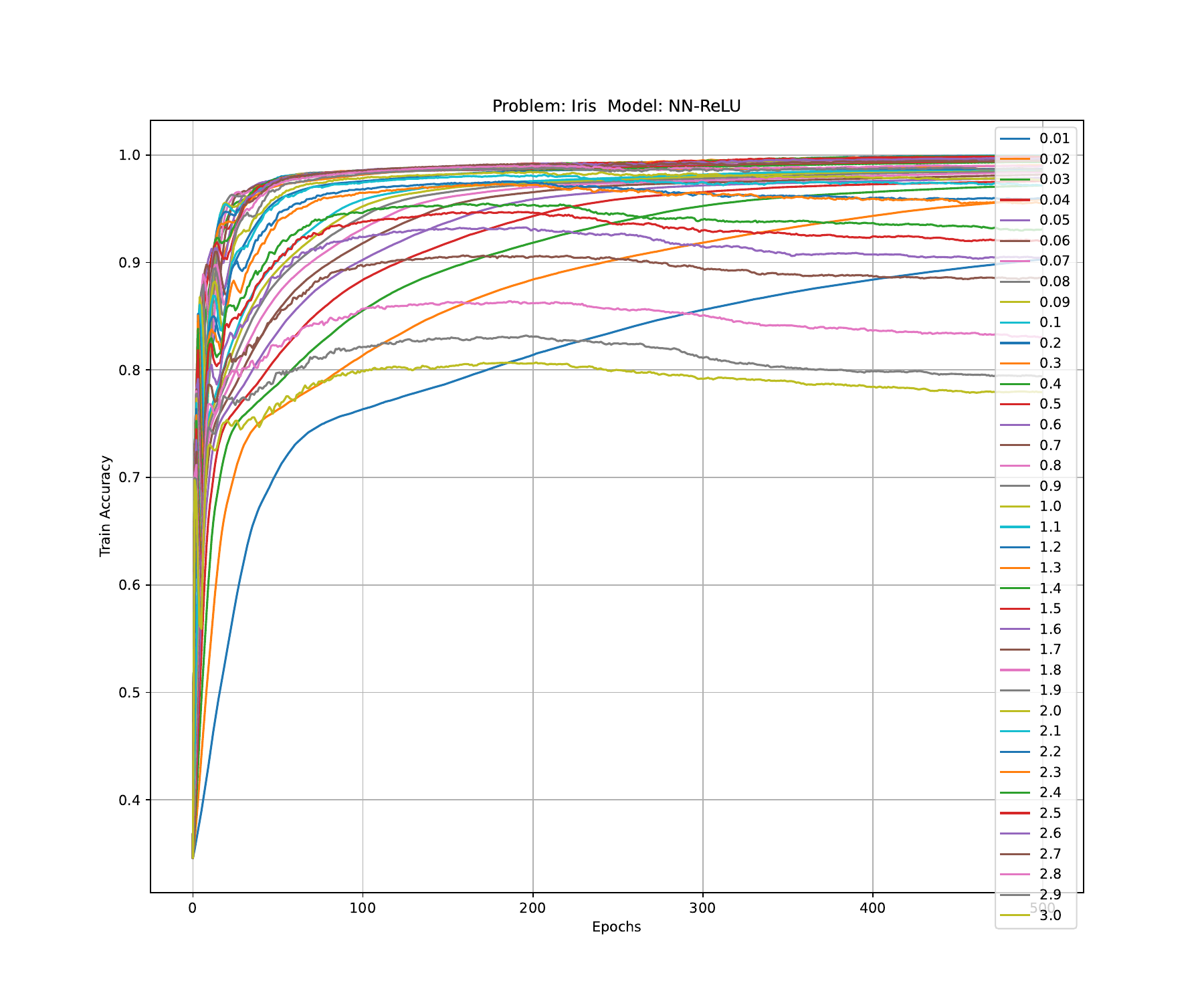} &
    \includegraphics[clip,trim=70 30 30 30,width=.52\linewidth]{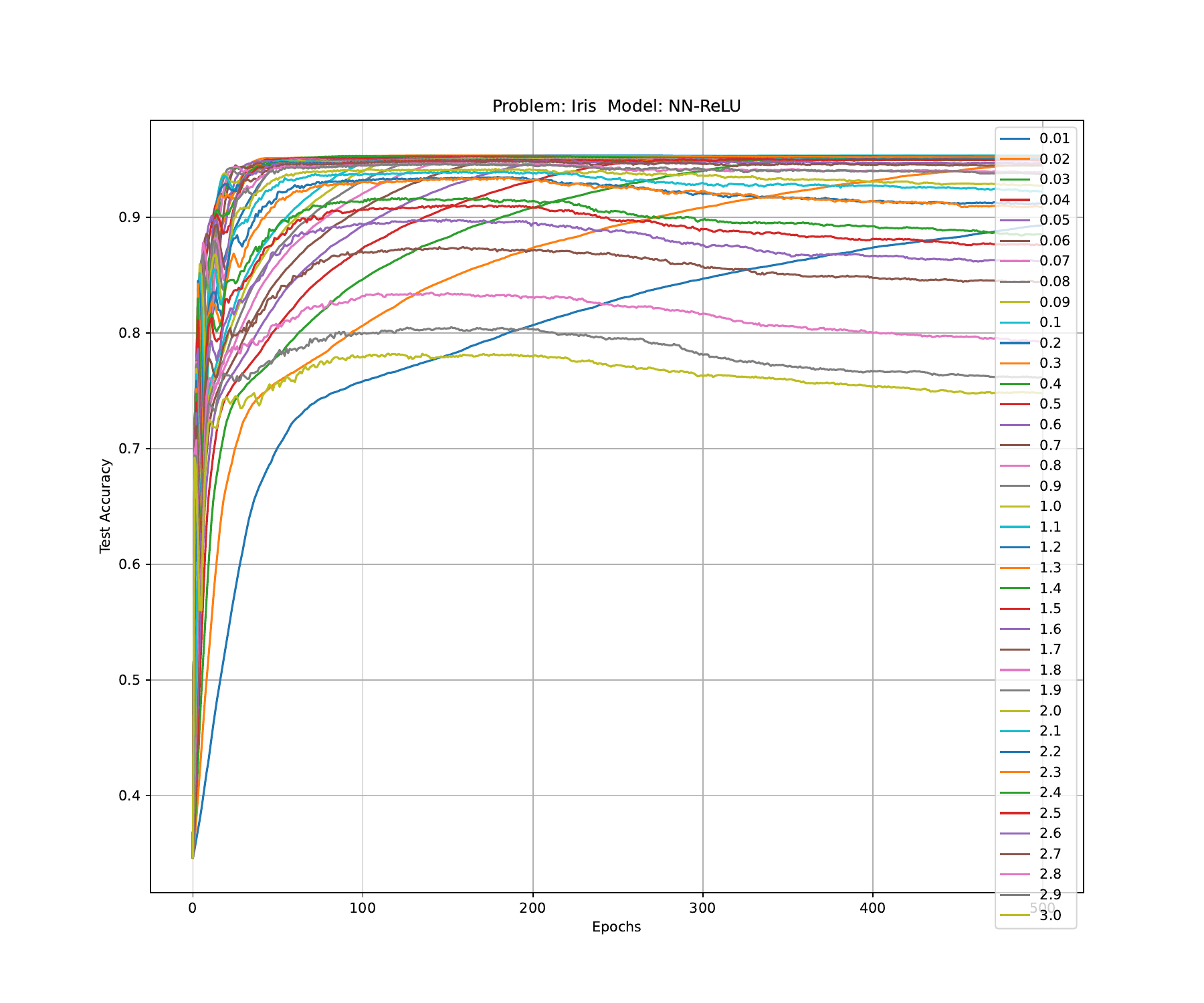} \\[-3mm]
    \includegraphics[clip,trim=70 30 30 30,width=.52\linewidth]{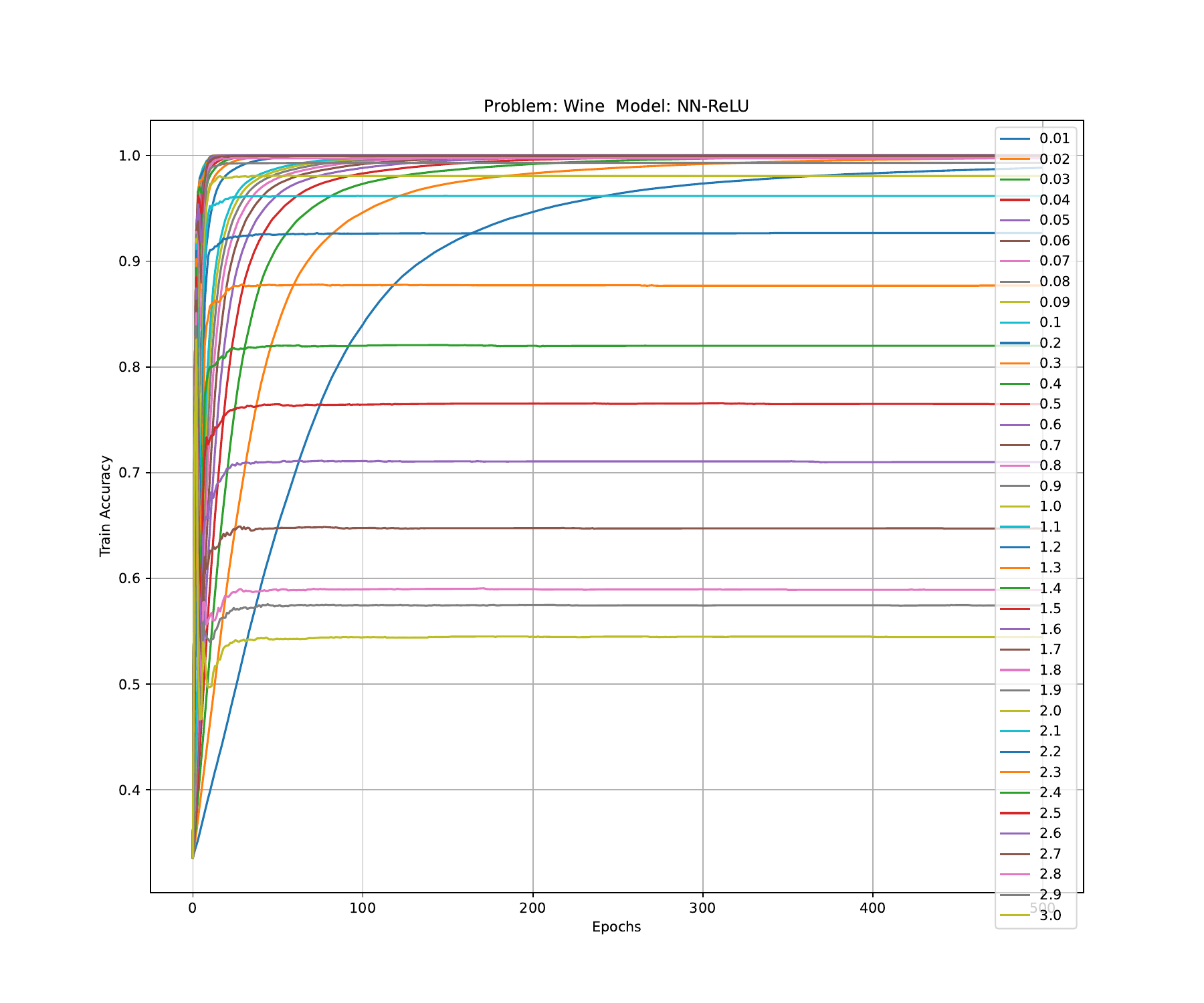} &
    \includegraphics[clip,trim=70 30 30 30,width=.52\linewidth]{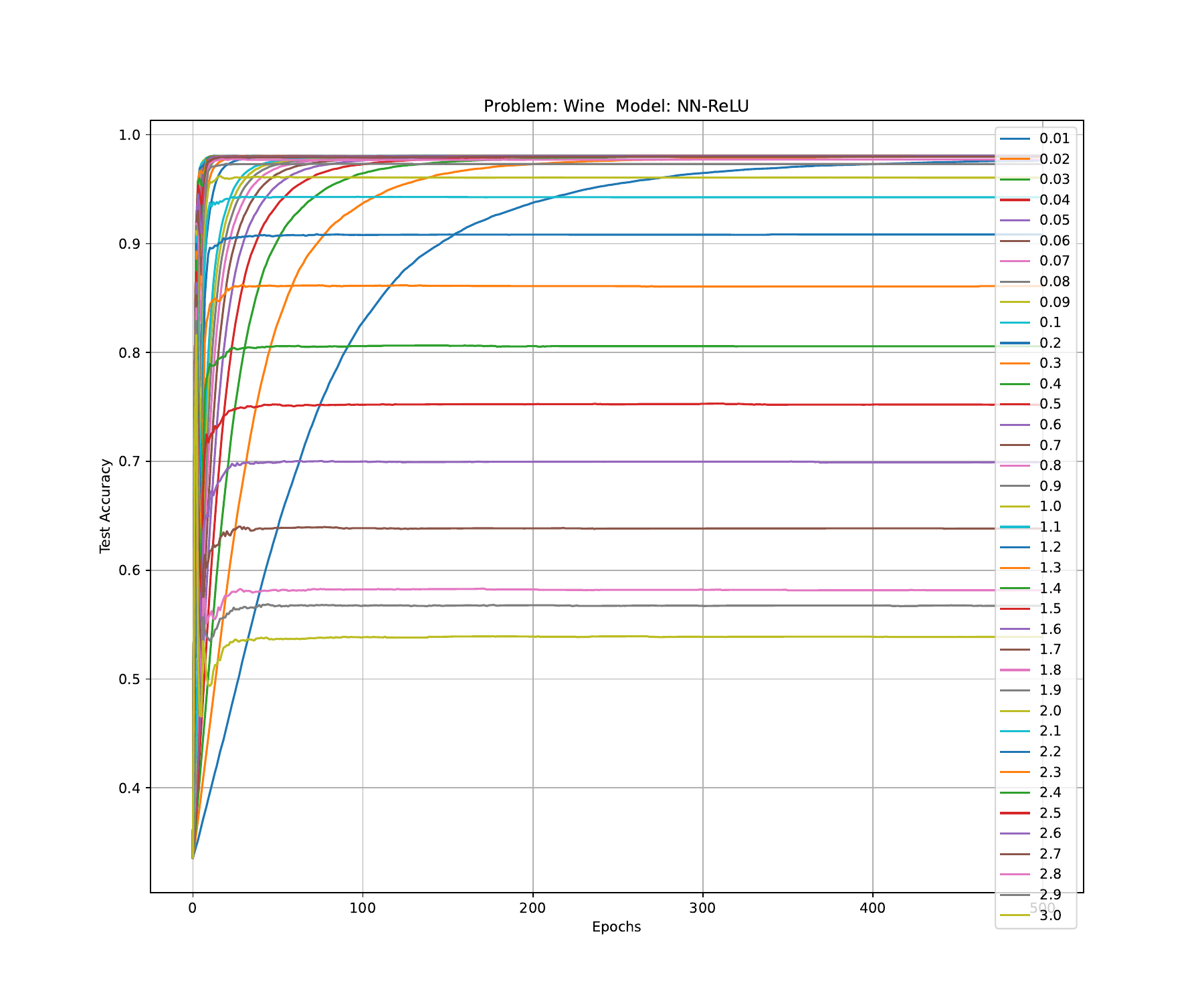} \\[-3mm]
  \end{tabular}
\caption{Accuracies for NN-ReLU model (continued).}
\end{figure*}

\medskip

Focusing first on \emph{training accuracy}, we see that for LR without regularisation
(Figure~\ref{fig:accuracies_LR_model}),
the mean accuracy trajectories are reasonably similar across problems,
all approaching an asymptote with the higher learning rates tending to
produce faster convergence towards it.  However, in many other models,
including NN-ReLU (Figure~\ref{fig:accuracies_NN_ReLU_model}), we see
a more complex dynamics, that, as discussed in
Section~\ref{sec:chosen-test-problems}, depends crucially on the
learning rate. More specifically, for smaller learning rates, we still
see a monotonic increase eventually saturating to a high common value
of accuracy. However, for bigger learning rates, mean accuracies do
not grow monotonically, often having a rapid initial increase followed
by a sudden decrease, finally settling for a different asymptote
characterised by lower accuracy than the one associated with smaller
learning rates.

With some models, such as LDA, LR and SVM with regularisation (see SM,
Section~\ref{app:mean_accuracies}), certain learning rates produce
very marked oscillations in mean accuracies. In some cases the
oscillations have low frequency (e.g., see LR-0.1 model on the DNA
problem in Figure~\ref{fig:accuracies_LR_0.1_model}) while in others
have very high-frequency (e.g., see LR-0.01 model on the Iris problem
in Figure~\ref{fig:accuracies_LR_0.01_model}). The amplitude of the
oscillations ranges from very small to catastrophically large. In many
(but not all) cases, the oscillations finally stop, with accuracy
settling for a constant, but disappointingly low, level.

As we discussed in
Section~\ref{sec:char-object-funct}, for GD with large enough learning
rates one can expect oscillations in individual runs also on strictly
convex error surfaces. However, the fact that we see \emph{large
oscillations at the level of averages}, suggests that there
is significant phase-alignment in the periodic GD cycles around the
optimum of individual runs.%
\footnote{We have not explored the origin of this phase-alignment. It
  might be due to a ``synchronising'' effect associated with the shape
  of basins of attraction, to the fact that all gradient descents
  start near the origin in weight space, or some other reason.}

\medskip

Focusing now on \emph{validation accuracy}, in general, we see a very
\emph{significant similarity} between the dynamics of training accuracies and
the corresponding dynamics of validation accuracies. There is one key
difference though: \emph{the end-of-run validation accuracy is lower
  than the training accuracy}. That is, even if both accuracies have
been obtained in cross-validation, some degree of \emph{overfitting}
is always present irrespective of model, problem and learning rate
(e.g., see LR without regularisation on the DNA problem in
Figure~\ref{fig:accuracies_LR_model}), although this tends to be
smaller for some problems (e.g., Wine) than for others (e.g., DNA).

With several problems, models and $\eta$ values, even when training
accuracy is monotonically increasing, validation accuracy peaks well
before the maximum number of epochs, to later slowly slightly
decrease, indicating, not unexpectedly, an \emph{increasing degree of overfitting}. An
example of this is, of course, the highly over-parametrised DNN-ReLU (see
Figure~\ref{fig:accuracies_DNN_ReLU_model}), but it is also present in
other models, e.g., for LR-inf on the Ecoli problem 
(Figure~\ref{fig:accuracies_LR_model}).
However, \emph{for many model-problem pairs, mean validation accuracy is a
non-decreasing function of the number of epochs}.

\subsection{Success Probabilities}
\label{sec:succ-probabilities}

In this section we look at how the success probability, $P(i,a,\eta)$, estimated as
indicated in Section~\ref{sec:impl-grad-steps}, varies as the accuracy
threshold $a$ (for success) is varied between 0.75 and 1.0 in steps of
0.005. We tested two cases: (1) when the accuracy threshold is applied
  on training accuracy resulting in a \emph{training success
  probability}, and (2) when the threshold is applied to the
  validation accuracy resulting in a \emph{validation success
  probability}.

Figures~\ref{fig:success_probability_LR_Inf_model_0.9}
and~\ref{fig:success_probability_NN_ReLU_model_0.9} show how the success
probabilities, for an accuracy-threshold $a=0.9$, vary as a function of
epochs numbers and learning rate $\eta$ for the two models LR-Inf (i.e., without
regularisation) and
NN-ReLU. Section~\ref{app:succ_probabilities} of SM reports full
results for all other models and a representative sample of values of
$a$; specifically, $a\in\{0.85, 0.9, 0.95\}$.

In the figures we see that the \emph{success probabilities} (whether
for training or validation) are \emph{non-decreasing functions}.  This
is not surprising as, as explained in
Section~\ref{sec:hyperp-optim-ace}, $P(i,a,\eta)$ is the probability
of solving a problem (reaching accuracy $a$) by epoch $i$ and not at
exactly epoch $i$.

Naturally, the dynamics of $P(i,a,\eta)$ is determined by the upper
tail of $f_{\mathcal{A}'_{i, \eta}}(x)$, the pdf of $\mathcal{A}'_{i,
  \eta}$, as described in Section.
For any reasonably-high threshold $a$ and for very small values of $i$
(epochs), $P(i,a,\eta)=0$ because not only
$\mathbb{E}[\mathcal{A}'_{i, \eta}]$ but the whole pdf of
$\mathcal{A}'_{i, \eta}$ is below $a$. When
$\mathbb{E}[\mathcal{A}'_{i, \eta}]$ shifts upwards enough, some of
the upper tail of the distribution is above $a$ and, so,
$P(i,a,\eta)>0$, continuing to progressively increase as $i$
increases, until \emph{$P(i,a,\eta)$ reaches a plateau}, $p(a,\eta)$, which depends on
of learning rate $\eta$ and the accuracy threshold~$a$, and, of
course, also the problem-model pair.

The specific value of the success-probability plateau $p(a,\eta)$ is important  as
if $p(a,\eta)\ge z$ ($z$ being the confidence parameter, which we set
to 0.99 in our experiments), that implies that a single run is
almost certain  to achieve an accuracy $a$. 

In Section~\ref{sec:mean_train_valid}, we discussed cases where mean
accuracies $\mathbb{E}[\mathcal{A}_{i, \eta}]$ grow very high (in many
cases eventually reaching 100\%). Because it is always the case that
$\mathbb{E}[\mathcal{A}'_{i, \eta}] \ge \mathbb{E}[\mathcal{A}_{i,
  \eta}]$, in such cases we should expect the plateau for the success
probabilities to be close to, or even reach, 100\%. In other cases, we
saw that $\mathbb{E}[\mathcal{A}_{i, \eta}]$ never reached values
close to $a$ (whether in its initial increases or late plateauing). In
such cases, we should expect success probabilities to plateau at
values less than 100\%.
Indeed, we see both these situations in
Figures~\ref{fig:success_probability_LR_Inf_model_0.9}
and~\ref{fig:success_probability_NN_ReLU_model_0.9} (and
Section~\ref{app:succ_probabilities} of SM).
Finally, if $a$ too high for a problem-model pair, the success
probabilities can even be zero, irrespective of the number of
epochs. For instance, for $a=0.95$ and Ecoli, $P(i,a,\eta)=0$ for all
$i$ and most models, see
Figures~\ref{fig:success_probability_NN_ReLU_model_0.95}---\ref{fig:success_probability_SVM_0.1_model_0.95}
of SM.

\begin{figure*}[p]
  \centering
  \begin{tabular}{c@{}c}
    Training Success Probability & Validation Success Probability \\
    \includegraphics[clip,trim=0 0 0 0,width=.52\linewidth]{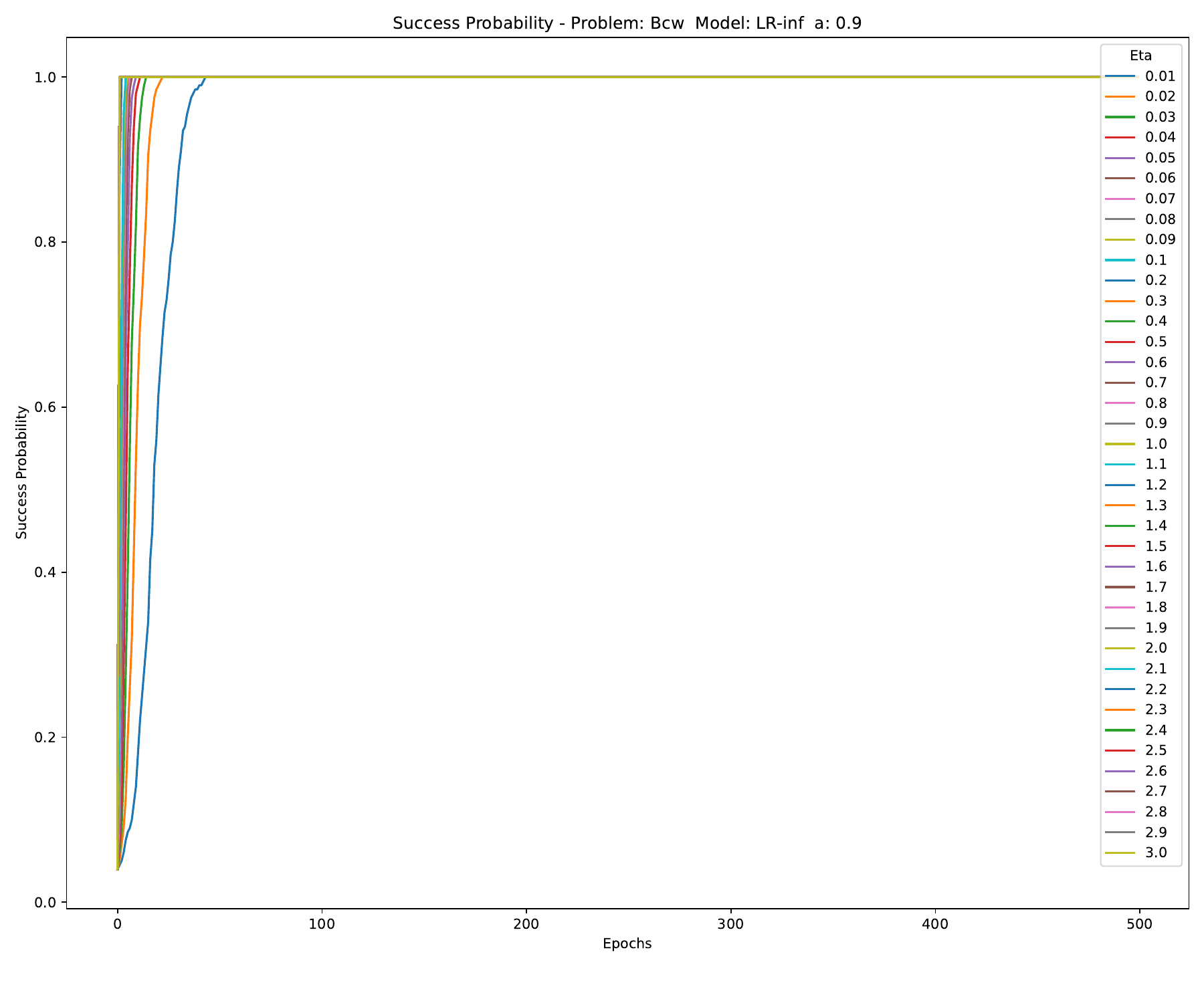} &
    \includegraphics[clip,trim=0 0 0 0,width=.52\linewidth]{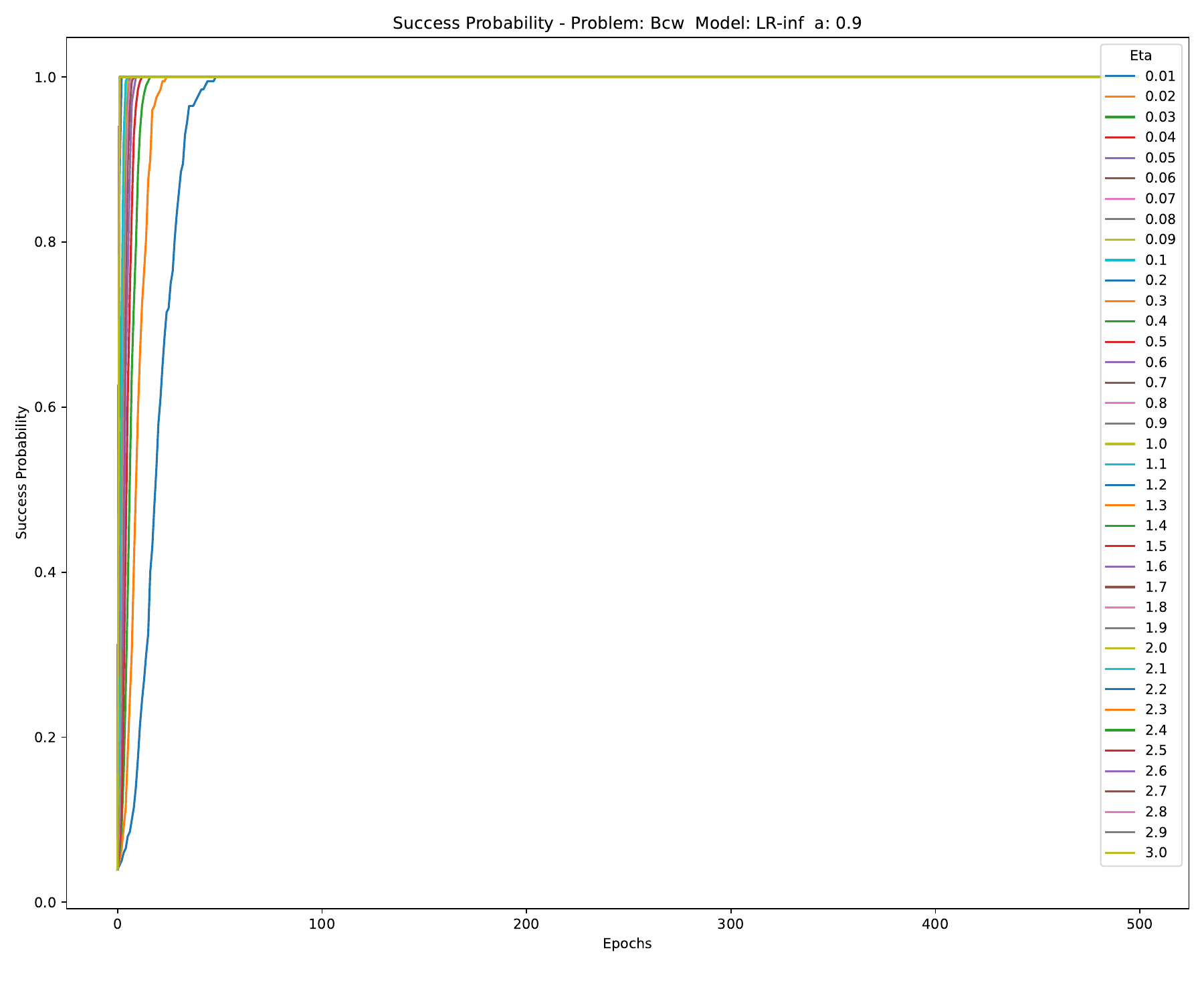} \\[-3mm]
    \includegraphics[clip,trim=0 0 0 0,width=.52\linewidth]{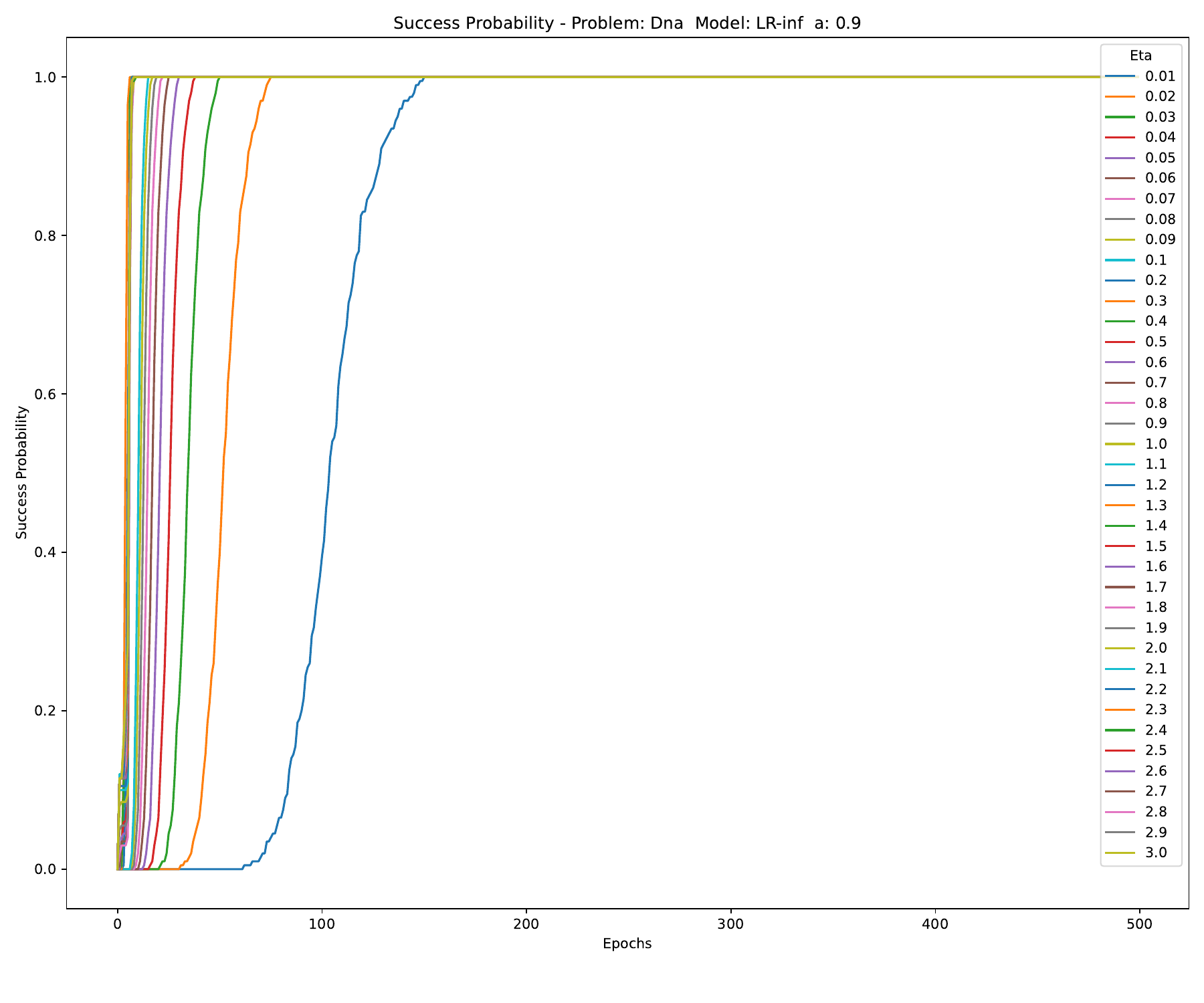} &
    \includegraphics[clip,trim=0 0 0 0,width=.52\linewidth]{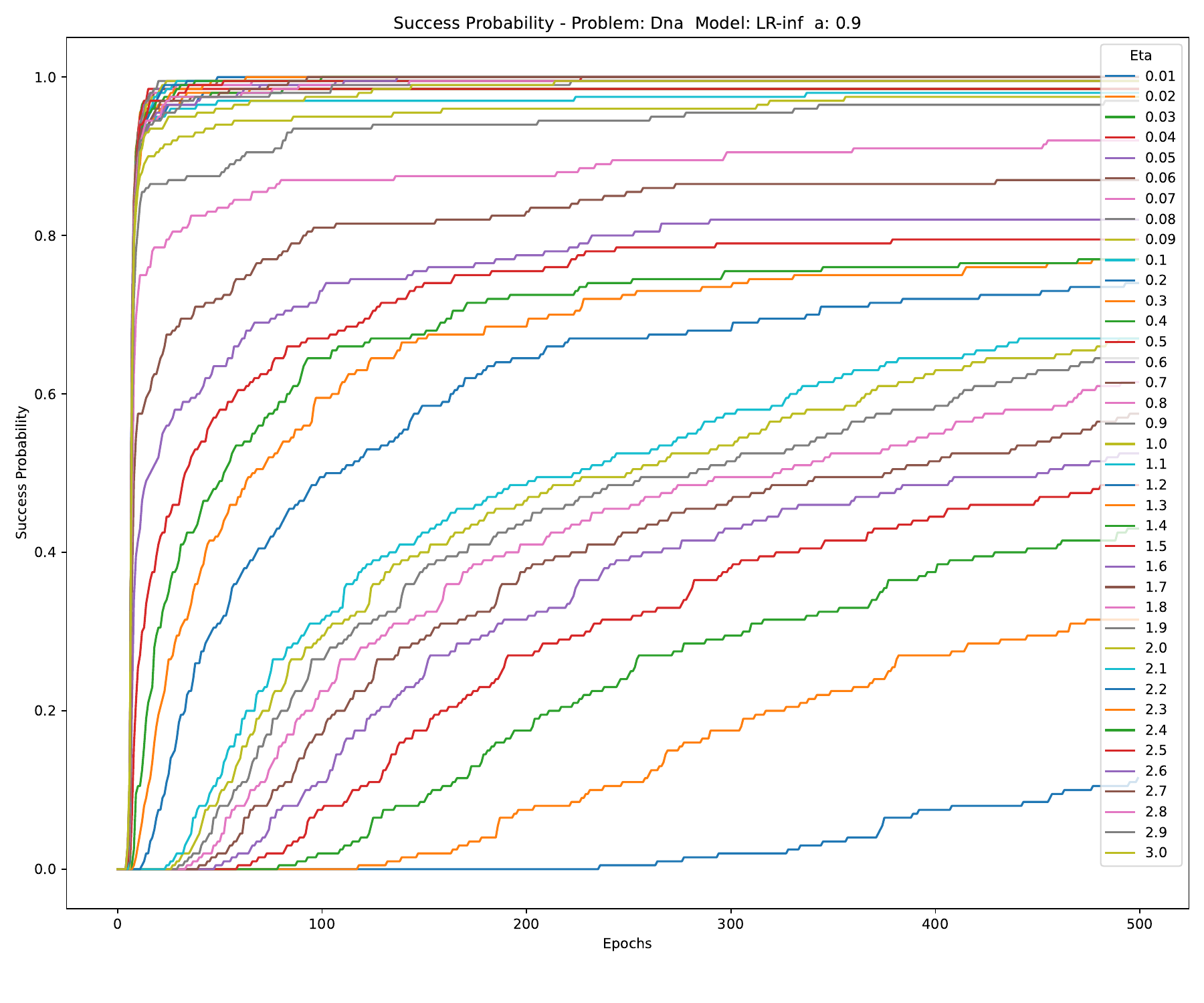} \\[-3mm]
    \includegraphics[clip,trim=0 0 0 0,width=.52\linewidth]{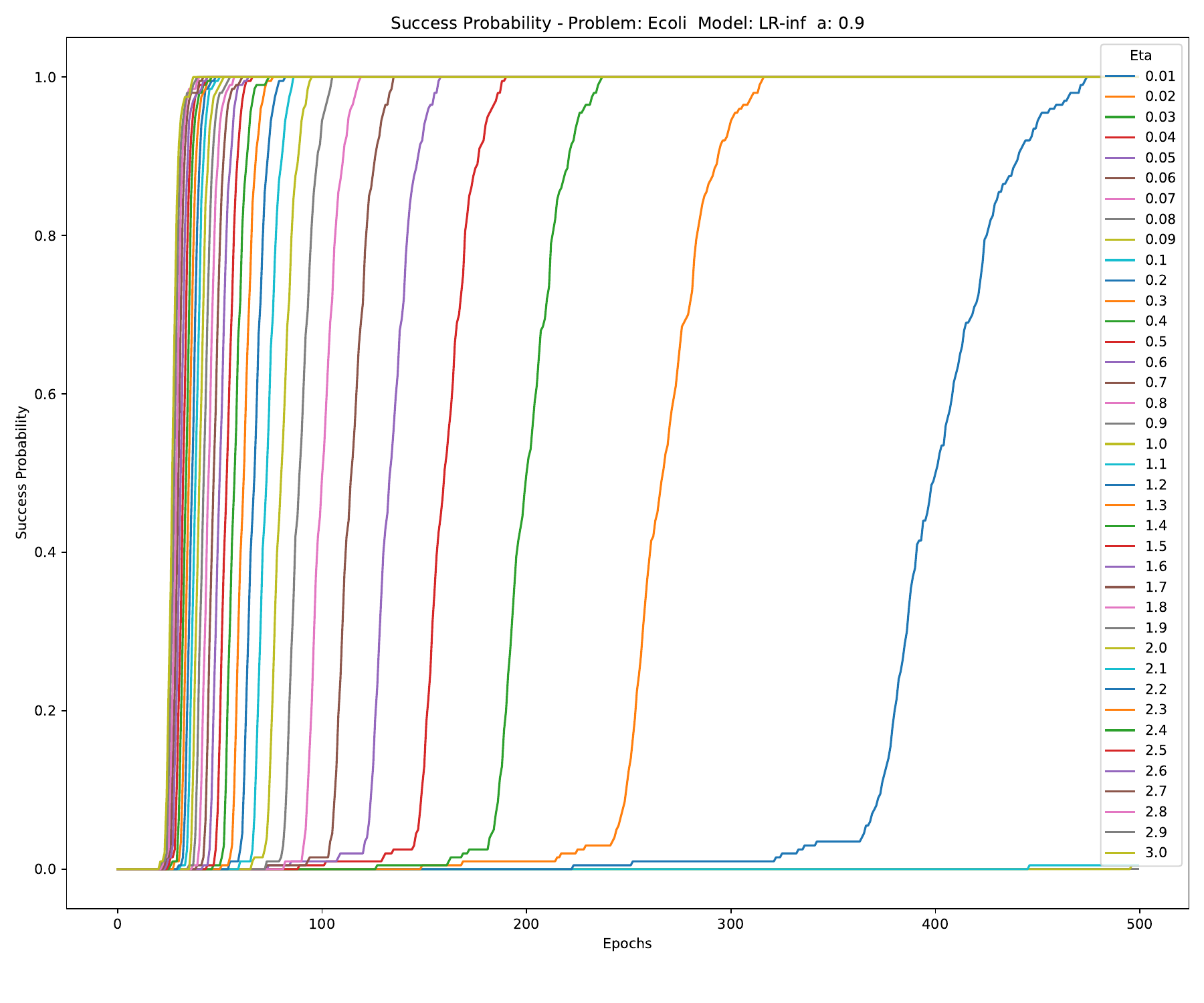} &
    \includegraphics[clip,trim=0 0 0 0,width=.52\linewidth]{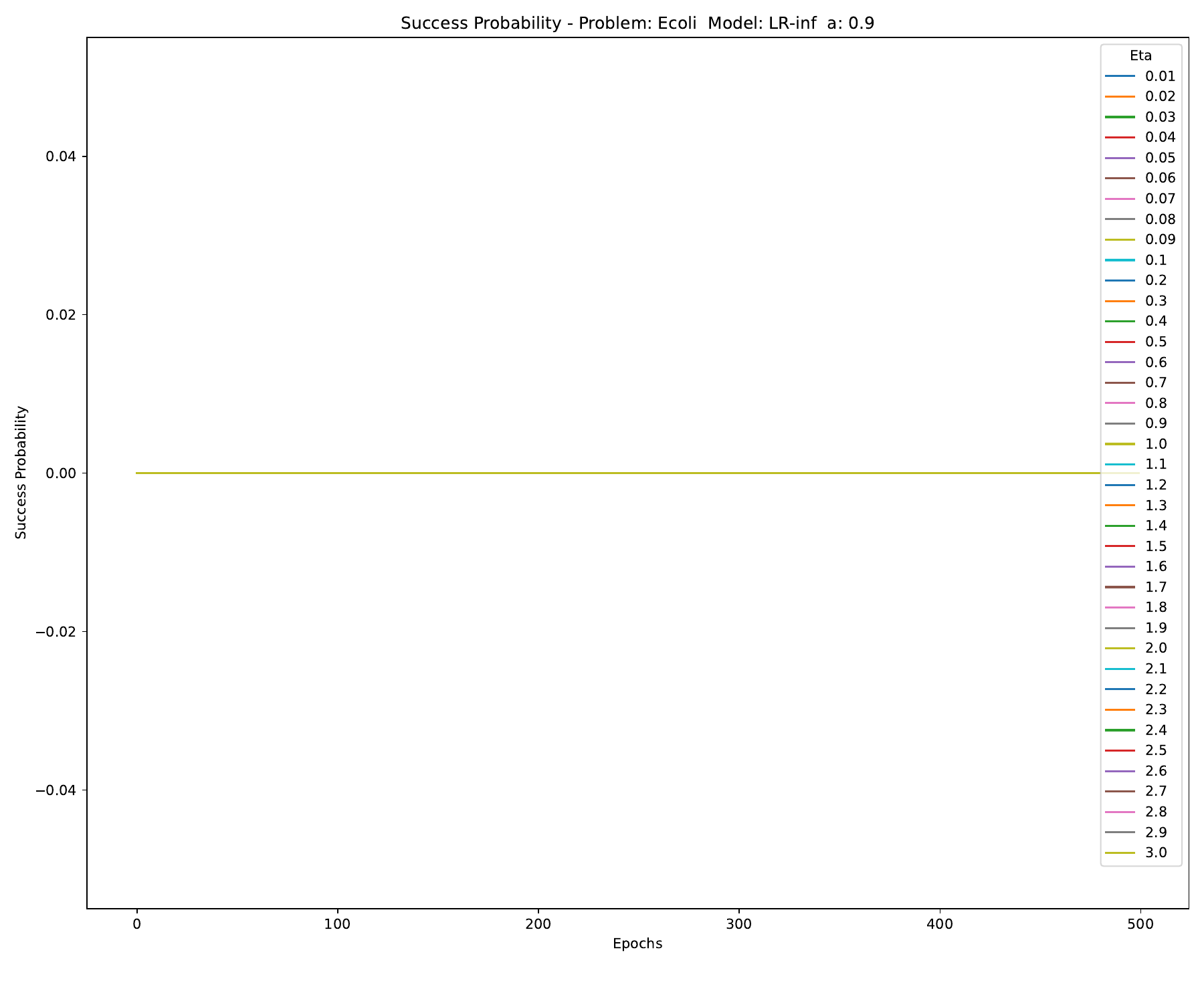} \\[-3mm]
  \end{tabular}
\caption{Success probability ($a=0.9$) for LR-Inf model.}
  \label{fig:success_probability_LR_Inf_model_0.9}
\end{figure*}

\begin{figure*}[t]
  \centering
  \ContinuedFloat
  \begin{tabular}{c@{}c}
    Training Success Probability & Validation Success Probability \\
    \includegraphics[clip,trim=0 0 0 0,width=.52\linewidth]{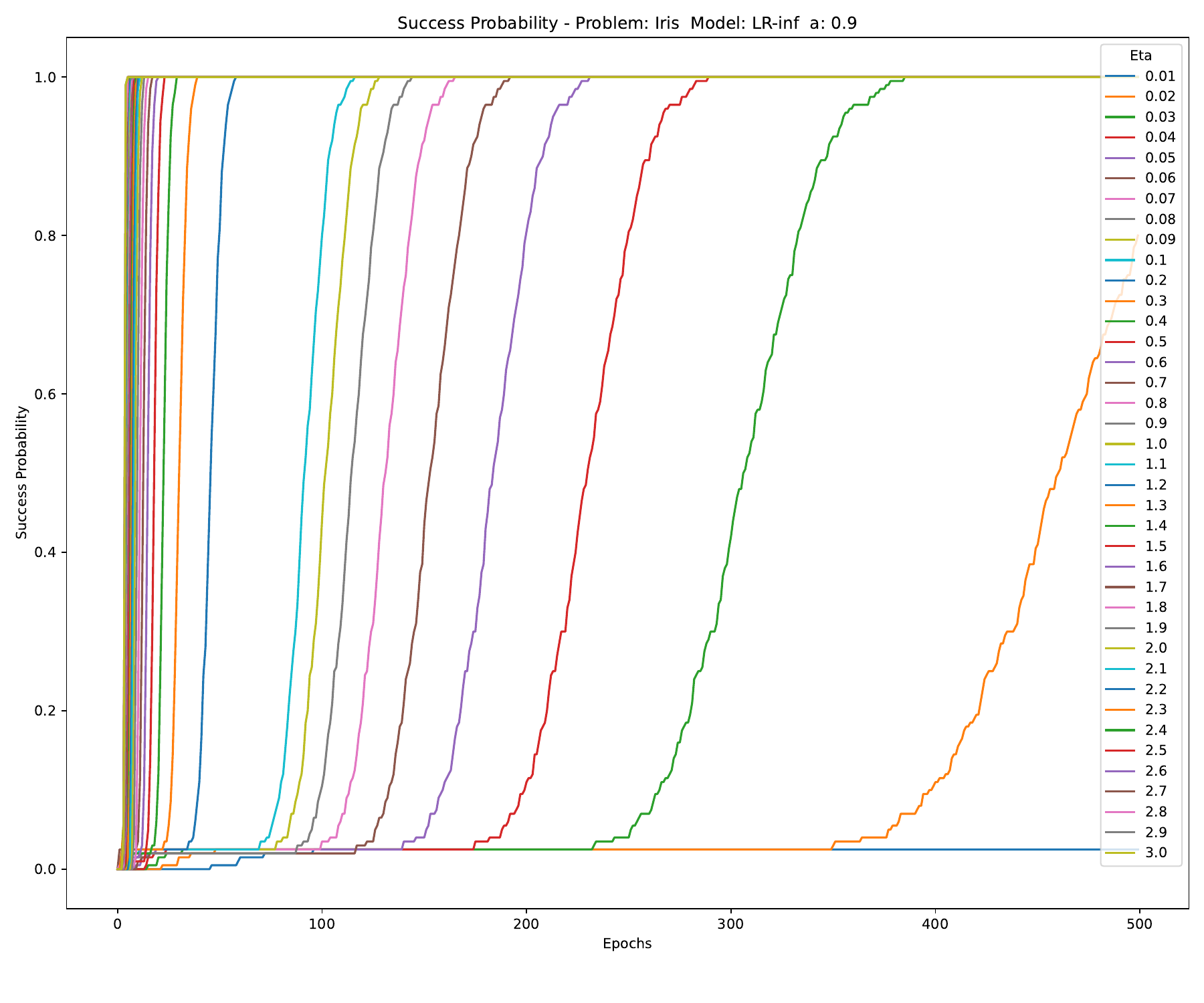} &
    \includegraphics[clip,trim=0 0 0 0,width=.52\linewidth]{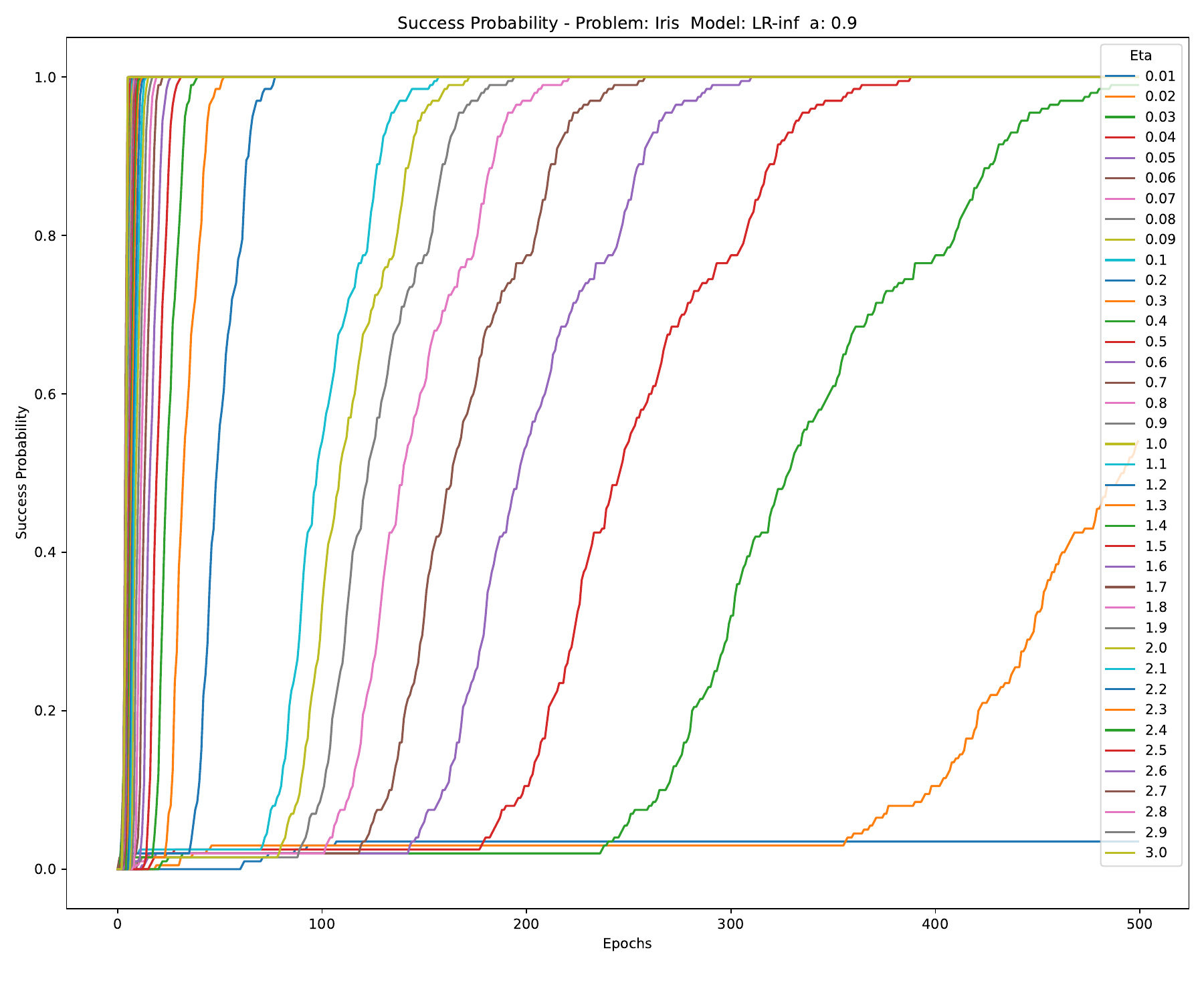} \\[-3mm]
    \includegraphics[clip,trim=0 0 0 0,width=.52\linewidth]{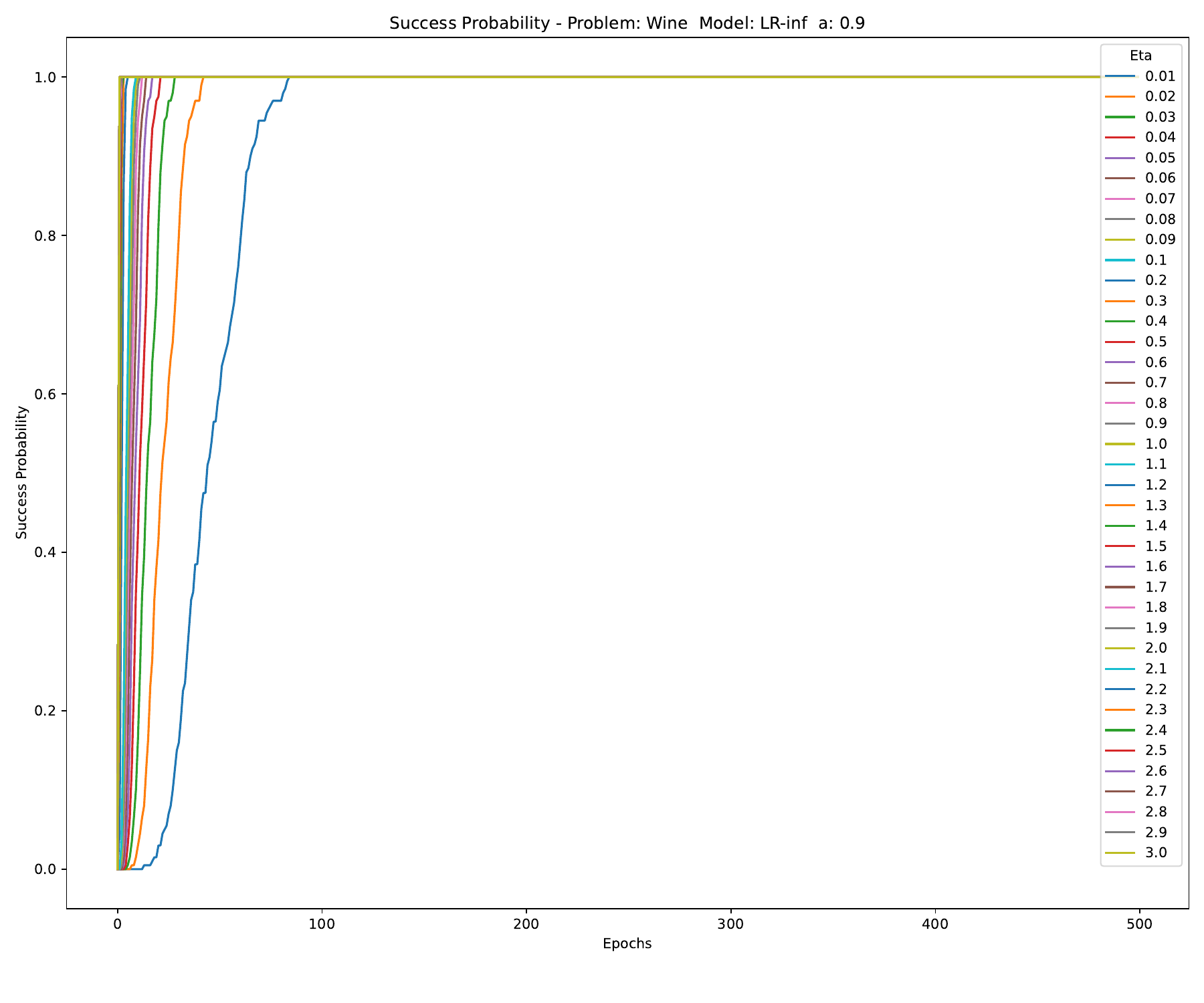} &
    \includegraphics[clip,trim=0 0 0 0,width=.52\linewidth]{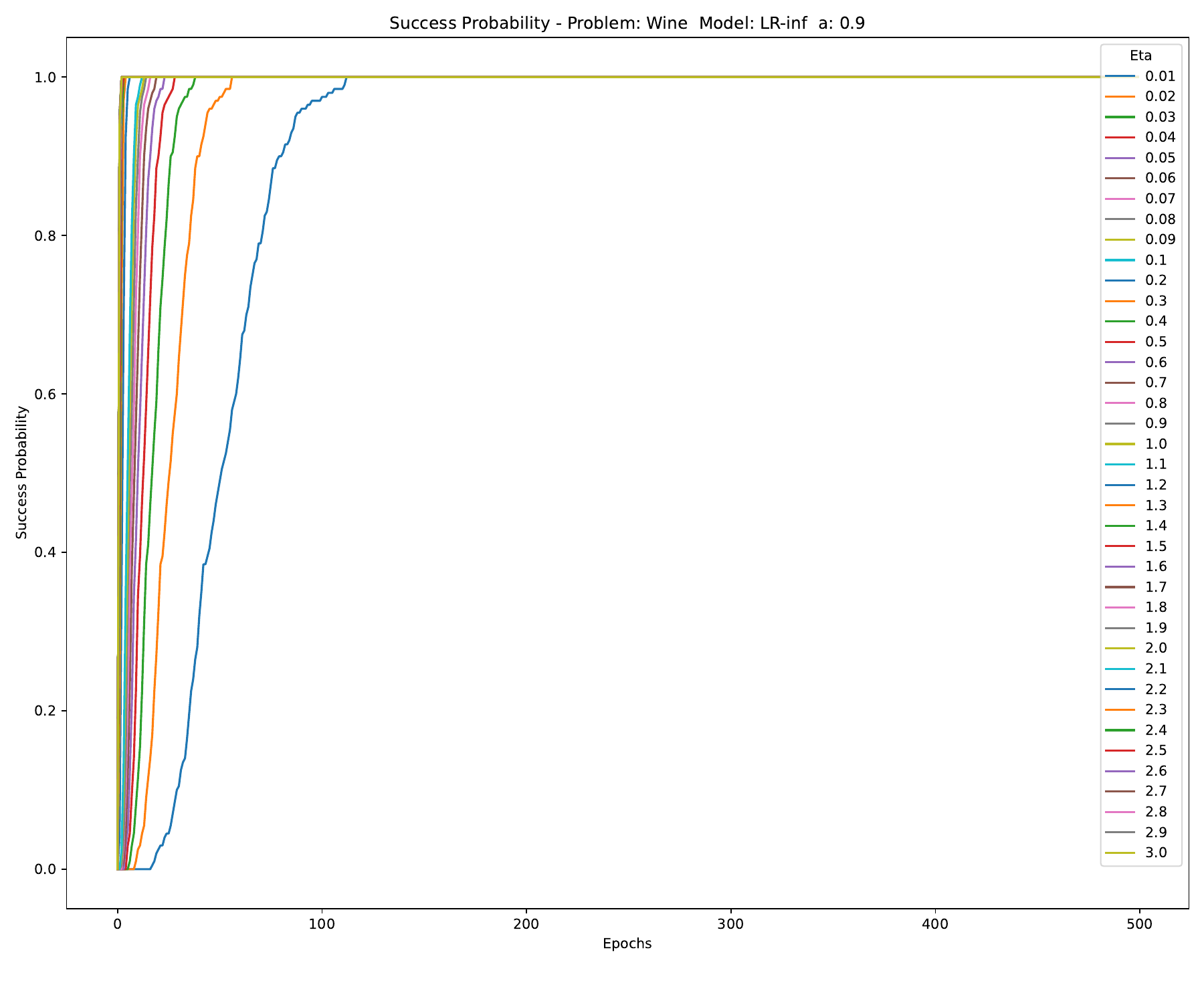} \\[-3mm]
  \end{tabular}
\caption{Success probability ($a=0.9$) for LR-Inf model (continued).}
\end{figure*}

\begin{figure*}[p]
  \centering
  \begin{tabular}{c@{}c}
    Training Success Probability & Validation Success Probability \\
    \includegraphics[clip,trim=0 0 0 0,width=.52\linewidth]{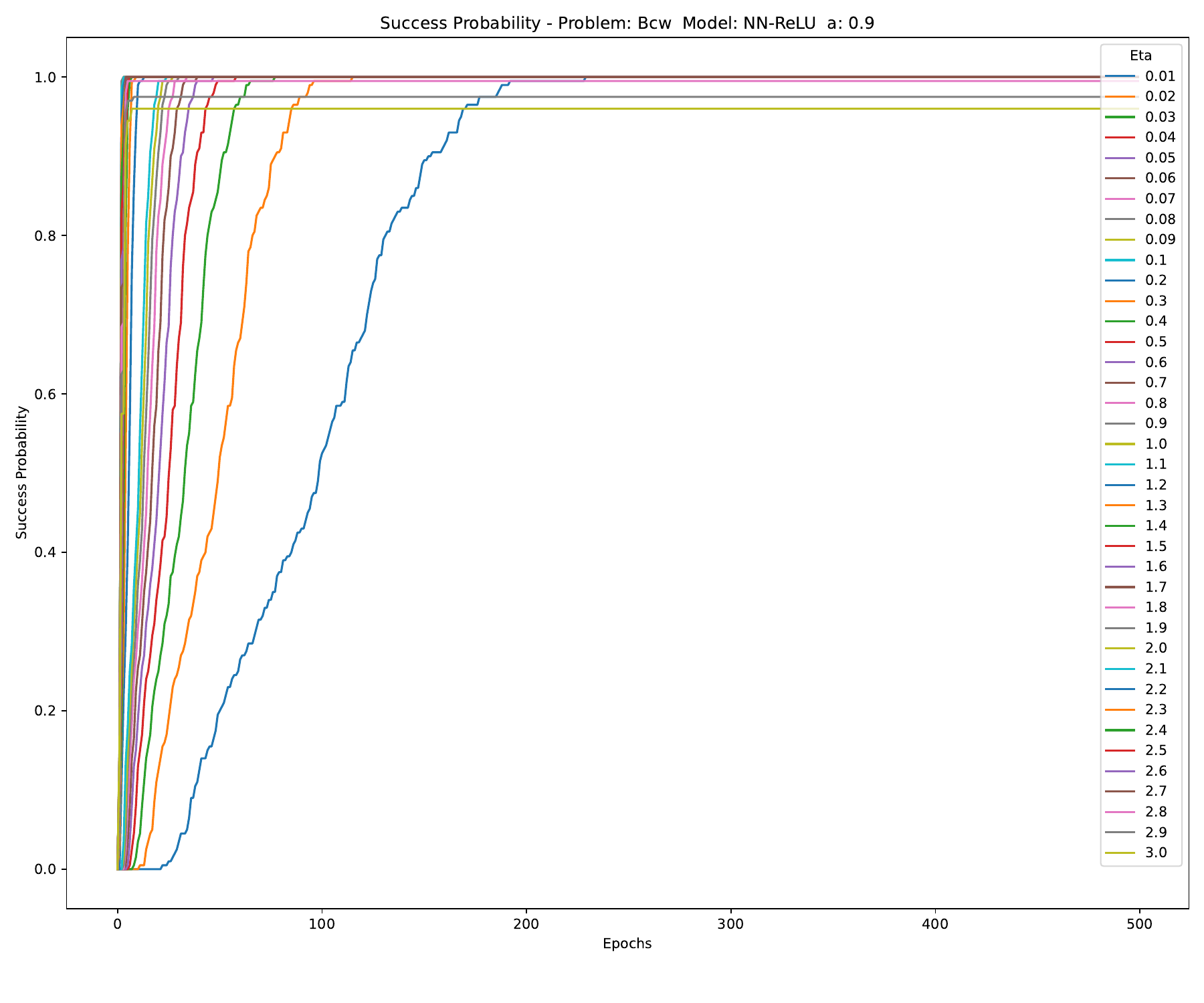} &
    \includegraphics[clip,trim=0 0 0 0,width=.52\linewidth]{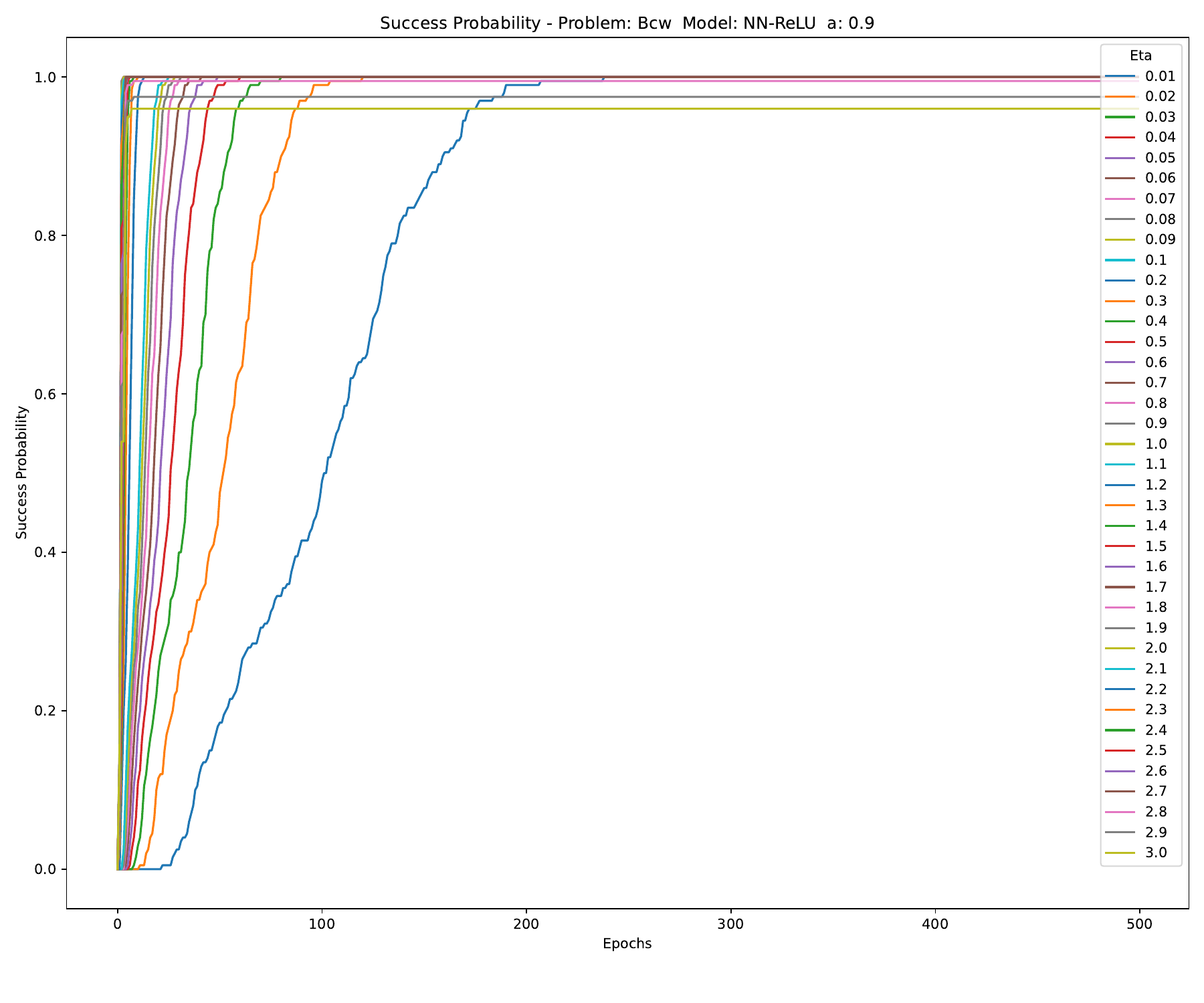} \\[-3mm]
    \includegraphics[clip,trim=0 0 0 0,width=.52\linewidth]{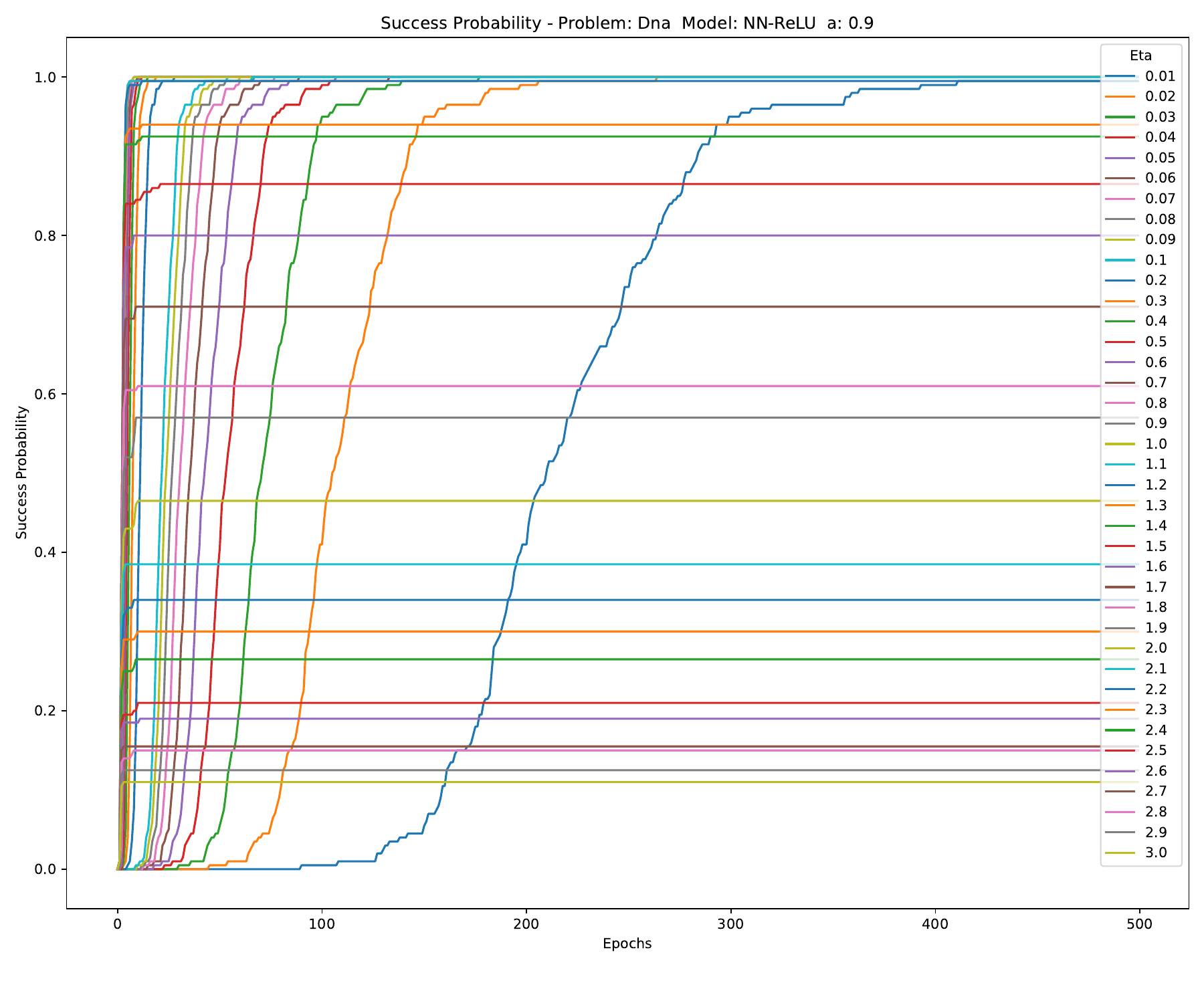} &
    \includegraphics[clip,trim=0 0 0 0,width=.52\linewidth]{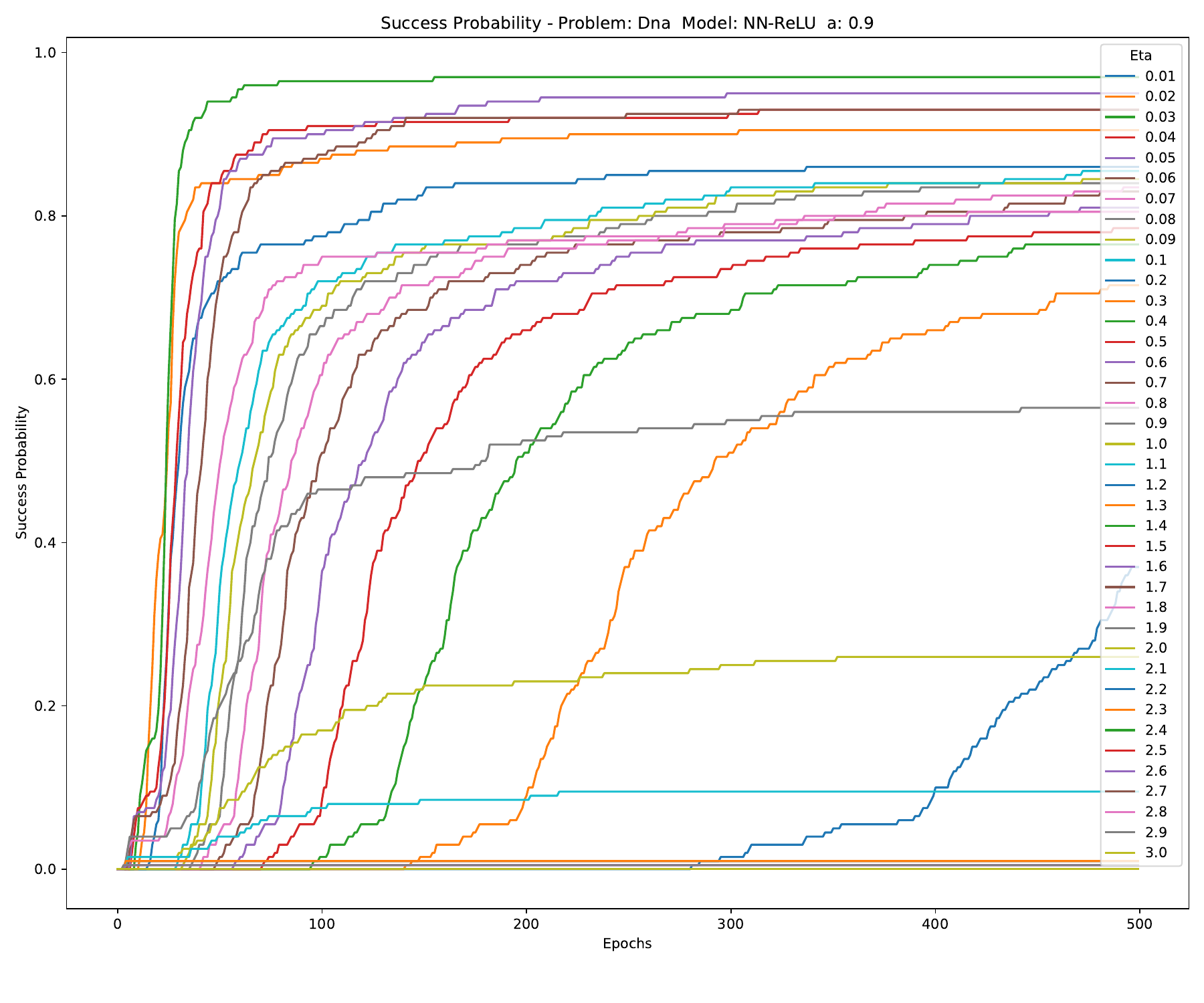} \\[-3mm]
    \includegraphics[clip,trim=0 0 0 0,width=.52\linewidth]{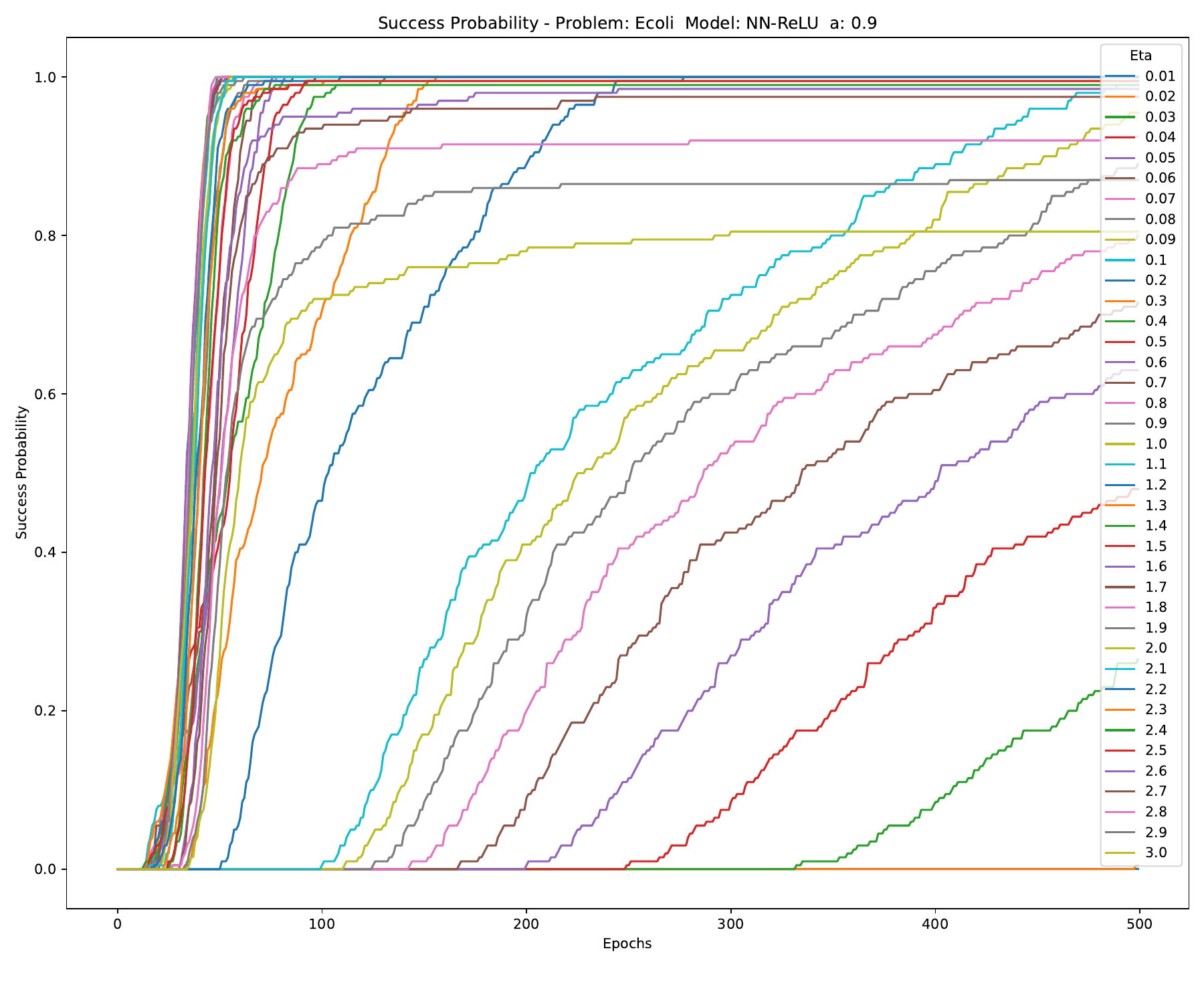} &
    \includegraphics[clip,trim=0 0 0 0,width=.52\linewidth]{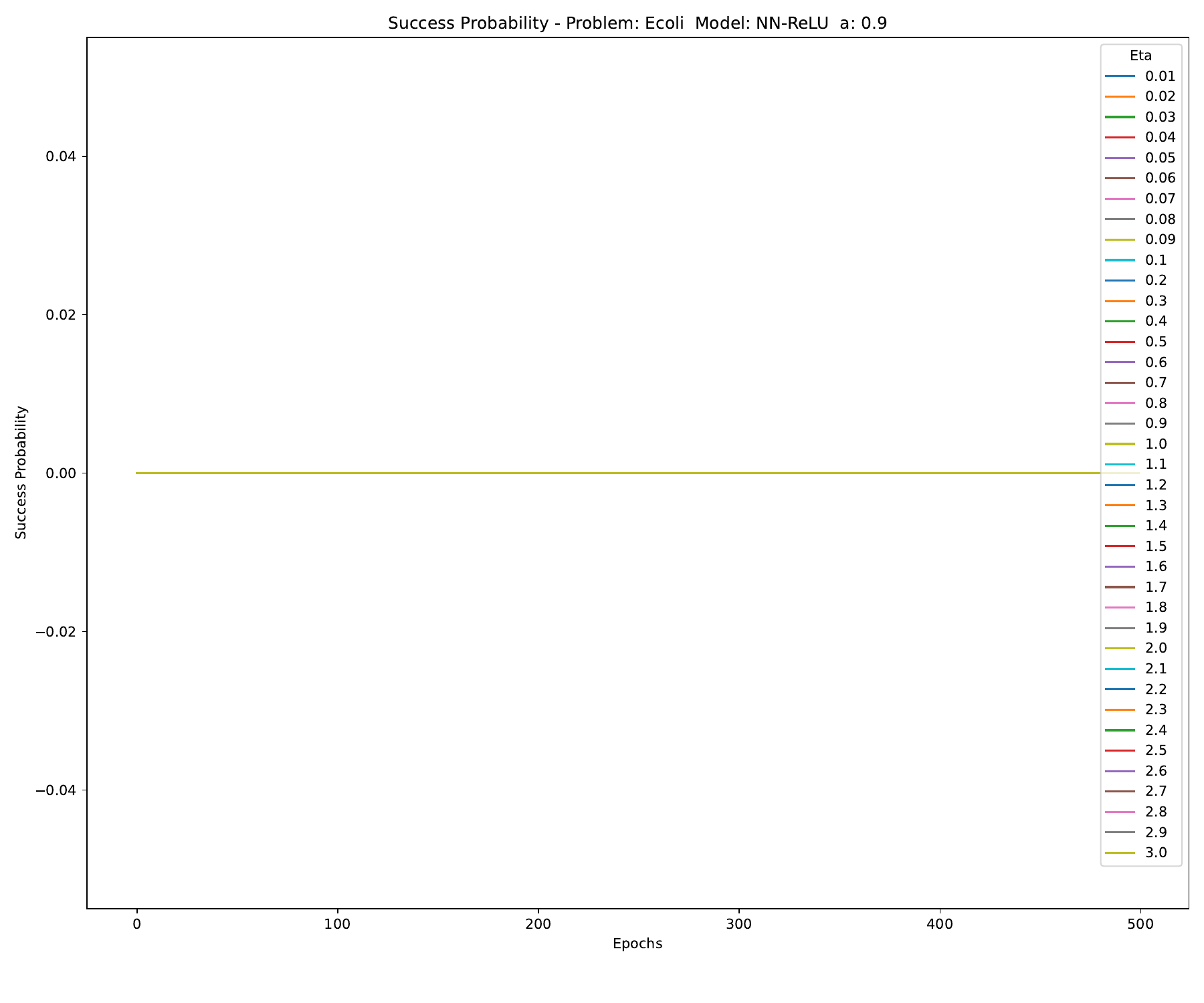} \\[-3mm]
  \end{tabular}
\caption{Success probability ($a=0.9$) for NN-ReLU model.}
  \label{fig:success_probability_NN_ReLU_model_0.9}
\end{figure*}

\begin{figure*}[t]
  \centering
  \ContinuedFloat
  \begin{tabular}{c@{}c}
    Training Success Probability & Validation Success Probability \\
    \includegraphics[clip,trim=0 0 0 0,width=.52\linewidth]{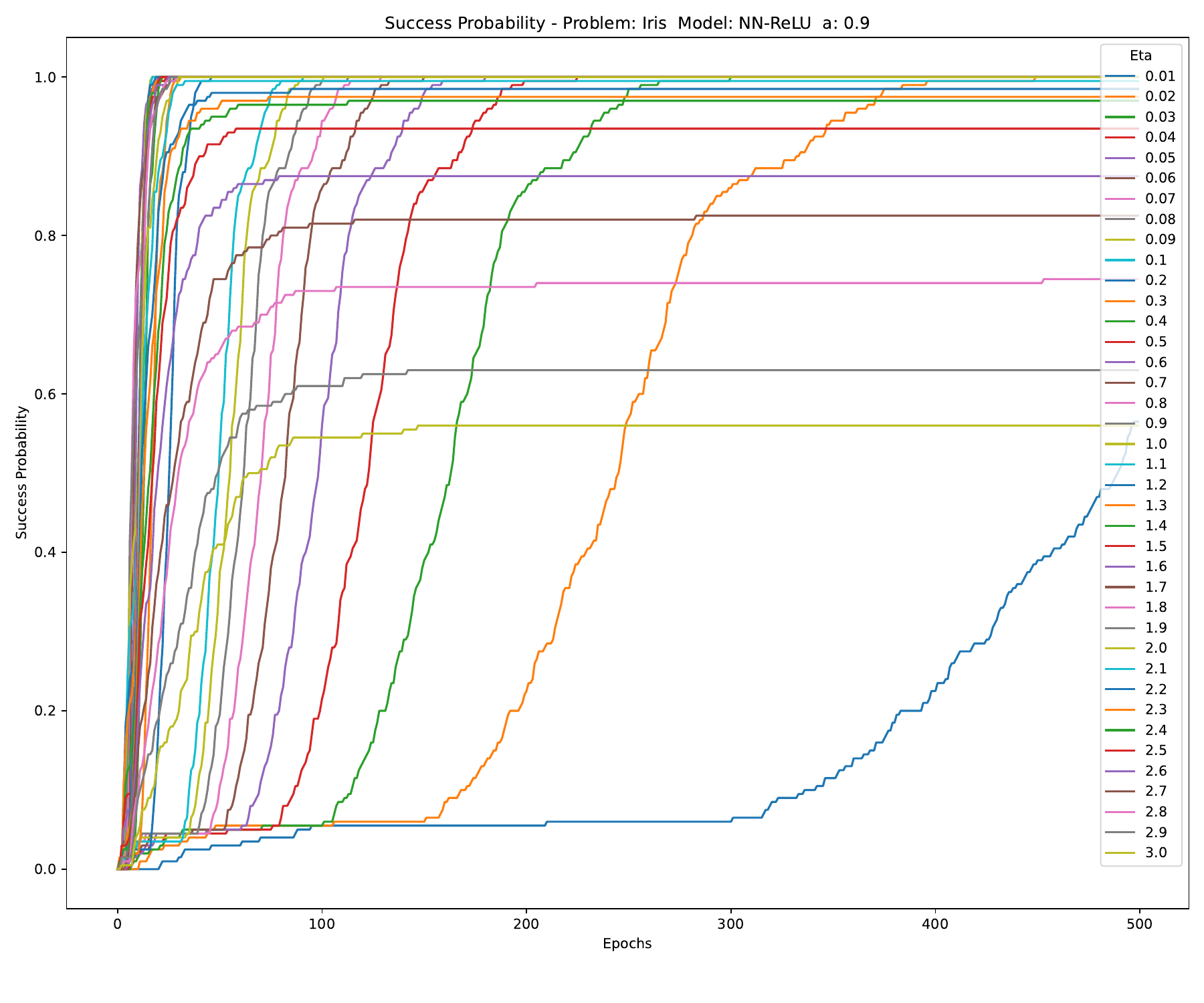} &
    \includegraphics[clip,trim=0 0 0 0,width=.52\linewidth]{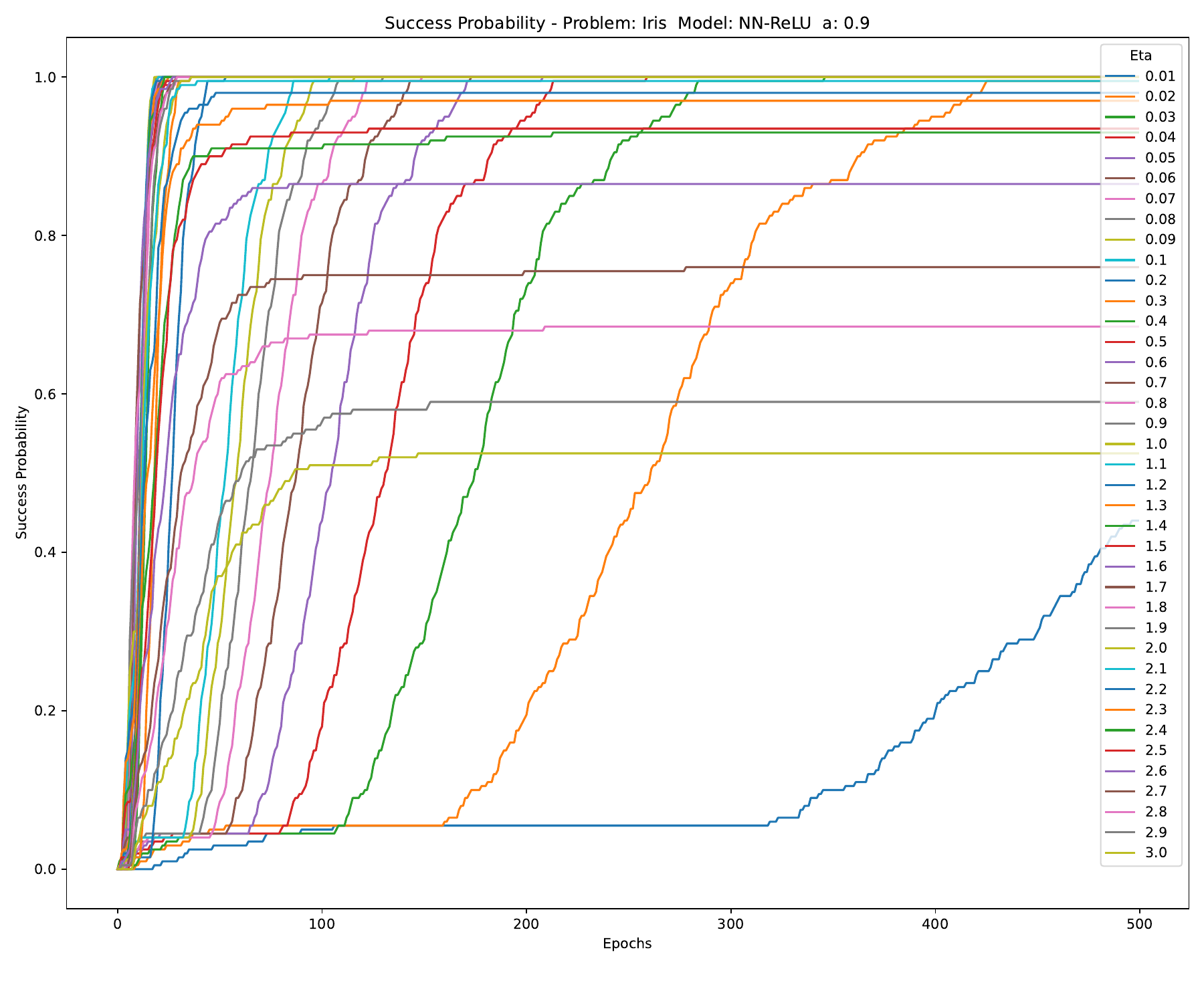} \\[-3mm]
    \includegraphics[clip,trim=0 0 0 0,width=.52\linewidth]{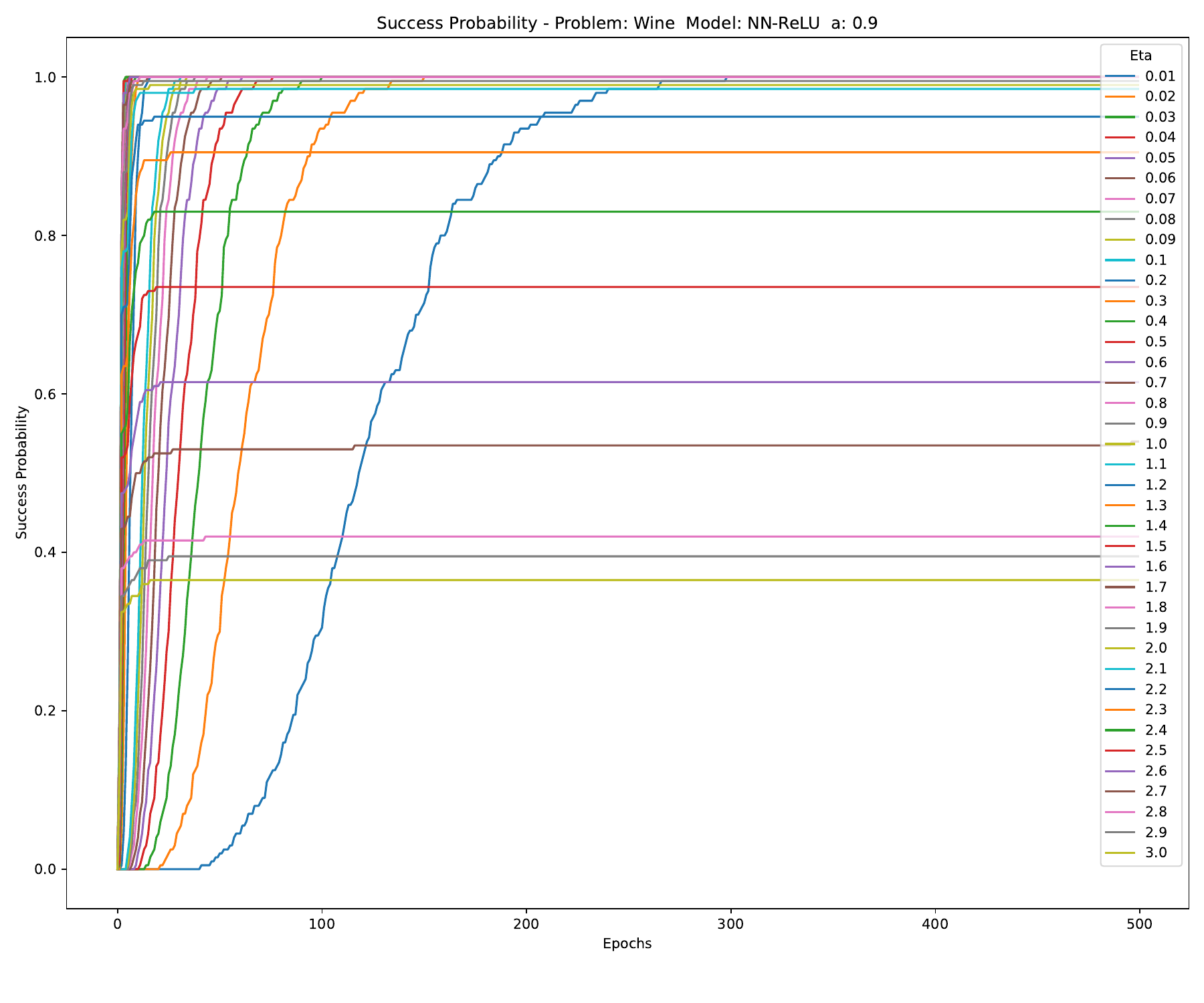} &
    \includegraphics[clip,trim=0 0 0 0,width=.52\linewidth]{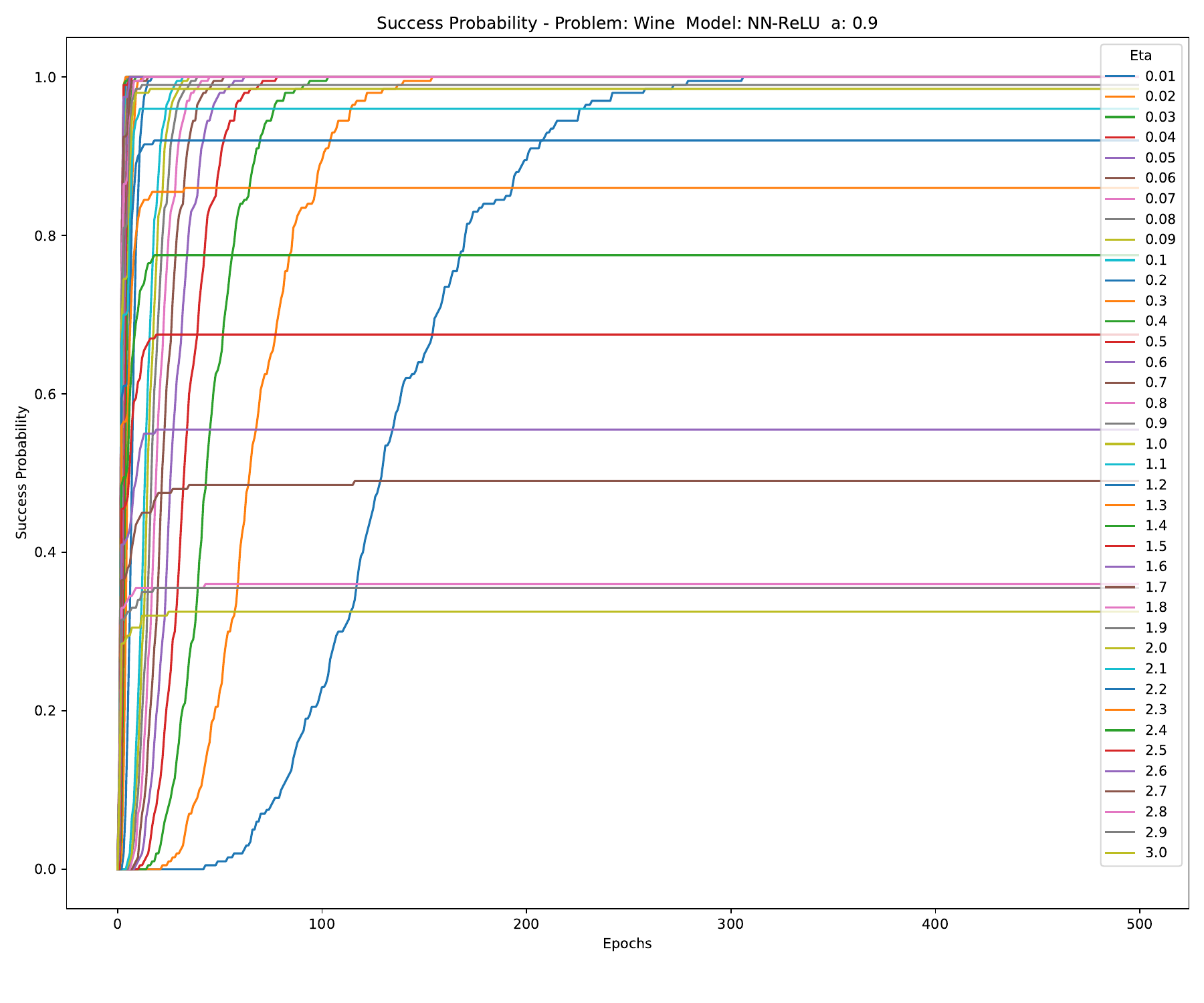} \\[-3mm]
  \end{tabular}
\caption{Success probability ($a=0.9$) for NN-ReLU model (continued).}
\end{figure*}

\medskip

Turning our attention to the \emph{validation success probabilities},
we see that the overfitting identified when looking at validation accuracies
(Section~\ref{sec:mean_train_valid}) manifests itself also at the
level of validation success probabilities for every fixed value of
$a$. So, normally,
for a given learning rate, \emph{validation success probabilities are lower
than for training} (e.g., case of LR without regularisation and DNA
problem in 
Figure~\ref{fig:success_probability_LR_Inf_model_0.9}).
However, when
overfitting is limited (e.g., in BCW, see top row in
Figure~\ref{fig:accuracies_NN_ReLU_model}), training and validation
test probabilities can be surprisingly similar (see top row in
Figure~\ref{fig:success_probability_NN_ReLU_model_0.9}).

\subsection{Computational Effort}
\label{sec:train-test-comp}

Based on the observations of the previous sections, one has to expect
significant variability in the computational effort associated with
different problems,  models, learning rates, number of
iterations and acceptance thresholds $a$.

Figure~\ref{fig:comp_effort_LR_Inf_model_0.9}
and~\ref{fig:comp_effort_NN_ReLU_model_0.9} show how the (training and
validation) computational efforts for $a=0.9$ vary as a function of
epochs numbers $i$ and learning rate $\eta$ for LR-Inf and NN-ReLU.
The cyan dots in the plots represent the
\emph{minimum computational effort for each learning rate}, while the
red dot represents the overall \emph{minimum
  computational effort} $E^*$ across all $\eta$'s (see
Section~\ref{sec:hyperp-optim-ace}). Full results for $a\in\{0.85, 0.9, 0.95\}$ are available in
SM's Section~\ref{app:traintest_comp}.%
\footnote{Following Koza, in all calculations of the computational effort we
set the desired probability of success in solving the problem to the
value $z=99\%$.}

\begin{figure*}[p]
  \centering
  \begin{tabular}{c@{}c}
    Training Computational Effort & Validation Computational Effort \\
    \includegraphics[clip,trim=0 0 0 0,width=.52\linewidth]{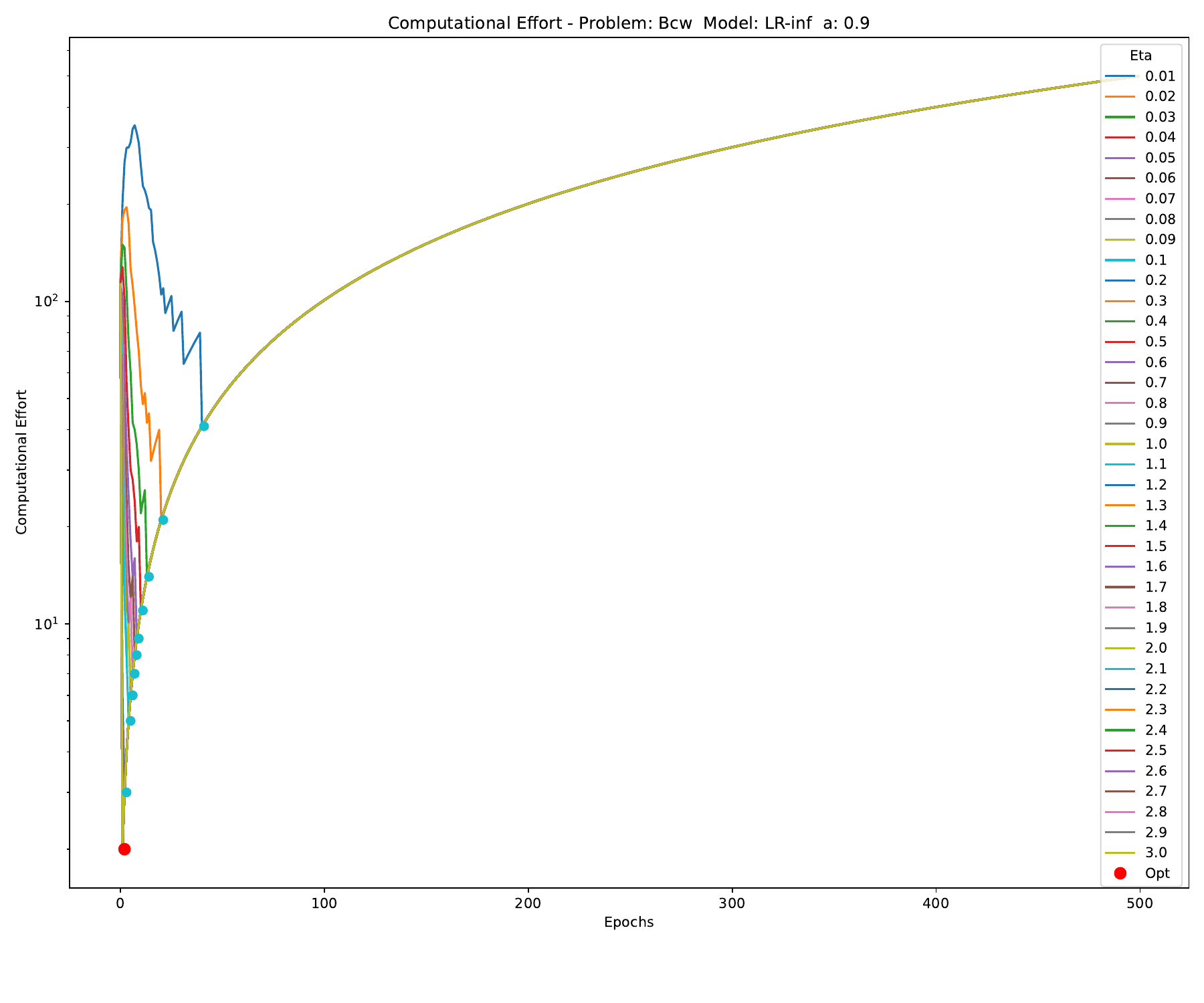} &
    \includegraphics[clip,trim=0 0 0 0,width=.52\linewidth]{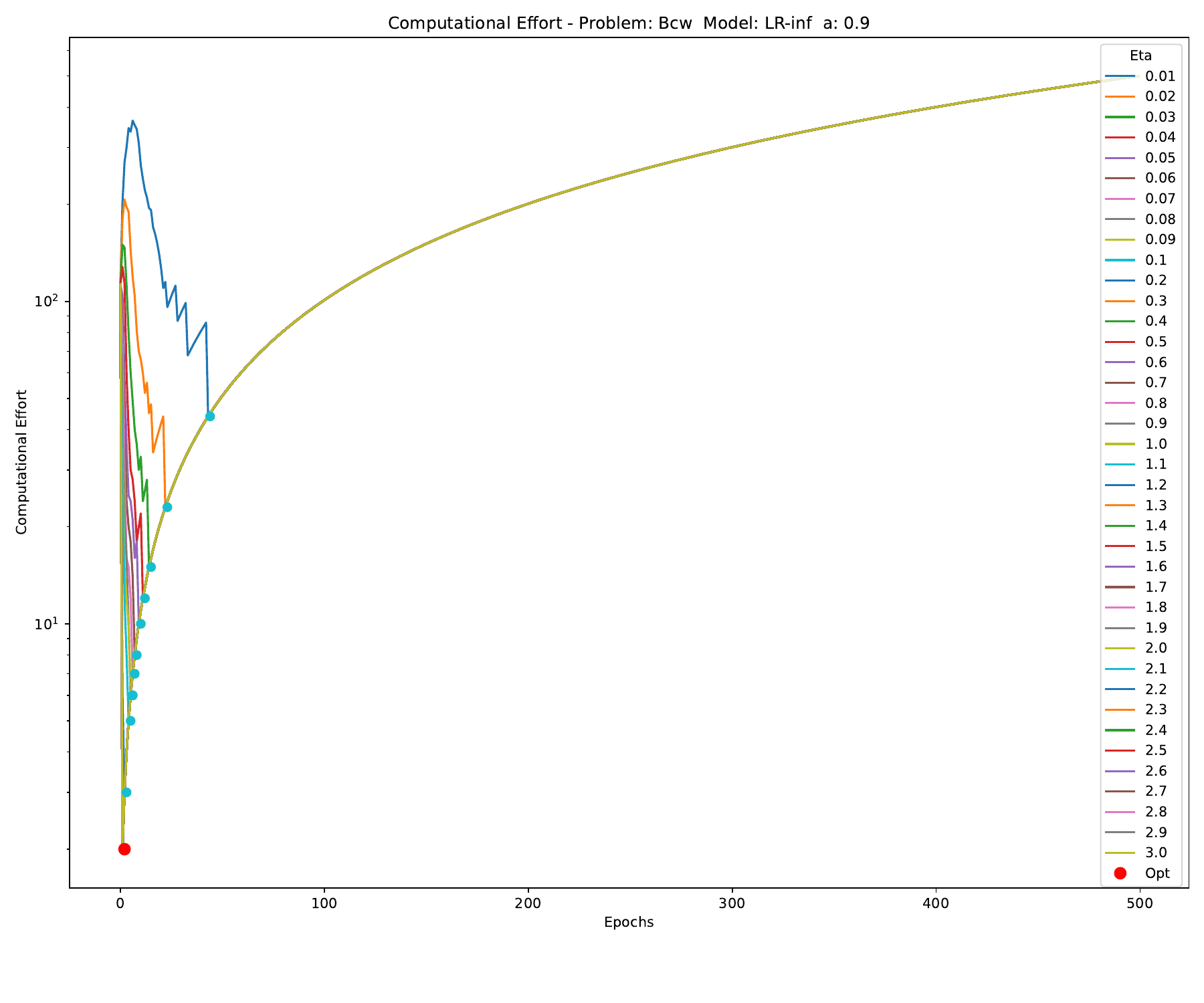} \\[-3mm]
    \includegraphics[clip,trim=0 0 0 0,width=.52\linewidth]{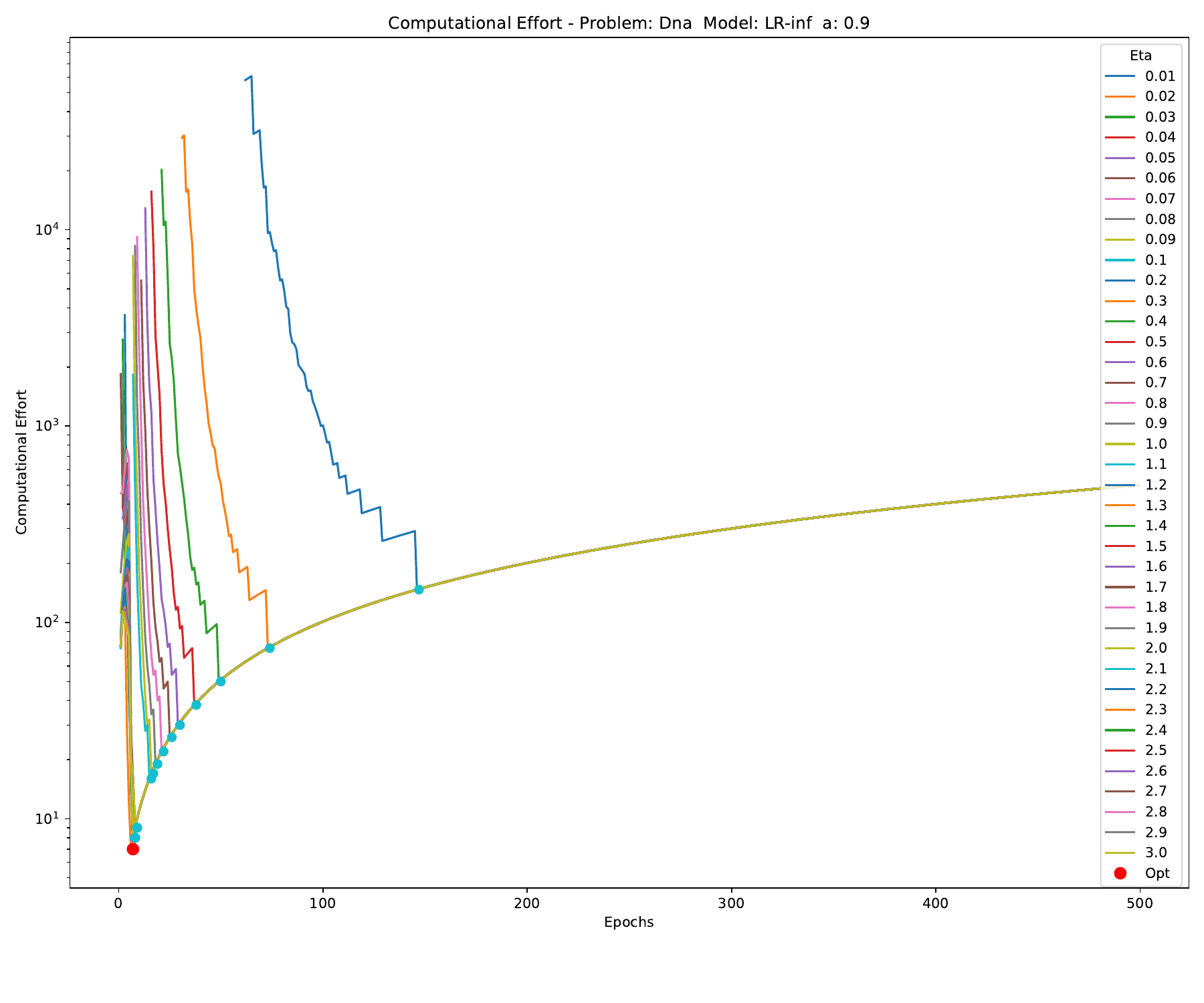} &
    \includegraphics[clip,trim=0 0 0 0,width=.52\linewidth]{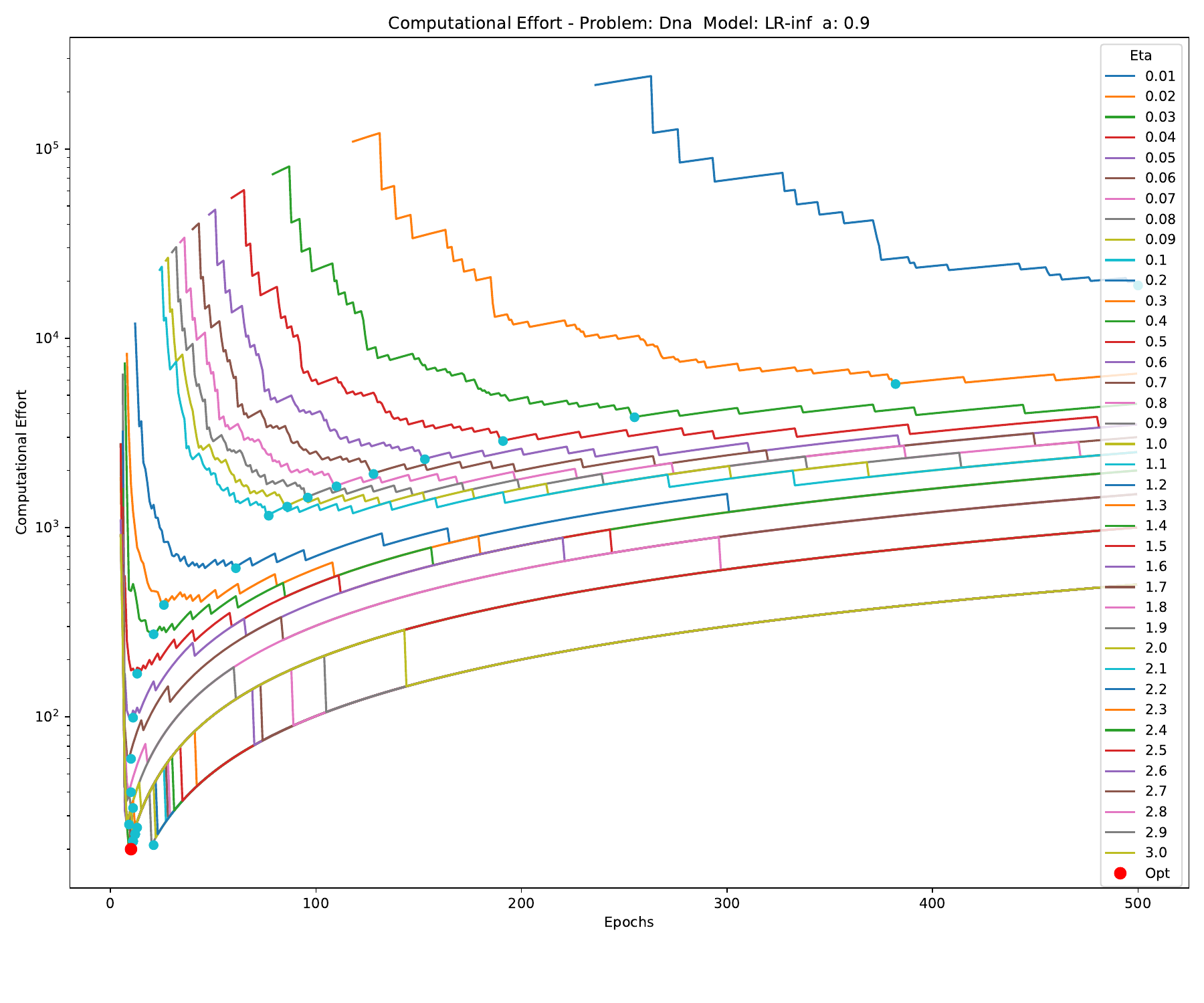} \\[-3mm]
    \includegraphics[clip,trim=0 0 0 0,width=.52\linewidth]{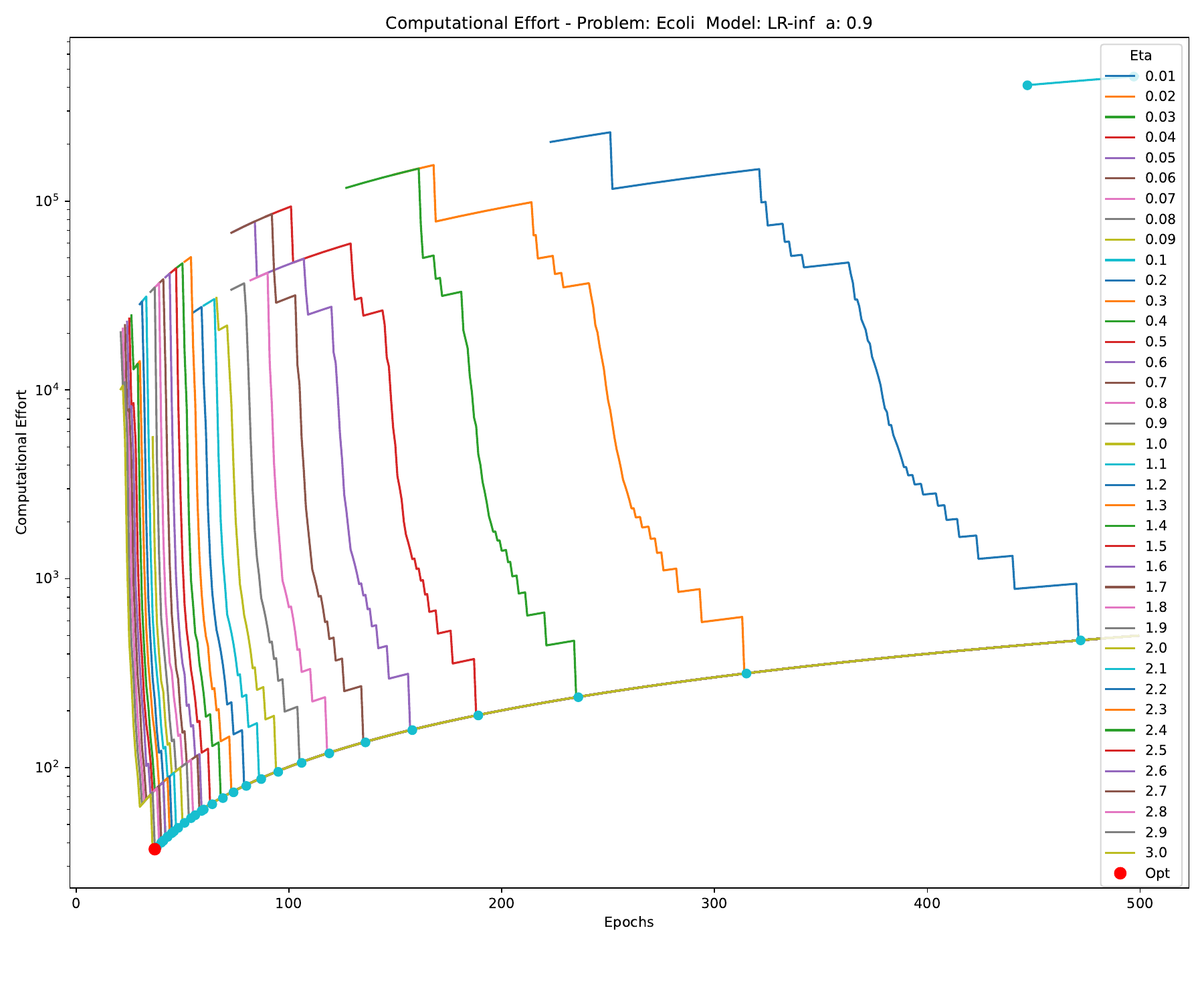} &
    \includegraphics[clip,trim=0 0 0 0,width=.52\linewidth]{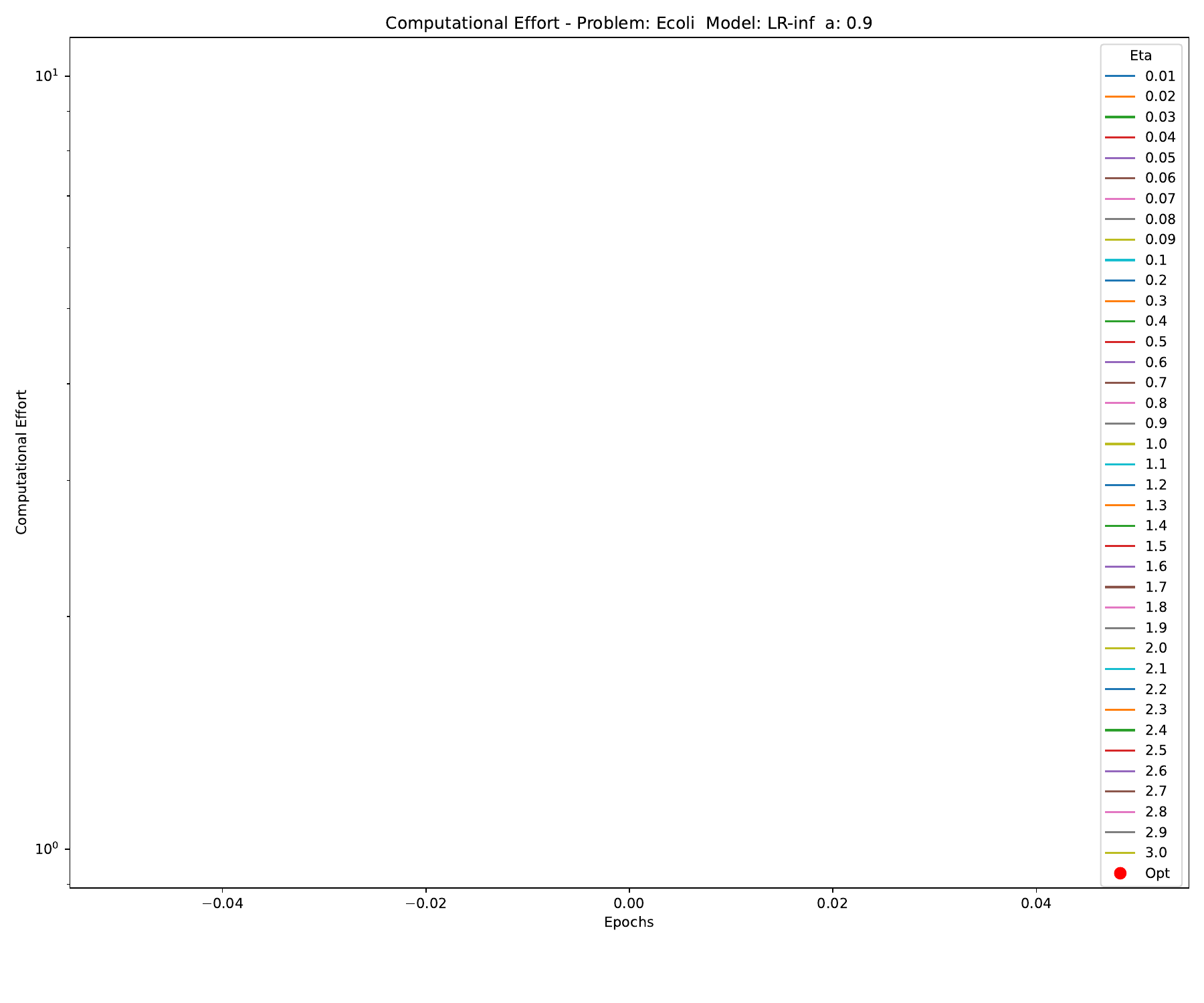} \\[-3mm]
  \end{tabular}
\caption{Computational effort ($a=0.9$) for LR-Inf model.}
  \label{fig:comp_effort_LR_Inf_model_0.9}
\end{figure*}

\begin{figure*}[t]
  \centering
  \ContinuedFloat
  \begin{tabular}{c@{}c}
    Training Computational Effort & Validation Computational Effort \\
    \includegraphics[clip,trim=0 0 0 0,width=.52\linewidth]{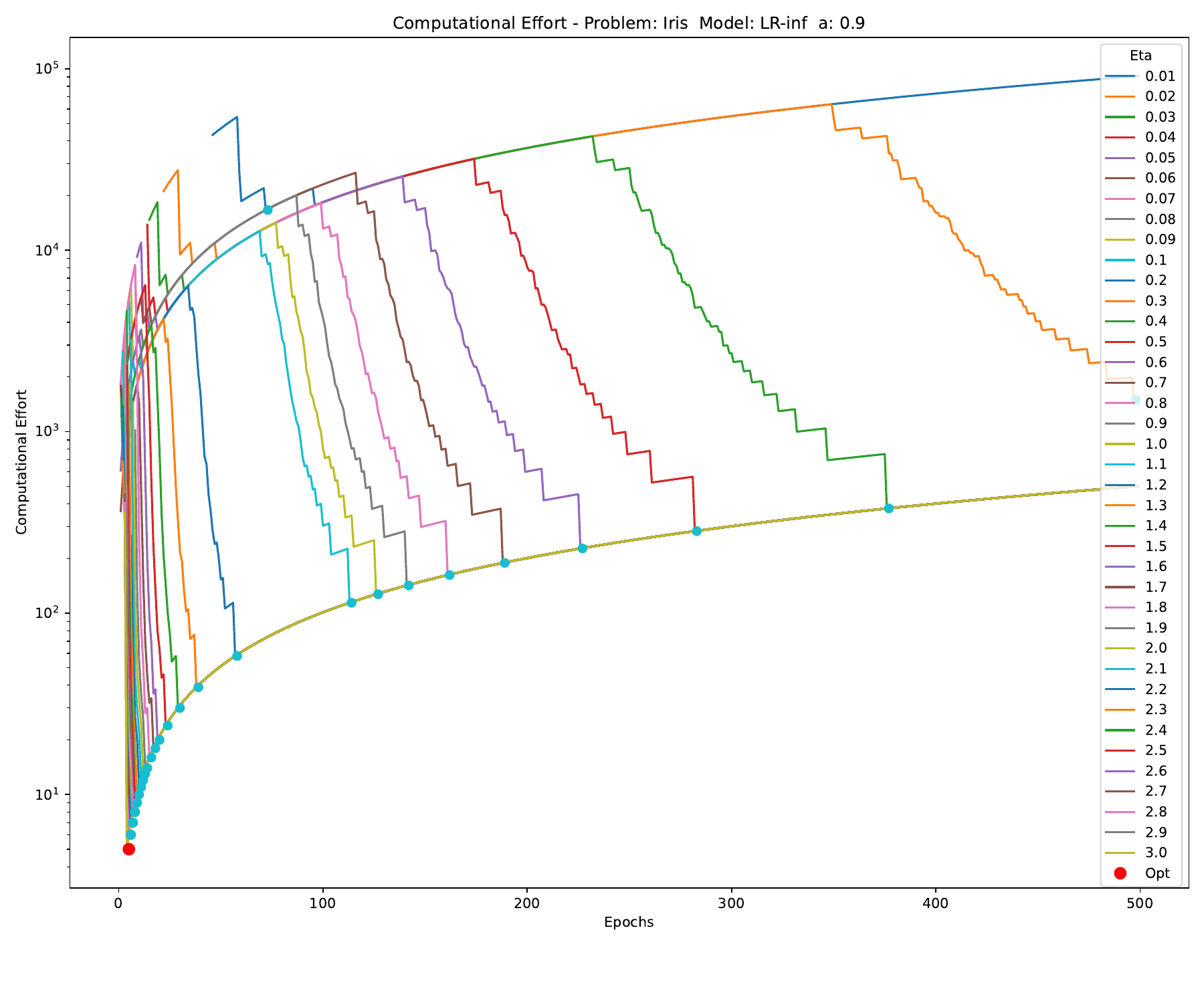} &
    \includegraphics[clip,trim=0 0 0 0,width=.52\linewidth]{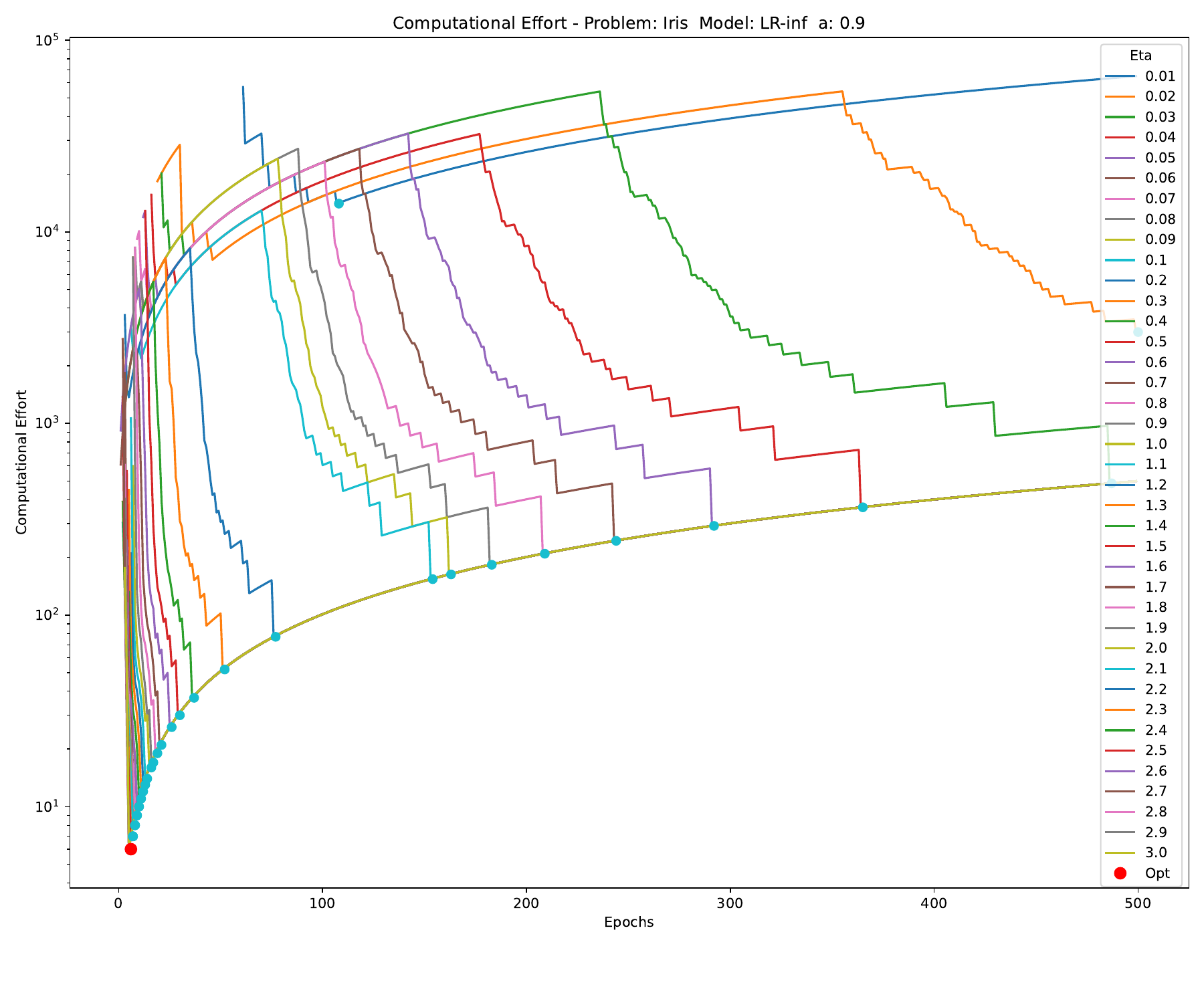} \\[-3mm]
    \includegraphics[clip,trim=0 0 0 0,width=.52\linewidth]{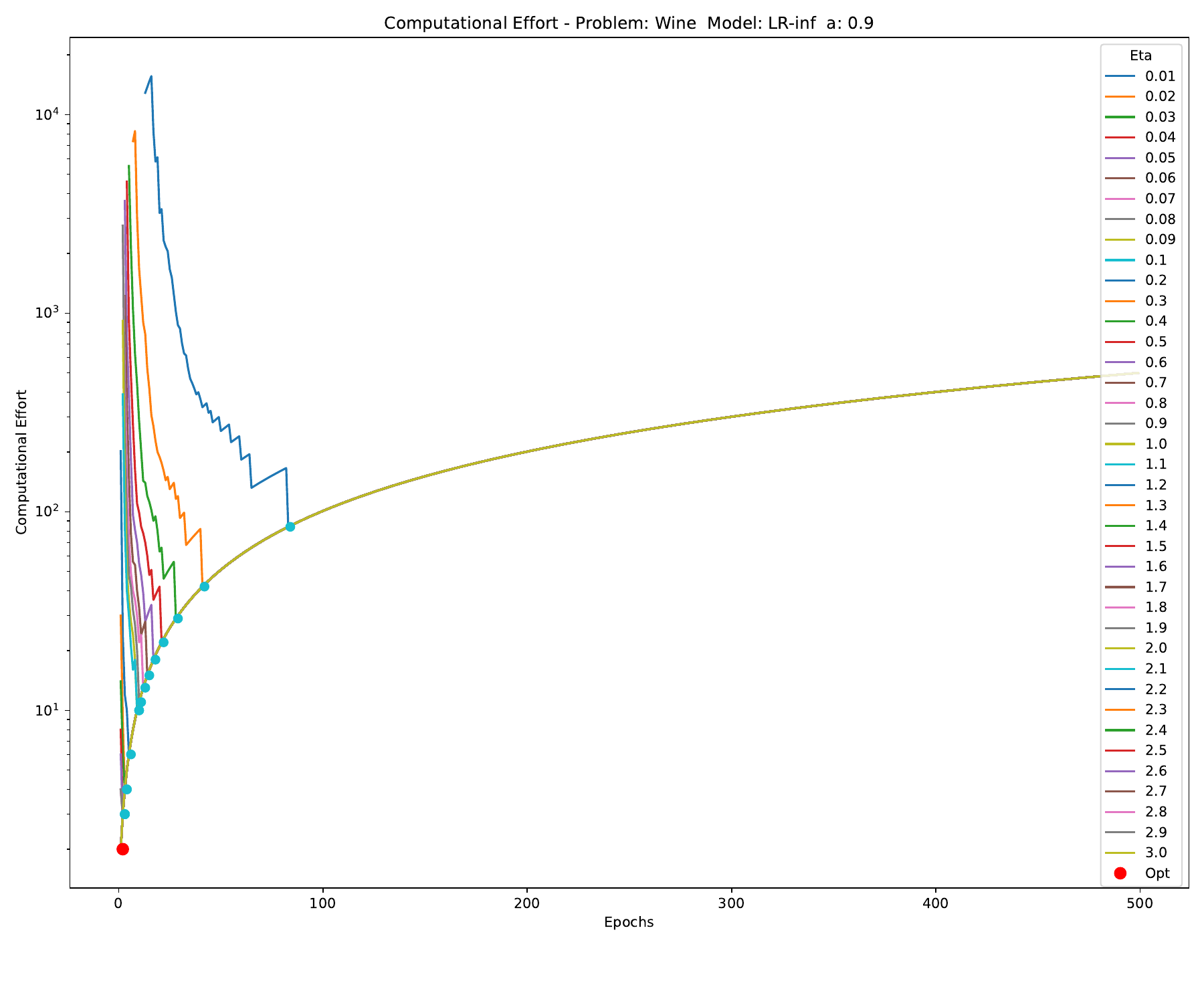} &
    \includegraphics[clip,trim=0 0 0 0,width=.52\linewidth]{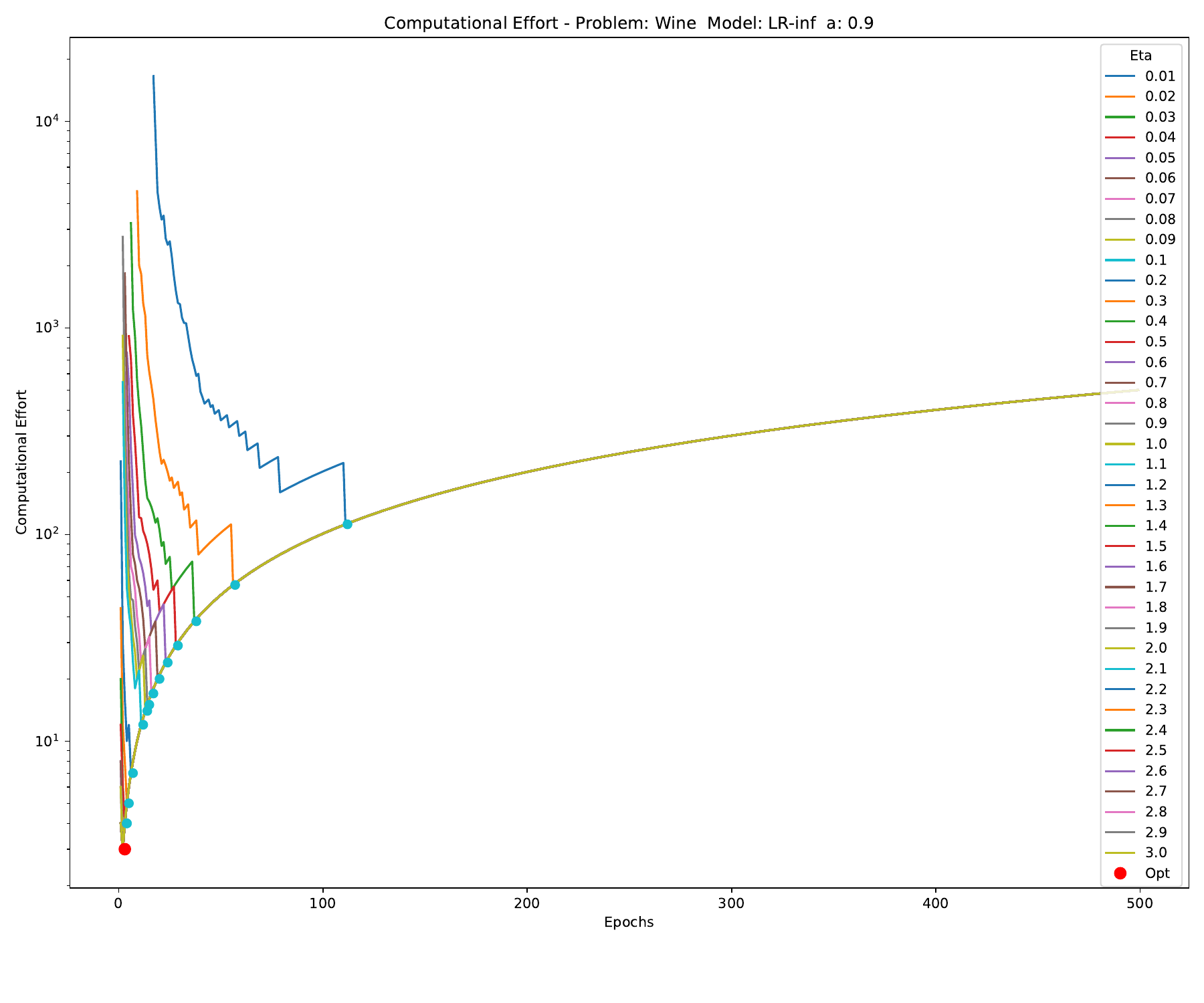} \\[-3mm]
  \end{tabular}
\caption{Computational effort ($a=0.9$) for LR-Inf model (continued).}
\end{figure*}

\begin{figure*}[p]
  \centering
  \begin{tabular}{c@{}c}
    Training Computational Effort & Validation Computational Effort \\
    \includegraphics[clip,trim=0 0 0 0,width=.52\linewidth]{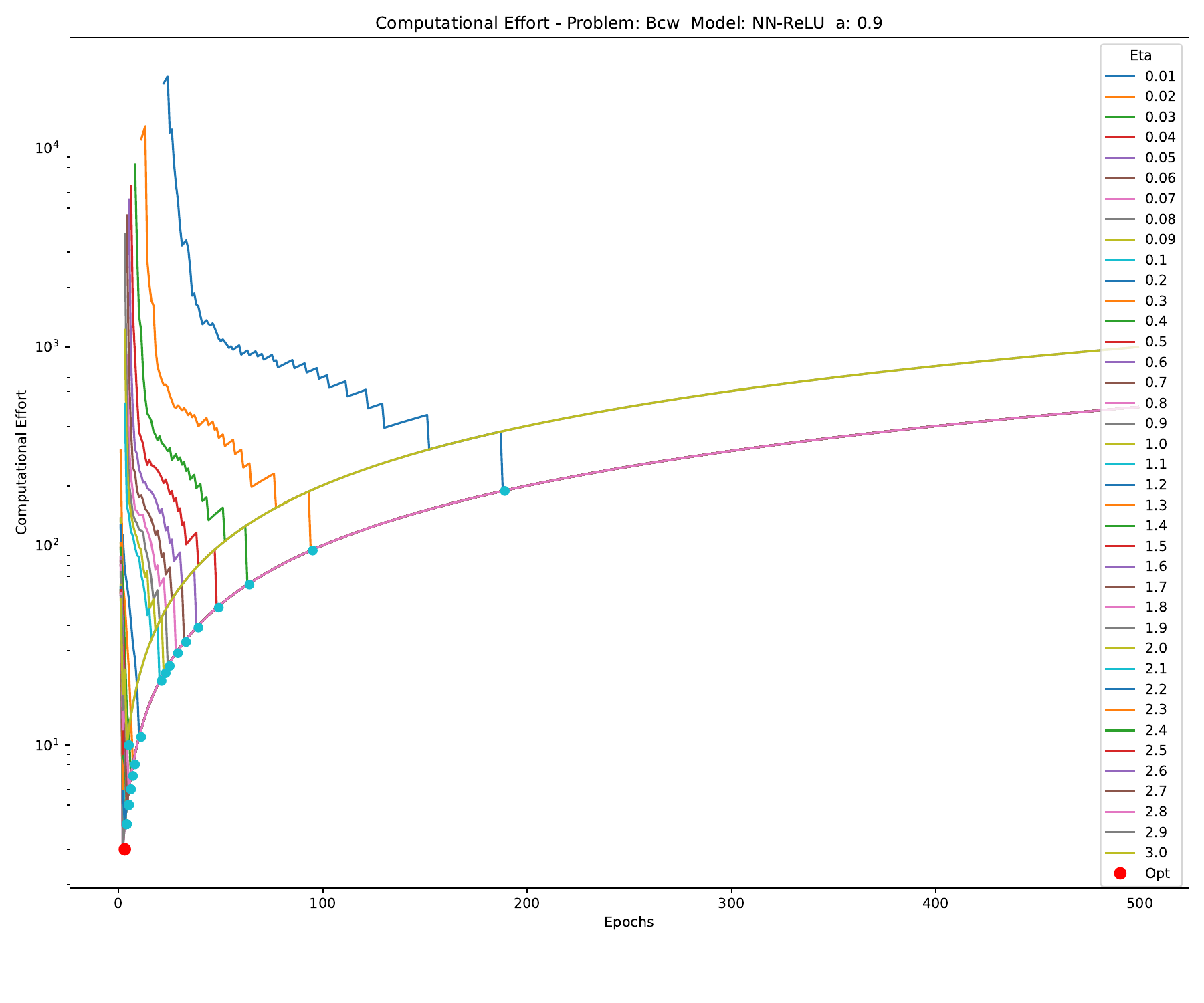} &
    \includegraphics[clip,trim=0 0 0 0,width=.52\linewidth]{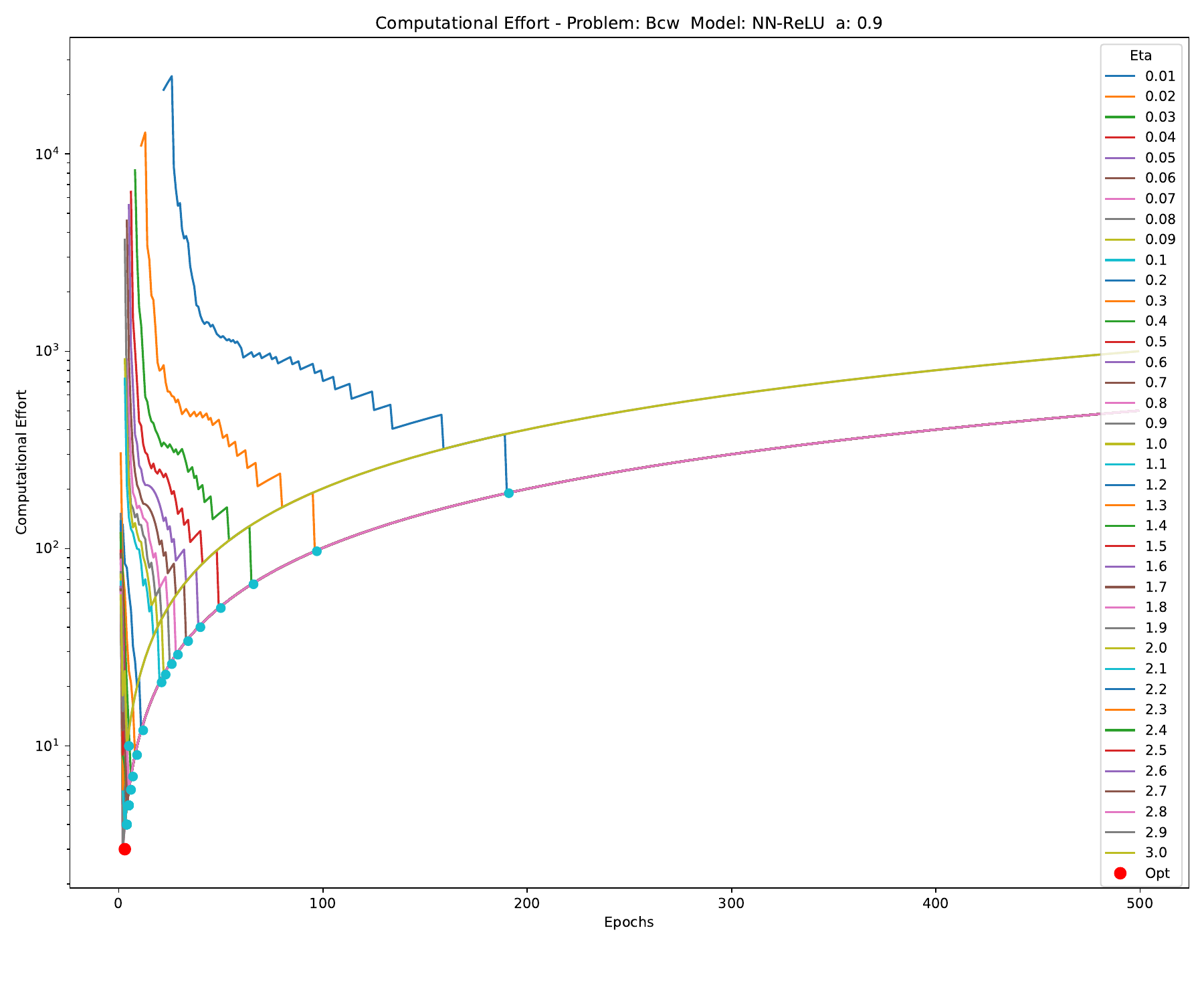} \\[-3mm]
    \includegraphics[clip,trim=0 0 0 0,width=.52\linewidth]{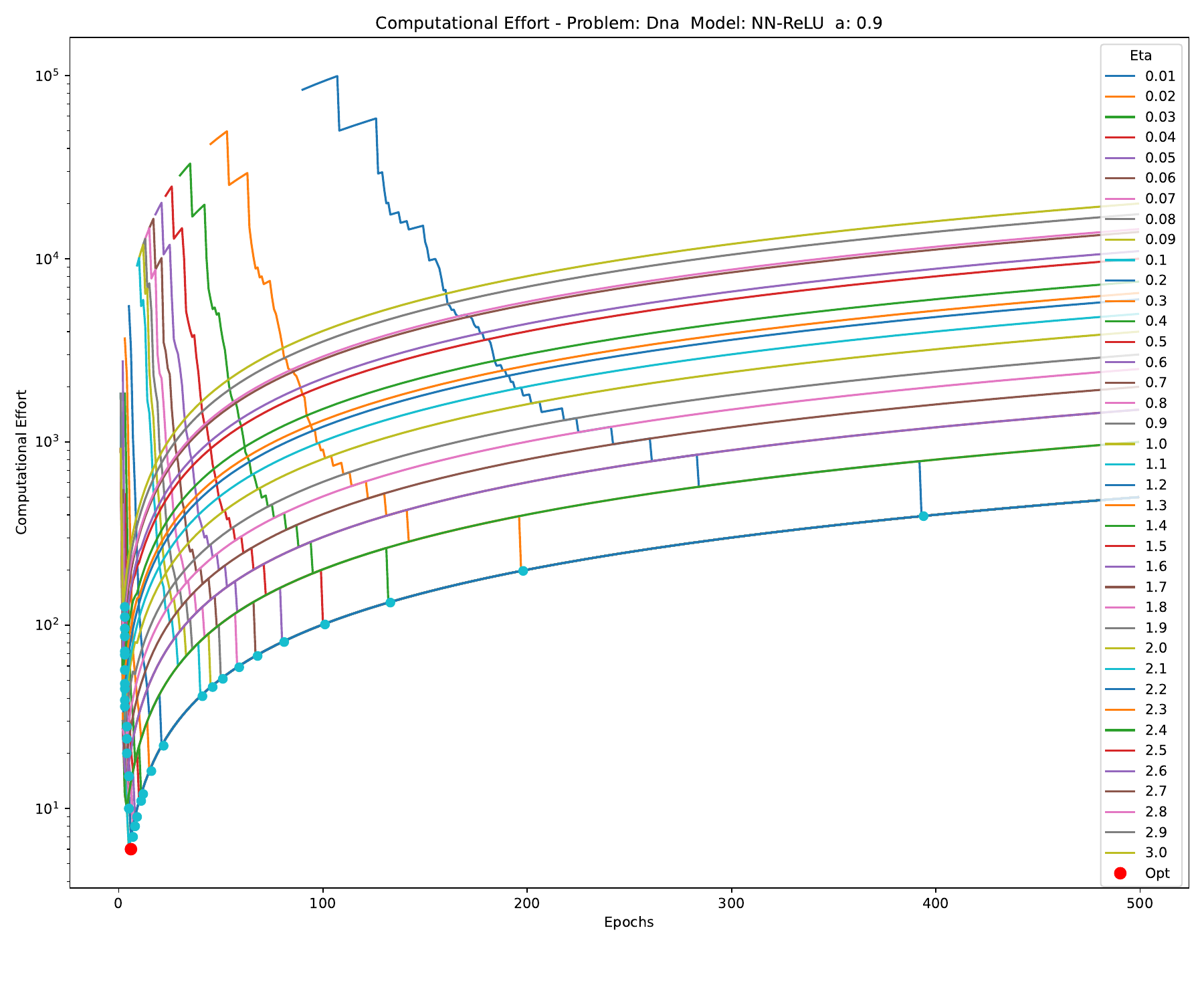} &
    \includegraphics[clip,trim=0 0 0 0,width=.52\linewidth]{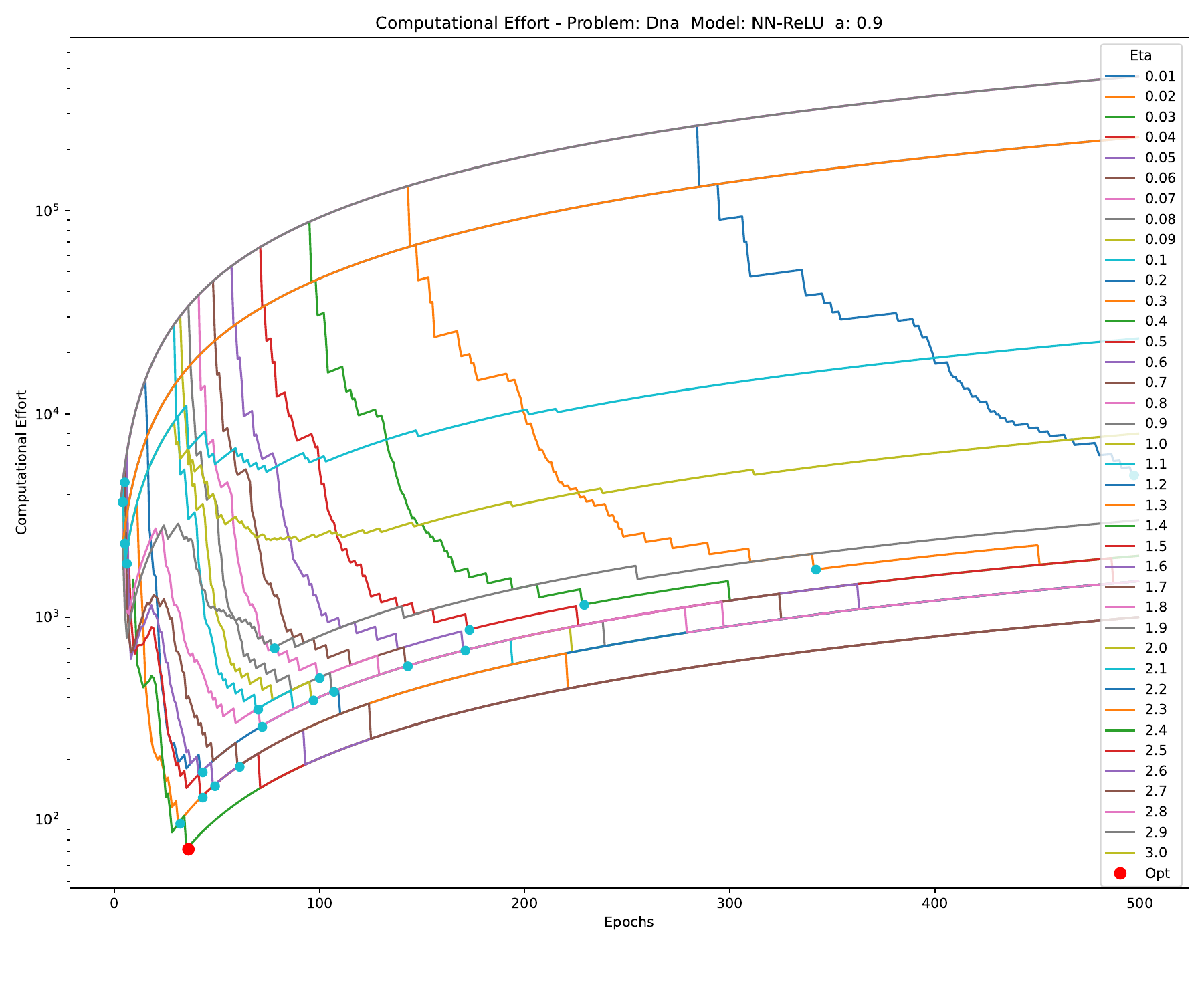} \\[-3mm]
    \includegraphics[clip,trim=0 0 0 0,width=.52\linewidth]{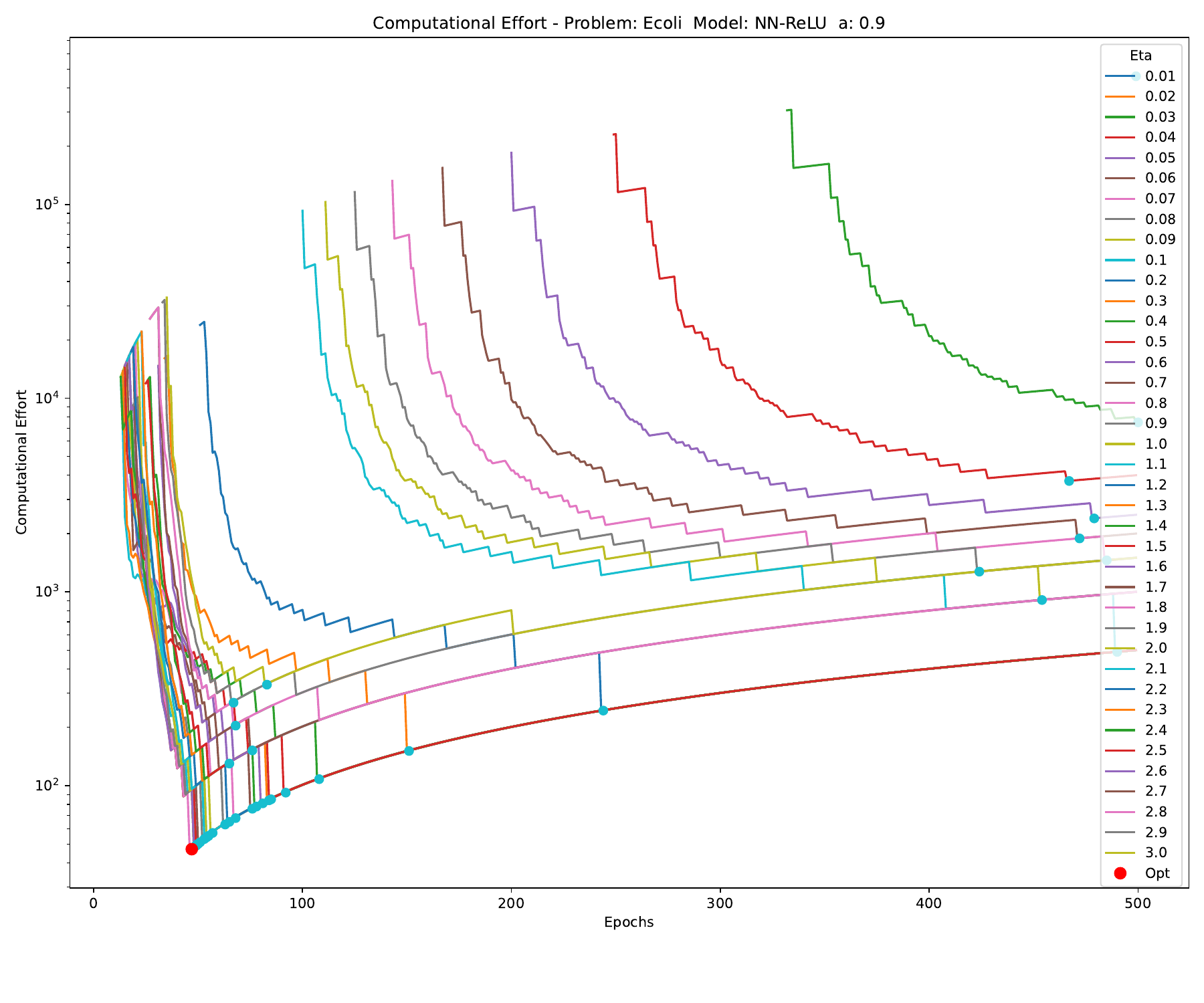} &
    \includegraphics[clip,trim=0 0 0 0,width=.52\linewidth]{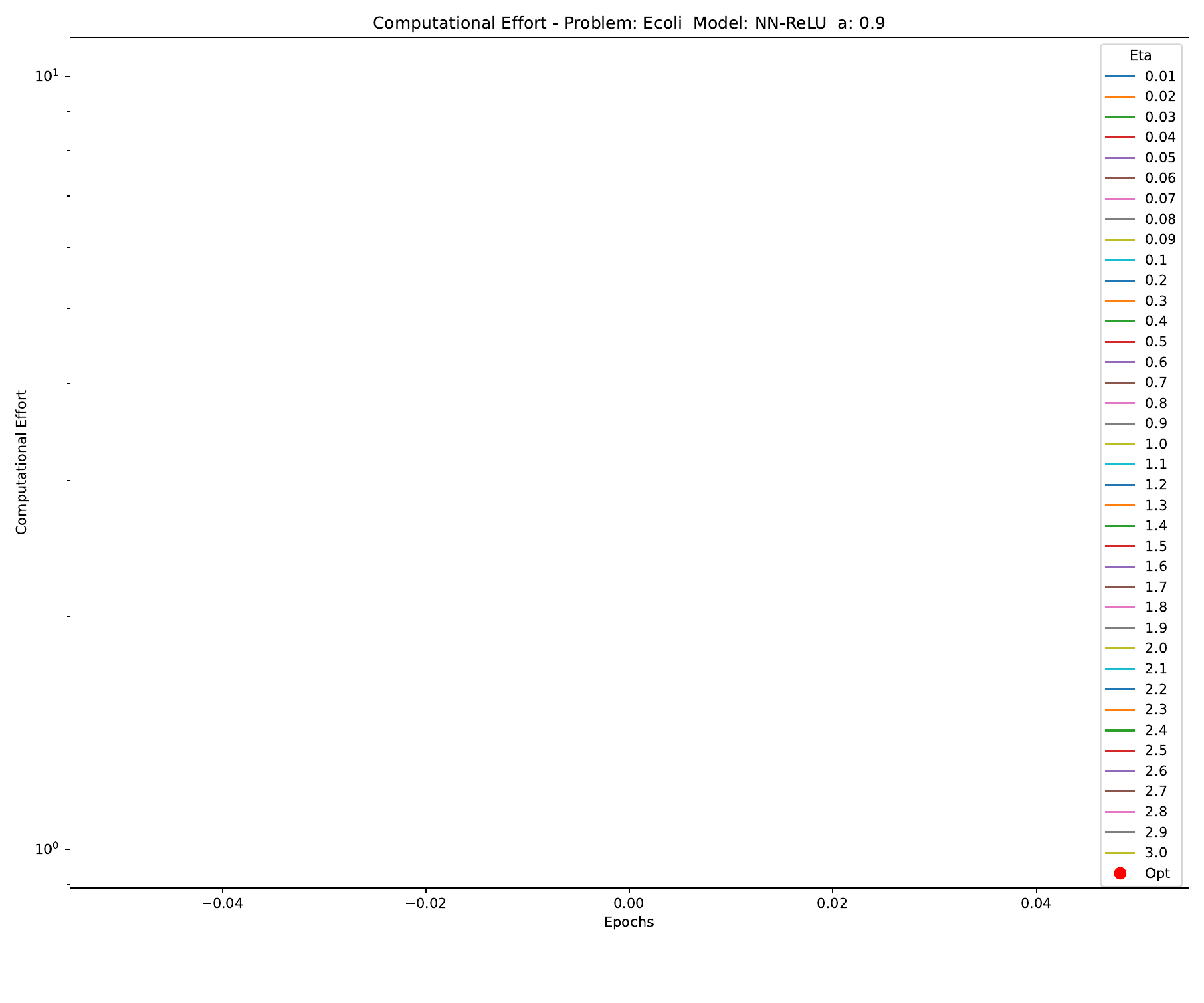} \\[-3mm]
  \end{tabular}
\caption{Computational effort ($a=0.9$) for NN-ReLU model.}
  \label{fig:comp_effort_NN_ReLU_model_0.9}
\end{figure*}

\begin{figure*}[t]
  \centering
  \ContinuedFloat
  \begin{tabular}{c@{}c}
    Training Computational Effort & Validation Computational Effort \\
    \includegraphics[clip,trim=0 0 0 0,width=.52\linewidth]{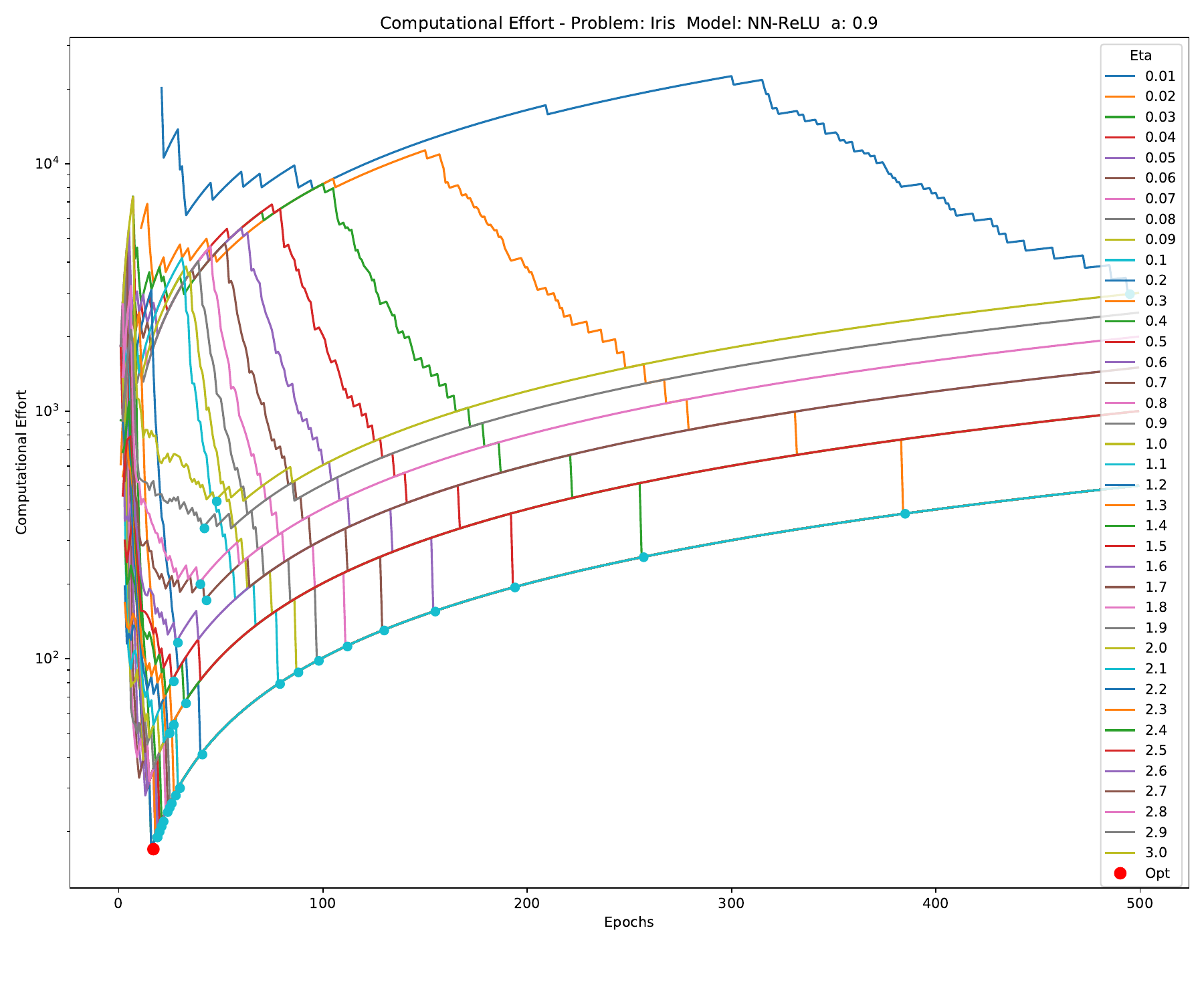} &
    \includegraphics[clip,trim=0 0 0 0,width=.52\linewidth]{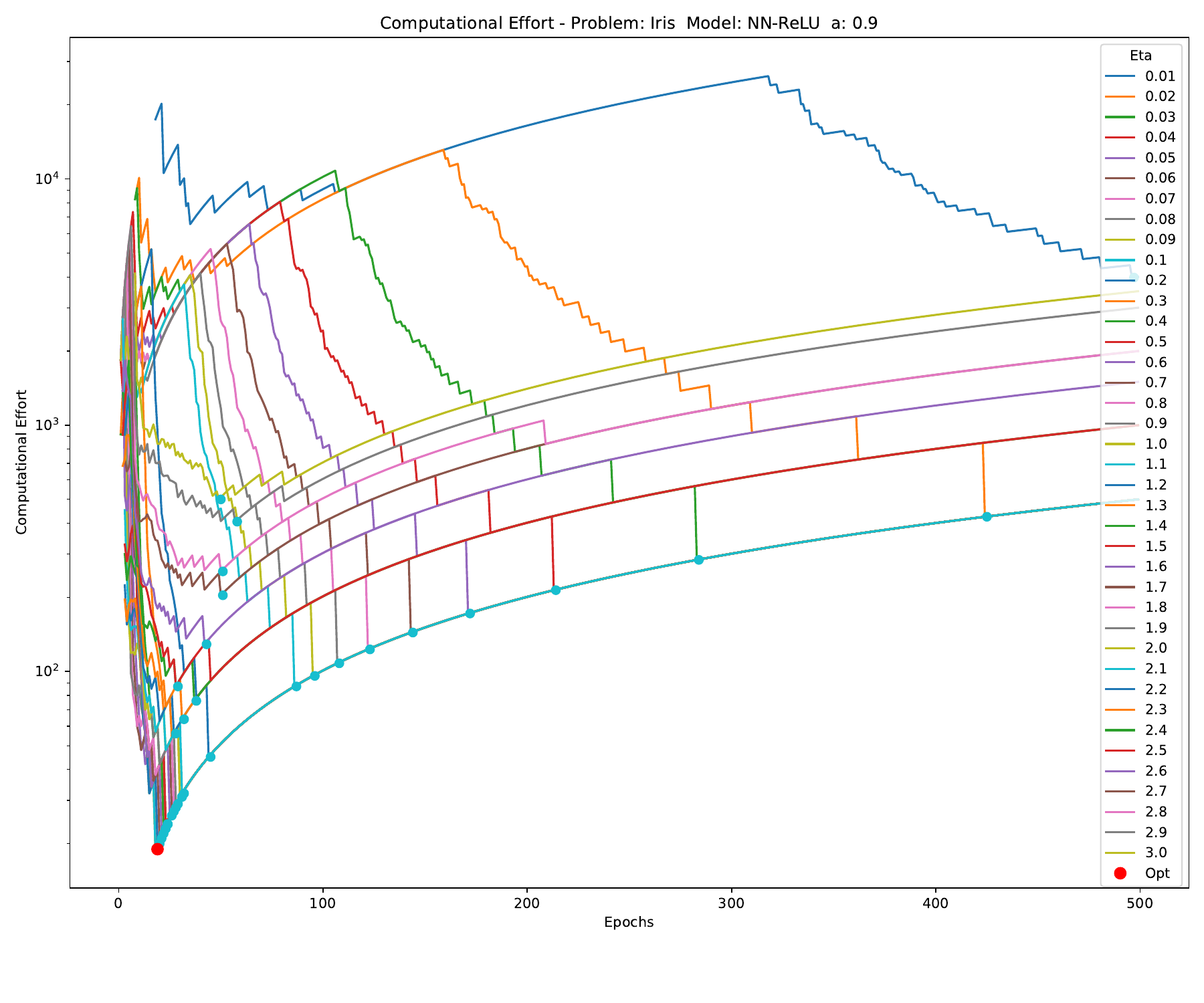} \\[-3mm]
    \includegraphics[clip,trim=0 0 0 0,width=.52\linewidth]{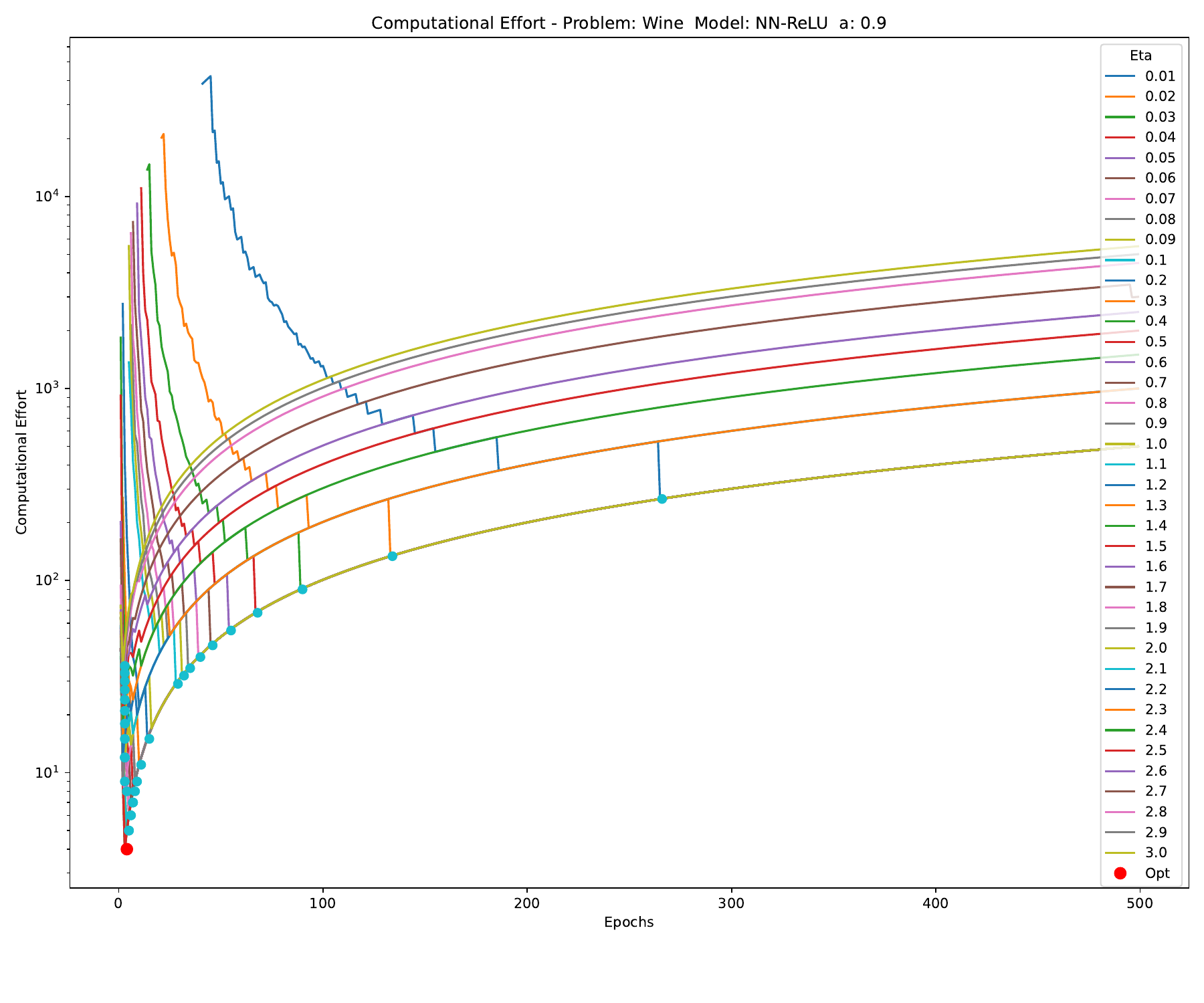} &
    \includegraphics[clip,trim=0 0 0 0,width=.52\linewidth]{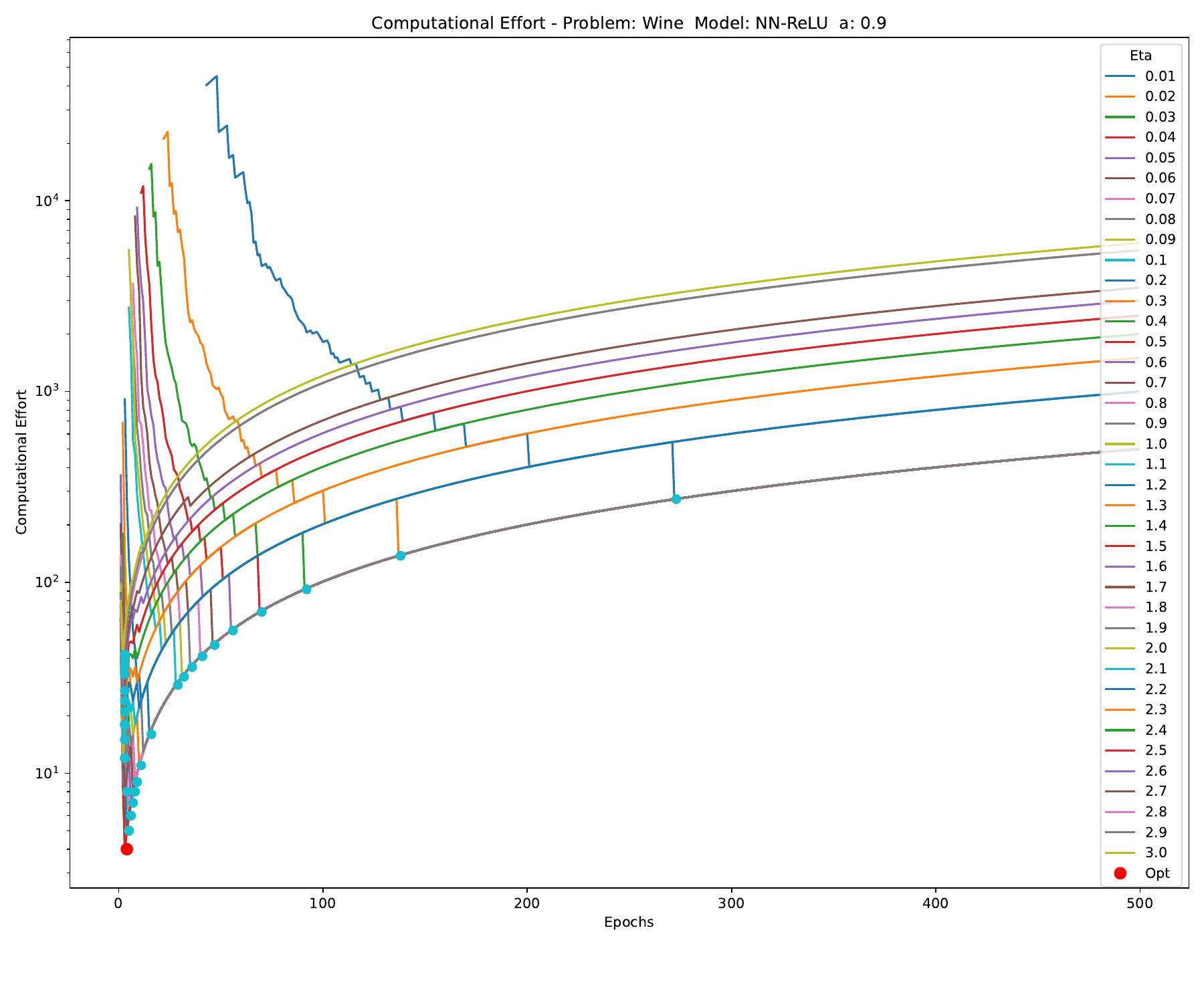} \\[-3mm]
  \end{tabular}
\caption{Computational effort ($a=0.9$) for NN-ReLU model (continued).}
\end{figure*}

\subsubsection{Training Computational Effort Dynamics}
\label{sec:train-comp-effort-dynam}

The plots confirm that the training computational effort $E$ is a
non-monotonic function of the number of epochs as we suggested in
Section~\ref{sec:hyperp-optim-ace}. They also reveal the presence of a
phase transition between the single- to multi-run regimes when the
condition $P(i, a, \eta) = z$ is met, as predicted theoretically in
the same section (see Equation~\ref{eq:phase_transition_condition}).

Indeed, they present the following three phases.  The transition between
the second and the third of these is where the single- to multi-run
phase-transition condition,   $P(i, a, \eta) = z$,   
we  predicted theoretically 
in Section~\ref{sec:hyperp-optim-ace} (Equation~\ref{eq:phase_transition_condition}) is triggered.

\paragraph{Infinite Phase}
Virtually all plots start with $E$ being $\infty$ when the number of
training epochs $i$ is too low with a given learning rate for the
model to reach the accuracy threshold $a$ in any of the 200
runs. Because infinite ordinate values cannot be plotted in a figure,
we simply omit the corresponding points in the plots.  For instance,
this can be observed for the \emph{training} computational effort, for
LR without regularisation on the DNA problem (see
Figure~\ref{fig:comp_effort_LR_Inf_model_0.9}), where the line for
$\eta=0.01$ starts around epoch 60 and the line for $\eta=0.02$ starts
around epoch 30, the computational effort being $\infty$ for all
preceding epochs.

\paragraph{Multi-Run Phase}
Past the aforementioned phase, that is, as soon as $E$ is finite,%
\footnote{With 200 runs and a maximum of 500 epochs, the
  largest finite $E$ is always smaller than 460,000 as indicated in
  Section~\ref{sec:impl-grad-steps}.}  $E$ tends to \emph{decrease}
overall, but in a \emph{zigzagging fashion}, i.e., presenting
\emph{several local minima} (e.g., in
Figure~\ref{fig:comp_effort_LR_Inf_model_0.9}, for LR-inf and DNA,
between epochs 60 and 150 for $\eta=0.01$ and epochs 30 and 70 for
$\eta=0.02$). This is due to short periods of monotonic ascent%
\footnote{In fact, the ascent is \emph{linear}, although this is not
  obvious from the plots as we use a logarithmic scale.}  associated
with the fact that the success probability $P(i,a,\eta)$ is increasing
with every additional epoch, and so the magnitude of the denominator
within the ceiling operation in
Equations~\eqref{app:eq:R_as_func_of_z_and_p_i}
and~\eqref{eq:comp_effort_ACE} increases. These ascent periods
continue until $P(i,a,\eta)$ has grown sufficiently big
to be equal to $z$, and so (as predicted in
Section~\ref{sec:hyperp-optim-ace}) the number of runs $R$ drops by 1,
with a final sudden reduction of the computational effort associated with
this discontinuity. This is the where the \emph{global minimum of $E$,
  $E^*$, for that $\eta$} is (cyan dots in each plot).

\paragraph{Single-Run Phase}
From there, any further increase in the number of epochs simply adds
more expenditure of gradient descent steps, without any tangible
(based on our success criterion) additional benefit. So, a
\emph{monotonically increasing phase} starts for $E$. In some cases,
in this phase \emph{the curves of $E$ for different $\eta$'s all
  collapse onto one line} (e.g., for LR without regularisation and the
BCW and Wine problems for $a=0.9$, see
Figure~\ref{fig:comp_effort_LR_Inf_model_0.9}). This happens when
success probabilities (see
Figure~\ref{fig:success_probability_LR_Inf_model_0.9}) saturate at a
common value, which does not depend on the learning rate.  In other
cases, the success probabilities for different learning rates may
saturate at different values (e.g., this happens for the NN-ReLU
model, see Figure~\ref{fig:success_probability_NN_ReLU_model_0.9}), so
\emph{the computational effort curves may converge to multiple
  asymptotes} (e.g., this happens for NN-ReLU for all problems, see
Figure~\ref{fig:comp_effort_NN_ReLU_model_0.9}).%
\footnote{The only exceptions to this
  monotonically-increasing-dynamics phase are cases where the effort
  has a \emph{middle hump} (see peak in training effort at around
  epoch 300 for NN-ReLU with $\eta=0.01$ on Iris for $a=0.9$ in
  Figure~\ref{fig:comp_effort_NN_ReLU_model_0.9} or peak for LR-Inf
  at the same epoch, $\eta$ and on the same problem but for $a=0.85$
  in Figure~\ref{fig:comp_effort_LR_Inf_model_0.85}) which is the
  result of the corresponding success probabilities plateauing at
  constants values for many epochs, before eventually restarting their
  ascent.}

\subsubsection{Validation Computational Effort Dynamics}
\label{sec:valid-comp-effort-dynam}

For problems and models where an accuracy of \emph{$a$ is comfortably
  achievable}, the plots of the validation \emph{computational effort
  are often identical or very close to those of the training
  computational effort}. So, validation computational efforts, too,
present the \emph{infinite, multi-run and single-run
  phases} discussed in the previous section.  The reason is that for
low enough values of $a$ the training and validation accuracy
distributions are essentially identical, resulting in almost identical
success probabilities.  For $a=0.85$, examples of this include the
BCW, Ecoli, Iris and Wine problems solved via NN-ReLU and LR without
regularisation (see SM's
Figures~\ref{fig:comp_effort_NN_ReLU_model_0.85}
and~\ref{fig:comp_effort_LR_Inf_model_0.85}).  For $a=0.9$, this is
still true for BCW, Iris and Wine solved via LR without regularisation
(see Figure~\ref{fig:comp_effort_LR_Inf_model_0.9}) and NN-ReLU (see
Figure~\ref{fig:comp_effort_NN_ReLU_model_0.9}).  So, in all such
problems also the training and validation optimal $E$ for each $\eta$
(cyan dots) tend to be close (or
identical), and so do the overall optimal $E^*$'s
(red dots).

However, for problems and models \emph{where achieving an accuracy $a$
  is more difficult, then the training and validation computational
  efforts can diverge significantly} with the validation computational
effort always being higher than the training computational effort,
which obviously applies also to the optimal validation effort
values. The reason for this is that in these circumstances some degree
of \emph{overfitting} has started to hit runs, and reaching the required
value of $a$ requires more epochs (hence higher $E$) for validation
accuracy than for training accuracy. Essentially, the bigger the
overfitting, the bigger the differences. For instance, for $a=0.9$,
examples of divergence between training and validation computational
efforts include the DNA problem solved via LR without regularisation
(Figure~\ref{fig:comp_effort_LR_Inf_model_0.9}) and NN-ReLU
(Figure~\ref{fig:comp_effort_NN_ReLU_model_0.9}).

In extreme cases, \emph{validation accuracy may never reach $a$}
irrespective of learning rate, so the validation computational effort
is always infinite (meaning the probability of success is zero) while
the training computational effort is finite.  For $a= 0.9$, this is
true for the Ecoli problem solved via LR without regularisation and
NN-ReLU, so the corresponding validation plots in
Figures~\ref{fig:comp_effort_LR_Inf_model_0.9}
and~\ref{fig:comp_effort_NN_ReLU_model_0.9} are empty.

Naturally, for even higher values of $a$, also training
computational effort can become infinite. This happens for instance,
for $a=0.95$, on Ecoli, with all the LDA models, and the two
regularised LR models (e.g., see Figure~\ref{fig:comp_effort_LDA_Inf_model_0.95}).

\subsubsection{Effect of Hyper-Parameter Choices on  Computational Effort}
\label{sec:effects-hyper-param-on-comp-effort}

Looking at the computational effort plots across models, problems and accuracy
thresholds $a$, we see that with computational efforts, the difference
between an optimal choice of learning rate and number of epochs and a
really bad one can be enormous, easily \emph{reaching or even
  exceeding 4 orders of magnitude}.  Examples of
this for $a=0.9$ are the case of LR without regularisation for DNA and
Iris (see Figure~\ref{fig:comp_effort_LR_Inf_model_0.9}) or the case
of NN-ReLU when applied to DNA and Ecoli (see
Figure~\ref{fig:comp_effort_NN_ReLU_model_0.9}) where the
computational effort $E$ ranges from under 10 
to  $\approx$100,000 gradient descent steps. Very many other examples
of this are shown in SM's Section~\ref{app:traintest_comp}.

With any learning rate (including bad ones), optimising the epoch number has a
significant impact on effort, but typically not exceeding 2 orders of
magnitude. Similarly, with a bad choice of number of epochs, the
difference between a good and a bad learning rate is typically between
1.5 and 2
orders of magnitude (see vertical spread of cyan dots in each plot in
the figures in Section~\ref{app:traintest_comp} of SM).

ACE co-optimises learning rate and number of epochs together resulting
in the \emph{minimum computational effort} indicated by the
red dots in the figures. Overall the
benefits of using ACE to optimise the learning rate and stopping
iteration are very clear if one compare the computational efforts
corresponding to the red dots with any other point in the graphs.

Note that all the plots in this section and those in
Section~\ref{app:traintest_comp} have been obtained using the
\emph{optimum number of runs for each configuration}. Had we not used
this (e.g., had we used the traditional approach of setting $R=1$),
\emph{the disparity of computational efforts between the optimal
configurations and bad ones would have been even greater}.

\medskip

In the following section we look at the optimal hyper-parameters
identified by ACE (including the values of the optimum number of runs,
which cannot be inferred from the plots discussed above) across all
values of $a$ tested.

\subsection{Optimal Hyper-parameters Identified by ACE}
\label{sec:optim-learn-hyperp}

In this section, we focus on the optimal hyper-parameters identified
by ACE and examine how they vary as a function of the threshold accuracy
$a$, which ranges from 0.75 to 1.0 in steps of 0.005.

To explore these variations, we employ plots that report the optimal
computational effort, learning rate, number of runs, and number of
epochs. Figures~\ref{fig:E_vs_accuracy_Pareto_LR_Inf_model}
and~\ref{fig:E_vs_accuracy_Pareto_NN_ReLU_model} illustrate these plots
for the LR-Inf and NN-ReLU models; however, full results are presented in
SM's Section~\ref{app:optim-learn-hyperp}.
In these figures, we depict the \emph{minimum computational effort}
$E^*$ using red stars and corresponding red lines. The \emph{optimal
number of epochs $i^*$} is denoted by green squares and green lines,
while the \emph{optimal number of runs $R^{*}$} is represented by blue
circles and blue lines. Additionally, we plot the \emph{optimal
learning rate $\eta^*$} using orange circles and orange lines.

\medskip

Before delving into the results, we want to make two general
observations:
\begin{itemize}
\item Firstly, the plots of $E^*$ (red lines) effectively represent
  the trade-off between the accuracy threshold $a$ and computational
  effort for optimally parametrised models. So, they are \emph{Pareto
    frontiers}. In other words, for any given $a$, $E^*$ is the best
  possible effort one can obtain across all parameter settings
  (learning rate, epochs, and number of runs) for a given model and
  problem. Naturally, as one would expect, in all such plots $E^*$
  \emph{is a non-decreasing function of $a$}.
\item Secondly, glancing at the figures, one immediately notices that,
  for a significant range of $a$ values, the \emph{lines representing}
  $E^*$ \emph{overlap with those representing}~$i^*$. This happens for
  all values of the accuracy threshold $a$ for which $R^{*}=1$, that
  is when \emph{single} (optimally hyper-parametrised) \emph{runs are
    highly successful} (reaching the required accuracy $a$ with at
  least 99\% probability) and, so,  $E^*=i^*$ (more on this below).
\end{itemize}

\begin{figure*}[p]
  \centering
  \begin{tabular}{c@{}c}
     Optimal Computational Effort and  & Optimal Computational Effort and   \\
     Hyper-Parameters (Training)  & Hyper-Parameters (Validation)  \\
    \includegraphics[clip,trim=70 30 30 30,width=.52\linewidth]{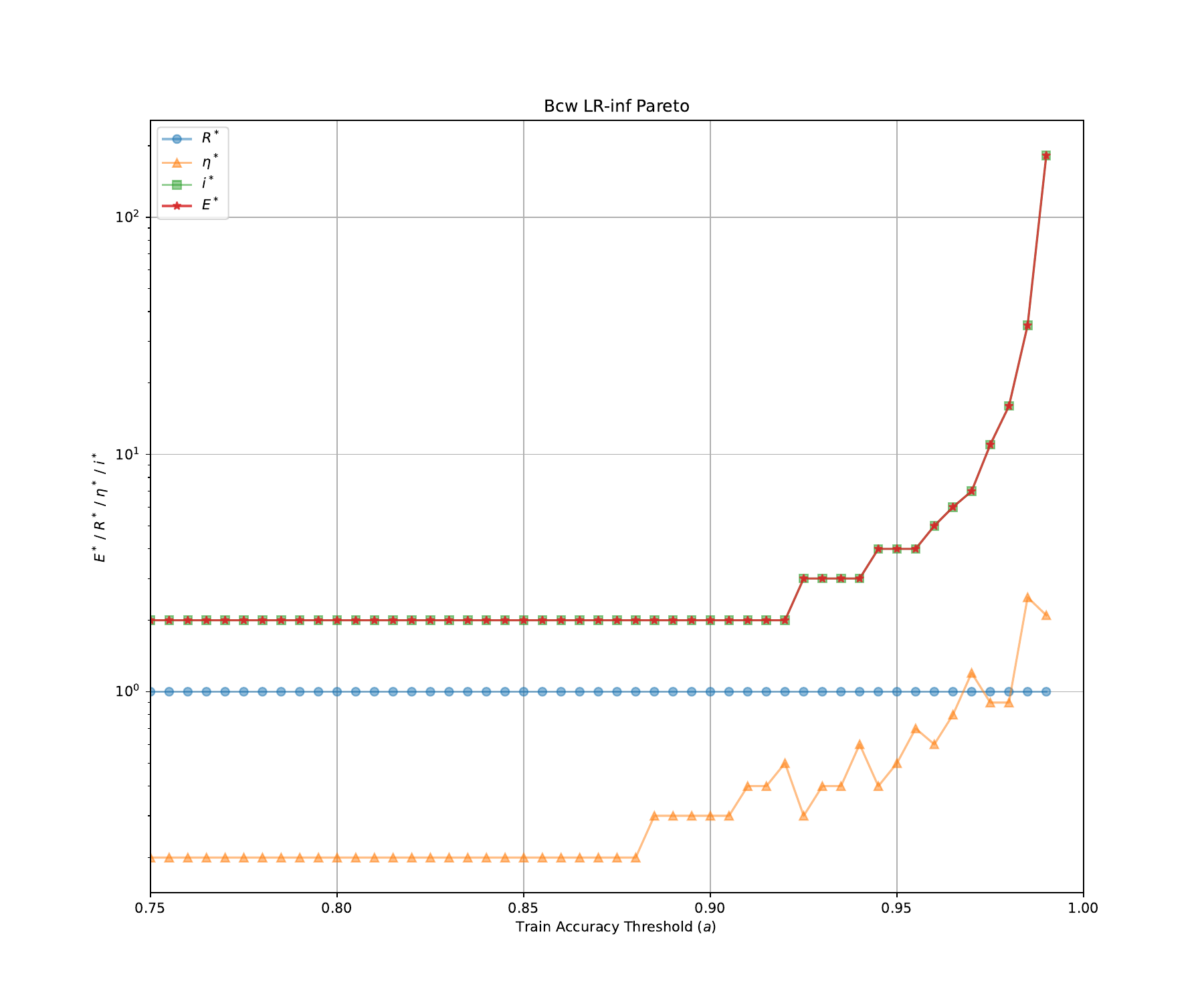} &
    \includegraphics[clip,trim=70 30 30 30,width=.52\linewidth]{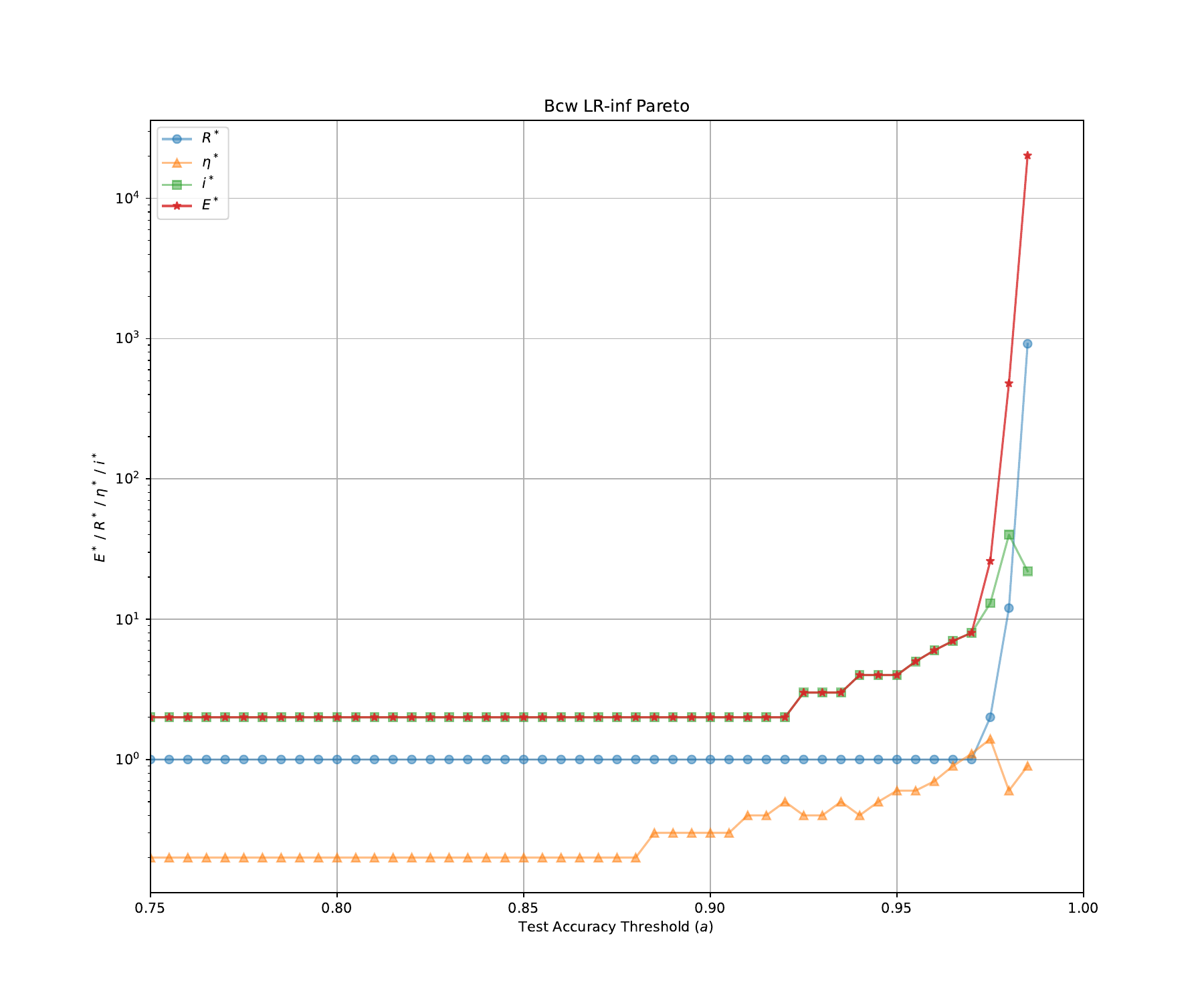} \\[-3mm]
    \includegraphics[clip,trim=70 30 30 30,width=.52\linewidth]{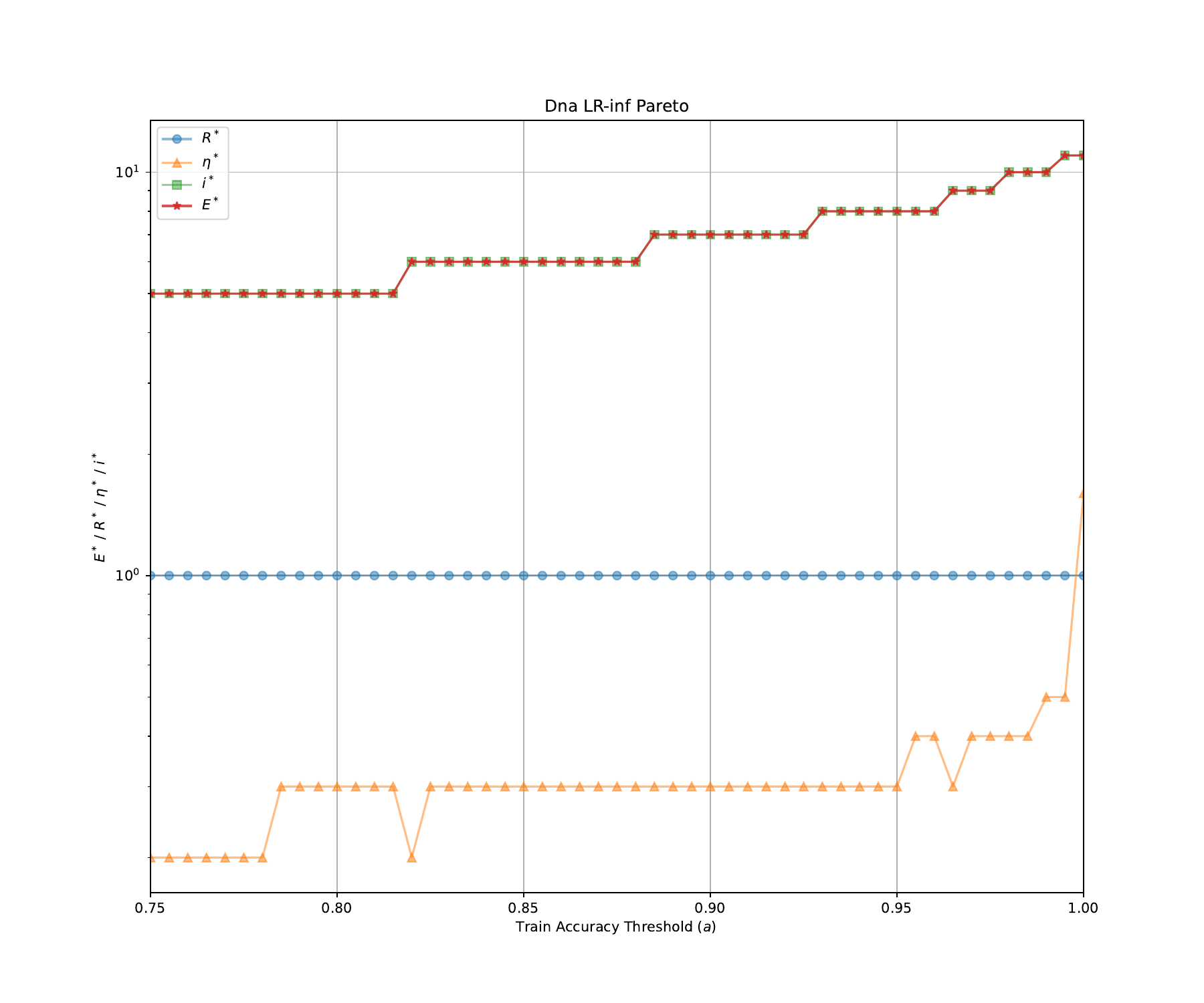} &
    \includegraphics[clip,trim=70 30 30 30,width=.52\linewidth]{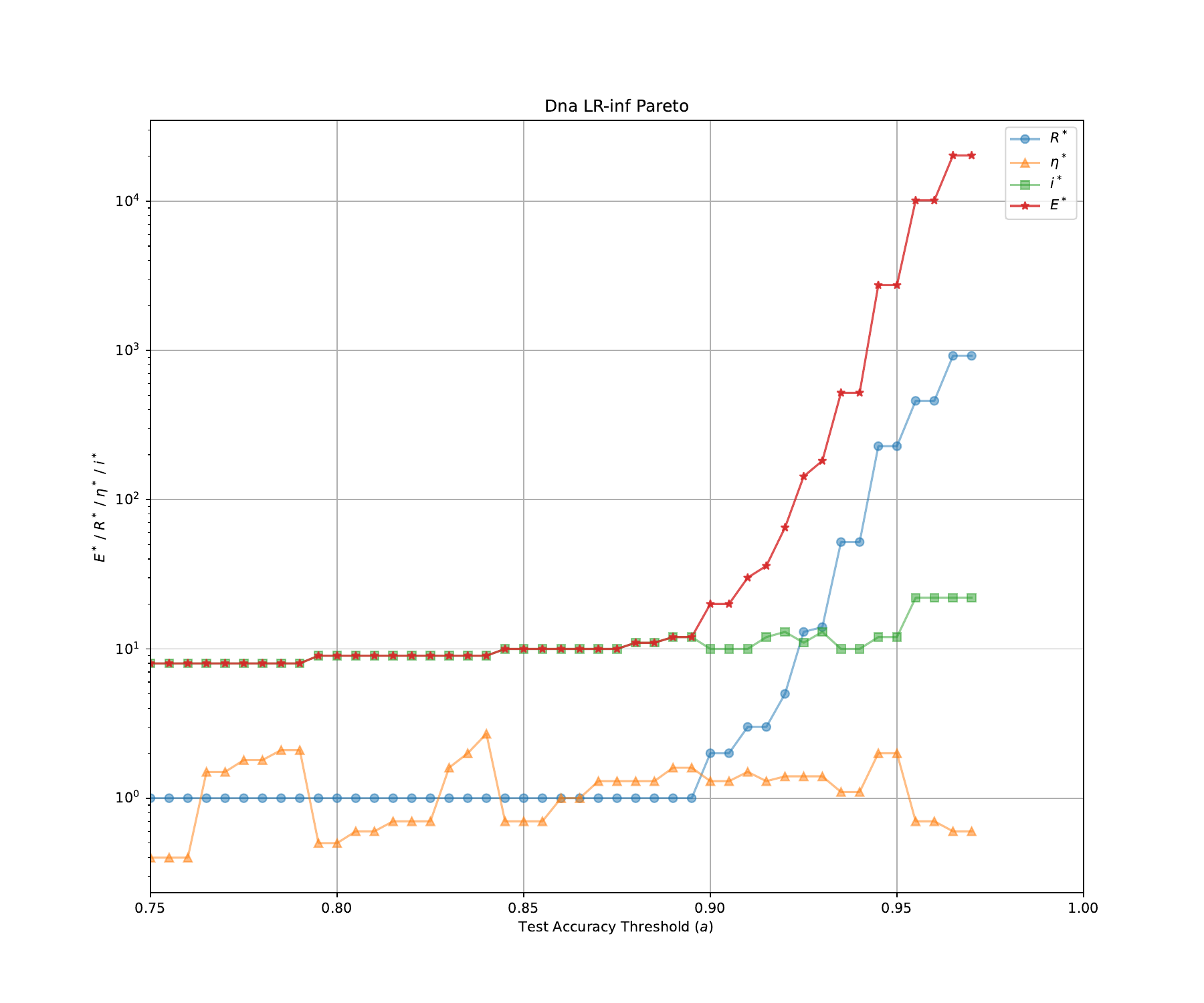} \\[-3mm]
    \includegraphics[clip,trim=70 30 30 30,width=.52\linewidth]{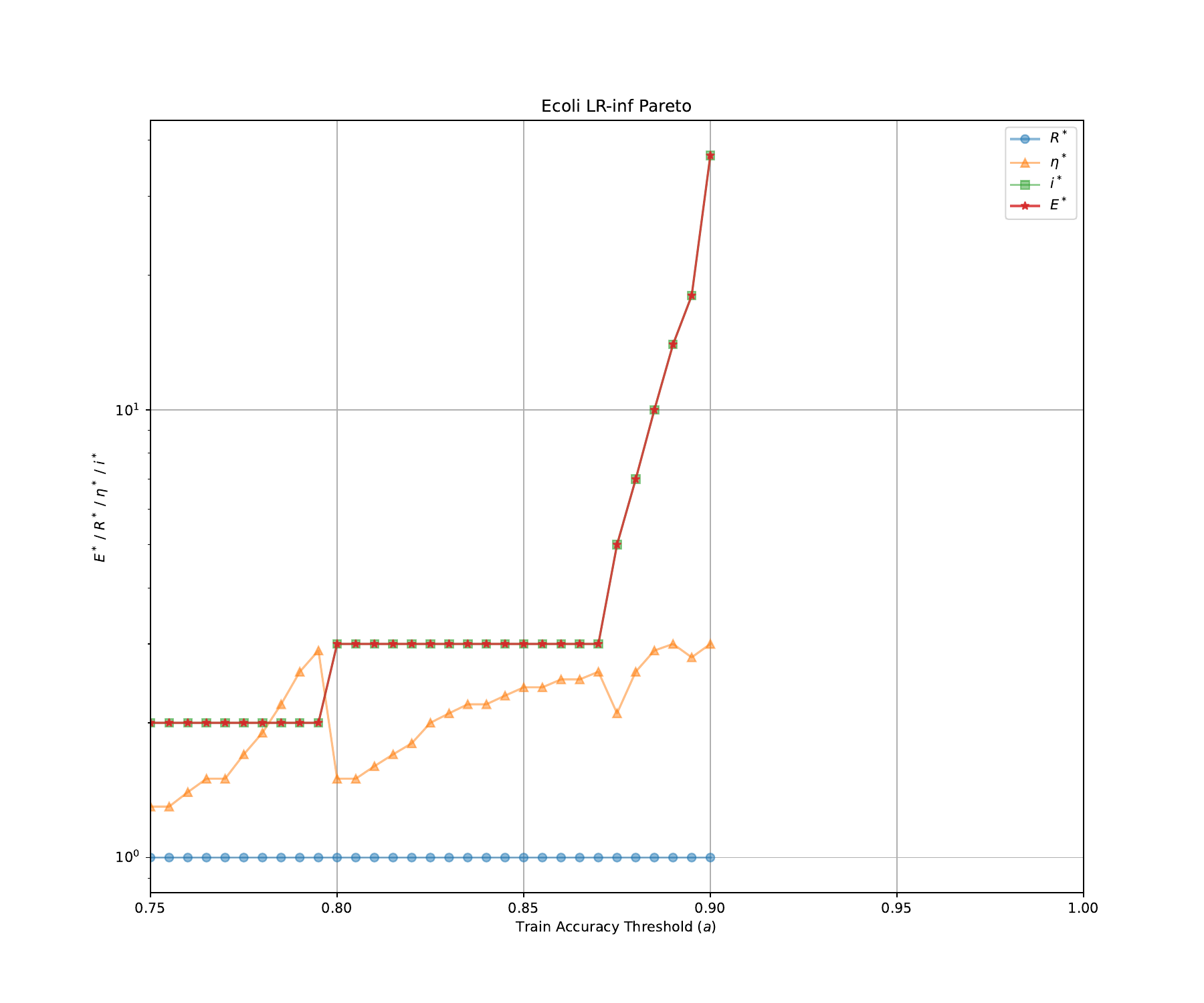} &
    \includegraphics[clip,trim=70 30 30 30,width=.52\linewidth]{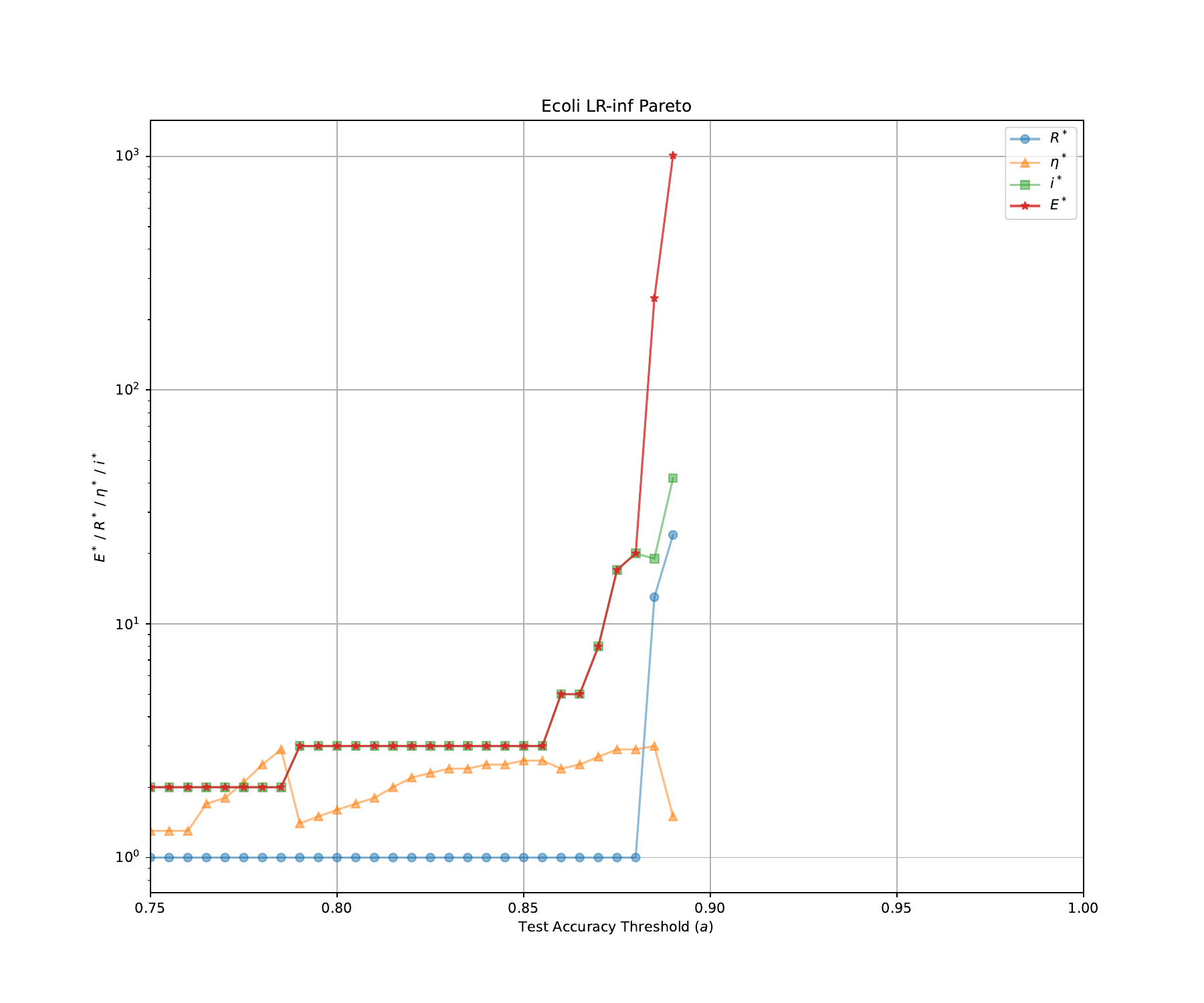} \\[-3mm]
  \end{tabular}
\caption{Optimal computational effort and hyper-parameters for different values of $a$ for LR-Inf model.}
  \label{fig:E_vs_accuracy_Pareto_LR_Inf_model}
\end{figure*}

\begin{figure*}[t!]
  \centering
  \ContinuedFloat
  \begin{tabular}{c@{}c}
     Optimal Computational Effort and  & Optimal Computational Effort and   \\
     Hyper-Parameters (Training)  & Hyper-Parameters (Validation)  \\
    \includegraphics[clip,trim=70 30 30 30,width=.52\linewidth]{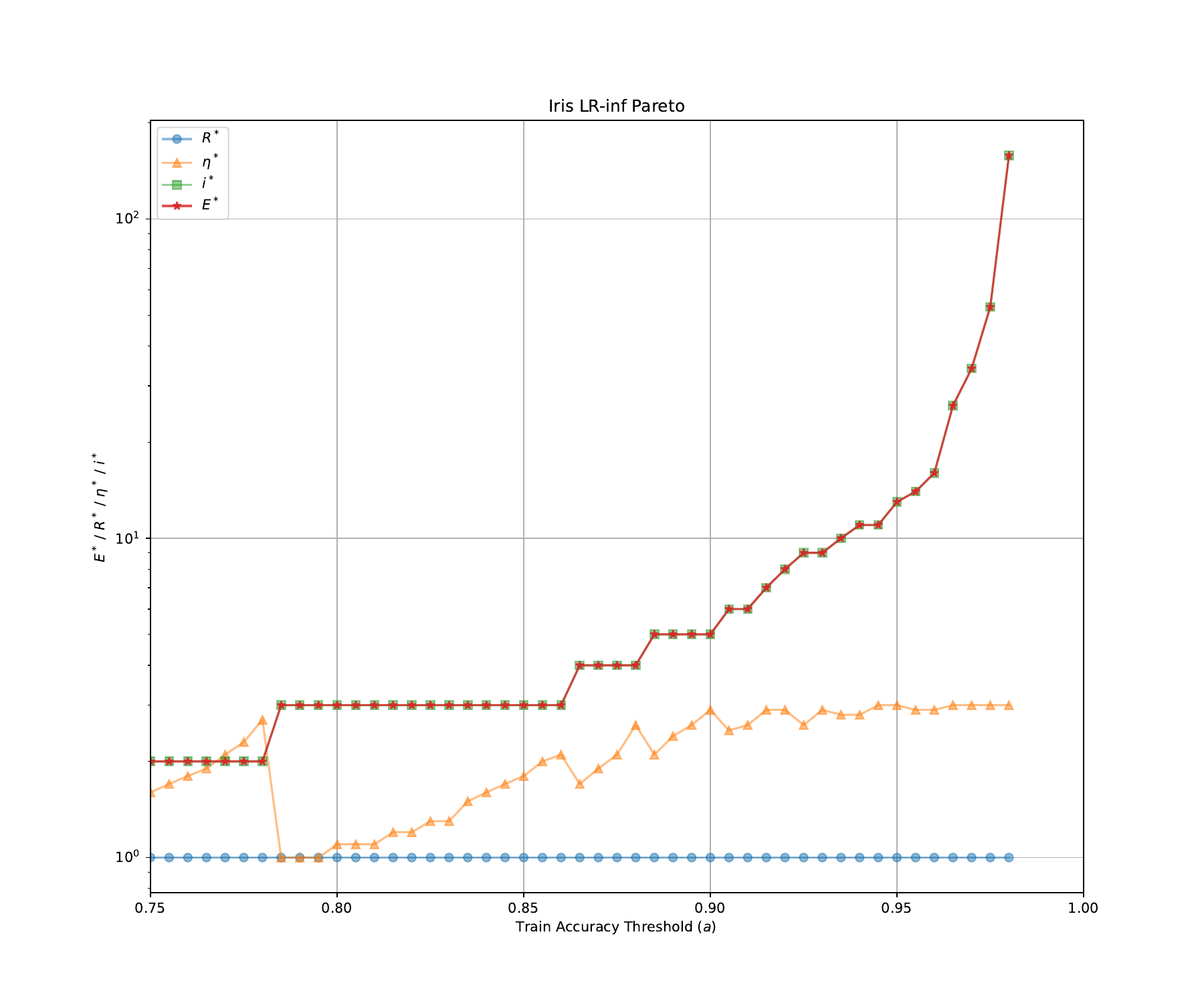} &
    \includegraphics[clip,trim=70 30 30 30,width=.52\linewidth]{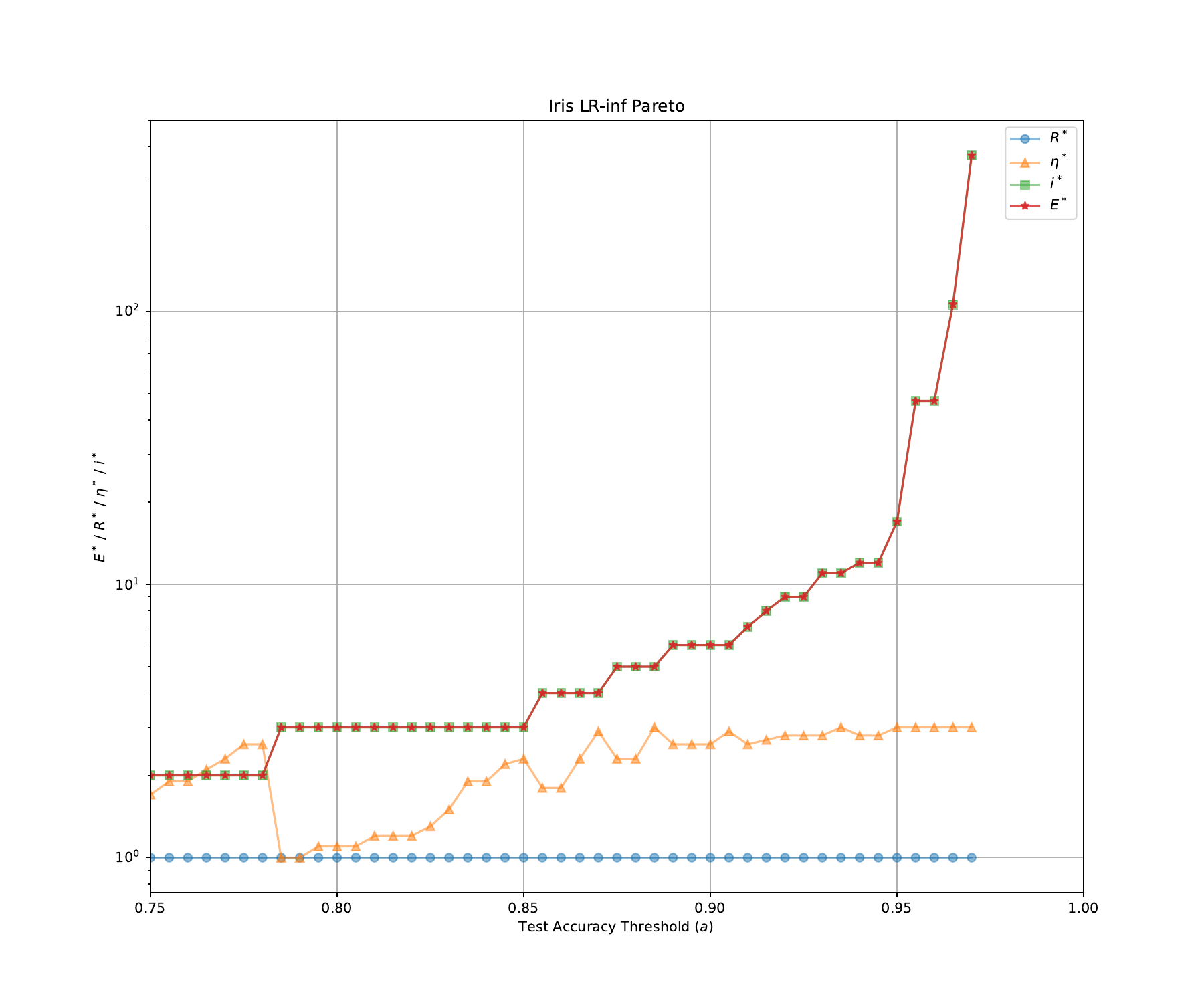} \\[-3mm]
    \includegraphics[clip,trim=70 30 30 30,width=.52\linewidth]{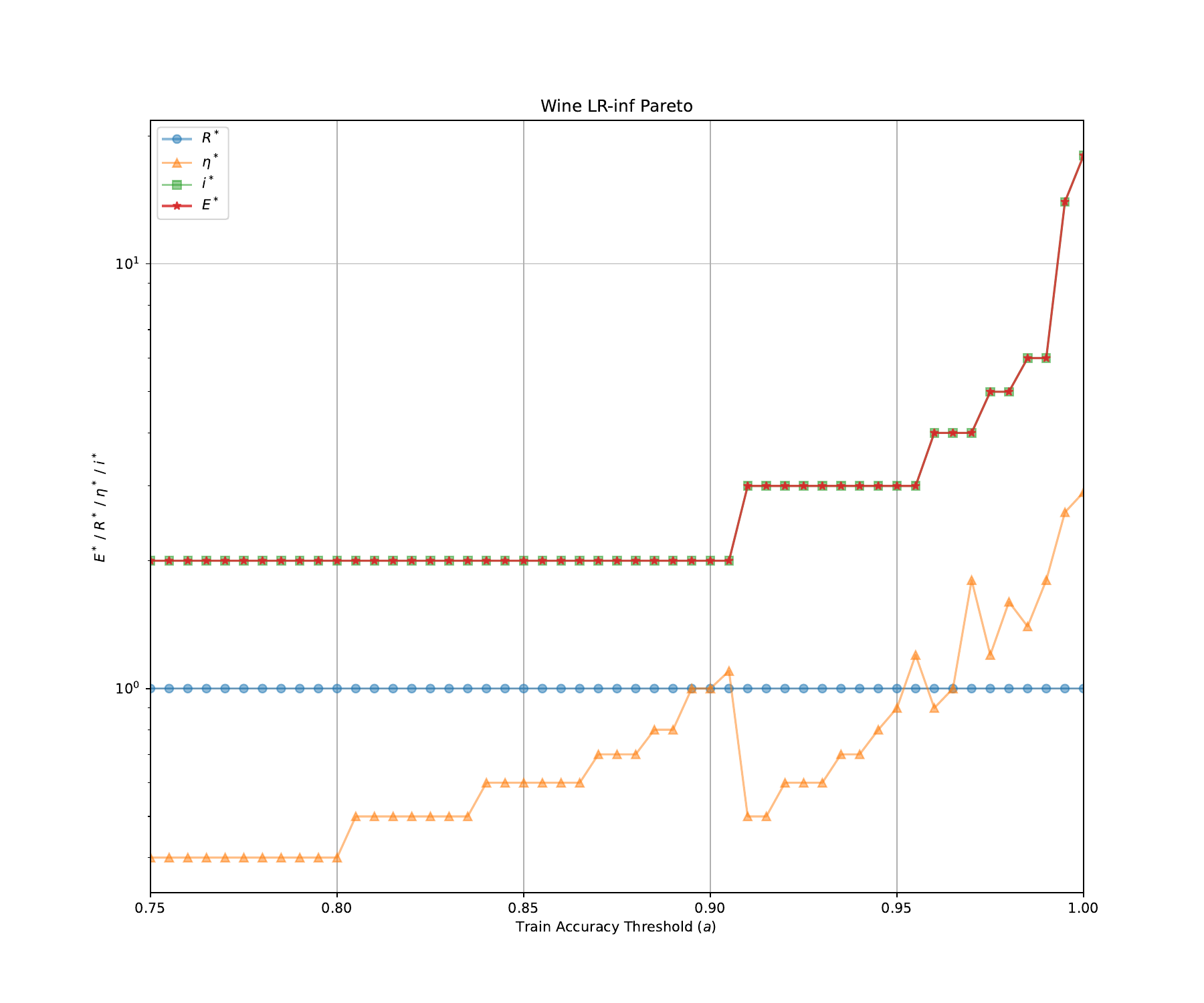} &
    \includegraphics[clip,trim=70 30 30 30,width=.52\linewidth]{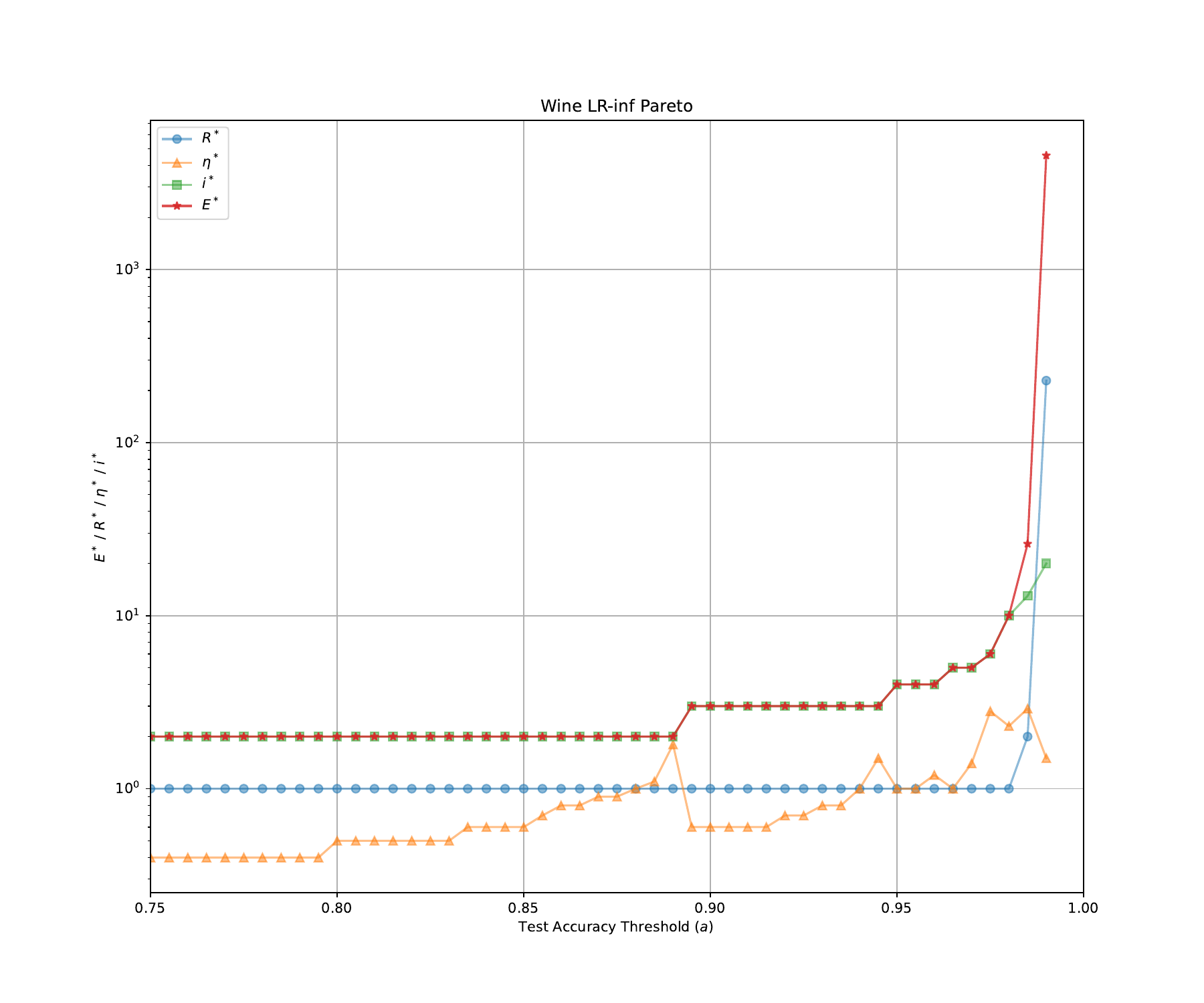} \\[-3mm]
  \end{tabular}
\caption{Optimal computational effort and hyper-parameters for different values of $a$ for LR-Inf model (continued).}
\end{figure*}

\begin{figure*}[p]
  \centering
  \begin{tabular}{c@{}c}
    Optimal Computational Effort and  & Optimal Computational Effort and   \\
    Hyper-Parameters (Training)  & Hyper-Parameters (Validation)  \\
    \includegraphics[clip,trim=70 30 30 30,width=.52\linewidth]{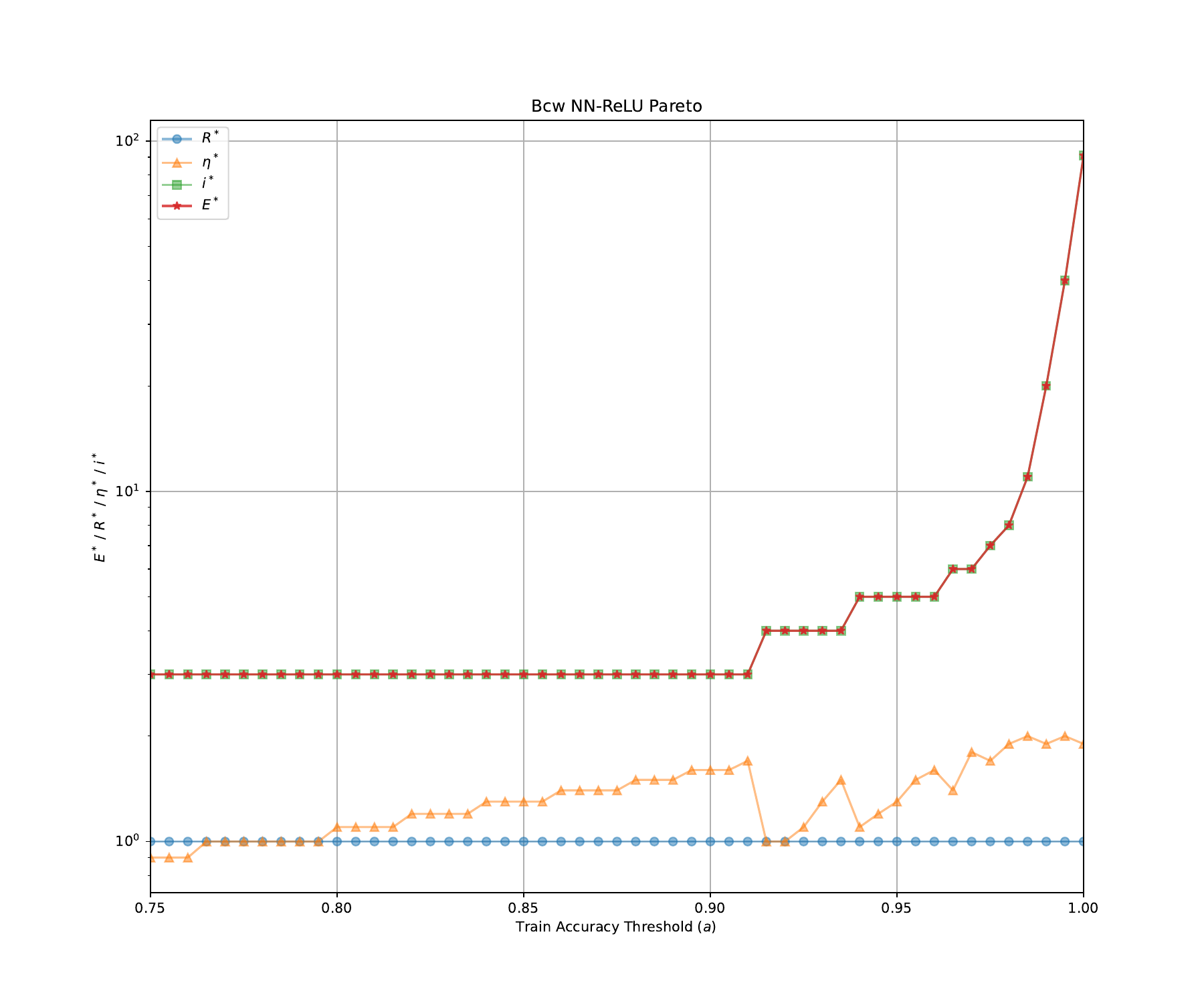} &
    \includegraphics[clip,trim=70 30 30 30,width=.52\linewidth]{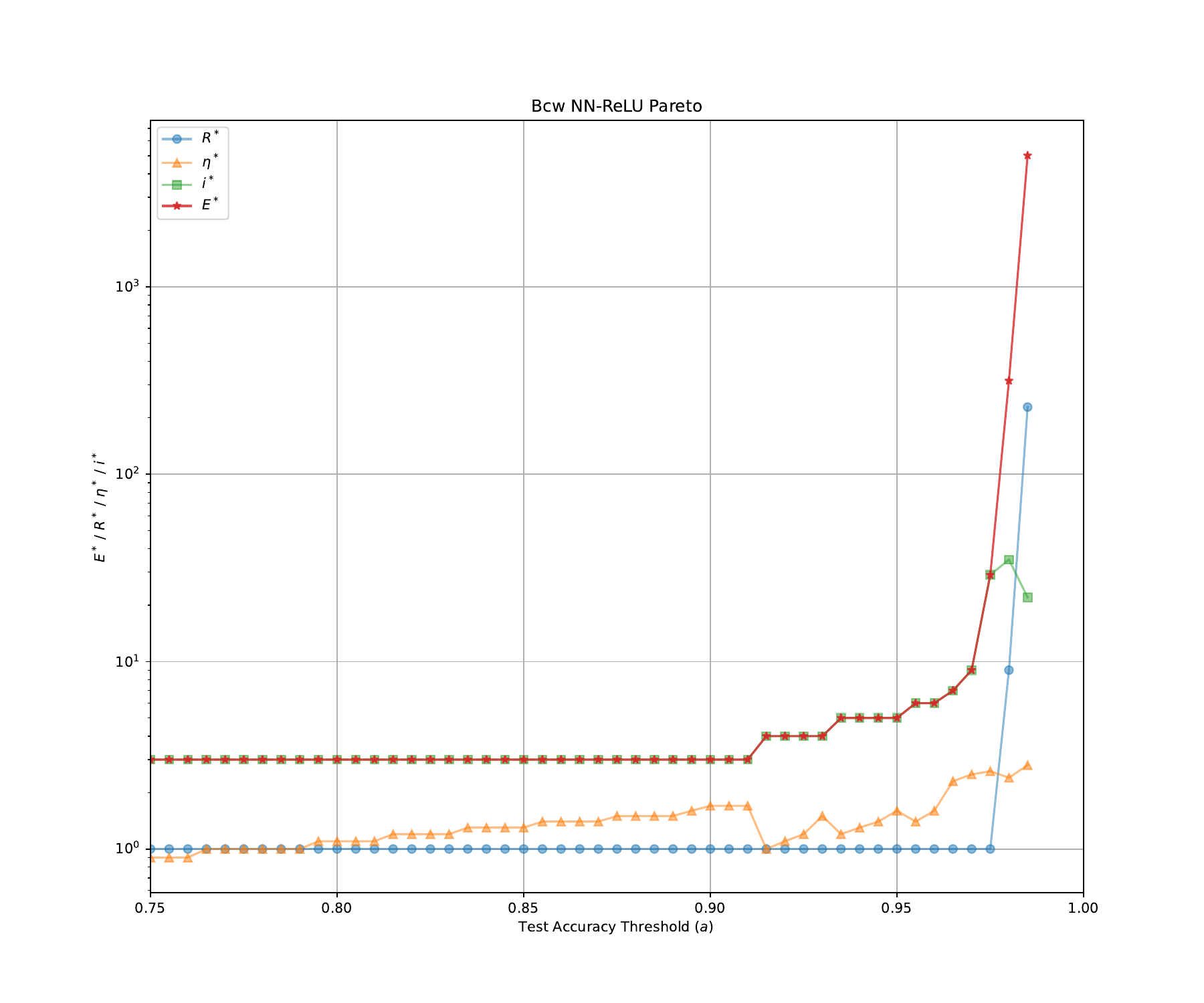} \\[-3mm]
    \includegraphics[clip,trim=70 30 30 30,width=.52\linewidth]{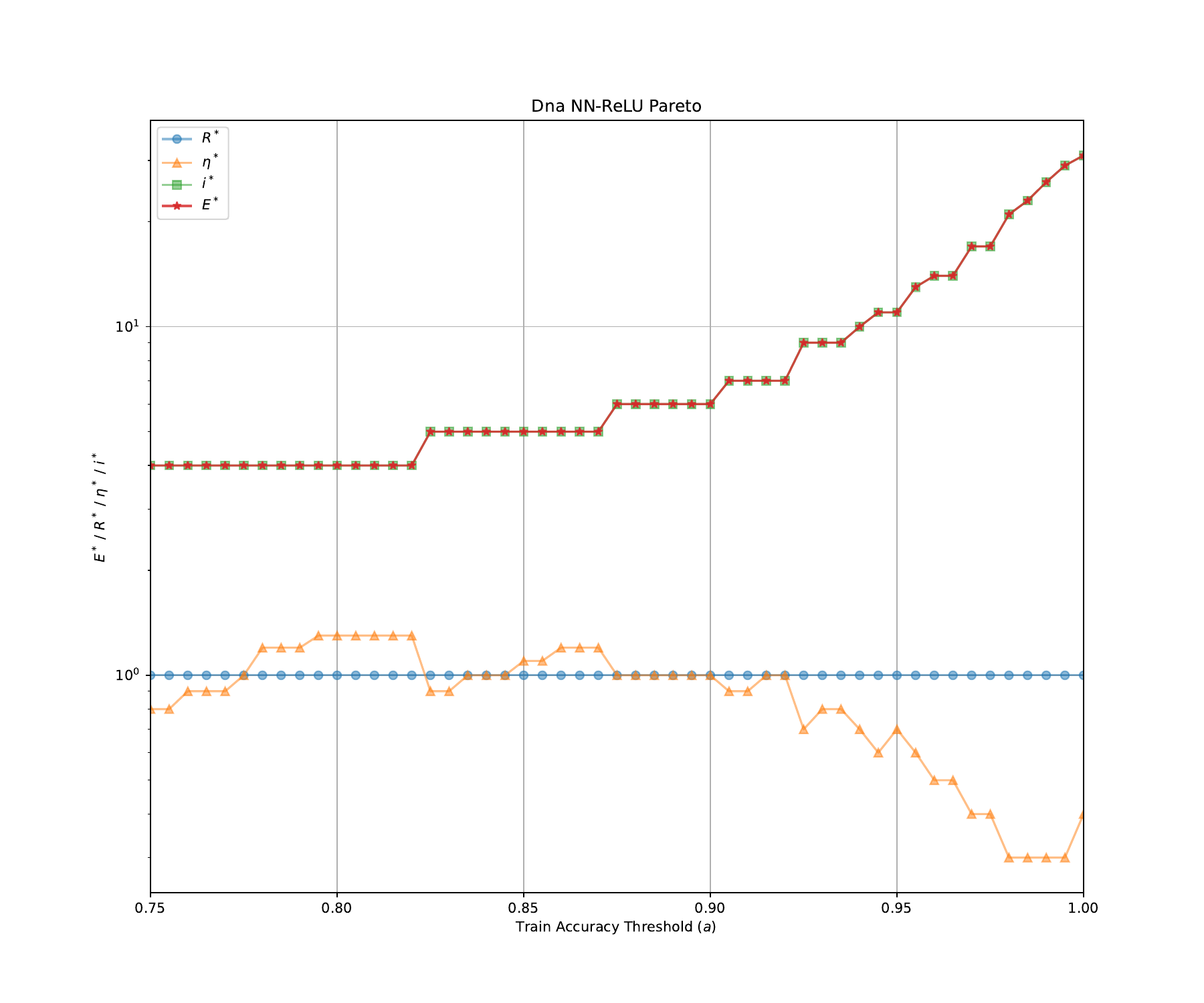} &
    \includegraphics[clip,trim=70 30 30 30,width=.52\linewidth]{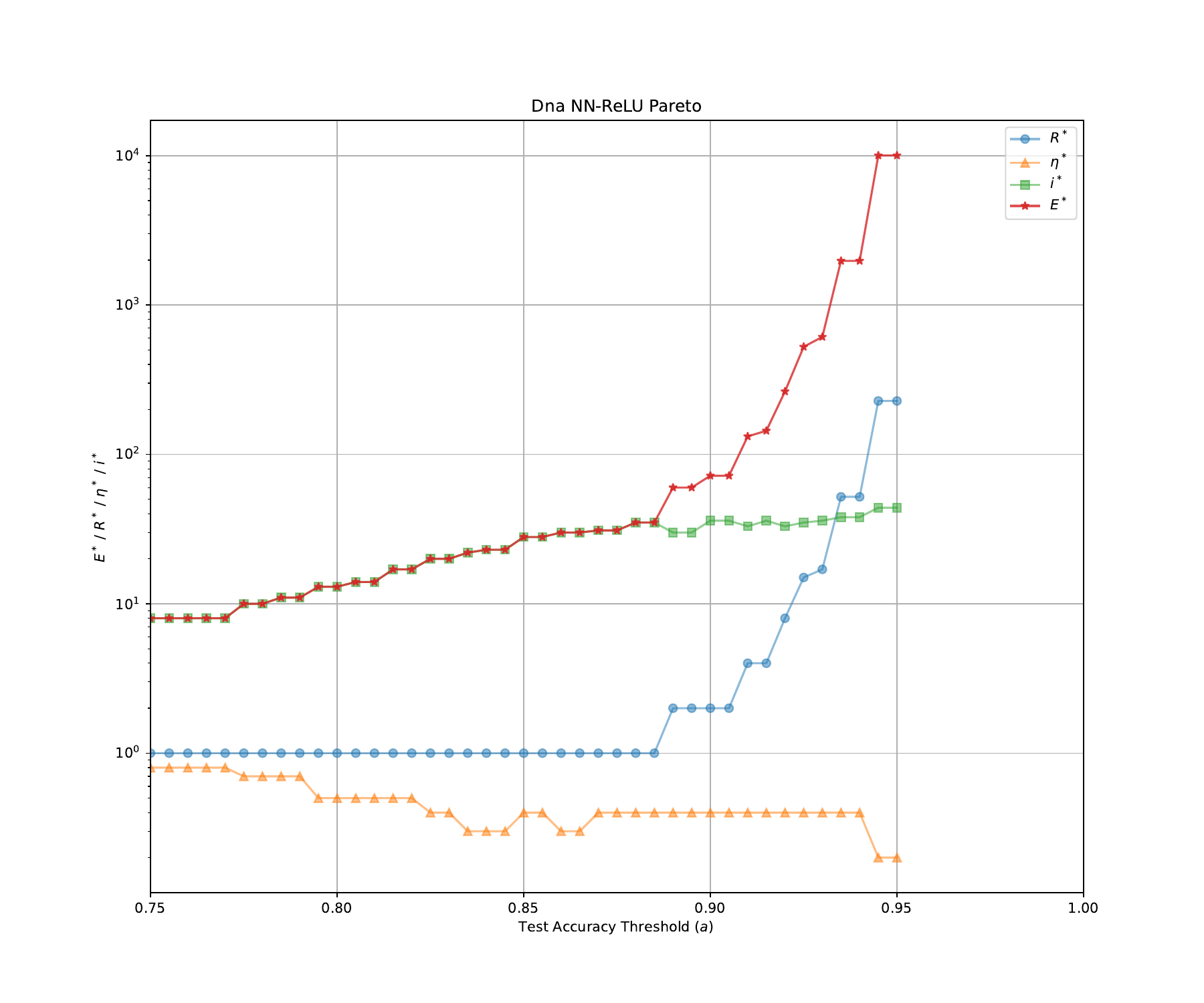} \\[-3mm]
    \includegraphics[clip,trim=70 30 30 30,width=.52\linewidth]{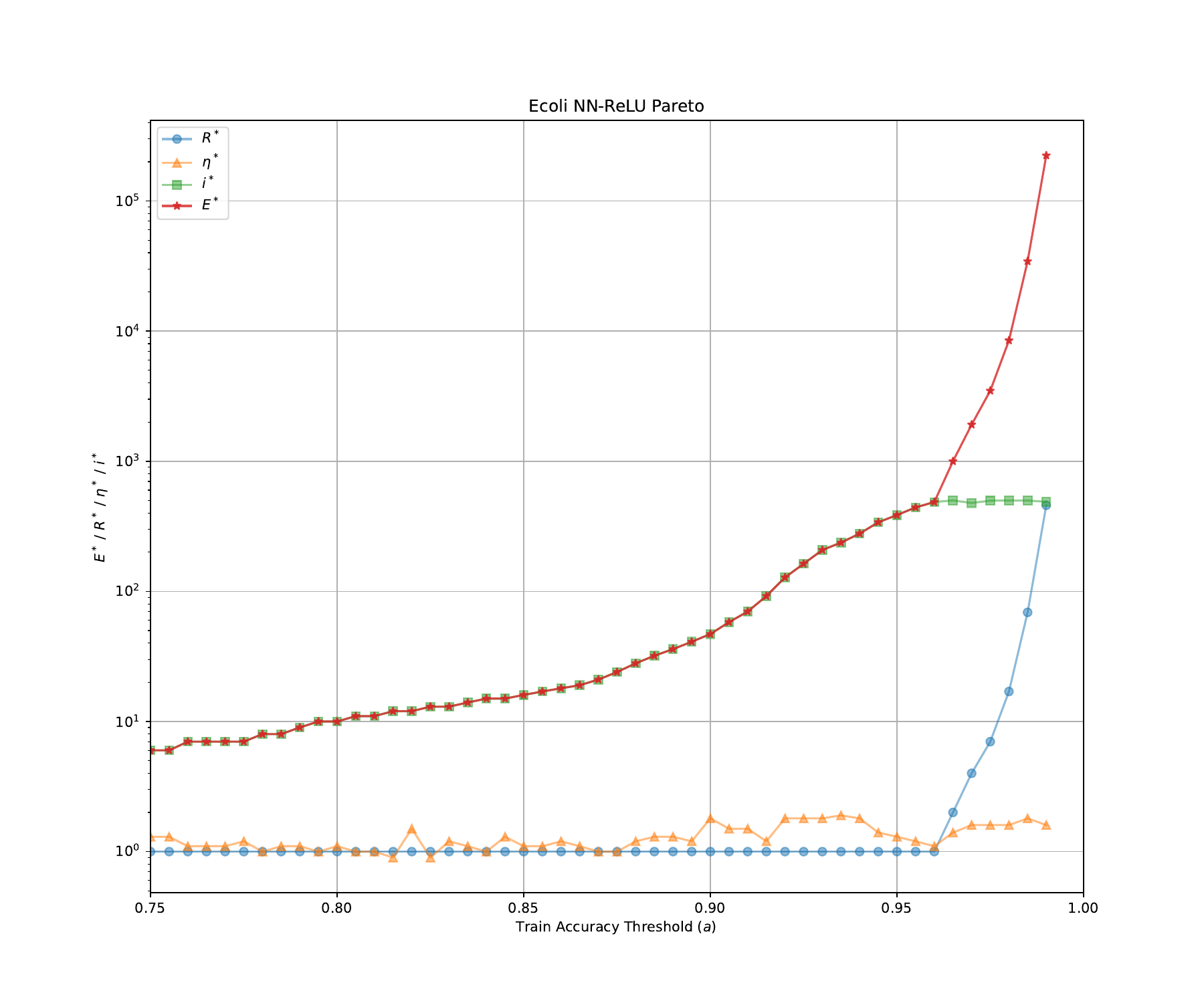} &
    \includegraphics[clip,trim=70 30 30 30,width=.52\linewidth]{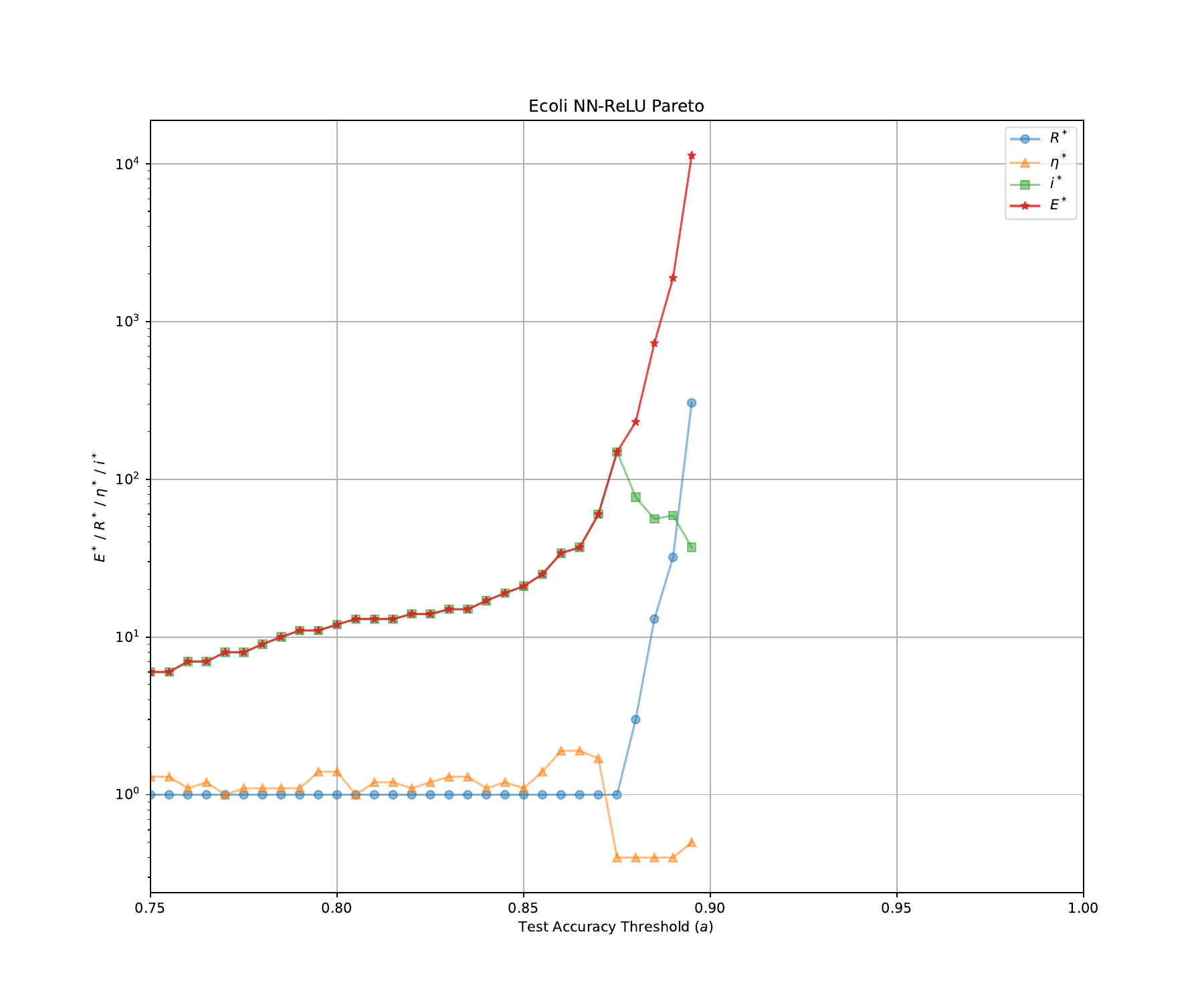} \\[-3mm]
  \end{tabular}
\caption{Optimal computational effort and hyper-parameters for different values of $a$ for NN-ReLU model.}
  \label{fig:E_vs_accuracy_Pareto_NN_ReLU_model}
\end{figure*}

\begin{figure*}[!t]
  \centering
  \ContinuedFloat
  \begin{tabular}{c@{}c}
    Optimal Computational Effort and  & Optimal Computational Effort and   \\
    Hyper-Parameters (Training)  & Hyper-Parameters (Validation)  \\
    \includegraphics[clip,trim=70 30 30 30,width=.52\linewidth]{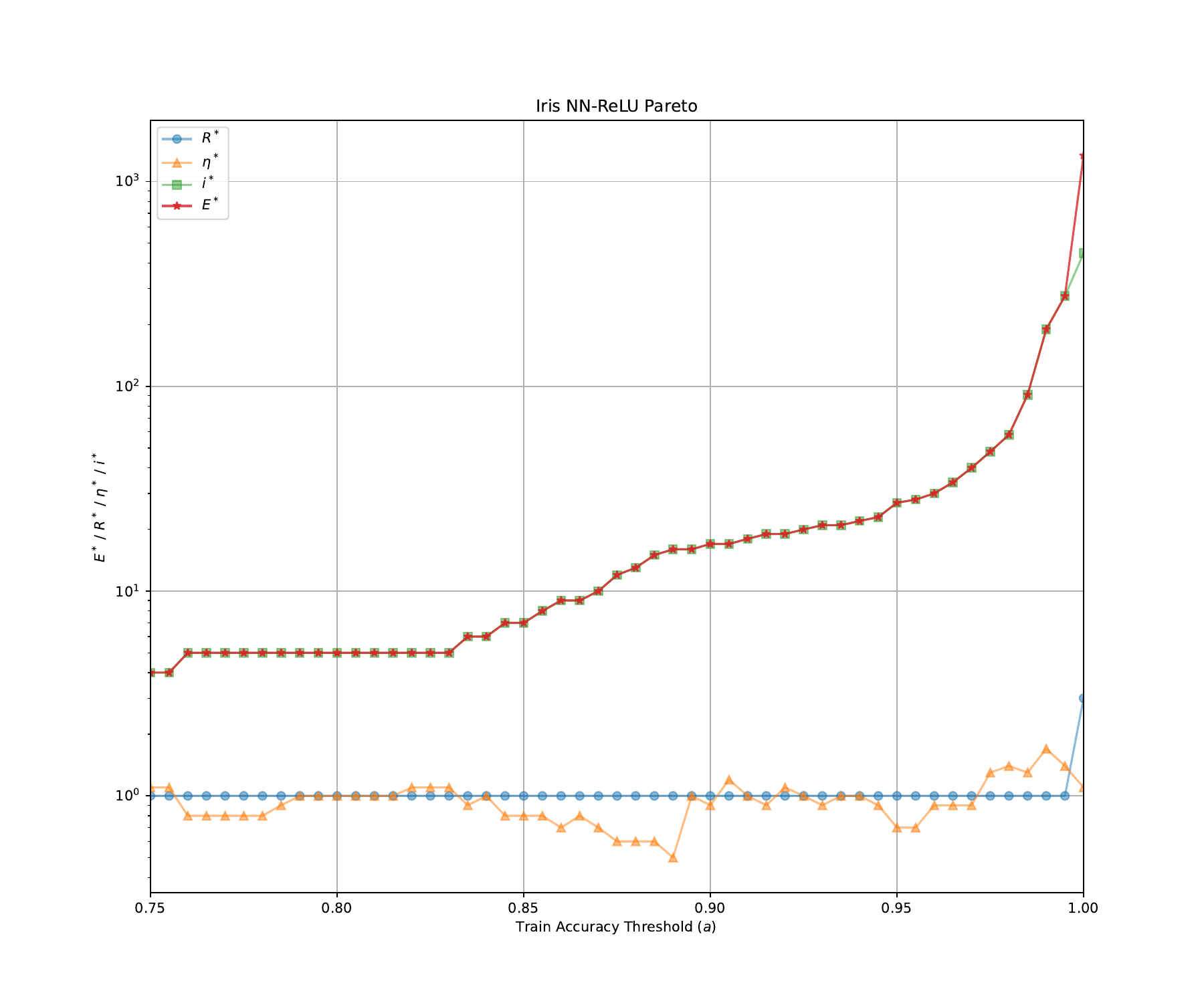} &
    \includegraphics[clip,trim=70 30 30 30,width=.52\linewidth]{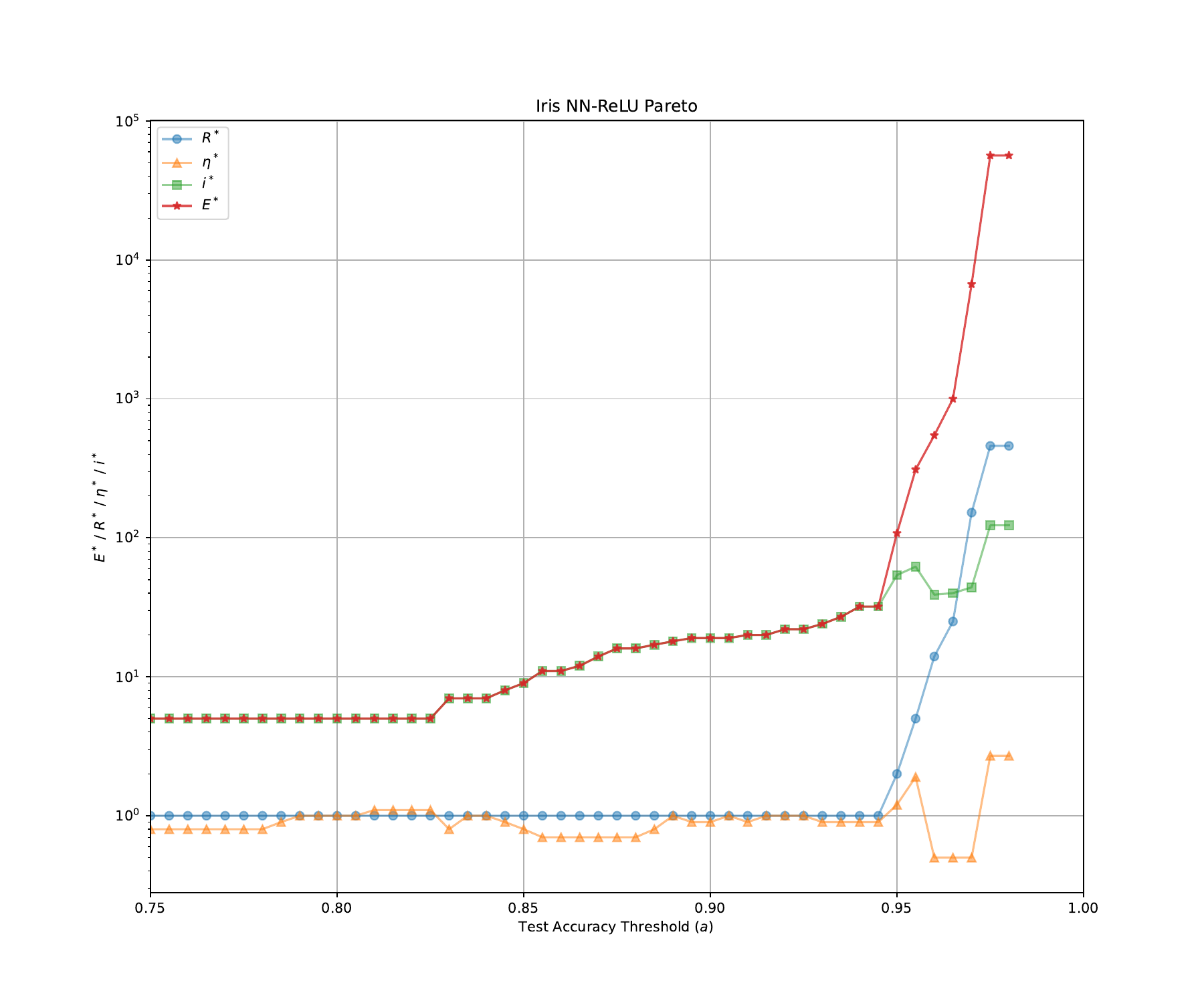} \\[-3mm]
    \includegraphics[clip,trim=70 30 30 30,width=.52\linewidth]{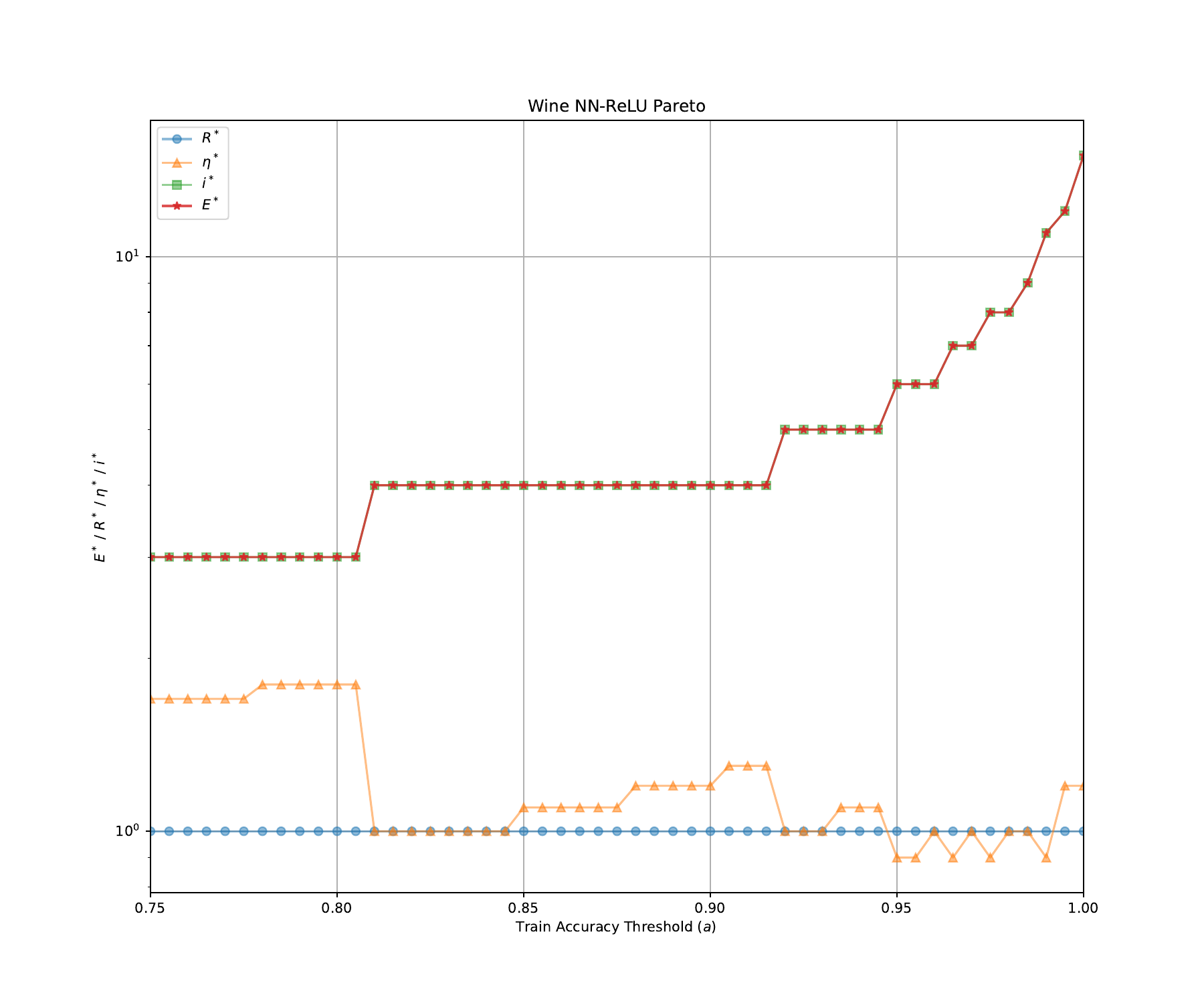} &
    \includegraphics[clip,trim=70 30 30 30,width=.52\linewidth]{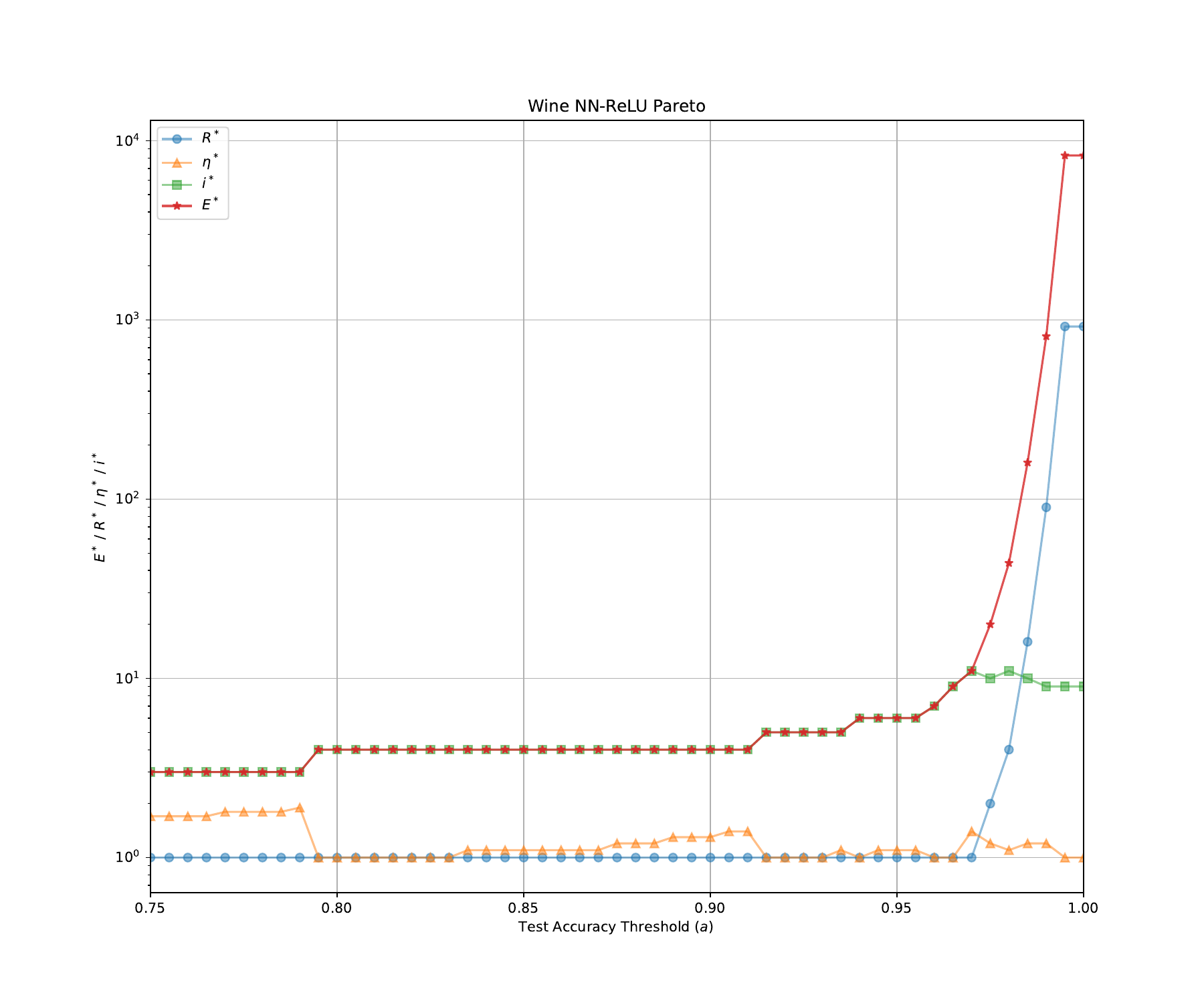} \\[-3mm]
  \end{tabular}
\caption{Optimal computational effort and hyper-parameters for different values of $a$ for NN-ReLU model (continued).}
\end{figure*}

\subsubsection{Optimal Training Hyper-parameters}
\label{sec:optim-train-hyper}

Like the computational effort dynamics
(Section~\ref{sec:train-comp-effort-dynam}), also the \emph{optimal
  training hyper-parameters}, illustrated in
Figures~\ref{fig:E_vs_accuracy_Pareto_LR_Inf_model}
and~\ref{fig:E_vs_accuracy_Pareto_NN_ReLU_model} (left columns) for
LR-Inf and NN-ReLU (and provided in
Section~\ref{app:optim-learn-hyperp} for all other models),
\emph{present several phases separated by important phase transitions
  occurring at particular values of $a$}.%
\footnote{These are represented by ranges of values for $a$ and are
  linked to different phenomena than in
  Section~\ref{sec:train-comp-effort-dynam}, where the phases and
  phase transitions were associated to different values of the number
  of epochs $i$.} We describe them below.

\paragraph{Easier-Success Single-Run  Phase}

This first phase comprises sufficiently low values of $a$ that result
in \emph{both} $R^{*}=1$ (which, as indicated above, means that single runs
are highly successful) \emph{and $E^*$ being small} (easy success)
\emph{and nearly constant} (as a function of~$a$). Exactly when this
phase ends depends on the model and problem but it is widespread as
almost all models can achieve training accuracies of $a= 0.8$ (or
more) within less that 10 gradient descent steps irrespective of
problem (when $\eta$ is optimal).%
\footnote{This is particularly remarkable given that 4 out of 5 of our
  problems are multi-class ($K>2$) (so chance-level accuracy is way
  less than 50\%).}  Naturally, while $E^*$ and $i^*$ are semi-static
in this phase, the learning rate may show much more variation to meet
the requirements of ever increasing $a$'s. For instance, this happens
for E-coli, Iris and Wine for LR-Inf (see
Figure~\ref{fig:E_vs_accuracy_Pareto_LR_Inf_model}) and for BCW and Wine
for NN-ReLU (see Figure~\ref{fig:E_vs_accuracy_Pareto_NN_ReLU_model}).

\paragraph{Harder-Success Single-Run  Phase}

This second phase comprises higher values of $a$ that still can be
best achieved with \emph{a single run} (so $R^{*}=1$) but now
\emph{the minimum computational effort grows often exponentially or
  super-exponentially} with $a$.%
\footnote{Respectively represented by \emph{a constant slope} and
  \emph{a positive curvature} in the $E^*$ plots as the ordinate
  scales are logarithmic.} In this phase, often \emph{the learning rate}
$\eta$ \emph{changes significantly} from the values associated with the first
phase to cope with the more stringent demands imposed by the higher
$a$'s. As shown in
Figures~\ref{fig:E_vs_accuracy_Pareto_LR_Inf_model}
and~\ref{fig:E_vs_accuracy_Pareto_NN_ReLU_model} (left columns),
exponential growths in $E^*$ occur with Ecoli for LR-Inf and with DNA
for NN-ReLU, while super-exponential growths occur with BCW, Iris and
Wine for LR-Inf and with BCW and Iris for NN-ReLU.

\paragraph{Multiple-Runs Phase}

For some model-problem pairs, at higher values of $a$, \emph{the 
  phase transition corresponding to} $ P(i, a, \eta) = z$ (see
Equation~\eqref{eq:phase_transition_condition} in
Section~\ref{sec:hyperp-optim-ace})  occurs leading to a third phase
where $ P(i, a, \eta) < z$ and, so, $E^*\ne i^*$, with $i^*$ suddenly stopping to increase and the
number of runs $R^*$ \emph{ramping up exponentially} (or
super-exponentially).%
\footnote{Given the limitations of our simulations indicated in
  Section~\ref{sec:impl-grad-steps}, in all cases $R^*<920$.}  One
such case is shown in
Figure~\ref{fig:E_vs_accuracy_Pareto_NN_ReLU_model}, for NN-ReLU on
the Ecoli problem when $a>0.96$, where $R^*$, from being 1 at $a=0.96$
rapidly reaches almost 500 for $a=0.99$. Correspondingly, also $E^*$
grows exponentially from about 500 gradient descent steps for $a=0.96$
to almost 250,000 gradient descent steps for $a=0.99$.

This phase transition from the regime $R^*=1$ to the exponential
growth in $R^*$ and $E^*$ corresponds to the situation where
\emph{irrespective of the learning rate, a single gradient descent is
 not  the best approach} for the learning algorithm to satisfy the demands
imposed by very high $a$'s. Instead, ACE finds that a form of
\emph{random search in the space of GD trajectories on the error
  surface}, different initialisations (random seeds) leading to
end-trajectories of different quality (as discussed in
Section~\ref{sec:char-object-funct}), \emph{is more efficient}.

While the multiple-runs phase transition is present only in one
problem for NN-ReLU and no problems for LR-Inf in
Figures~\ref{fig:E_vs_accuracy_Pareto_LR_Inf_model}
and~\ref{fig:E_vs_accuracy_Pareto_NN_ReLU_model}, \emph{it is quite
  common}.  Indeed, when looking at the training hyper-parameters
across all models in Section~\ref{app:optim-learn-hyperp} of SM, we
find that \emph{it occurs in over 50\% of the 55 model--problem
  pairs}, particularly with Ecoli and Iris, but also, to a lesser
extent, BCW and Wine (while it is virtually absent in the case of the
DNA dataset).  That is, multiple runs are optimal not only in models
with non-convex error surfaces (where one might assume that GD could
be trapped by local optima), but also in linear and regularised
models, which have convex and strictly convex error surfaces.

\paragraph{Infinite-Effort Phase}

In 60\% of the model--problem pairs, \emph{a final phase transition
  occurs for even higher values of $a$, where both $R^*$ and $E^*$
  become $\infty$}. This is an indication that \emph{none} of the
7,800 ($=$ 200 seeds $\times$ 39 learning rates) training attempts led
to achieving the target training accuracy. The last value of $a$
before $R^* = E^* = \infty$ thus represents the \emph{training-set
  learnability limit} for a particular model and problem for
full-batch GD (with restarts).

\medskip

\paragraph{Optimal Learning Rates for Training} 

While above we describe in detail how $R^*$, $i^*$ and $E^*$ vary,
here we want to look at the optimal learning rates.

In our experiments we tested values of $\eta$ between 0.01 and 3.0,
i.e., covering a range of $2 \, \sfrac 1 3$ orders of
magnitude. However, the results presented in this section show that
irrespective of phase, the \emph{optimal learning rate $\eta^*$ for
  training most often varies within the $[0.1, 1]$}, which
  represents \emph{one} order of magnitude.

These \emph{values of $\eta^*$ are large} in comparison to those
suggested in both the scientific literature and ML/AutoML libraries,
which tend to set $\eta \ll 1$. For instance, values of $\eta$ in the
range $[0.00001,0.01]$ were used in~\citep{smith2017cyclical}, while
initial values of 0.01, 0.001 and 0.002 were suggested
in~\citep{ruder2016overview} for Adagrad, RMSprop and AdaMax,
respectively. Learning rates in the range $[10^{-6},0.05]$ were used
in~\citep{sutskever2013importance}. Also, within standard ML
libraries, the default value of $\eta$ used in the \texttt{GD} class
in Keras is 0.01, while in Pytorch it is 0.001. The value 0.001 is
also used as the initial $\eta$ value in the \texttt{Adam} class in
both frameworks and also in the \texttt{MLPClassifier} class in
Scikit-learn (which uses Adam by default). Finally, within the AutoML
community, values of $\eta$ in the range $[0.001,0.1]$ were suggested
for Auto-Net-2.0~\cite[Chapter 7]{hutter2019automated} and values in
the range $[10^{-4},10^{-2}]$ are suggested in the Keras Tuner
documentation~\citep{omalley2019kerastuner,keras_tuner_docs}.
  
So, it appears as if, in our experiments, ACE effectively discovered
that
\emph{superconvergence}~\citep{smith2019super,oymak2021provable,smith2021origin,smith2018disciplined,li2019towards,mohtashami2023special}
which we discussed in Section~\ref{sec:train-valid-error-surfaces},
can be extremely useful to achieve high training accuracies  in GD
with restarts, not just to provide better generalisation.

\subsubsection{Optimal Validation Hyper-parameters}
\label{sec:optim-valid-hyper}

In this section we 
focus now on the \emph{optimal hyper-parameters for validation}
reported in the right columns of
Figures~\ref{fig:E_vs_accuracy_Pareto_LR_Inf_model}
and~\ref{fig:E_vs_accuracy_Pareto_NN_ReLU_model} and the figures in
Section~\ref{app:optim-learn-hyperp}.

\paragraph{Phases}

It is easy to see from the figures that the phase transitions
described in the previous section also occur for validation, although
the generally they happen at lower values of $a$ than for training.

For instance, \emph{in all but one case} out of 55 the easier- and
harder-success single-run phases, where virtually all single,
optimally hyper-parametrised runs attain the required accuracy $a$
(i.e., $R^*=1$), are shorter than for training. 

In other words, the multi-run phase, associated with an
exponential growth of both $E^*$ and $R^*$, \emph{starts sooner} (for
a smaller $a$). This phase  is also \emph{more widespread} than for
training, being present in more than 90\% of the 55 model--problem
pairs.

Correspondingly, also the infinite-effort phase (where both $R^*$ and
$E^*$ become infinity) \emph{occurs for smaller values of $a$ and
  happens more frequently}%
\footnote{More specifically, this happens in nearly 95\% of the
  model--problem pairs, the only exceptions being the Wine problem for
  NN-ReLU and for the SVMs with $C=0.1$ and $C=0.01$.}  than for
training.

The last value of $a$ before $R^* = E^* = \infty$, often requires
hundreds of runs to be achieved. So, the accuracy $a$ could only be
achieved by runs that sample the extreme upper tail of the validation
accuracy distribution, in other words, using extreme
``cherry-picking'', which is a frowned-upon practice
in ML~\citep{henderson2018deep,kapoor2022leakage}.
So, we could call
the corresponding value of $a$ as the \emph{cherry-picking limit} for
a model-problem pair.

To avoid falling pray of this replicability issue, one would need take
the highest $a$ value such that $R^*=1$ as a realistic limit for
validation accuracy.

\paragraph{Optimal Learning Rates for Validation}

Finally, turning to the learning rate, for any given model and problem, the
\emph{optimal learning rates $\eta^*$ for validation follow a very
  similar dynamics than for training and are most often limited to the
  range 0.1--1.0}. So, here too \emph{learning rates are large},
resulting in the well-known generalisation benefits of superconvergence.

So,  $\eta\in [0.1,1]$ is a better choice for both training and validation
when using GD with restarts.

\subsubsection{Superconvergence and Computational Effort}
\label{sec:superc-comp-effort}

It is important to contextualise our findings regarding large learning
rates and short runs within the existing machine learning literature,
particularly alongside the phenomenon of \emph{superconvergence}. As
previously discussed, superconvergence demonstrates that
large learning rates act as a powerful regulariser, promoting
generalisation by forcing the optimiser out of sharp, narrow valleys
and into wider, flatter minima. However, our findings indicate that
\emph{this regime is not only beneficial for generalisation, but is also
strictly optimal for minimising training computational effort}.

These two observations are highly complementary. Large learning rates
enable a rapid, aggressive traversal of the error surface. While this
aggressiveness risks divergence or overshooting in a traditional
single-run paradigm, the ACE framework—which calculates the expected
effort across multiple independent restarts—fundamentally shifts the
optimisation calculus. Statistically, it is vastly more efficient to
execute numerous short, aggressive runs that have a small but non-zero
probability of hitting the target accuracy very early, rather than
executing a single, conservative run with a small learning rate that
ensures convergence but requires an order of magnitude more gradient
steps. Thus, the rapid landscape exploration that superconvergence
leverages to find generalisable flat minima is the exact same
mechanism that ACE exploits to statistically minimise the expected
hitting time for a training threshold.

\subsection{Model Selection}
\label{sec:model-selection}

\subsubsection{Model-dependent Problem Difficulty for Gradient Descent}
\label{sec:probl-diff-grad}

As indicated at the beginning of the article, Koza used the minimum
computational effort as a measure of problem difficulty for genetic
programming and also to compare different forms of GP (including GP
with automatically-defined functions and without them).
In the same
vein, we feel that our plots of $E^*$ as a function of $a$ can help
assess \emph{problem difficulty for different forms of machine
  learning based on GD}, both from the training and the validation
points of view.

While for sufficiently low values of $a$, for both training and
validation, most problems are easy --- requiring 10 or fewer gradient
descent steps ($E^*\le 10$) for most models --- \emph{bigger differences in
difficulty exist} at the other end of the spectrum, that is, for high
values of $a$.

This is illustrated, for instance, by the case of LR-Inf
(Figure~\ref{fig:E_vs_accuracy_Pareto_LR_Inf_model}), where looking at
\emph{training difficulty} \emph{for high $a$'s}, DNA and Wine are
easiest requiring ${\approx} 10^1$ gradient-descent steps, followed by
BCW, E-coli and Iris requiring ${\approx} 10^2$ steps. The case of
NN-ReLU (Figure~\ref{fig:E_vs_accuracy_Pareto_NN_ReLU_model}) has a
wider range of problem difficulties, DNA and Wine
still being the easiest problems (${\approx} 10^1$ gradient-descent
steps), followed by BCW (${\approx} 10^2$ steps), Iris
(${\approx} 10^3$ steps) and Ecoli (${\approx} 10^5$ steps).  While
all our problems are all considered standard test problems in ML, their
difficulty is perceived very differently by LR-Inf and
NN-ReLU. \emph{For LR-Inf all problems are similarly difficult}, the
minimum number of gradient descent steps required for reliable
training ($E^*$) spanning 1 order of magnitude. On the contrary,
\emph{for NN-ReLU some problems are trivial while others are
  difficult}, the effort spanning 4 orders of magnitude.

Looking at \emph{validation difficulty} for high $a$'s, we see that
these are generally higher than the corresponding efforts for training
and go to infinity for many more values of $a$ than for training, both
of which was expected.  For LR-Inf, Ecoli and Iris are on par
(${\approx} 10^3$ steps) with BCW, DNA and Wine are all one order of
magnitude harder (${\approx} 10^4$ steps). For NN-ReLU, BCW, DNA,
E-coli and Wine are on par (${\approx} 10^4$ steps) while Iris is one
order of magnitude harder (${\approx} 10^5$ steps). So, we see a
\emph{compression in the range of problem difficulties for validation}.

The bottom lines are: (1) not all problems are equally difficult for
all models and (2) relative problem difficulty varies between training and
validation. Thus, irrespective of whether one is concerned with
training or validation, it stands to reason that \emph{one should do
  model selection by picking the model(s) for which a problem is
  easiest from the computational effort standpoint}.

\medskip

In the following sub-sections we consider different ways to
do so.

\subsubsection{Model Selection based on Target Accuracy}
\label{sec:model-select-based-on_a}

To look at model selection in more detail, in
Figure~\ref{fig:model_selection} we report the optimal computational
effort recorded across problems for all 11 ML models considered and
all values of $a$ between 0.75 and 1.0. The plots in the figure have
being obtained by simply redrawing the $E^*$ plots from
Figures~\ref{fig:E_vs_accuracy_Pareto_LR_Inf_model}
and~\ref{fig:E_vs_accuracy_Pareto_NN_ReLU_model} (as well as the
corresponding figures for all others models reported in
Section~\ref{app:optim-learn-hyperp} of SM) into \emph{one plot per
  problem} (for both training and validation). This way of looking at
the results makes it possible to see how difficult each problem is for
each model for different values of $a$.

The black dashed lines at the bottom of each plot show the \emph{lower
  envelope} for the efforts plots, representing the \emph{minimum
  computational effort across ML models} for each particular
problem. So, this represents a \emph{Pareto frontier}, i.e., the
best that can be achieved in terms of effort for any given $a$ for a
given problem, irrespective of model. 

From just visually inspecting the plots, it is clear that for certain
models there is no value of $a$ where they overlap with the Pareto
frontiers. Thus such models can be \emph{deselected} (i.e., excluded
from consideration) straight away as there are other models that would
achieve the target accuracy at a lower cost in terms of gradient
descent steps. For instance, for the \emph{training} results, this is
the case for DNN-ReLU on BCW and DNA, and also for LDA-Inf on DNA and
Iris.

In relation to which model should be selected for a given problem and
required accuracy, it is clear that \emph{model selection should be
  based on how close a model is to the Pareto frontier}, the ones on
the frontier being optimal.

As many lines overlap near or on the frontier, for each value of $a$
in Figure~\ref{fig:non_dominated_models} we explicitly report which
models provide the frontier's minimum effort --- and thus are
\emph{non-dominated} by other models --- across all problems, making
selection of a model based a target accuracy $a$ much easier. In the
figure, \emph{the blue dots at any given chosen accuracy $a$ represent
  equivalent models on the Pareto frontier}. That is, they all provide
at least the require accuracy with the same effort. Any model not
presenting a blue dot at that accuracy is suboptimal.

Looking specifically at the \emph{training results} in the figure, we
see clearly that for some problems and ranges of $a$ values, there are
many equally-good models (e.g., all LRs and all SVMs are non-dominated
for $a\le 0.94$ on training accuracy for BCW and $a\le 0.90$ for
Wine).  On the contrary for other problems, only fewer models are
non-dominated, the extreme case being DNA where, for all but the
highest values of $a$, only NN-ReLU and LR-0.01 are non-dominated.

We also see that, other than for DNA, rarely the NN-based models are
competitive, except for some problems and with the highest values of
$a$, e.g., NN-ReLU dominates all other models for $a \ge 0.98$ on BCW
and Iris, for $a \ge 0.905$ on E-coli, and for $a=1$ for Wine.

Turning now to the \emph{validation results} in the figure, we see an
overall similarity with the training results. This is encouraging as
it indicates that a model selected using training results is likely to
provide non-dominated performance also in validation.  There are,
however, some notable exceptions.  For instance, in the validation
results, NN-ReLU is \emph{dominated} by LR models and, to a lesser
extend, by SVMs, on the DNA problem (when it was non-dominated w.r.t.\
training effort), and so it is suboptimal.  Interesting, DNN-ReLU,
which was dominated across all problems in the training results, in
validation, it dominates all other models for $a> 0.99$ on BCW and
$a=0.995$ on Iris.

\subsubsection{Model Selection based on Affordable Effort}
\label{sec:model-select-based-on_E}

\begin{figure*}[p]
  \centering
  \begin{tabular}{c@{}c}
    Non-dominated  Models (Training)  & Non-dominated  Models (Validation)   \\
    \includegraphics[clip,trim=0 5 0 5,width=.51\linewidth]{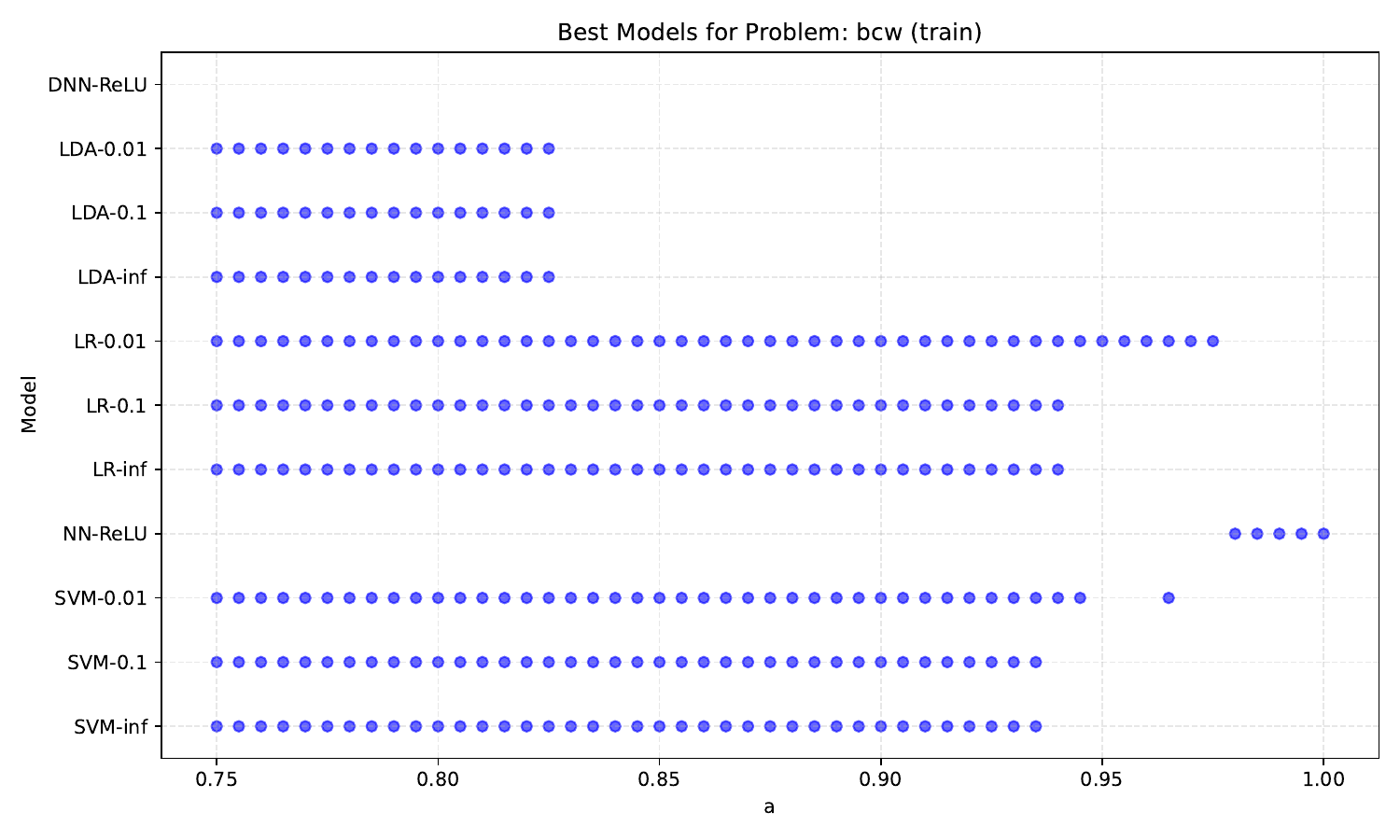} &
    \includegraphics[clip,trim=0 5 0 5,width=.51\linewidth]{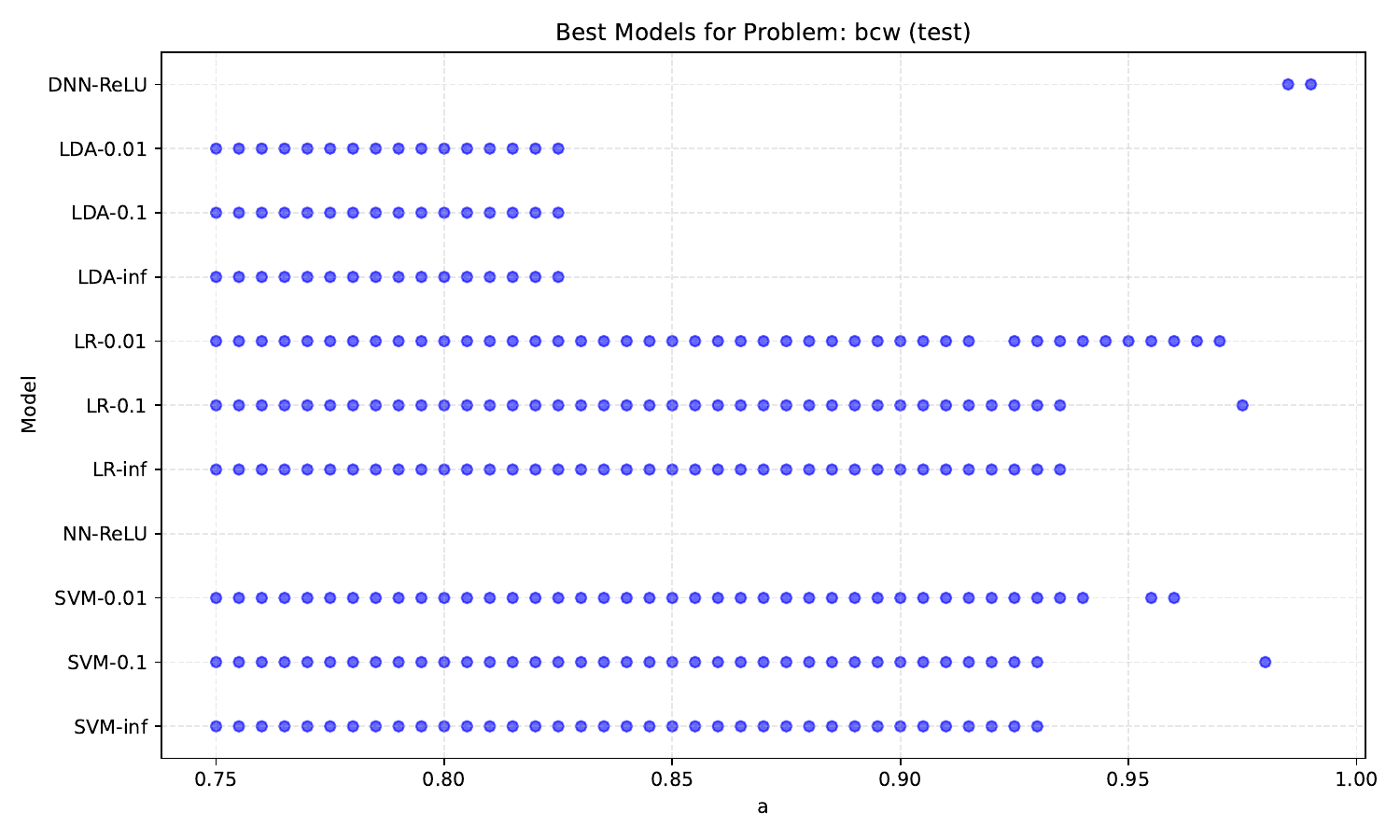} \\[-2mm]
    \includegraphics[clip,trim=0 5 0 5,width=.51\linewidth]{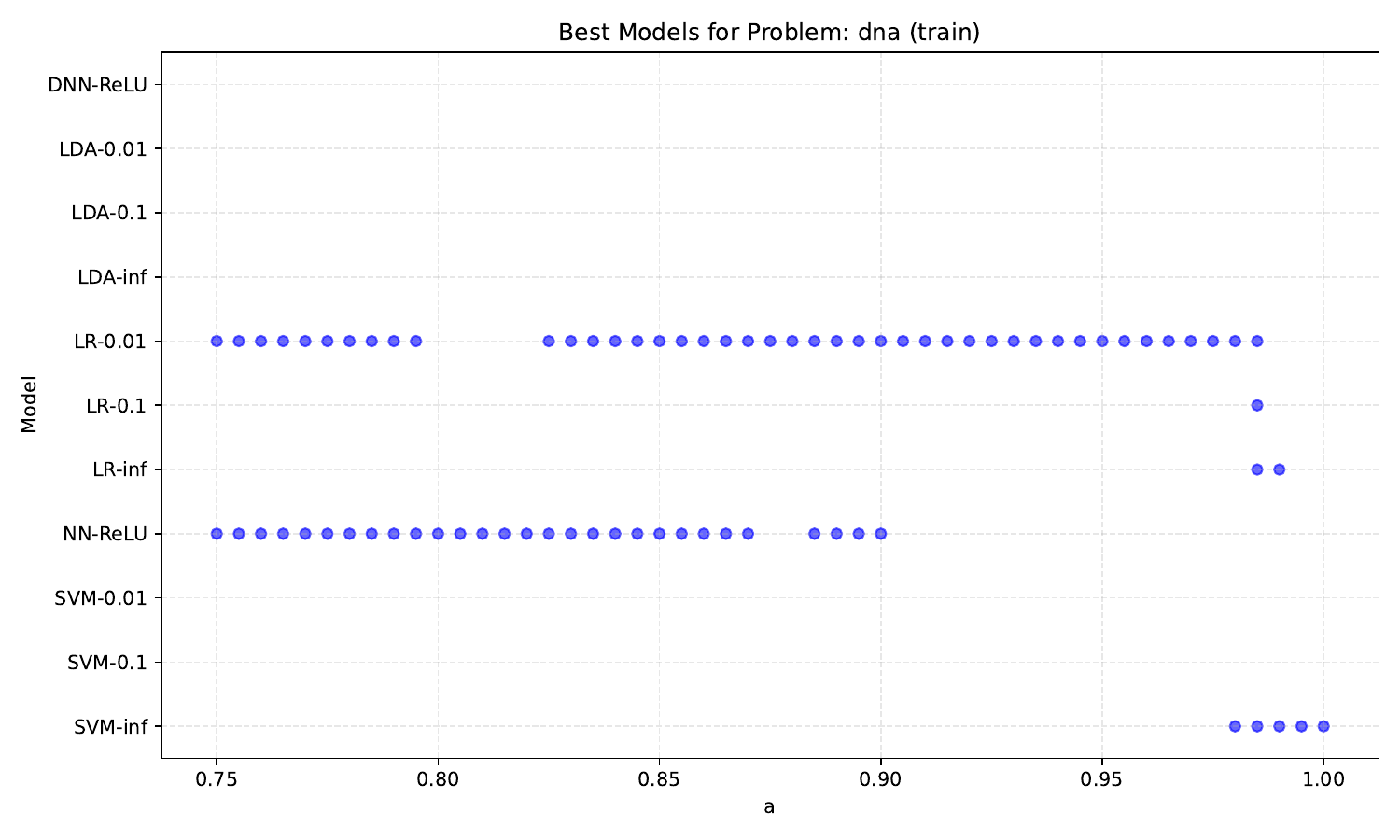} &
    \includegraphics[clip,trim=0 5 0 5,width=.51\linewidth]{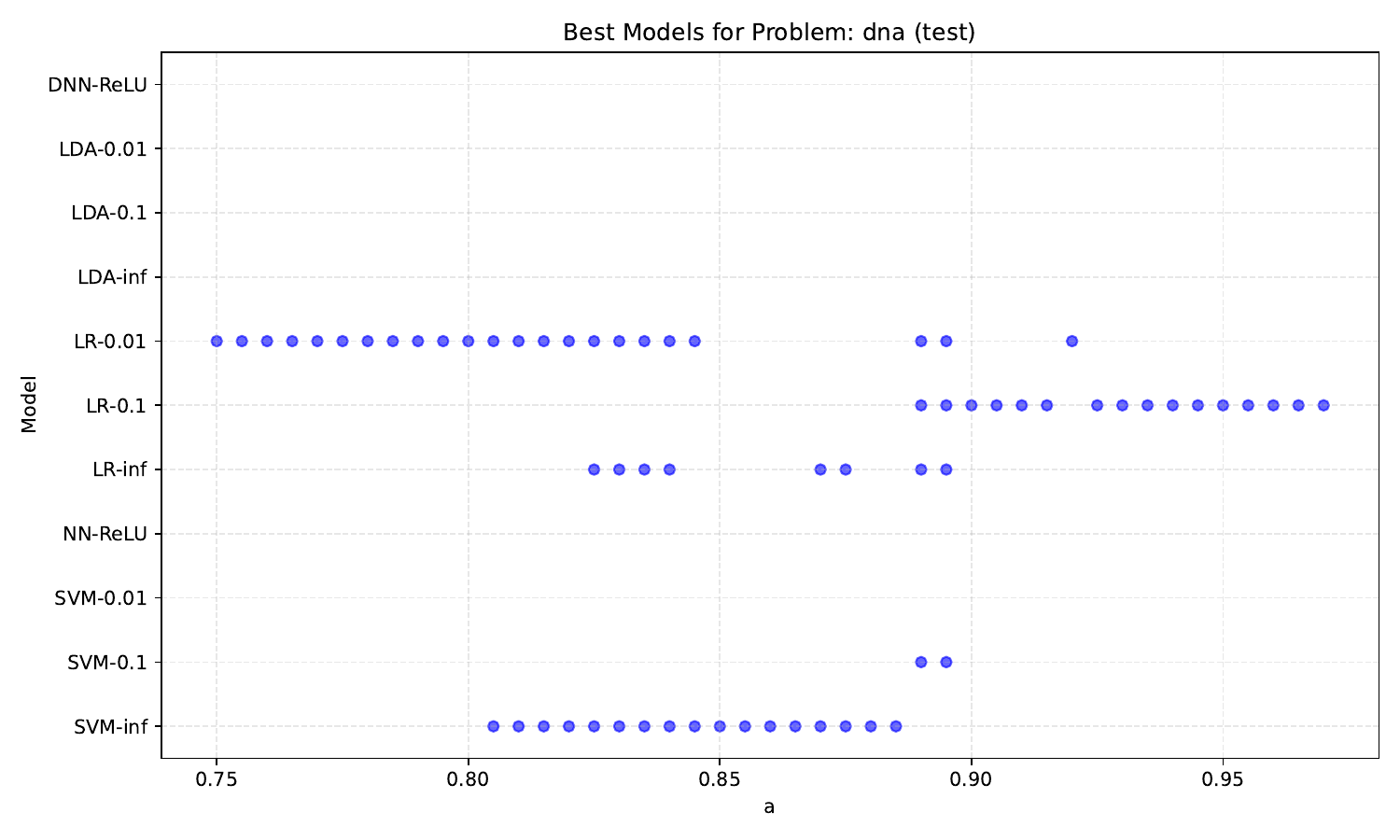} \\[-2mm]
    \includegraphics[clip,trim=0 5 0 5,width=.51\linewidth]{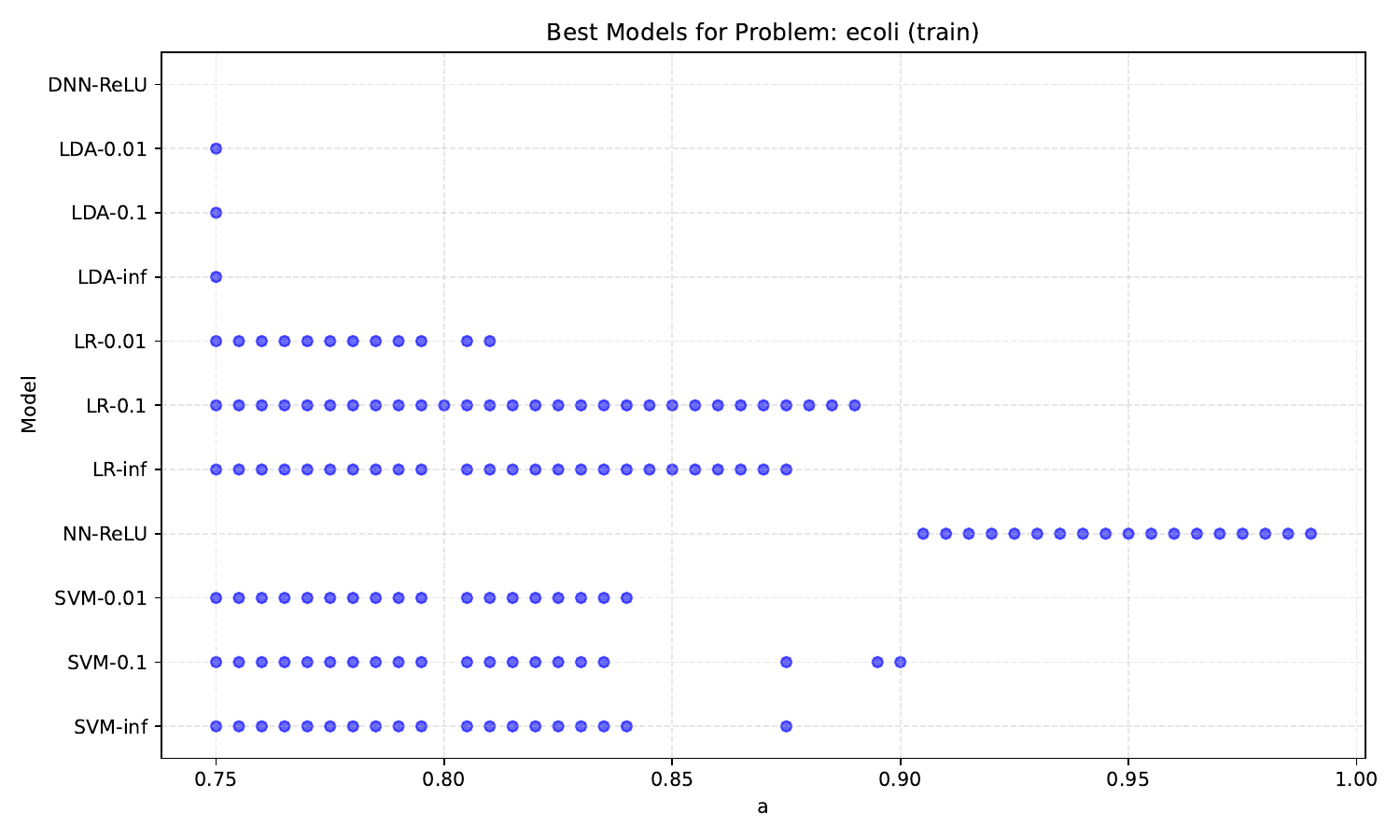} &
    \includegraphics[clip,trim=0 5 0 5,width=.51\linewidth]{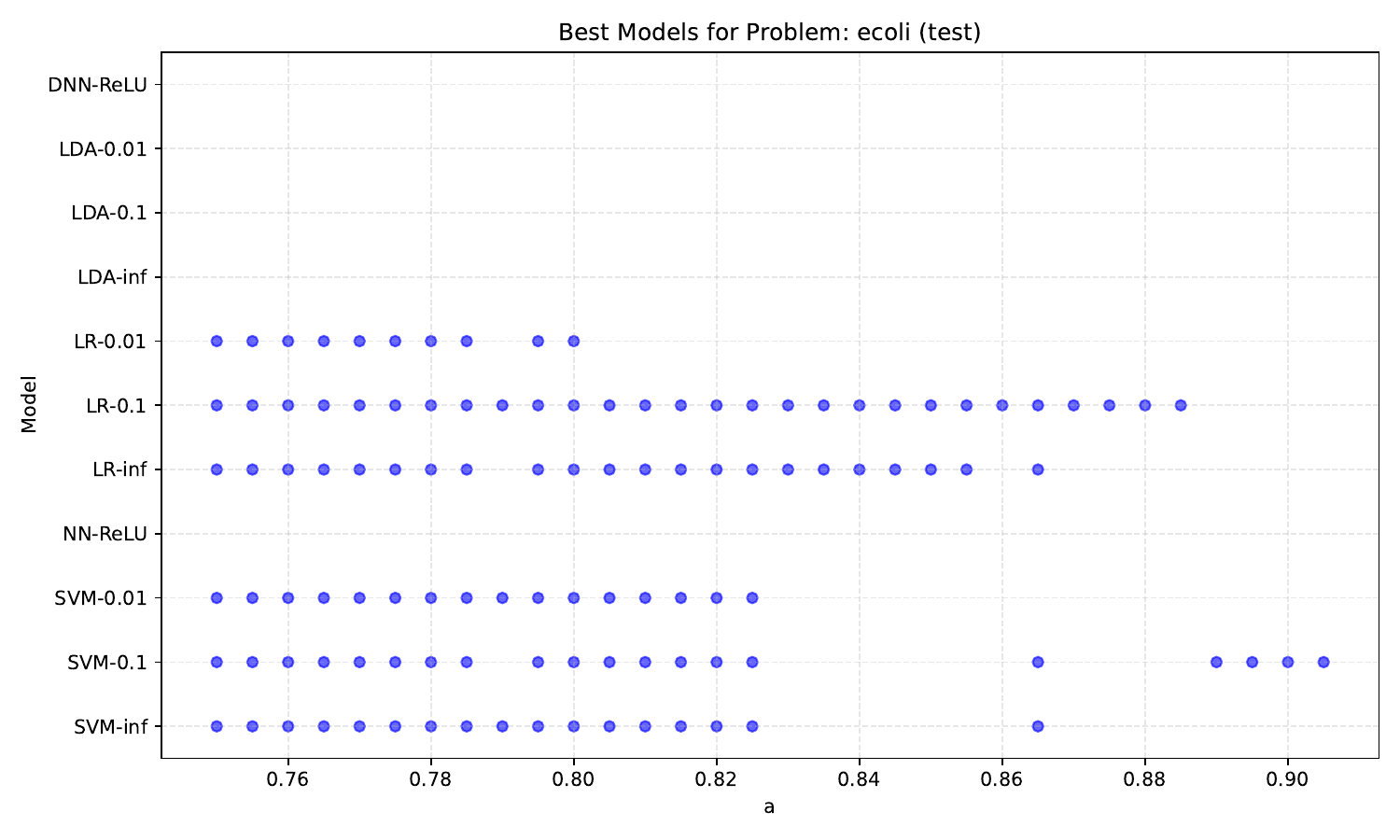} \\[-2mm]
    \includegraphics[clip,trim=0 5 0 5,width=.51\linewidth]{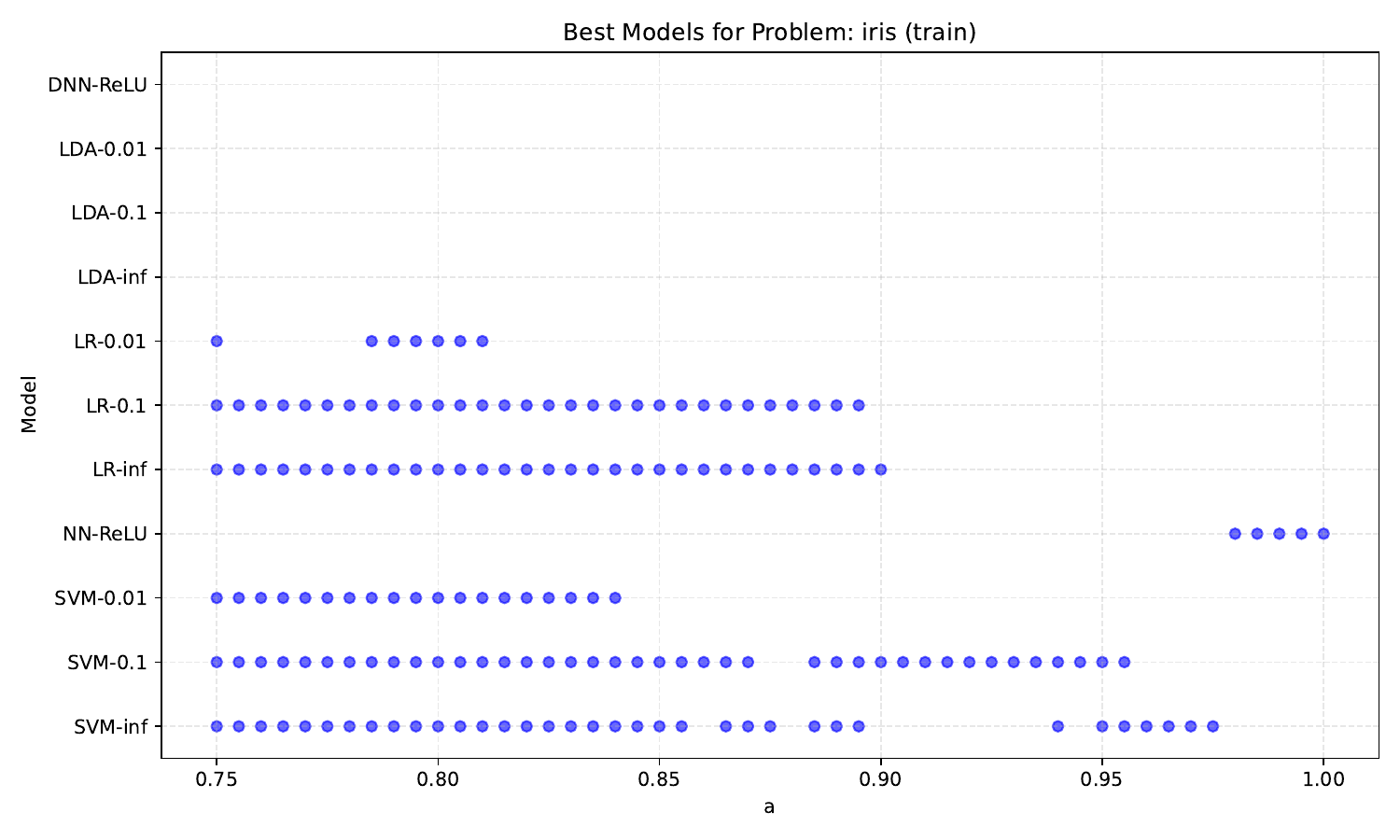} &
    \includegraphics[clip,trim=0 5 0 5,width=.51\linewidth]{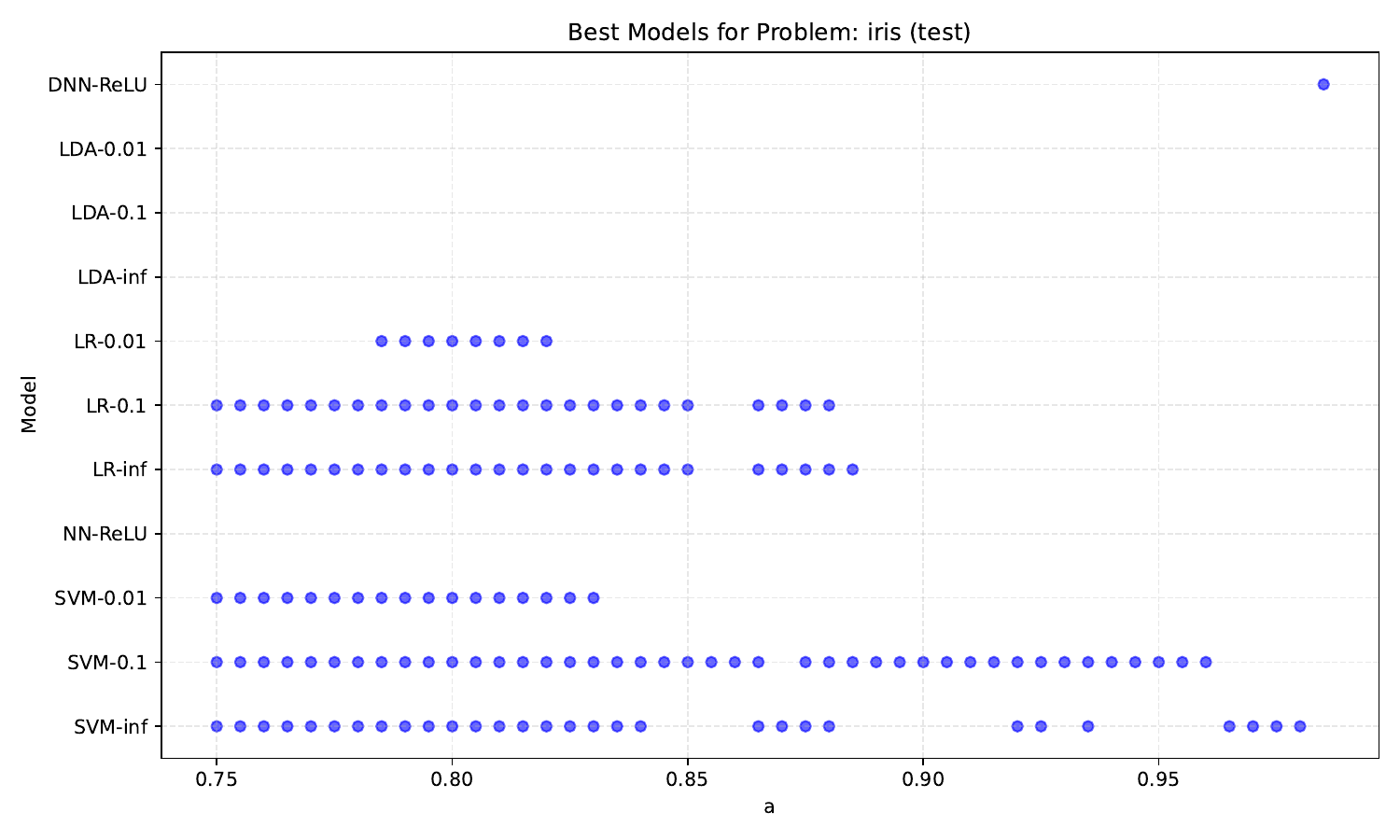} \\[-2mm]
    \includegraphics[clip,trim=0 5 0 5,width=.51\linewidth]{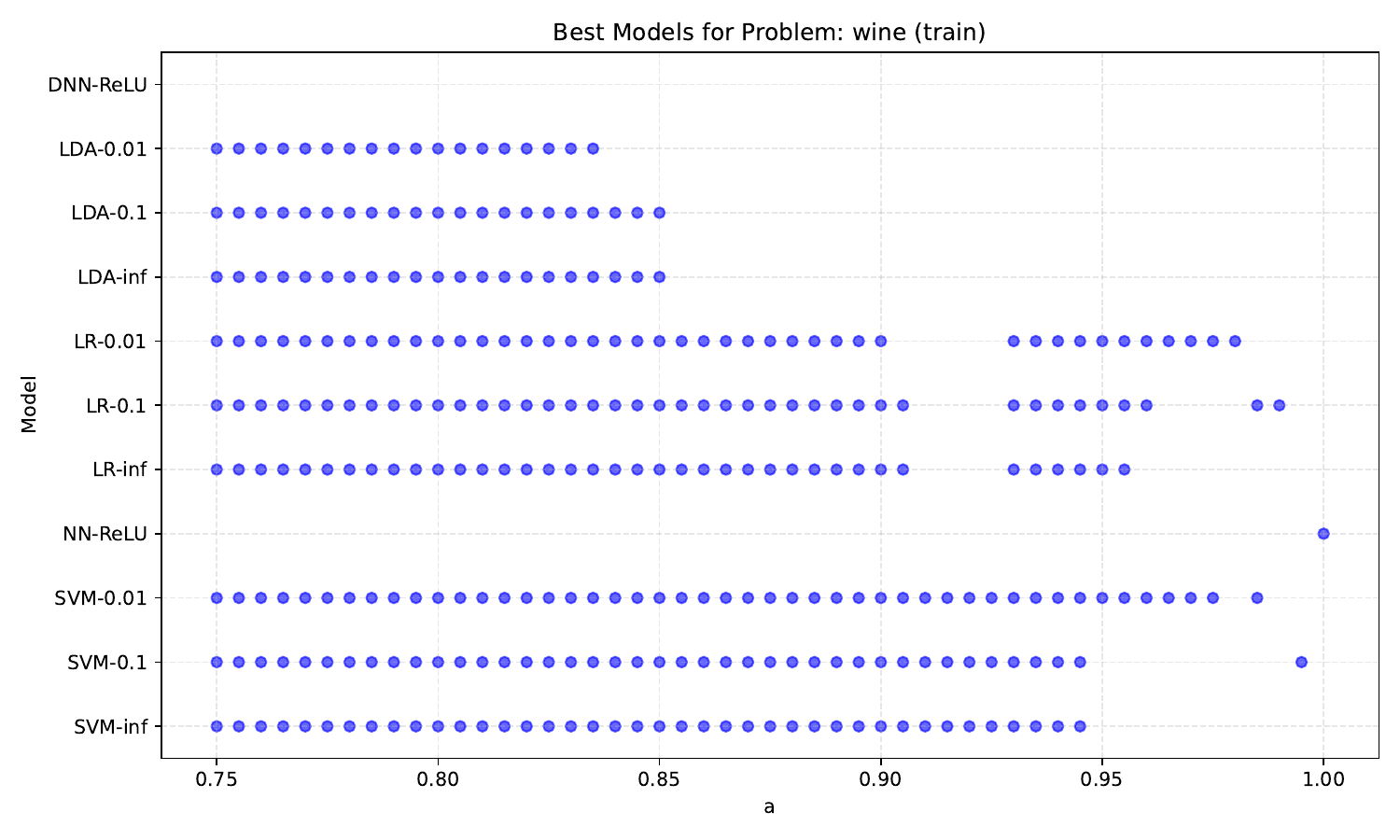} &
    \includegraphics[clip,trim=0 5 0 5,width=.51\linewidth]{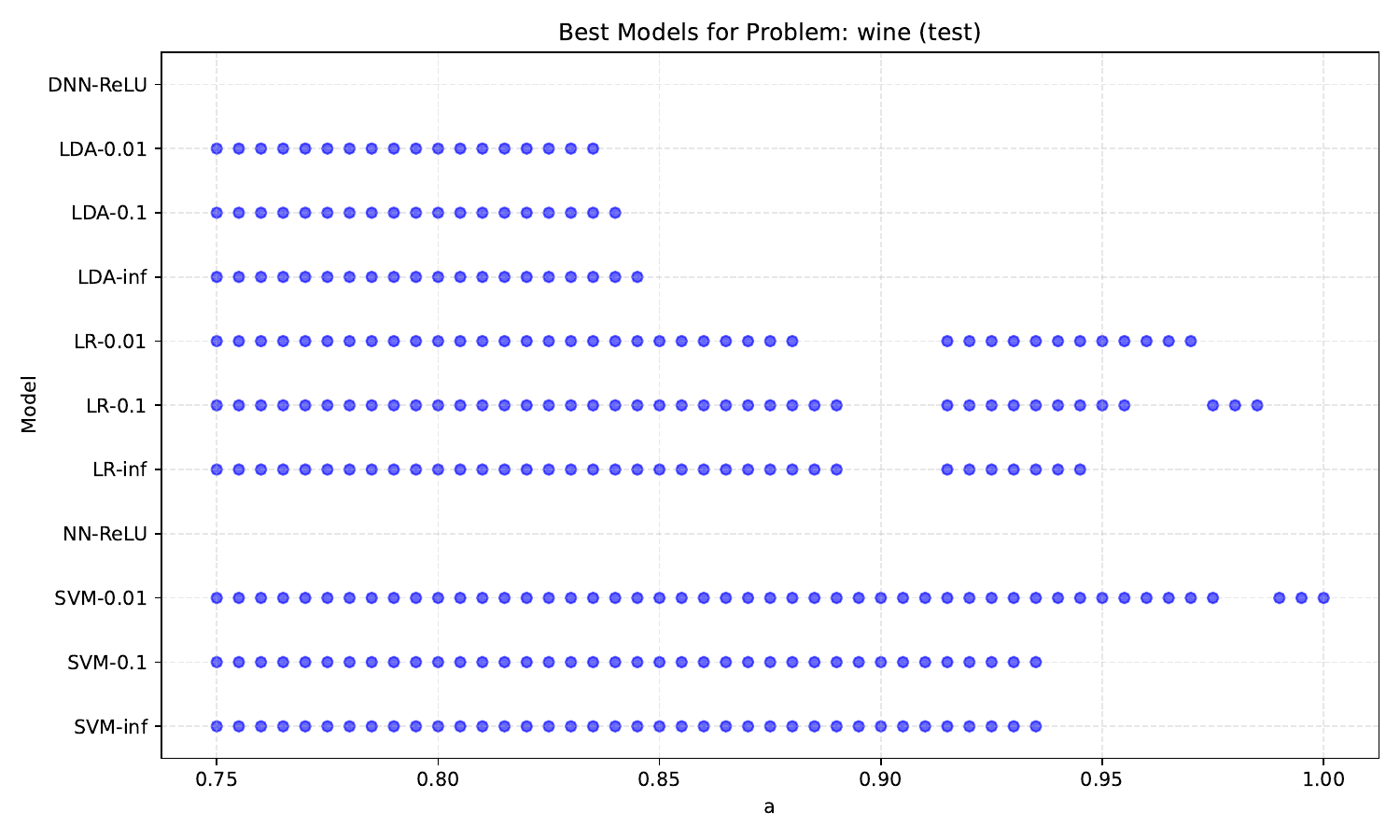} \\[-2mm]
  \end{tabular}
\caption{Models providing the minimum  effort (and thus are non-dominated) for different values of $a$ across problems.}
  \label{fig:non_dominated_models}
\end{figure*}

The models identified as non-dominated in
Figure~\ref{fig:non_dominated_models} for a given value of $a$ are all
equivalent in the sense that they all provide an accuracy of \underline{at
  least} $a$ with the \emph{same effort}. However, some models may
just hit the accuracy $a$, while others may significantly exceed
it. So, this \emph{excess accuracy} could be used as a tie breaker
between non-dominated models. That is, one could select the model with the
highest accuracy out of the ones that achieve at least the target
accuracy $a$ with equal effort.

This observation suggests an alternative way of looking at model
selection: \emph{getting the best bang for the buck}. In other words,
in cases where computational effort \emph{is an issue} but there is
some freedom as to what the acceptable accuracy for a problem can be,
one might want to get the best possible accuracy (bang) for the number
of gradient descent steps that one can be expended on the problem (buck).

\begin{figure*}[p]
  \centering
  \begin{tabular}{c@{}c}
    Non-dominated  Models (Training)  & Non-dominated  Models (Validation)   \\
    \includegraphics[clip,trim=0 5 0 5,width=.51\linewidth]{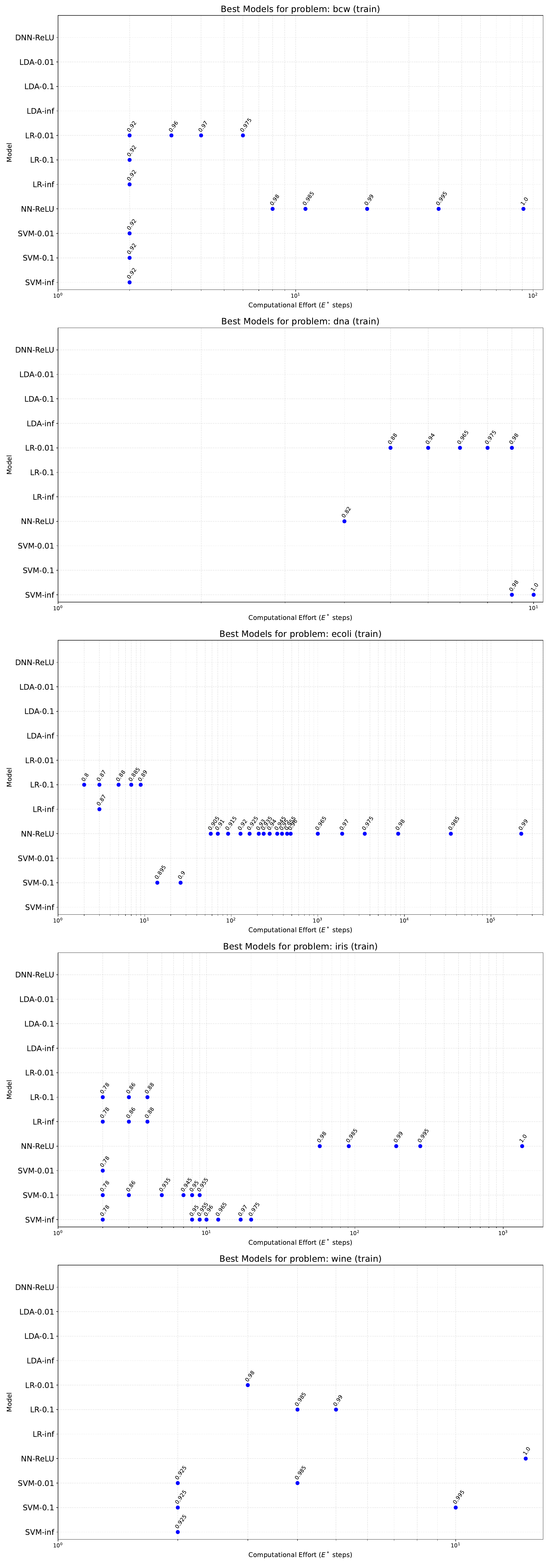} &
    \includegraphics[clip,trim=0 5 0 5,width=.51\linewidth]{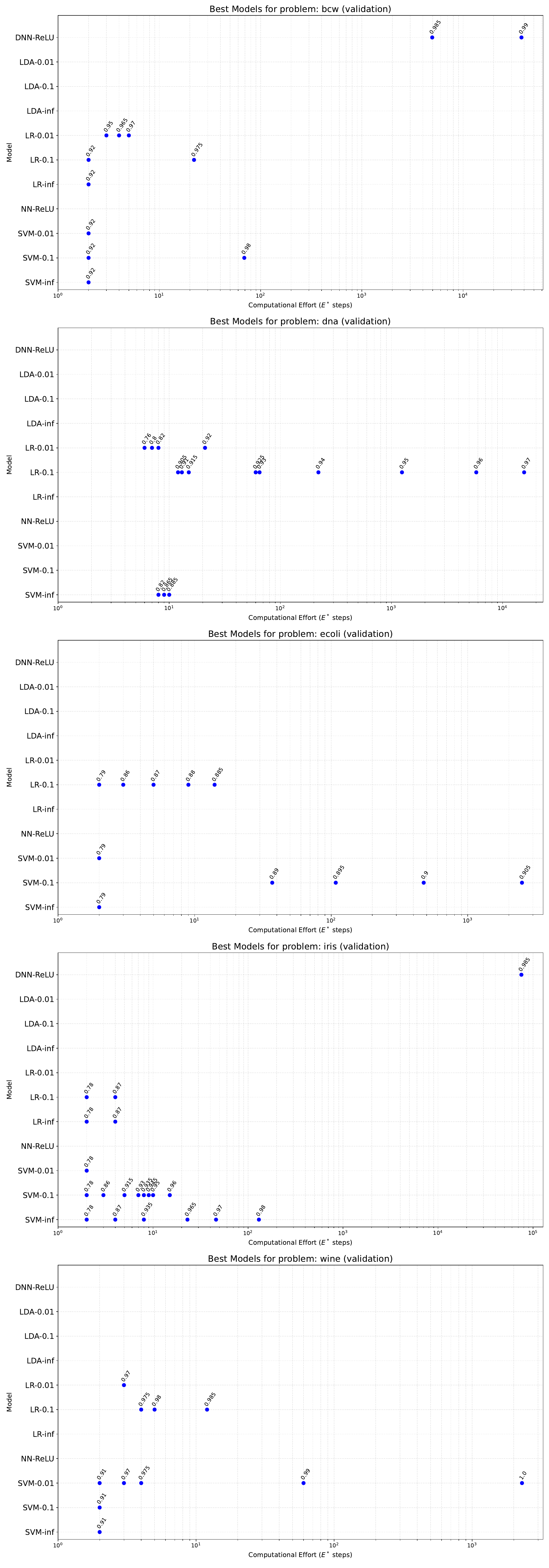} \\
  \end{tabular}
\caption{Models achieving the best training and validation
  accuracy for each budget of gradient descent steps for each problem.}
  \label{fig:non_dominated_models_with_budget}
\end{figure*}

In Figure~\ref{fig:non_dominated_models_with_budget} we report the
same data shown in Figure~\ref{fig:non_dominated_models} but
explicitly showing the computational effort required and \emph{excluding
  the models that \underline{do not} provide the best accuracy for each value of the
  effort}.
The accuracy achieved by each model is reported next to the corresponding data
point.

As expected, this figure makes model selection much easier,
essentially only offering very few models to choose from for any level
of accuracy and making clear how many more gradient-descent steps
would be necessary to get to that level. For instance, for E-coli,
except for a training accuracy of 87\% (where 2 equivalent models are
available), there is only exactly one model one can choose, the choice being
determined by the affordable effort. So, if $10^3$ gradient descent
steps were the limit, one could get 96.5\% accuracy, with every
further 0.05\% requiring approximately a doubling  of the
computational effort.

Interestingly, in the figure there is a clear dominance of LR, SVM and
to some degree NN-ReLU, while the LDA models never   offer the best
accuracy/effort trade off.

\subsection{Relative Problem Difficulty: An Algorithm's Perspective}
\label{sec:relative-difficulty}
  
Thanks to the results provided and discussed in
Section~\ref{sec:model-select-based-on_a}, we are now in a position to
look at the \emph{relative} difficulty of problems.
More specifically, it is arguable that
shapes and positions of the Pareto frontiers shown in
Figure~\ref{fig:model_selection} are first order approximations of the
relative difficulty of the corresponding problems irrespective of the
model used to train them.%
\footnote{Better approximations would require recomputing the
  frontiers using many more of the ML models available in the
  literature.}

For instance, looking at the \emph{training Pareto frontiers}, it is
clear that Wine and DNA are the easiest problems, as one can train a
classifier with just 10 gradient-descent steps for almost all
attainable $a$'s. These are followed by the slightly more difficult
BCW, where still only ${\approx}10^2$ steps are required for all
attainable $a$'s, the harder Iris which requires ${\approx}10^3$
steps. The Ecoli problem is clearly the hardest as it may require
${\approx}10^5$ steps for the highest values of $a$.

If, instead, we consider the \emph{validation Pareto frontiers}, Wine
and E-coli are the easiest problems (${\approx}10^3$ steps), followed
by BCW and DNA (${\approx}10^4$ steps), Iris being clearly the hardest
(${\approx}10^5$ steps) for the highest values of $a$.
Comparing relative problem difficulty across both training and
validation, we see convergence for Wine (easy for both), BCW
(intermediate difficulty for both) and Iris (difficult for
both). However, DNA is easy for training but intermediate/hard for
validation and E-coli is hardest for training but easiest for
validation.

The observed decoupling between training and validation difficulty
yields a crucial insight: problem difficulty is not a monolithic,
intrinsic property of a dataset.
Rather, it is a relative metric---the difficulty of a problem is
strictly defined by how it is perceived by a specific algorithm striving
to reach a predefined acceptable level of accuracy ($a$).
This perceived difficulty comprises two distinct
dimensions: \emph{optimisability} (how easily the algorithm's gradient
descent navigates the training error surface) and
\emph{generalisability} (the degree of alignment between the training
and validation error surfaces for that specific model).
For instance, a dataset like DNA is perceived as highly optimisable by
most models but struggles with generalisability, suggesting the primary
bottleneck is severe overfitting.
Conversely, E-coli presents significant
optimisation challenges during training, yet the sparse solutions
found tend to generalise well.
Recognising that difficulty is algorithm-dependent and accuracy-conditioned
is vital for practitioners. It demonstrates that Pareto frontiers of
computational effort are diagnostic lenses;
they reveal whether a specific algorithm deployed on a specific dataset
requires interventions focused on enhancing optimisation (e.g., learning
rate scheduling) or regularisation to bridge the generalisation gap.
\begin{figure*}[p]
  \centering
  \begin{tabular}{c@{}c}
    Optimal Computational Effort for  & Optimal Computational Effort for   \\
    All Models (Training)  & All Models (Validation)  \\
    \includegraphics[clip,trim=60 20 30 30,width=.52\linewidth]{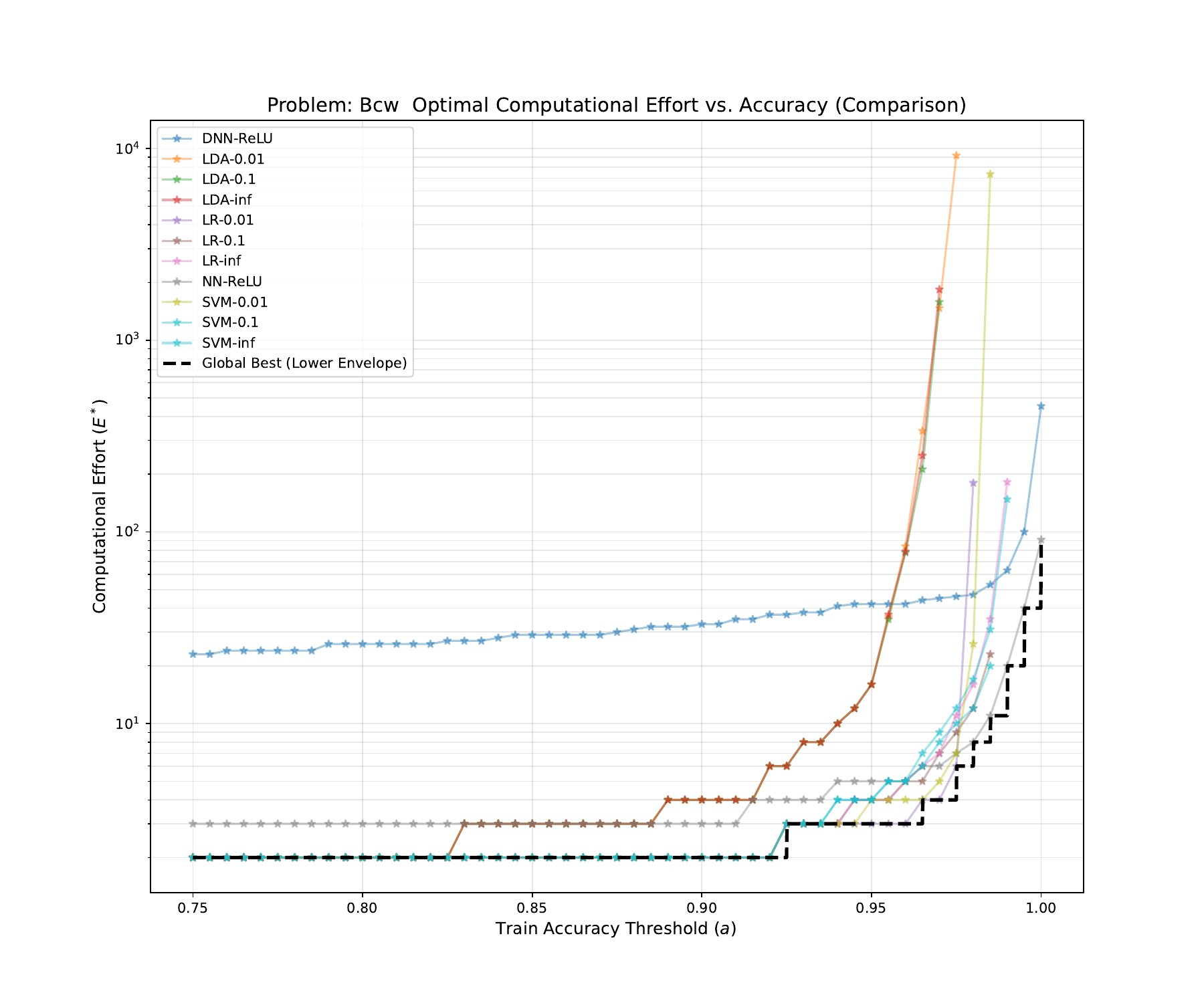} &
    \includegraphics[clip,trim=60 20 30 30,width=.52\linewidth]{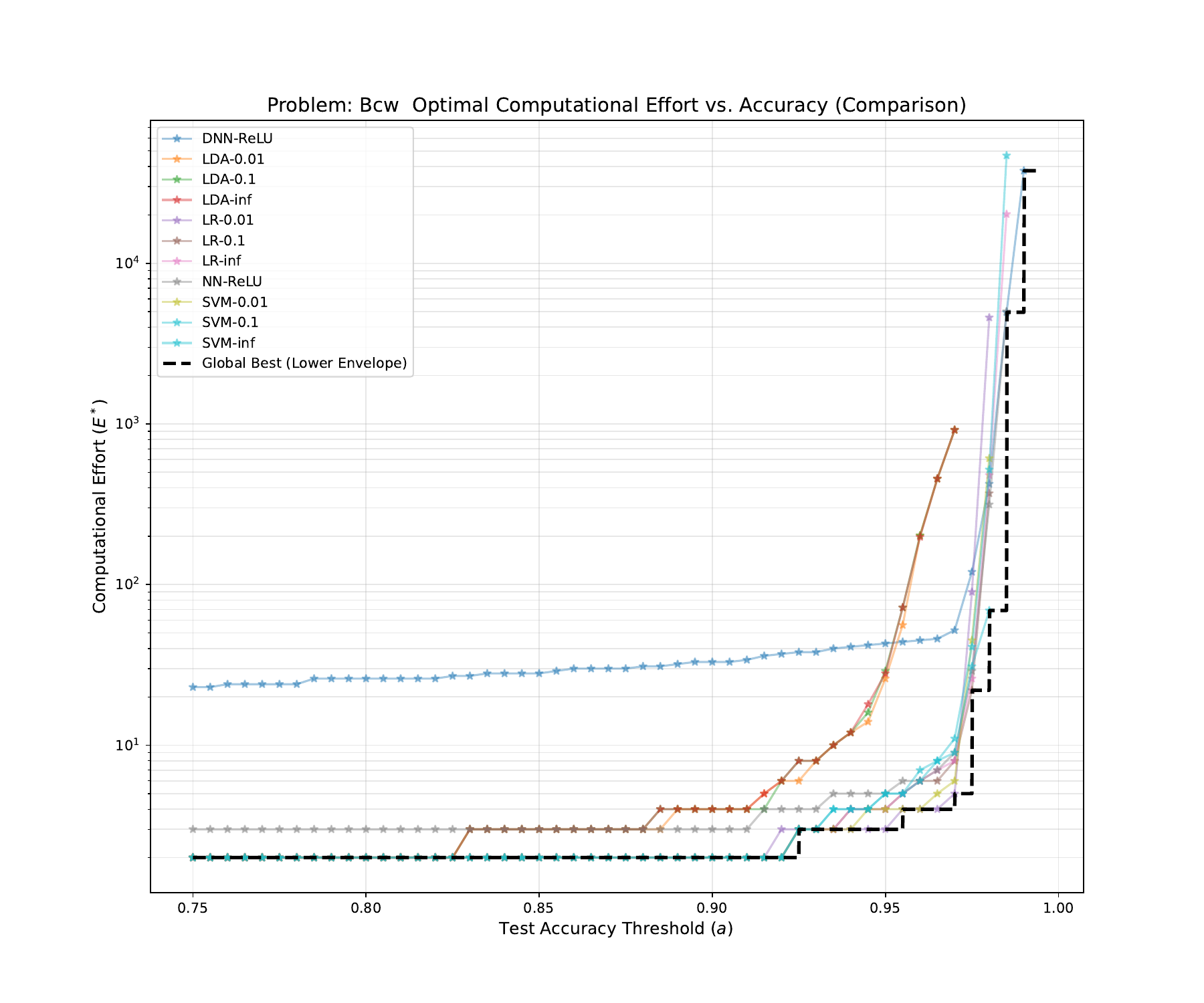} \\[-3mm]
    \includegraphics[clip,trim=60 20 30 30,width=.52\linewidth]{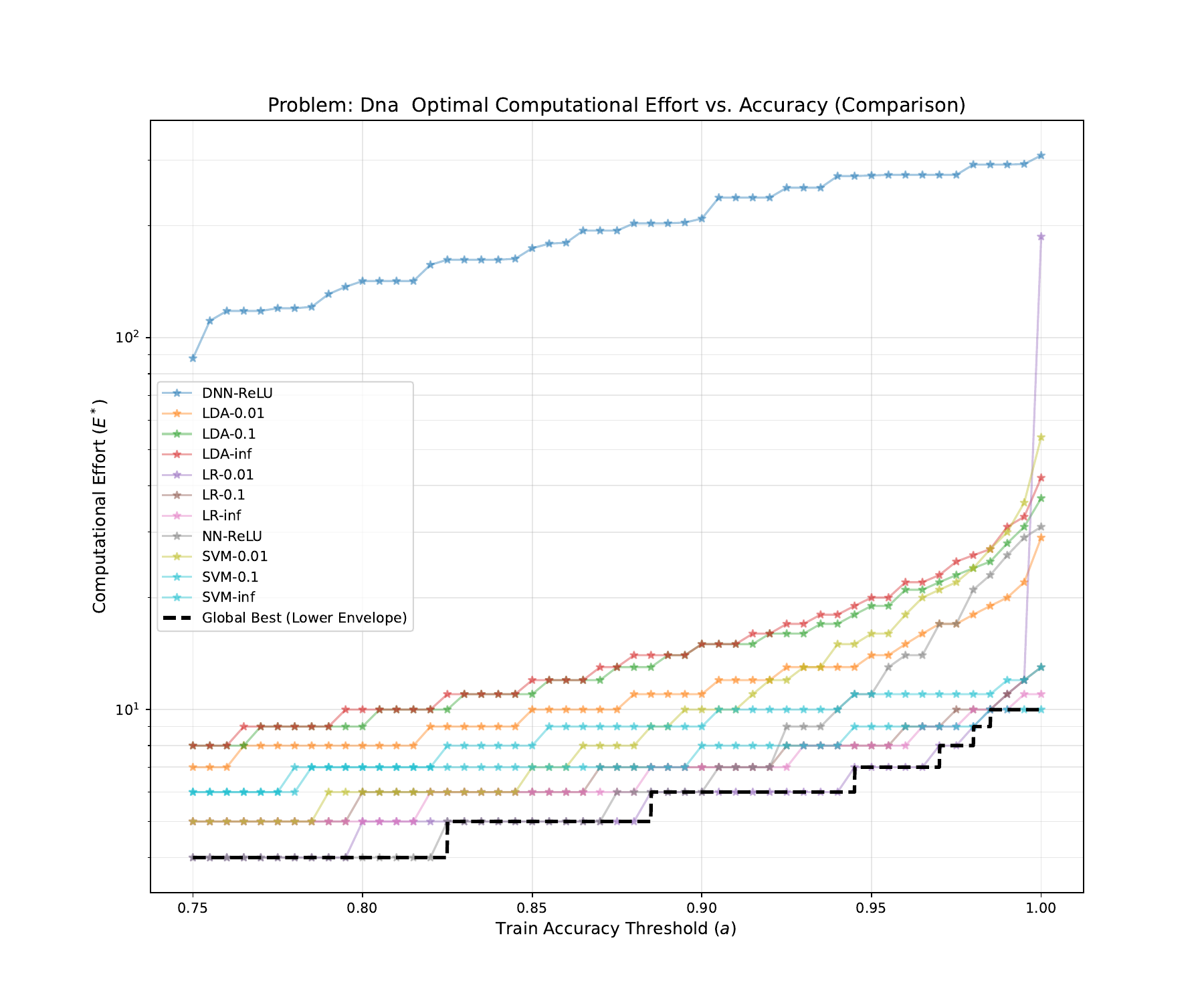} &
    \includegraphics[clip,trim=60 20 30 30,width=.52\linewidth]{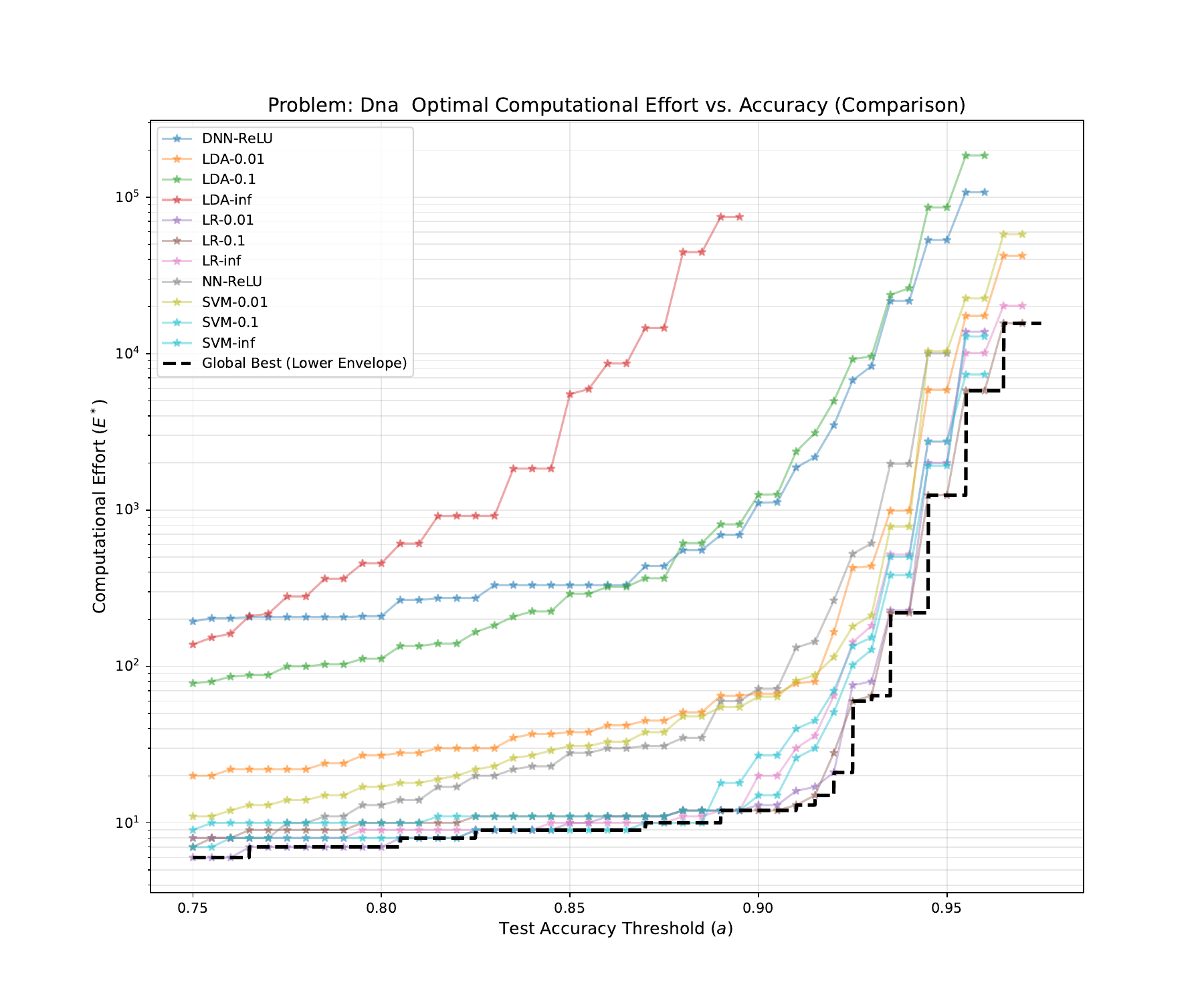} \\[-3mm]
    \includegraphics[clip,trim=60 20 30 30,width=.52\linewidth]{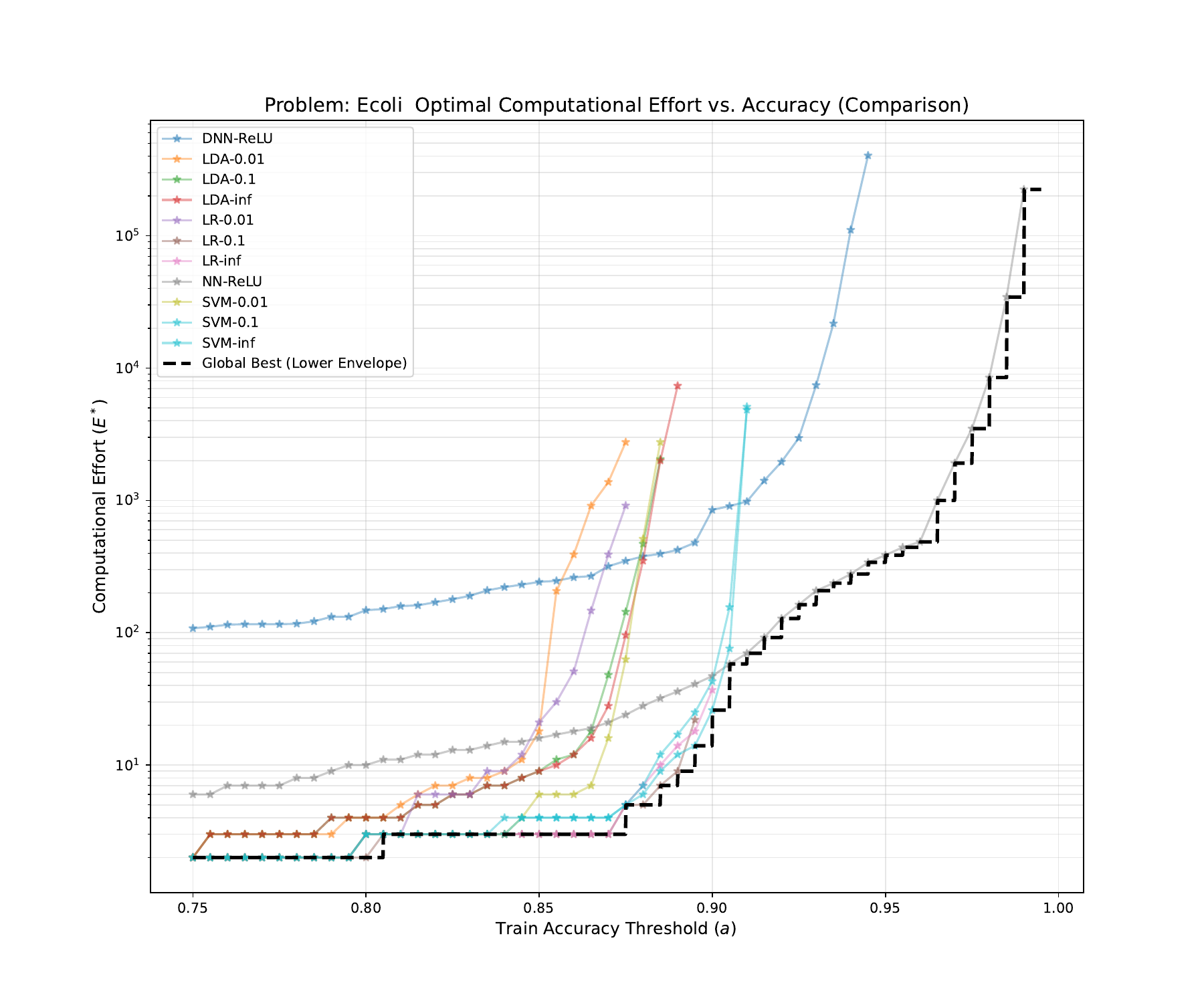} &
    \includegraphics[clip,trim=60 20 30 30,width=.52\linewidth]{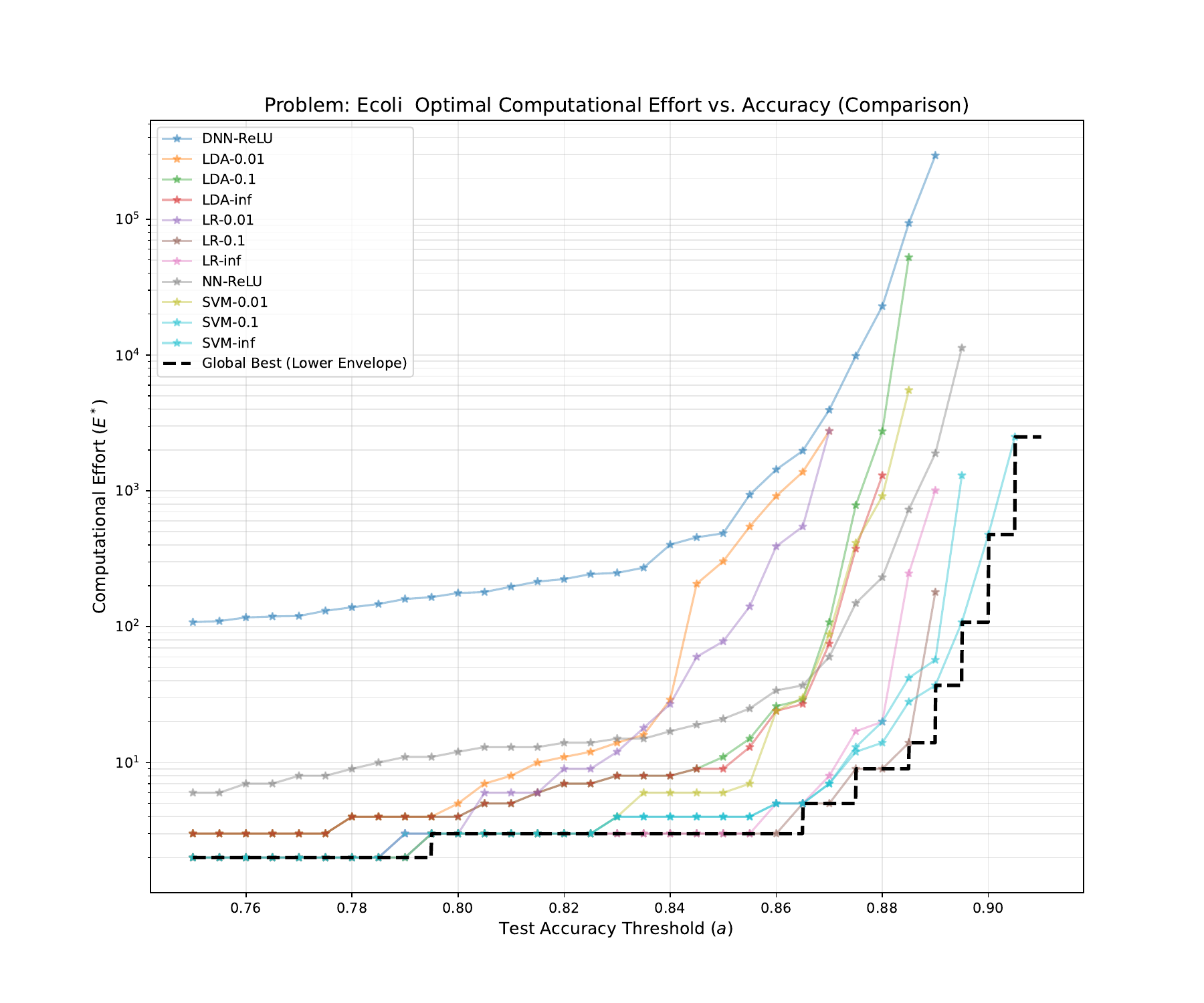} \\[-3mm]
  \end{tabular}
\caption{Optimal computational effort across problems for all 11 models and different values of $a$.}
  \label{fig:model_selection}
\end{figure*}

\begin{figure*}[p]
  \centering
  \ContinuedFloat
  \begin{tabular}{c@{}c}
    Optimal Computational Effort for  & Optimal Computational Effort for   \\
    All Models (Training)  & All Models (Validation)  \\
    \includegraphics[clip,trim=60 20 30 30,width=.52\linewidth]{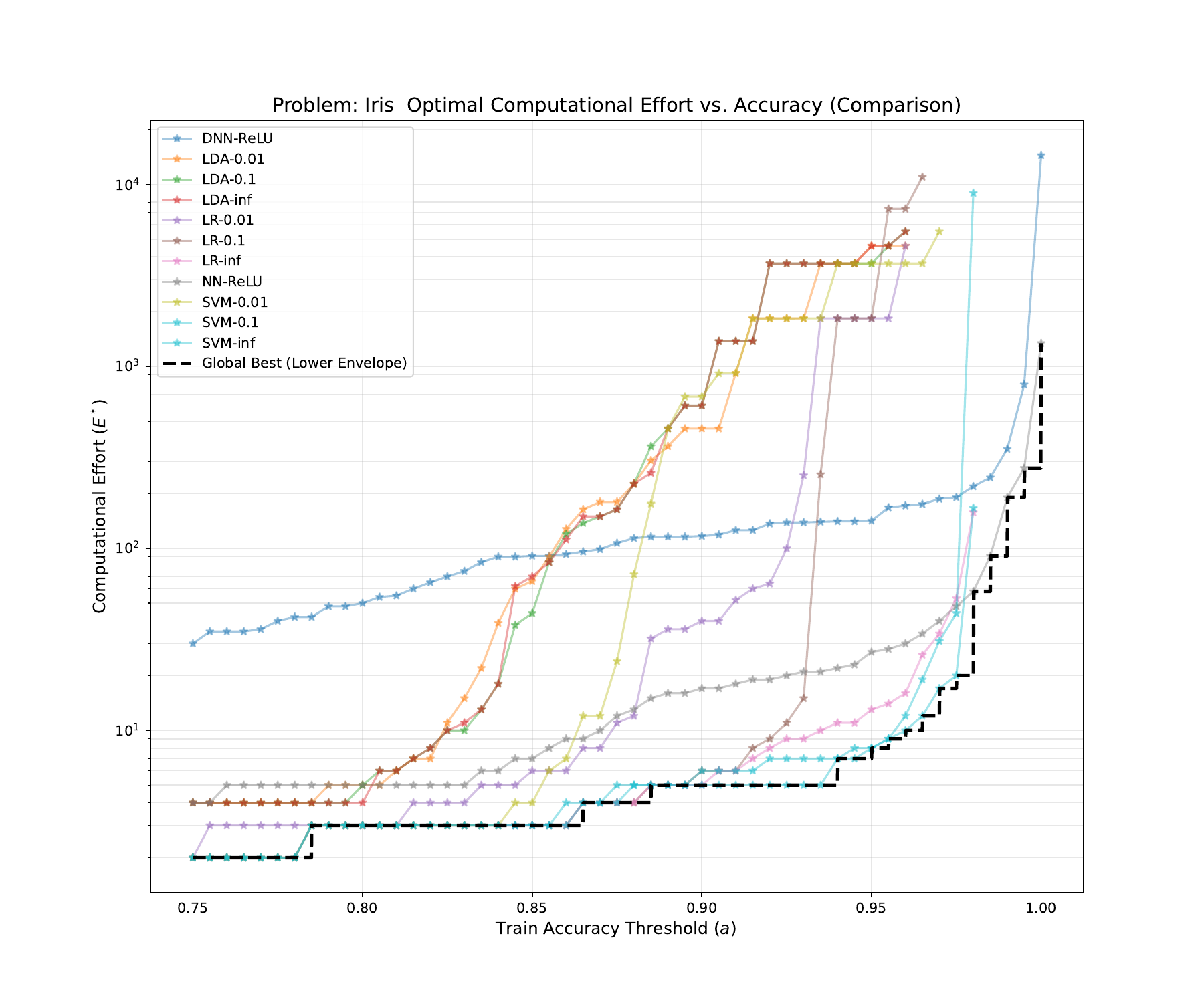} &
    \includegraphics[clip,trim=60 20 30 30,width=.52\linewidth]{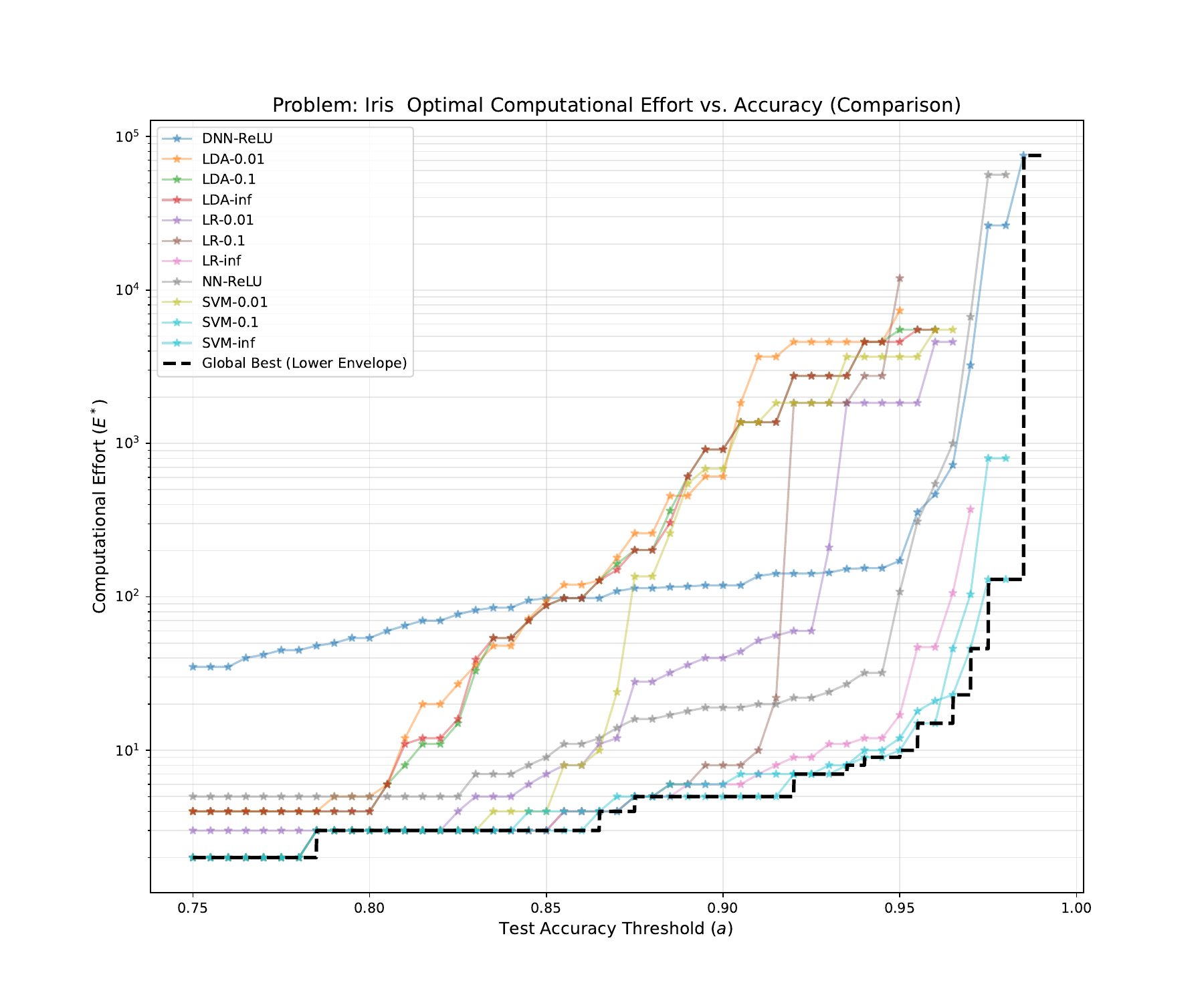} \\[-3mm]
    \includegraphics[clip,trim=60 20 30 30,width=.52\linewidth]{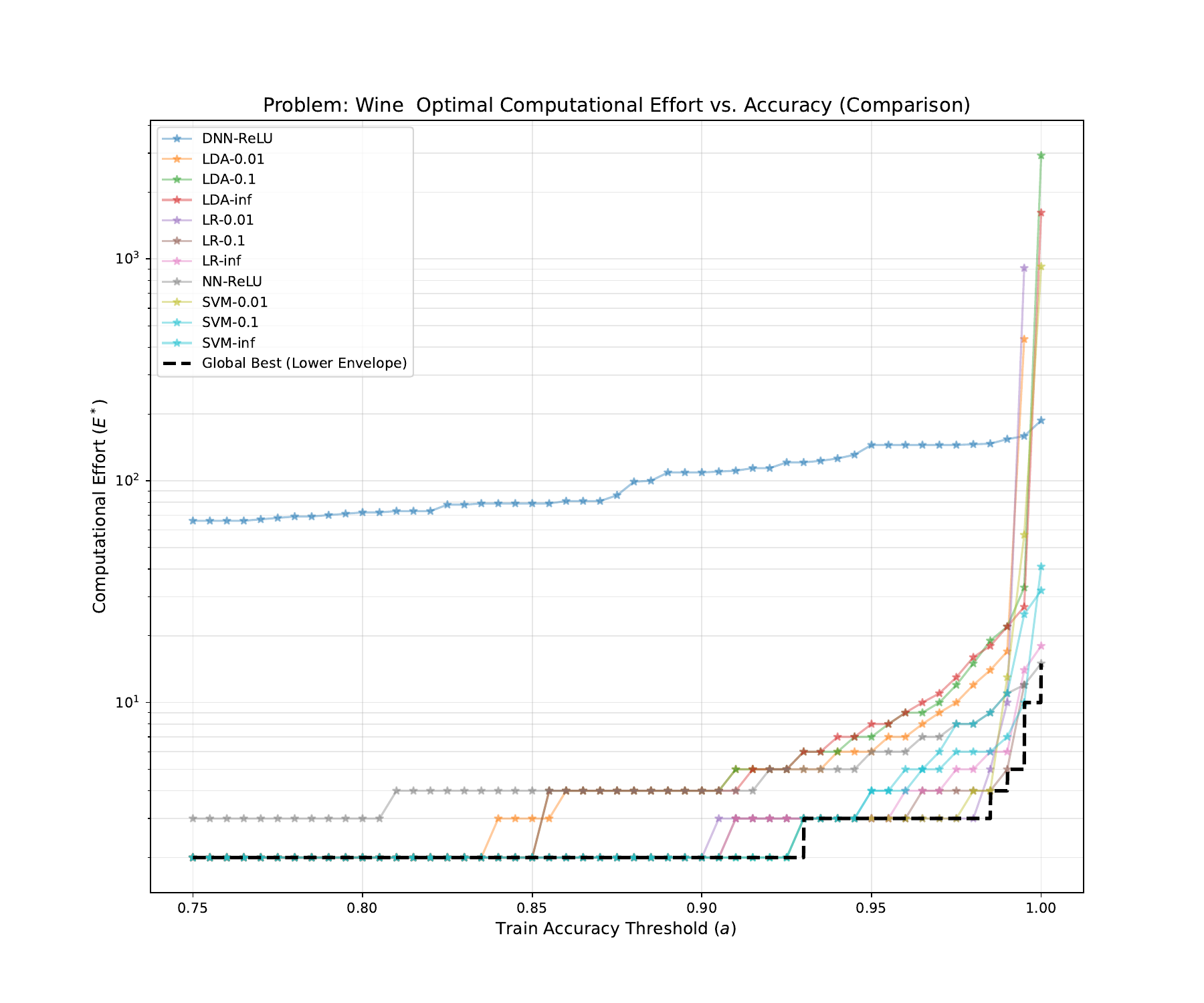} &
    \includegraphics[clip,trim=60 20 30 30,width=.52\linewidth]{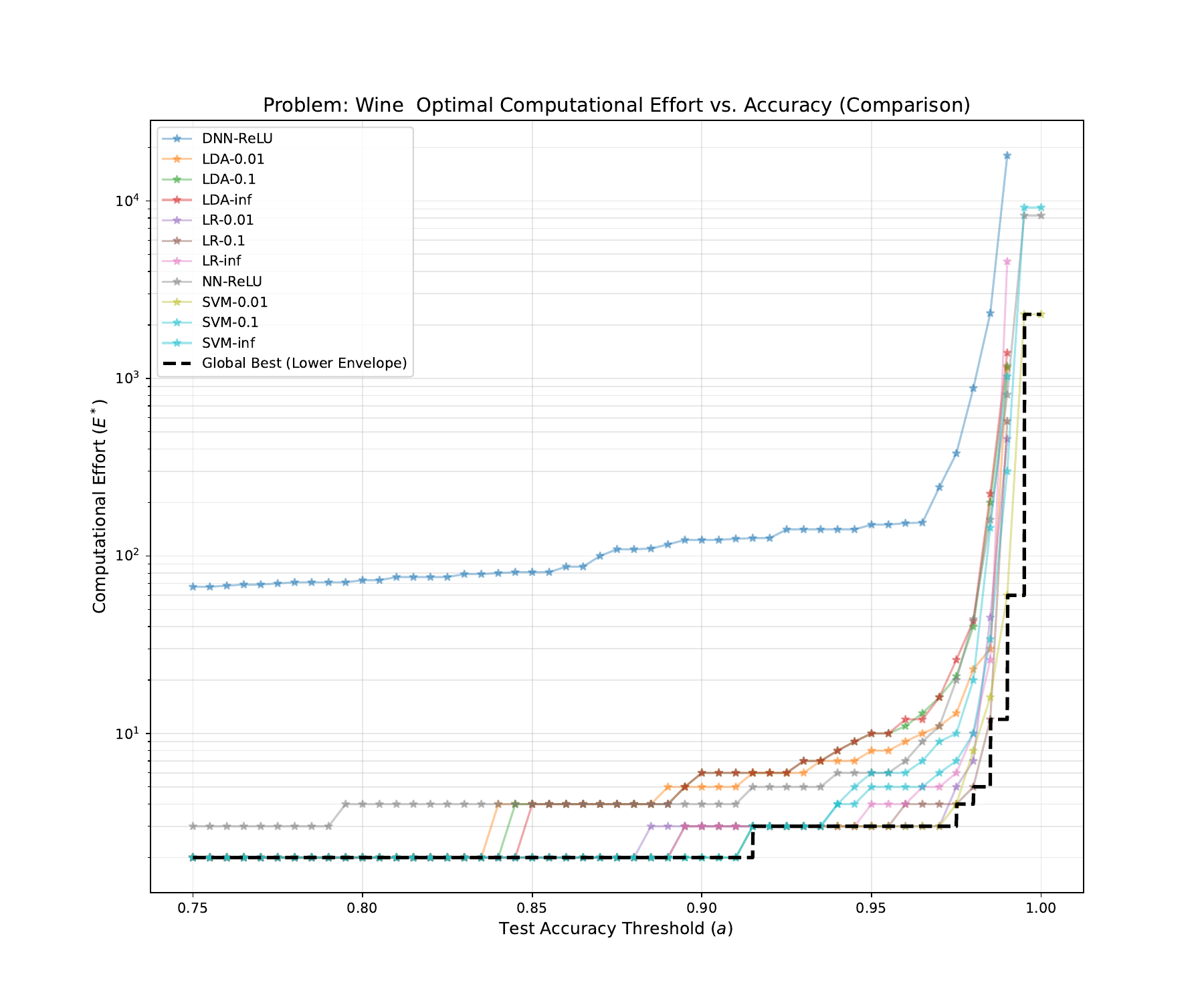} \\[-3mm]
  \end{tabular}
\caption{Optimal computational effort across problems for all 11 models and different values of $a$ (continued).}
\end{figure*}

\subsection{Effect on the Results Induced by Changes in  the Confidence Parameter $z$}

The theoretical implications of changes in the confidence parameter
$z$ were discussed in Section~\ref{sec:sensitivity-analysis-z}.  In
Section~\ref{sec:how-results-are-affected-by-z} of the SM, we provide
a qualitative analysis on how such changes would affect the results and
observations  provided in the previous subsections.

\section{Conclusions, Limitations and Future Work}
\label{sec:conclusions}

\subsection{Conclusions}

In this article, we have challenged the conventional machine learning
paradigm that equates optimal performance with the maximum achievable
average accuracy. Instead, we introduced a novel framework that
evaluates models based on the expected number of gradient-descent
steps required to solve a problem to an acceptable standard with 99\%
probability. By extending Koza's computational effort from the realm
of Genetic Programming to standard gradient-based machine learning, we
developed AutoML on Computational Effort (ACE)---an approach that
explicitly optimises learning rates, epoch limits, and the number of
independent restarts required to guarantee success.

Our extensive experiments across 11 diverse ML models and five
classification datasets yielded several unexpected insights. Most
notably, ACE consistently identified that optimal performance is
achieved using remarkably large learning rates and multiple short
runs. Crucially, as explored through its connection to
superconvergence, this approach demonstrates a powerful dual benefit:
the rapid, aggressive traversal of the error surface not only acts as
a strong regulariser for generalisation, but it is also statistically
optimal for minimising the expected computational effort during
training. Furthermore, our analysis uncovered distinct phase
transitions in the optimal search strategy depending on the target
accuracy. At lower accuracy thresholds, a single run is usually
sufficient; however, as the threshold approaches the intrinsic limits
of the model, the optimal strategy shifts dramatically towards
conducting numerous independent restarts to explore different
trajectories on the error surface. Finally, we demonstrated that
computational effort provides a highly effective metric for model
selection, allowing practitioners to rank models systematically based
on their efficiency in reaching a specified accuracy.

\subsection{Limitations and Future Work}

Despite its significant advantages, the current implementation of ACE
has certain limitations that present promising avenues for future research. 
Firstly, estimating the cumulative success probability currently relies on a 
computationally expensive grid-search methodology requiring hundreds of 
independent runs per hyper-parameter configuration. A primary objective for 
future work will be to replace this exhaustive search with more efficient AutoML 
strategies, such as Bayesian optimisation or evolutionary algorithms, which could 
drastically reduce the computational overhead. Secondly, our study focused 
entirely on full-batch gradient descent. While this provided a clear theoretical 
foundation for exact gradient steps, modern deep learning heavily relies on 
Stochastic Gradient Descent (SGD) and sophisticated adaptive optimisers. Future 
work must explore the integration of ACE with methods like Adam, studying how 
mini-batching alters the dynamics of computational effort and phase transitions. 
Finally, our empirical evaluation was constrained to classic, relatively small-scale 
tabular classification datasets and simple network architectures. It is crucial 
to scale this framework to larger datasets (e.g., CIFAR-10, ImageNet) and 
state-of-the-art architectures, including deep convolutional networks and transformers. 

Presently, our definition of computational effort is strictly formulated 
in terms of gradient descent steps. While this provides an elegant, 
hardware-agnostic measure of optimisability, it inherently abstracts away 
the stark disparities in computational complexity per step. Specifically, 
it does not account for the vast differences in parameter counts between 
architectures (e.g., highly over-parametrised deep neural networks versus 
simple linear models), the varying dimensionality of different problems, 
or the impact of the training set size. Future work must bridge this gap 
by translating the theoretical step-count into hardware-aware metrics, 
quantifying the empirical wall-clock time and the total floating-point 
operations (e.g., GigaFLOPS or PetaFLOPS) required to reach the target 
accuracy $a$ with probability $z$. This translation will provide 
practitioners with a more holistic cost-benefit analysis, allowing them to 
perform model selection based on an affordable budget of GPU time.

In tandem with exploring new architectures, it is essential to broaden
the scope of evaluated models and investigate the impact of different
parameter initialisation strategies. Because computational effort is
inherently tied to the initial position on the error surface, future
studies should quantify how advanced initialisation schemes shift the
effort-based Pareto frontiers. Specifically, we plan to extend this
analysis to deep neural networks employing logistic activation
functions.  To ensure competitive performance, these models will be
initialised utilising the negative-mean weight initialisation method
previously developed by the authors to systematically overcome the
vanishing gradient problem \citep{YilmazPoli2022}.%
\footnote{Our early version of ACE~\citep{yilmaz2021uncovering}
  previously mentioned, was used with \emph{shallow} NNs with logistic
  activation, and so the gradient-vanishing problem was absent.}
Comparing the effort dynamics of these networks against the standard
ReLU models tested here could yield profound theoretical insights into
the complex interactions between activation functions, initialisation,
and gradient descent hitting times.

Furthermore, as introduced in
Section~\ref{sec:stat-reliability-effort}, we must acknowledge the
statistical uncertainty inherent in empirically estimating the
cumulative success probability $P(i, a, \eta)$ from a finite number of
runs. Because our identified phase transitions and cherry-picking
limits often occur precisely at the probability extremes where
relative estimation error spikes \citep{barrero2011empirical,
  barrero2015study, noori2026statistical}, the resulting estimates of
$E^*$ are susceptible to sampling noise, which could perturb the model
selection rankings.

In this article, we relied on a fixed, large-scale grid search to
estimate these probabilities. However, a major objective for future
iterations of the ACE framework will be to replace this static
methodology with the adaptive sampling algorithm recently proposed by
\citet{noori2026statistical}. By dynamically adjusting the number of
runs based on real-time confidence intervals, ACE will be able to
drastically reduce the computational overhead on ``easy''
hyper-parameter configurations while automatically allocating more
runs to boundary conditions, ensuring that model selection remains
both computationally efficient and statistically robust even when
evaluating highly unstable configurations.
\newpage

\bibliography{Bibliography}

\begin{thebibliography}{60}
\providecommand{\natexlab}[1]{#1}
\providecommand{\url}[1]{\texttt{#1}}
\expandafter\ifx\csname urlstyle\endcsname\relax
  \providecommand{\doi}[1]{doi: #1}\else
  \providecommand{\doi}{doi: \begingroup \urlstyle{rm}\Url}\fi

\bibitem[Assun{\c{c}}{\~a}o et~al.(2019)Assun{\c{c}}{\~a}o, Louren{\c{c}}o,
  Machado, and Ribeiro]{assunccao2019denser}
Filipe Assun{\c{c}}{\~a}o, Nuno Louren{\c{c}}o, Penousal Machado, and
  Bernardete Ribeiro.
\newblock Denser: deep evolutionary network structured representation.
\newblock \emph{Genetic Programming and Evolvable Machines}, 20:\penalty0
  5--35, 2019.

\bibitem[Baldock et~al.(2021)Baldock, Maennel, and Neyshabur]{baldock2021deep}
Robert J.~N. Baldock, Hartmut Maennel, and Behnam Neyshabur.
\newblock Deep learning through the lens of example difficulty.
\newblock \emph{arXiv preprint arXiv:2106.09647}, 2021.
\newblock URL \url{https://doi.org/10.48550/arxiv.2106.09647}.

\bibitem[Barrero et~al.(2011)Barrero, R-Moreno, Casta{\~n}o, and
  Camacho]{barrero2011empirical}
David~F Barrero, Mar{\'\i}a~D R-Moreno, Bonifacio Casta{\~n}o, and David
  Camacho.
\newblock An empirical study on the accuracy of computational effort in genetic
  programming.
\newblock In \emph{2011 IEEE Congress on Evolutionary Computation (CEC)}, pages
  1164--1171. IEEE, 2011.

\bibitem[Barrero et~al.(2015)Barrero, Casta{\~n}o, R-Moreno, and
  Camacho]{barrero2015study}
David~F Barrero, Bonifacio Casta{\~n}o, Mar{\'\i}a~D R-Moreno, and David
  Camacho.
\newblock A study on koza's performance measures.
\newblock \emph{Genetic Programming and Evolvable Machines}, 16:\penalty0
  327--353, 2015.

\bibitem[Bethard(2022)]{bethard2022we}
Steven Bethard.
\newblock We need to talk about random seeds.
\newblock \emph{arXiv preprint arXiv:2210.13393}, 2022.

\bibitem[Calle et~al.(2025)Calle, Bates, Reynolds, Liu, Cui, Ly, Wang, Zhang,
  de~Armendi, Shettar, Fung, Tang, and Pan]{calle2025integration}
P.~Calle, A.~Bates, J.~C. Reynolds, Y.~Liu, H.~Cui, S.~Ly, C.~Wang, Q.~Zhang,
  A.~J. de~Armendi, S.~S. Shettar, K.~M. Fung, Q.~Tang, and C.~Pan.
\newblock Integration of nested cross-validation, automated hyperparameter
  optimization, high-performance computing to reduce and quantify the variance
  of test performance estimation of deep learning models.
\newblock \emph{arXiv preprint arXiv:2501.109063}, 2025.
\newblock URL \url{https://doi.org/10.1016/j.cmpb.2025.109063}.

\bibitem[Castle and Johnson(2012)]{castle2012evolving}
Tom Castle and Colin~G Johnson.
\newblock Evolving high-level imperative program trees with strongly formed
  genetic programming.
\newblock In \emph{European Conference on Genetic Programming}, pages 1--12.
  Springer, 2012.

\bibitem[Choromanska et~al.(2015)Choromanska, Henaff, Mathieu, Arous, and
  LeCun]{choromanska2015loss}
Anna Choromanska, Mikael Henaff, Michael Mathieu, G{\'e}rard~Ben Arous, and
  Yann LeCun.
\newblock The loss surfaces of multilayer networks.
\newblock In \emph{Artificial intelligence and statistics}, pages 192--204.
  PMLR, 2015.

\bibitem[Christensen and Oppacher(2002)]{christensen2002analysis}
Steffen Christensen and Franz Oppacher.
\newblock An analysis of koza’s computational effort statistic for genetic
  programming.
\newblock In \emph{Genetic Programming: 5th European Conference, EuroGP 2002
  Kinsale, Ireland, April 3--5, 2002 Proceedings 5}, pages 182--191. Springer,
  2002.

\bibitem[Dauphin et~al.(2014)Dauphin, Pascanu, Gulcehre, Cho, Ganguli, and
  Bengio]{dauphin2014identifying}
Yann~N Dauphin, Razvan Pascanu, Caglar Gulcehre, Kyunghyun Cho, Surya Ganguli,
  and Yoshua Bengio.
\newblock Identifying and attacking the saddle point problem in
  high-dimensional non-convex optimization.
\newblock \emph{Advances in neural information processing systems}, 27, 2014.

\bibitem[Dua et~al.(2017)]{Dua2019}
Graff~Casey Dua, Dheeru et~al.
\newblock {UCI} machine learning repository, 2017.
\newblock URL \url{http://archive.ics.uci.edu/ml}.

\bibitem[Forstenlechner et~al.(2018)Forstenlechner, Fagan, Nicolau, and
  O'Neill]{forstenlechner2018towards}
Stefan Forstenlechner, David Fagan, Miguel Nicolau, and Michael O'Neill.
\newblock Towards understanding and refining the general program synthesis
  benchmark suite with genetic programming.
\newblock In \emph{2018 IEEE Congress on Evolutionary Computation (CEC)}, pages
  1--6. IEEE, 2018.

\bibitem[Hastie et~al.(2009)Hastie, Tibshirani, and
  Friedman]{hastie2009elements}
Trevor Hastie, Robert Tibshirani, and Jerome Friedman.
\newblock \emph{The Elements of Statistical Learning}.
\newblock Springer, 2 edition, 2009.

\bibitem[He et~al.(2021)He, Zhao, and Chu]{he2021automl}
Xin He, Kaiyong Zhao, and Xiaowen Chu.
\newblock Automl: A survey of the state-of-the-art.
\newblock \emph{Knowledge-based systems}, 212:\penalty0 106622, 2021.

\bibitem[Henderson et~al.(2018)Henderson, Islam, Bachman, Pineau, Precup, and
  Meger]{henderson2018deep}
Peter Henderson, Riashat Islam, Philip Bachman, Joelle Pineau, Doina Precup,
  and David Meger.
\newblock Deep reinforcement learning that matters.
\newblock In \emph{Proceedings of the AAAI conference on artificial
  intelligence}, volume~32, 2018.

\bibitem[Ho(2002)]{ho2002complexity}
Tin~Kam Ho.
\newblock Complexity of classification problems and comparative advantage of
  classifiers.
\newblock \emph{Pattern Recognition Letters}, 23\penalty0 (7):\penalty0
  803--810, 2002.

\bibitem[Hutter et~al.(2019)Hutter, Kotthoff, and
  Vanschoren]{hutter2019automated}
Frank Hutter, Lars Kotthoff, and Joaquin Vanschoren.
\newblock \emph{Automated machine learning: methods, systems, challenges}.
\newblock Springer Nature, Cham, 2019.

\bibitem[Kapoor and Narayanan(2022)]{kapoor2022leakage}
Sayash Kapoor and Arvind Narayanan.
\newblock Leakage and the reproducibility crisis in ml-based science.
\newblock \emph{arXiv preprint arXiv:2207.07048}, 2022.

\bibitem[Karl et~al.(2023)Karl, Pielok, Moosbauer, Pfisterer, Coors, Binder,
  Schneider, Thomas, Richter, Lang, Garrido-Merchan, Branke, and
  Bischl]{karl2023multi}
Florian Karl, Tobias Pielok, Julia Moosbauer, Florian Pfisterer, Stefan Coors,
  Martin Binder, Lennart Schneider, Janek Thomas, Jakob Richter, Michel Lang,
  Eduardo~C. Garrido-Merchan, Juergen Branke, and Bernd Bischl.
\newblock Multi-objective hyperparameter optimization in machine learning--an
  overview.
\newblock \emph{ACM Transactions on Evolutionary Learning and Optimization},
  3\penalty0 (4):\penalty0 1--50, 2023.

\bibitem[Karmaker et~al.(2021)Karmaker, Hassan, Smith, Xu, Zhai, and
  Veeramachaneni]{karmaker2021automl}
Shubhra~Kanti Karmaker, Md~Mahadi Hassan, Micah~J Smith, Lei Xu, Chengxiang
  Zhai, and Kalyan Veeramachaneni.
\newblock {AutoML} to date and beyond: Challenges and opportunities.
\newblock \emph{ACM Computing Surveys (CSUR)}, 54\penalty0 (8):\penalty0 1--36,
  2021.

\bibitem[{Keras Team}(2024)]{keras_tuner_docs}
{Keras Team}.
\newblock Keras tuner documentation.
\newblock \url{https://keras.io/keras_tuner/}, 2024.
\newblock Accessed: 2026-04-21.

\bibitem[Kolmogorov(1965)]{kolmogorov1965three}
Andrey~N. Kolmogorov.
\newblock Three approaches to the quantitative definition of information.
\newblock \emph{Problems of Information Transmission}, 1\penalty0 (1):\penalty0
  1--7, 1965.

\bibitem[Koza(1992)]{koza92}
John~R. Koza.
\newblock \emph{Genetic Programming: On the Programming of Computers by Means
  of Natural Selection}.
\newblock MIT Press, Cambridge, MA, 1992.

\bibitem[Koza(1994)]{koza1994genetic}
John~R Koza.
\newblock Genetic programming as a means for programming computers by natural
  selection.
\newblock \emph{Statistics and computing}, 4\penalty0 (2):\penalty0 87--112,
  1994.

\bibitem[Koza et~al.(1999)Koza, Bennett~III, Andre, and Keane]{koza1999gp3}
John~R. Koza, Forrest~H Bennett~III, David Andre, and Martin~A. Keane.
\newblock \emph{Genetic Programming III: Darwinian Invention and Problem
  Solving}.
\newblock Morgan Kaufmann, San Francisco, CA, 1999.
\newblock ISBN 1-55860-543-6.

\bibitem[Koza et~al.(2003)Koza, Keane, Streeter, Mydlowec, Yu, and
  Lanza]{koza2003gp4}
John~R. Koza, Martin~A. Keane, Matthew~J. Streeter, William Mydlowec, Jessen
  Yu, and Guido Lanza.
\newblock \emph{Genetic Programming IV: Routine Human-Competitive Machine
  Intelligence}.
\newblock Kluwer Academic Publishers, Norwell, MA, 2003.
\newblock ISBN 1-4020-7446-8.

\bibitem[Langdon and Poli(2002)]{langdon2002foundations}
William~B Langdon and Riccardo Poli.
\newblock \emph{Foundations of genetic programming}, volume~90.
\newblock Springer, 2002.

\bibitem[Lewkowycz et~al.(2020)Lewkowycz, Bahri, Dyer, Sohl-Dickstein, and
  Gur-Ari]{lewkowycz2020large}
Aitor Lewkowycz, Yasaman Bahri, Ethan Dyer, Jascha Sohl-Dickstein, and Guy
  Gur-Ari.
\newblock The large learning rate phase of deep learning: the catapult
  mechanism.
\newblock \emph{arXiv preprint arXiv:2003.02218}, 2020.

\bibitem[Li and Vitanyi(2008)]{li2008kolmogorov}
Ming Li and Paul Vitanyi.
\newblock \emph{An Introduction to Kolmogorov Complexity and Its Applications}.
\newblock Springer, 3rd edition, 2008.

\bibitem[Li et~al.(2019)Li, Wei, and Ma]{li2019towards}
Yuanzhi Li, Colin Wei, and Tengyu Ma.
\newblock Towards explaining the regularization effect of initial large
  learning rate in training neural networks.
\newblock \emph{Advances in neural information processing systems}, 32, 2019.

\bibitem[Lorena et~al.(2019)Lorena, Garcia, Lehmann, Souto, and
  Ho]{lorena2019complexity}
Ana~Carolina Lorena, Luciano~J. Garcia, Jens Lehmann, Maria Carolina P.~de
  Souto, and Tin~Kam Ho.
\newblock How complex is your classification problem? a survey on measuring
  classification complexity.
\newblock \emph{ACM Computing Surveys}, 52\penalty0 (5):\penalty0 1--34, 2019.

\bibitem[Mendoza et~al.(2016)Mendoza, Klein, Feurer, Springenberg, and
  Hutter]{mendoza2016towards}
Hector Mendoza, Aaron Klein, Matthias Feurer, Jost~Tobias Springenberg, and
  Frank Hutter.
\newblock Towards automatically-tuned neural networks.
\newblock In \emph{Workshop on automatic machine learning}, pages 58--65. PMLR,
  2016.

\bibitem[Miller and Smith(2006)]{miller2006redundancy}
Julian~F Miller and Stephen~L Smith.
\newblock Redundancy and computational efficiency in cartesian genetic
  programming.
\newblock \emph{IEEE Transactions on evolutionary computation}, 10\penalty0
  (2):\penalty0 167--174, 2006.

\bibitem[Miller et~al.(1999)]{miller1999empirical}
Julian~F Miller et~al.
\newblock An empirical study of the efficiency of learning boolean functions
  using a cartesian genetic programming approach.
\newblock In \emph{Proceedings of the genetic and evolutionary computation
  conference}, volume~2, pages 1135--1142, 1999.

\bibitem[Minsky and Papert(1969)]{minsky1969perceptrons}
Marvin Minsky and Seymour Papert.
\newblock \emph{Perceptrons: An Introduction to Computational Geometry}.
\newblock MIT Press, 1969.

\bibitem[Minsky and Papert(1988)]{minsky1988perceptrons}
Marvin Minsky and Seymour Papert.
\newblock \emph{Perceptrons: Expanded Edition}.
\newblock MIT Press, 1988.

\bibitem[Mohtashami et~al.(2023)Mohtashami, Jaggi, and
  Stich]{mohtashami2023special}
Amirkeivan Mohtashami, Martin Jaggi, and Sebastian~U Stich.
\newblock Special properties of gradient descent with large learning rates.
\newblock In \emph{International Conference on Machine Learning}, pages
  25082--25104. PMLR, 2023.

\bibitem[Niehaus and Banzhaf(2003)]{niehaus2003more}
Jens Niehaus and Wolfgang Banzhaf.
\newblock More on computational effort statistics for genetic programming.
\newblock In \emph{European conference on genetic programming}, pages 164--172.
  Springer, 2003.

\bibitem[Noori et~al.(2026)Noori, Valiante, Rozada, Van~Vaerenbergh, and
  Mohseni]{noori2026statistical}
Moslem Noori, Elisabetta Valiante, Ignacio Rozada, Thomas Van~Vaerenbergh, and
  Masoud Mohseni.
\newblock A statistical analysis for per-instance evaluation of stochastic
  optimizers: Avoiding unreliable conclusions.
\newblock \emph{arXiv preprint arXiv:2503.16589v2}, 2026.

\bibitem[Novikoff(1962)]{novikoff1962convergence}
Albert B.~J. Novikoff.
\newblock On convergence proofs on perceptrons.
\newblock \emph{Proceedings of the Symposium on the Mathematical Theory of
  Automata}, pages 615--622, 1962.

\bibitem[O'Malley et~al.(2019)O'Malley, Bursztein, Long, Chollet, Jin,
  Invernizzi, et~al.]{omalley2019kerastuner}
Tom O'Malley, Elie Bursztein, James Long, Fran\c{c}ois Chollet, Haifeng Jin,
  Luca Invernizzi, et~al.
\newblock Kerastuner.
\newblock \url{https://github.com/keras-team/keras-tuner}, 2019.

\bibitem[Oymak(2021)]{oymak2021provable}
Samet Oymak.
\newblock Provable super-convergence with a large cyclical learning rate.
\newblock \emph{IEEE Signal Processing Letters}, 28:\penalty0 1645--1649, 2021.

\bibitem[Pedregosa et~al.(2011)Pedregosa, Varoquaux, Gramfort, Michel, Thirion,
  Grisel, Blondel, Prettenhofer, Weiss, Dubourg, et~al.]{scikit-learn}
Fabian Pedregosa, Ga{\"e}l Varoquaux, Alexandre Gramfort, Vincent Michel,
  Bertrand Thirion, Olivier Grisel, Mathieu Blondel, Peter Prettenhofer, Ron
  Weiss, Vincent Dubourg, et~al.
\newblock Scikit-learn: Machine learning in {P}ython.
\newblock \emph{Journal of Machine Learning Research}, 12:\penalty0 2825--2830,
  2011.

\bibitem[Poli(1996{\natexlab{a}})]{poli1996parallel}
Riccardo Poli.
\newblock \emph{Parallel distributed genetic programming}.
\newblock University of Birmingham, Cognitive Science Research Centre
  Birmingham, UK, 1996{\natexlab{a}}.

\bibitem[Poli(1996{\natexlab{b}})]{poli1996some}
Riccardo Poli.
\newblock Some steps towards a form of parallel distributed genetic
  programming.
\newblock In \emph{Proceedings of the first on-line workshop on soft
  computing}, pages 290--295, 1996{\natexlab{b}}.

\bibitem[Poli(1997)]{poli1997evolution}
Riccardo Poli.
\newblock Evolution of graph-like programs with parallel distributed genetic
  programming.
\newblock In \emph{ICGA}, pages 346--353, 1997.

\bibitem[Poli et~al.(2008)Poli, Langdon, and McPhee]{Poli2008fieldguide}
Riccardo Poli, William~B Langdon, and Nicholas~Freitag McPhee.
\newblock \emph{Field guide to genetic programming}.
\newblock Lulu Press, www.lulu.com, 2008.

\bibitem[Pratama et~al.(2024)Pratama, Rinanda, and
  Satria]{pratama2024comparative}
A.~Pratama, F.~Rinanda, and R.~Satria.
\newblock Comparative analysis of convolutional neural network and densenet121
  transfer learning in agriculture focusing on crop leaf disease
  identification.
\newblock \emph{Applied Computing and Informatics}, 20\penalty0 (2):\penalty0
  115--128, 2024.
\newblock \doi{10.1108/ACI-03-2024-0132}.

\bibitem[Rinanda et~al.(2026)Rinanda, Mohammad, and Ali]{rinanda2025deep}
F.~Rinanda, M.~Mohammad, and S.~Ali.
\newblock Intelligent systems in neuroimaging: Pioneering ai techniques for
  brain tumor detection.
\newblock \emph{arXiv preprint arXiv:2601.17655}, 2026.
\newblock URL \url{https://arxiv.org/abs/2601.17655}.

\bibitem[Rosenfeld and Vanneschi(2025)]{rosenfeld2025survey}
Liah Rosenfeld and Leonardo Vanneschi.
\newblock A survey on batch training in genetic programming.
\newblock \emph{Genetic Programming and Evolvable Machines}, 26\penalty0
  (1):\penalty0 2, 2025.

\bibitem[Ruder(2016)]{ruder2016overview}
Sebastian Ruder.
\newblock An overview of gradient descent optimization algorithms.
\newblock \emph{arXiv preprint arXiv:1609.04747}, 2016.

\bibitem[Smith(2017)]{smith2017cyclical}
Leslie~N Smith.
\newblock Cyclical learning rates for training neural networks.
\newblock In \emph{2017 IEEE winter conference on applications of computer
  vision (WACV)}, pages 464--472. IEEE, 2017.

\bibitem[Smith(2018)]{smith2018disciplined}
Leslie~N Smith.
\newblock A disciplined approach to neural network hyper-parameters: Part
  1--learning rate, batch size, momentum, and weight decay.
\newblock \emph{arXiv preprint arXiv:1803.09820}, 2018.

\bibitem[Smith and Topin(2019)]{smith2019super}
Leslie~N Smith and Nicholay Topin.
\newblock Super-convergence: Very fast training of neural networks using large
  learning rates.
\newblock In \emph{Artificial intelligence and machine learning for
  multi-domain operations applications}, volume 11006, pages 369--386. SPIE,
  2019.

\bibitem[Smith et~al.(2021)Smith, Dherin, Barrett, and De]{smith2021origin}
Samuel~L Smith, Benoit Dherin, David~GT Barrett, and Soham De.
\newblock On the origin of implicit regularization in stochastic gradient
  descent.
\newblock \emph{arXiv preprint arXiv:2101.12176}, 2021.

\bibitem[Sutskever et~al.(2013)Sutskever, Martens, Dahl, and
  Hinton]{sutskever2013importance}
Ilya Sutskever, James Martens, George Dahl, and Geoffrey Hinton.
\newblock On the importance of initialization and momentum in deep learning.
\newblock In \emph{International conference on machine learning}, pages
  1139--1147. PMLR, 2013.

\bibitem[Vapnik(1998)]{vapnik1998statistical}
Vladimir~N. Vapnik.
\newblock \emph{Statistical Learning Theory}.
\newblock Wiley, 1998.

\bibitem[Yang and Shami(2020)]{yang2020hyperparameter}
Li~Yang and Abdallah Shami.
\newblock On hyperparameter optimization of machine learning algorithms: Theory
  and practice.
\newblock \emph{Neurocomputing}, 415:\penalty0 295--316, 2020.

\bibitem[Yilmaz(2021)]{yilmaz2021uncovering}
Ahmet Yilmaz.
\newblock \emph{Uncovering Efficient Learning and Initialisation Algorithms for
  Neural Networks Using Evolutionary Algorithms and Theoretical Analyses}.
\newblock PhD thesis, University of Essex, 2021.
\newblock Supervisor: Prof Riccardo Poli.

\bibitem[Yilmaz and Poli(2022)]{YilmazPoli2022}
Ahmet Yilmaz and Riccardo Poli.
\newblock Successfully and efficiently training deep multi-layer perceptrons
  with logistic activation function simply requires initializing the weights
  with an appropriate negative mean.
\newblock \emph{Neural Networks}, 153:\penalty0 87--103, 2022.
\newblock ISSN 0893-6080.
\newblock \doi{https://doi.org/10.1016/j.neunet.2022.05.030}.
\newblock URL
  \url{https://www.sciencedirect.com/science/article/pii/S0893608022002040}.

\end{thebibliography}

\newpage
\appendix
\begin{center}
  \Large\textbf{Supplementary Material}
\end{center}

\addcontentsline{toc}{section}{Supplementary Material}

\renewcommand{\thesection}{S\arabic{section}}
\setcounter{section}{0}

\renewcommand{\thepage}{S\arabic{page}}
\setcounter{page}{1}

\setcounter{figure}{0}
\renewcommand{\thefigure}{S\arabic{figure}}

\setcounter{table}{0}
\renewcommand{\thetable}{S\arabic{table}}

\makeatletter
\renewcommand\@oddfoot{\hfill\footnotesize\thepage\hfill}
\renewcommand\@evenfoot{\hfill\footnotesize\thepage\hfill}
\makeatother

\section{Test Problems and Degree of Overparametrisation}
\label{sec:test-problems-degree-overparam}

The \emph{Iris}, \emph{Wine}, and \emph{BCW} datasets were
sourced from the Scikit-learn library \citep{scikit-learn}, while
\emph{E-coli} and \emph{DNA} were obtained from the UCI Machine
Learning Repository \citep{Dua2019}. The DNA sequences were
pre-processed using one-hot encoding, resulting in a 228-dimensional
feature vector. Detailed information regarding these
datasets --- including the number of instances ($N$), features ($F$), and
classes ($K$) --- is provided in Table~\ref{table:data}.

\begin{table}[h]
  \caption{Information on the data sets used in this study.}
  \label{table:data}
  \begin{center}
      \begin{small}
          \begin{tabular}{|l|c|c|c|}
            \hline
            \emph{Problem} & \emph{Instances} ($N$) & \emph{Features} ($F$) & \emph{Classes} ($K$)  \\
            \hline
            BCW     & 569  & 30  & 2 \\
            DNA     & 106  & 228 & 2 \\
            E-coli  & 327  & 7   & 5 \\
            Iris    & 150  & 4   & 3 \\
            Wine    & 178  & 13  & 3 \\
            \hline
          \end{tabular}
      \end{small}
  \end{center}
\end{table}

The degree of over-parametrisation for the models and problems mentioned in
Section~\ref{sec:test-problems-and-models}, as quantified by the ratio between the
number of parameters, $P$, in a model and the number of instances, $N$, in
the training set is reported Table~\ref{tab:overparameterization_analysis}. 

\begin{table}[h]
  \caption{Analysis of the degree of overparametrisation $P/N$, where
    $P$ is the number of parameters (computed based on the number of
    features $F$ and the number of classes $K$
    from Table~\ref{table:data}, as well as the architecture of the ML model)
    and $N$ the number of instances (also from
    Table~\ref{table:data}).}
\label{tab:overparameterization_analysis}
\begin{center}
\begin{small}
  \begin{tabular}{|l|p{1.3in}|p{1.8in}|p{1.8in}|}
    \cline{2-4}
\multicolumn{1}{c|}{}& \multicolumn{3}{|c|}{\emph{Models}} \\
\hline
\emph{Dataset}  & \emph{LDA, LR, SVM} ($C=\infty$) & \emph{NN-ReLU} & \emph{DNN-ReLU} \\
  \hline
 & $P \approx 60 \Rightarrow$ & $P \approx 5{,}150 \Rightarrow$ & $P \approx 20{,}100 \Rightarrow$ \\ 
BCW & $P/N \approx 0.1$ & $P/N \approx 10$ & $P/N \approx 35$ \\ 
 & (Underparametrised) & (Moderately overparametrised) & (Highly overparametrised) \\ \hline 

 & $P \approx 460 \Rightarrow$ & $P \approx 20{,}100 \Rightarrow$ & $P \approx 33{,}000 \Rightarrow$ \\ 
DNA & $P/N \approx 4.3$ & $P/N \approx 170$ & $P/N \approx 300$ \\ 
 & (Overparametrised) & (Extremely overparametrised) & (Extremely overparametrised) \\ \hline 

 & $P \approx 40 \Rightarrow$ & $P \approx 3{,}900 \Rightarrow$ & $P \approx 18{,}900 \Rightarrow$ \\ 
E-coli & $P/N \approx 0.1$ & $P/N \approx 10$ & $P/N \approx 58$ \\ 
 & (Underparametrised) & (Highly overparametrised) & (Highly overparametrised) \\ \hline

 & $P \approx 15 \Rightarrow$ & $P \approx 3{,}500 \Rightarrow$ & $P \approx 18{,}500 \Rightarrow$ \\ 
Iris & $P/N \approx 0.1$ & $P/N \approx 25$ & $P/N \approx 120$ \\ 
 & (Underparametrised) & (Highly overparametrised) & (Extremely overparametrised) \\ \hline 

 & $P \approx 40 \Rightarrow$ & $P \approx 4{,}100 \Rightarrow$ & $P \approx 19{,}100 \Rightarrow$ \\ 
Wine & $P/N \approx 0.2$ & $P/N \approx 25$ & $P/N \approx 107$ \\ 
 & (Underparametrised) & (Highly overparametrised) & (Extremely overparametrised) \\ \hline
\end{tabular}
\end{small}
\end{center}
\end{table}

\FloatBarrier
\vfill
\mbox{}
\newpage

\section{Effect on the Results Induced by Changes in  the Confidence Parameter $z$}
\label{sec:how-results-are-affected-by-z}

All the results reported in  Section~\ref{sec:valid-perf-gener} were obtained using the
conventional value of $z = 0.99$. We discussed the theoretical
implications of changing $z$ in
Section~\ref{sec:sensitivity-analysis-z}. Based on that analysis,
in this section we look at the  effects that changing $z$  have
on the results reported in Sections~\ref{sec:train-test-comp}---\ref{sec:relative-difficulty}.%
\footnote{The results reported in Sections~\ref{sec:mean_train_valid}
  and~\ref{sec:succ-probabilities} are unaffected as they do not depend
on $z$.}

\subsection{Horizontal Displacement of Phase Transitions}
\label{sec:horiz-displ-phase-trans}

Because the single-run boundary is strictly bounded by the condition
$P(i, a, \eta) \ge z$, adjusting $z$ causes the \emph{phase transitions}
described in Sections~\ref{sec:train-test-comp}
and~\ref{sec:optim-learn-hyperp} to \emph{slide horizontally} along the
accuracy threshold ($a$) axis.

If $z$ is reduced (e.g., $z = 0.90$), the easier- and harder-success
single-run phases  expand significantly to the right. Because a
model only requires a 90\% chance of hitting the threshold in a single
run to avoid reinitialisation, the multi-run restart phase is delayed
and pushed to much higher accuracy thresholds. For inherently easier
datasets, the restart strategy may be entirely suppressed, rendering
the statistical benefit of exploiting path-dependent trajectories less
critical.

Conversely, if $z$ is increased (e.g., $z = 0.999$), the multi-run
phase transition triggers much sooner, shifting sharply to the left on
the $a$-axis. Under this regime, even if a single run boasts a highly
reliable 98\% success probability, it fails the $P(i, a, \eta) \ge z$
threshold, immediately forcing $R^* > 1$. Furthermore, the absolute
learnability and cherry-picking limits ($E^* \to \infty$) contract
horizontally to the left; requiring absolute certainty from a highly
 stochastic process  causes $R^*$ to explode non-linearly
except where the empirical success probability are extremely close to 1.

\subsection{The Influence of the Confidence Parameter $z$ on Model Selection}
\label{sec:influence-z-model-selection}

In our results in Section~\ref{sec:model-selection}, model selection was evaluated under the
assumption of a stringent confidence requirement ($z =
0.99$). However, relaxing or tightening this parameter does not merely
scale the computational effort uniformly across the board; it fundamentally alters the 
shape of the Pareto frontiers, thereby influencing which models are deemed
optimal for a given dataset and value of $a$. The influence of $z$ on model selection
manifests in the following ways.

\subsubsection{Relative Stability in the Multi-Run Regime}
\label{sec:relat-stab-multi}

For accuracy thresholds ($a$) where multiple models are \emph{already  in
the multi-run phase} ($R^* > 1$), changing $z$ applies a uniform
logarithmic multiplier to the expected number of runs across all
such models (as demonstrated in Section~\ref{sec:sensitivity-analysis-z}). 
Because this scaling factor is applied globally to these models, if $z$ is varied, their 
\emph{relative ranking} remains stable w.r.t.\ what was reported in
Section~\ref{sec:model-select-based-on_a}. 

\subsubsection{Rank-Shifting at the Phase Boundaries}
\label{sec:rank-shifting-at-phase-bound}

The most profound impact of $z$ on model selection occurs at the
boundary between the single-run and multi-run phases. This is because
\emph{models that were in the single-run phase ($P \ge z$) with
  $z=0.99$ and are still in that phase with the new $z$ are immune to
  the scaling factor} ($R$ remains $1$). However, models transitioning to
the multi-run phase as a result of the new $z$ value will likely see their computational
effort change. So, results on model dominance-relationships and model
selection based on $a$ (Section~\ref{sec:model-select-based-on_a}) are
likely to change if $z\ne 0.99$.

\subsubsection{Confidence-Accuracy Trade-off with Affordable Effort Budgets}
\label{sec:conf-accur-tradeoff-budget}

Finally, altering $z$ also impacts model selection under strict
computational budgets (explored
Section~\ref{sec:model-select-based-on_E}). In most cases, raising $z$
shifts computational efforts upward and, so, a fixed
budget of gradient descent steps yields a diminishing return in
achievable accuracy, but with almost certainty. On the contrary,
reducing $z$ makes it possible to reach higher accuracies with the
same budget, but with less confidence.

\subsection{The Impact of $z$ on the Perception of Problem Difficulty}
\label{sec:impact-z-relative-difficulty}

In the main text, we discussed problem difficulty from two
complementary perspectives: first, as a \emph{model-dependent} metric
used to rank the difficulty of different datasets for a specific
algorithm (Section~\ref{sec:probl-diff-grad}), and second, as a
\emph{relative} metric decomposed into the dual dimensions of
optimisability and generalisability
(Section~\ref{sec:relative-difficulty}). The confidence parameter $z$
plays a critical role in how both of these aspects are mathematically
perceived and visualised via the Pareto frontiers. Adjusting $z$
influences the perception of problem difficulty in the following ways.

\subsubsection{Invariance of Dataset Rankings in the Multi-Run Regime}
\label{sec:invar-relat-diffic-multi-run}

As established in Section~\ref{sec:influence-z-model-selection},
altering $z$ applies a uniform multiplier to the expected number of
runs. Therefore, when evaluating model-dependent problem difficulty
(Section~\ref{sec:probl-diff-grad}), the \emph{relative rank ordering}
of datasets for a given model is strictly preserved within the
multi-run phase. Consequently, in this phase, the fundamental
hierarchy of which datasets a specific model finds mathematically
``easy'' or ``hard'' is robust and invariant to the practitioner's
chosen confidence level.

\subsubsection{Exacerbation of Difficulty Gaps at Phase Boundaries}
While rank ordering among datasets is preserved in the multi-run
phase, the absolute magnitude of the perceived gap between easy and
hard problems is highly sensitive to $z$ due to the phase transition
boundary ($P(i, a, \eta) \ge z$). Demanding extreme confidence (e.g.,
$z=0.999$) acts as a magnifying glass for dataset difficulty. A
dataset with highly stable optimisability (such as Wine) may possess a
cumulative success probability $P$ so close to 1 that it easily
absorbs the stricter $z$ requirement, remaining safely within the
single-run phase and keeping its optimal computational effort entirely
unchanged. Conversely, a harder dataset with higher optimiser variance
(such as Iris) may fail the condition. Forced into the multi-run
regime, its effort is suddenly subjected to the logarithmic scaling
penalty associated with multiple restarts. Thus, raising $z$
drastically widens the perceived computational gap between stable and
unstable datasets,
 because the penalty for variance is applied asymmetrically across
problems.

\subsubsection{Stress-Testing the Dimensions of Relative Difficulty}
\label{sec:stress-test-dimens-rel-diff}

Finally, the choice of $z$ dictates the maximum accuracy threshold
($a$) at which a problem is deemed ``solvable'' by gradient descent
(the point where $E^* \to \infty$). Lowering the confidence
requirement (e.g., $z=0.90$) compresses the difficulty between
datasets by extending their Pareto frontiers further to the right.

Crucially, when viewed through the lens of relative difficulty
(Section~\ref{sec:relative-difficulty}), \emph{$z$ acts as a
  stress-test for learnability bottlenecks} irrespective of which
dimension --- optimisability (training) or generalisability
(validation) --- is the primary issue. For instance, a dataset that is
extremely difficult to generalise will hit its cherry-picking limit
for validation accuracy at $a=0.95$ under strict confidence. However,
it might appear learnable up to $a=0.99$ if the practitioner is
willing to accept a 10\% risk of failure ($z=0.90$). Conversely,
consider a dataset where the primary bottleneck is optimisability,
presenting a treacherous error surface where gradient descent frequently
stalls or progresses too slowly. Under strict confidence ($z=0.99$), the model 
will hit an early learnability limit for training accuracy because too 
few runs successfully reach the required accuracy threshold $a$. Yet, by relaxing $z$, 
the practitioner can rely on the rare, ``lucky'' random initialisation 
that successfully navigates the landscape rapidly enough to hit $a$, making the training problem 
appear deceptively solvable at much higher accuracies. Therefore, a high
value of $z$ exposes a dataset's underlying learnability
deficiencies---whether that is an inability to converge efficiently (poor
optimisability) or a proneness to overfitting (poor generalisability)---while 
a lower value may mask these bottlenecks.

\vfill
\mbox{}
\pagebreak

\section{Accuracies}
\label{app:mean_accuracies}

In this section we provide plots of the mean training (left) and
validation (right) accuracies for all the problems and models used in
this article.

\begin{figure*}[p]
  \centering
  \begin{tabular}{c@{}c}
    Training Accuracy & Validation Accuracy \\
    \includegraphics[clip,trim=70 30 30 30,width=.52\linewidth]{arXiv-figures/bcw_NN-ReLU_average_accuracy_train.pdf} &
    \includegraphics[clip,trim=70 30 30 30,width=.52\linewidth]{arXiv-figures/bcw_NN-ReLU_average_accuracy_test.pdf} \\[-3mm]
   \includegraphics[clip,trim=70 30 30 30,width=.52\linewidth]{arXiv-figures/dna_NN-ReLU_average_accuracy_train.pdf} &
    \includegraphics[clip,trim=70 30 30 30,width=.52\linewidth]{arXiv-figures/dna_NN-ReLU_average_accuracy_test.pdf} \\[-3mm]
    \includegraphics[clip,trim=70 30 30 30,width=.52\linewidth]{arXiv-figures/ecoli_NN-ReLU_average_accuracy_train.pdf} &
    \includegraphics[clip,trim=70 30 30 30,width=.52\linewidth]{arXiv-figures/ecoli_NN-ReLU_average_accuracy_test.pdf} \\[-3mm]
  \end{tabular}
\caption{Accuracies for NN-ReLU model.}
  \label{fig:accuracies_NN_ReLU_model_duplicate}
\end{figure*}

\begin{figure*}[p]
  \centering
  \ContinuedFloat
  \begin{tabular}{c@{}c}
    Training Accuracy & Validation Accuracy \\
    \includegraphics[clip,trim=70 30 30 30,width=.52\linewidth]{arXiv-figures/iris_NN-ReLU_average_accuracy_train.pdf} &
    \includegraphics[clip,trim=70 30 30 30,width=.52\linewidth]{arXiv-figures/iris_NN-ReLU_average_accuracy_test.pdf} \\[-3mm]
    \includegraphics[clip,trim=70 30 30 30,width=.52\linewidth]{arXiv-figures/wine_NN-ReLU_average_accuracy_train.pdf} &
    \includegraphics[clip,trim=70 30 30 30,width=.52\linewidth]{arXiv-figures/wine_NN-ReLU_average_accuracy_test.pdf} \\[-3mm]
  \end{tabular}
\caption{Accuracies for NN-ReLU model (continued).}
\end{figure*}

\begin{figure*}[p]
  \centering
  \begin{tabular}{c@{}c}
    Training Accuracy & Validation Accuracy \\
    \includegraphics[clip,trim=70 30 30 30,width=.52\linewidth]{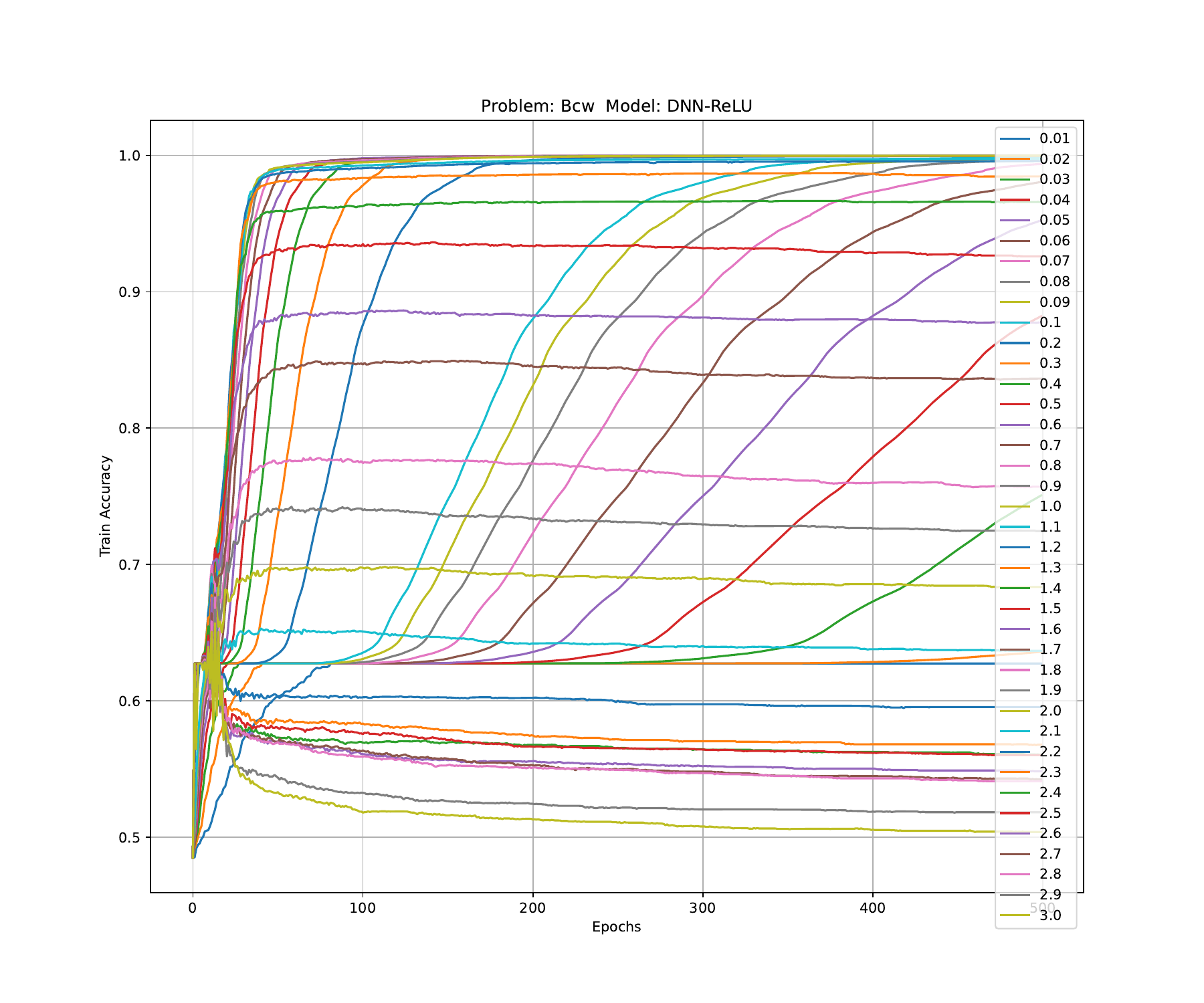} &
    \includegraphics[clip,trim=70 30 30 30,width=.52\linewidth]{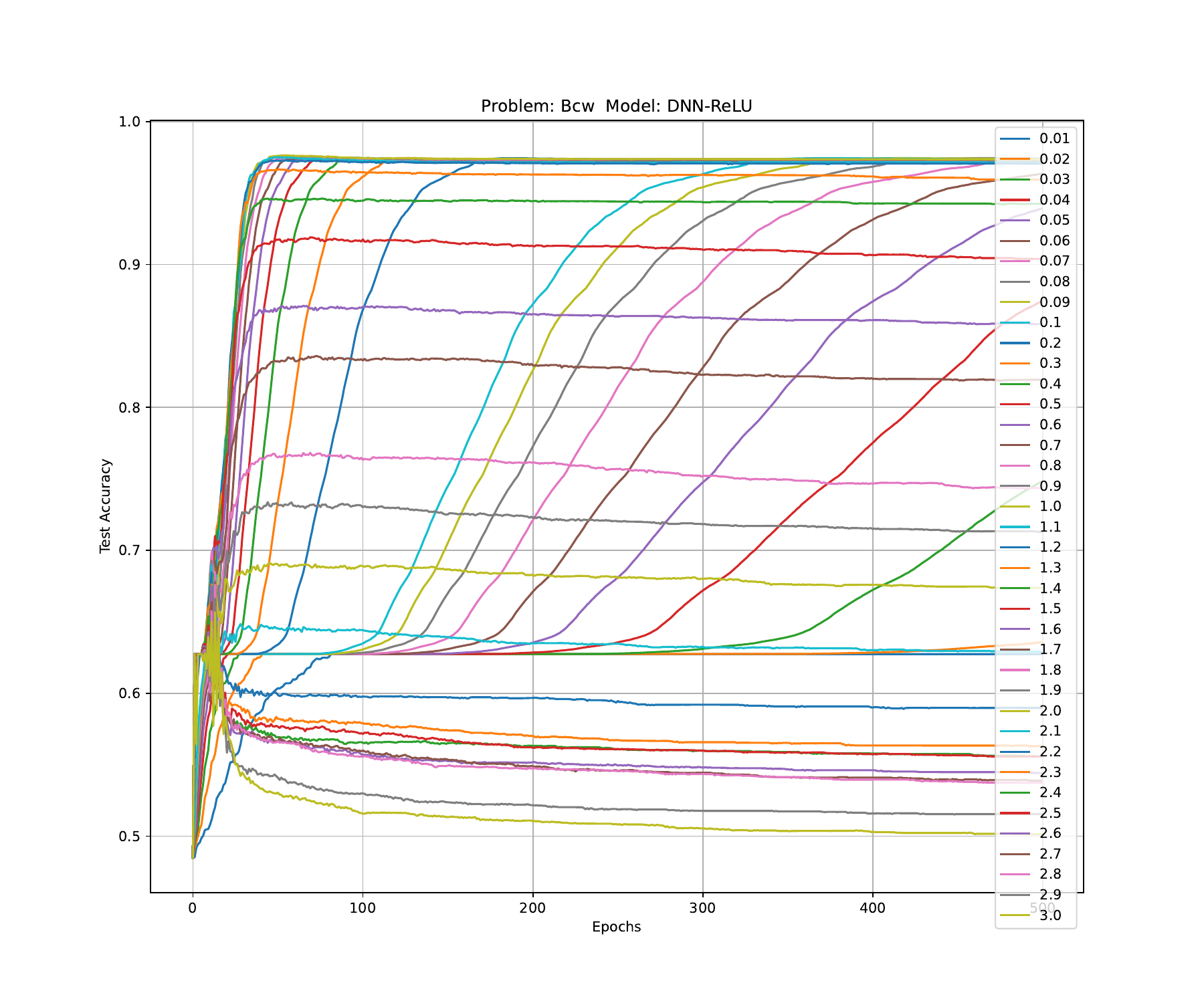} \\[-3mm]
    \includegraphics[clip,trim=70 30 30 30,width=.52\linewidth]{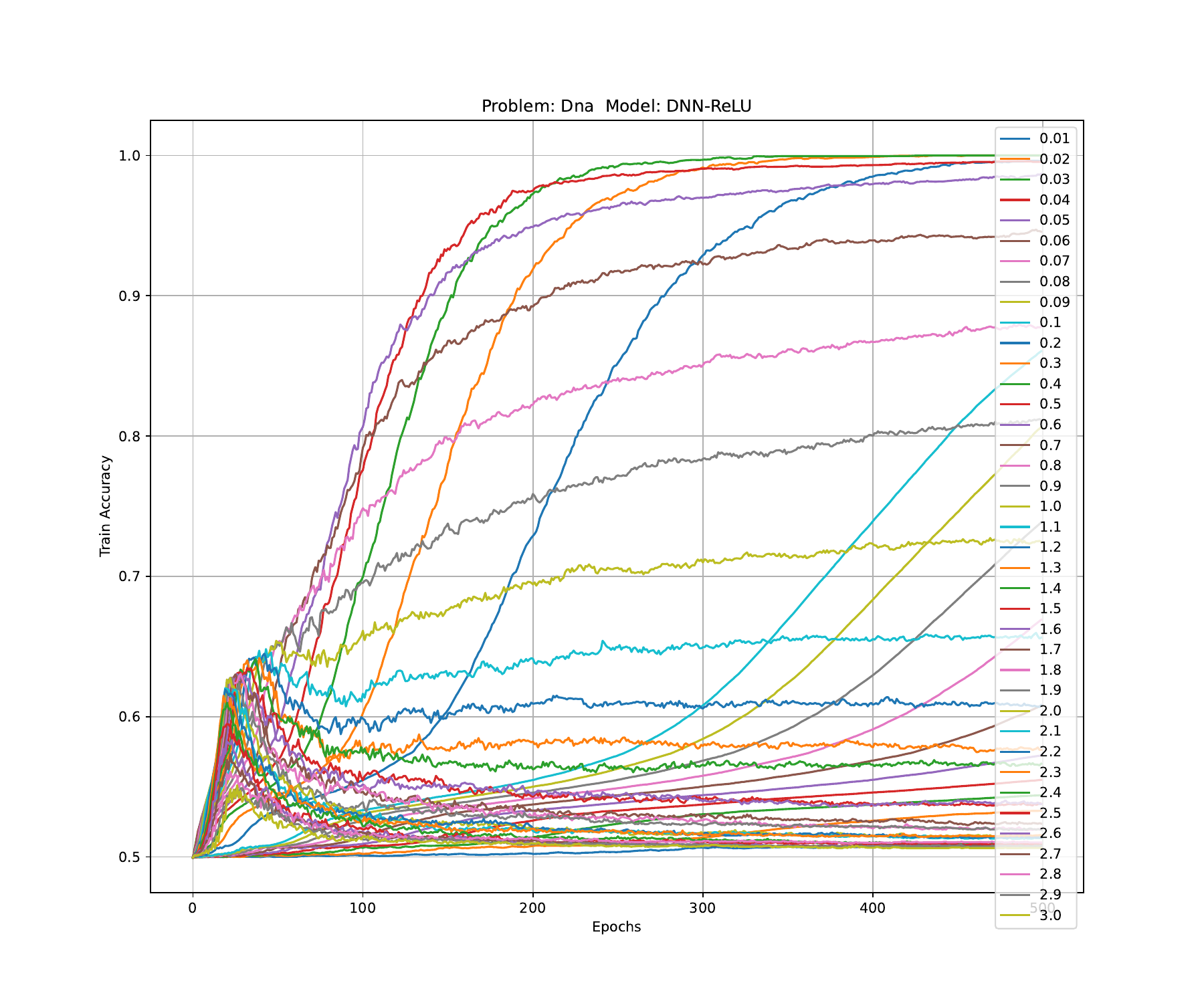} &
    \includegraphics[clip,trim=70 30 30 30,width=.52\linewidth]{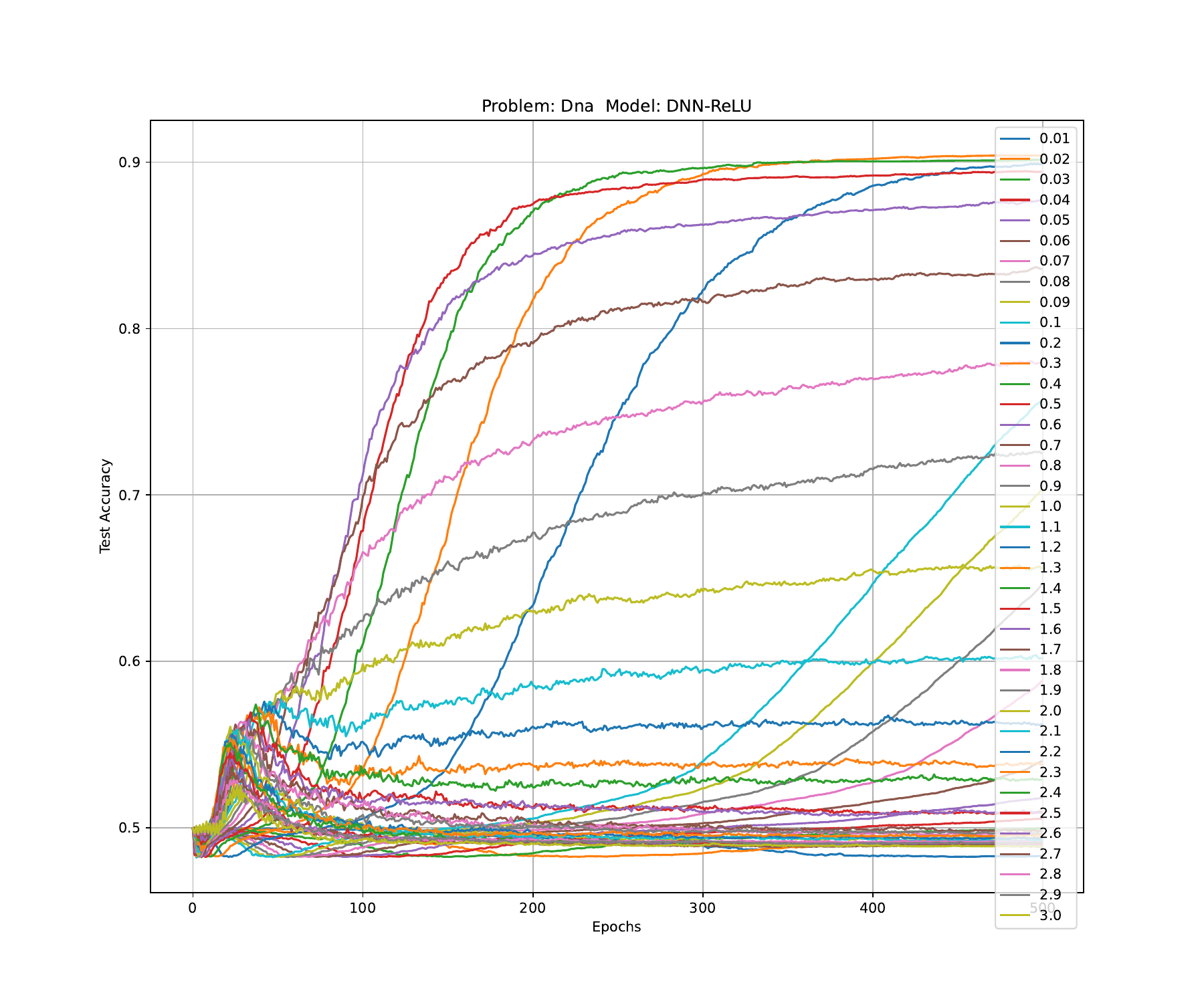} \\[-3mm]
    \includegraphics[clip,trim=70 30 30 30,width=.52\linewidth]{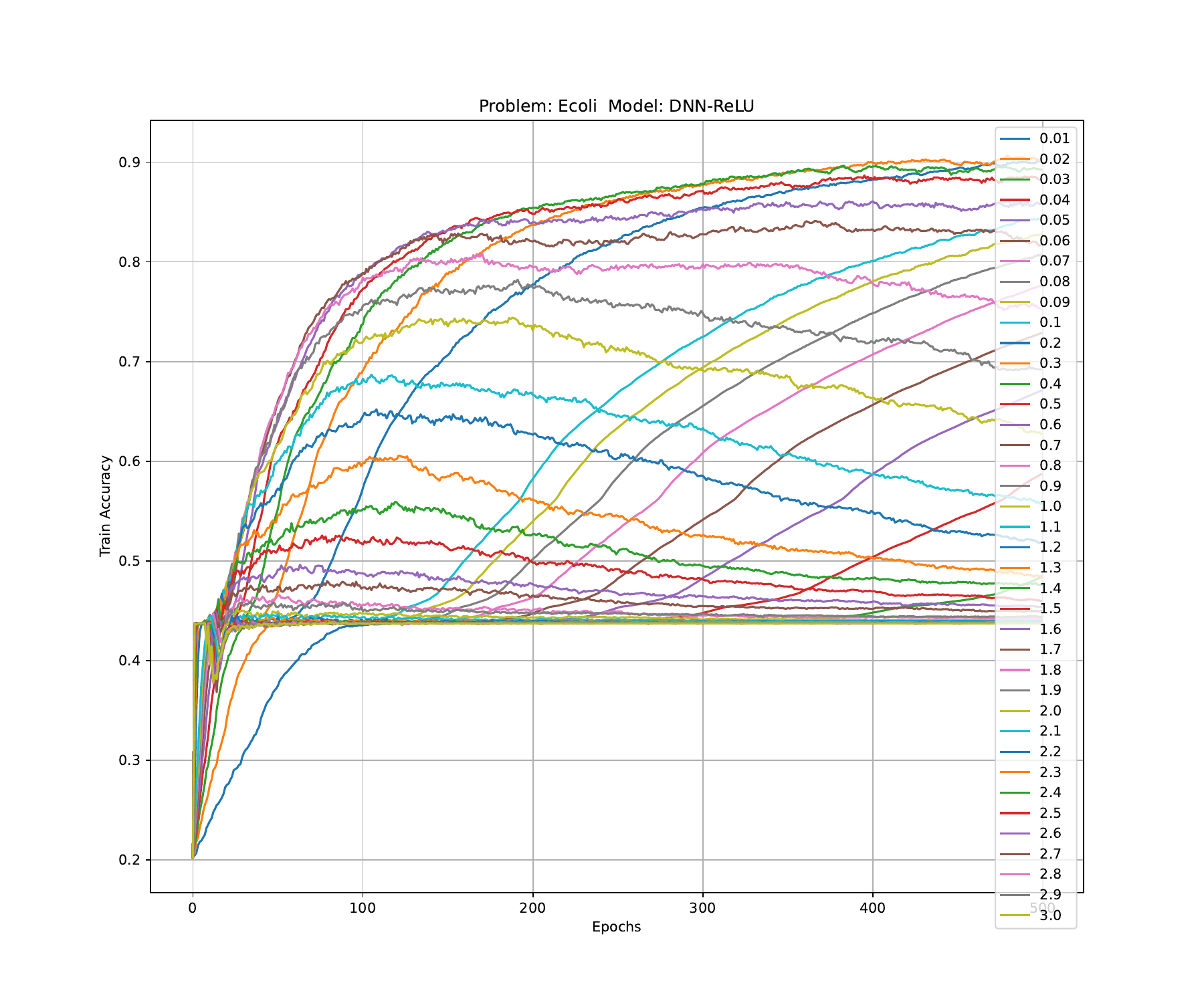} &
    \includegraphics[clip,trim=70 30 30 30,width=.52\linewidth]{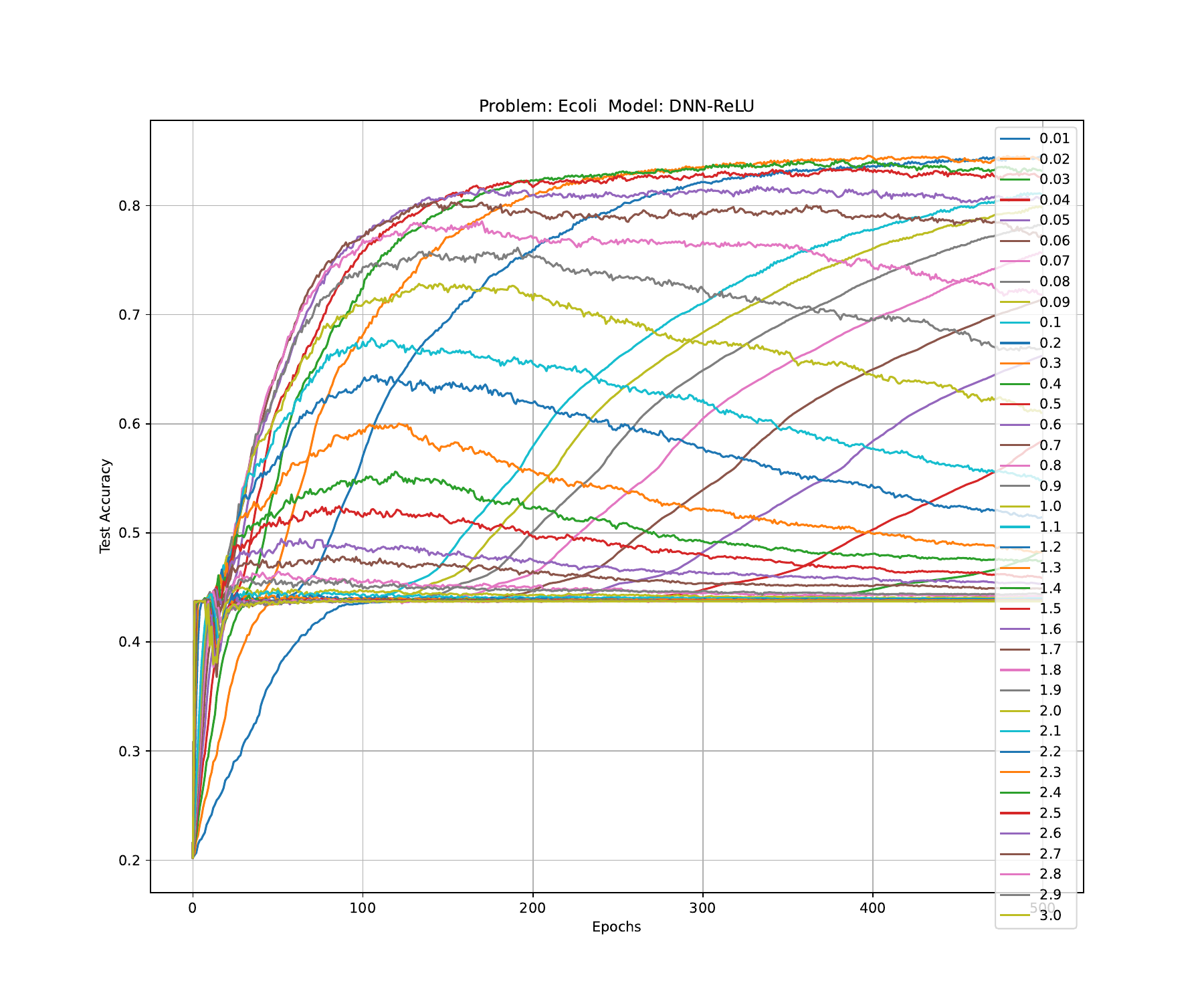} \\[-3mm]
  \end{tabular}
\caption{Accuracies for DNN-ReLU model.}
  \label{fig:accuracies_DNN_ReLU_model}
\end{figure*}

\begin{figure*}[p]
  \centering
  \ContinuedFloat
  \begin{tabular}{c@{}c}
    Training Accuracy & Validation Accuracy \\
    \includegraphics[clip,trim=70 30 30 30,width=.52\linewidth]{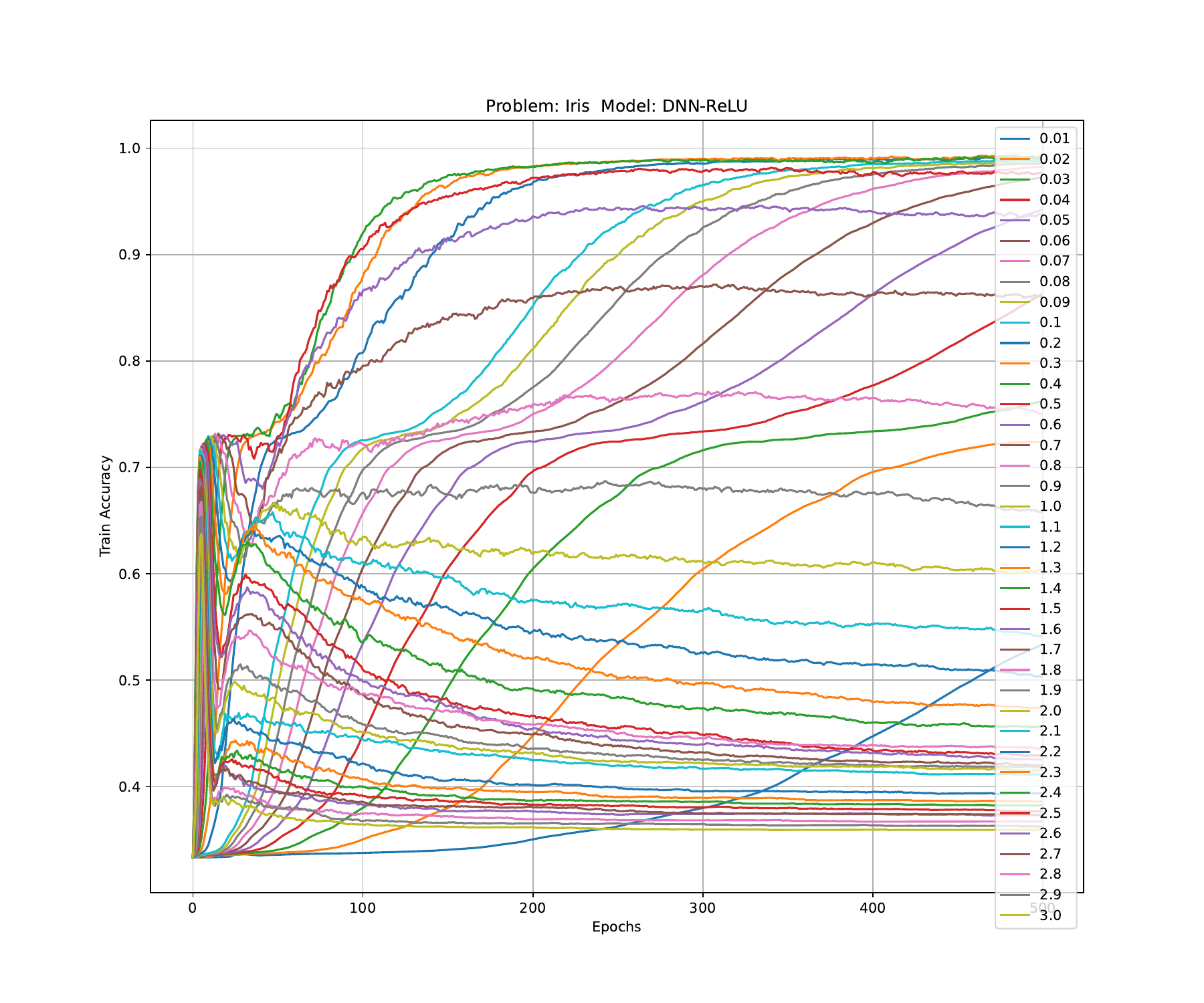} &
    \includegraphics[clip,trim=70 30 30 30,width=.52\linewidth]{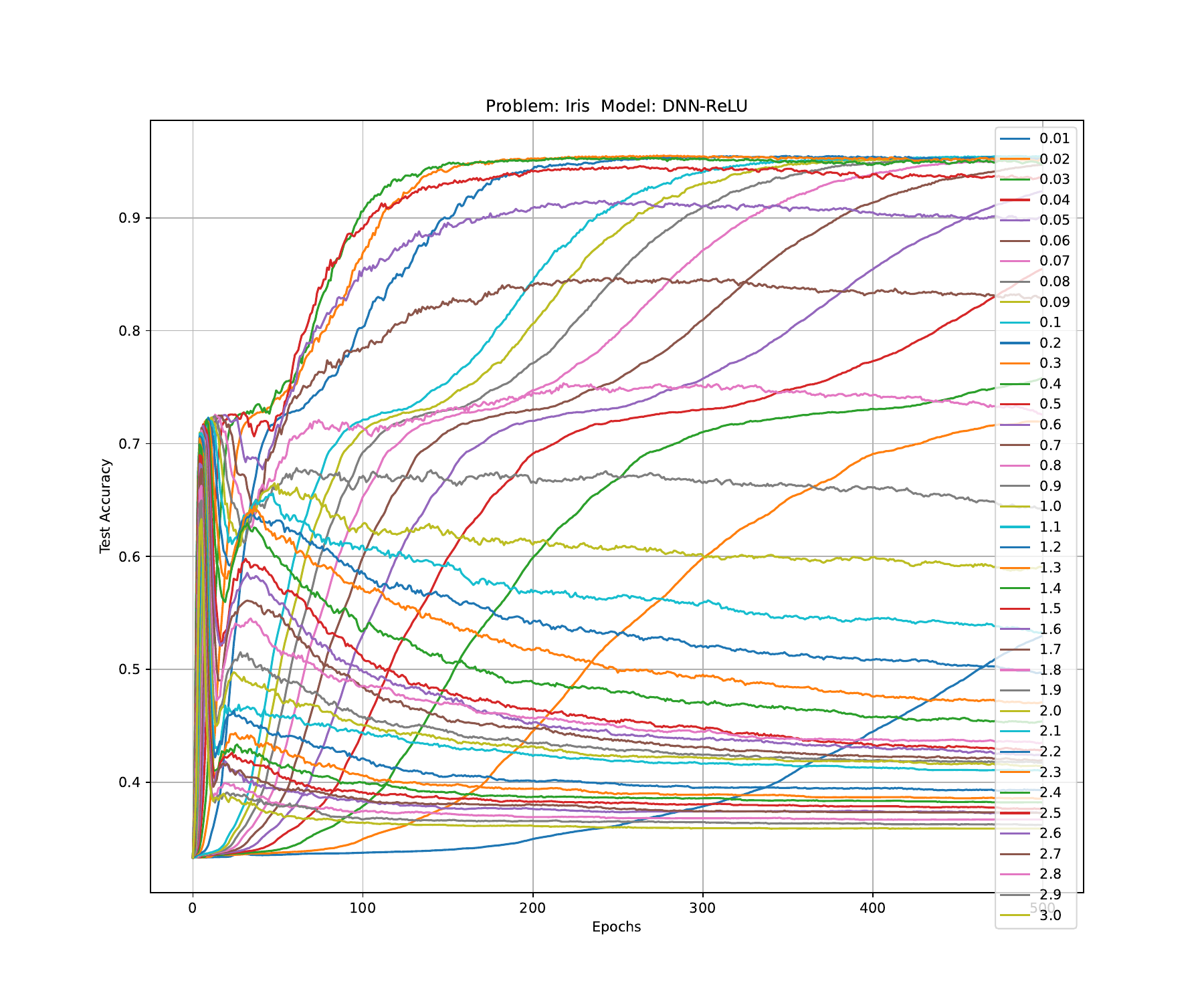} \\[-3mm]
    \includegraphics[clip,trim=70 30 30 30,width=.52\linewidth]{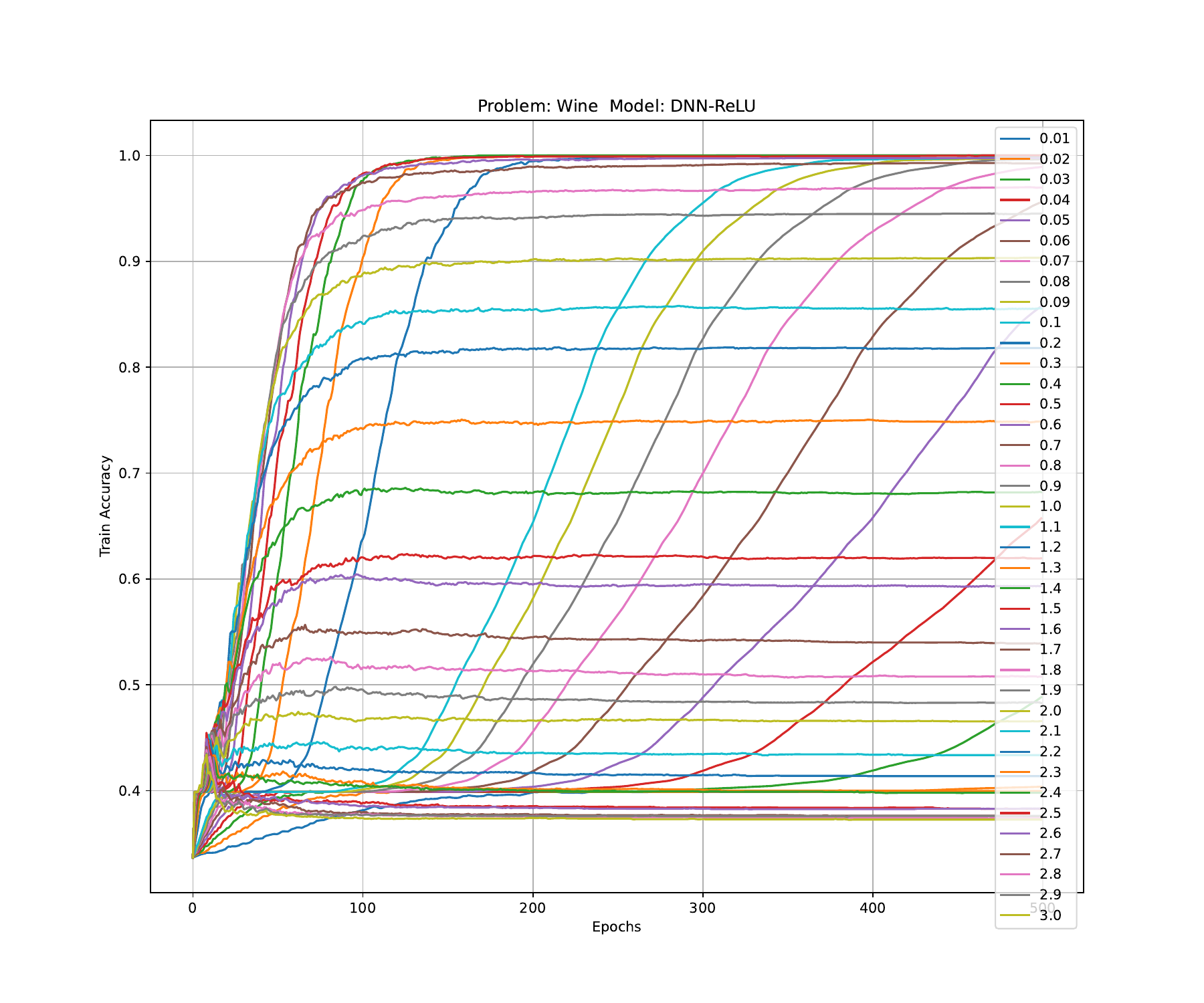} &
    \includegraphics[clip,trim=70 30 30 30,width=.52\linewidth]{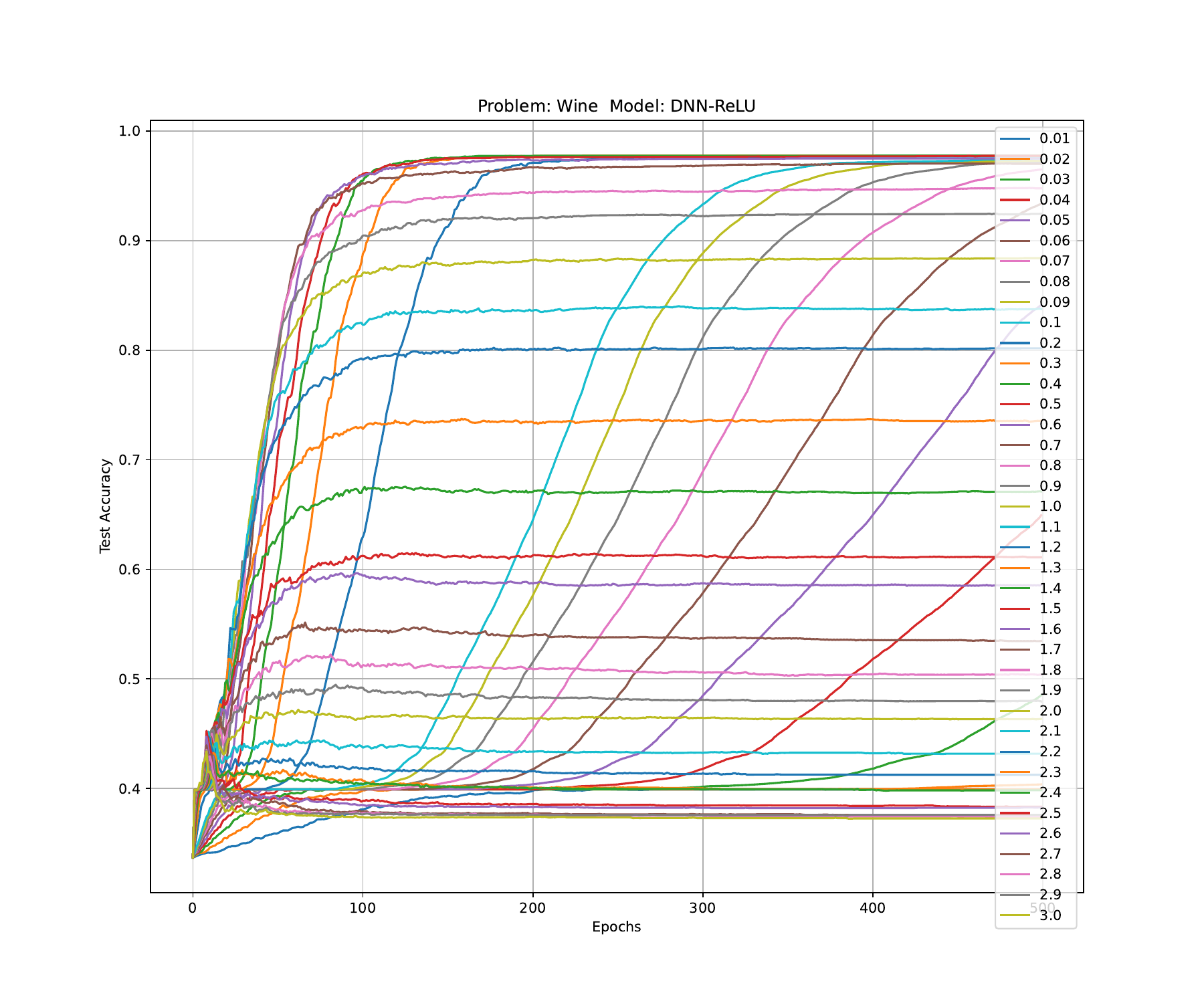} \\[-3mm]
  \end{tabular}
\caption{Accuracies for DNN-ReLU model (continued).}
\end{figure*}

\begin{figure*}[p]
  \centering
  \begin{tabular}{c@{}c}
    Training Accuracy & Validation Accuracy \\
    \includegraphics[clip,trim=70 30 30 30,width=.52\linewidth]{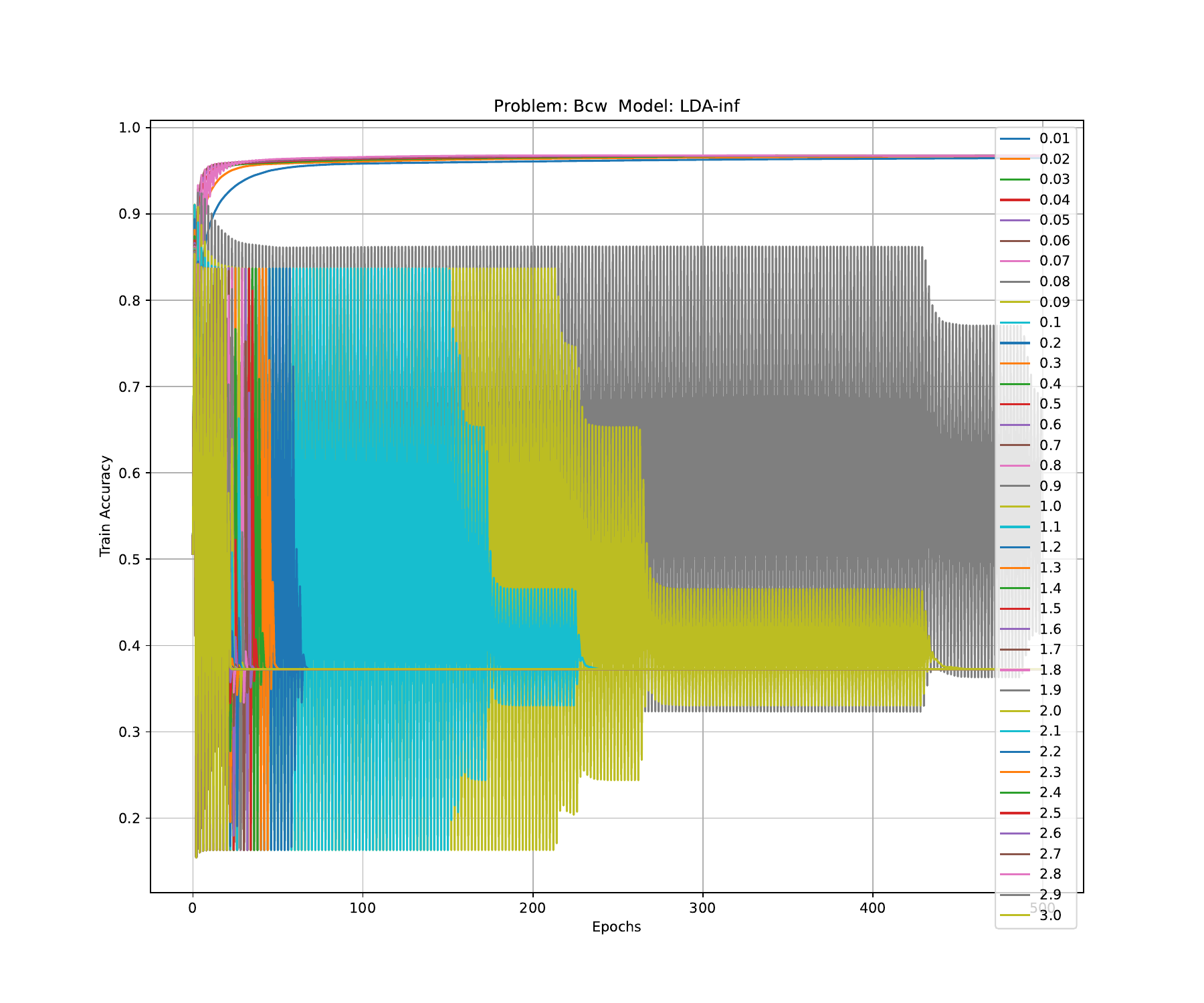} &
    \includegraphics[clip,trim=70 30 30 30,width=.52\linewidth]{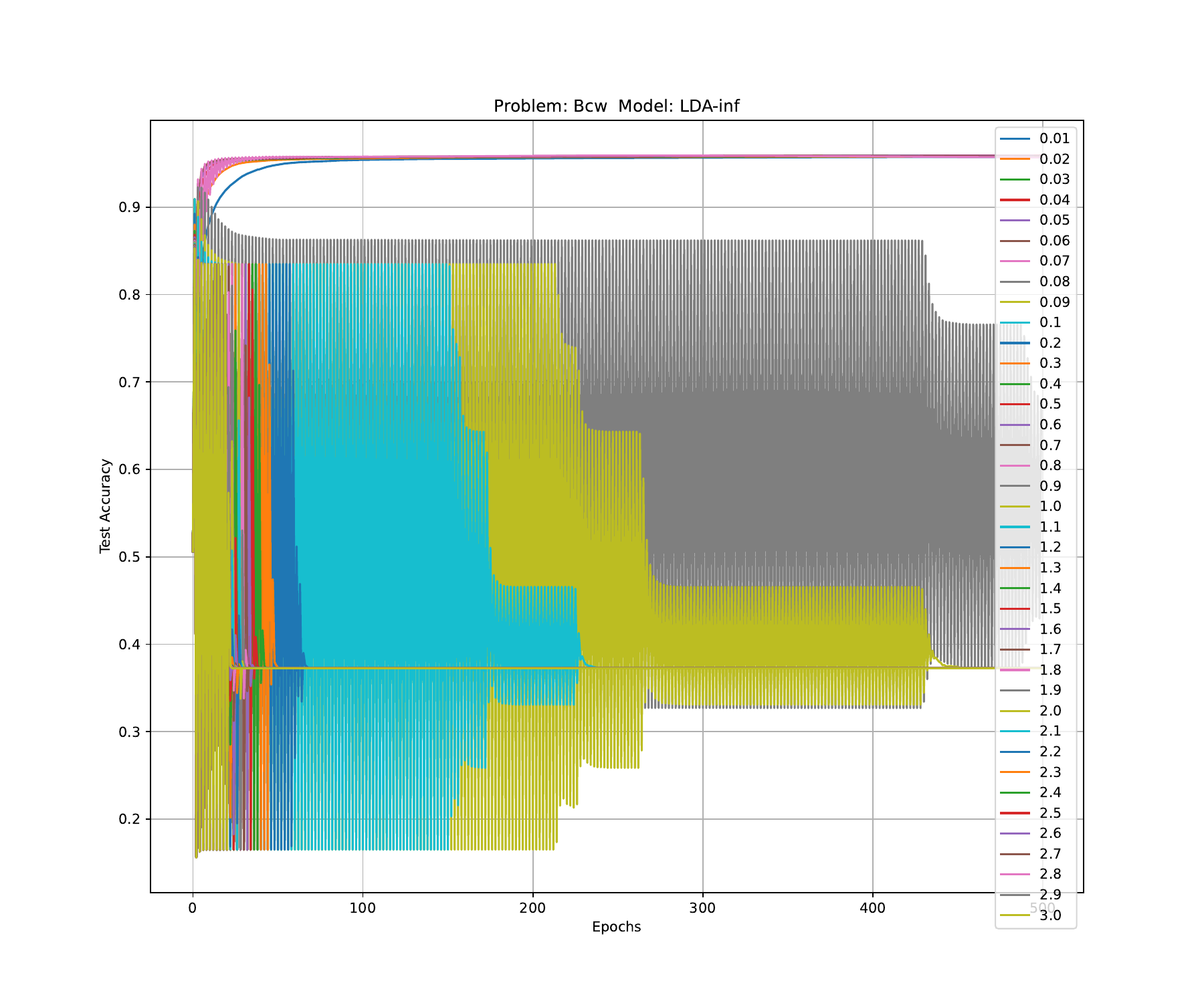} \\[-3mm]
    \includegraphics[clip,trim=70 30 30 30,width=.52\linewidth]{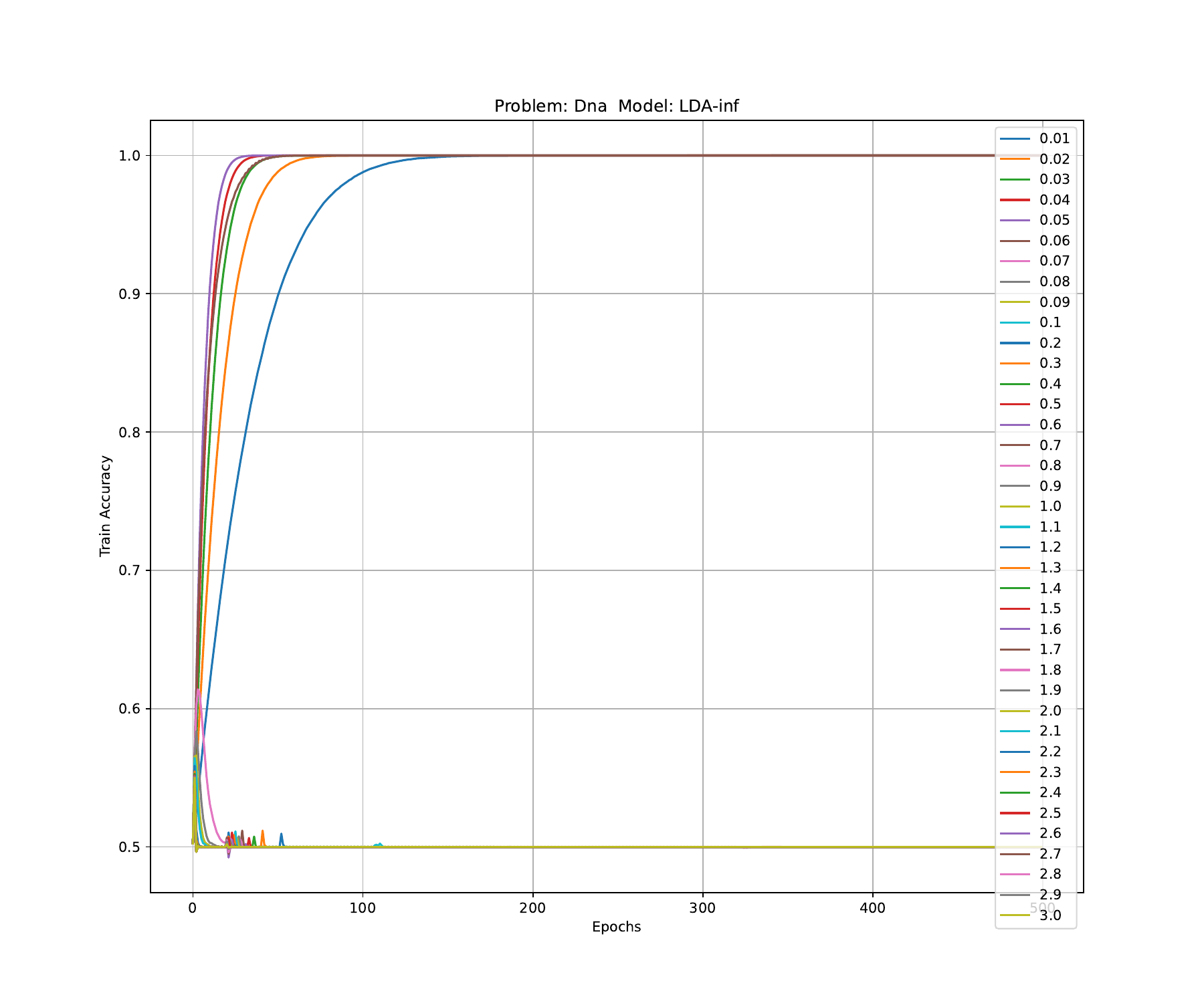} &
    \includegraphics[clip,trim=70 30 30 30,width=.52\linewidth]{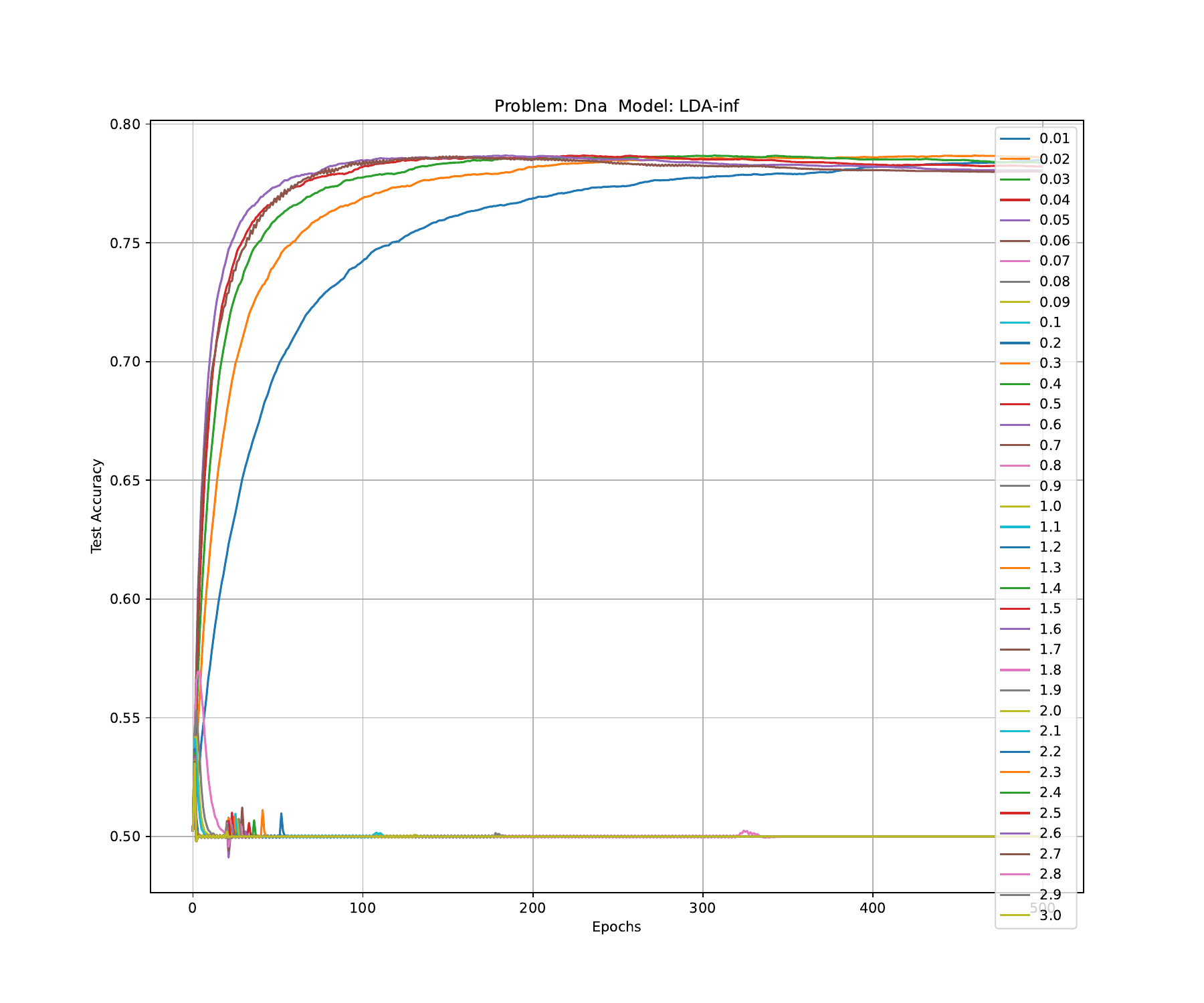} \\[-3mm]
    \includegraphics[clip,trim=70 30 30 30,width=.52\linewidth]{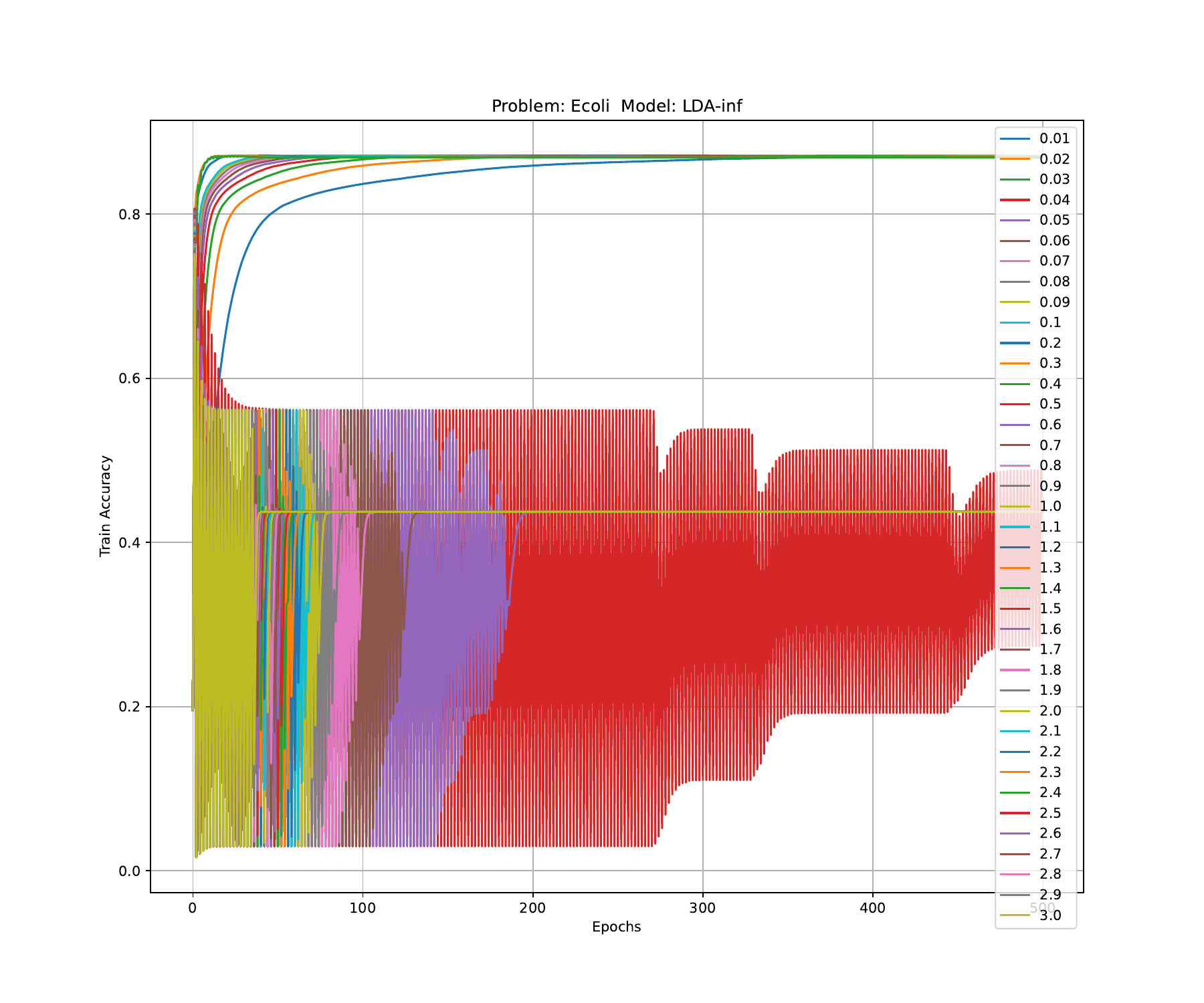} &
    \includegraphics[clip,trim=70 30 30 30,width=.52\linewidth]{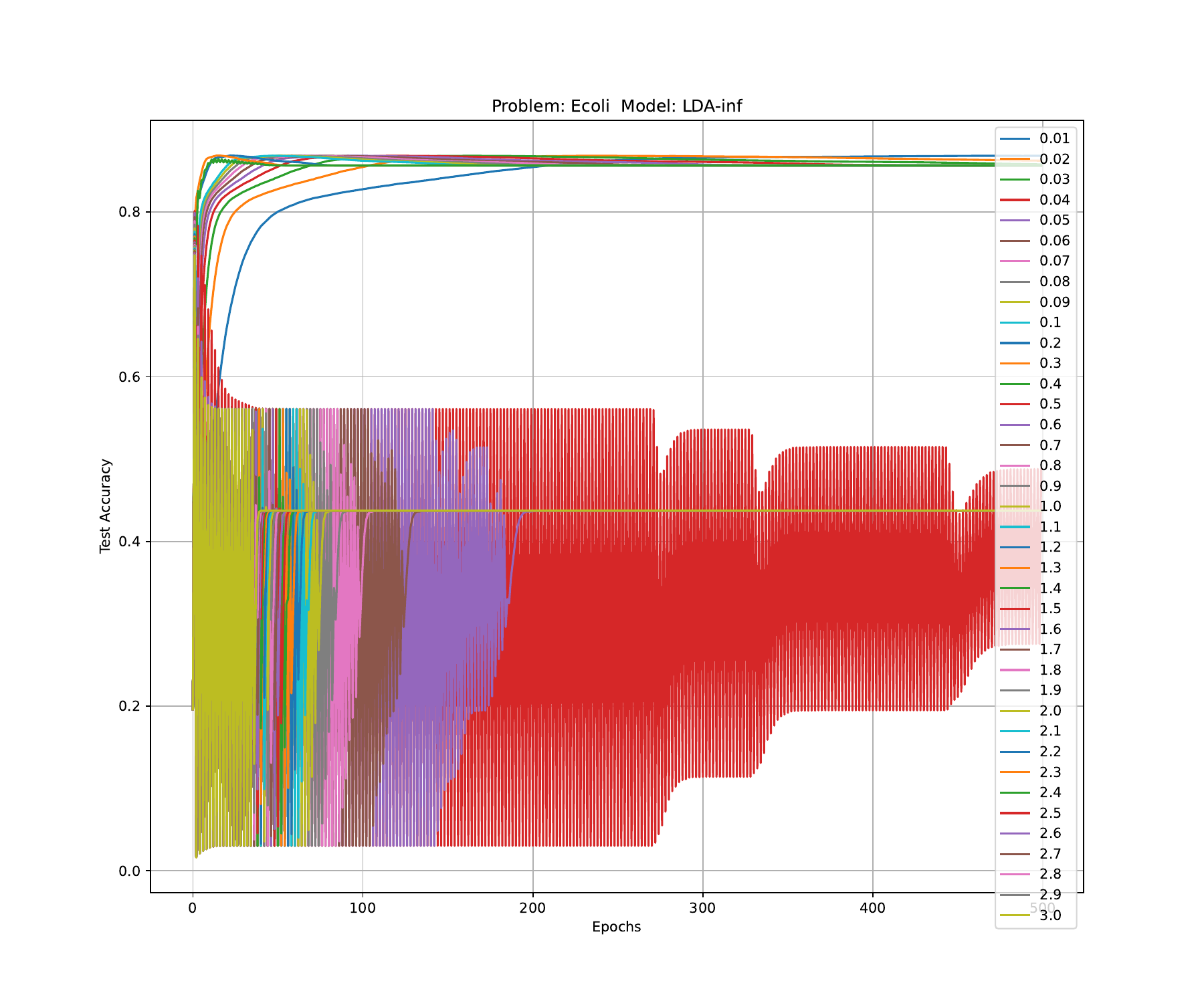} \\[-3mm]
  \end{tabular}
\caption{Accuracies for LDA-Inf model.}
  \label{fig:accuracies_LDA_Inf_model}
\end{figure*}

\begin{figure*}[p]
  \centering
  \ContinuedFloat
  \begin{tabular}{c@{}c}
    Training Accuracy & Validation Accuracy \\
    \includegraphics[clip,trim=70 30 30 30,width=.52\linewidth]{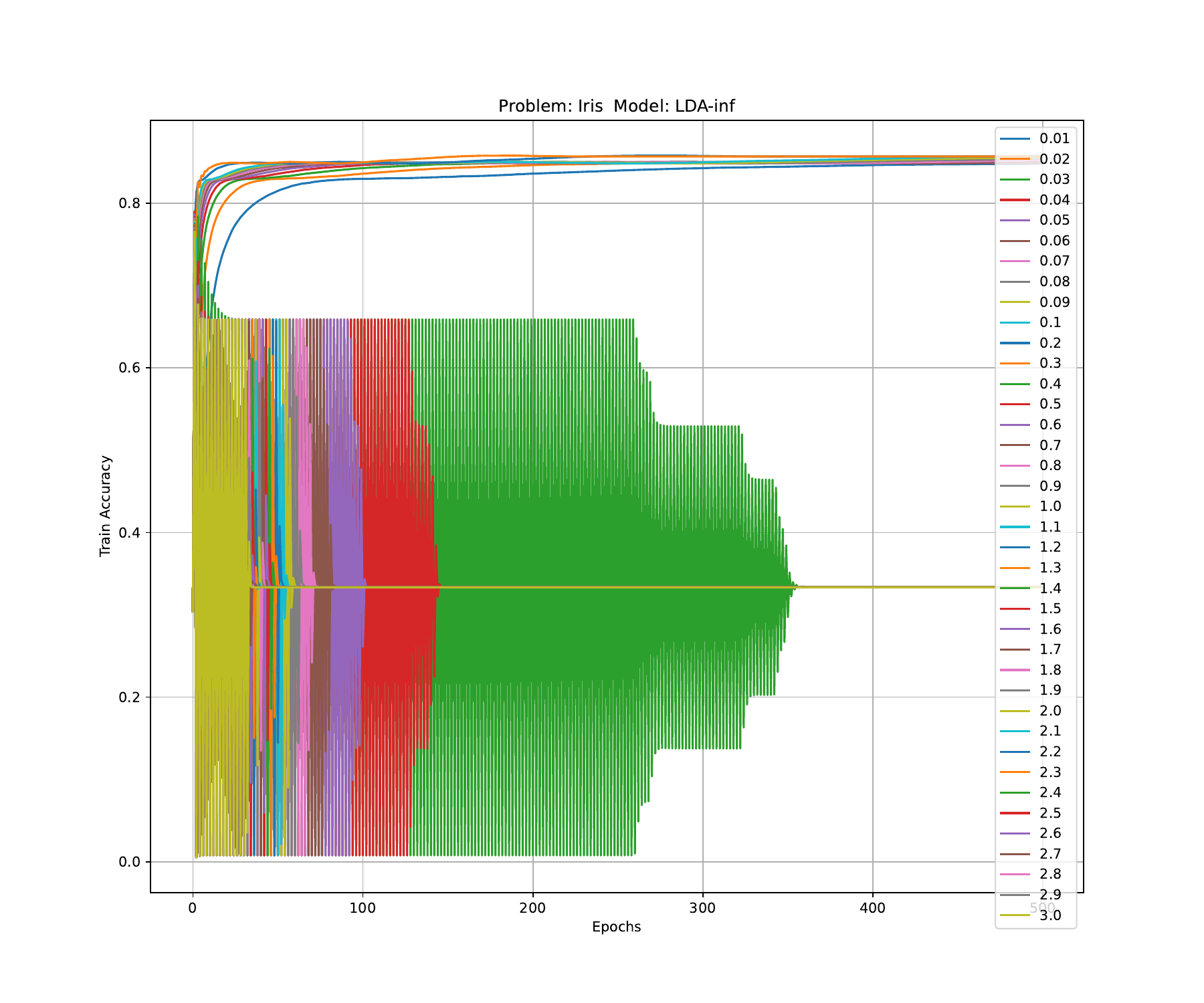} &
    \includegraphics[clip,trim=70 30 30 30,width=.52\linewidth]{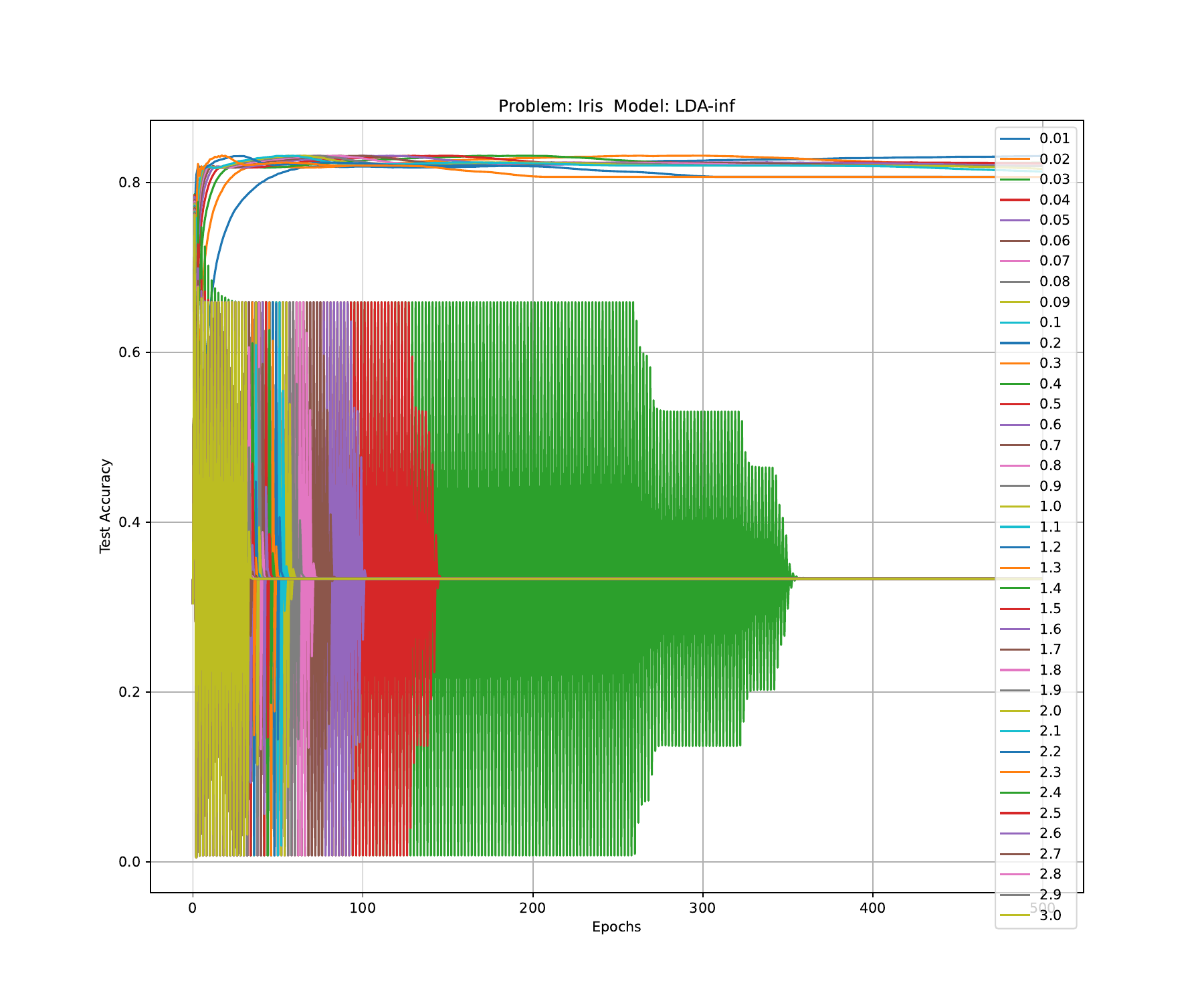} \\[-3mm]
    \includegraphics[clip,trim=70 30 30 30,width=.52\linewidth]{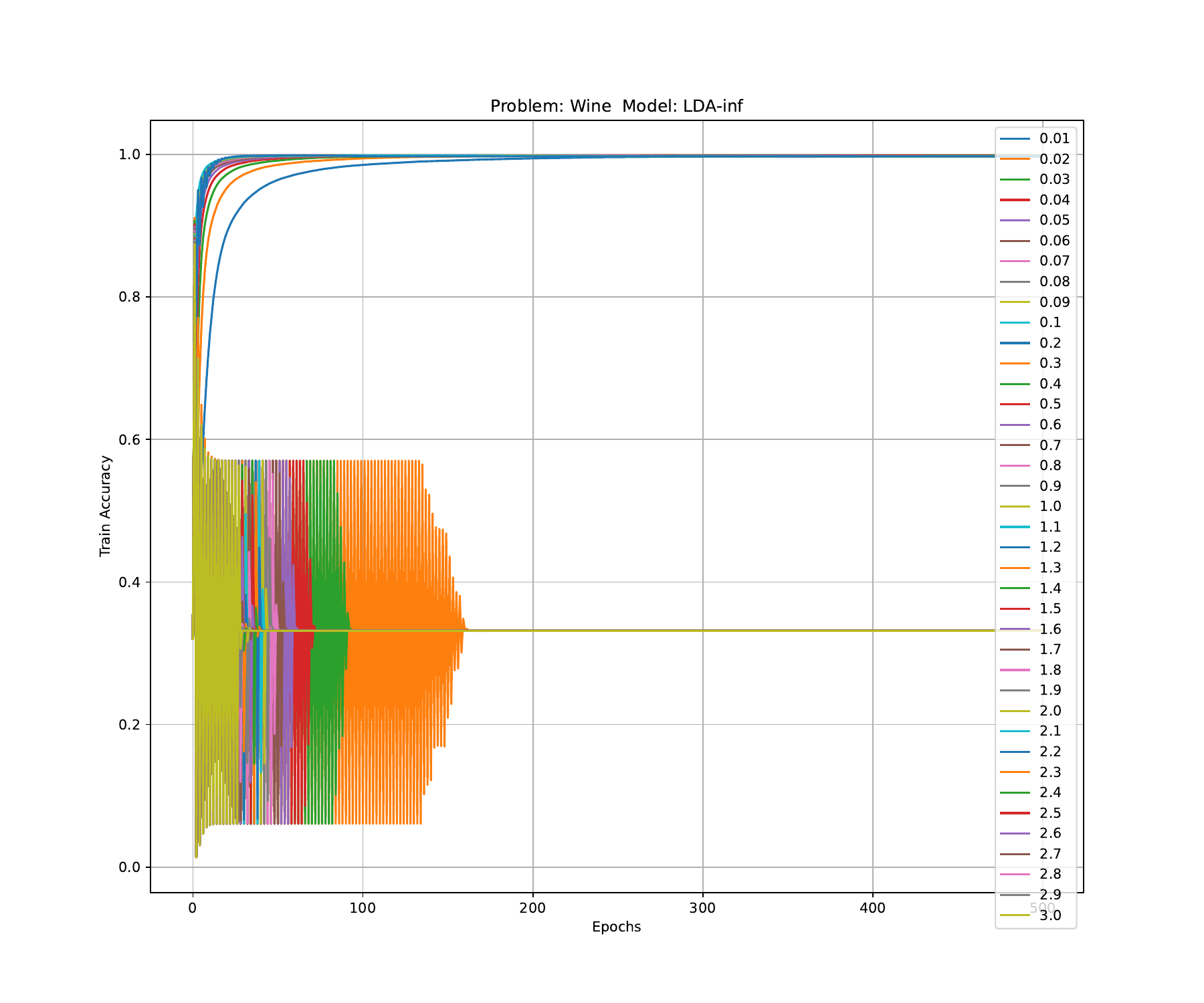} &
    \includegraphics[clip,trim=70 30 30 30,width=.52\linewidth]{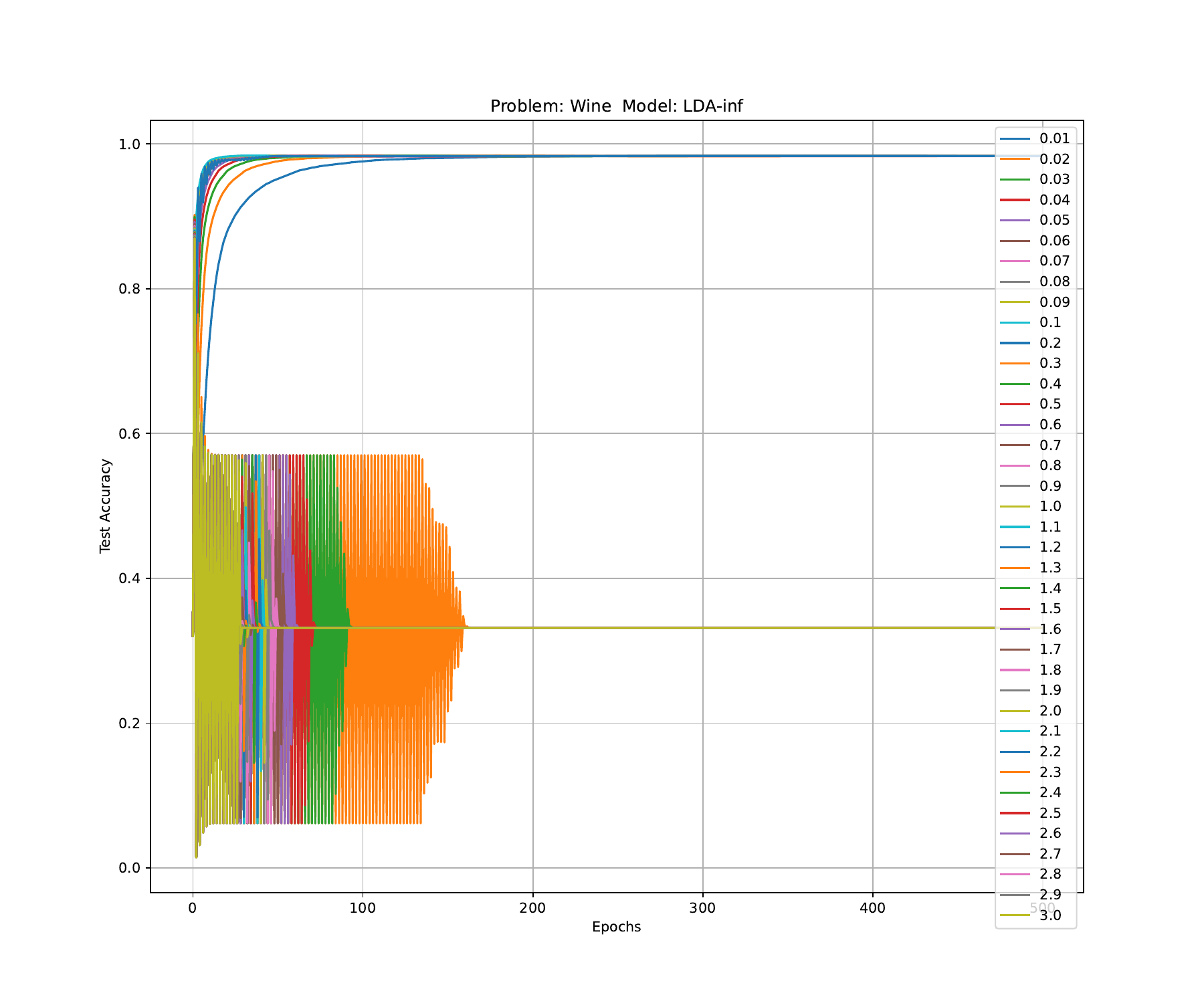} \\[-3mm]
  \end{tabular}
\caption{Accuracies for LDA-Inf model (continued).}
\end{figure*}

\begin{figure*}[p]
  \centering
  \begin{tabular}{c@{}c}
    Training Accuracy & Validation Accuracy \\
    \includegraphics[clip,trim=70 30 30 30,width=.52\linewidth]{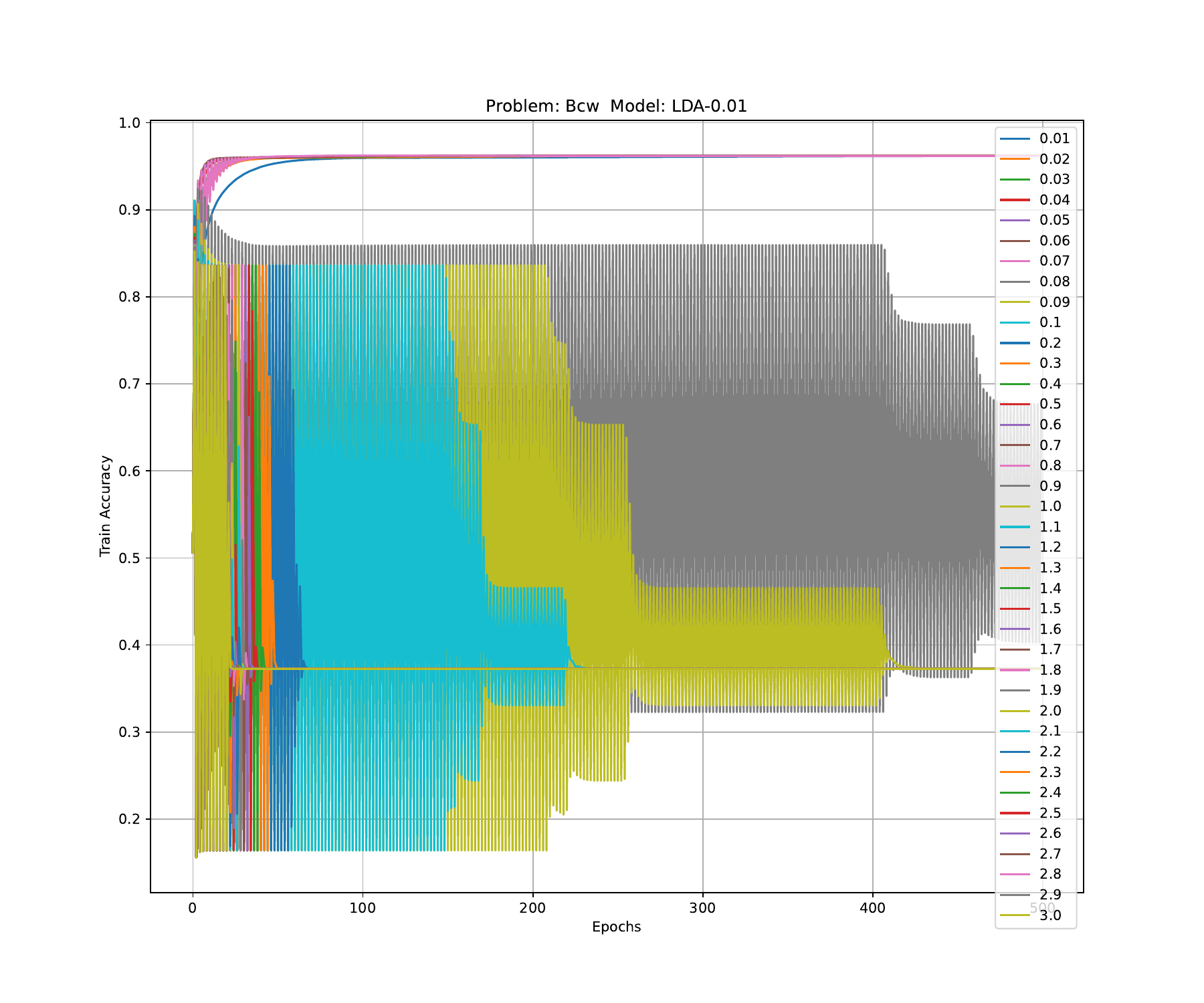} &
    \includegraphics[clip,trim=70 30 30 30,width=.52\linewidth]{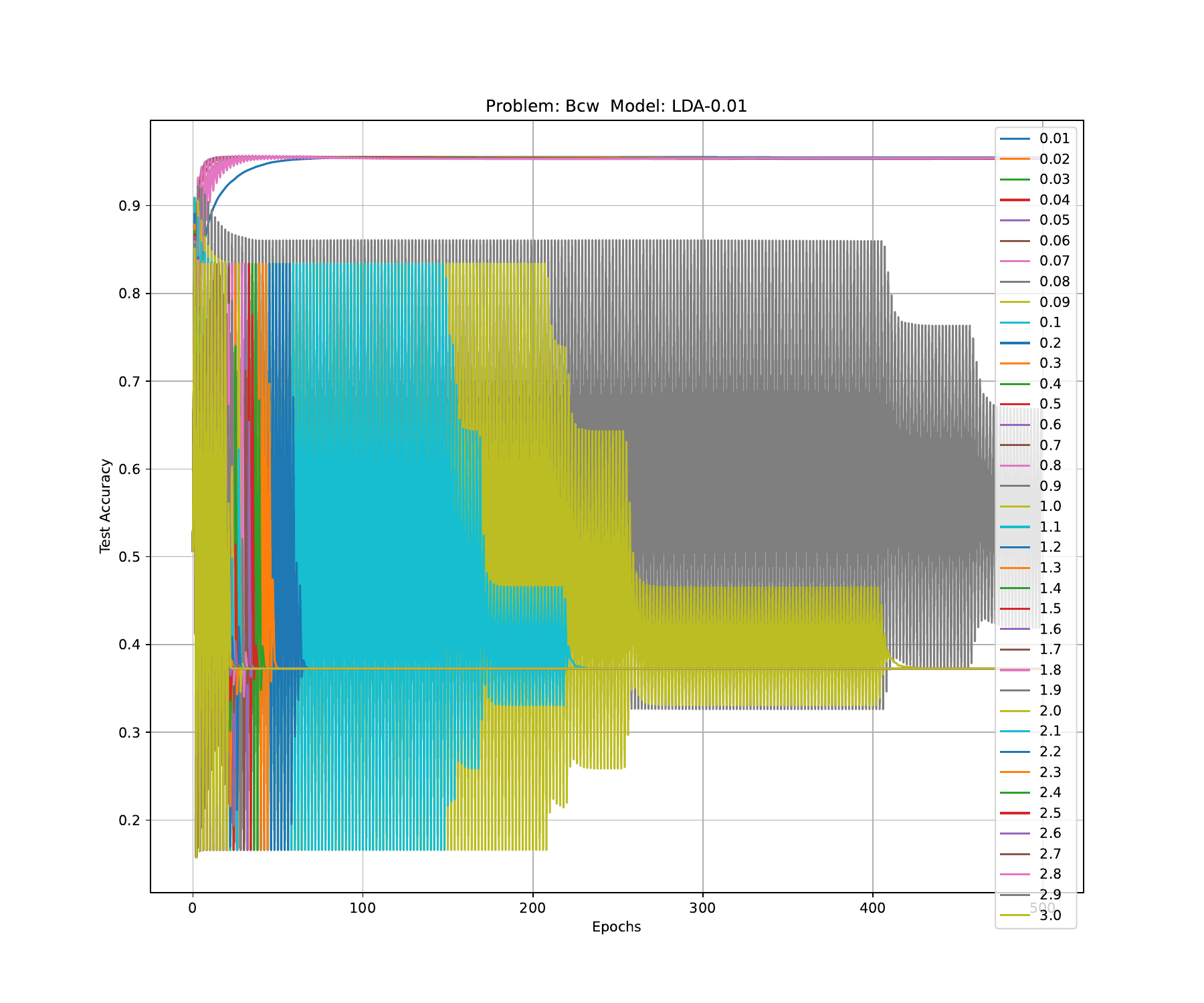} \\[-3mm]
    \includegraphics[clip,trim=70 30 30 30,width=.52\linewidth]{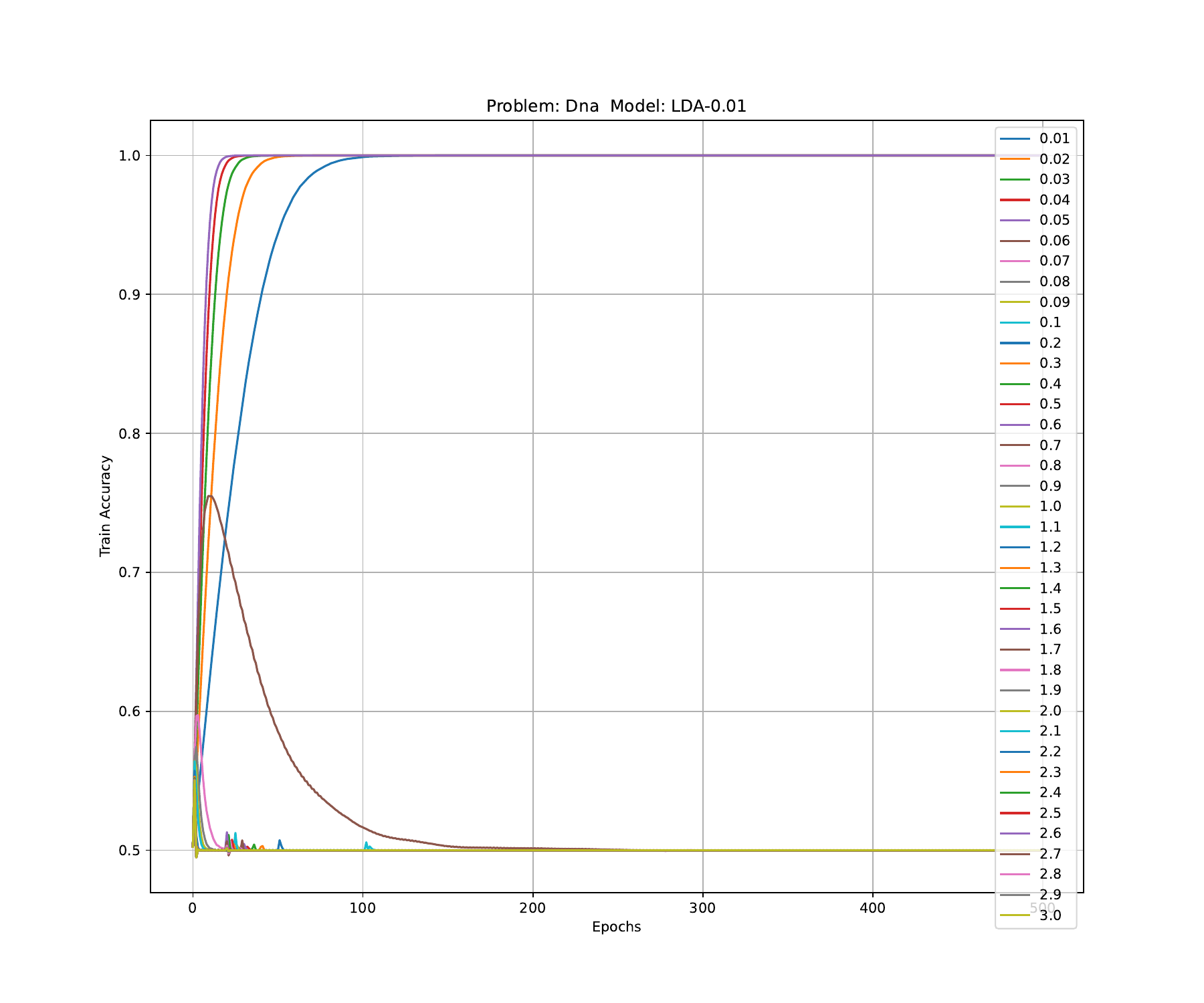} &
    \includegraphics[clip,trim=70 30 30 30,width=.52\linewidth]{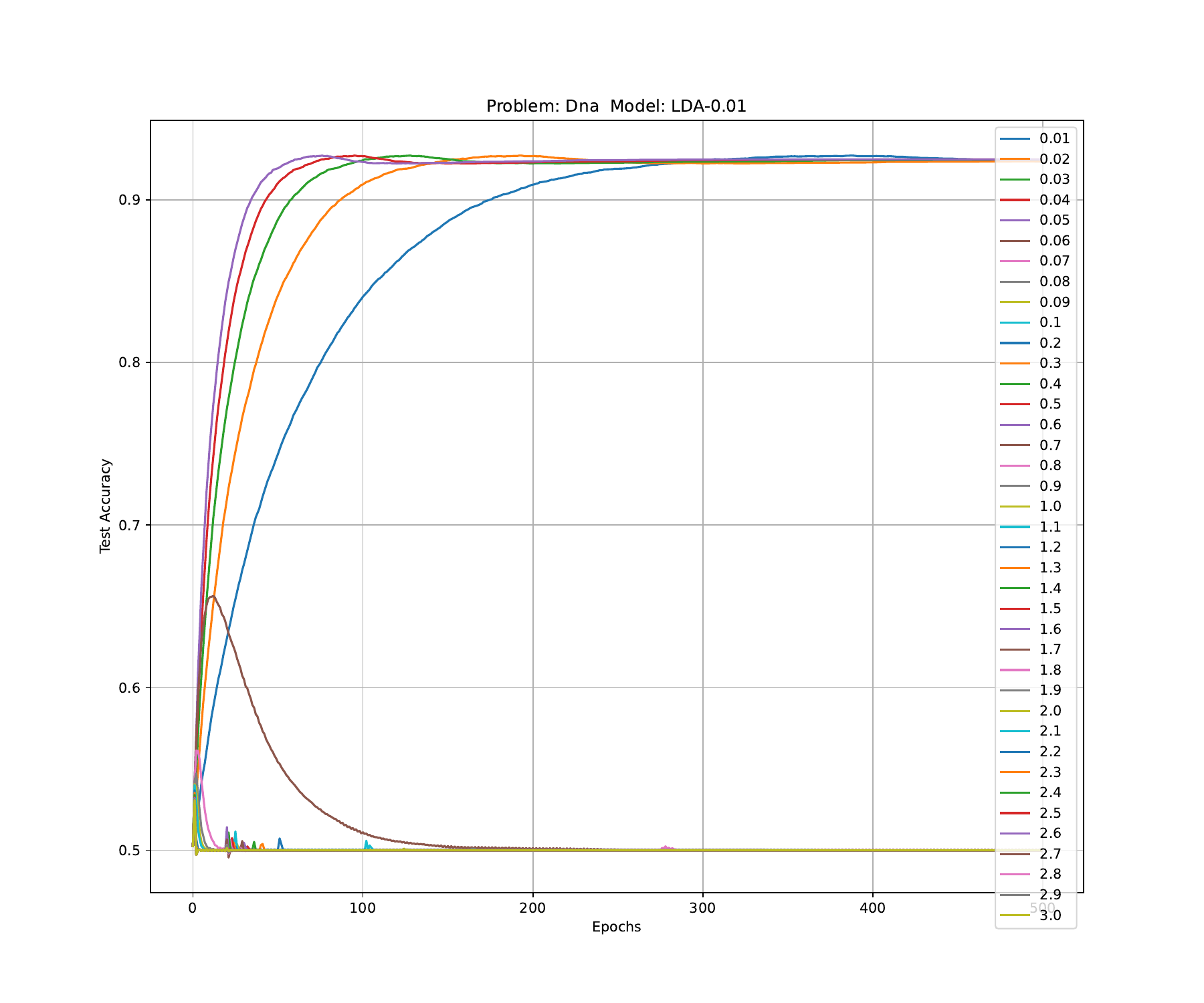} \\[-3mm]
    \includegraphics[clip,trim=70 30 30 30,width=.52\linewidth]{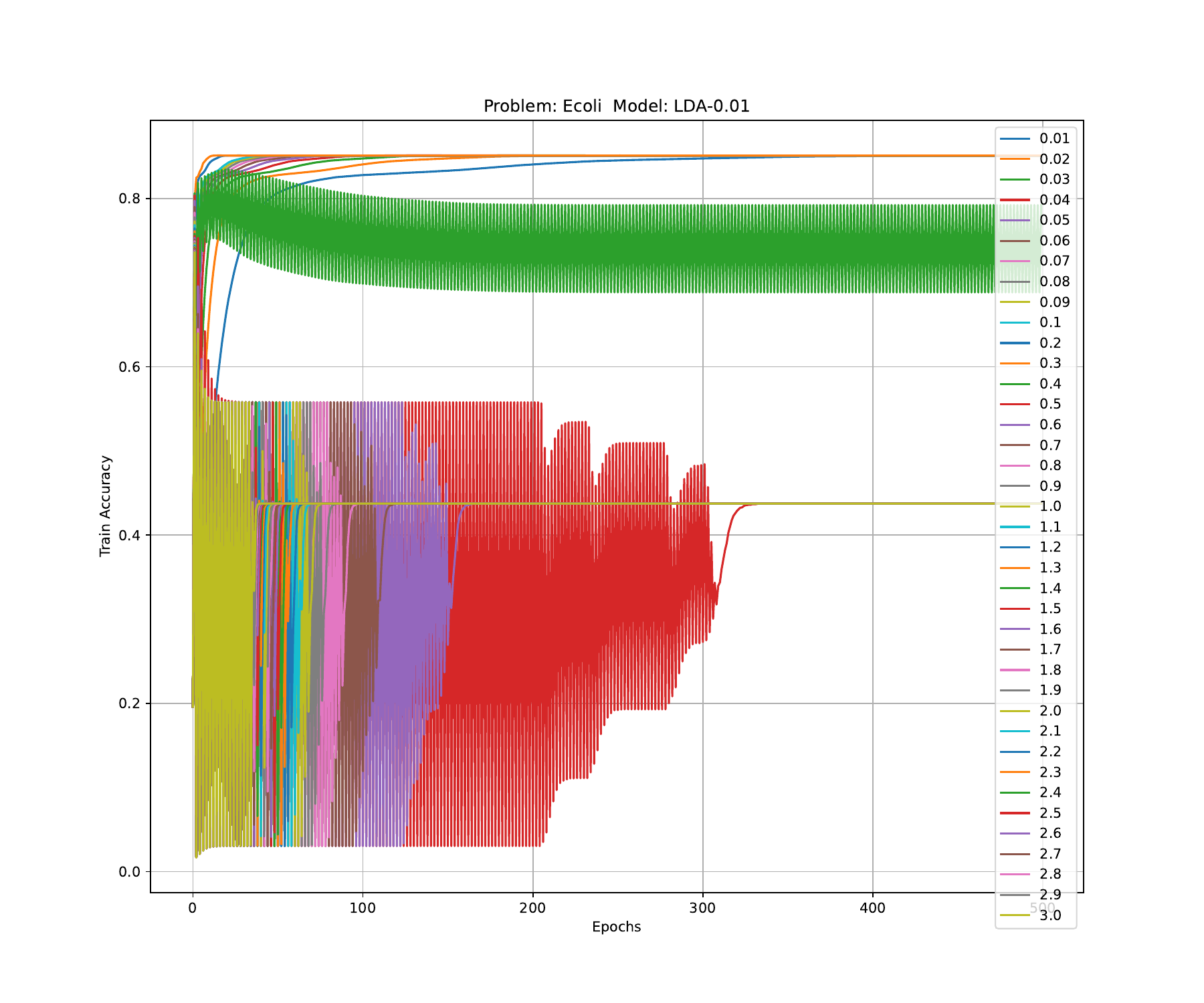} &
    \includegraphics[clip,trim=70 30 30 30,width=.52\linewidth]{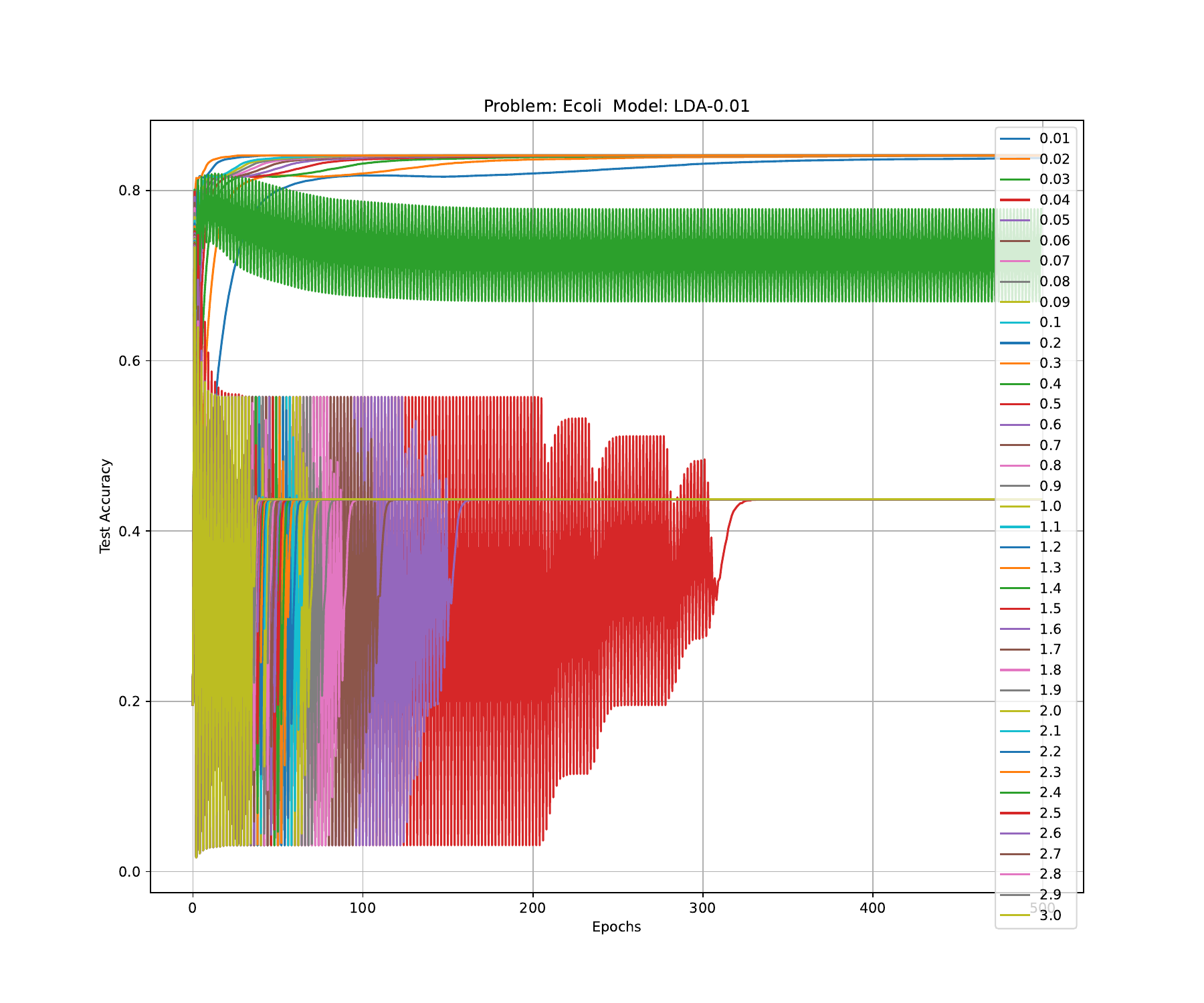} \\[-3mm]
  \end{tabular}
\caption{Accuracies for LDA-0.01 model.}
  \label{fig:accuracies_LDA_0.01_model}
\end{figure*}

\begin{figure*}[p]
  \centering
  \ContinuedFloat
  \begin{tabular}{c@{}c}
    Training Accuracy & Validation Accuracy \\
    \includegraphics[clip,trim=70 30 30 30,width=.52\linewidth]{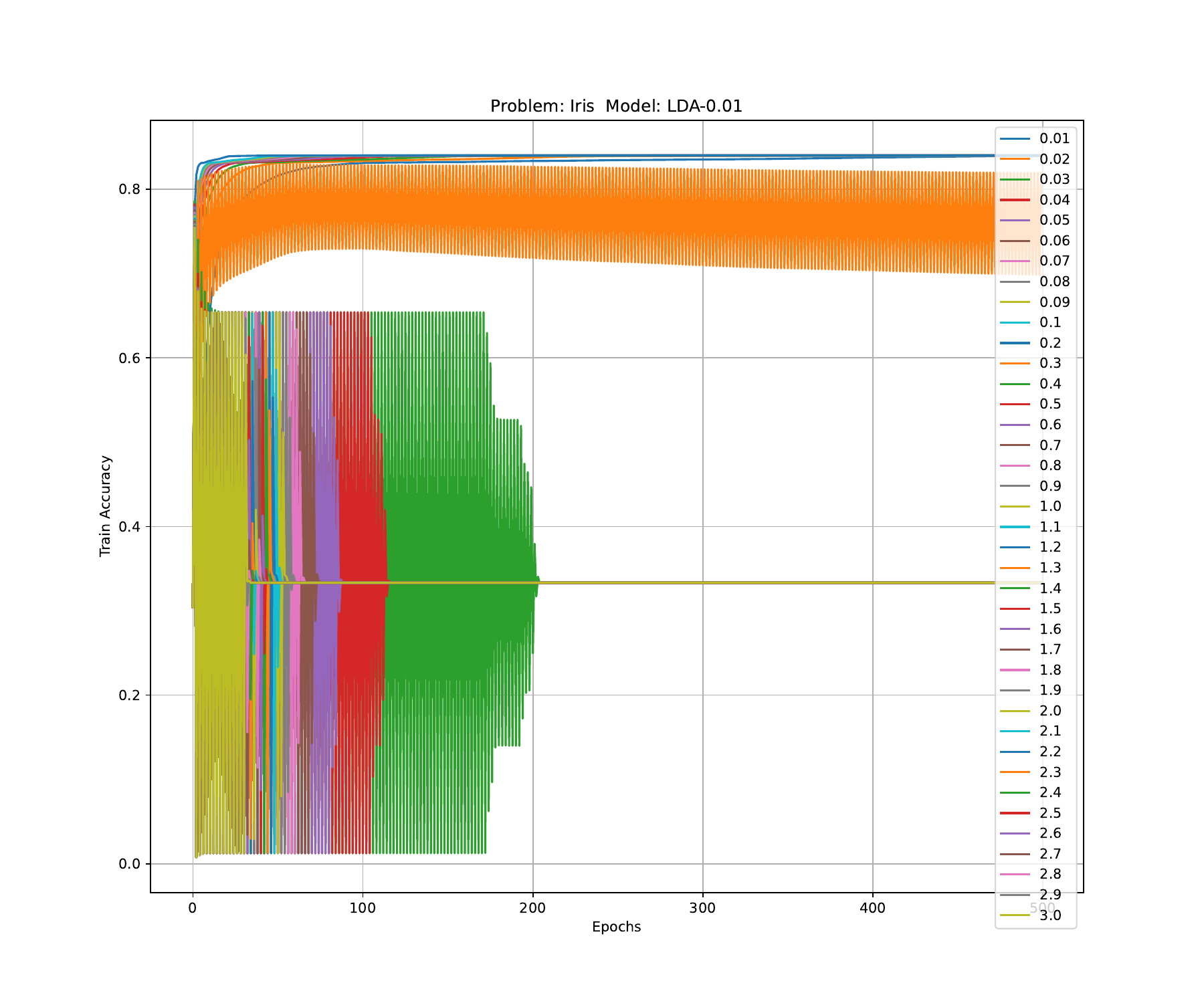} &
    \includegraphics[clip,trim=70 30 30 30,width=.52\linewidth]{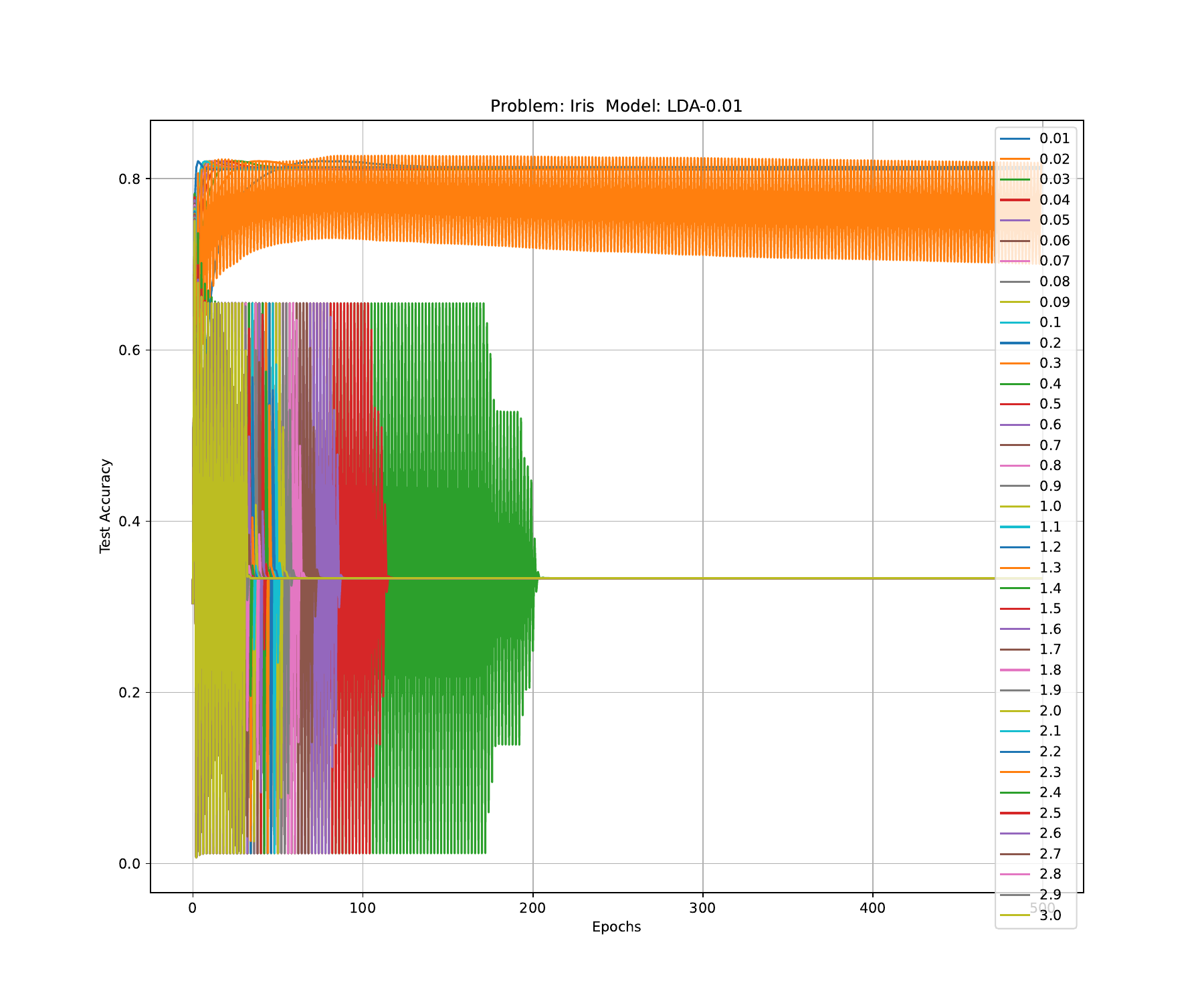} \\[-3mm]
    \includegraphics[clip,trim=70 30 30 30,width=.52\linewidth]{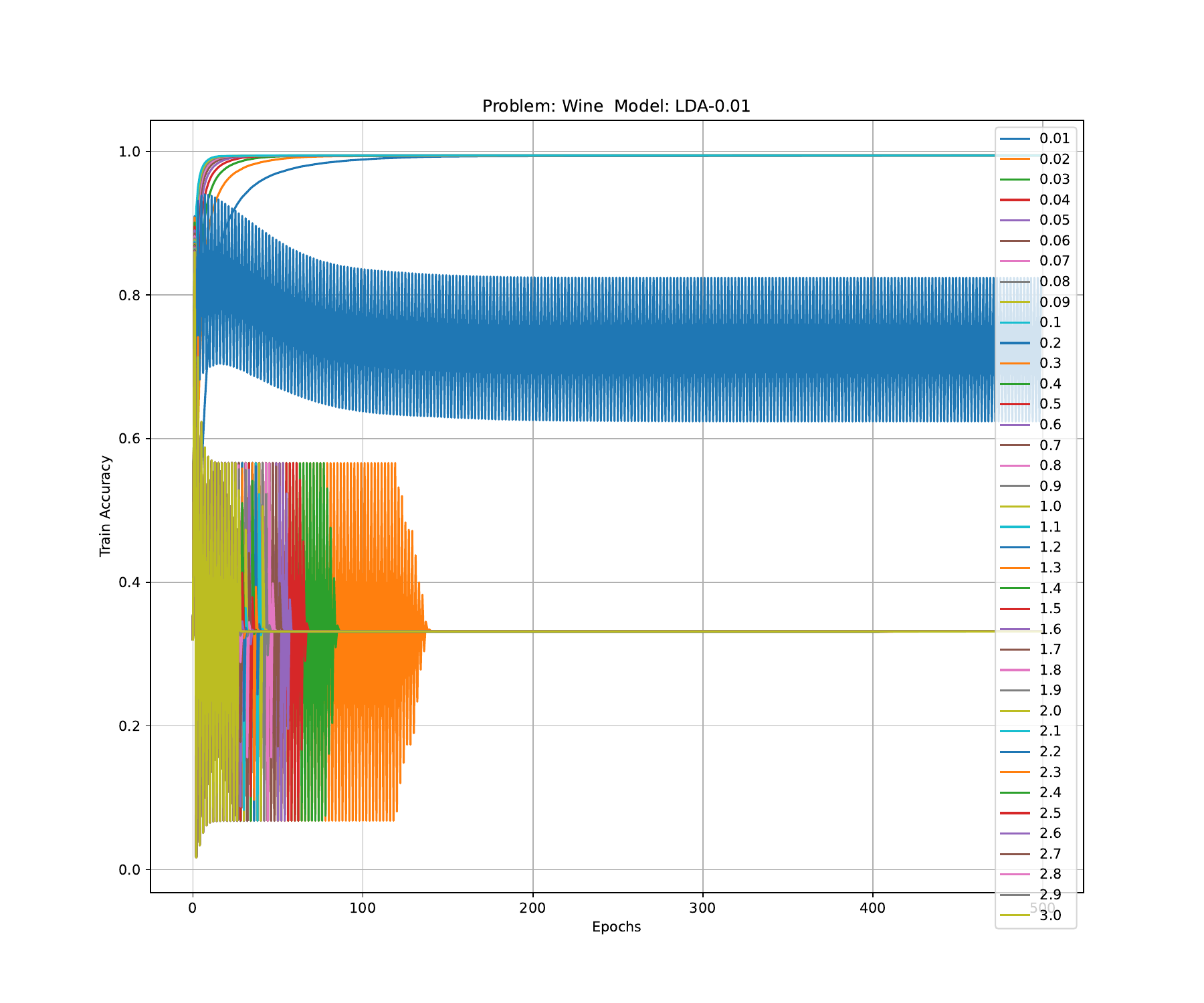} &
    \includegraphics[clip,trim=70 30 30 30,width=.52\linewidth]{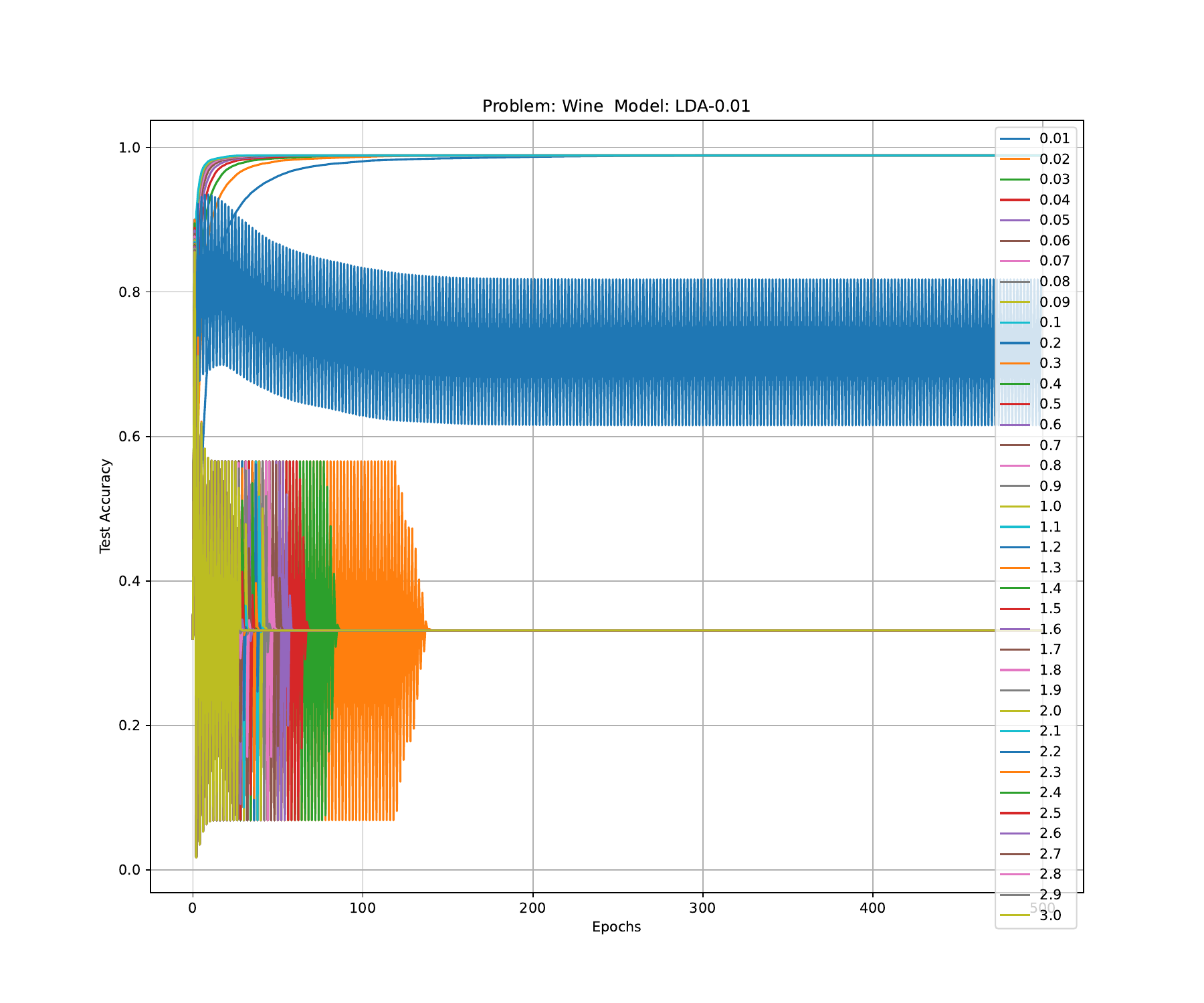} \\[-3mm]
  \end{tabular}
\caption{Accuracies for LDA-0.01 model (continued).}
\end{figure*}

\begin{figure*}[p]
  \centering
  \begin{tabular}{c@{}c}
    Training Accuracy & Validation Accuracy \\
    \includegraphics[clip,trim=70 30 30 30,width=.52\linewidth]{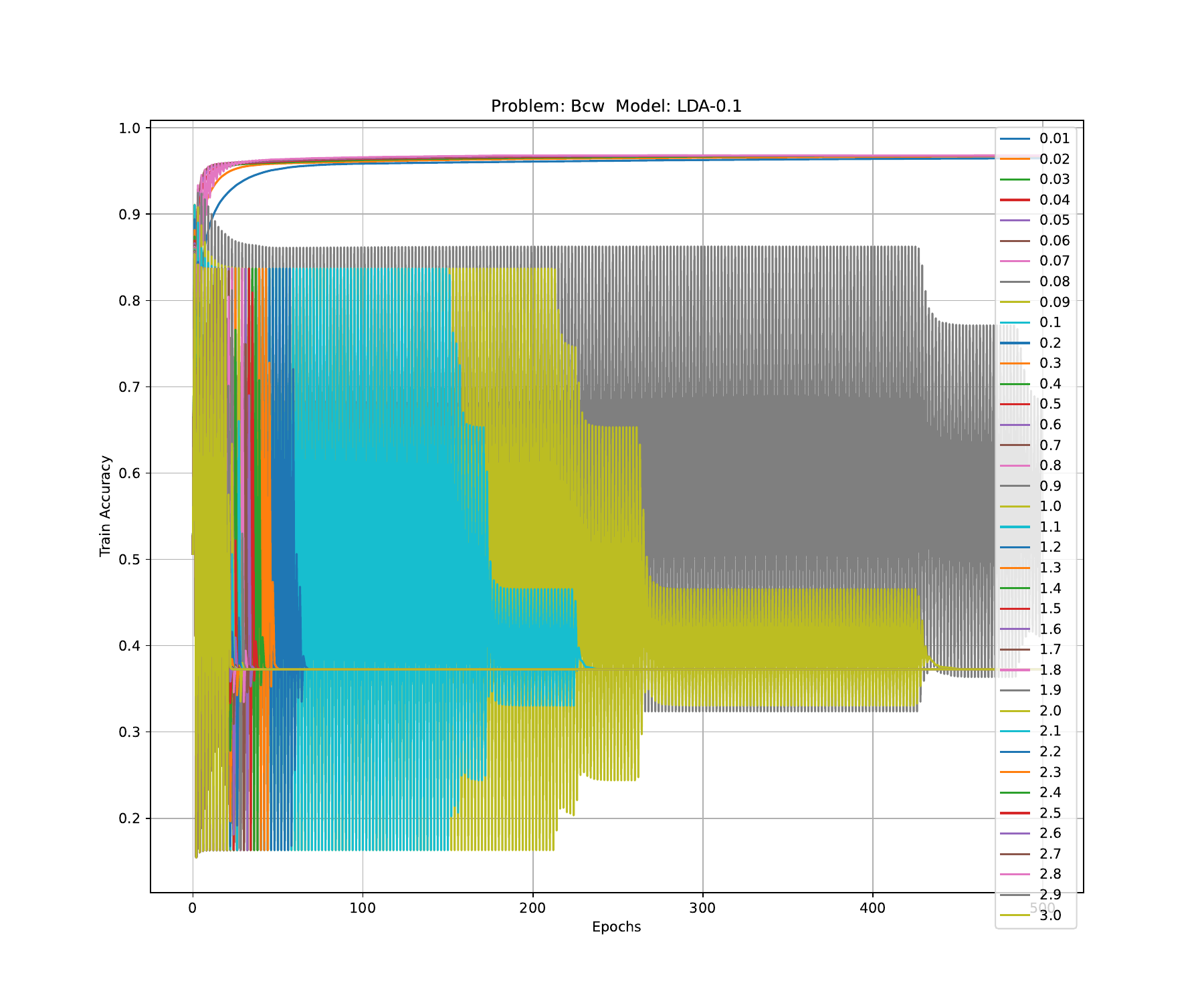} &
    \includegraphics[clip,trim=70 30 30 30,width=.52\linewidth]{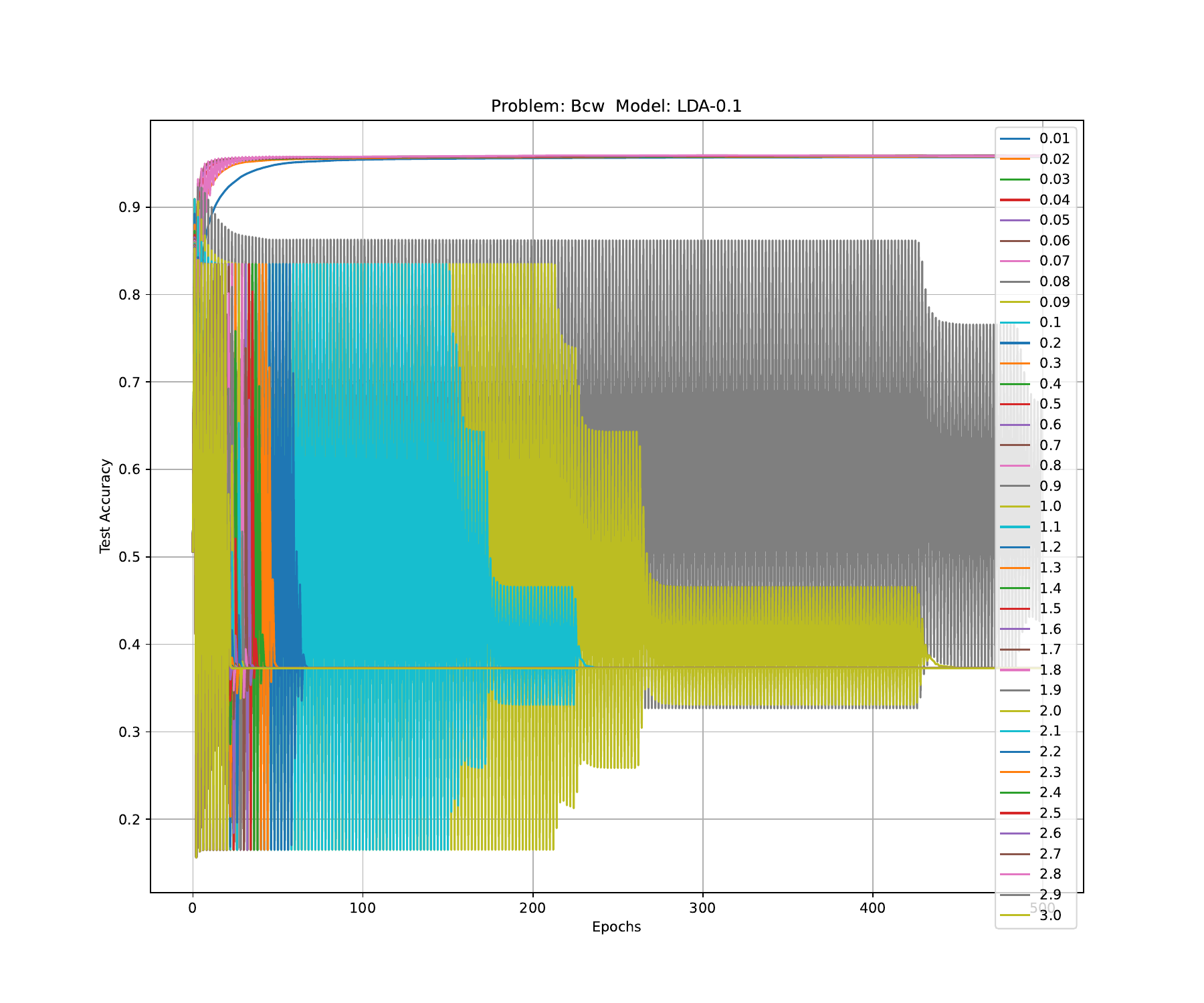} \\[-3mm]
    \includegraphics[clip,trim=70 30 30 30,width=.52\linewidth]{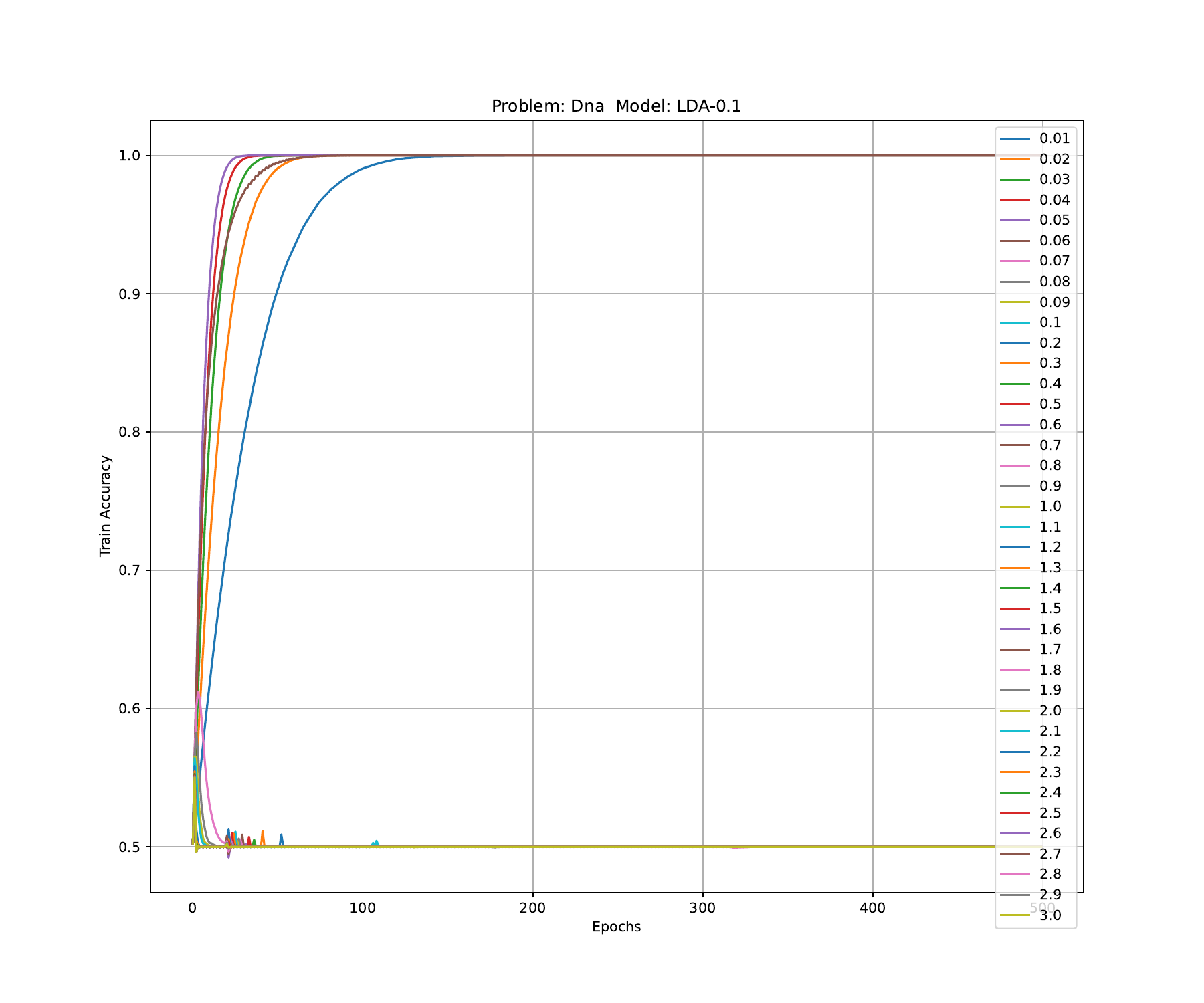} &
    \includegraphics[clip,trim=70 30 30 30,width=.52\linewidth]{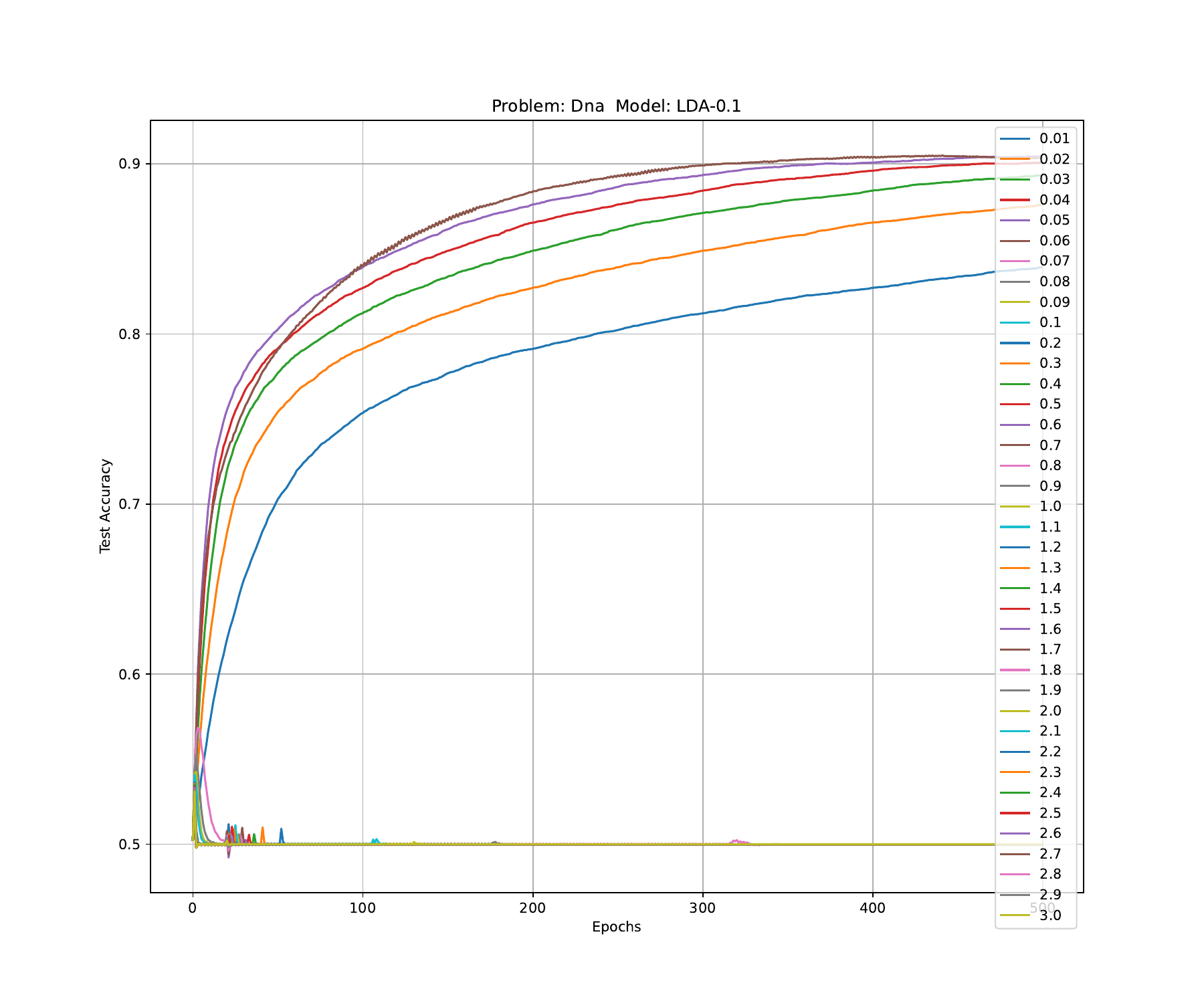} \\[-3mm]
    \includegraphics[clip,trim=70 30 30 30,width=.52\linewidth]{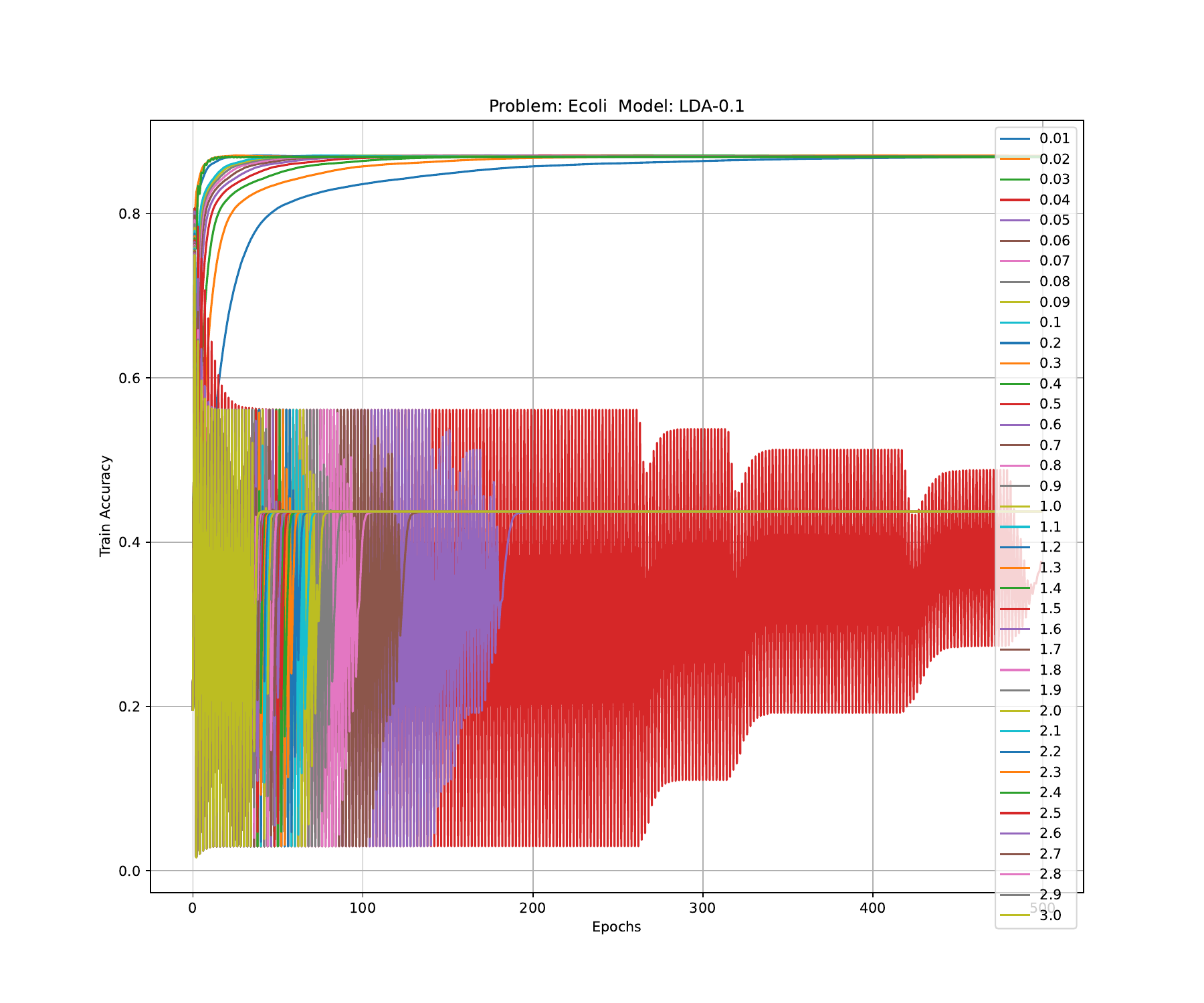} &
    \includegraphics[clip,trim=70 30 30 30,width=.52\linewidth]{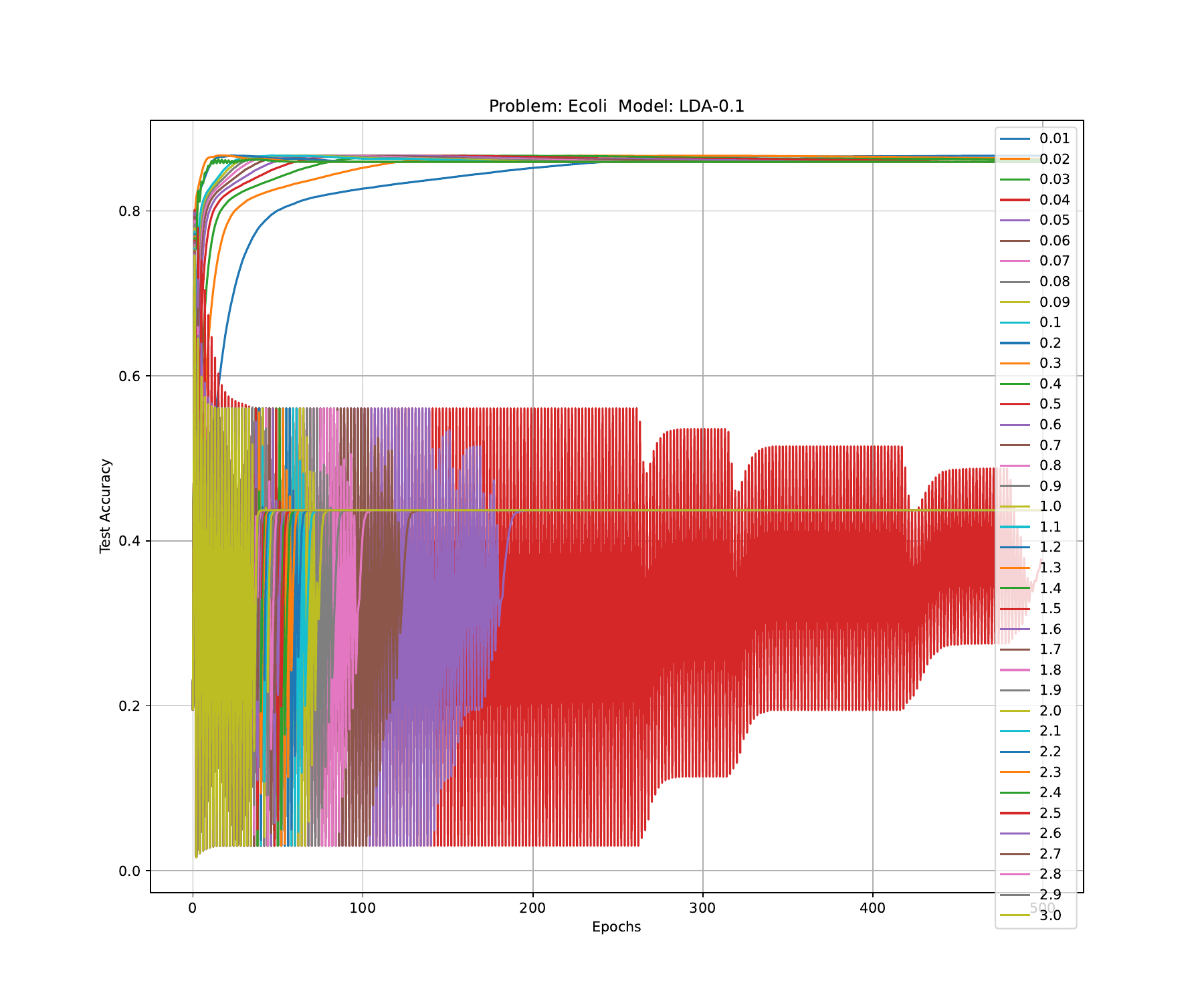} \\[-3mm]
  \end{tabular}
\caption{Accuracies for LDA-0.1 model.}
  \label{fig:accuracies_LDA_0.1_model}
\end{figure*}

\begin{figure*}[p]
  \centering
  \ContinuedFloat
  \begin{tabular}{c@{}c}
    Training Accuracy & Validation Accuracy \\
    \includegraphics[clip,trim=70 30 30 30,width=.52\linewidth]{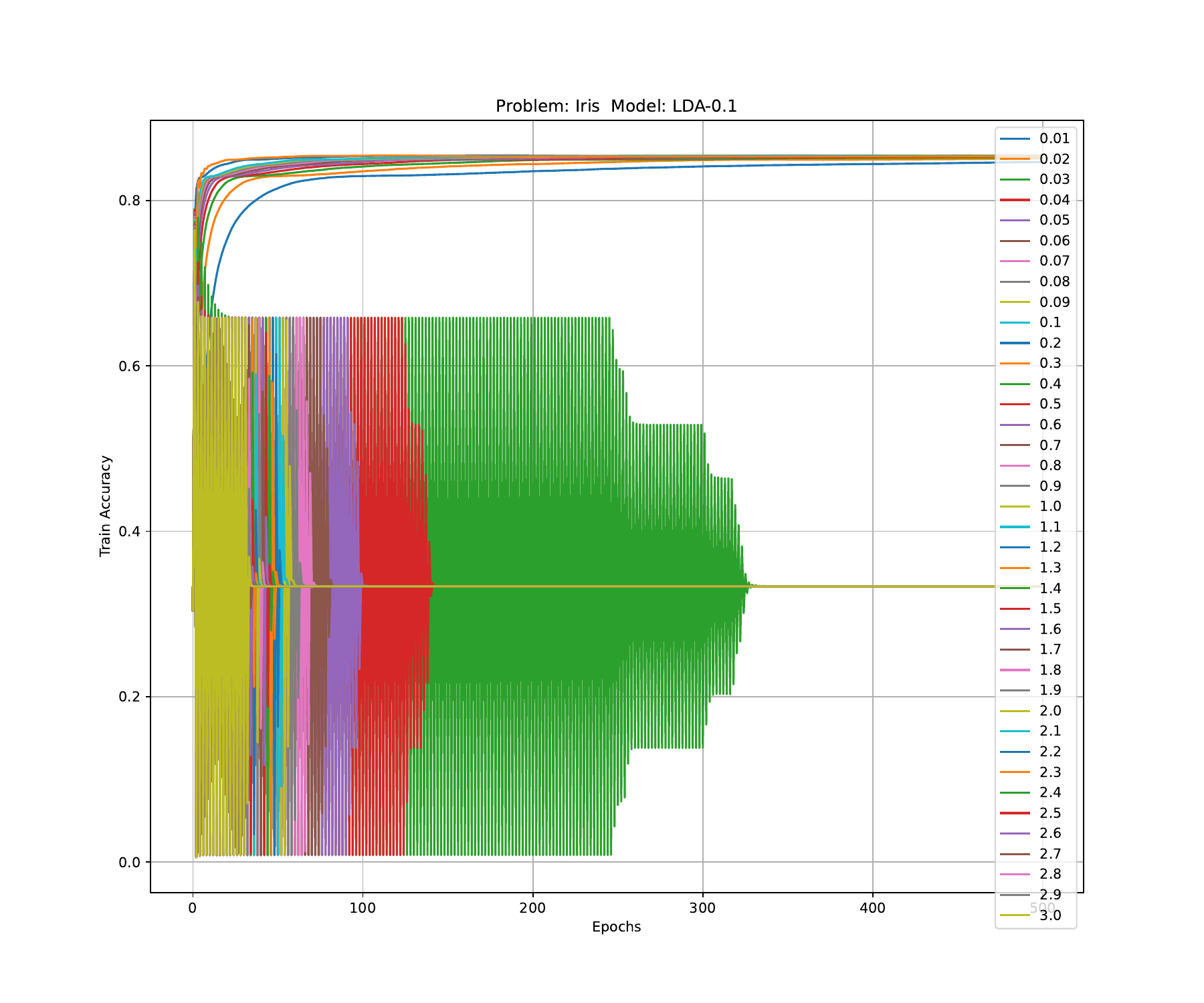} &
    \includegraphics[clip,trim=70 30 30 30,width=.52\linewidth]{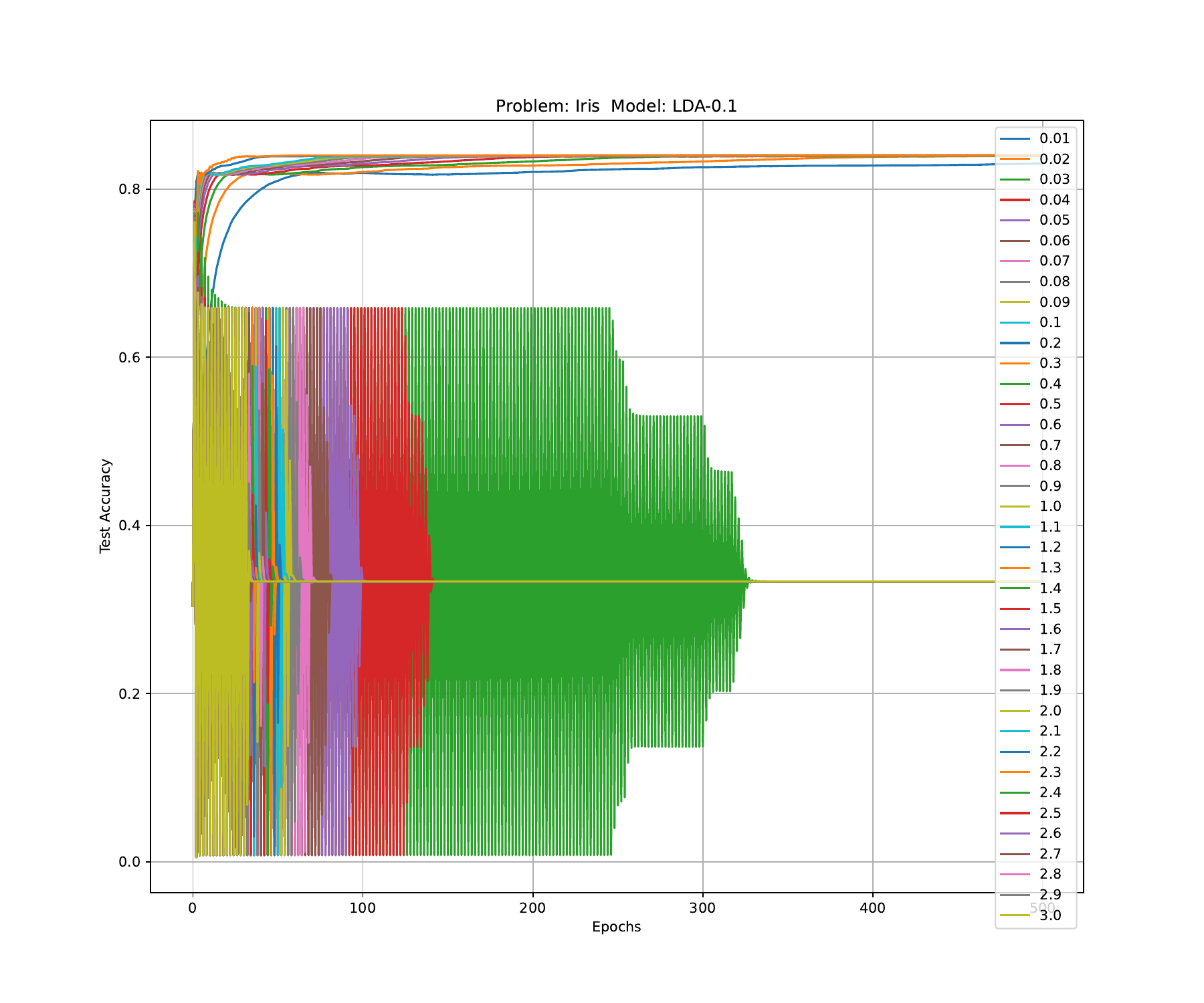} \\[-3mm]
    \includegraphics[clip,trim=70 30 30 30,width=.52\linewidth]{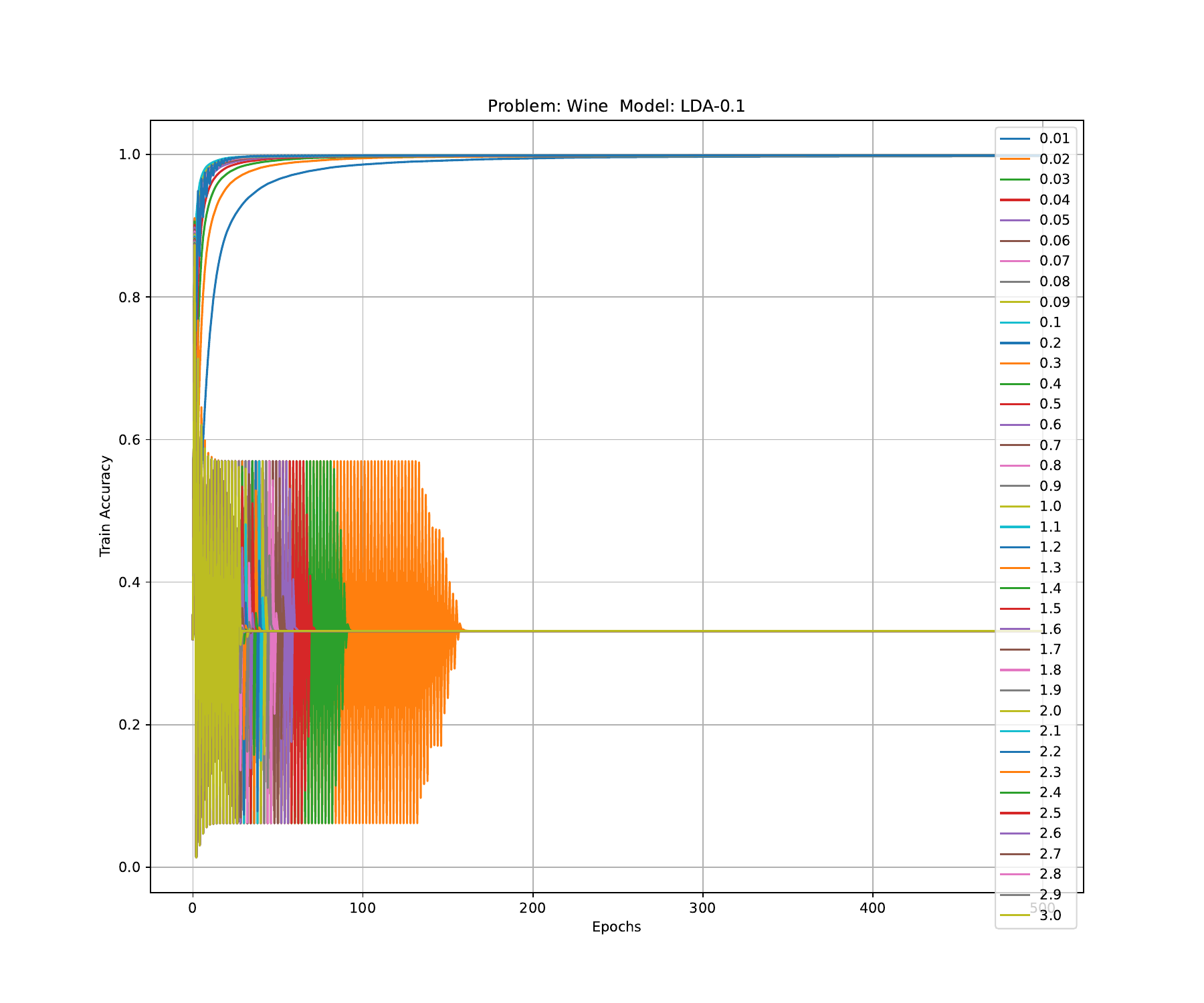} &
    \includegraphics[clip,trim=70 30 30 30,width=.52\linewidth]{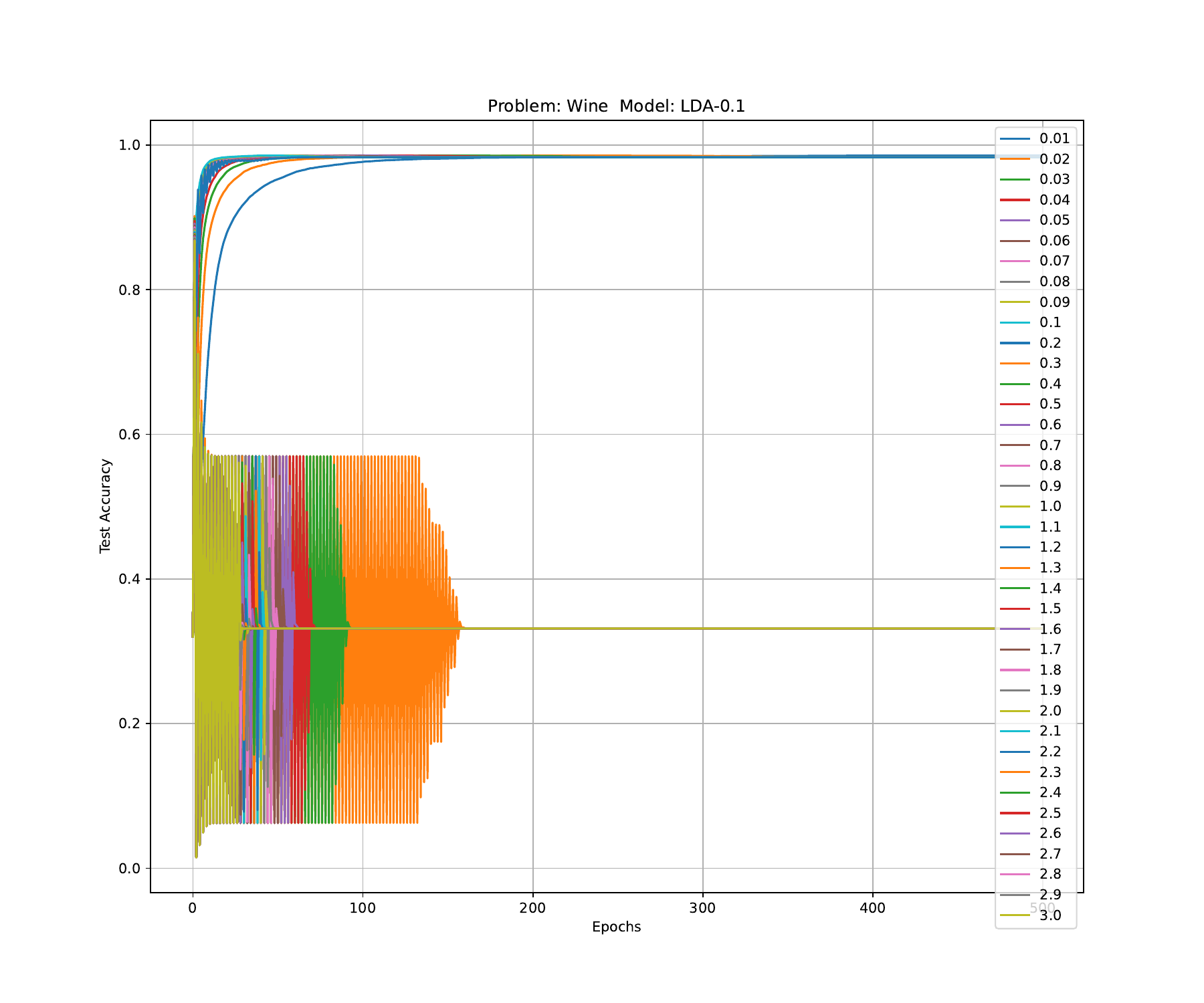} \\[-3mm]
  \end{tabular}
\caption{Accuracies for LDA-0.1 model (continued).}
\end{figure*}

\begin{figure*}[h!]
  \centering
  \begin{tabular}{c@{}c}
    Training Accuracy & Validation Accuracy \\
    \includegraphics[clip,trim=70 30 30 30,width=.52\linewidth]{arXiv-figures/bcw_LR-inf_average_accuracy_train.pdf} &
    \includegraphics[clip,trim=70 30 30 30,width=.52\linewidth]{arXiv-figures/bcw_LR-inf_average_accuracy_test.pdf} \\[-3mm]
    \includegraphics[clip,trim=70 30 30 30,width=.52\linewidth]{arXiv-figures/dna_LR-inf_average_accuracy_train.pdf} &
    \includegraphics[clip,trim=70 30 30 30,width=.52\linewidth]{arXiv-figures/dna_LR-inf_average_accuracy_test.pdf} \\[-3mm]
    \includegraphics[clip,trim=70 30 30 30,width=.52\linewidth]{arXiv-figures/ecoli_LR-inf_average_accuracy_train.pdf} &
    \includegraphics[clip,trim=70 30 30 30,width=.52\linewidth]{arXiv-figures/ecoli_LR-inf_average_accuracy_test.pdf} \\[-3mm]
  \end{tabular}
  \caption{Accuracies for LR model without regularisation.}
  \label{fig:accuracies_LR_model_duplicate}
\end{figure*}

\begin{figure*}[h!]
  \centering
  \ContinuedFloat
  \begin{tabular}{c@{}c}
    Training Accuracy & Validation Accuracy \\
    \includegraphics[clip,trim=70 30 30 30,width=.52\linewidth]{arXiv-figures/iris_LR-inf_average_accuracy_train.pdf} &
    \includegraphics[clip,trim=70 30 30 30,width=.52\linewidth]{arXiv-figures/iris_LR-inf_average_accuracy_test.pdf} \\[-3mm]
    \includegraphics[clip,trim=70 30 30 30,width=.52\linewidth]{arXiv-figures/wine_LR-inf_average_accuracy_train.pdf} &
    \includegraphics[clip,trim=70 30 30 30,width=.52\linewidth]{arXiv-figures/wine_LR-inf_average_accuracy_test.pdf} \\[-3mm]
  \end{tabular}
  \caption{Accuracies for LR model without regularisation (continued).}
\end{figure*}

\begin{figure*}[p]
  \centering
  \begin{tabular}{c@{}c}
    Training Accuracy & Validation Accuracy \\
    \includegraphics[clip,trim=70 30 30 30,width=.52\linewidth]{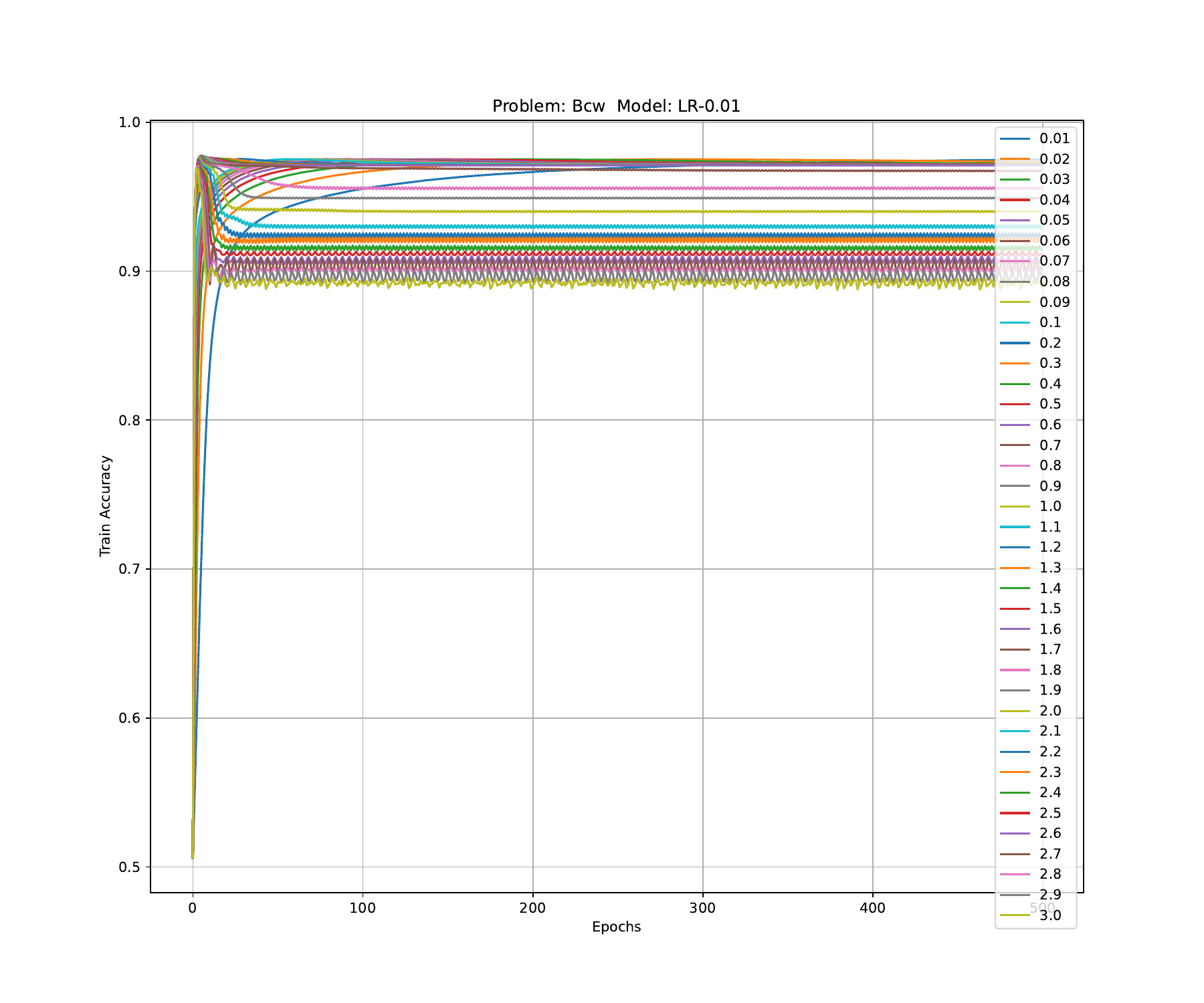} &
    \includegraphics[clip,trim=70 30 30 30,width=.52\linewidth]{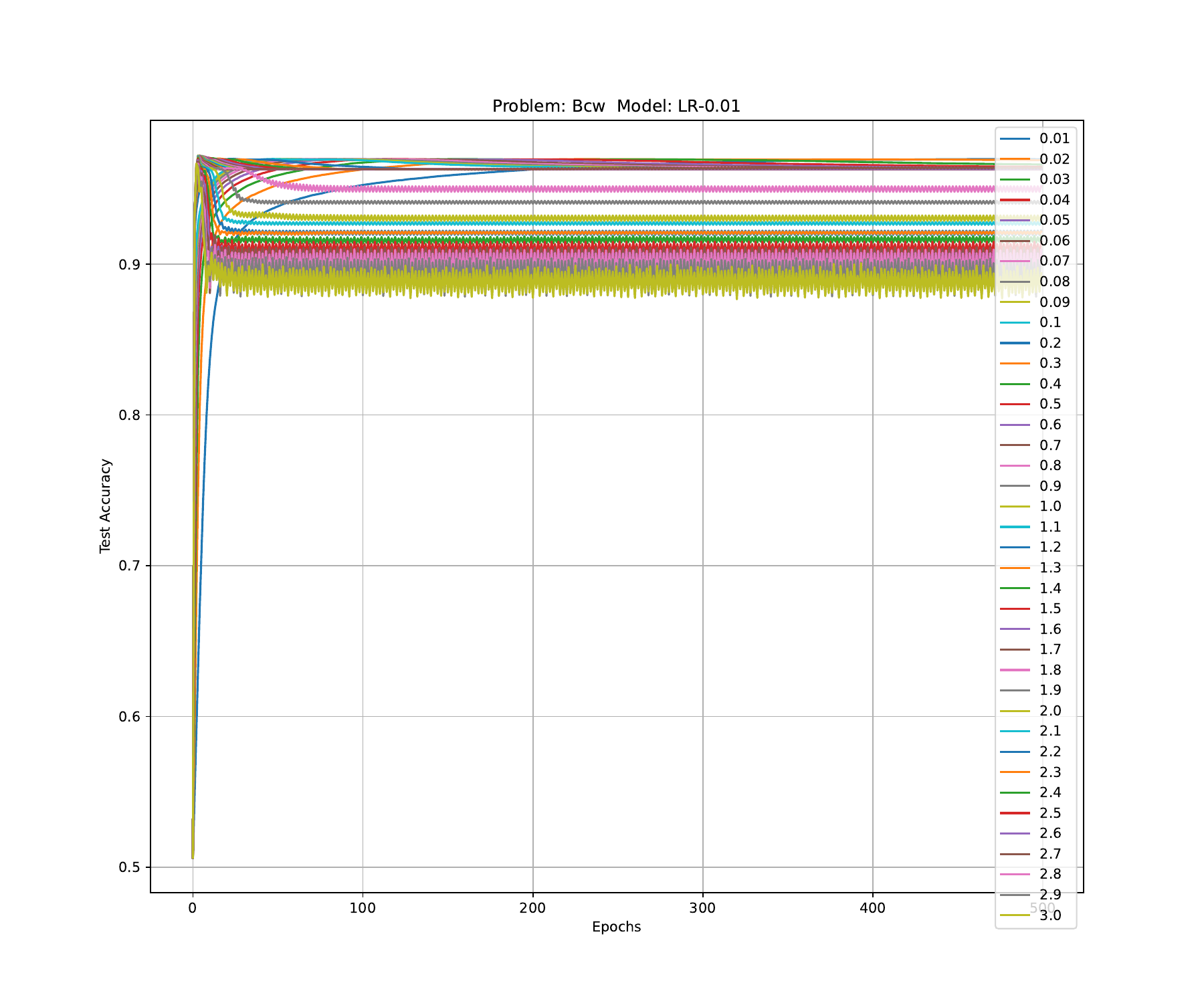} \\[-3mm]
    \includegraphics[clip,trim=70 30 30 30,width=.52\linewidth]{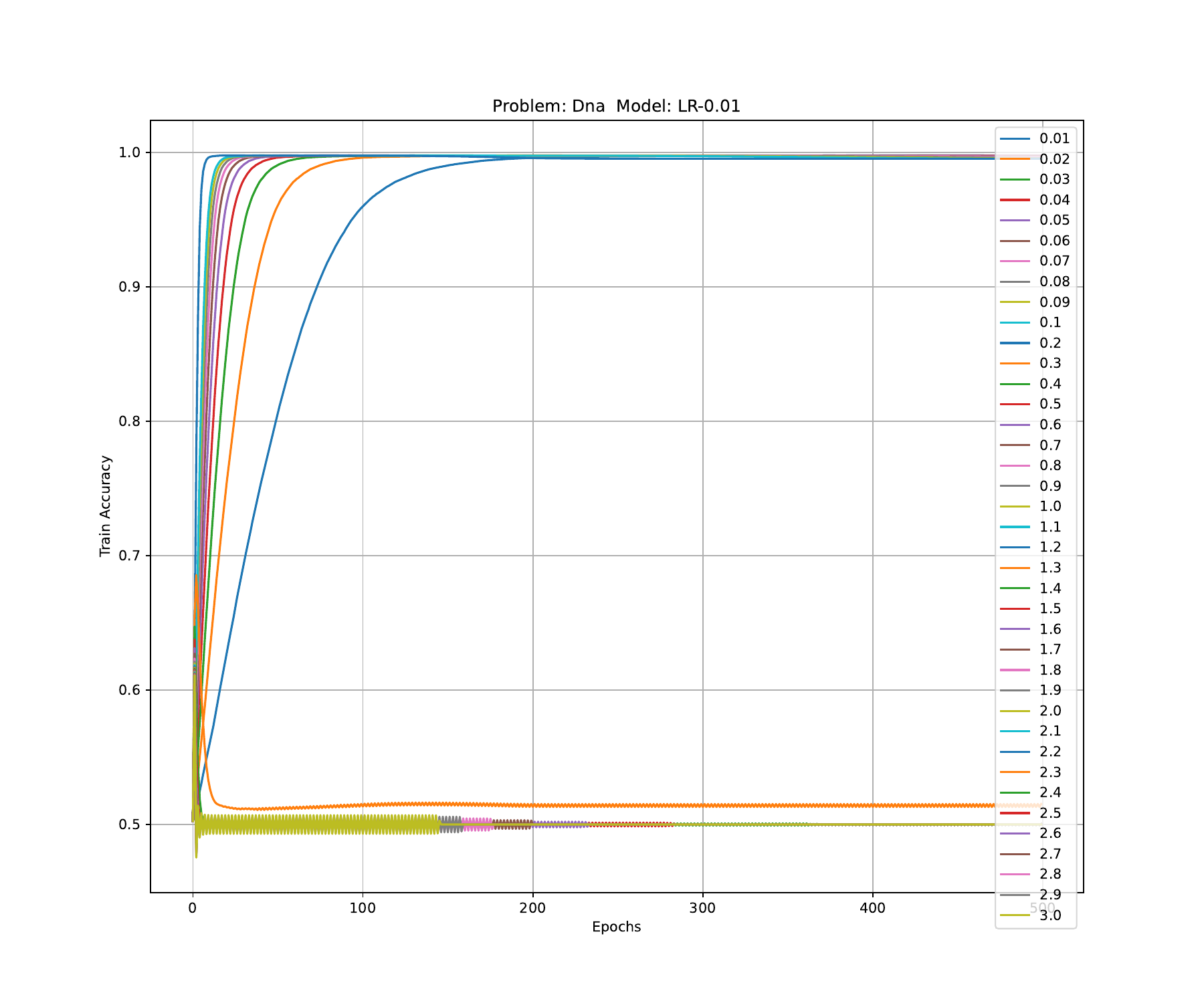} &
    \includegraphics[clip,trim=70 30 30 30,width=.52\linewidth]{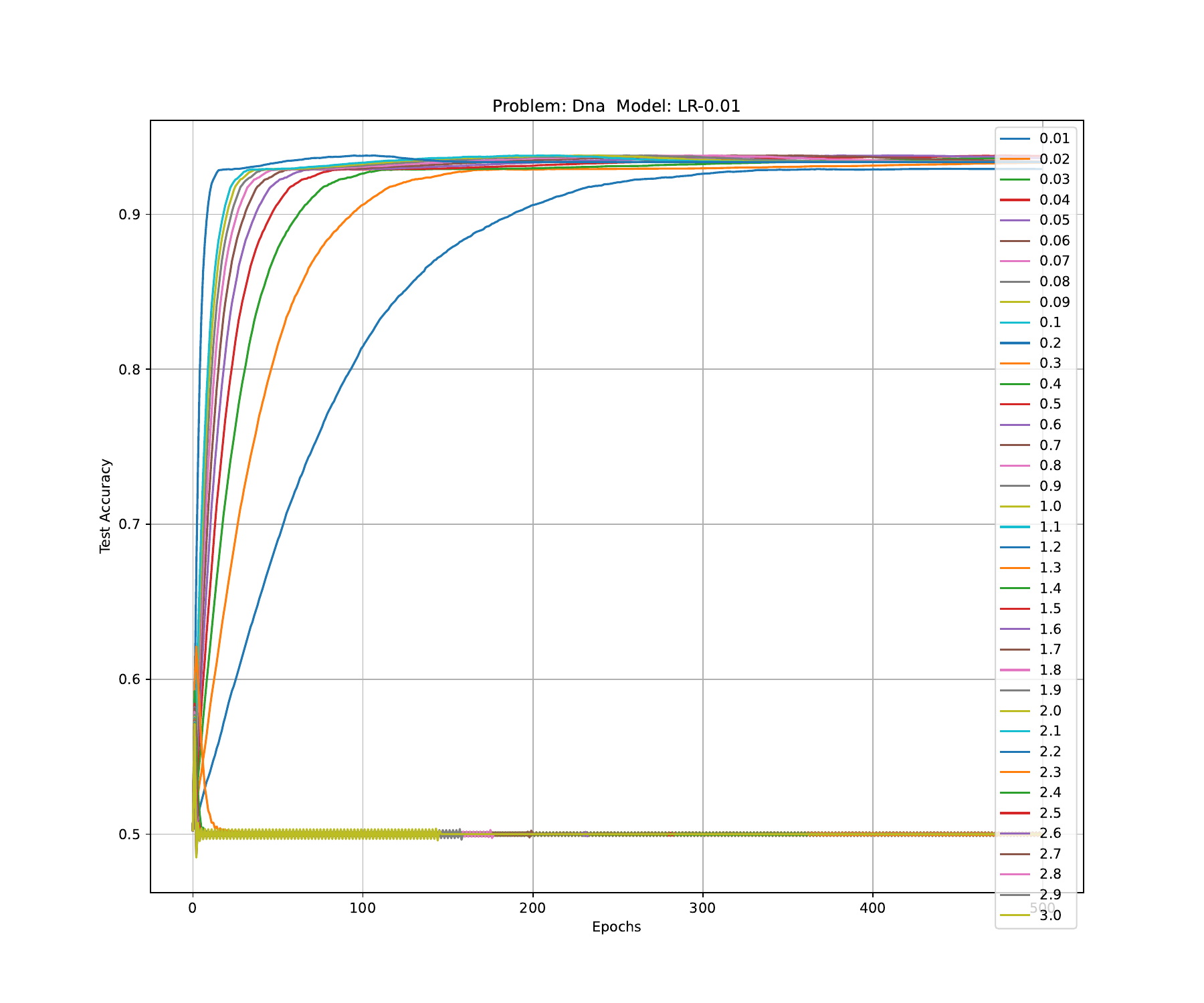} \\[-3mm]
    \includegraphics[clip,trim=70 30 30 30,width=.52\linewidth]{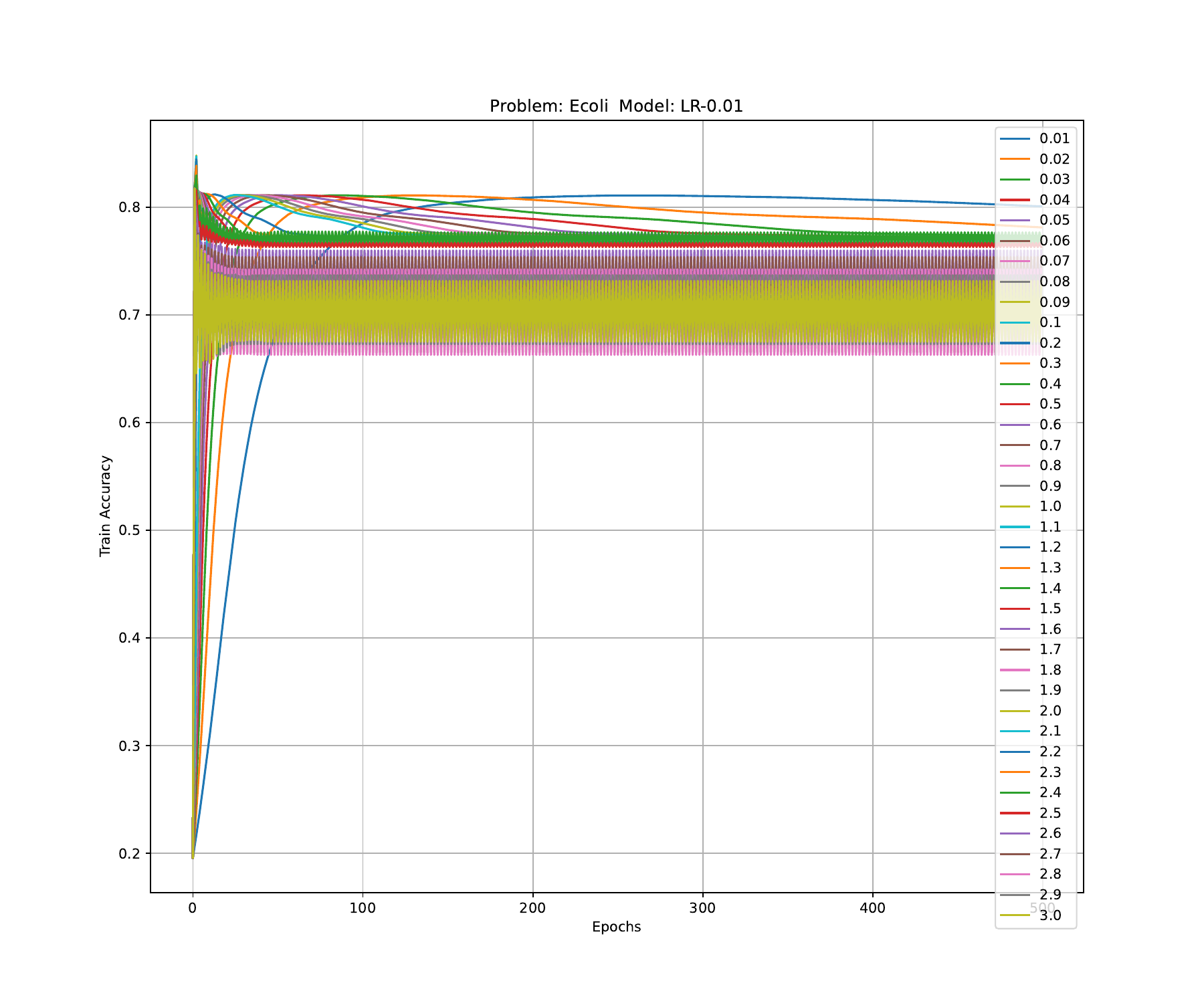} &
    \includegraphics[clip,trim=70 30 30 30,width=.52\linewidth]{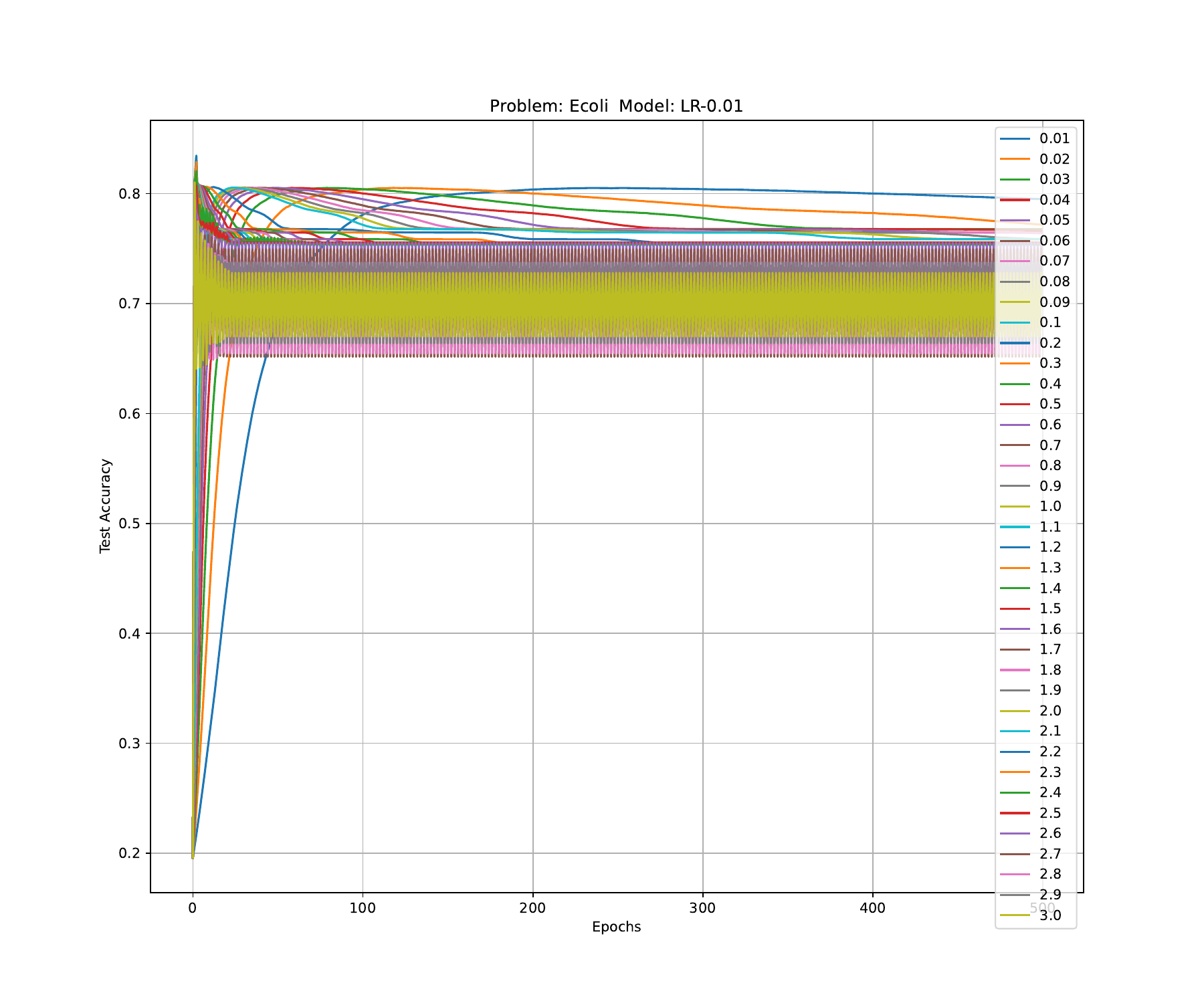} \\[-3mm]
  \end{tabular}
\caption{Accuracies for LR-0.01 model.}
  \label{fig:accuracies_LR_0.01_model}
\end{figure*}

\begin{figure*}[p]
  \centering
  \ContinuedFloat
  \begin{tabular}{c@{}c}
    Training Accuracy & Validation Accuracy \\
    \includegraphics[clip,trim=70 30 30 30,width=.52\linewidth]{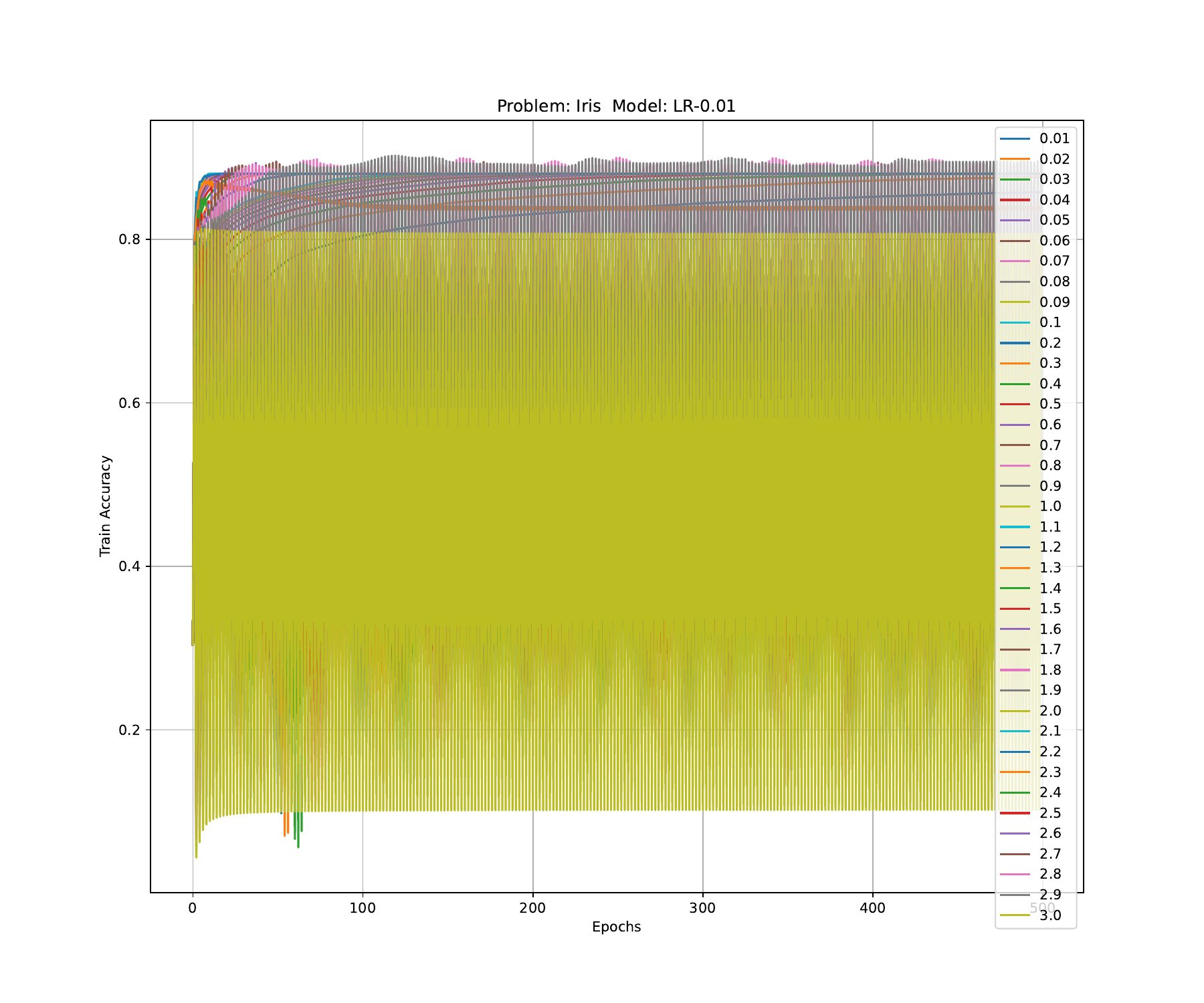} &
    \includegraphics[clip,trim=70 30 30 30,width=.52\linewidth]{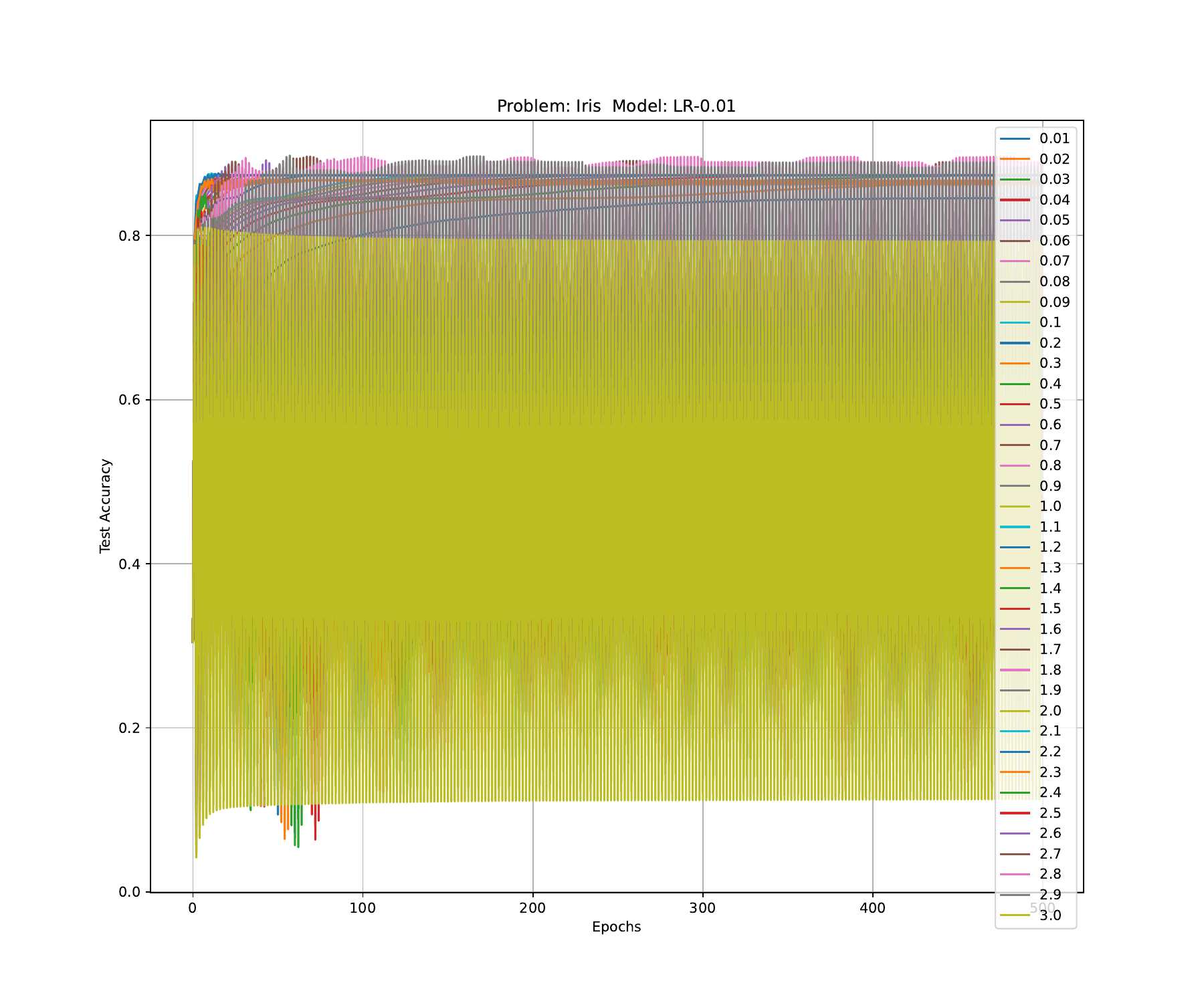} \\[-3mm]
    \includegraphics[clip,trim=70 30 30 30,width=.52\linewidth]{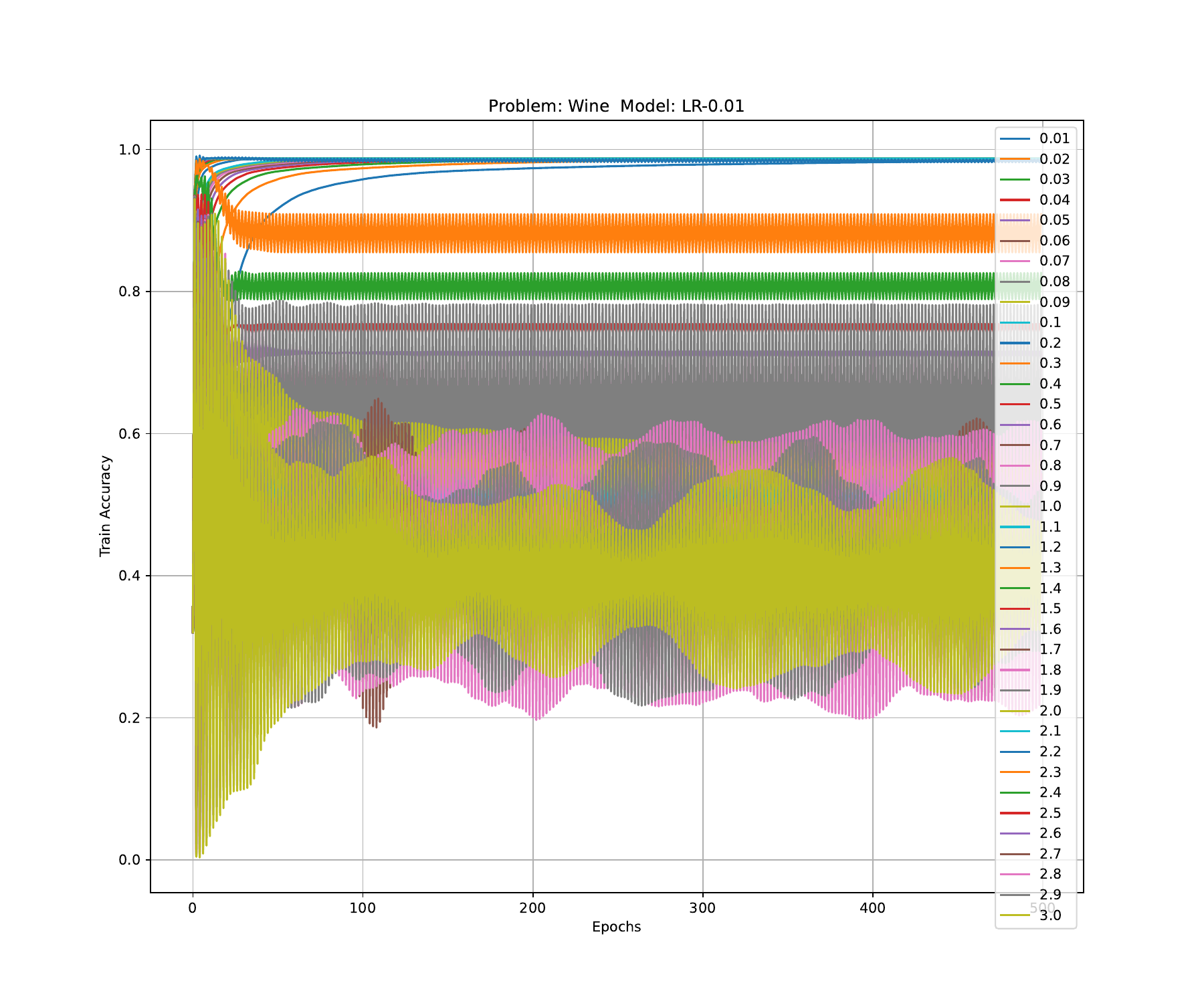} &
    \includegraphics[clip,trim=70 30 30 30,width=.52\linewidth]{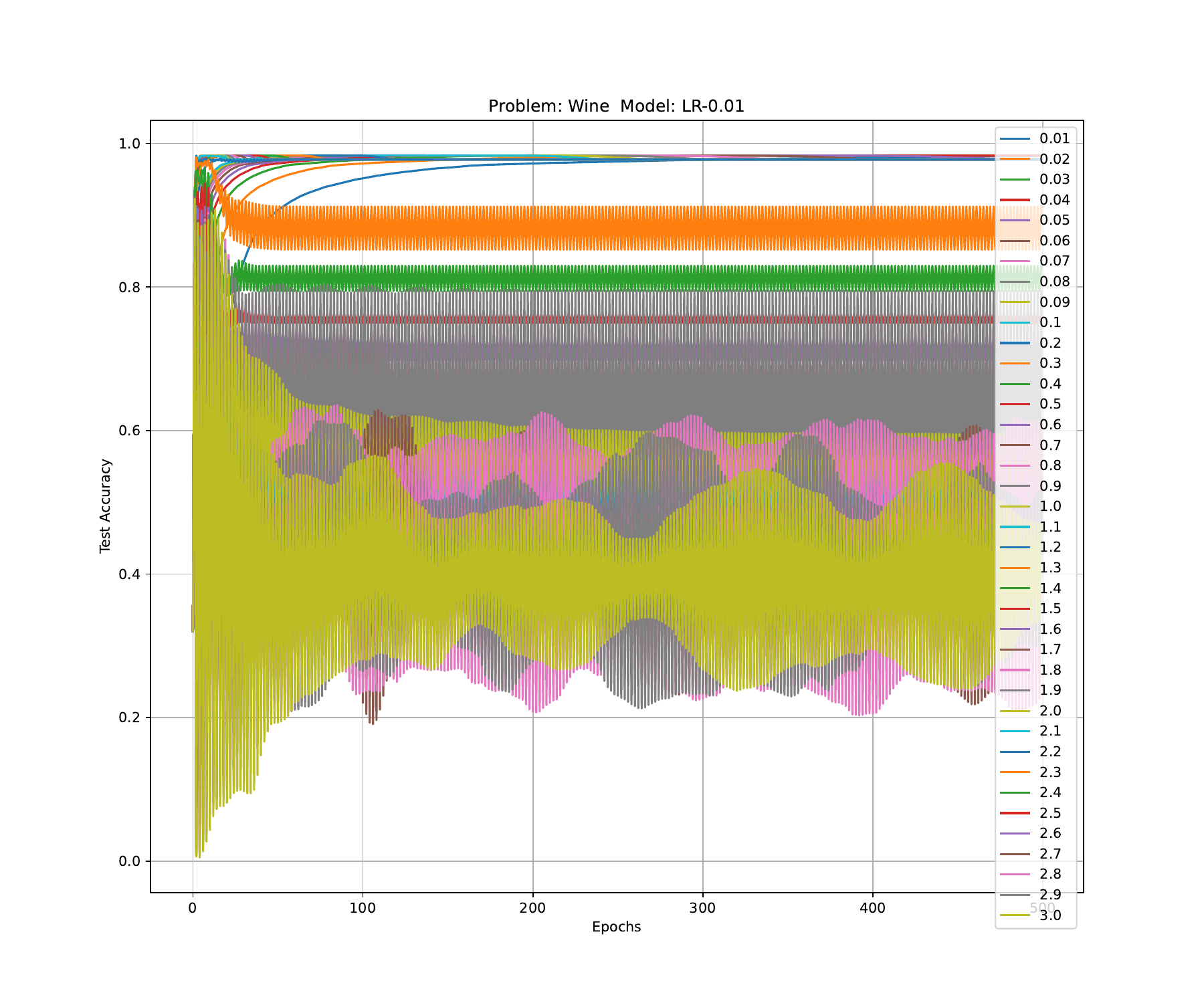} \\[-3mm]
  \end{tabular}
\caption{Accuracies for LR-0.01 model (continued).}
\end{figure*}

\begin{figure*}[p]
  \centering
  \begin{tabular}{c@{}c}
    Training Accuracy & Validation Accuracy \\
    \includegraphics[clip,trim=70 30 30 30,width=.52\linewidth]{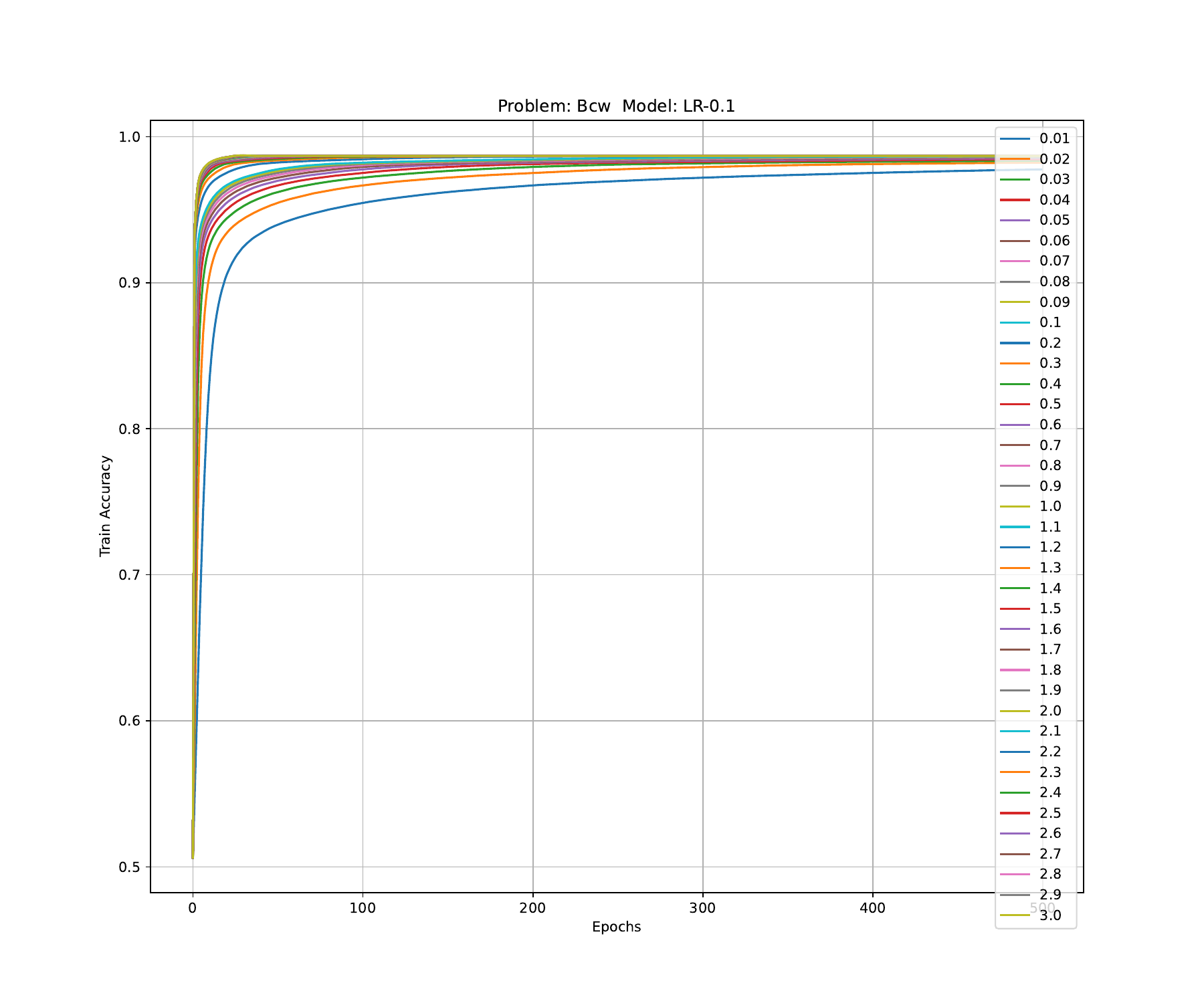} &
    \includegraphics[clip,trim=70 30 30 30,width=.52\linewidth]{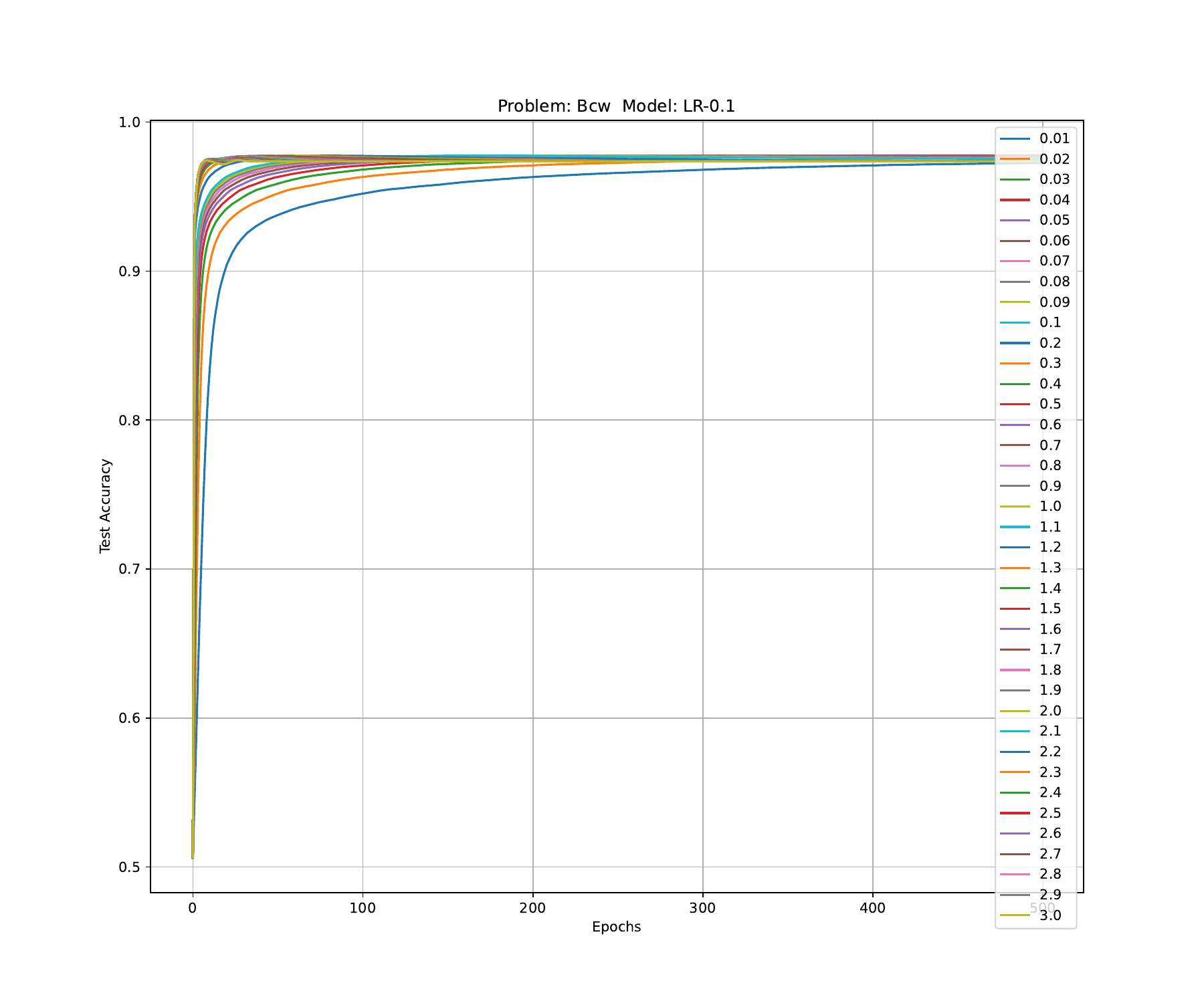} \\[-3mm]
    \includegraphics[clip,trim=70 30 30 30,width=.52\linewidth]{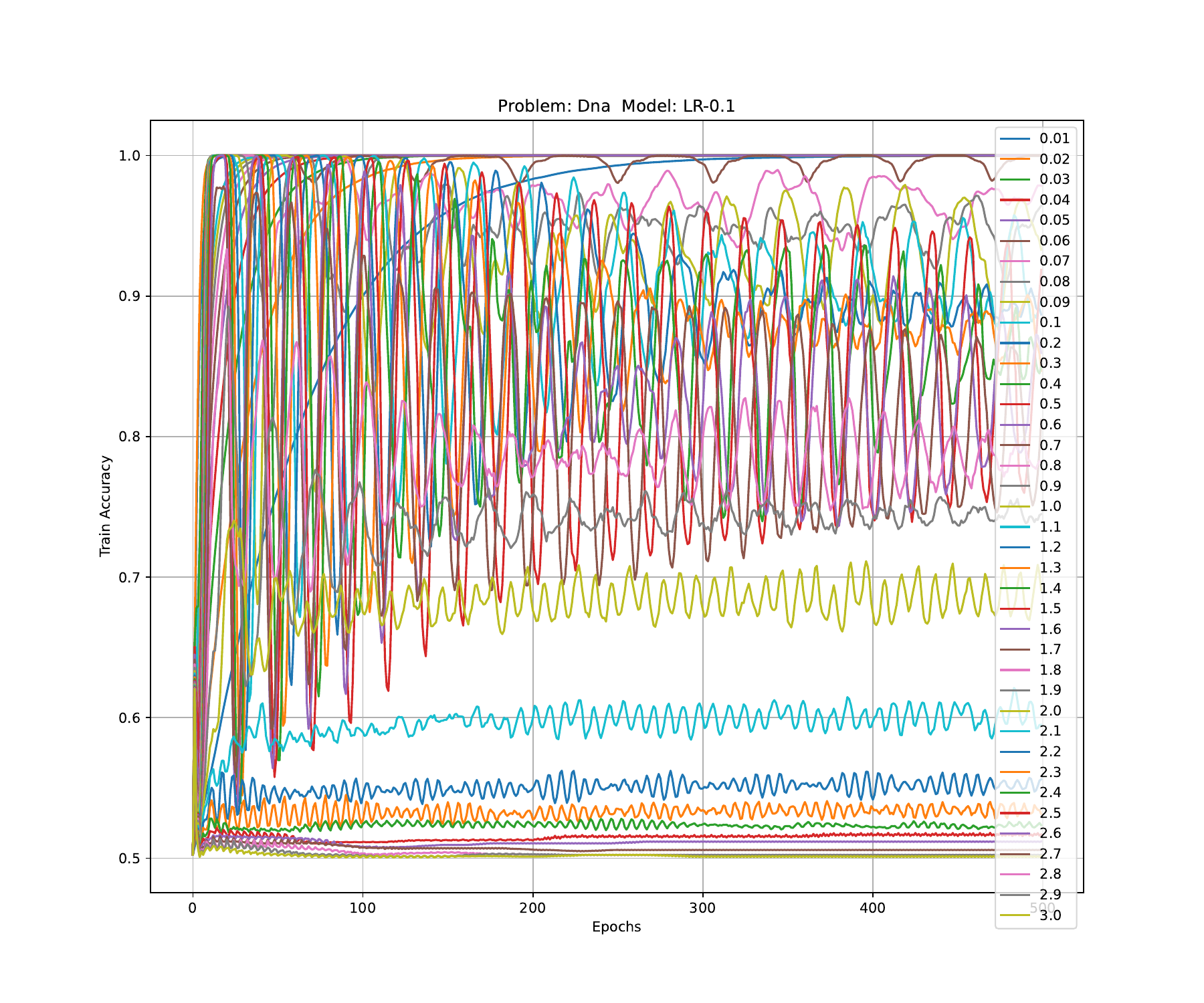} &
    \includegraphics[clip,trim=70 30 30 30,width=.52\linewidth]{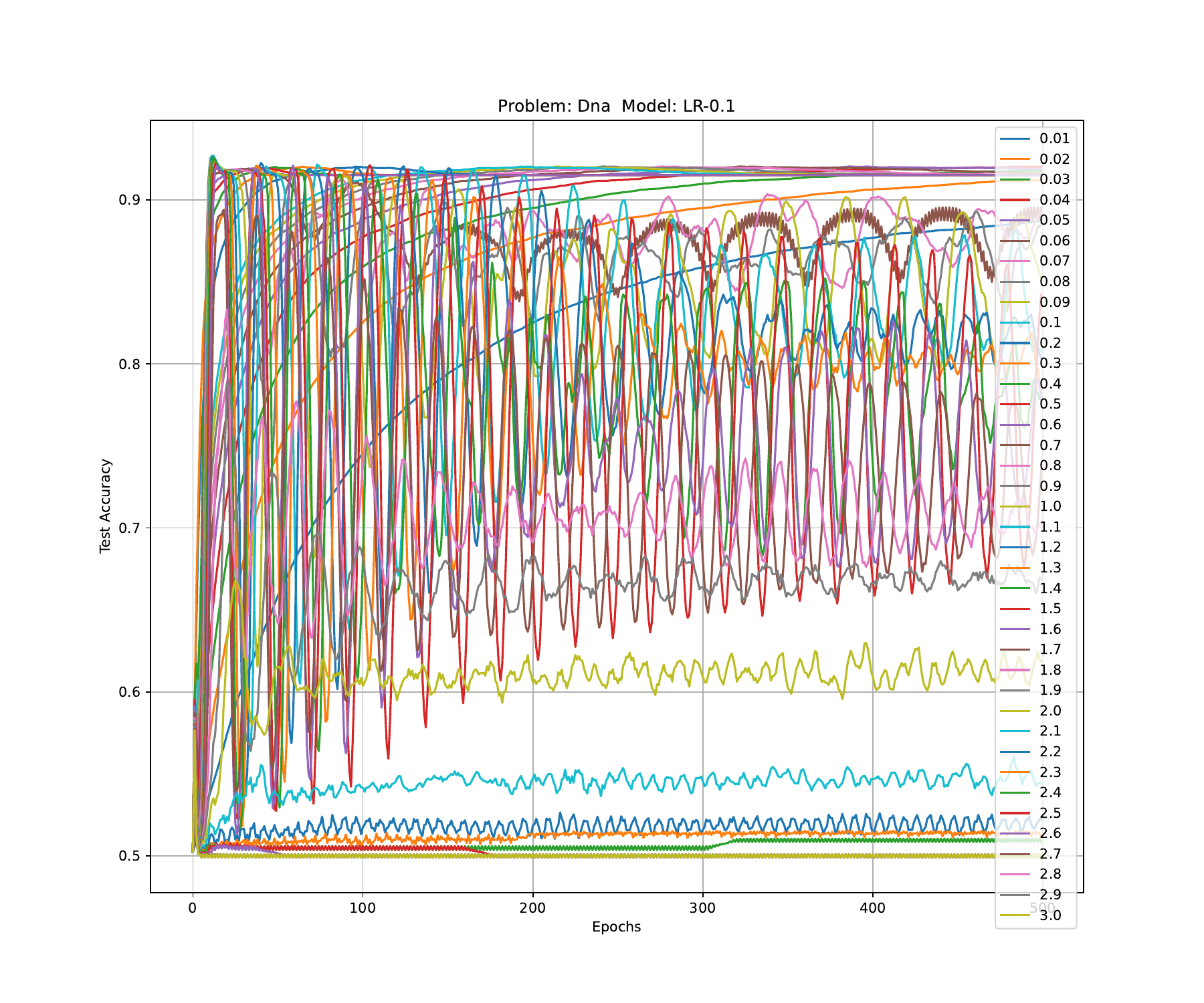} \\[-3mm]
    \includegraphics[clip,trim=70 30 30 30,width=.52\linewidth]{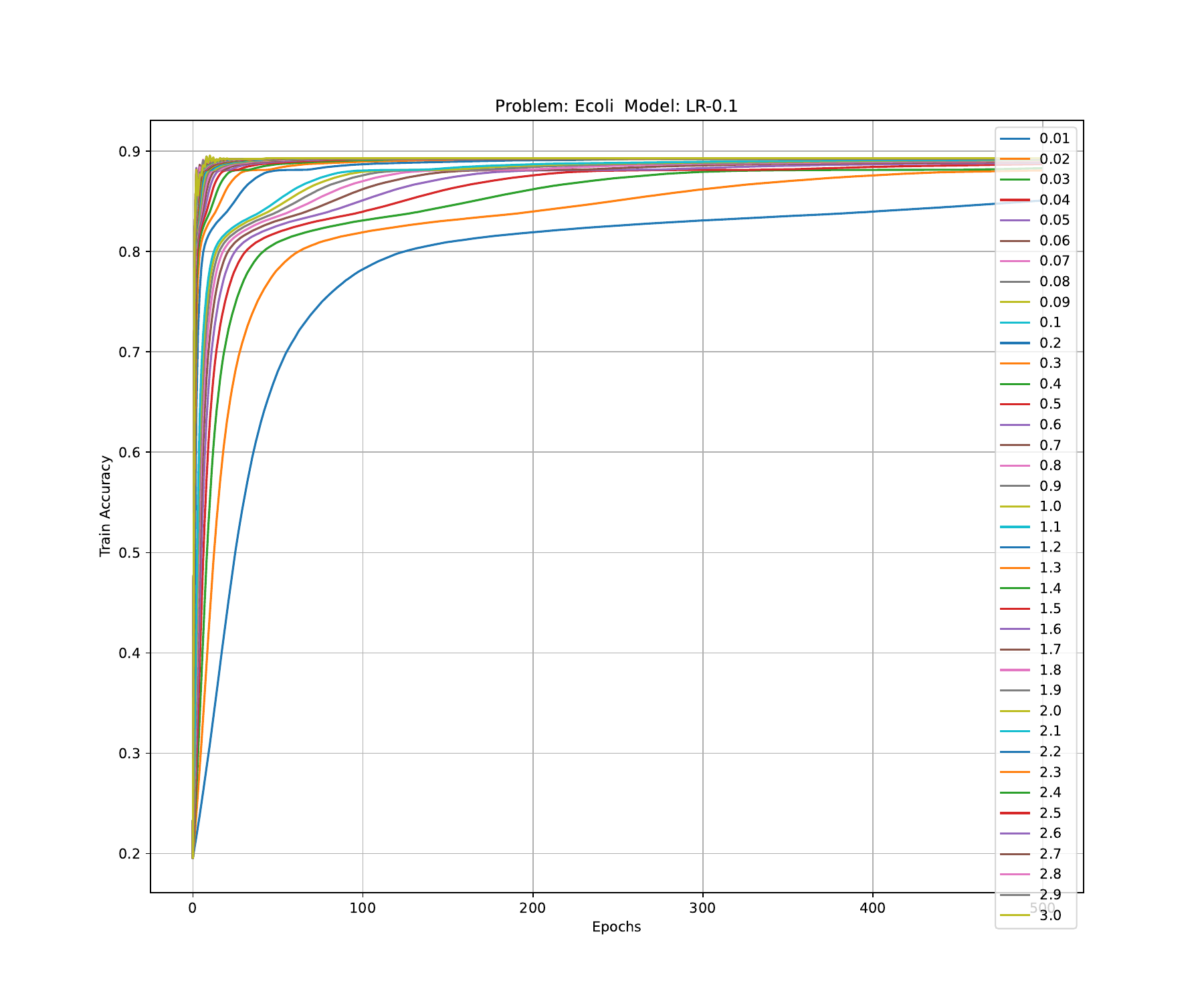} &
    \includegraphics[clip,trim=70 30 30 30,width=.52\linewidth]{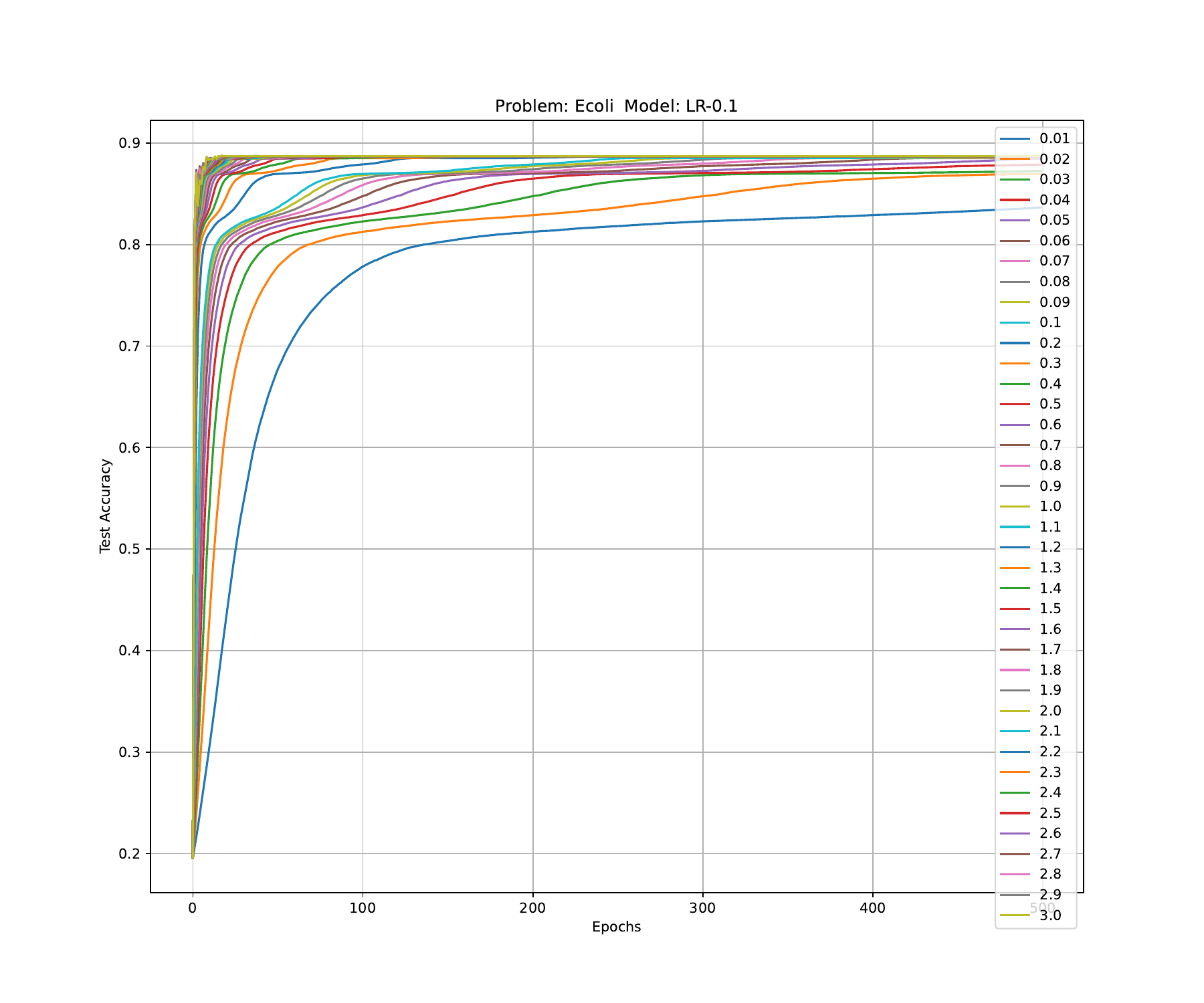} \\[-3mm]
  \end{tabular}
\caption{Accuracies for LR-0.1 model.}
  \label{fig:accuracies_LR_0.1_model}
\end{figure*}

\begin{figure*}[p]
  \centering
  \ContinuedFloat
  \begin{tabular}{c@{}c}
    Training Accuracy & Validation Accuracy \\
    \includegraphics[clip,trim=70 30 30 30,width=.52\linewidth]{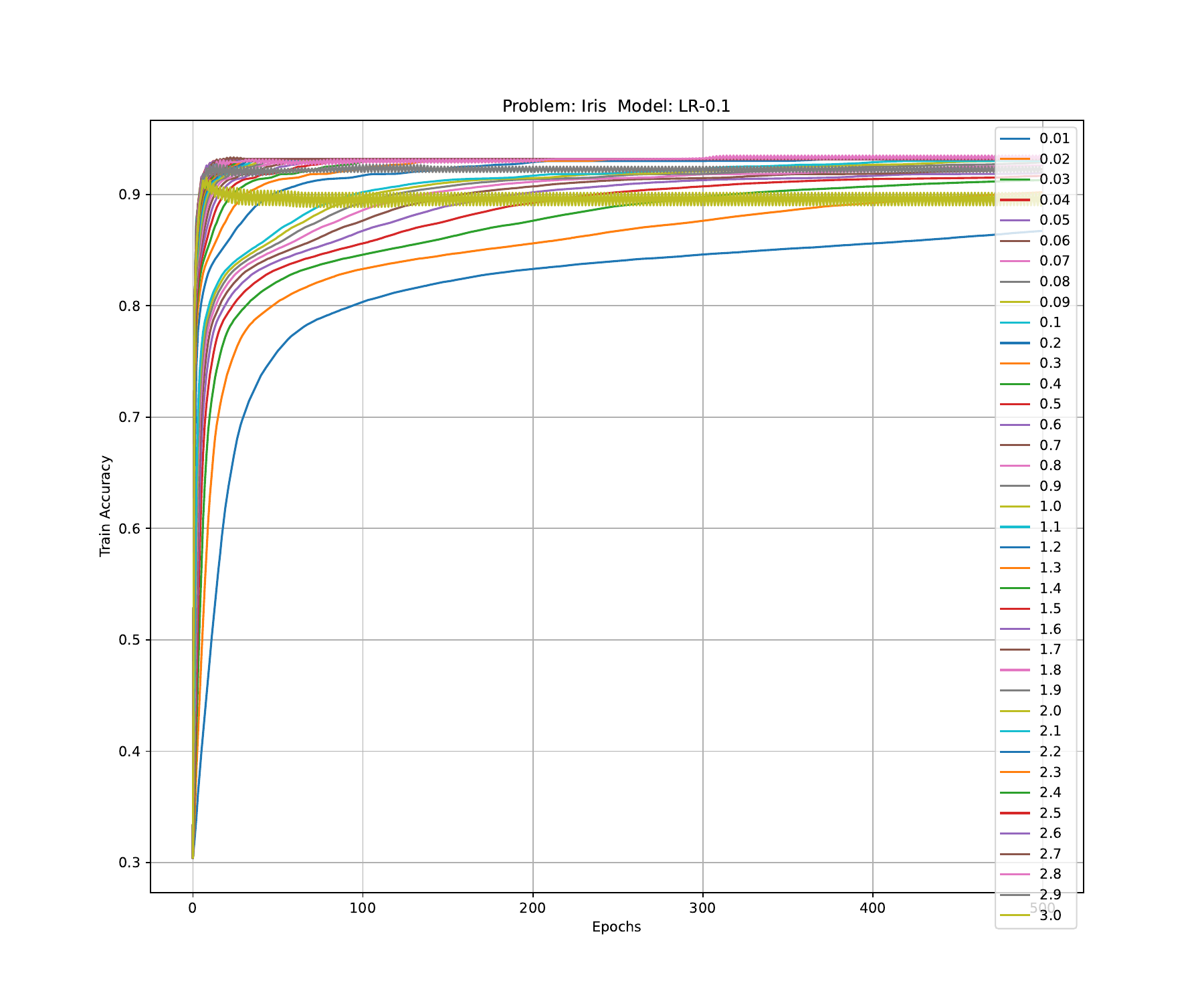} &
    \includegraphics[clip,trim=70 30 30 30,width=.52\linewidth]{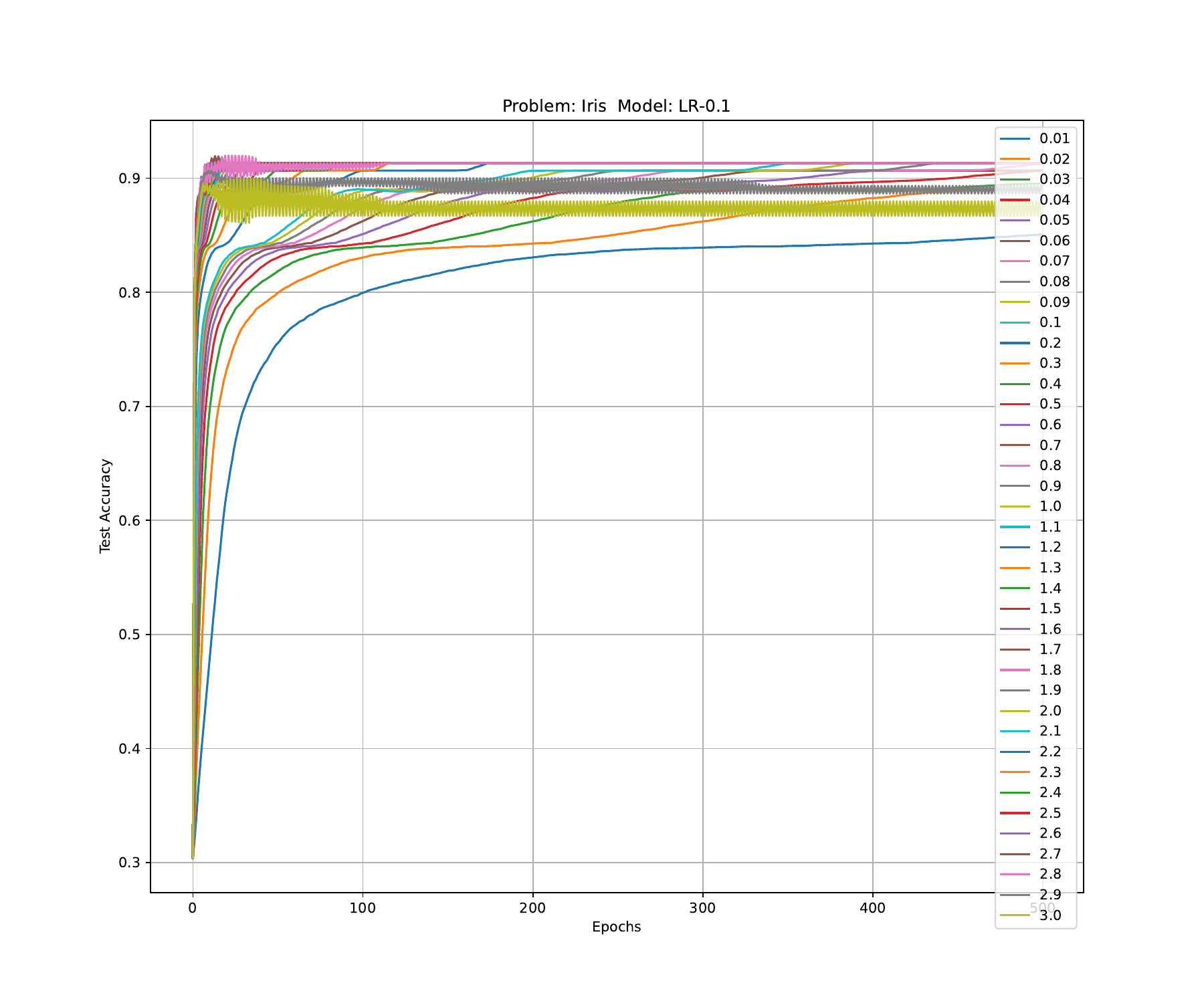} \\[-3mm]
    \includegraphics[clip,trim=70 30 30 30,width=.52\linewidth]{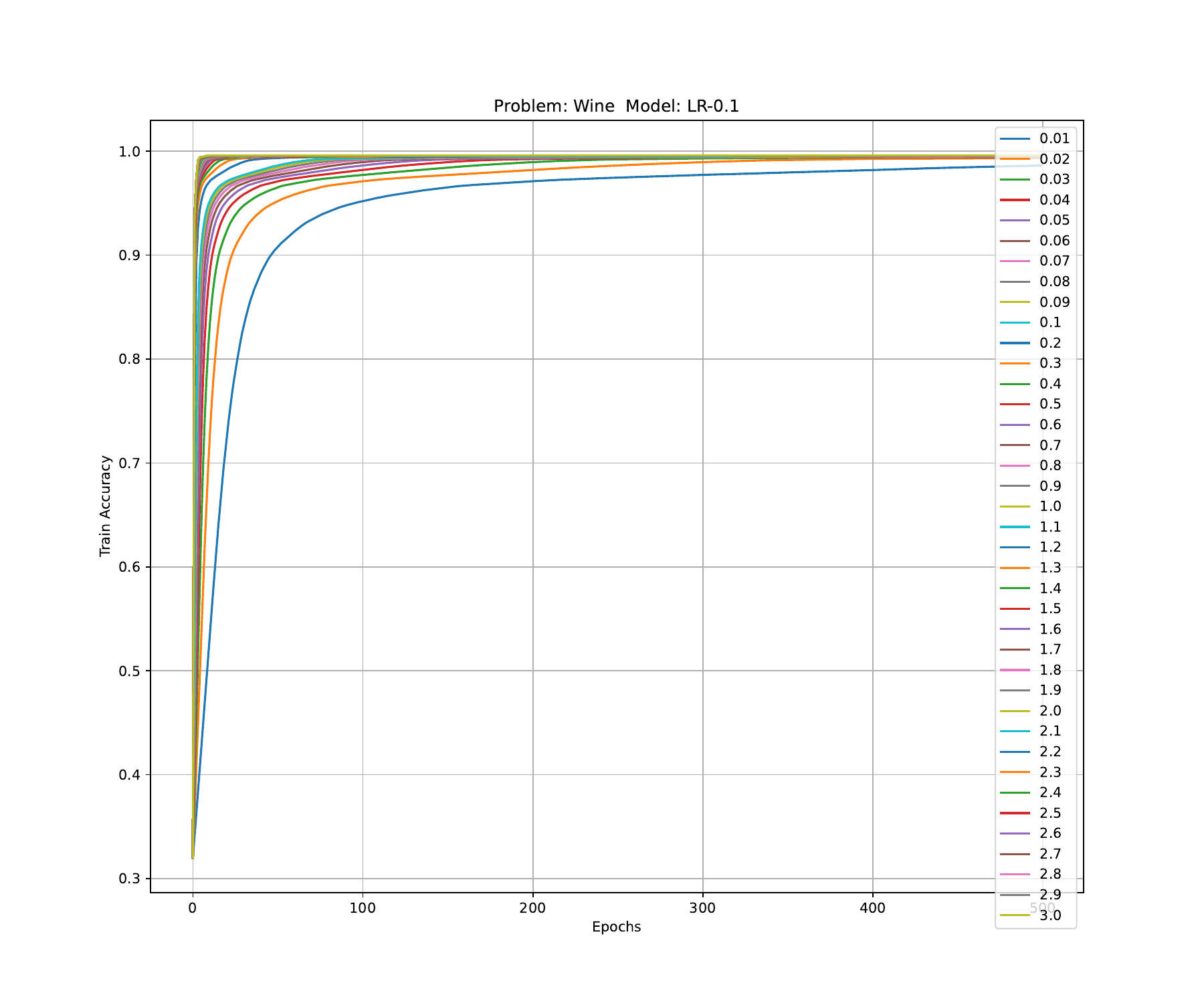} &
    \includegraphics[clip,trim=70 30 30 30,width=.52\linewidth]{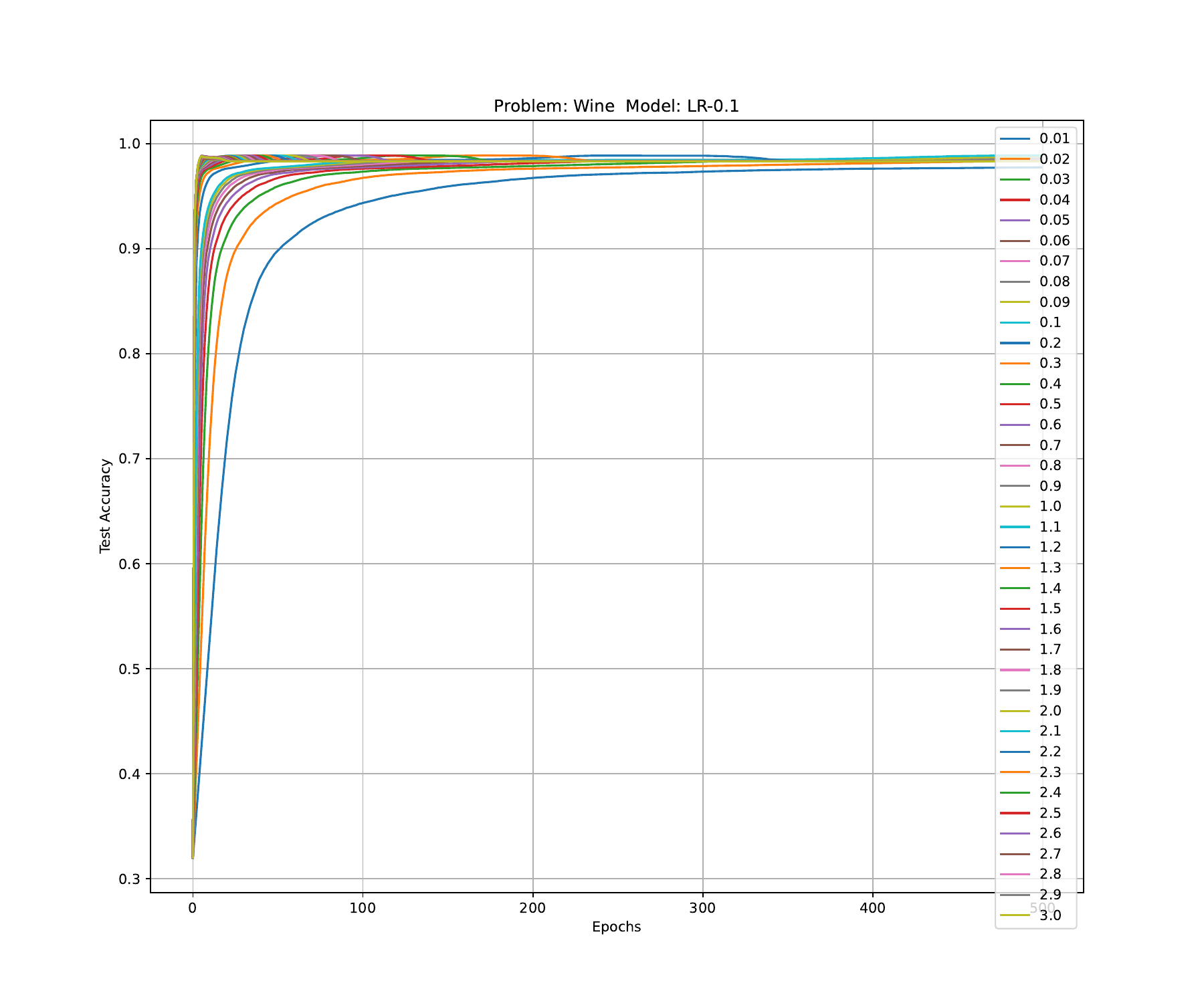} \\[-3mm]
  \end{tabular}
\caption{Accuracies for LR-0.1 model (continued).}
\end{figure*}

\begin{figure*}[p]
  \centering
  \begin{tabular}{c@{}c}
    Training Accuracy & Validation Accuracy \\
    \includegraphics[clip,trim=70 30 30 30,width=.52\linewidth]{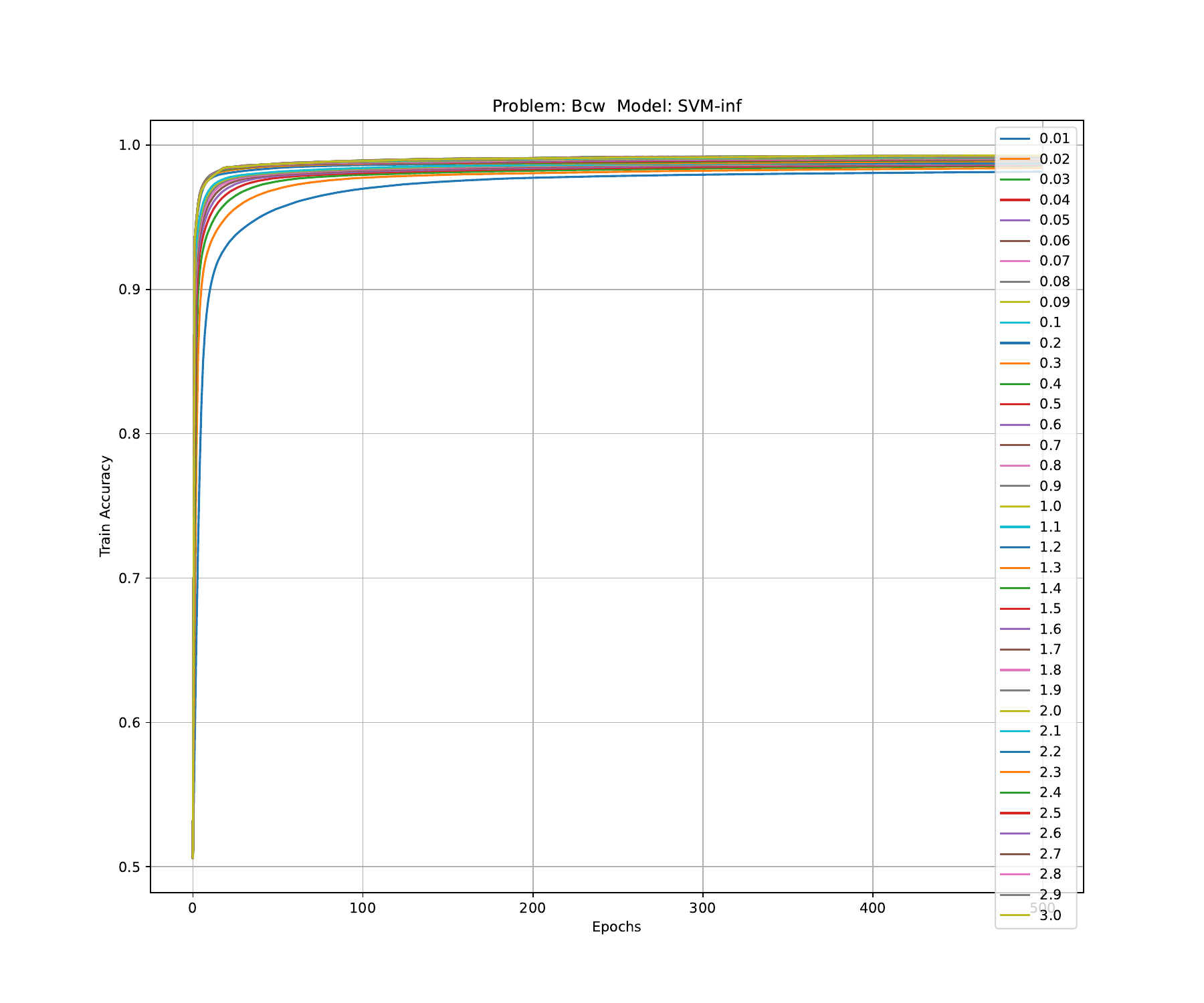} &
    \includegraphics[clip,trim=70 30 30 30,width=.52\linewidth]{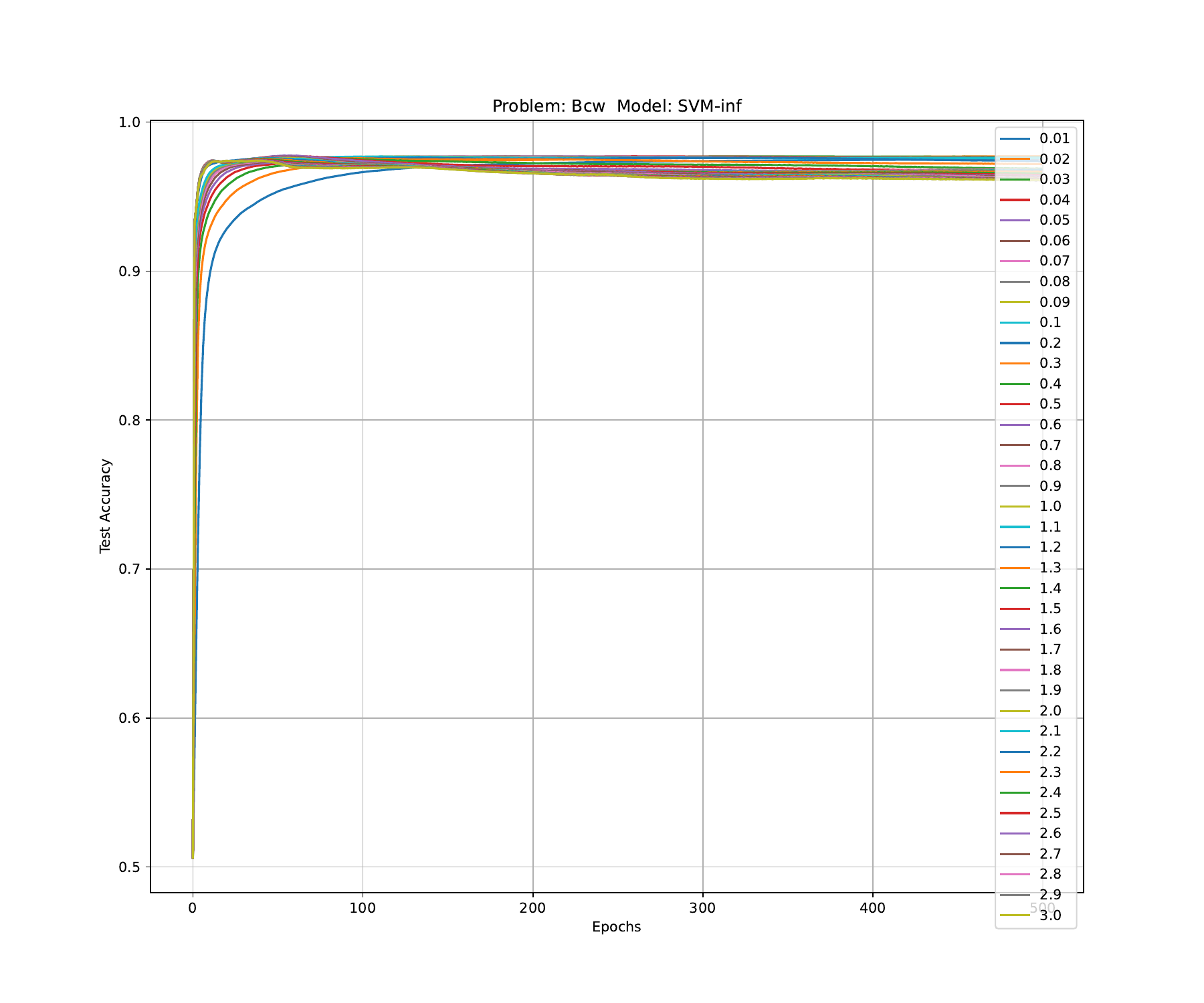} \\[-3mm]
    \includegraphics[clip,trim=70 30 30 30,width=.52\linewidth]{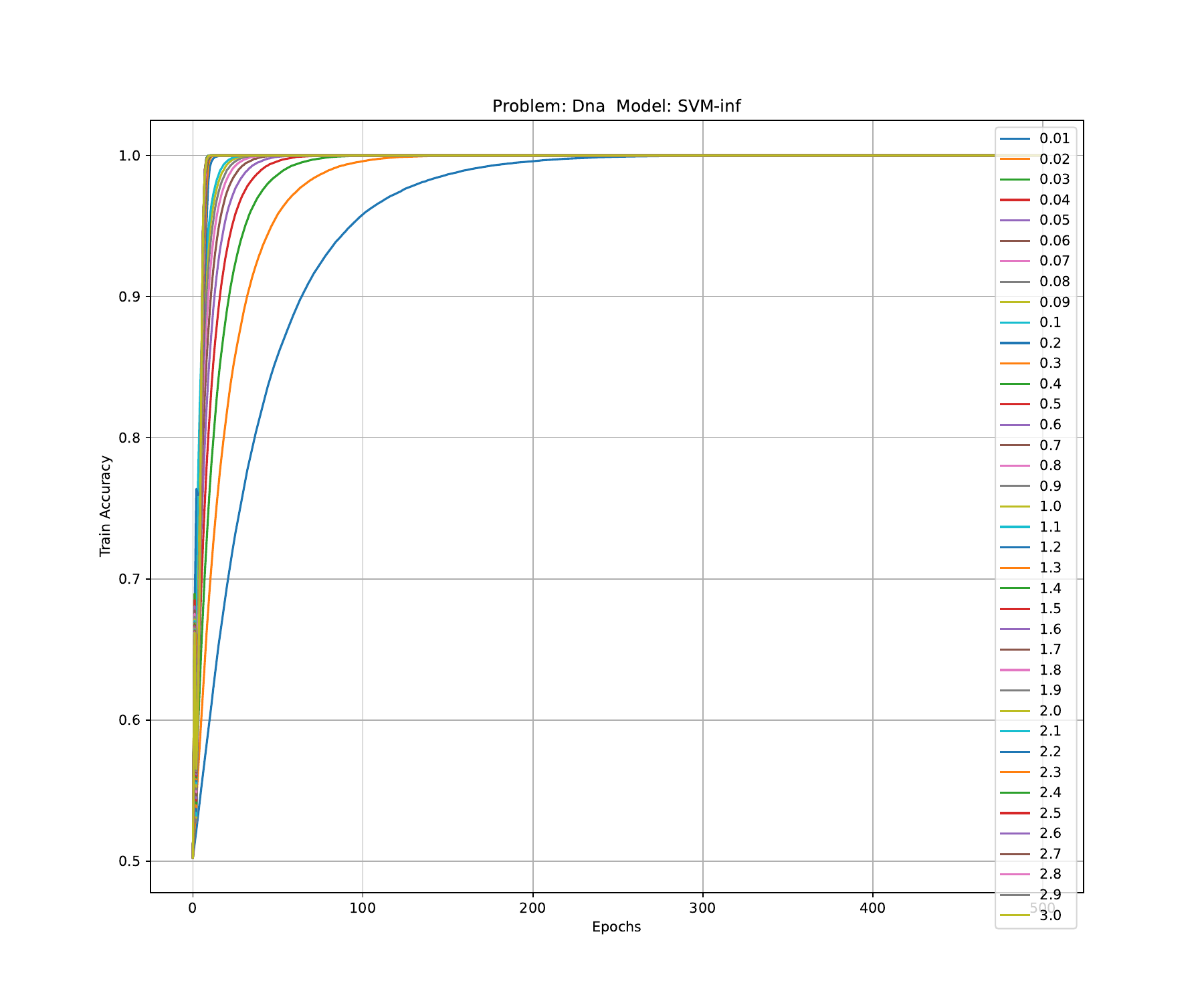} &
    \includegraphics[clip,trim=70 30 30 30,width=.52\linewidth]{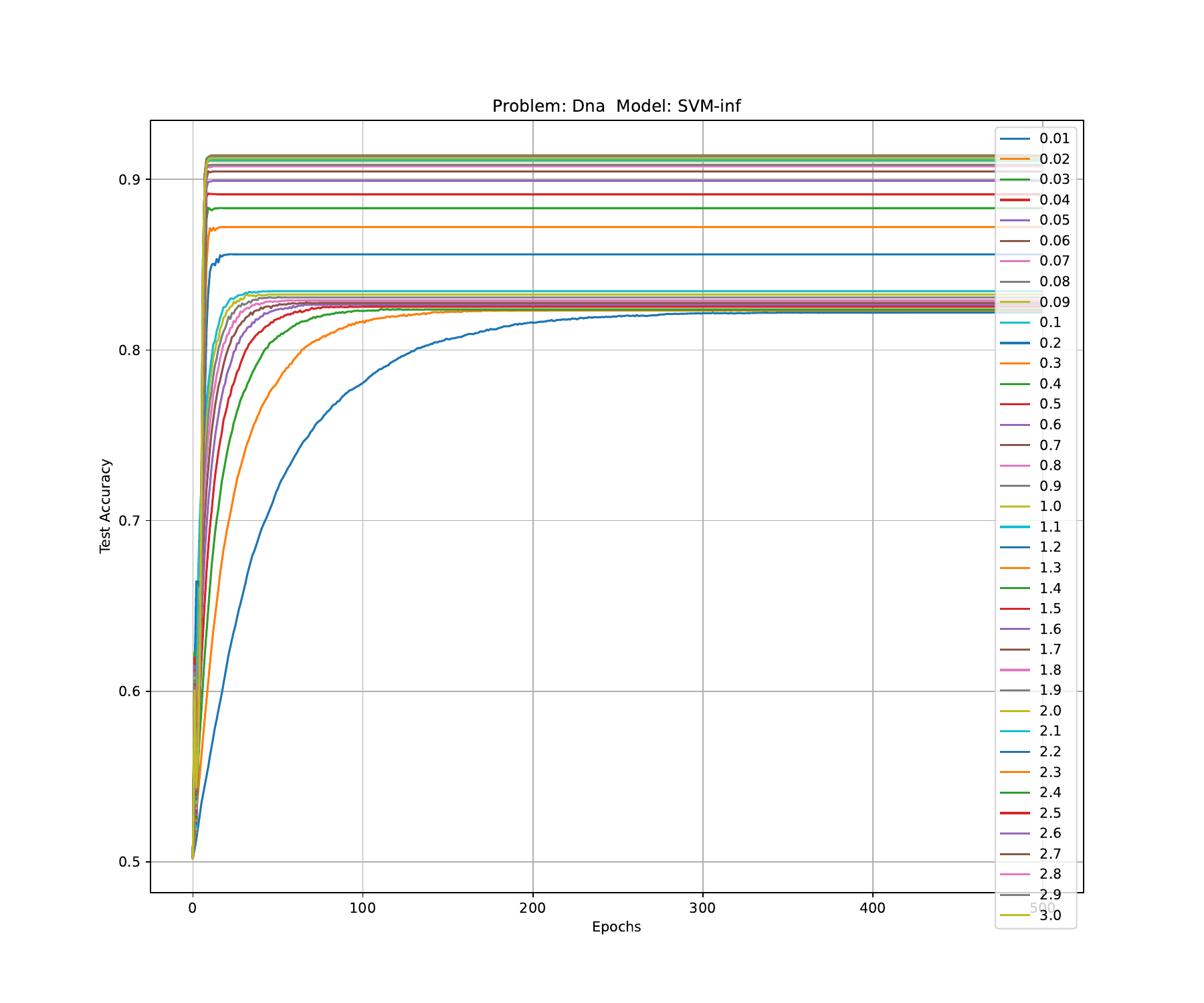} \\[-3mm]
    \includegraphics[clip,trim=70 30 30 30,width=.52\linewidth]{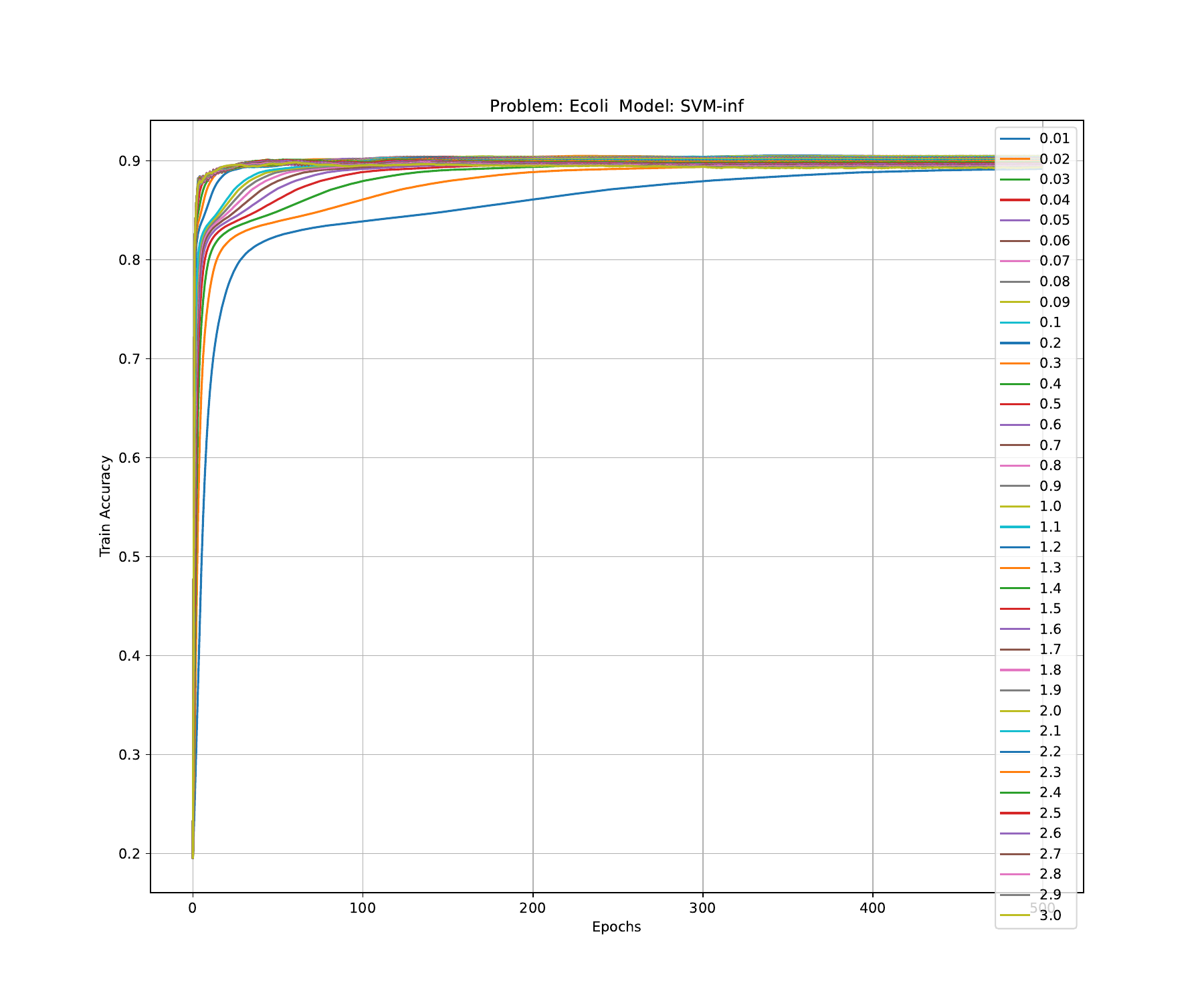} &
    \includegraphics[clip,trim=70 30 30 30,width=.52\linewidth]{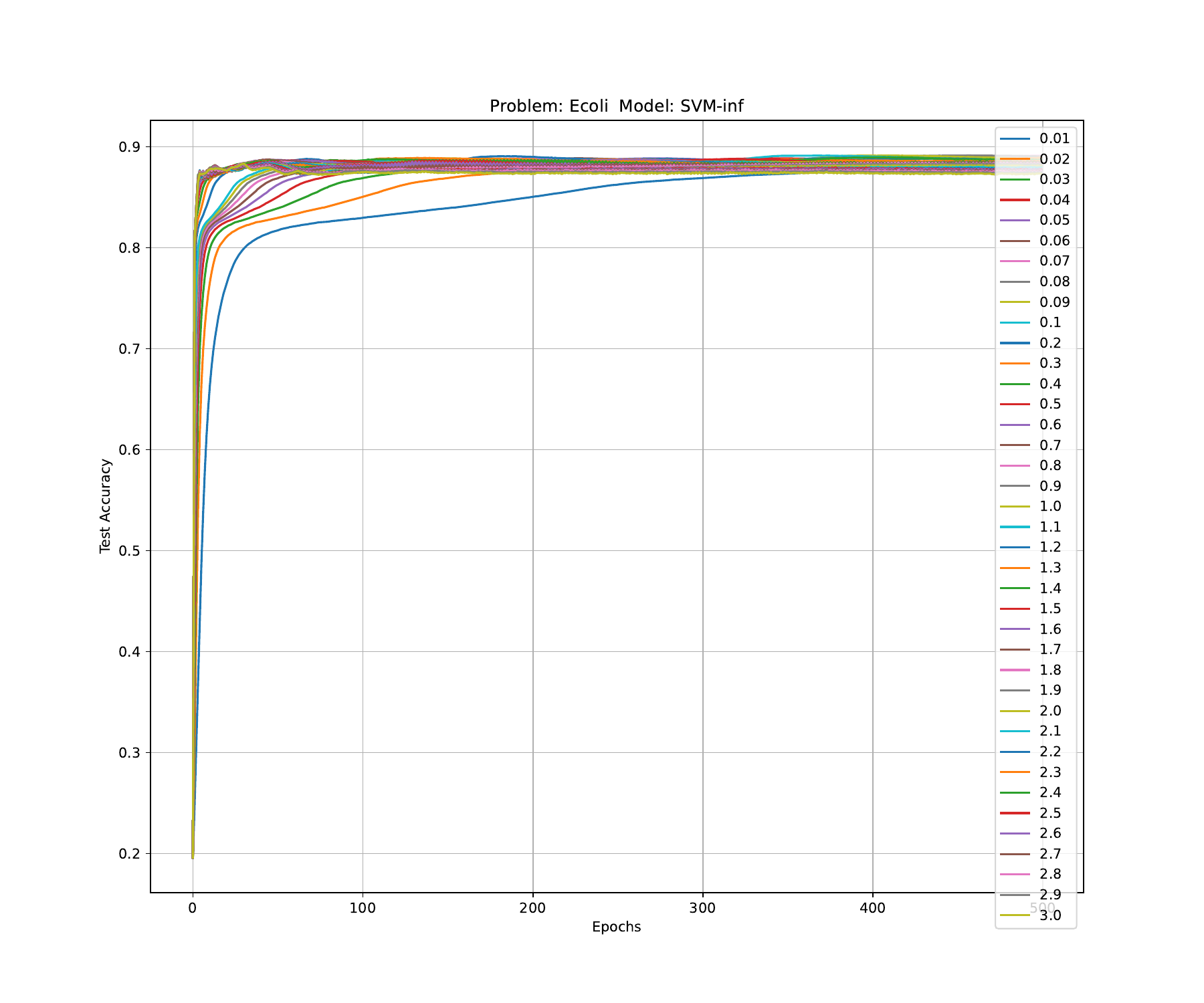} \\[-3mm]
  \end{tabular}
\caption{Accuracies for SVM-Inf model.}
  \label{fig:accuracies_SVM_Inf_model}
\end{figure*}

\begin{figure*}[p]
  \centering
  \ContinuedFloat
  \begin{tabular}{c@{}c}
    Training Accuracy & Validation Accuracy \\
    \includegraphics[clip,trim=70 30 30 30,width=.52\linewidth]{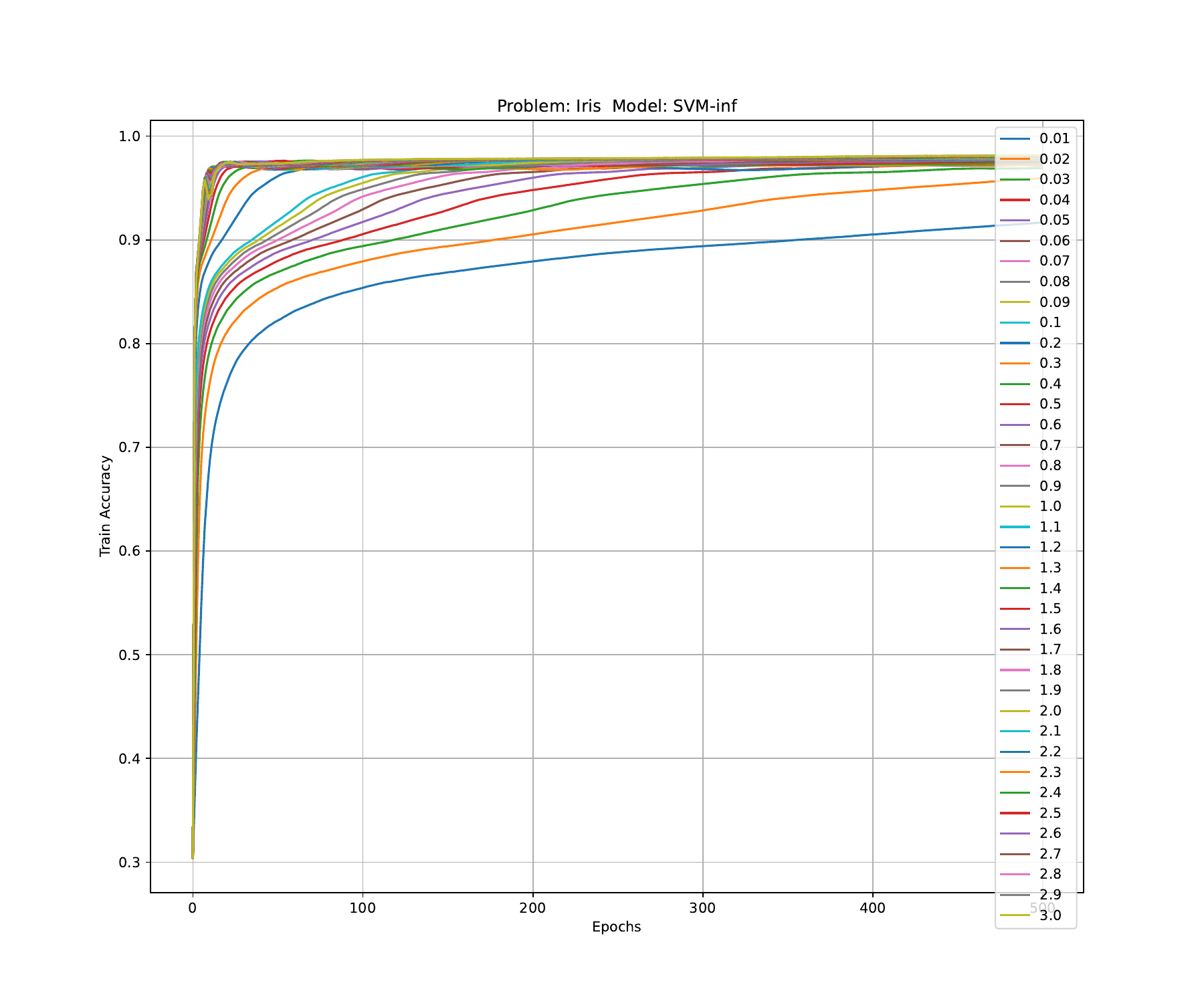} &
    \includegraphics[clip,trim=70 30 30 30,width=.52\linewidth]{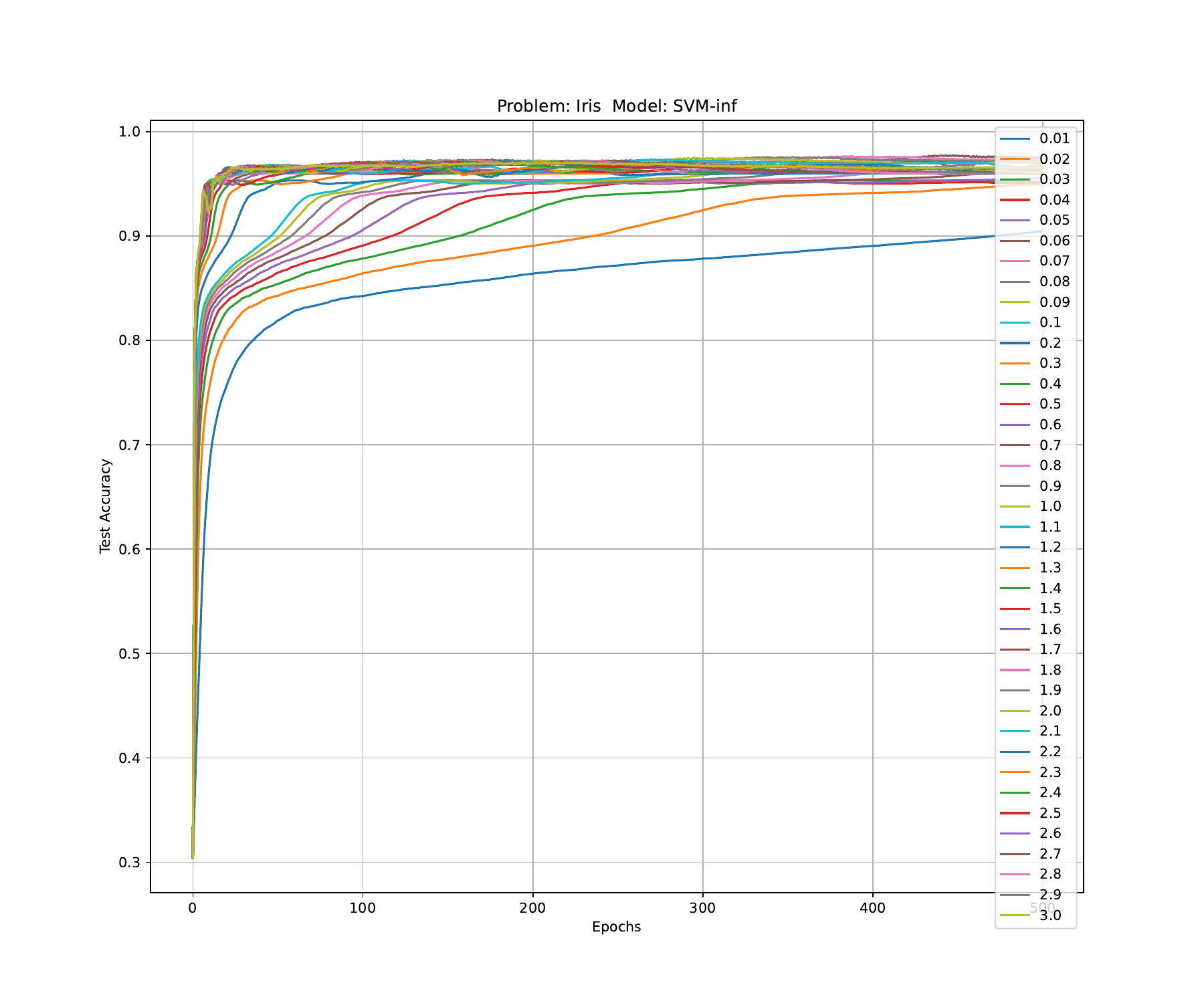} \\[-3mm]
    \includegraphics[clip,trim=70 30 30 30,width=.52\linewidth]{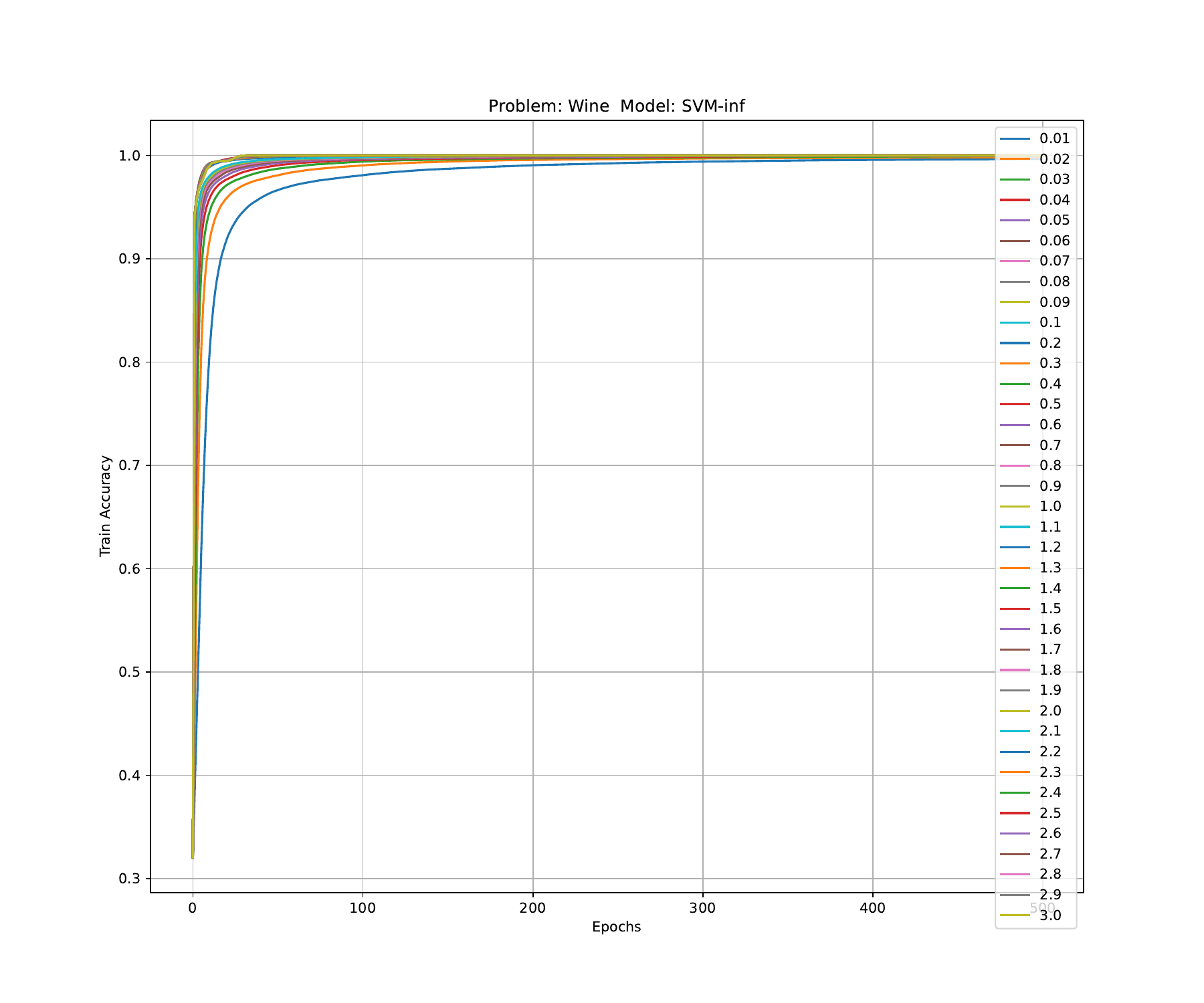} &
    \includegraphics[clip,trim=70 30 30 30,width=.52\linewidth]{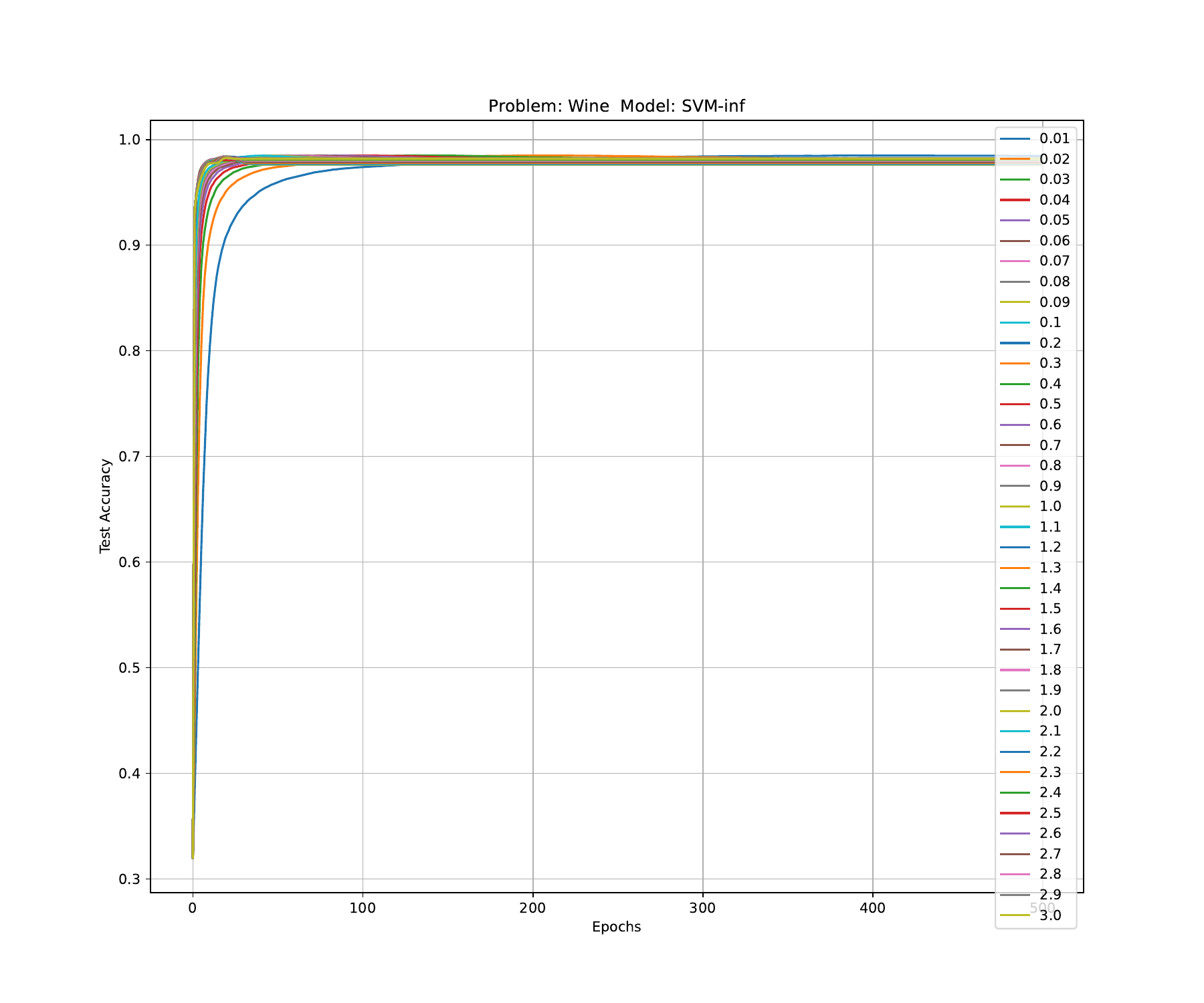} \\[-3mm]
  \end{tabular}
\caption{Accuracies for SVM-Inf model (continued).}
\end{figure*}

\begin{figure*}[p]
  \centering
  \begin{tabular}{c@{}c}
    Training Accuracy & Validation Accuracy \\
    \includegraphics[clip,trim=70 30 30 30,width=.52\linewidth]{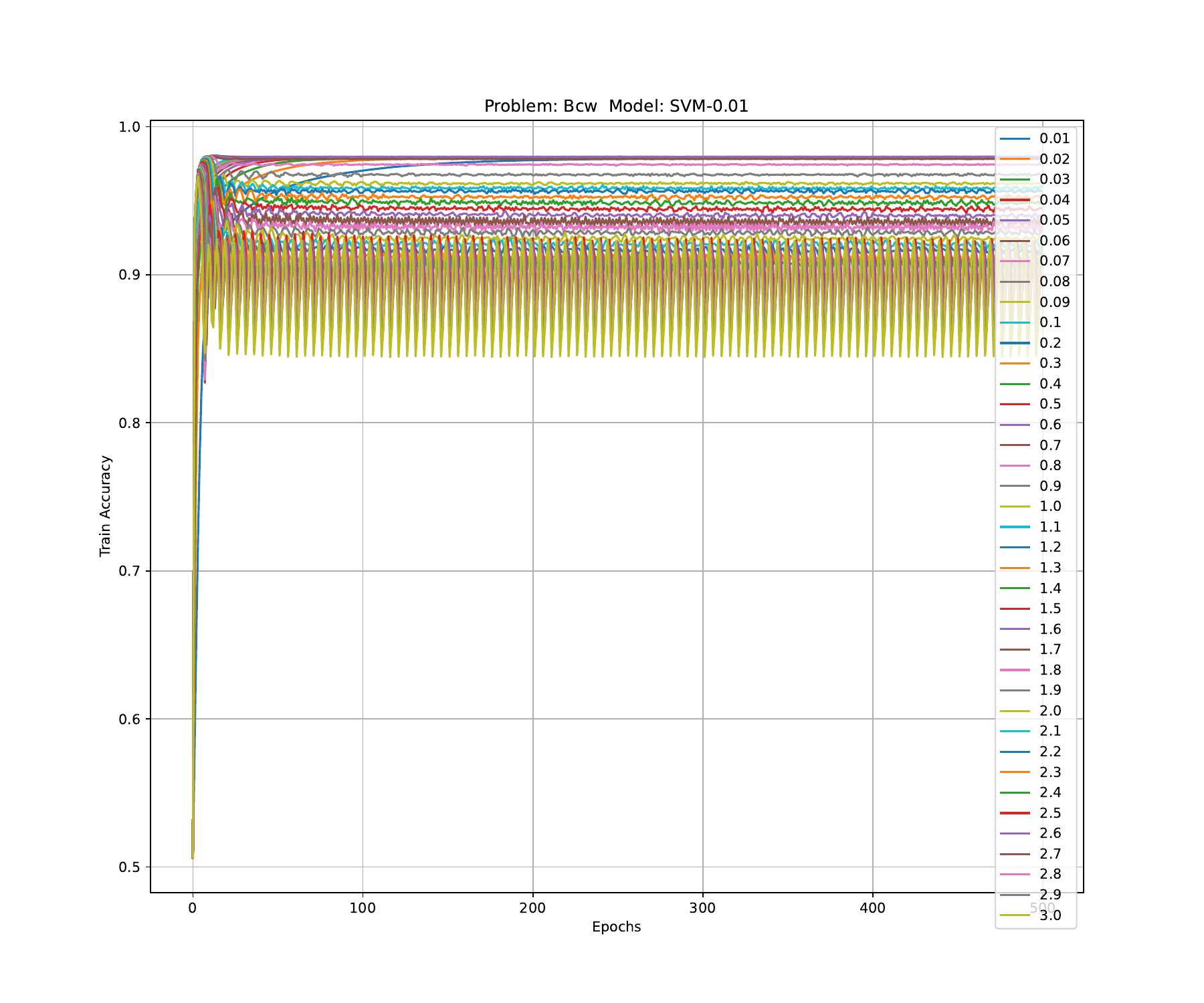} &
    \includegraphics[clip,trim=70 30 30 30,width=.52\linewidth]{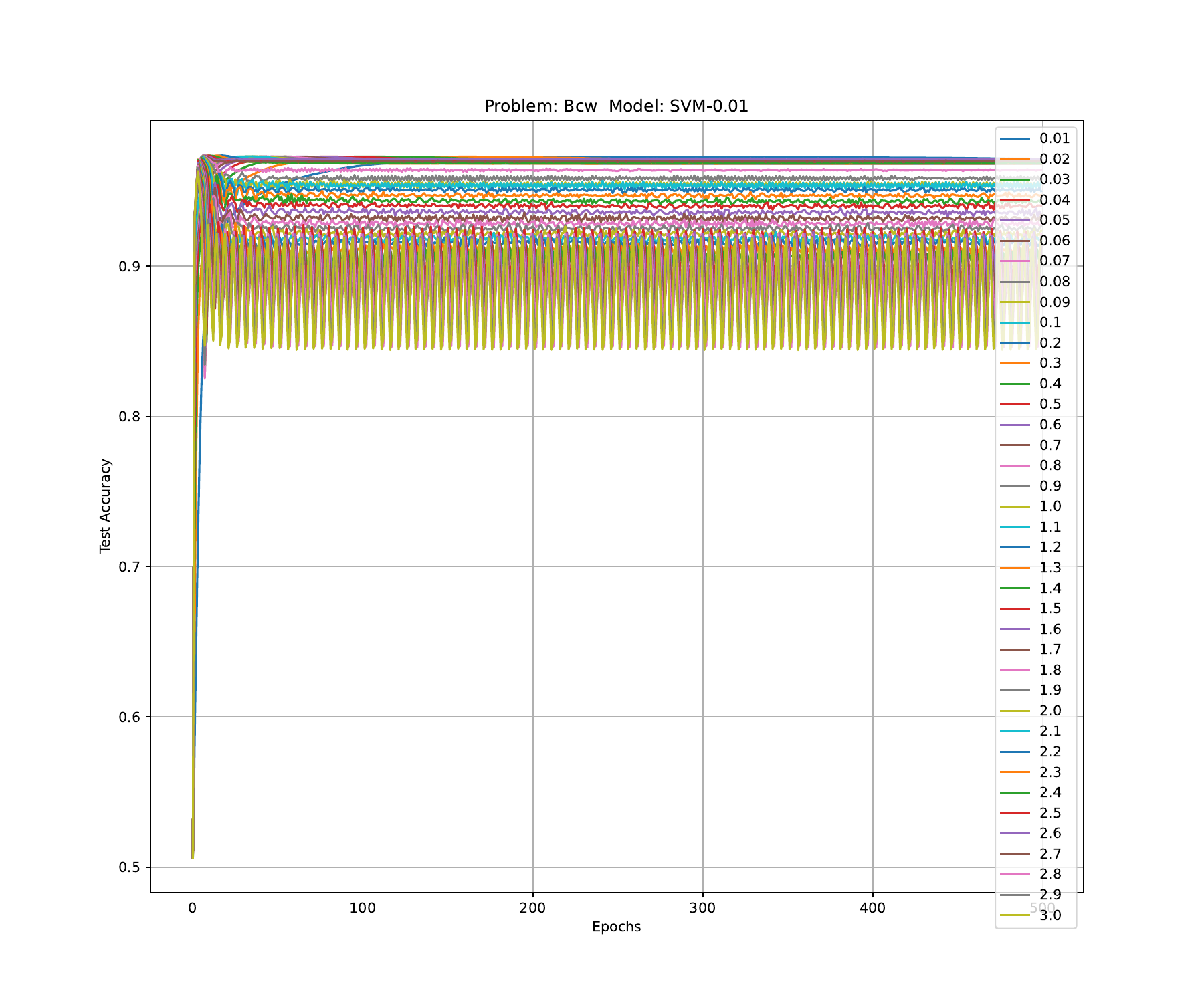} \\[-3mm]
    \includegraphics[clip,trim=70 30 30 30,width=.52\linewidth]{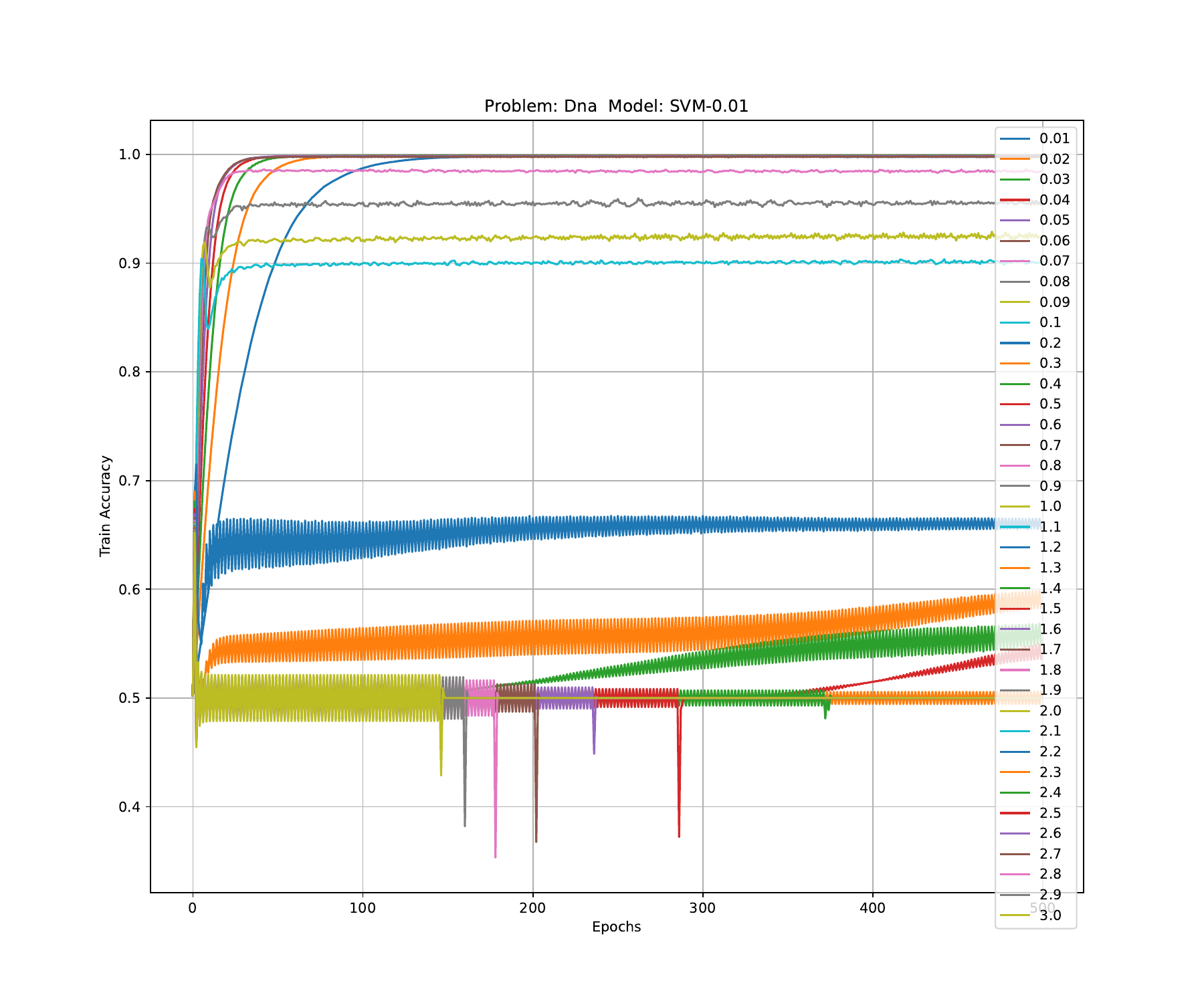} &
    \includegraphics[clip,trim=70 30 30 30,width=.52\linewidth]{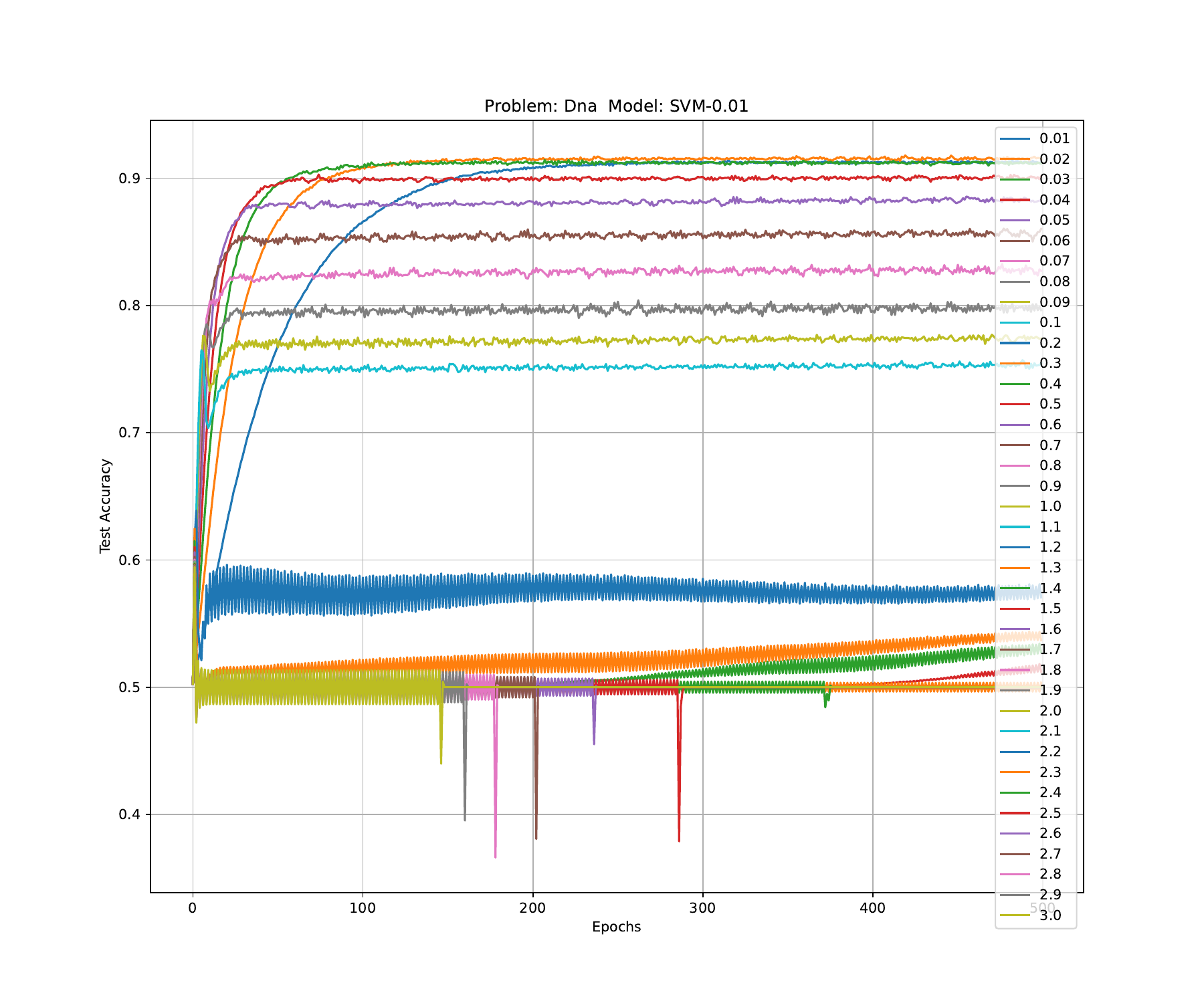} \\[-3mm]
    \includegraphics[clip,trim=70 30 30 30,width=.52\linewidth]{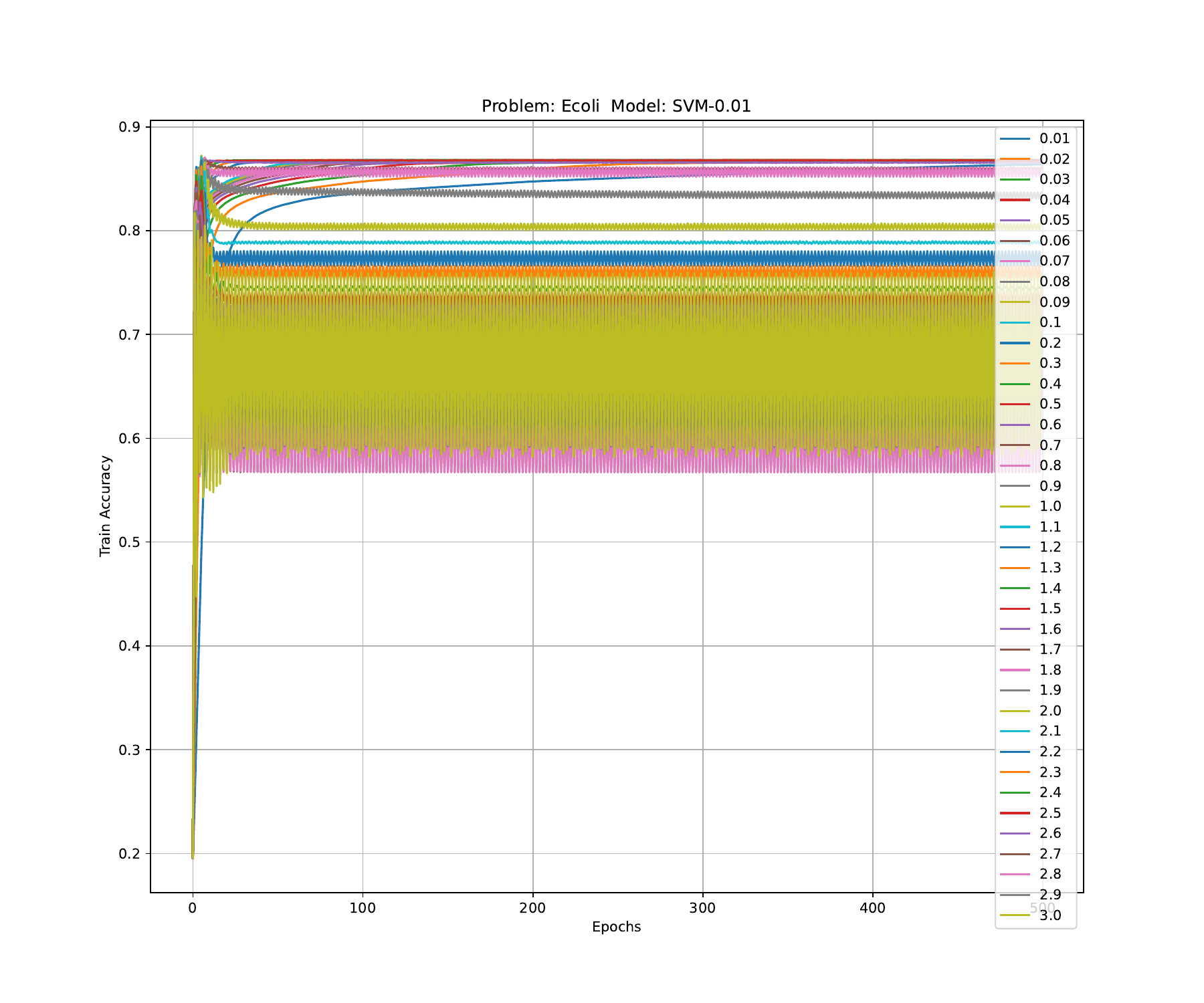} &
    \includegraphics[clip,trim=70 30 30 30,width=.52\linewidth]{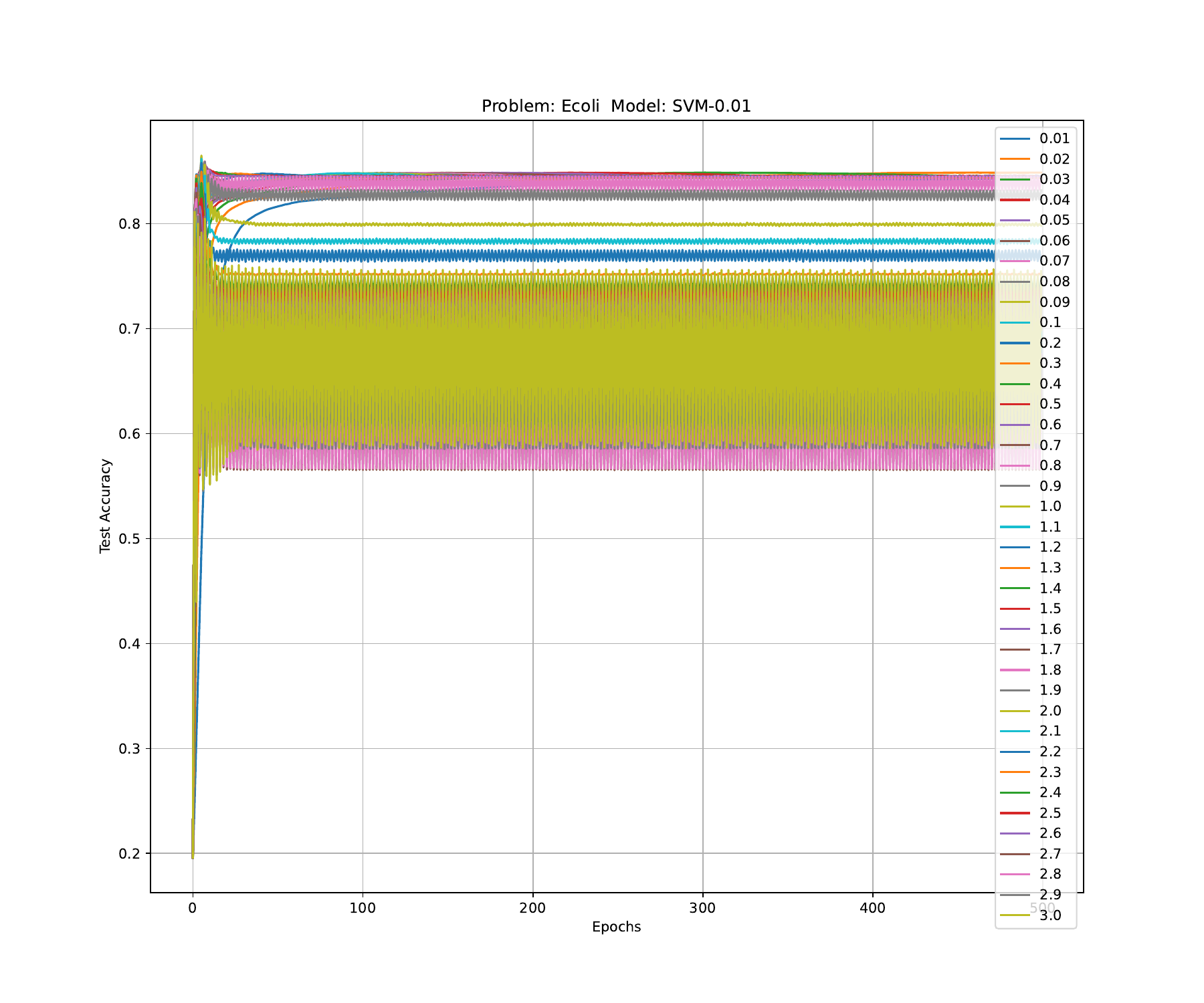} \\[-3mm]
  \end{tabular}
\caption{Accuracies for SVM-0.01 model.}
  \label{fig:accuracies_SVM_0.01_model}
\end{figure*}

\begin{figure*}[p]
  \centering
  \ContinuedFloat
  \begin{tabular}{c@{}c}
    Training Accuracy & Validation Accuracy \\
    \includegraphics[clip,trim=70 30 30 30,width=.52\linewidth]{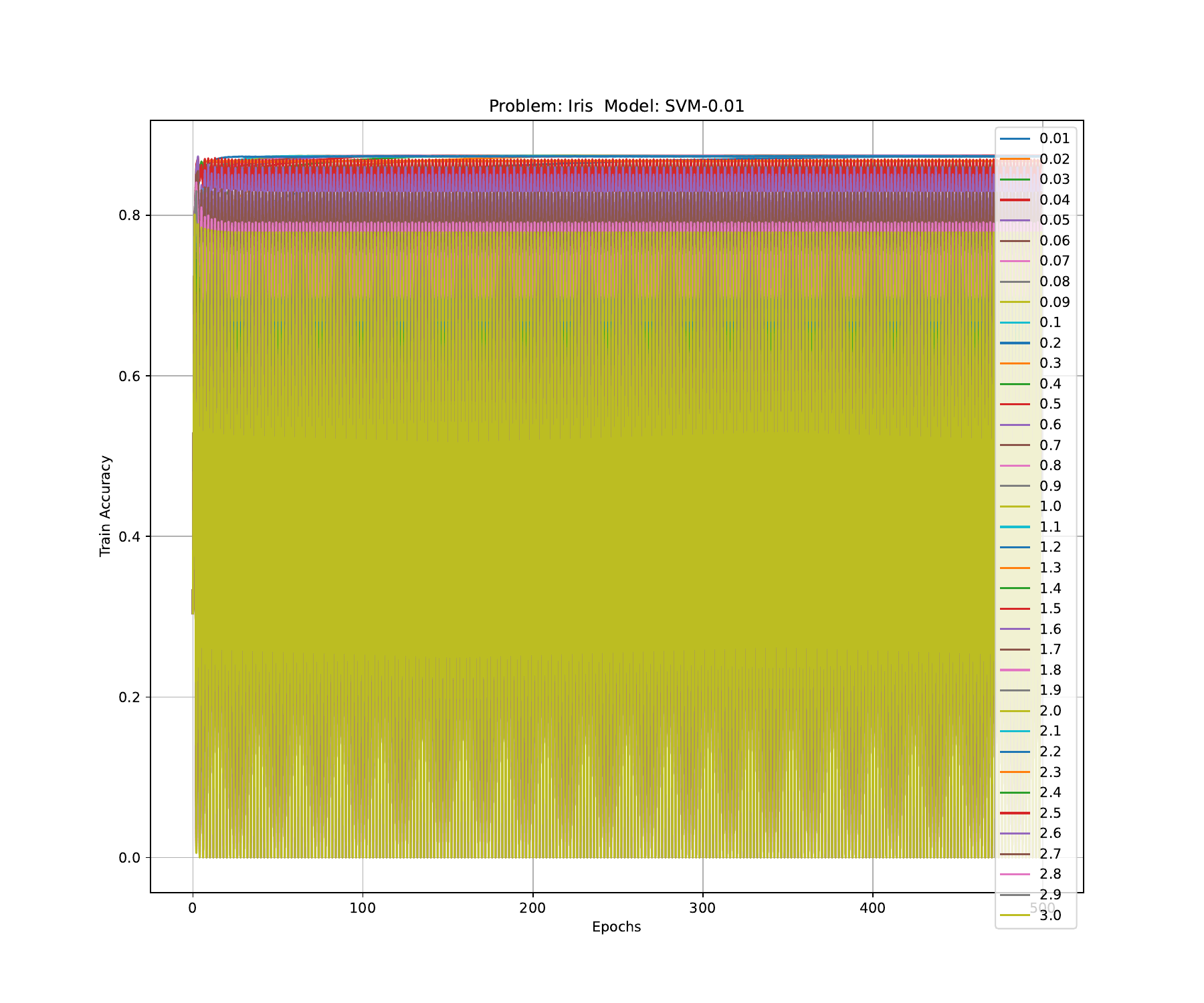} &
    \includegraphics[clip,trim=70 30 30 30,width=.52\linewidth]{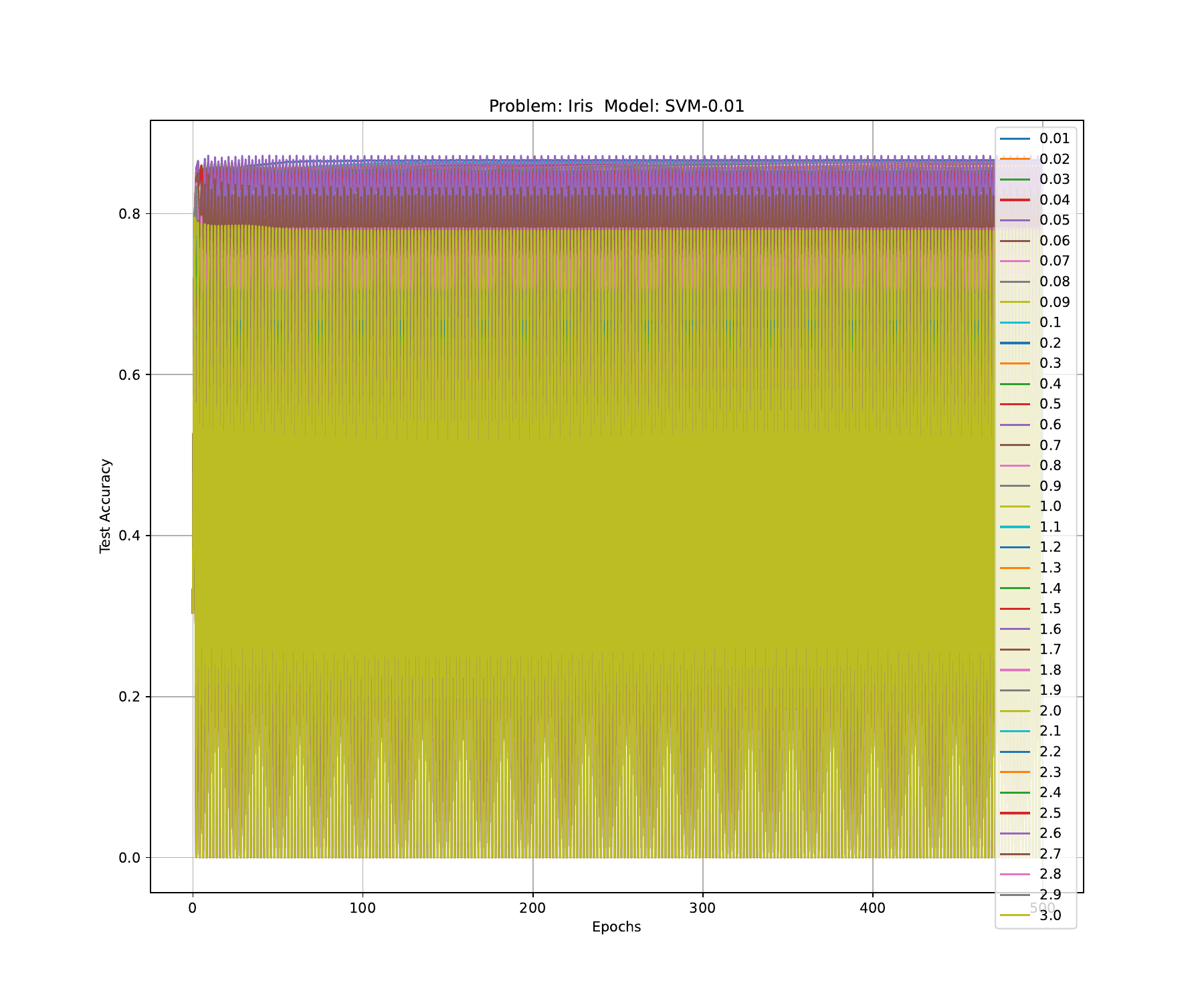} \\[-3mm]
    \includegraphics[clip,trim=70 30 30 30,width=.52\linewidth]{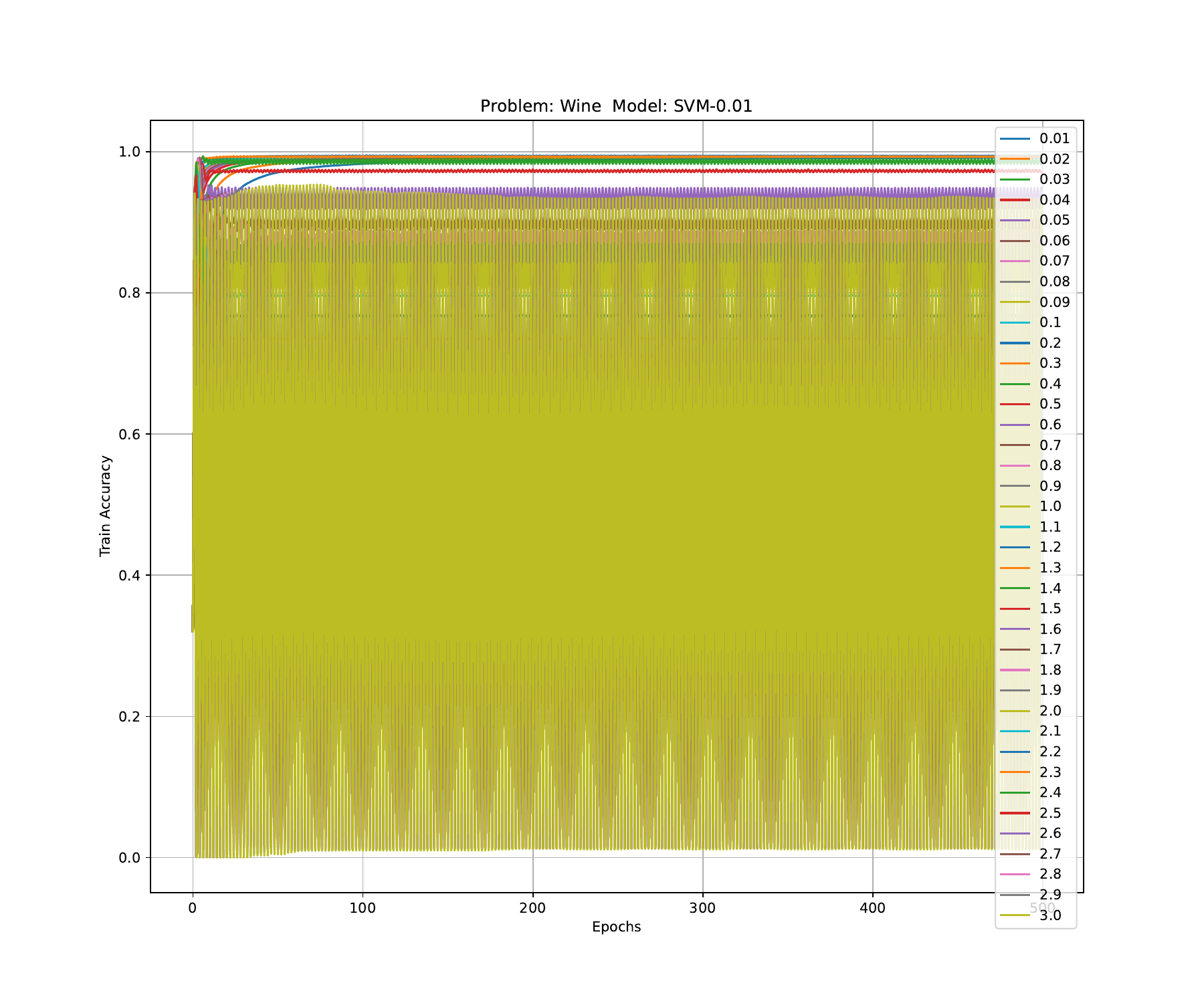} &
    \includegraphics[clip,trim=70 30 30 30,width=.52\linewidth]{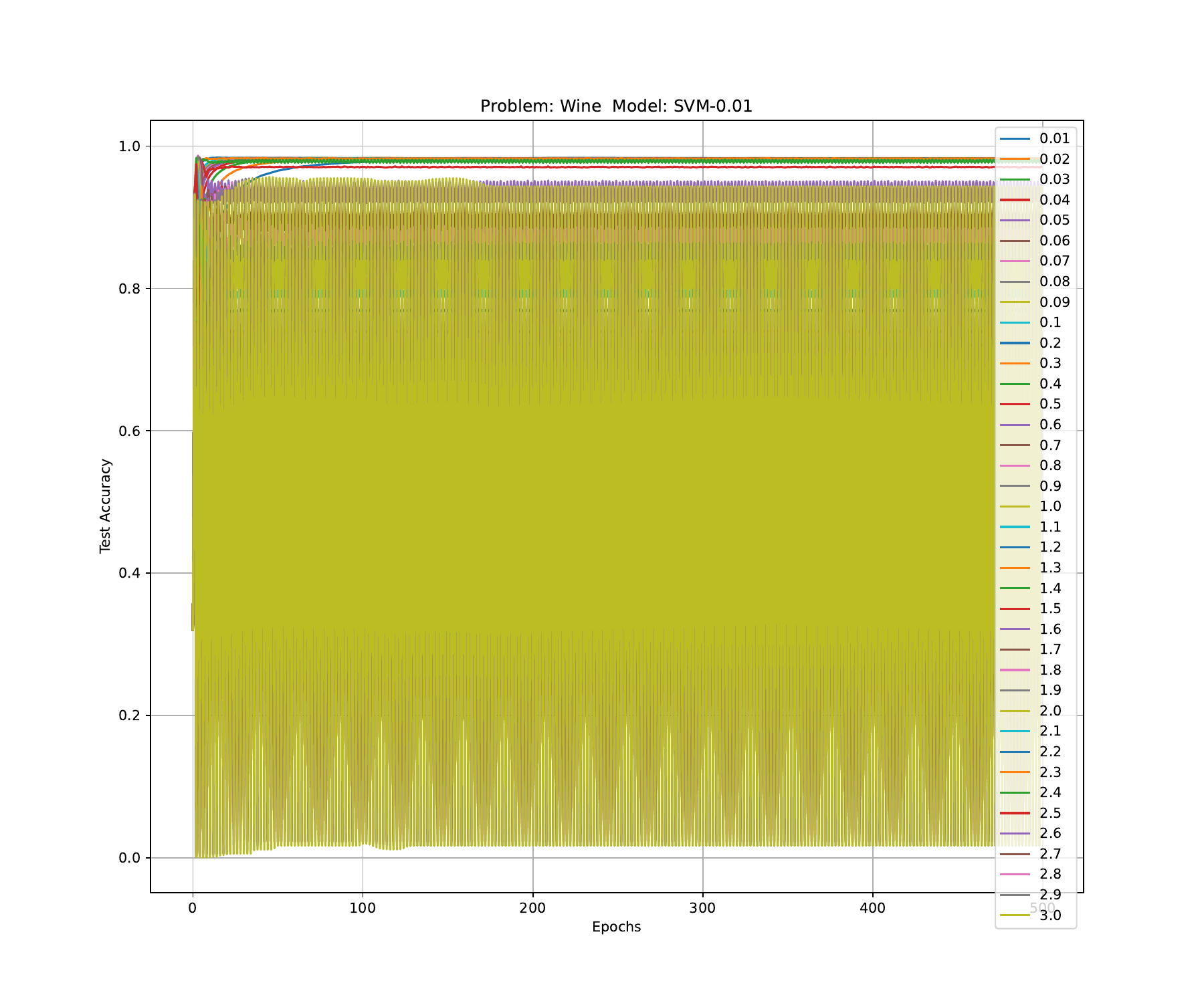} \\[-3mm]
  \end{tabular}
\caption{Accuracies for SVM-0.01 model (continued).}
\end{figure*}

\begin{figure*}[p]
  \centering
  \begin{tabular}{c@{}c}
    Training Accuracy & Validation Accuracy \\
    \includegraphics[clip,trim=70 30 30 30,width=.52\linewidth]{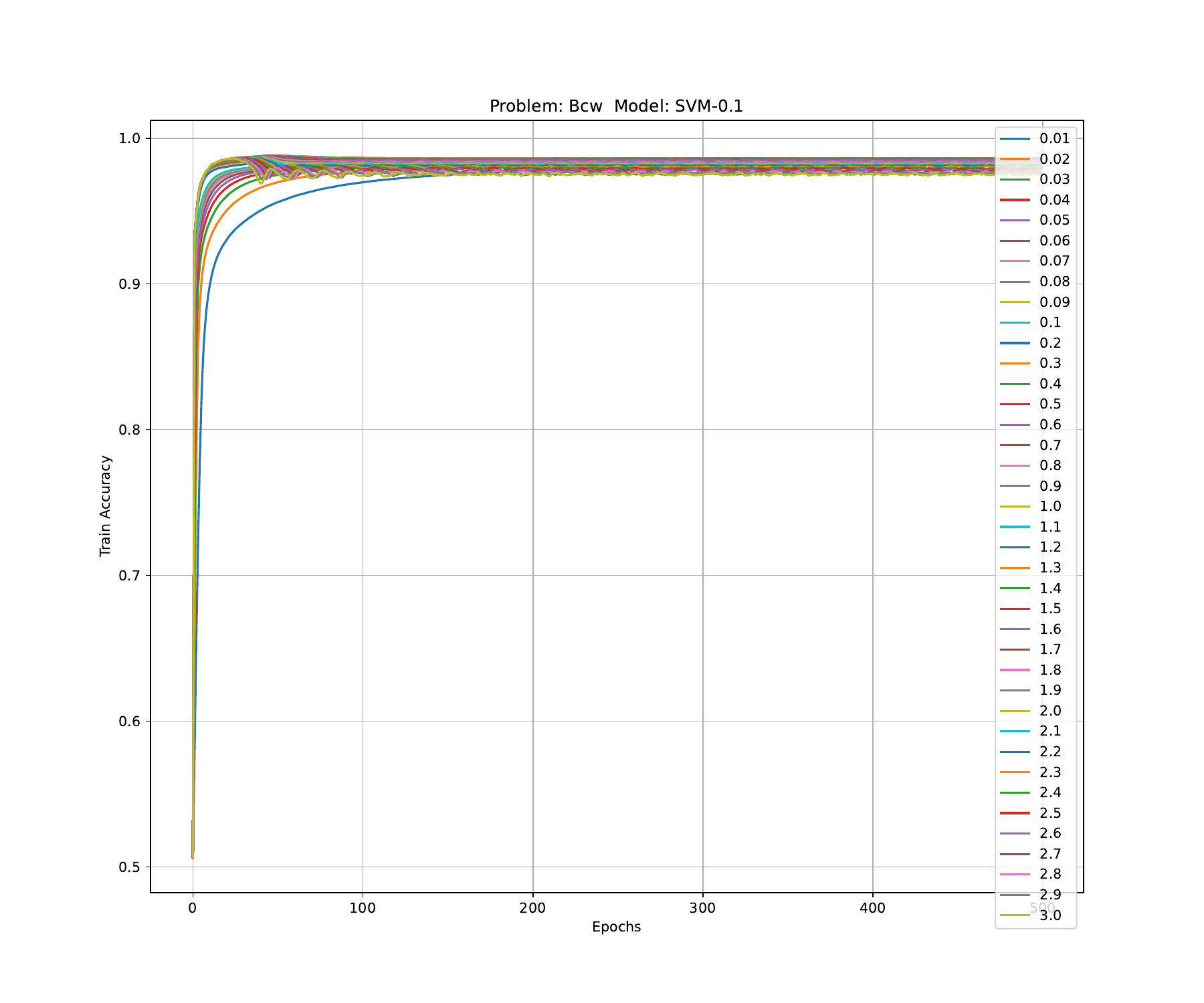} &
    \includegraphics[clip,trim=70 30 30 30,width=.52\linewidth]{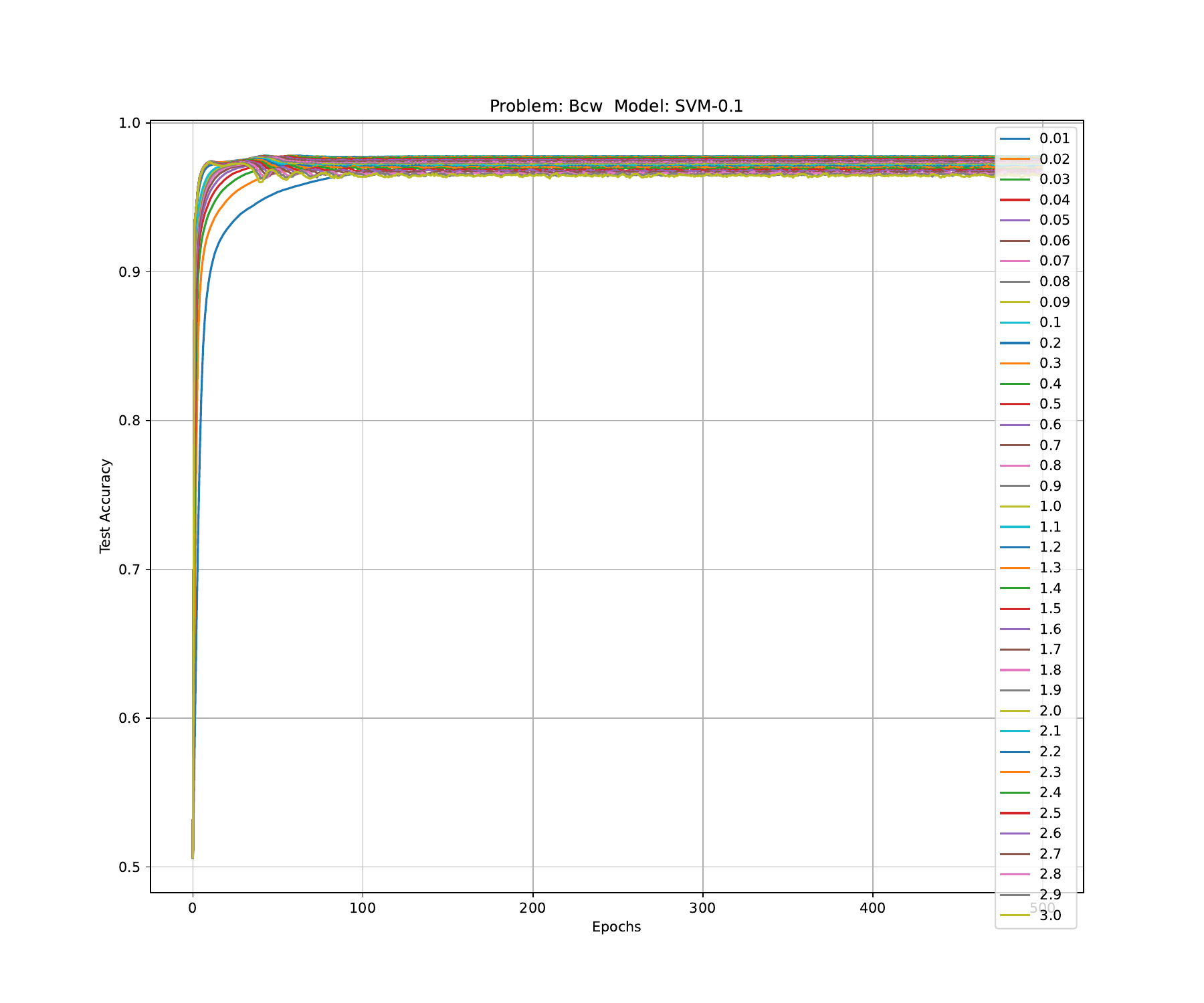} \\[-3mm]
    \includegraphics[clip,trim=70 30 30 30,width=.52\linewidth]{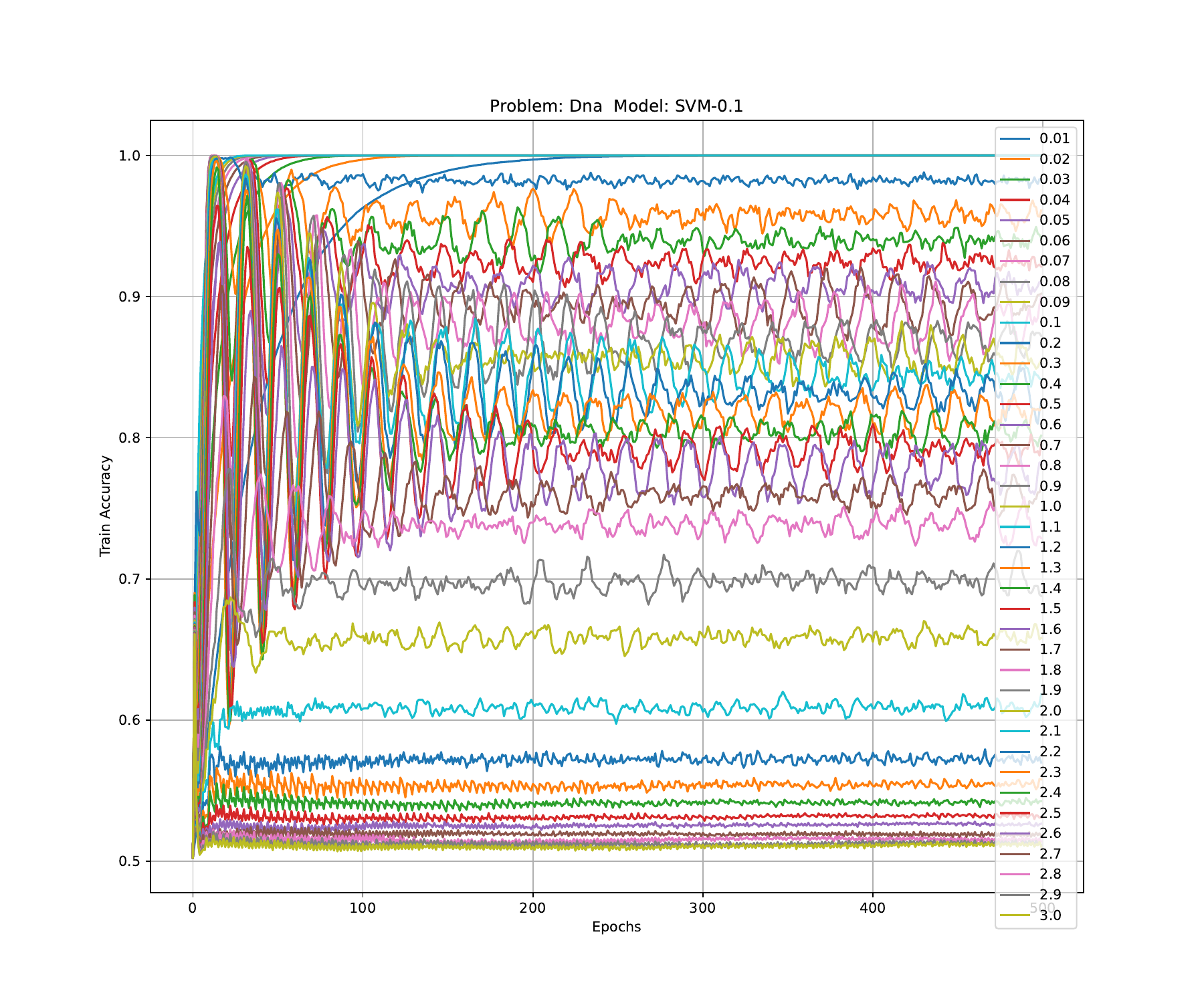} &
    \includegraphics[clip,trim=70 30 30 30,width=.52\linewidth]{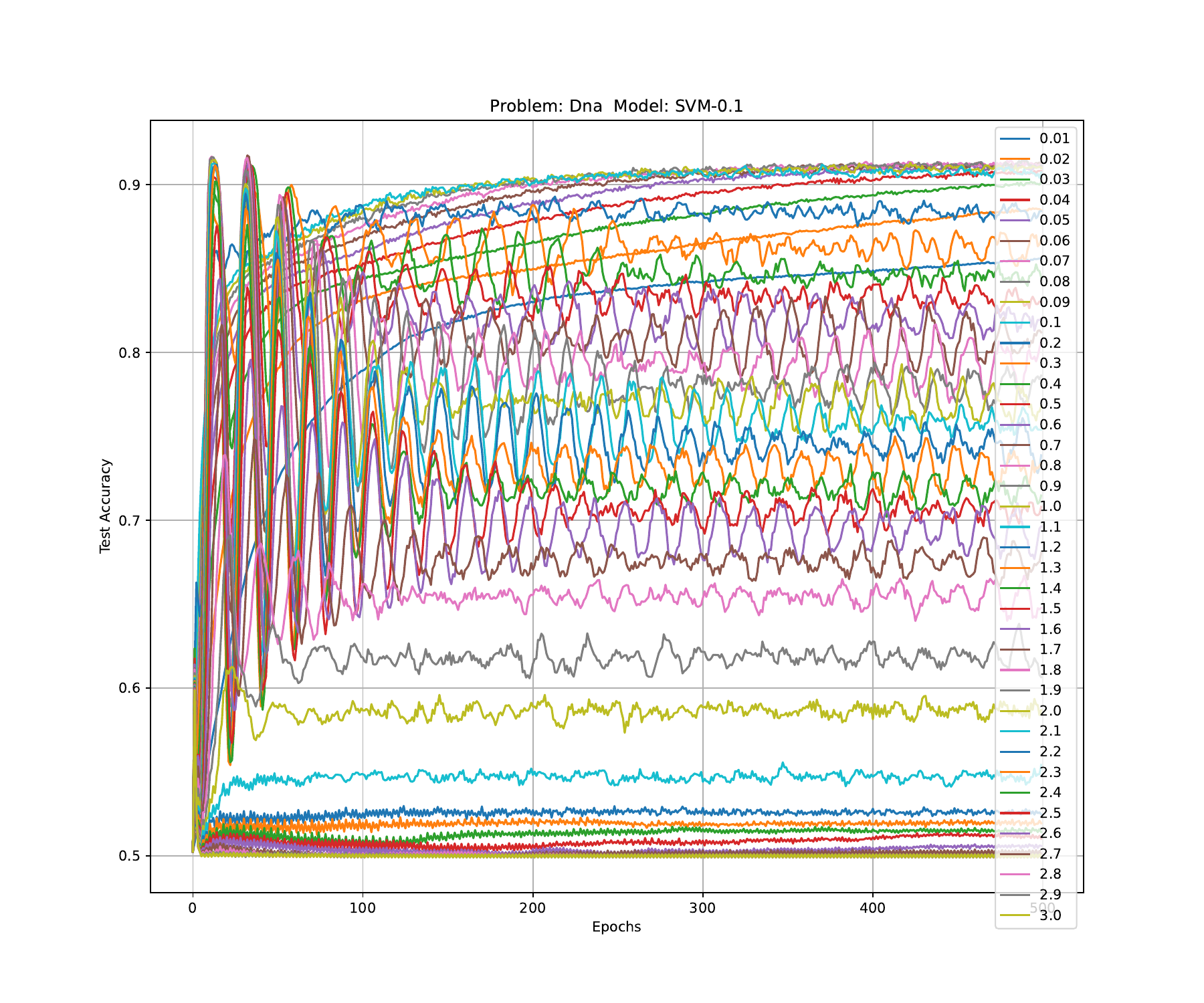} \\[-3mm]
    \includegraphics[clip,trim=70 30 30 30,width=.52\linewidth]{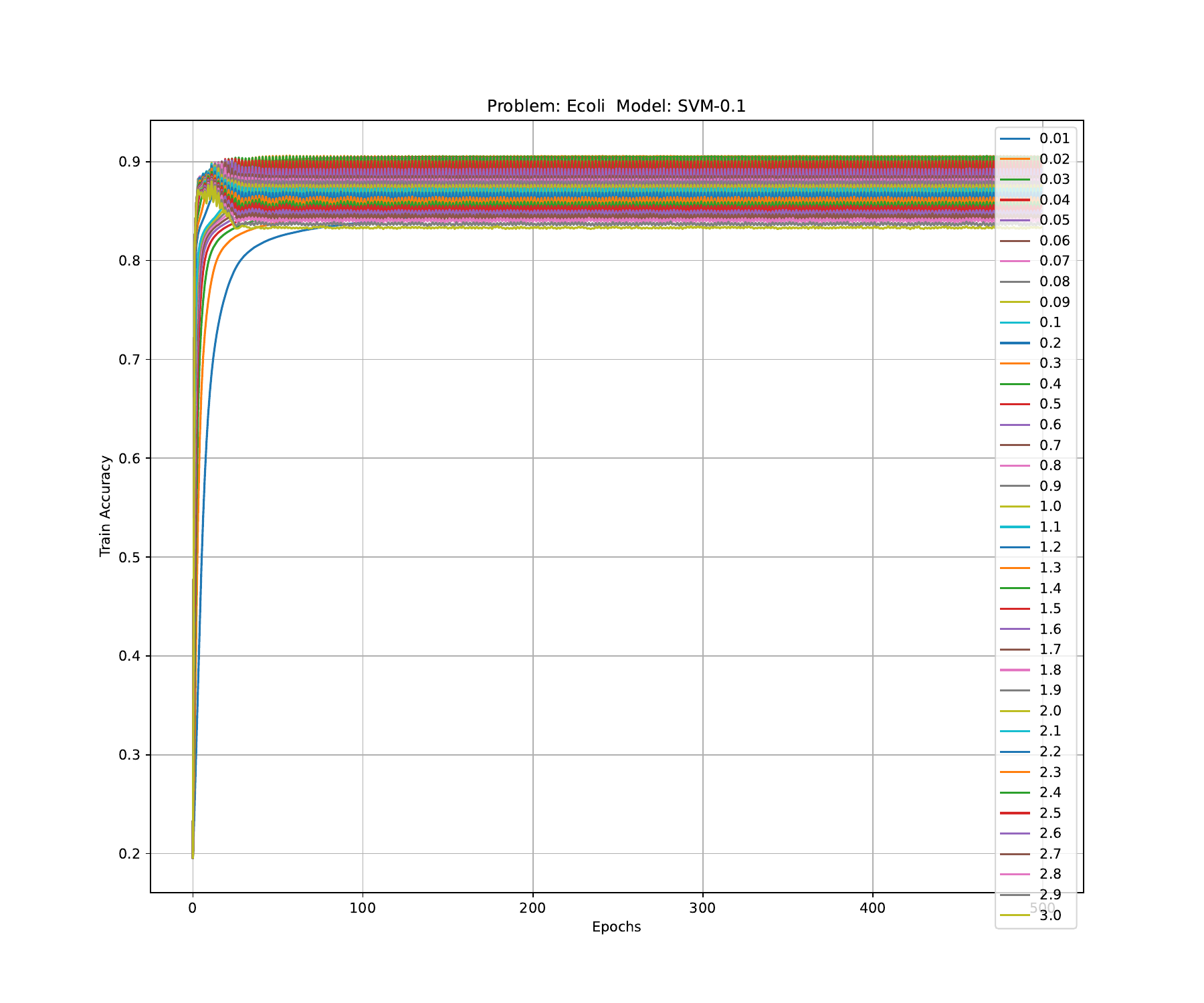} &
    \includegraphics[clip,trim=70 30 30 30,width=.52\linewidth]{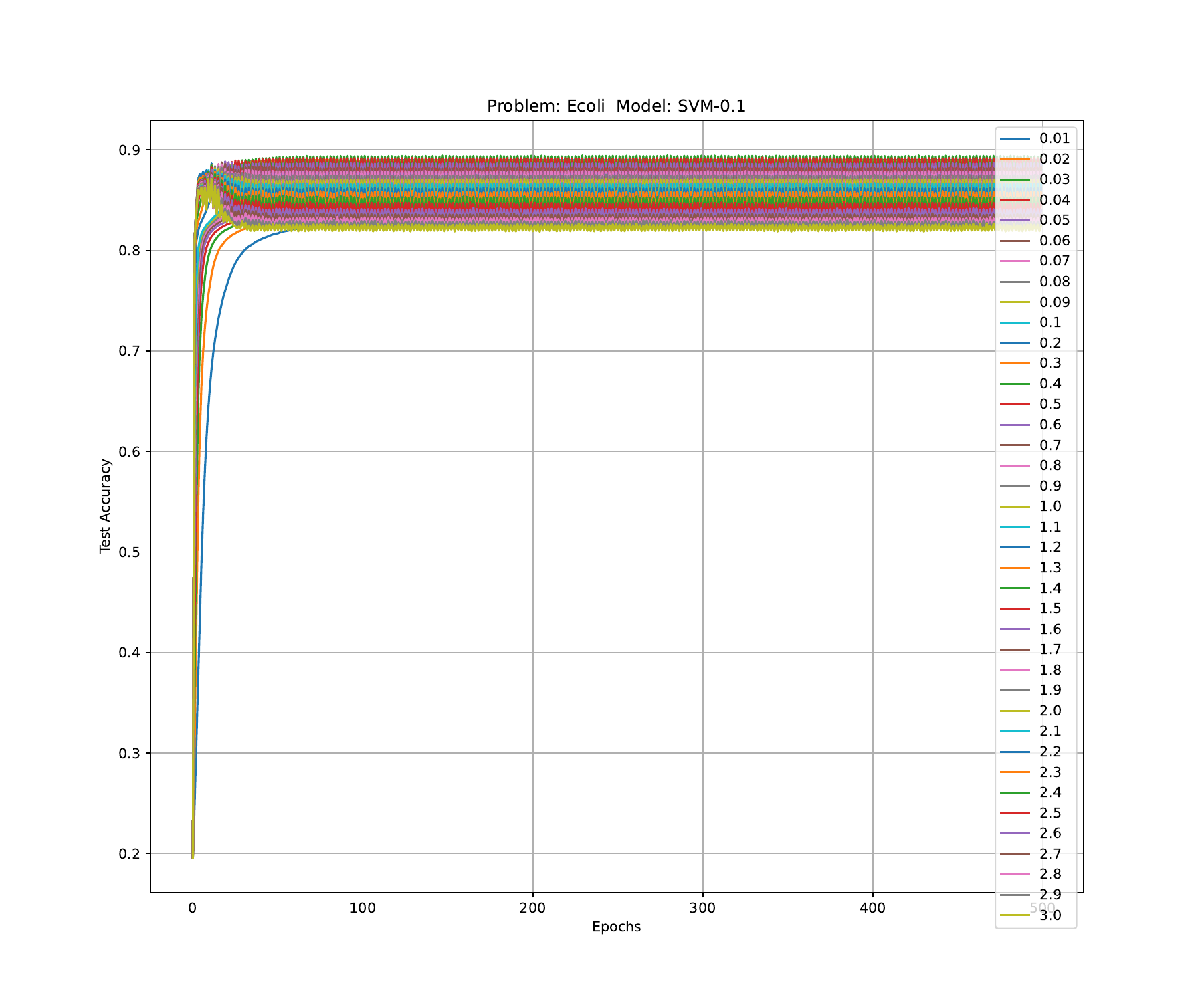} \\[-3mm]
  \end{tabular}
\caption{Accuracies for SVM-0.1 model.}
  \label{fig:accuracies_SVM_0.1_model}
\end{figure*}

\begin{figure*}[p]
  \centering
  \ContinuedFloat
  \begin{tabular}{c@{}c}
    Training Accuracy & Validation Accuracy \\
    \includegraphics[clip,trim=70 30 30 30,width=.52\linewidth]{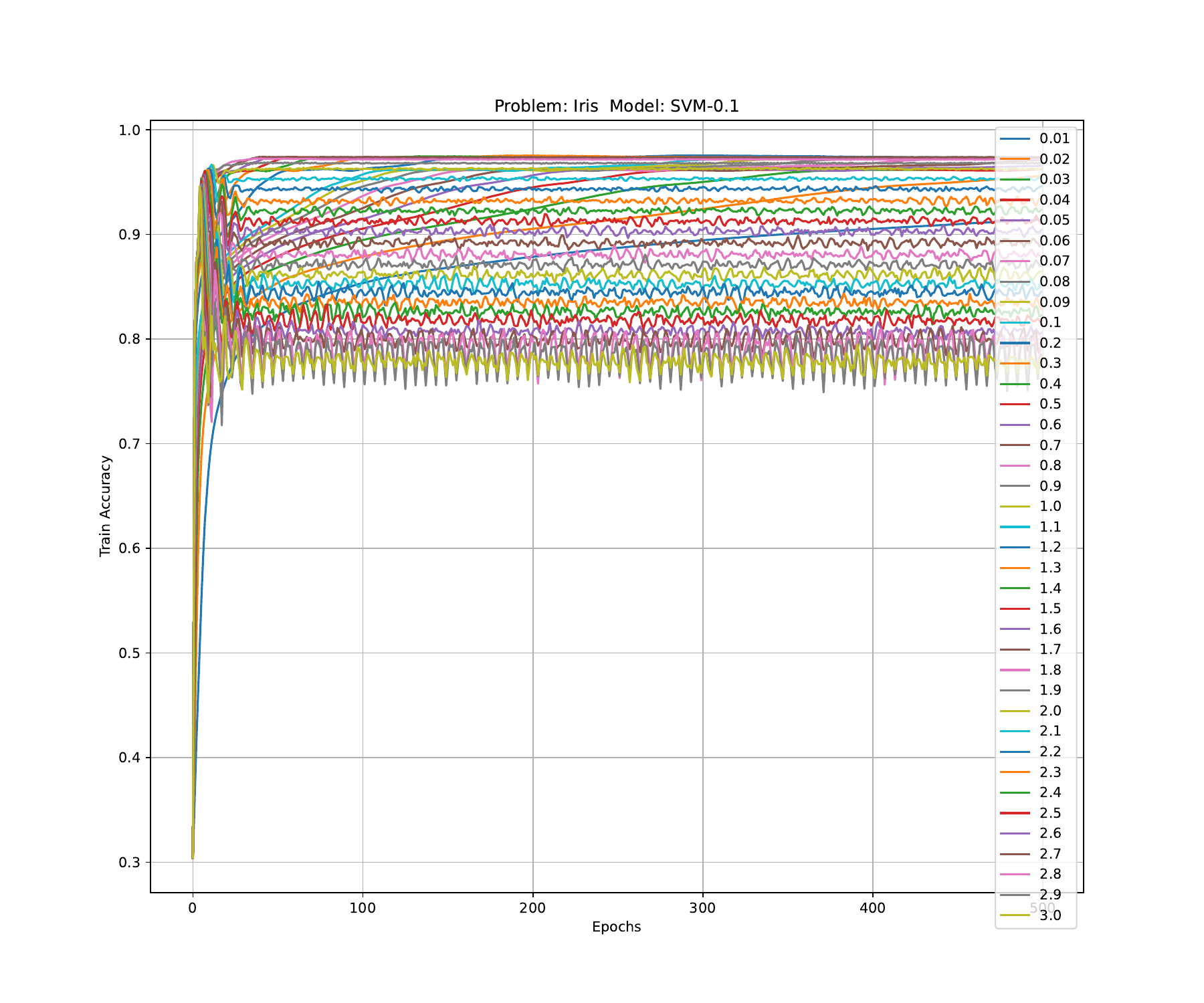} &
    \includegraphics[clip,trim=70 30 30 30,width=.52\linewidth]{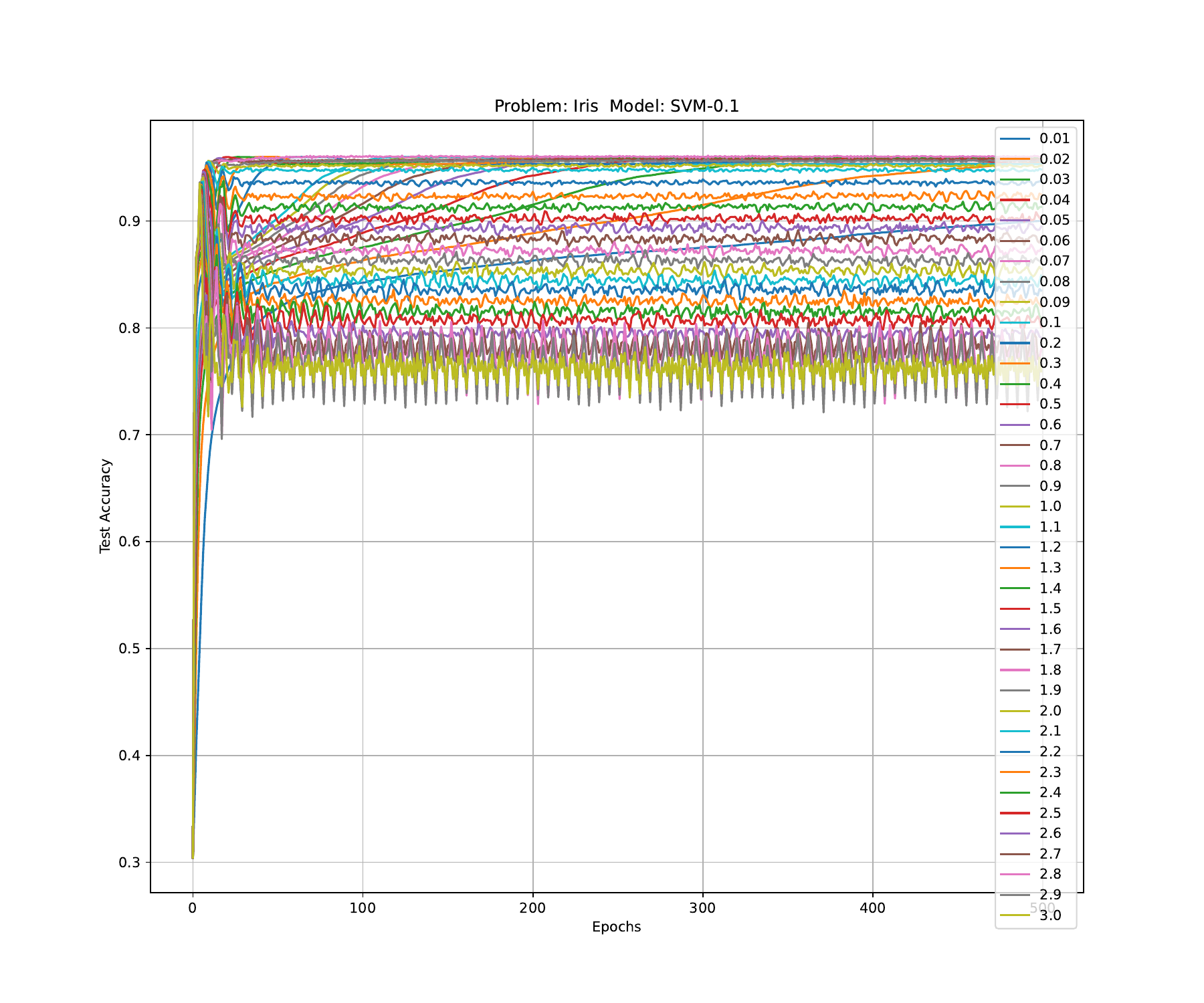} \\[-3mm]
    \includegraphics[clip,trim=70 30 30 30,width=.52\linewidth]{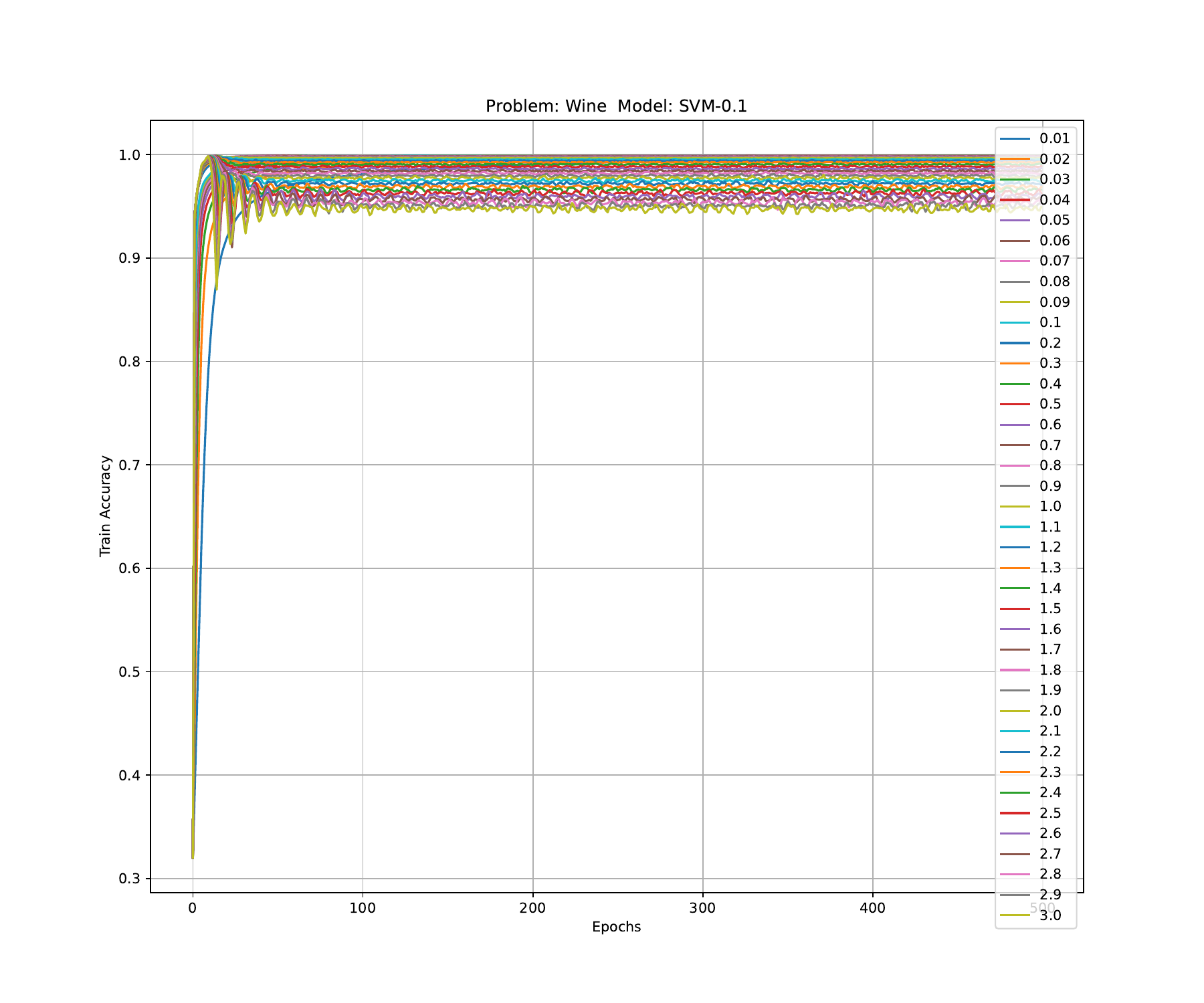} &
    \includegraphics[clip,trim=70 30 30 30,width=.52\linewidth]{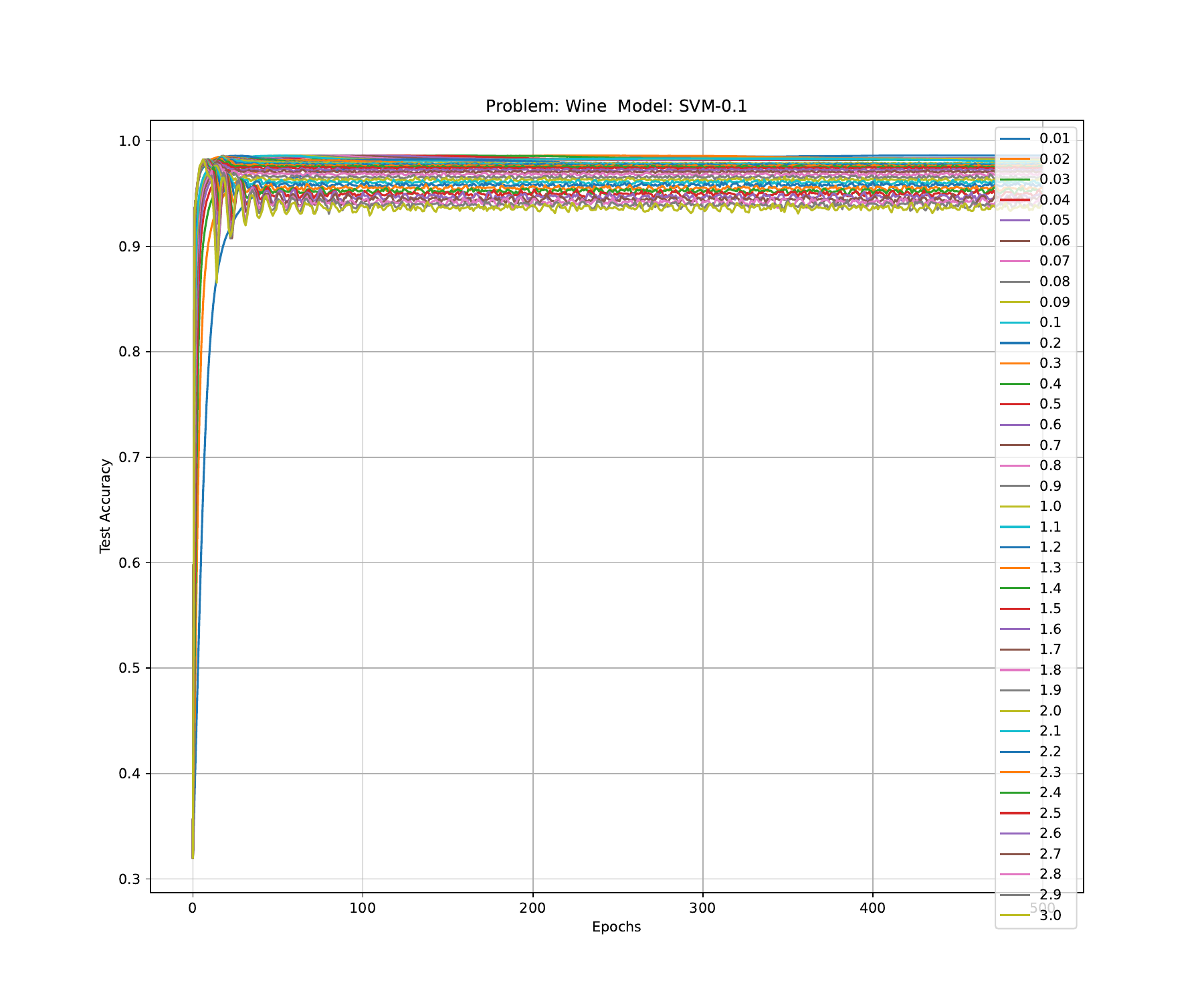} \\[-3mm]
  \end{tabular}
\caption{Accuracies for SVM-0.1 model (continued).}
\end{figure*}

\FloatBarrier

\pagebreak
\section{Success Probabilities}
\label{app:succ_probabilities}

In this section we report the success probabilities $P(i, a, \eta)$,
for $a\in\{0.85, 0.9, 0.95\}$ for all the problems and models used in
this article. We do have the plots for all other values of $a$ between
0.75 and 1.0 (in steps of 0.005), which we would be happy to provide
upon request.

\subsection{Success Probabilities for $a=0.85$}

In Figures~\ref{fig:success_probability_NN_ReLU_model_0.85}---\ref{fig:success_probability_SVM_0.1_model_0.85} we present plots of the success probabilities
$P(i, 0.85, \eta)$ (that is for $a=0.85$) for all the problems and models
used in this article.

\begin{figure*}[p]
  \centering
\begin{tabular}{c@{}c}
    Training Success Probability & Validation Success Probability \\
    \includegraphics[clip,trim=0 0 0 0,width=.52\linewidth]{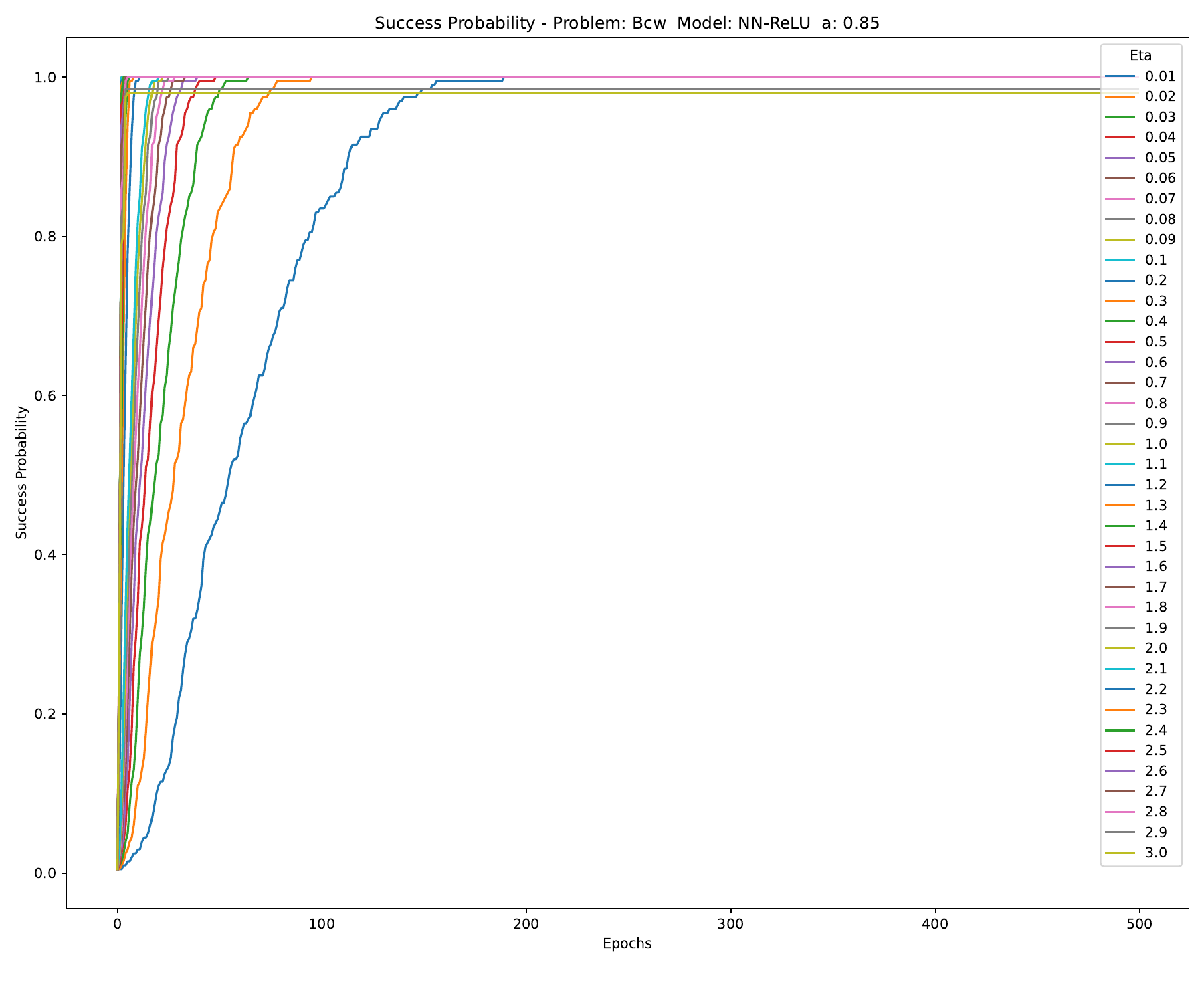} &
    \includegraphics[clip,trim=0 0 0 0,width=.52\linewidth]{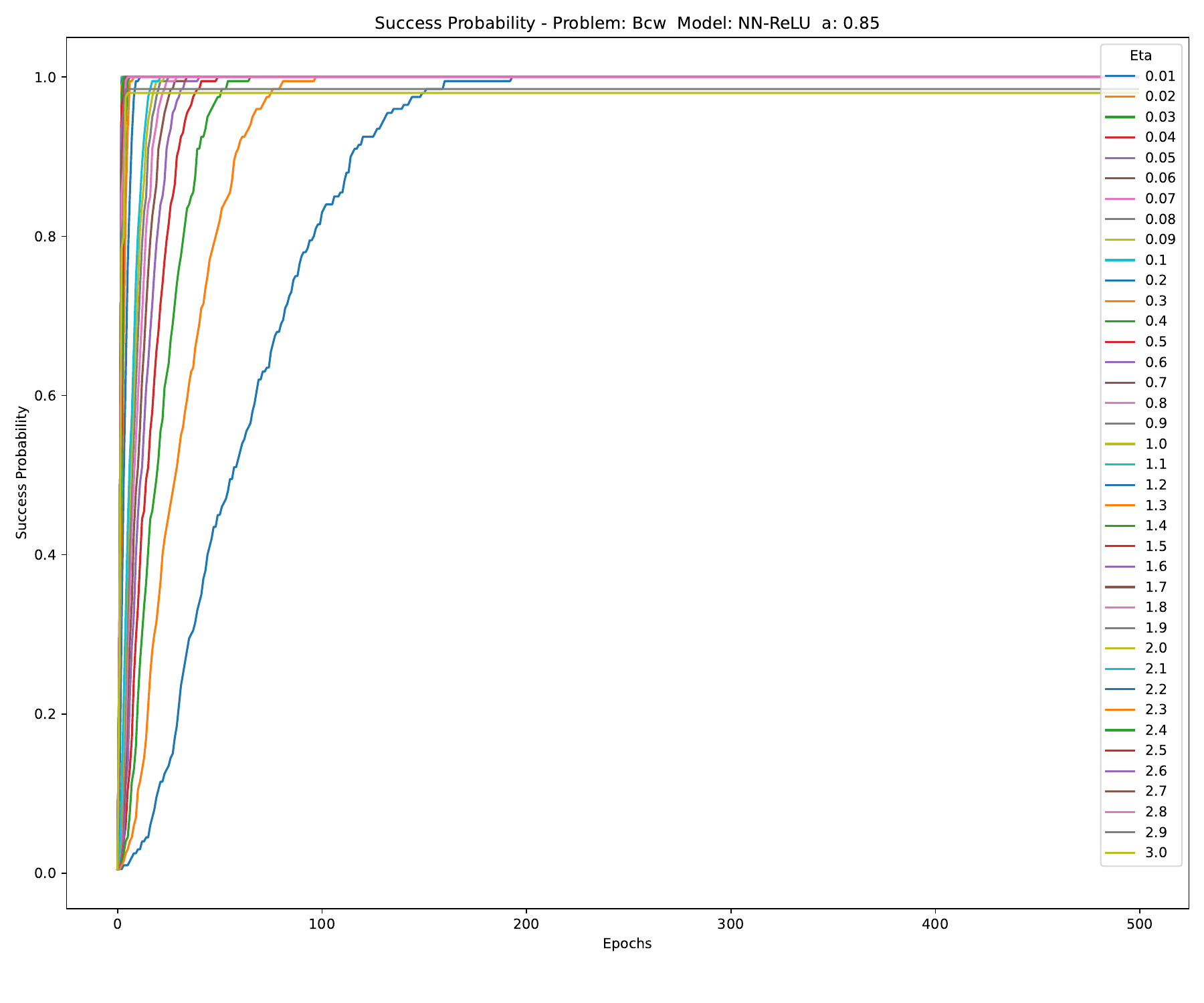} \\[-3mm]
    \includegraphics[clip,trim=0 0 0 0,width=.52\linewidth]{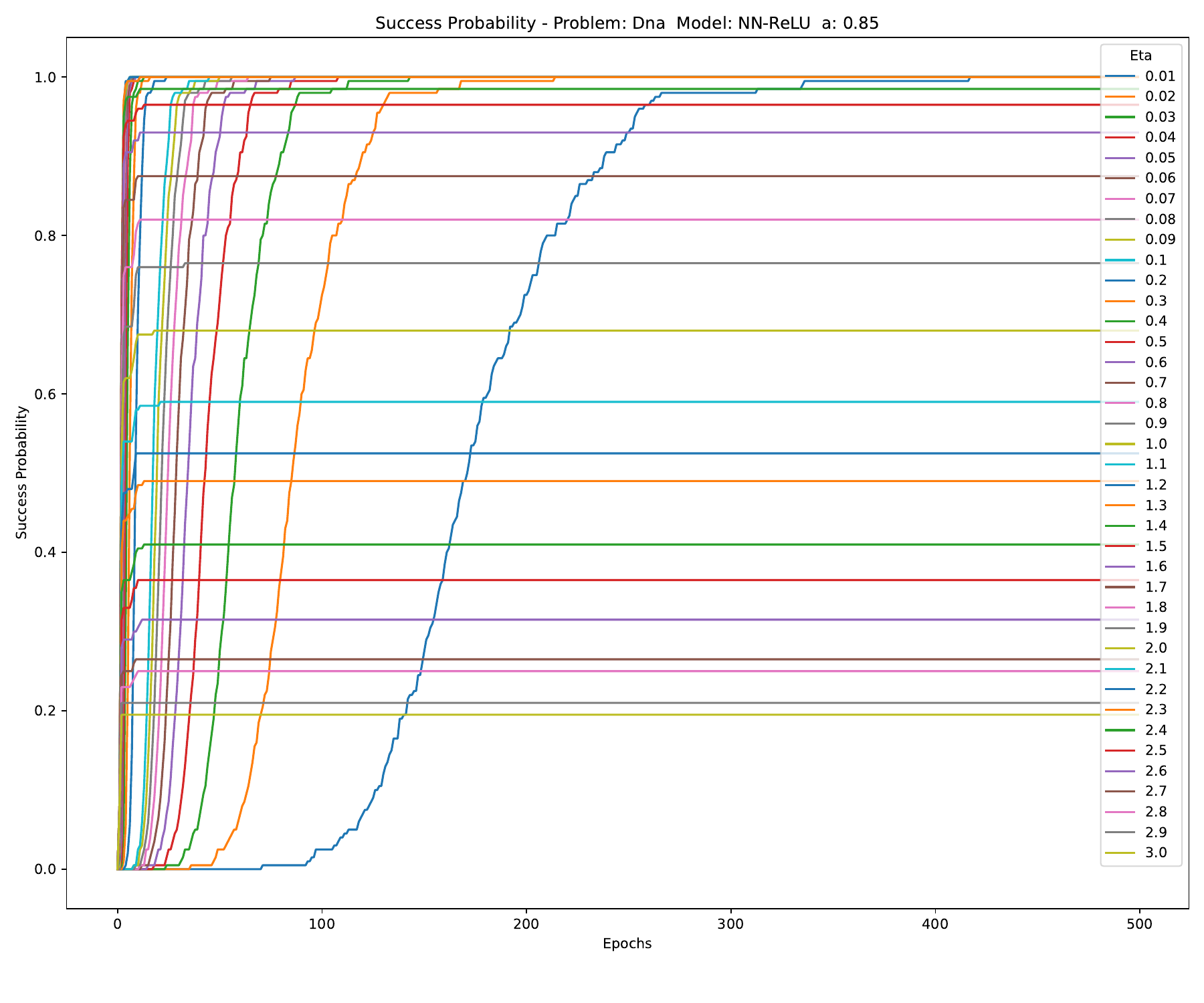} &
    \includegraphics[clip,trim=0 0 0 0,width=.52\linewidth]{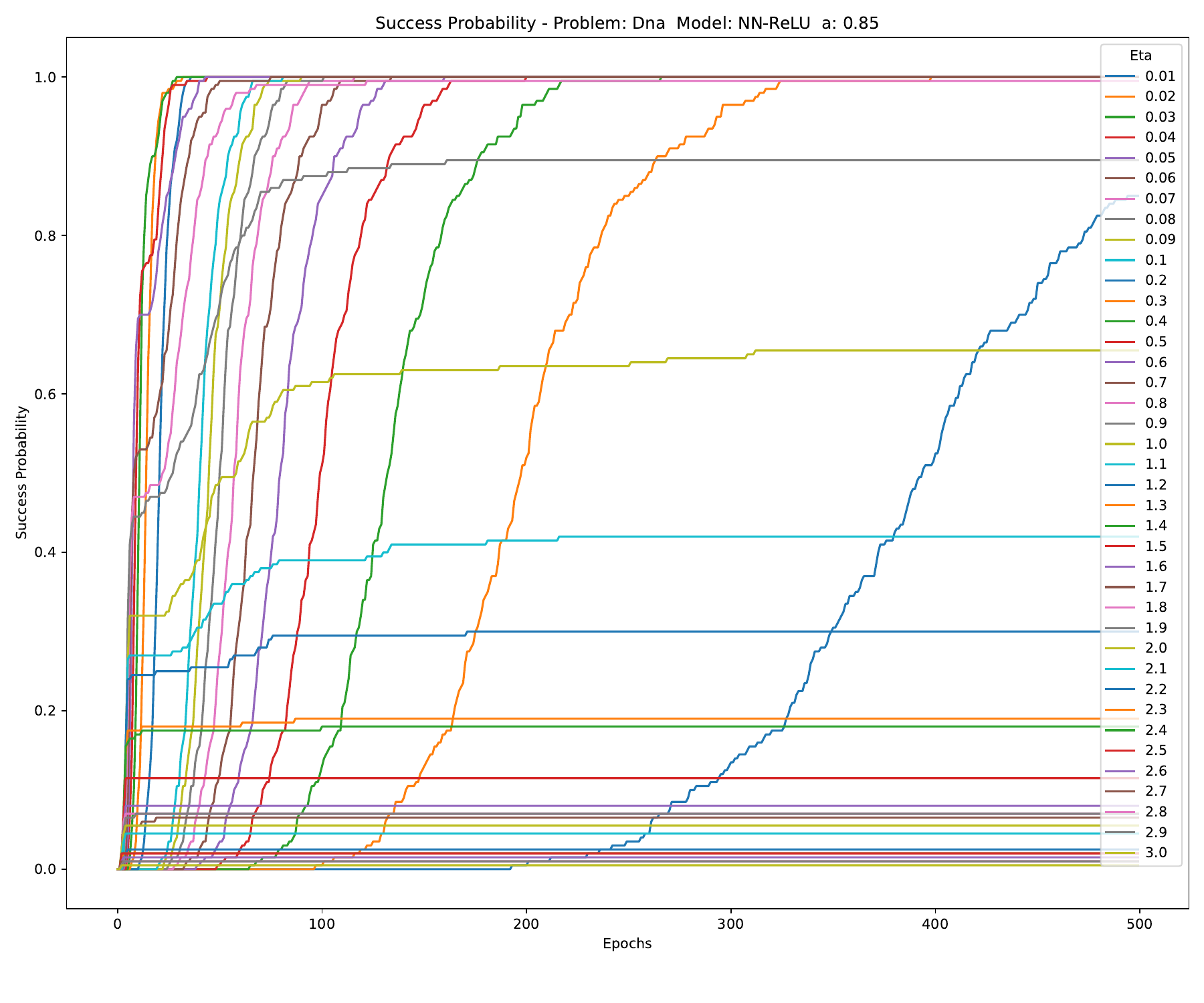} \\[-3mm]
    \includegraphics[clip,trim=0 0 0 0,width=.52\linewidth]{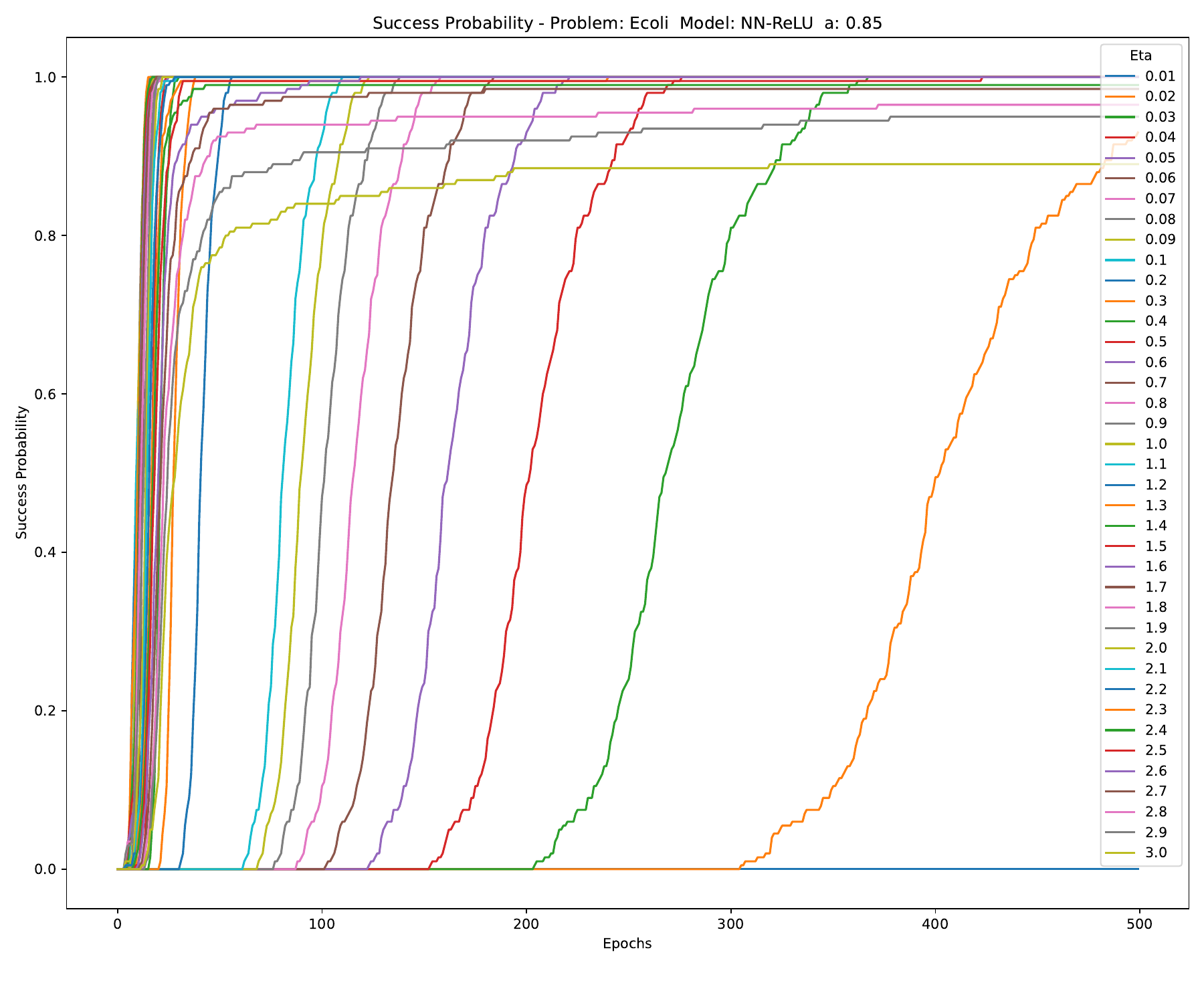} &
    \includegraphics[clip,trim=0 0 0 0,width=.52\linewidth]{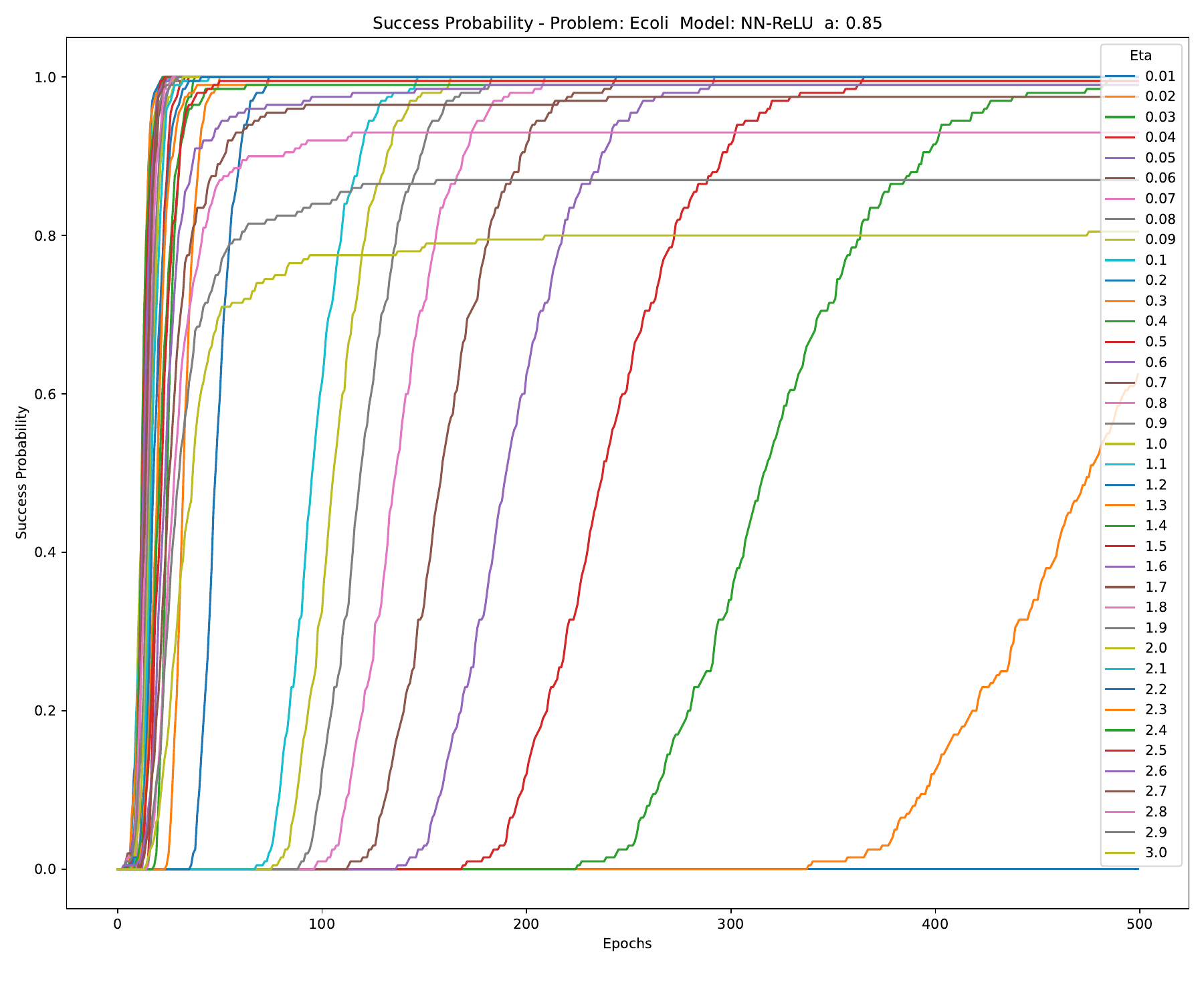} \\[-3mm]
  \end{tabular}
\caption{Success probability ($a=0.85$) for NN-ReLU model.}
  \label{fig:success_probability_NN_ReLU_model_0.85}
\end{figure*}

\begin{figure*}[p]
  \centering
  \ContinuedFloat
  \begin{tabular}{c@{}c}
    Training Success Probability & Validation Success Probability \\
    \includegraphics[clip,trim=0 0 0 0,width=.52\linewidth]{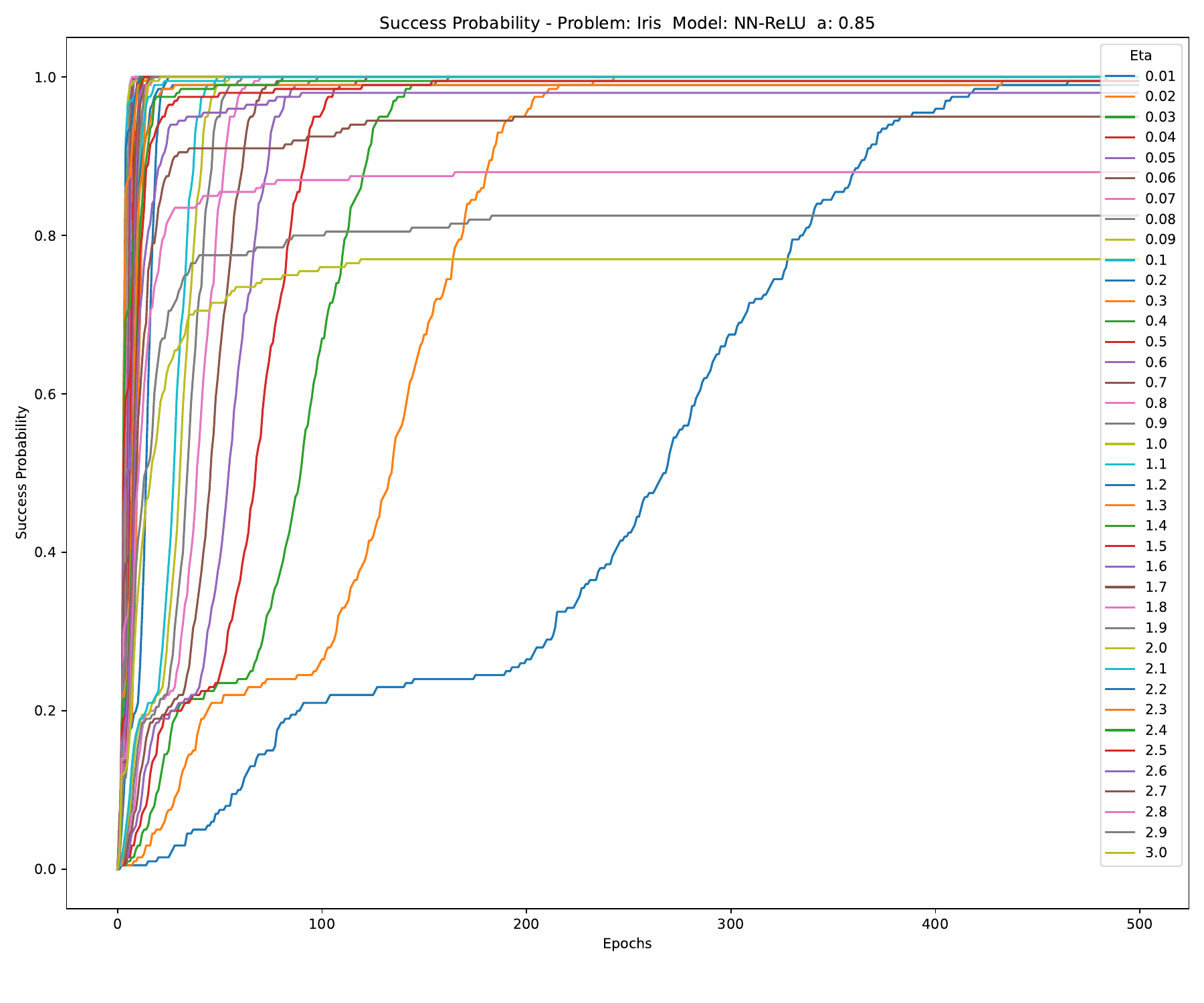} &
    \includegraphics[clip,trim=0 0 0 0,width=.52\linewidth]{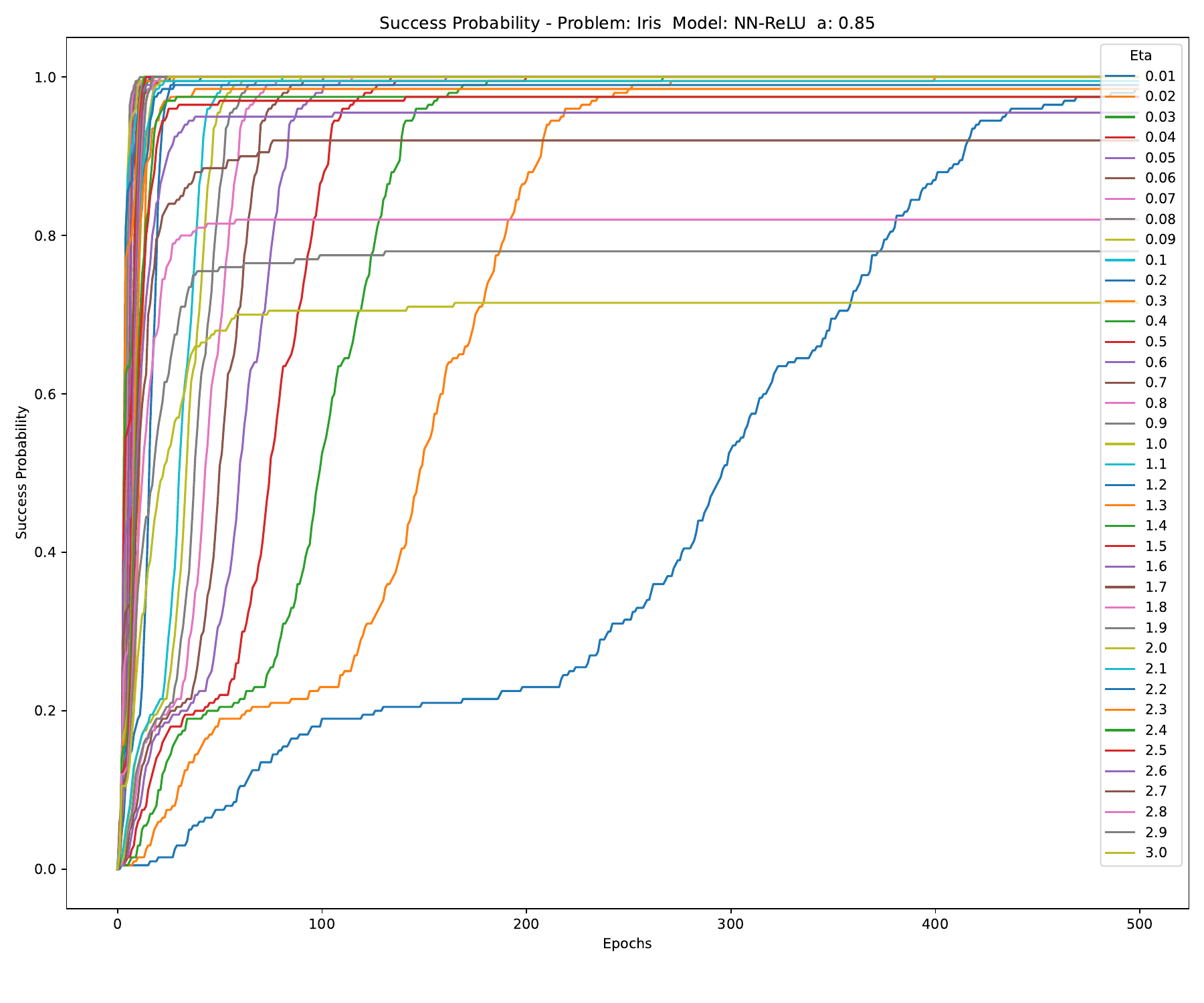} \\[-3mm]
    \includegraphics[clip,trim=0 0 0 0,width=.52\linewidth]{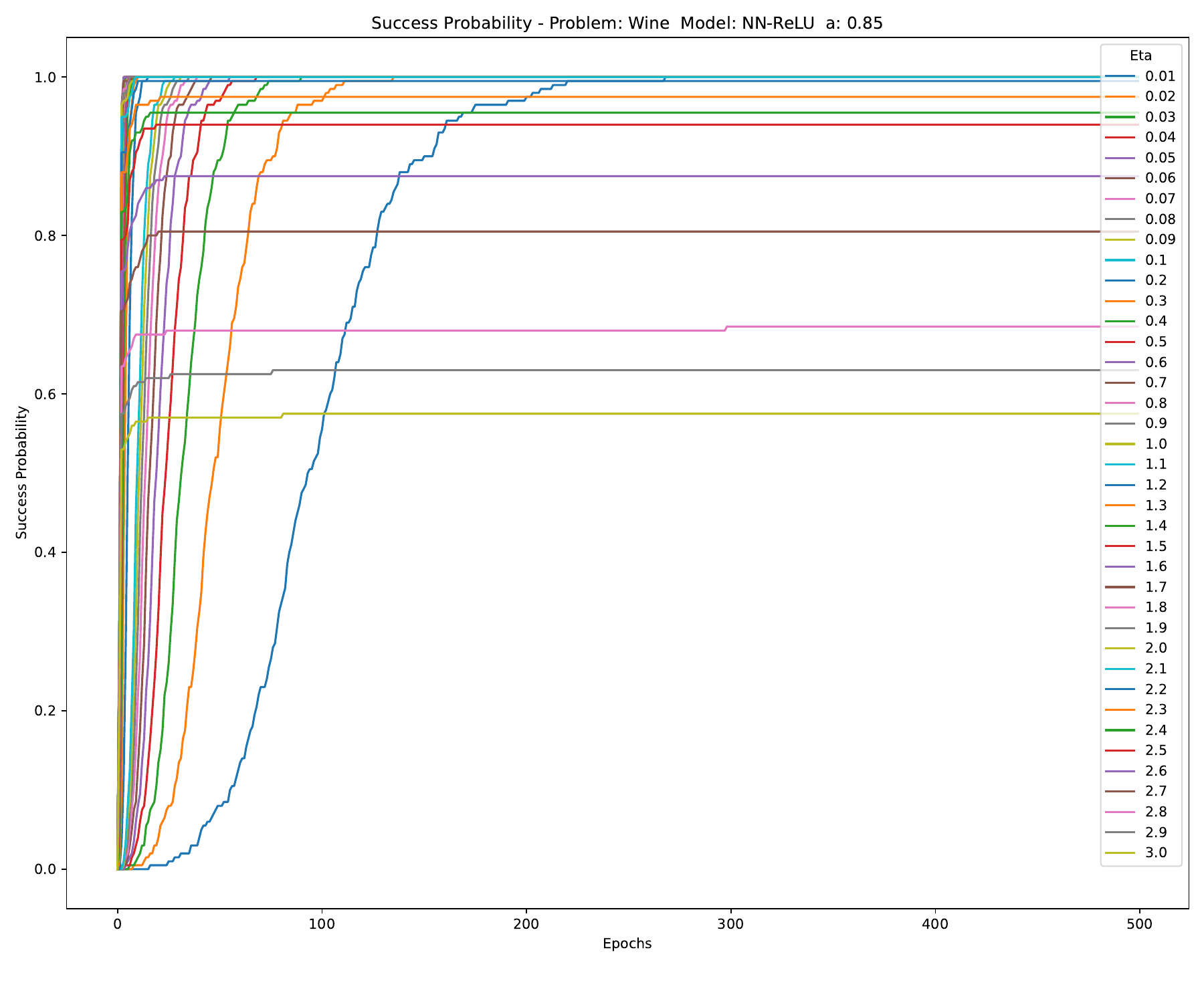} &
    \includegraphics[clip,trim=0 0 0 0,width=.52\linewidth]{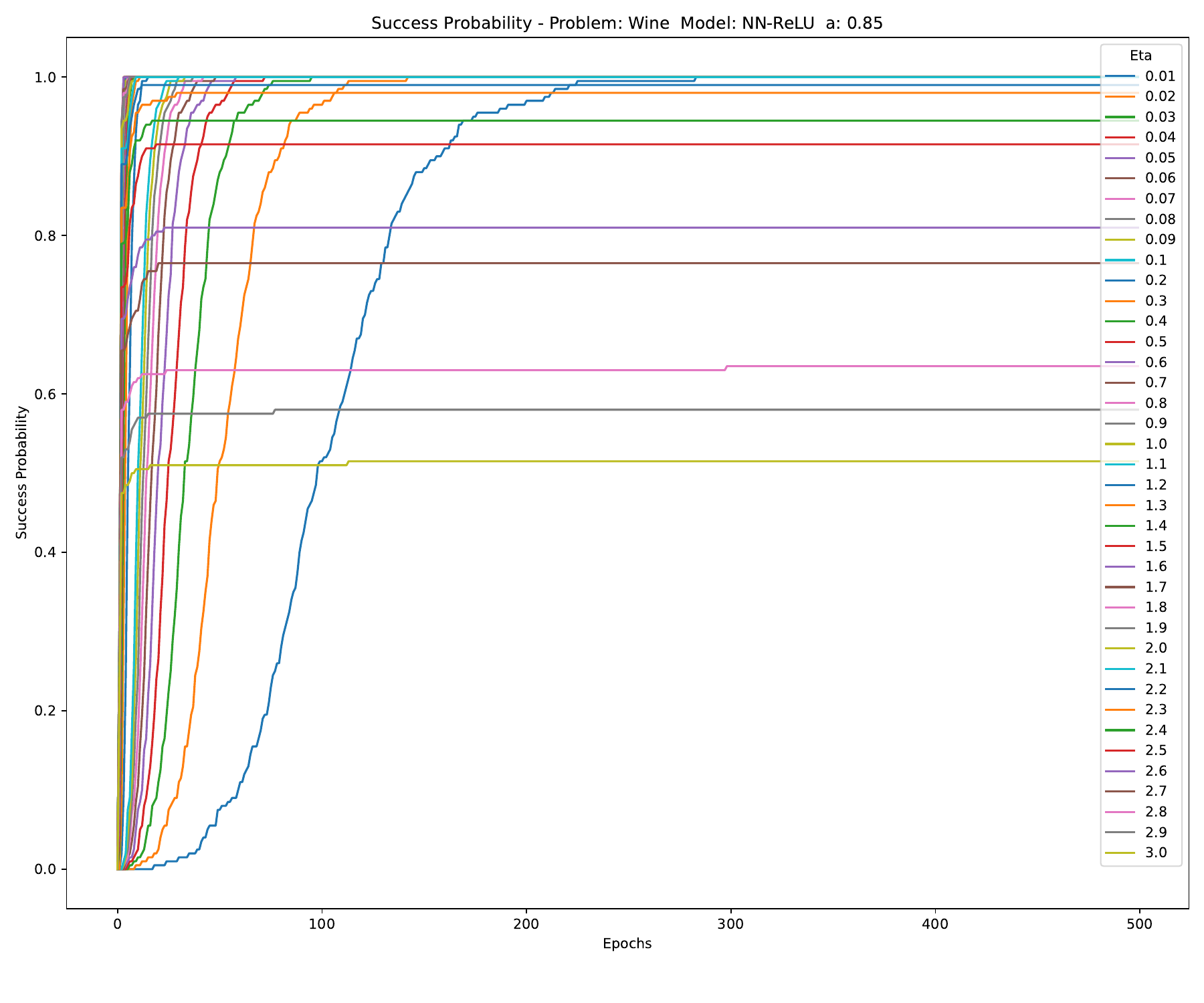} \\[-3mm]
  \end{tabular}
\caption{Success probability ($a=0.85$) for NN-ReLU model (continued).}
\end{figure*}

\begin{figure*}[p]
  \centering
  \begin{tabular}{c@{}c}
    Training Success Probability & Validation Success Probability \\
    \includegraphics[clip,trim=0 0 0 0,width=.52\linewidth]{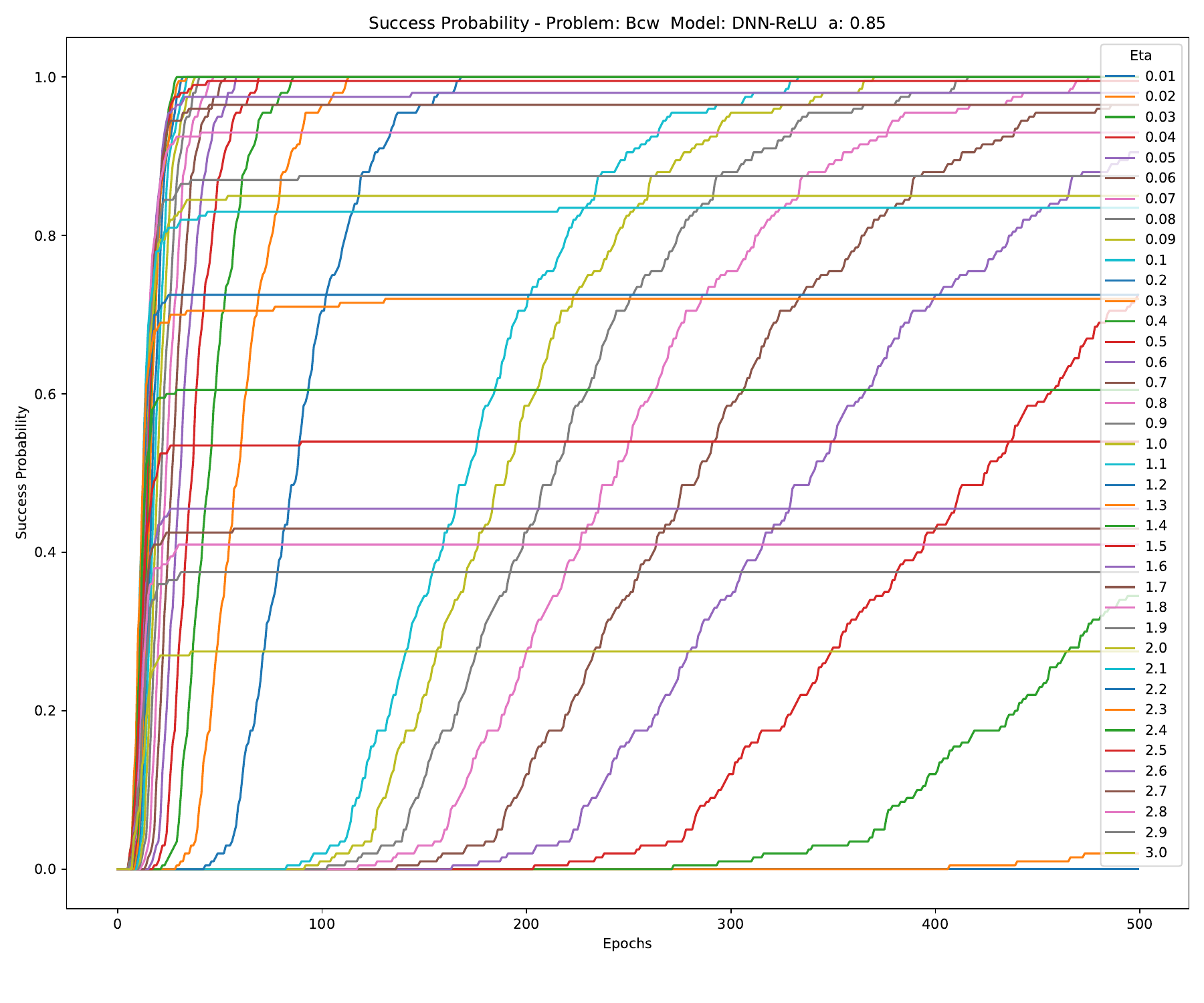} &
    \includegraphics[clip,trim=0 0 0 0,width=.52\linewidth]{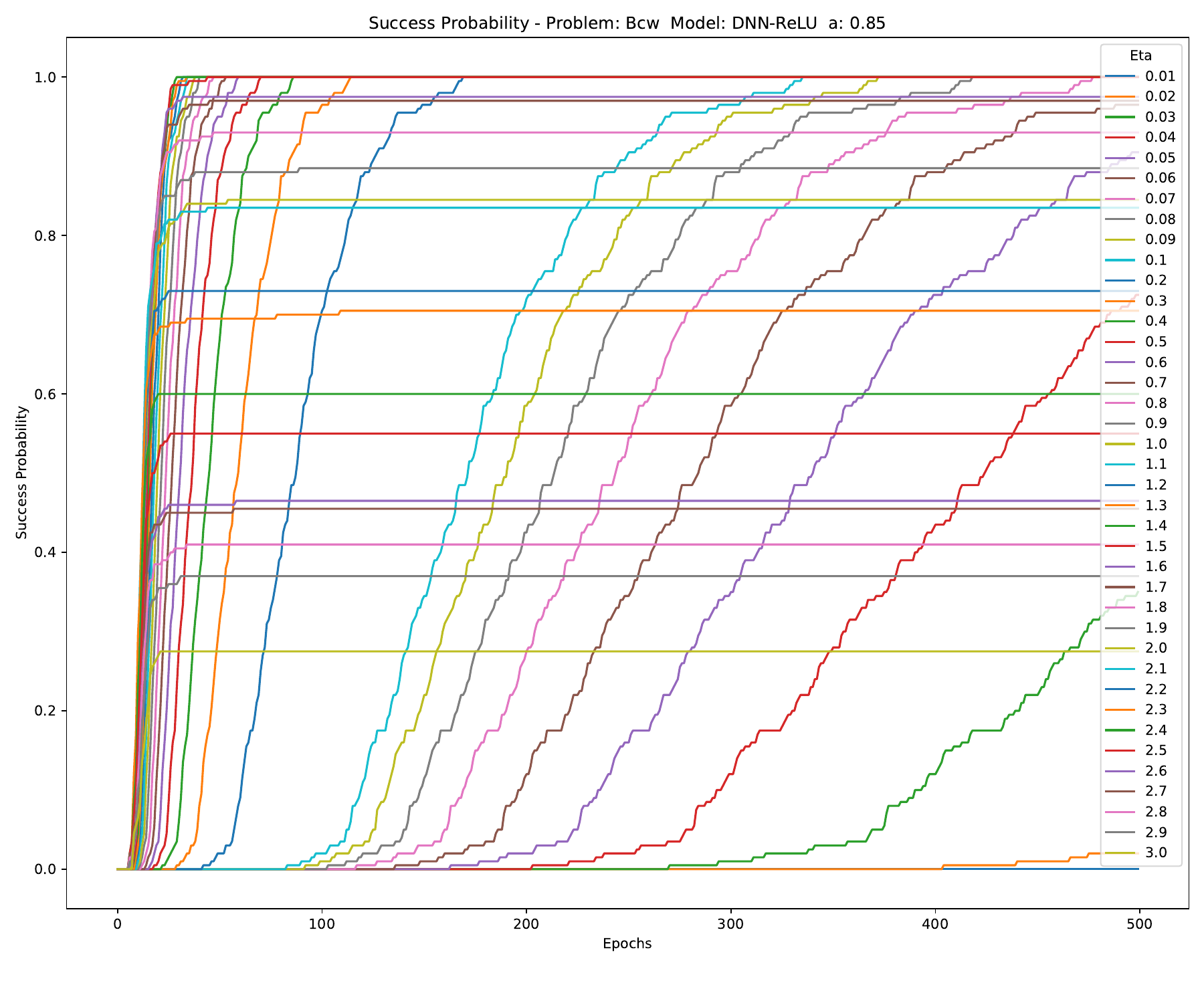} \\[-3mm]
    \includegraphics[clip,trim=0 0 0 0,width=.52\linewidth]{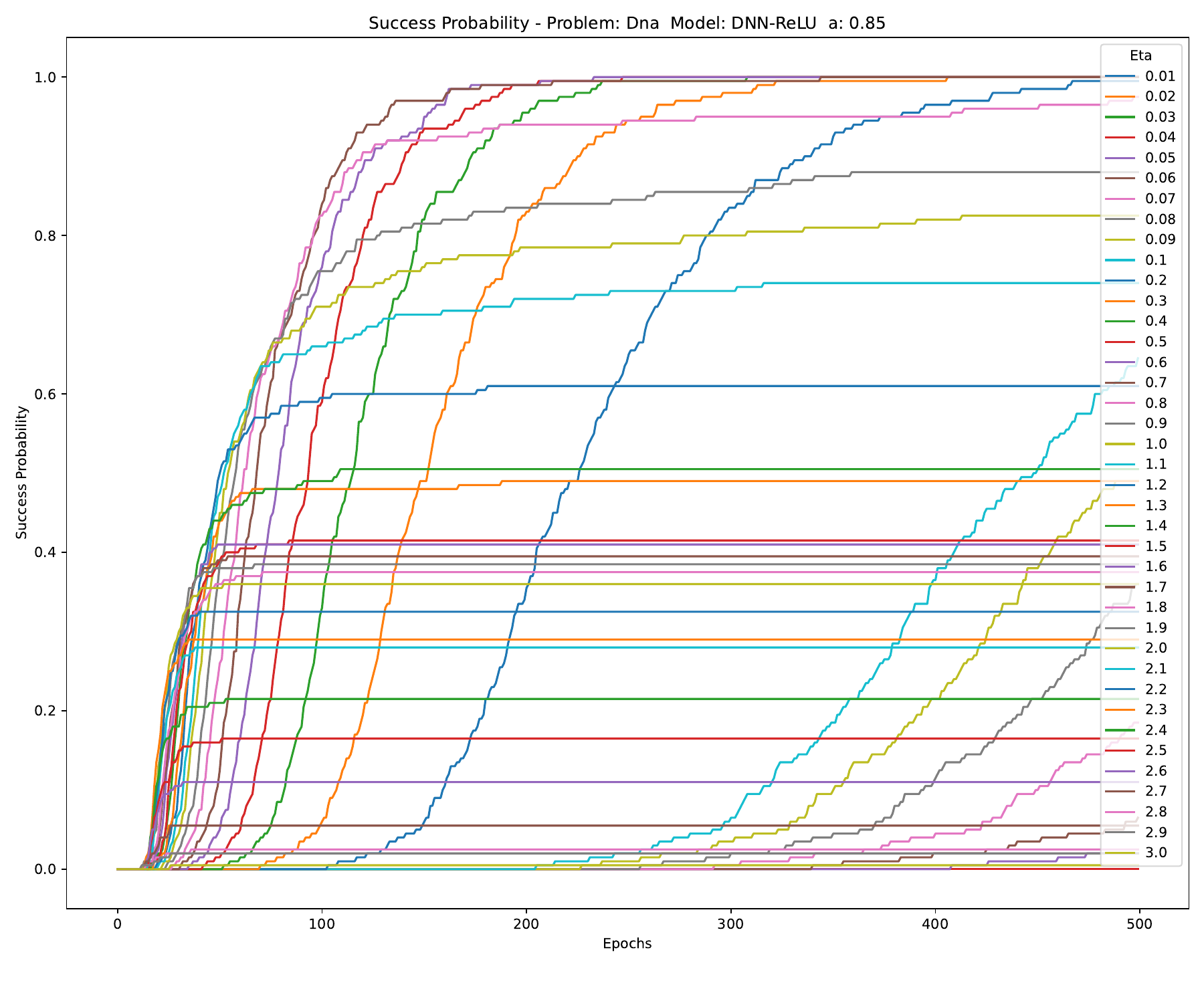} &
    \includegraphics[clip,trim=0 0 0 0,width=.52\linewidth]{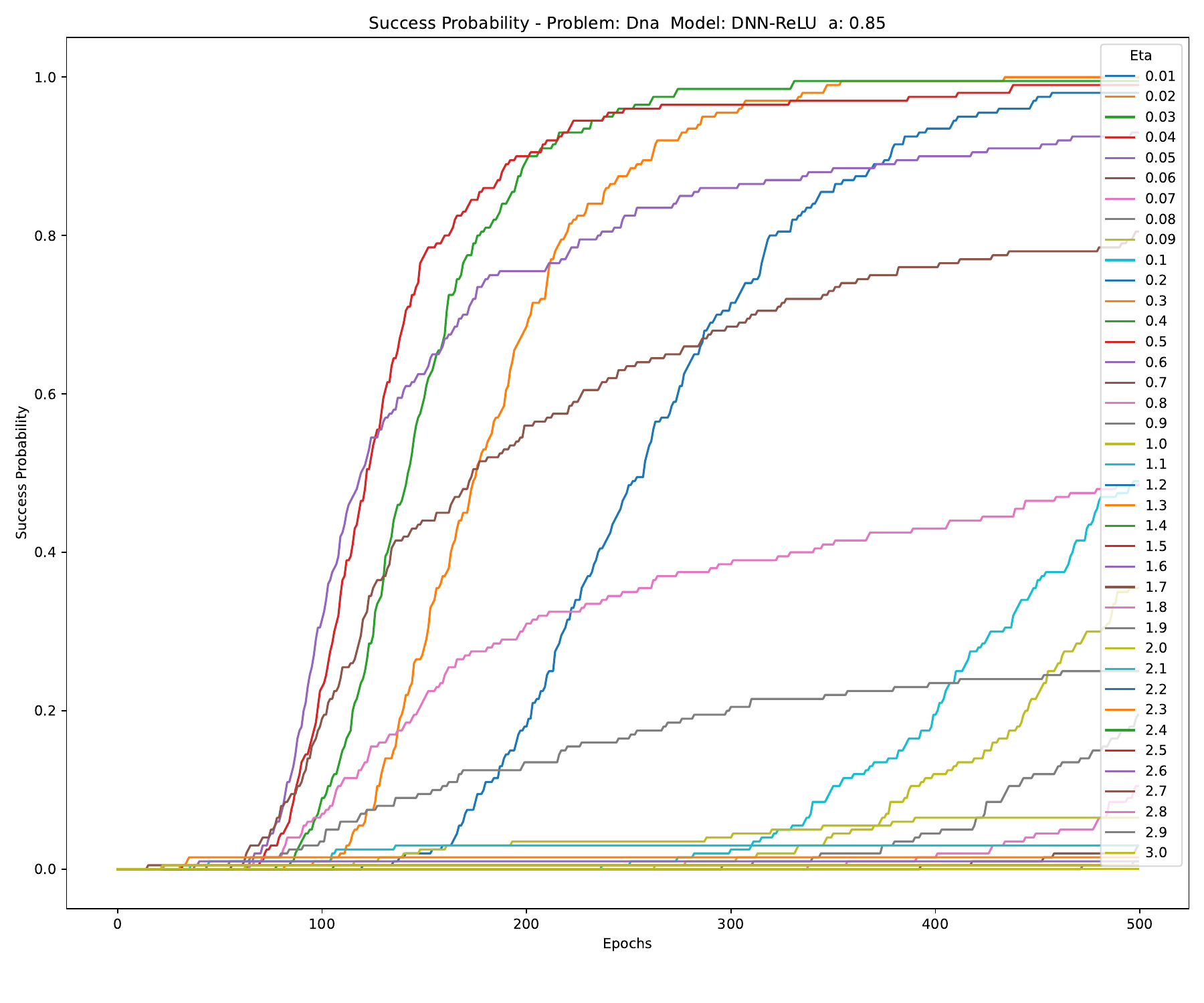} \\[-3mm]
    \includegraphics[clip,trim=0 0 0 0,width=.52\linewidth]{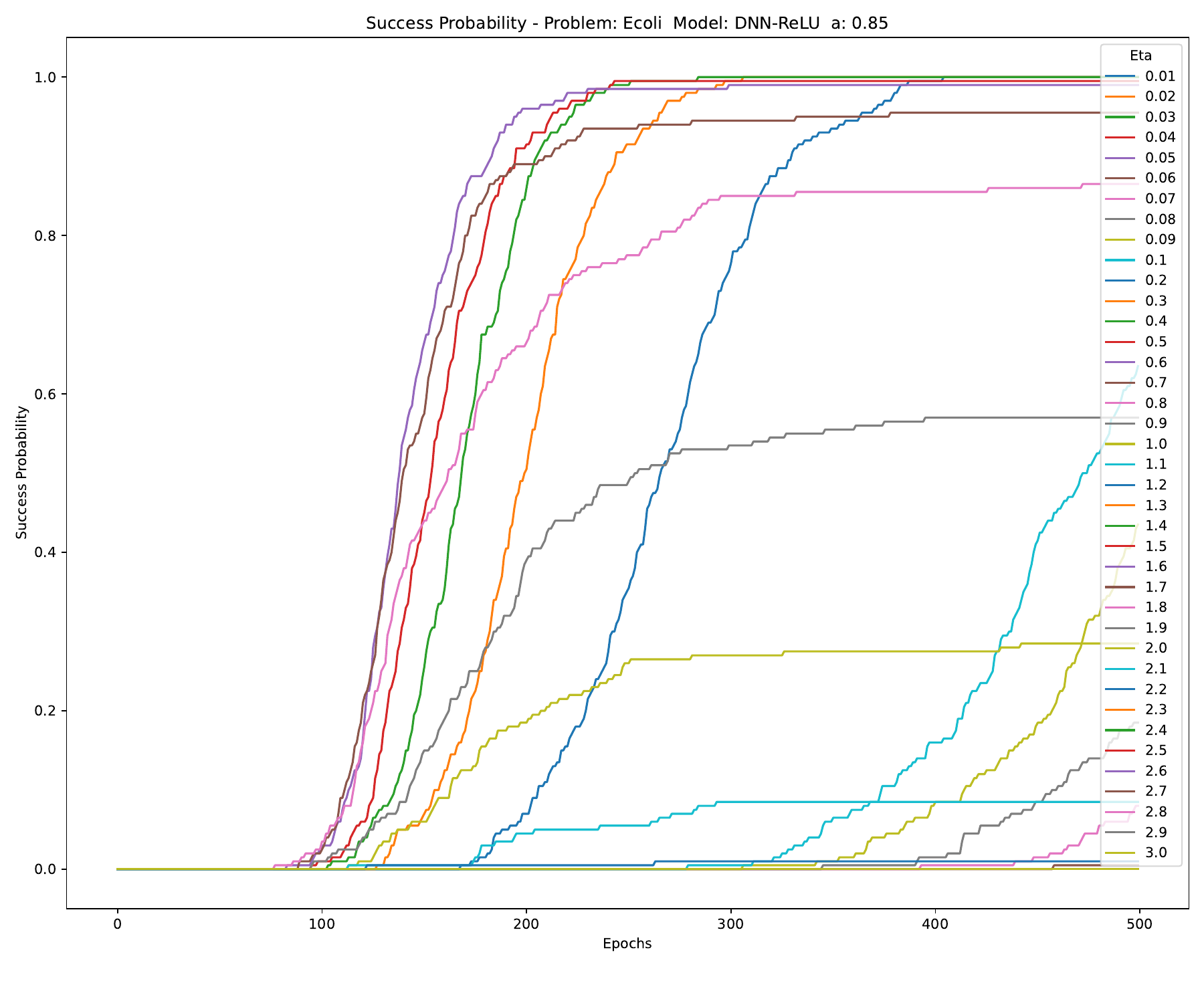} &
    \includegraphics[clip,trim=0 0 0 0,width=.52\linewidth]{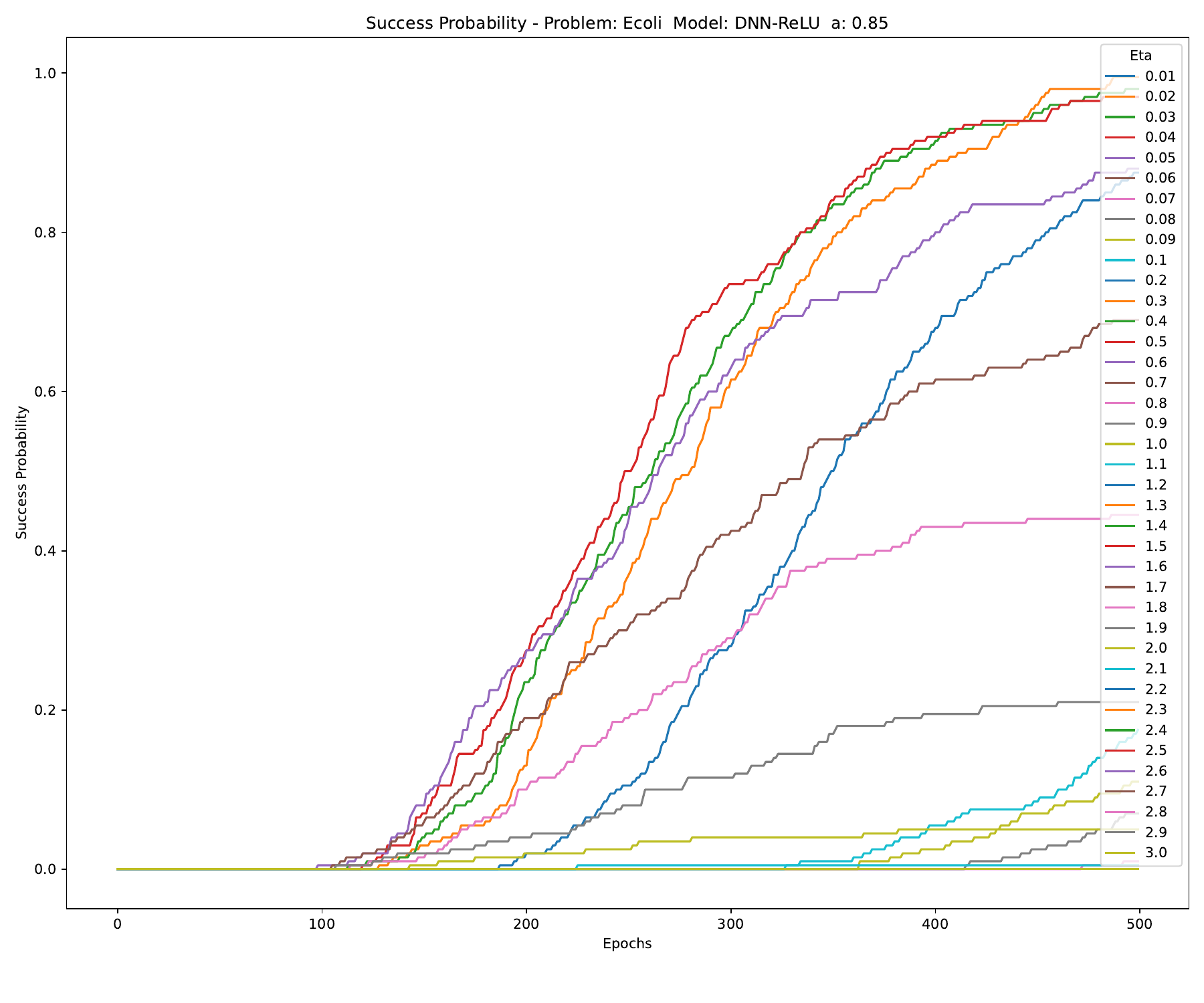} \\[-3mm]
  \end{tabular}
\caption{Success probability ($a=0.85$) for DNN-ReLU model.}
  \label{fig:success_probability_DNN_ReLU_model_0.85}
\end{figure*}

\begin{figure*}[p]
  \centering
  \ContinuedFloat
  \begin{tabular}{c@{}c}
    Training Success Probability & Validation Success Probability \\
    \includegraphics[clip,trim=0 0 0 0,width=.52\linewidth]{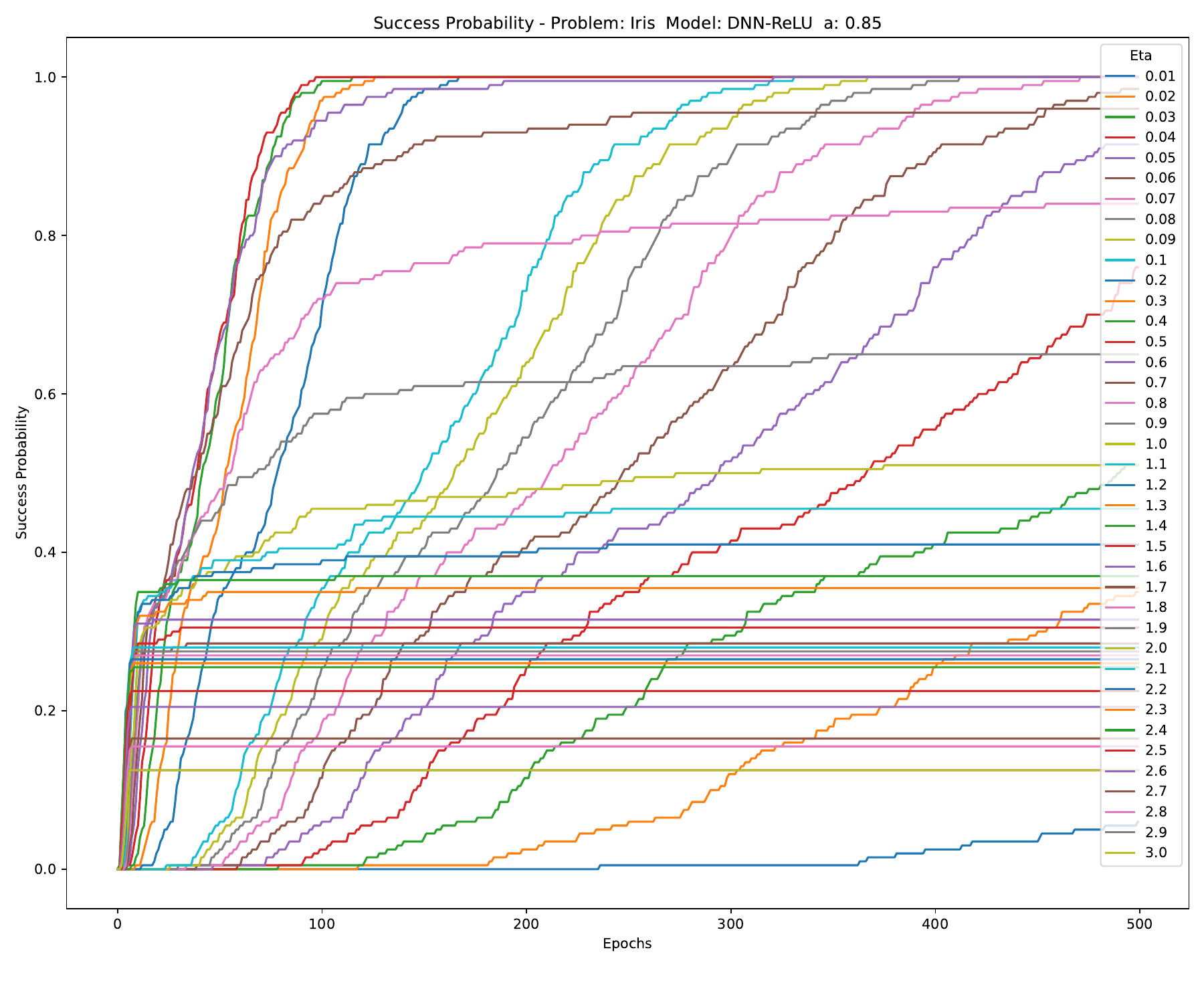} &
    \includegraphics[clip,trim=0 0 0 0,width=.52\linewidth]{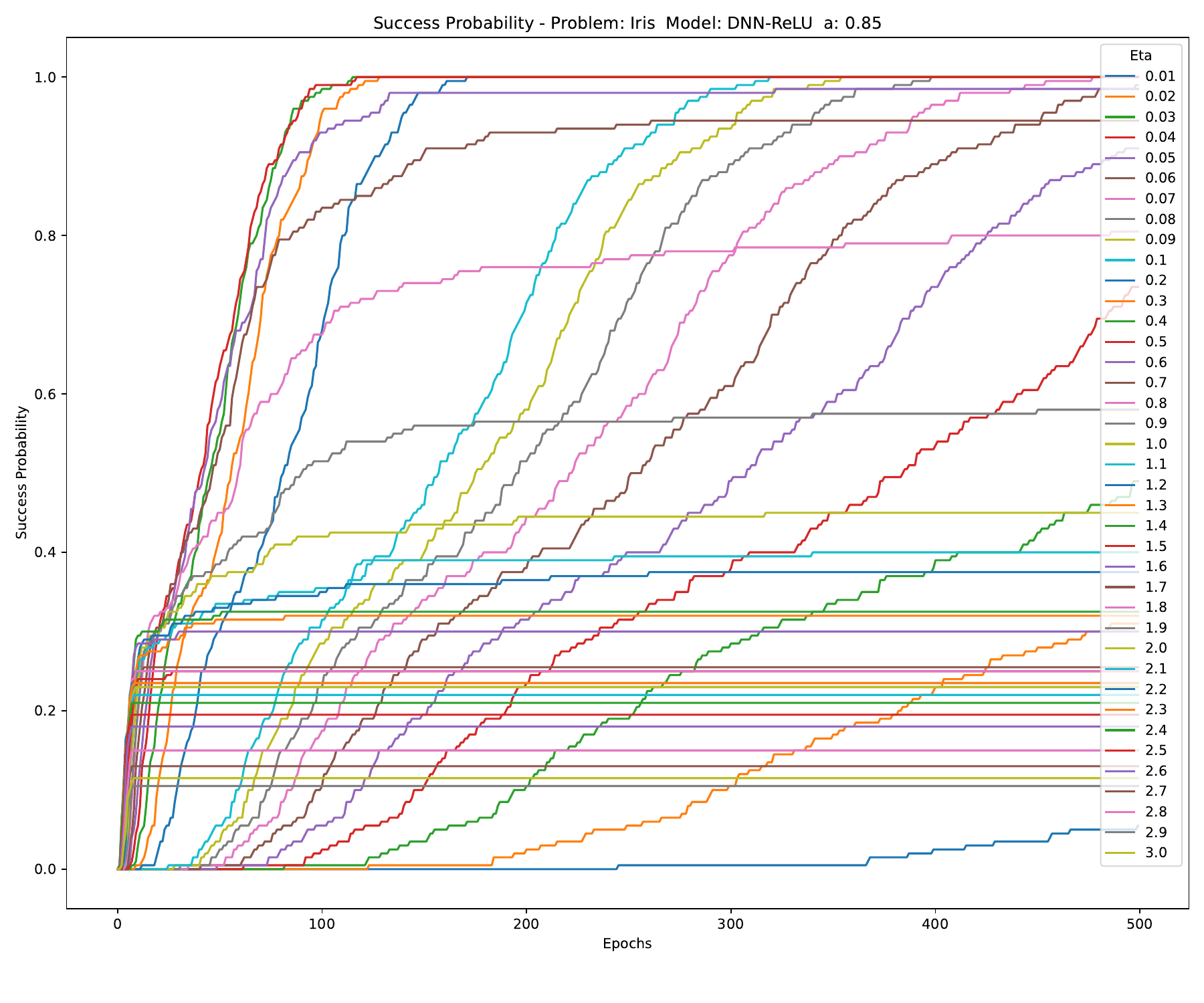} \\[-3mm]
    \includegraphics[clip,trim=0 0 0 0,width=.52\linewidth]{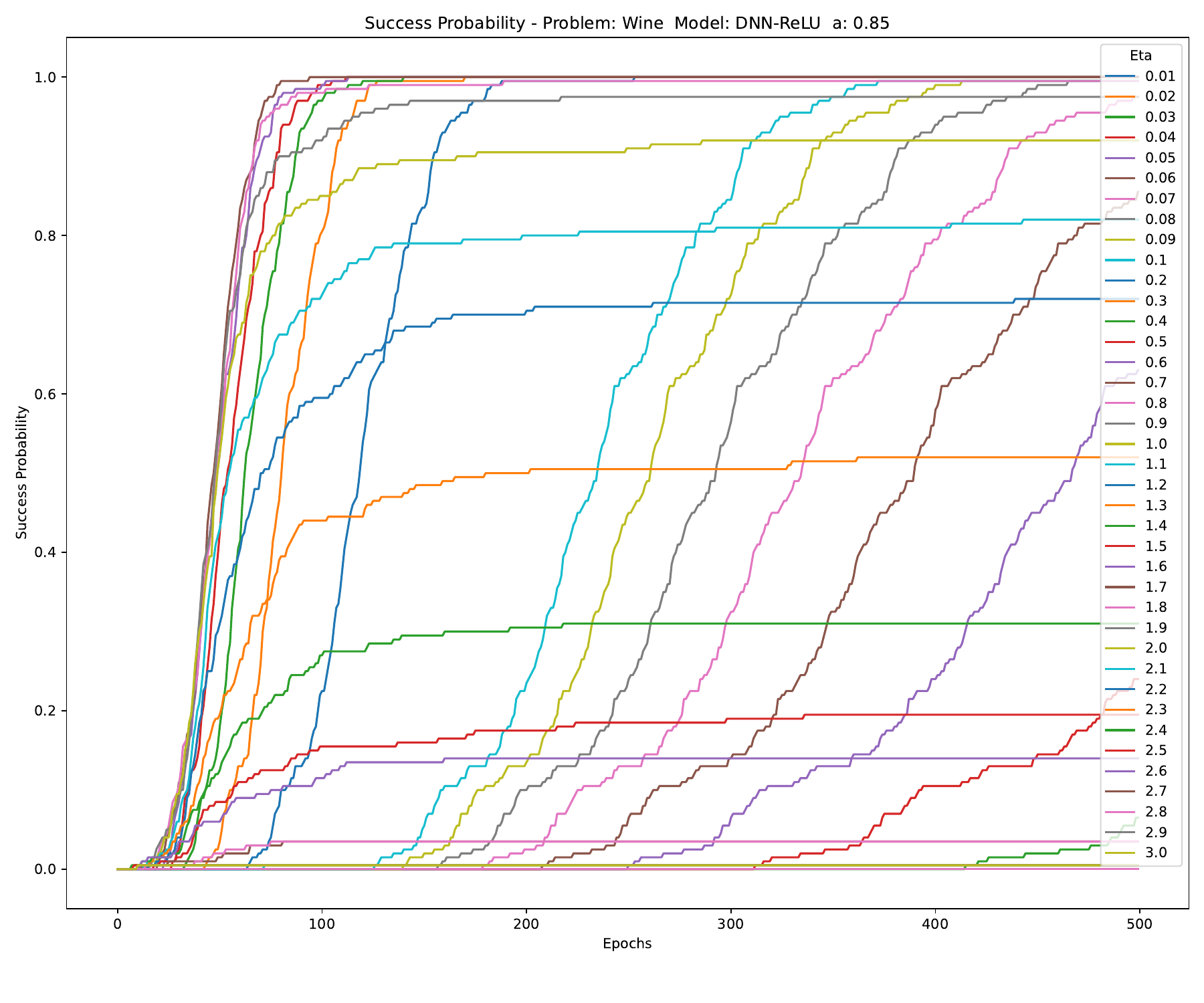} &
    \includegraphics[clip,trim=0 0 0 0,width=.52\linewidth]{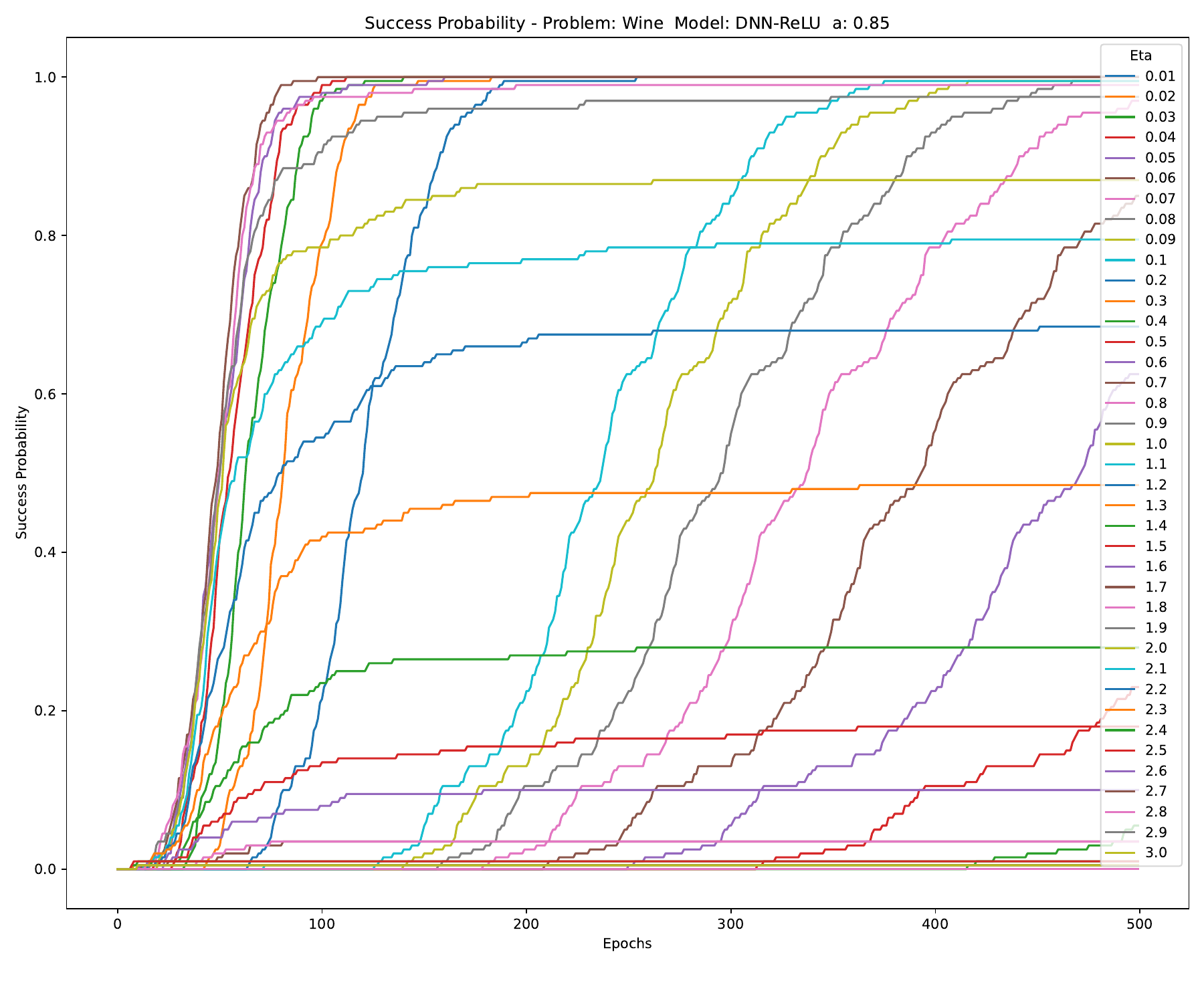} \\[-3mm]
  \end{tabular}
\caption{Success probability ($a=0.85$) for DNN-ReLU model (continued).}
\end{figure*}

\begin{figure*}[p]
  \centering
  \begin{tabular}{c@{}c}
    Training Success Probability & Validation Success Probability \\
    \includegraphics[clip,trim=0 0 0 0,width=.52\linewidth]{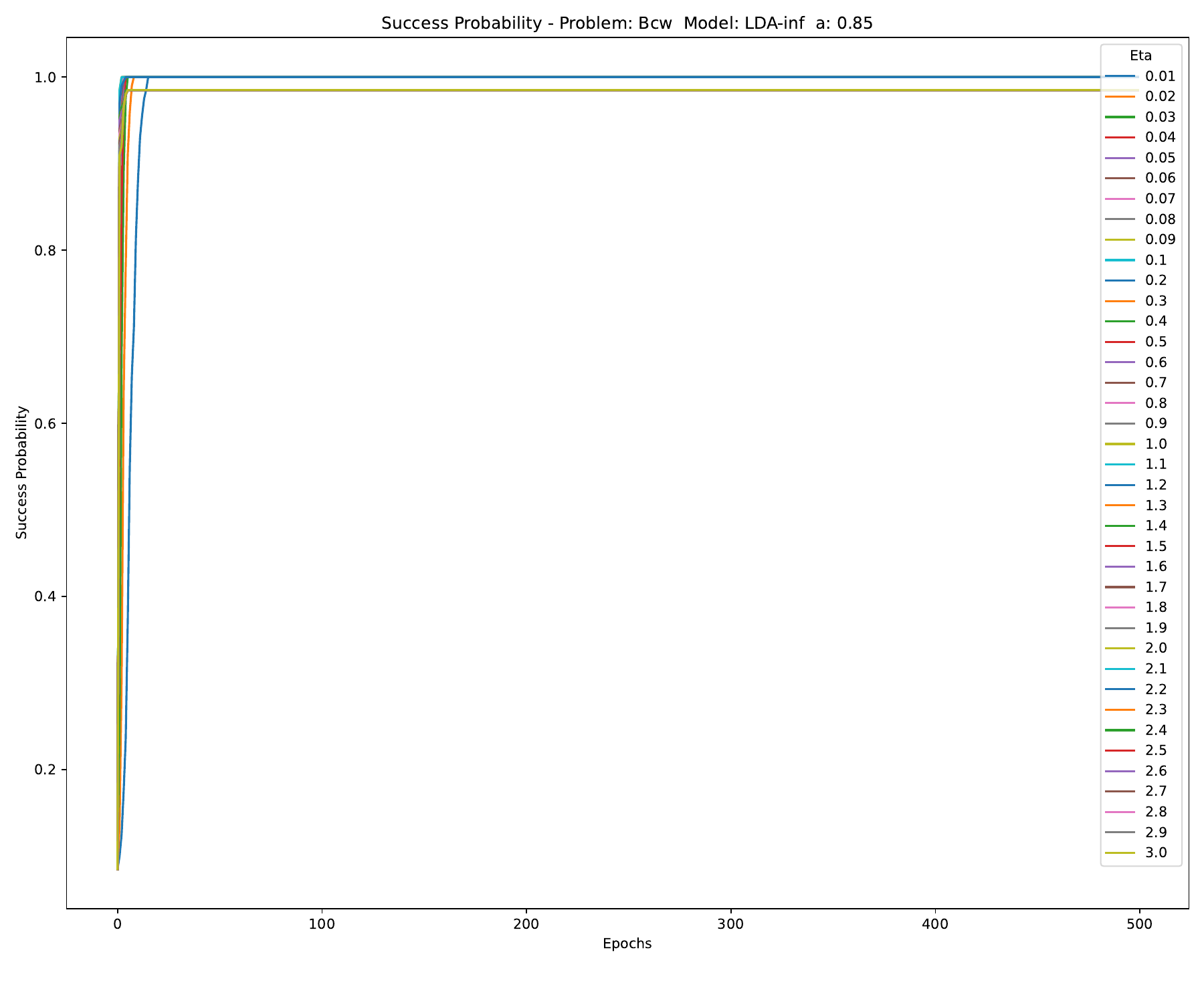} &
    \includegraphics[clip,trim=0 0 0 0,width=.52\linewidth]{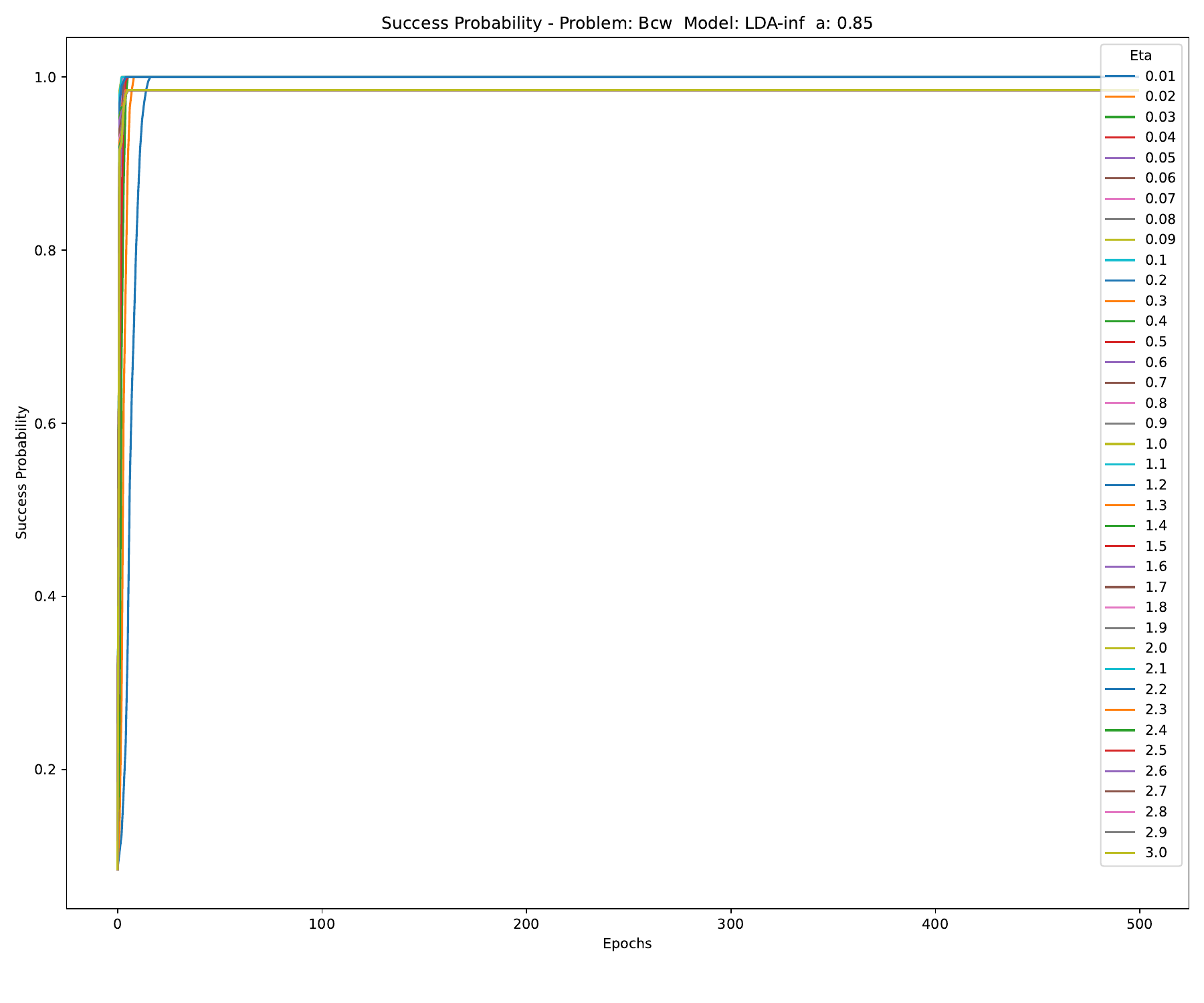} \\[-3mm]
    \includegraphics[clip,trim=0 0 0 0,width=.52\linewidth]{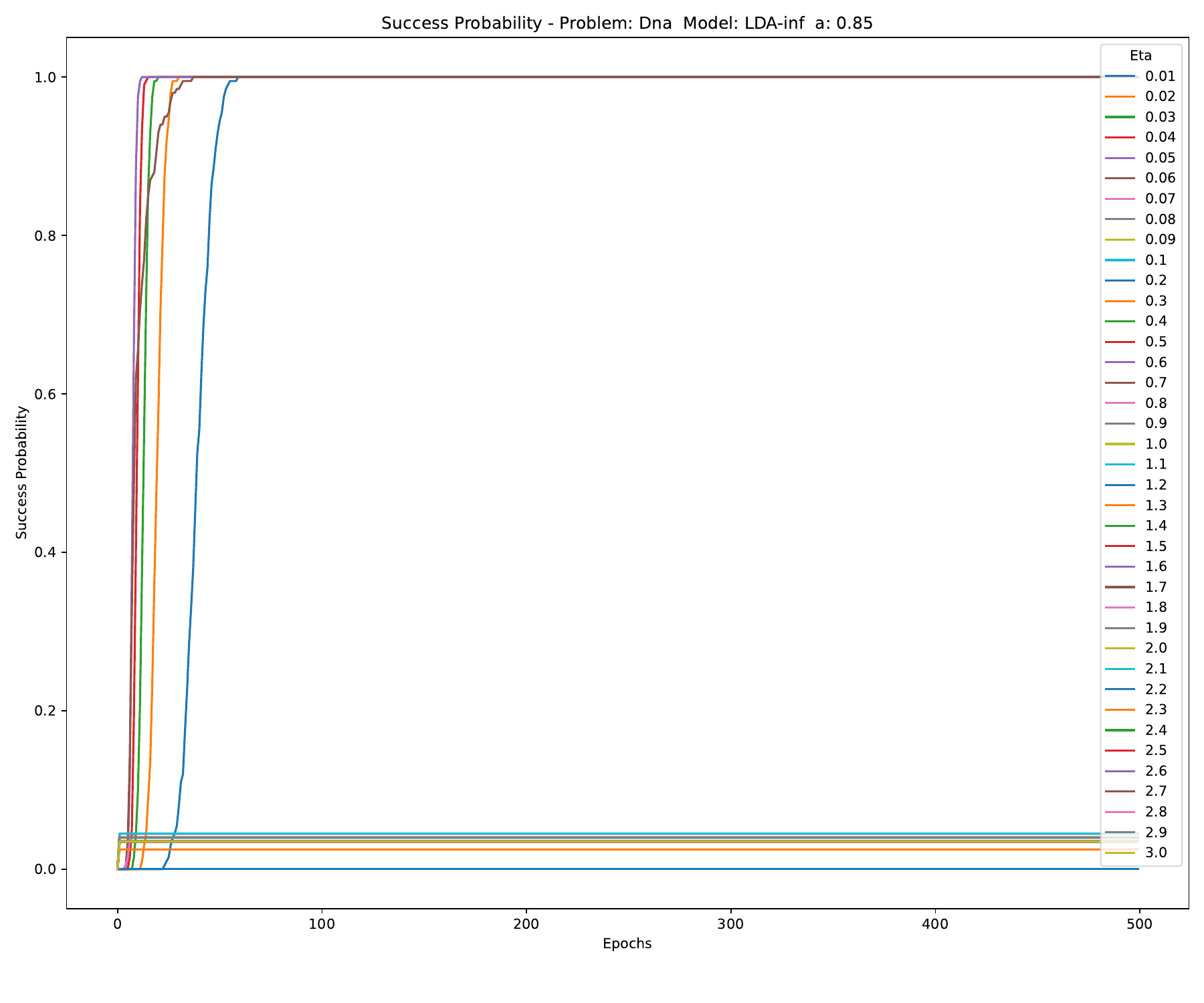} &
    \includegraphics[clip,trim=0 0 0 0,width=.52\linewidth]{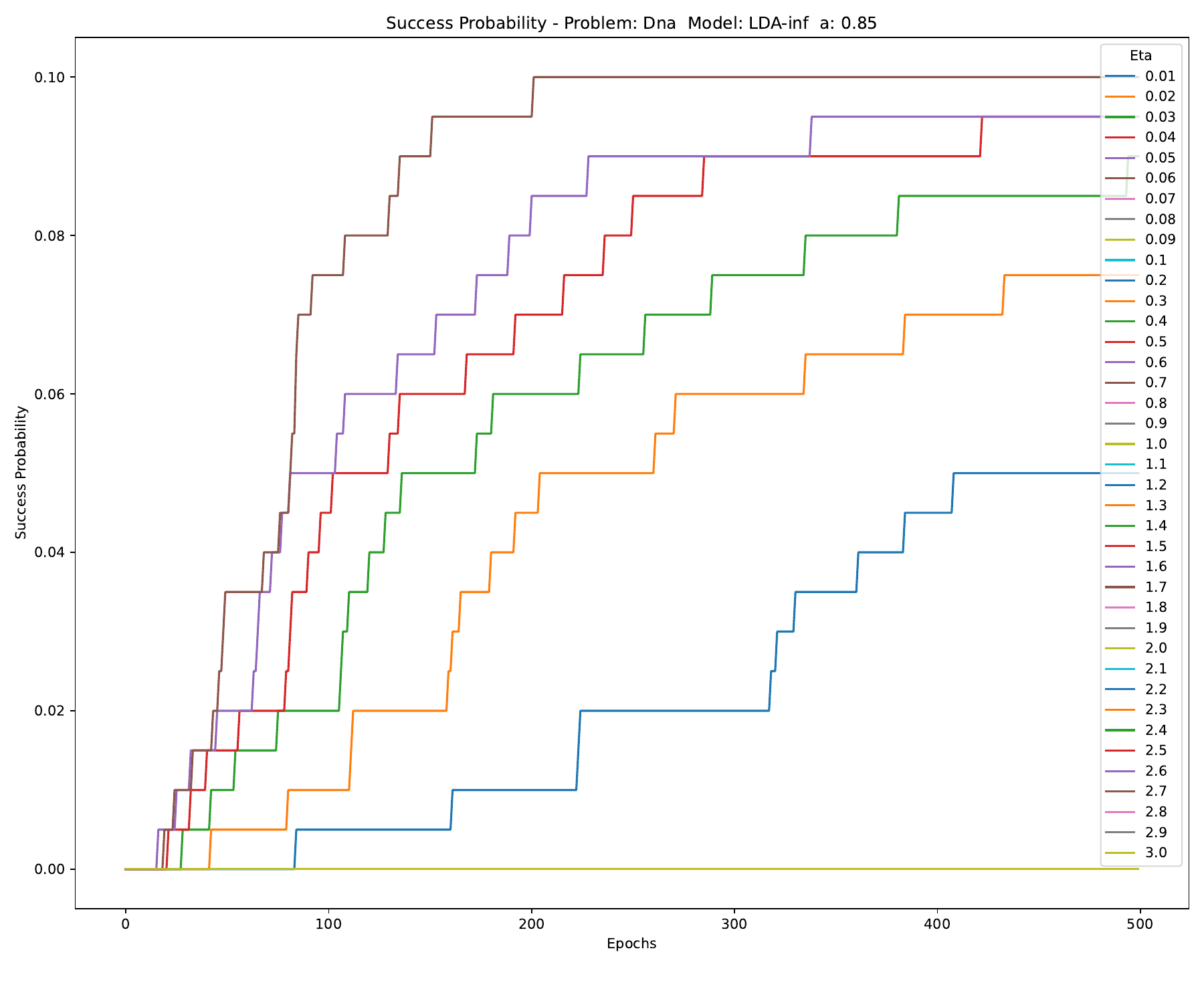} \\[-3mm]
    \includegraphics[clip,trim=0 0 0 0,width=.52\linewidth]{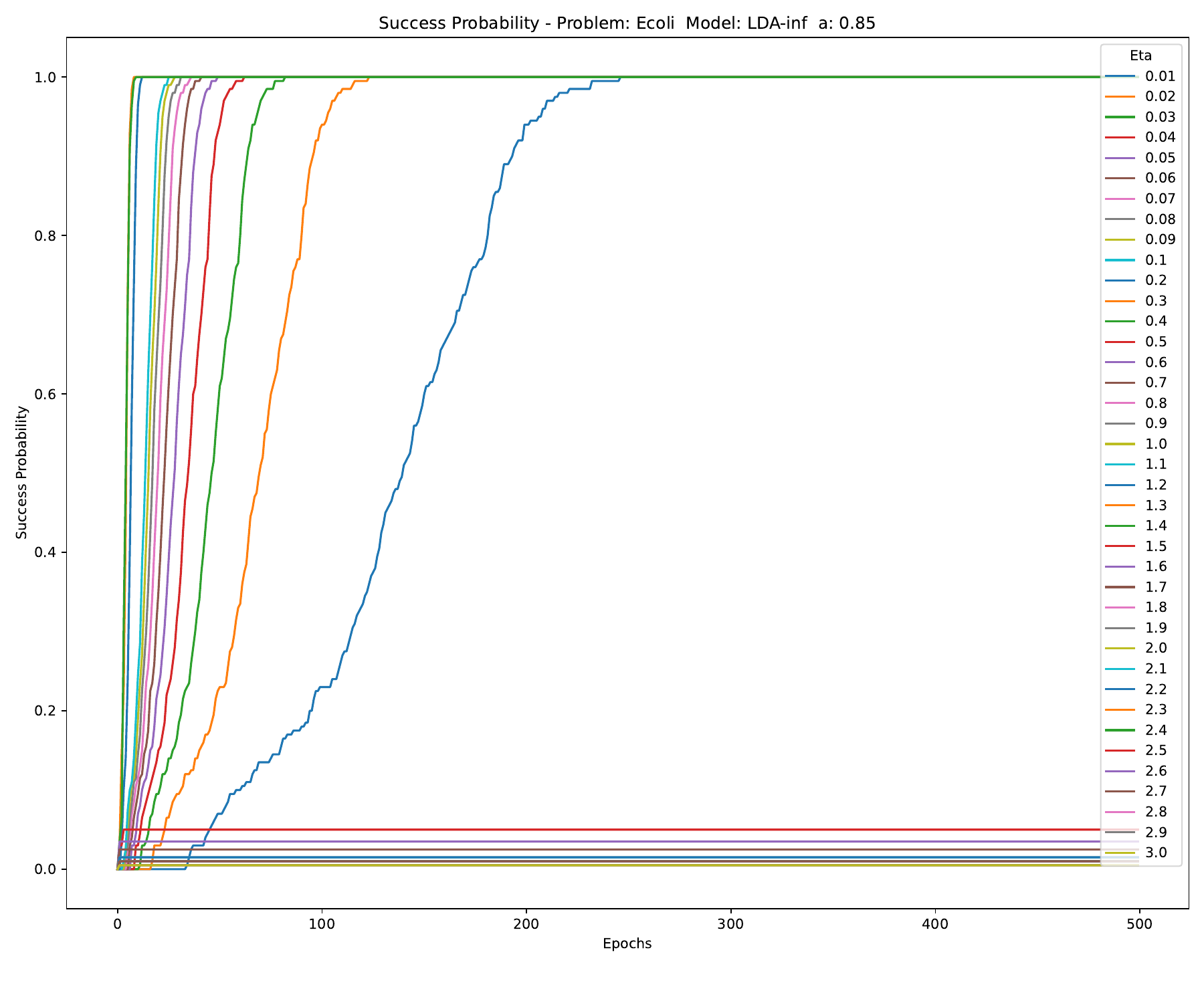} &
    \includegraphics[clip,trim=0 0 0 0,width=.52\linewidth]{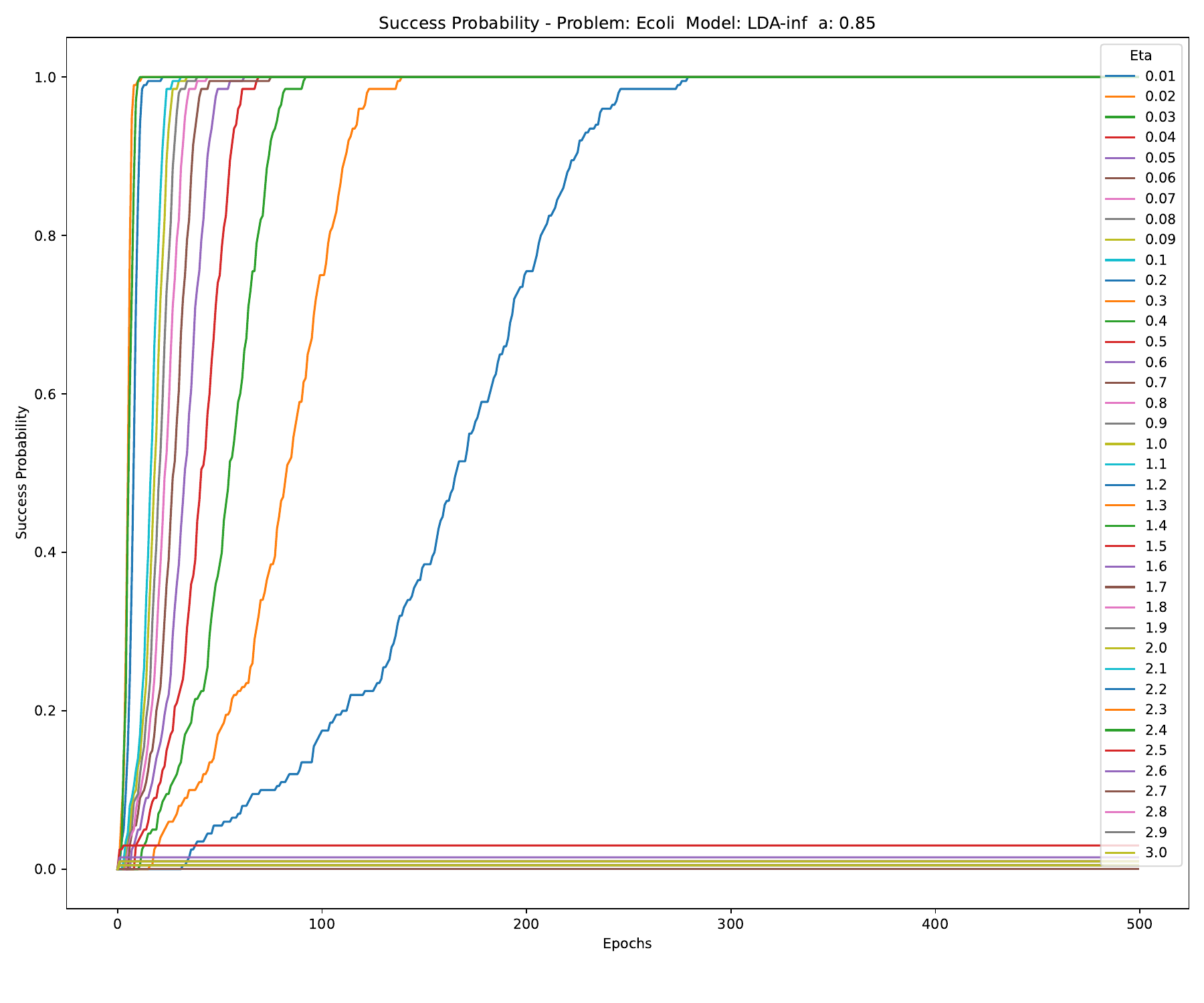} \\[-3mm]
  \end{tabular}
\caption{Success probability ($a=0.85$) for LDA-Inf model.}
  \label{fig:success_probability_LDA_Inf_model_0.85}
\end{figure*}

\begin{figure*}[p]
  \centering
  \ContinuedFloat
  \begin{tabular}{c@{}c}
    Training Success Probability & Validation Success Probability \\
    \includegraphics[clip,trim=0 0 0 0,width=.52\linewidth]{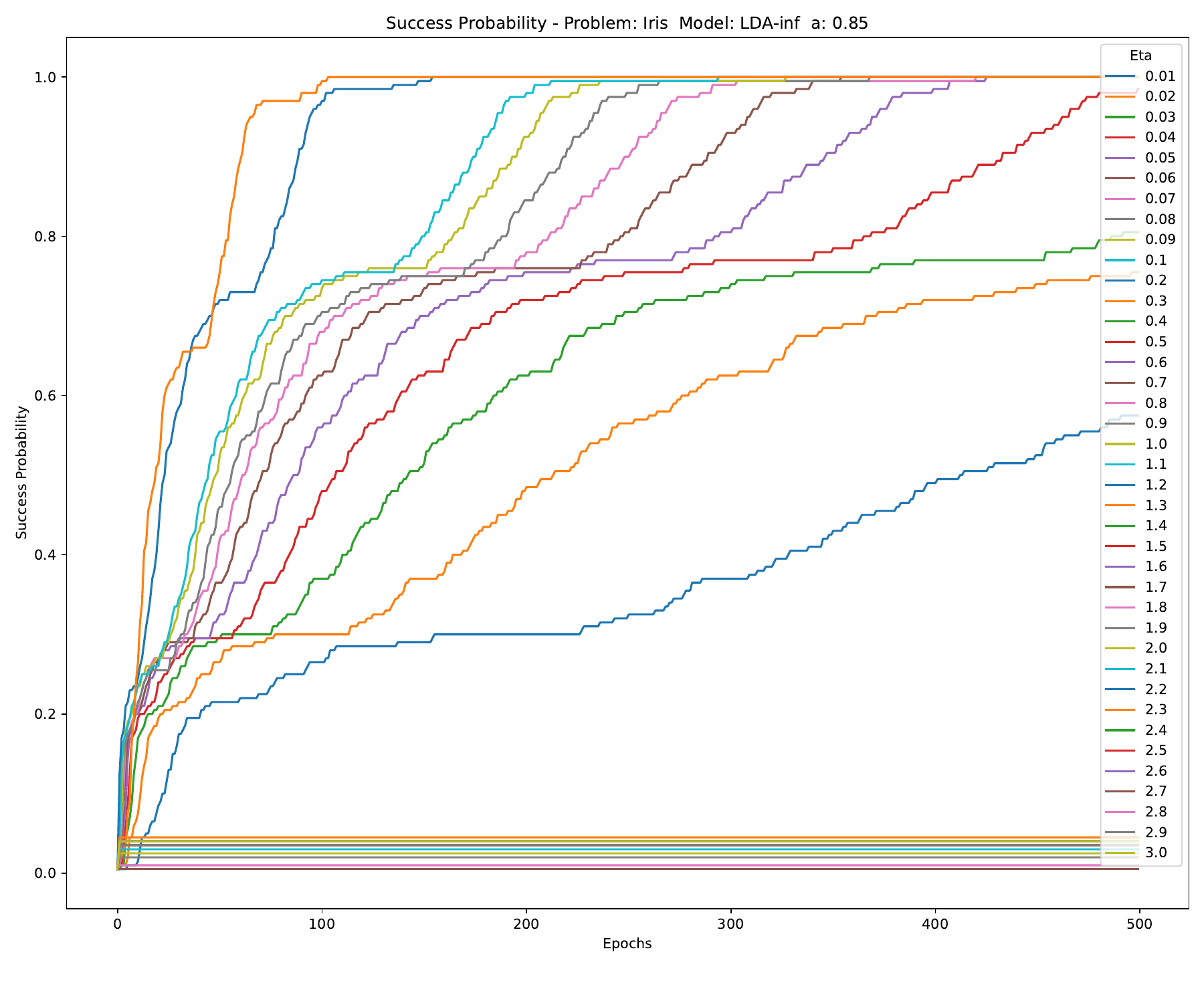} &
    \includegraphics[clip,trim=0 0 0 0,width=.52\linewidth]{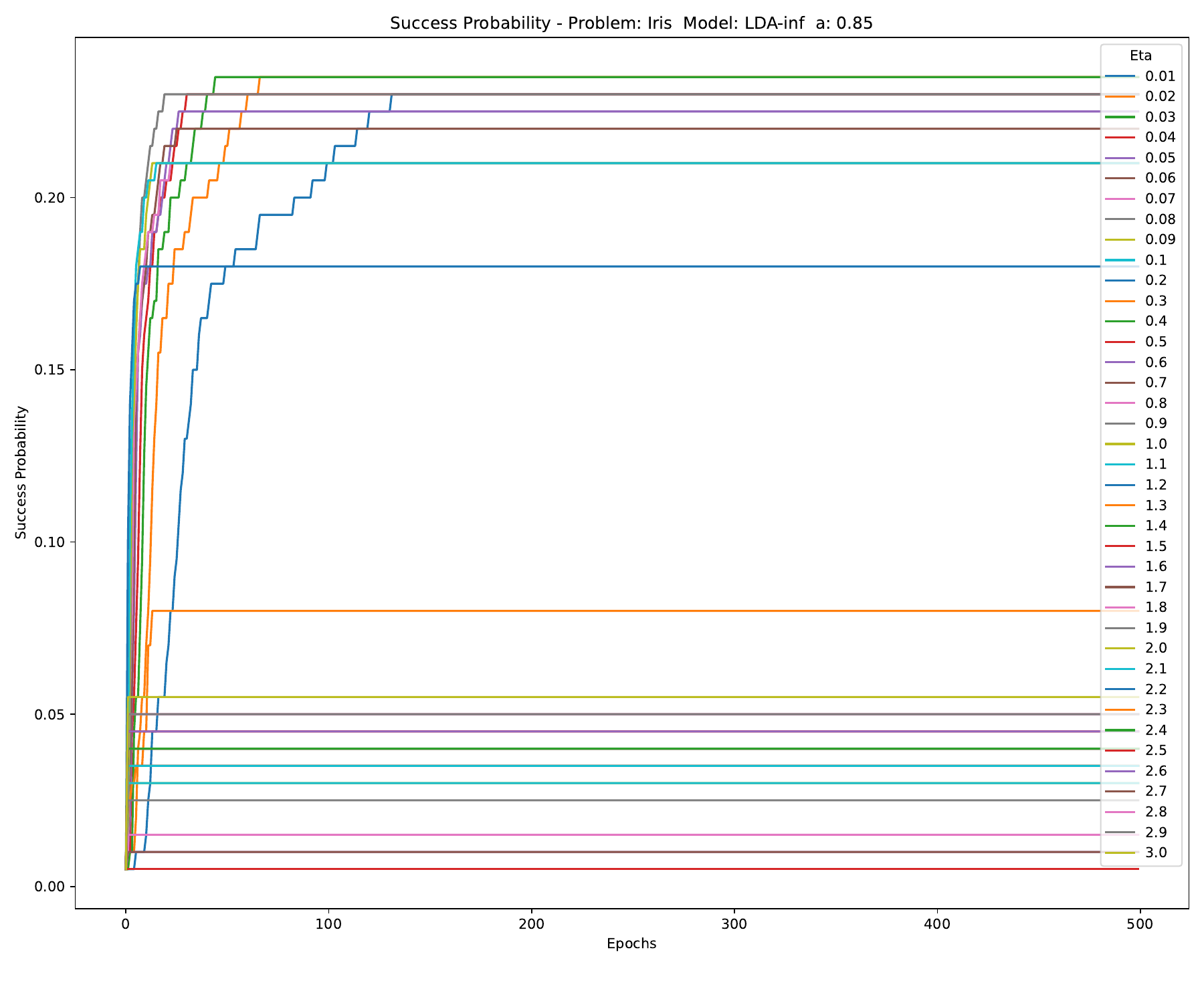} \\[-3mm]
    \includegraphics[clip,trim=0 0 0 0,width=.52\linewidth]{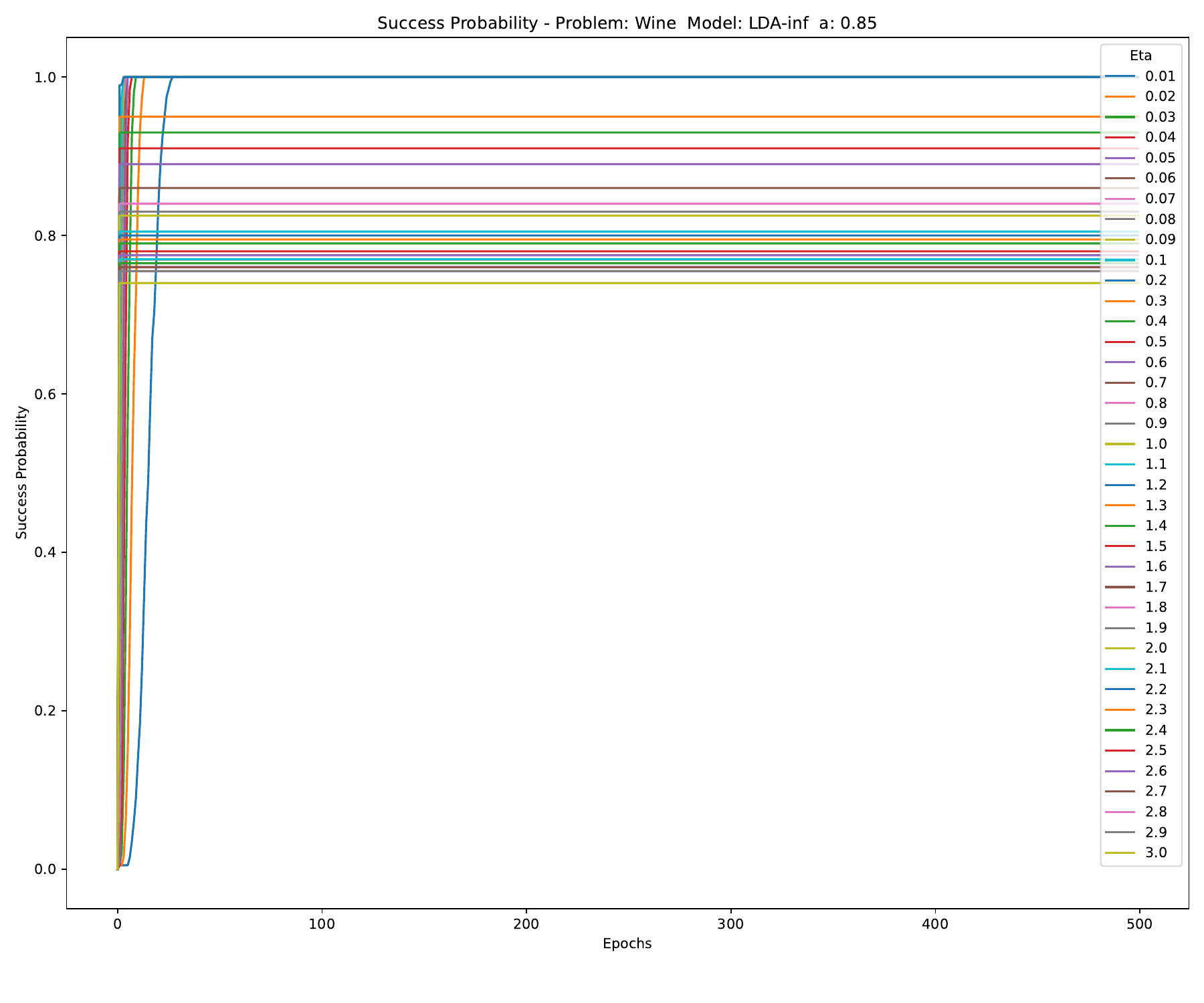} &
    \includegraphics[clip,trim=0 0 0 0,width=.52\linewidth]{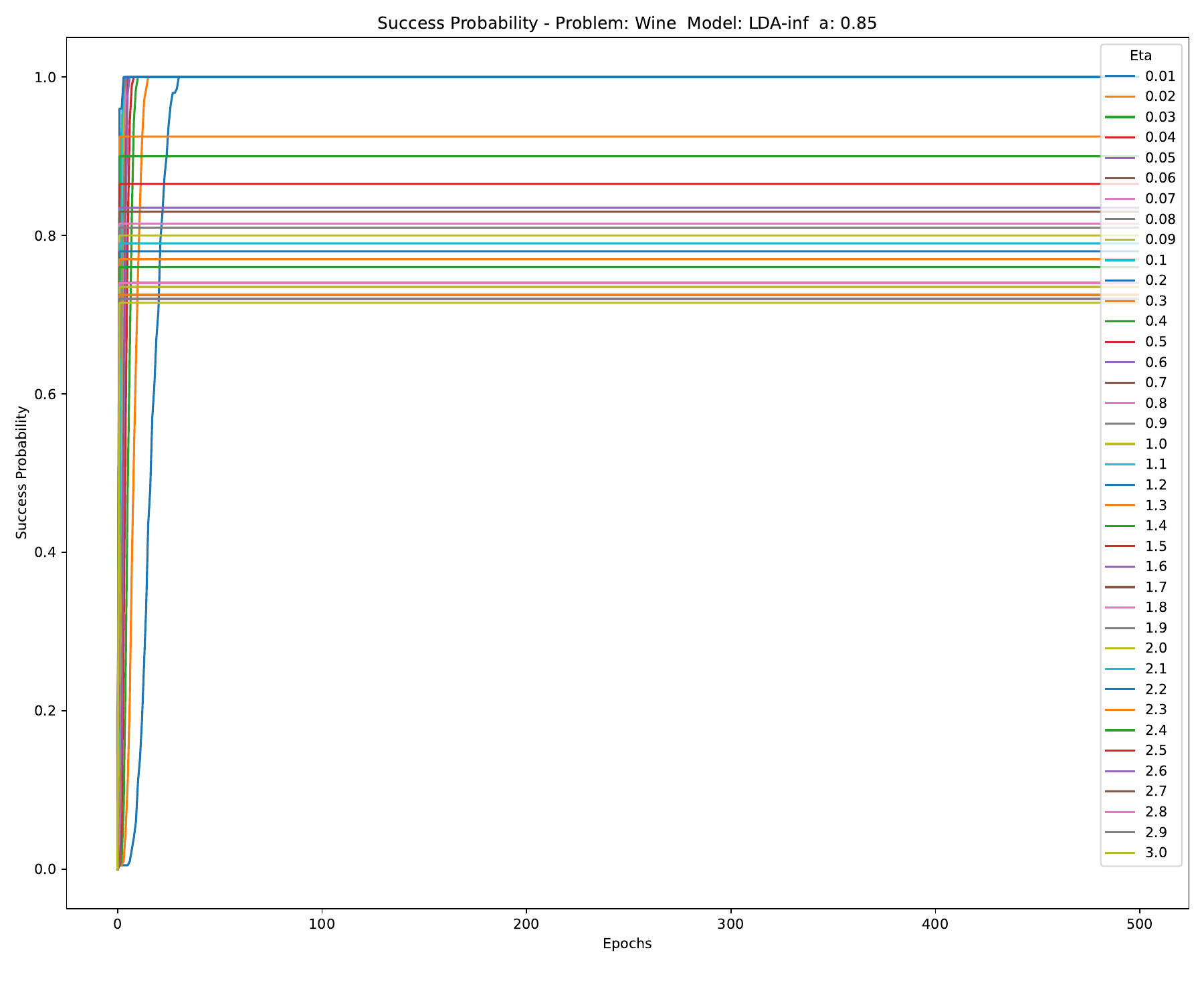} \\[-3mm]
  \end{tabular}
\caption{Success probability ($a=0.85$) for LDA-Inf model (continued).}
\end{figure*}

\begin{figure*}[p]
  \centering
  \begin{tabular}{c@{}c}
    Training Success Probability & Validation Success Probability \\
    \includegraphics[clip,trim=0 0 0 0,width=.52\linewidth]{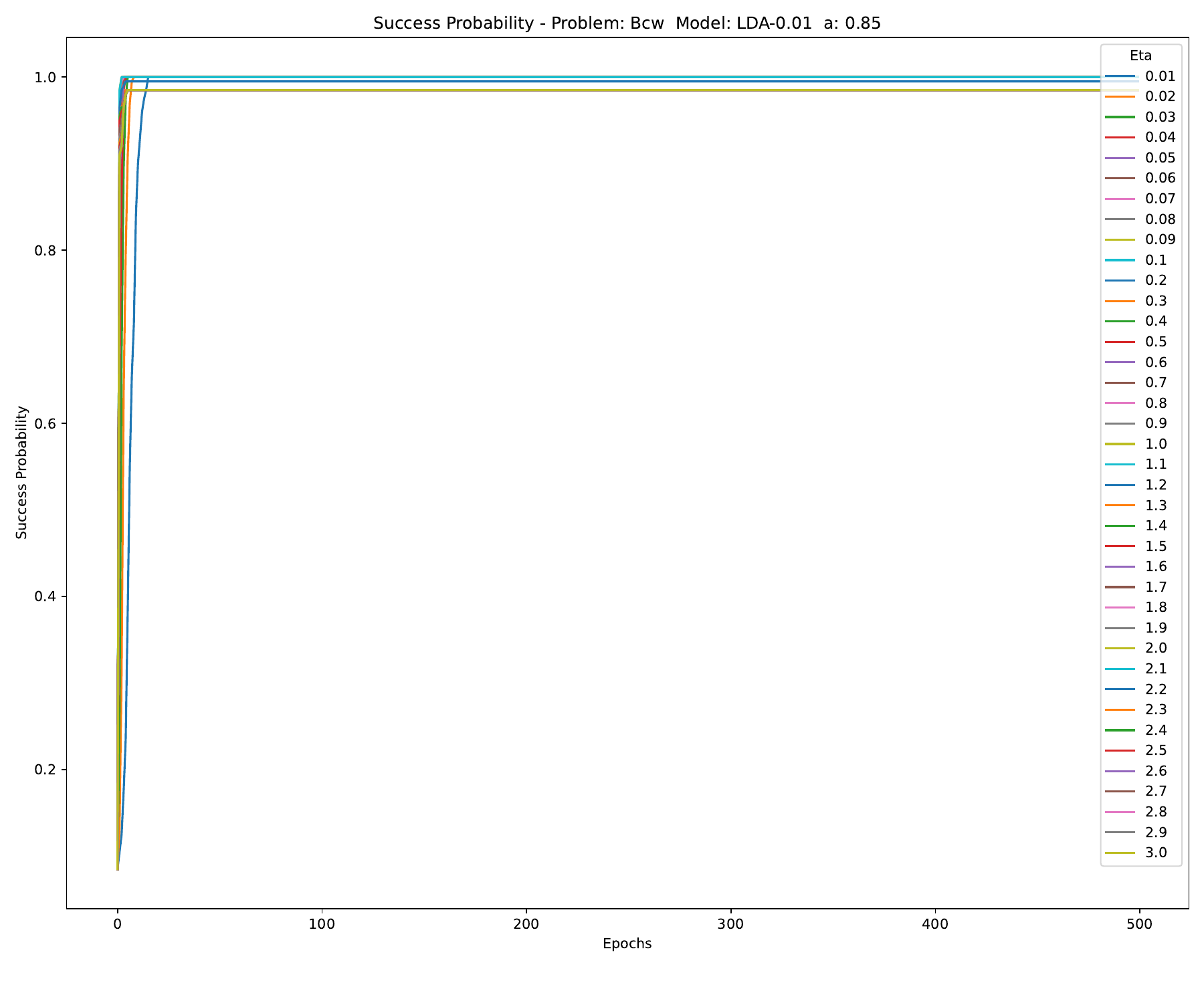} &
    \includegraphics[clip,trim=0 0 0 0,width=.52\linewidth]{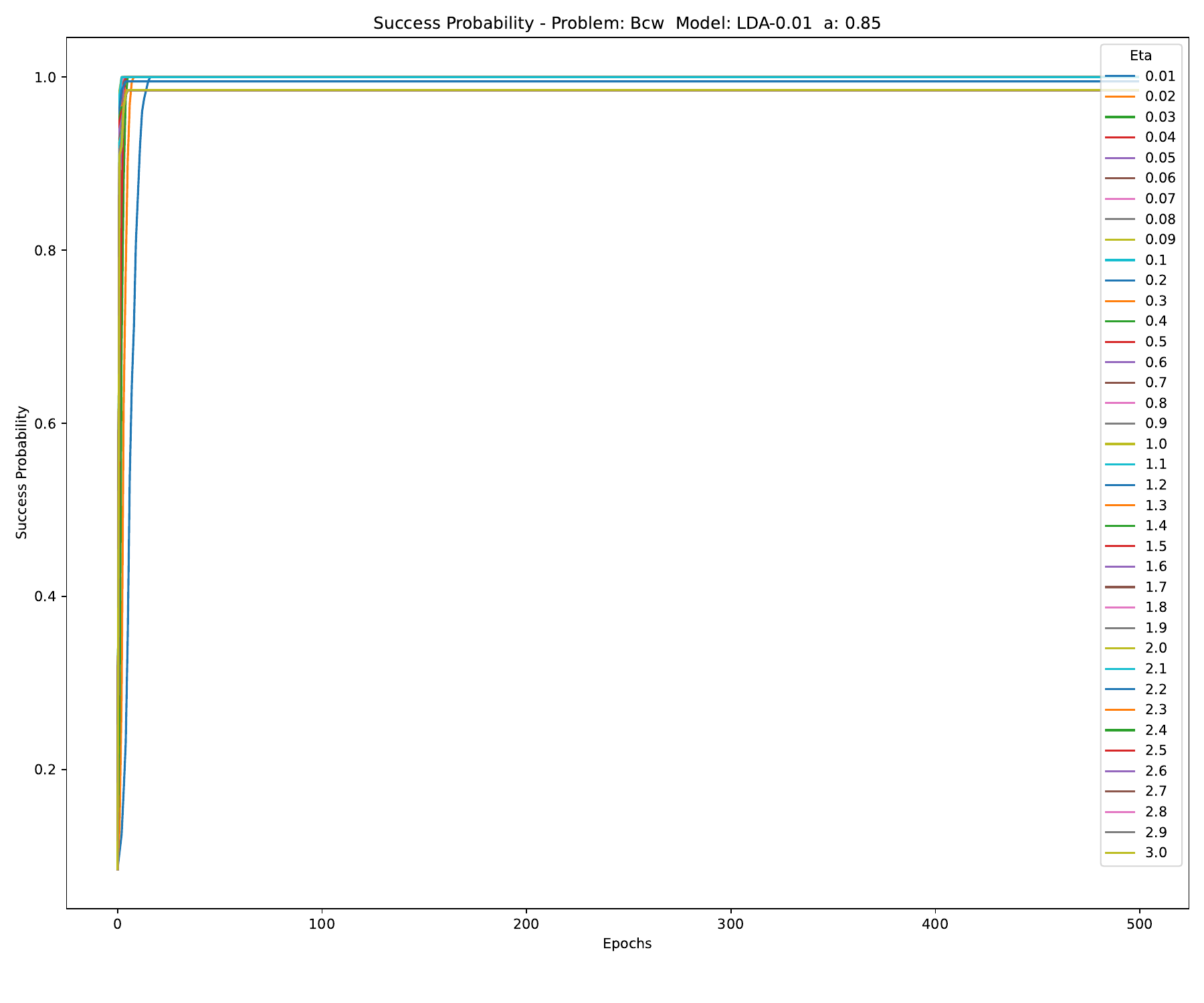} \\[-3mm]
    \includegraphics[clip,trim=0 0 0 0,width=.52\linewidth]{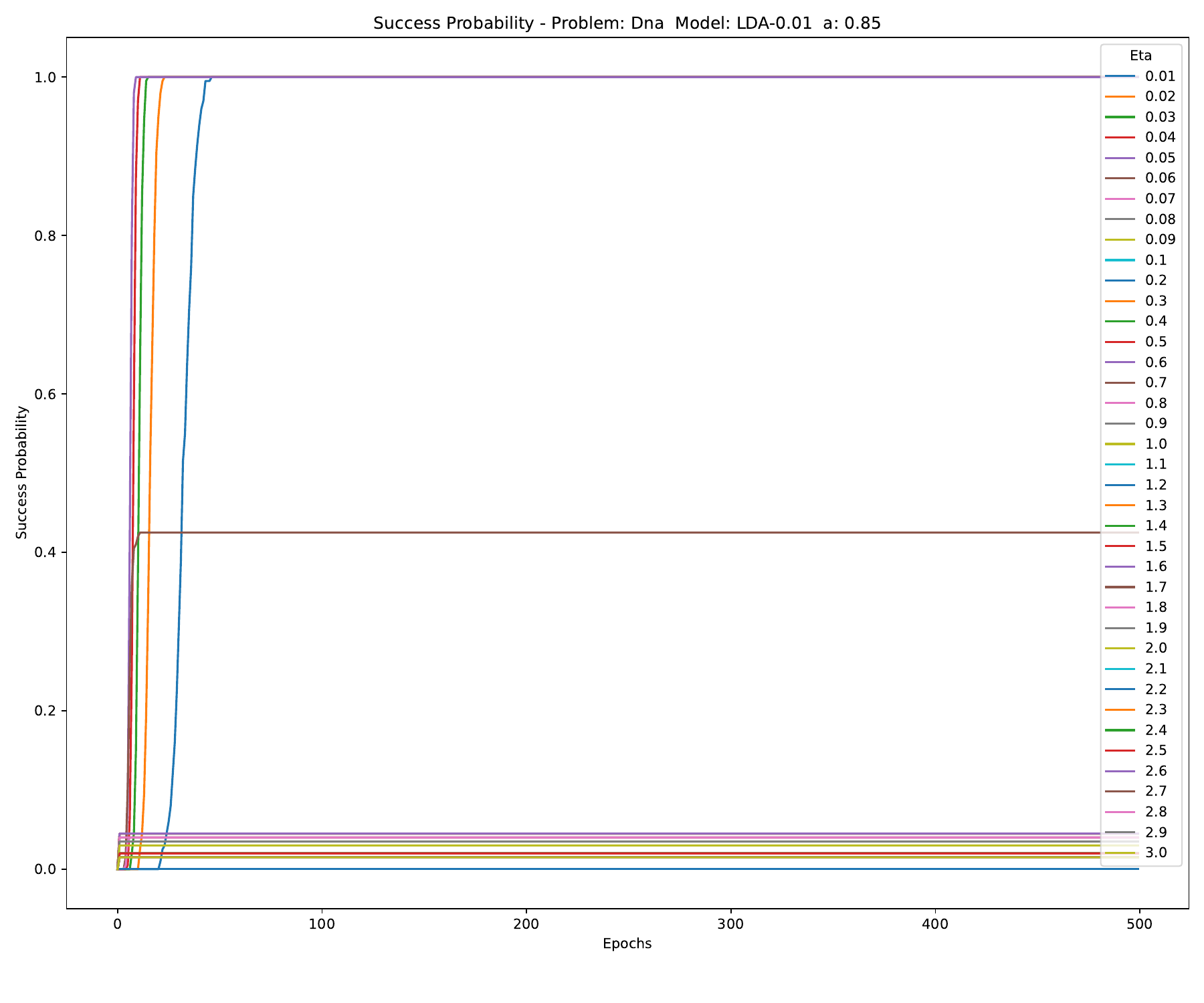} &
    \includegraphics[clip,trim=0 0 0 0,width=.52\linewidth]{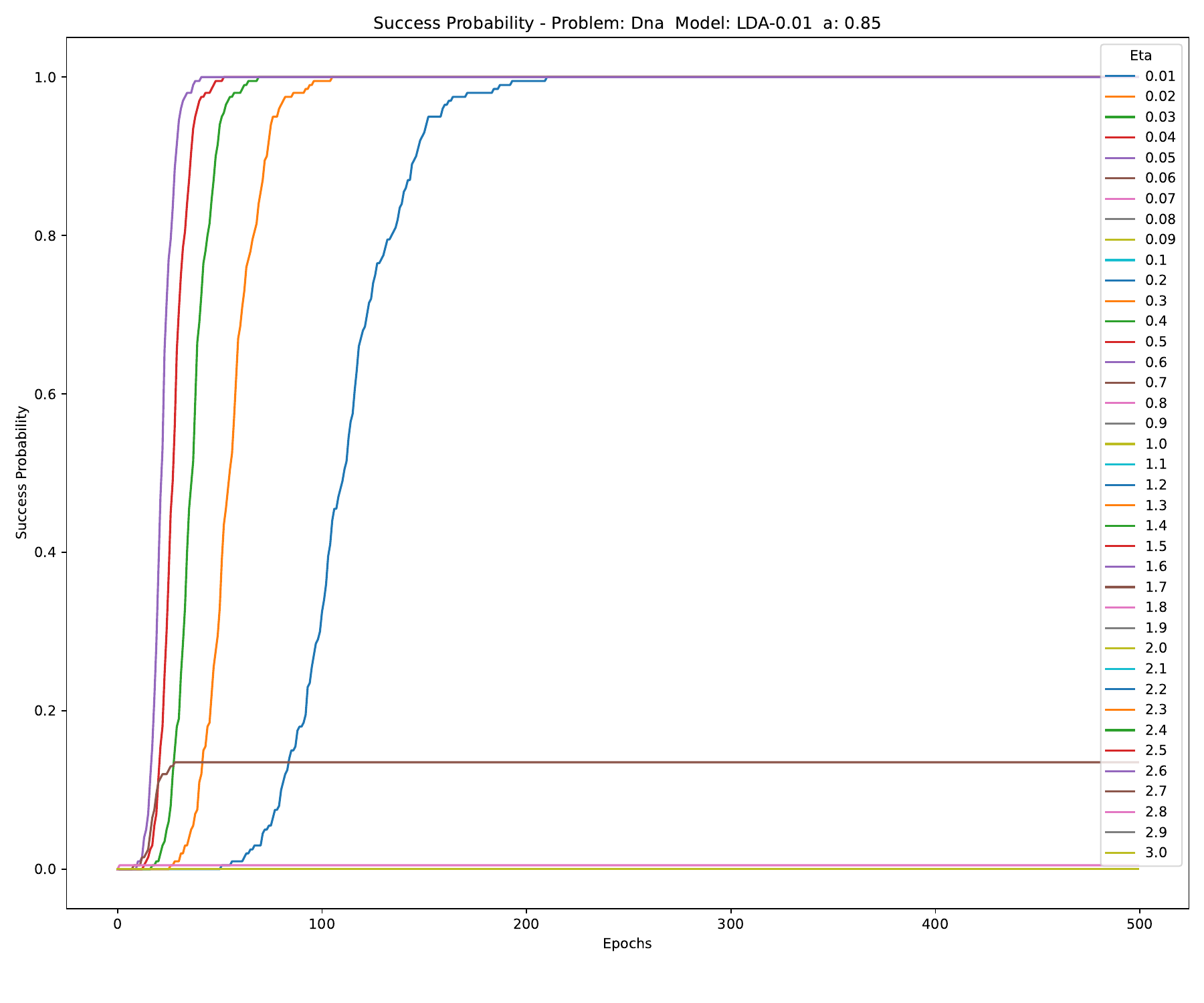} \\[-3mm]
    \includegraphics[clip,trim=0 0 0 0,width=.52\linewidth]{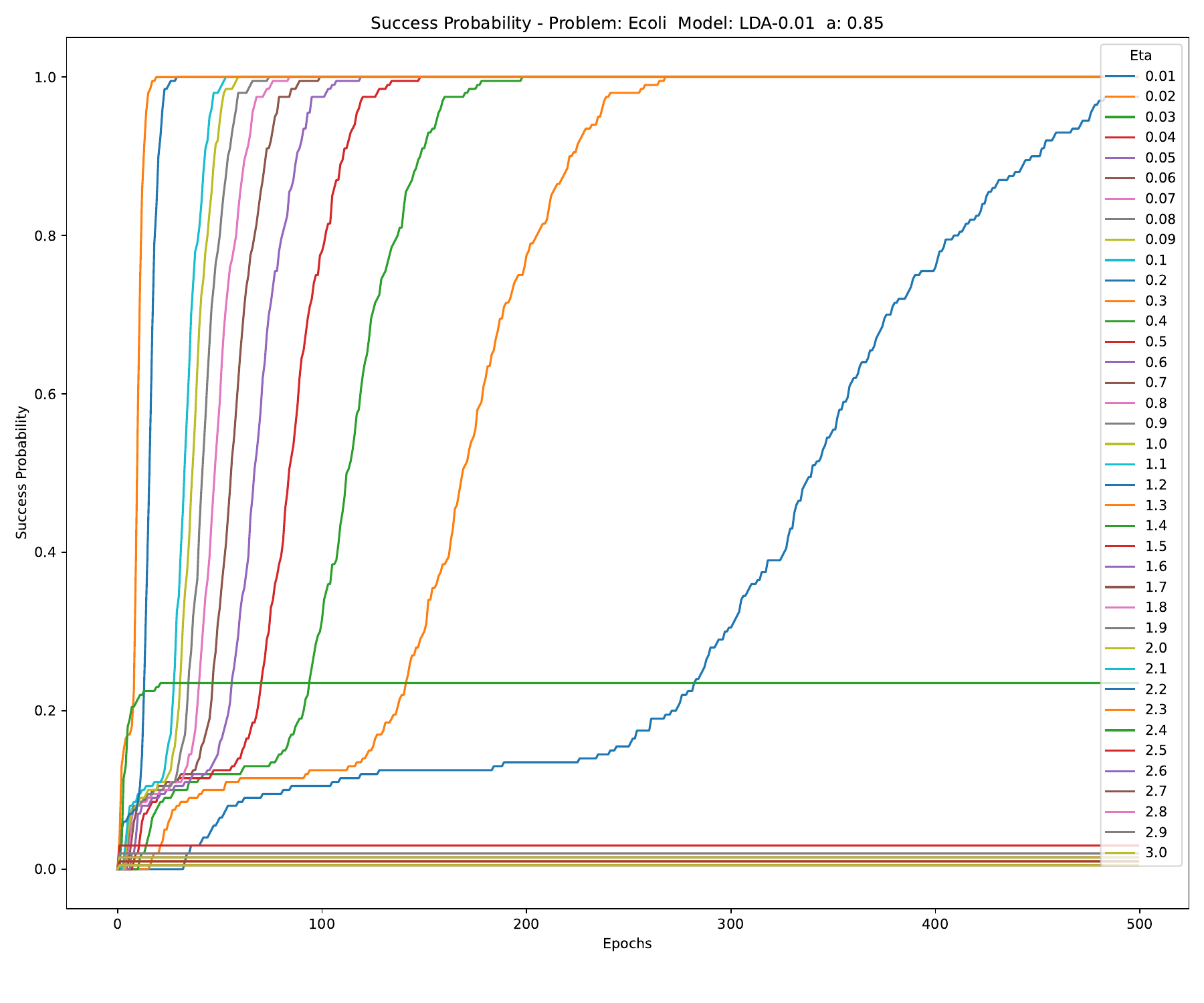} &
    \includegraphics[clip,trim=0 0 0 0,width=.52\linewidth]{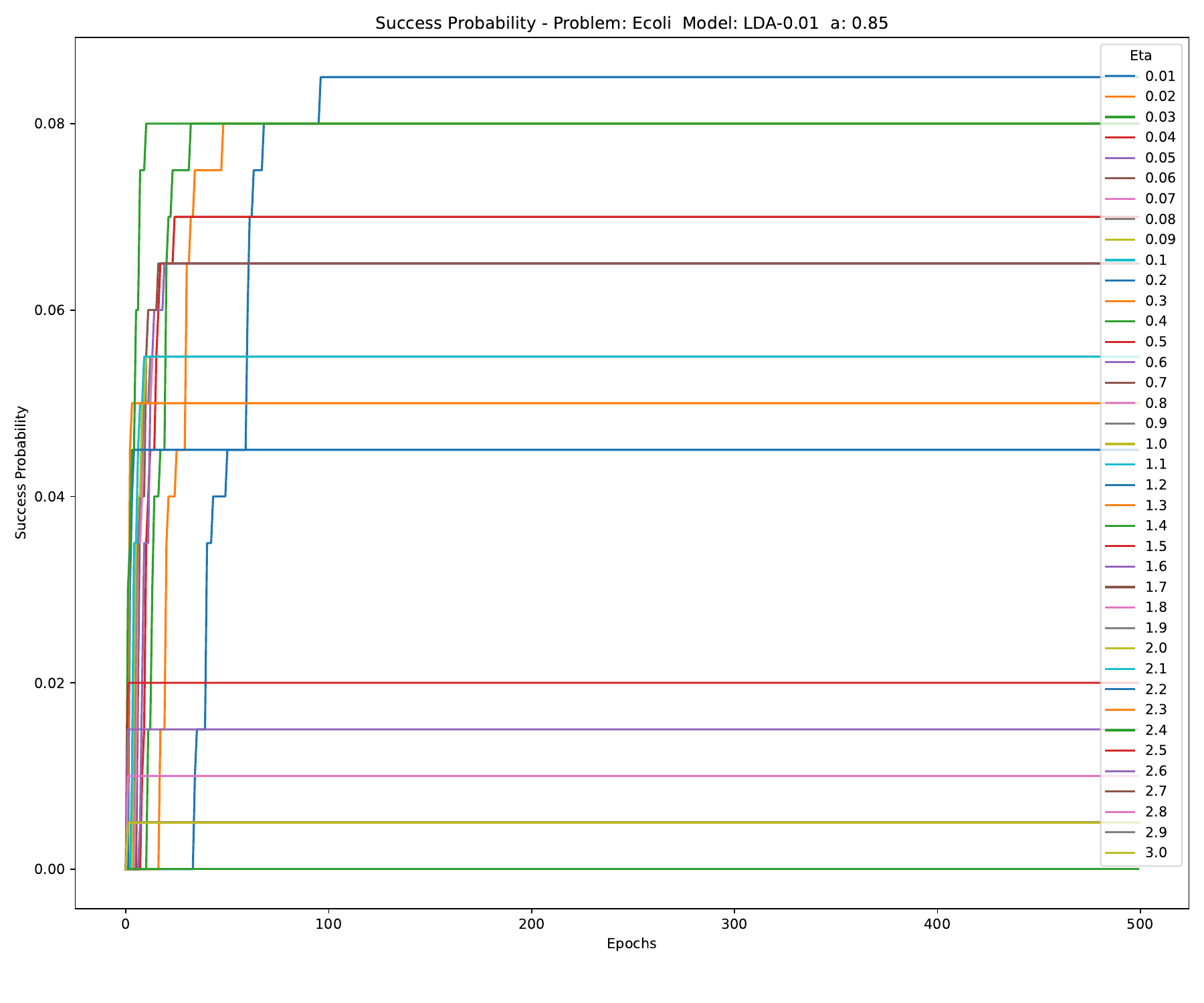} \\[-3mm]
  \end{tabular}
\caption{Success probability ($a=0.85$) for LDA-0.01 model.}
  \label{fig:success_probability_LDA_0.01_model_0.85}
\end{figure*}

\begin{figure*}[p]
  \centering
  \ContinuedFloat
  \begin{tabular}{c@{}c}
    Training Success Probability & Validation Success Probability \\
    \includegraphics[clip,trim=0 0 0 0,width=.52\linewidth]{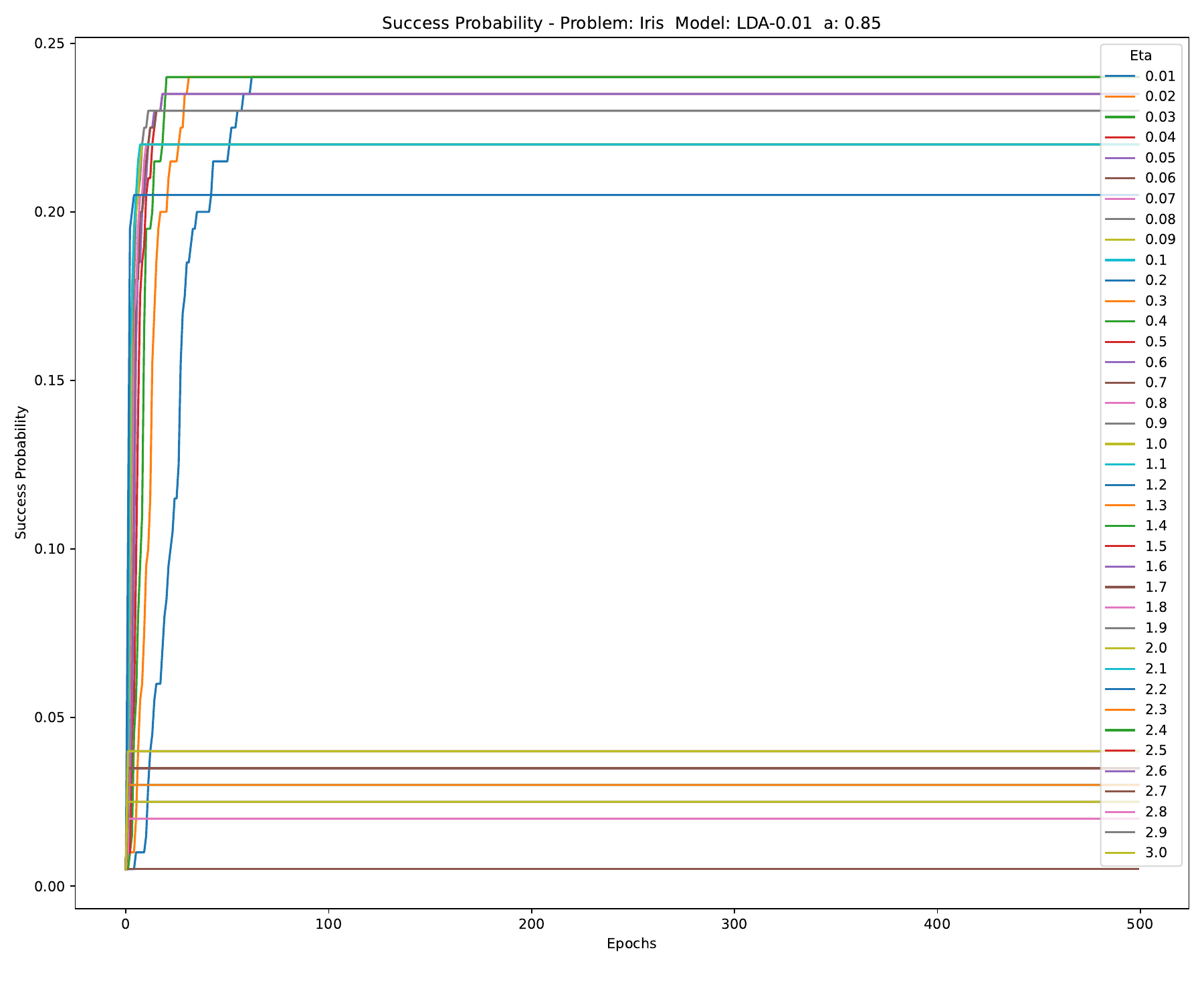} &
    \includegraphics[clip,trim=0 0 0 0,width=.52\linewidth]{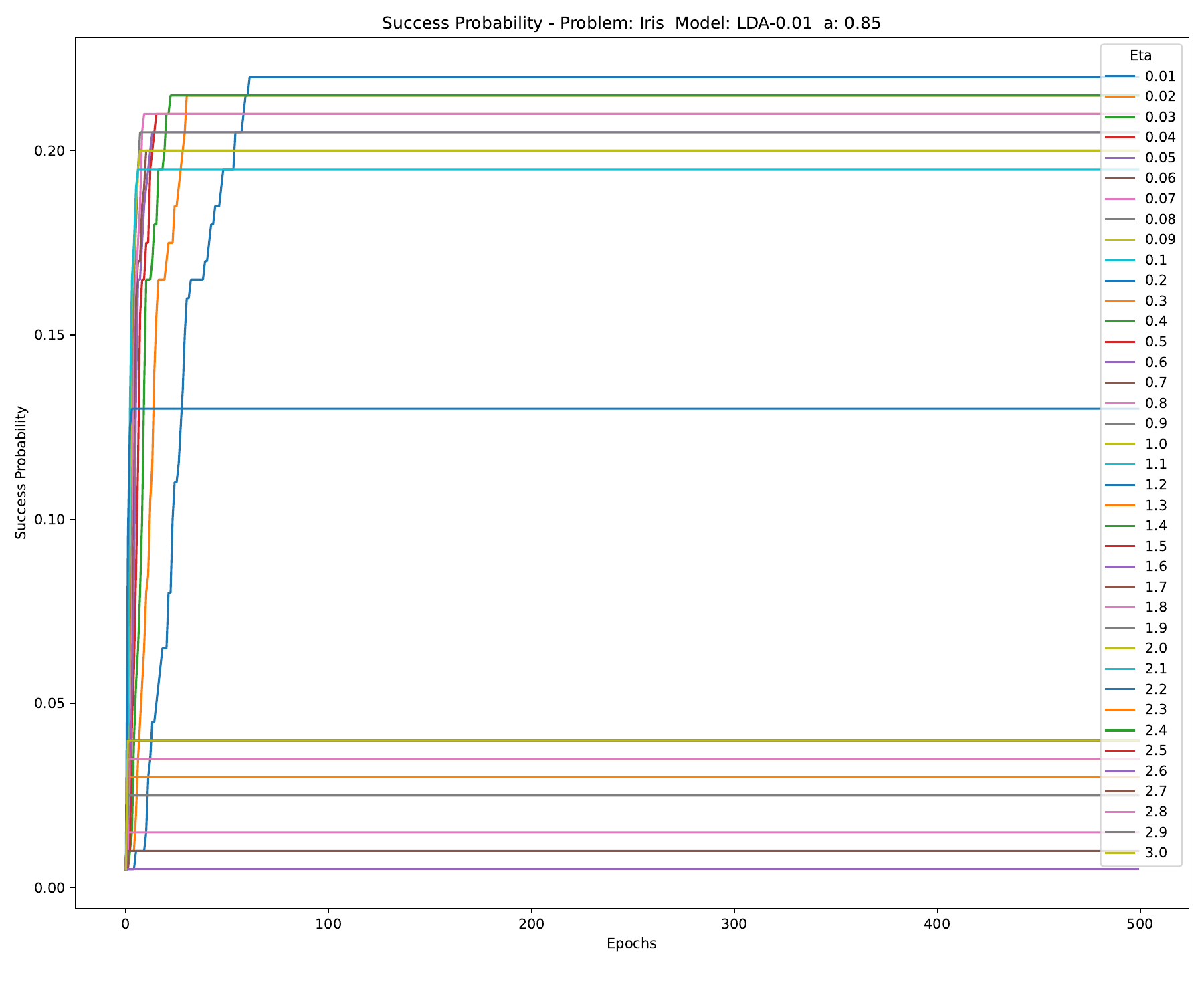} \\[-3mm]
    \includegraphics[clip,trim=0 0 0 0,width=.52\linewidth]{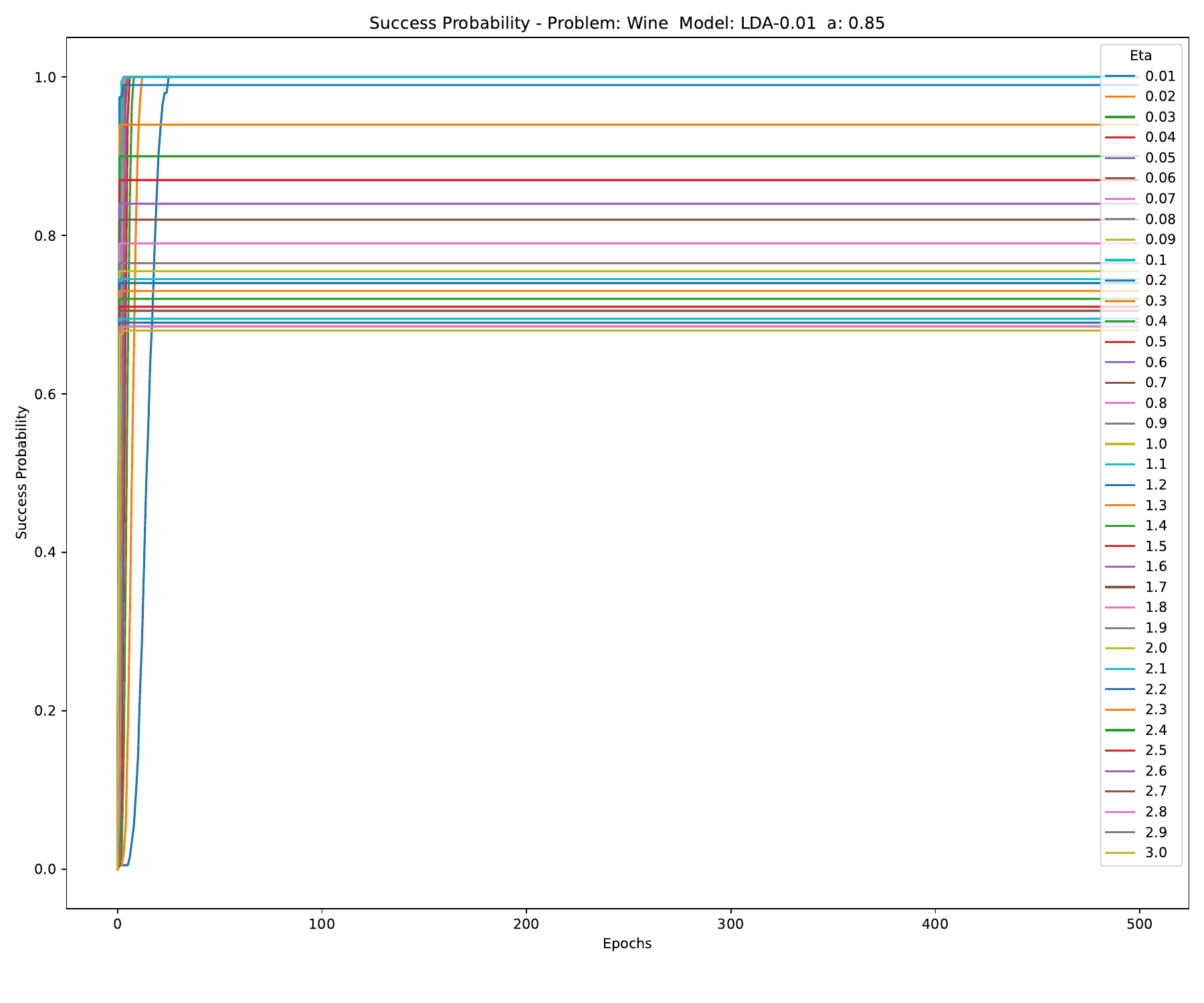} &
    \includegraphics[clip,trim=0 0 0 0,width=.52\linewidth]{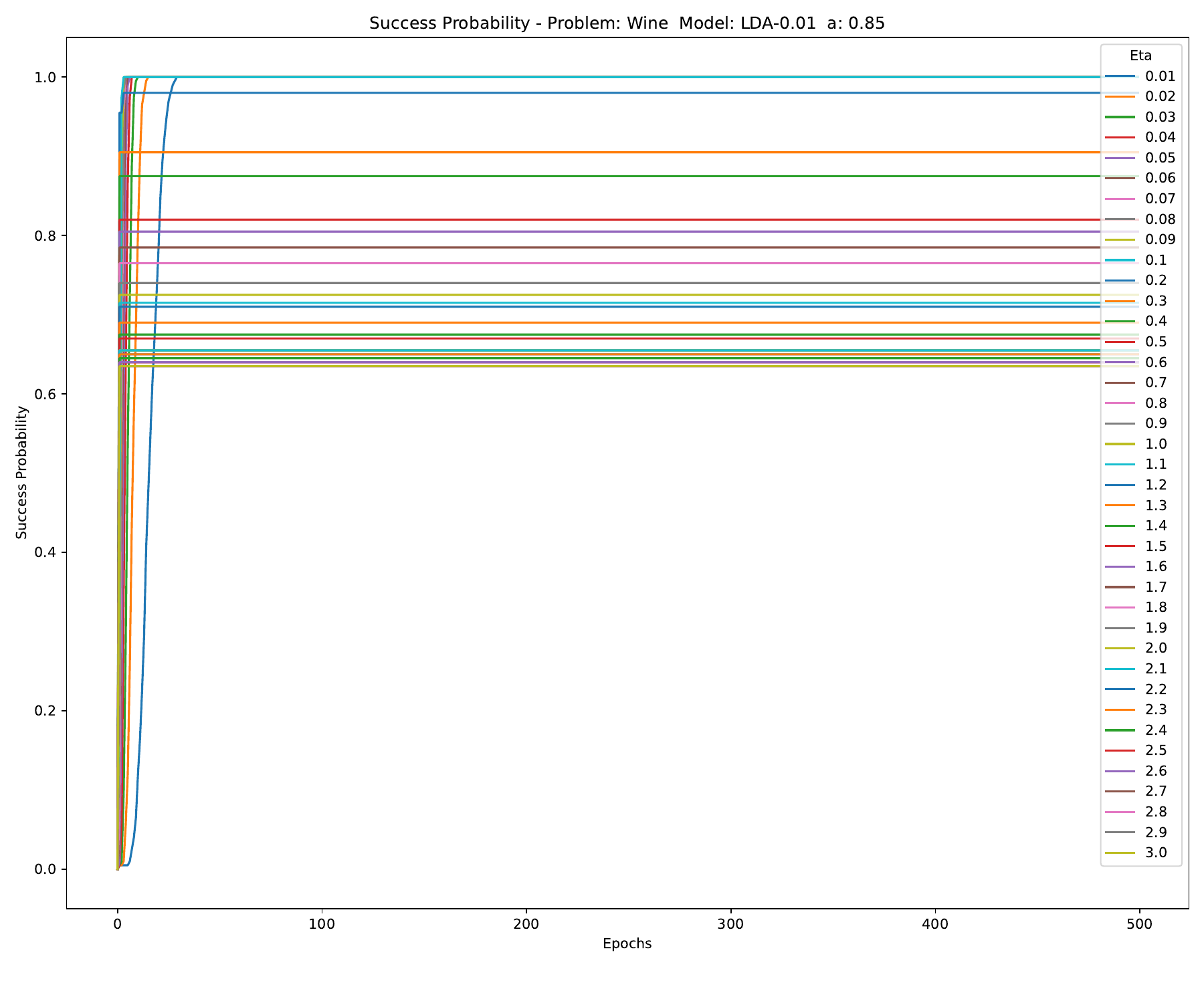} \\[-3mm]
  \end{tabular}
\caption{Success probability ($a=0.85$) for LDA-0.01 model (continued).}
\end{figure*}

\begin{figure*}[p]
  \centering
  \begin{tabular}{c@{}c}
    Training Success Probability & Validation Success Probability \\
    \includegraphics[clip,trim=0 0 0 0,width=.52\linewidth]{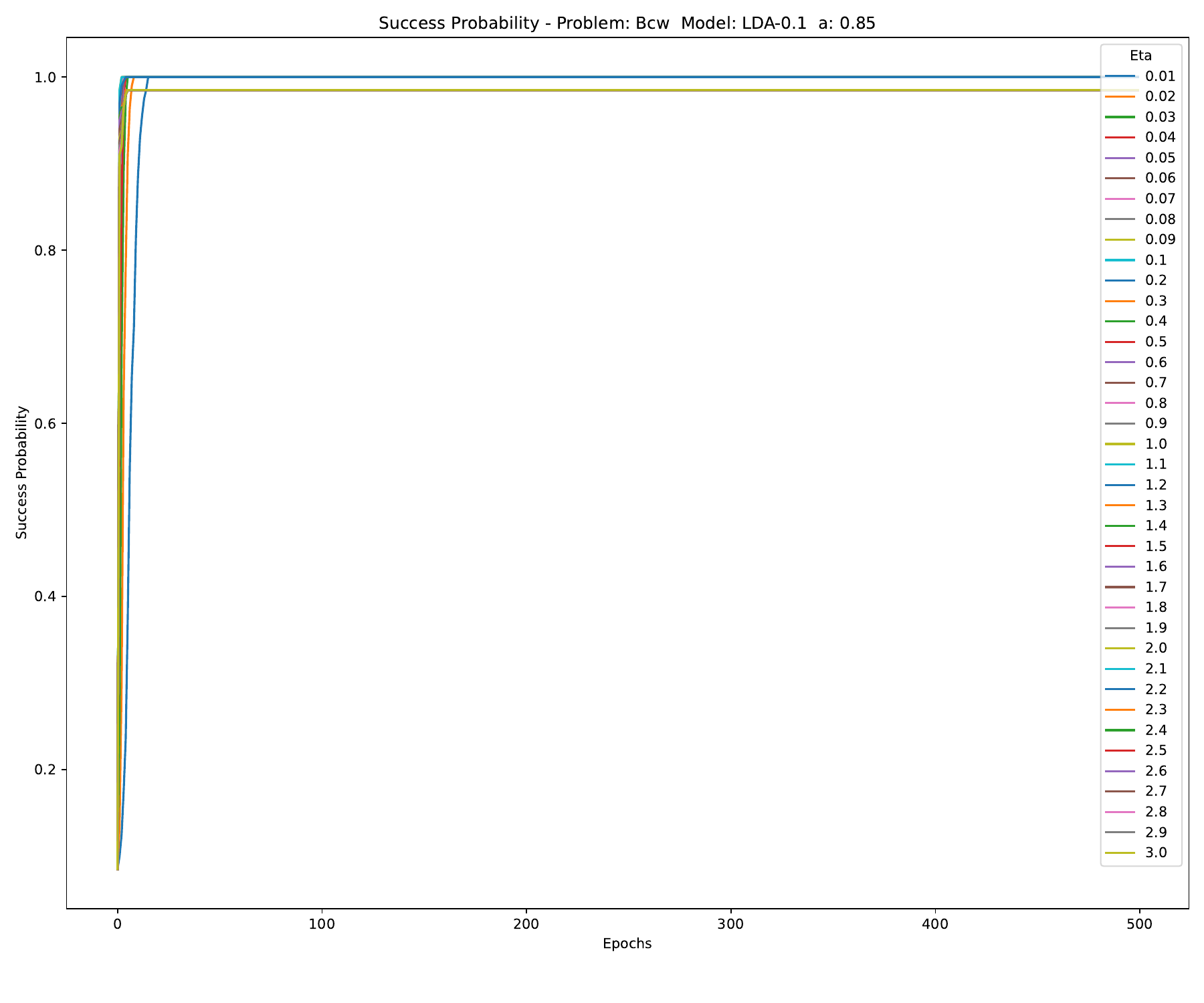} &
    \includegraphics[clip,trim=0 0 0 0,width=.52\linewidth]{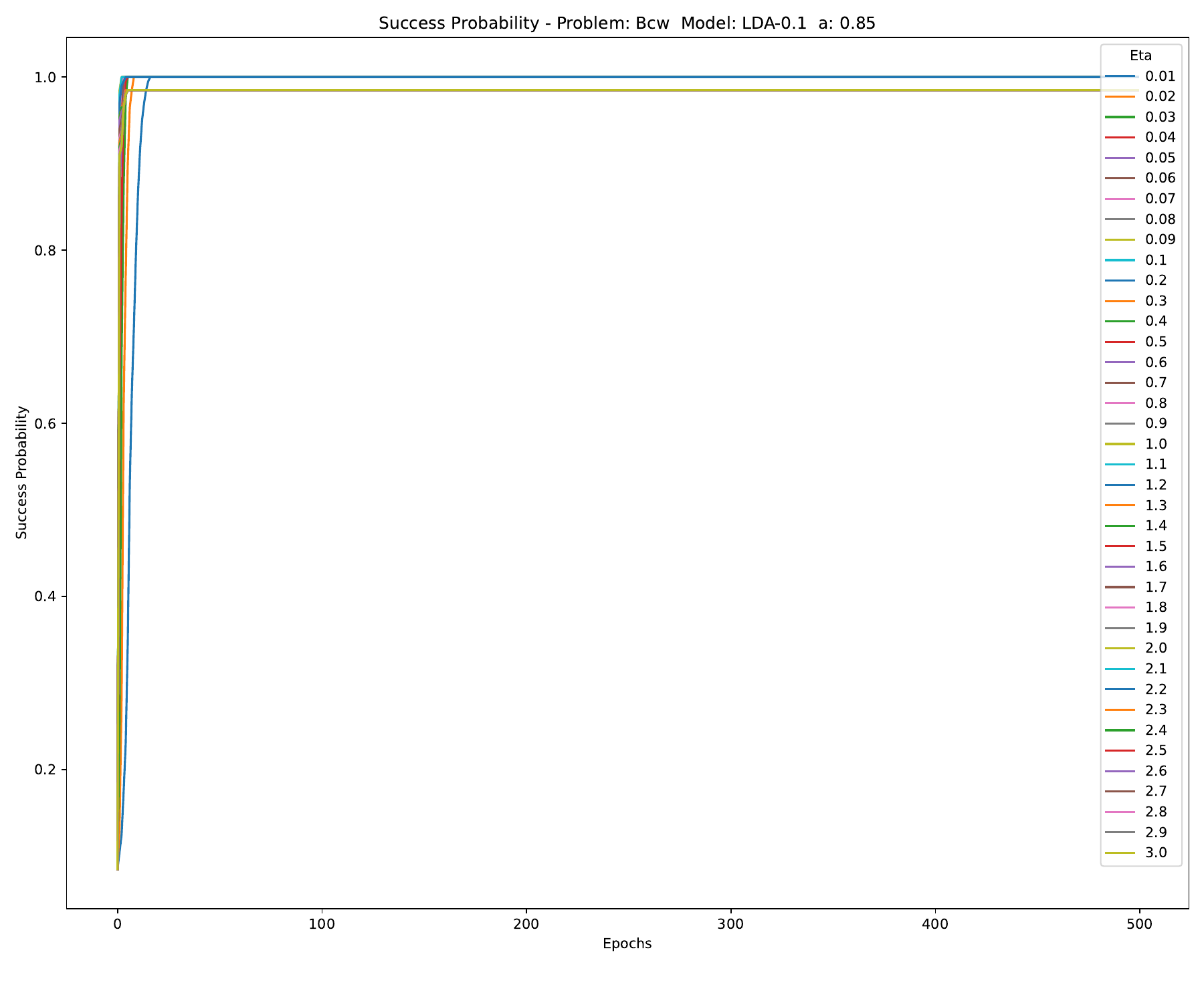} \\[-3mm]
    \includegraphics[clip,trim=0 0 0 0,width=.52\linewidth]{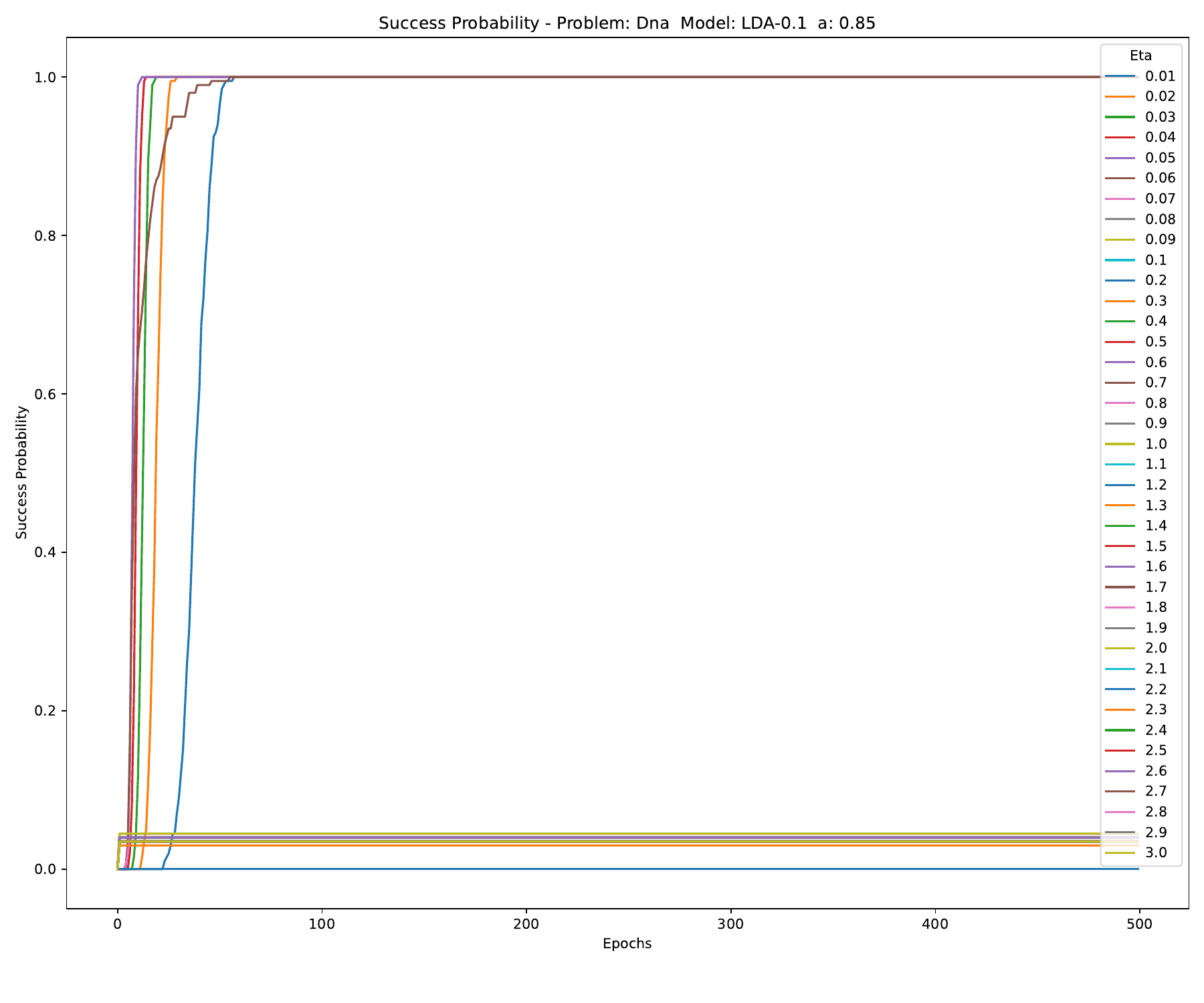} &
    \includegraphics[clip,trim=0 0 0 0,width=.52\linewidth]{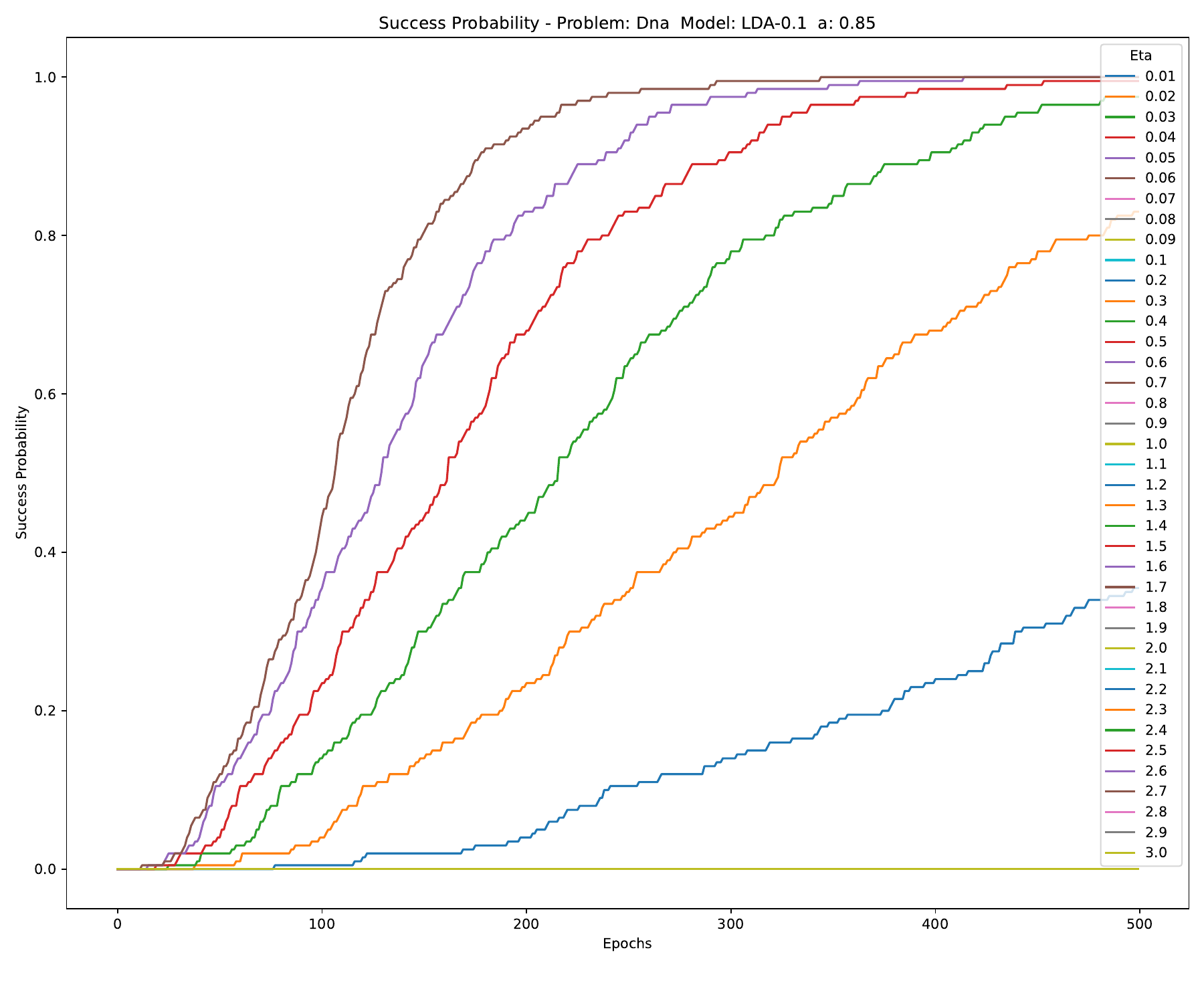} \\[-3mm]
    \includegraphics[clip,trim=0 0 0 0,width=.52\linewidth]{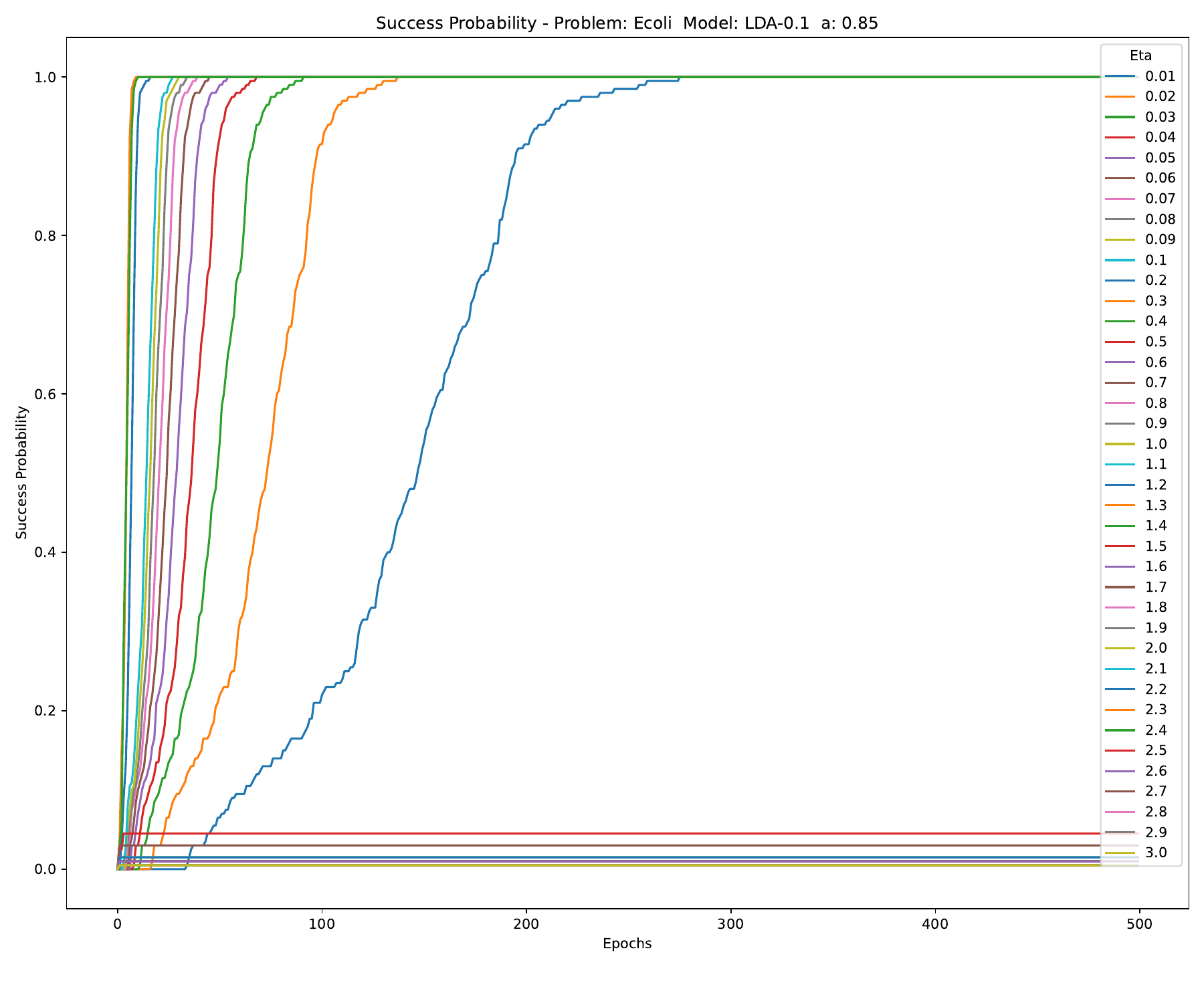} &
    \includegraphics[clip,trim=0 0 0 0,width=.52\linewidth]{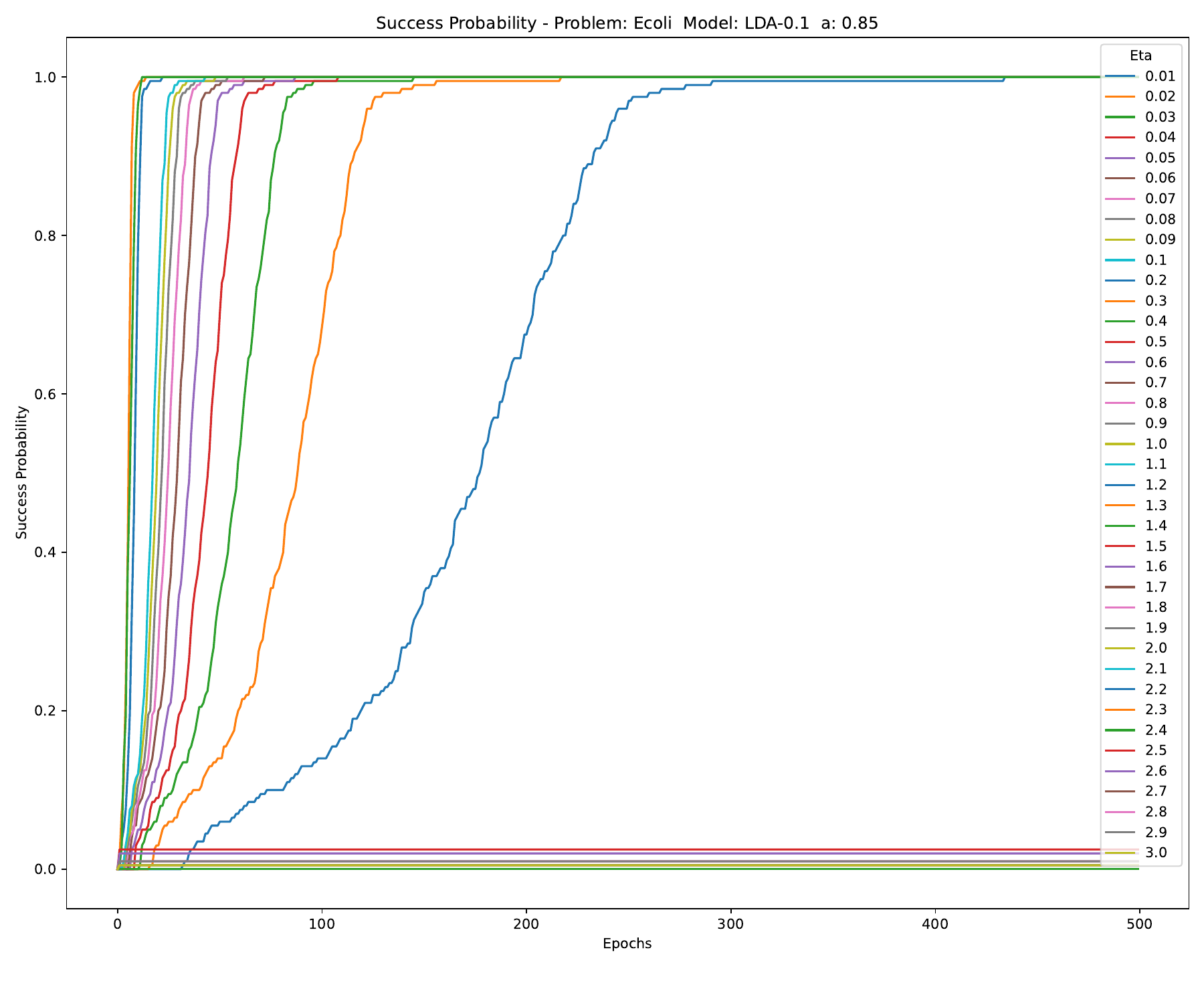} \\[-3mm]
  \end{tabular}
\caption{Success probability ($a=0.85$) for LDA-0.1 model.}
  \label{fig:success_probability_LDA_0.1_model_0.85}
\end{figure*}

\begin{figure*}[p]
  \centering
  \ContinuedFloat
  \begin{tabular}{c@{}c}
    Training Success Probability & Validation Success Probability \\
    \includegraphics[clip,trim=0 0 0 0,width=.52\linewidth]{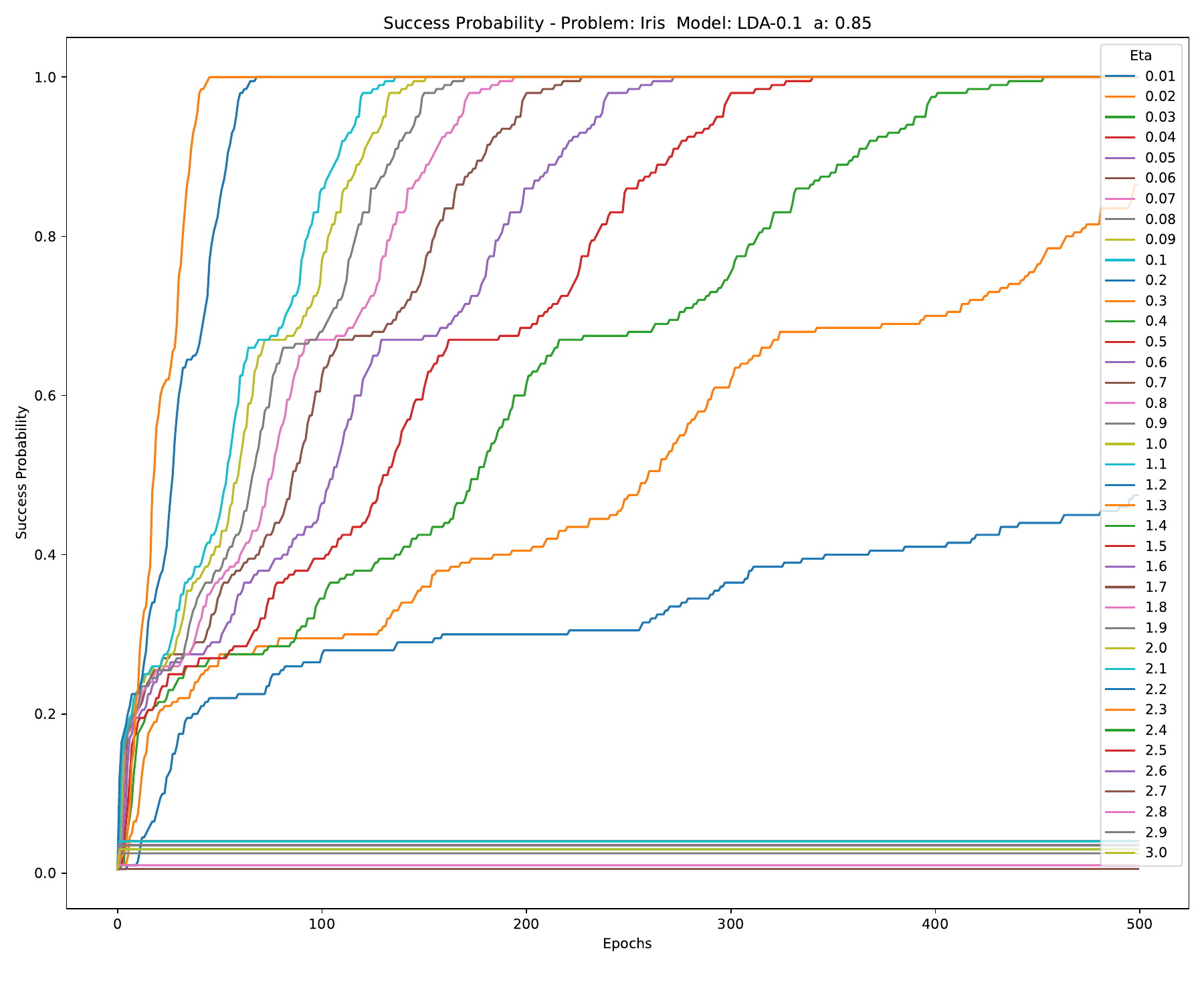} &
    \includegraphics[clip,trim=0 0 0 0,width=.52\linewidth]{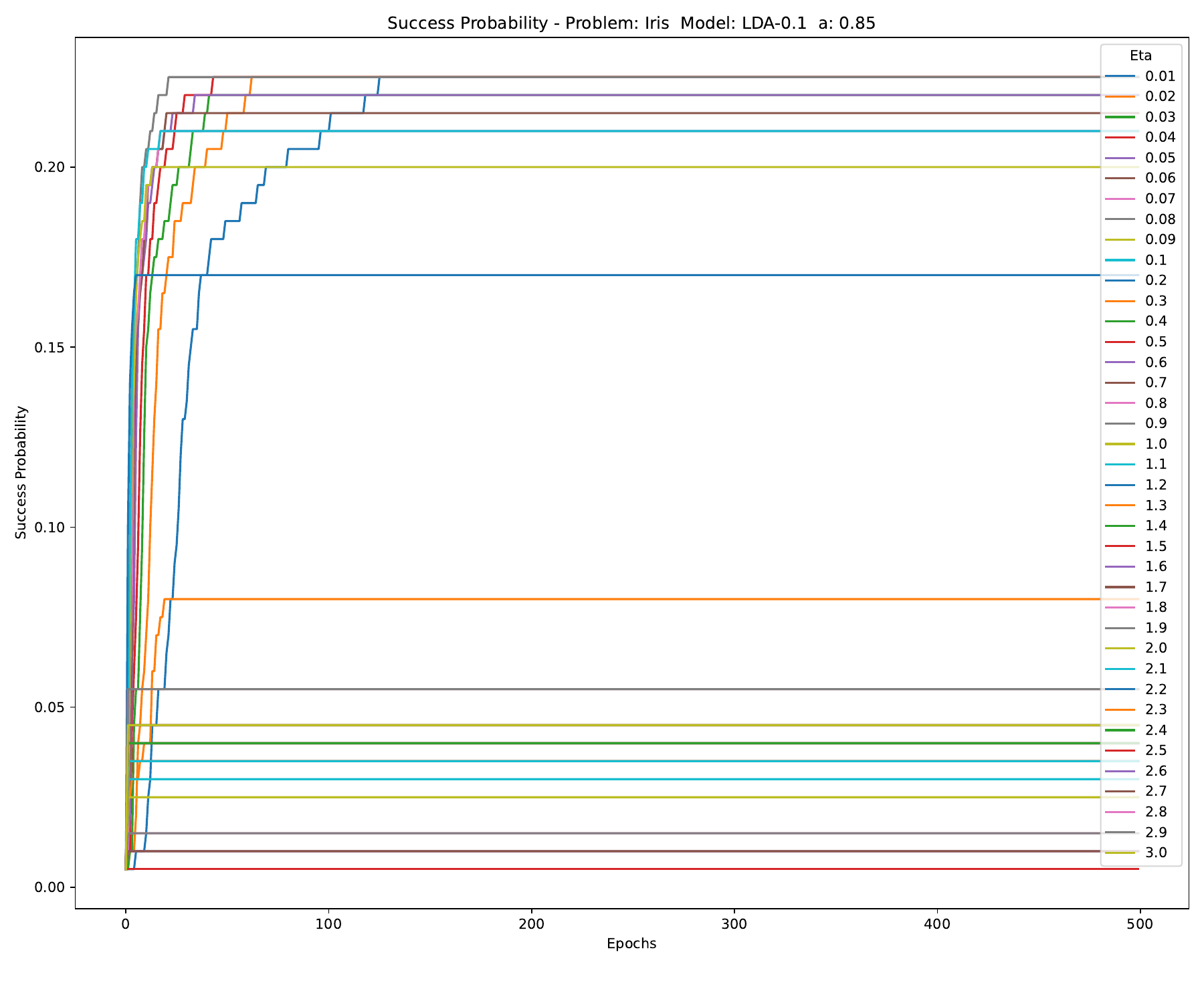} \\[-3mm]
    \includegraphics[clip,trim=0 0 0 0,width=.52\linewidth]{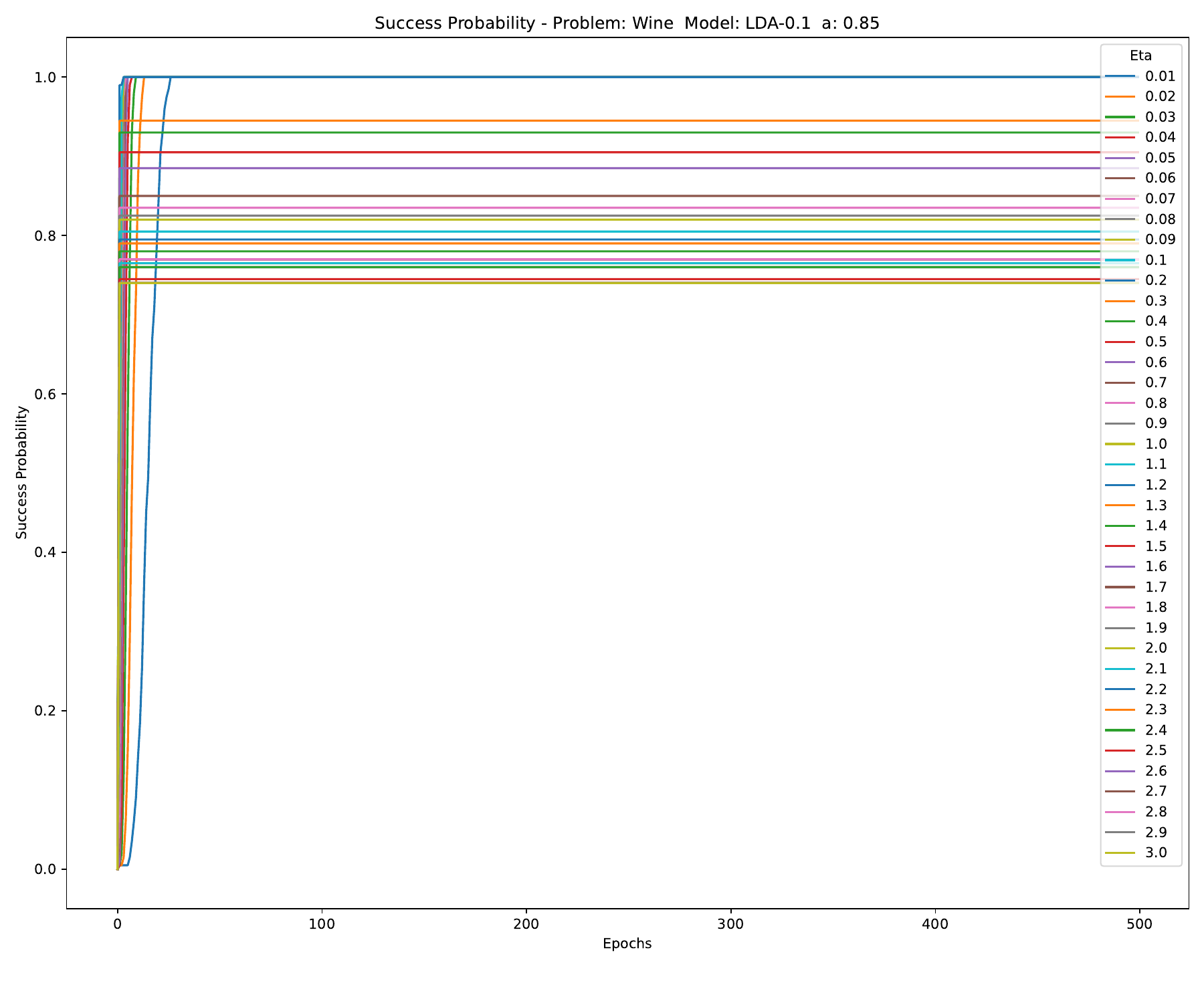} &
    \includegraphics[clip,trim=0 0 0 0,width=.52\linewidth]{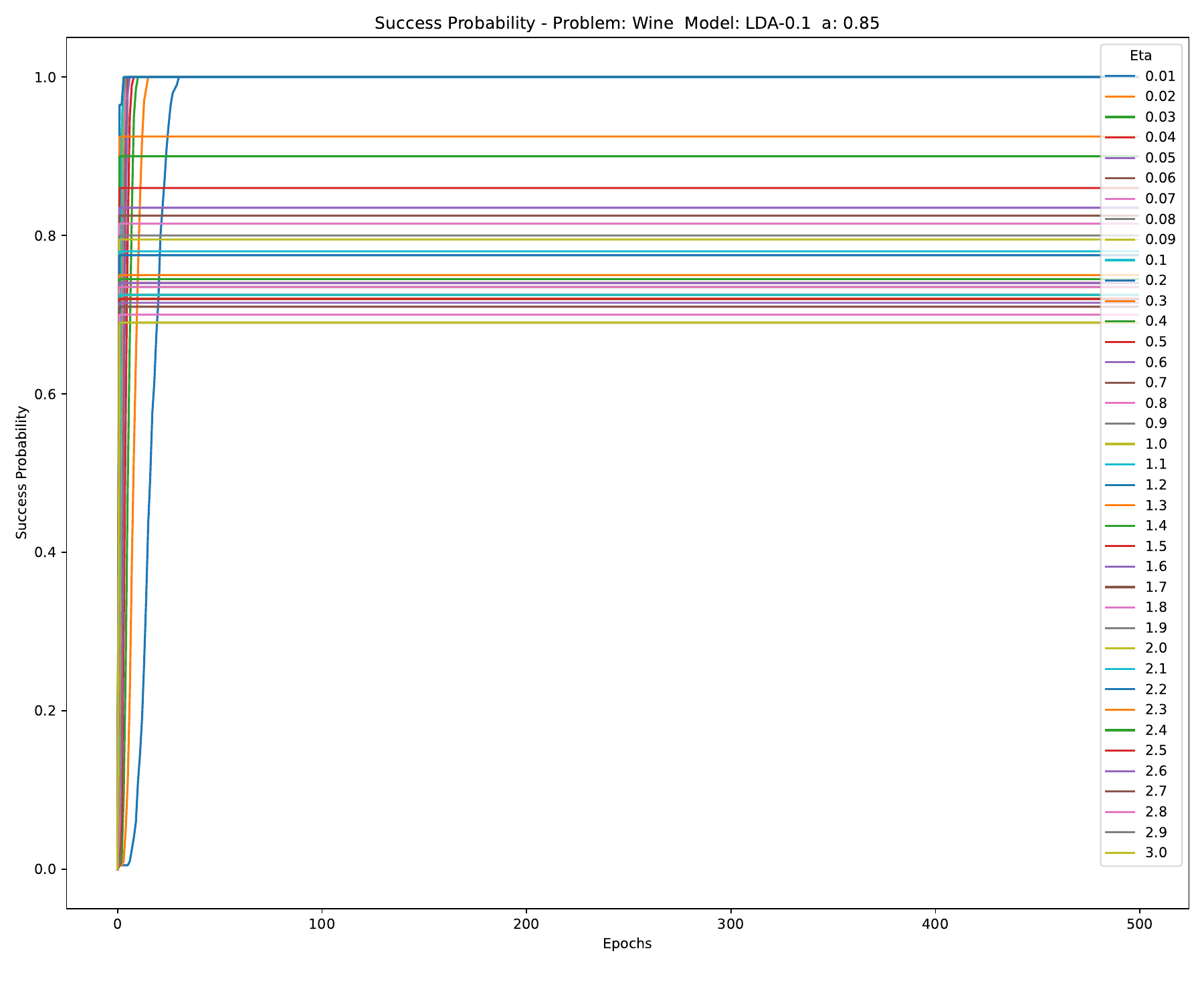} \\[-3mm]
  \end{tabular}
\caption{Success probability ($a=0.85$) for LDA-0.1 model (continued).}
\end{figure*}

\begin{figure*}[p]
  \centering
  \begin{tabular}{c@{}c}
    Training Success Probability & Validation Success Probability \\
    \includegraphics[clip,trim=0 0 0 0,width=.52\linewidth]{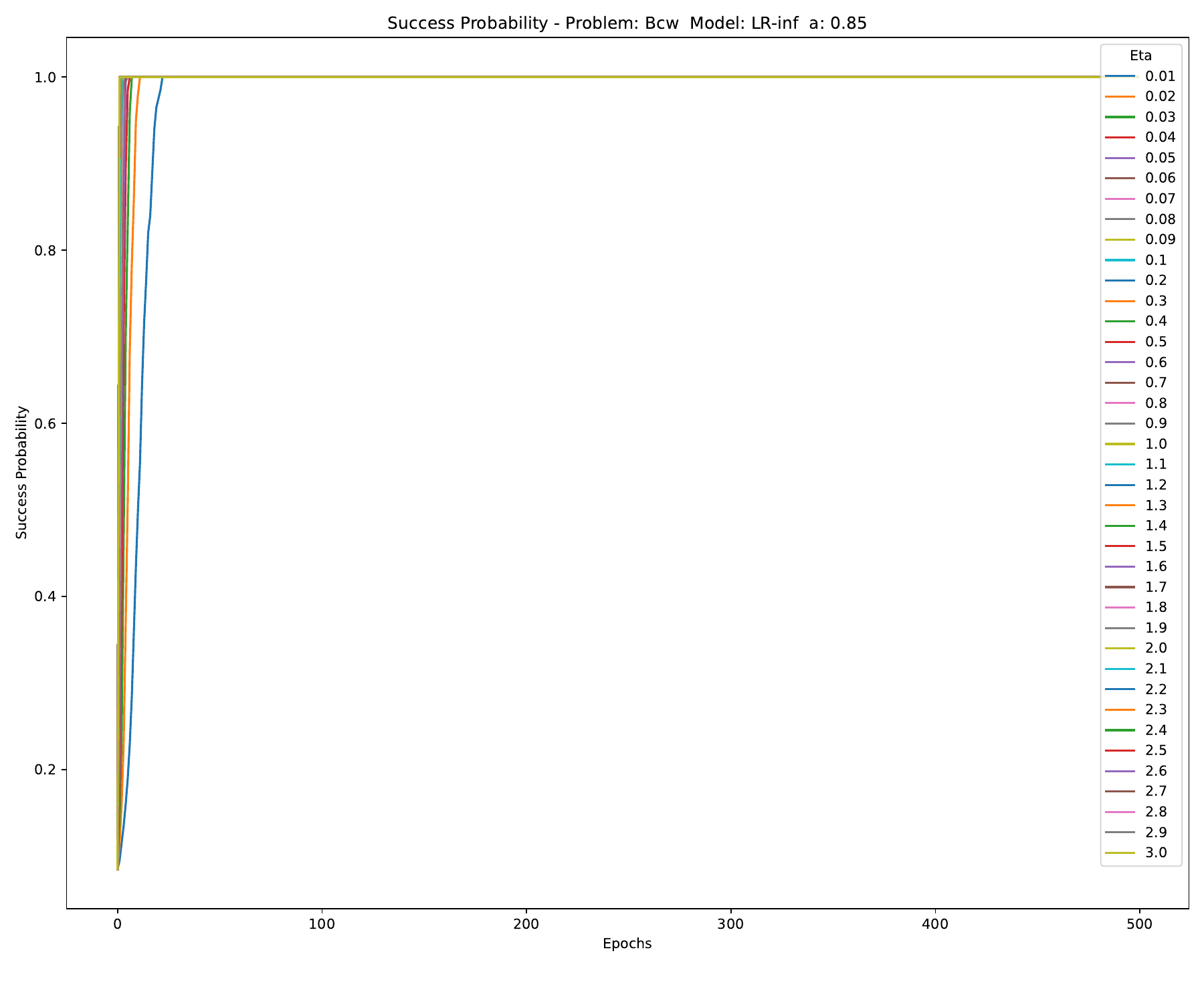} &
    \includegraphics[clip,trim=0 0 0 0,width=.52\linewidth]{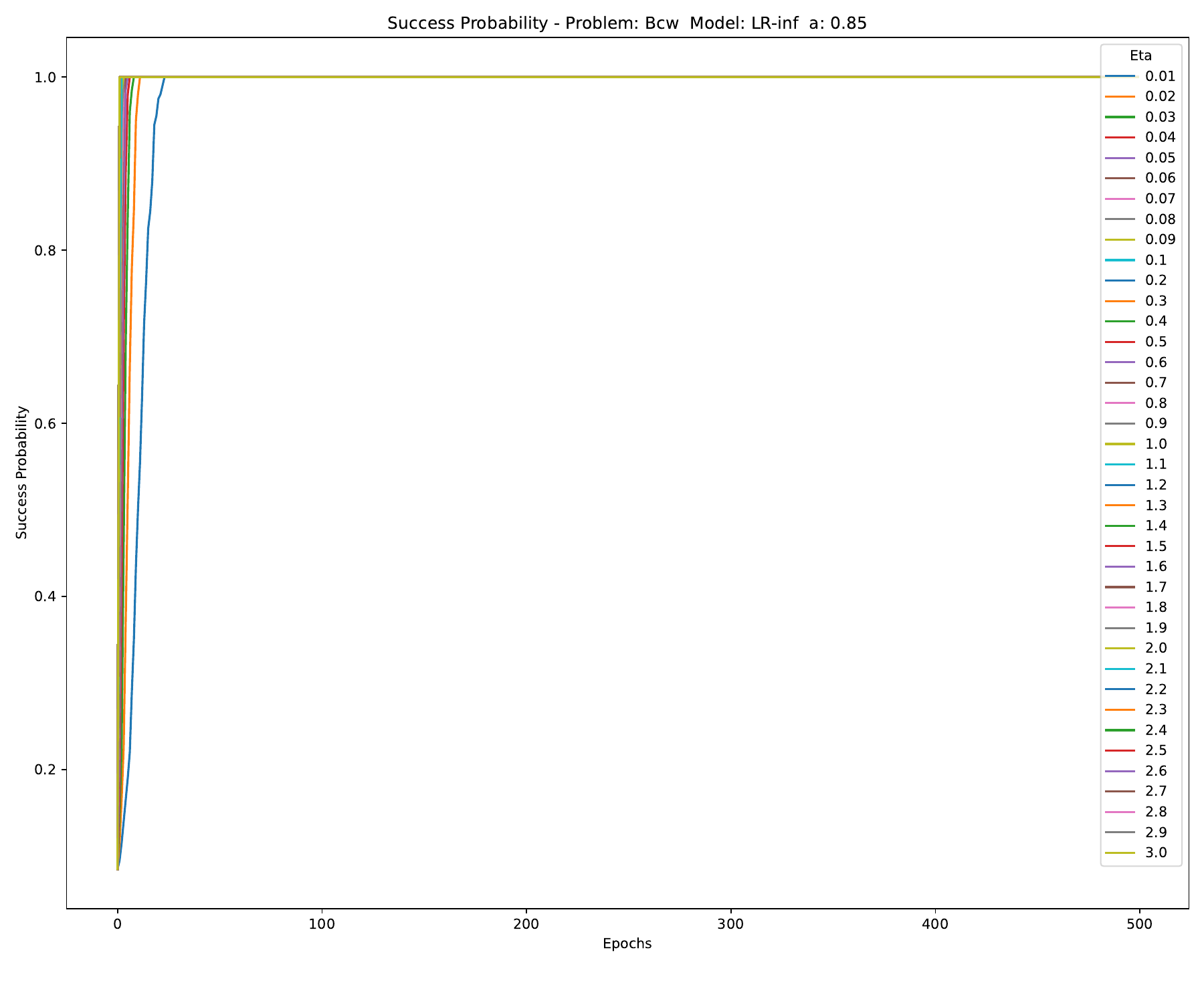} \\[-3mm]
    \includegraphics[clip,trim=0 0 0 0,width=.52\linewidth]{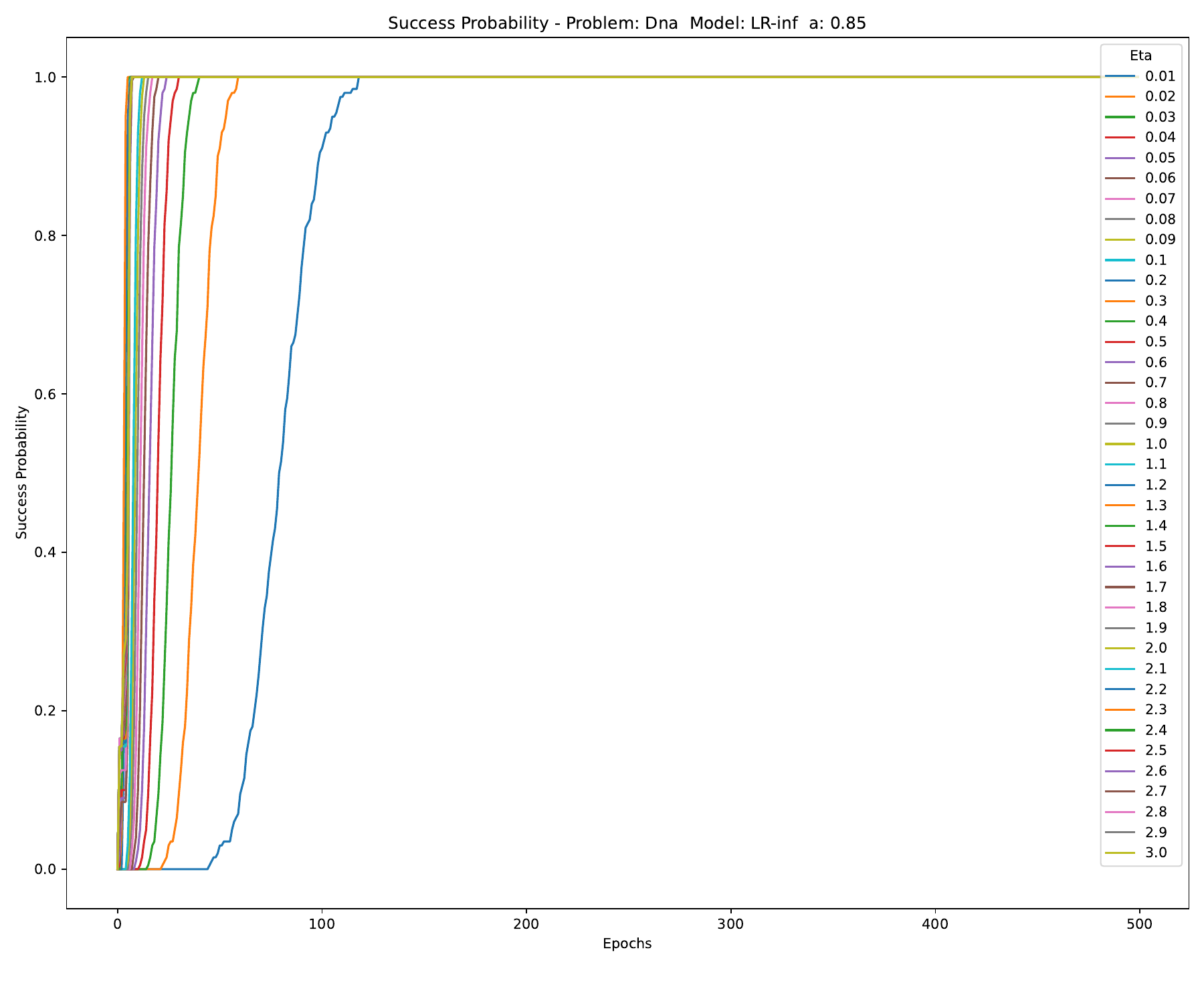} &
    \includegraphics[clip,trim=0 0 0 0,width=.52\linewidth]{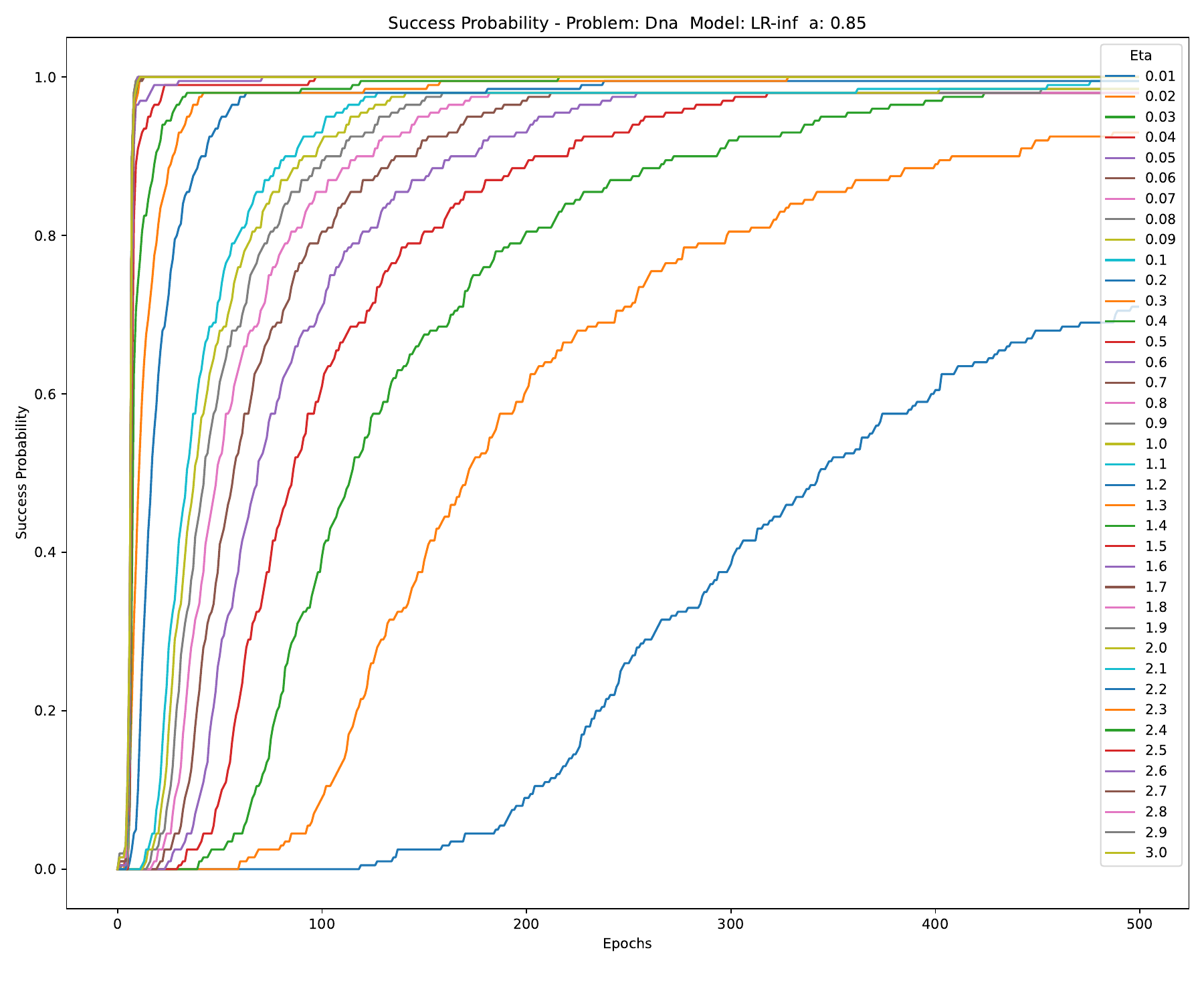} \\[-3mm]
    \includegraphics[clip,trim=0 0 0 0,width=.52\linewidth]{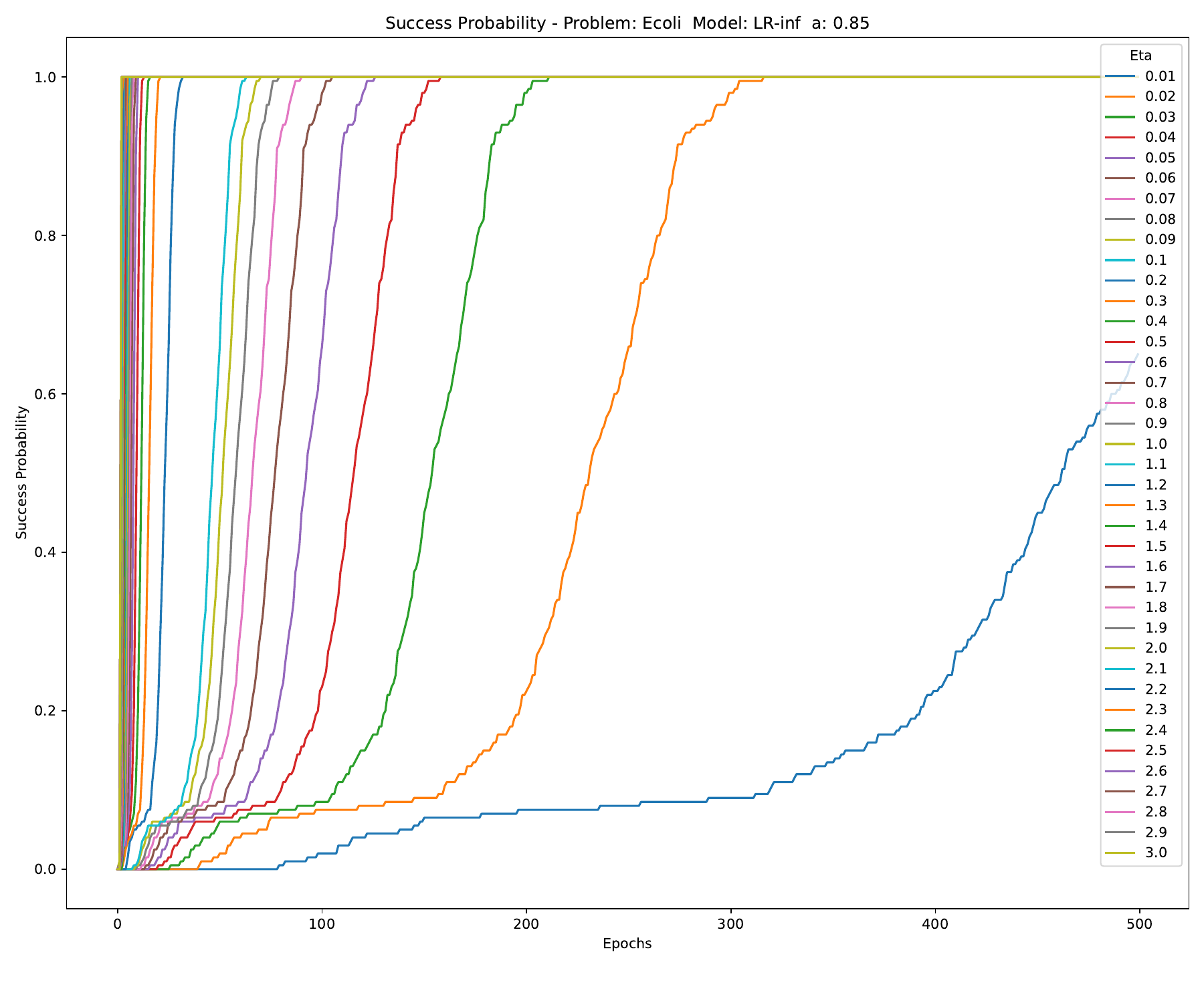} &
    \includegraphics[clip,trim=0 0 0 0,width=.52\linewidth]{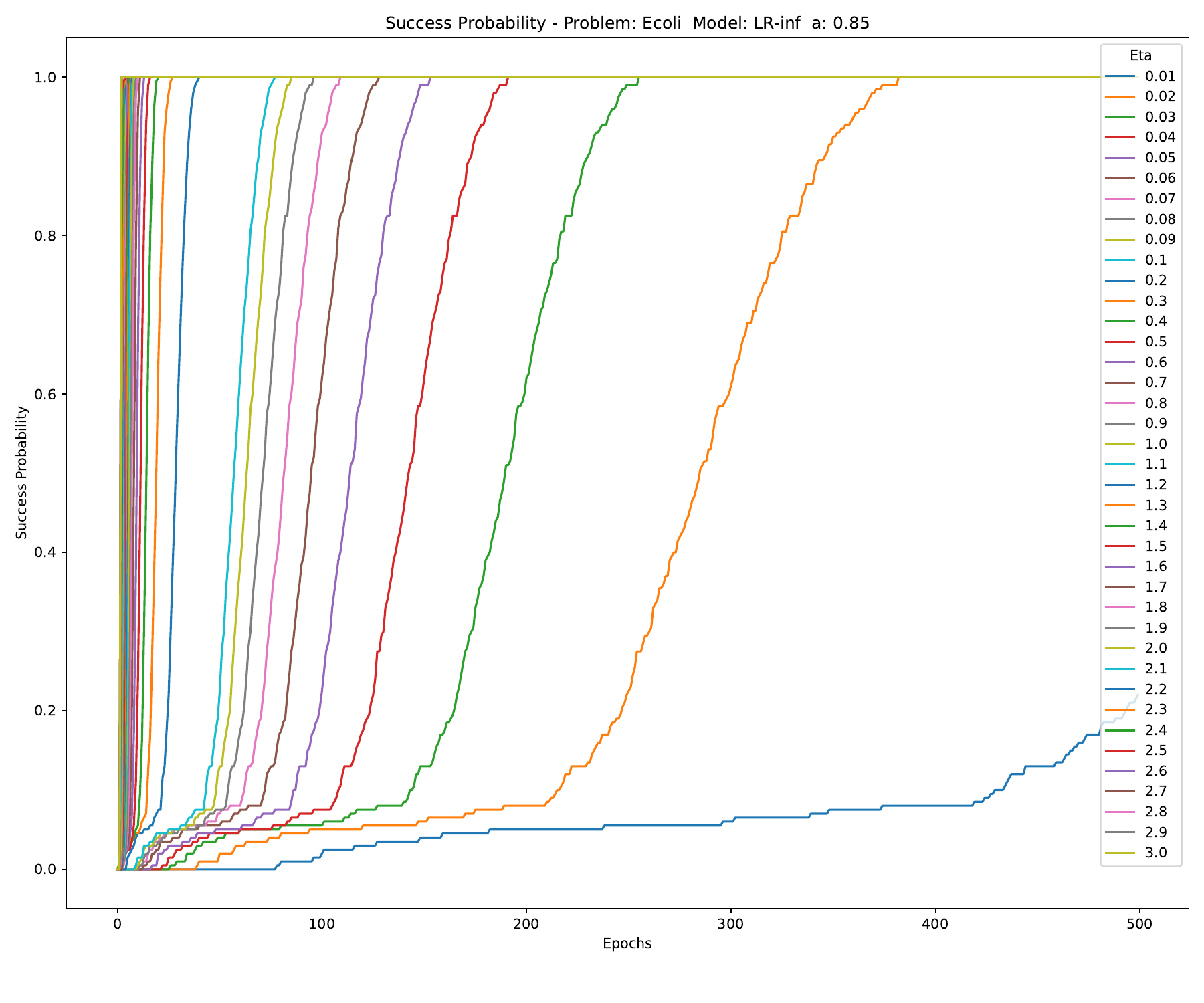} \\[-3mm]
  \end{tabular}
\caption{Success probability ($a=0.85$) for LR-Inf model.}
  \label{fig:success_probability_LR_Inf_model_0.85}
\end{figure*}

\begin{figure*}[p]
  \centering
  \ContinuedFloat
  \begin{tabular}{c@{}c}
    Training Success Probability & Validation Success Probability \\
    \includegraphics[clip,trim=0 0 0 0,width=.52\linewidth]{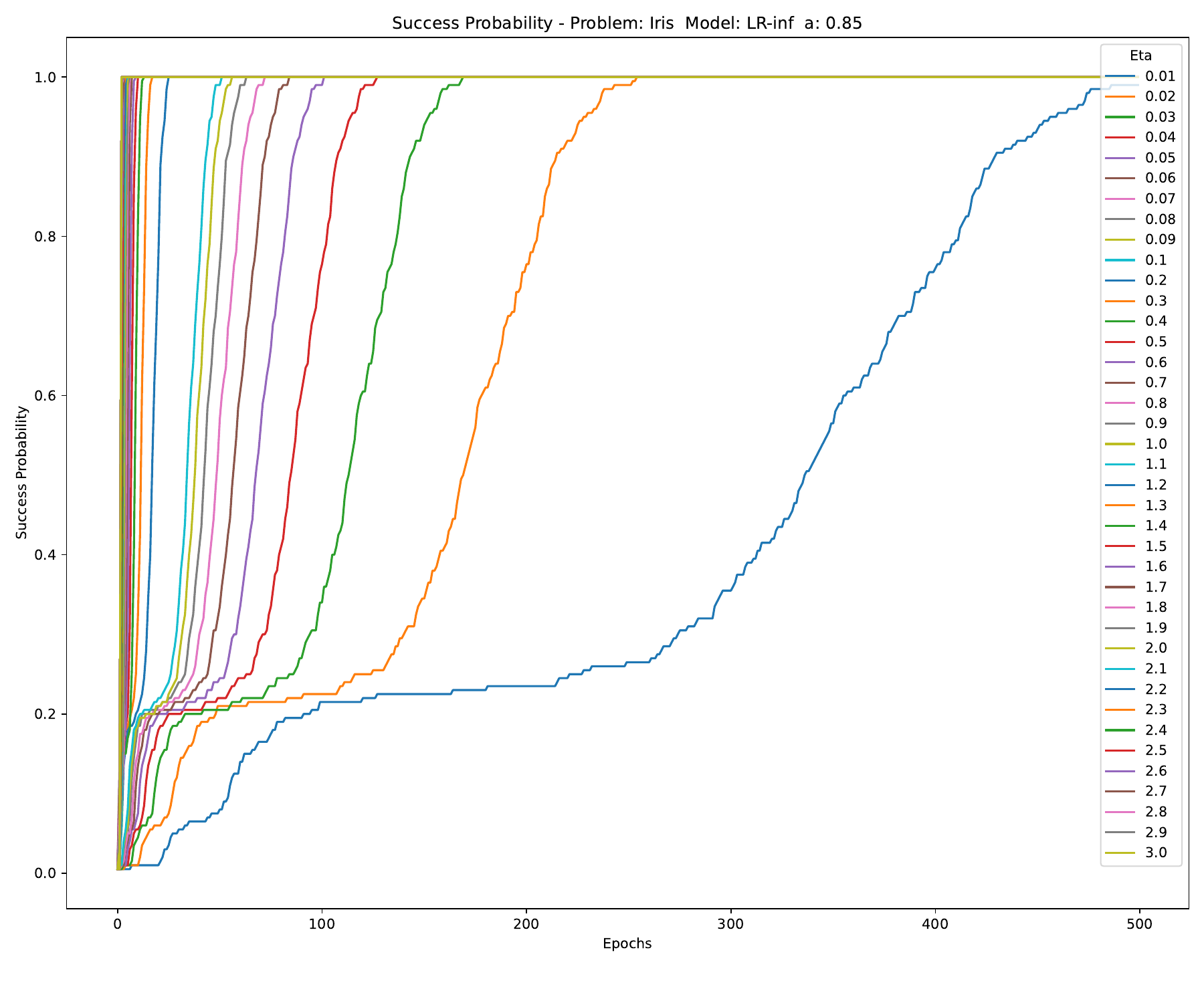} &
    \includegraphics[clip,trim=0 0 0 0,width=.52\linewidth]{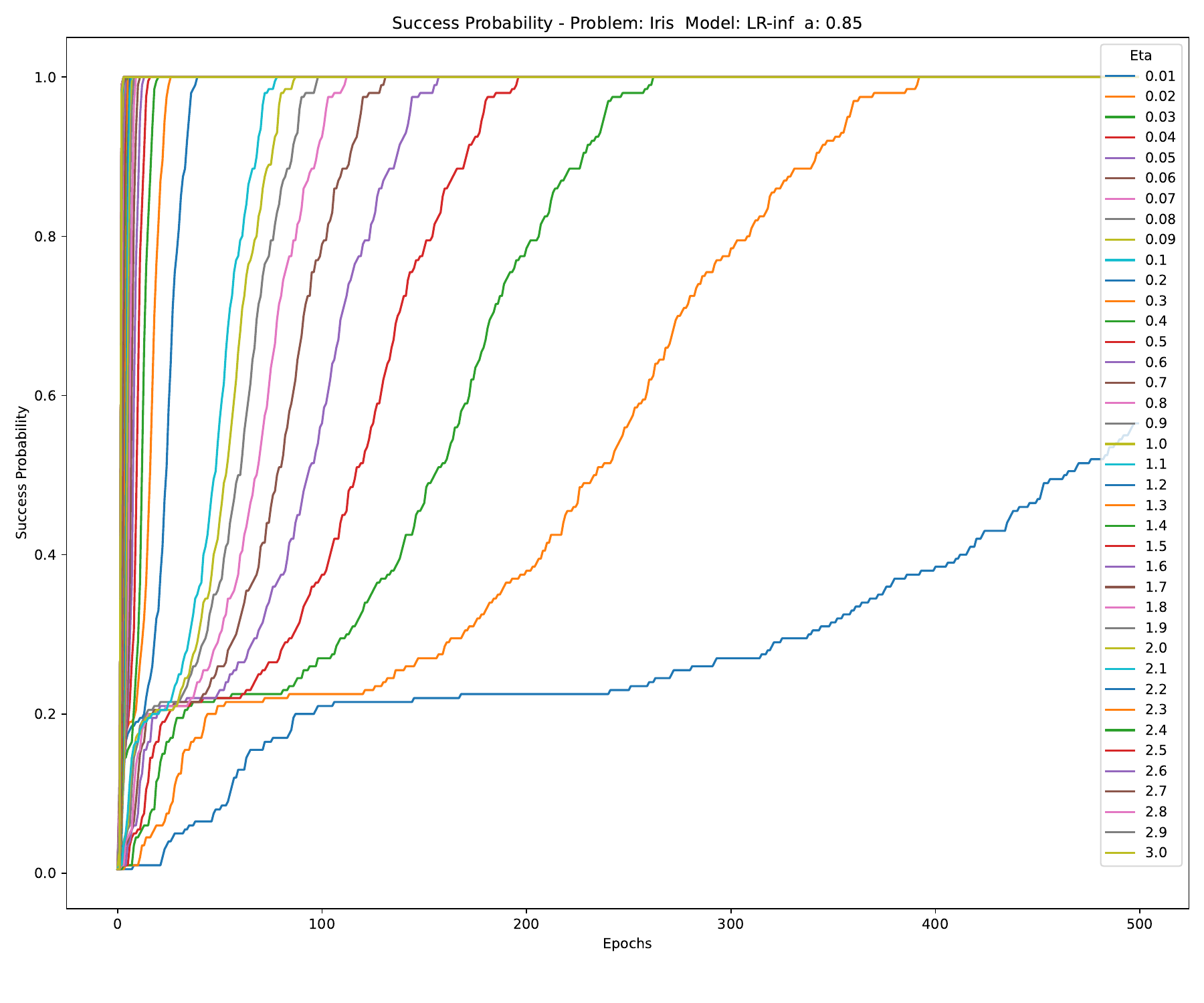} \\[-3mm]
    \includegraphics[clip,trim=0 0 0 0,width=.52\linewidth]{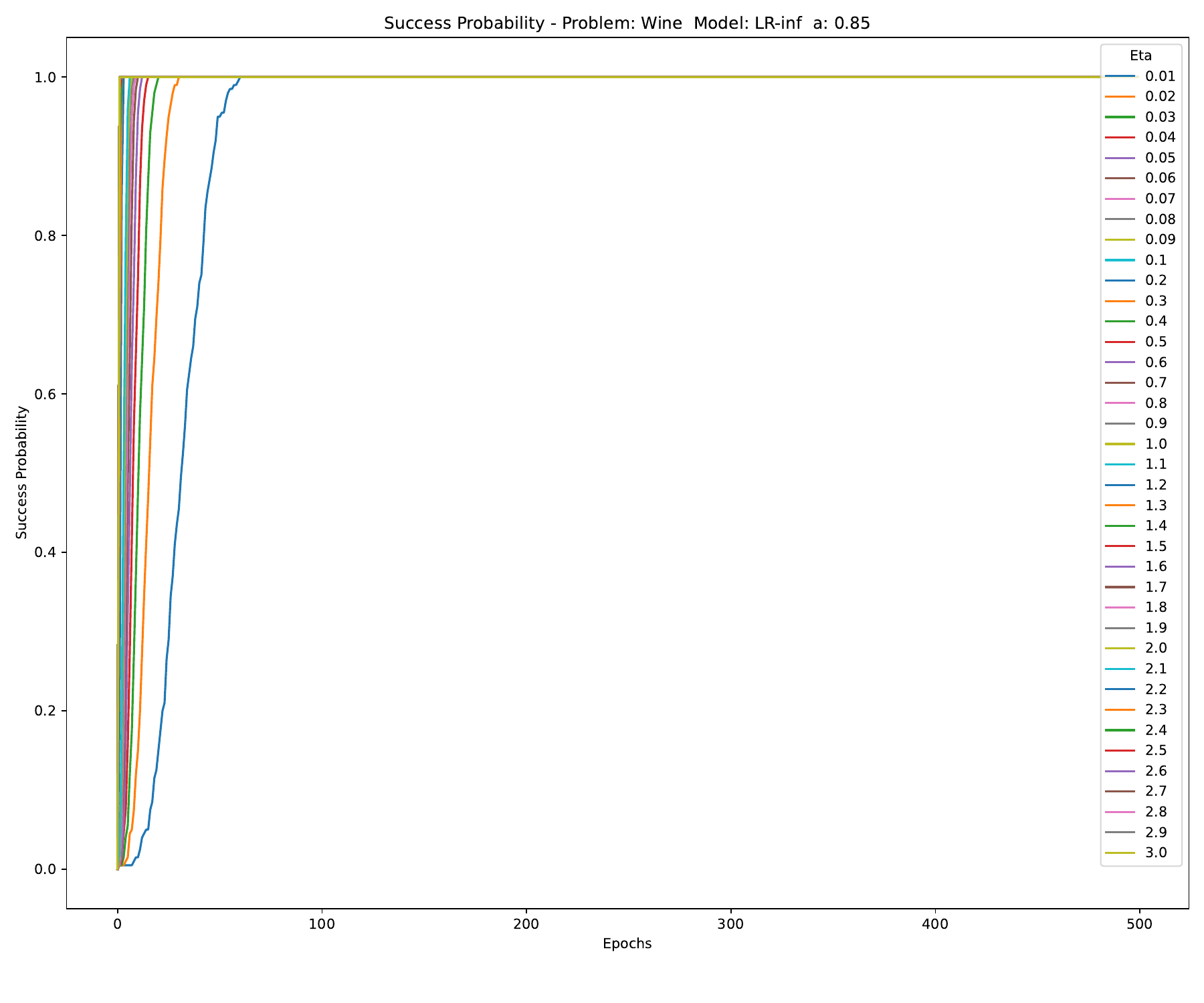} &
    \includegraphics[clip,trim=0 0 0 0,width=.52\linewidth]{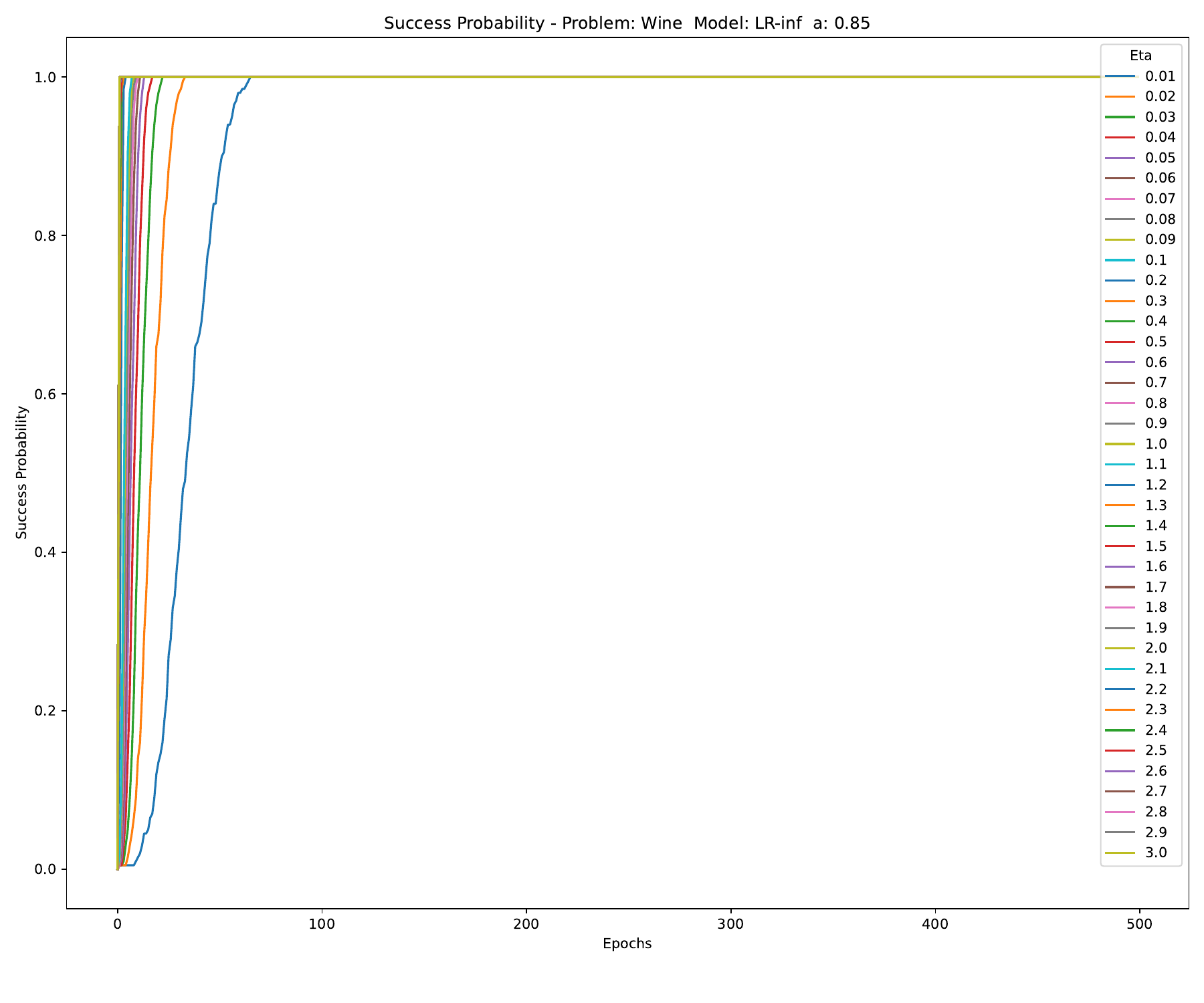} \\[-3mm]
  \end{tabular}
\caption{Success probability ($a=0.85$) for LR-Inf model (continued).}
\end{figure*}

\begin{figure*}[p]
  \centering
  \begin{tabular}{c@{}c}
    Training Success Probability & Validation Success Probability \\
    \includegraphics[clip,trim=0 0 0 0,width=.52\linewidth]{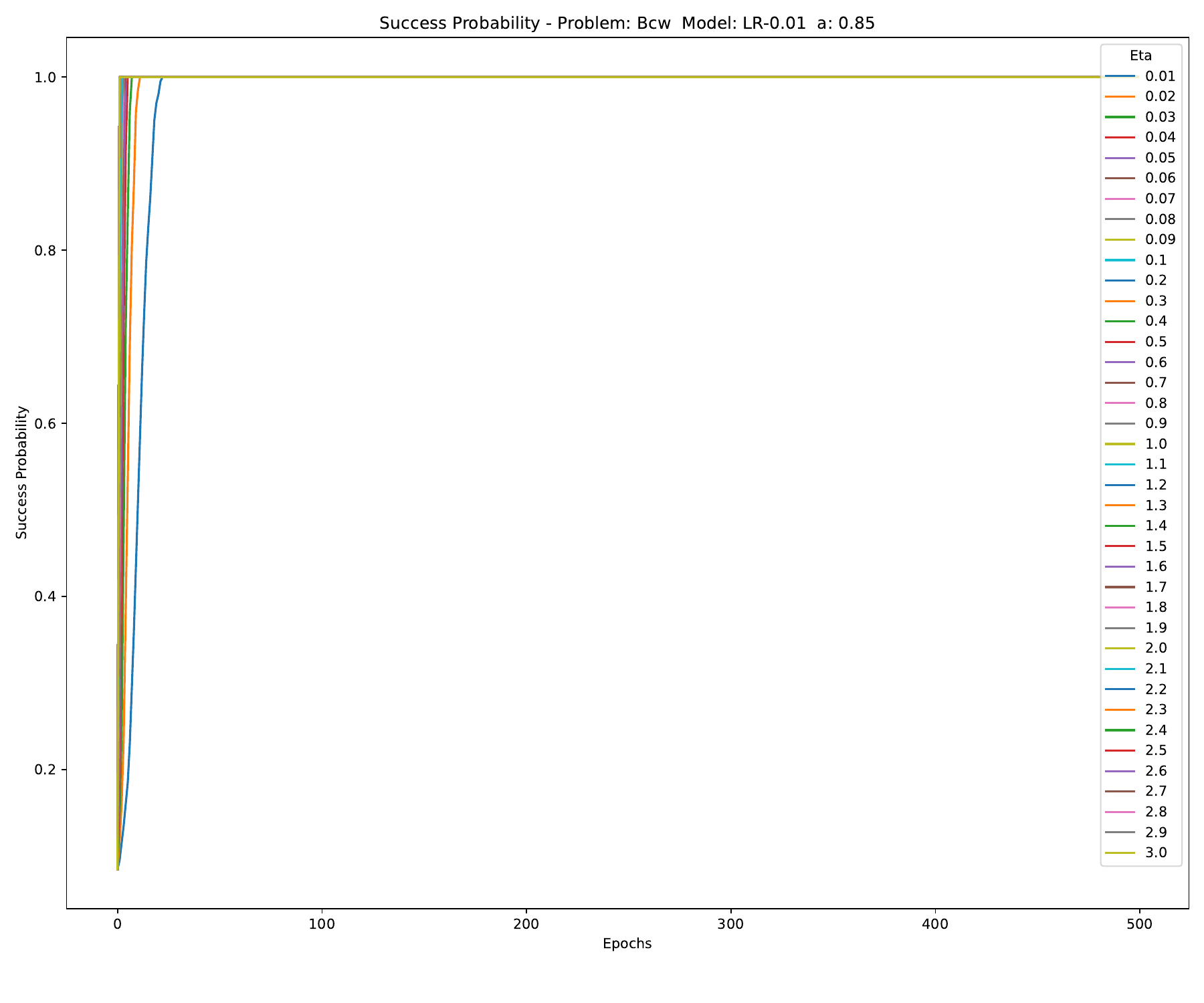} &
    \includegraphics[clip,trim=0 0 0 0,width=.52\linewidth]{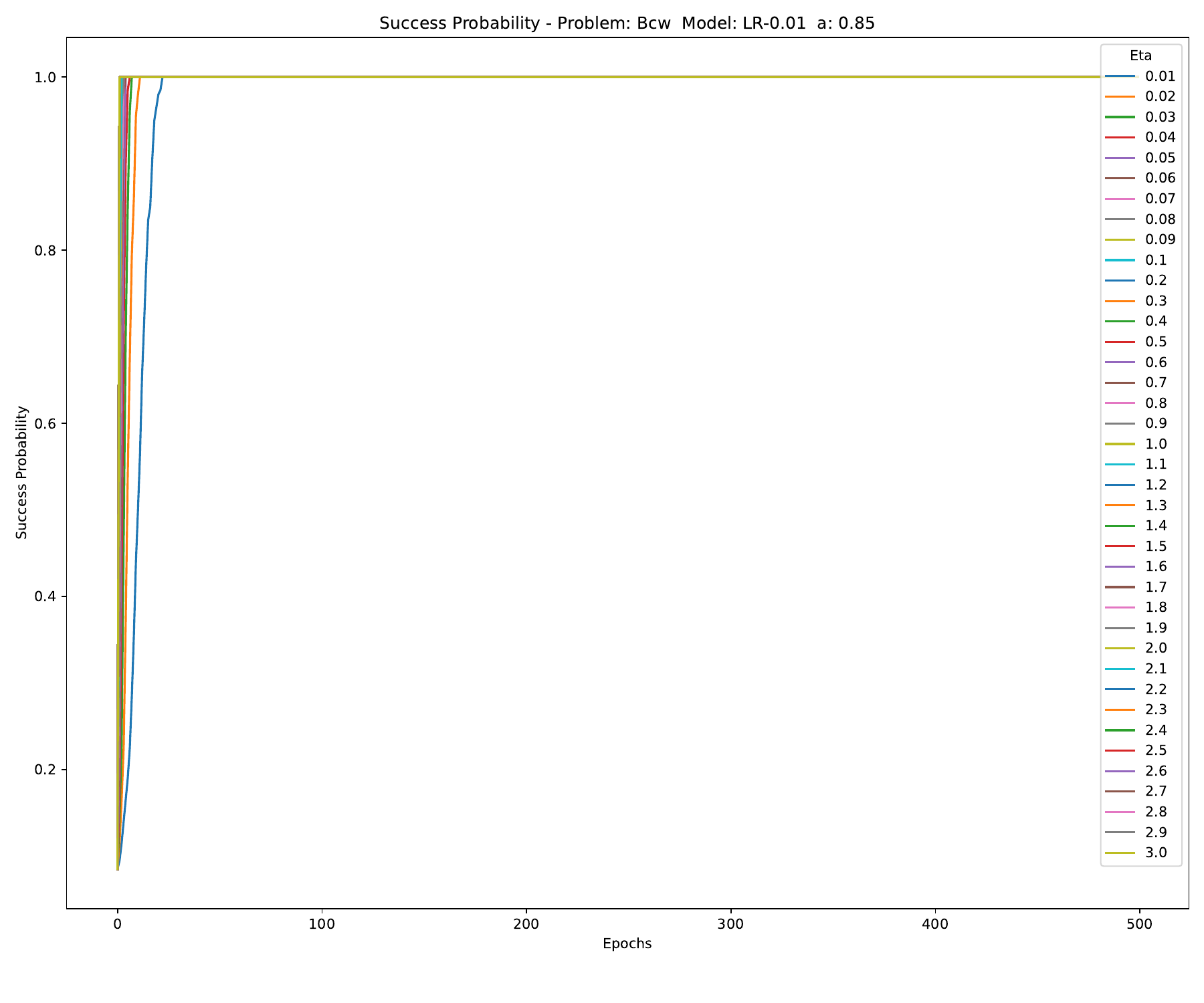} \\[-3mm]
    \includegraphics[clip,trim=0 0 0 0,width=.52\linewidth]{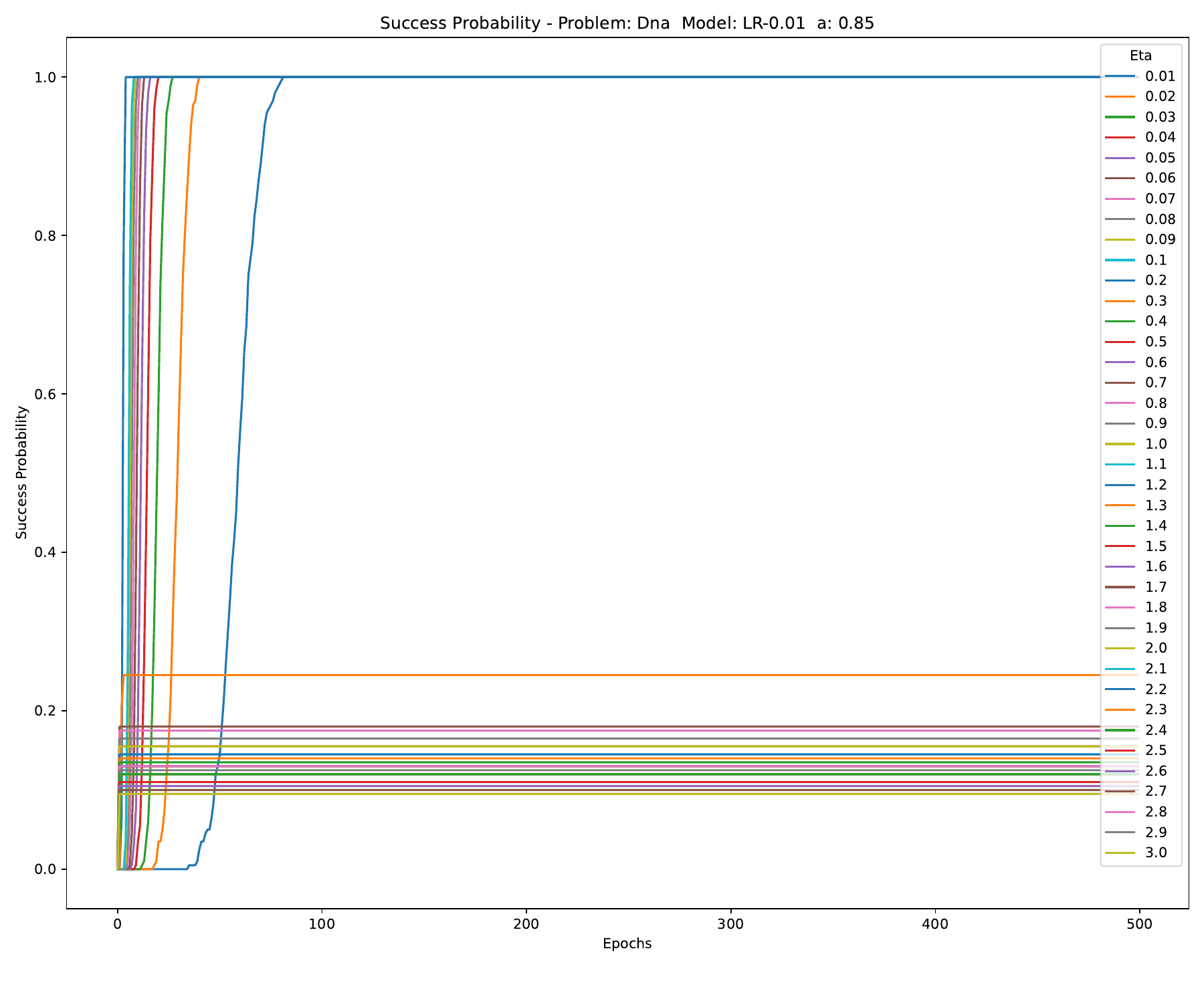} &
    \includegraphics[clip,trim=0 0 0 0,width=.52\linewidth]{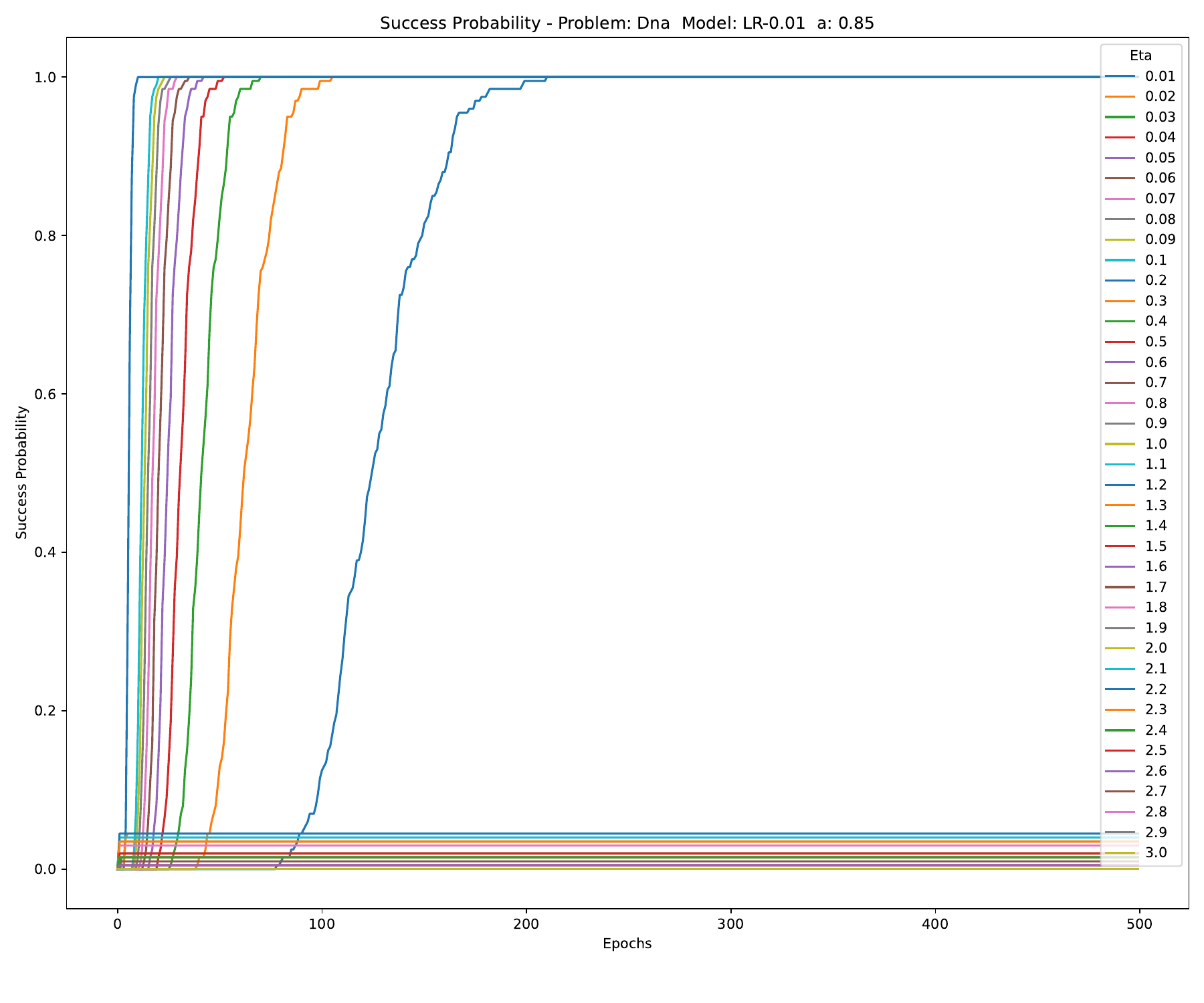} \\[-3mm]
    \includegraphics[clip,trim=0 0 0 0,width=.52\linewidth]{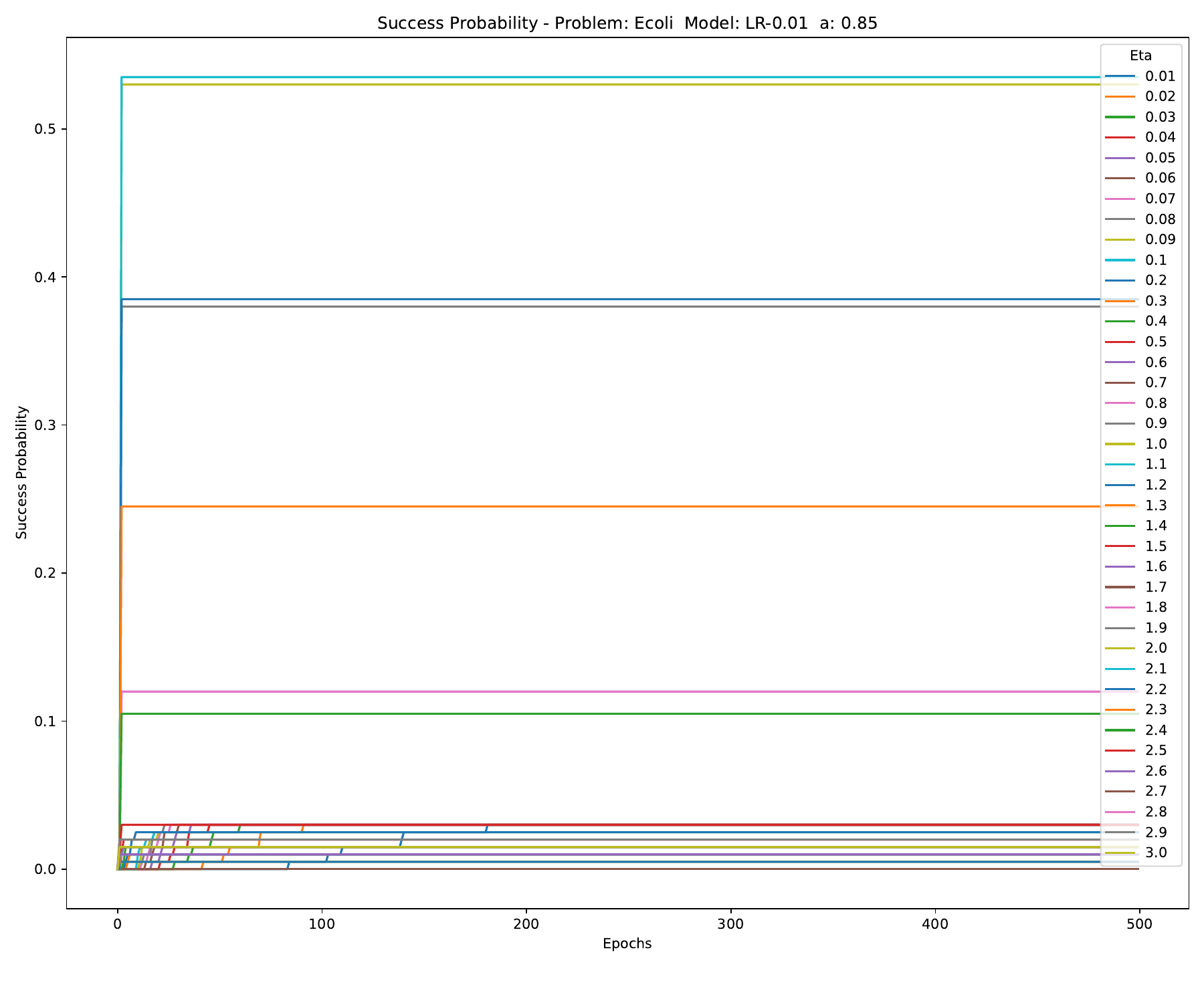} &
    \includegraphics[clip,trim=0 0 0 0,width=.52\linewidth]{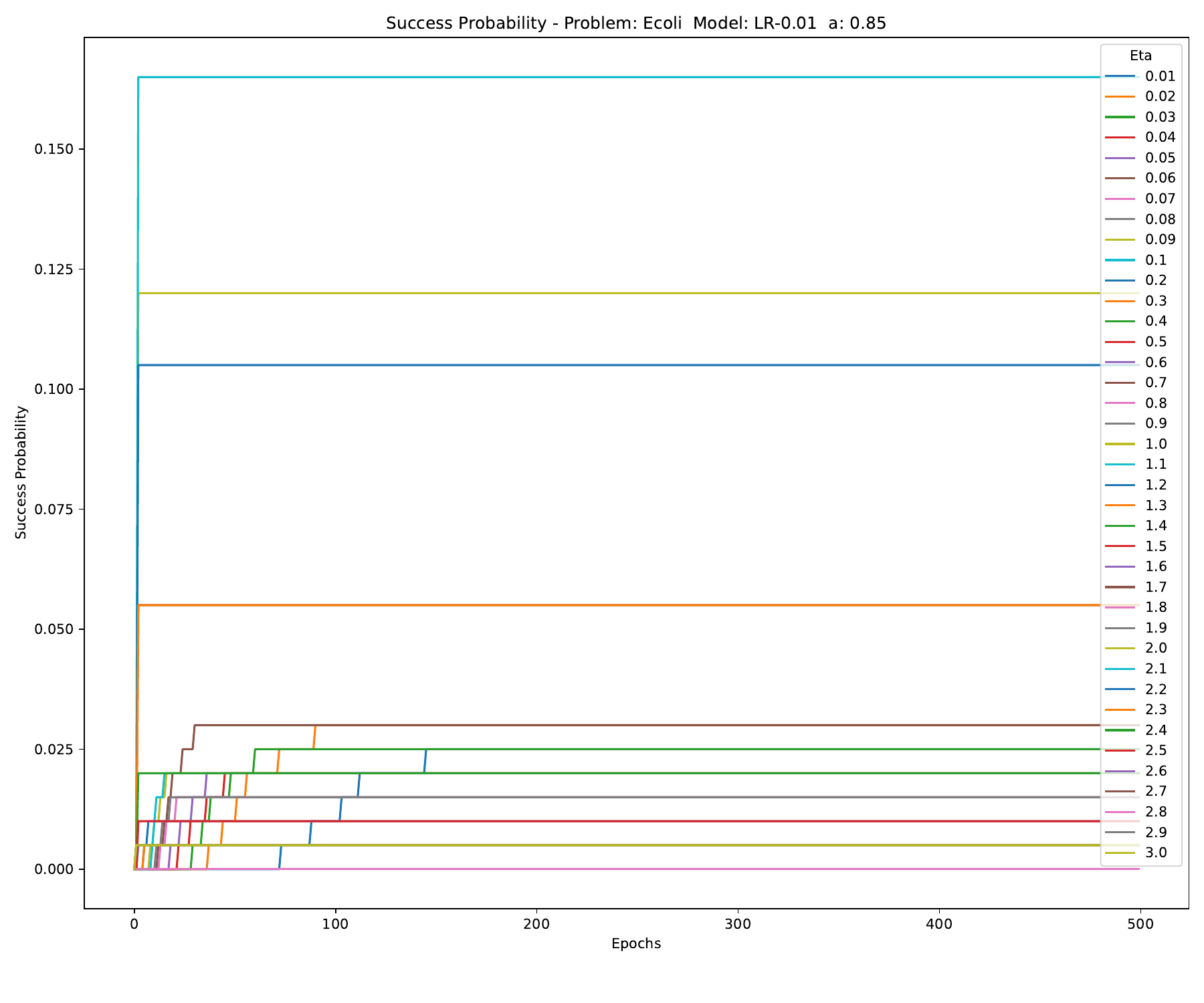} \\[-3mm]
  \end{tabular}
\caption{Success probability ($a=0.85$) for LR-0.01 model.}
  \label{fig:success_probability_LR_0.01_model_0.85}
\end{figure*}

\begin{figure*}[p]
  \centering
  \ContinuedFloat
  \begin{tabular}{c@{}c}
    Training Success Probability & Validation Success Probability \\
    \includegraphics[clip,trim=0 0 0 0,width=.52\linewidth]{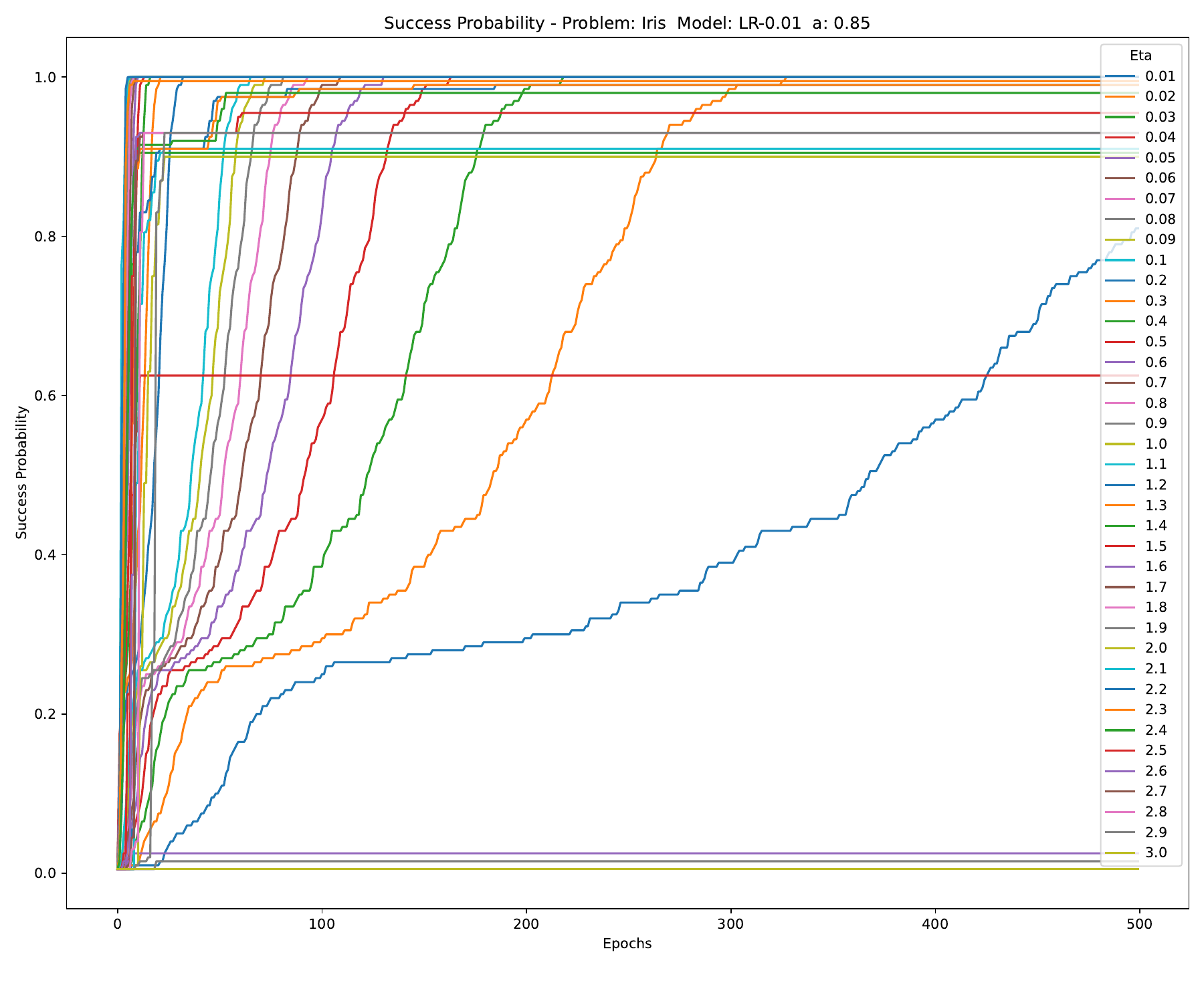} &
    \includegraphics[clip,trim=0 0 0 0,width=.52\linewidth]{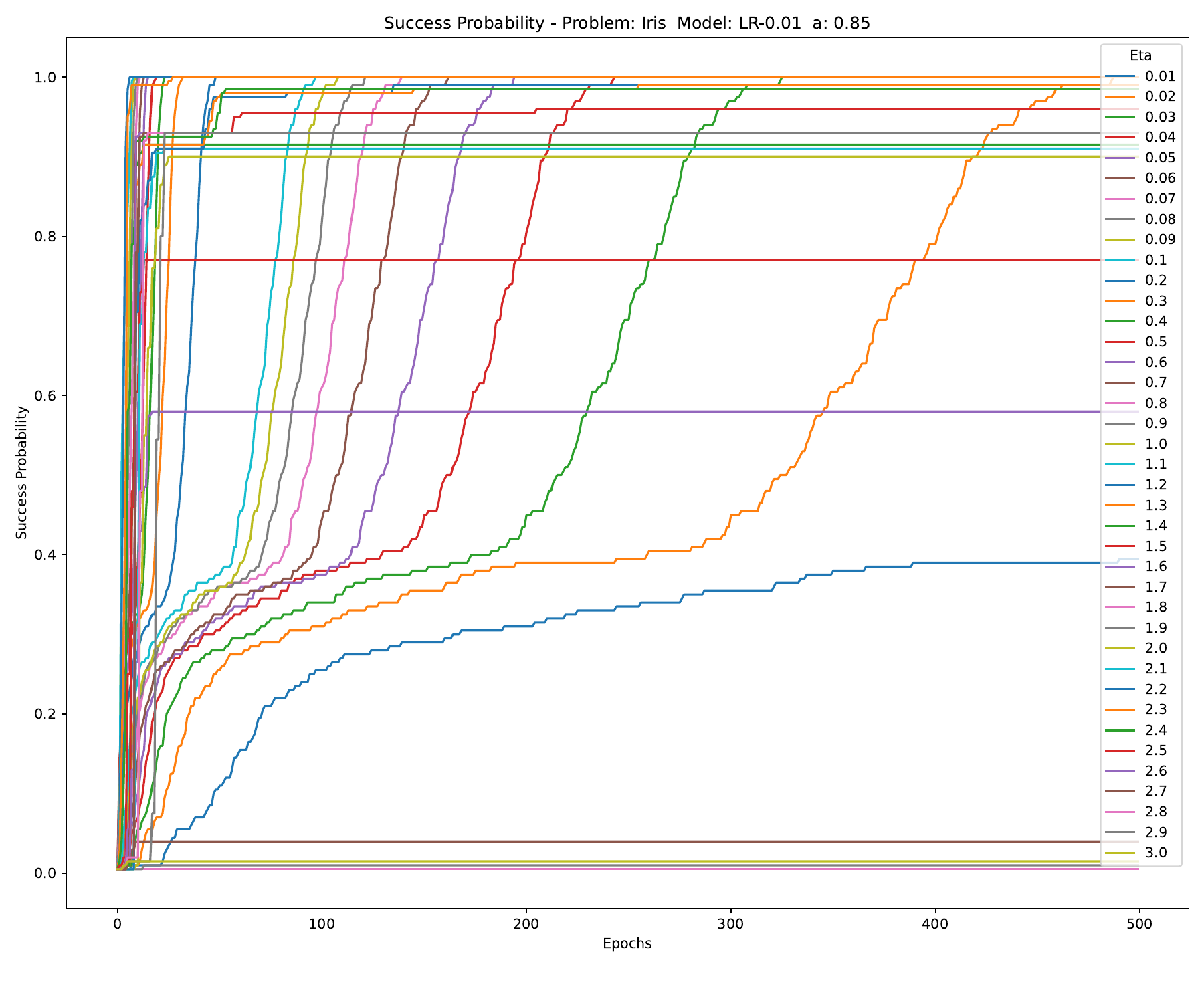} \\[-3mm]
    \includegraphics[clip,trim=0 0 0 0,width=.52\linewidth]{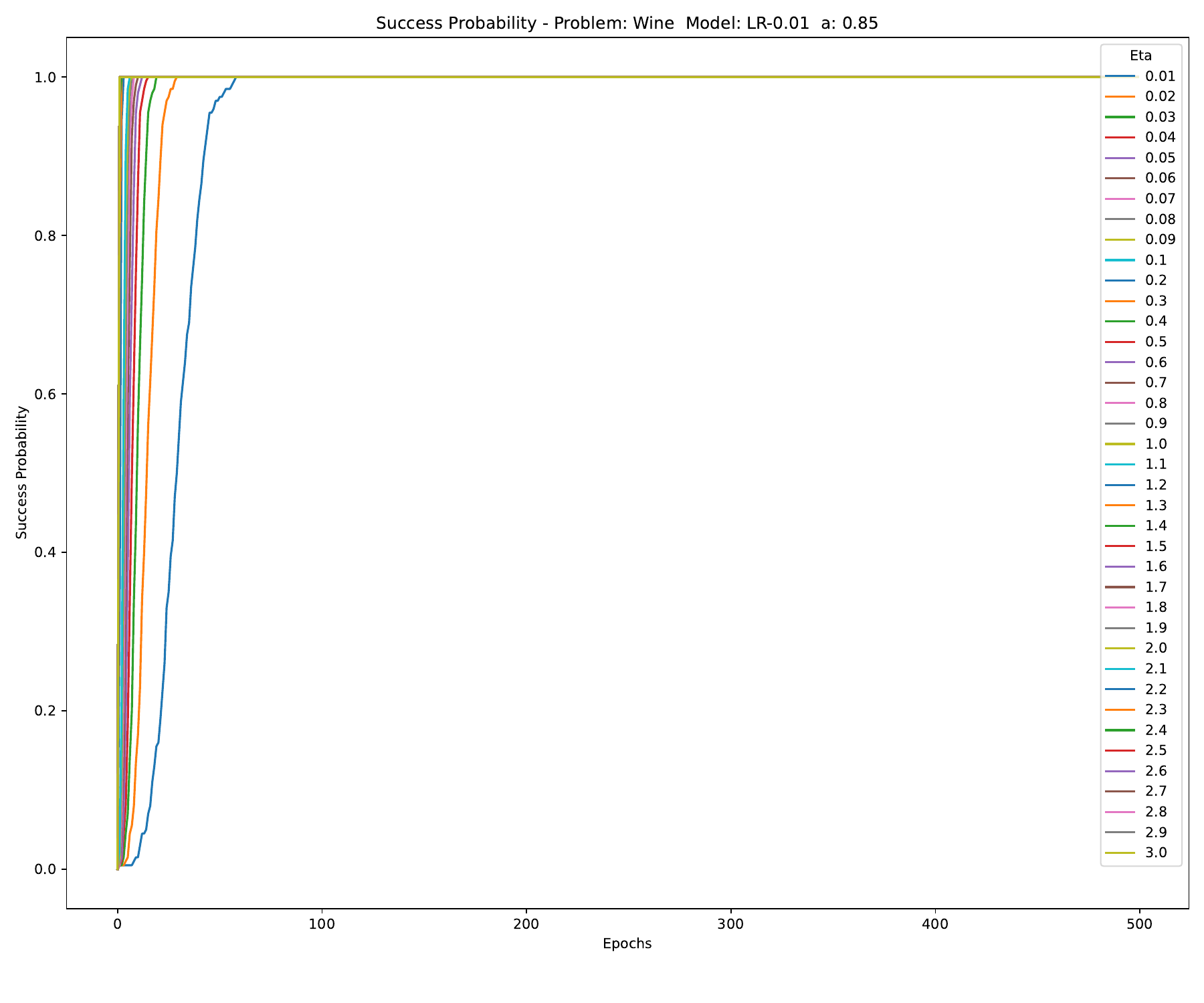} &
    \includegraphics[clip,trim=0 0 0 0,width=.52\linewidth]{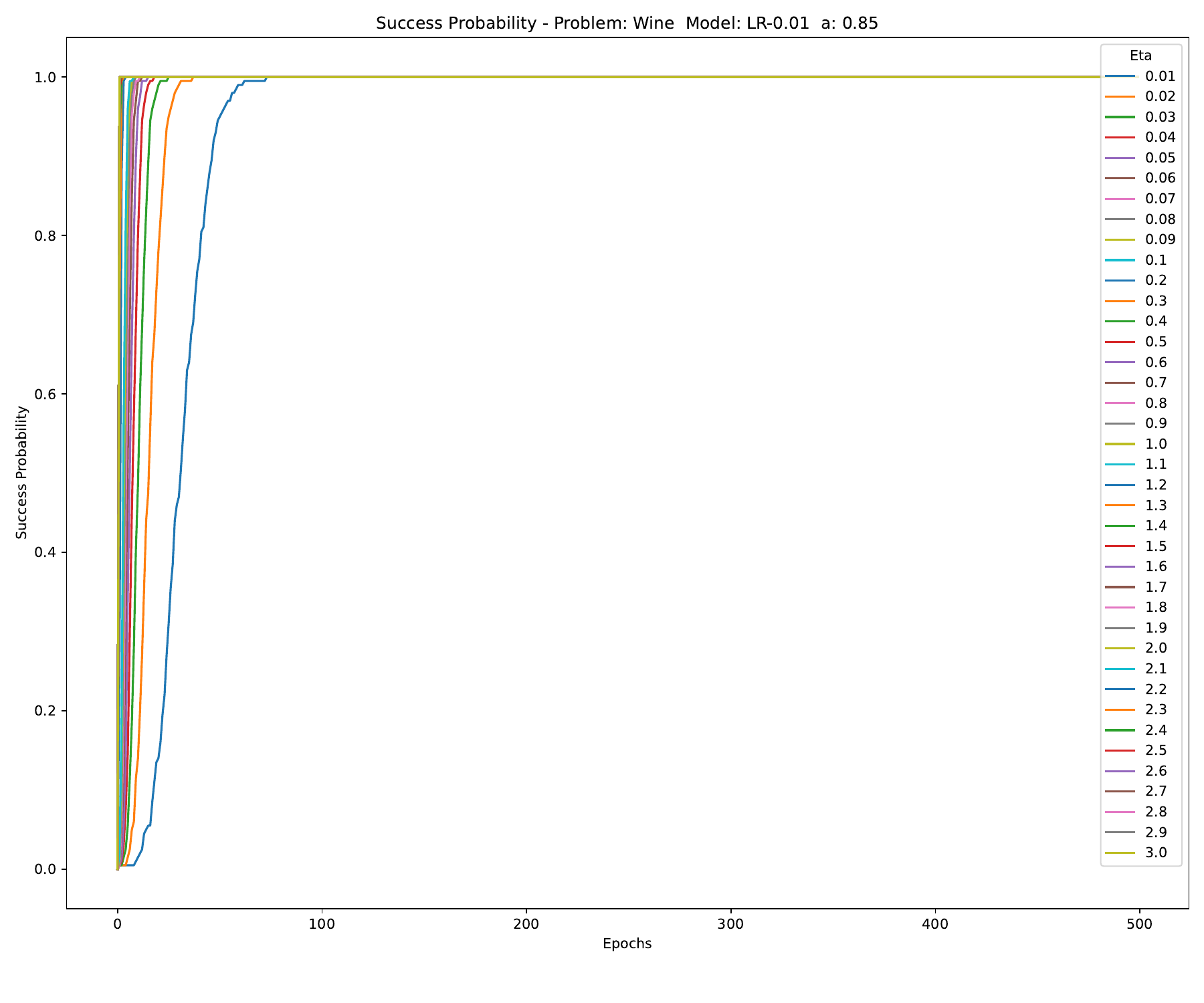} \\[-3mm]
  \end{tabular}
\caption{Success probability ($a=0.85$) for LR-0.01 model (continued).}
\end{figure*}

\begin{figure*}[p]
  \centering
  \begin{tabular}{c@{}c}
    Training Success Probability & Validation Success Probability \\
    \includegraphics[clip,trim=0 0 0 0,width=.52\linewidth]{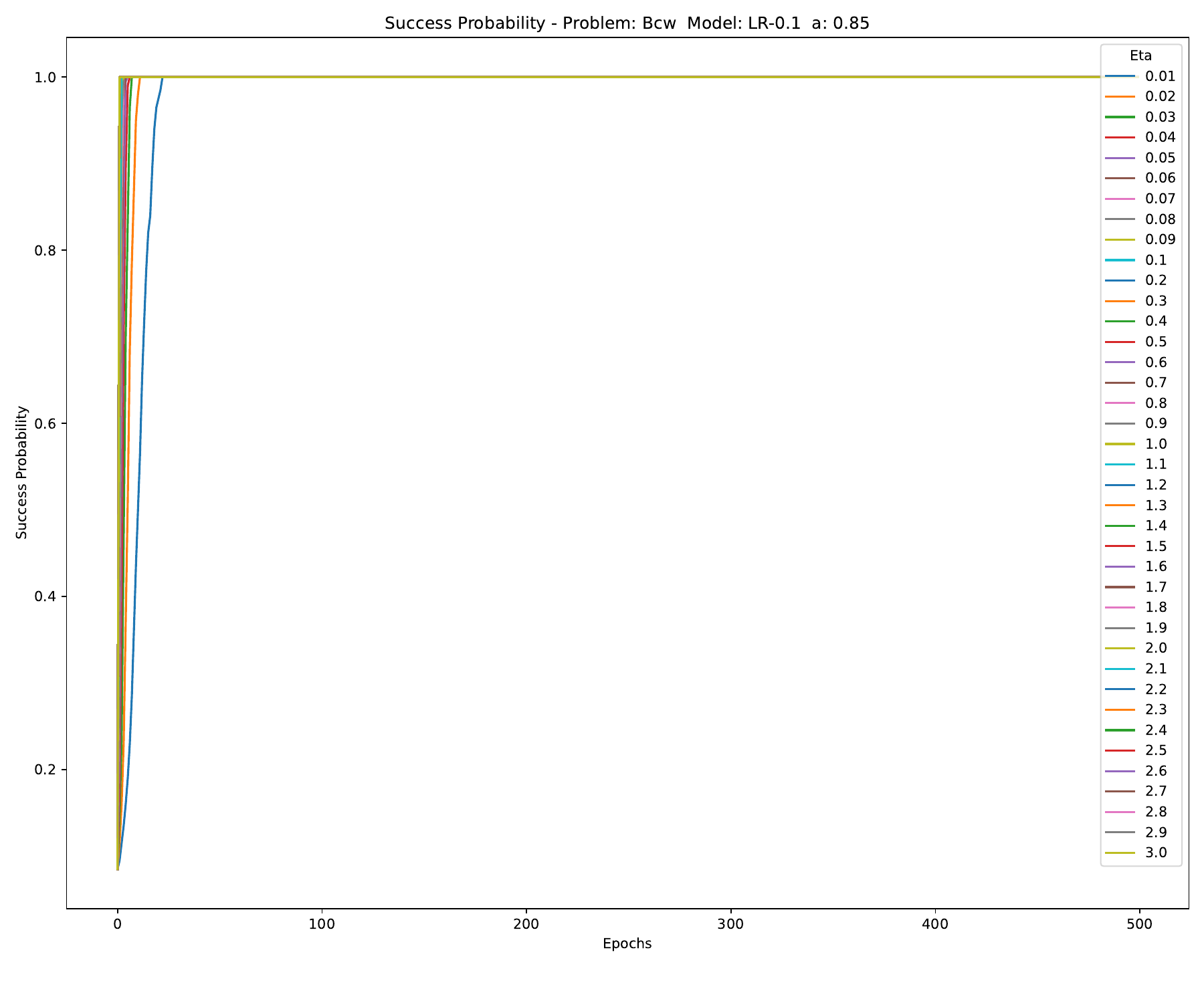} &
    \includegraphics[clip,trim=0 0 0 0,width=.52\linewidth]{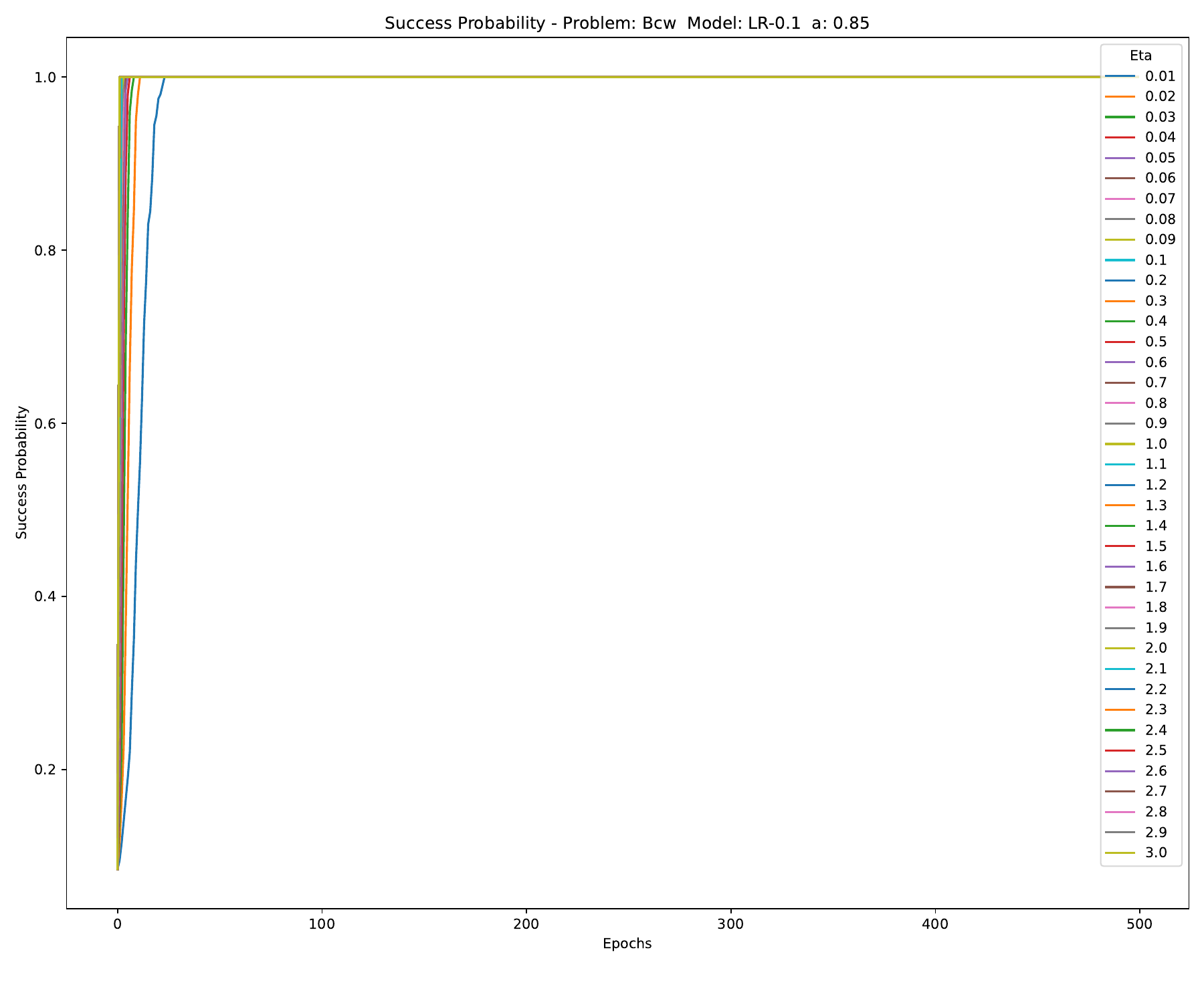} \\[-3mm]
    \includegraphics[clip,trim=0 0 0 0,width=.52\linewidth]{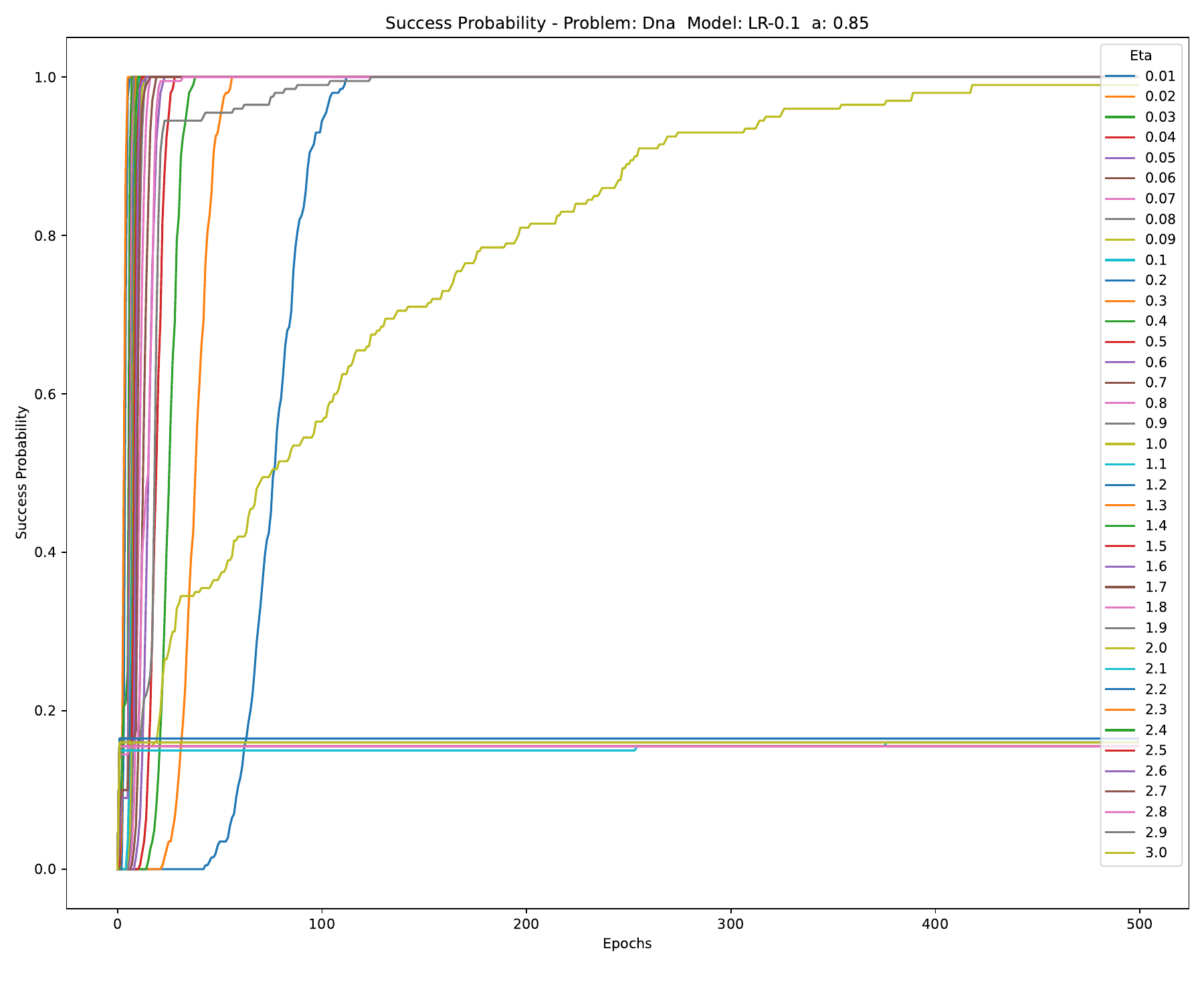} &
    \includegraphics[clip,trim=0 0 0 0,width=.52\linewidth]{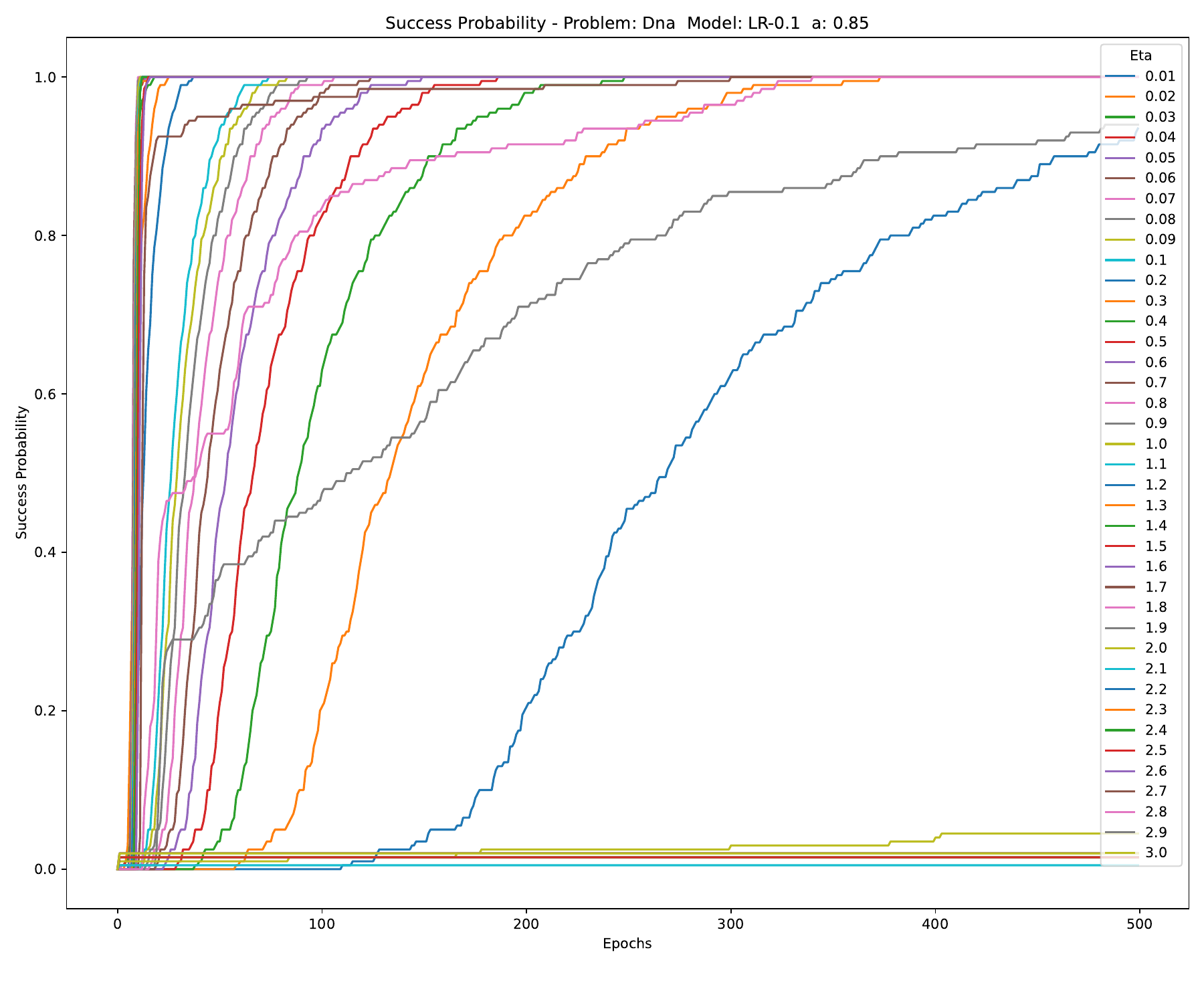} \\[-3mm]
    \includegraphics[clip,trim=0 0 0 0,width=.52\linewidth]{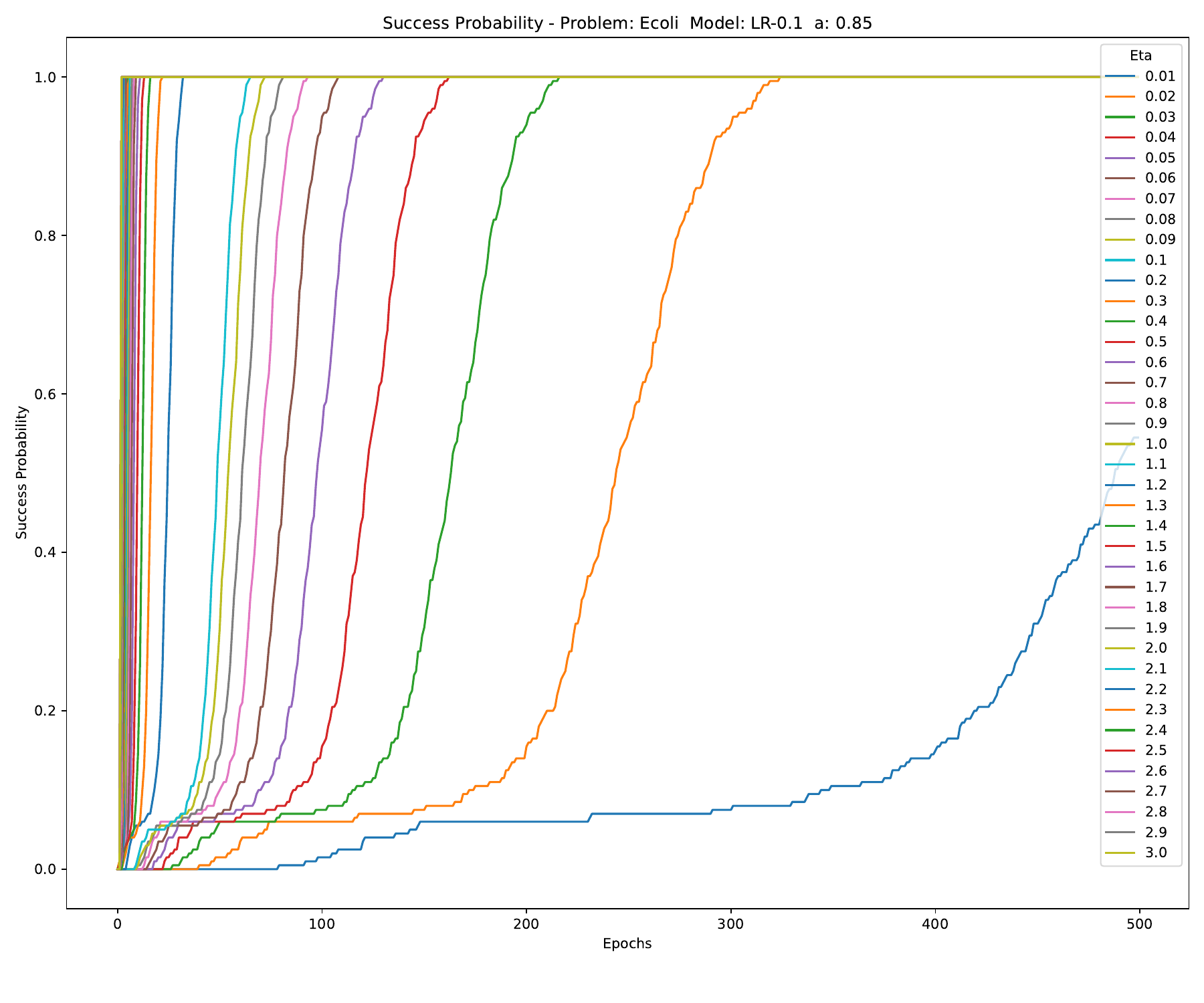} &
    \includegraphics[clip,trim=0 0 0 0,width=.52\linewidth]{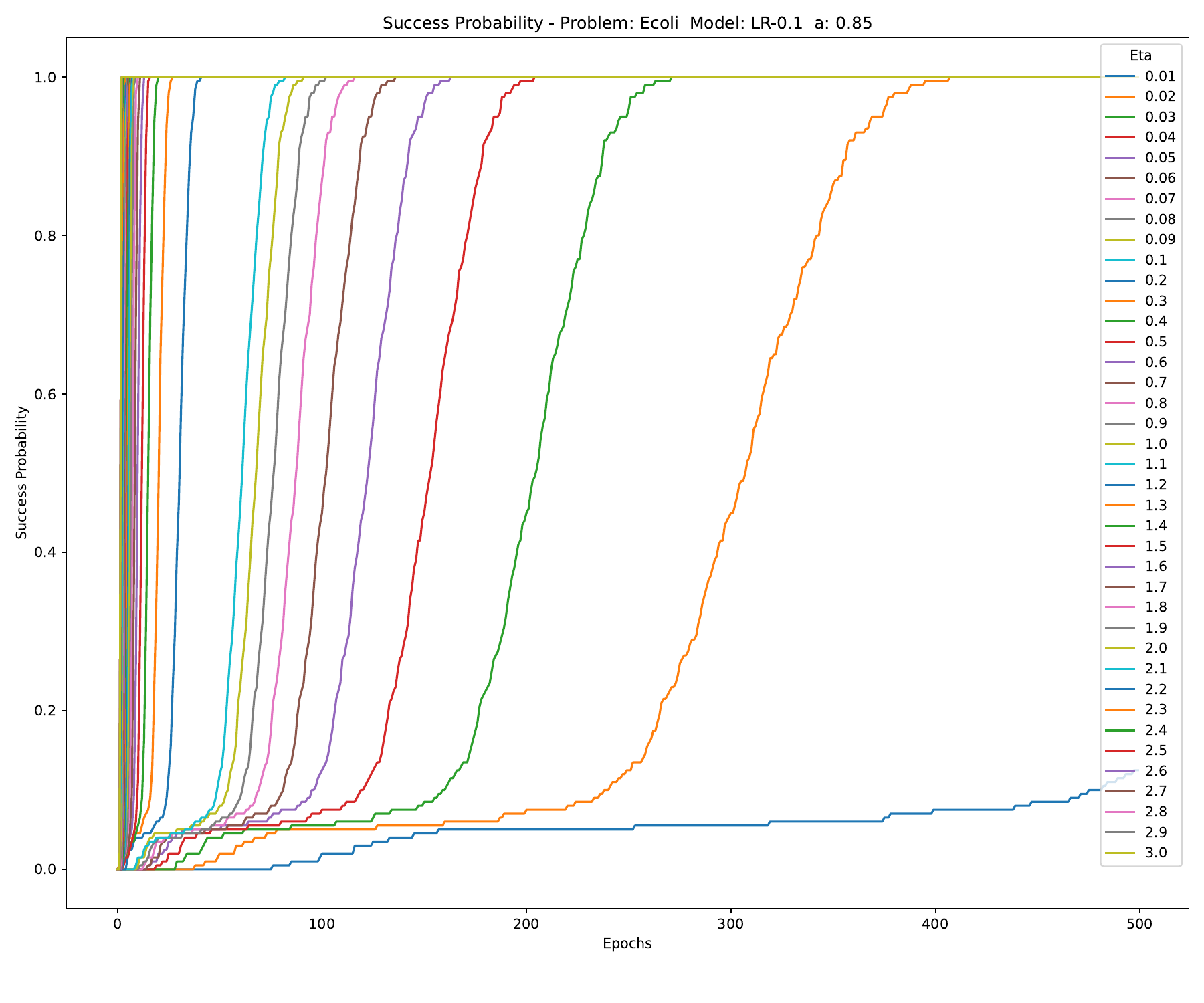} \\[-3mm]
  \end{tabular}
\caption{Success probability ($a=0.85$) for LR-0.1 model.}
  \label{fig:success_probability_LR_0.1_model_0.85}
\end{figure*}

\begin{figure*}[p]
  \centering
  \ContinuedFloat
  \begin{tabular}{c@{}c}
    Training Success Probability & Validation Success Probability \\
    \includegraphics[clip,trim=0 0 0 0,width=.52\linewidth]{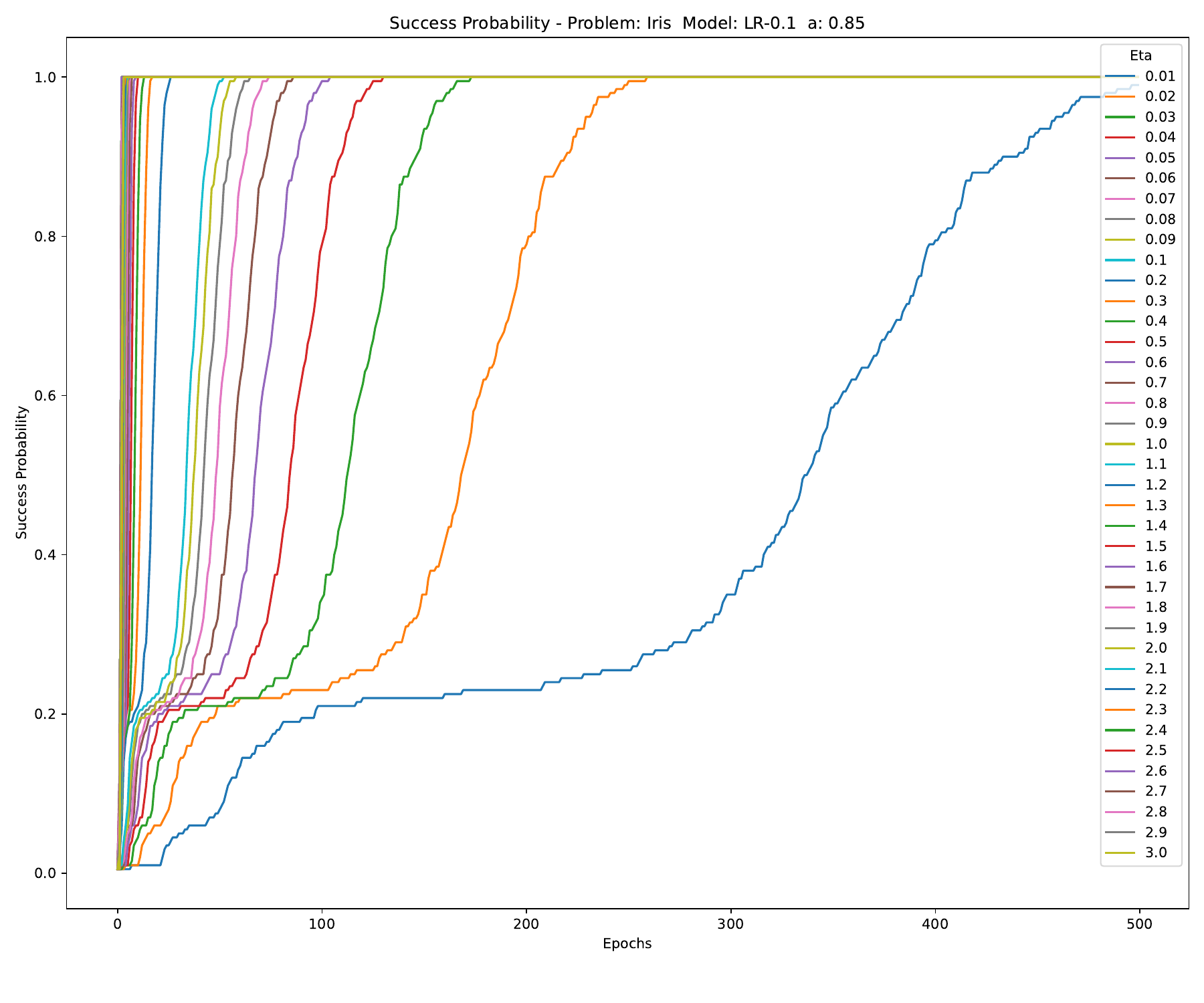} &
    \includegraphics[clip,trim=0 0 0 0,width=.52\linewidth]{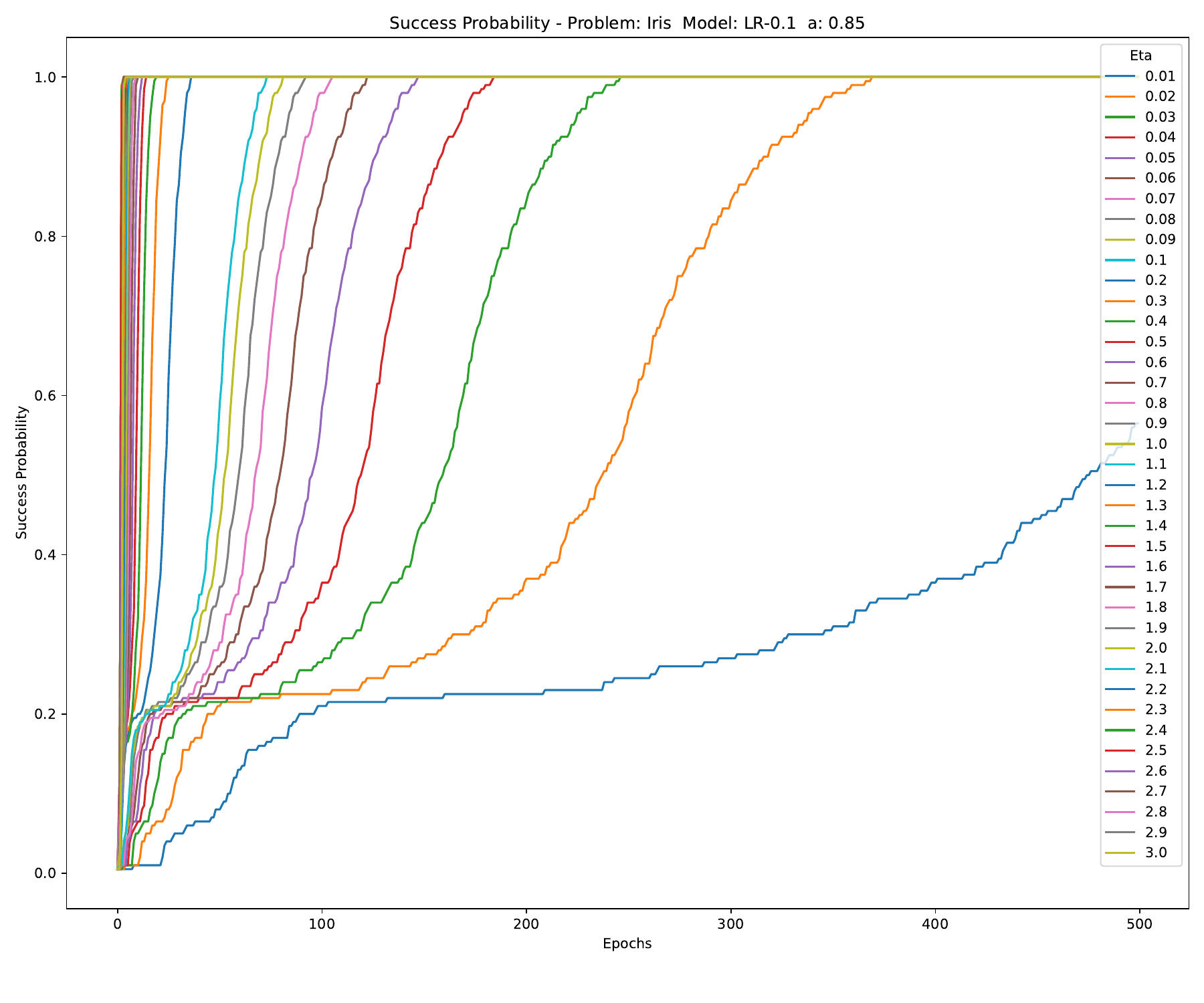} \\[-3mm]
    \includegraphics[clip,trim=0 0 0 0,width=.52\linewidth]{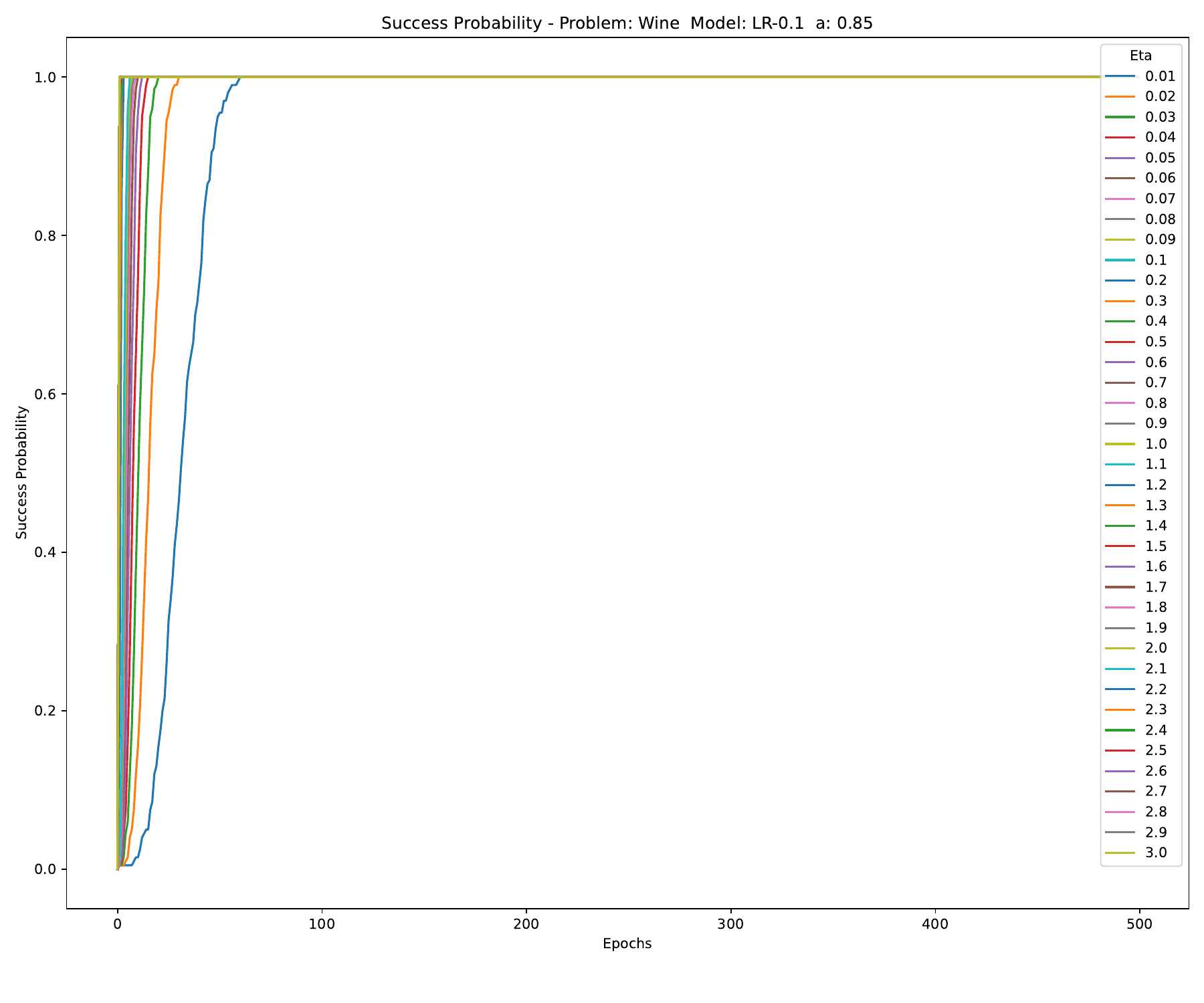} &
    \includegraphics[clip,trim=0 0 0 0,width=.52\linewidth]{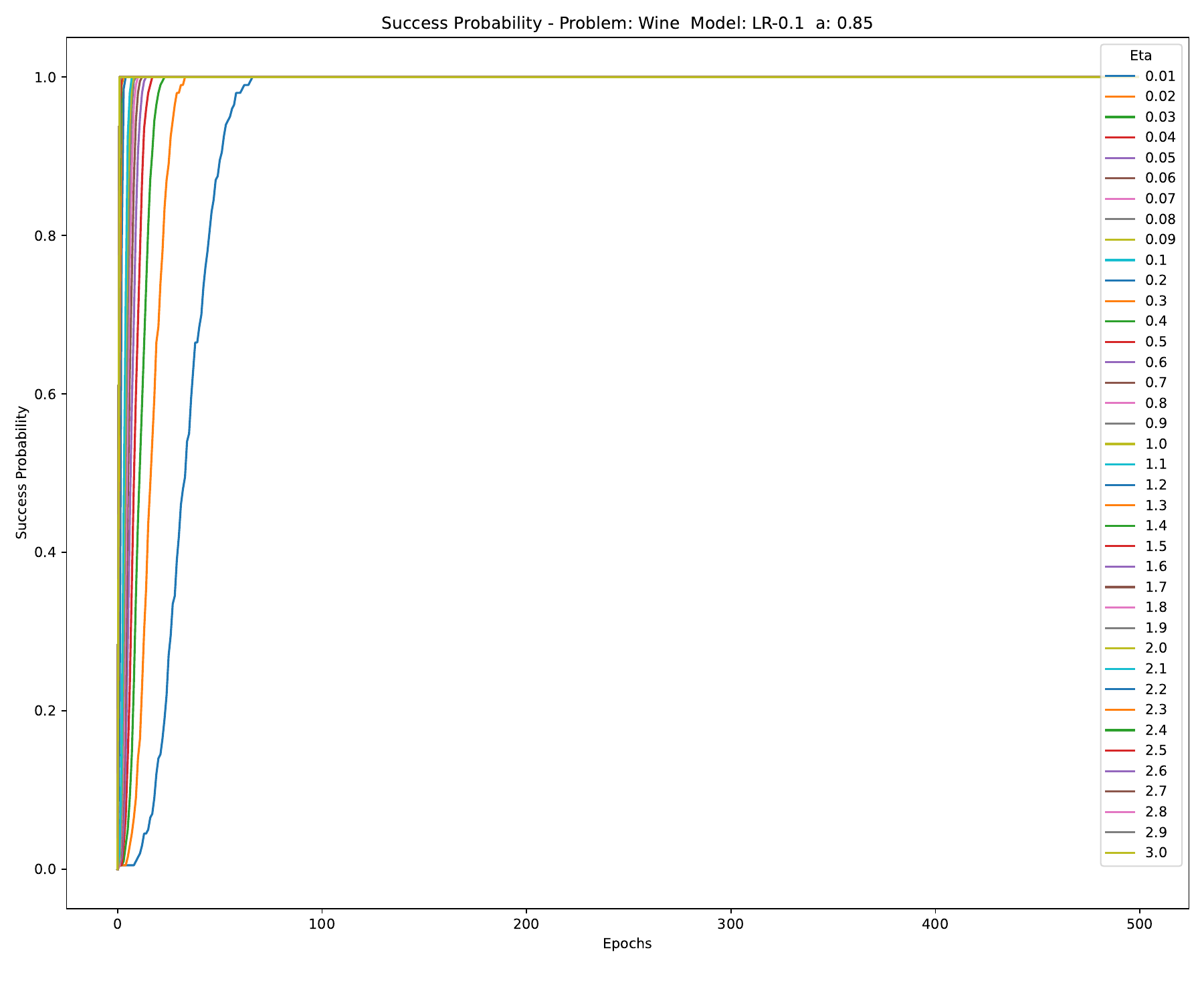} \\[-3mm]
  \end{tabular}
\caption{Success probability ($a=0.85$) for LR-0.1 model (continued).}
\end{figure*}

\begin{figure*}[p]
  \centering
  \begin{tabular}{c@{}c}
    Training Success Probability & Validation Success Probability \\
    \includegraphics[clip,trim=0 0 0 0,width=.52\linewidth]{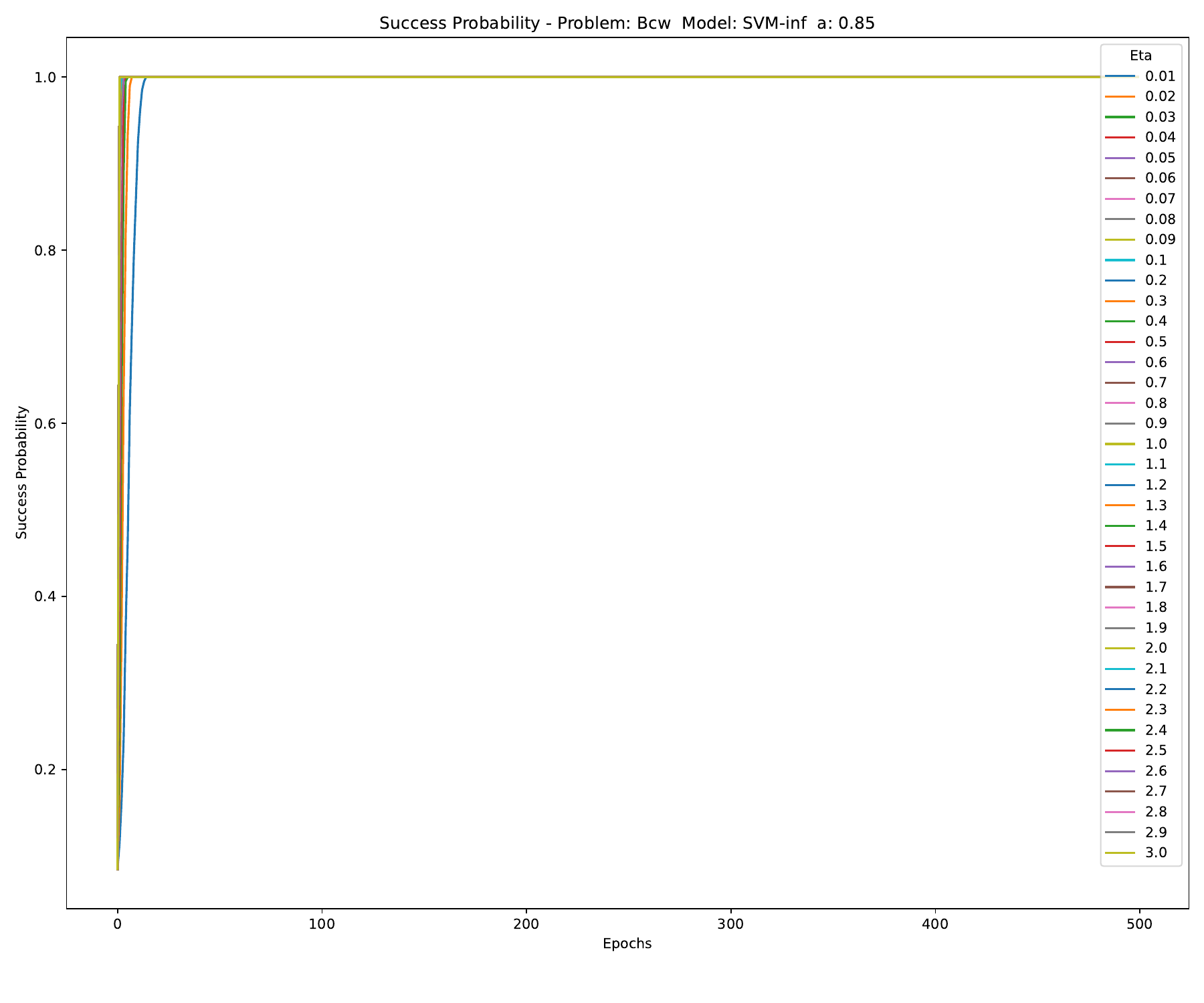} &
    \includegraphics[clip,trim=0 0 0 0,width=.52\linewidth]{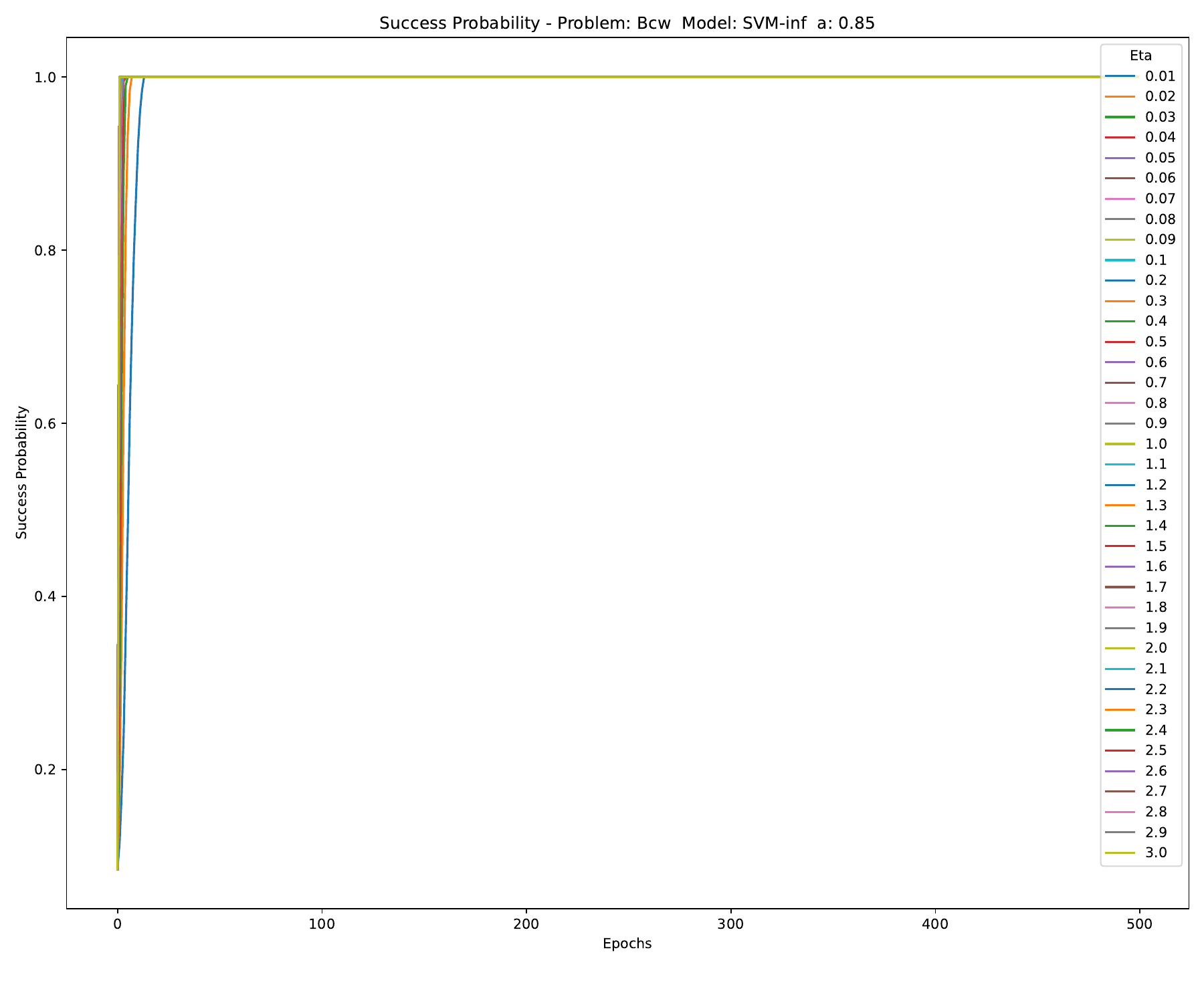} \\[-3mm]
    \includegraphics[clip,trim=0 0 0 0,width=.52\linewidth]{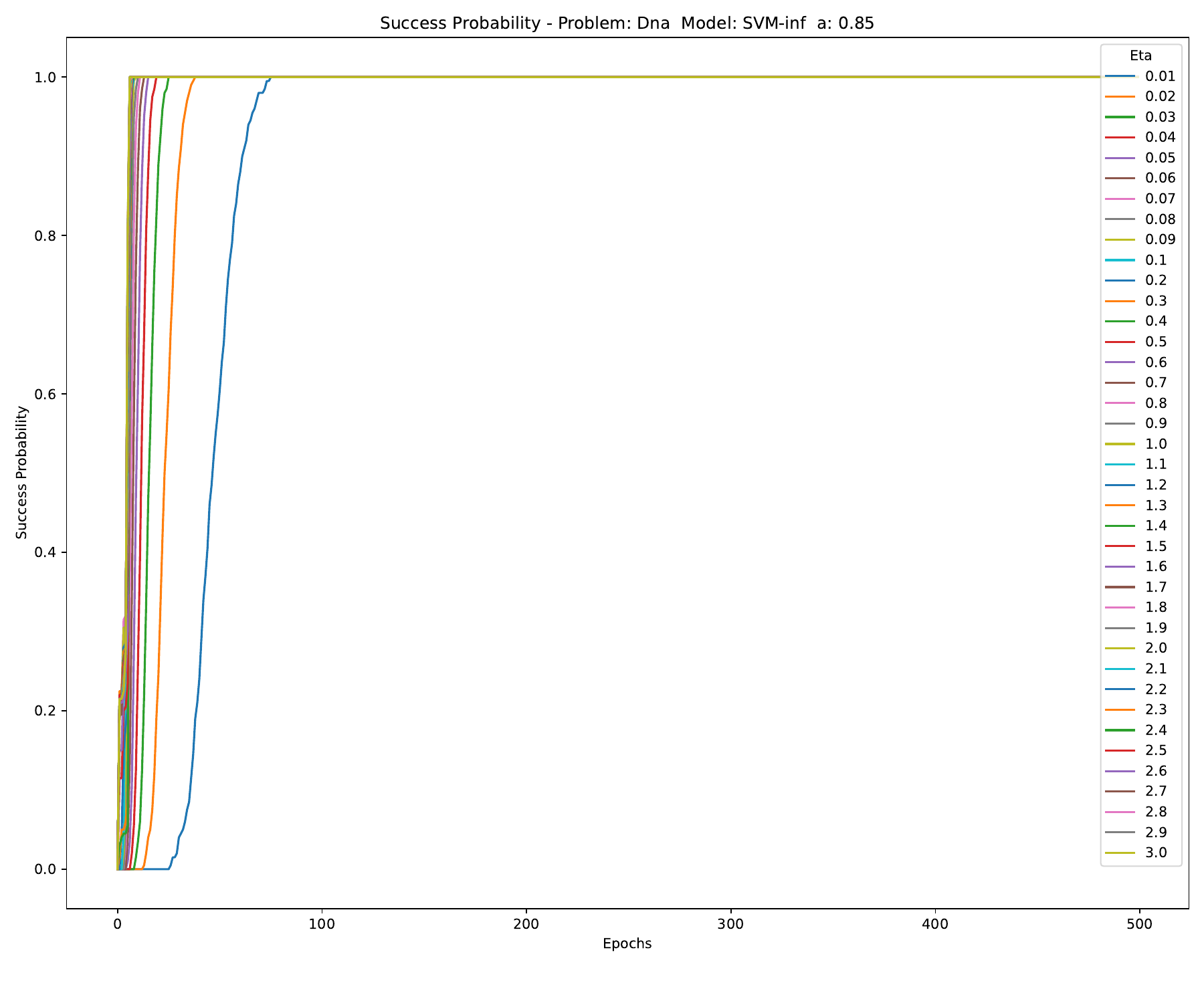} &
    \includegraphics[clip,trim=0 0 0 0,width=.52\linewidth]{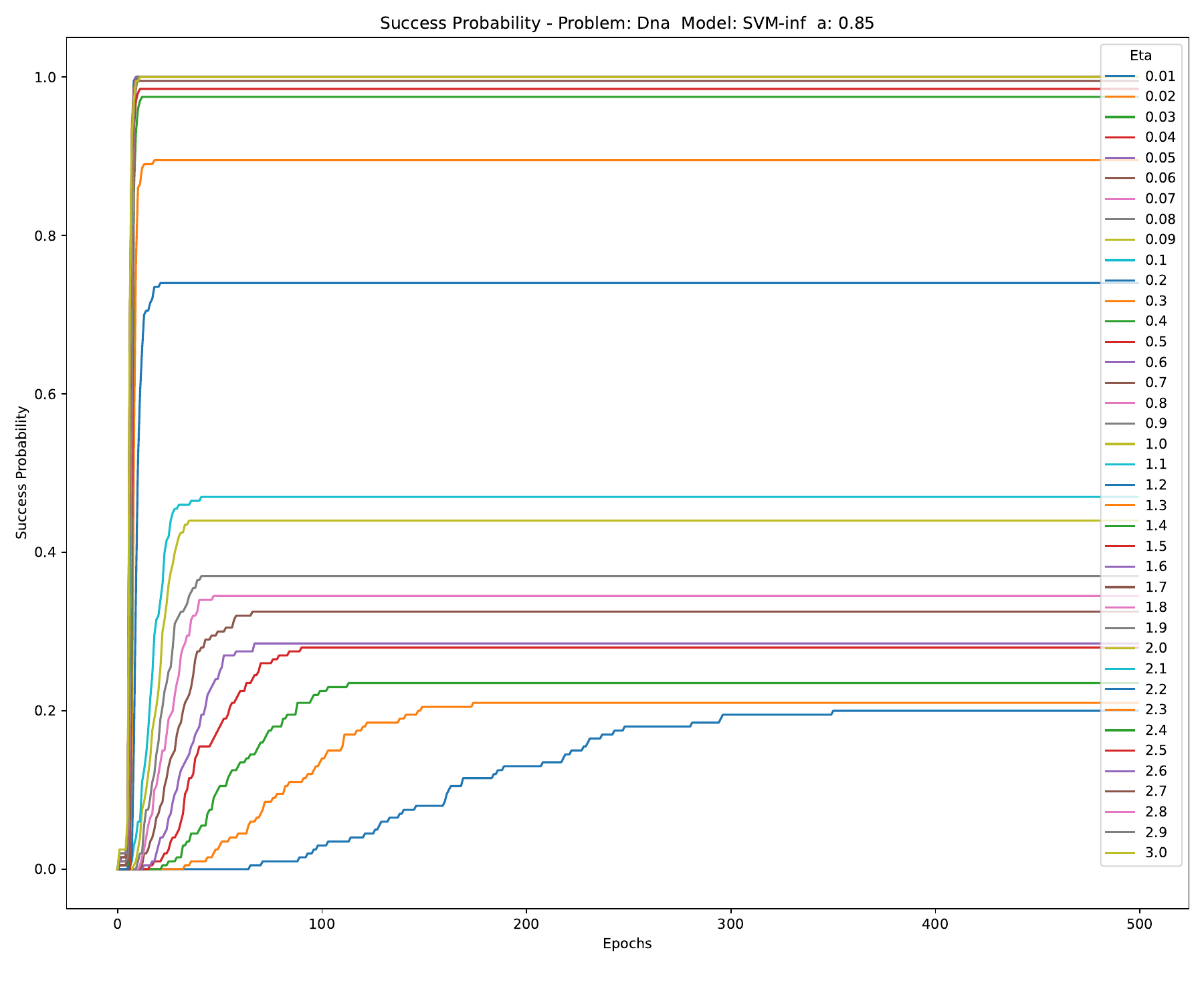} \\[-3mm]
    \includegraphics[clip,trim=0 0 0 0,width=.52\linewidth]{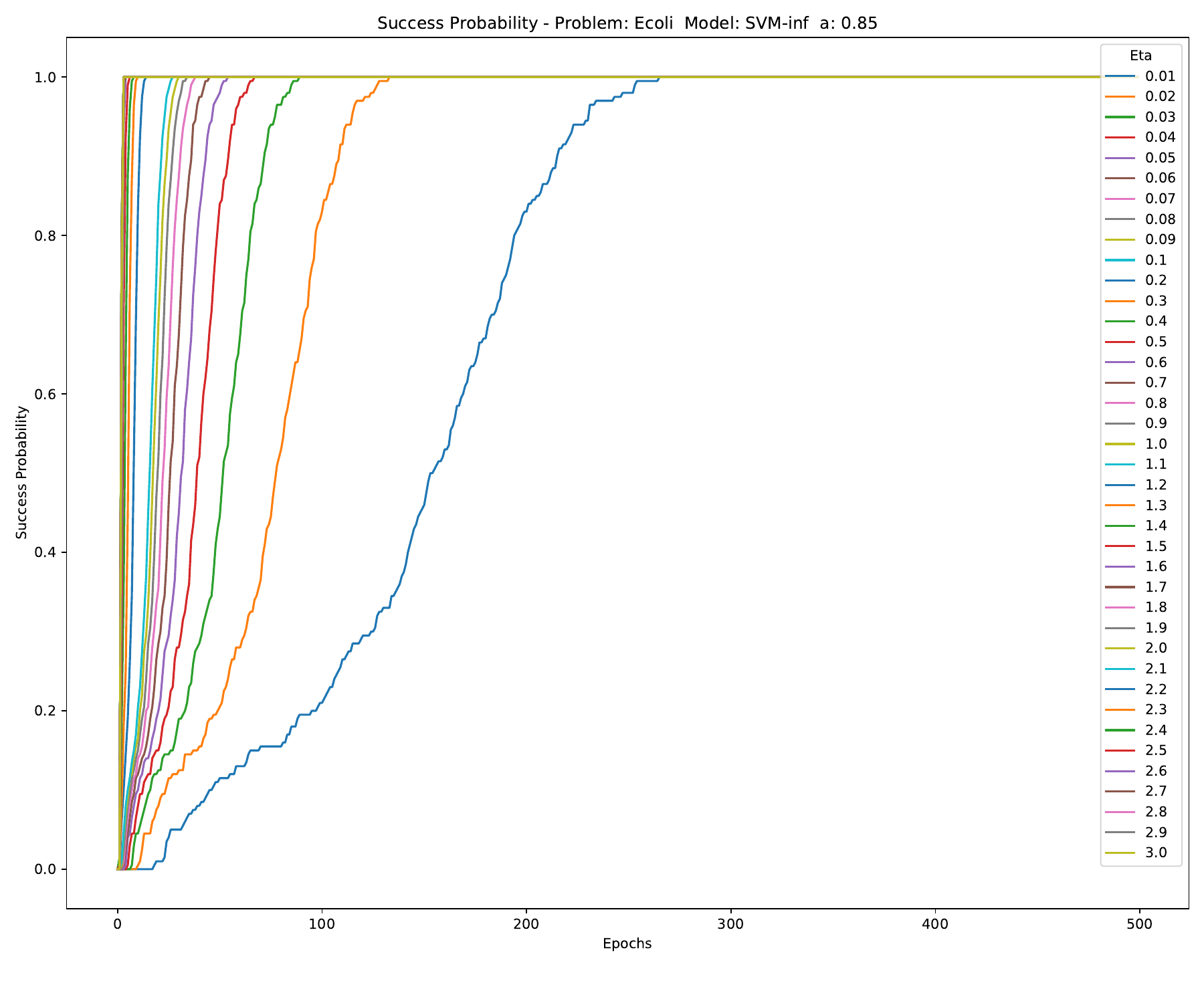} &
    \includegraphics[clip,trim=0 0 0 0,width=.52\linewidth]{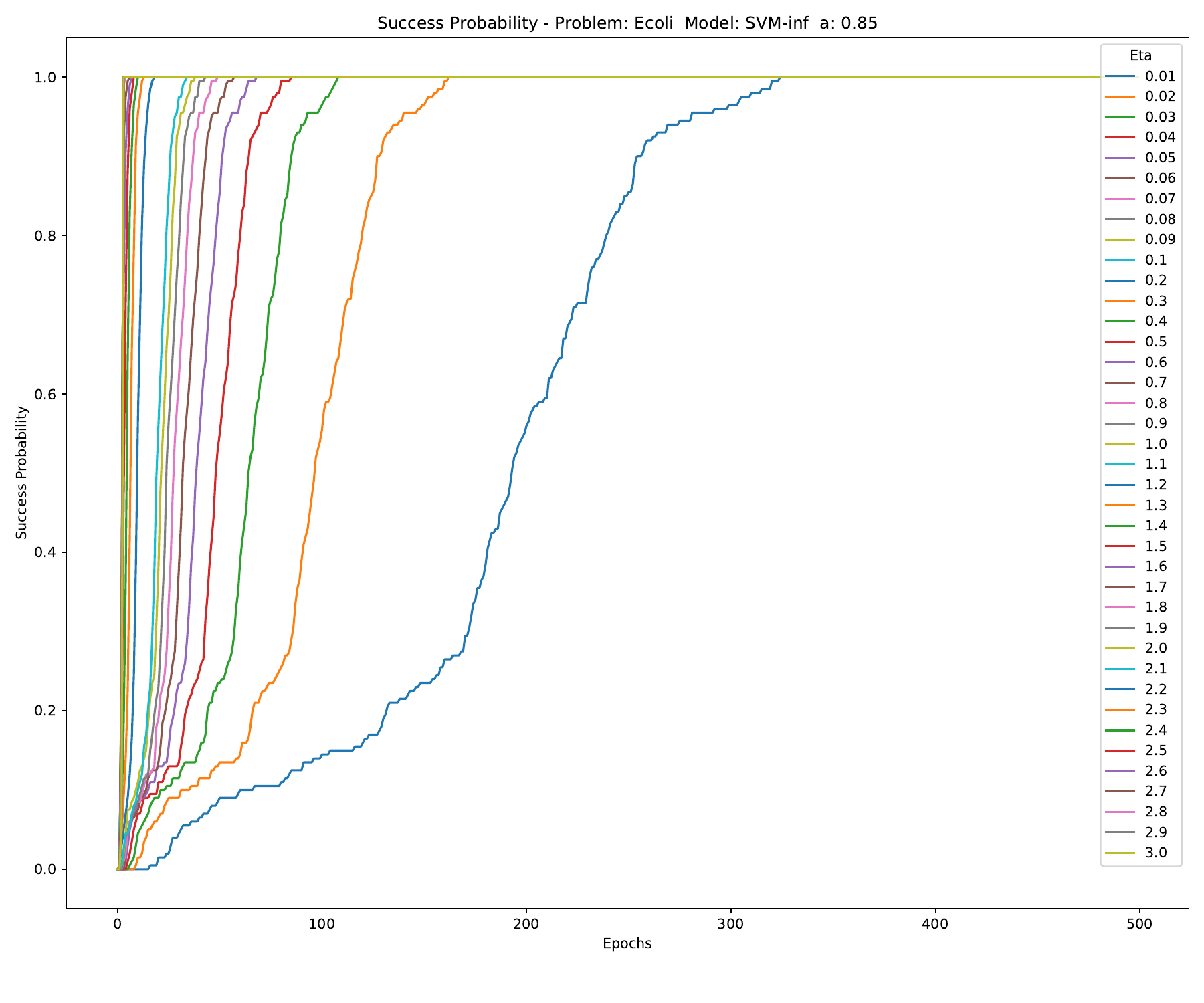} \\[-3mm]
  \end{tabular}
\caption{Success probability ($a=0.85$) for SVM-Inf model.}
  \label{fig:success_probability_SVM_Inf_model_0.85}
\end{figure*}

\begin{figure*}[p]
  \centering
  \ContinuedFloat
  \begin{tabular}{c@{}c}
    Training Success Probability & Validation Success Probability \\
    \includegraphics[clip,trim=0 0 0 0,width=.52\linewidth]{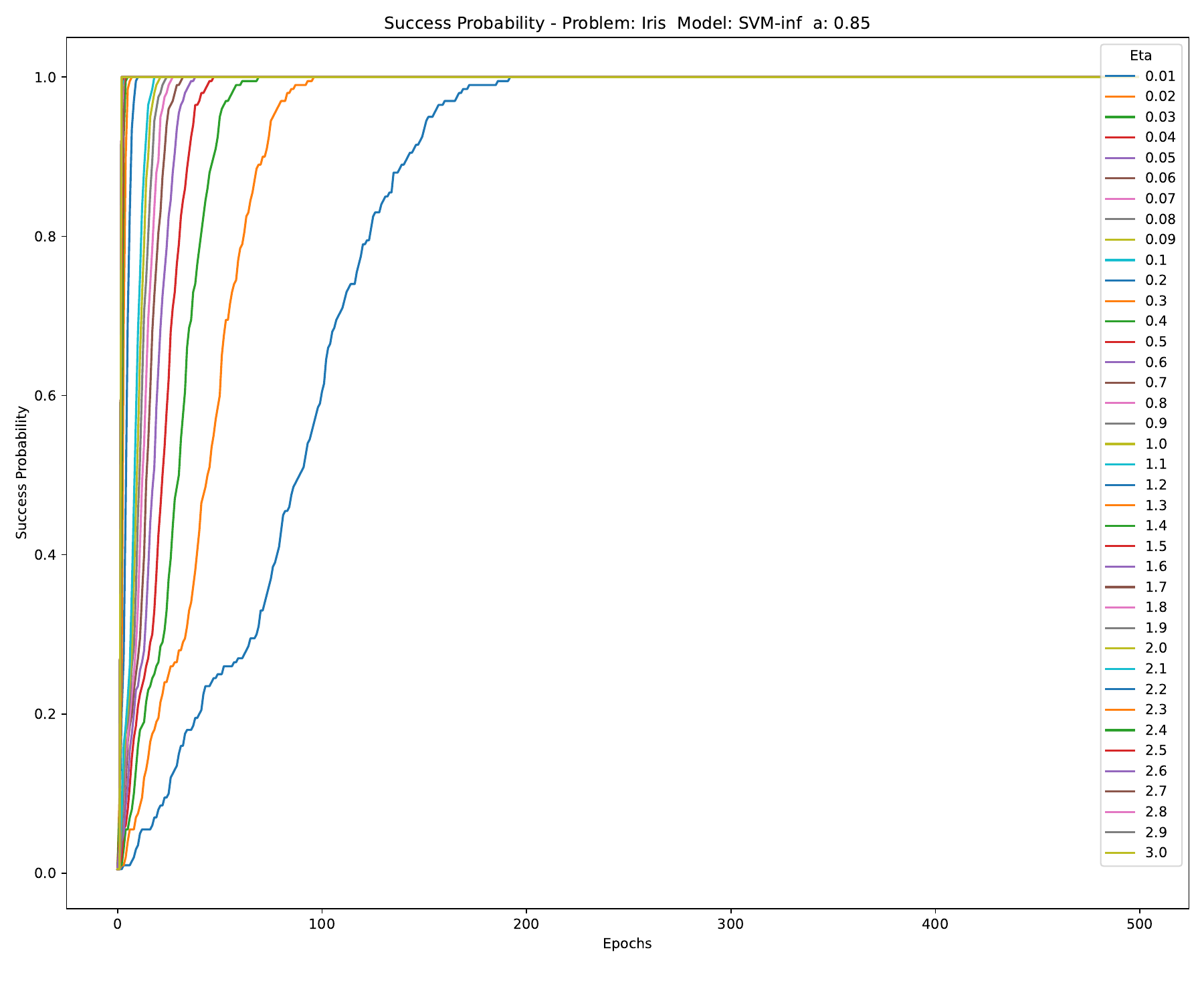} &
    \includegraphics[clip,trim=0 0 0 0,width=.52\linewidth]{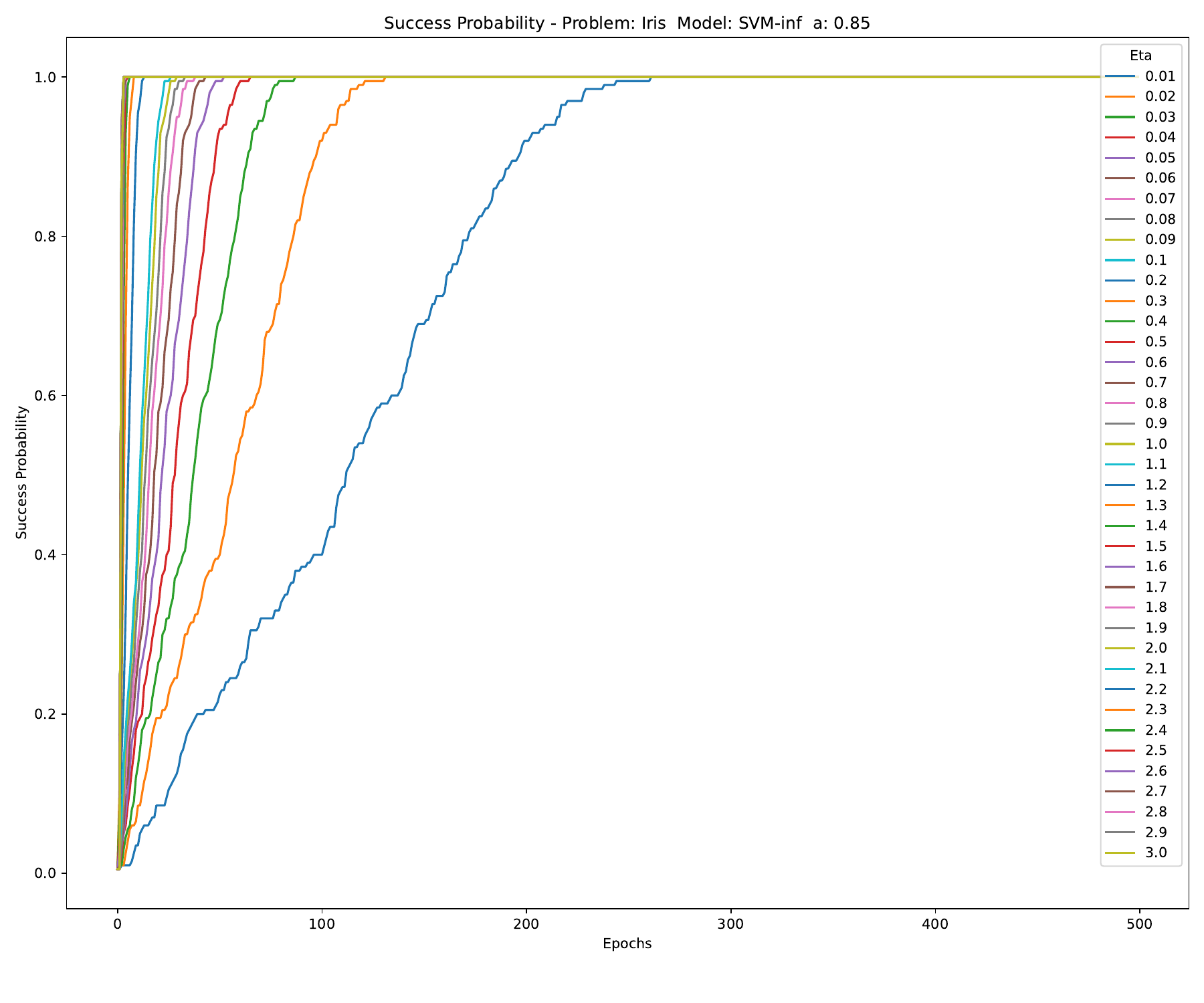} \\[-3mm]
    \includegraphics[clip,trim=0 0 0 0,width=.52\linewidth]{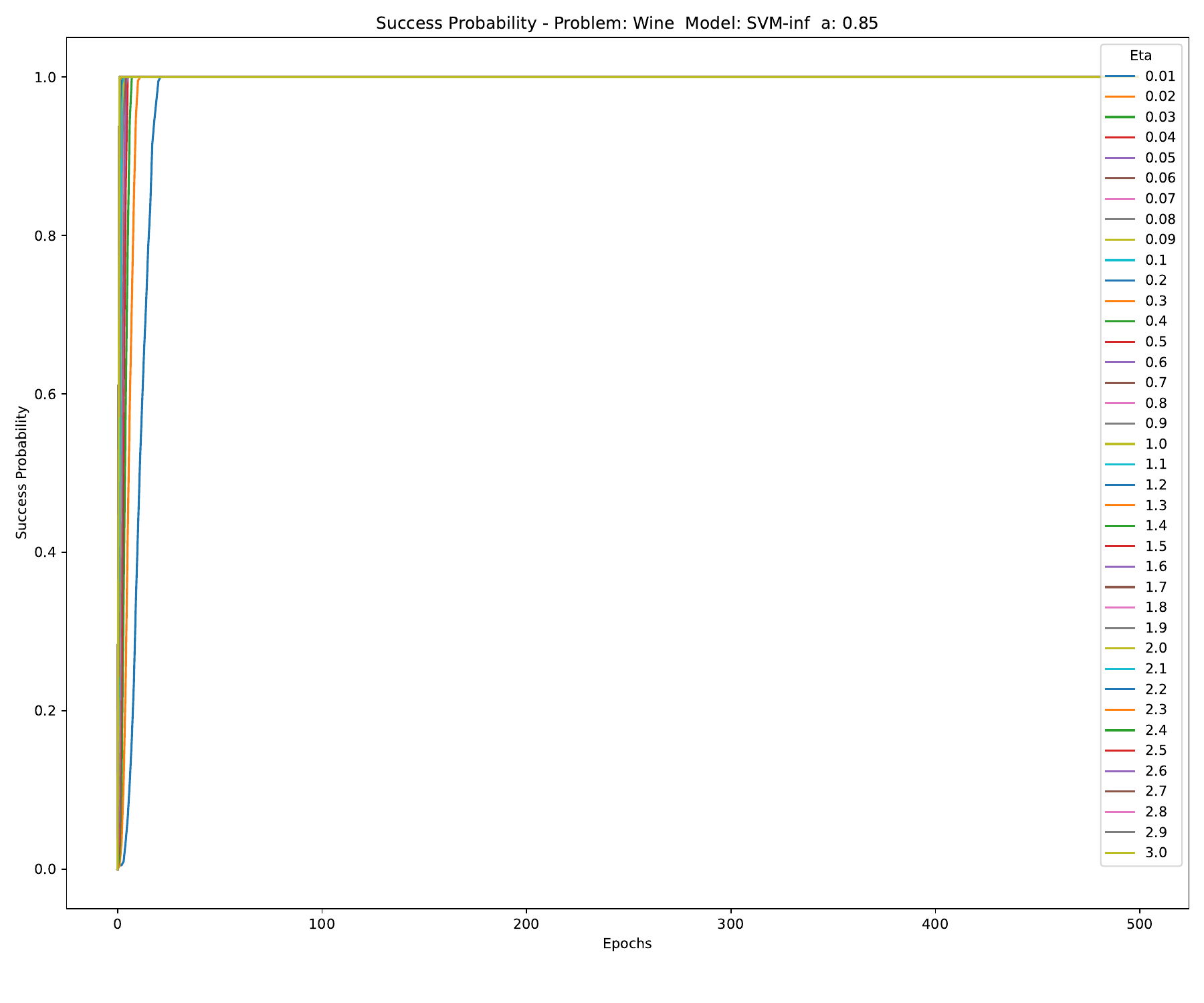} &
    \includegraphics[clip,trim=0 0 0 0,width=.52\linewidth]{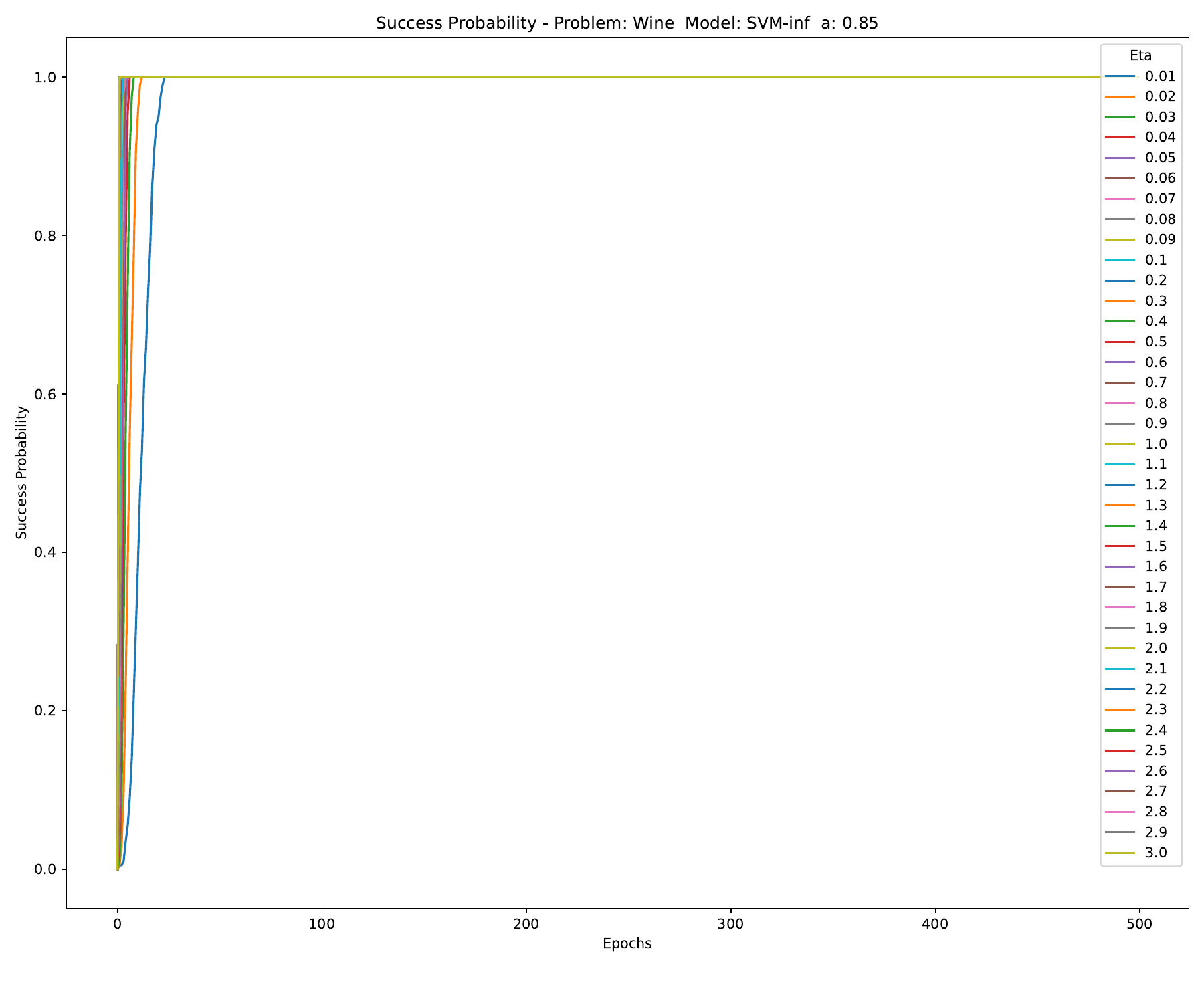} \\[-3mm]
  \end{tabular}
\caption{Success probability ($a=0.85$) for SVM-Inf model (continued).}
\end{figure*}

\begin{figure*}[p]
  \centering
  \begin{tabular}{c@{}c}
    Training Success Probability & Validation Success Probability \\
    \includegraphics[clip,trim=0 0 0 0,width=.52\linewidth]{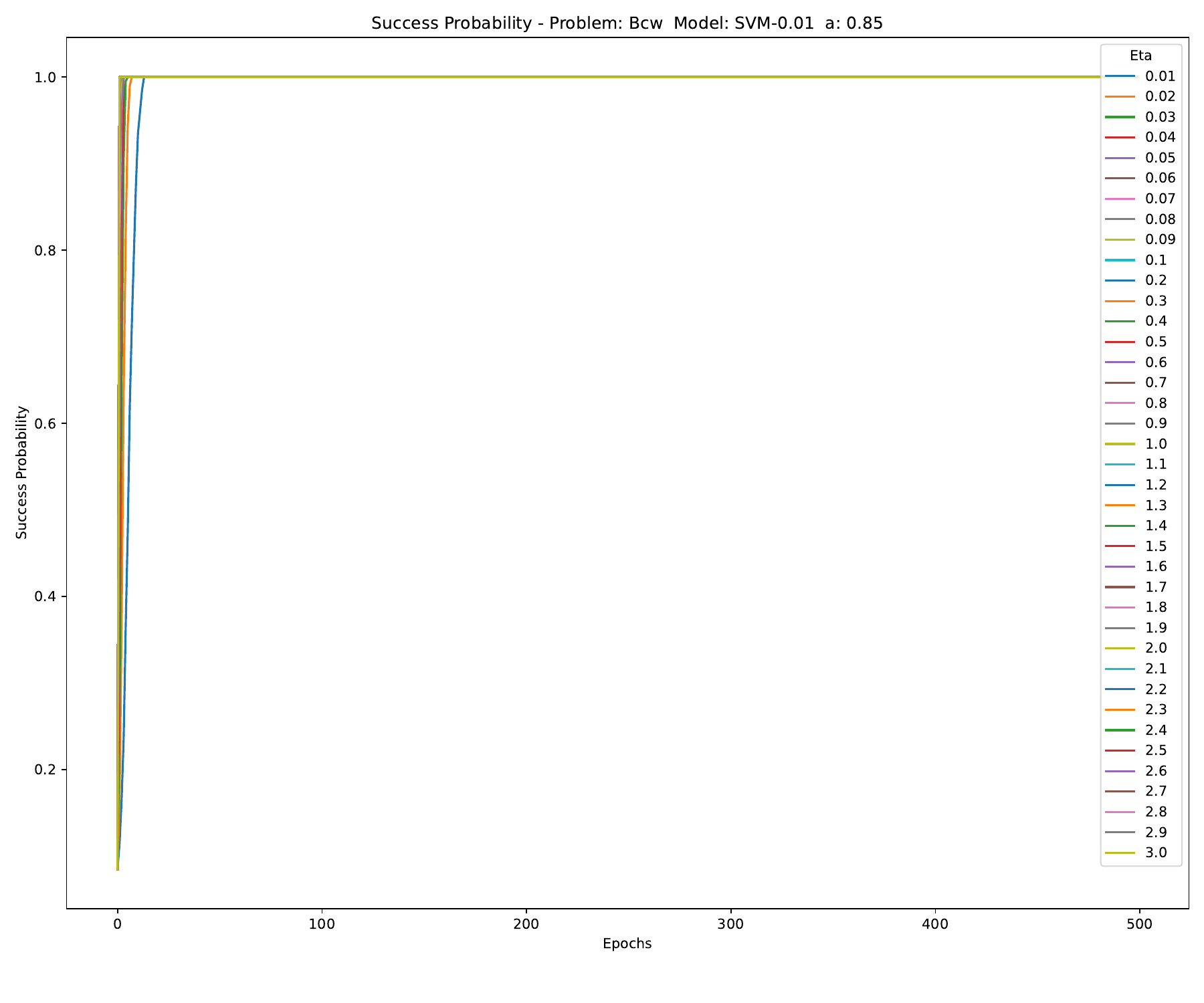} &
    \includegraphics[clip,trim=0 0 0 0,width=.52\linewidth]{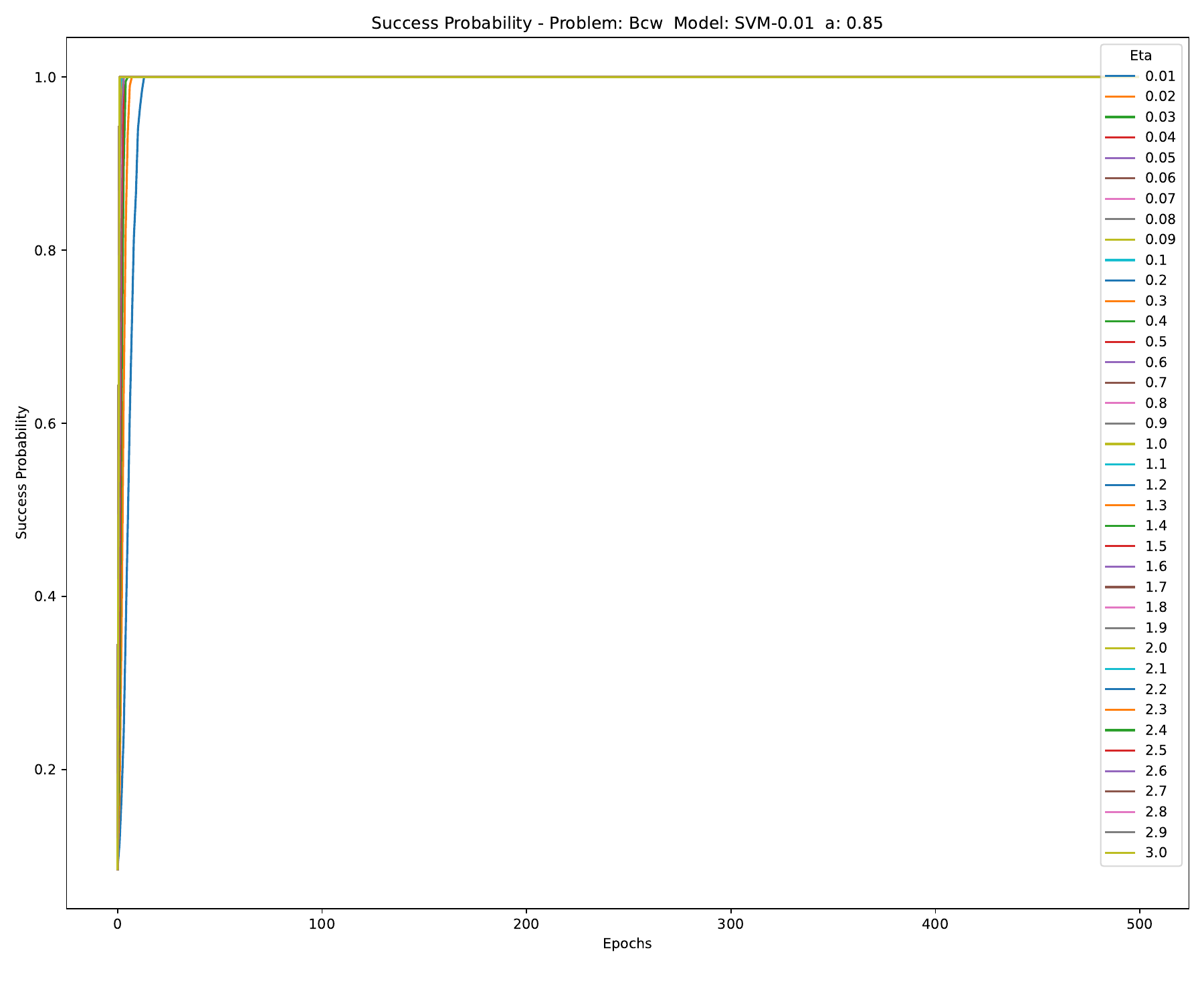} \\[-3mm]
    \includegraphics[clip,trim=0 0 0 0,width=.52\linewidth]{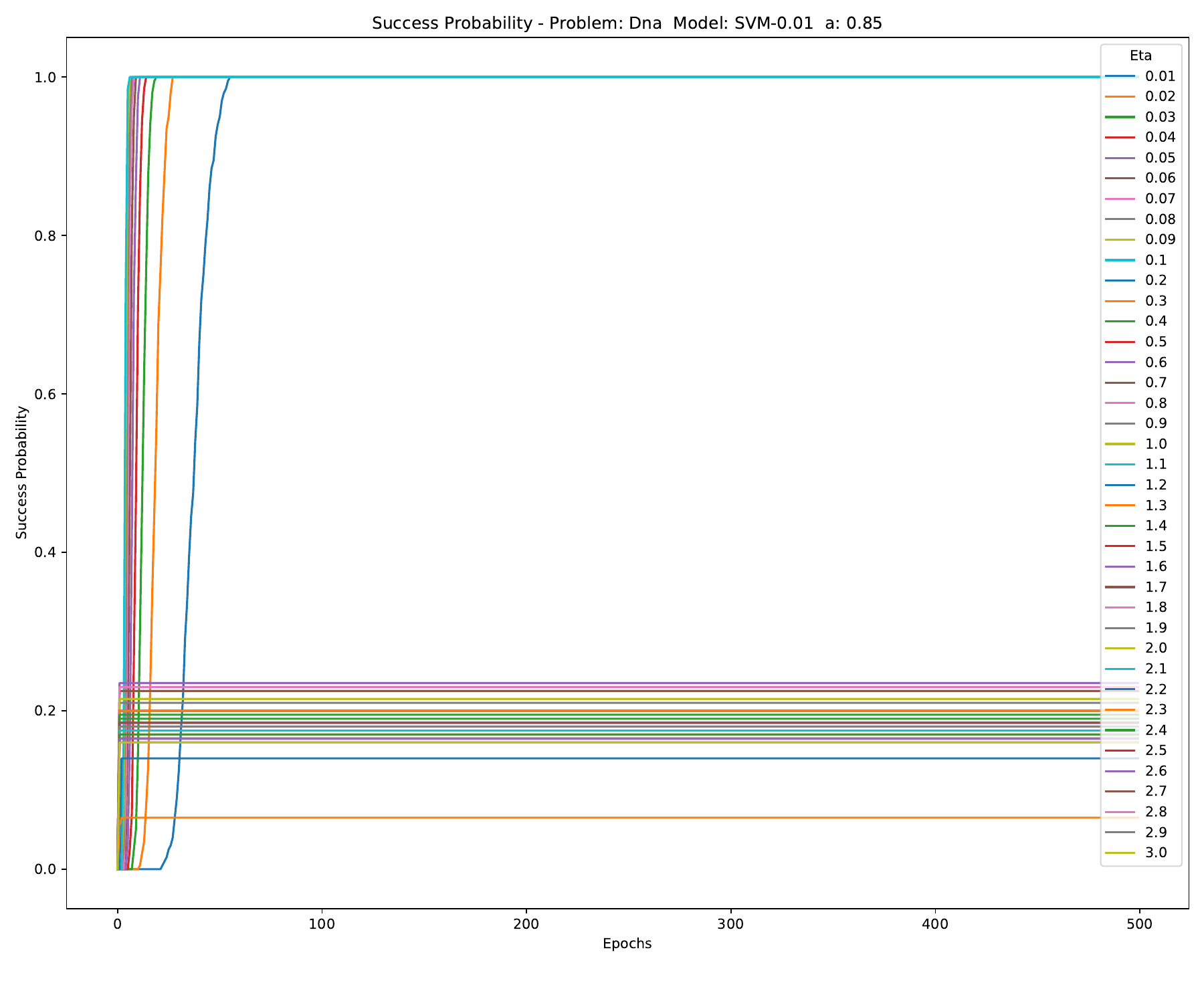} &
    \includegraphics[clip,trim=0 0 0 0,width=.52\linewidth]{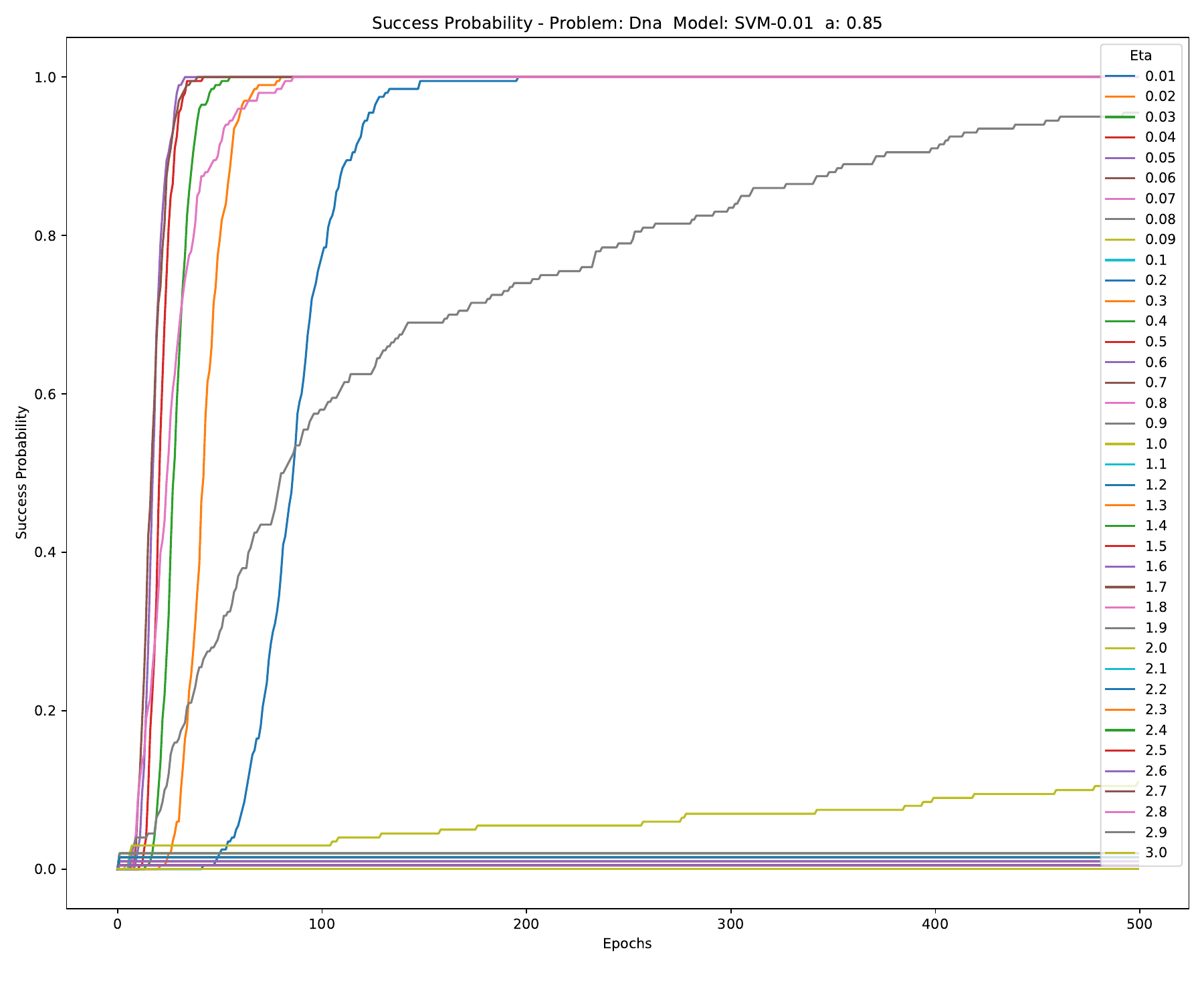} \\[-3mm]
    \includegraphics[clip,trim=0 0 0 0,width=.52\linewidth]{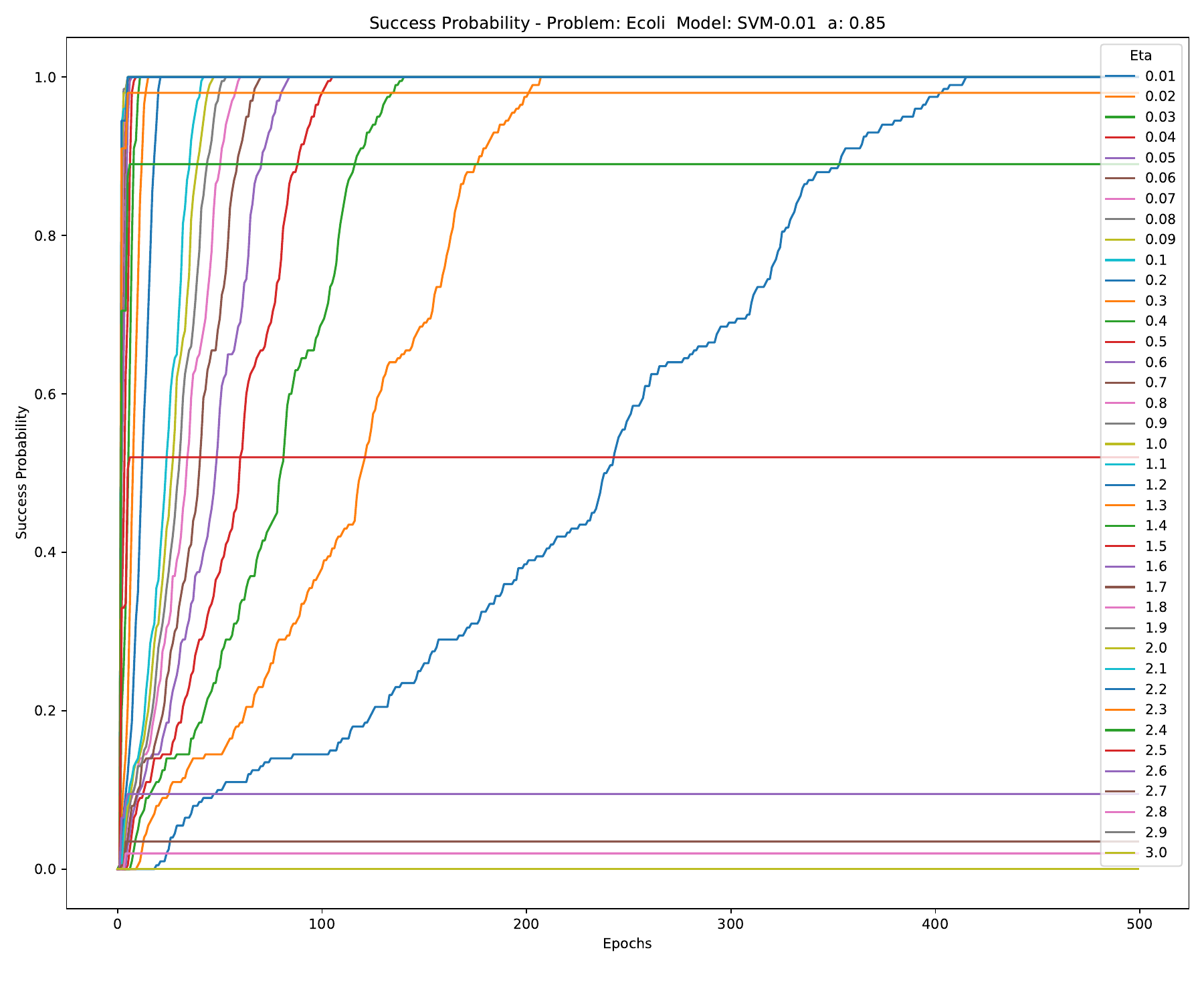} &
    \includegraphics[clip,trim=0 0 0 0,width=.52\linewidth]{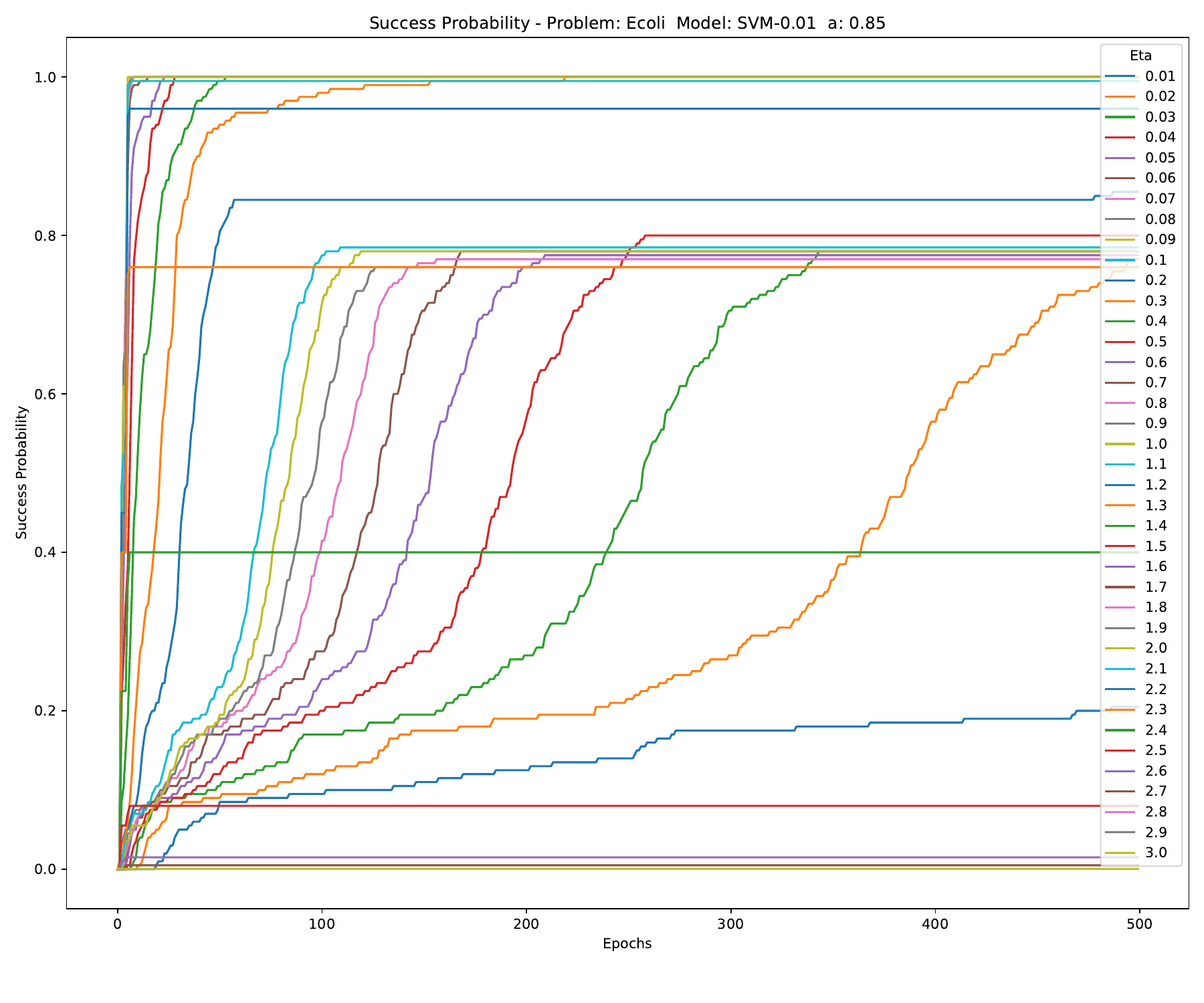} \\[-3mm]
  \end{tabular}
\caption{Success probability ($a=0.85$) for SVM-0.01 model.}
  \label{fig:success_probability_SVM_0.01_model_0.85}
\end{figure*}

\begin{figure*}[p]
  \centering
  \ContinuedFloat
  \begin{tabular}{c@{}c}
    Training Success Probability & Validation Success Probability \\
    \includegraphics[clip,trim=0 0 0 0,width=.52\linewidth]{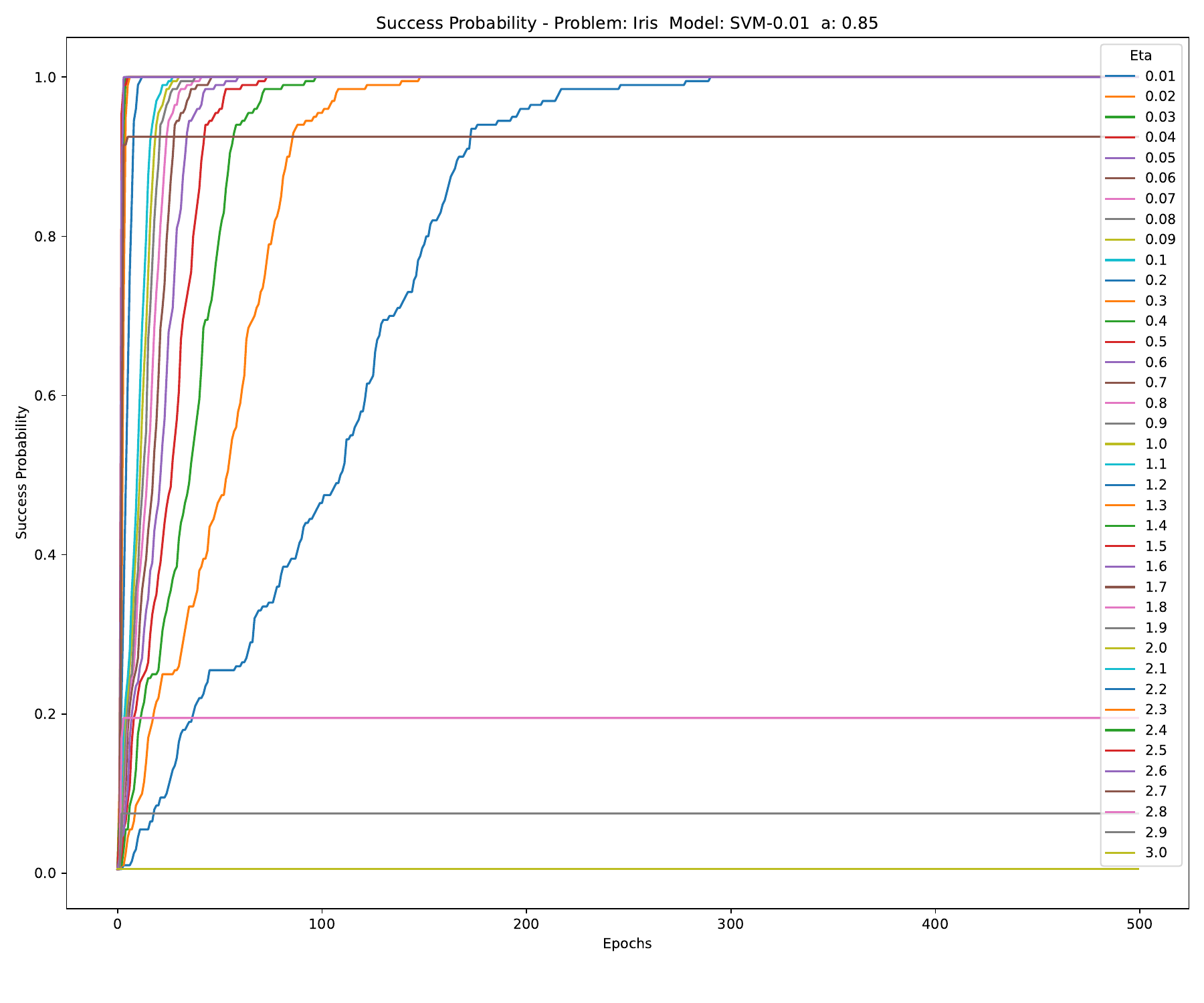} &
    \includegraphics[clip,trim=0 0 0 0,width=.52\linewidth]{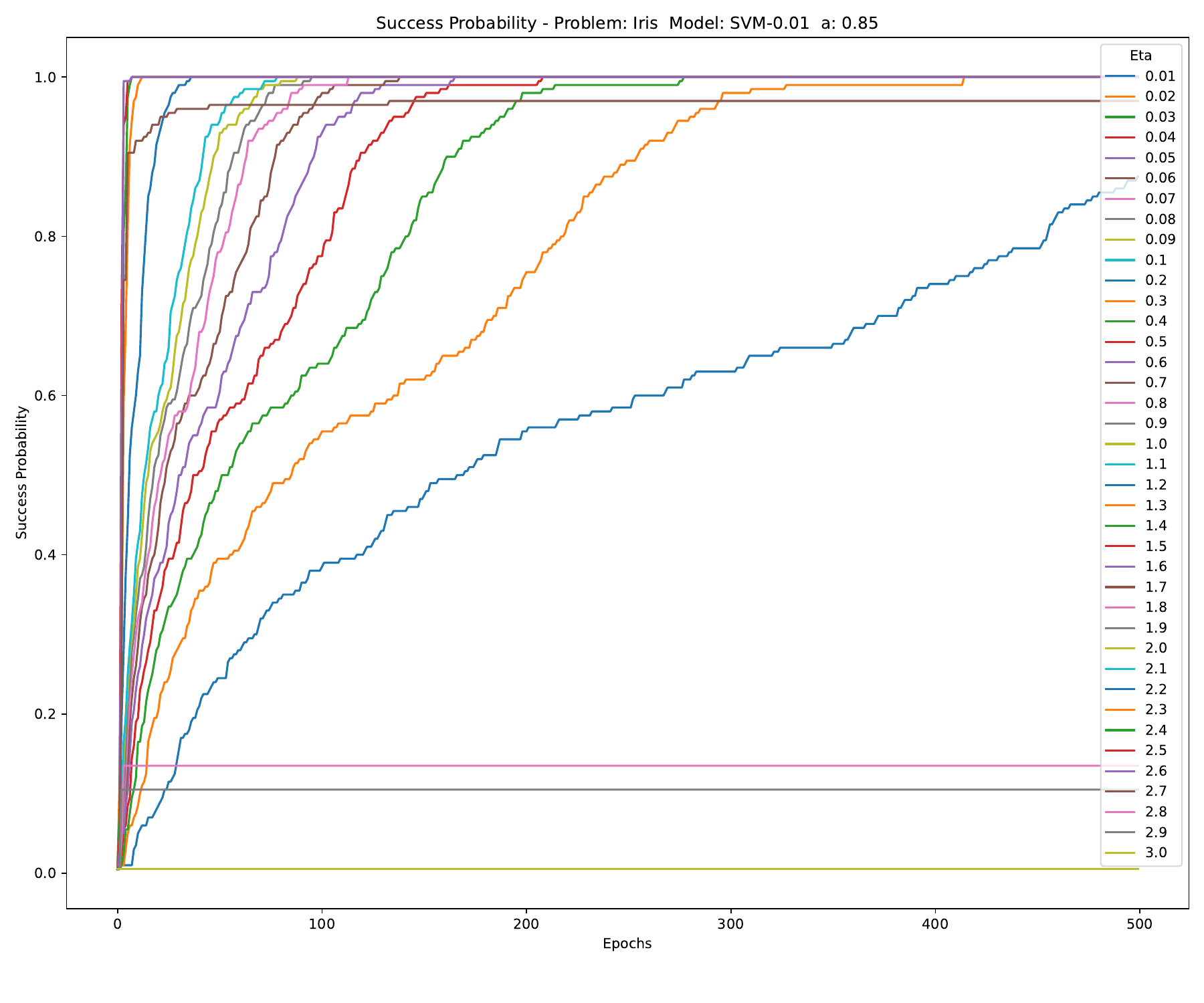} \\[-3mm]
    \includegraphics[clip,trim=0 0 0 0,width=.52\linewidth]{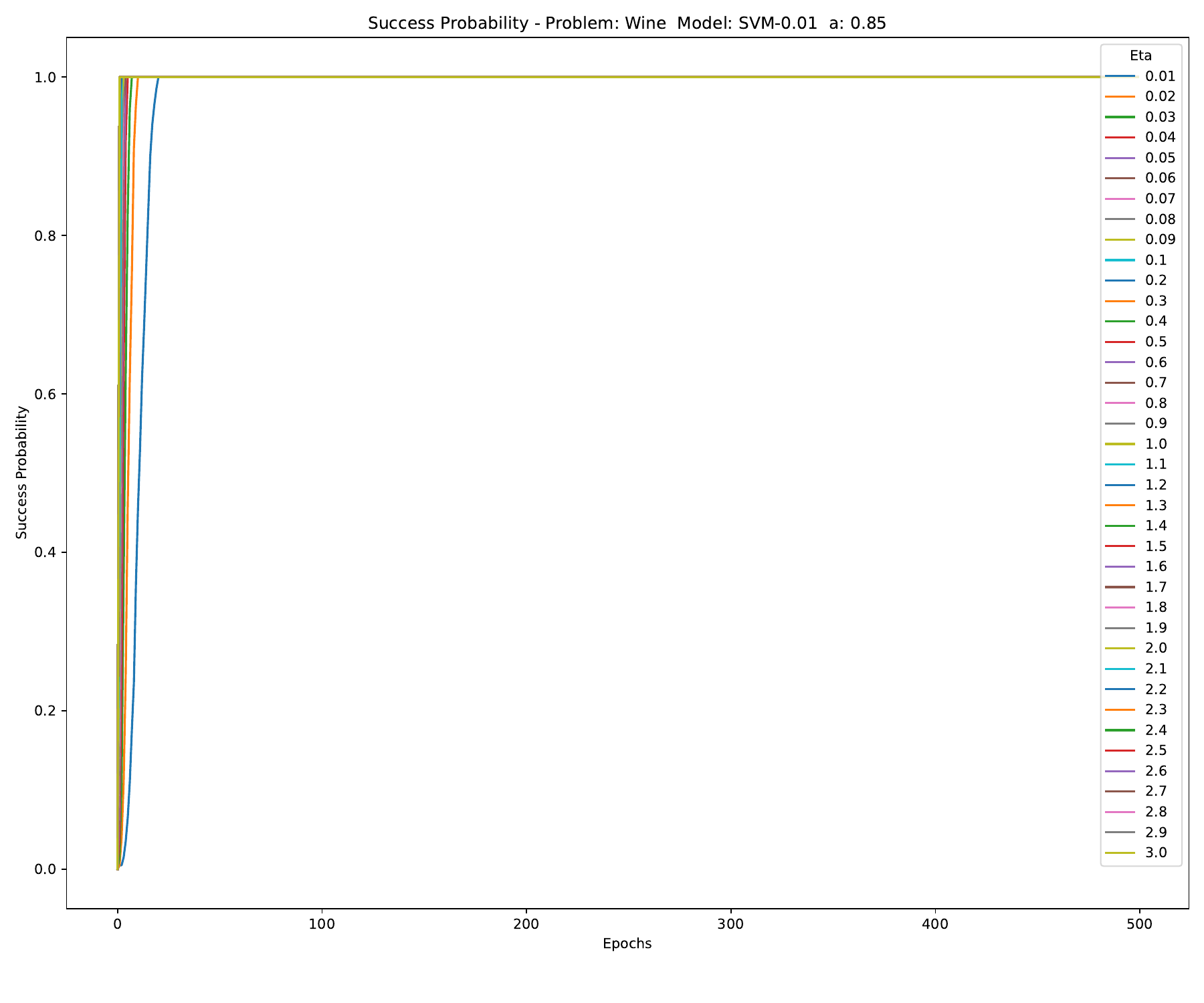} &
    \includegraphics[clip,trim=0 0 0 0,width=.52\linewidth]{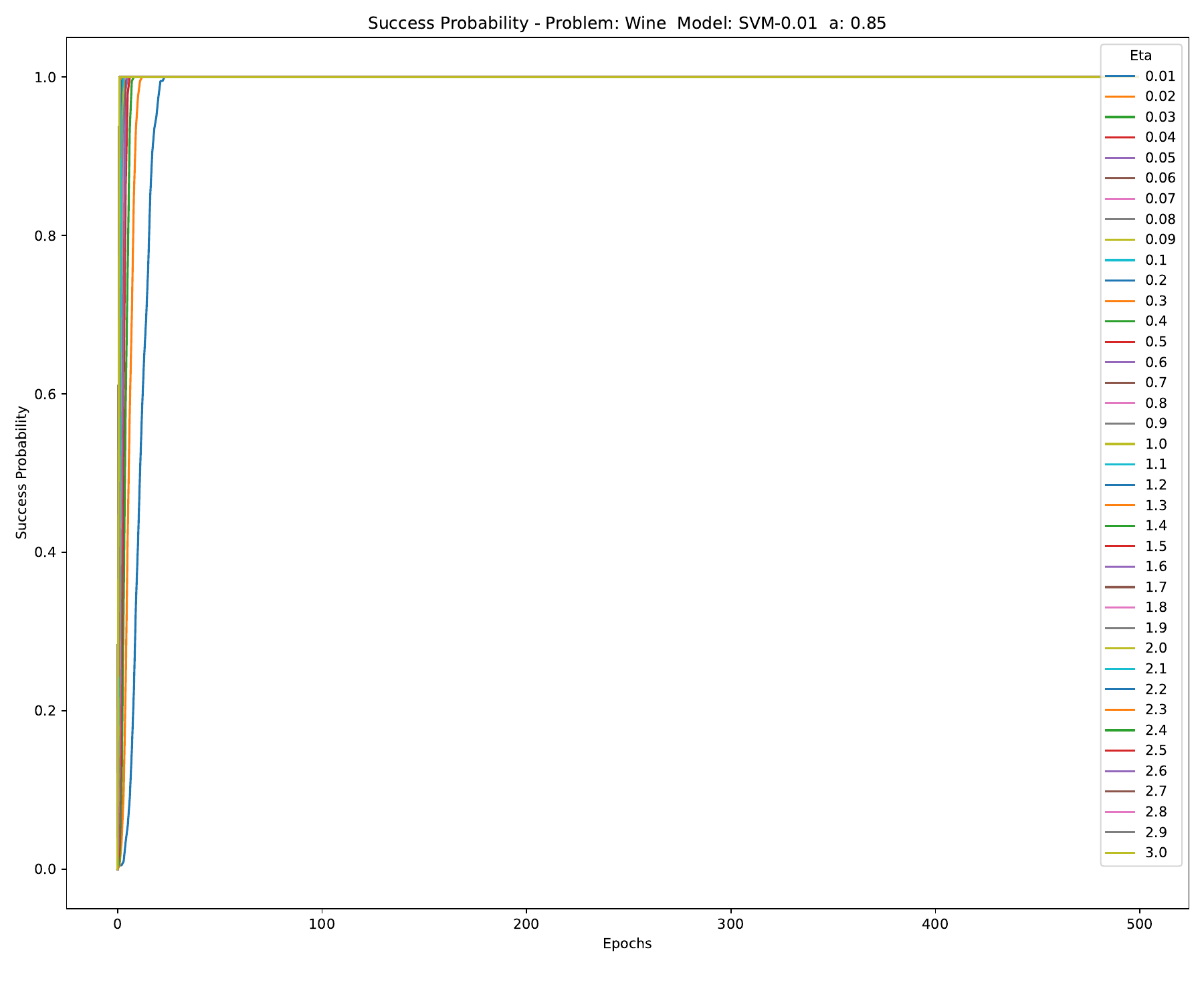} \\[-3mm]
  \end{tabular}
\caption{Success probability ($a=0.85$) for SVM-0.01 model (continued).}
\end{figure*}

\begin{figure*}[p]
  \centering
  \begin{tabular}{c@{}c}
    Training Success Probability & Validation Success Probability \\
    \includegraphics[clip,trim=0 0 0 0,width=.52\linewidth]{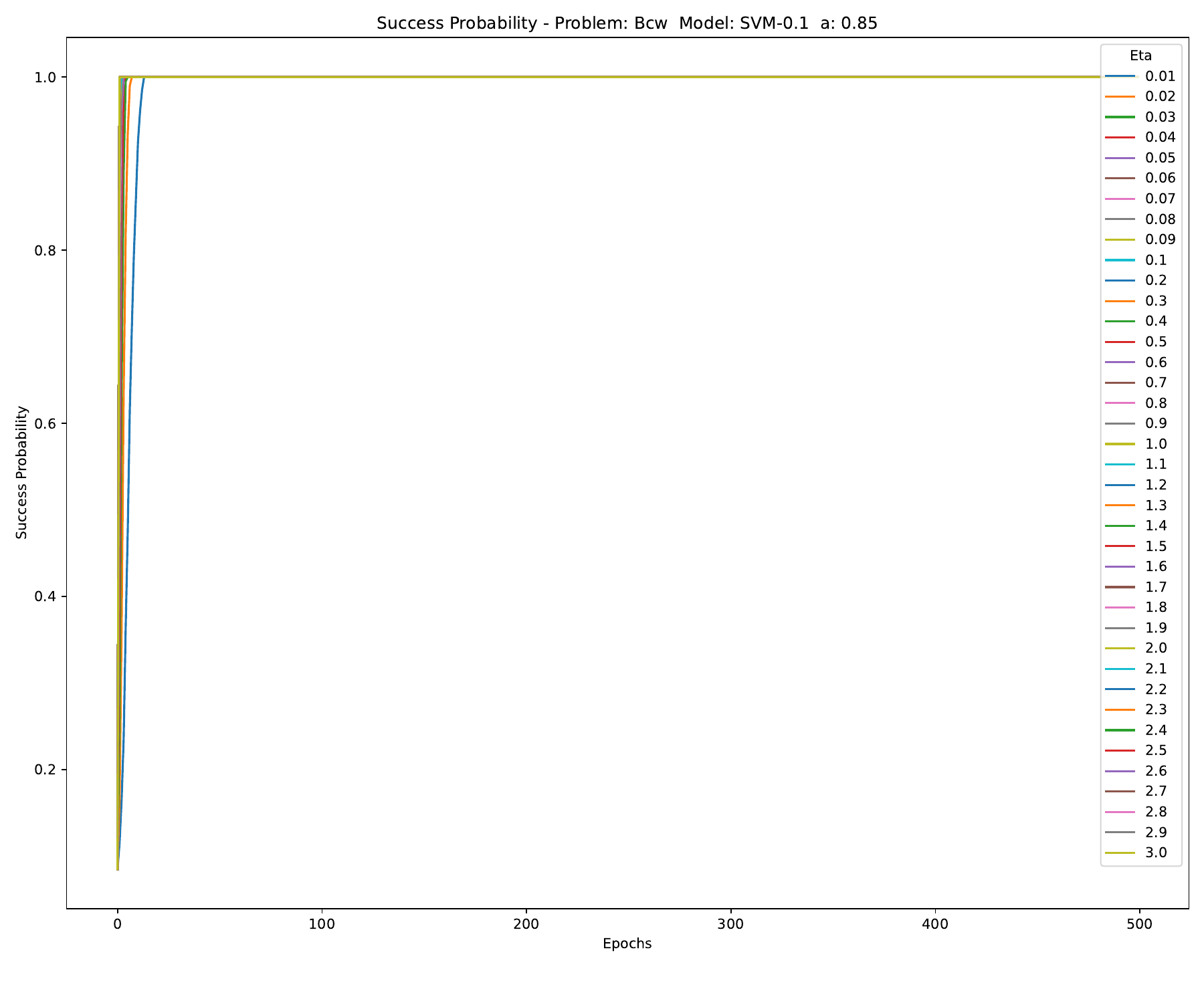} &
    \includegraphics[clip,trim=0 0 0 0,width=.52\linewidth]{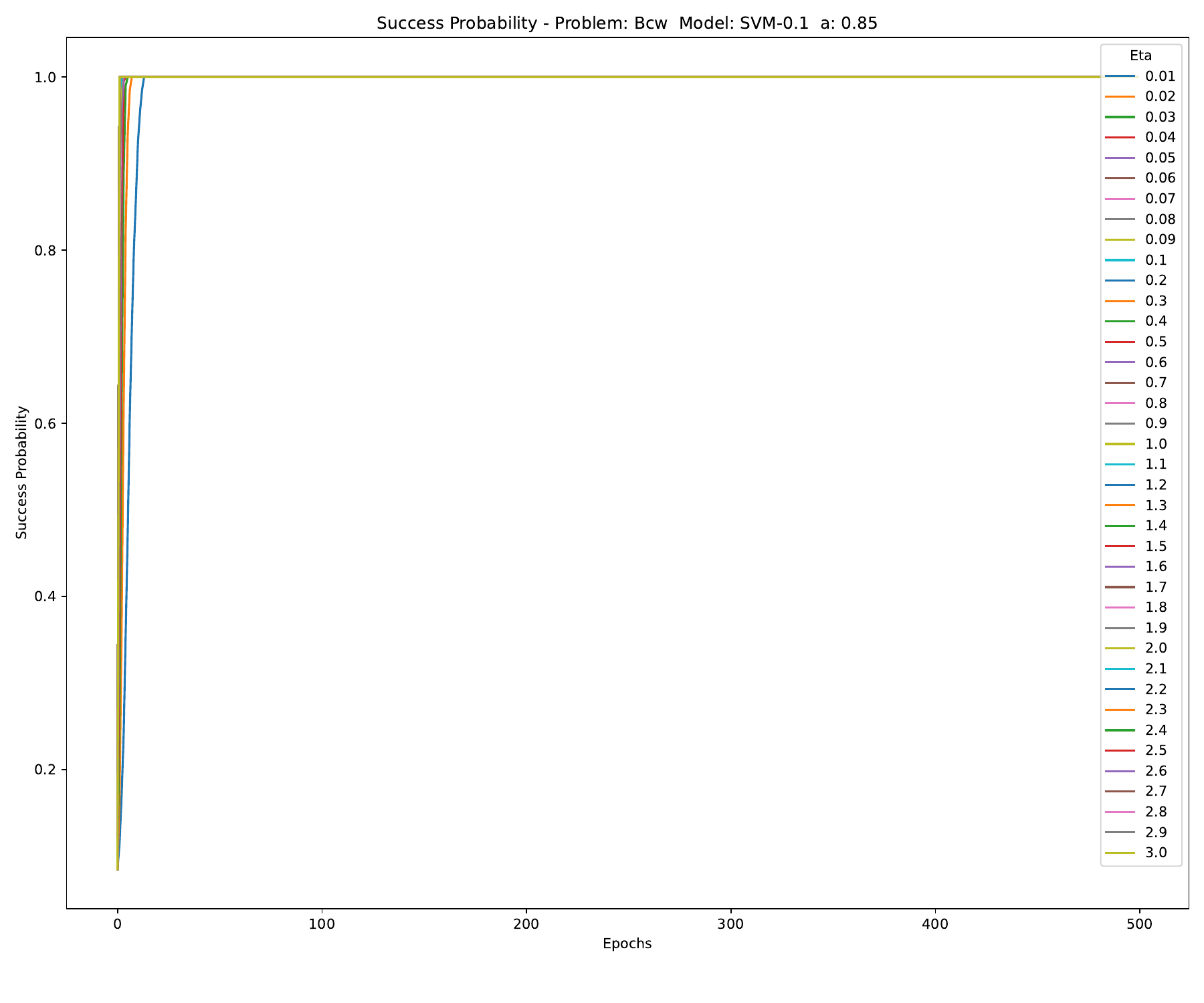} \\[-3mm]
    \includegraphics[clip,trim=0 0 0 0,width=.52\linewidth]{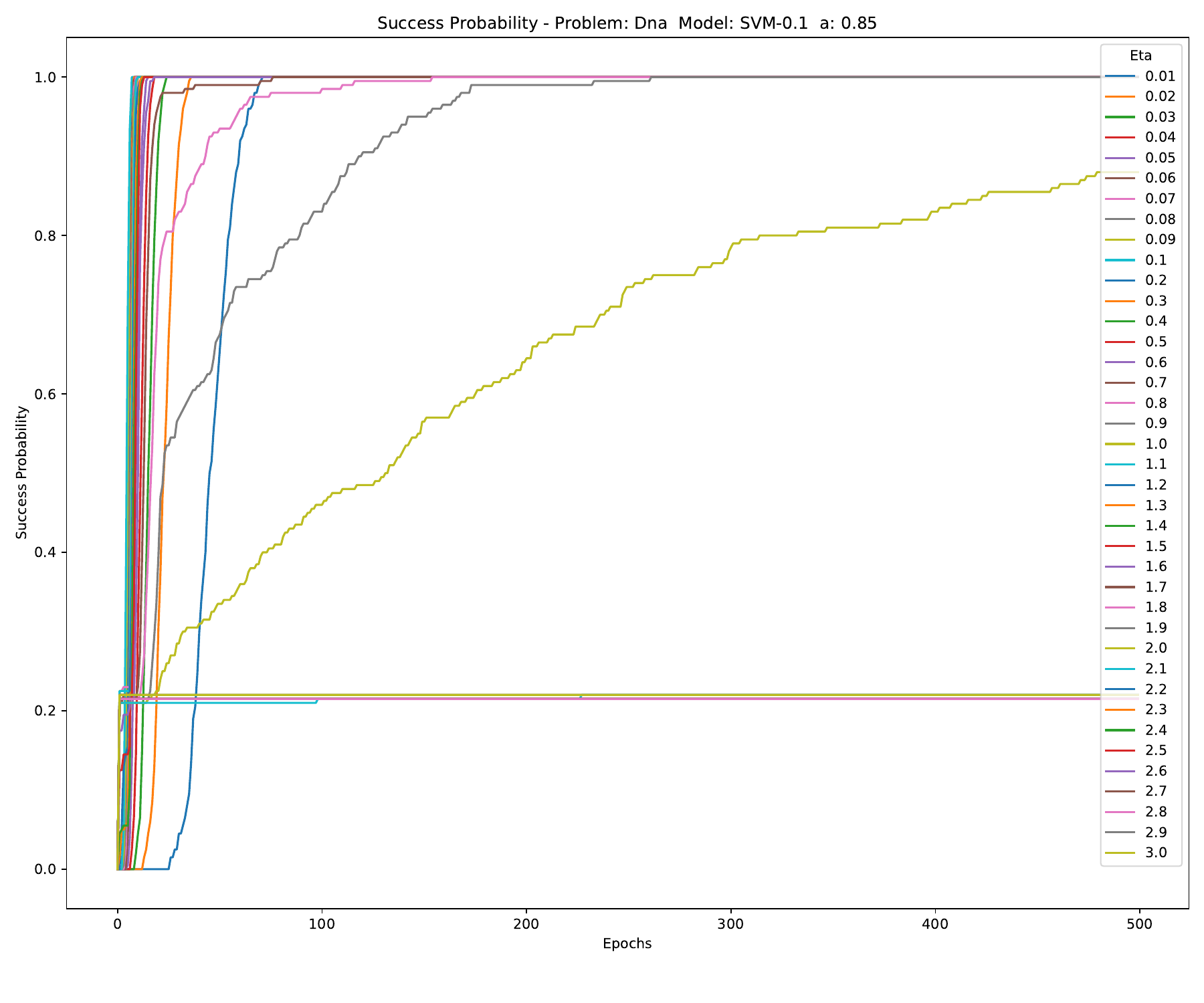} &
    \includegraphics[clip,trim=0 0 0 0,width=.52\linewidth]{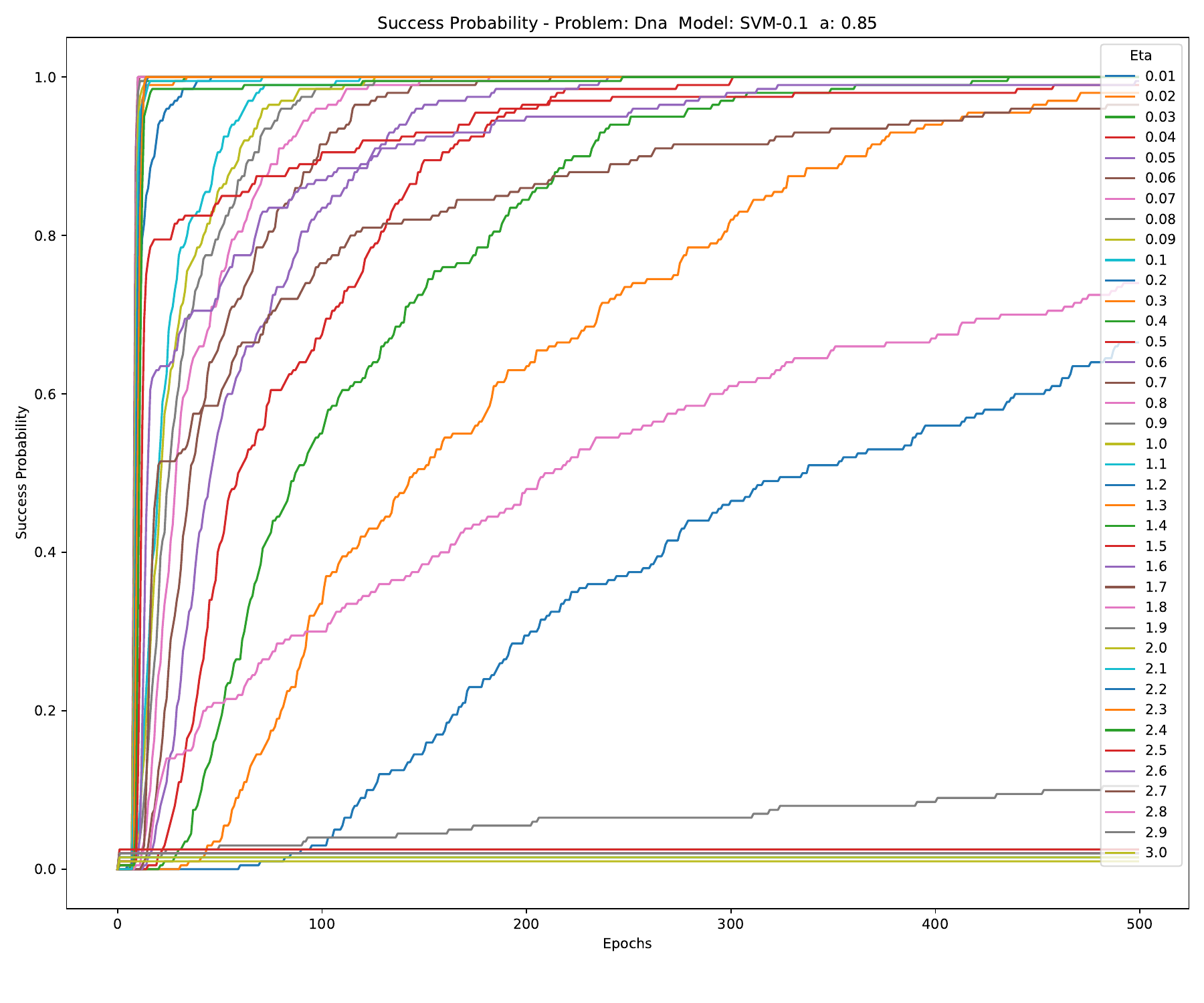} \\[-3mm]
    \includegraphics[clip,trim=0 0 0 0,width=.52\linewidth]{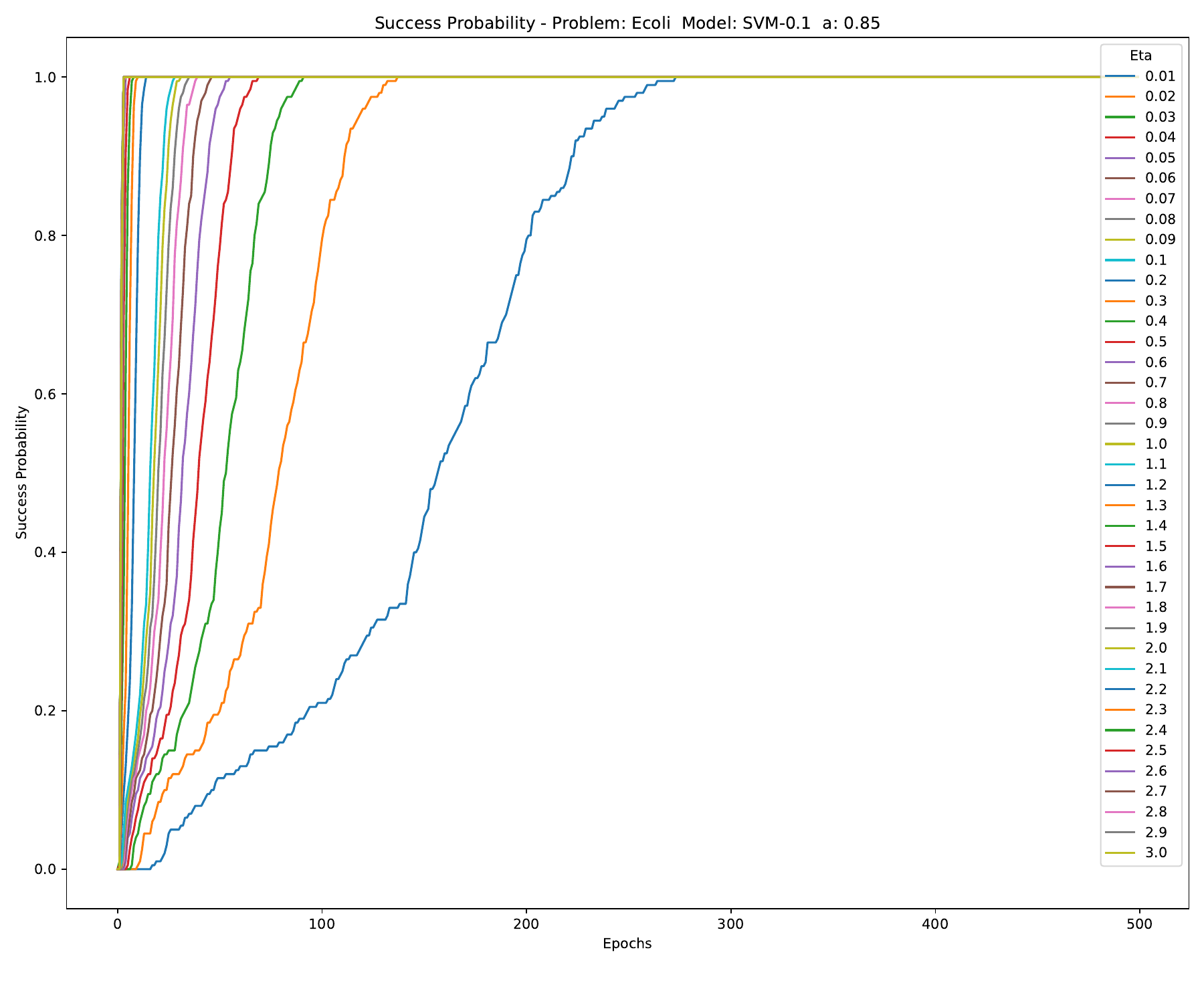} &
    \includegraphics[clip,trim=0 0 0 0,width=.52\linewidth]{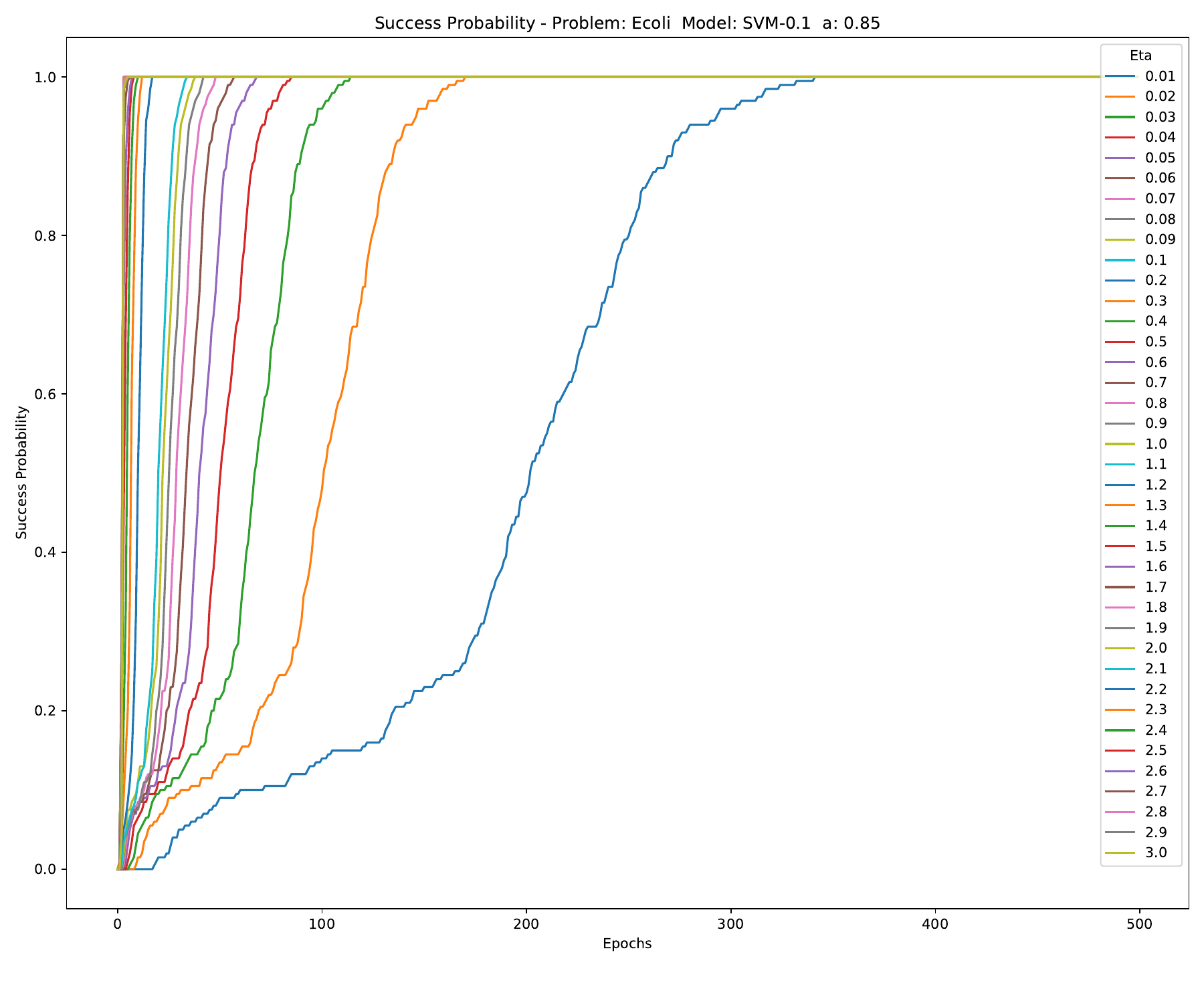} \\[-3mm]
  \end{tabular}
\caption{Success probability ($a=0.85$) for SVM-0.1 model.}
  \label{fig:success_probability_SVM_0.1_model_0.85}
\end{figure*}

\begin{figure*}[p]
  \centering
  \ContinuedFloat
  \begin{tabular}{c@{}c}
    Training Success Probability & Validation Success Probability \\
    \includegraphics[clip,trim=0 0 0 0,width=.52\linewidth]{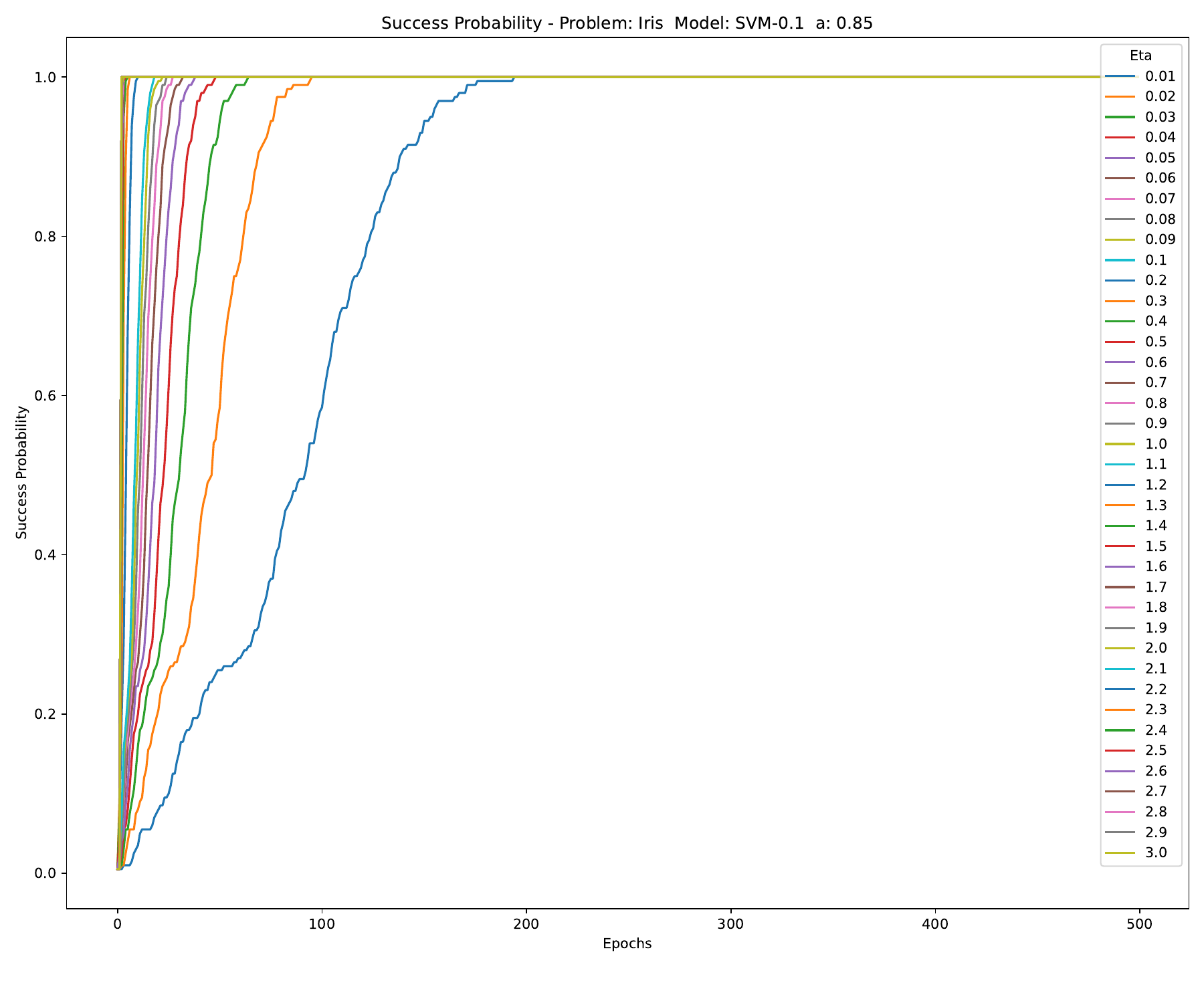} &
    \includegraphics[clip,trim=0 0 0 0,width=.52\linewidth]{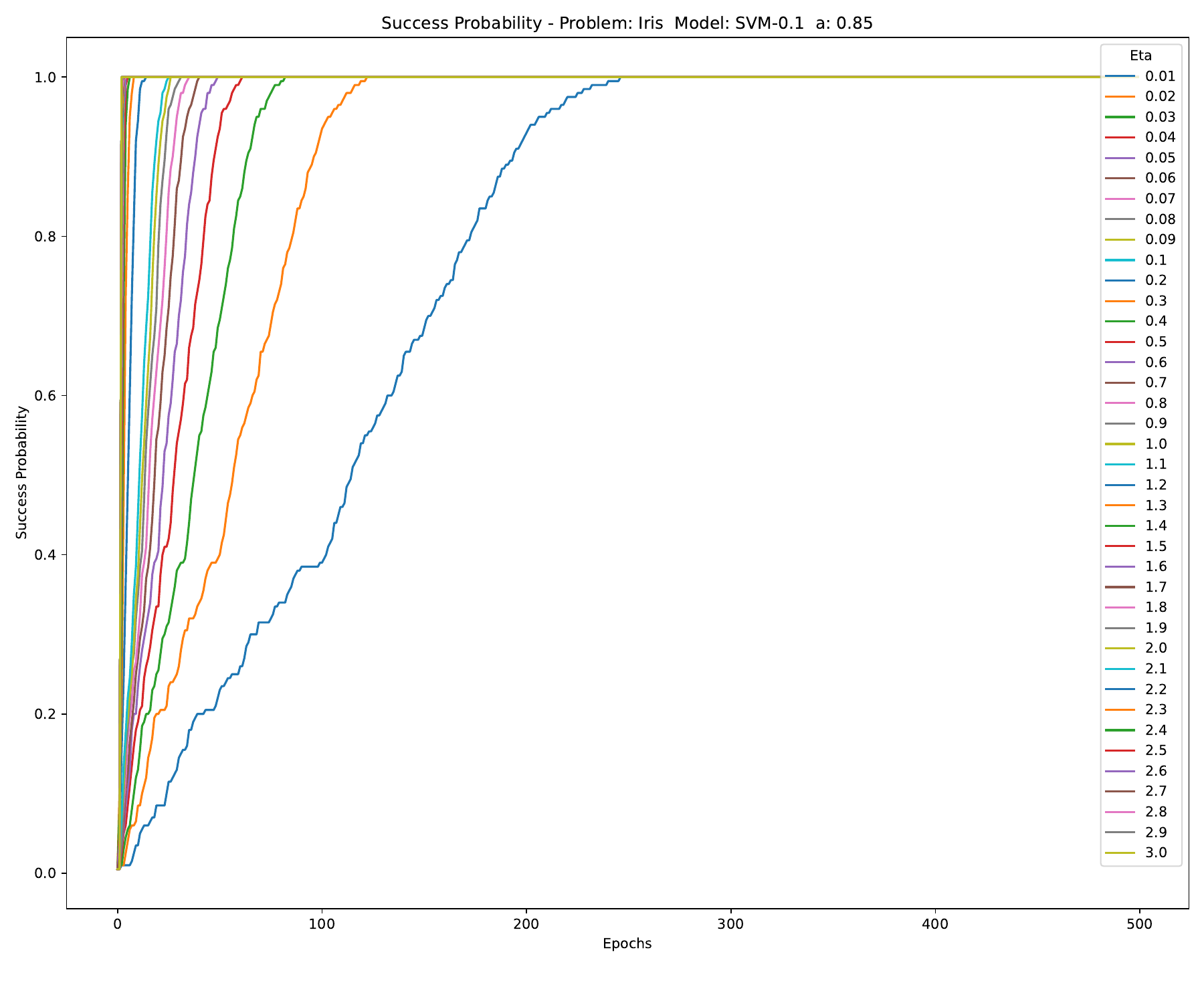} \\[-3mm]
    \includegraphics[clip,trim=0 0 0 0,width=.52\linewidth]{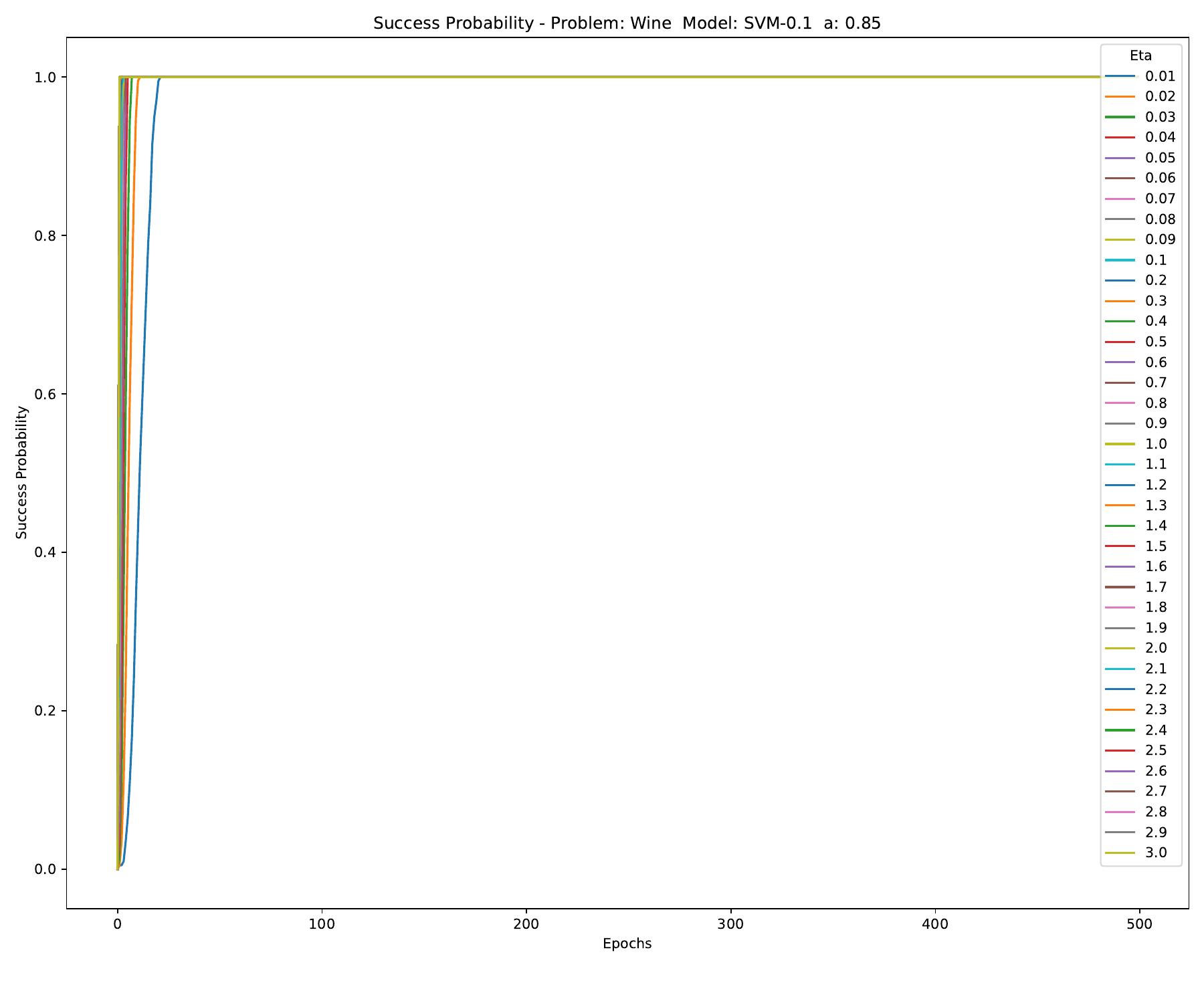} &
    \includegraphics[clip,trim=0 0 0 0,width=.52\linewidth]{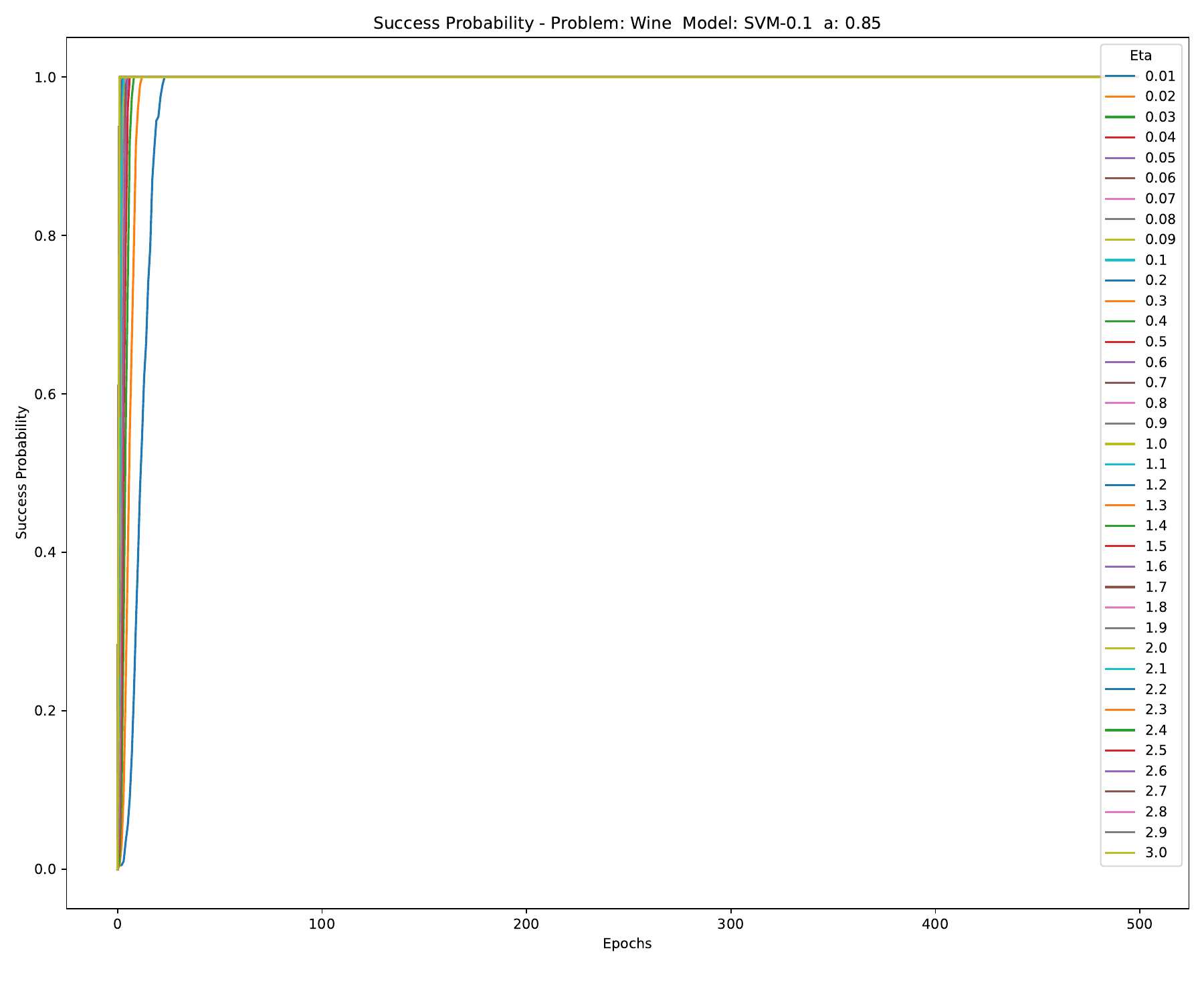} \\[-3mm]
  \end{tabular}
\caption{Success probability ($a=0.85$) for SVM-0.1 model (continued).}
\end{figure*}
\FloatBarrier

\subsection{Success Probabilities for  $a=0.9$}

In
Figures~\ref{fig:success_probability_NN_ReLU_model_0.9}---\ref{fig:success_probability_SVM_0.1_model_0.9}
we present plots of the success probabilities $P(i, 0.90, \eta)$ (that
is for $a=0.90$) for all the problems and models used in this article.

\begin{figure*}[p]
  \centering
  \begin{tabular}{c@{}c}
    Training Success Probability & Validation Success Probability \\
    \includegraphics[clip,trim=0 0 0 0,width=.52\linewidth]{arXiv-figures/bcw_NN-ReLU_a=0.9_success_prob_train_acc_thr.pdf} &
    \includegraphics[clip,trim=0 0 0 0,width=.52\linewidth]{arXiv-figures/bcw_NN-ReLU_a=0.9_success_prob_test_acc_thr.pdf} \\[-3mm]
    \includegraphics[clip,trim=0 0 0 0,width=.52\linewidth]{arXiv-figures/dna_NN-ReLU_a=0.9_success_prob_train_acc_thr.pdf} &
    \includegraphics[clip,trim=0 0 0 0,width=.52\linewidth]{arXiv-figures/dna_NN-ReLU_a=0.9_success_prob_test_acc_thr.pdf} \\[-3mm]
    \includegraphics[clip,trim=0 0 0 0,width=.52\linewidth]{arXiv-figures/ecoli_NN-ReLU_a=0.9_success_prob_train_acc_thr.pdf} &
    \includegraphics[clip,trim=0 0 0 0,width=.52\linewidth]{arXiv-figures/ecoli_NN-ReLU_a=0.9_success_prob_test_acc_thr.pdf} \\[-3mm]
  \end{tabular}
\caption{Success probability ($a=0.9$) for NN-ReLU model.}
  \label{fig:success_probability_NN_ReLU_model_0.9_duplicate}
\end{figure*}

\begin{figure*}[p]
  \centering
  \ContinuedFloat
  \begin{tabular}{c@{}c}
    Training Success Probability & Validation Success Probability \\
    \includegraphics[clip,trim=0 0 0 0,width=.52\linewidth]{arXiv-figures/iris_NN-ReLU_a=0.9_success_prob_train_acc_thr.pdf} &
    \includegraphics[clip,trim=0 0 0 0,width=.52\linewidth]{arXiv-figures/iris_NN-ReLU_a=0.9_success_prob_test_acc_thr.pdf} \\[-3mm]
    \includegraphics[clip,trim=0 0 0 0,width=.52\linewidth]{arXiv-figures/wine_NN-ReLU_a=0.9_success_prob_train_acc_thr.pdf} &
    \includegraphics[clip,trim=0 0 0 0,width=.52\linewidth]{arXiv-figures/wine_NN-ReLU_a=0.9_success_prob_test_acc_thr.pdf} \\[-3mm]
  \end{tabular}
\caption{Success probability ($a=0.9$) for NN-ReLU model (continued).}
\end{figure*}

\begin{figure*}[p]
  \centering
  \begin{tabular}{c@{}c}
    Training Success Probability & Validation Success Probability \\
    \includegraphics[clip,trim=0 0 0 0,width=.52\linewidth]{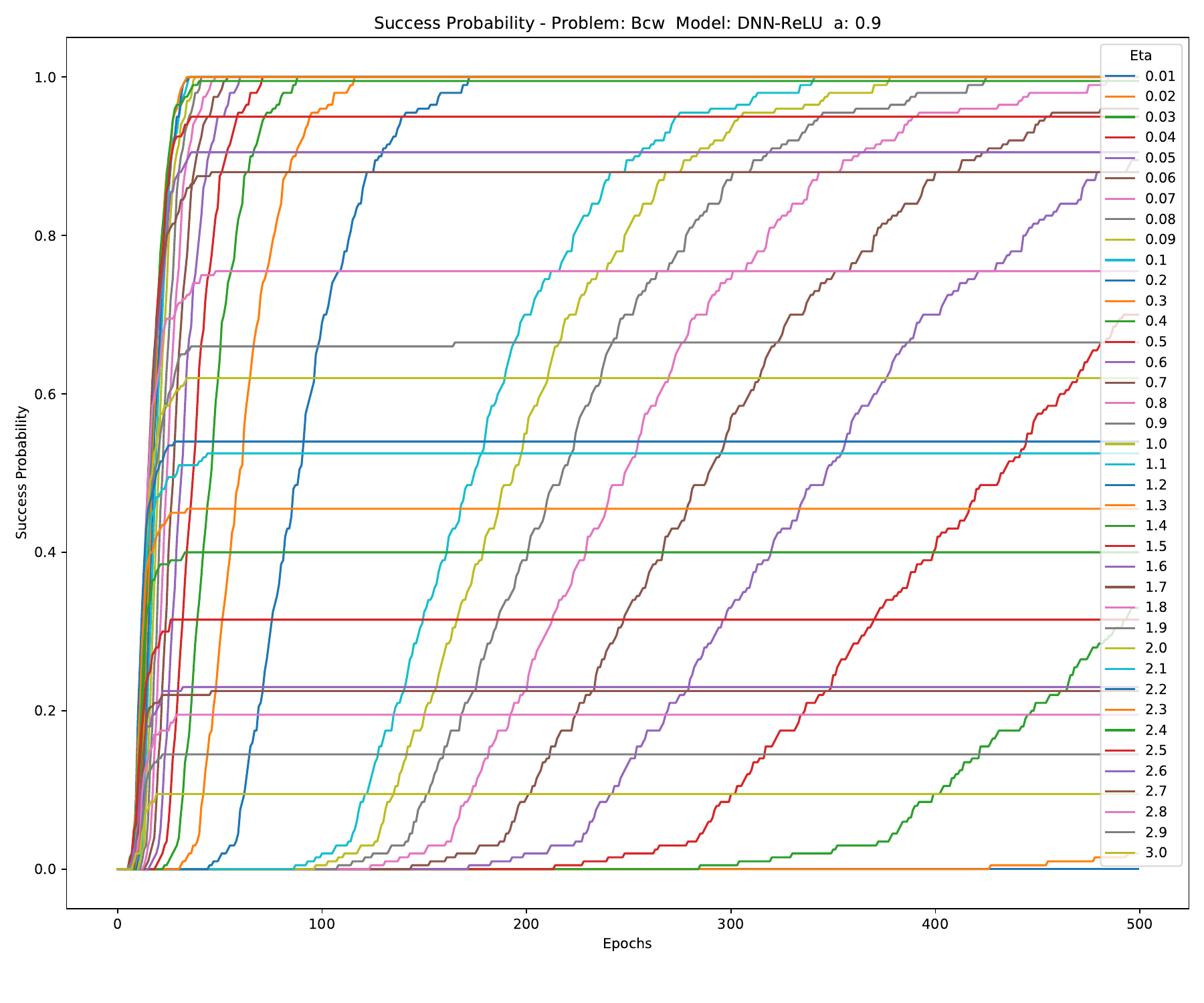} &
    \includegraphics[clip,trim=0 0 0 0,width=.52\linewidth]{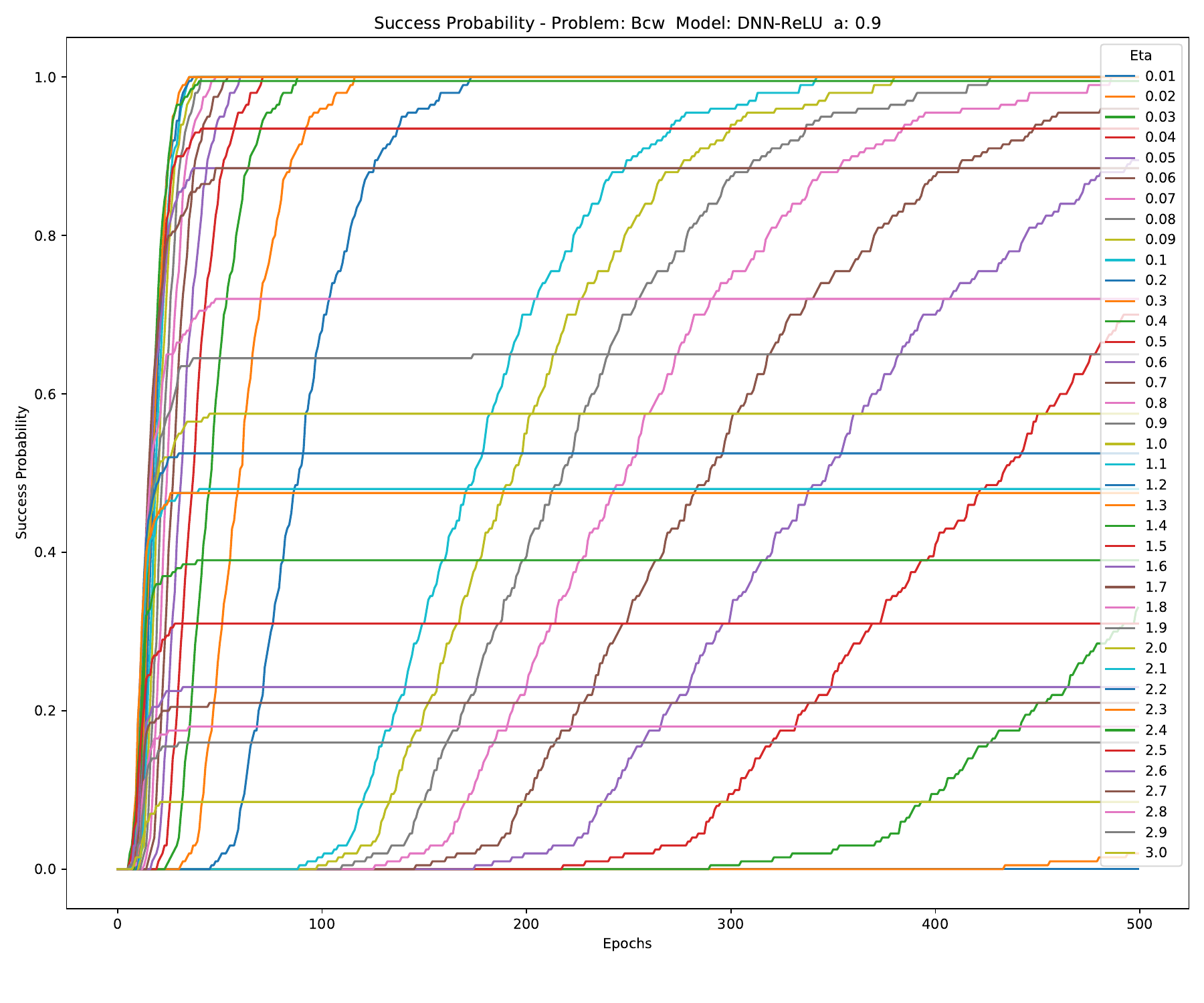} \\[-3mm]
    \includegraphics[clip,trim=0 0 0 0,width=.52\linewidth]{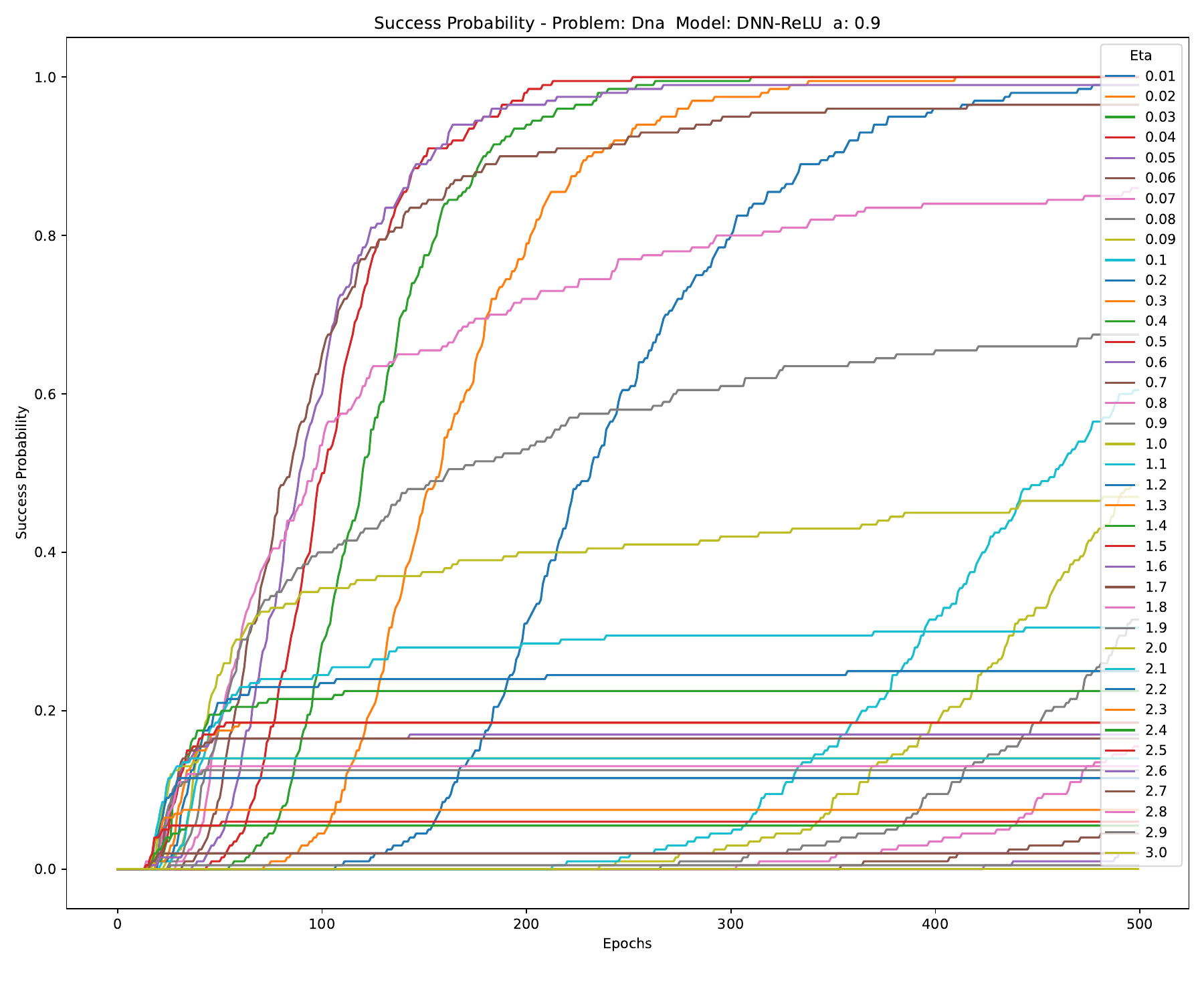} &
    \includegraphics[clip,trim=0 0 0 0,width=.52\linewidth]{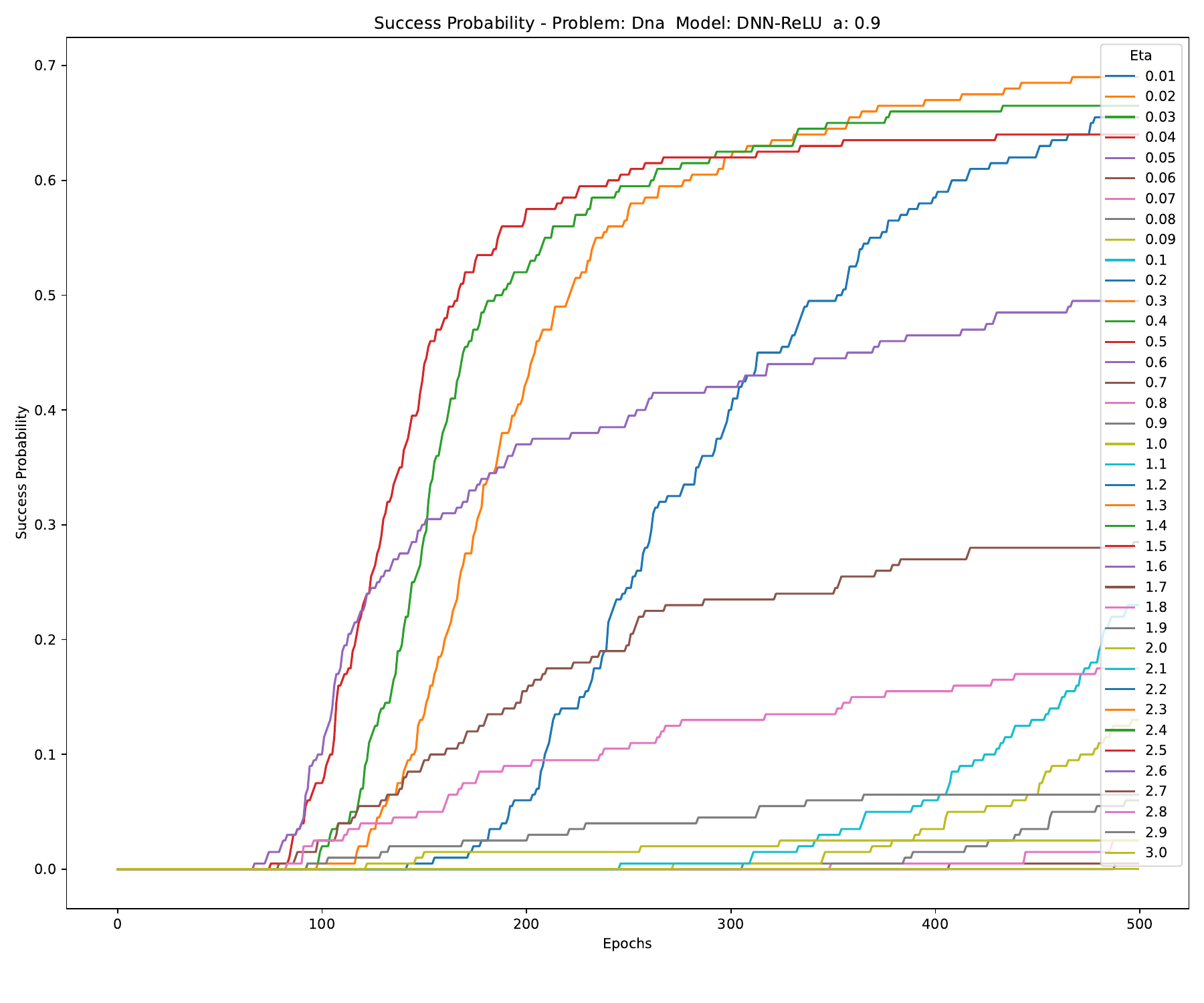} \\[-3mm]
    \includegraphics[clip,trim=0 0 0 0,width=.52\linewidth]{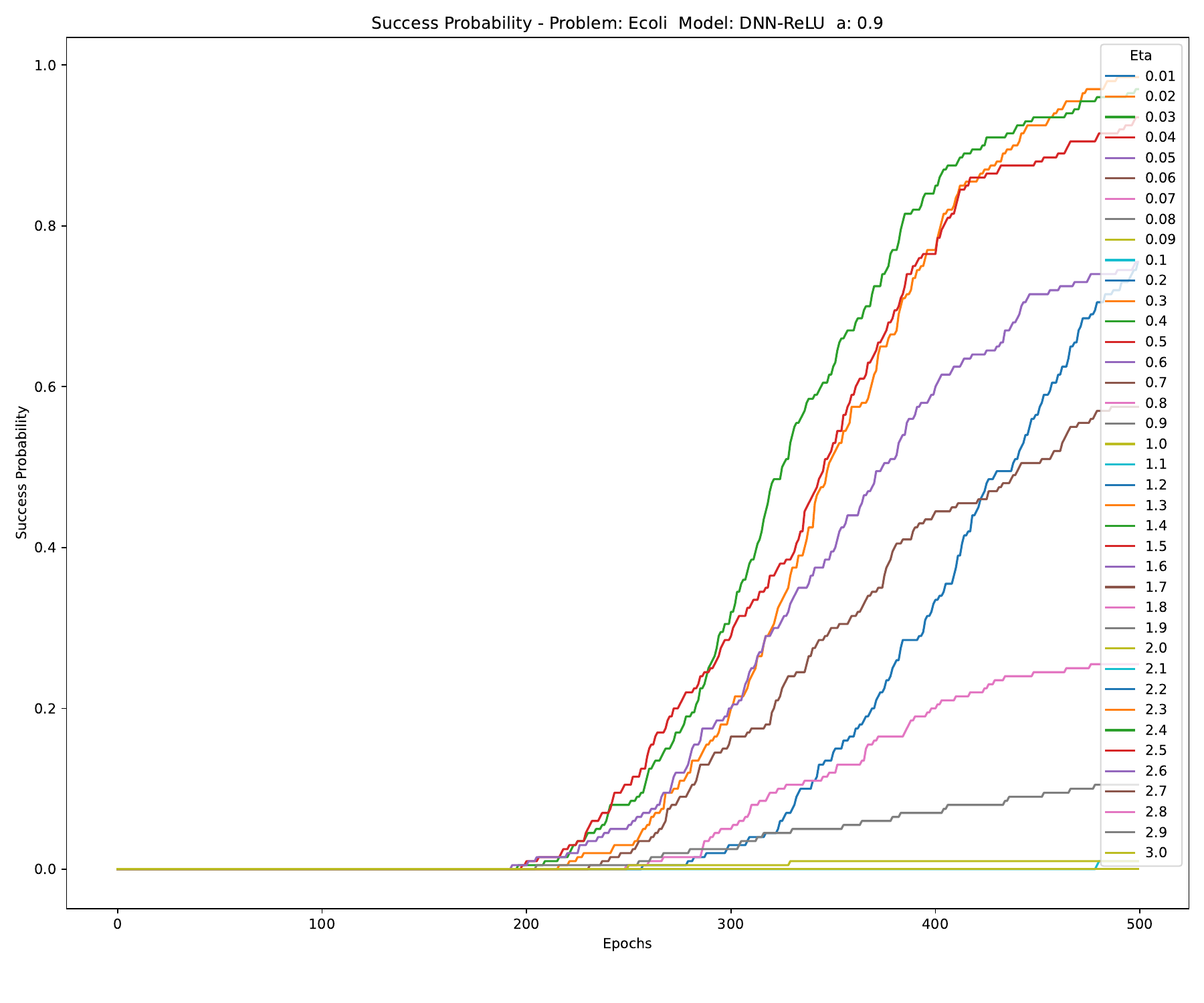} &
    \includegraphics[clip,trim=0 0 0 0,width=.52\linewidth]{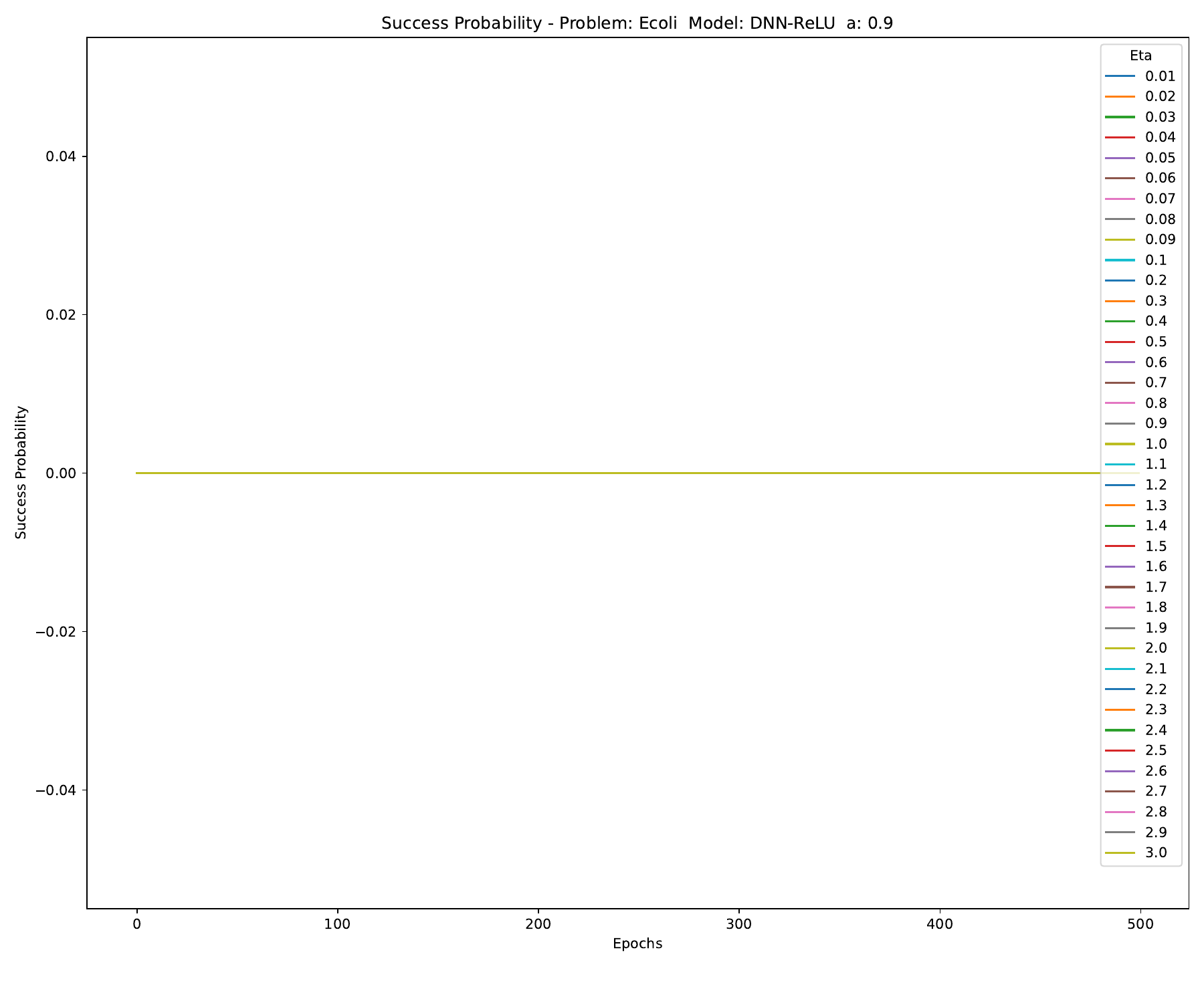} \\[-3mm]
  \end{tabular}
\caption{Success probability ($a=0.9$) for DNN-ReLU model.}
  \label{fig:success_probability_DNN_ReLU_model_0.9}
\end{figure*}

\begin{figure*}[p]
  \centering
  \ContinuedFloat
  \begin{tabular}{c@{}c}
    Training Success Probability & Validation Success Probability \\
    \includegraphics[clip,trim=0 0 0 0,width=.52\linewidth]{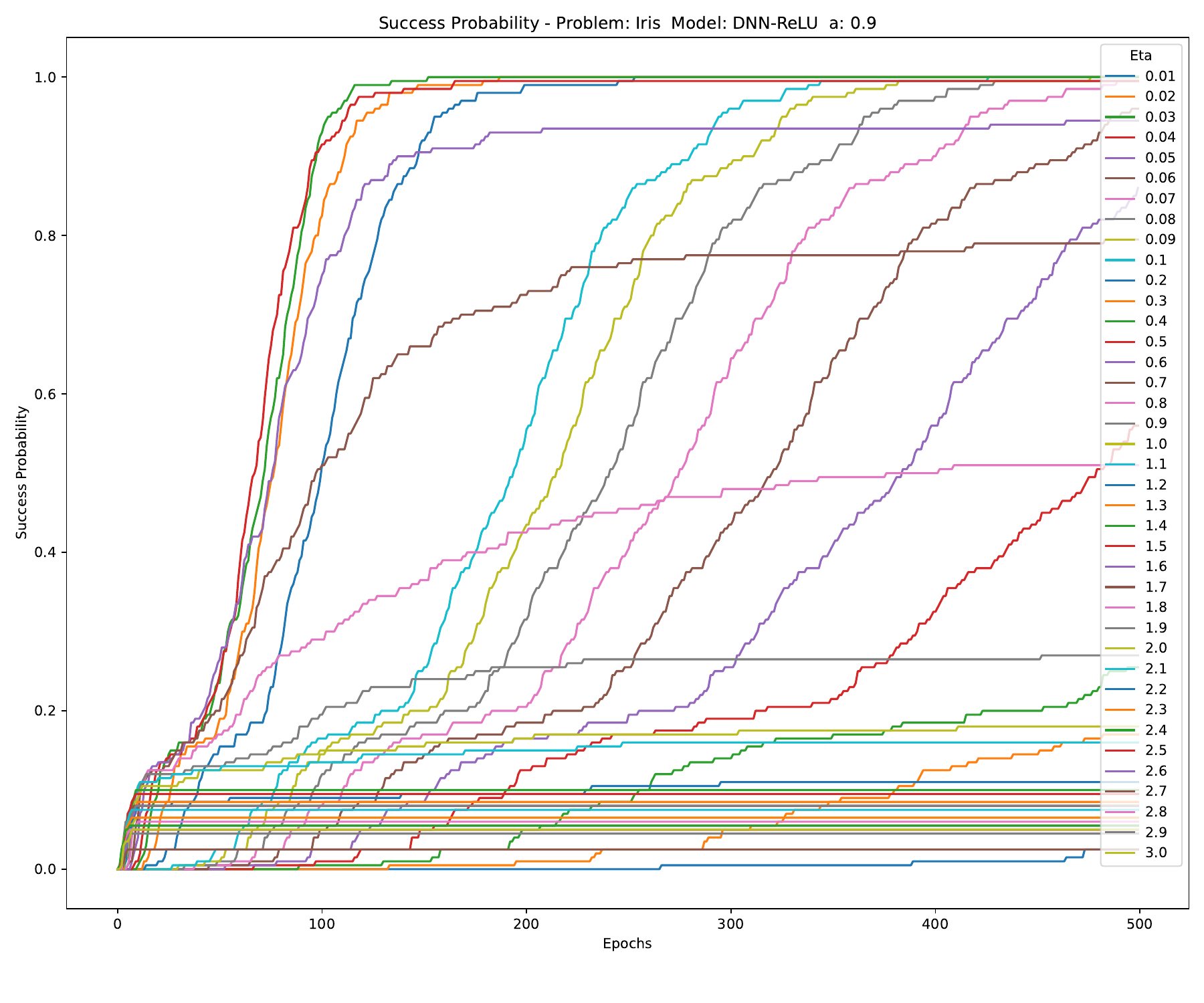} &
    \includegraphics[clip,trim=0 0 0 0,width=.52\linewidth]{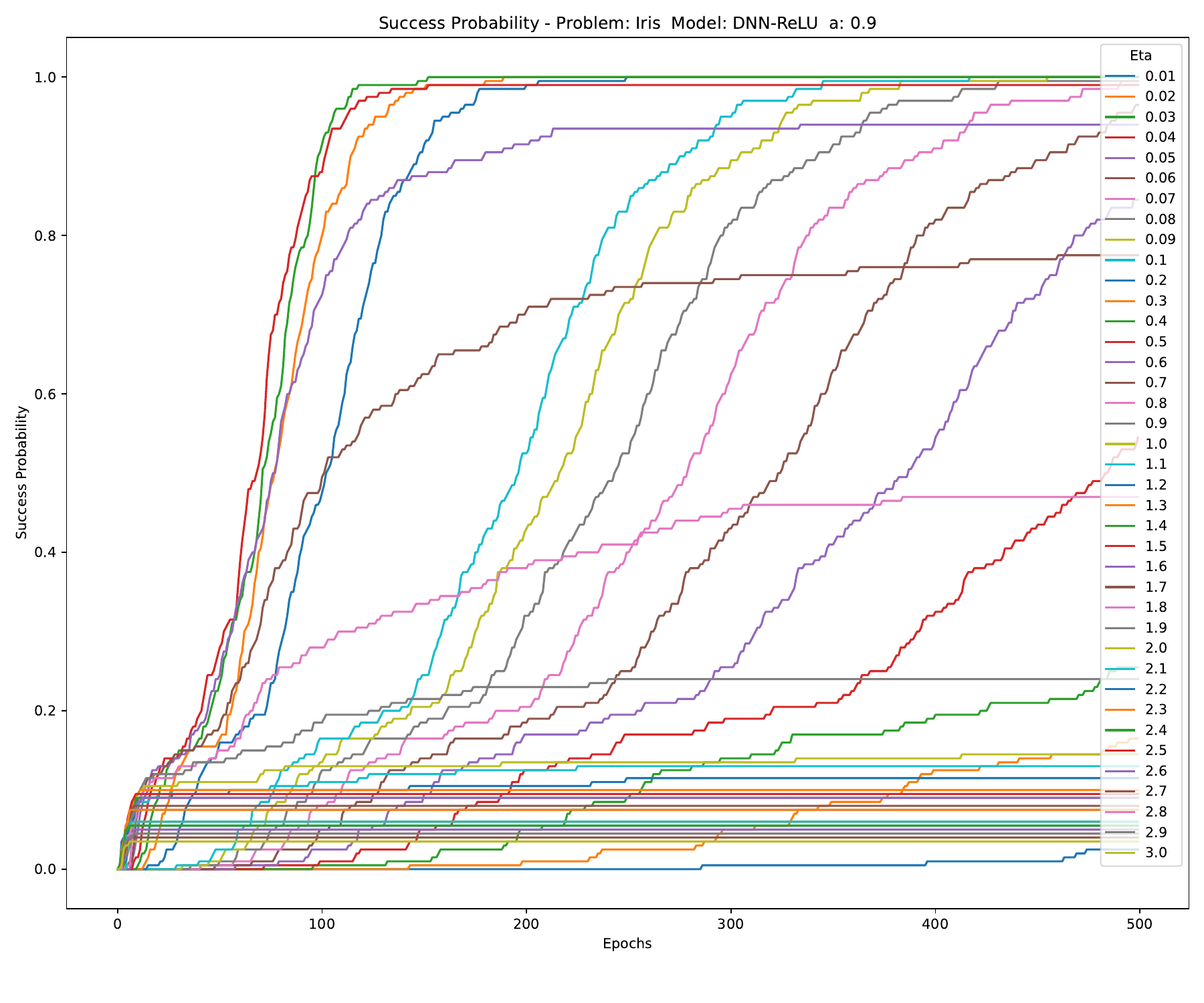} \\[-3mm]
    \includegraphics[clip,trim=0 0 0 0,width=.52\linewidth]{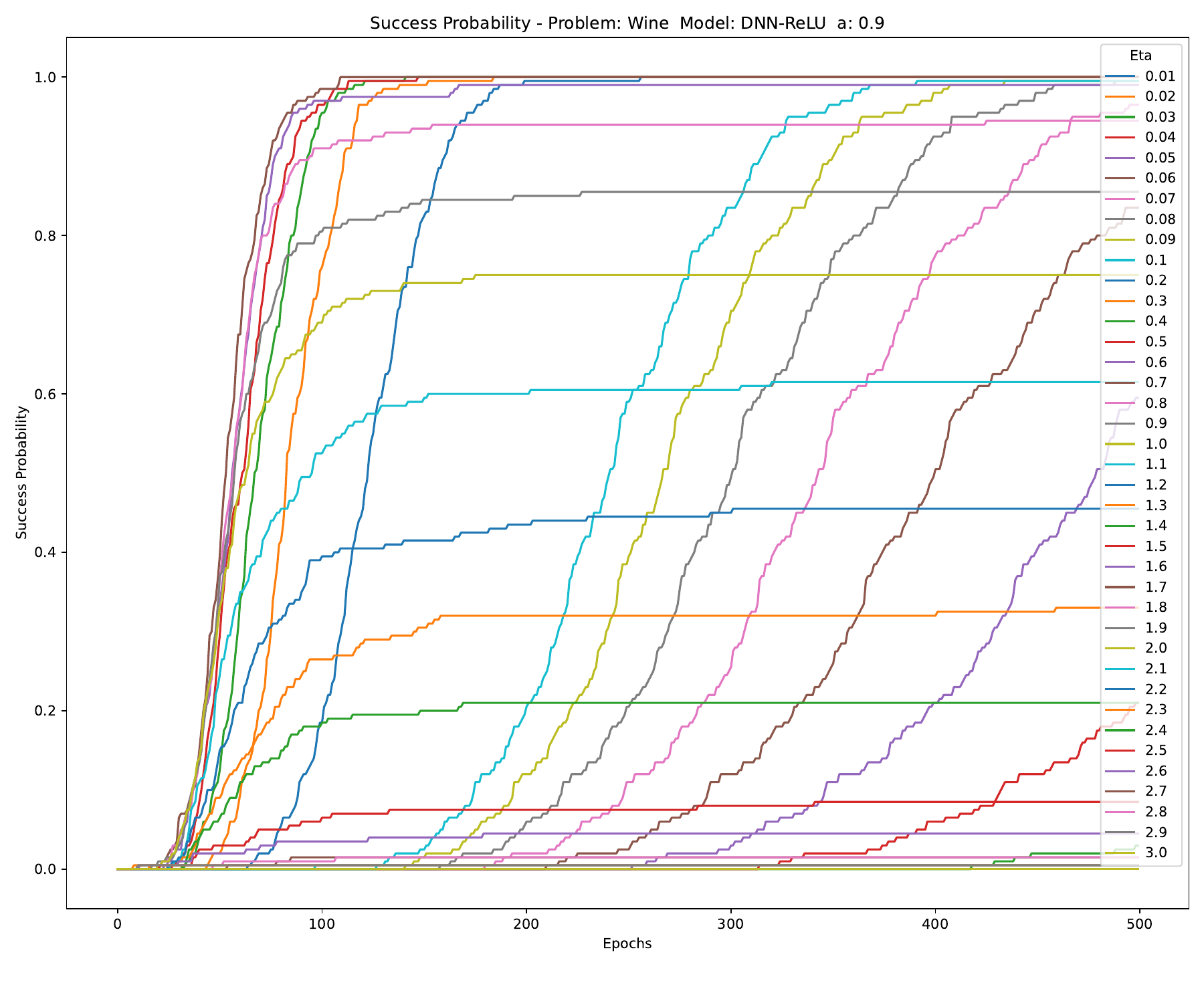} &
    \includegraphics[clip,trim=0 0 0 0,width=.52\linewidth]{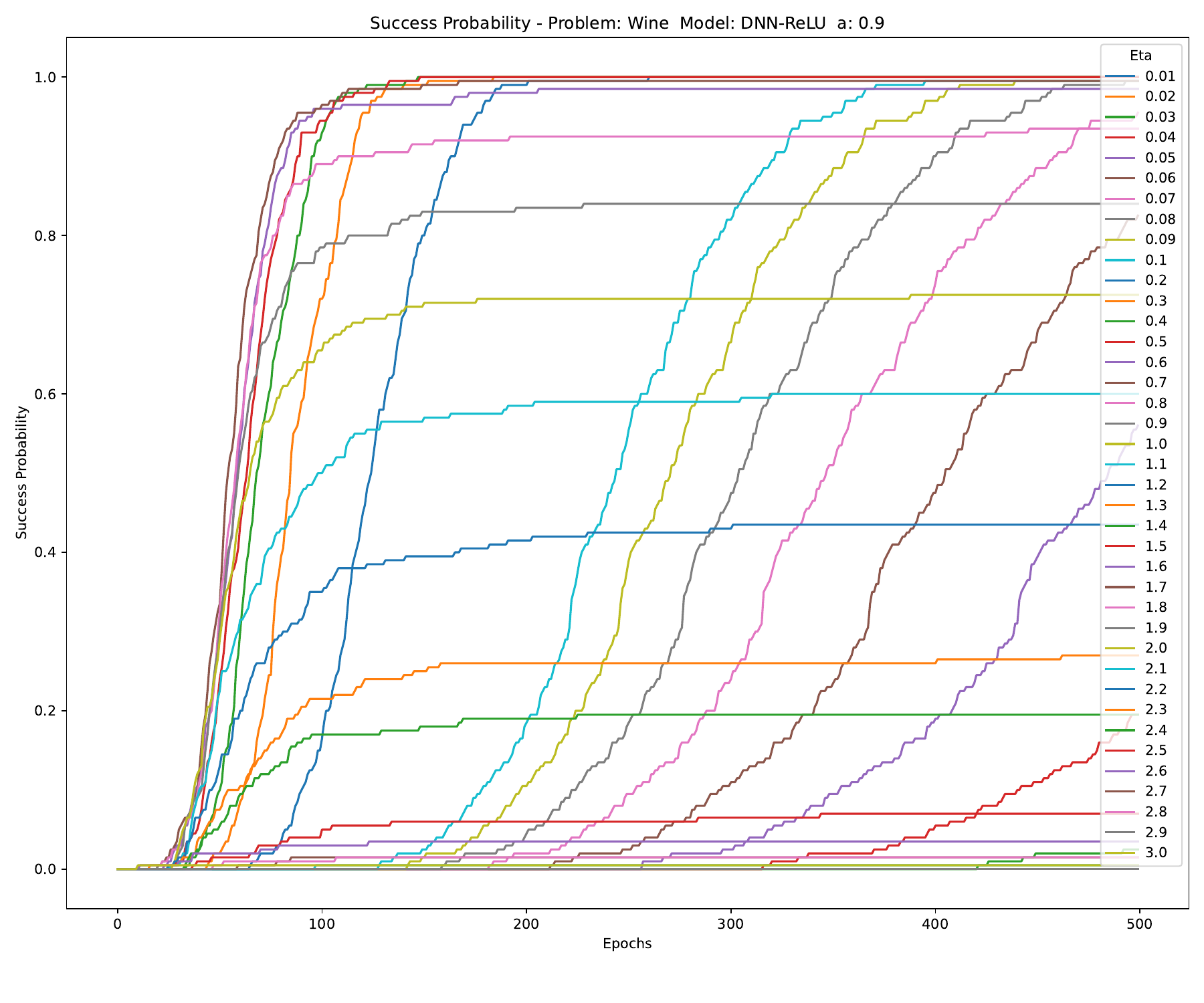} \\[-3mm]
  \end{tabular}
\caption{Success probability ($a=0.9$) for DNN-ReLU model (continued).}
\end{figure*}

\begin{figure*}[p]
  \centering
  \begin{tabular}{c@{}c}
    Training Success Probability & Validation Success Probability \\
    \includegraphics[clip,trim=0 0 0 0,width=.52\linewidth]{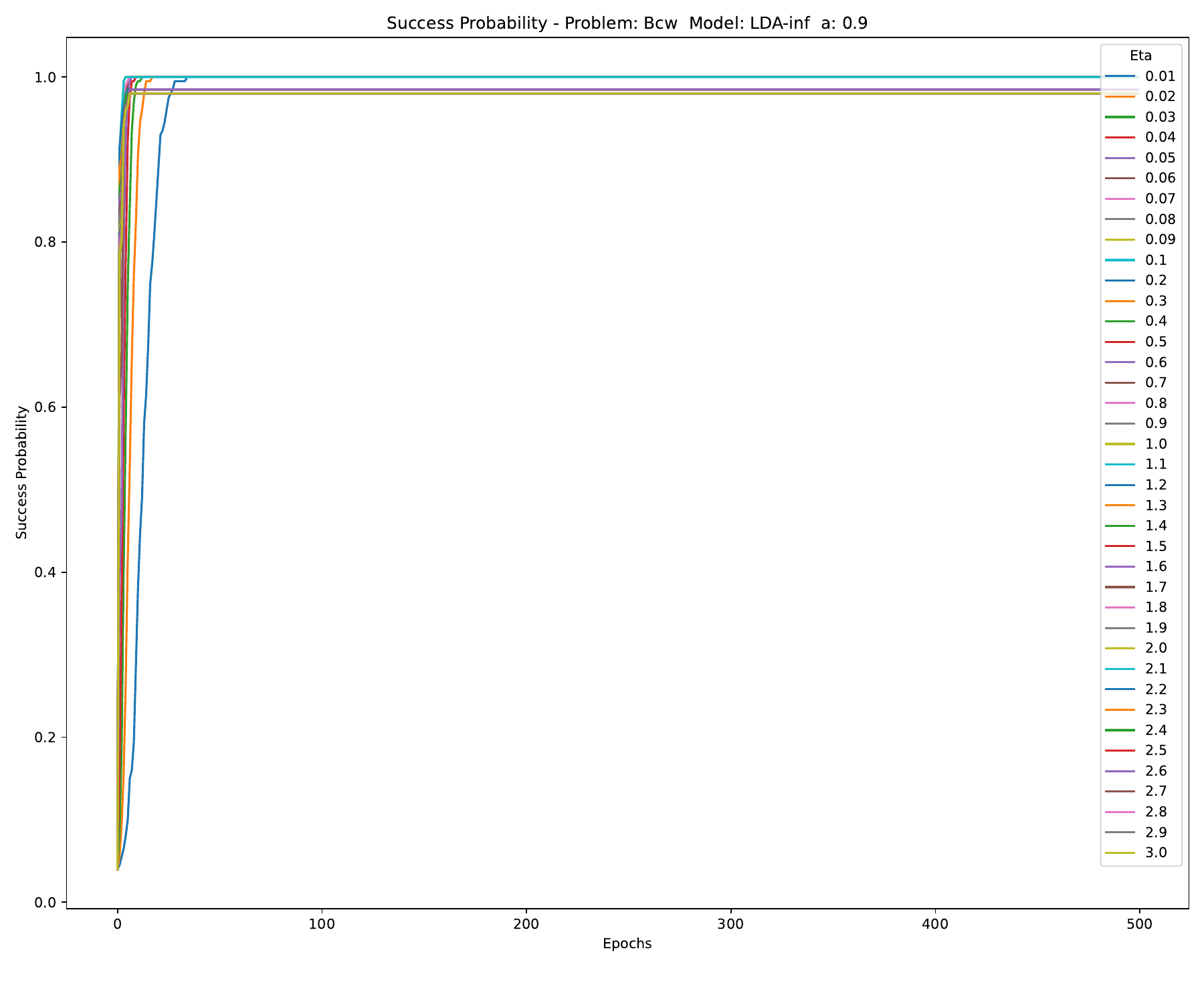} &
    \includegraphics[clip,trim=0 0 0 0,width=.52\linewidth]{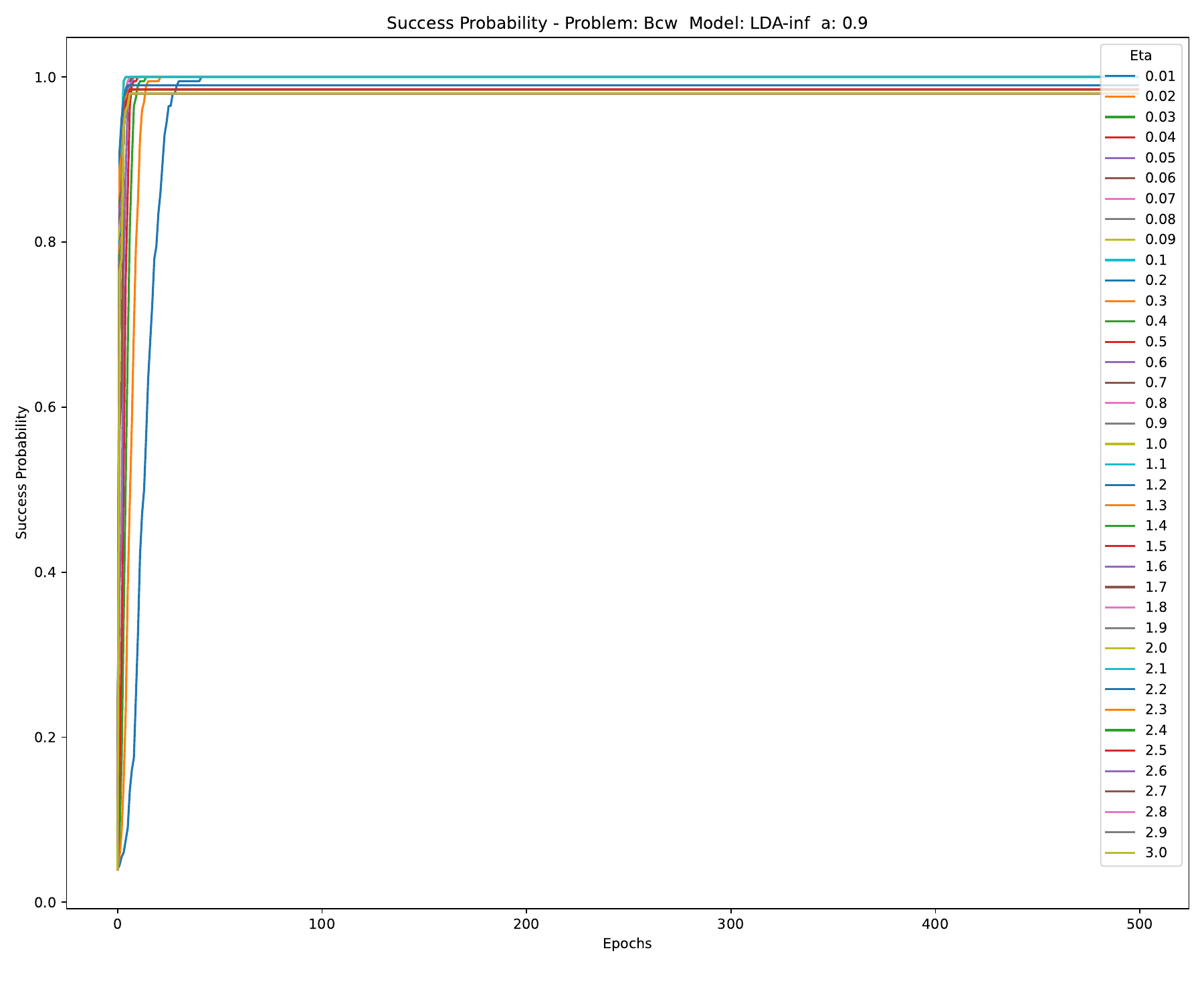} \\[-3mm]
    \includegraphics[clip,trim=0 0 0 0,width=.52\linewidth]{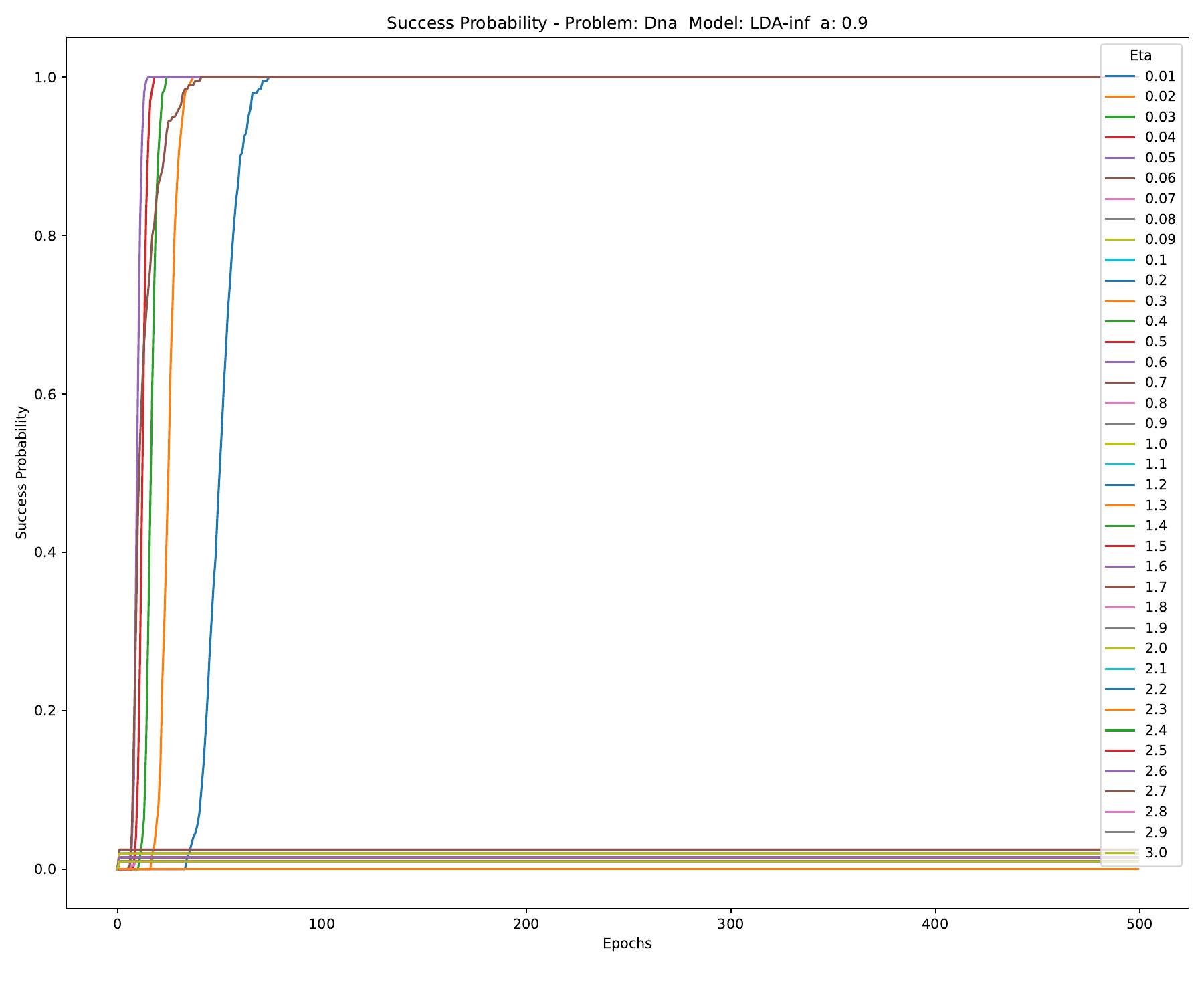} &
    \includegraphics[clip,trim=0 0 0 0,width=.52\linewidth]{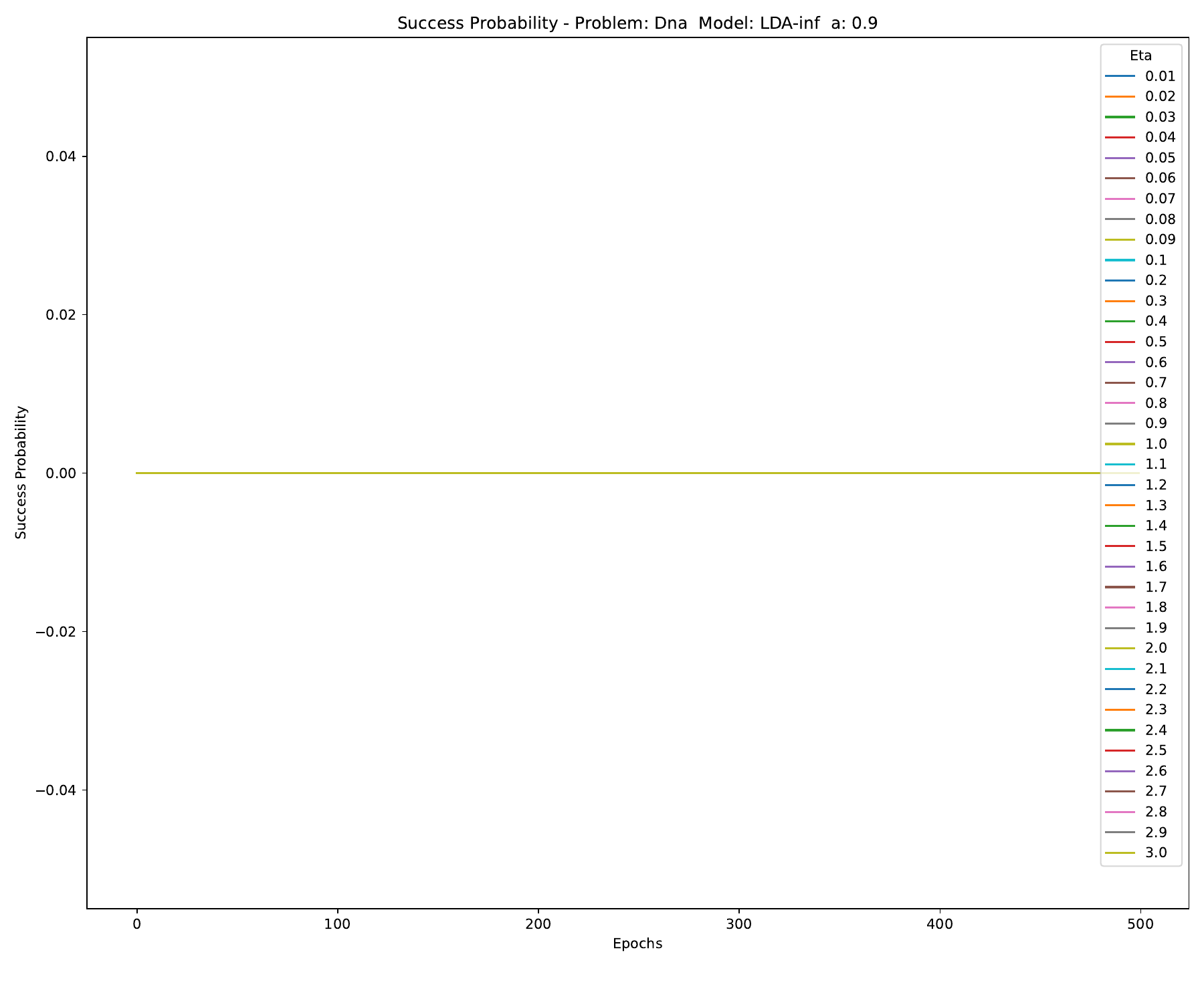} \\[-3mm]
    \includegraphics[clip,trim=0 0 0 0,width=.52\linewidth]{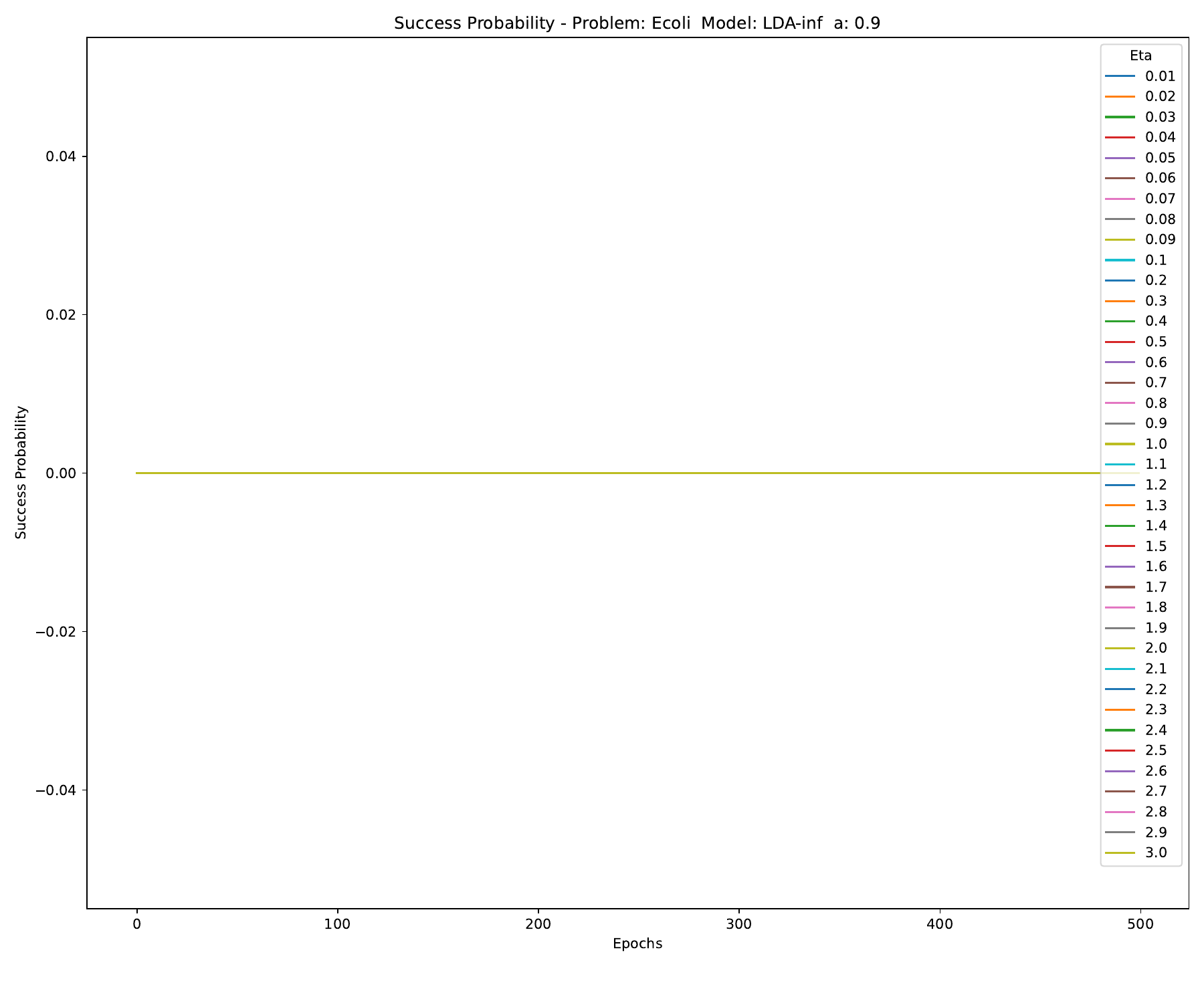} &
    \includegraphics[clip,trim=0 0 0 0,width=.52\linewidth]{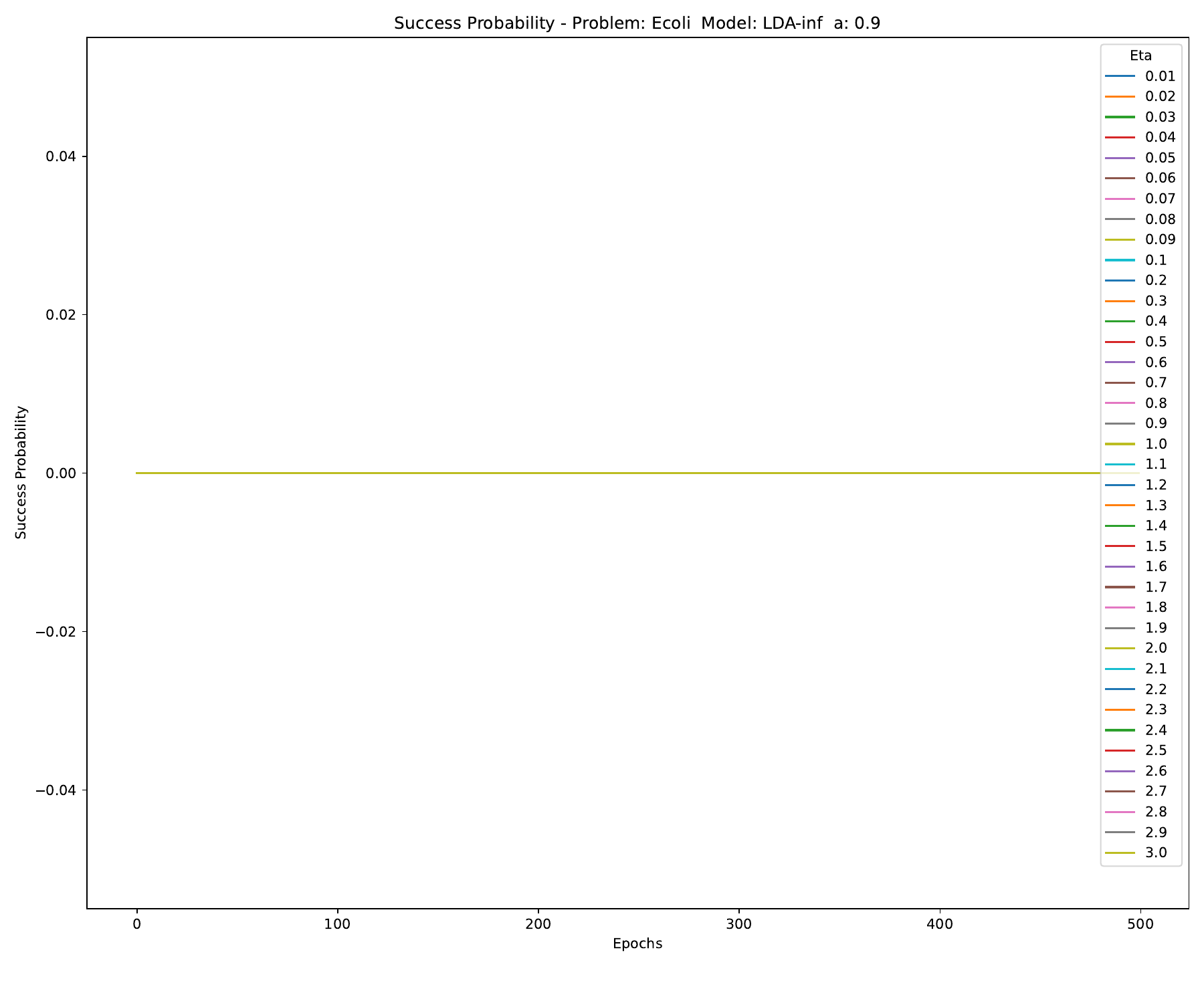} \\[-3mm]
  \end{tabular}
\caption{Success probability ($a=0.9$) for LDA-Inf model.}
  \label{fig:success_probability_LDA_Inf_model_0.9}
\end{figure*}

\begin{figure*}[p]
  \centering
  \ContinuedFloat
  \begin{tabular}{c@{}c}
    Training Success Probability & Validation Success Probability \\
    \includegraphics[clip,trim=0 0 0 0,width=.52\linewidth]{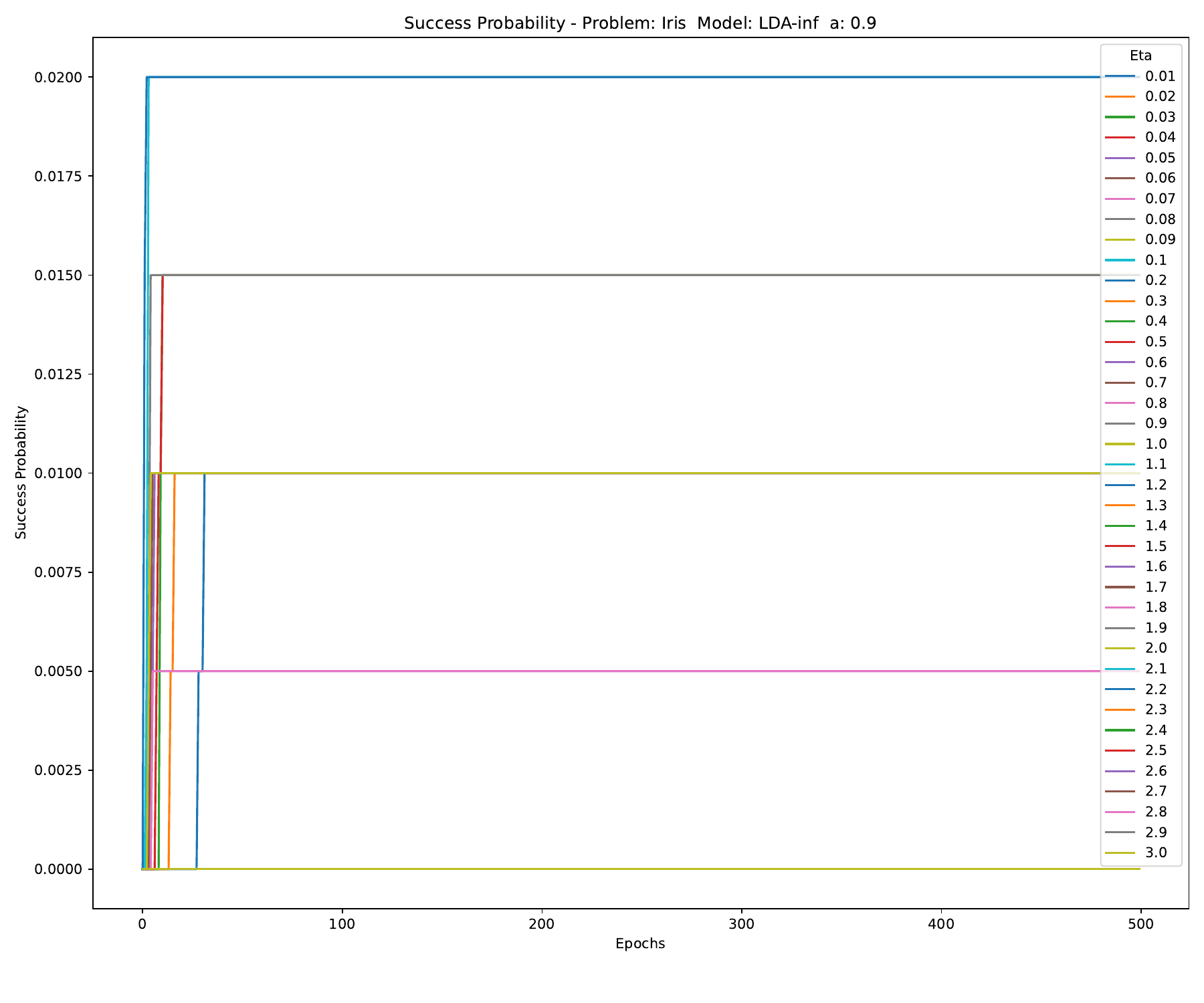} &
    \includegraphics[clip,trim=0 0 0 0,width=.52\linewidth]{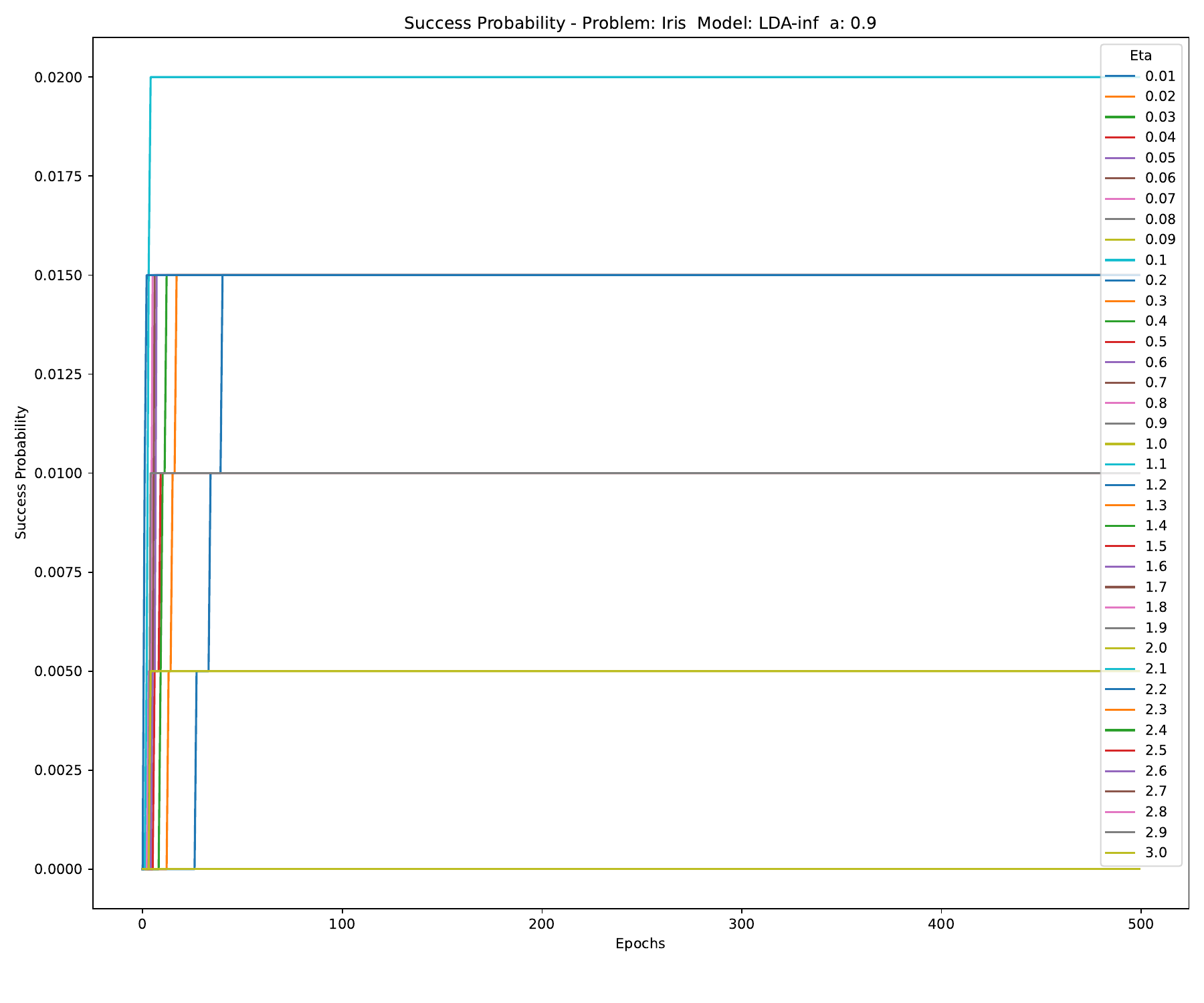} \\[-3mm]
    \includegraphics[clip,trim=0 0 0 0,width=.52\linewidth]{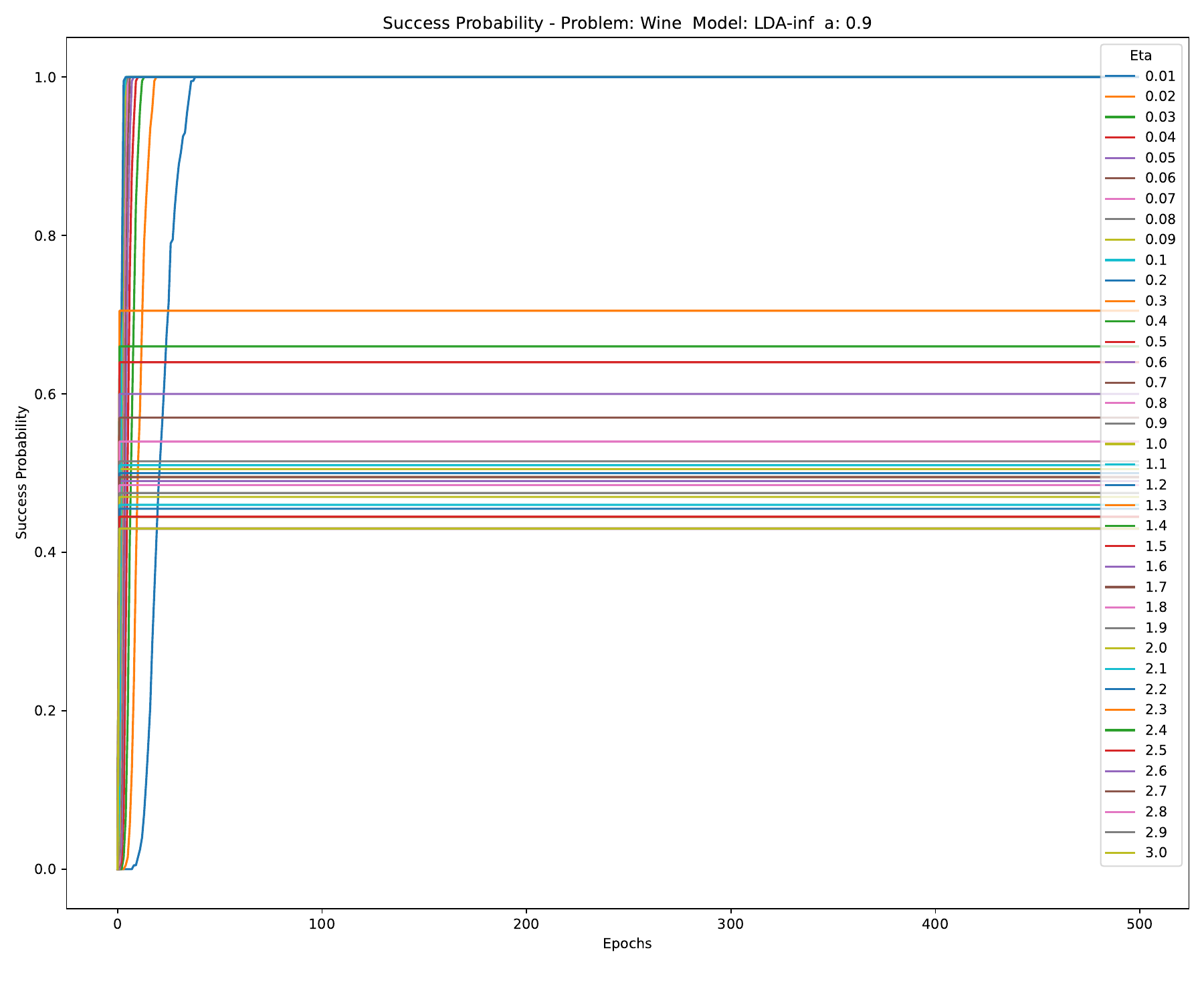} &
    \includegraphics[clip,trim=0 0 0 0,width=.52\linewidth]{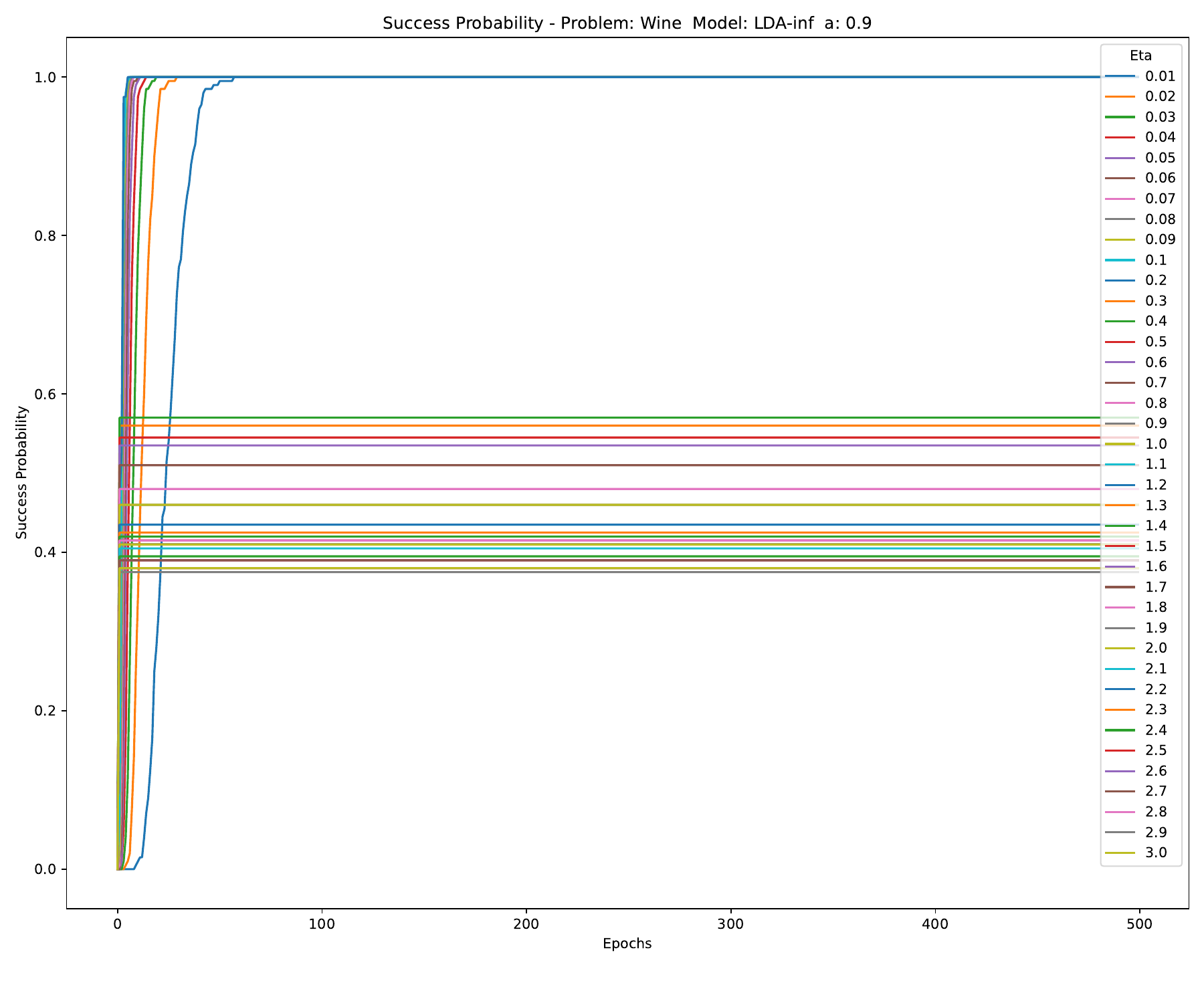} \\[-3mm]
  \end{tabular}
\caption{Success probability ($a=0.9$) for LDA-Inf model (continued).}
\end{figure*}

\begin{figure*}[p]
  \centering
  \begin{tabular}{c@{}c}
    Training Success Probability & Validation Success Probability \\
    \includegraphics[clip,trim=0 0 0 0,width=.52\linewidth]{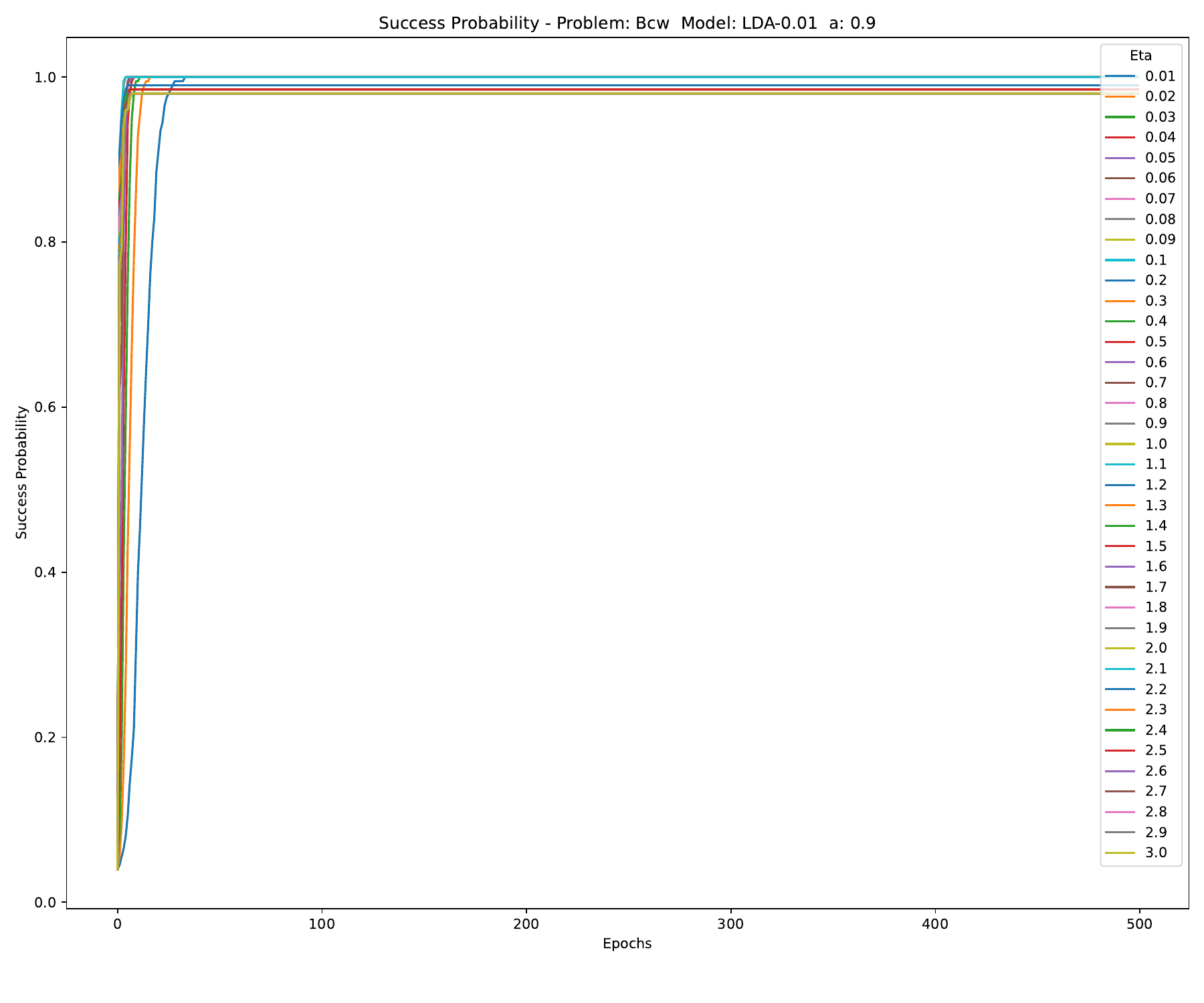} &
    \includegraphics[clip,trim=0 0 0 0,width=.52\linewidth]{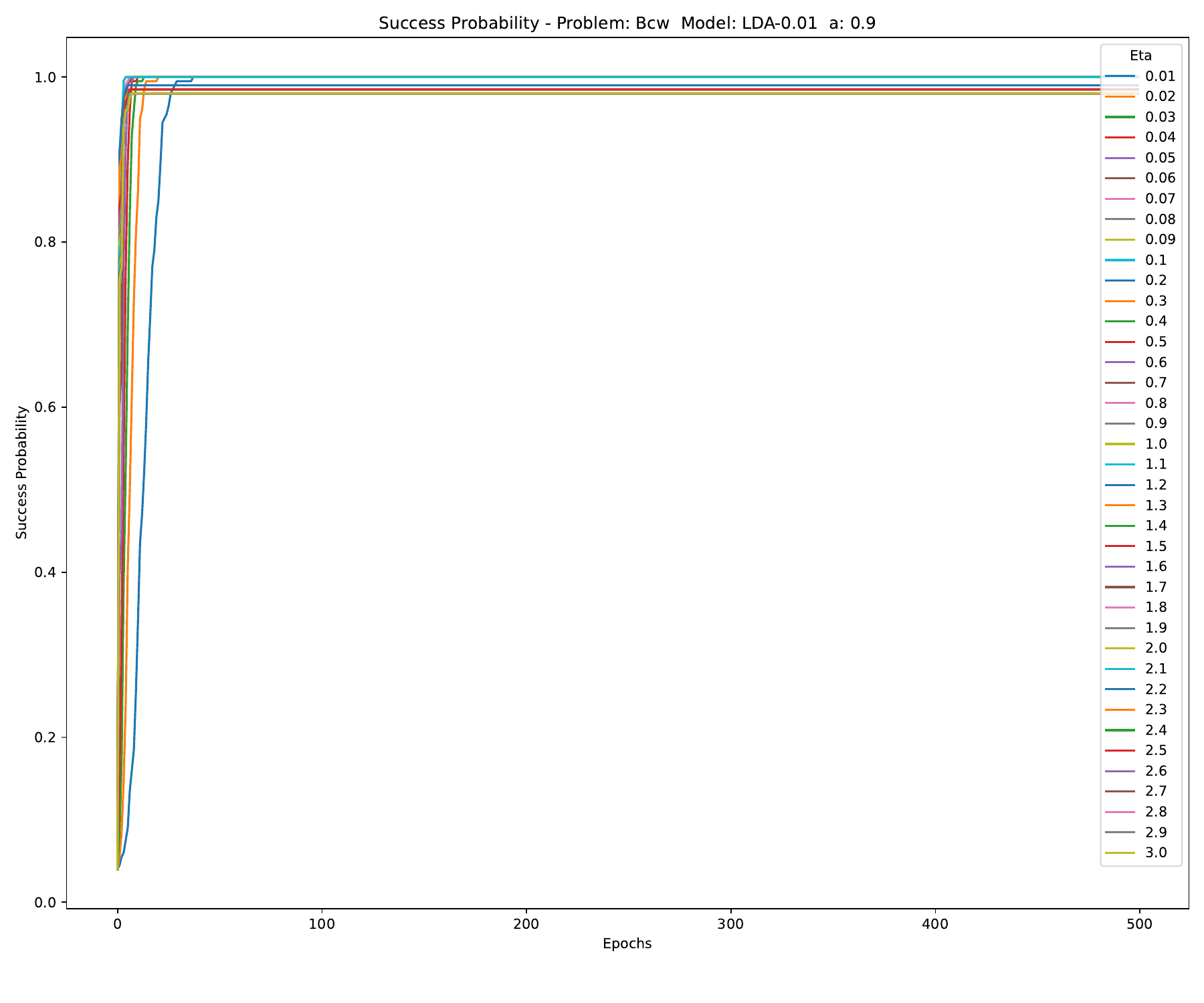} \\[-3mm]
    \includegraphics[clip,trim=0 0 0 0,width=.52\linewidth]{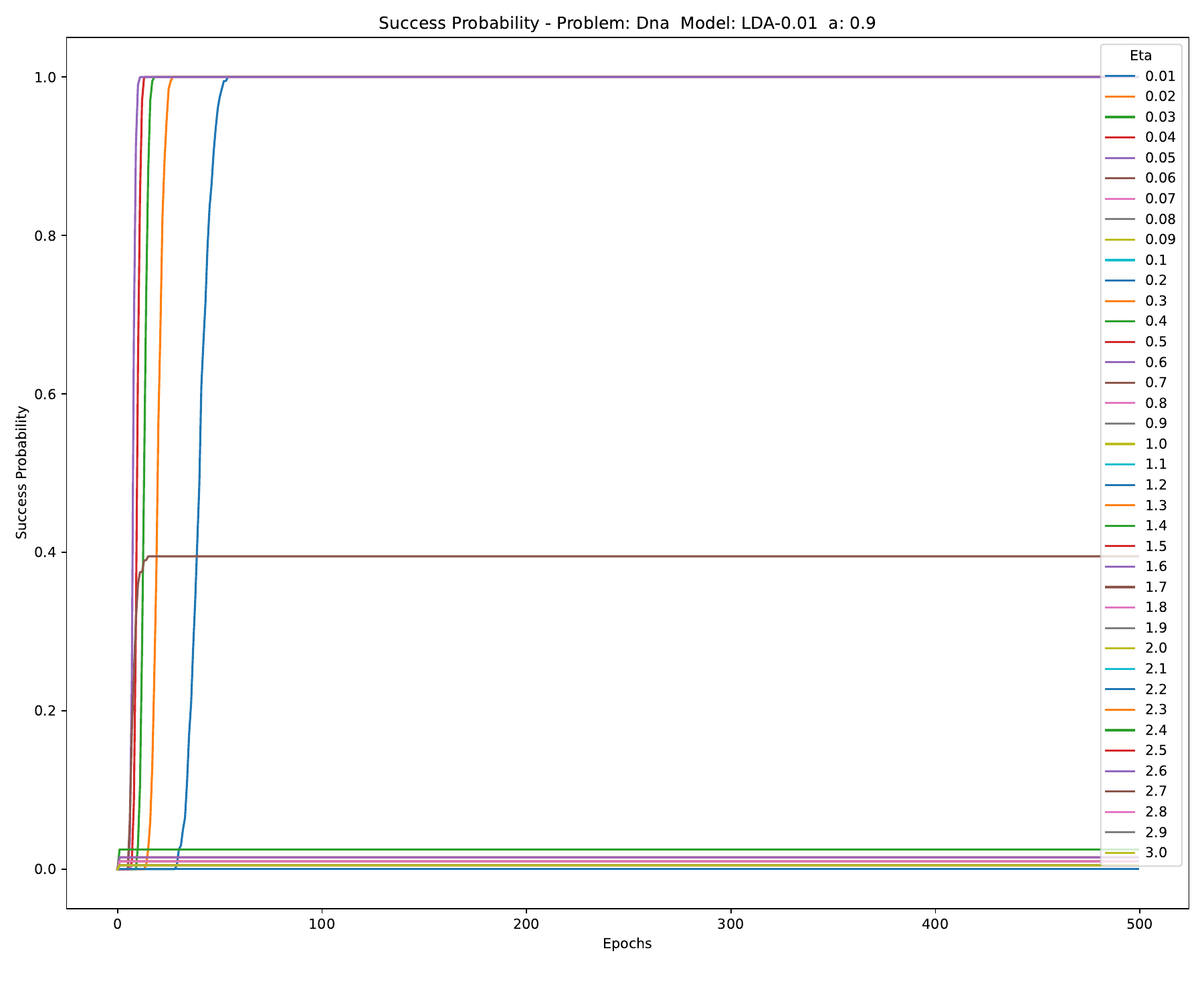} &
    \includegraphics[clip,trim=0 0 0 0,width=.52\linewidth]{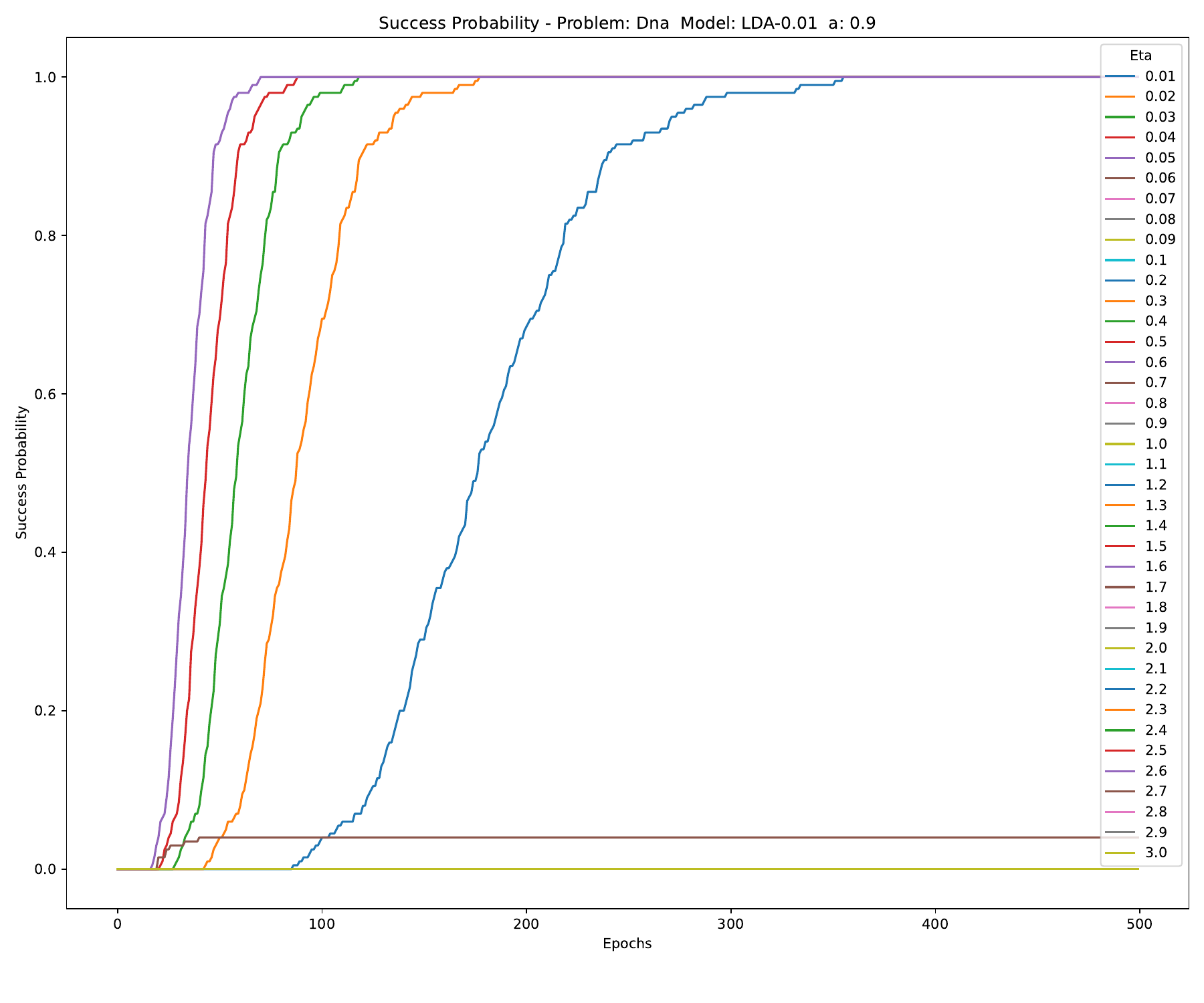} \\[-3mm]
    \includegraphics[clip,trim=0 0 0 0,width=.52\linewidth]{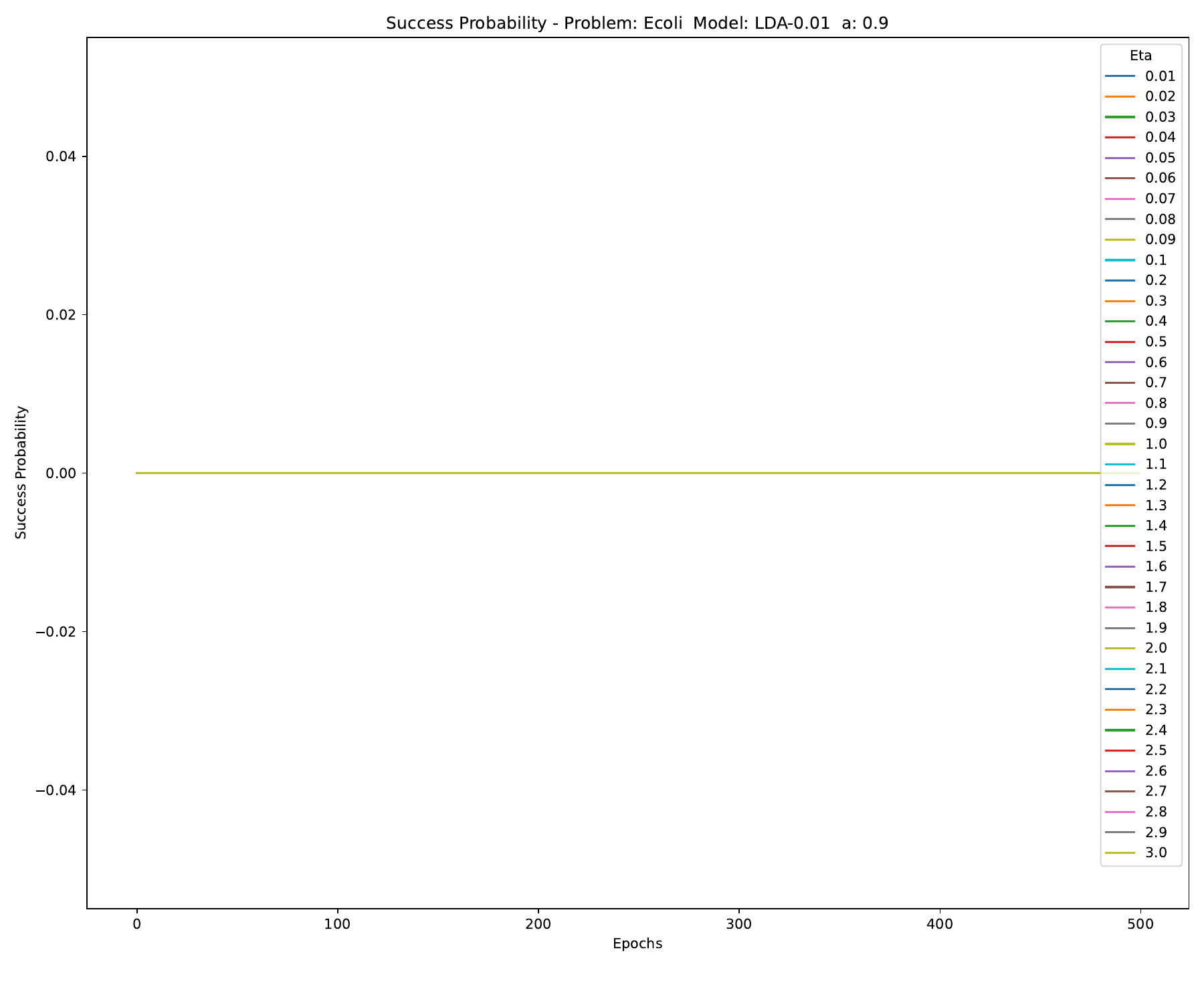} &
    \includegraphics[clip,trim=0 0 0 0,width=.52\linewidth]{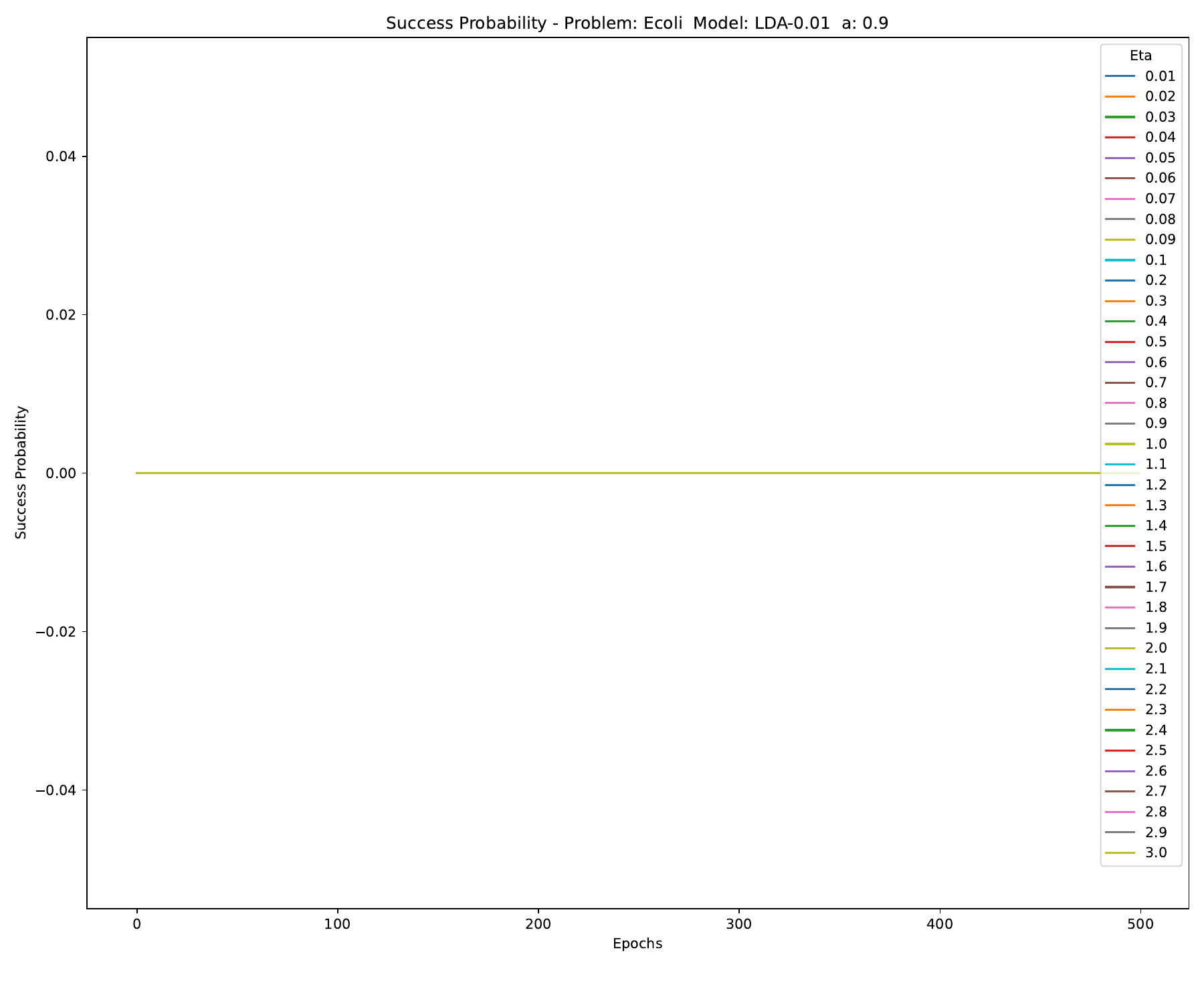} \\[-3mm]
  \end{tabular}
\caption{Success probability ($a=0.9$) for LDA-0.01 model.}
  \label{fig:success_probability_LDA_0.01_model_0.9}
\end{figure*}

\begin{figure*}[p]
  \centering
  \ContinuedFloat
  \begin{tabular}{c@{}c}
    Training Success Probability & Validation Success Probability \\
    \includegraphics[clip,trim=0 0 0 0,width=.52\linewidth]{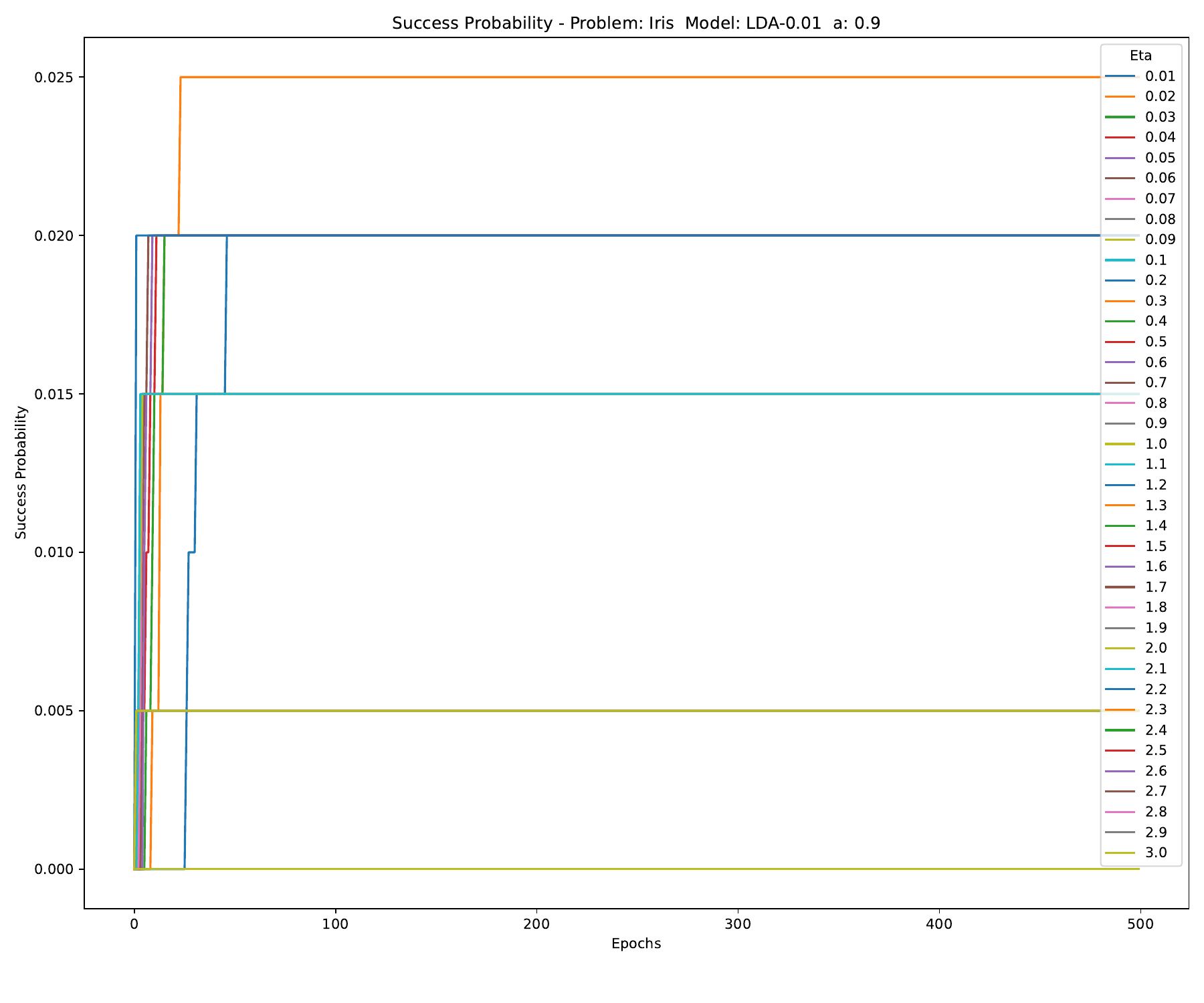} &
    \includegraphics[clip,trim=0 0 0 0,width=.52\linewidth]{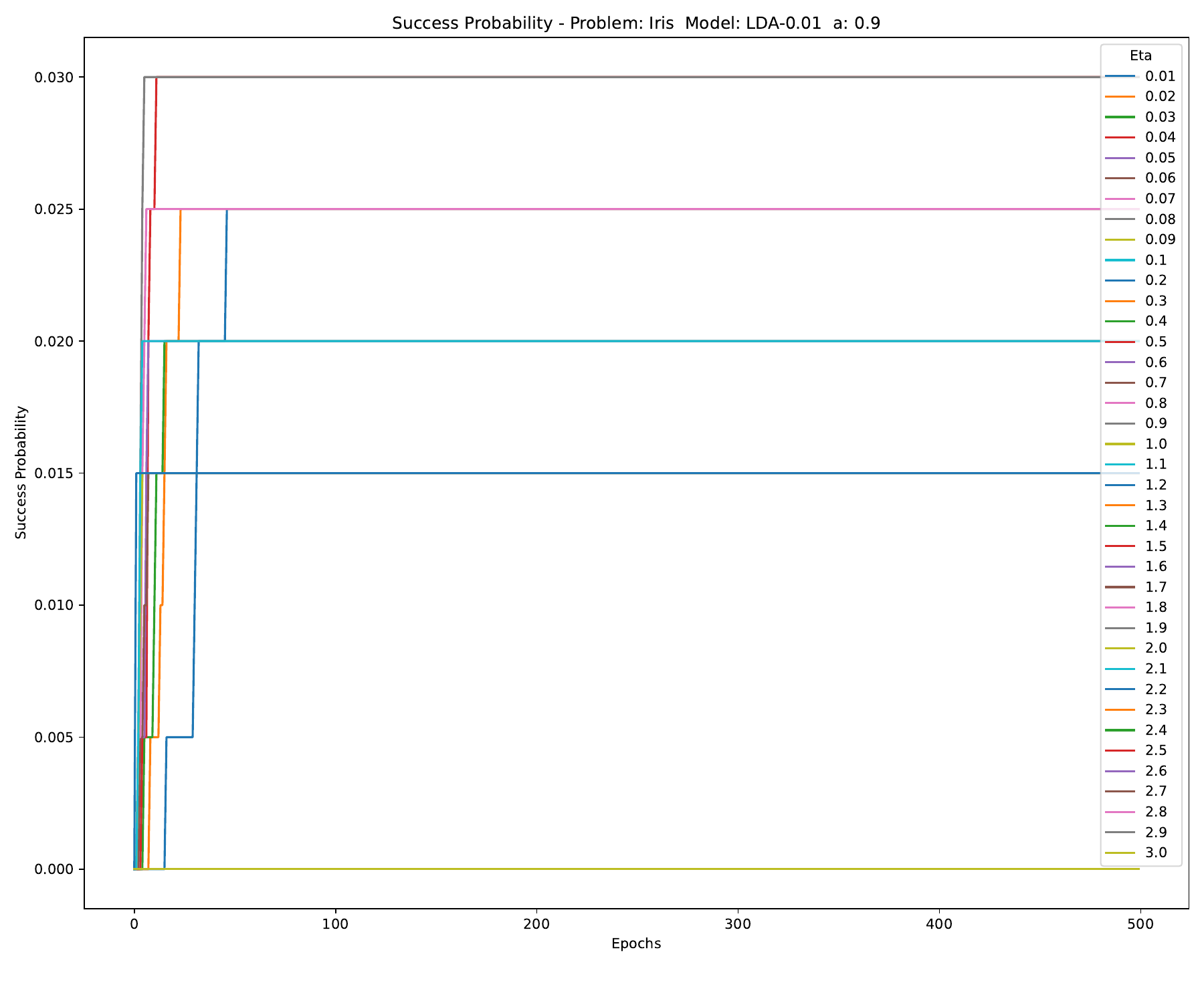} \\[-3mm]
    \includegraphics[clip,trim=0 0 0 0,width=.52\linewidth]{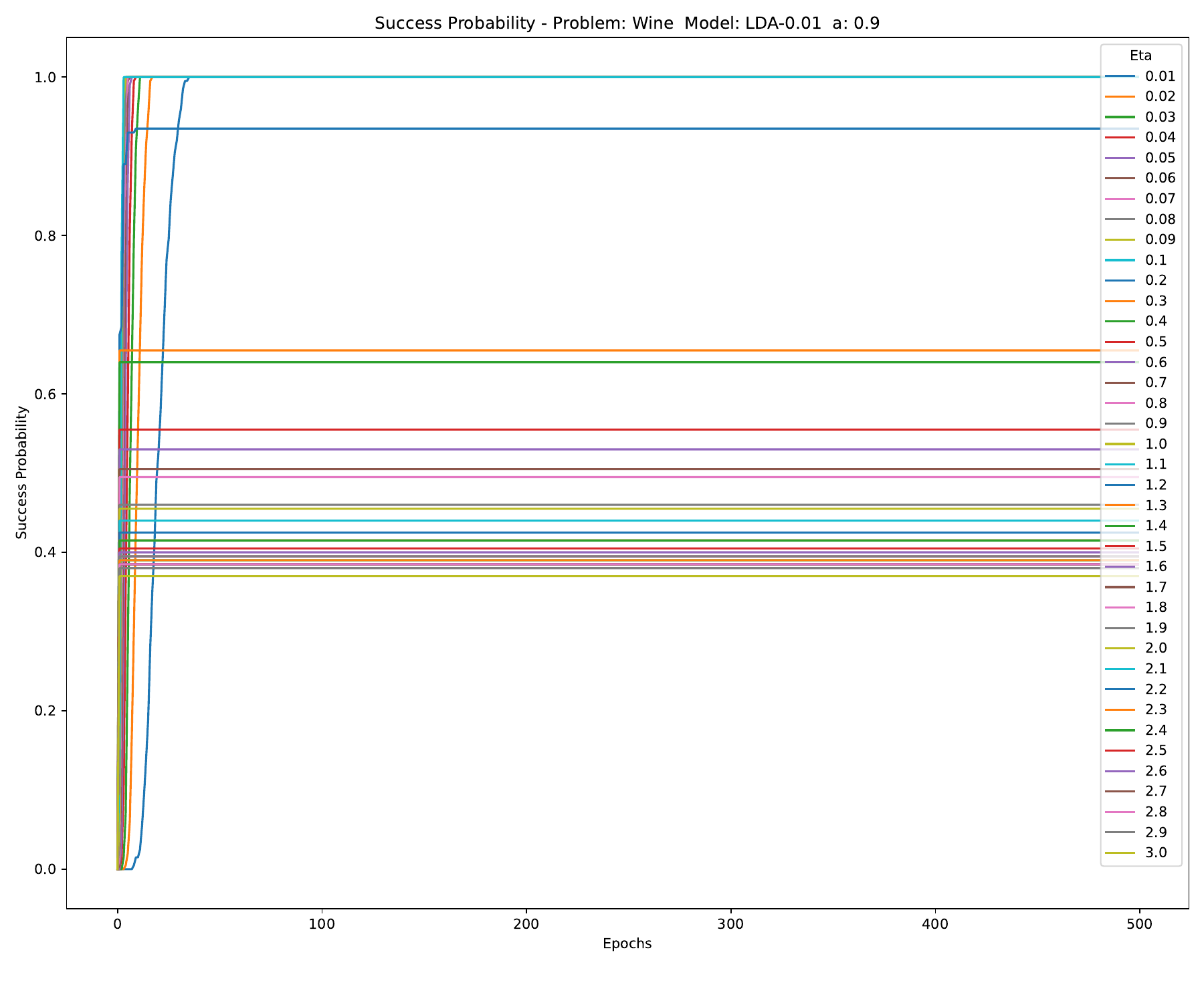} &
    \includegraphics[clip,trim=0 0 0 0,width=.52\linewidth]{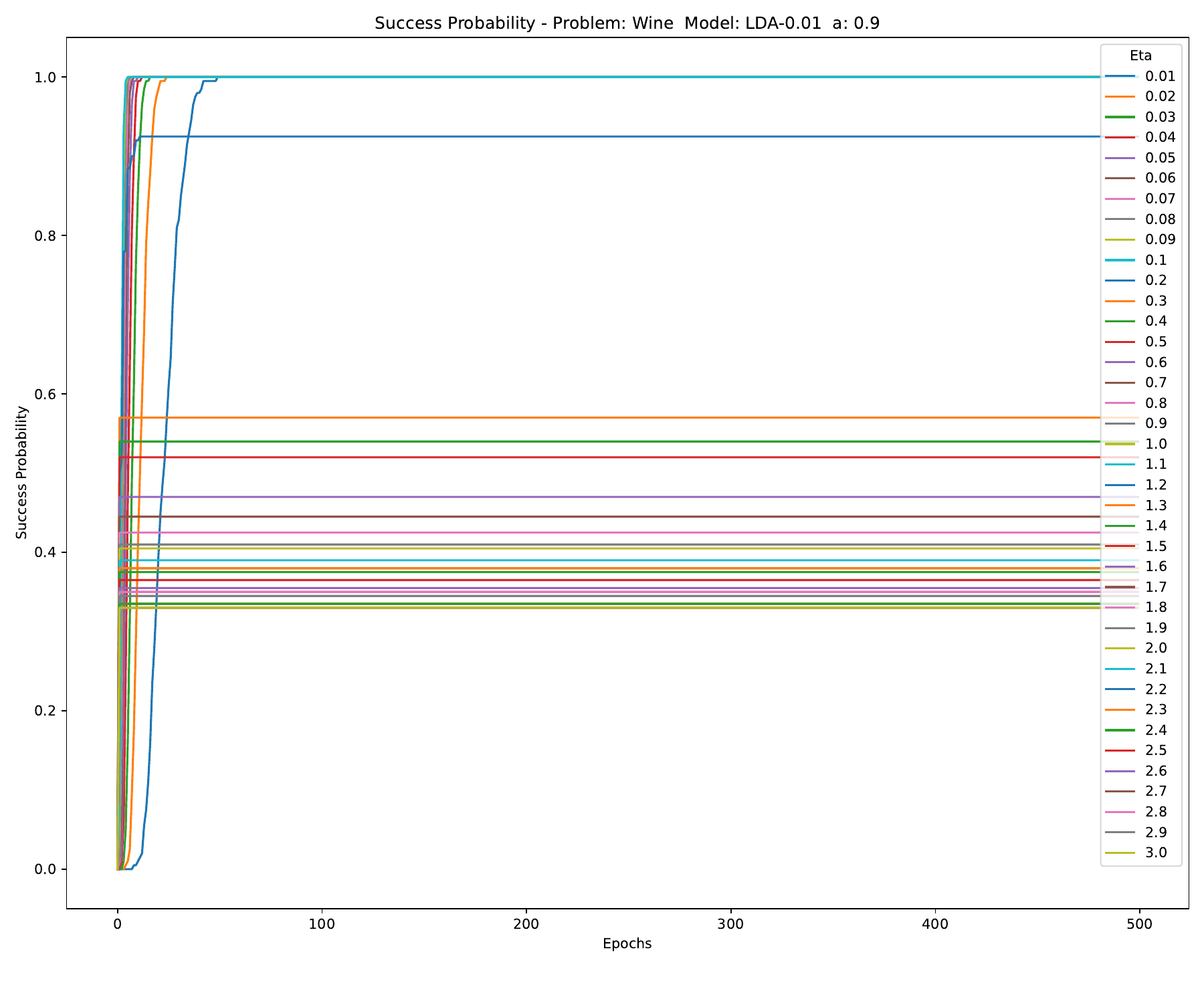} \\[-3mm]
  \end{tabular}
\caption{Success probability ($a=0.9$) for LDA-0.01 model (continued).}
\end{figure*}

\begin{figure*}[p]
  \centering
  \begin{tabular}{c@{}c}
    Training Success Probability & Validation Success Probability \\
    \includegraphics[clip,trim=0 0 0 0,width=.52\linewidth]{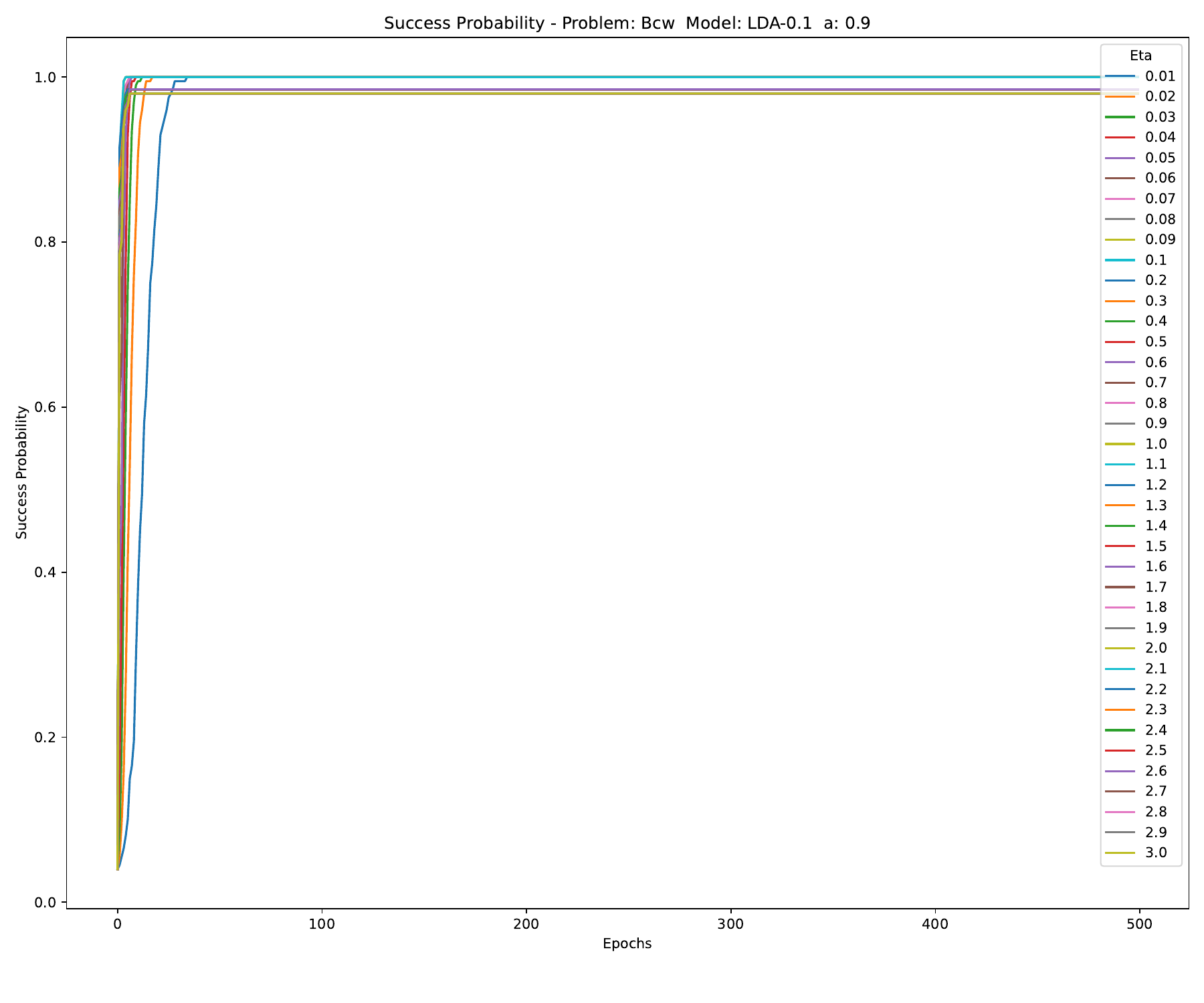} &
    \includegraphics[clip,trim=0 0 0 0,width=.52\linewidth]{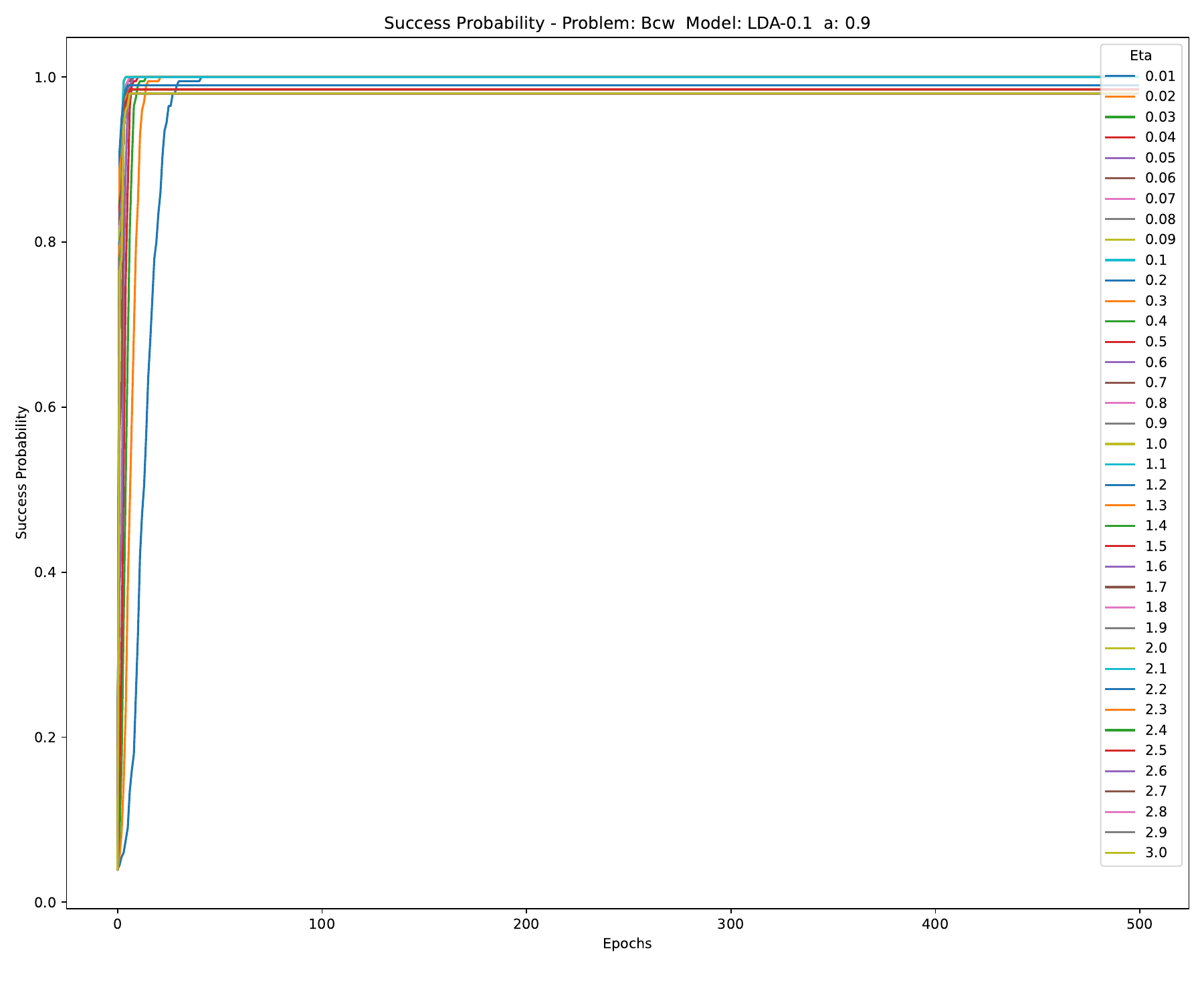} \\[-3mm]
    \includegraphics[clip,trim=0 0 0 0,width=.52\linewidth]{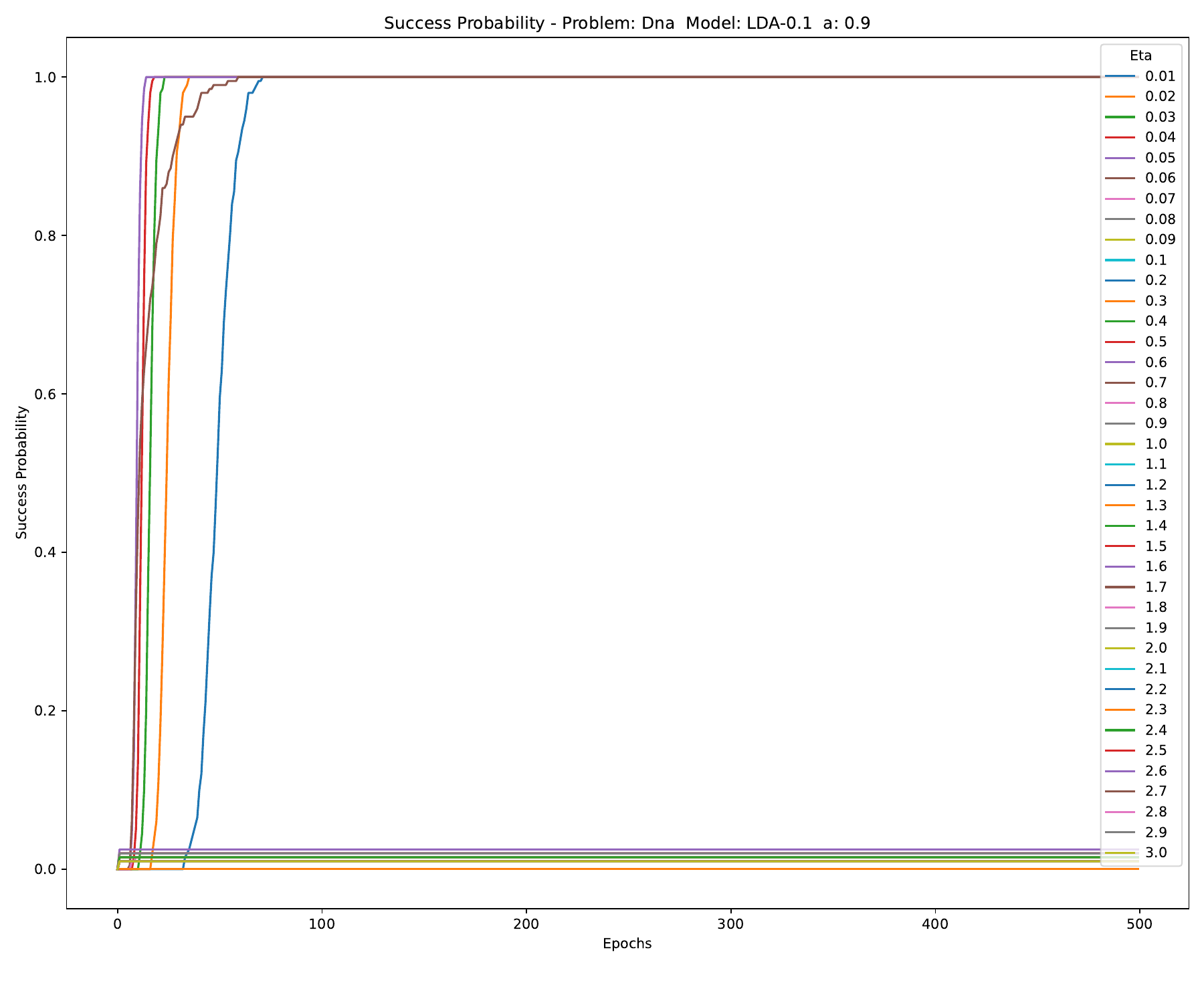} &
    \includegraphics[clip,trim=0 0 0 0,width=.52\linewidth]{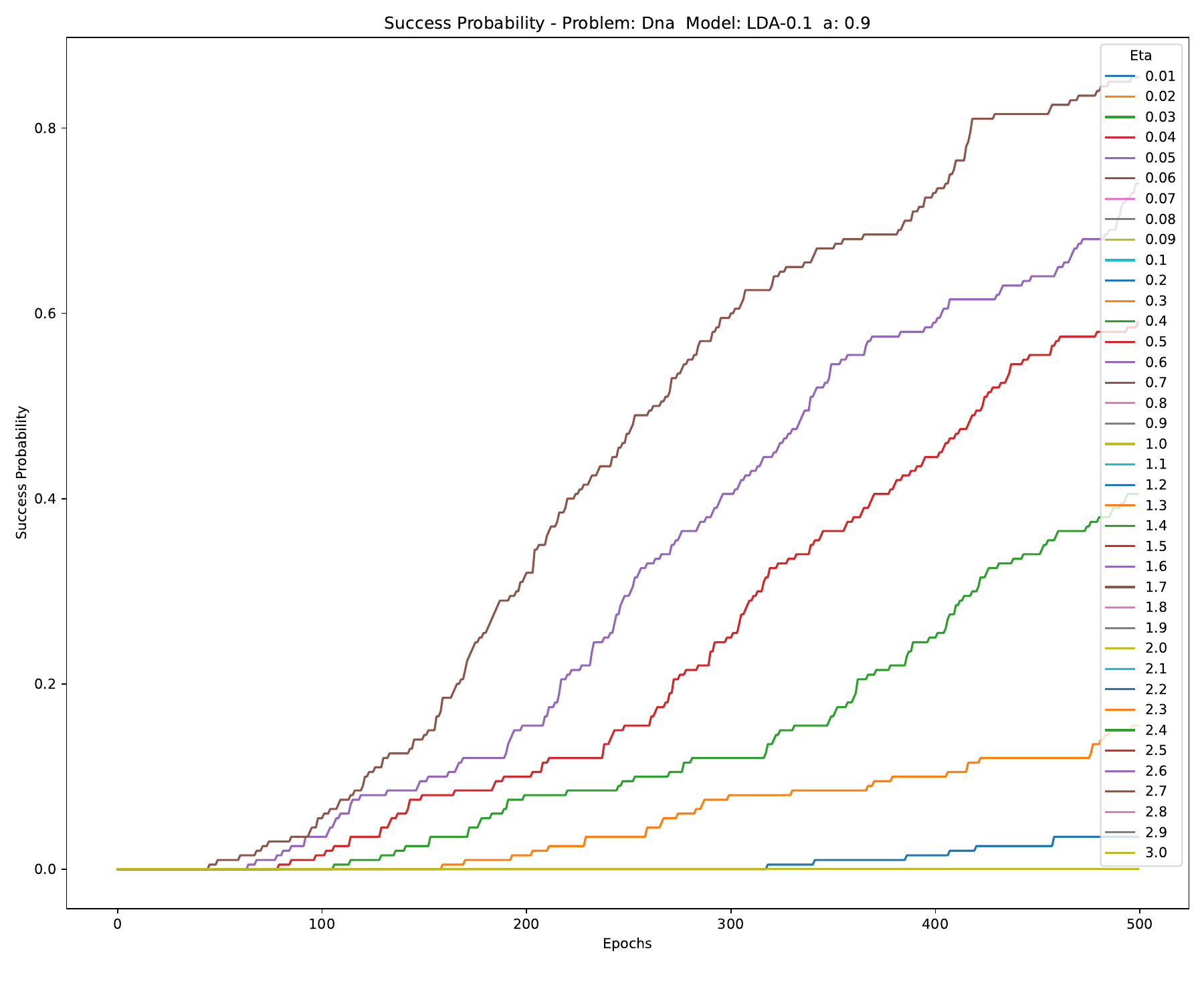} \\[-3mm]
    \includegraphics[clip,trim=0 0 0 0,width=.52\linewidth]{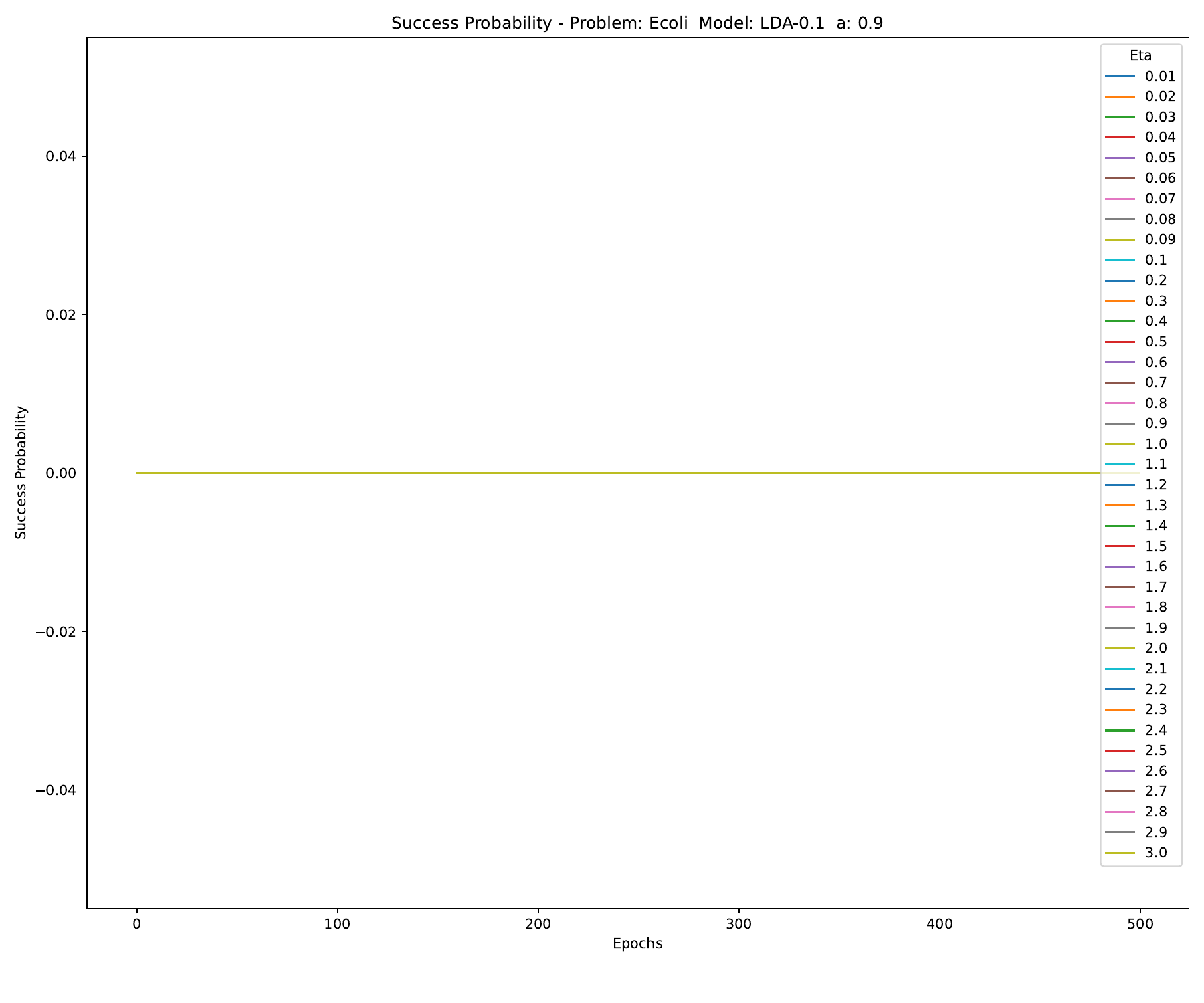} &
    \includegraphics[clip,trim=0 0 0 0,width=.52\linewidth]{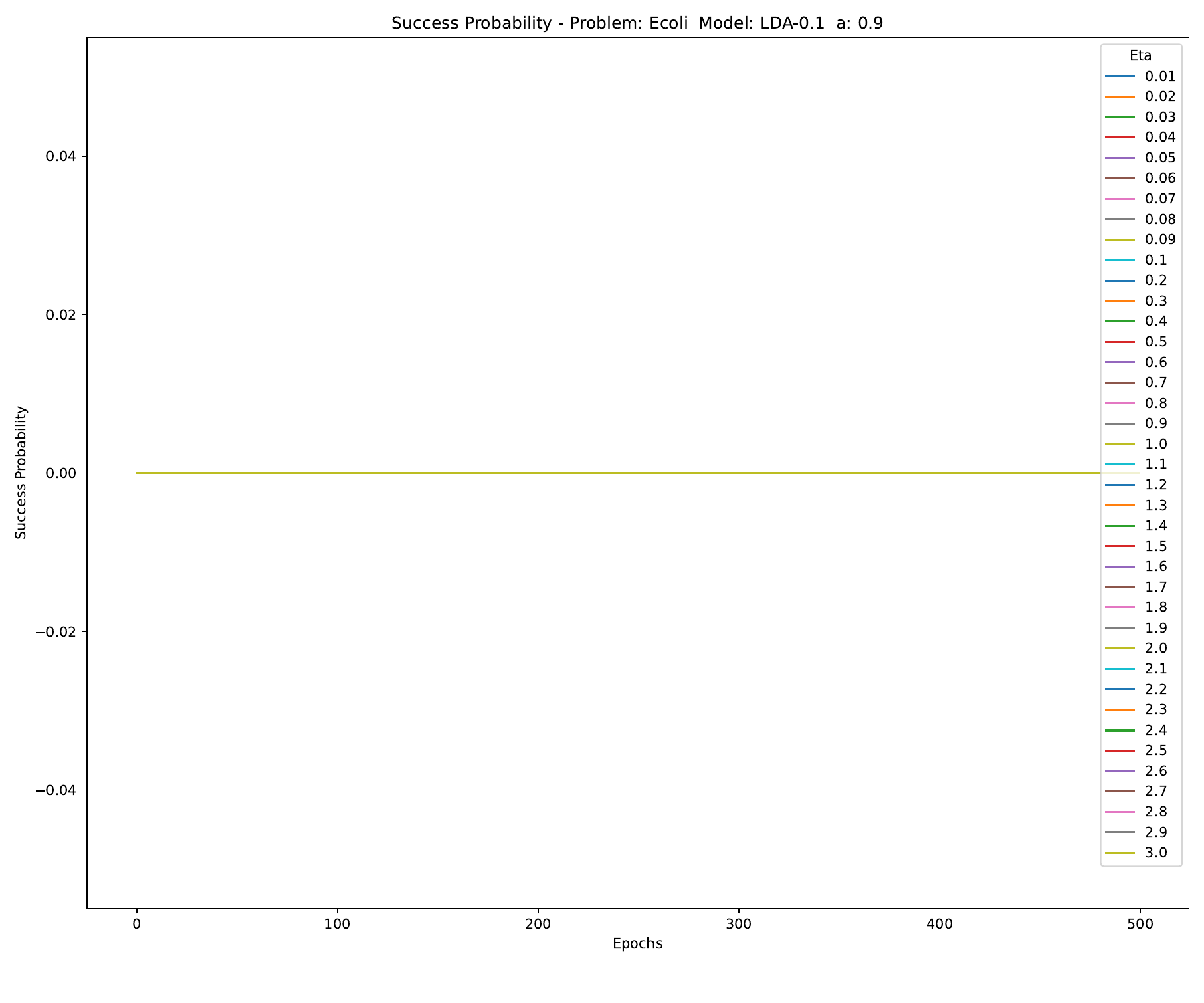} \\[-3mm]
  \end{tabular}
\caption{Success probability ($a=0.9$) for LDA-0.1 model.}
  \label{fig:success_probability_LDA_0.1_model_0.9}
\end{figure*}

\begin{figure*}[p]
  \centering
  \ContinuedFloat
  \begin{tabular}{c@{}c}
    Training Success Probability & Validation Success Probability \\
    \includegraphics[clip,trim=0 0 0 0,width=.52\linewidth]{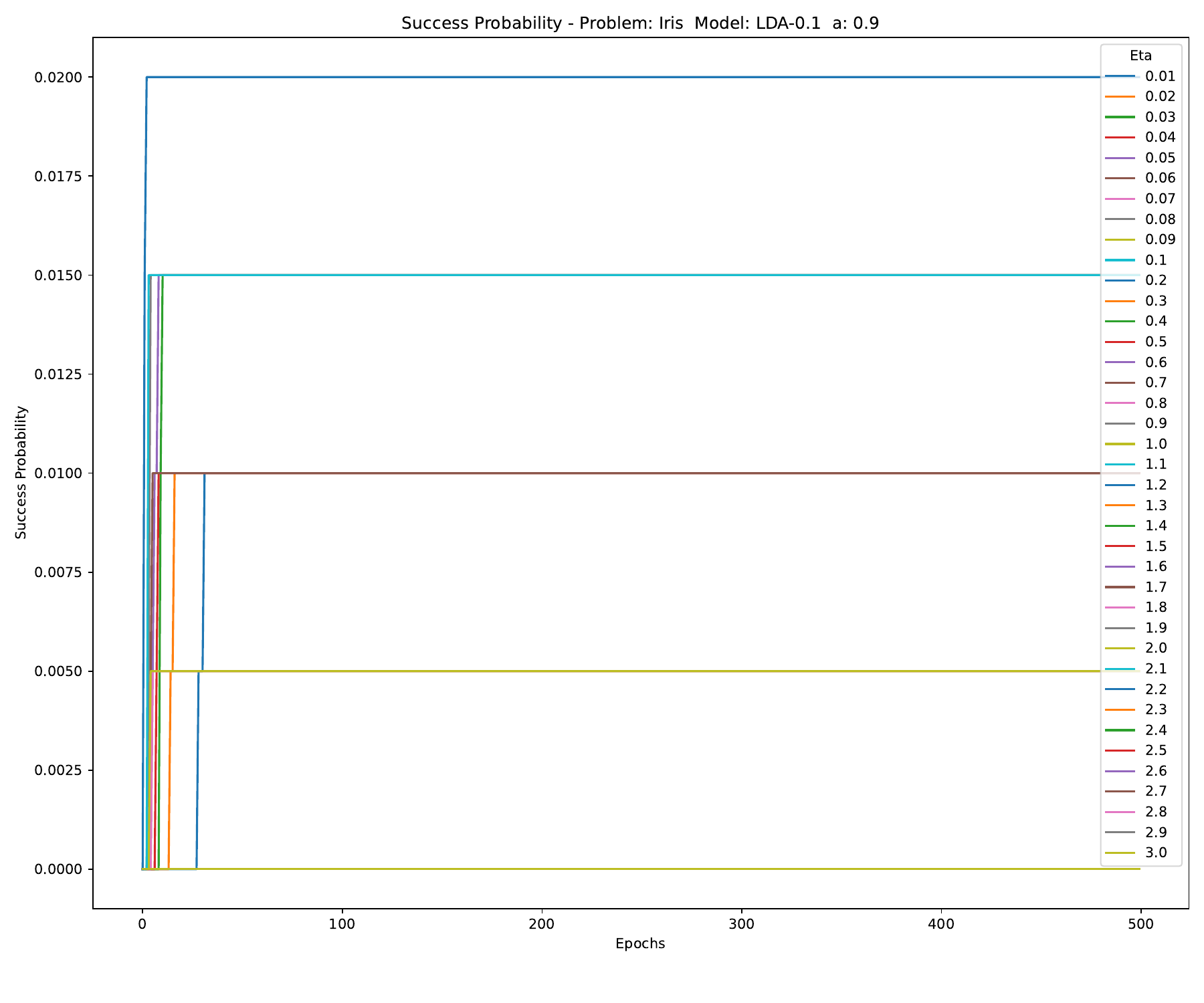} &
    \includegraphics[clip,trim=0 0 0 0,width=.52\linewidth]{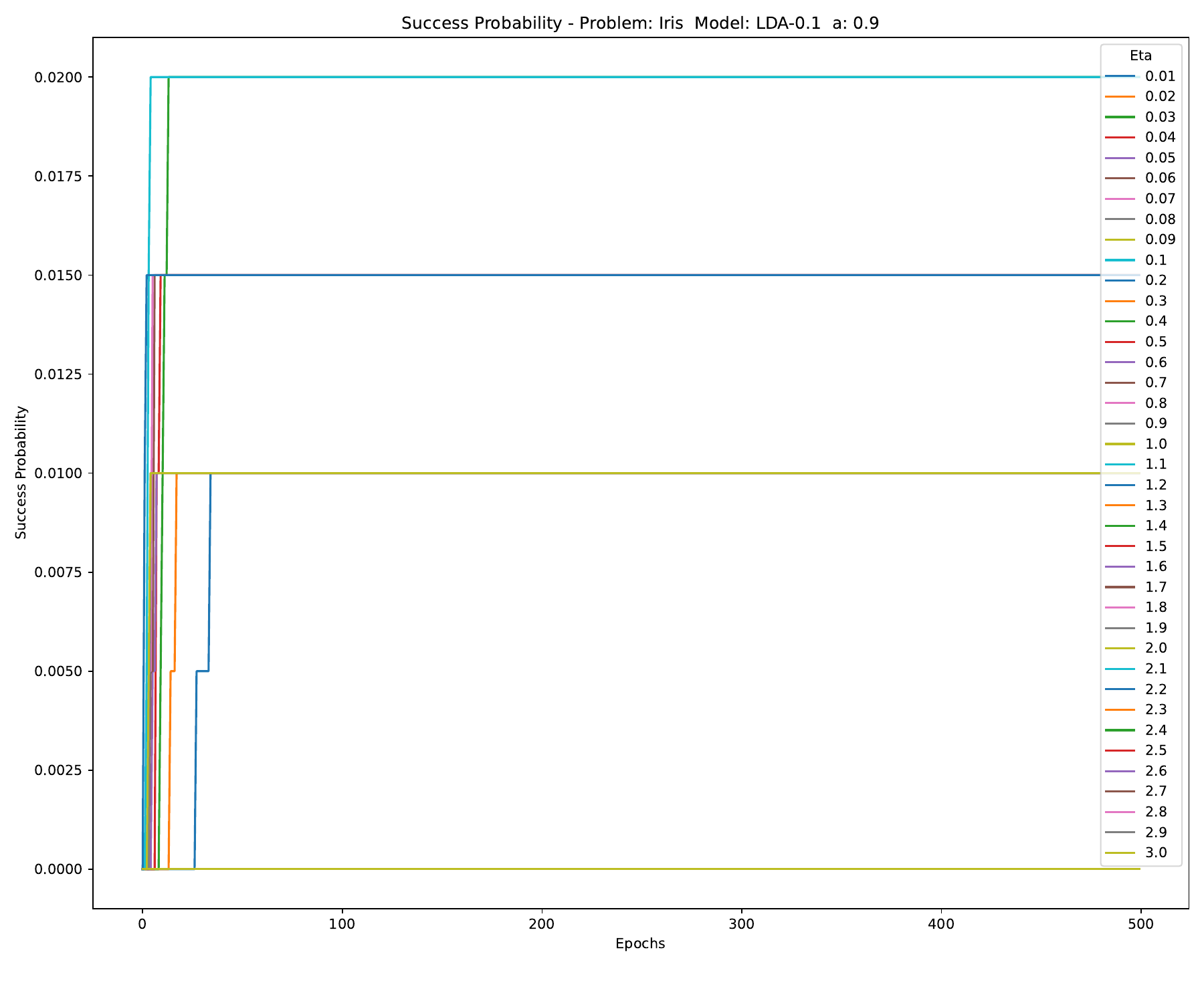} \\[-3mm]
    \includegraphics[clip,trim=0 0 0 0,width=.52\linewidth]{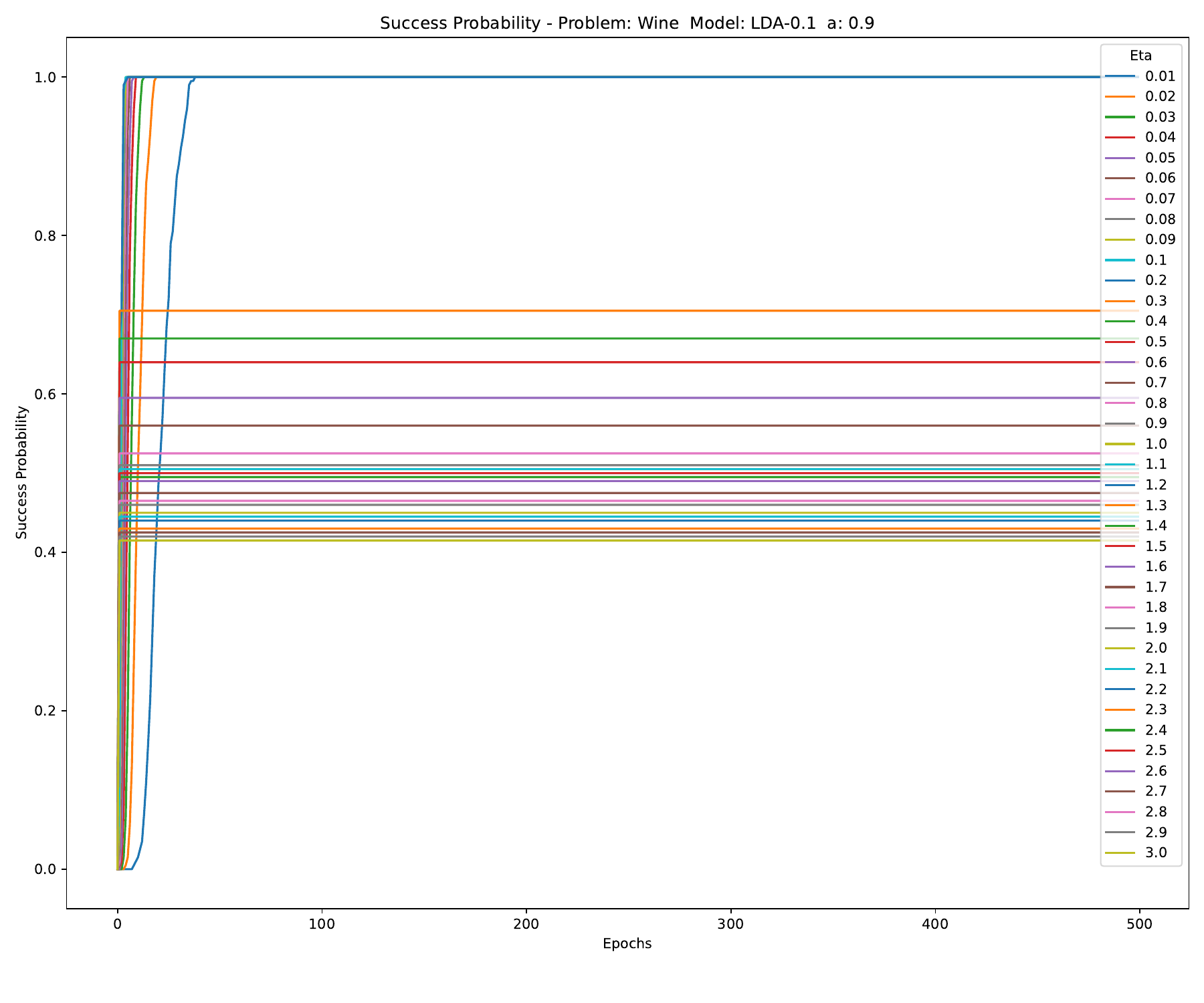} &
    \includegraphics[clip,trim=0 0 0 0,width=.52\linewidth]{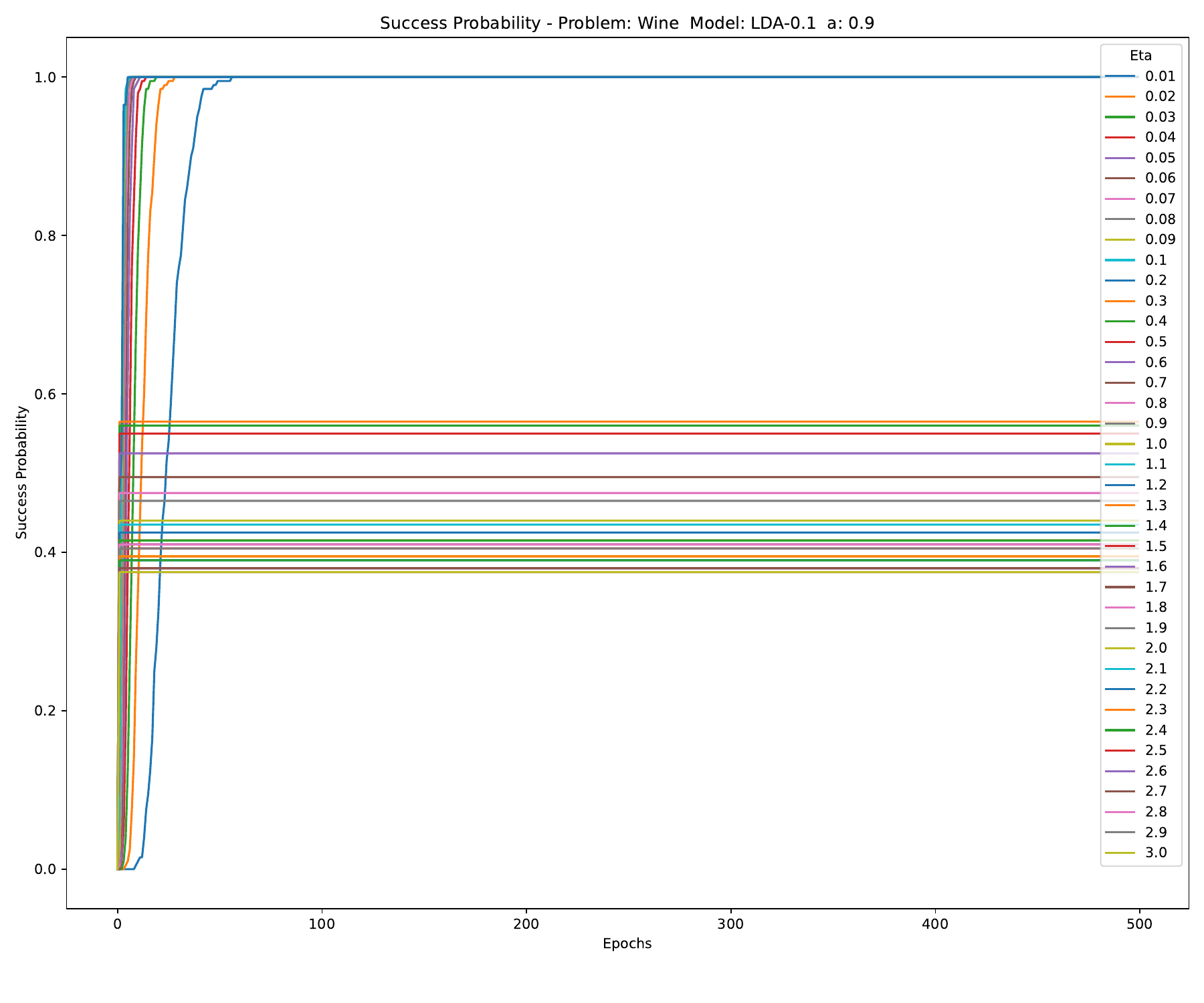} \\[-3mm]
  \end{tabular}
\caption{Success probability ($a=0.9$) for LDA-0.1 model (continued).}
\end{figure*}

\begin{figure*}[p]
  \centering
  \begin{tabular}{c@{}c}
    Training Success Probability & Validation Success Probability \\
    \includegraphics[clip,trim=0 0 0 0,width=.52\linewidth]{arXiv-figures/bcw_LR-Inf_a=0.9_success_prob_train_acc_thr.pdf} &
    \includegraphics[clip,trim=0 0 0 0,width=.52\linewidth]{arXiv-figures/bcw_LR-Inf_a=0.9_success_prob_test_acc_thr.pdf} \\[-3mm]
    \includegraphics[clip,trim=0 0 0 0,width=.52\linewidth]{arXiv-figures/dna_LR-Inf_a=0.9_success_prob_train_acc_thr.pdf} &
    \includegraphics[clip,trim=0 0 0 0,width=.52\linewidth]{arXiv-figures/dna_LR-Inf_a=0.9_success_prob_test_acc_thr.pdf} \\[-3mm]
    \includegraphics[clip,trim=0 0 0 0,width=.52\linewidth]{arXiv-figures/ecoli_LR-Inf_a=0.9_success_prob_train_acc_thr.pdf} &
    \includegraphics[clip,trim=0 0 0 0,width=.52\linewidth]{arXiv-figures/ecoli_LR-Inf_a=0.9_success_prob_test_acc_thr.pdf} \\[-3mm]
  \end{tabular}
\caption{Success probability ($a=0.9$) for LR-Inf model.}
  \label{fig:success_probability_LR_Inf_model_0.9_duplicate}
\end{figure*}

\begin{figure*}[p]
  \centering
  \ContinuedFloat
  \begin{tabular}{c@{}c}
    Training Success Probability & Validation Success Probability \\
    \includegraphics[clip,trim=0 0 0 0,width=.52\linewidth]{arXiv-figures/iris_LR-Inf_a=0.9_success_prob_train_acc_thr.pdf} &
    \includegraphics[clip,trim=0 0 0 0,width=.52\linewidth]{arXiv-figures/iris_LR-Inf_a=0.9_success_prob_test_acc_thr.pdf} \\[-3mm]
    \includegraphics[clip,trim=0 0 0 0,width=.52\linewidth]{arXiv-figures/wine_LR-Inf_a=0.9_success_prob_train_acc_thr.pdf} &
    \includegraphics[clip,trim=0 0 0 0,width=.52\linewidth]{arXiv-figures/wine_LR-Inf_a=0.9_success_prob_test_acc_thr.pdf} \\[-3mm]
  \end{tabular}
\caption{Success probability ($a=0.9$) for LR-Inf model (continued).}
\end{figure*}

\begin{figure*}[p]
  \centering
  \begin{tabular}{c@{}c}
    Training Success Probability & Validation Success Probability \\
    \includegraphics[clip,trim=0 0 0 0,width=.52\linewidth]{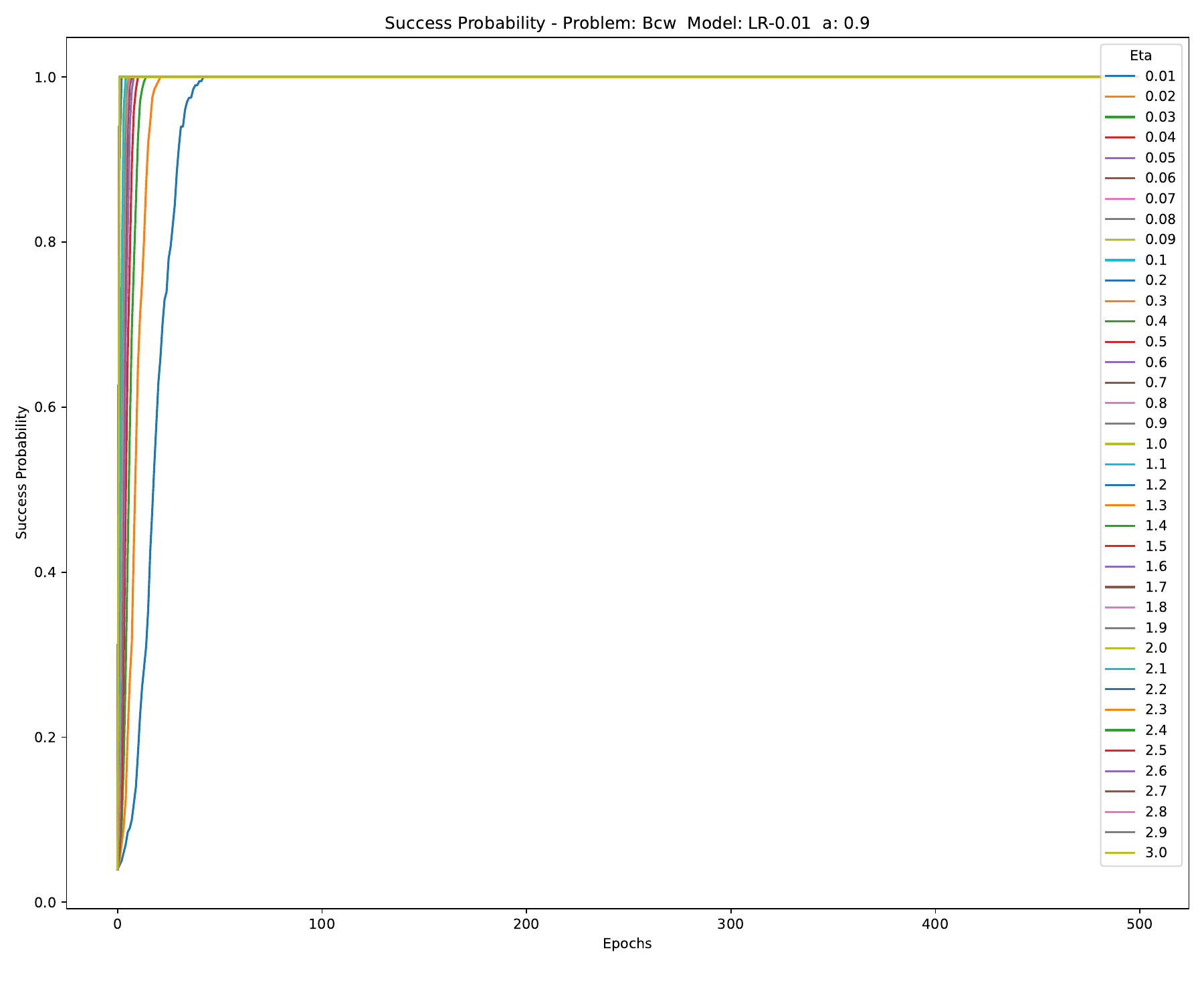} &
    \includegraphics[clip,trim=0 0 0 0,width=.52\linewidth]{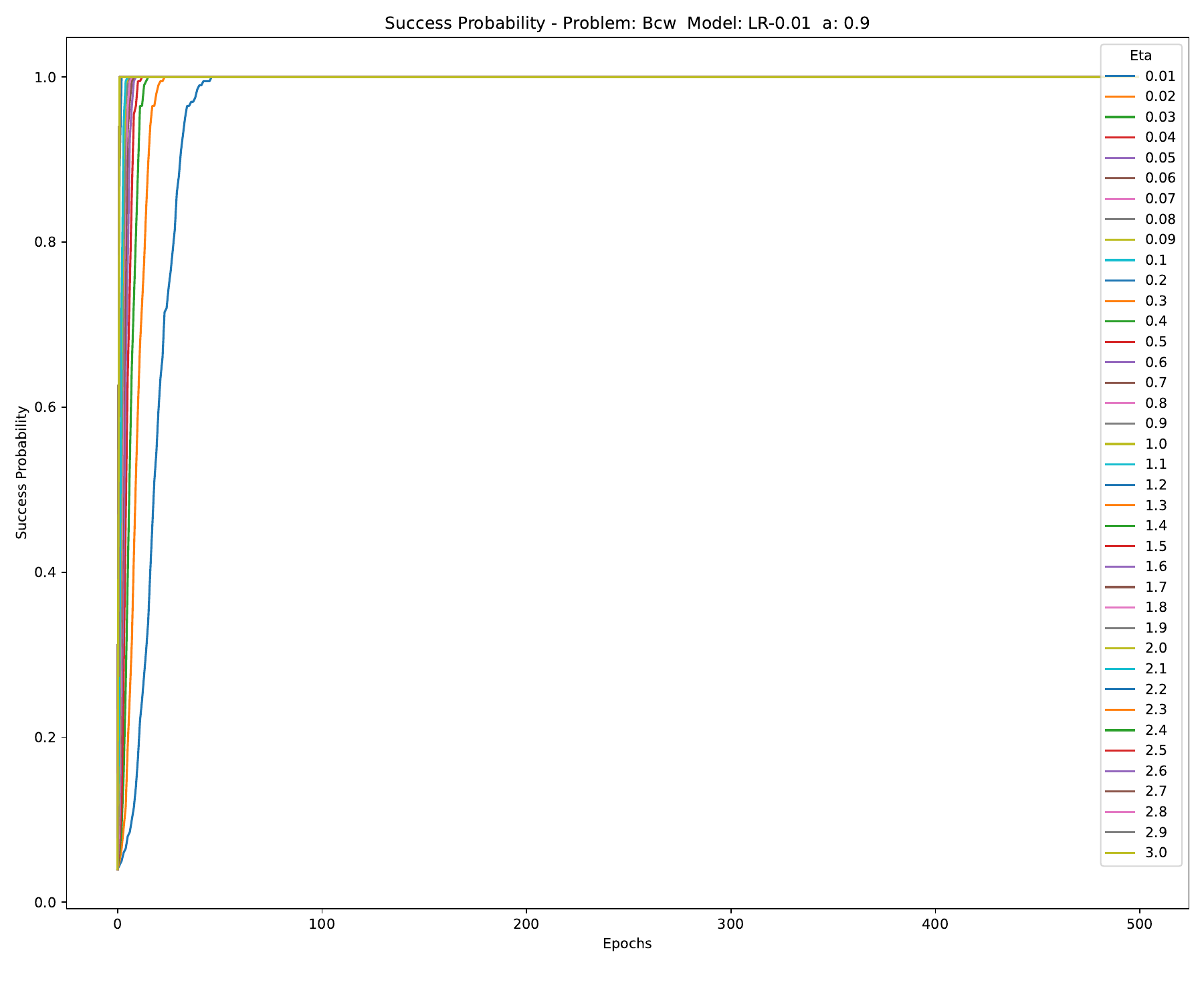} \\[-3mm]
    \includegraphics[clip,trim=0 0 0 0,width=.52\linewidth]{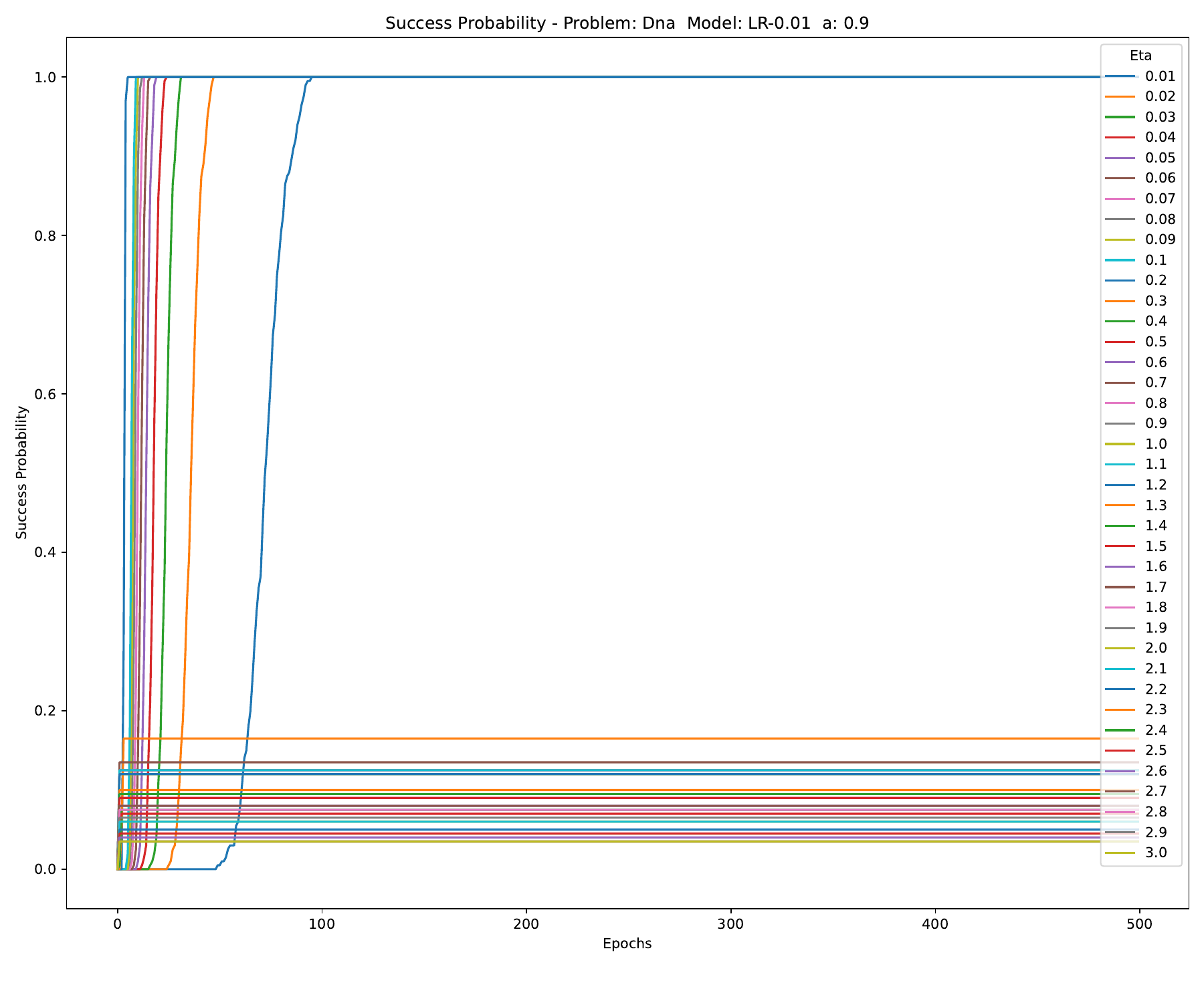} &
    \includegraphics[clip,trim=0 0 0 0,width=.52\linewidth]{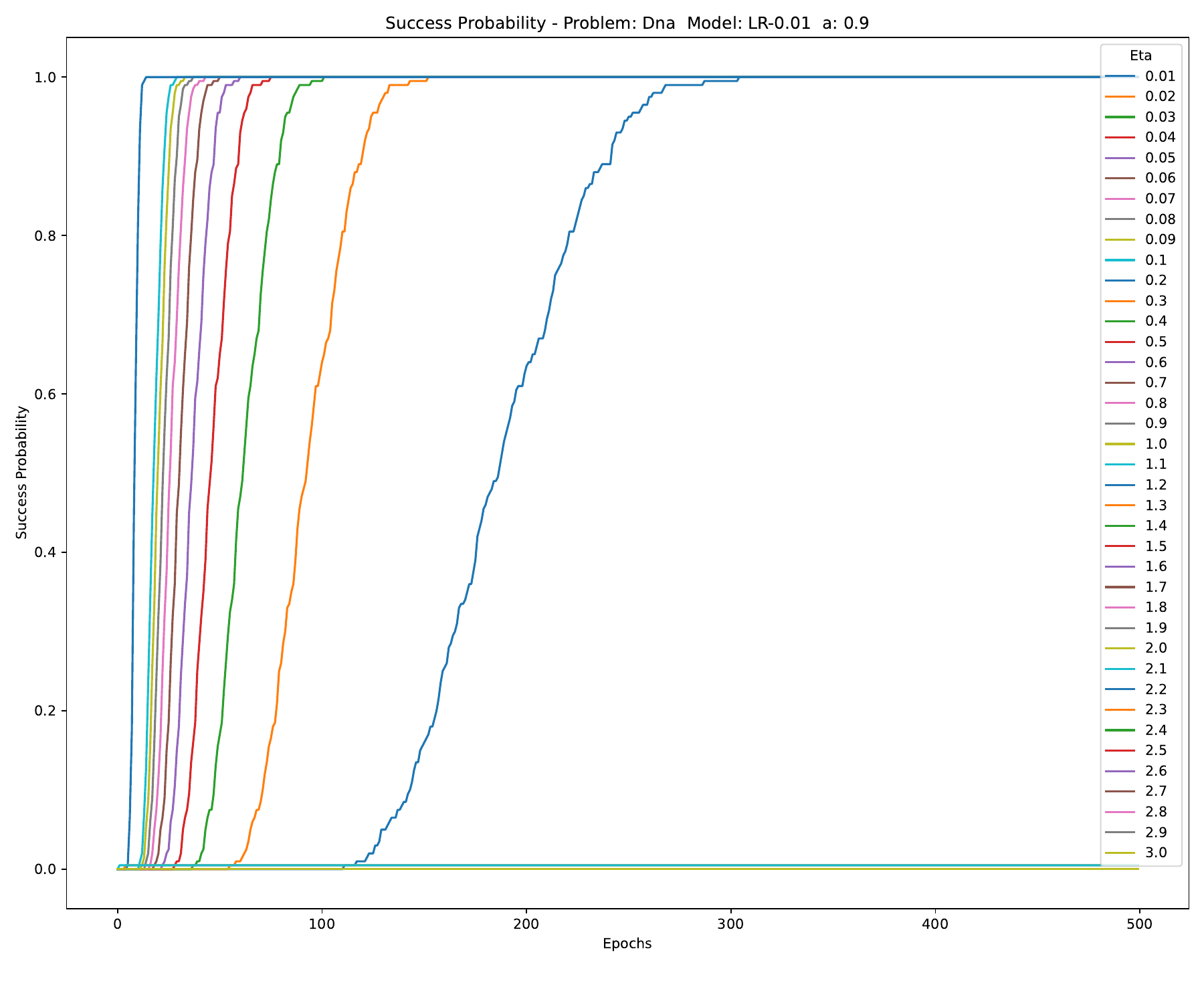} \\[-3mm]
    \includegraphics[clip,trim=0 0 0 0,width=.52\linewidth]{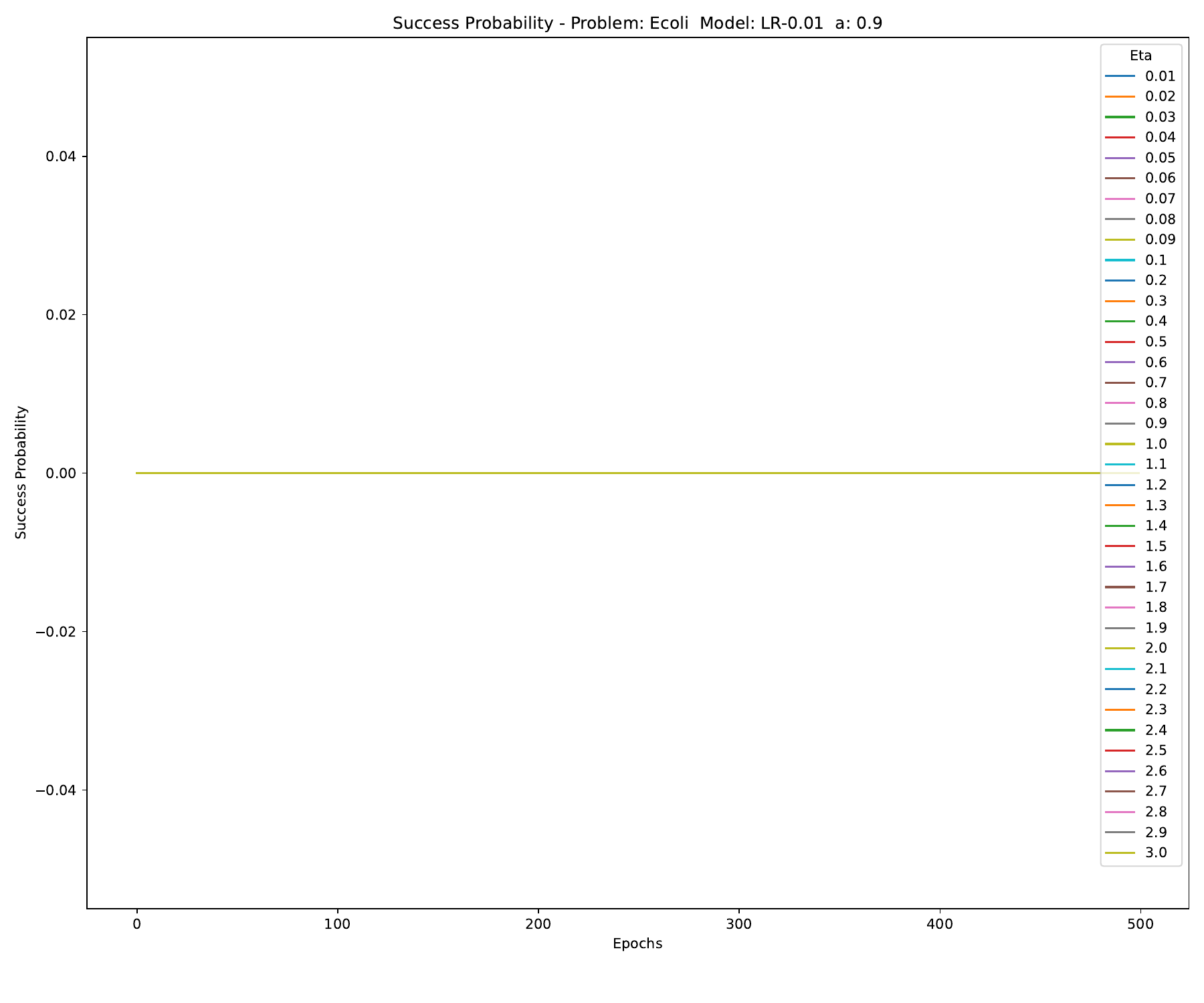} &
    \includegraphics[clip,trim=0 0 0 0,width=.52\linewidth]{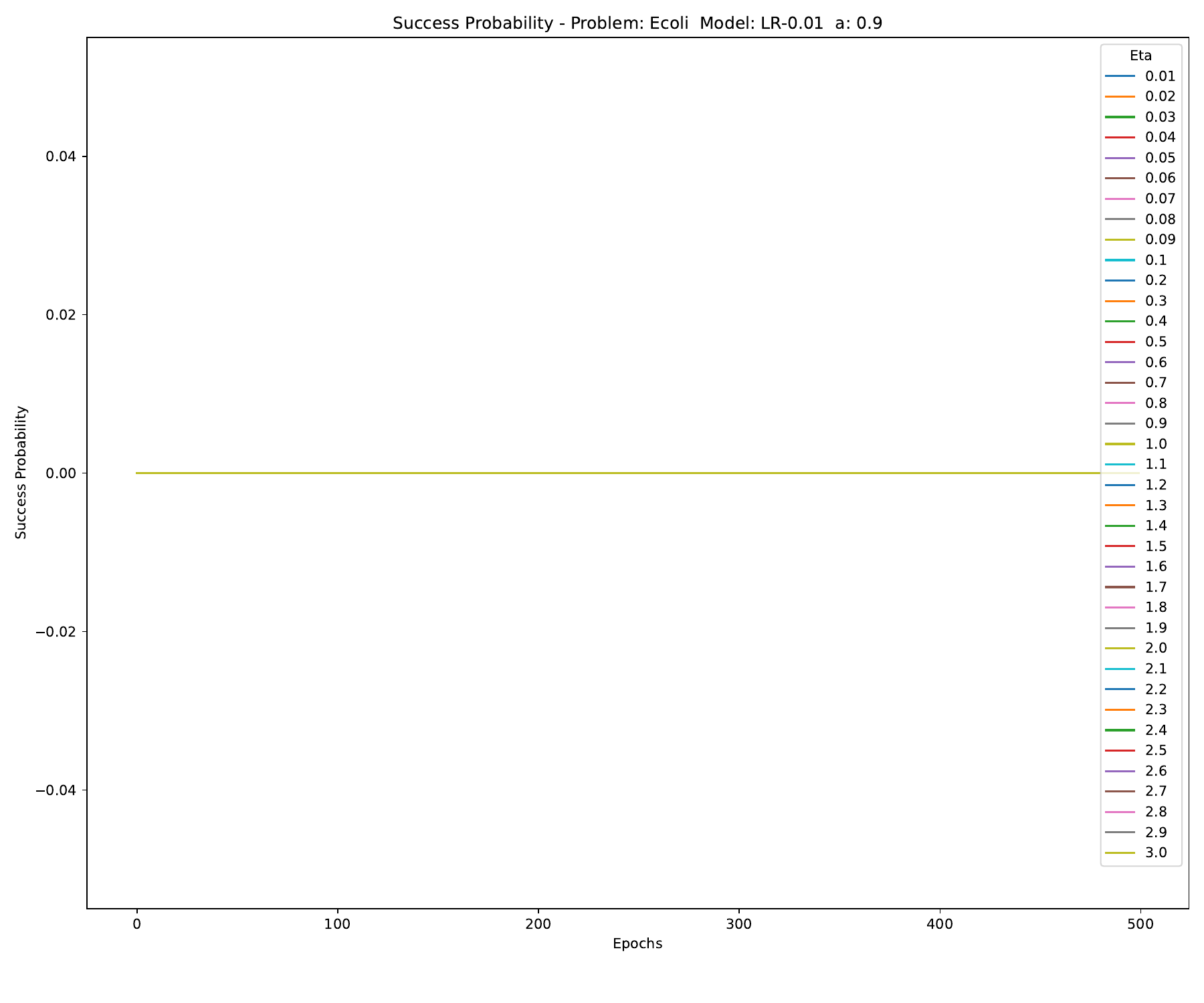} \\[-3mm]
  \end{tabular}
\caption{Success probability ($a=0.9$) for LR-0.01 model.}
  \label{fig:success_probability_LR_0.01_model_0.9}
\end{figure*}

\begin{figure*}[p]
  \centering
  \ContinuedFloat
  \begin{tabular}{c@{}c}
    Training Success Probability & Validation Success Probability \\
    \includegraphics[clip,trim=0 0 0 0,width=.52\linewidth]{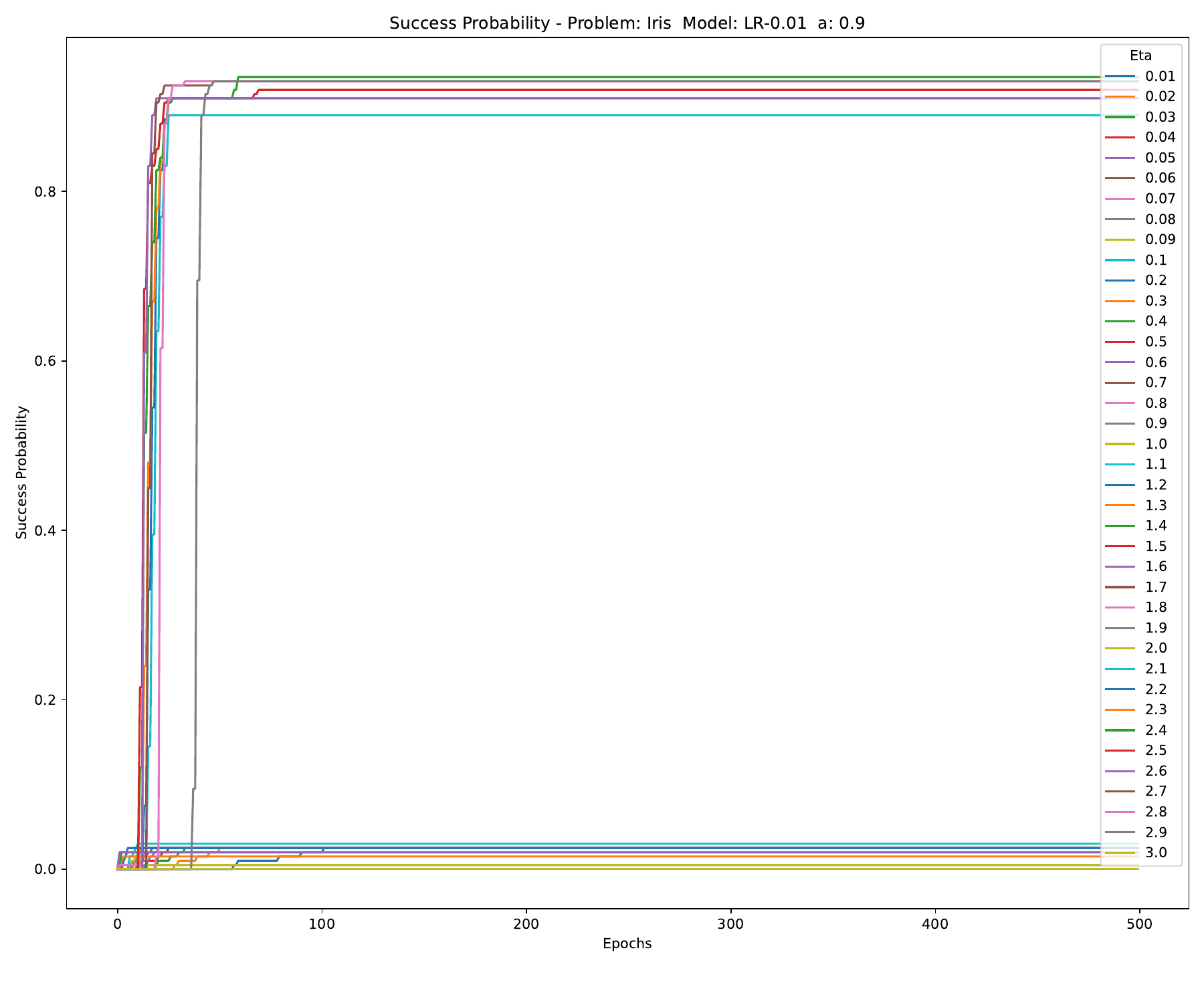} &
    \includegraphics[clip,trim=0 0 0 0,width=.52\linewidth]{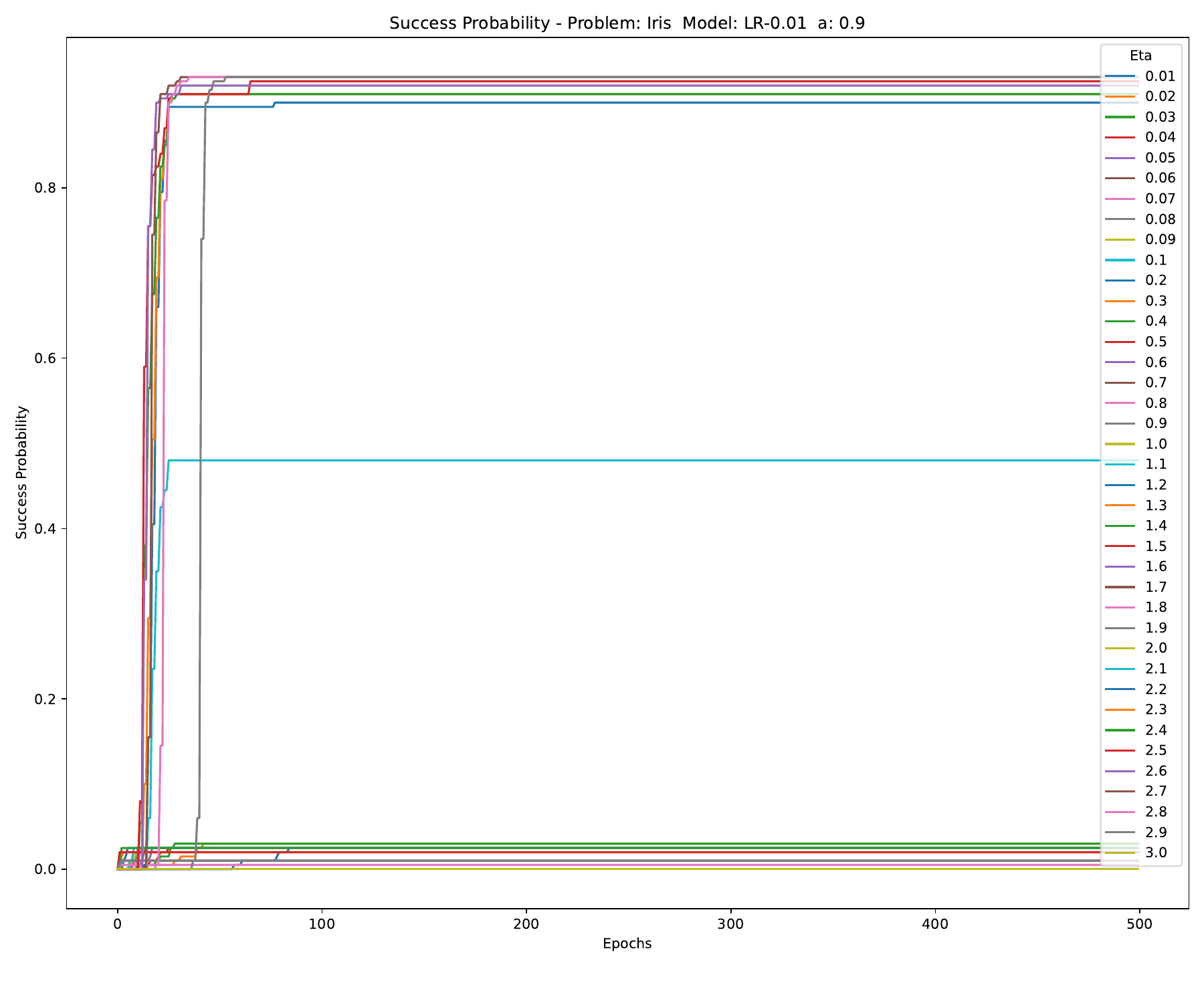} \\[-3mm]
    \includegraphics[clip,trim=0 0 0 0,width=.52\linewidth]{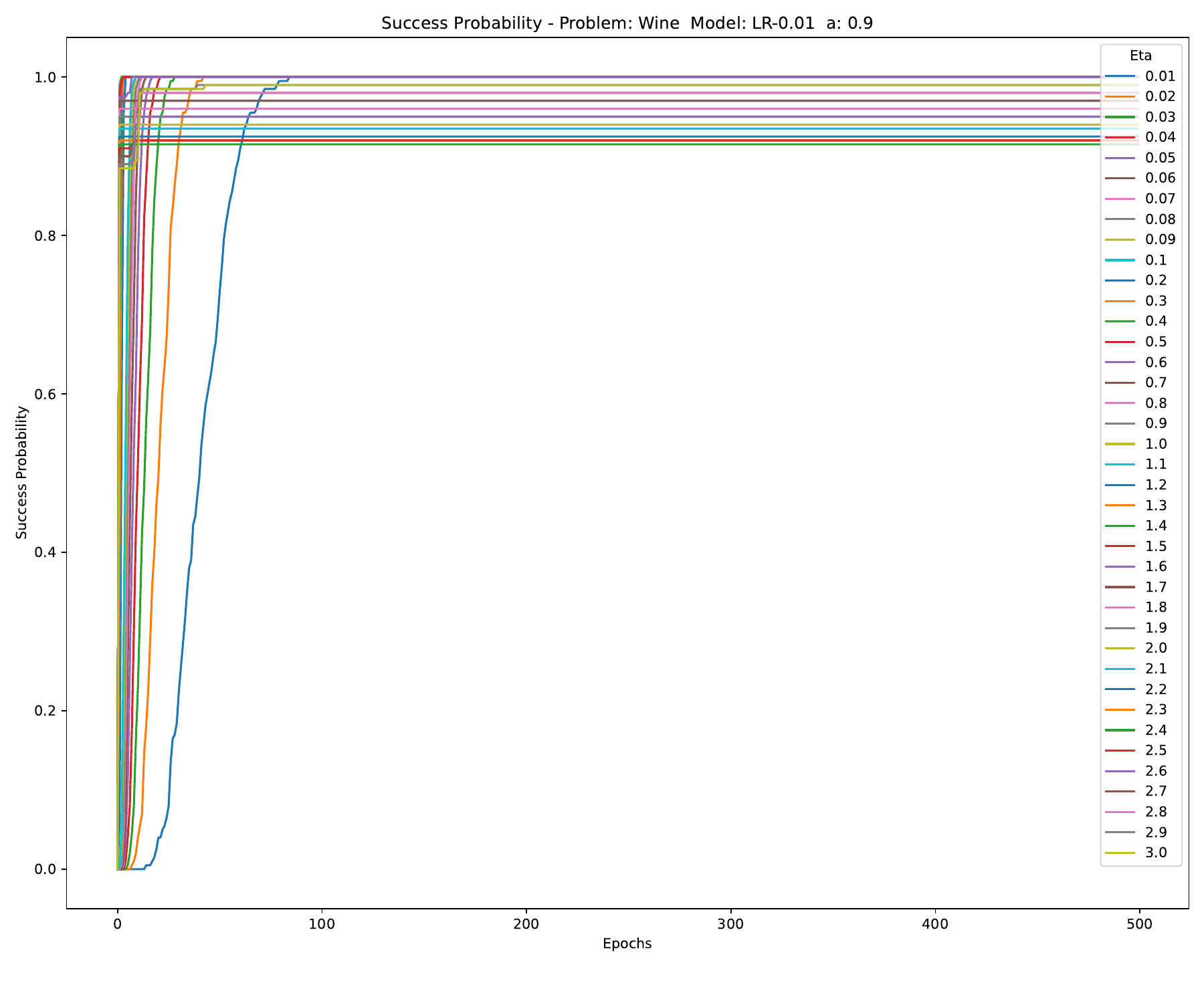} &
    \includegraphics[clip,trim=0 0 0 0,width=.52\linewidth]{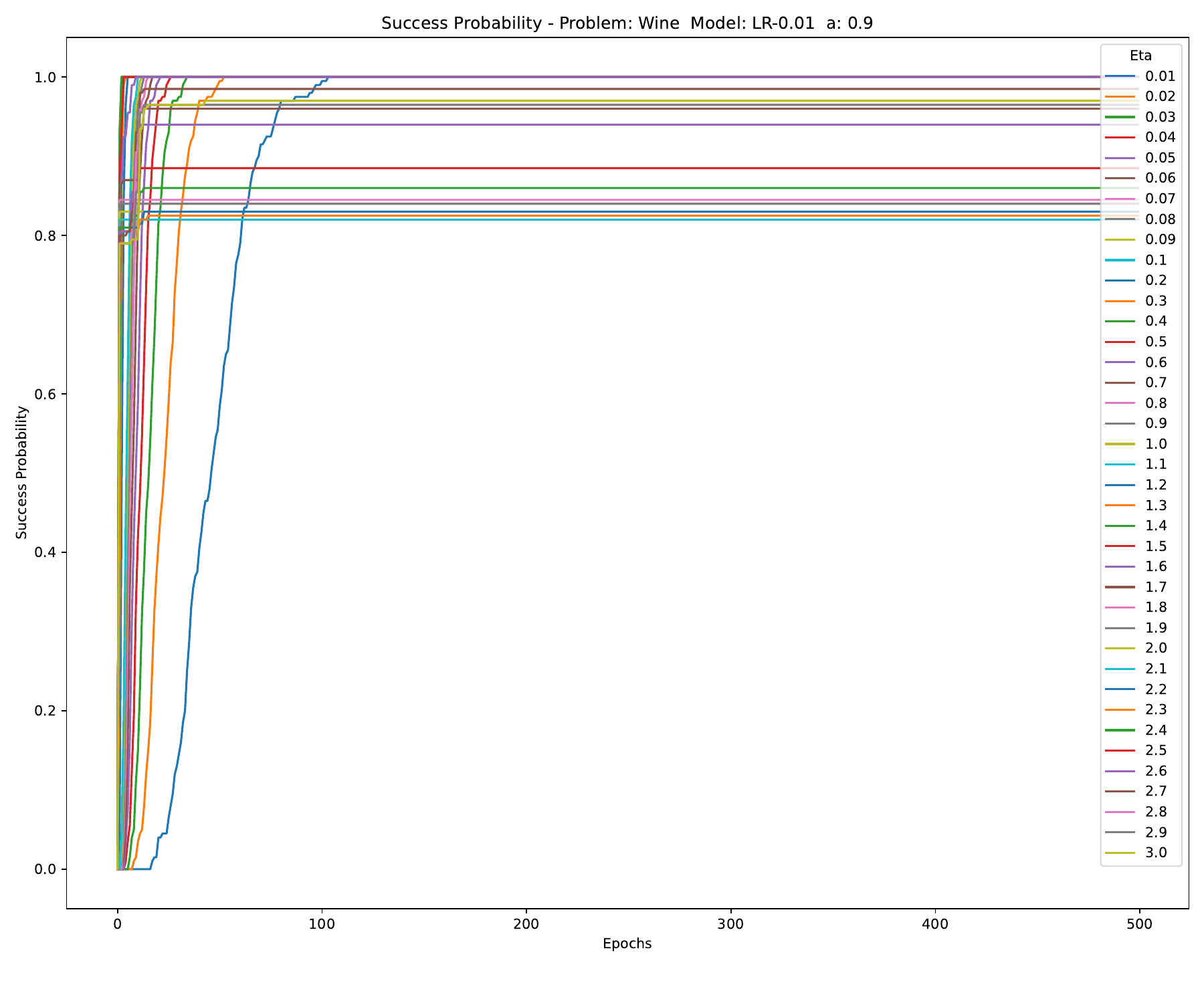} \\[-3mm]
  \end{tabular}
\caption{Success probability ($a=0.9$) for LR-0.01 model (continued).}
\end{figure*}

\begin{figure*}[p]
  \centering
  \begin{tabular}{c@{}c}
    Training Success Probability & Validation Success Probability \\
    \includegraphics[clip,trim=0 0 0 0,width=.52\linewidth]{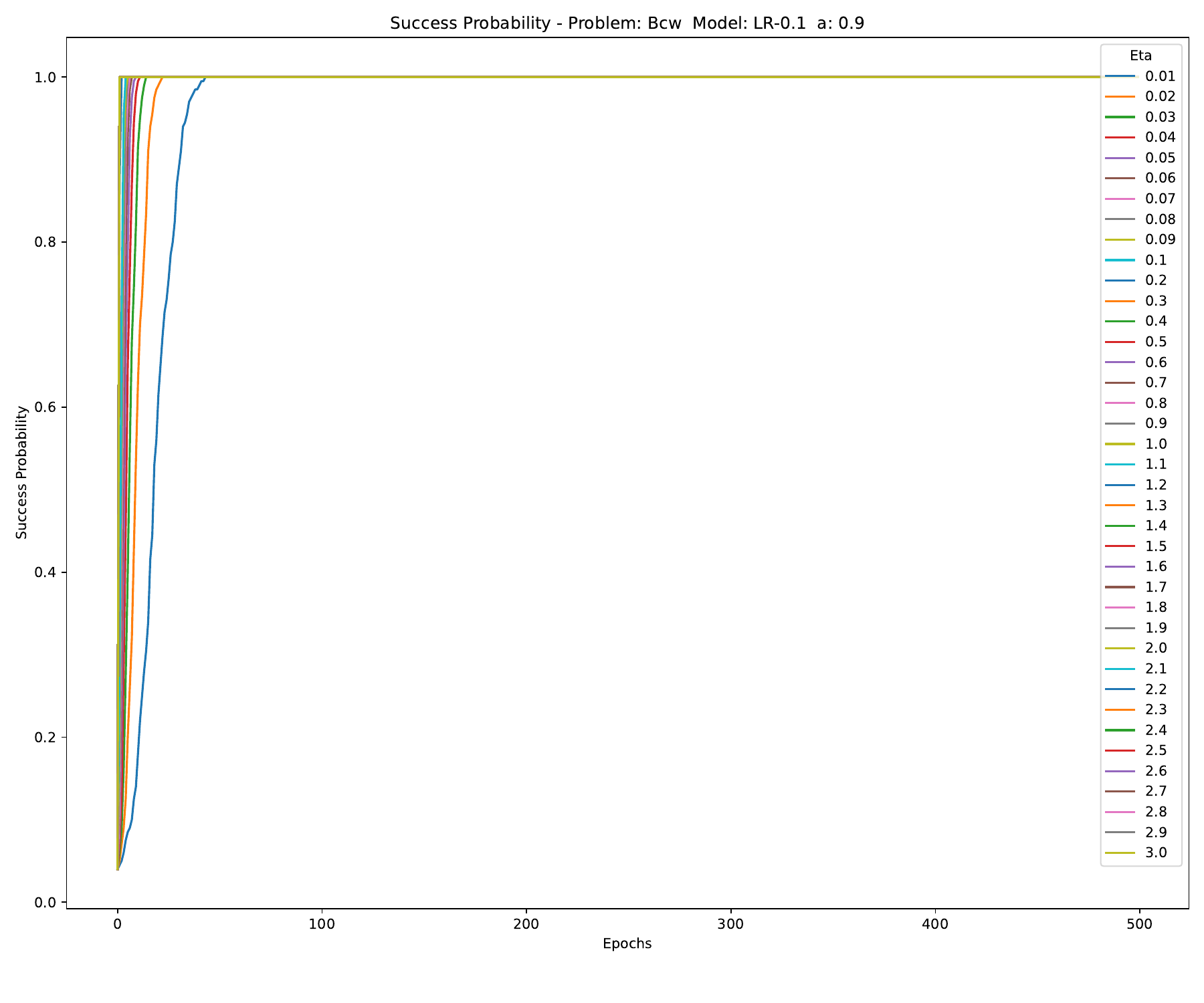} &
    \includegraphics[clip,trim=0 0 0 0,width=.52\linewidth]{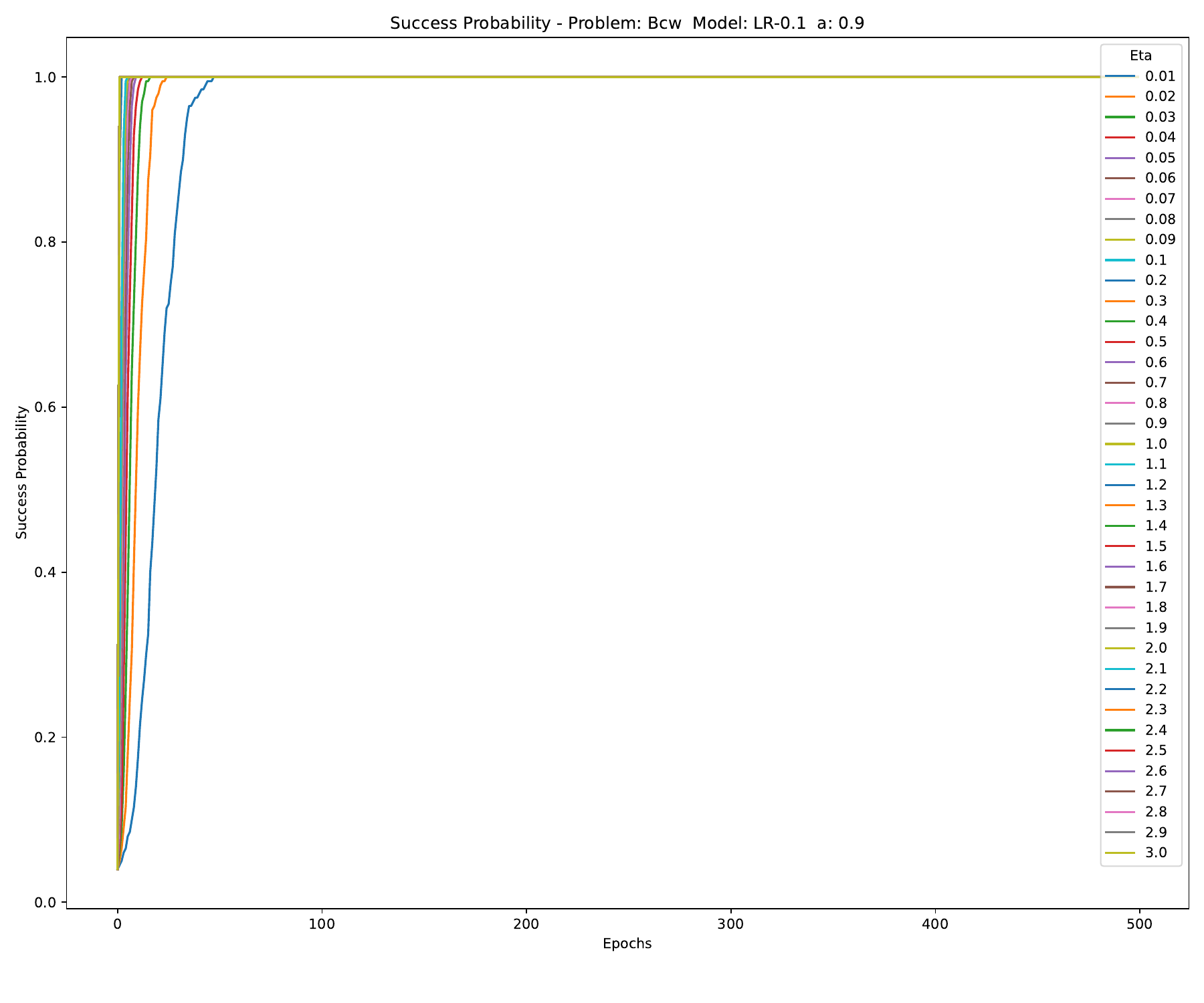} \\[-3mm]
    \includegraphics[clip,trim=0 0 0 0,width=.52\linewidth]{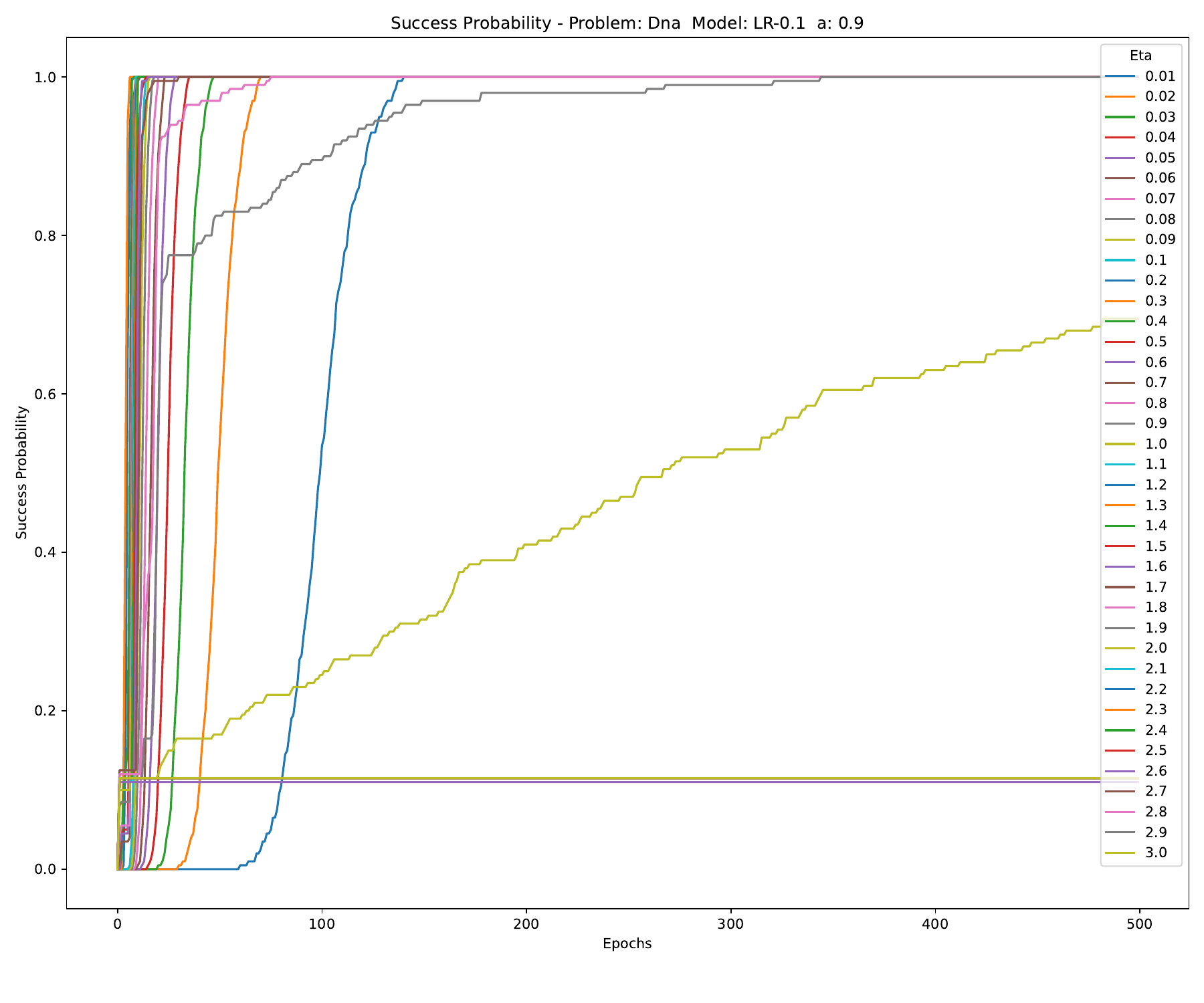} &
    \includegraphics[clip,trim=0 0 0 0,width=.52\linewidth]{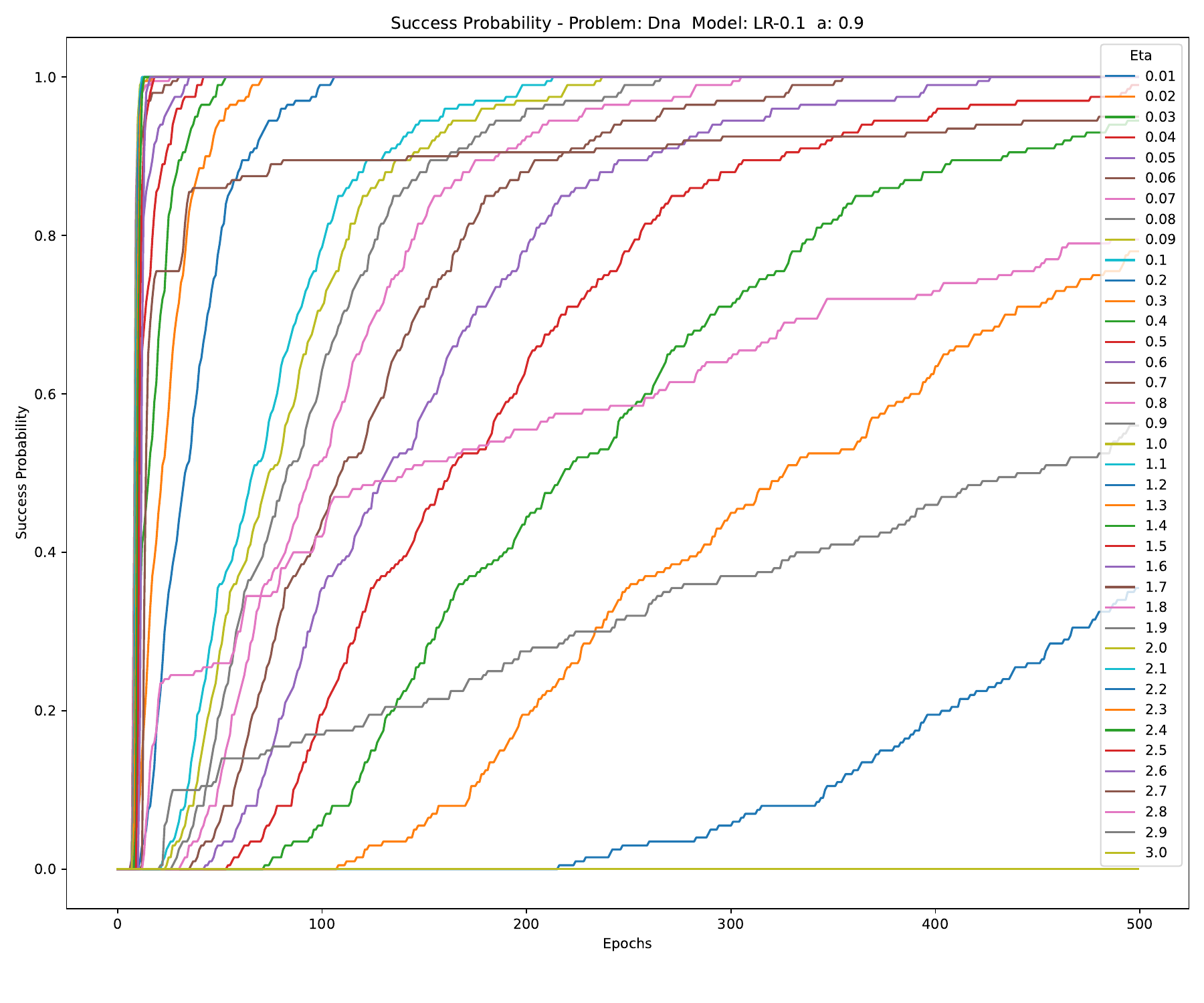} \\[-3mm]
    \includegraphics[clip,trim=0 0 0 0,width=.52\linewidth]{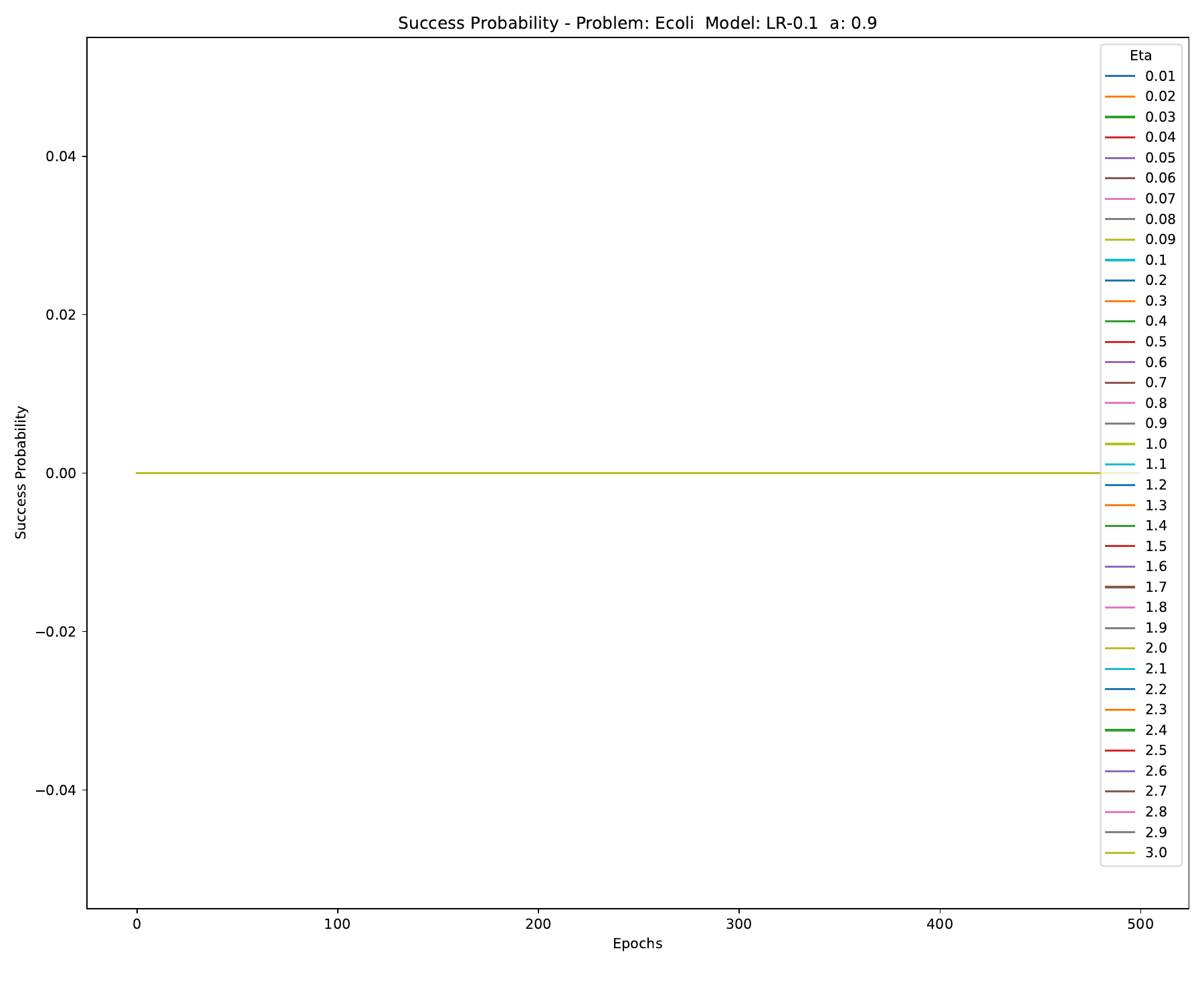} &
    \includegraphics[clip,trim=0 0 0 0,width=.52\linewidth]{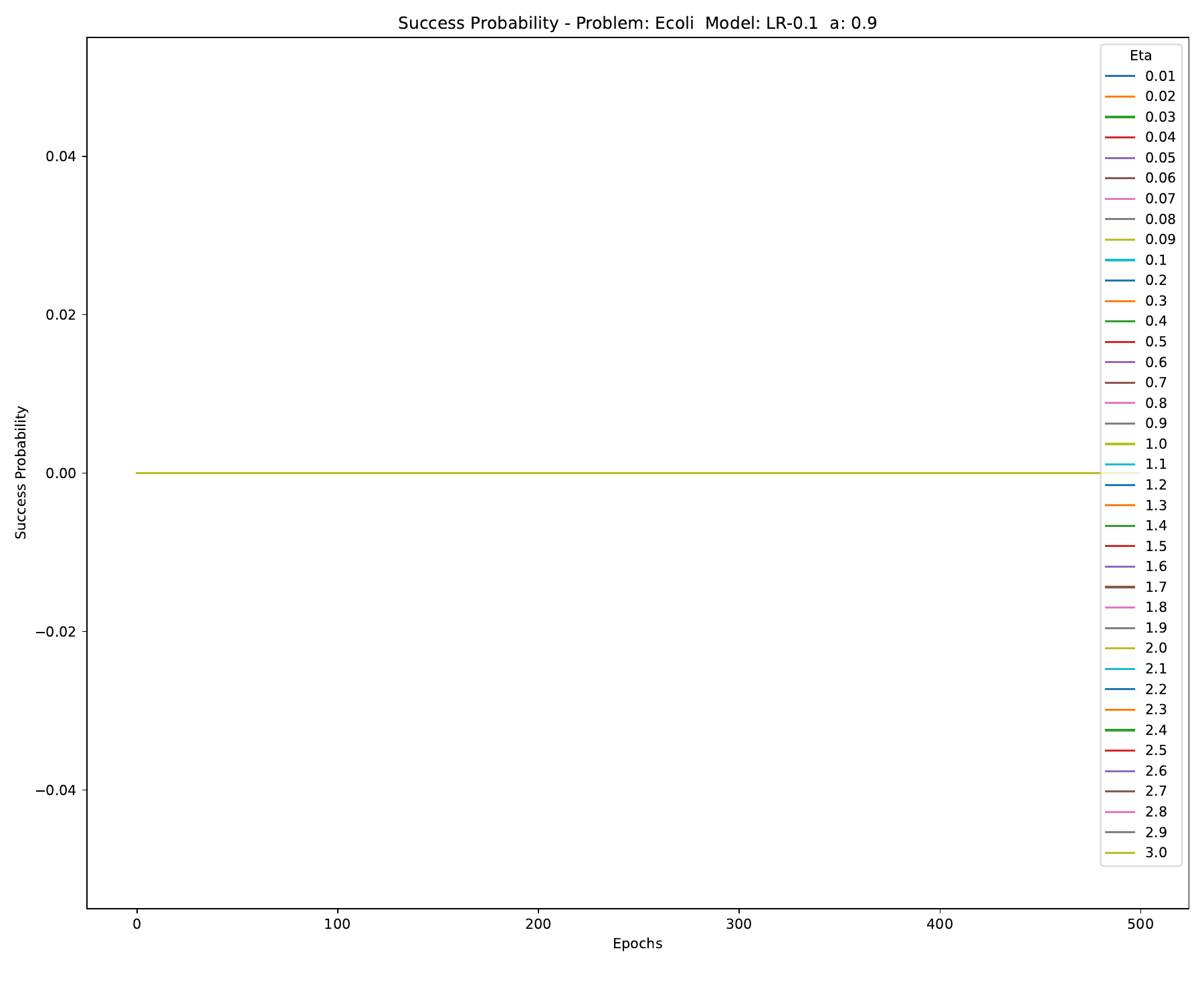} \\[-3mm]
  \end{tabular}
\caption{Success probability ($a=0.9$) for LR-0.1 model.}
  \label{fig:success_probability_LR_0.1_model_0.9}
\end{figure*}

\begin{figure*}[p]
  \centering
  \ContinuedFloat
  \begin{tabular}{c@{}c}
    Training Success Probability & Validation Success Probability \\
    \includegraphics[clip,trim=0 0 0 0,width=.52\linewidth]{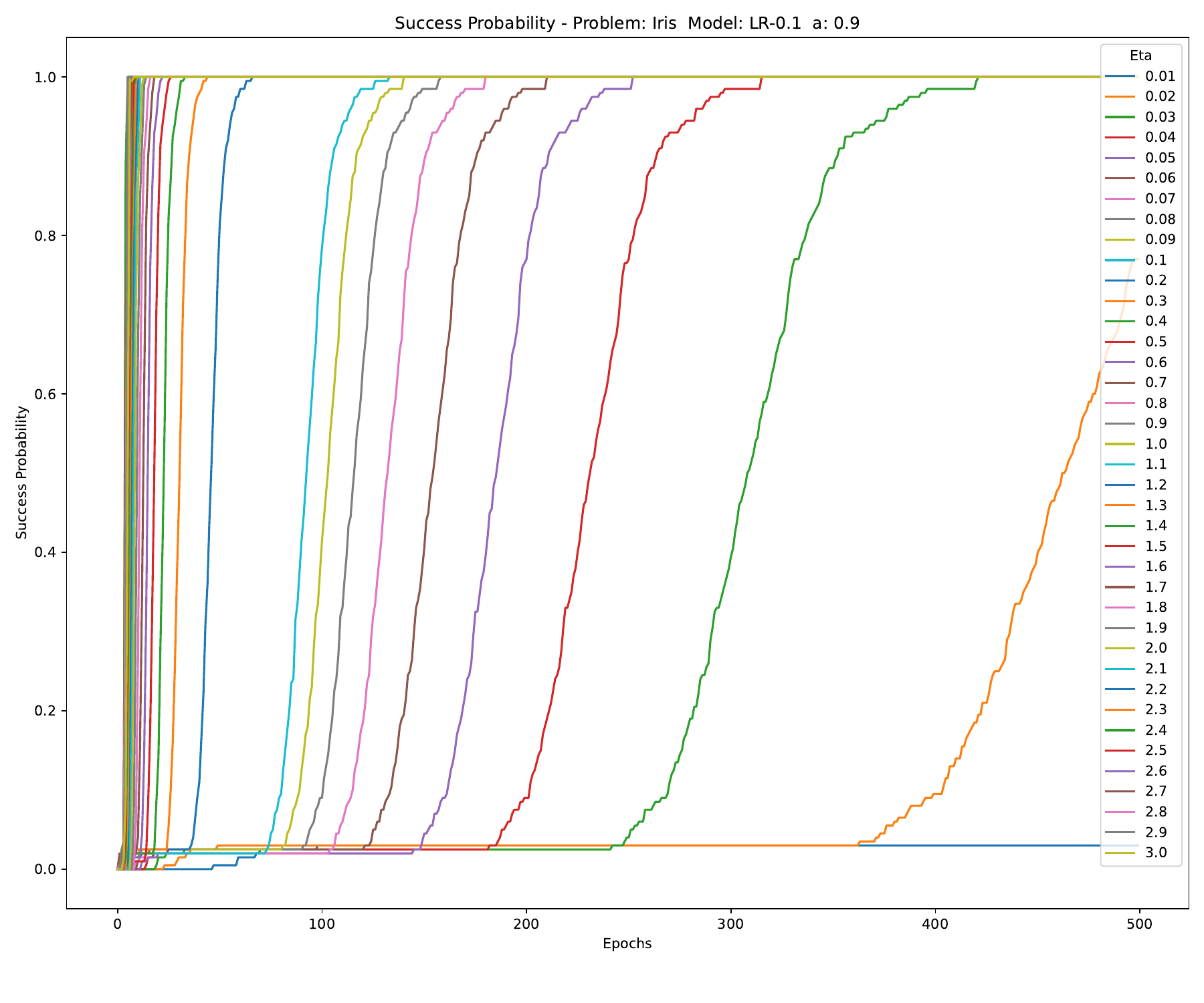} &
    \includegraphics[clip,trim=0 0 0 0,width=.52\linewidth]{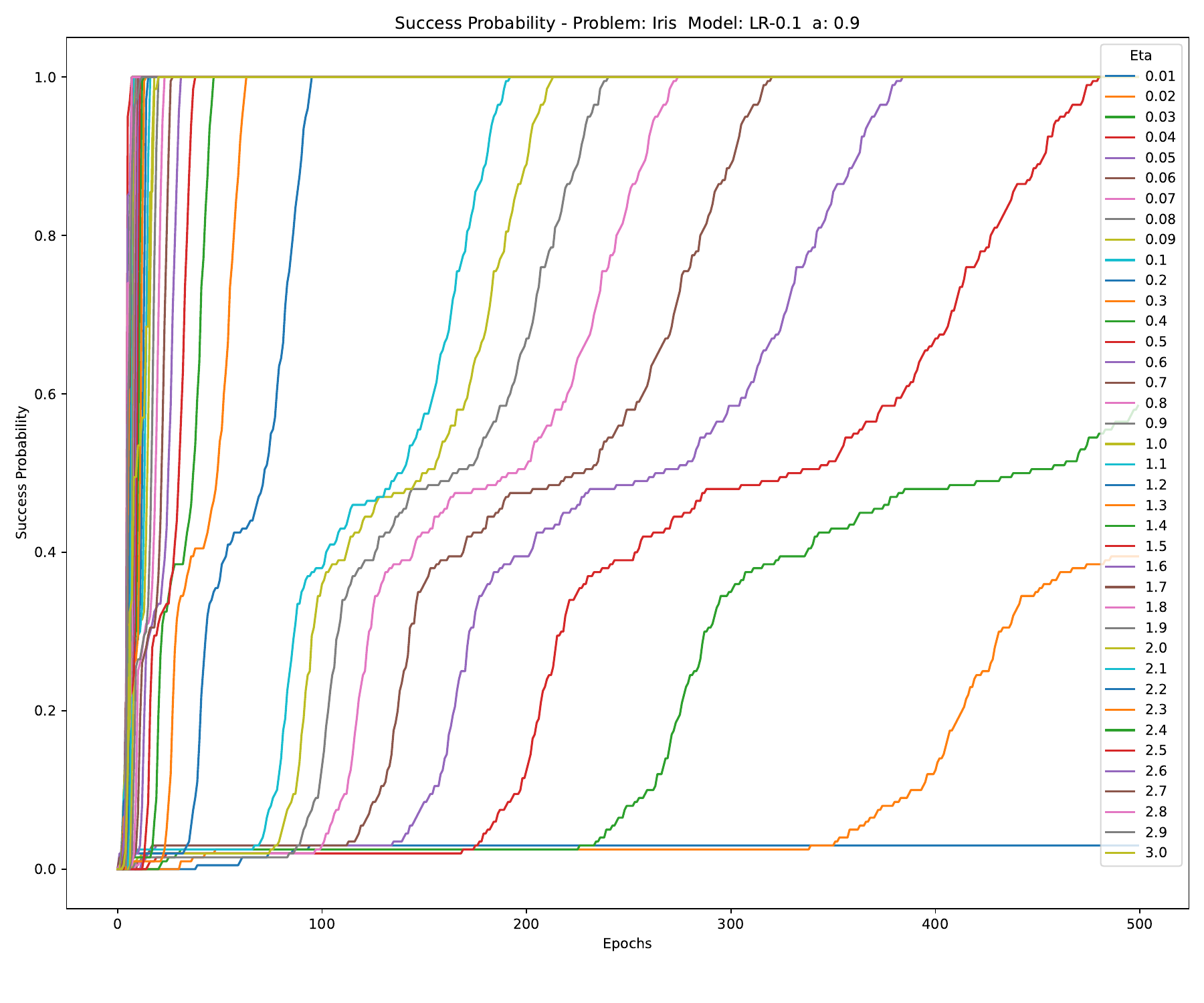} \\[-3mm]
    \includegraphics[clip,trim=0 0 0 0,width=.52\linewidth]{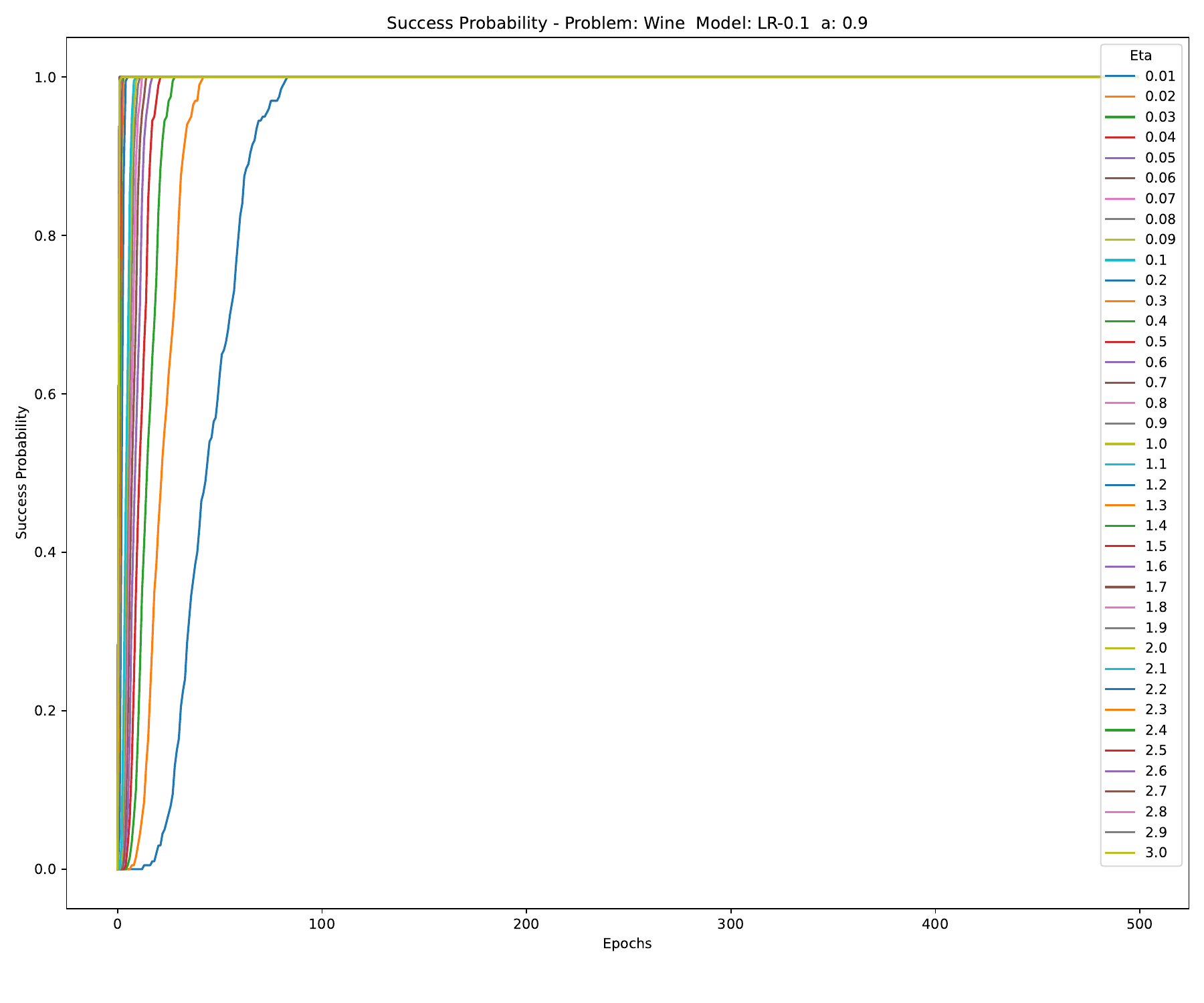} &
    \includegraphics[clip,trim=0 0 0 0,width=.52\linewidth]{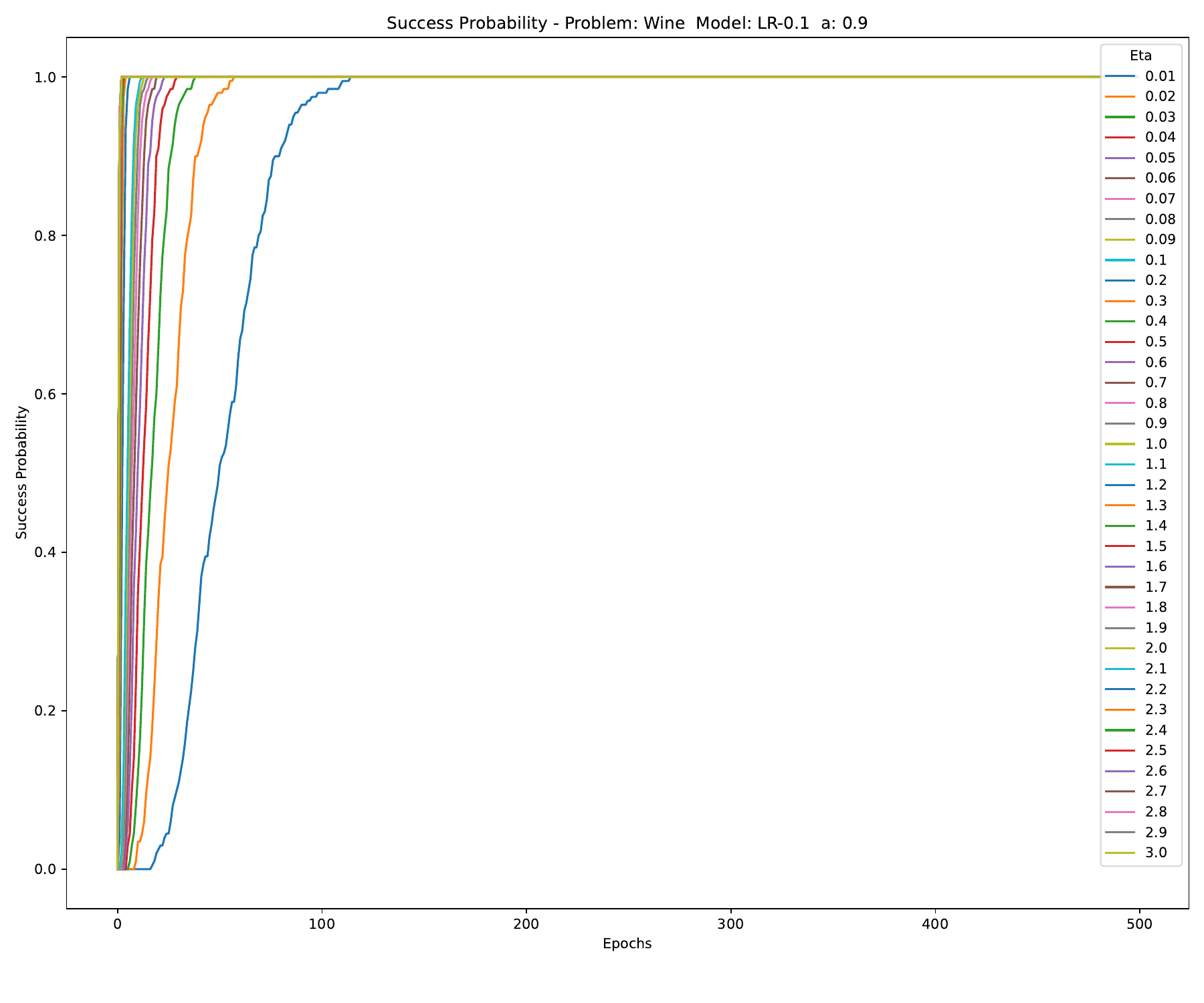} \\[-3mm]
  \end{tabular}
\caption{Success probability ($a=0.9$) for LR-0.1 model (continued).}
\end{figure*}

\begin{figure*}[p]
  \centering
  \begin{tabular}{c@{}c}
    Training Success Probability & Validation Success Probability \\
    \includegraphics[clip,trim=0 0 0 0,width=.52\linewidth]{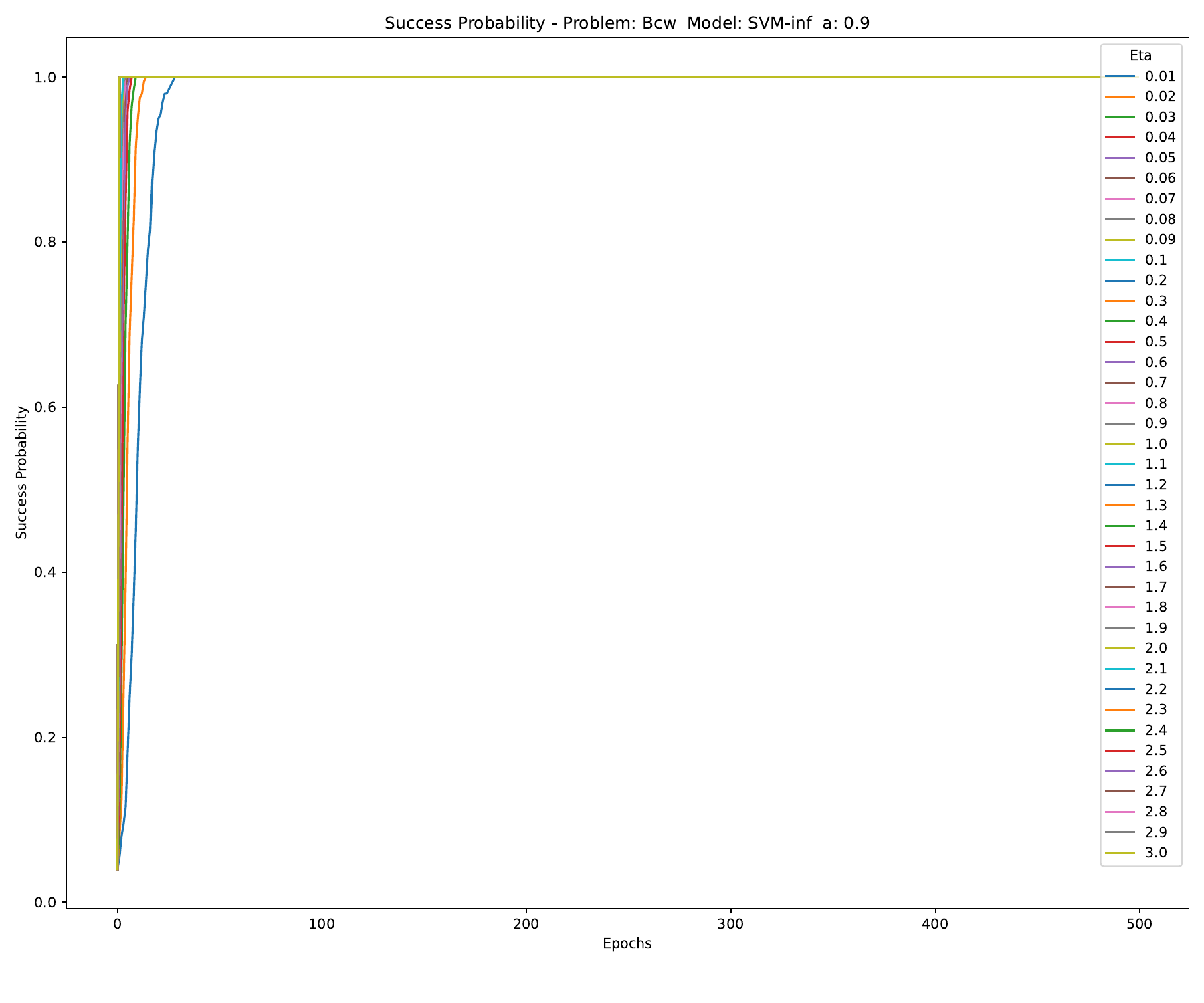} &
    \includegraphics[clip,trim=0 0 0 0,width=.52\linewidth]{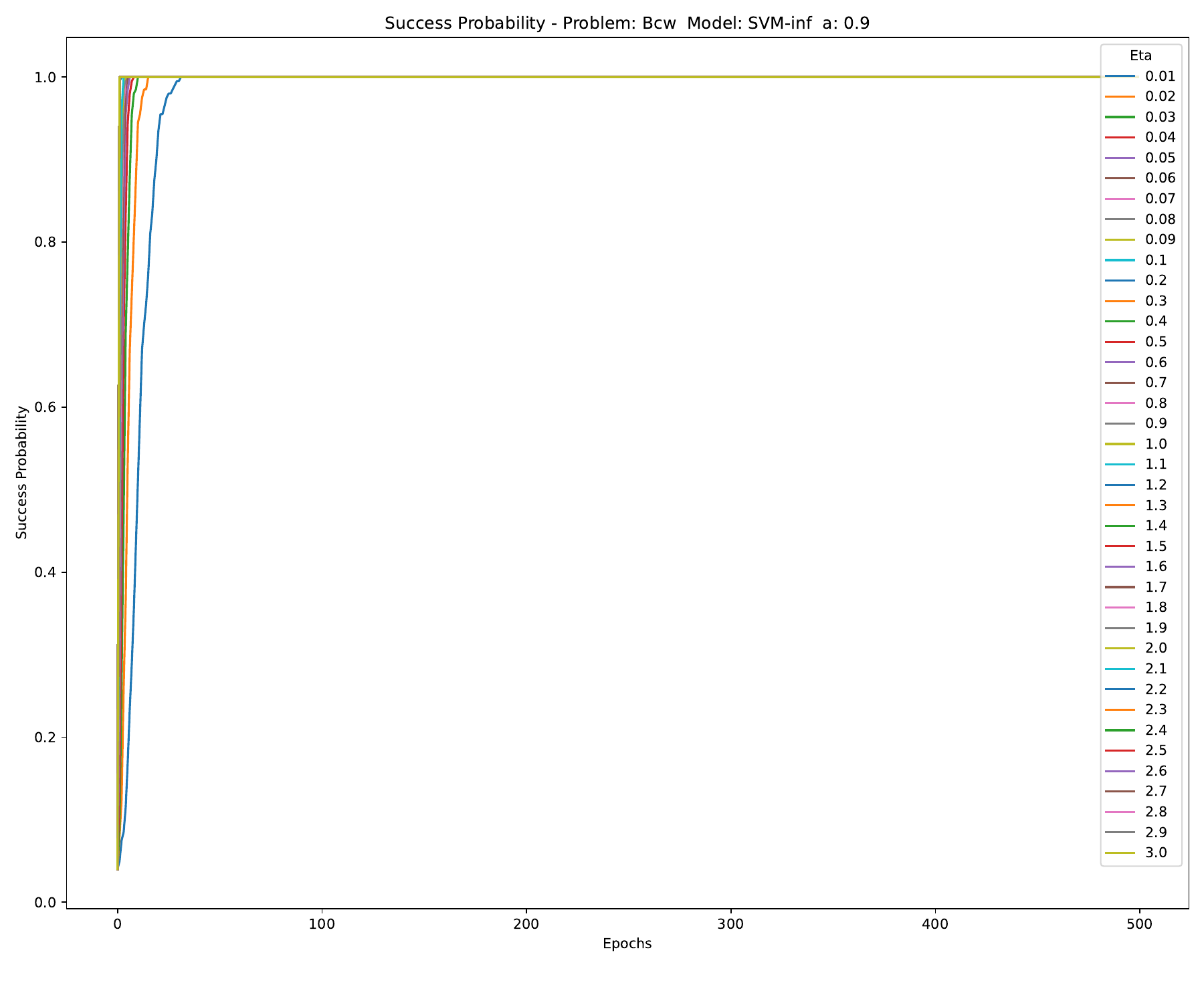} \\[-3mm]
    \includegraphics[clip,trim=0 0 0 0,width=.52\linewidth]{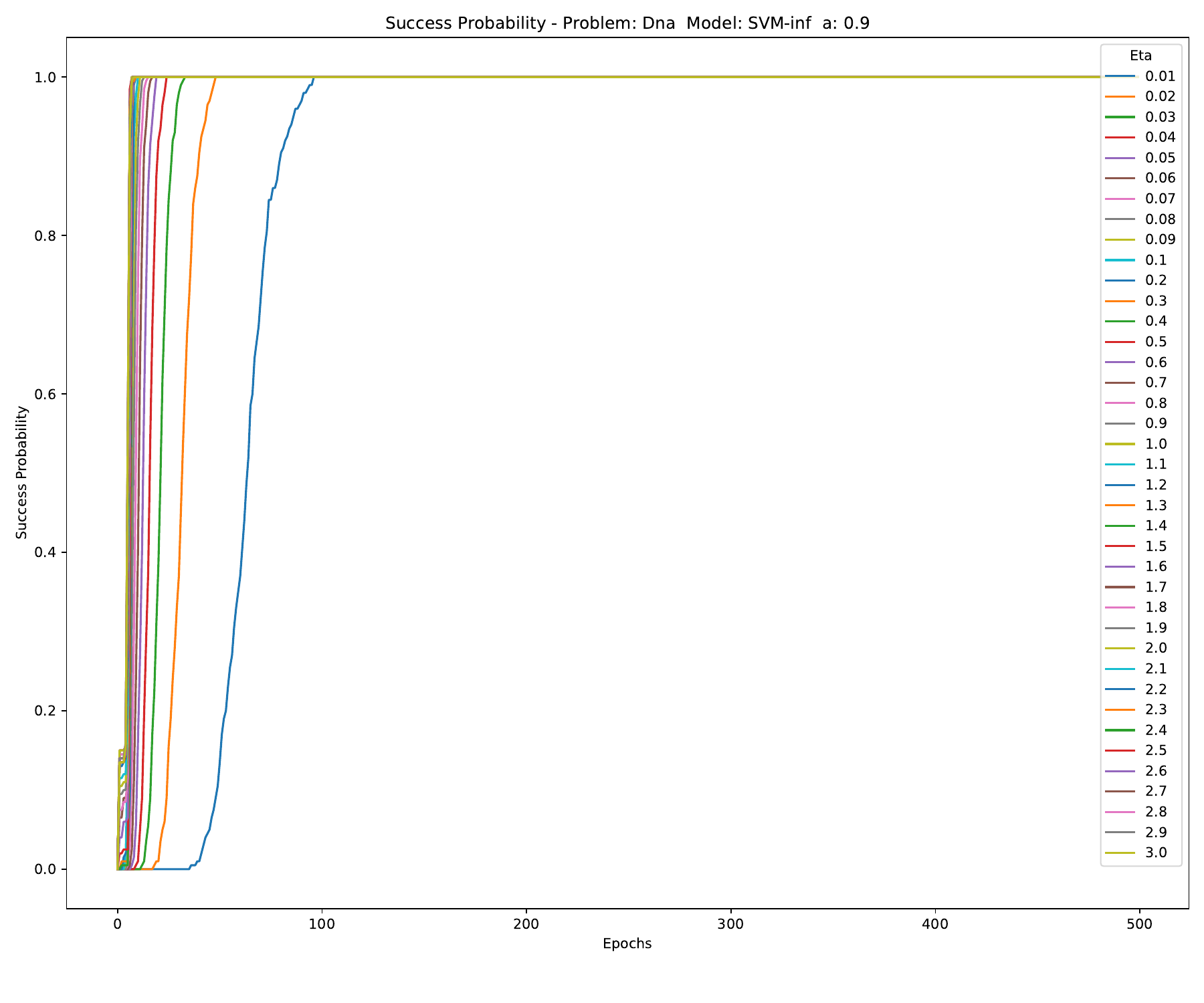} &
    \includegraphics[clip,trim=0 0 0 0,width=.52\linewidth]{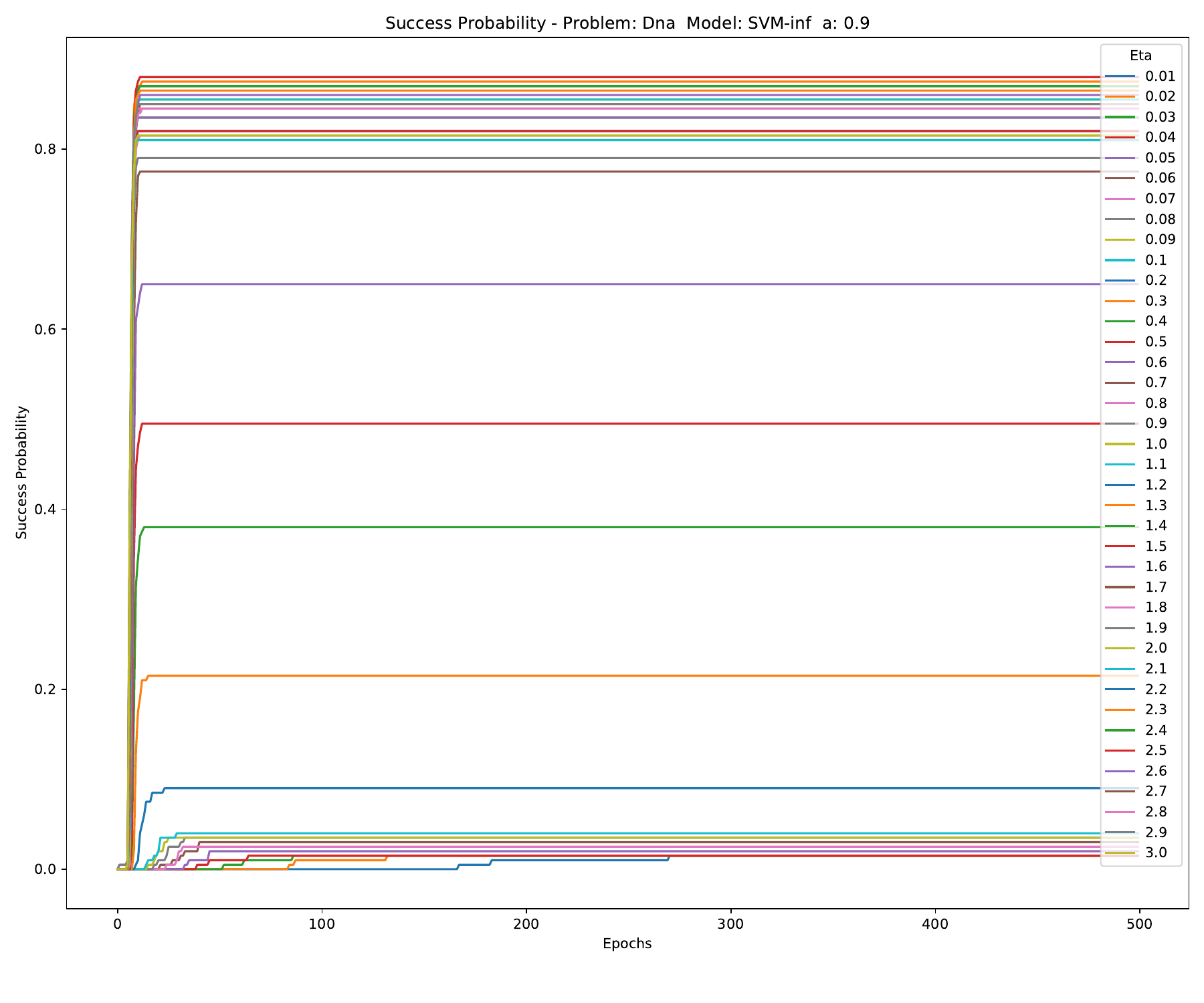} \\[-3mm]
    \includegraphics[clip,trim=0 0 0 0,width=.52\linewidth]{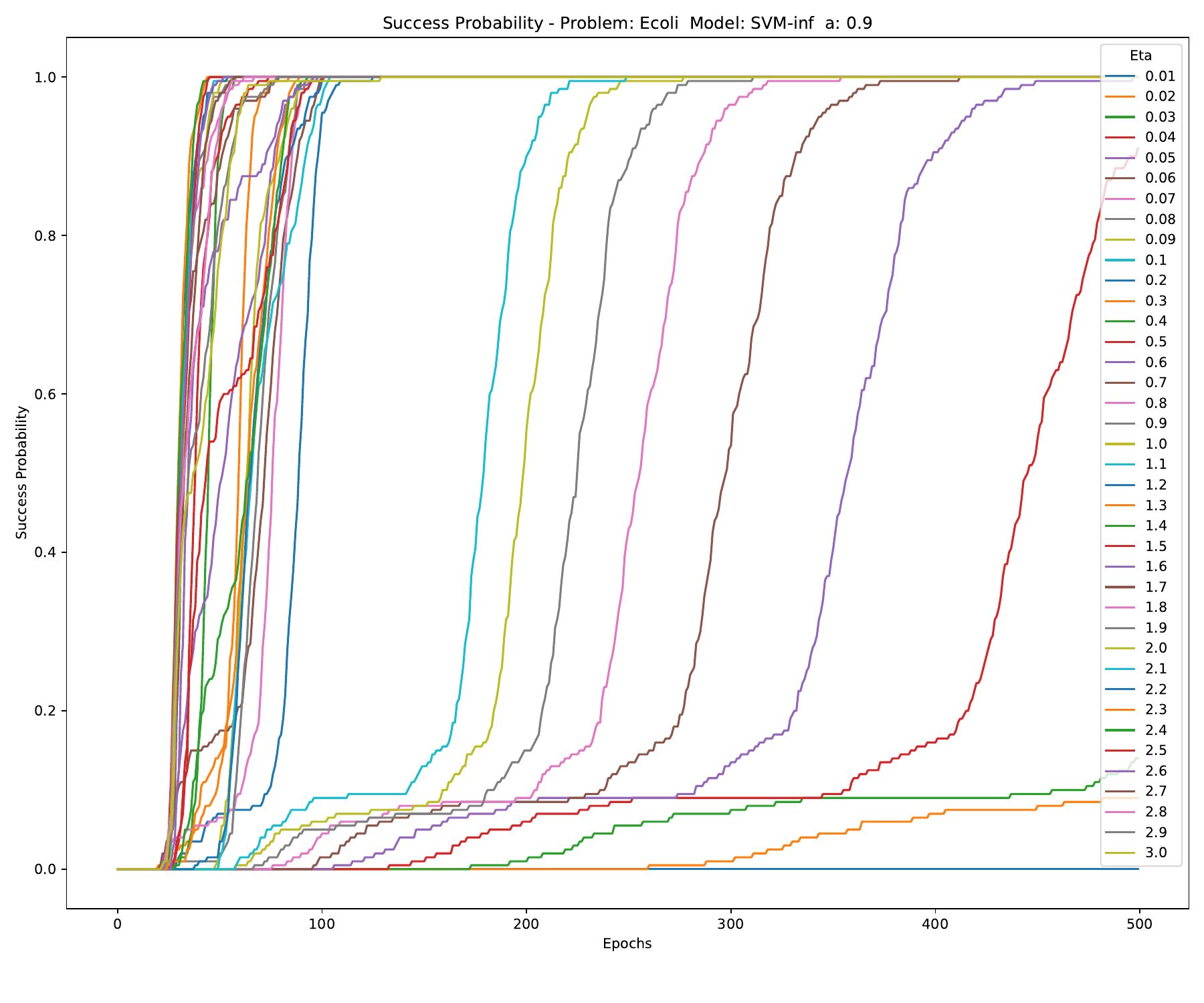} &
    \includegraphics[clip,trim=0 0 0 0,width=.52\linewidth]{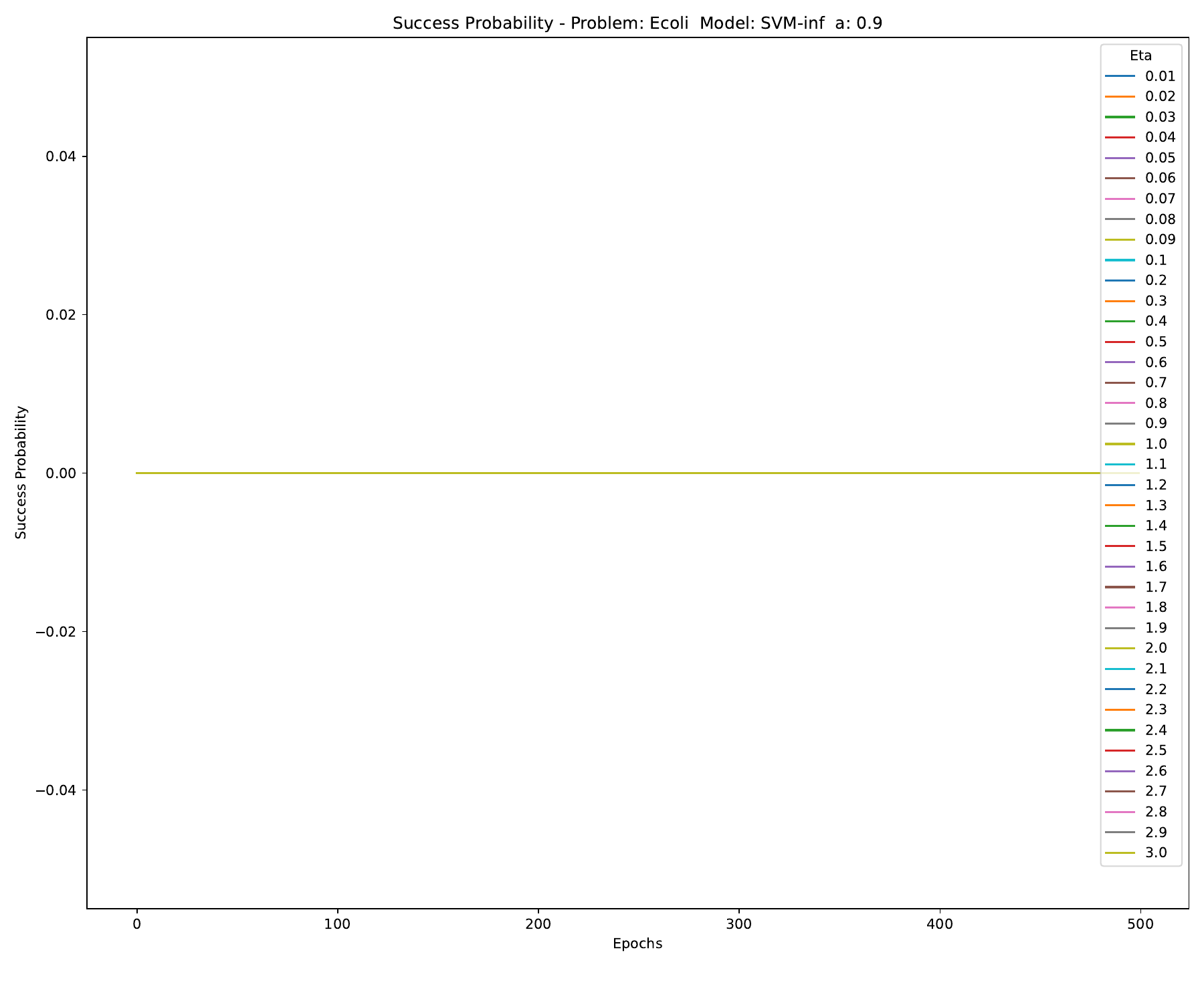} \\[-3mm]
  \end{tabular}
\caption{Success probability ($a=0.9$) for SVM-Inf model.}
  \label{fig:success_probability_SVM_Inf_model_0.9}
\end{figure*}

\begin{figure*}[p]
  \centering
  \ContinuedFloat
  \begin{tabular}{c@{}c}
    Training Success Probability & Validation Success Probability \\
    \includegraphics[clip,trim=0 0 0 0,width=.52\linewidth]{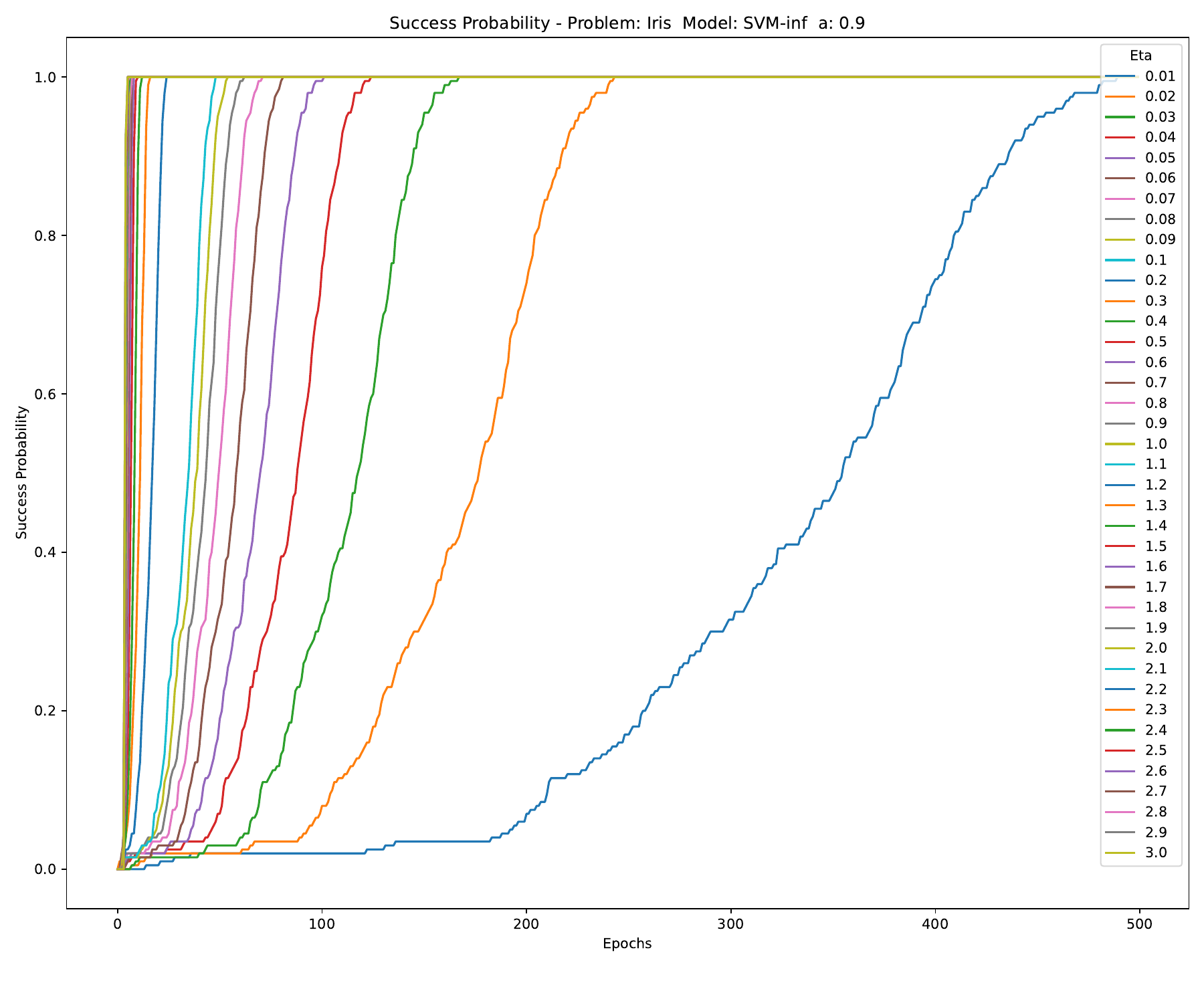} &
    \includegraphics[clip,trim=0 0 0 0,width=.52\linewidth]{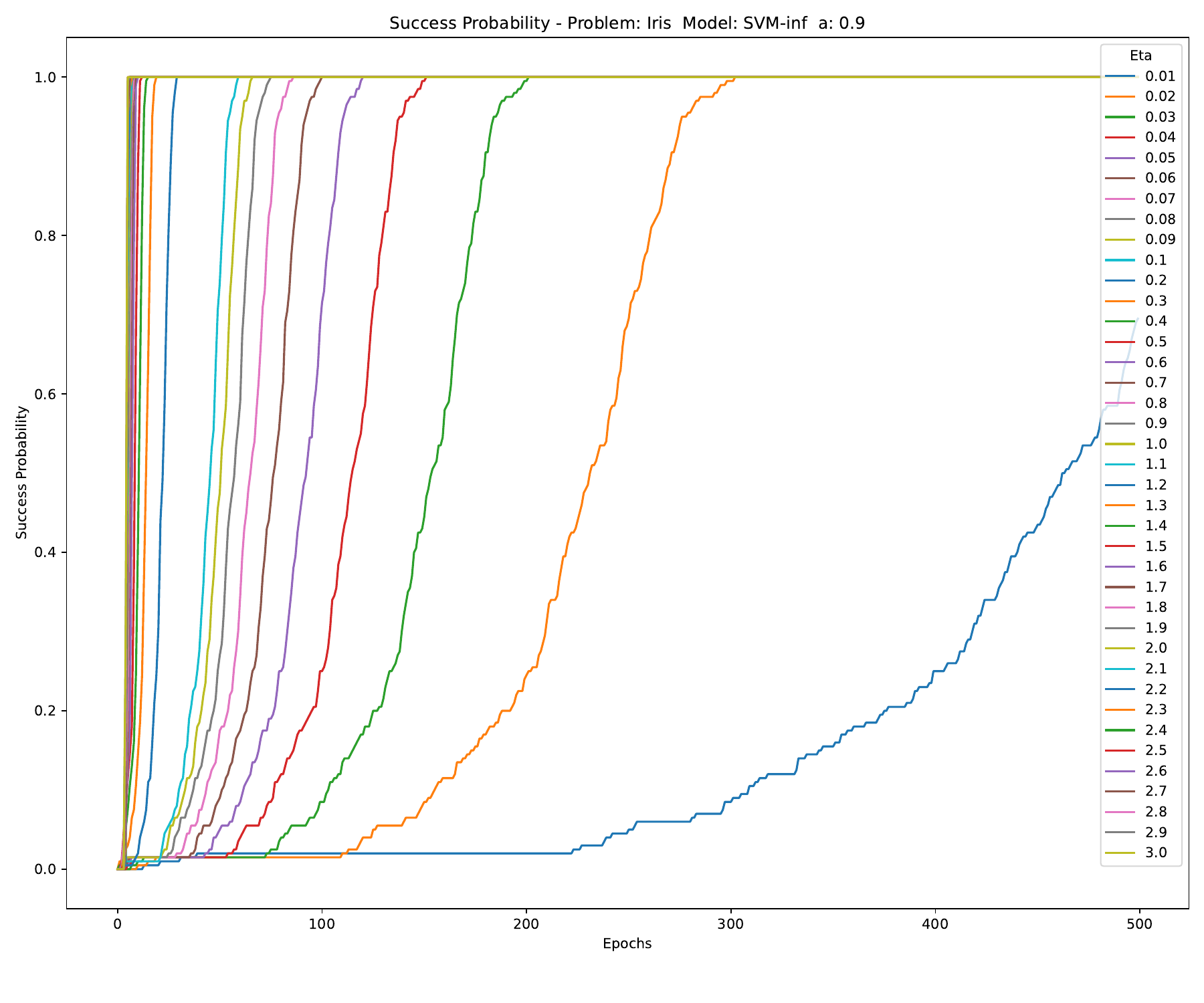} \\[-3mm]
    \includegraphics[clip,trim=0 0 0 0,width=.52\linewidth]{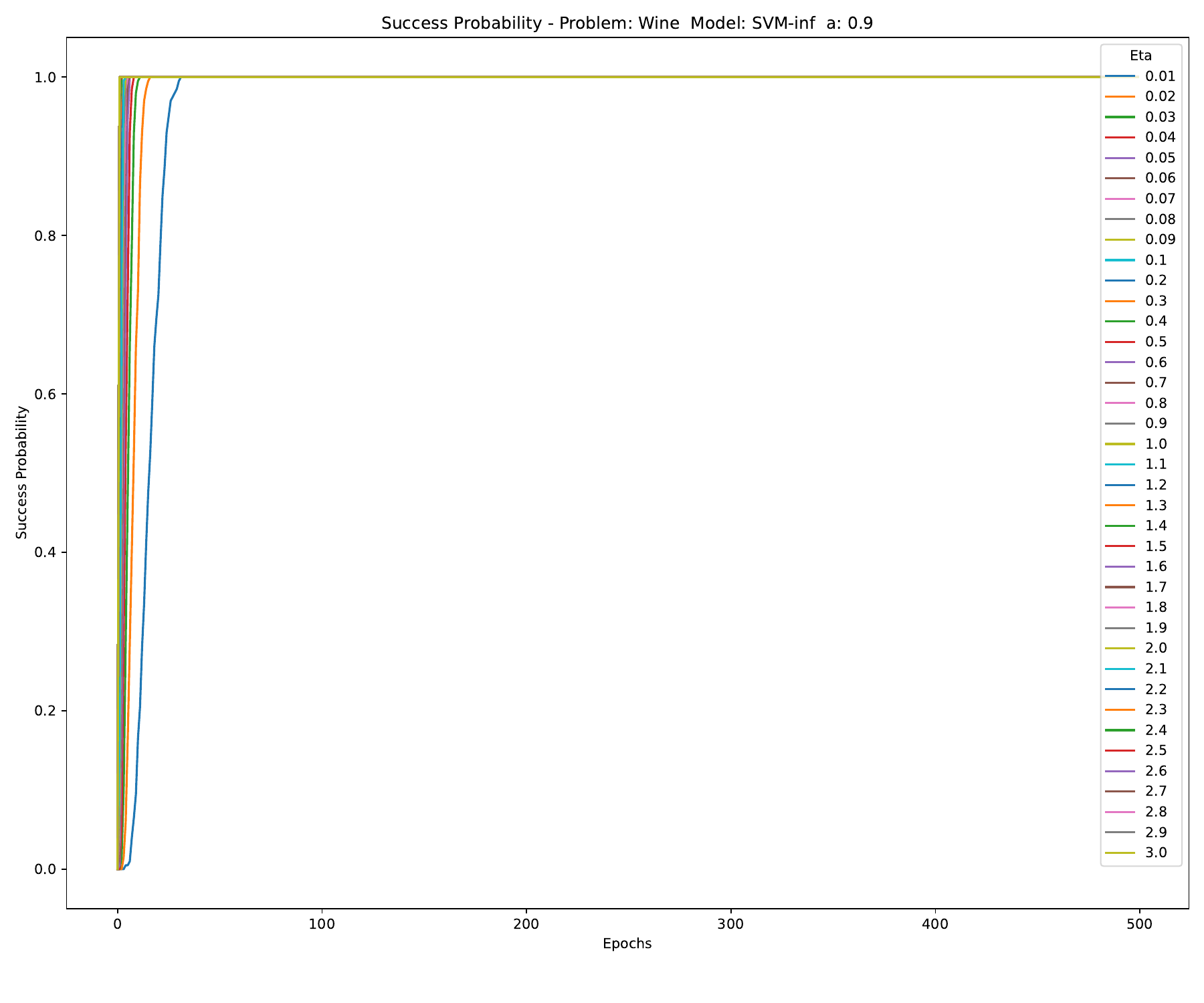} &
    \includegraphics[clip,trim=0 0 0 0,width=.52\linewidth]{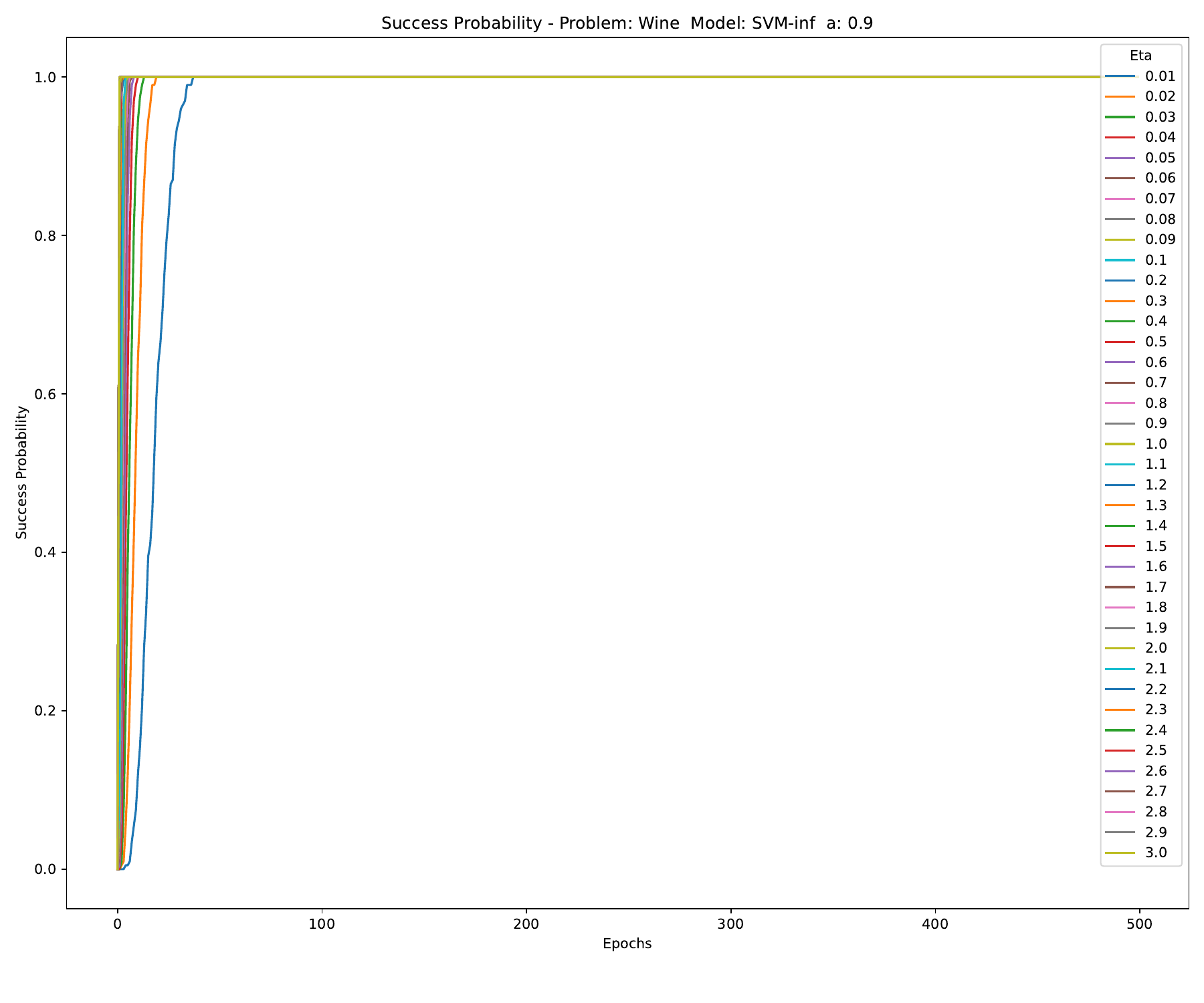} \\[-3mm]
  \end{tabular}
\caption{Success probability ($a=0.9$) for SVM-Inf model (continued).}
\end{figure*}

\begin{figure*}[p]
  \centering
  \begin{tabular}{c@{}c}
    Training Success Probability & Validation Success Probability \\
    \includegraphics[clip,trim=0 0 0 0,width=.52\linewidth]{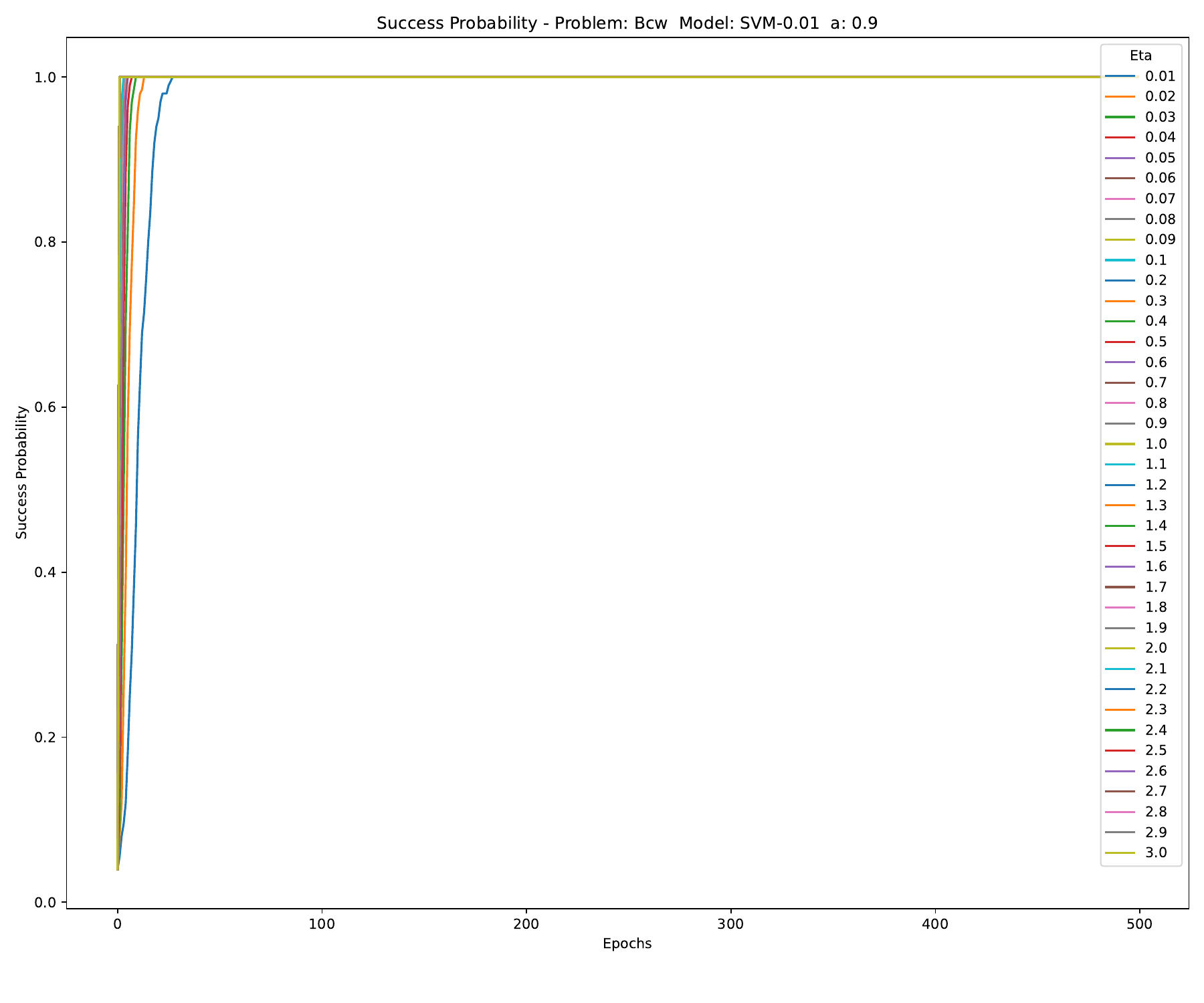} &
    \includegraphics[clip,trim=0 0 0 0,width=.52\linewidth]{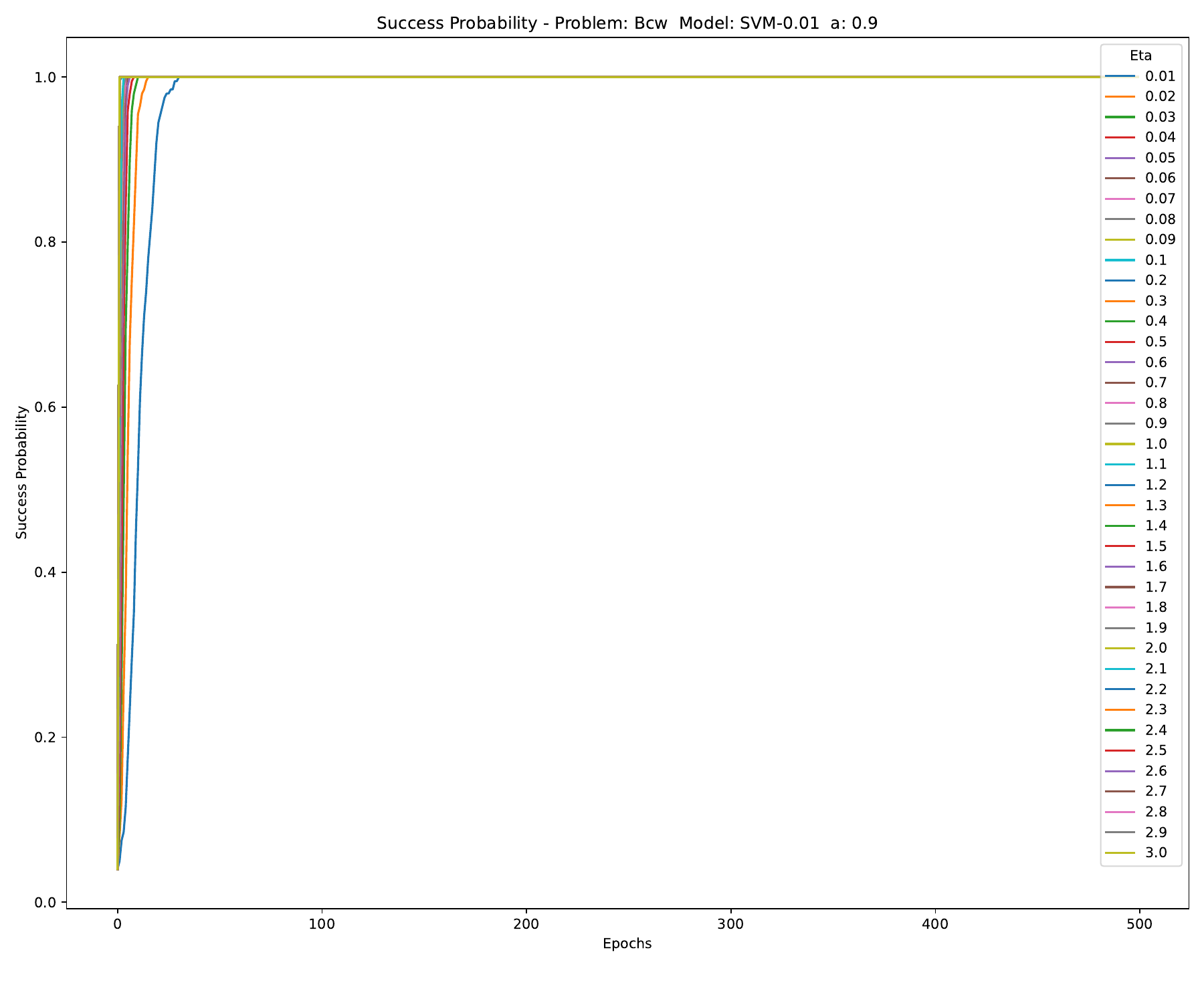} \\[-3mm]
    \includegraphics[clip,trim=0 0 0 0,width=.52\linewidth]{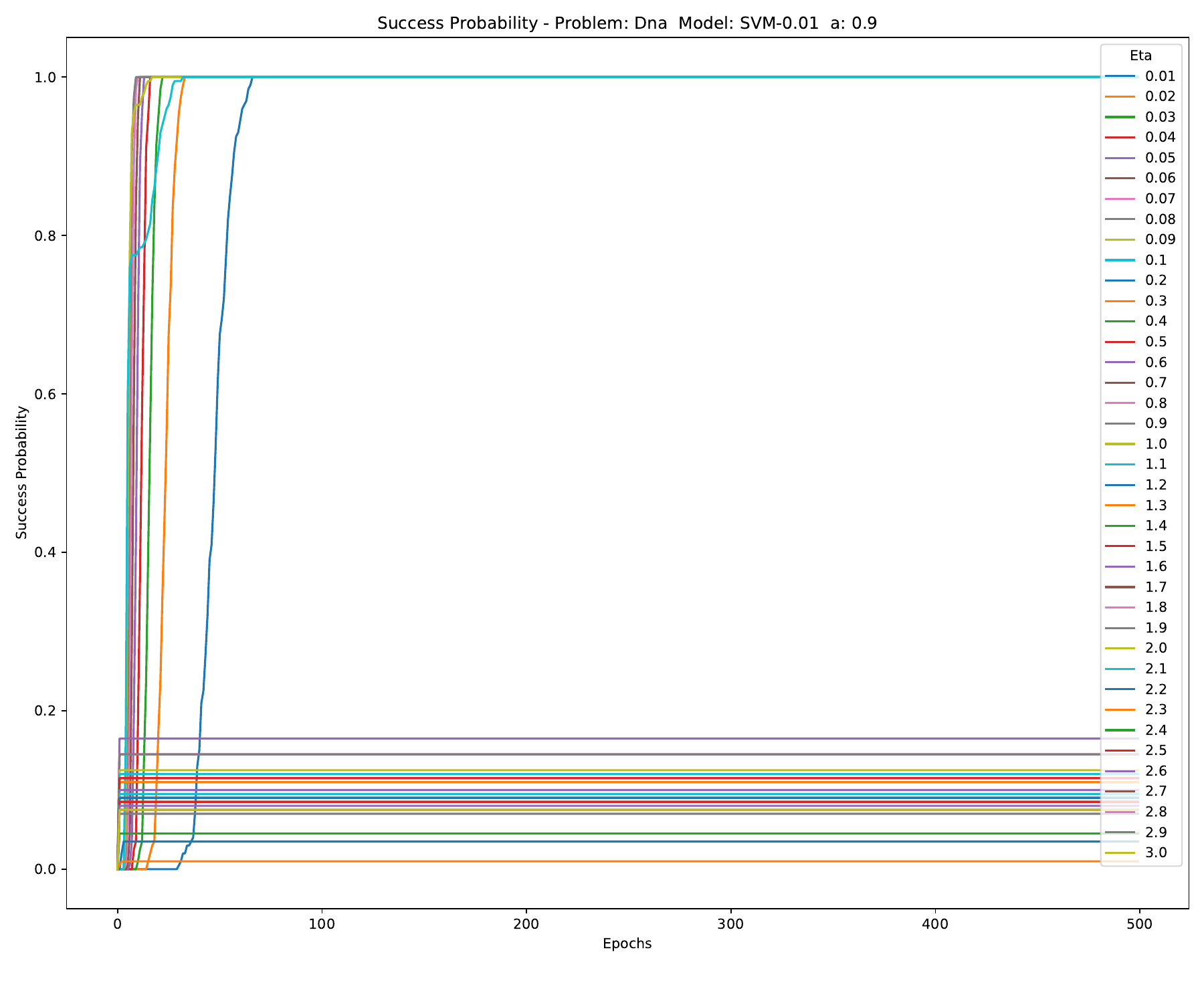} &
    \includegraphics[clip,trim=0 0 0 0,width=.52\linewidth]{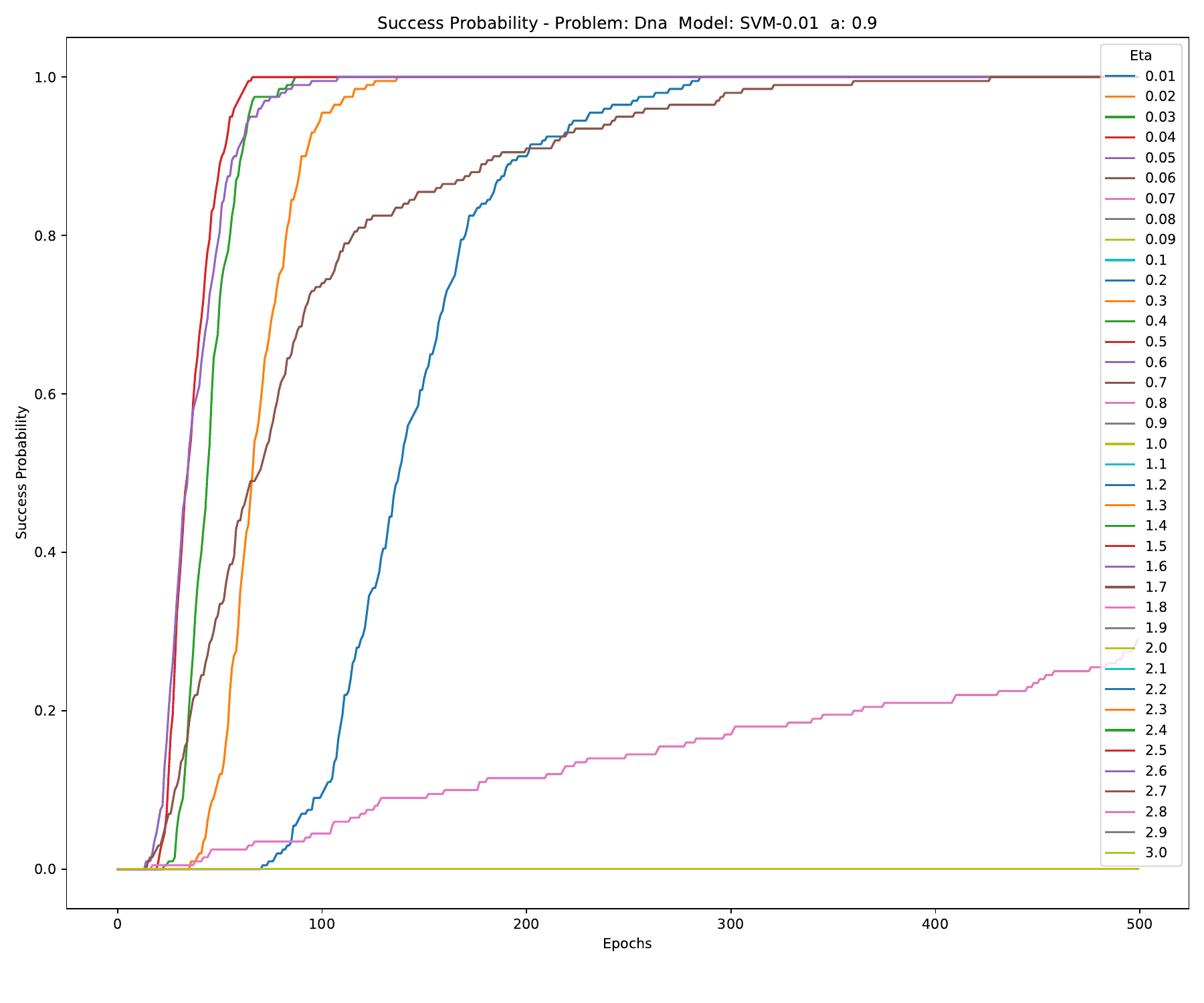} \\[-3mm]
    \includegraphics[clip,trim=0 0 0 0,width=.52\linewidth]{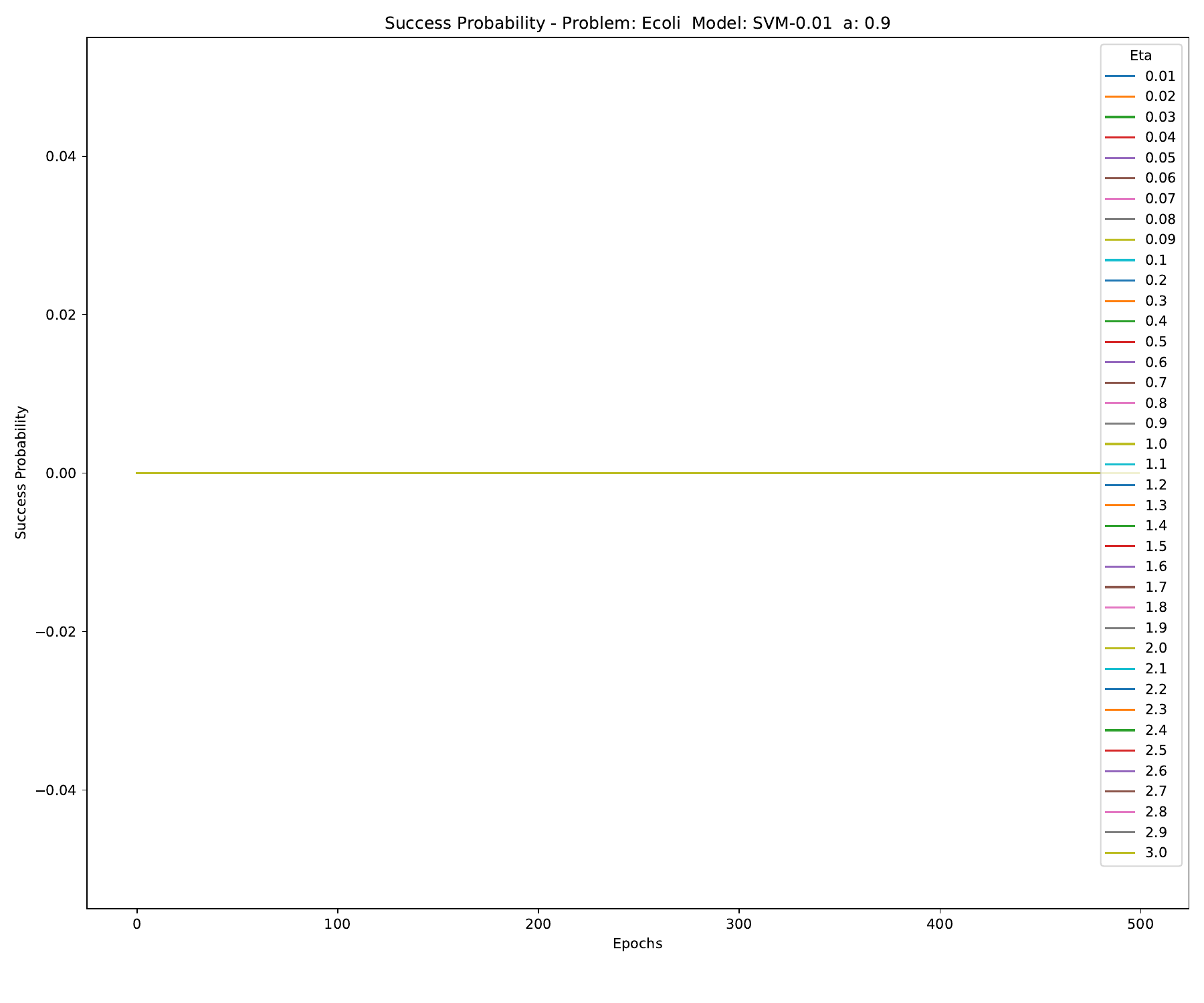} &
    \includegraphics[clip,trim=0 0 0 0,width=.52\linewidth]{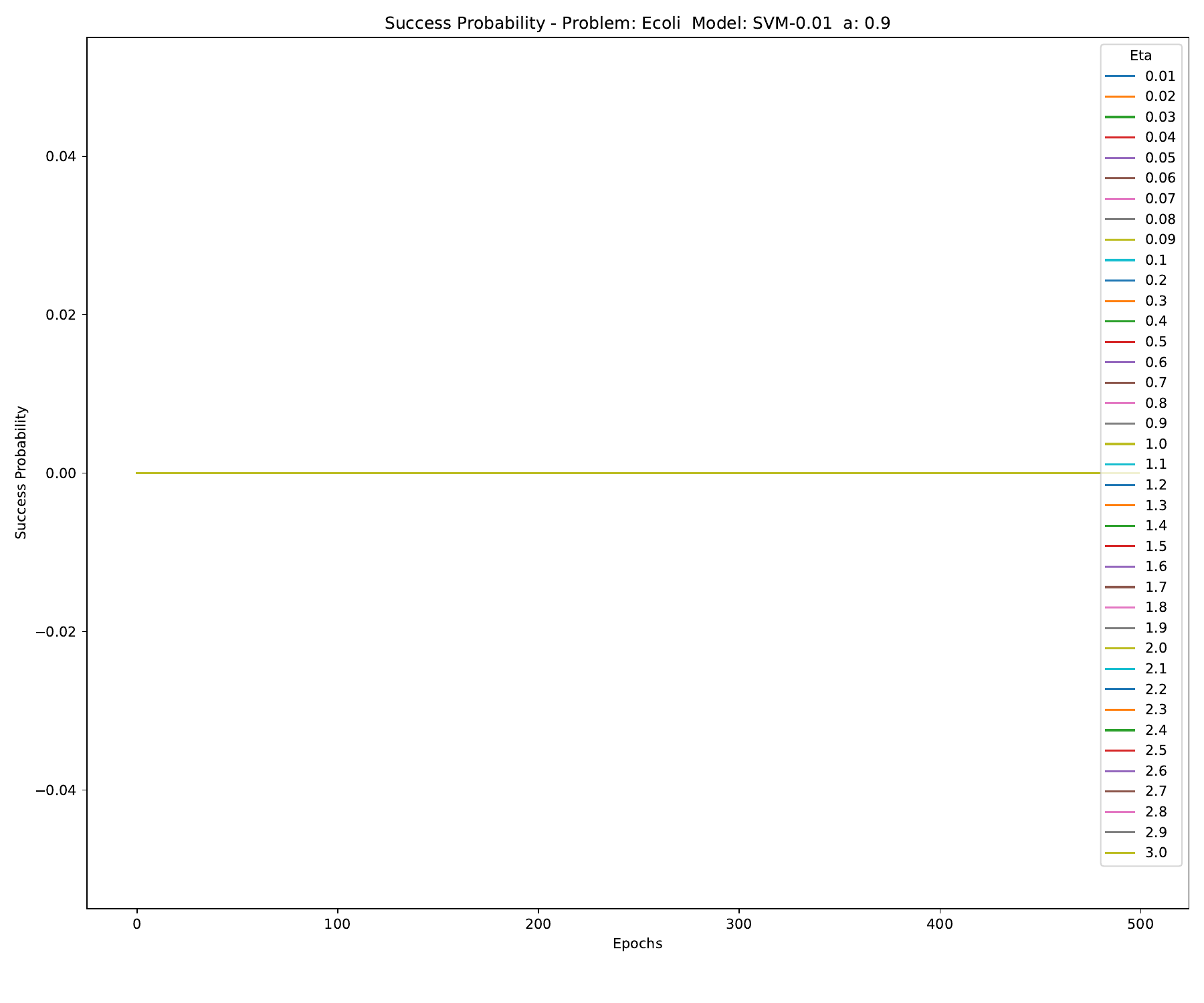} \\[-3mm]
  \end{tabular}
\caption{Success probability ($a=0.9$) for SVM-0.01 model.}
  \label{fig:success_probability_SVM_0.01_model_0.9}
\end{figure*}

\begin{figure*}[p]
  \centering
  \ContinuedFloat
  \begin{tabular}{c@{}c}
    Training Success Probability & Validation Success Probability \\
    \includegraphics[clip,trim=0 0 0 0,width=.52\linewidth]{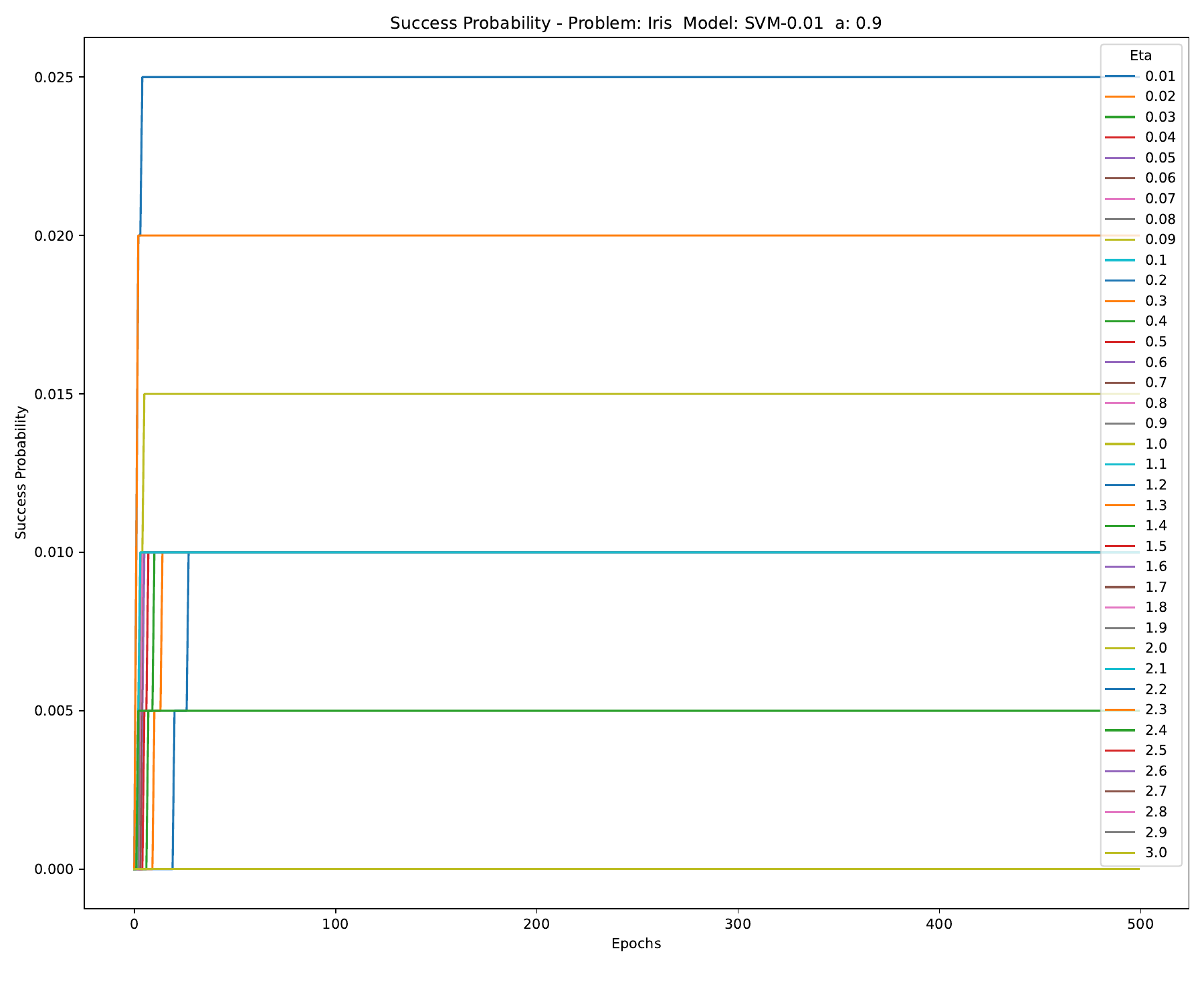} &
    \includegraphics[clip,trim=0 0 0 0,width=.52\linewidth]{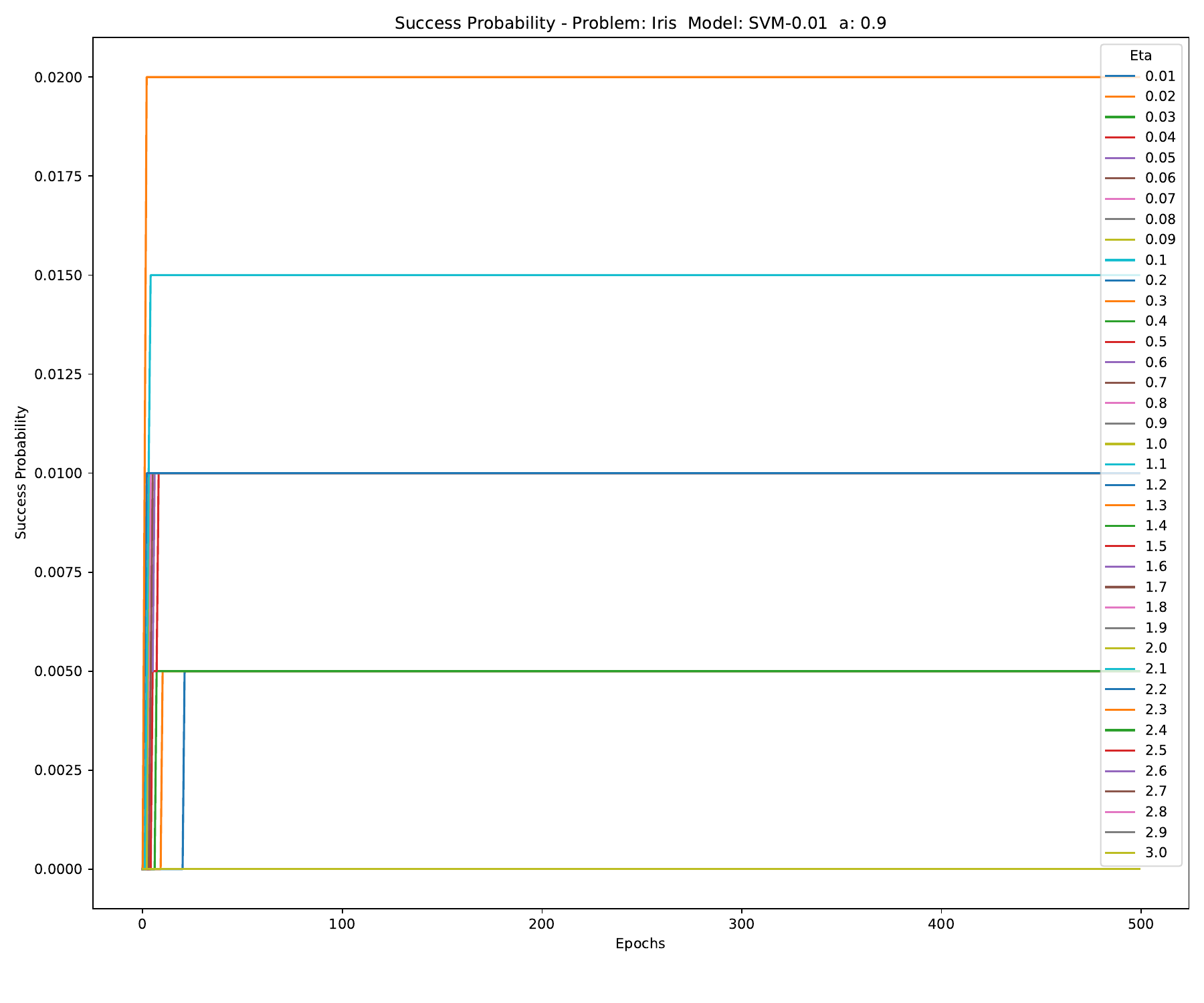} \\[-3mm]
    \includegraphics[clip,trim=0 0 0 0,width=.52\linewidth]{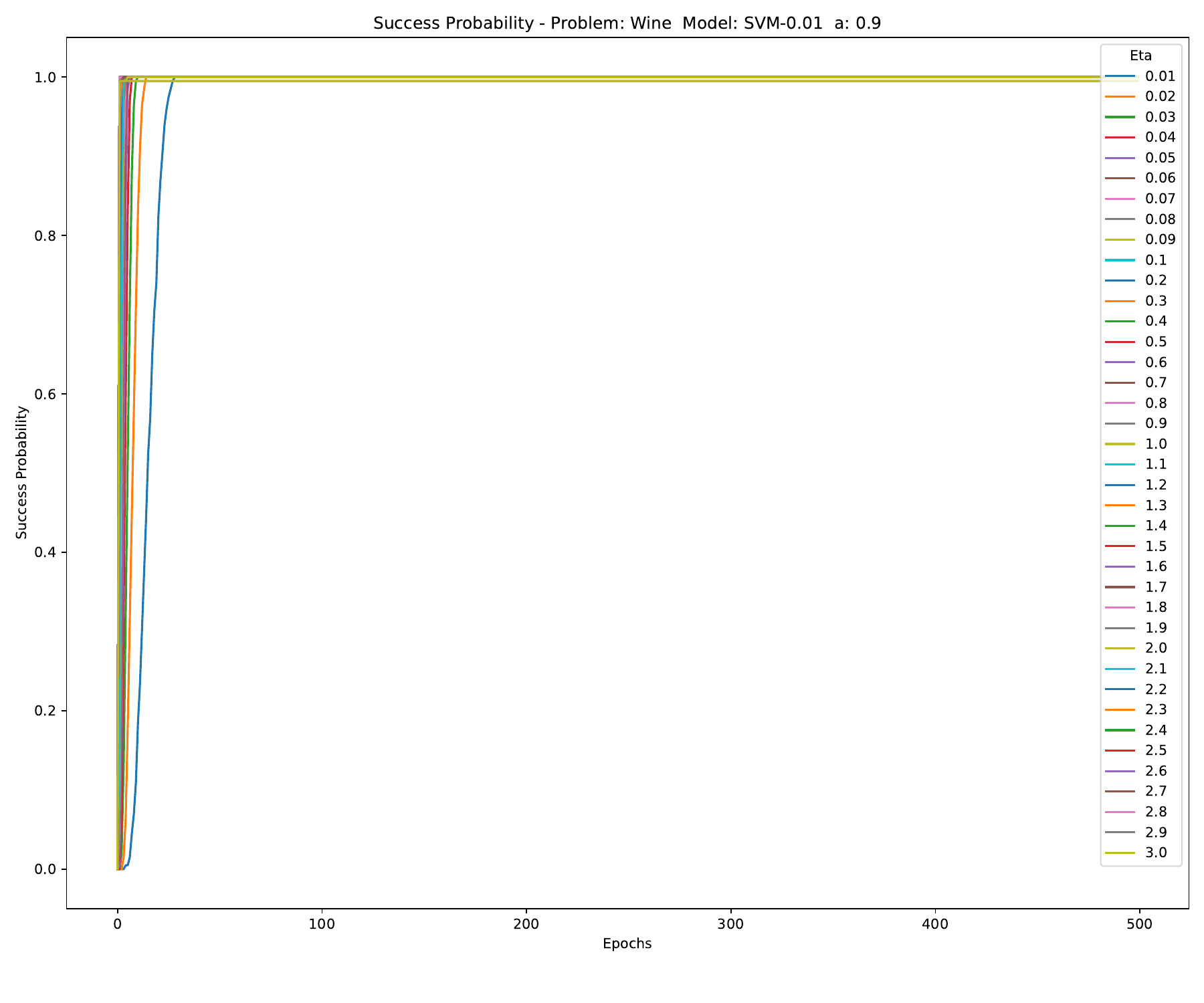} &
    \includegraphics[clip,trim=0 0 0 0,width=.52\linewidth]{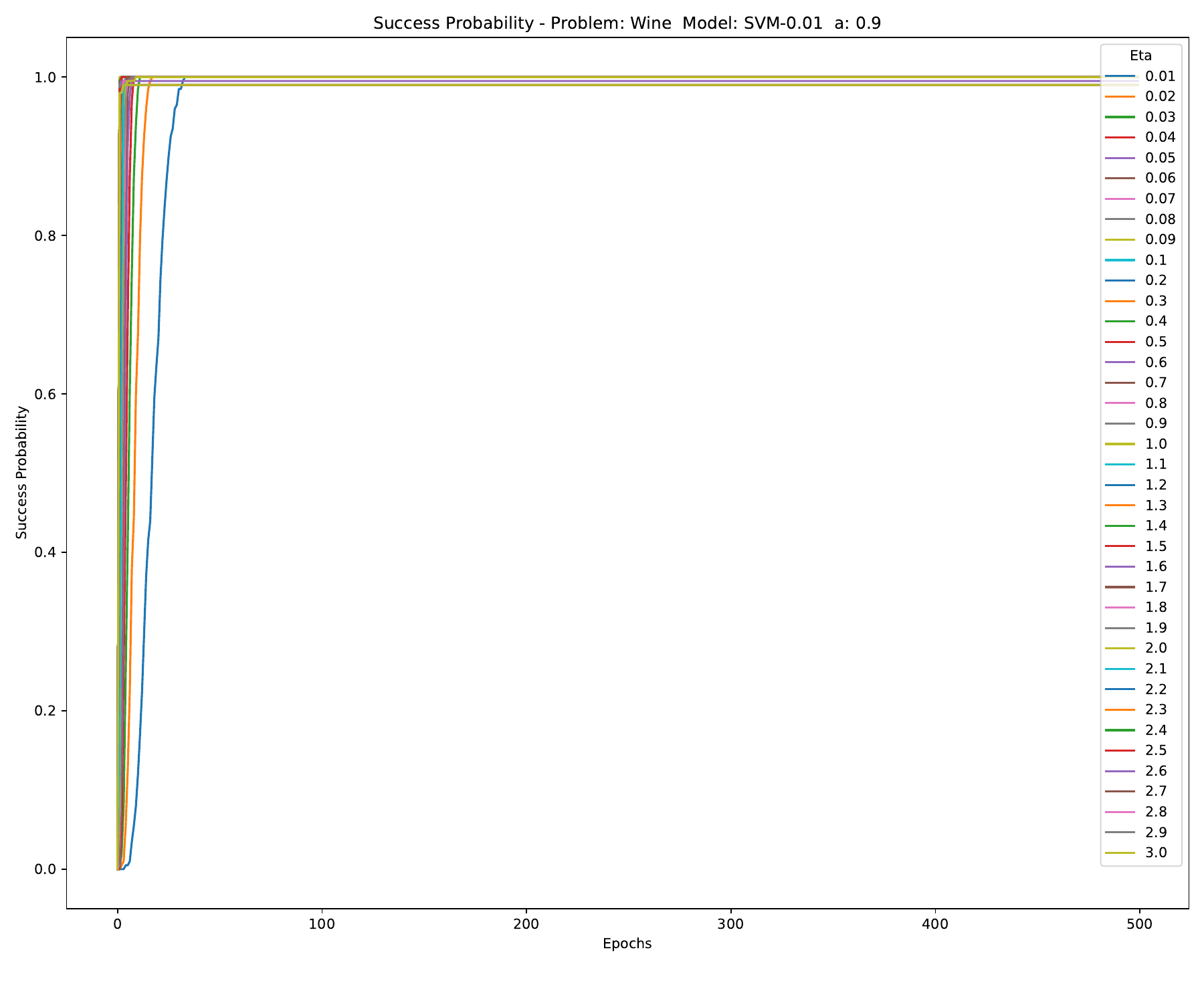} \\[-3mm]
  \end{tabular}
\caption{Success probability ($a=0.9$) for SVM-0.01 model (continued).}
\end{figure*}

\begin{figure*}[p]
  \centering
  \begin{tabular}{c@{}c}
    Training Success Probability & Validation Success Probability \\
    \includegraphics[clip,trim=0 0 0 0,width=.52\linewidth]{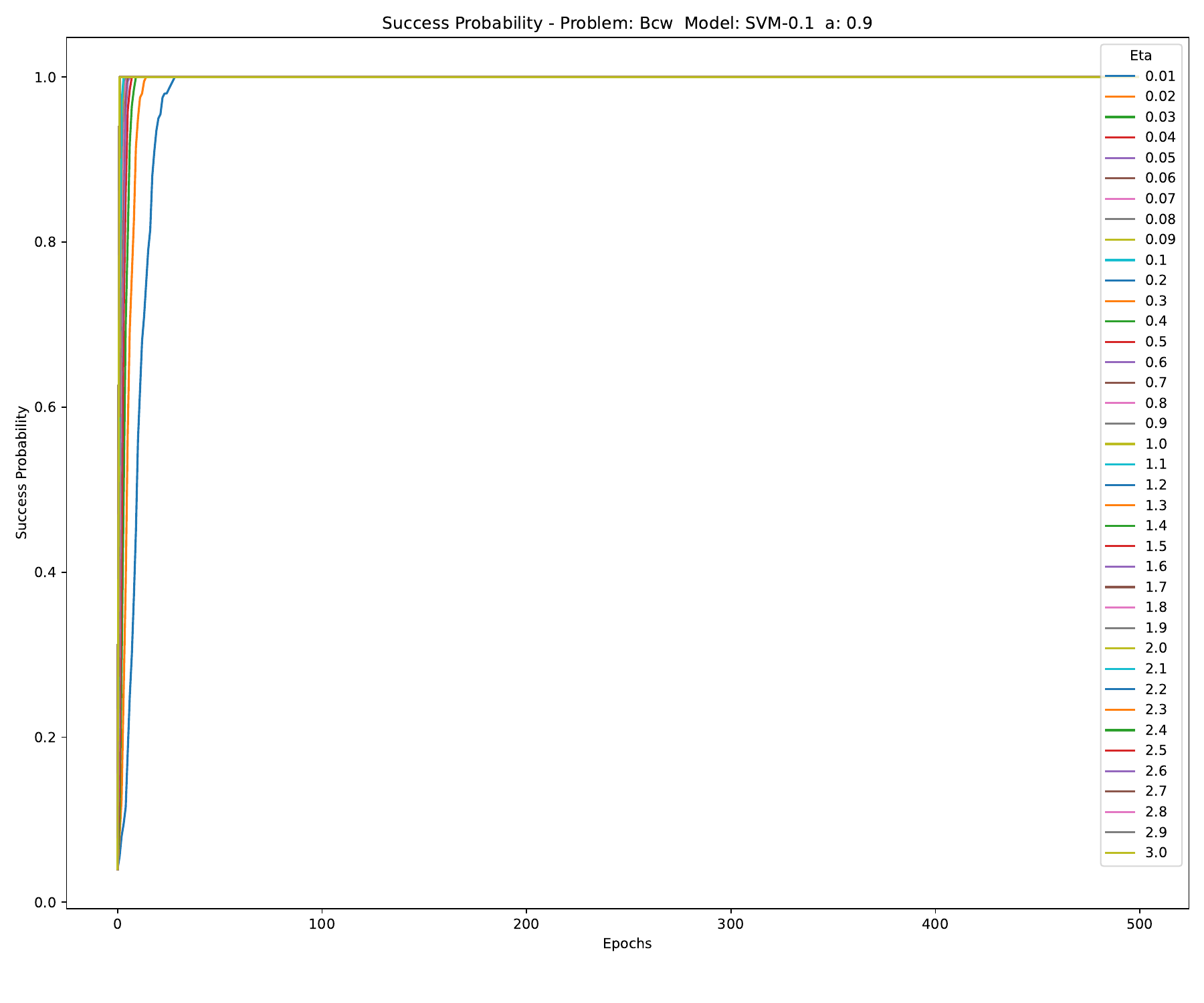} &
    \includegraphics[clip,trim=0 0 0 0,width=.52\linewidth]{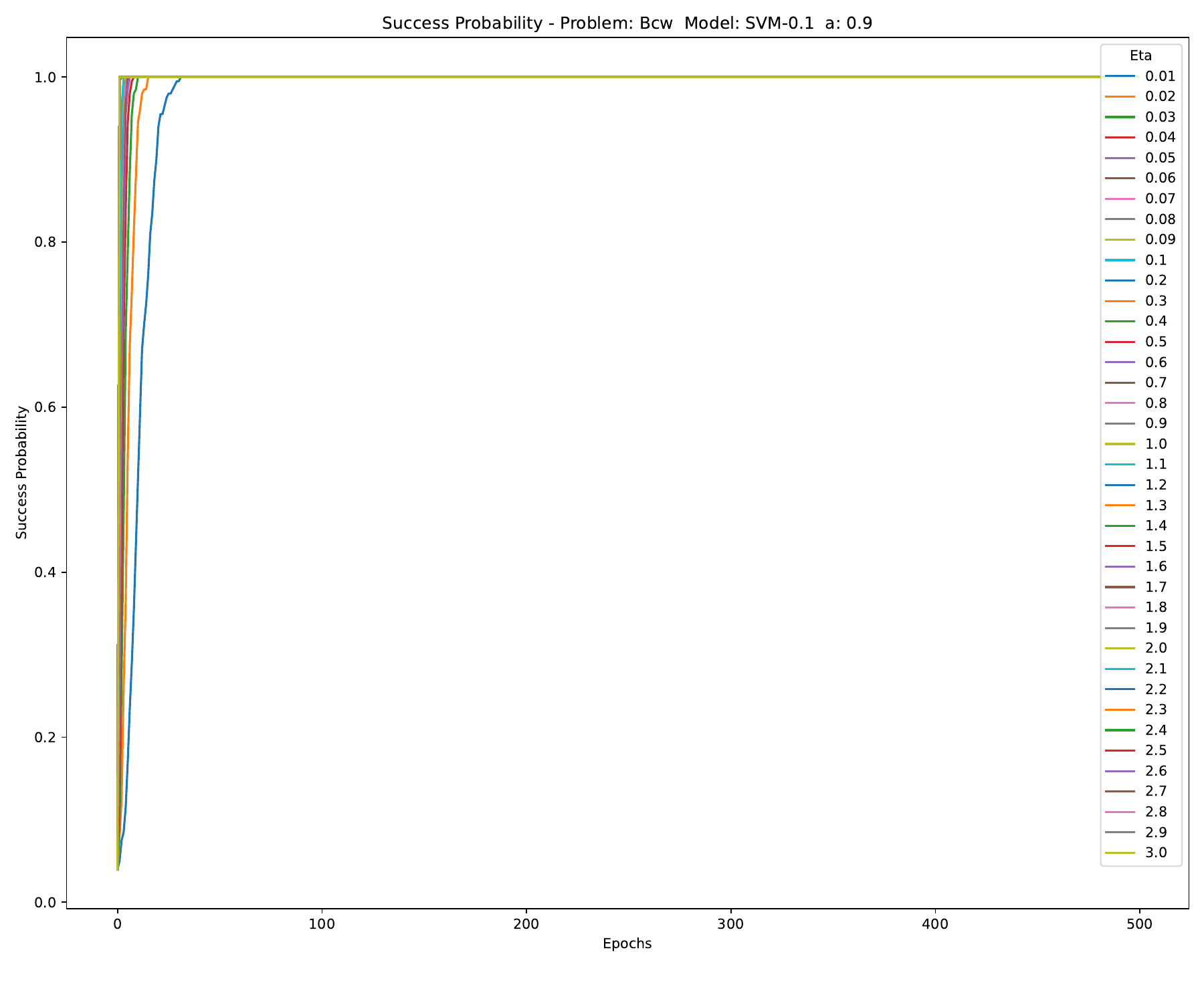} \\[-3mm]
    \includegraphics[clip,trim=0 0 0 0,width=.52\linewidth]{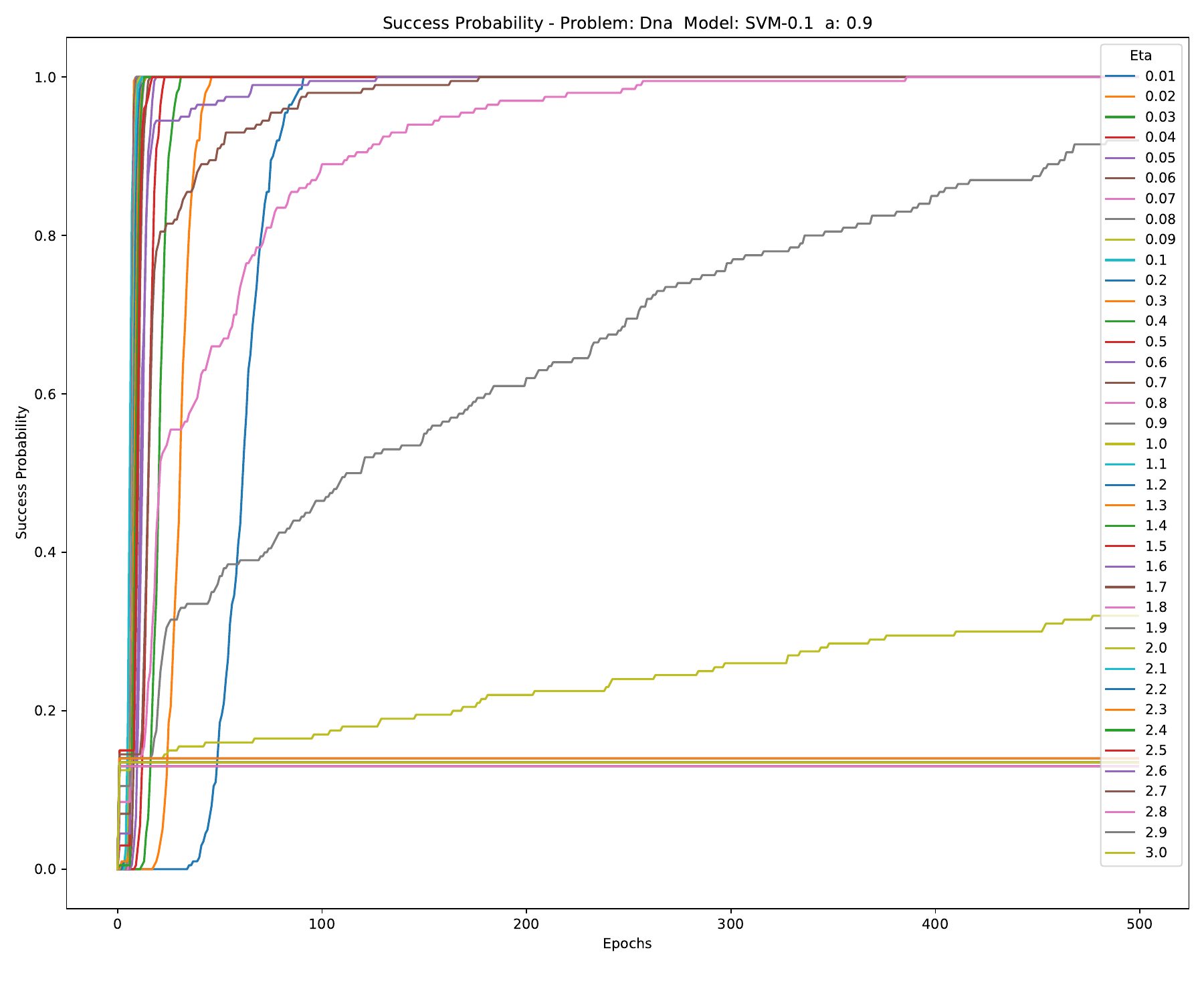} &
    \includegraphics[clip,trim=0 0 0 0,width=.52\linewidth]{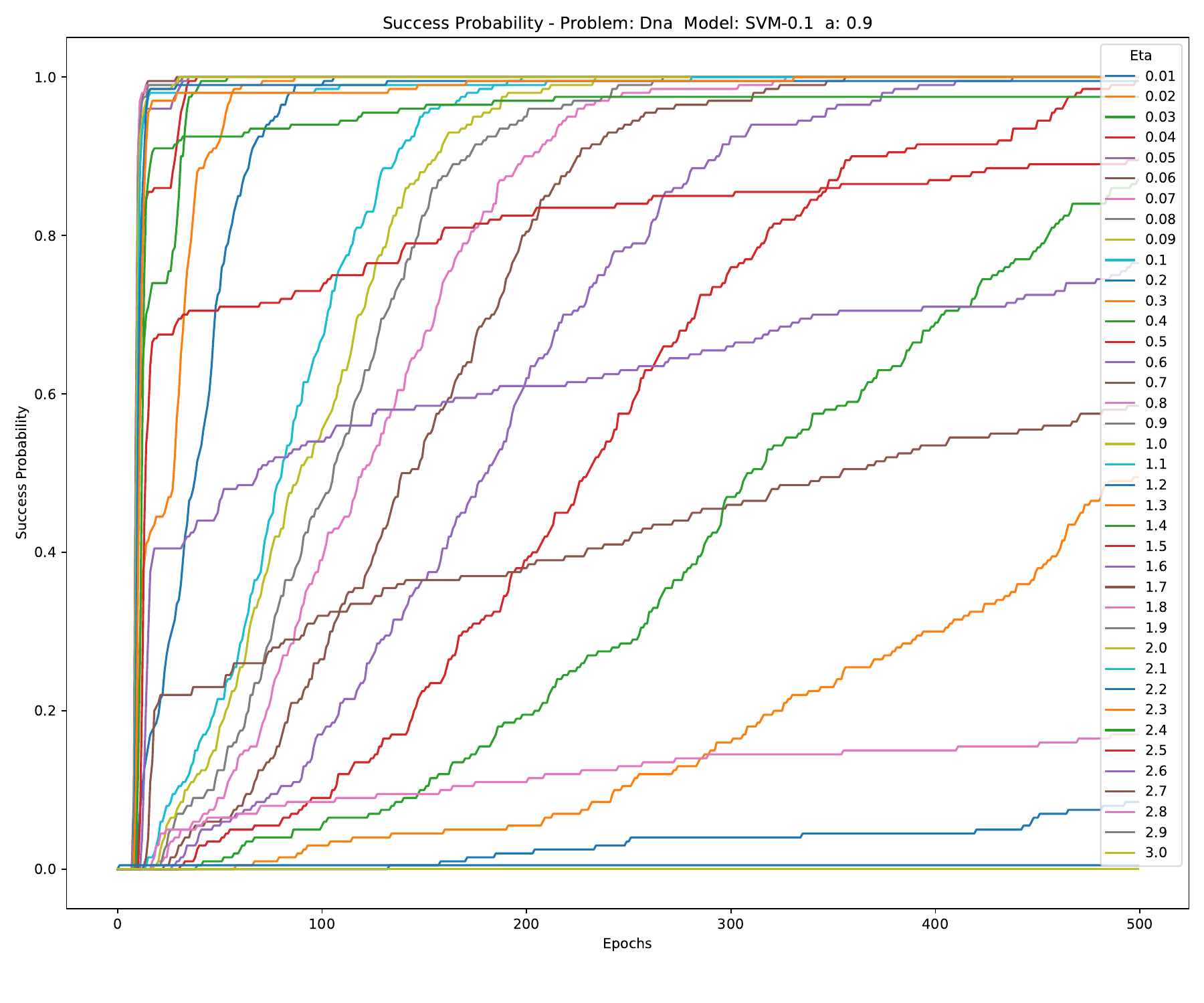} \\[-3mm]
    \includegraphics[clip,trim=0 0 0 0,width=.52\linewidth]{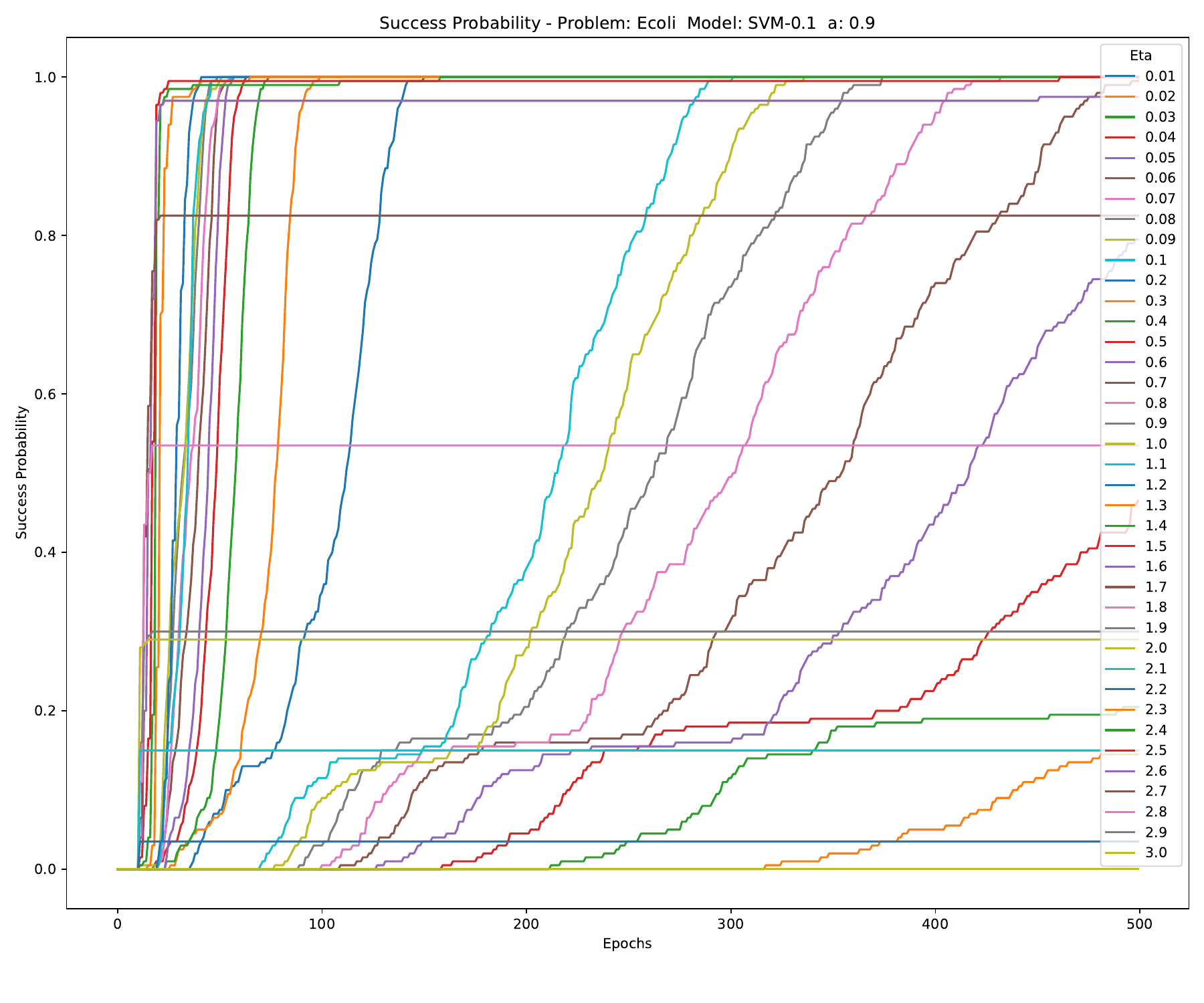} &
    \includegraphics[clip,trim=0 0 0 0,width=.52\linewidth]{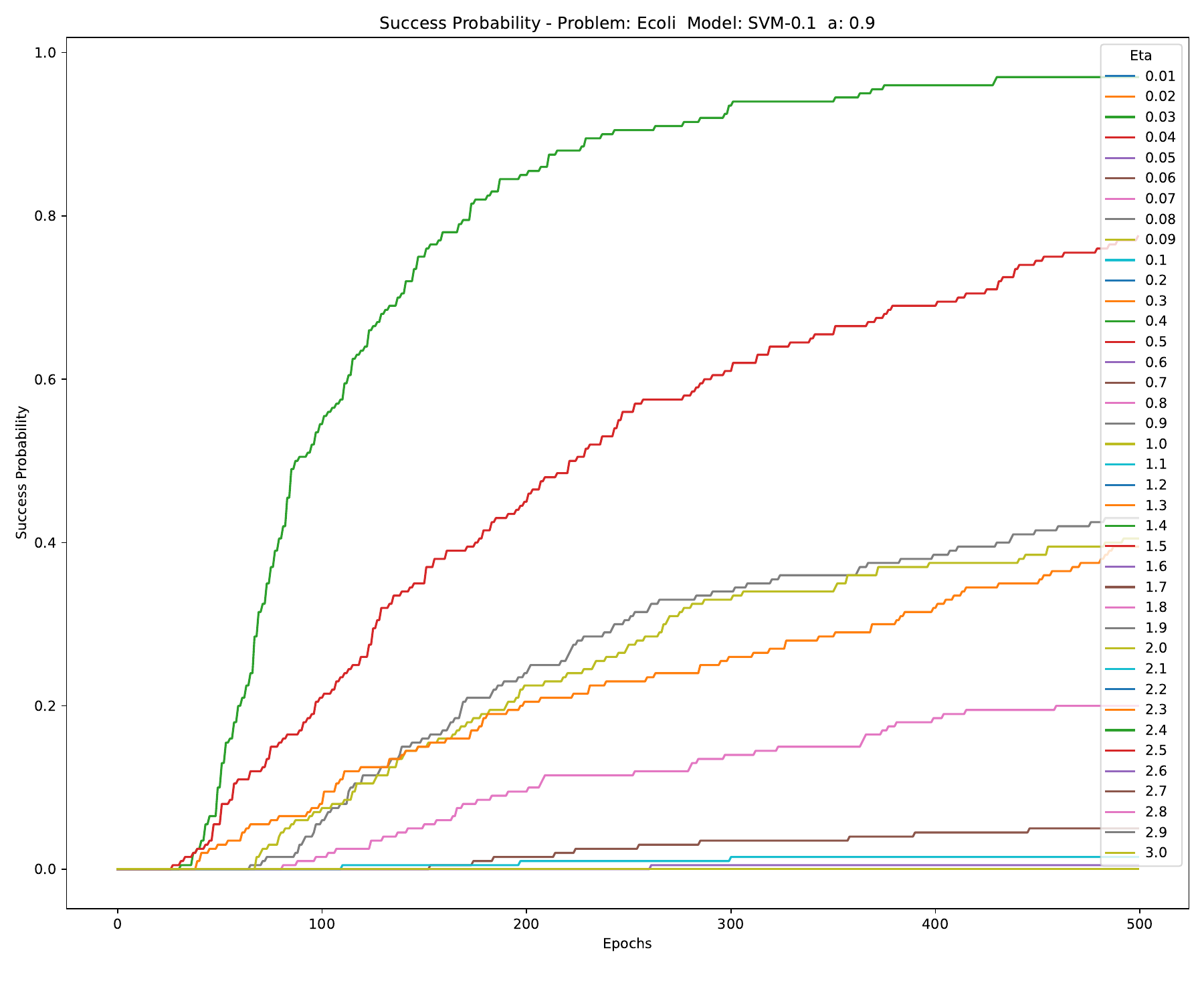} \\[-3mm]
  \end{tabular}
\caption{Success probability ($a=0.9$) for SVM-0.1 model.}
  \label{fig:success_probability_SVM_0.1_model_0.9}
\end{figure*}

\begin{figure*}[p]
  \centering
  \ContinuedFloat
  \begin{tabular}{c@{}c}
    Training Success Probability & Validation Success Probability \\
    \includegraphics[clip,trim=0 0 0 0,width=.52\linewidth]{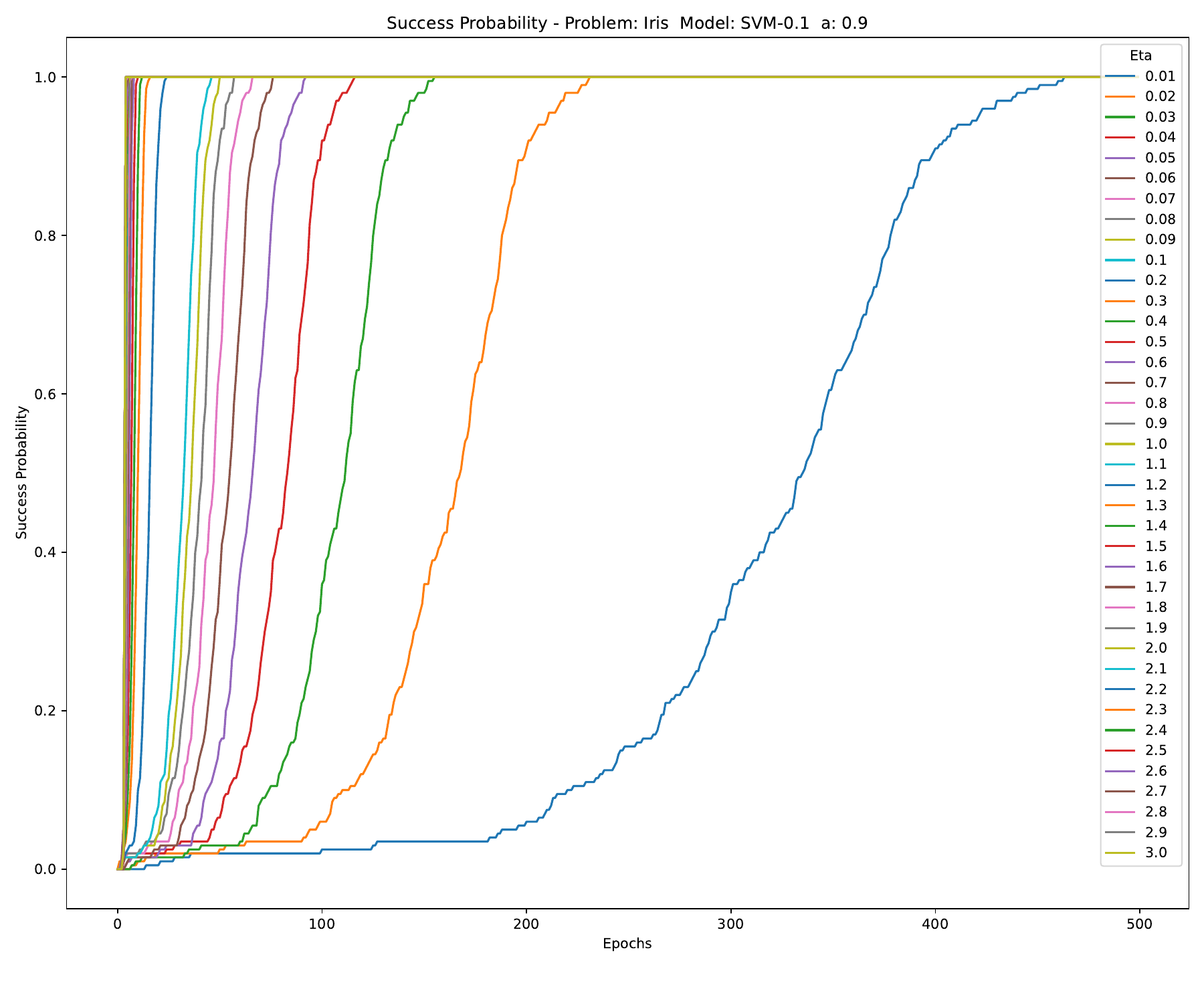} &
    \includegraphics[clip,trim=0 0 0 0,width=.52\linewidth]{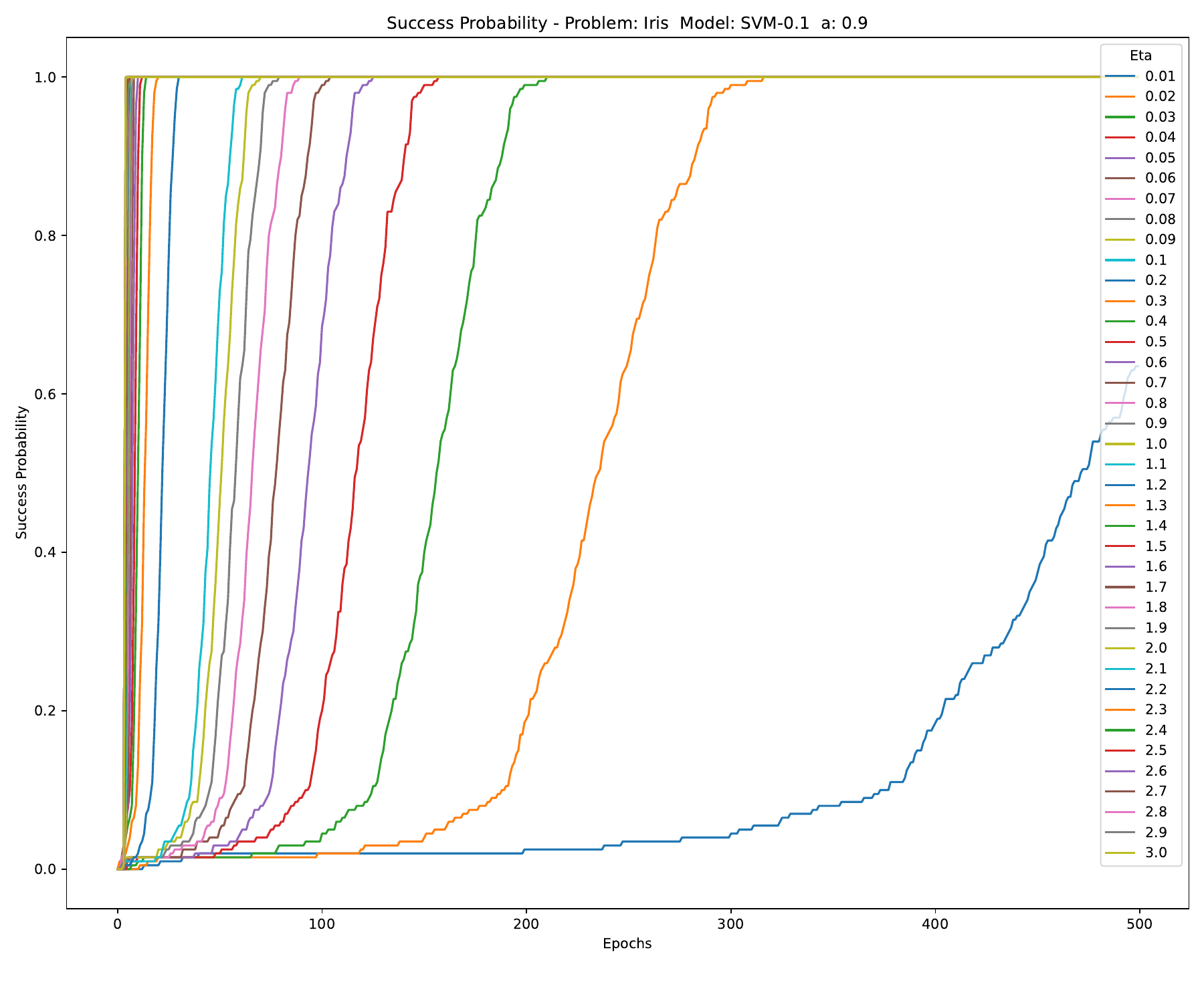} \\[-3mm]
    \includegraphics[clip,trim=0 0 0 0,width=.52\linewidth]{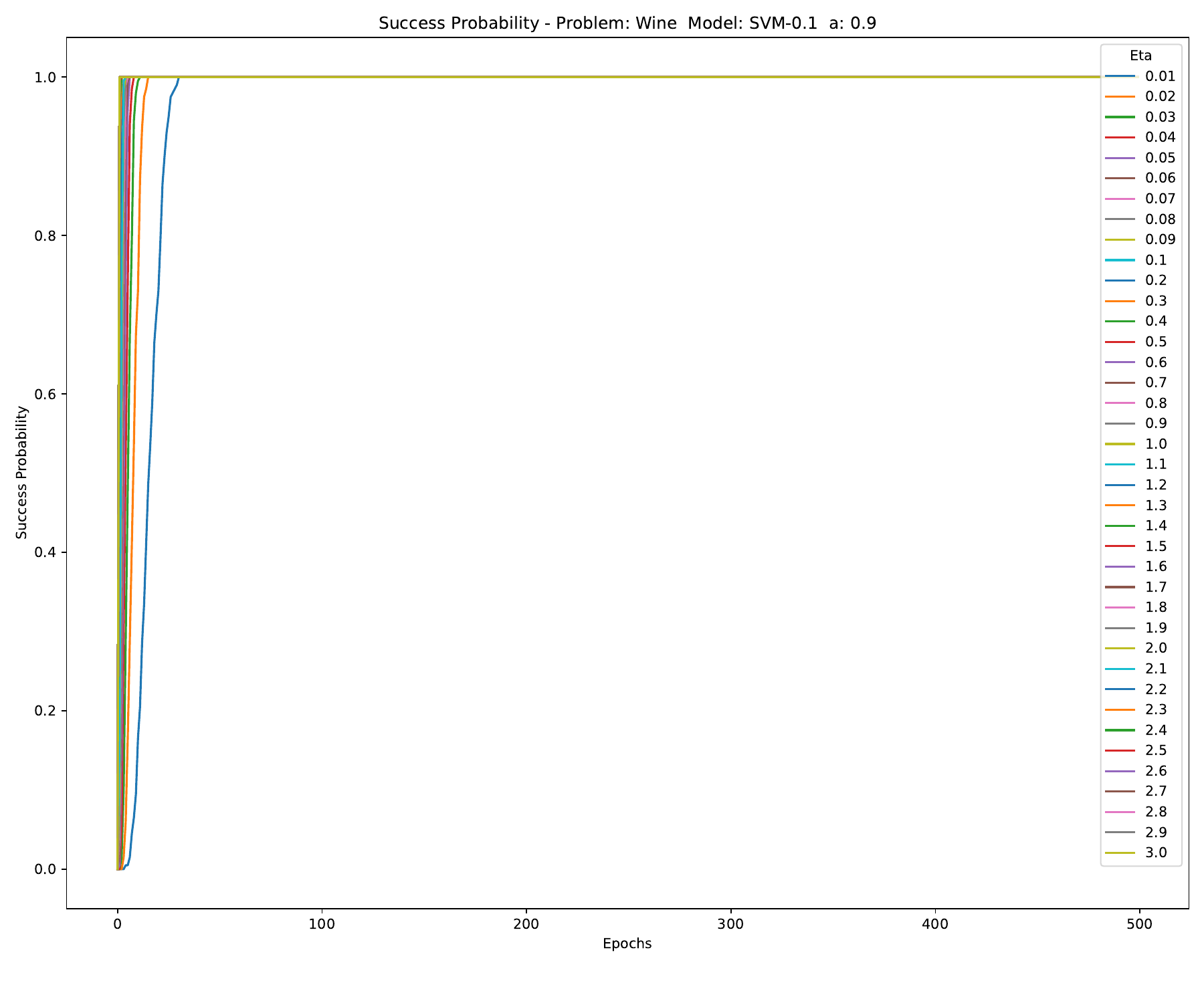} &
    \includegraphics[clip,trim=0 0 0 0,width=.52\linewidth]{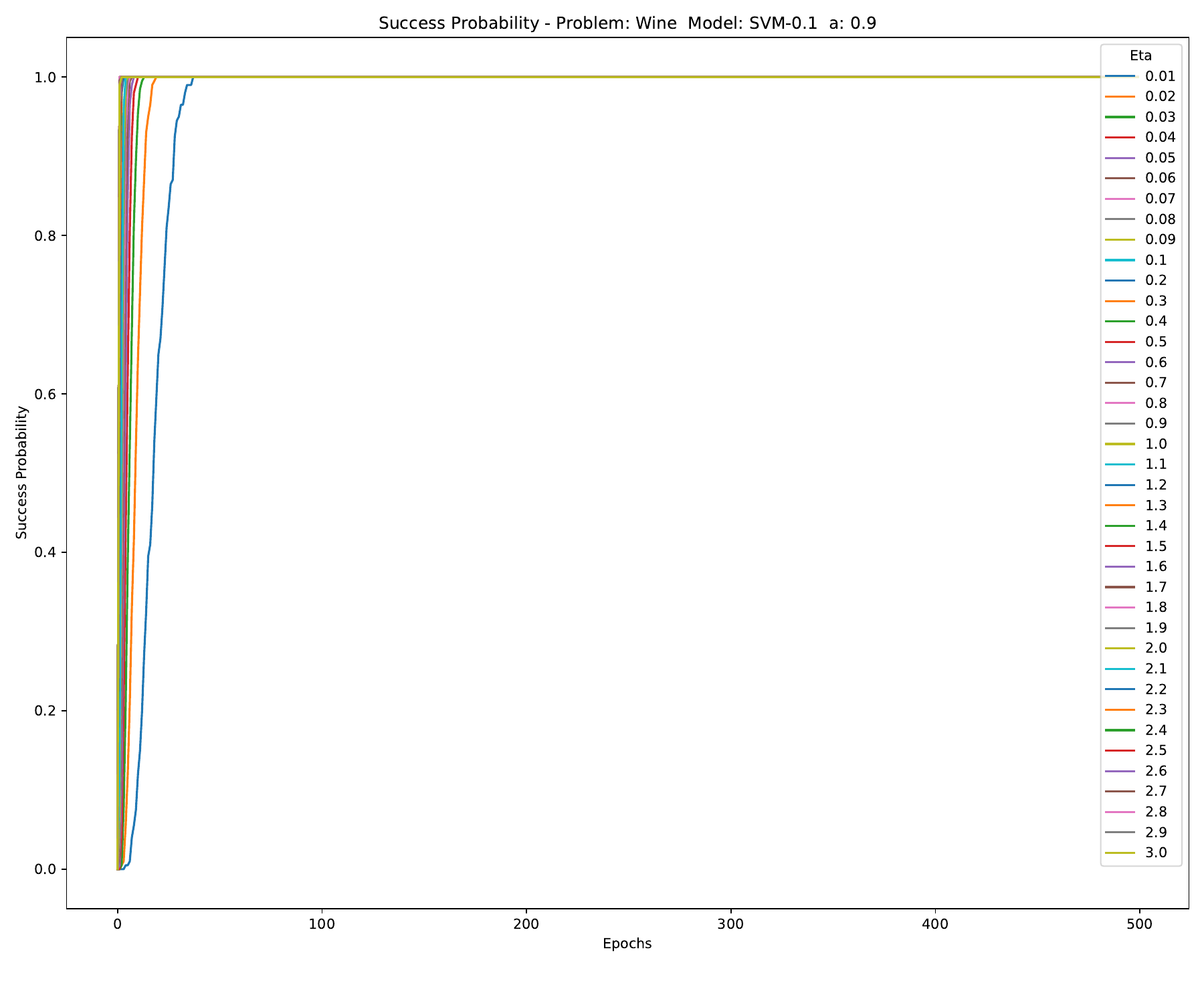} \\[-3mm]
  \end{tabular}
\caption{Success probability ($a=0.9$) for SVM-0.1 model (continued).}
\end{figure*}
\FloatBarrier

\subsection{Success Probabilities for $a=0.95$}

In
Figures~\ref{fig:success_probability_NN_ReLU_model_0.95}---\ref{fig:success_probability_SVM_0.1_model_0.95}
we present plots of the success probabilities $P(i, 0.95, \eta)$ (that
is for $a=0.95$) for all the problems and models used in this article.

\begin{figure*}[p]
  \centering
  \begin{tabular}{c@{}c}
    Training Success Probability & Validation Success Probability \\
    \includegraphics[clip,trim=0 0 0 0,width=.52\linewidth]{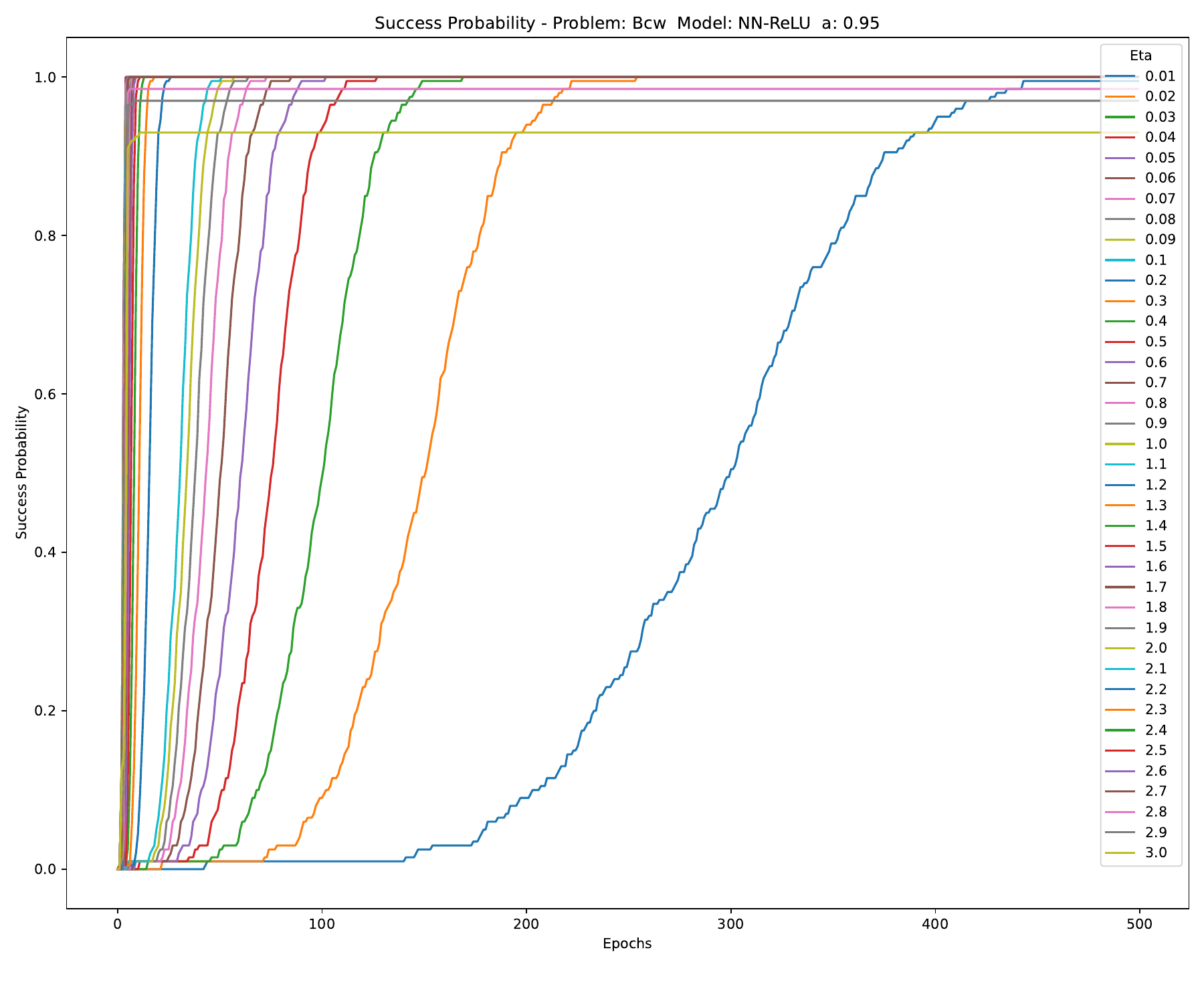} &
    \includegraphics[clip,trim=0 0 0 0,width=.52\linewidth]{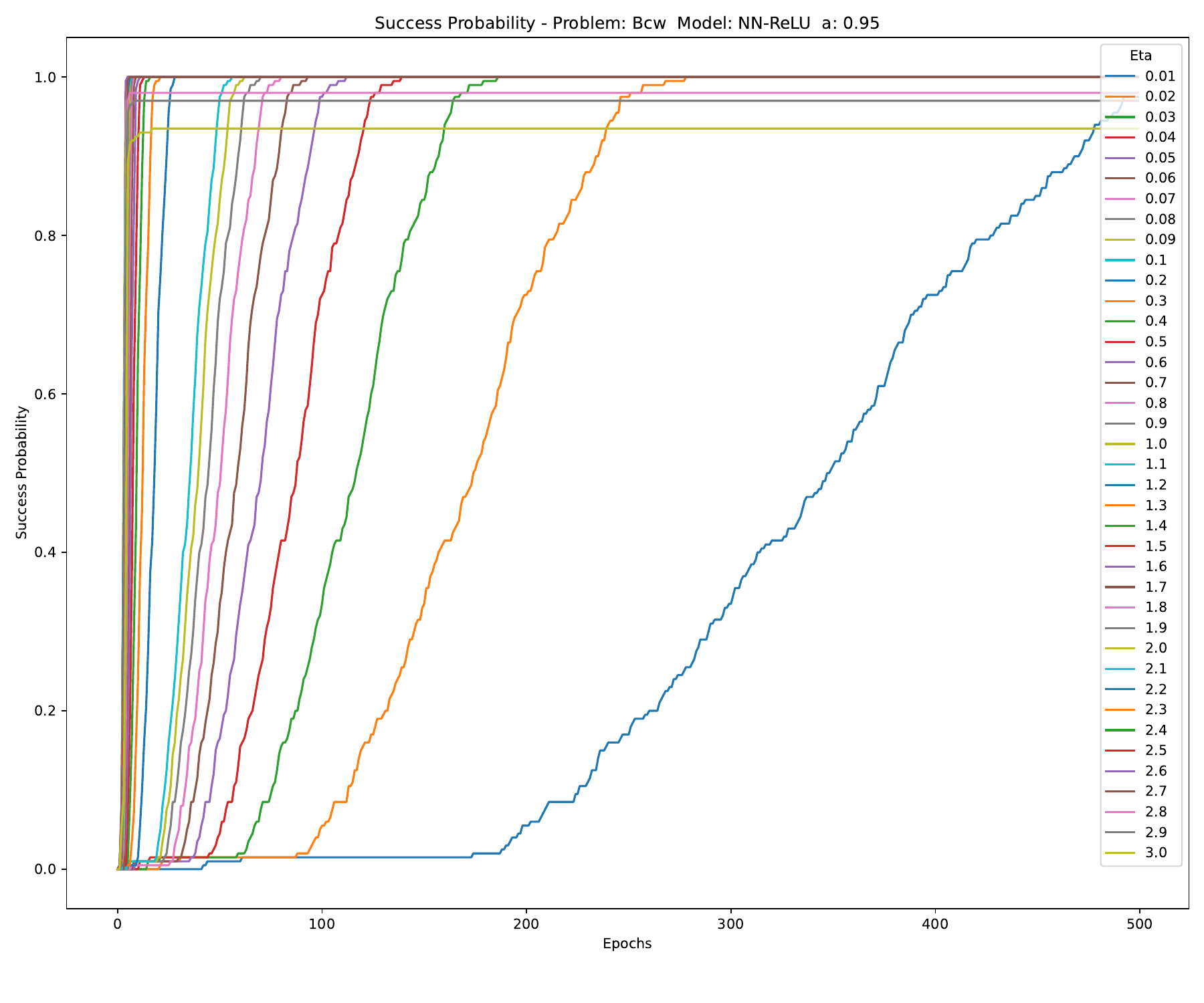} \\[-3mm]
    \includegraphics[clip,trim=0 0 0 0,width=.52\linewidth]{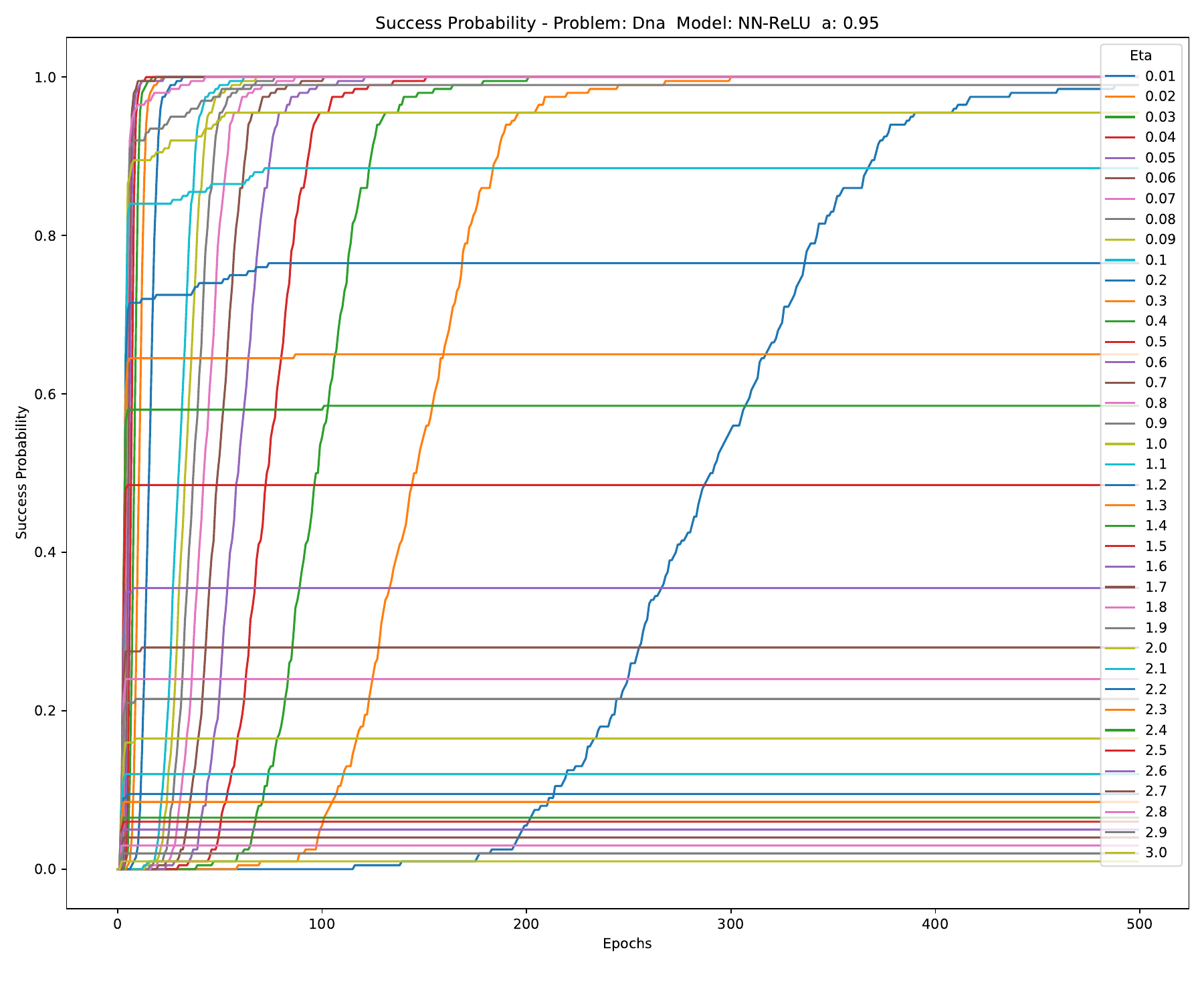} &
    \includegraphics[clip,trim=0 0 0 0,width=.52\linewidth]{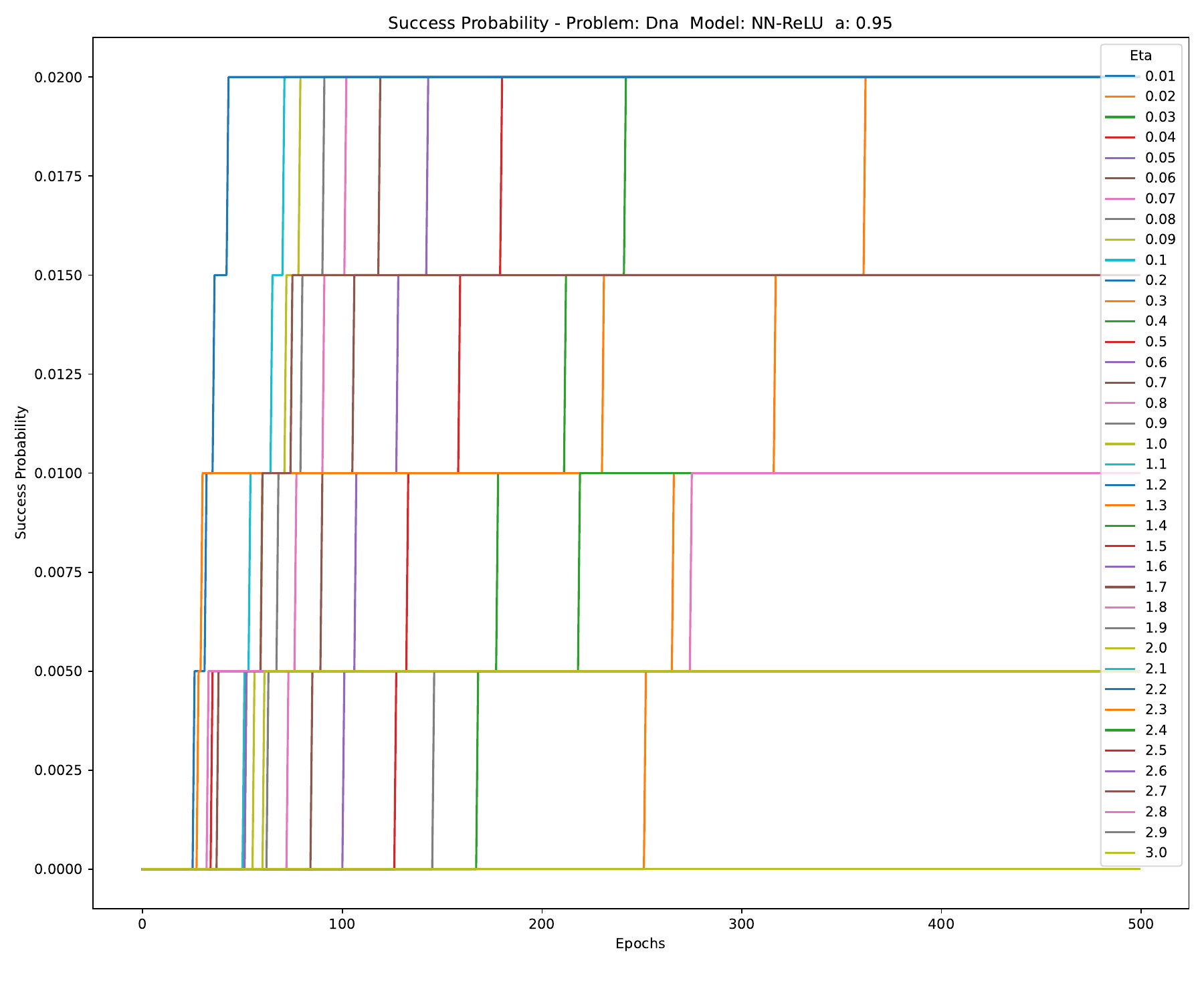} \\[-3mm]
    \includegraphics[clip,trim=0 0 0 0,width=.52\linewidth]{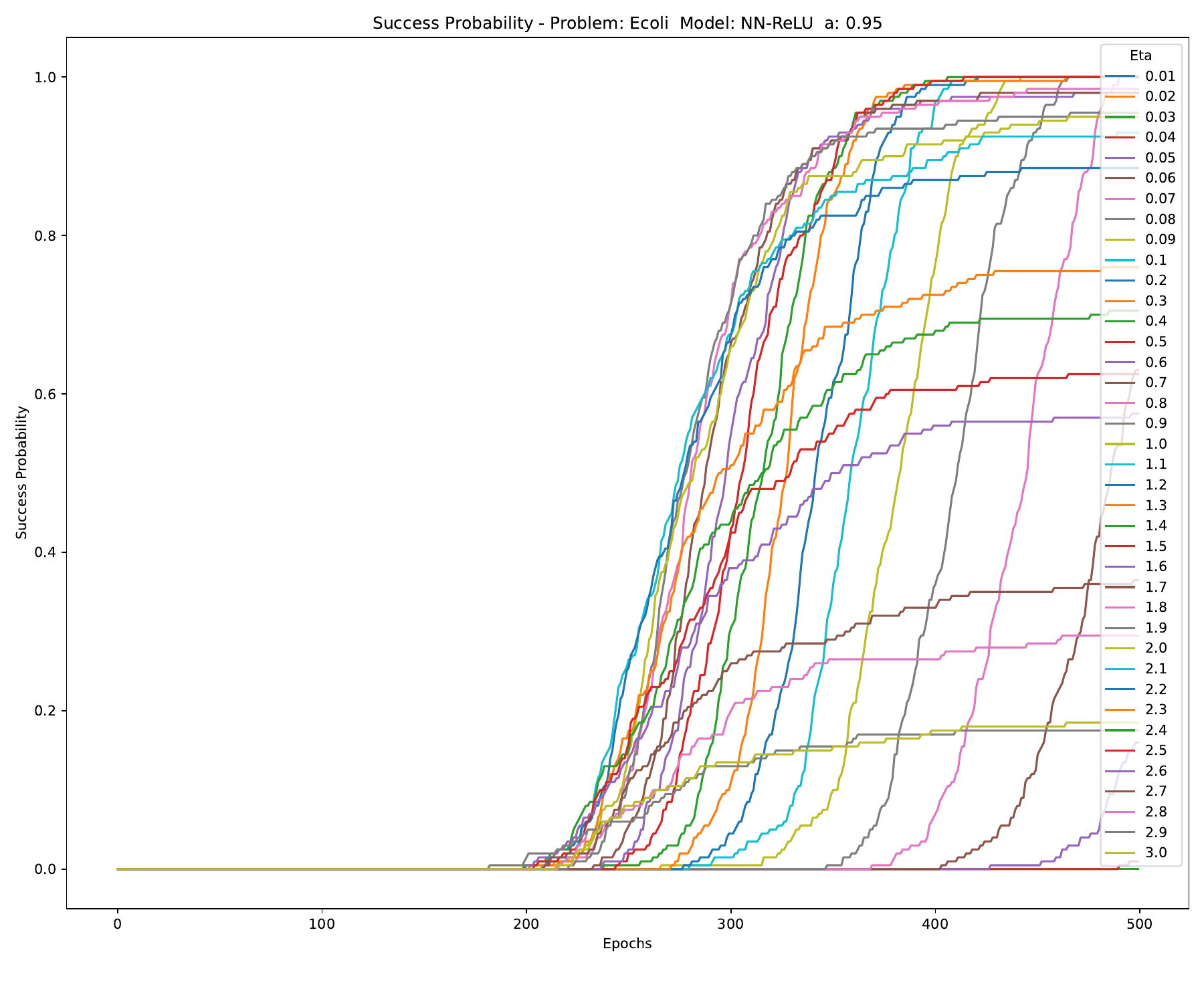} &
    \includegraphics[clip,trim=0 0 0 0,width=.52\linewidth]{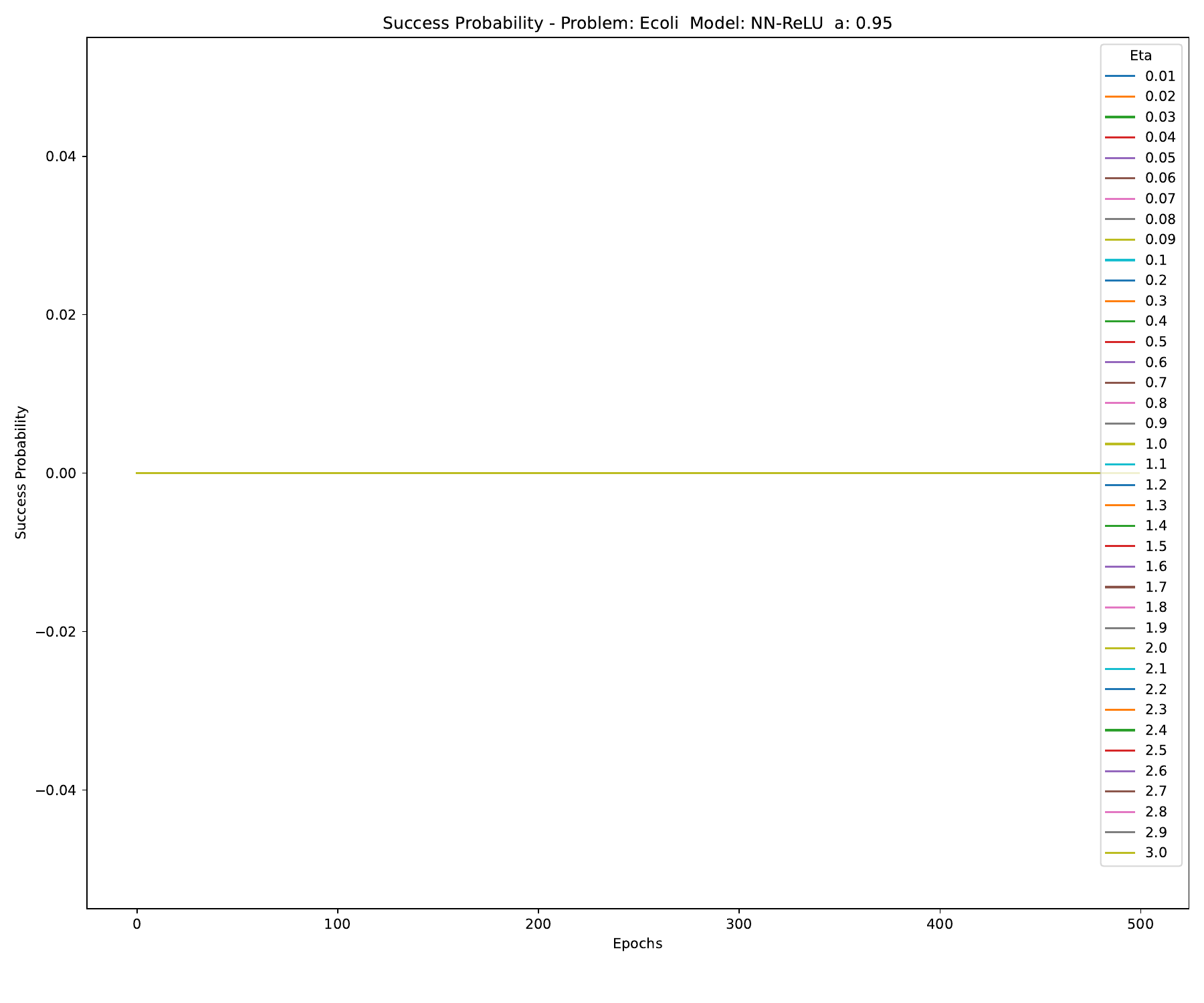} \\[-3mm]
  \end{tabular}
\caption{Success probability ($a=0.95$) for NN-ReLU model.}
  \label{fig:success_probability_NN_ReLU_model_0.95}
\end{figure*}

\begin{figure*}[p]
  \centering
  \ContinuedFloat
  \begin{tabular}{c@{}c}
    Training Success Probability & Validation Success Probability \\
    \includegraphics[clip,trim=0 0 0 0,width=.52\linewidth]{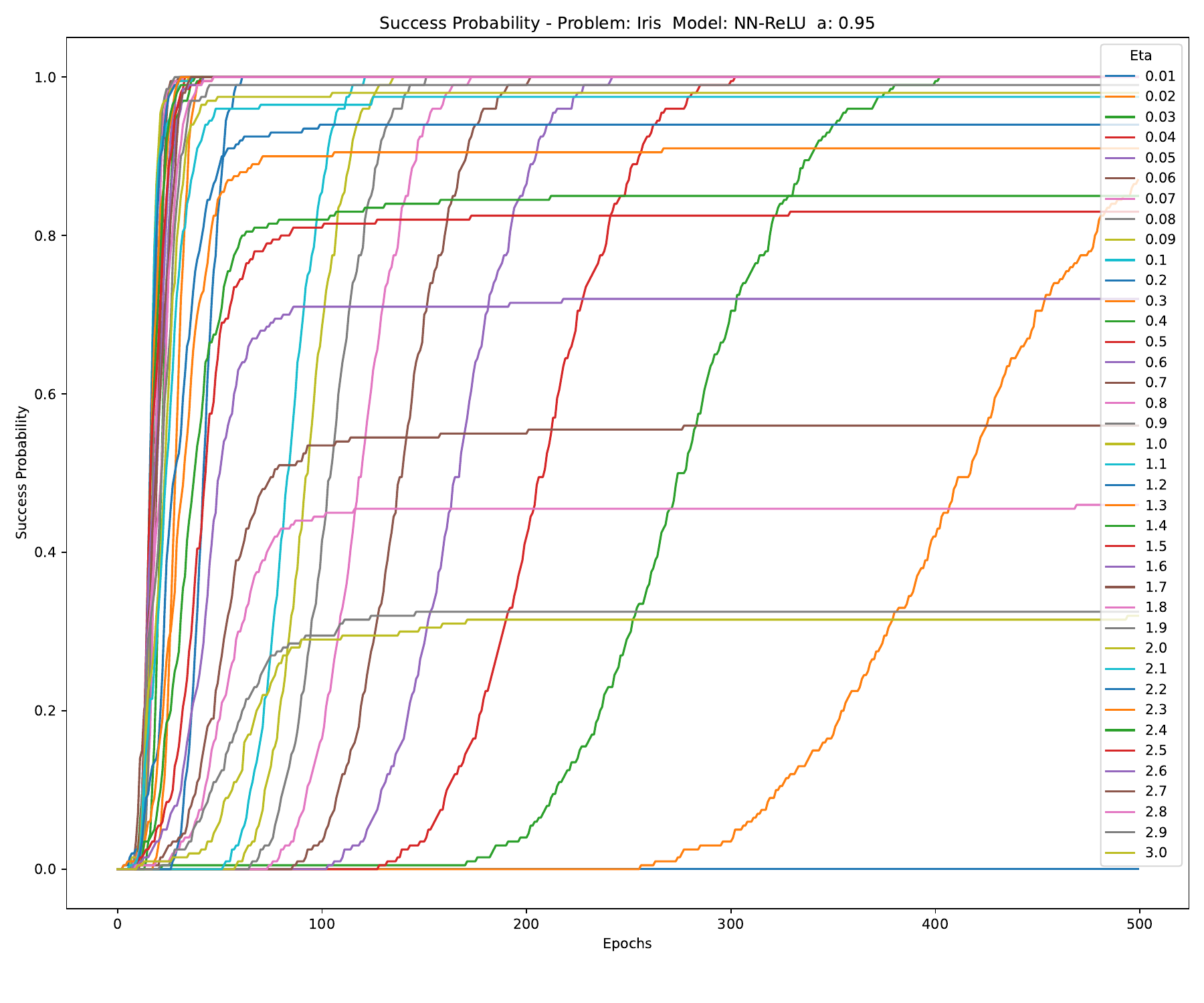} &
    \includegraphics[clip,trim=0 0 0 0,width=.52\linewidth]{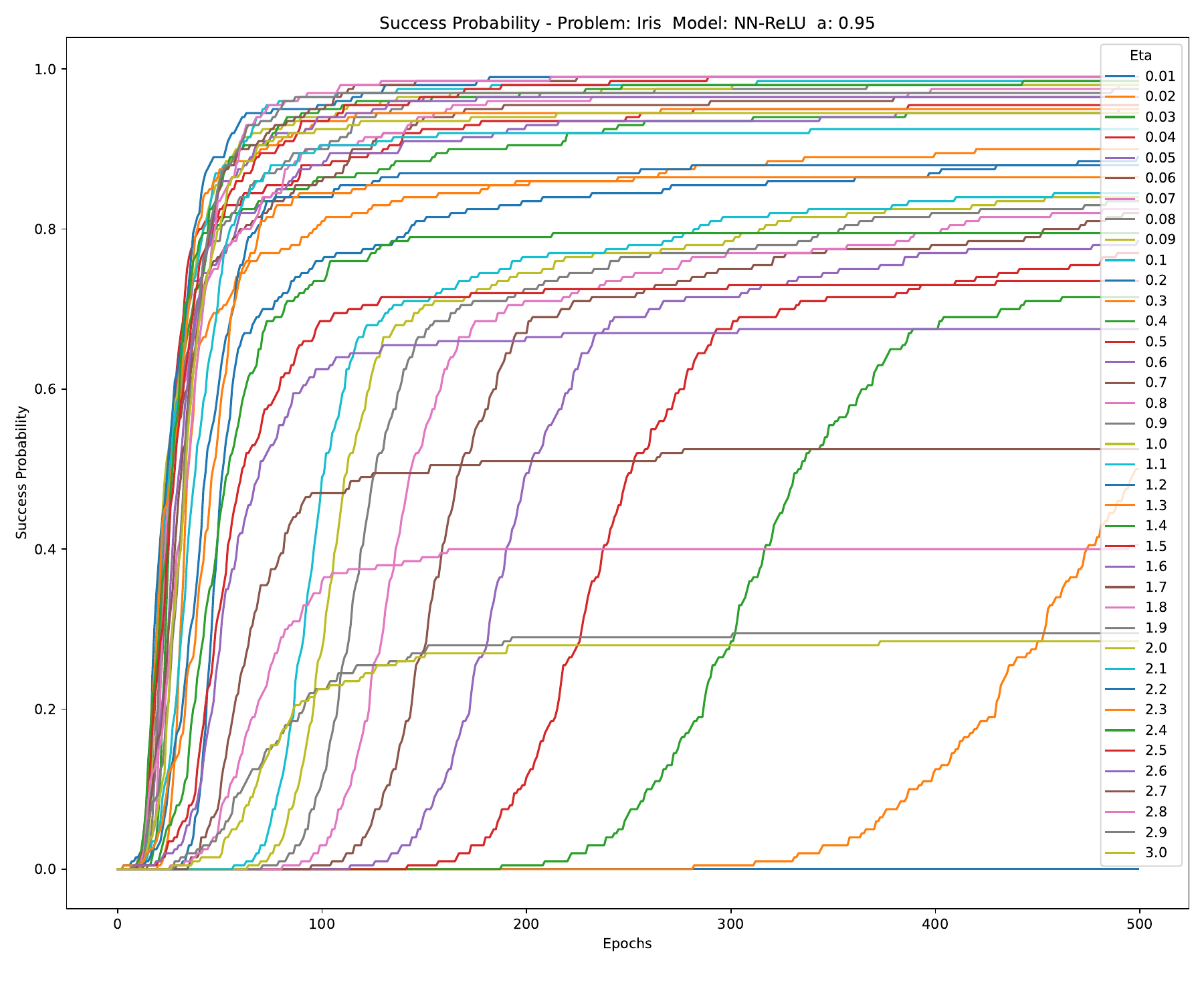} \\[-3mm]
    \includegraphics[clip,trim=0 0 0 0,width=.52\linewidth]{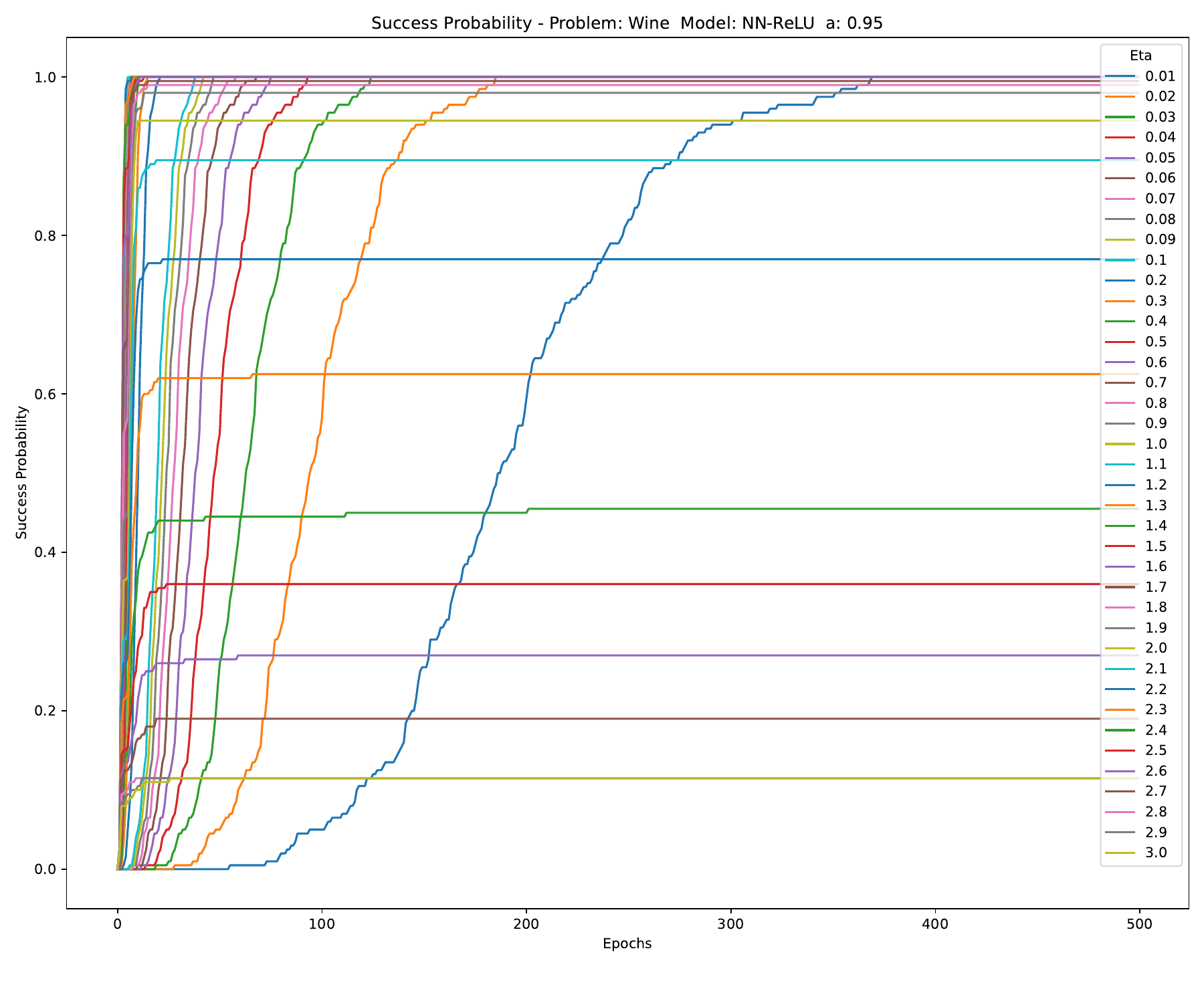} &
    \includegraphics[clip,trim=0 0 0 0,width=.52\linewidth]{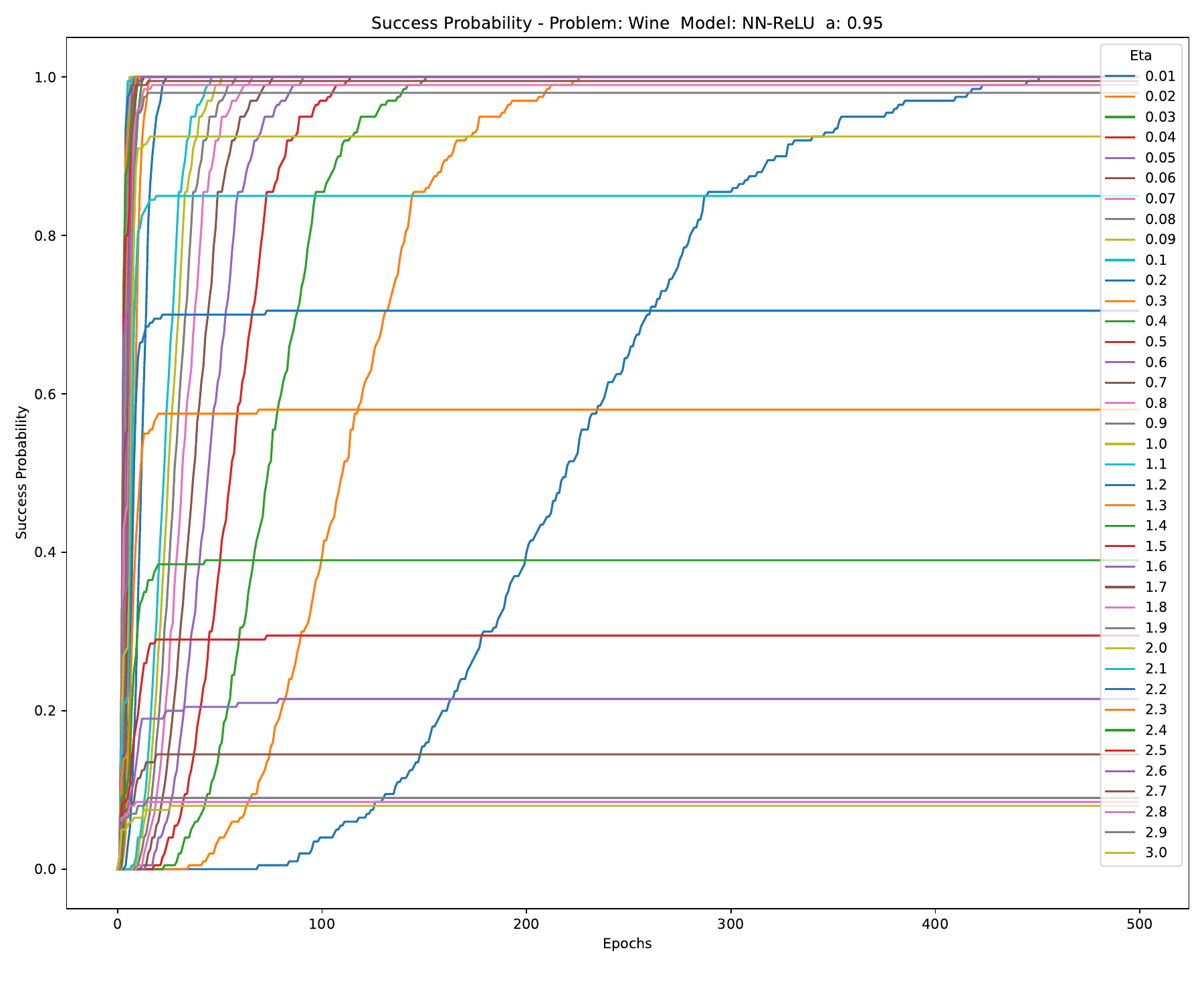} \\[-3mm]
  \end{tabular}
\caption{Success probability ($a=0.95$) for NN-ReLU model (continued).}
\end{figure*}

\begin{figure*}[p]
  \centering
  \begin{tabular}{c@{}c}
    Training Success Probability & Validation Success Probability \\
    \includegraphics[clip,trim=0 0 0 0,width=.52\linewidth]{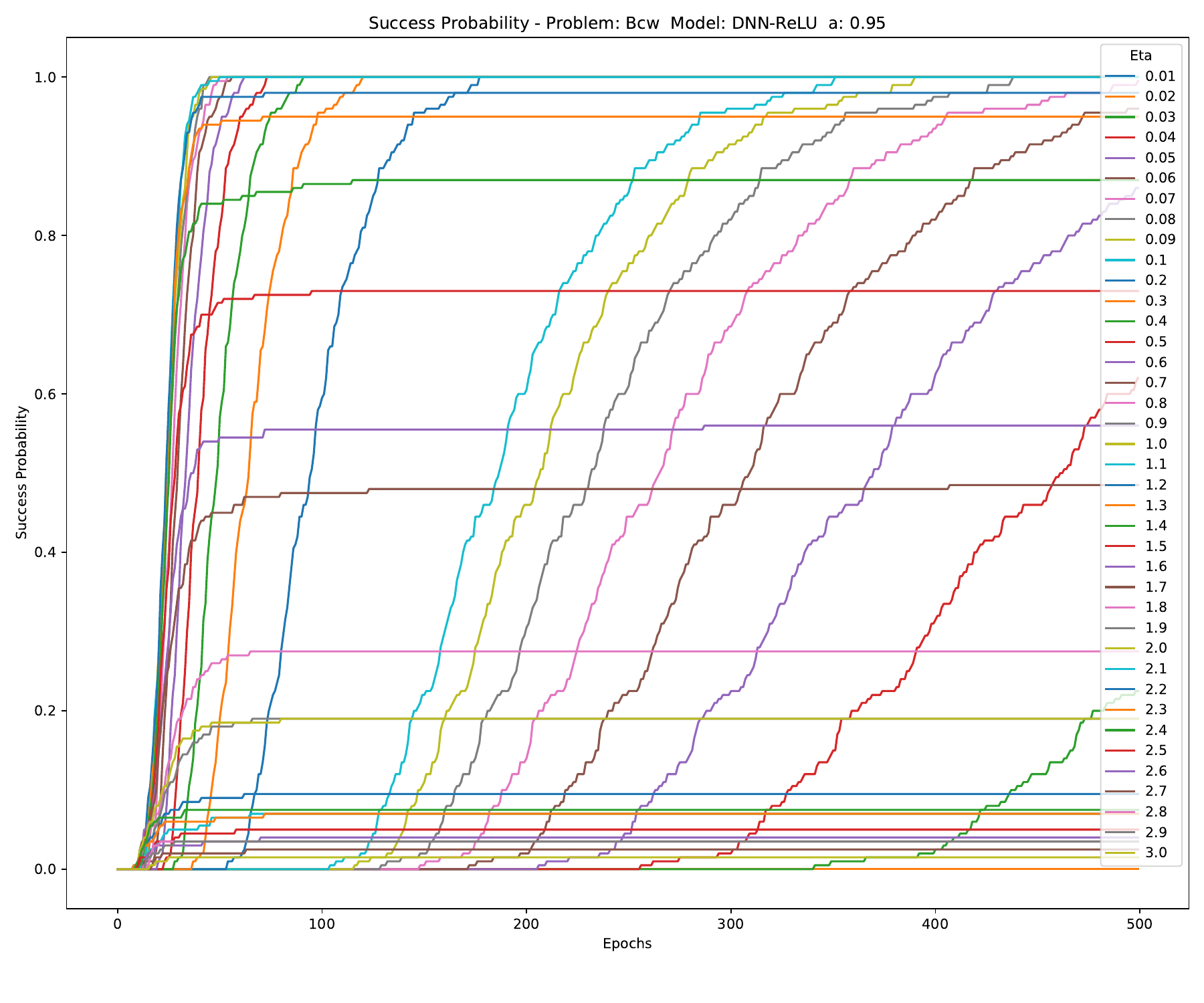} &
    \includegraphics[clip,trim=0 0 0 0,width=.52\linewidth]{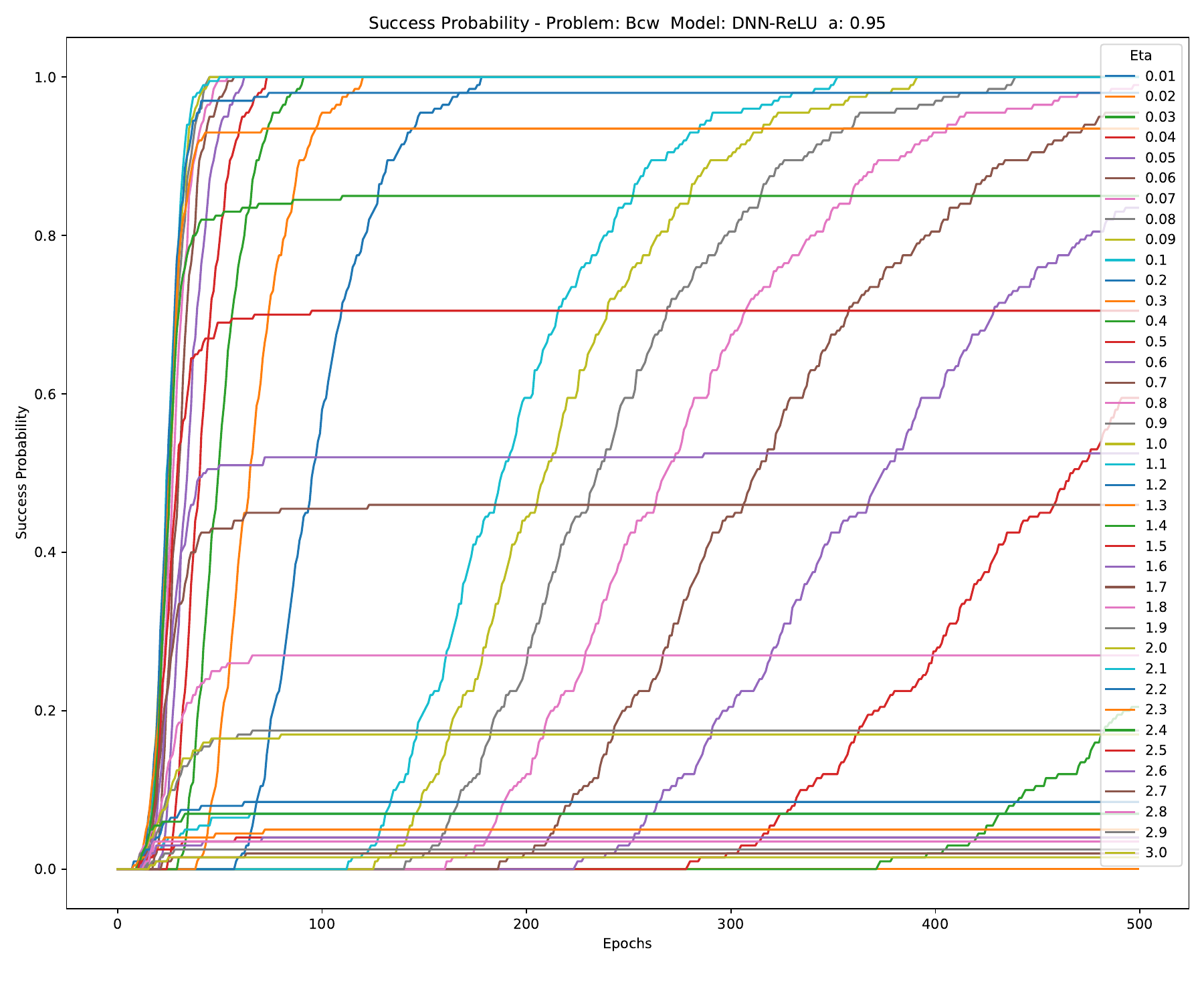} \\[-3mm]
    \includegraphics[clip,trim=0 0 0 0,width=.52\linewidth]{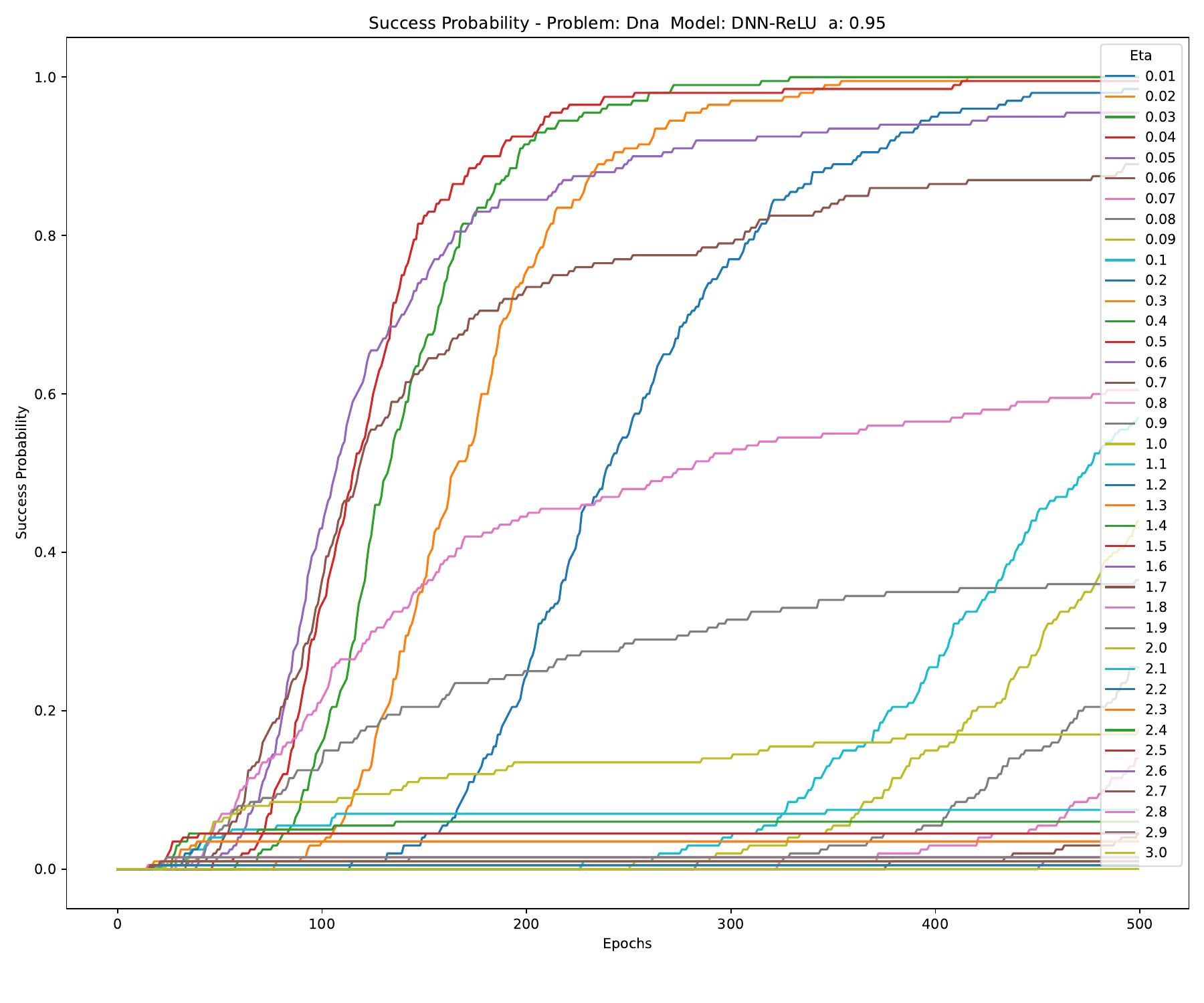} &
    \includegraphics[clip,trim=0 0 0 0,width=.52\linewidth]{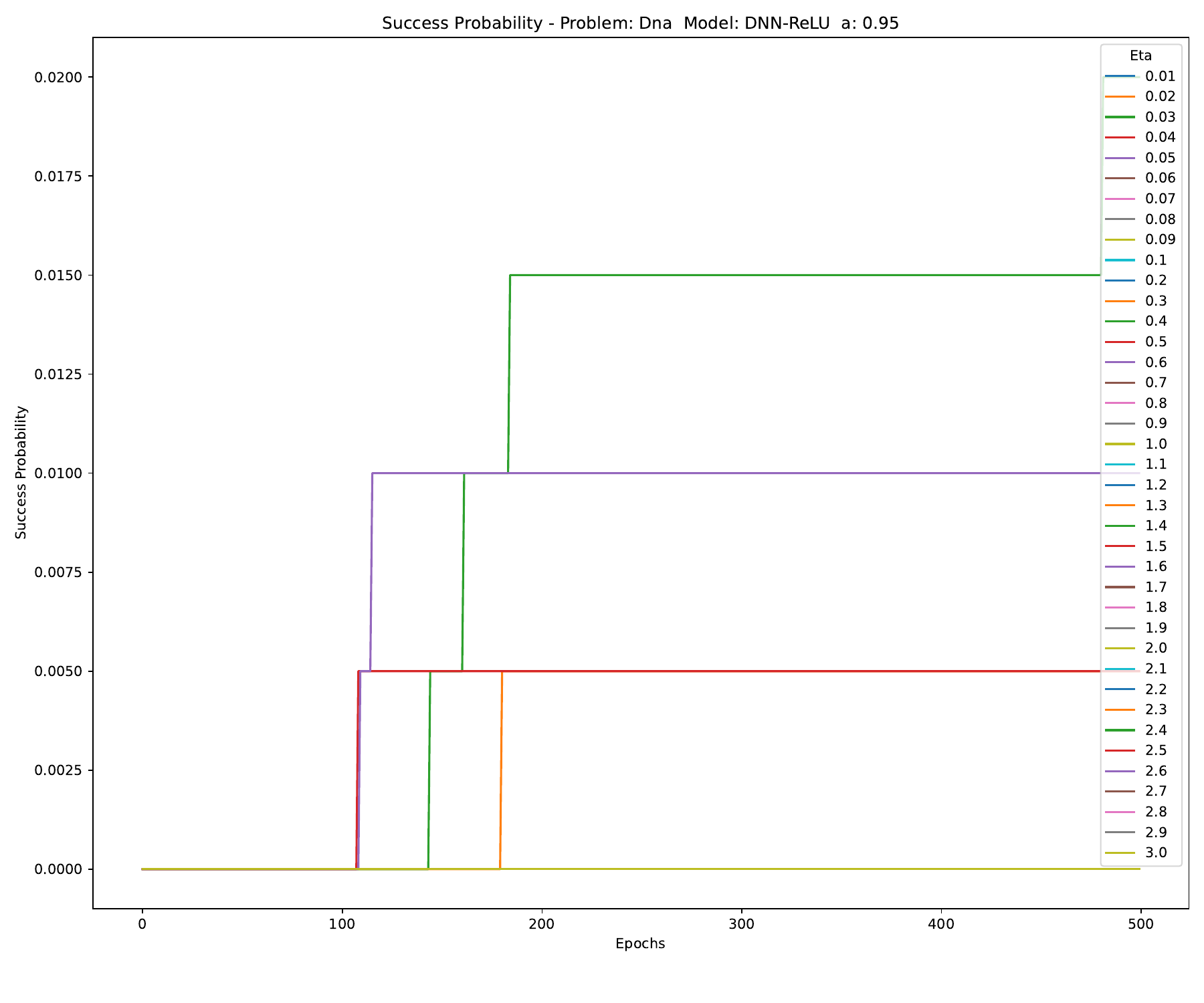} \\[-3mm]
    \includegraphics[clip,trim=0 0 0 0,width=.52\linewidth]{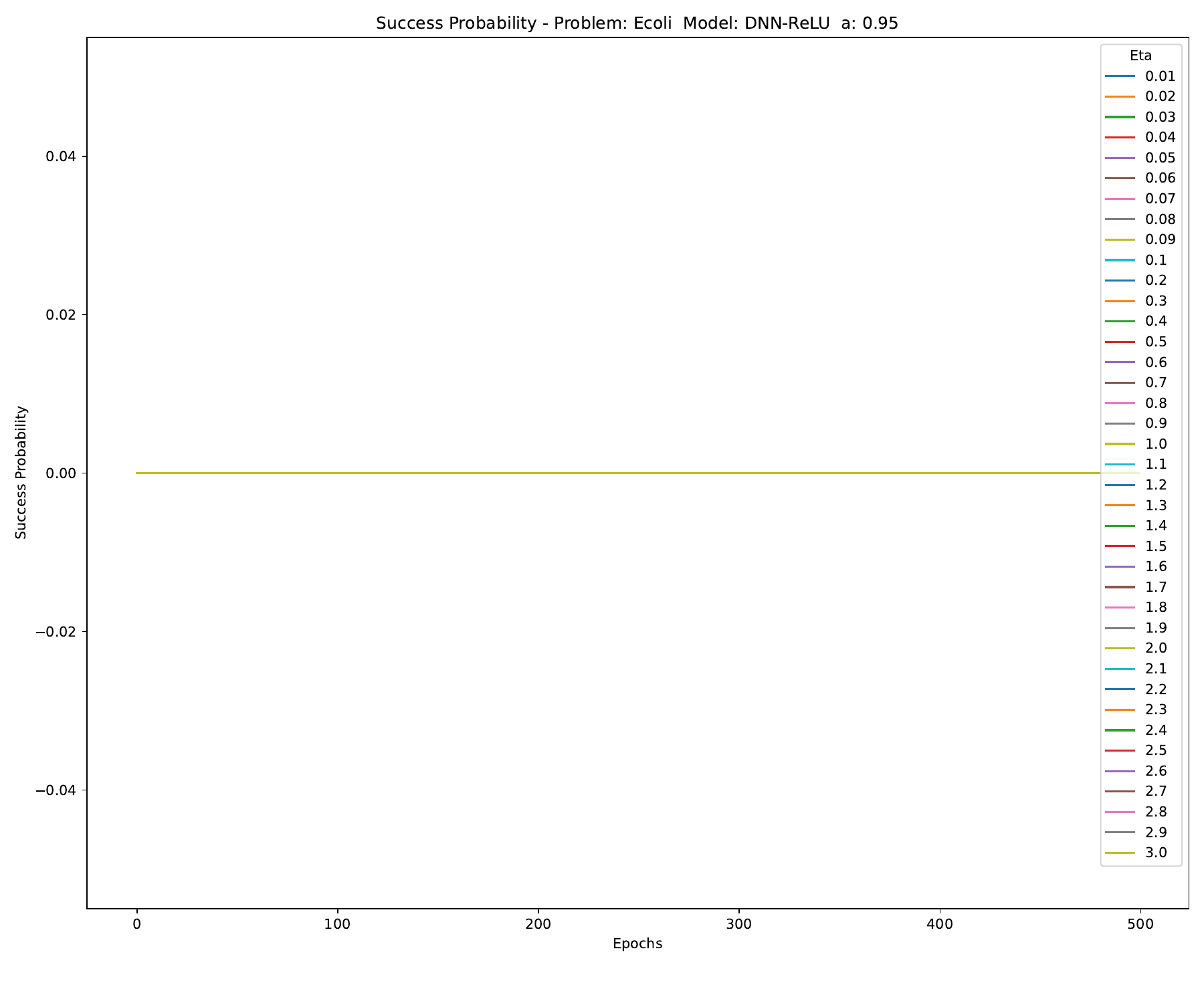} &
    \includegraphics[clip,trim=0 0 0 0,width=.52\linewidth]{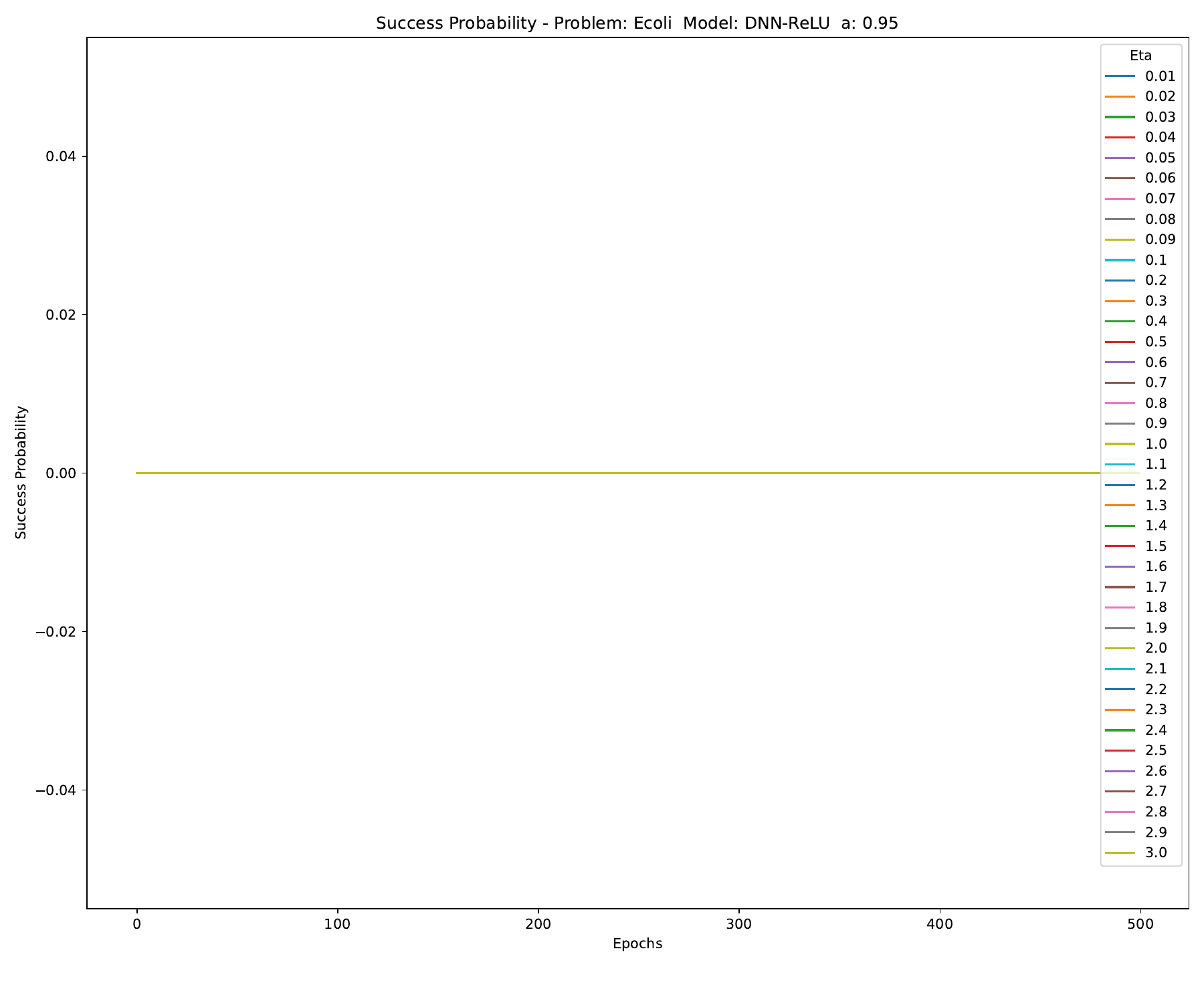} \\[-3mm]
  \end{tabular}
\caption{Success probability ($a=0.95$) for DNN-ReLU model.}
  \label{fig:success_probability_DNN_ReLU_model_0.95}
\end{figure*}

\begin{figure*}[p]
  \centering
  \ContinuedFloat
  \begin{tabular}{c@{}c}
    Training Success Probability & Validation Success Probability \\
    \includegraphics[clip,trim=0 0 0 0,width=.52\linewidth]{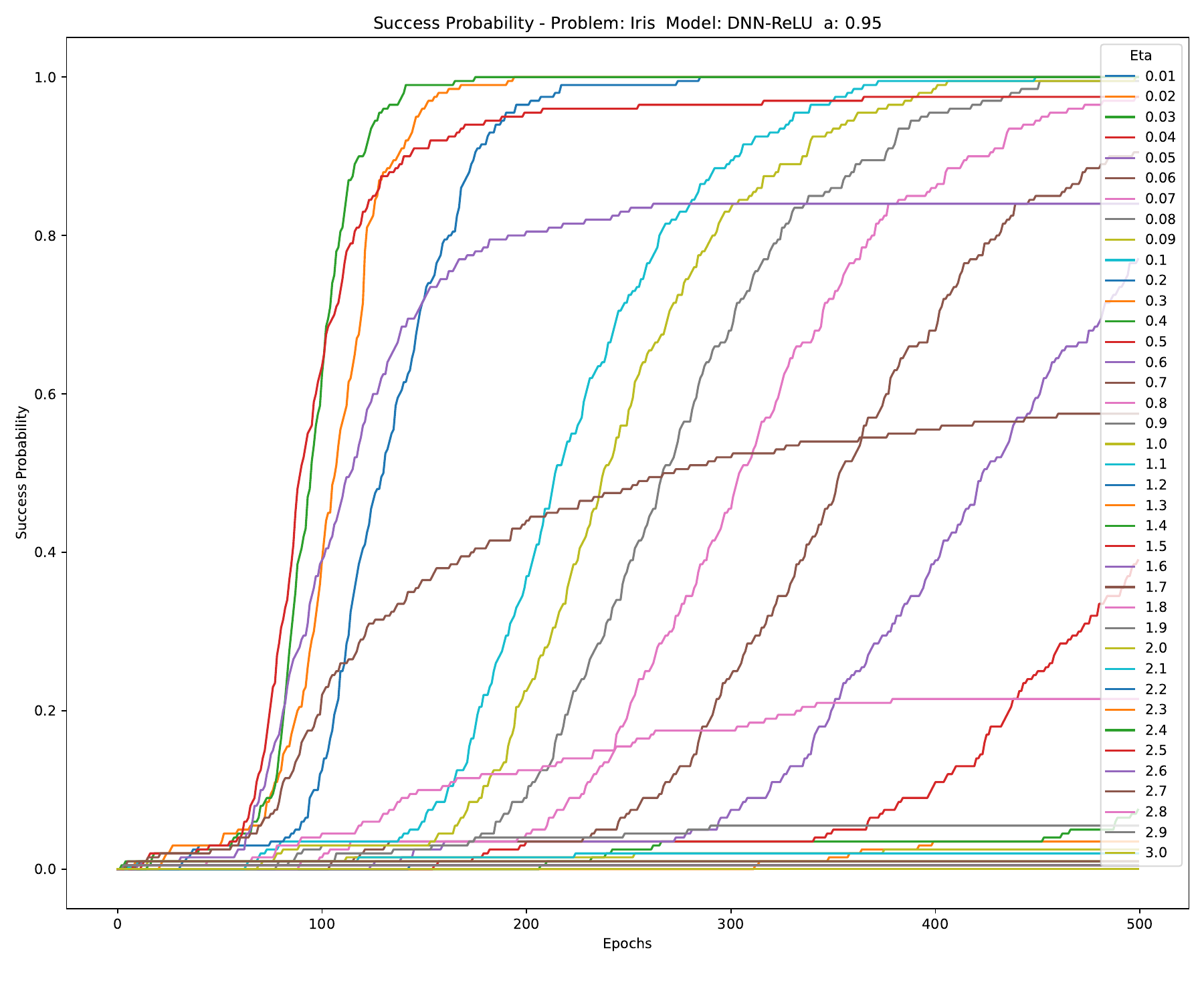} &
    \includegraphics[clip,trim=0 0 0 0,width=.52\linewidth]{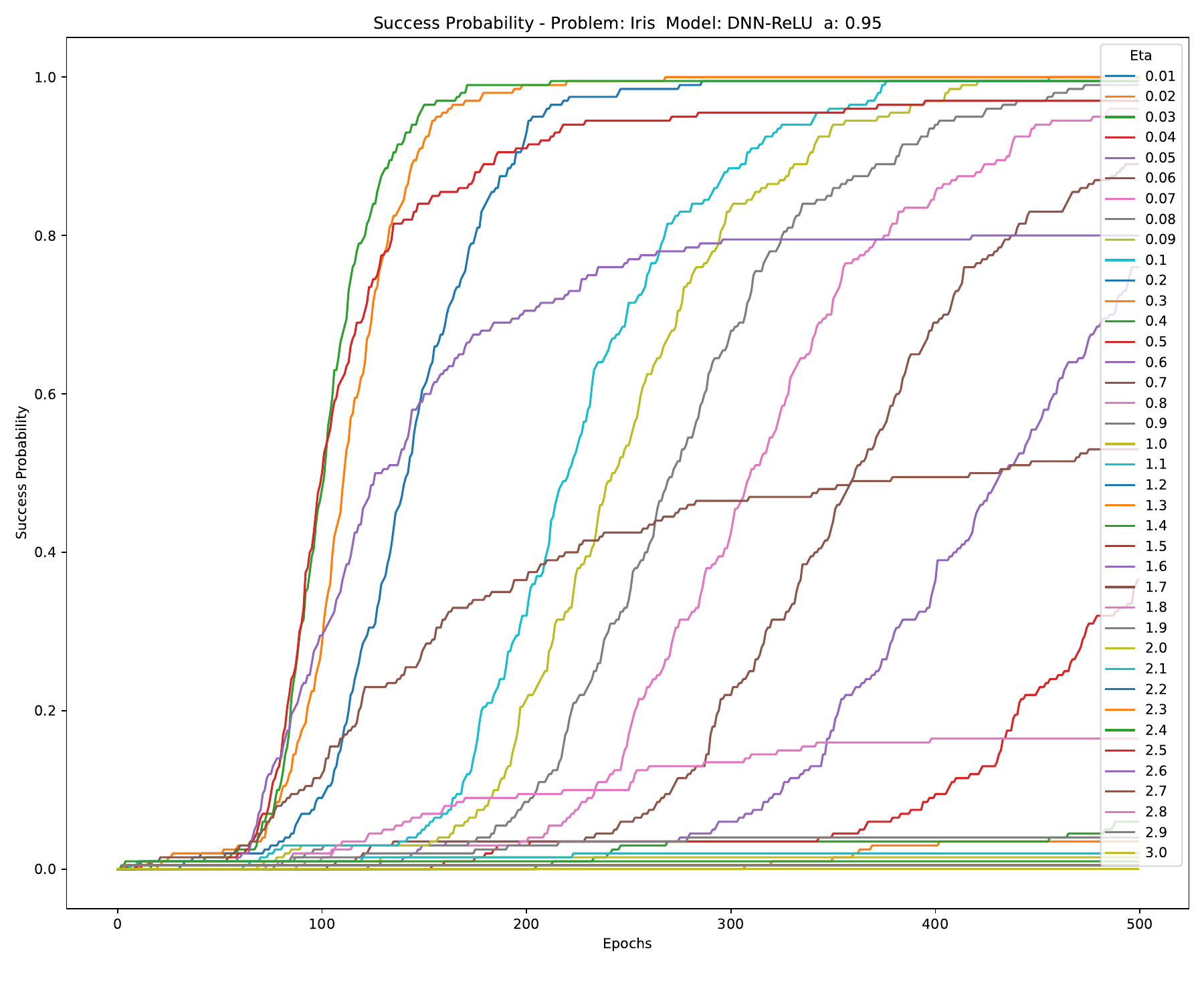} \\[-3mm]
    \includegraphics[clip,trim=0 0 0 0,width=.52\linewidth]{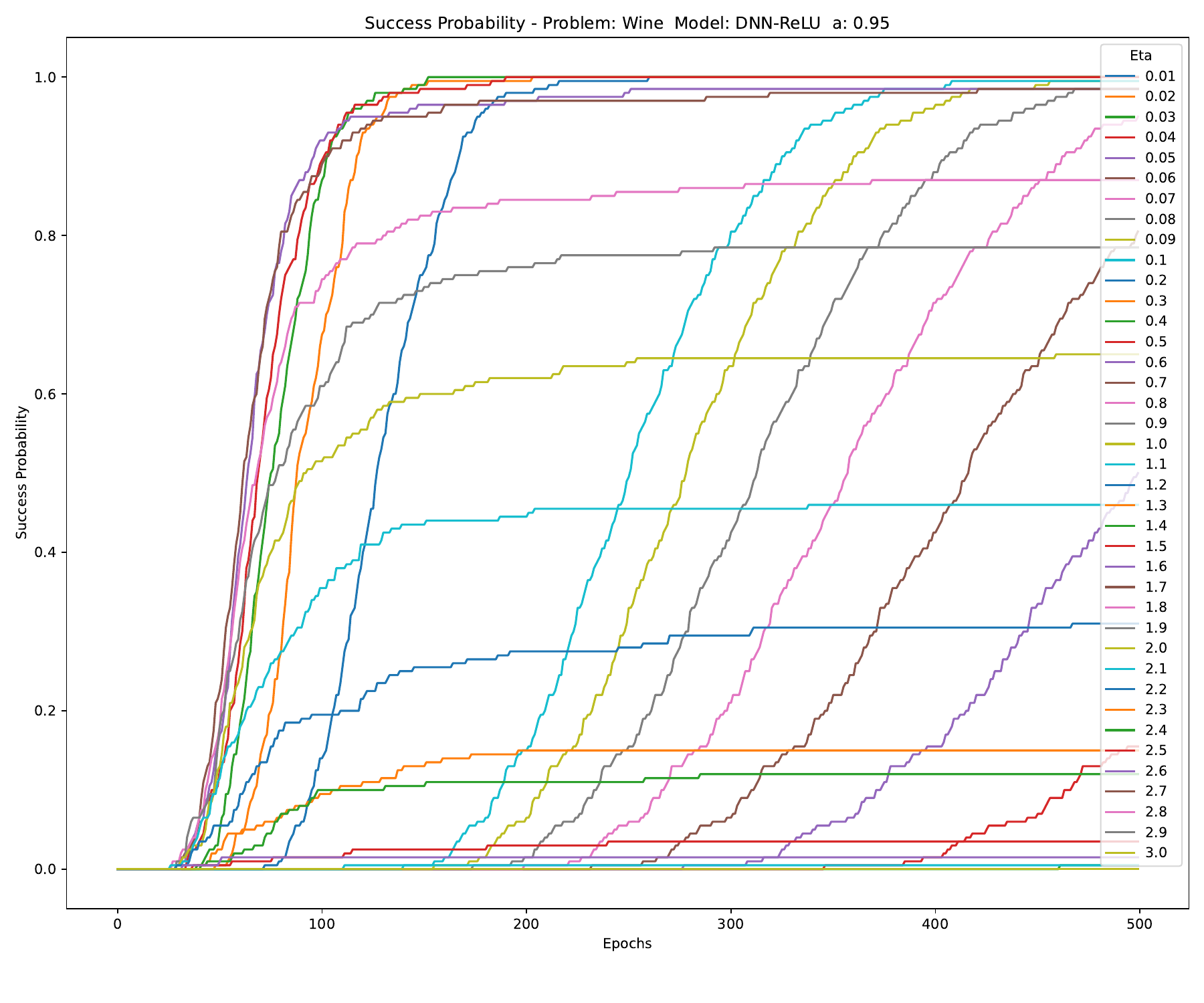} &
    \includegraphics[clip,trim=0 0 0 0,width=.52\linewidth]{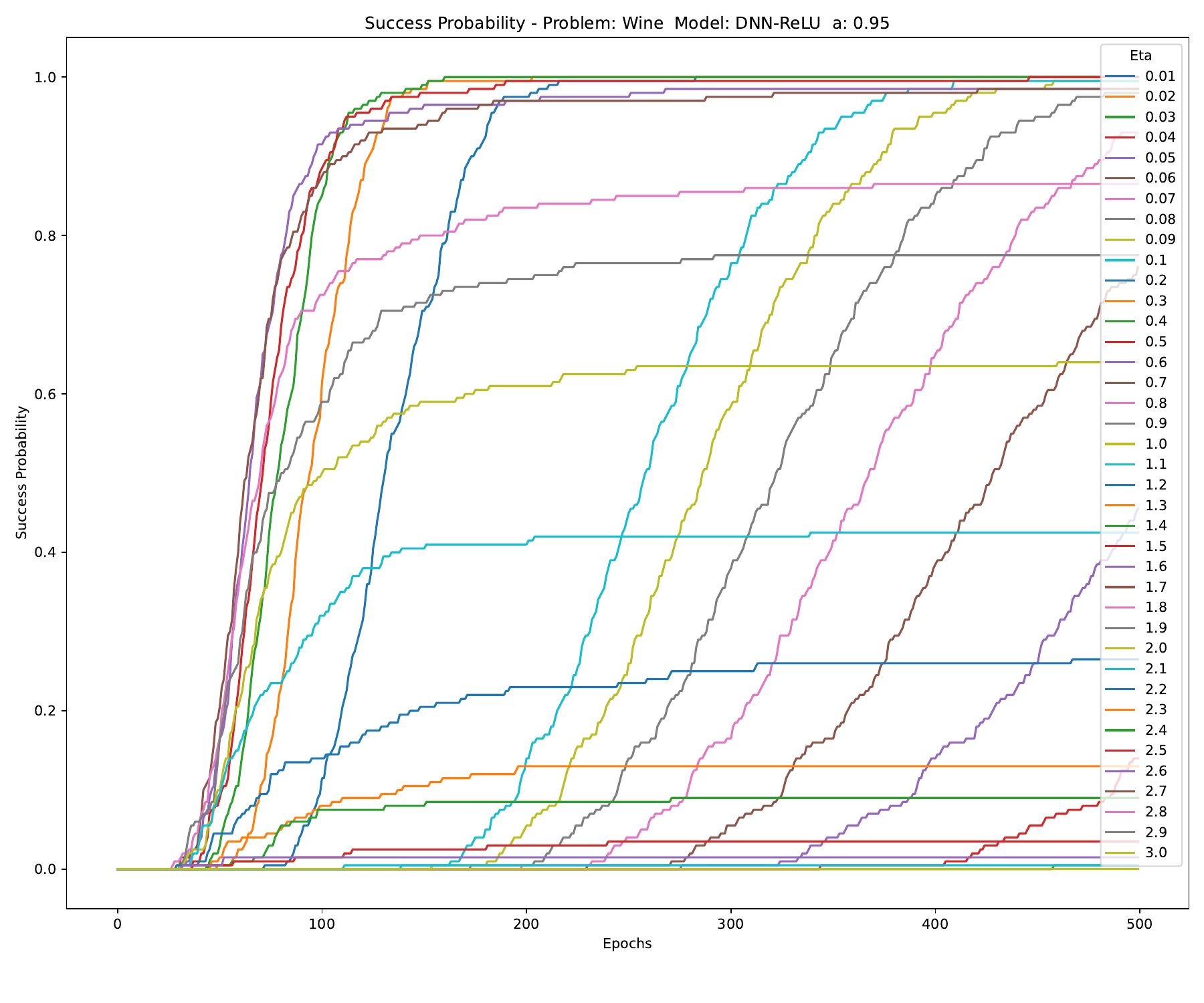} \\[-3mm]
  \end{tabular}
\caption{Success probability ($a=0.95$) for DNN-ReLU model (continued).}
\end{figure*}

\begin{figure*}[p]
  \centering
  \begin{tabular}{c@{}c}
    Training Success Probability & Validation Success Probability \\
    \includegraphics[clip,trim=0 0 0 0,width=.52\linewidth]{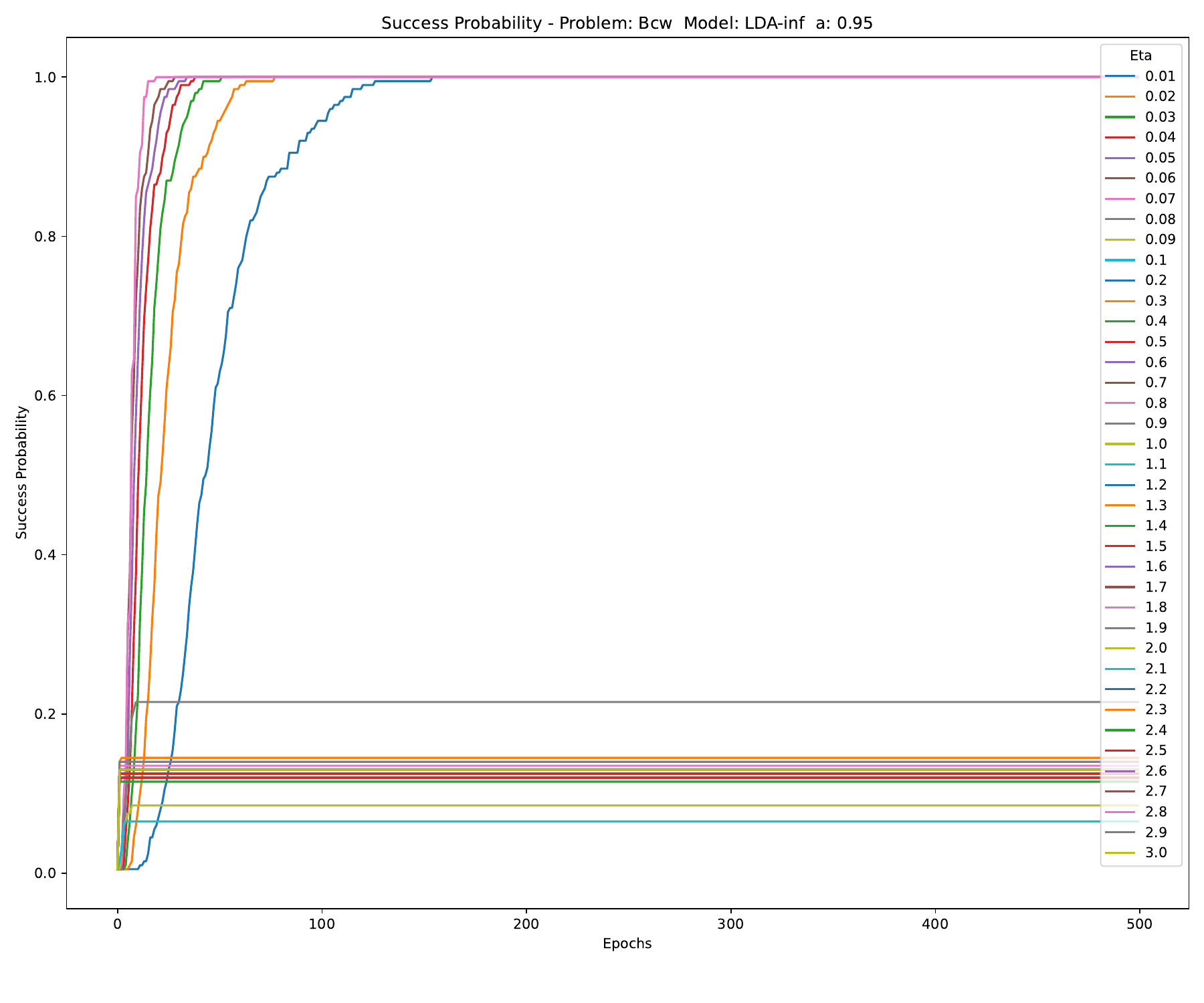} &
    \includegraphics[clip,trim=0 0 0 0,width=.52\linewidth]{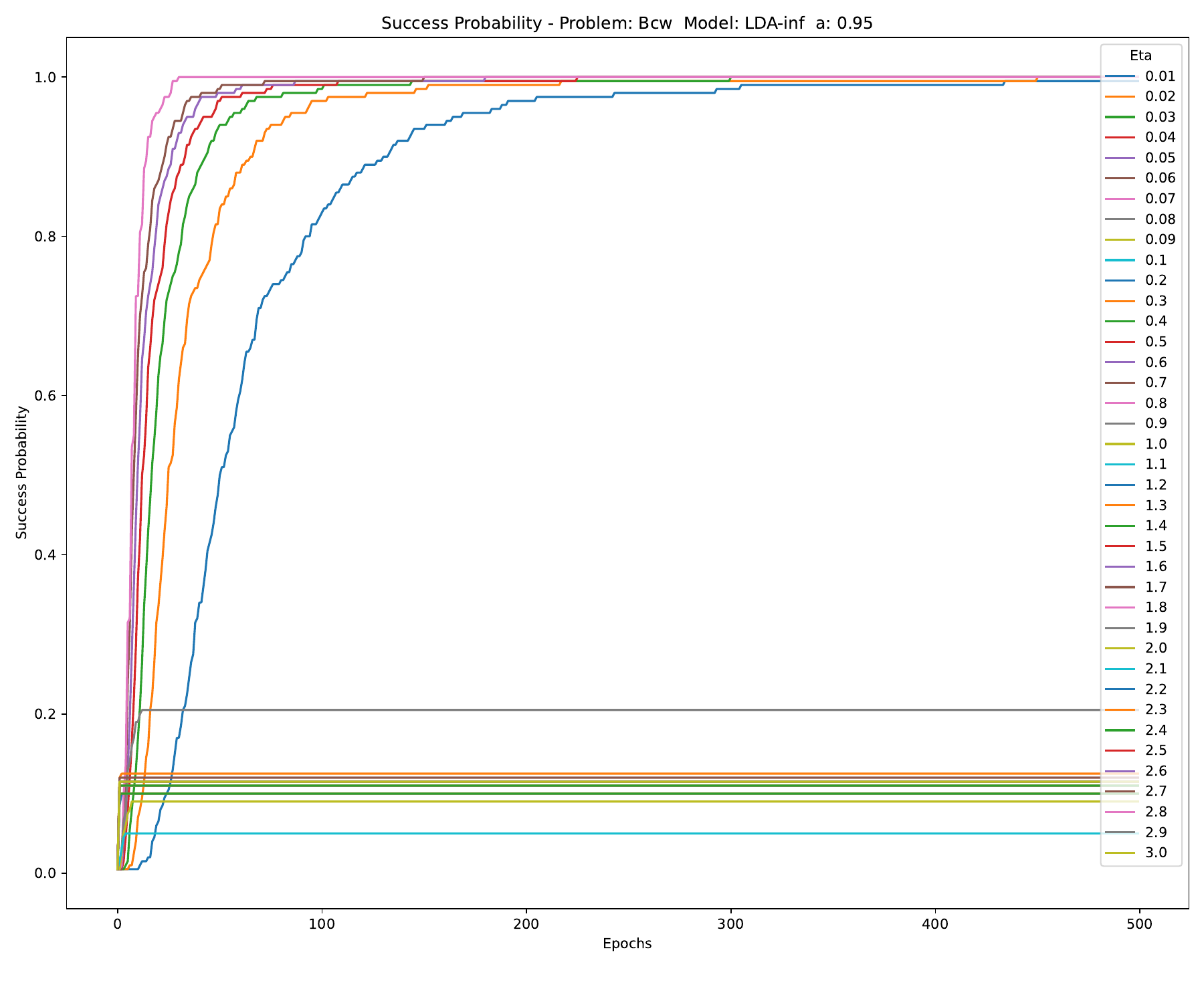} \\[-3mm]
    \includegraphics[clip,trim=0 0 0 0,width=.52\linewidth]{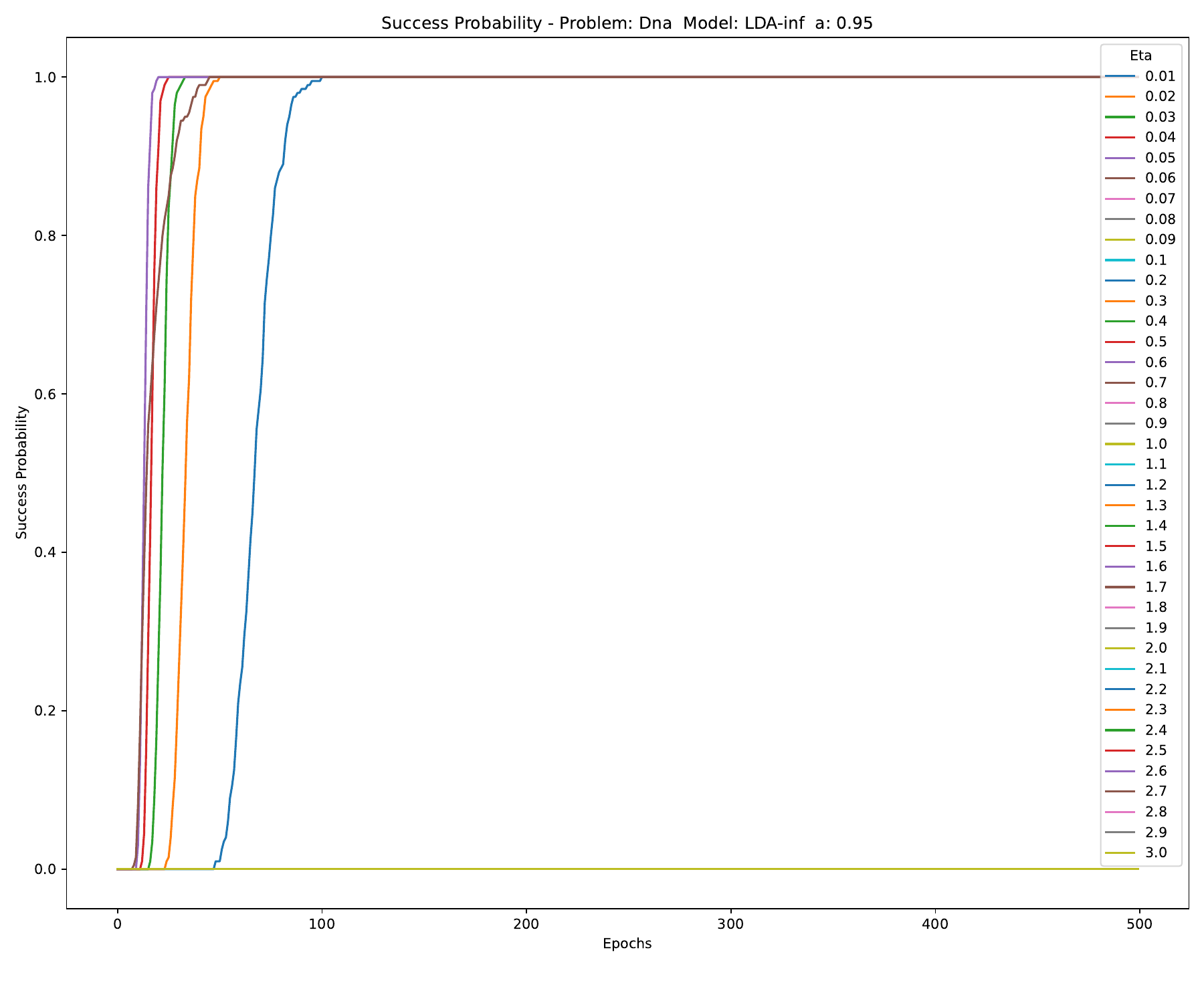} &
    \includegraphics[clip,trim=0 0 0 0,width=.52\linewidth]{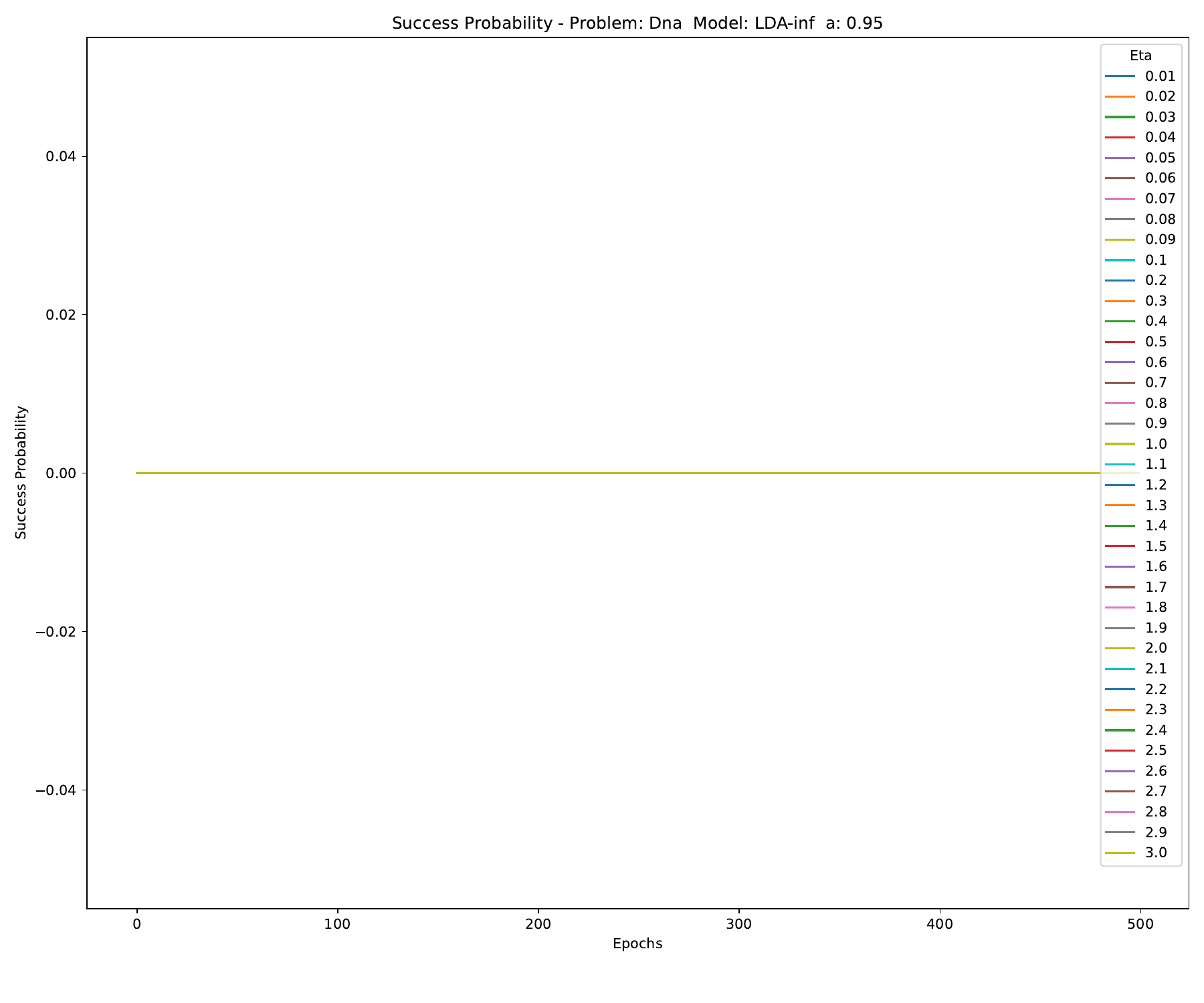} \\[-3mm]
    \includegraphics[clip,trim=0 0 0 0,width=.52\linewidth]{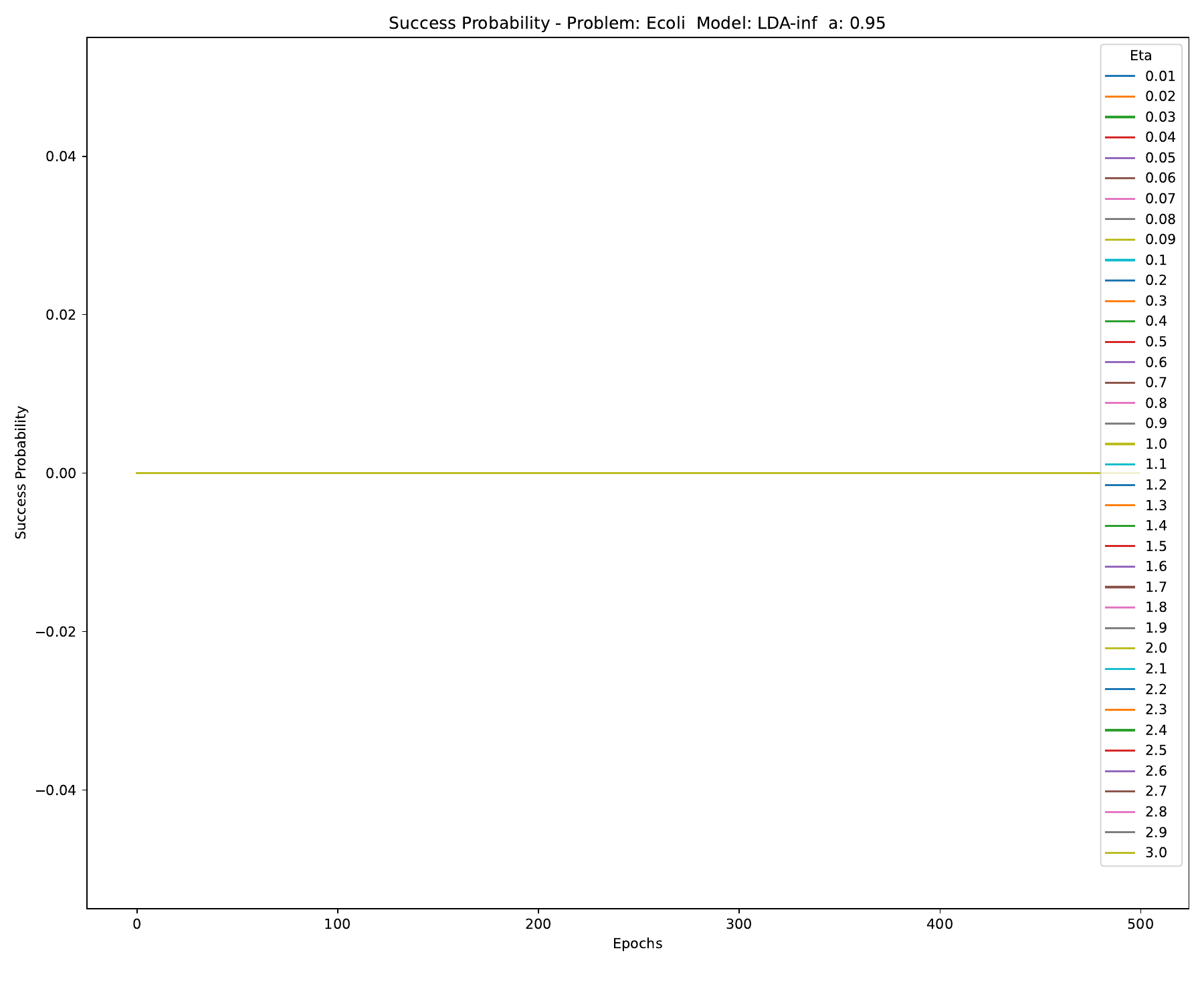} &
    \includegraphics[clip,trim=0 0 0 0,width=.52\linewidth]{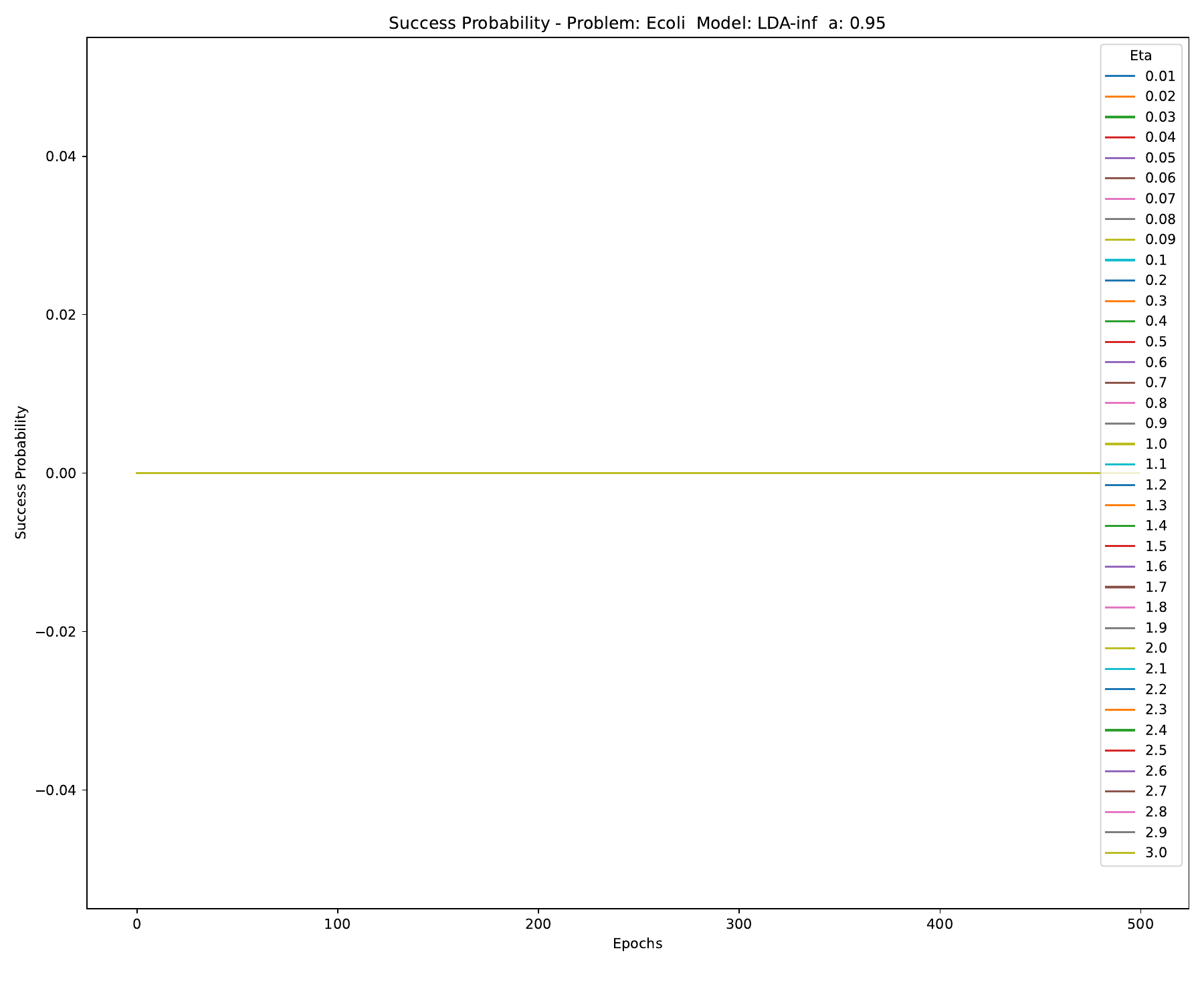} \\[-3mm]
  \end{tabular}
\caption{Success probability ($a=0.95$) for LDA-Inf model.}
  \label{fig:success_probability_LDA_Inf_model_0.95}
\end{figure*}

\begin{figure*}[p]
  \centering
  \ContinuedFloat
  \begin{tabular}{c@{}c}
    Training Success Probability & Validation Success Probability \\
    \includegraphics[clip,trim=0 0 0 0,width=.52\linewidth]{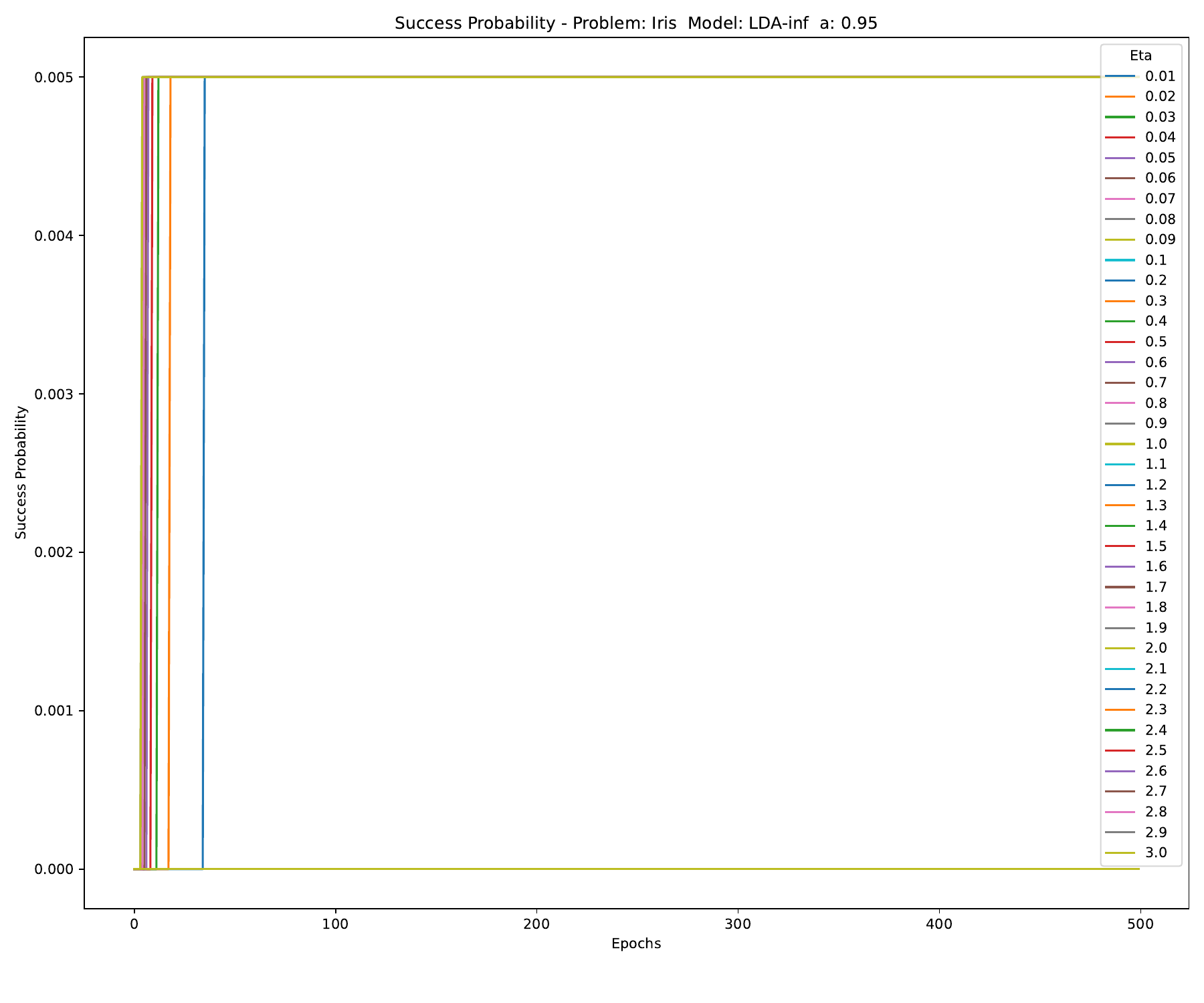} &
    \includegraphics[clip,trim=0 0 0 0,width=.52\linewidth]{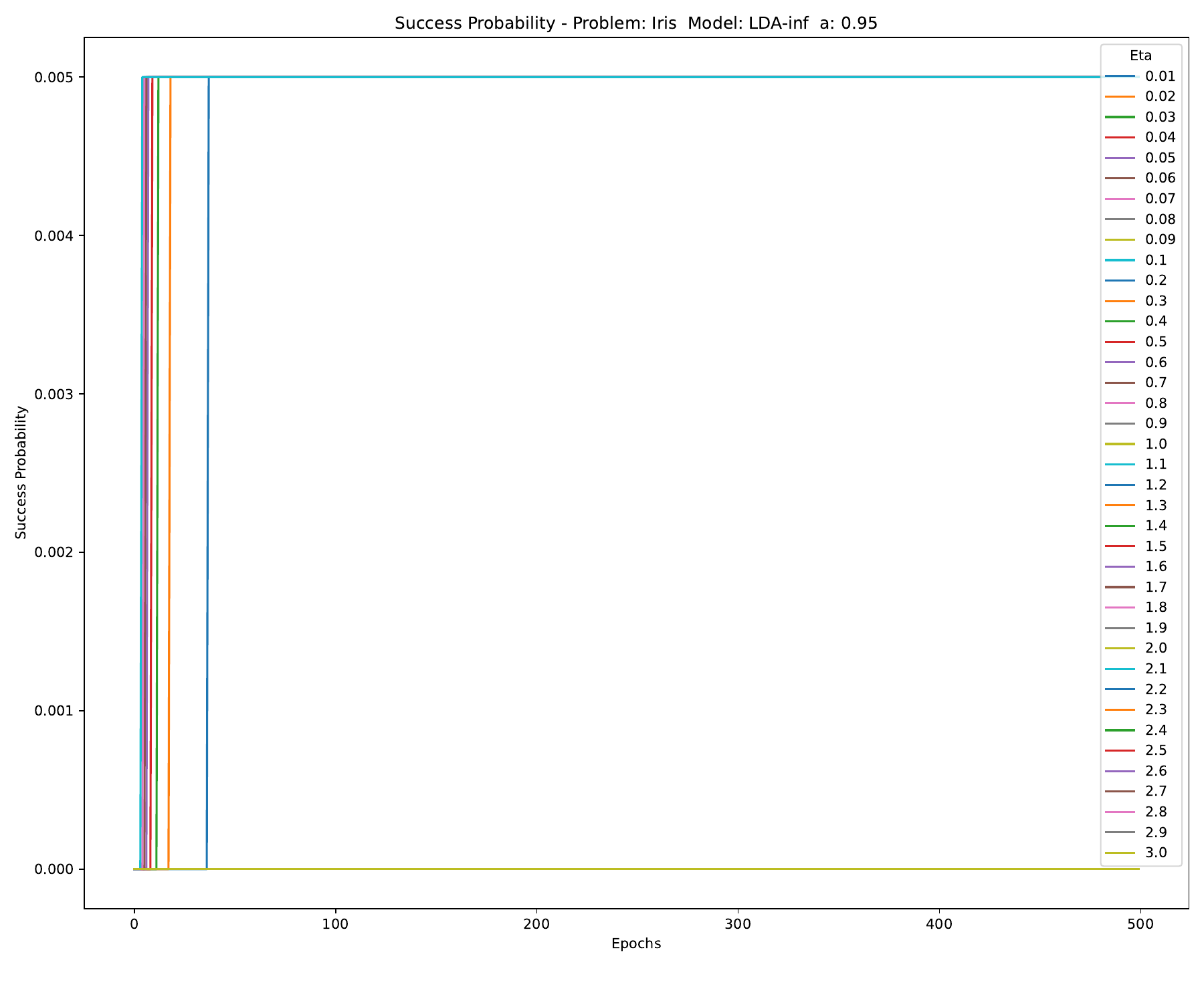} \\[-3mm]
    \includegraphics[clip,trim=0 0 0 0,width=.52\linewidth]{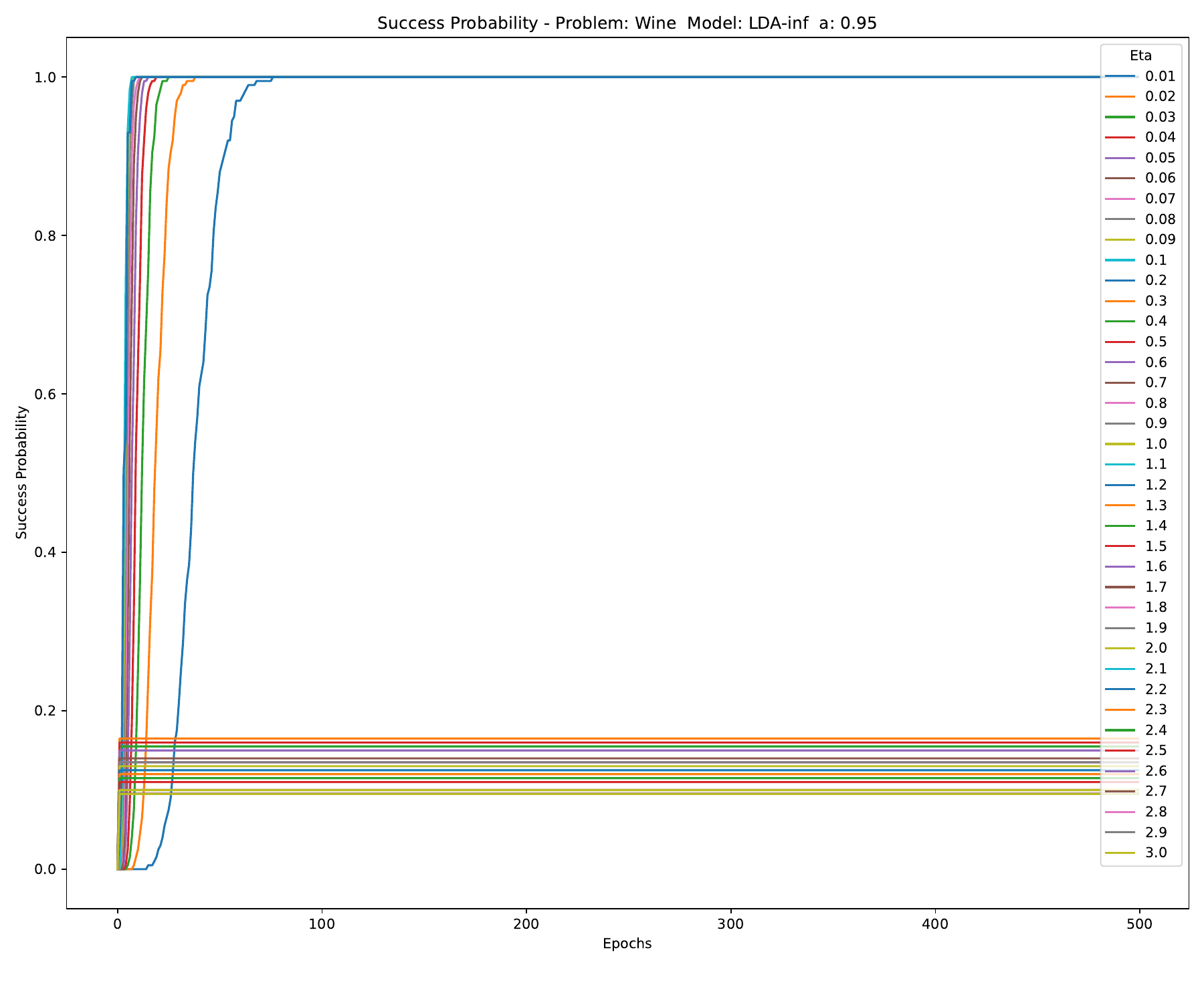} &
    \includegraphics[clip,trim=0 0 0 0,width=.52\linewidth]{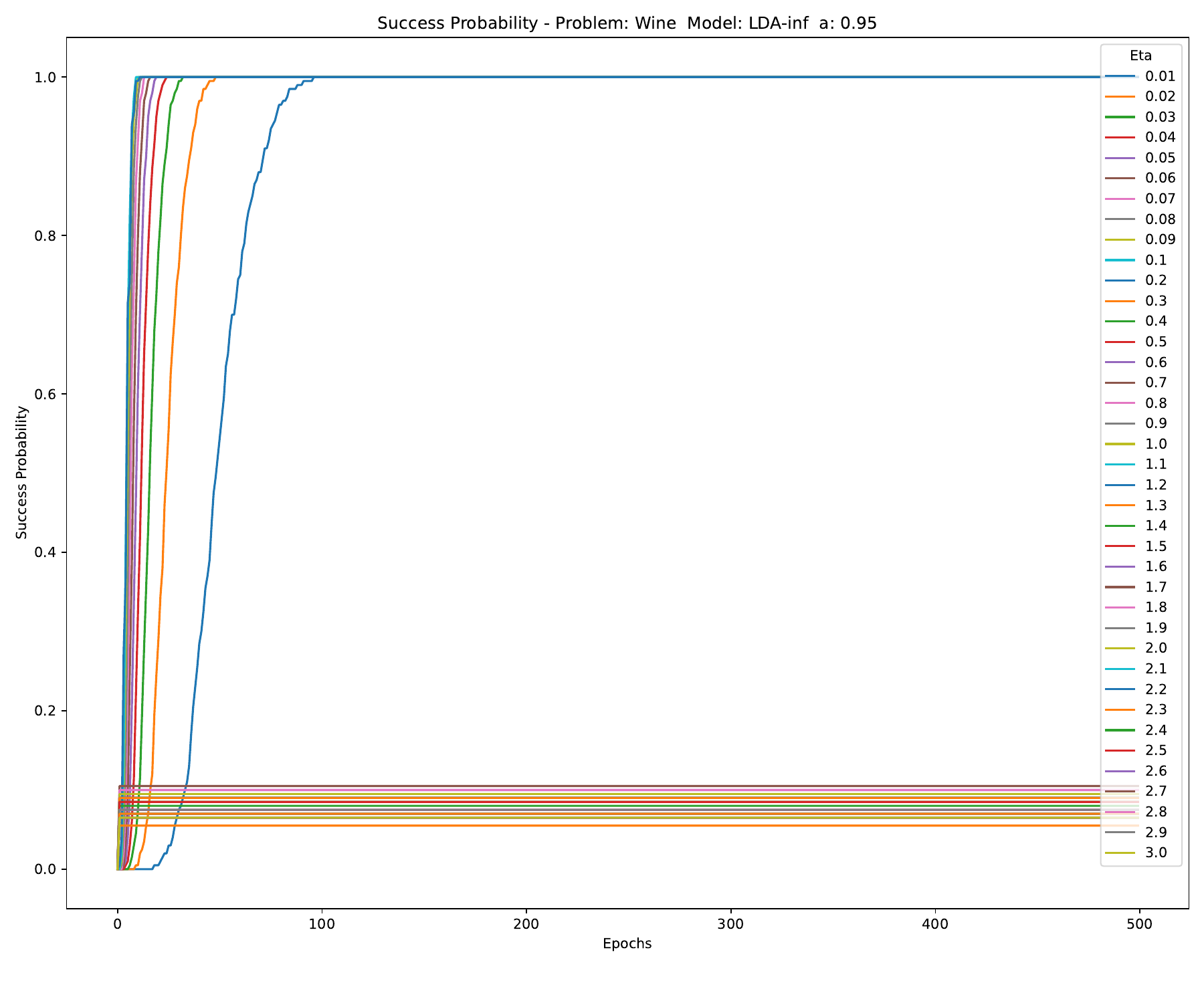} \\[-3mm]
  \end{tabular}
\caption{Success probability ($a=0.95$) for LDA-Inf model (continued).}
\end{figure*}

\begin{figure*}[p]
  \centering
  \begin{tabular}{c@{}c}
    Training Success Probability & Validation Success Probability \\
    \includegraphics[clip,trim=0 0 0 0,width=.52\linewidth]{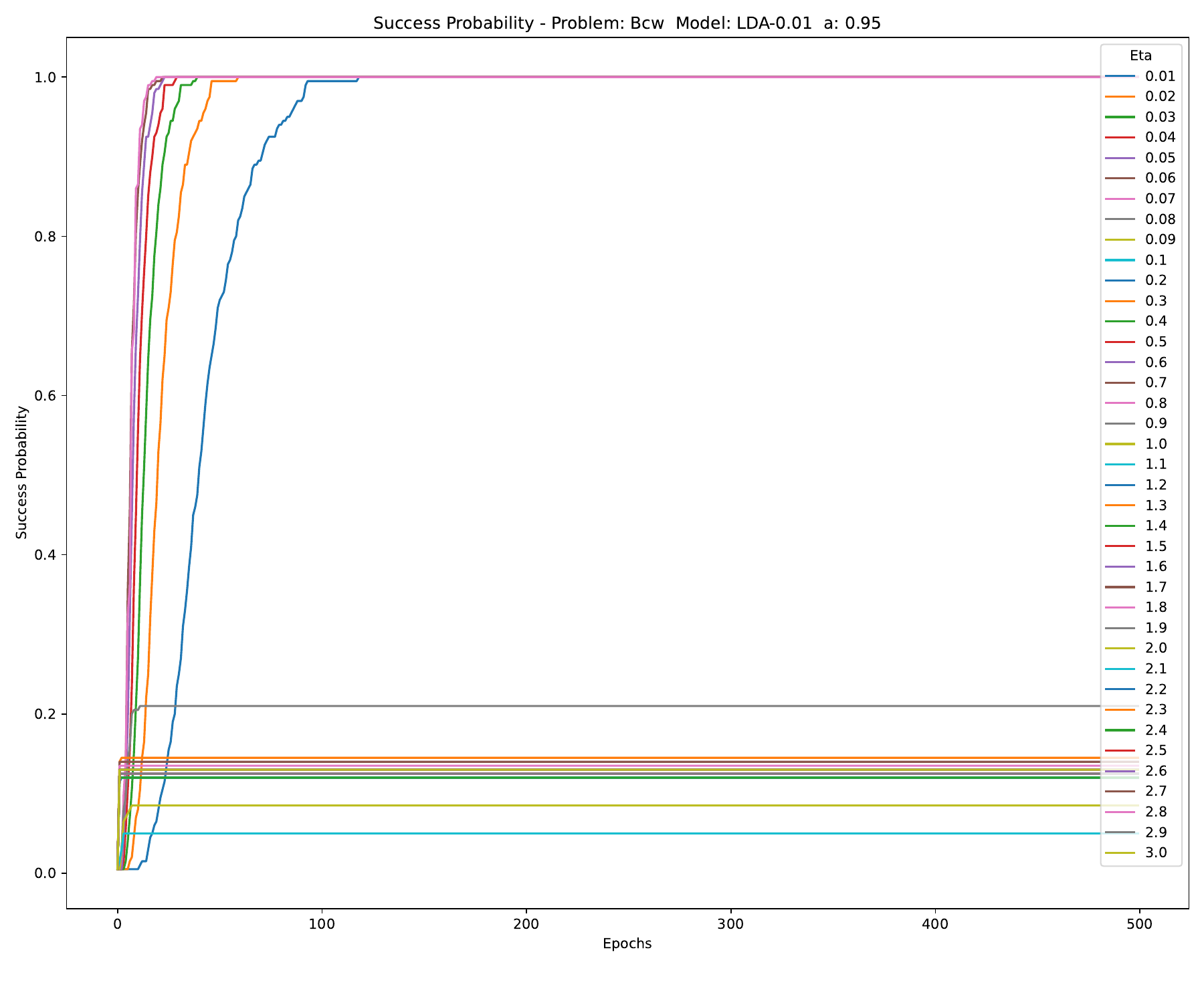} &
    \includegraphics[clip,trim=0 0 0 0,width=.52\linewidth]{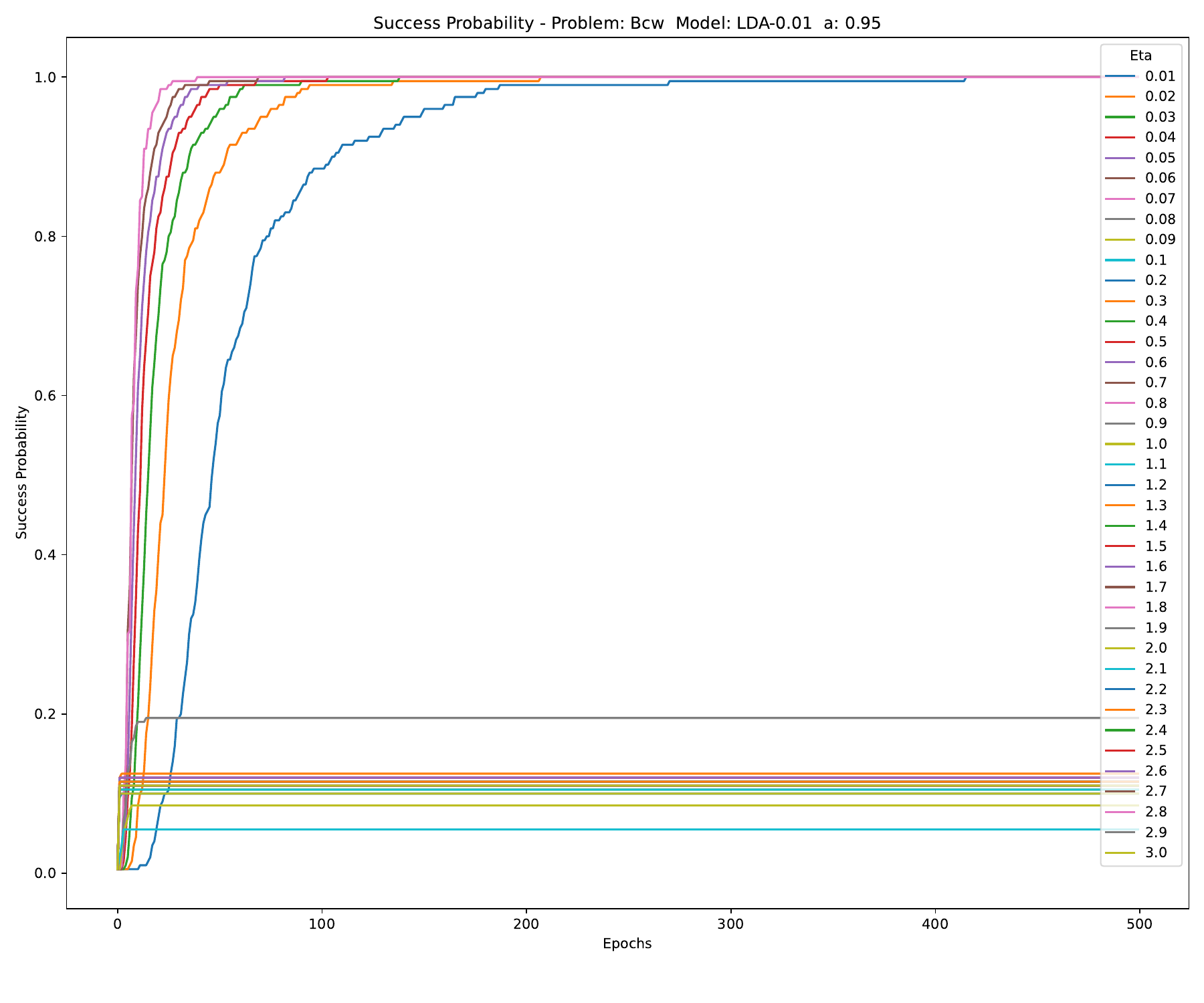} \\[-3mm]
    \includegraphics[clip,trim=0 0 0 0,width=.52\linewidth]{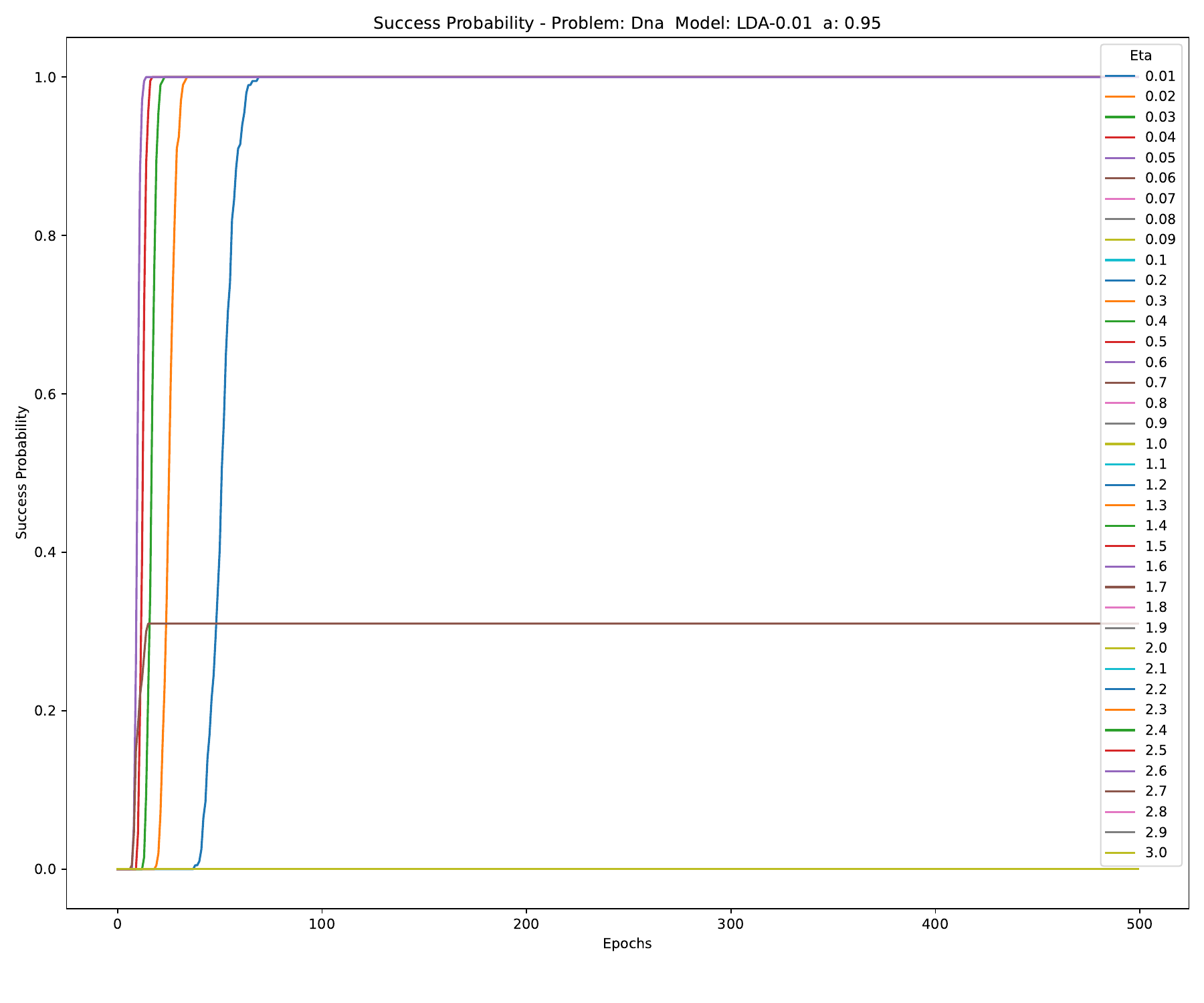} &
    \includegraphics[clip,trim=0 0 0 0,width=.52\linewidth]{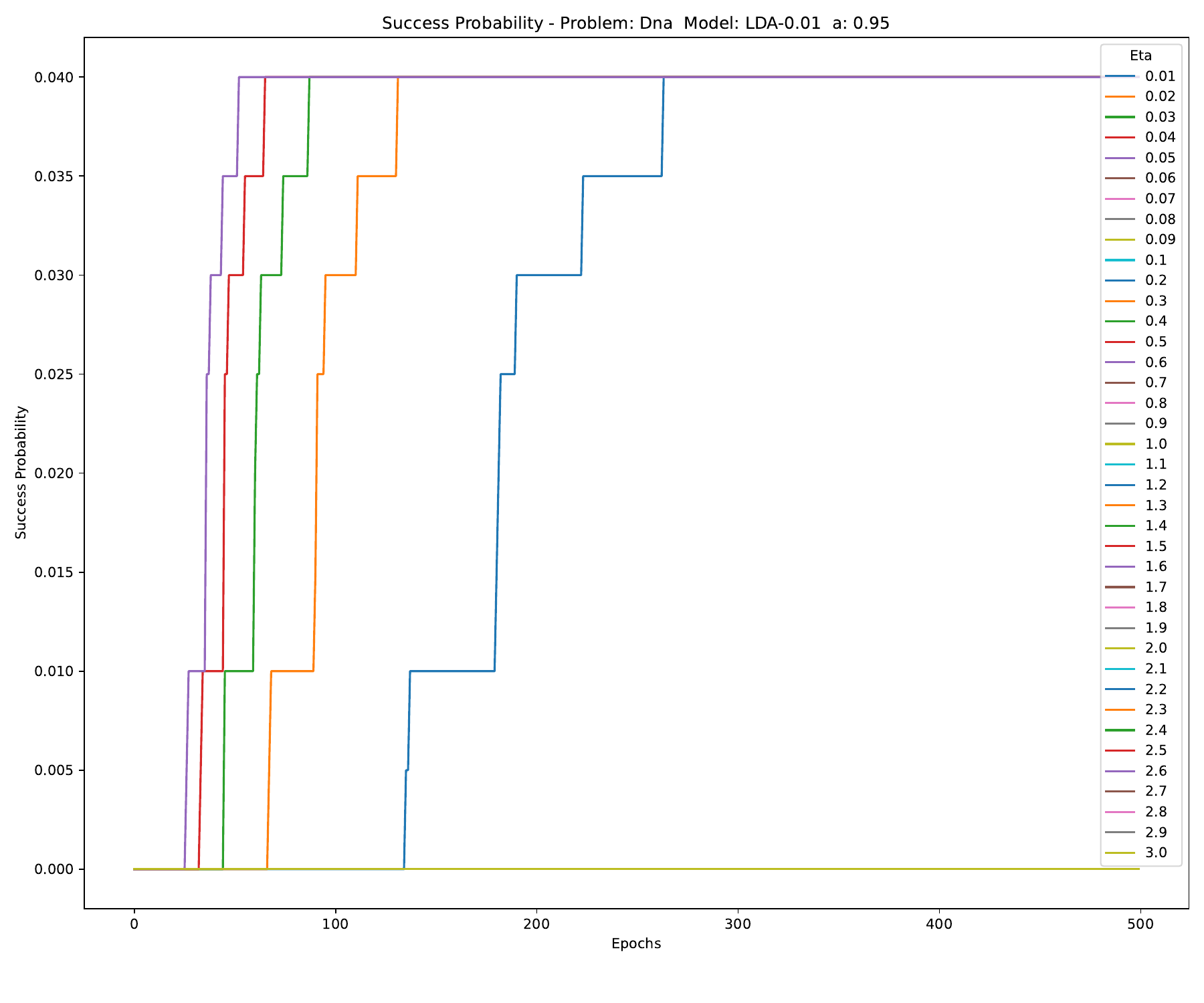} \\[-3mm]
    \includegraphics[clip,trim=0 0 0 0,width=.52\linewidth]{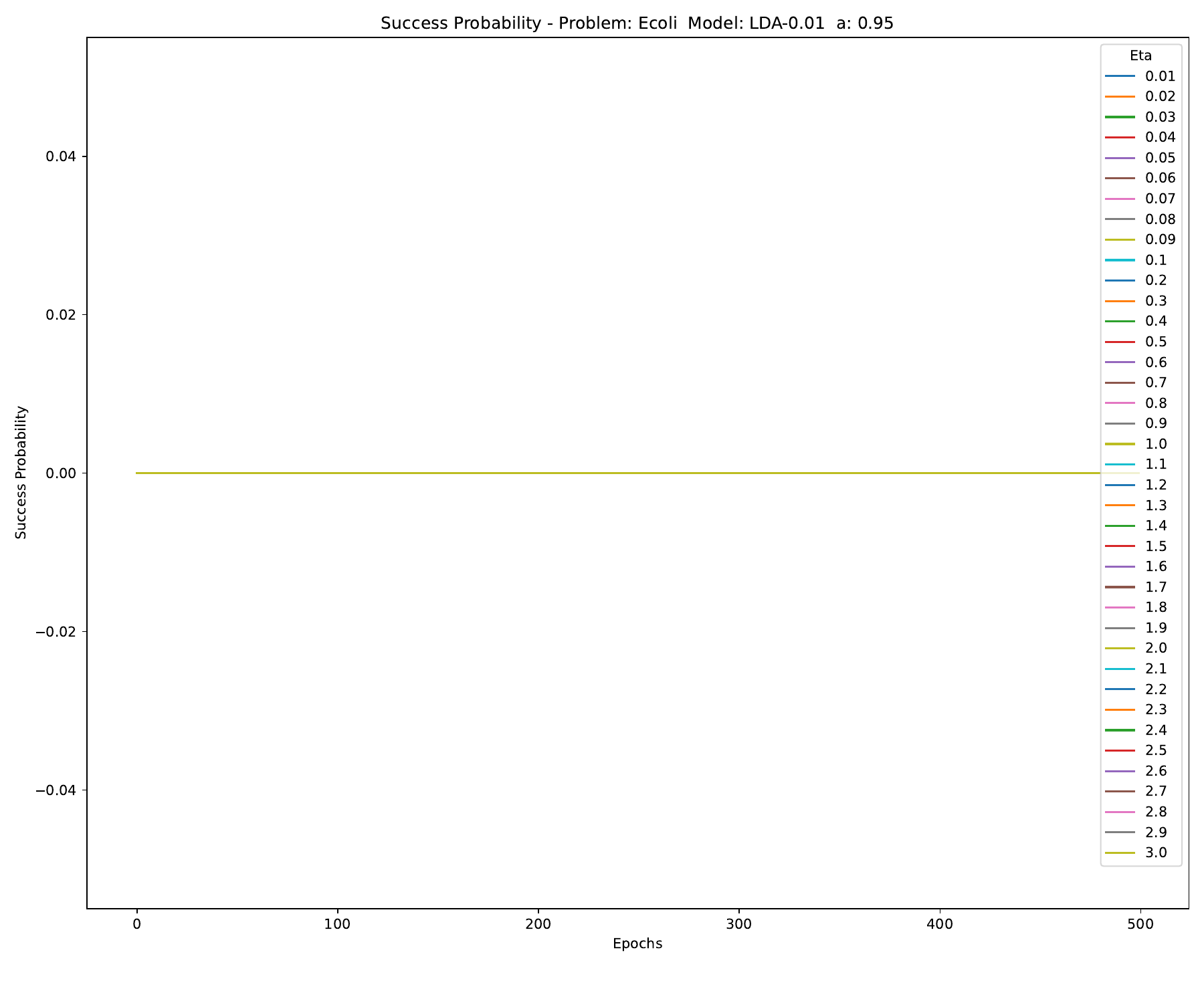} &
    \includegraphics[clip,trim=0 0 0 0,width=.52\linewidth]{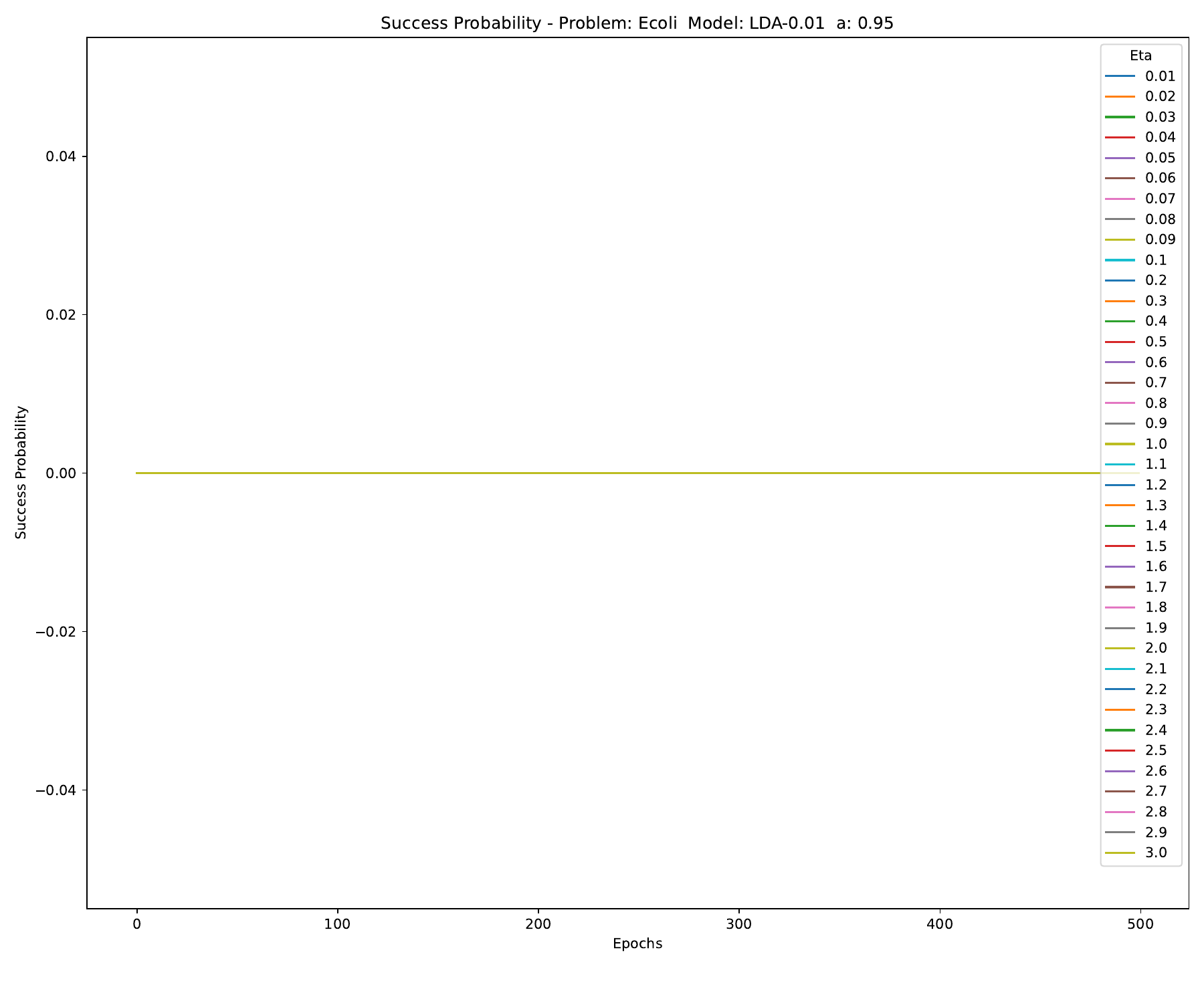} \\[-3mm]
  \end{tabular}
\caption{Success probability ($a=0.95$) for LDA-0.01 model.}
  \label{fig:success_probability_LDA_0.01_model_0.95}
\end{figure*}

\begin{figure*}[p]
  \centering
  \ContinuedFloat
  \begin{tabular}{c@{}c}
    Training Success Probability & Validation Success Probability \\
    \includegraphics[clip,trim=0 0 0 0,width=.52\linewidth]{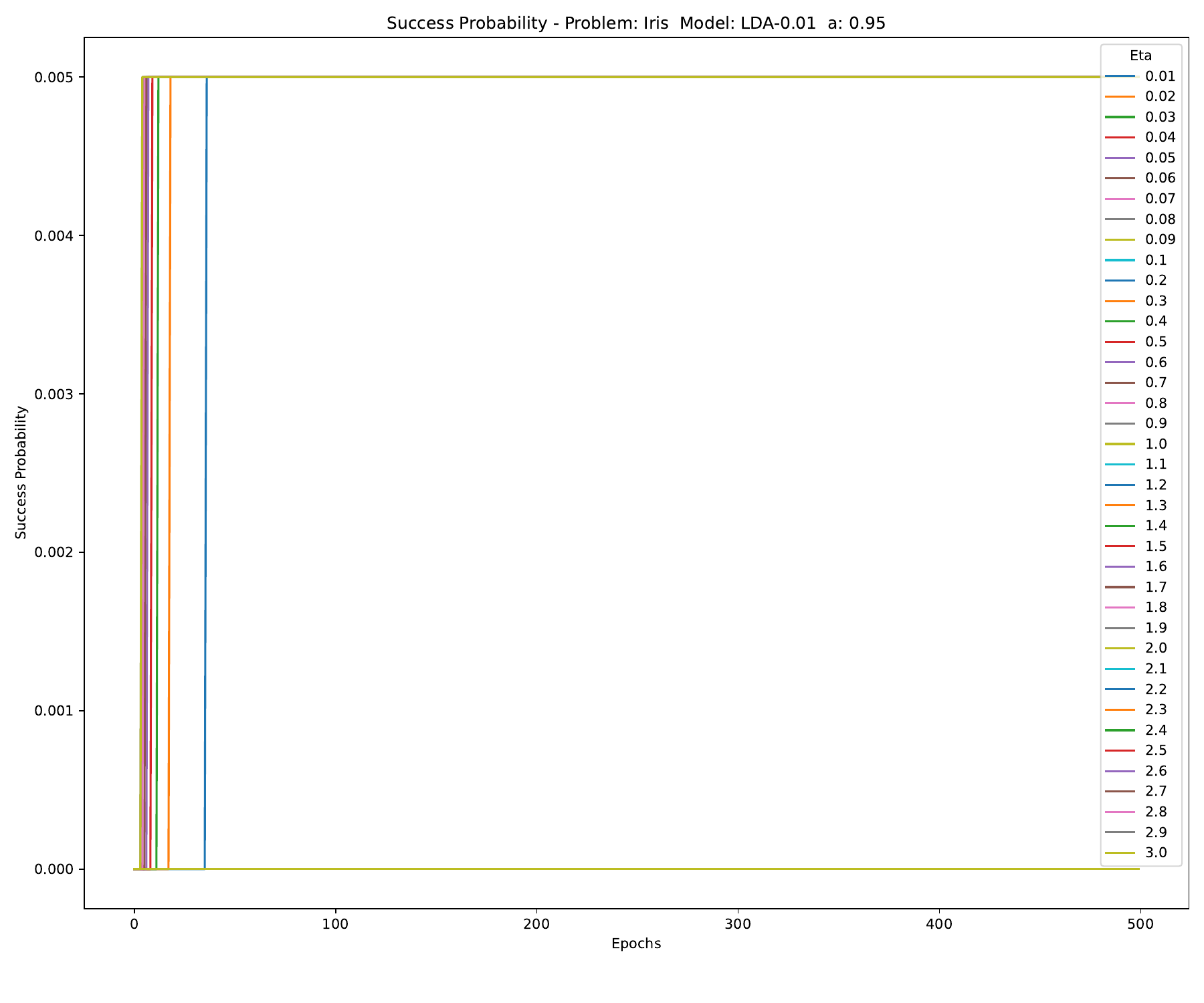} &
    \includegraphics[clip,trim=0 0 0 0,width=.52\linewidth]{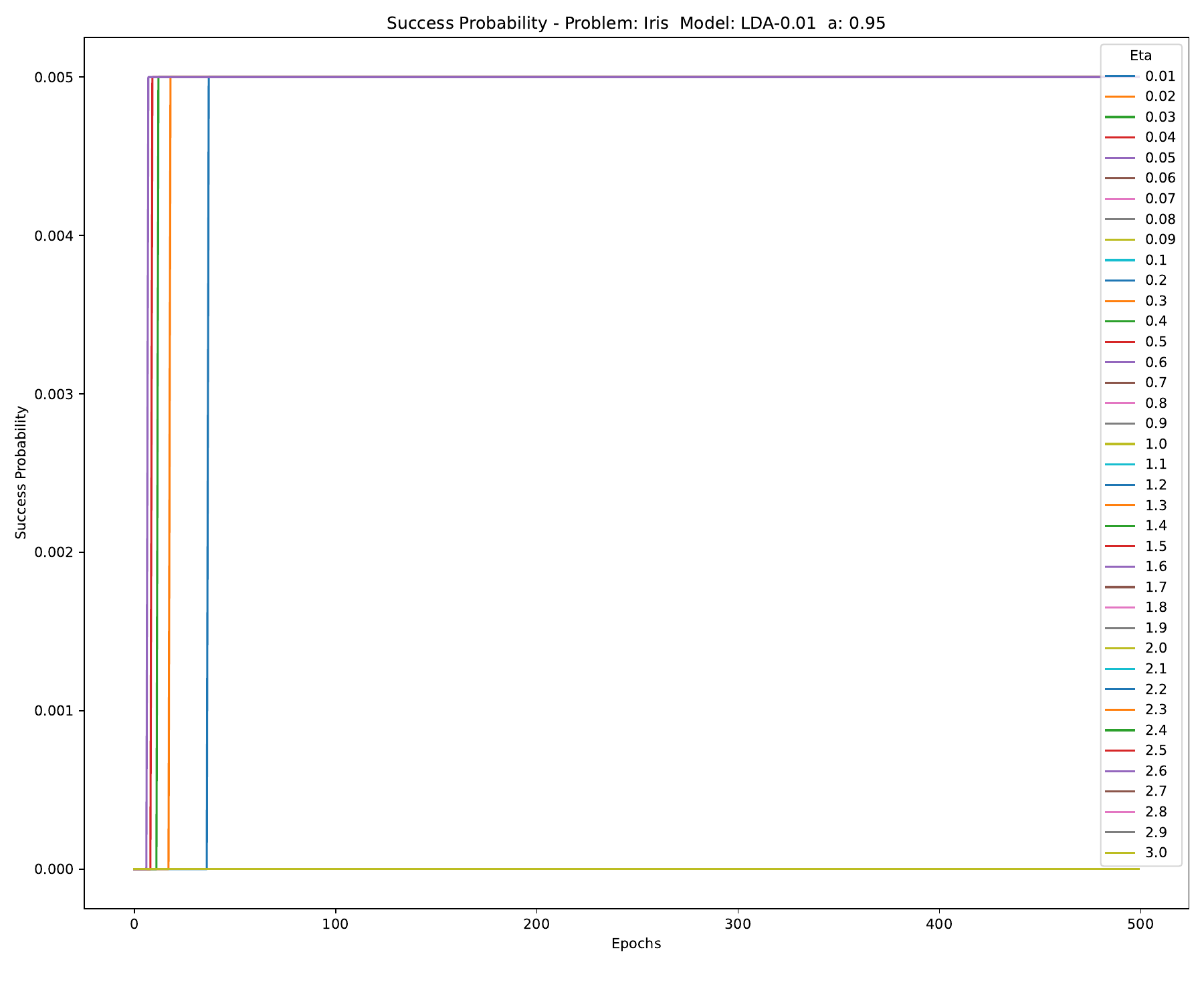} \\[-3mm]
    \includegraphics[clip,trim=0 0 0 0,width=.52\linewidth]{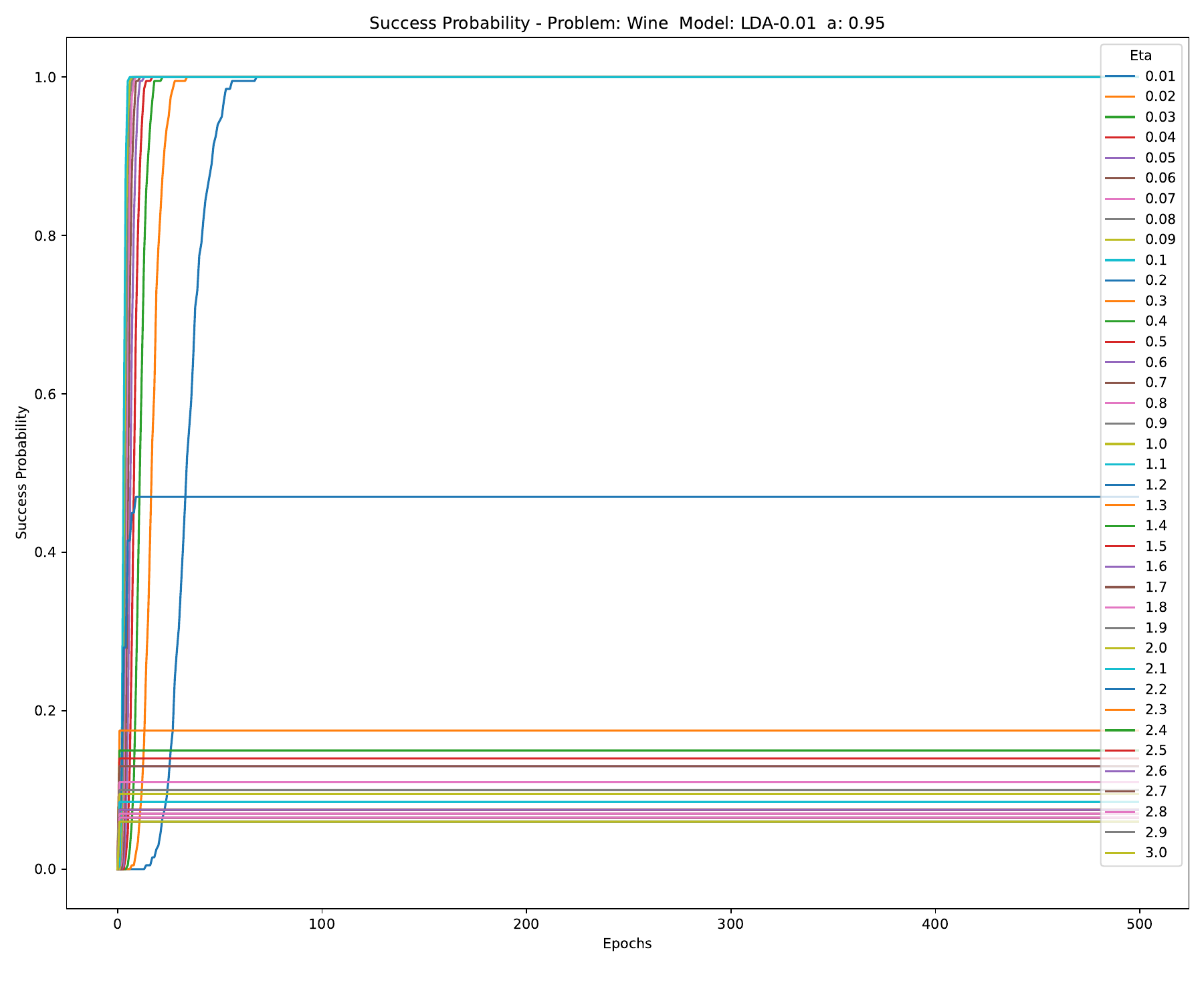} &
    \includegraphics[clip,trim=0 0 0 0,width=.52\linewidth]{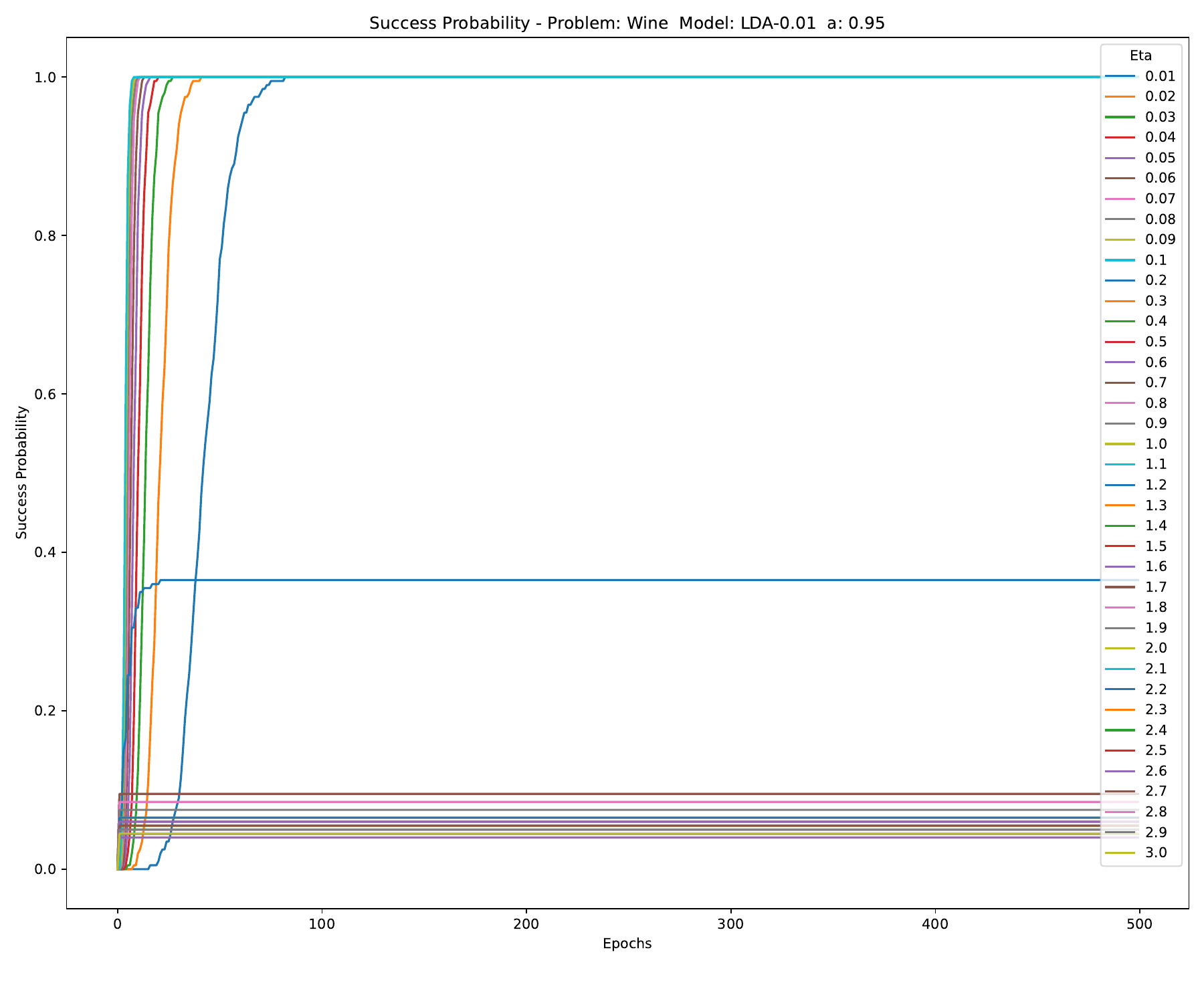} \\[-3mm]
  \end{tabular}
\caption{Success probability ($a=0.95$) for LDA-0.01 model (continued).}
\end{figure*}

\begin{figure*}[p]
  \centering
  \begin{tabular}{c@{}c}
    Training Success Probability & Validation Success Probability \\
    \includegraphics[clip,trim=0 0 0 0,width=.52\linewidth]{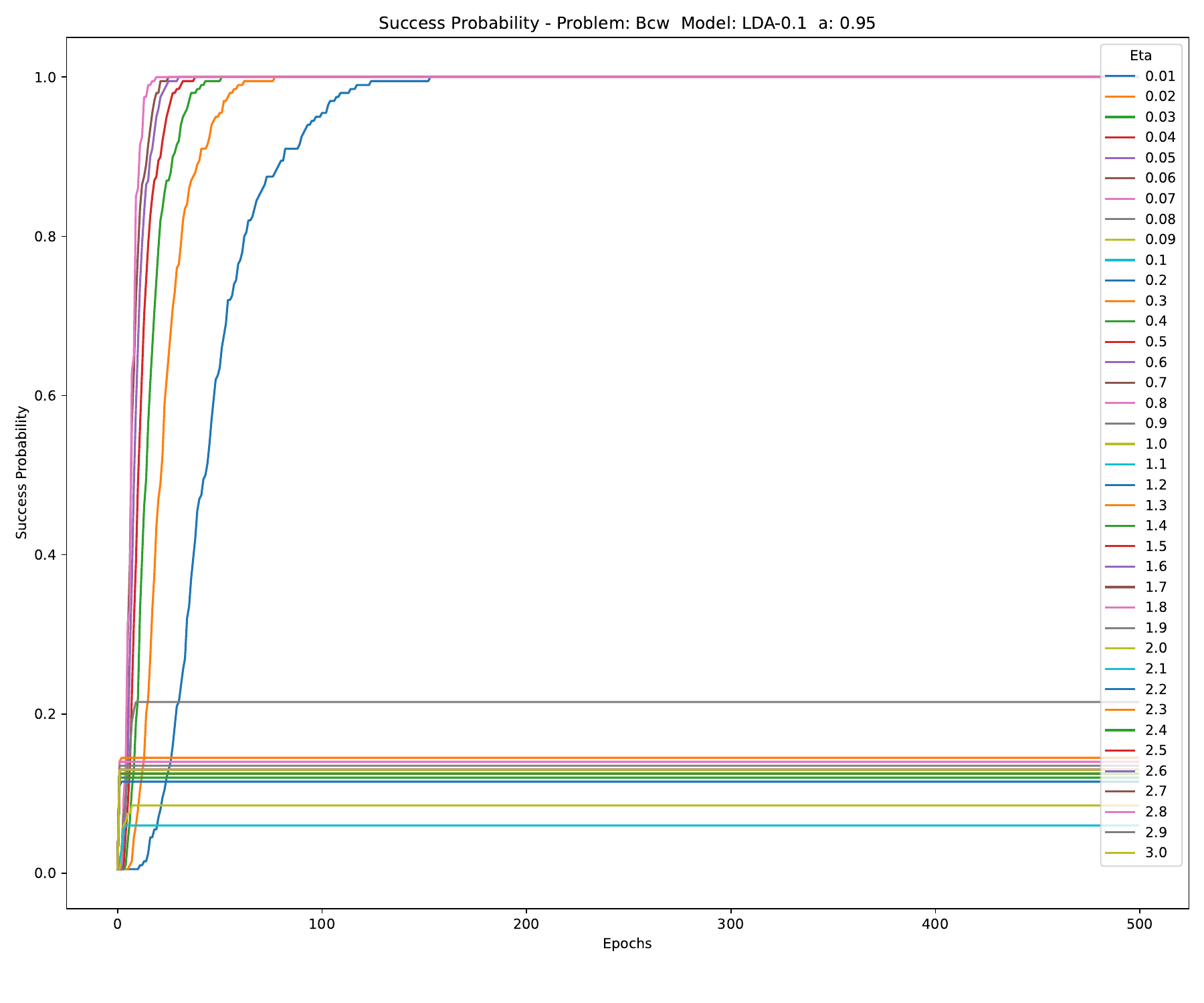} &
    \includegraphics[clip,trim=0 0 0 0,width=.52\linewidth]{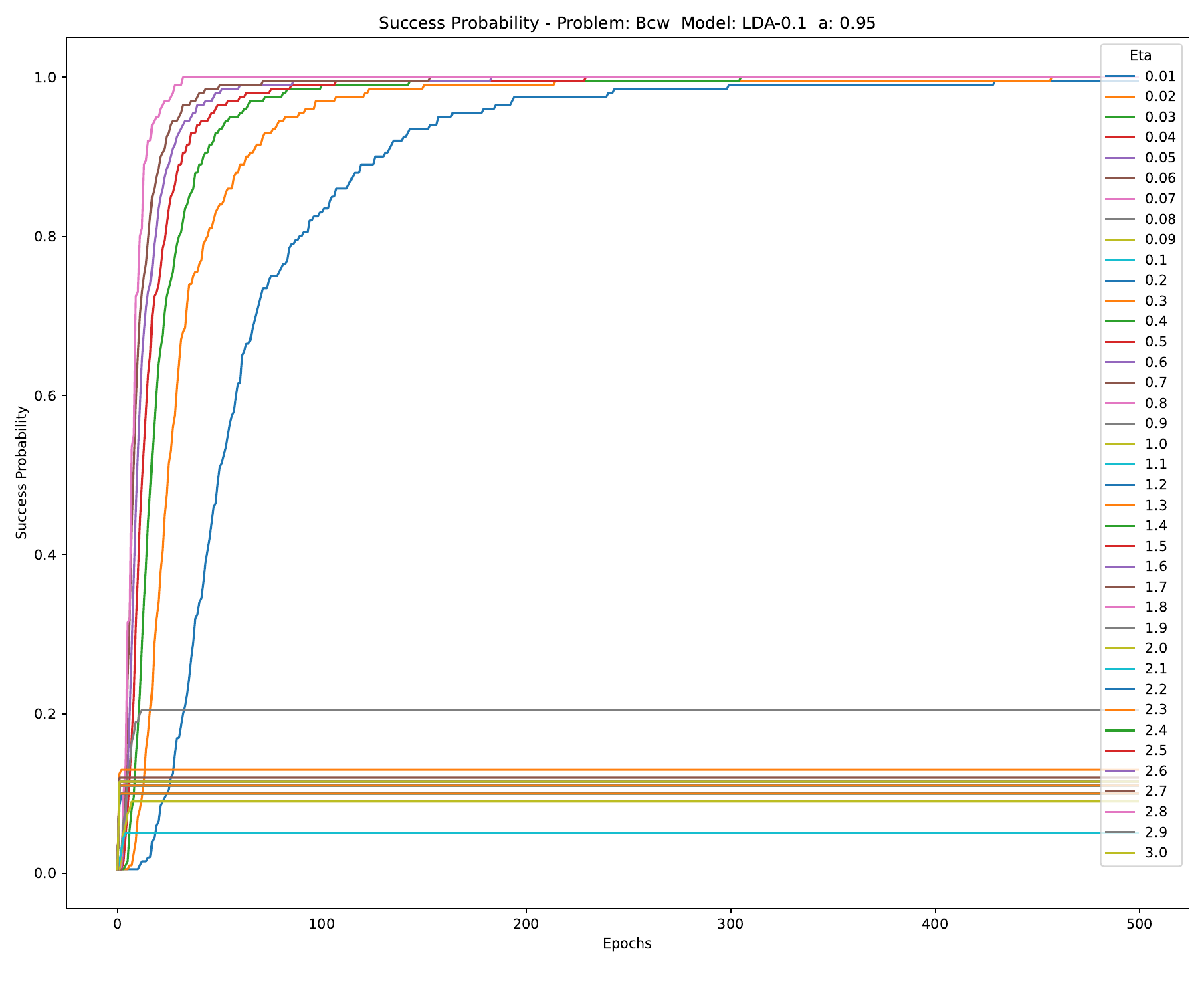} \\[-3mm]
    \includegraphics[clip,trim=0 0 0 0,width=.52\linewidth]{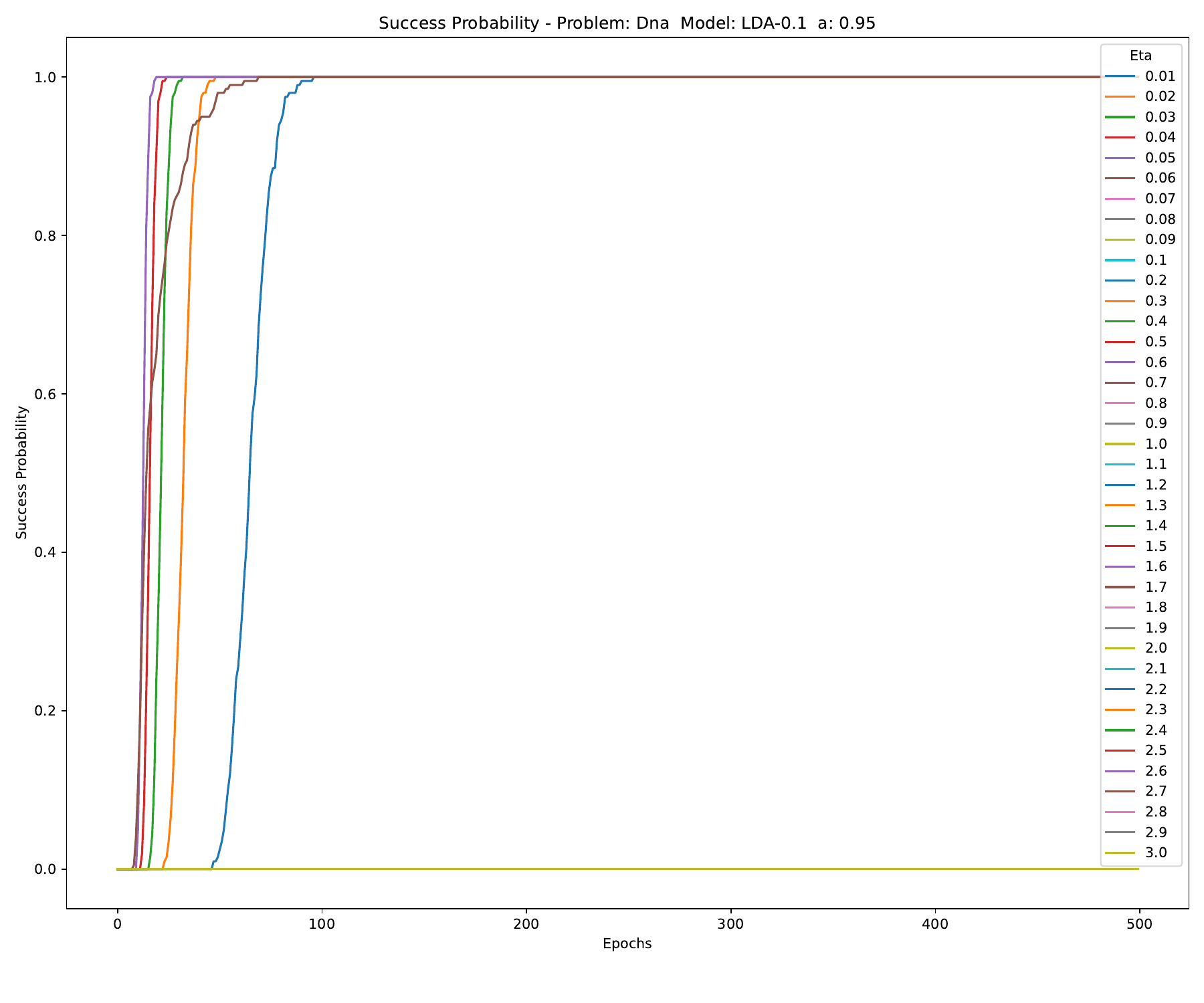} &
    \includegraphics[clip,trim=0 0 0 0,width=.52\linewidth]{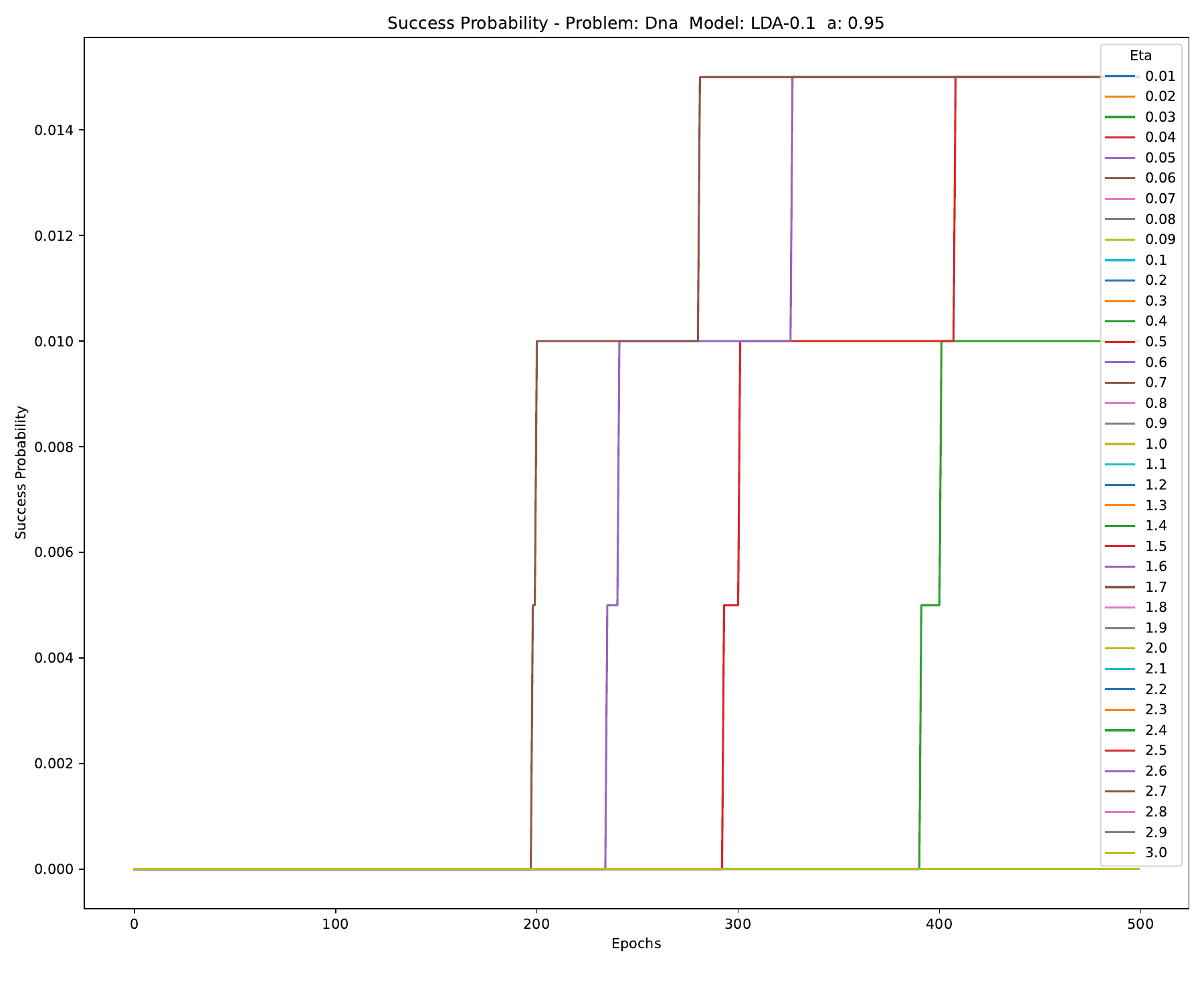} \\[-3mm]
    \includegraphics[clip,trim=0 0 0 0,width=.52\linewidth]{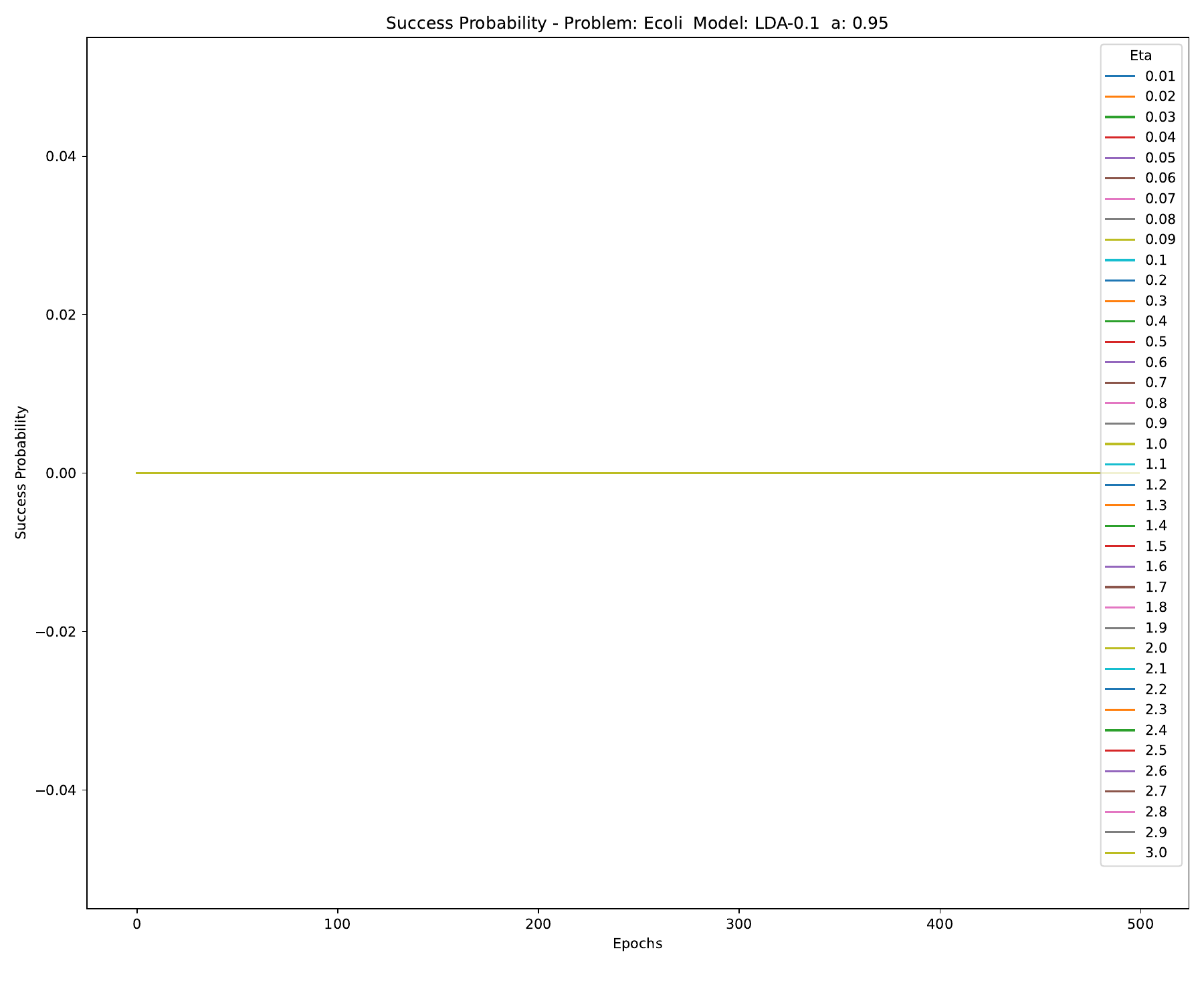} &
    \includegraphics[clip,trim=0 0 0 0,width=.52\linewidth]{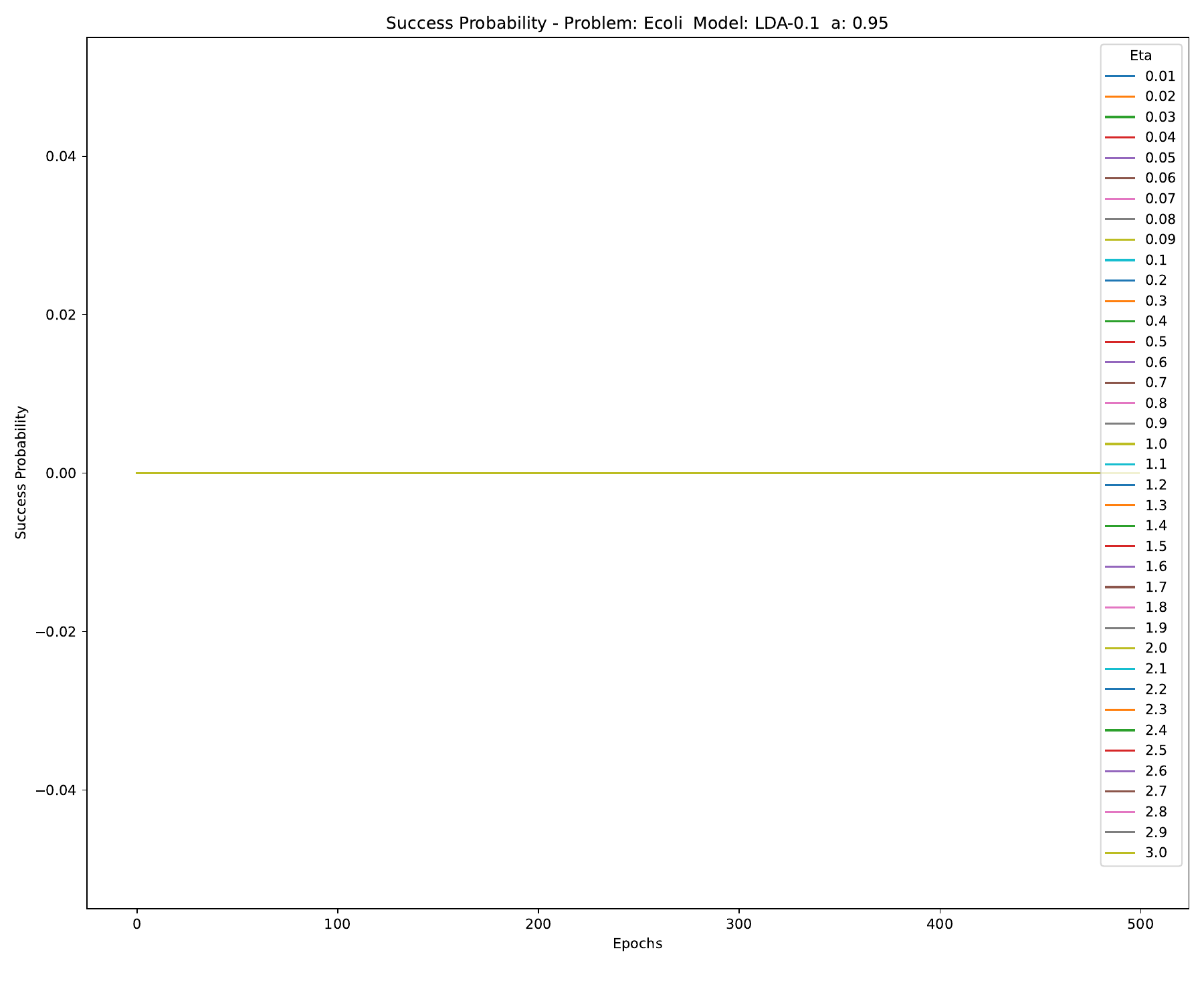} \\[-3mm]
  \end{tabular}
\caption{Success probability ($a=0.95$) for LDA-0.1 model.}
  \label{fig:success_probability_LDA_0.1_model_0.95}
\end{figure*}

\begin{figure*}[p]
  \centering
  \ContinuedFloat
  \begin{tabular}{c@{}c}
    Training Success Probability & Validation Success Probability \\
    \includegraphics[clip,trim=0 0 0 0,width=.52\linewidth]{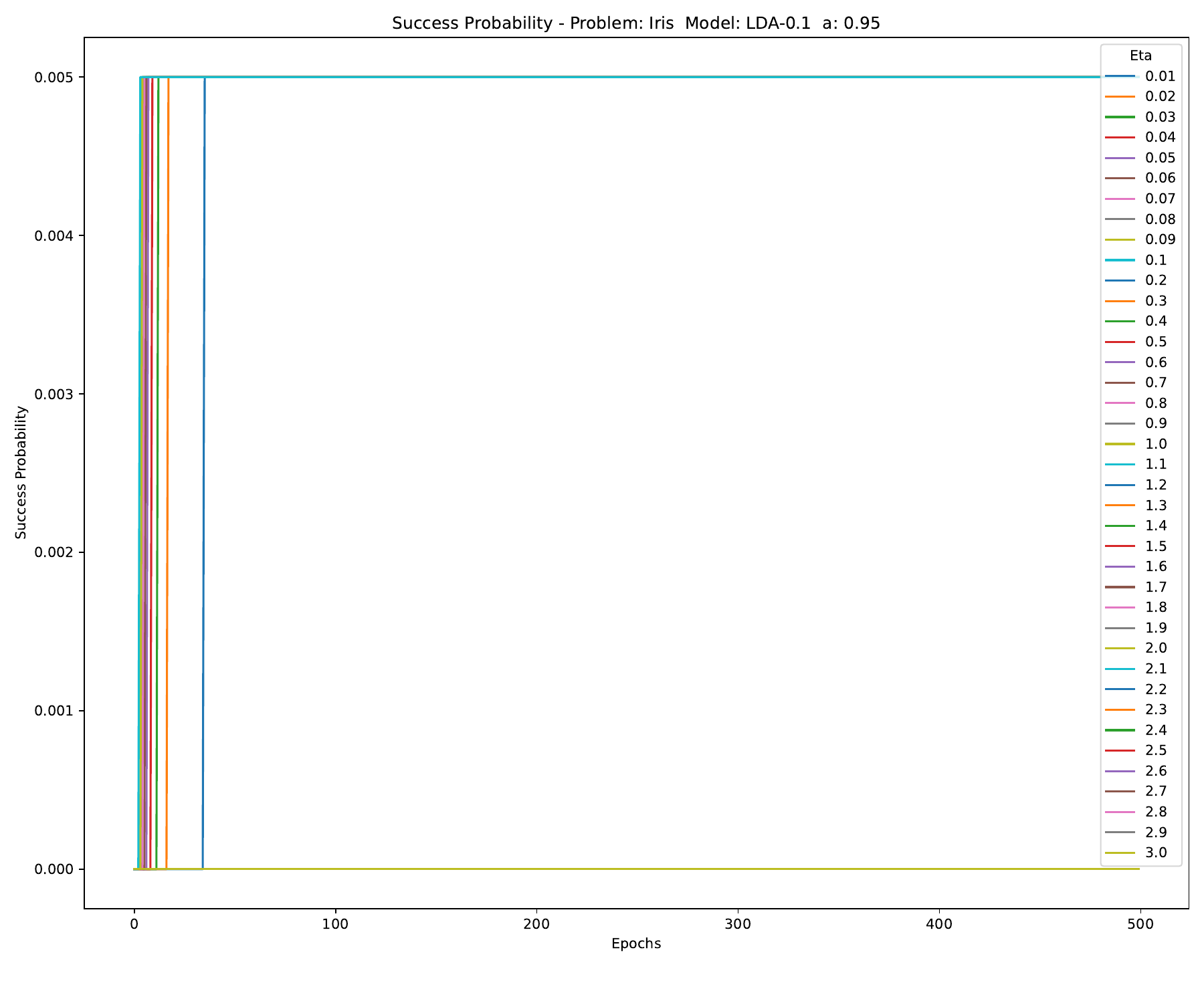} &
    \includegraphics[clip,trim=0 0 0 0,width=.52\linewidth]{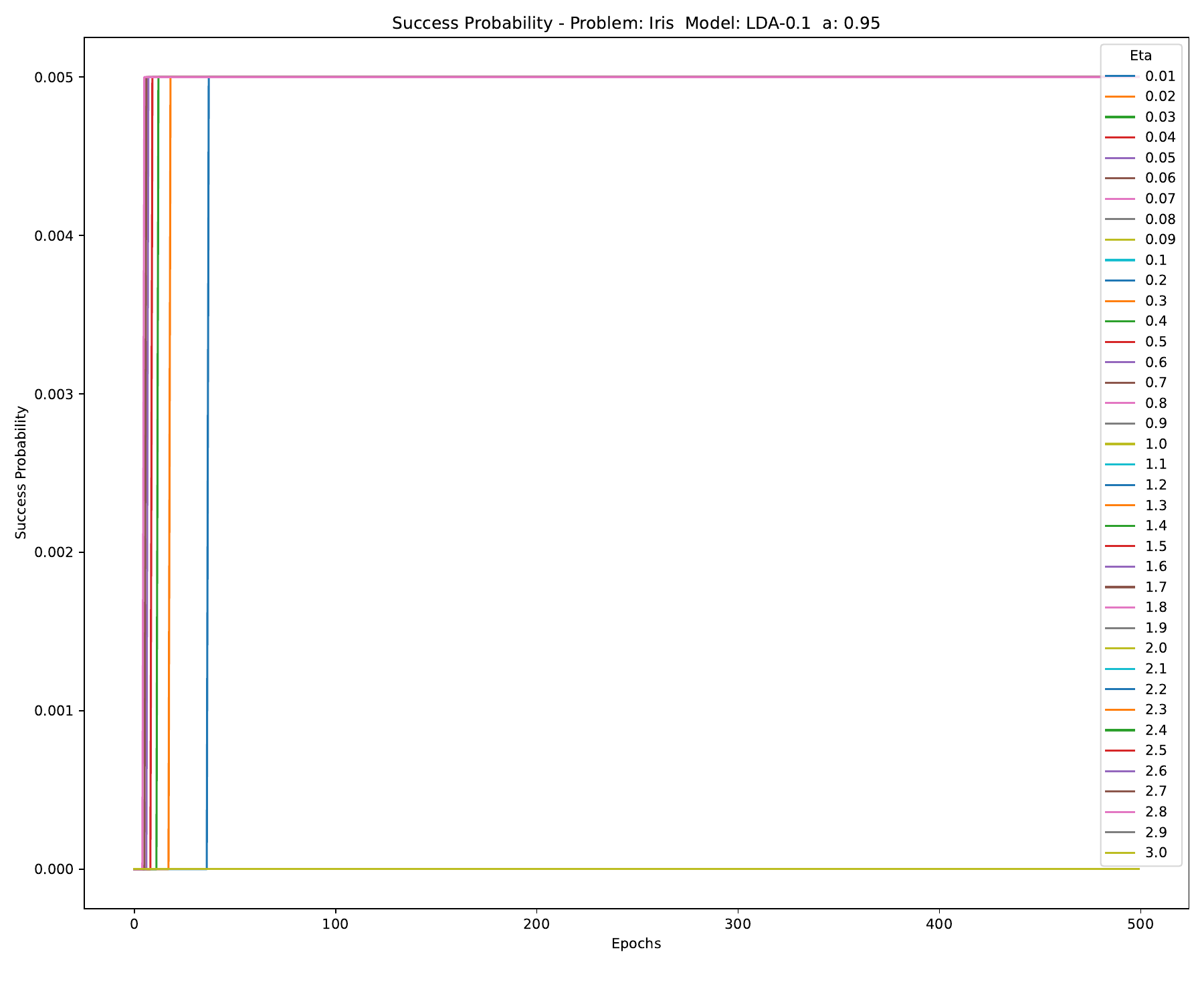} \\[-3mm]
    \includegraphics[clip,trim=0 0 0 0,width=.52\linewidth]{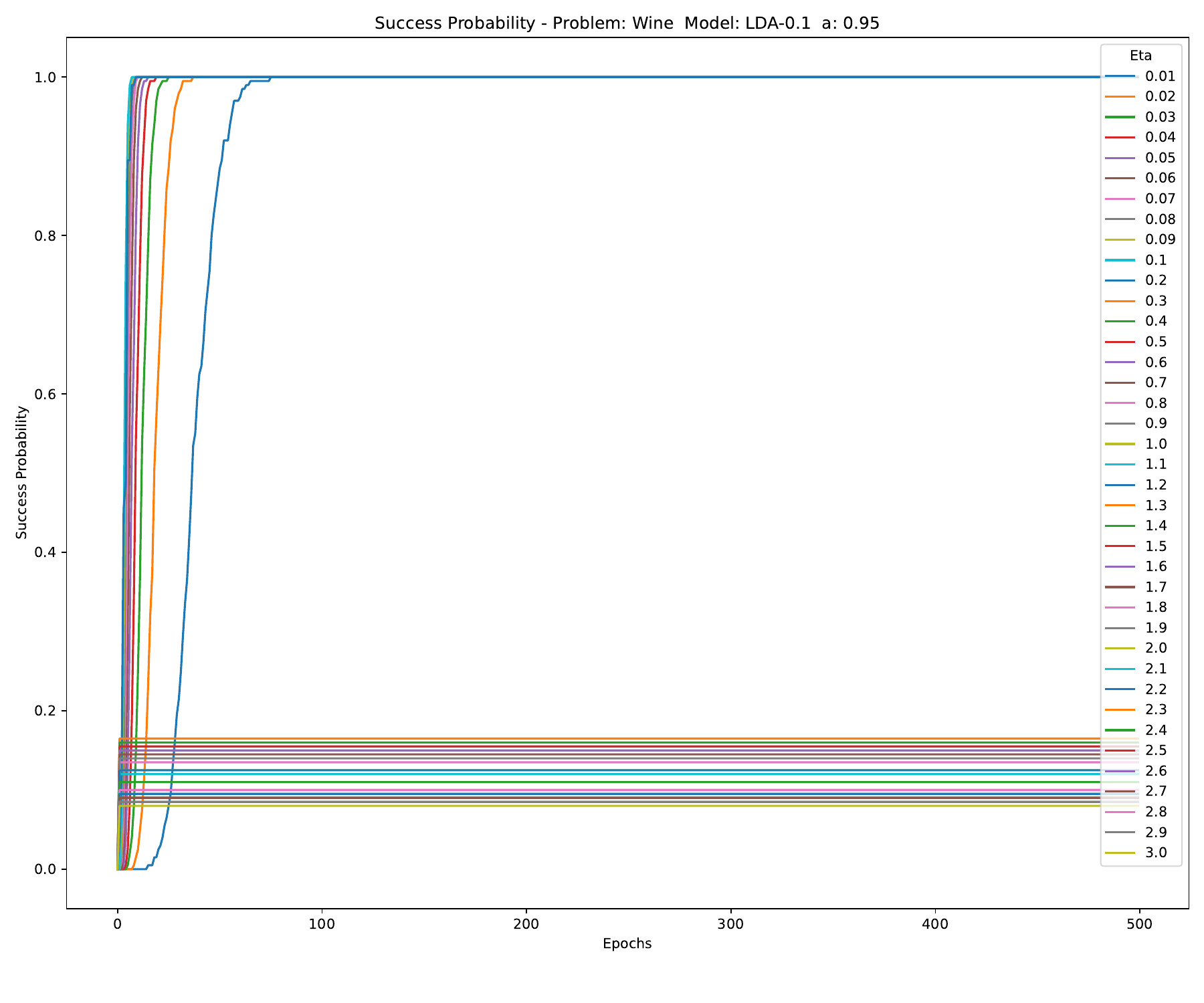} &
    \includegraphics[clip,trim=0 0 0 0,width=.52\linewidth]{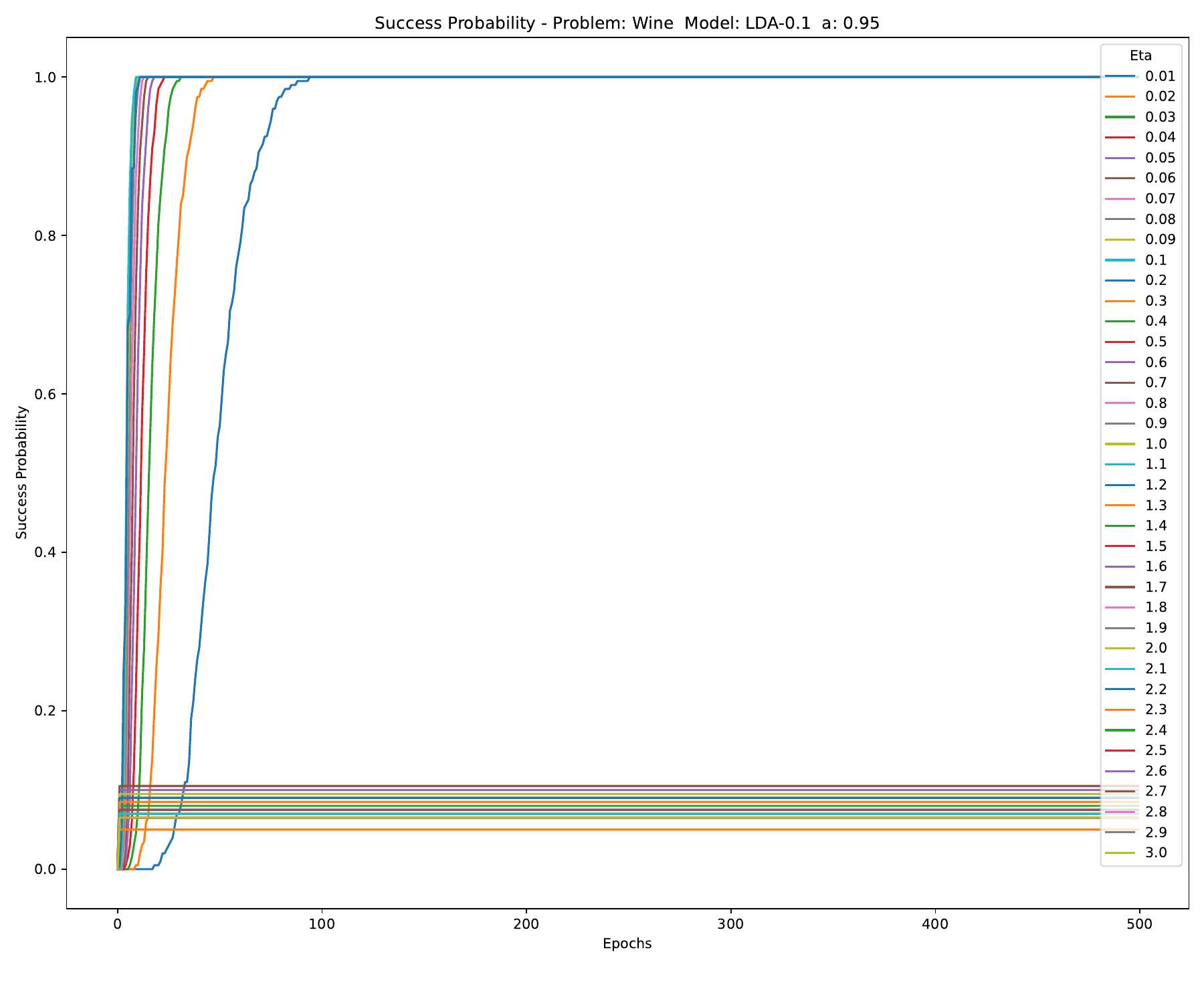} \\[-3mm]
  \end{tabular}
\caption{Success probability ($a=0.95$) for LDA-0.1 model (continued).}
\end{figure*}

\begin{figure*}[p]
  \centering
  \begin{tabular}{c@{}c}
    Training Success Probability & Validation Success Probability \\
    \includegraphics[clip,trim=0 0 0 0,width=.52\linewidth]{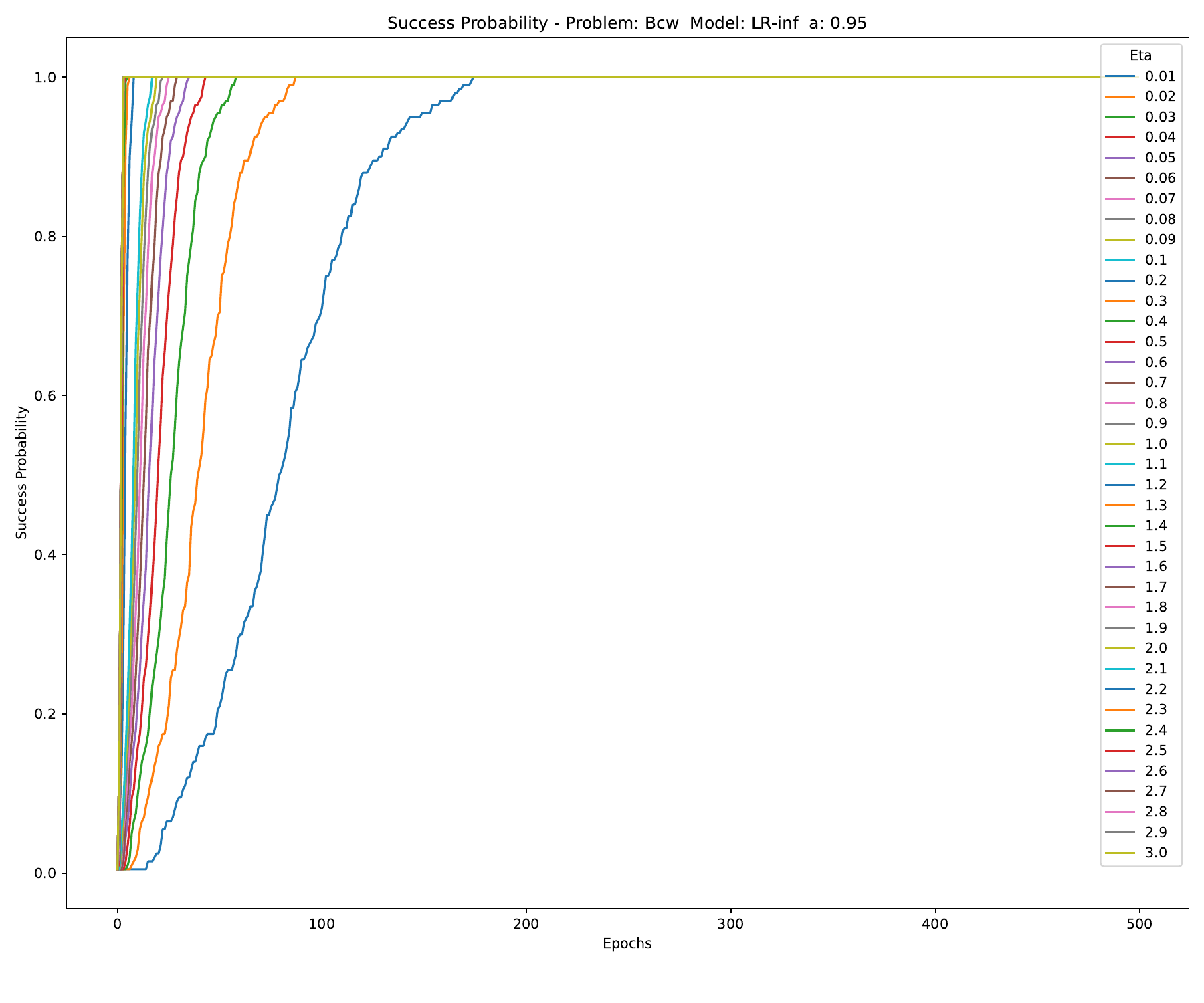} &
    \includegraphics[clip,trim=0 0 0 0,width=.52\linewidth]{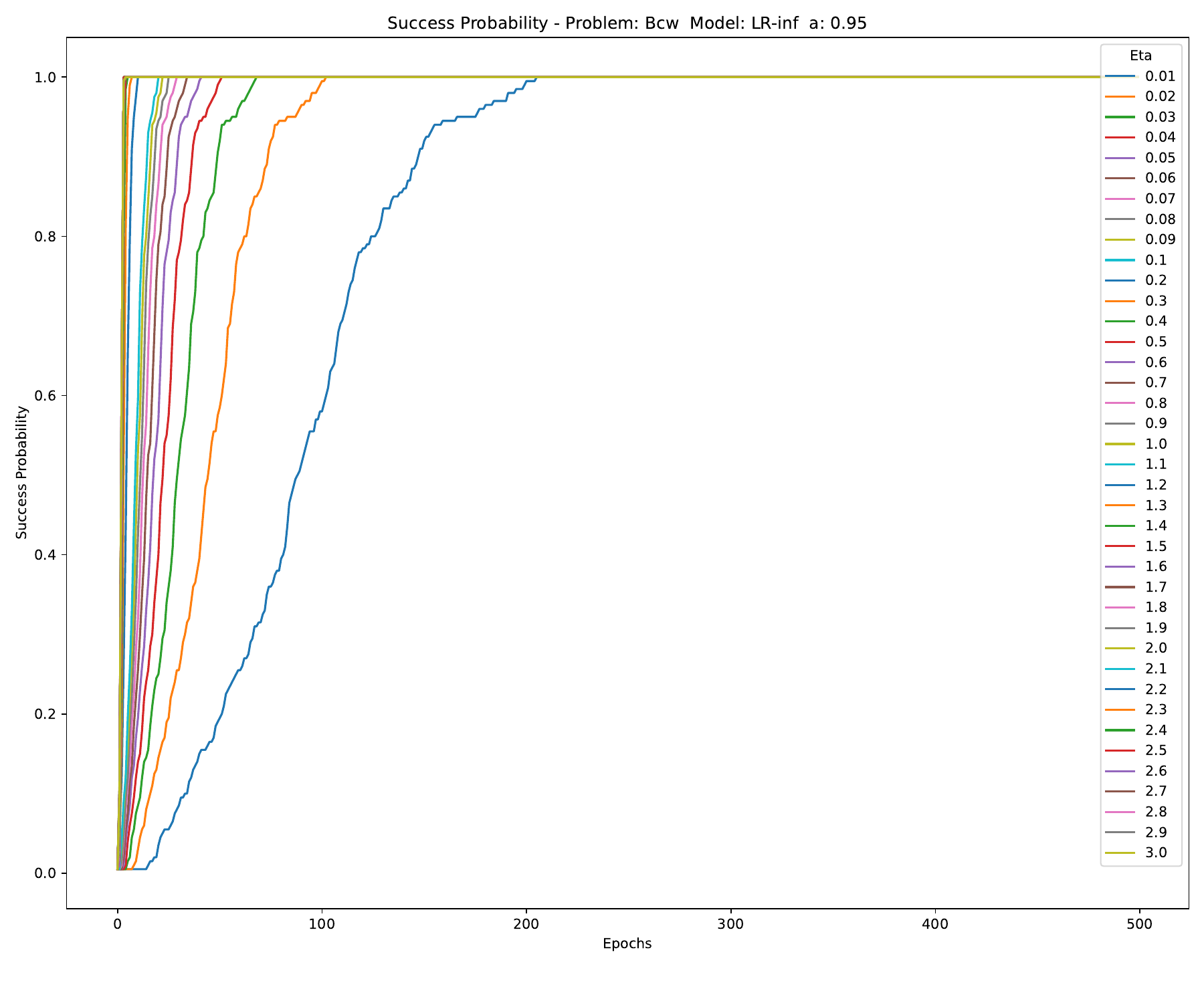} \\[-3mm]
    \includegraphics[clip,trim=0 0 0 0,width=.52\linewidth]{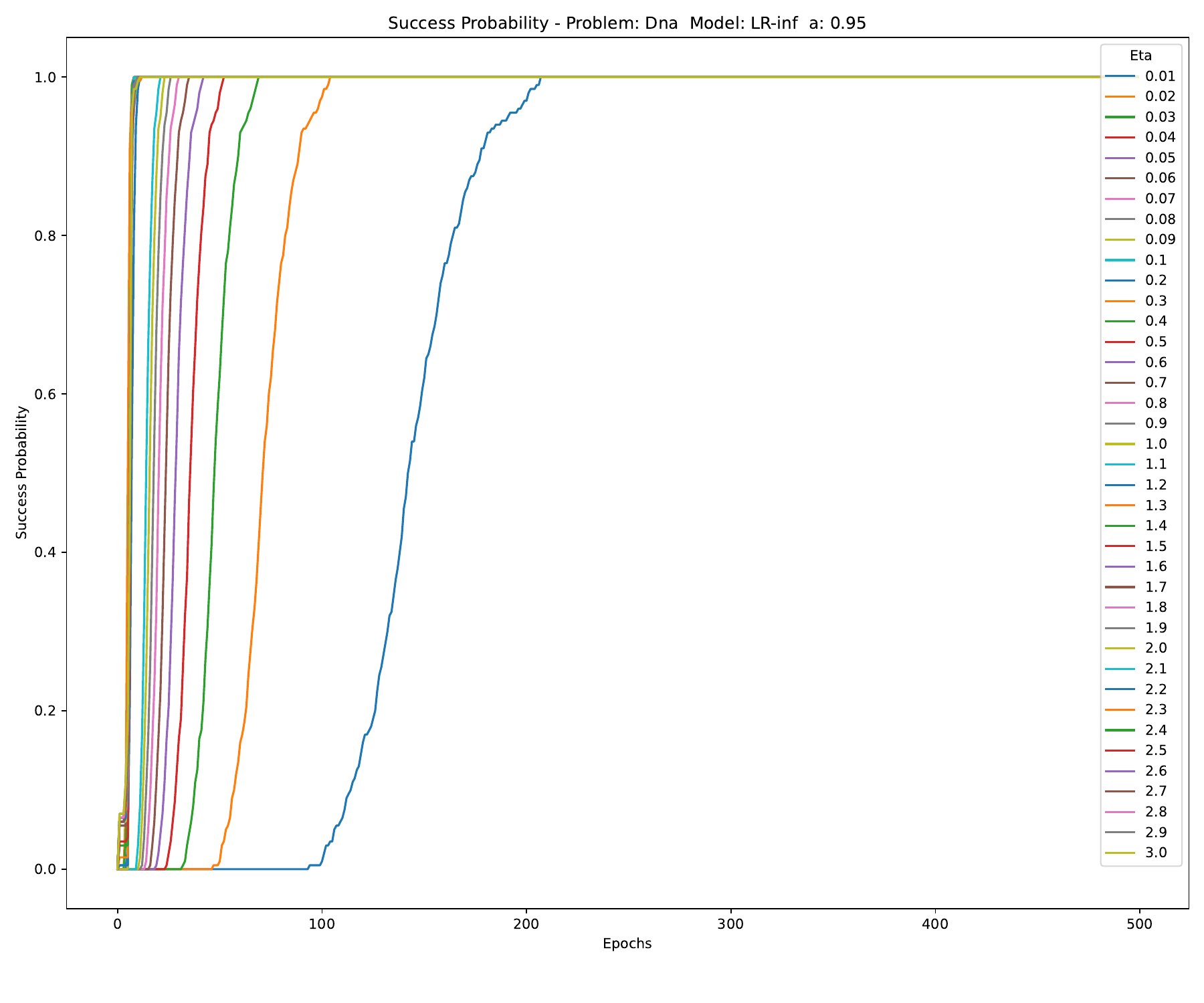} &
    \includegraphics[clip,trim=0 0 0 0,width=.52\linewidth]{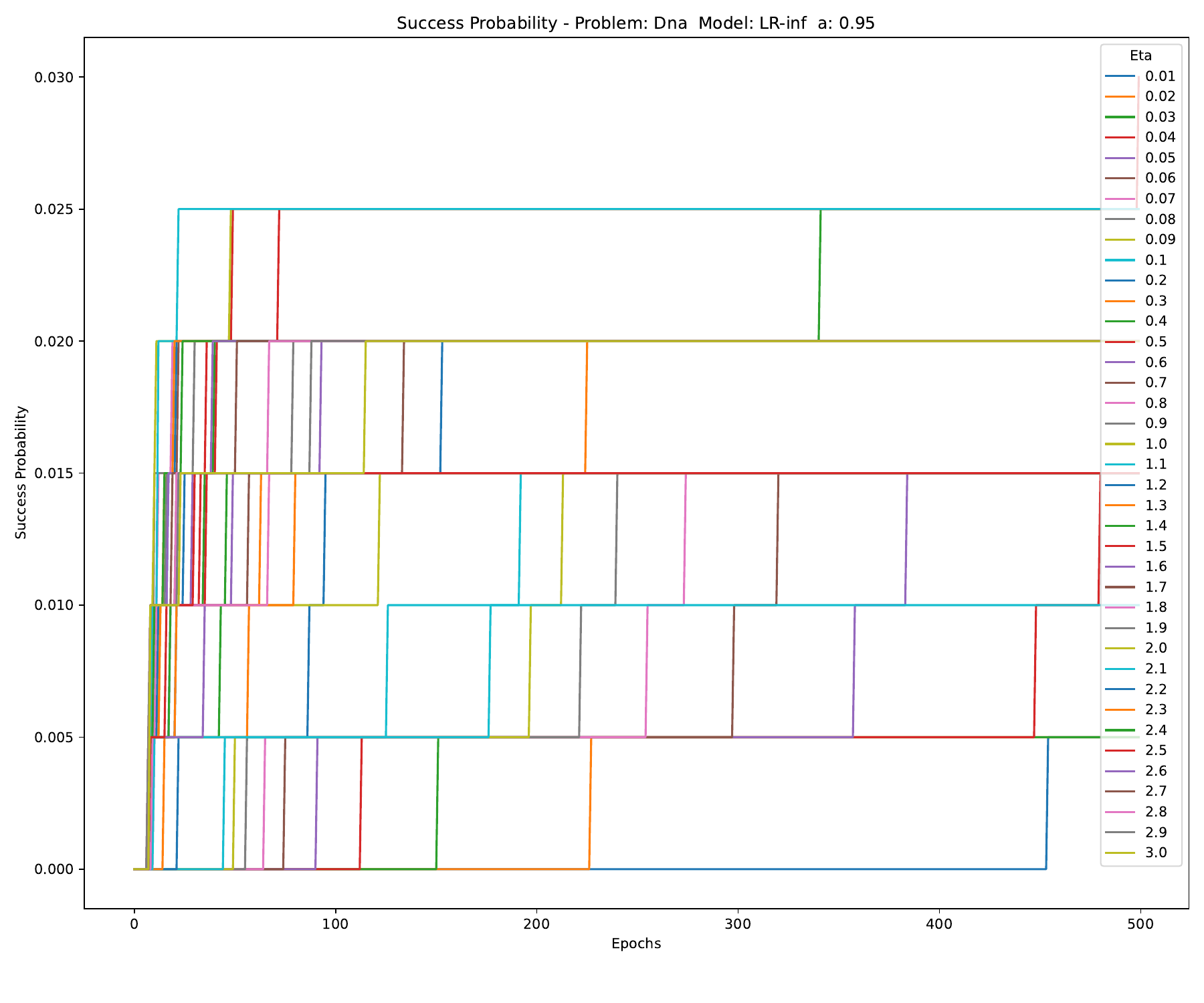} \\[-3mm]
    \includegraphics[clip,trim=0 0 0 0,width=.52\linewidth]{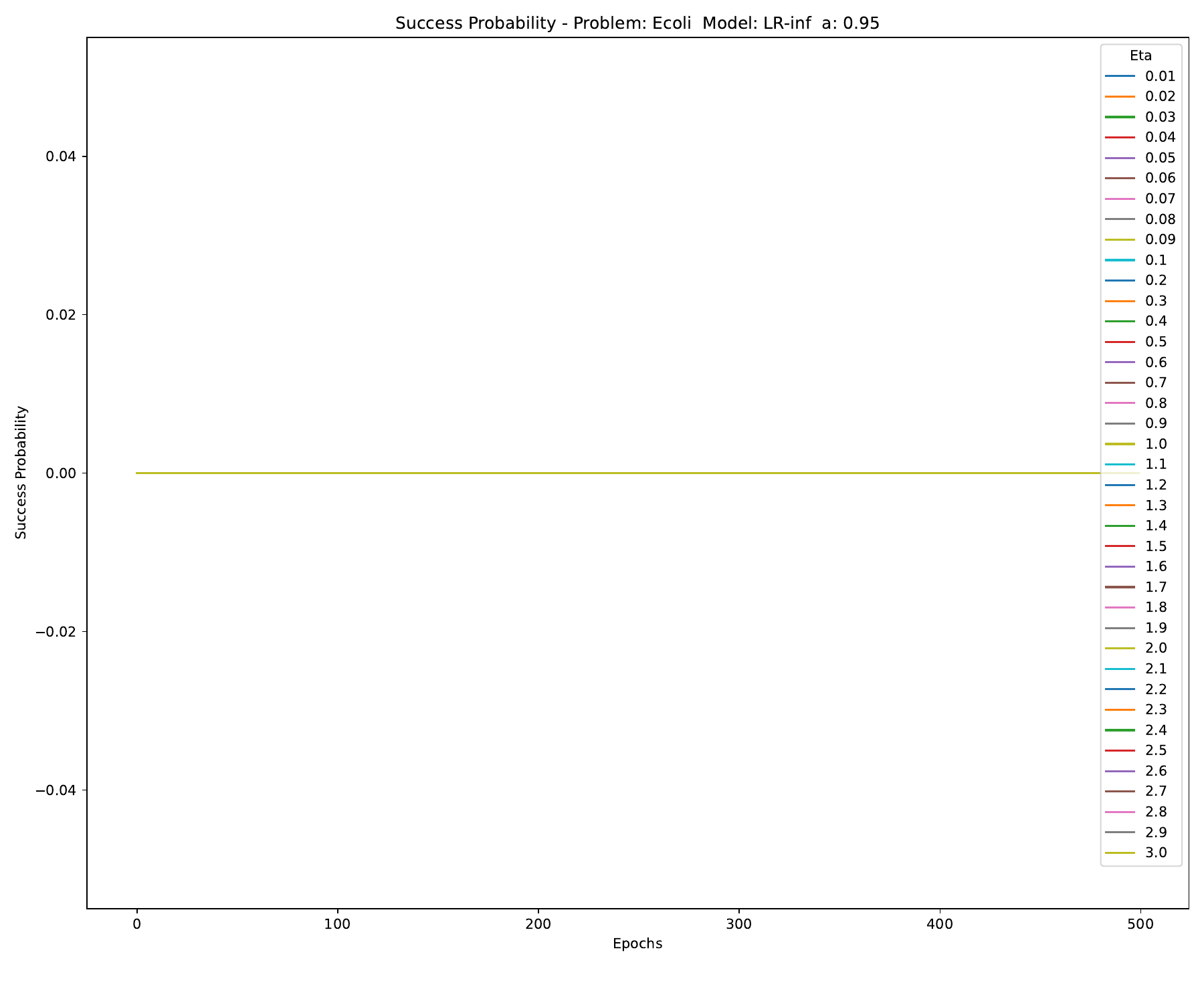} &
    \includegraphics[clip,trim=0 0 0 0,width=.52\linewidth]{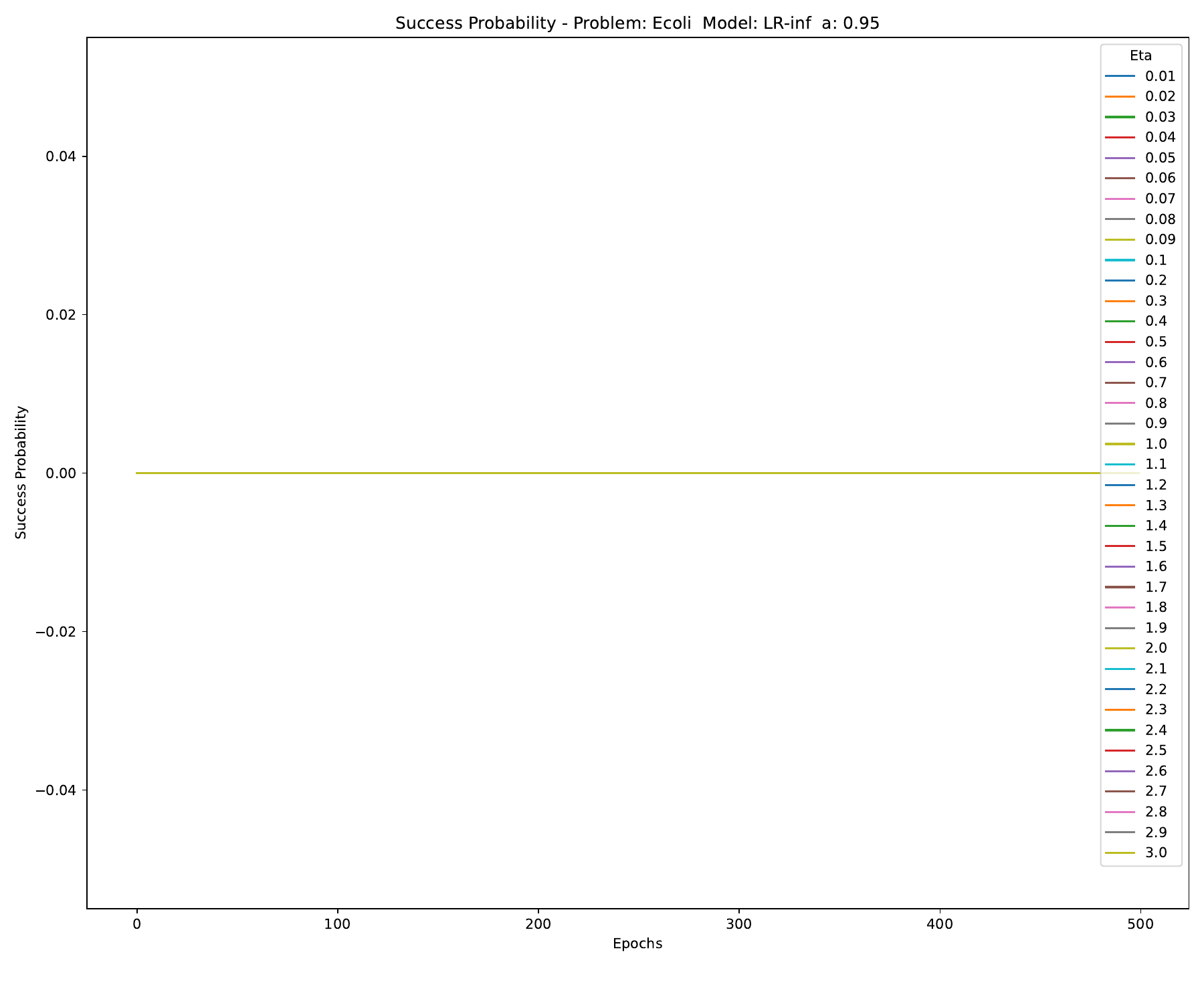} \\[-3mm]
  \end{tabular}
\caption{Success probability ($a=0.95$) for LR-Inf model.}
  \label{fig:success_probability_LR_Inf_model_0.95}
\end{figure*}

\begin{figure*}[p]
  \centering
  \ContinuedFloat
  \begin{tabular}{c@{}c}
    Training Success Probability & Validation Success Probability \\
    \includegraphics[clip,trim=0 0 0 0,width=.52\linewidth]{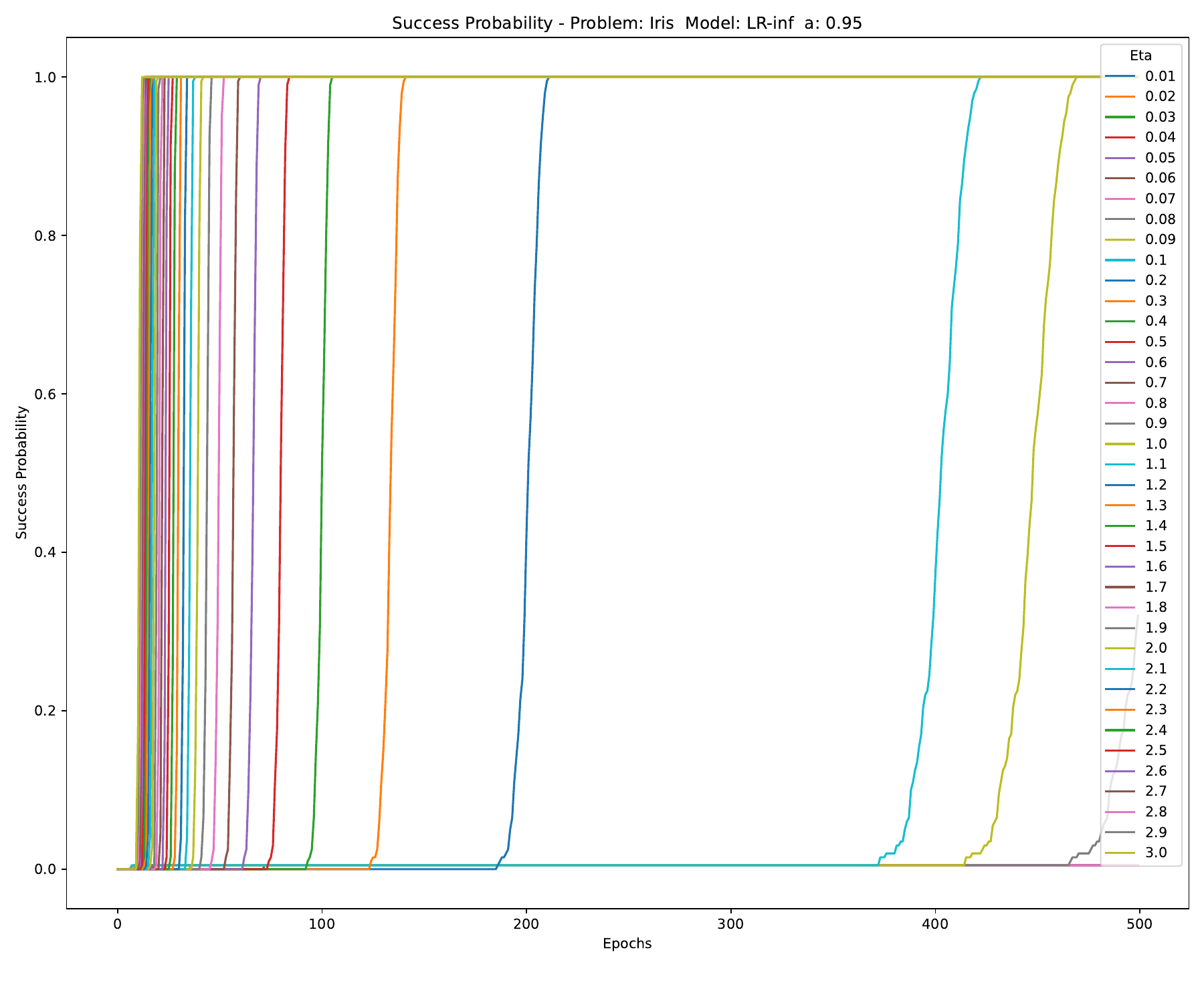} &
    \includegraphics[clip,trim=0 0 0 0,width=.52\linewidth]{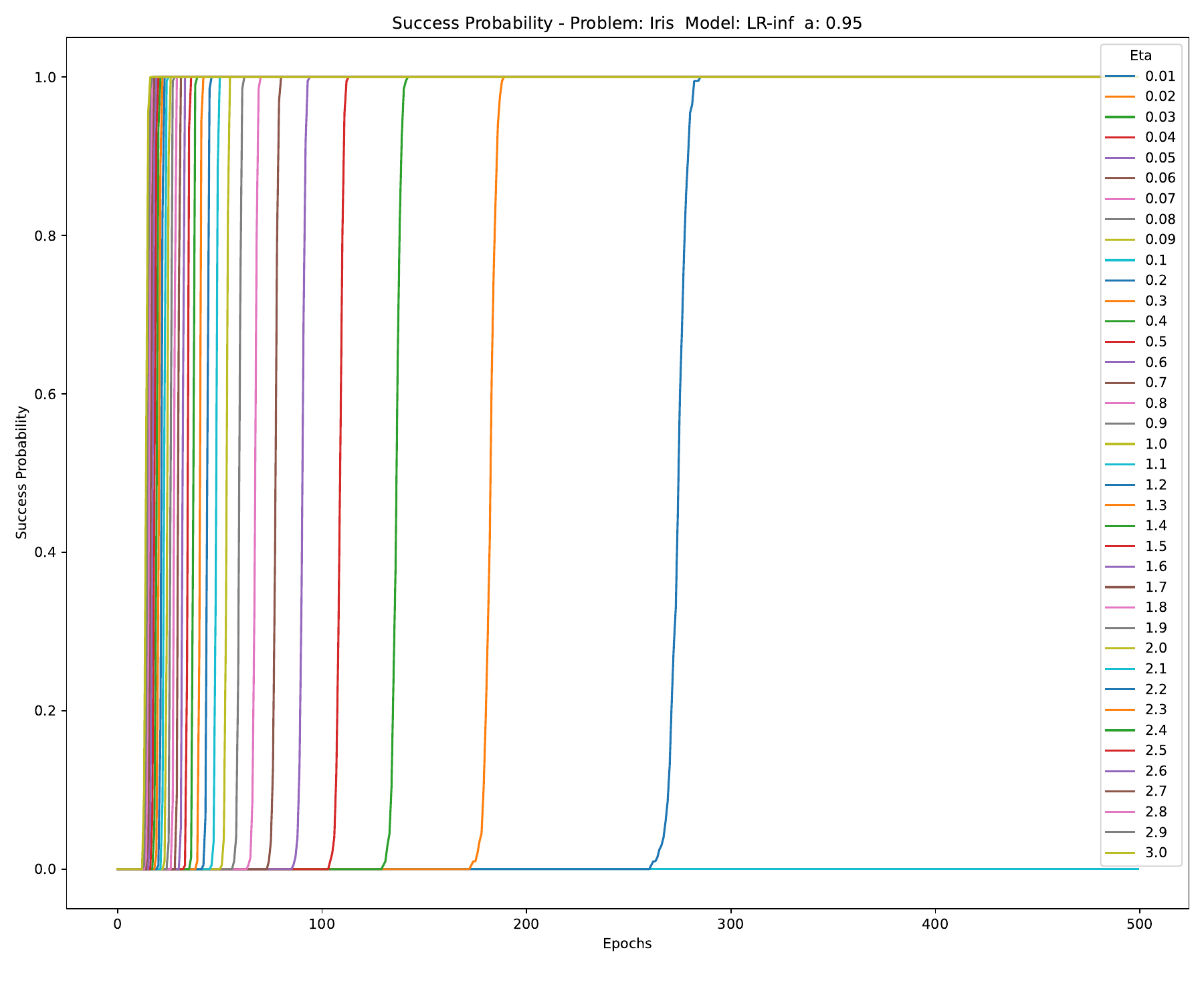} \\[-3mm]
    \includegraphics[clip,trim=0 0 0 0,width=.52\linewidth]{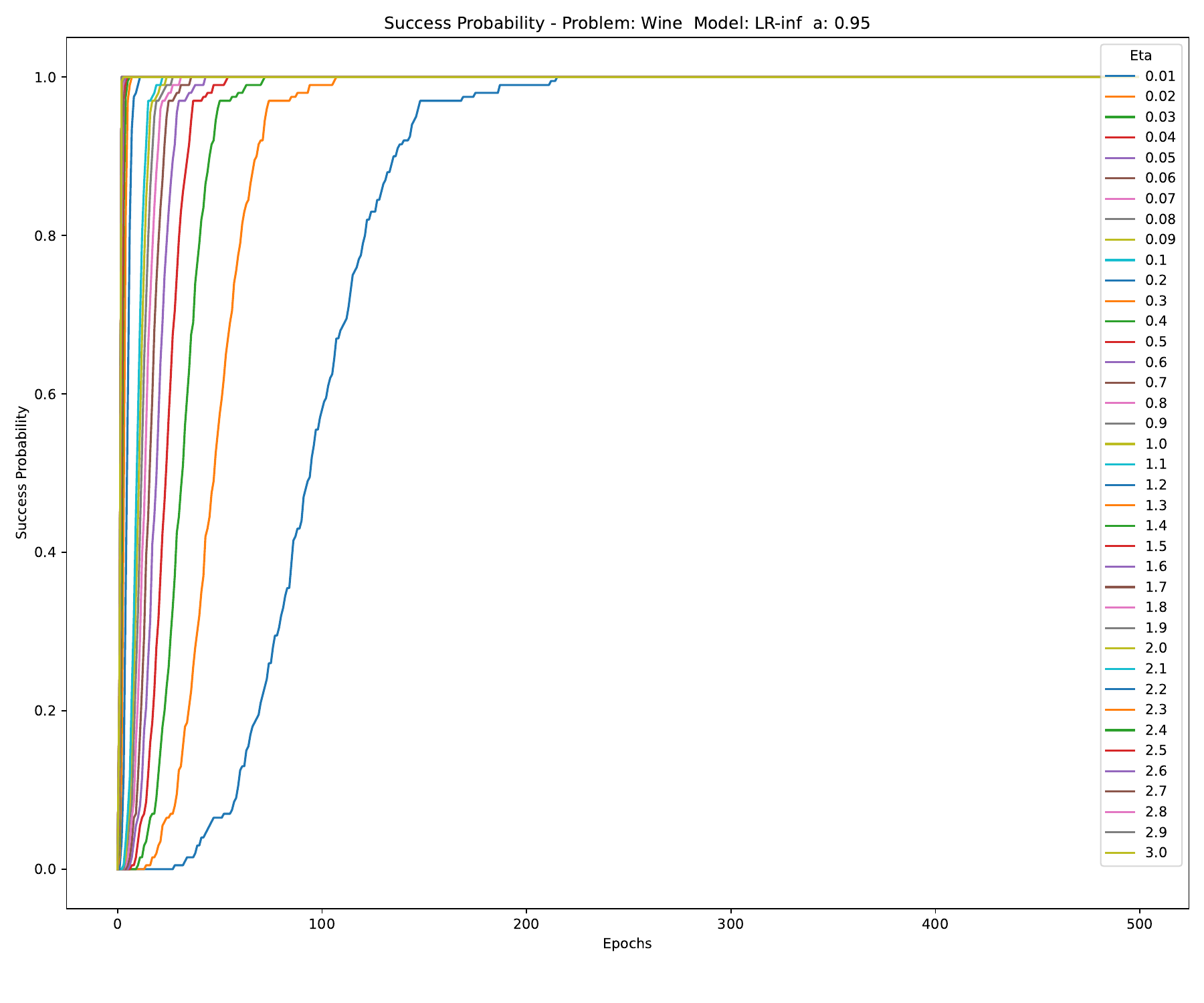} &
    \includegraphics[clip,trim=0 0 0 0,width=.52\linewidth]{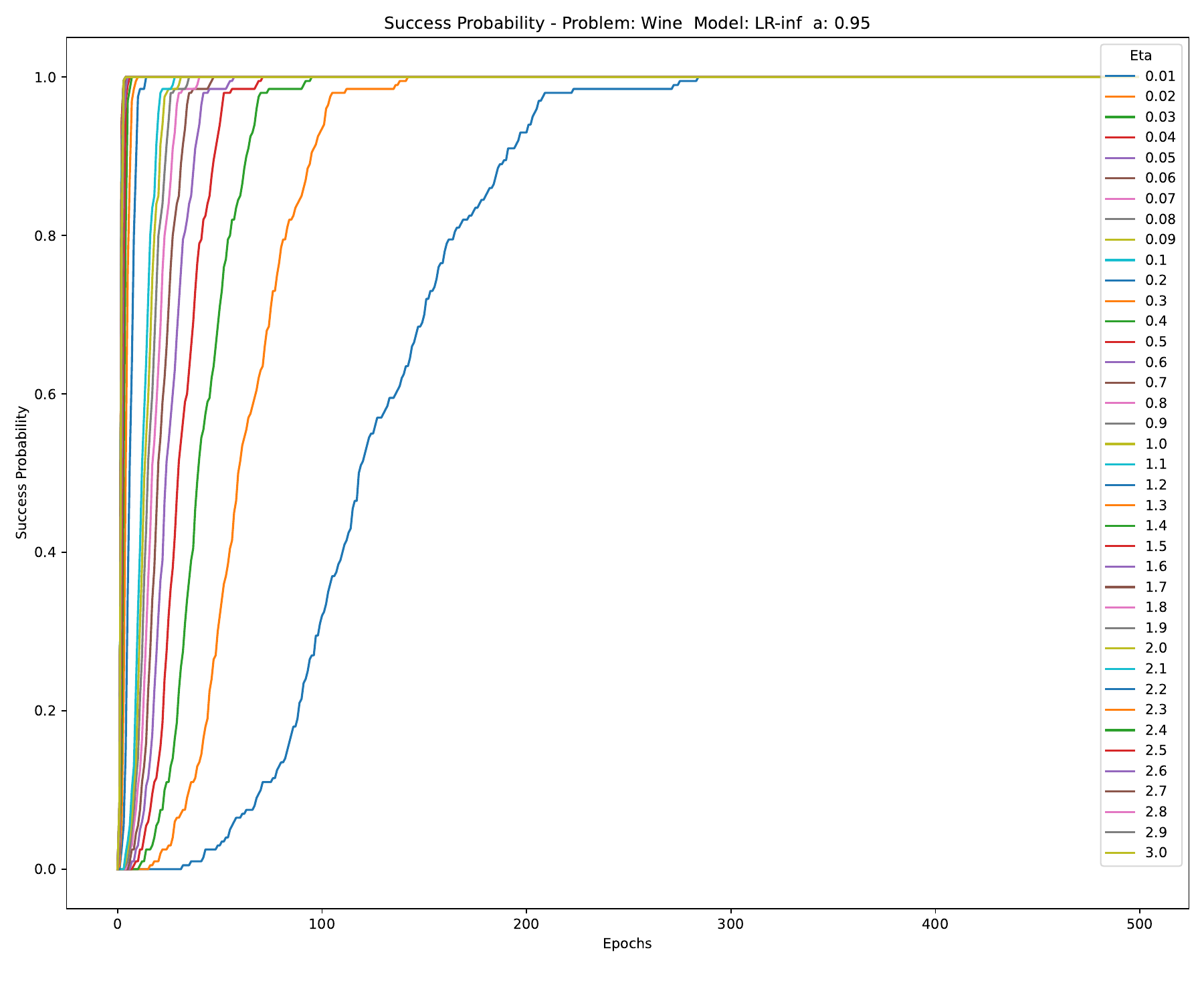} \\[-3mm]
  \end{tabular}
\caption{Success probability ($a=0.95$) for LR-Inf model (continued).}
\end{figure*}

\begin{figure*}[p]
  \centering
  \begin{tabular}{c@{}c}
    Training Success Probability & Validation Success Probability \\
    \includegraphics[clip,trim=0 0 0 0,width=.52\linewidth]{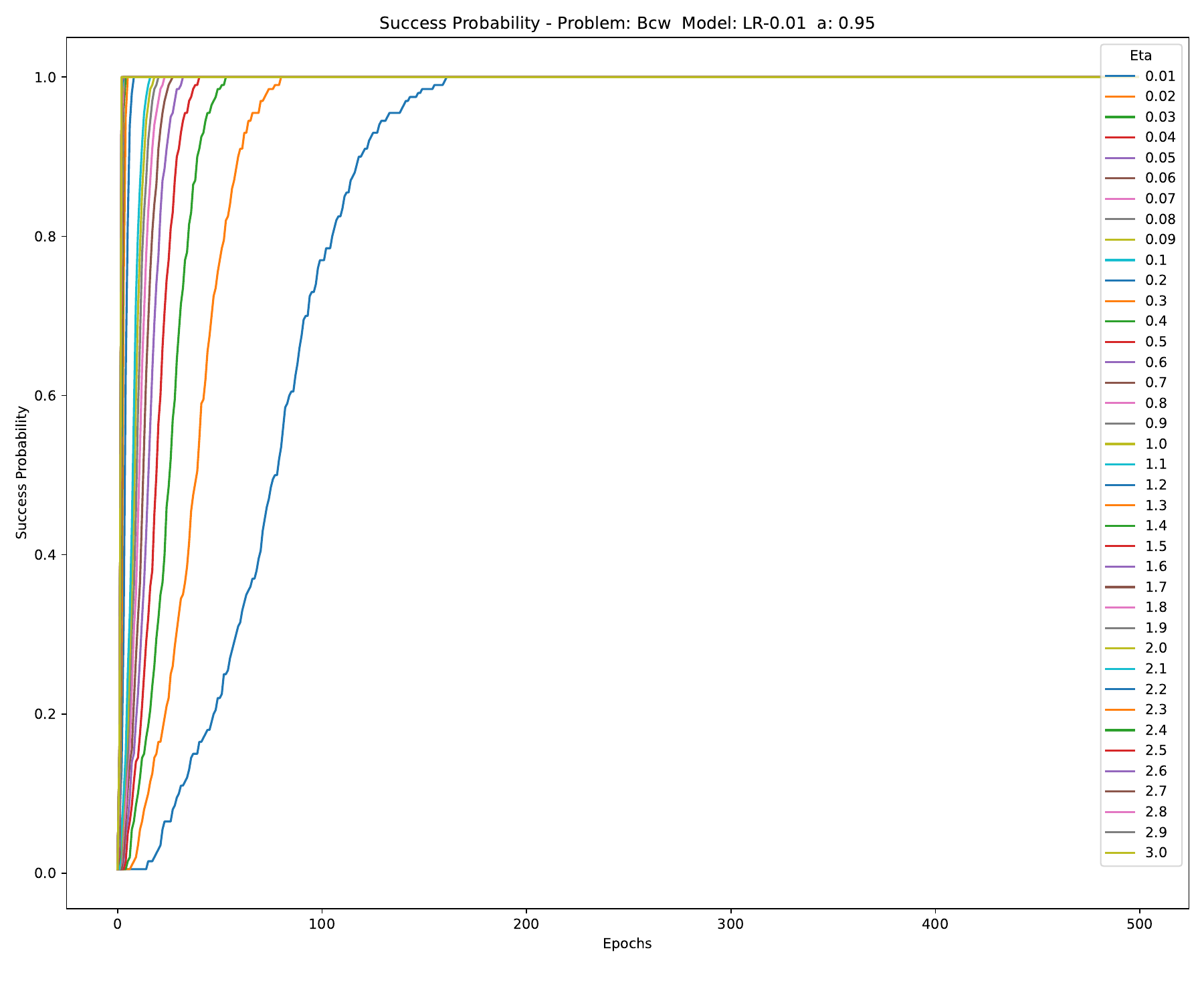} &
    \includegraphics[clip,trim=0 0 0 0,width=.52\linewidth]{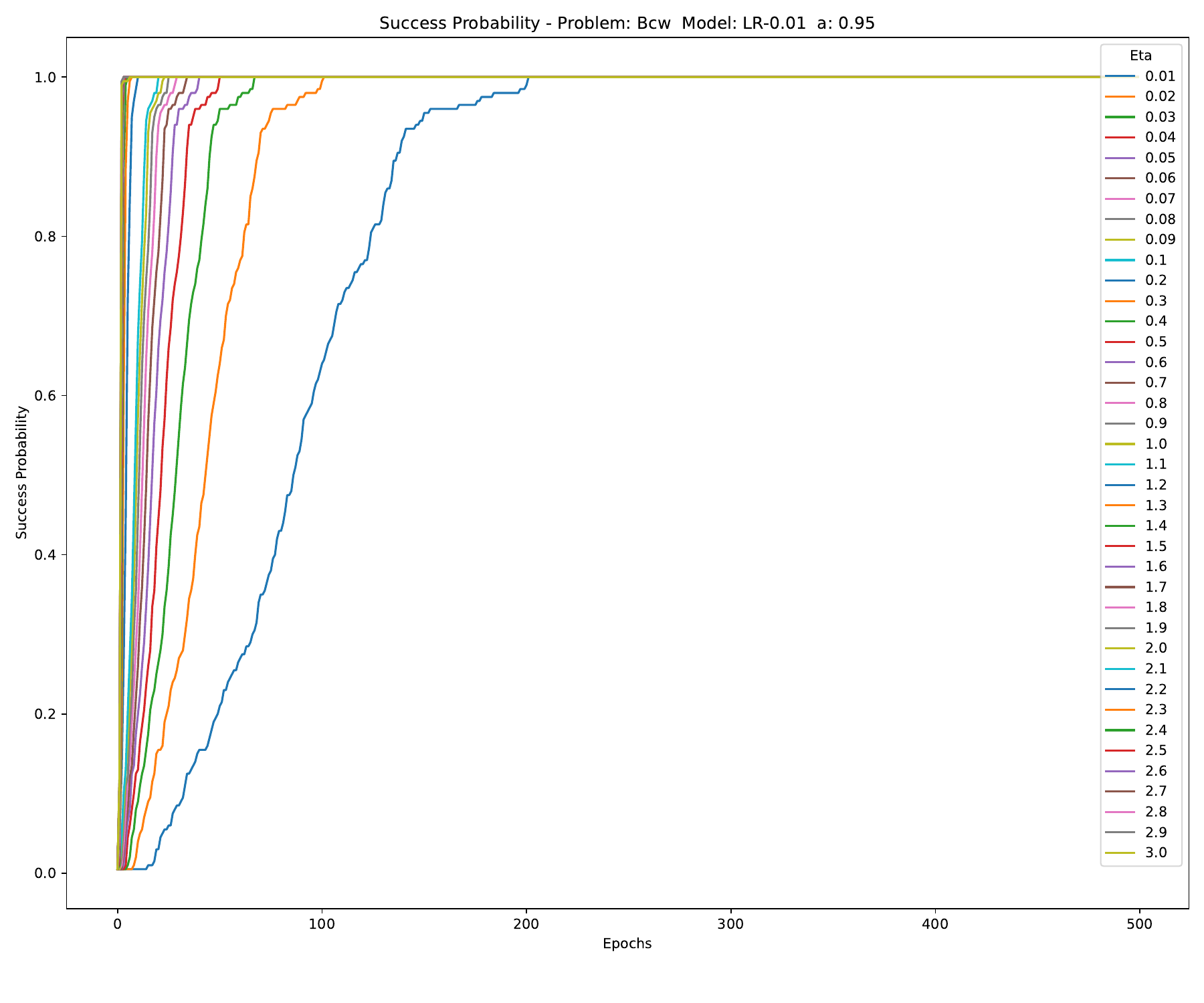} \\[-3mm]
    \includegraphics[clip,trim=0 0 0 0,width=.52\linewidth]{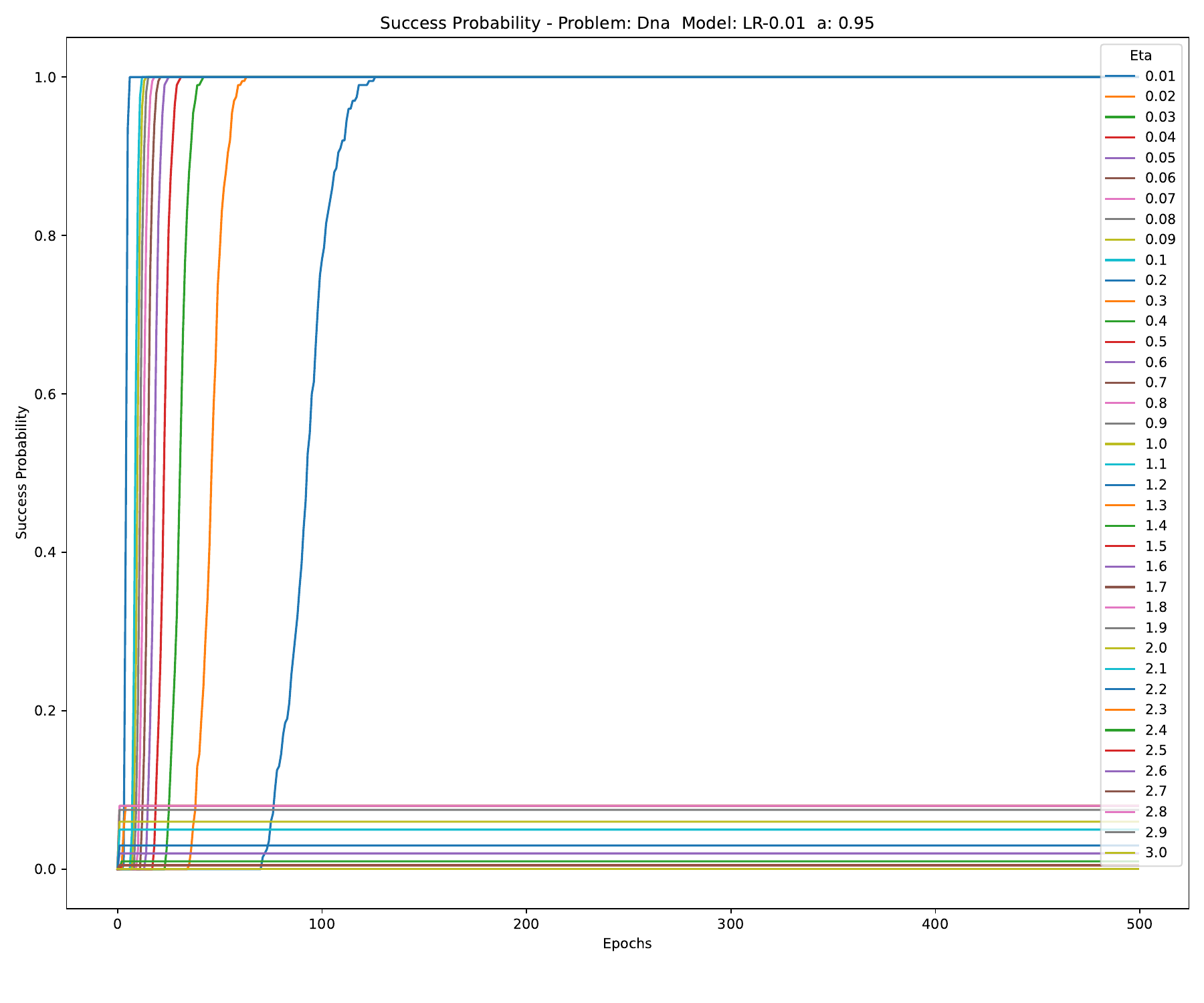} &
    \includegraphics[clip,trim=0 0 0 0,width=.52\linewidth]{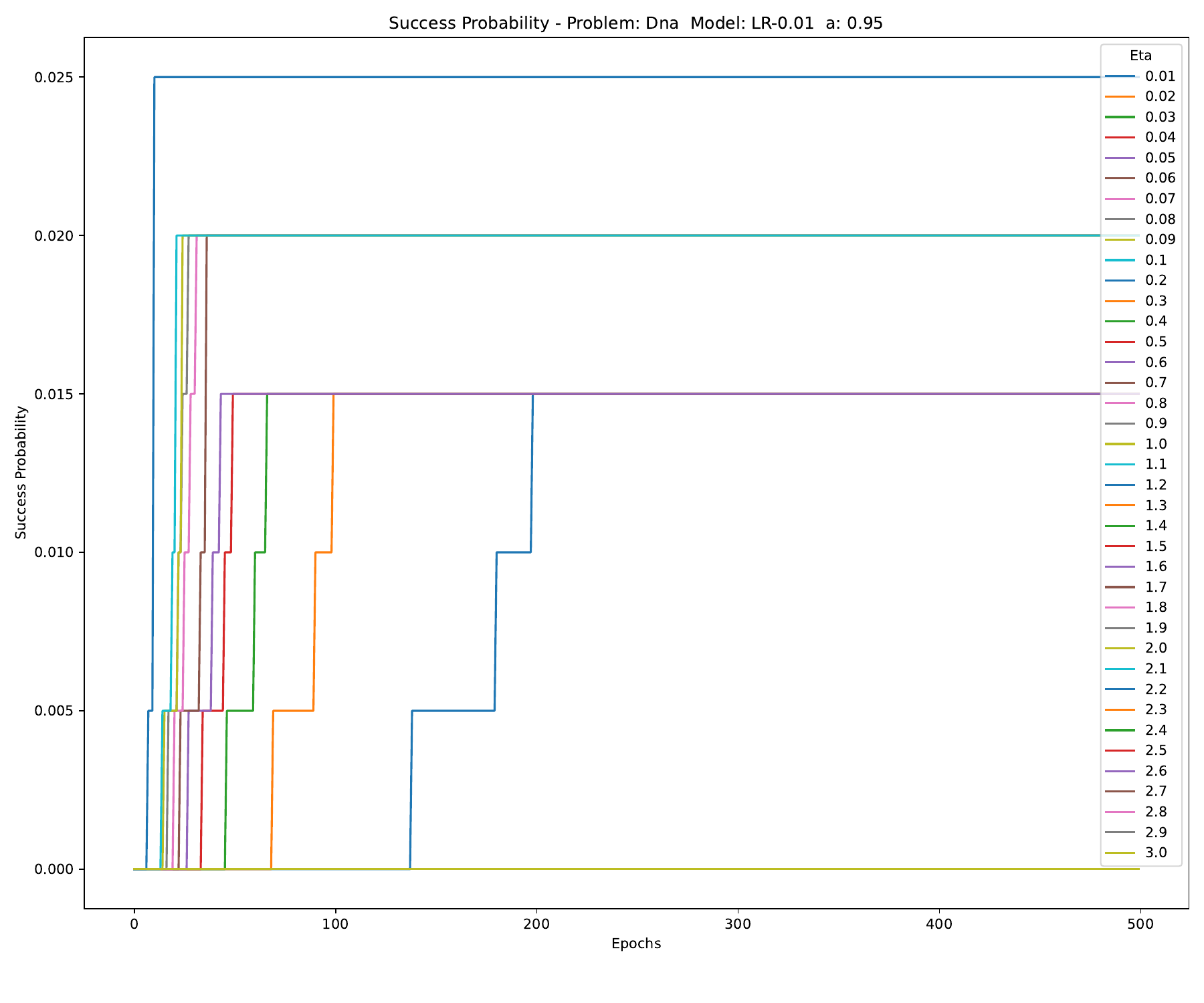} \\[-3mm]
    \includegraphics[clip,trim=0 0 0 0,width=.52\linewidth]{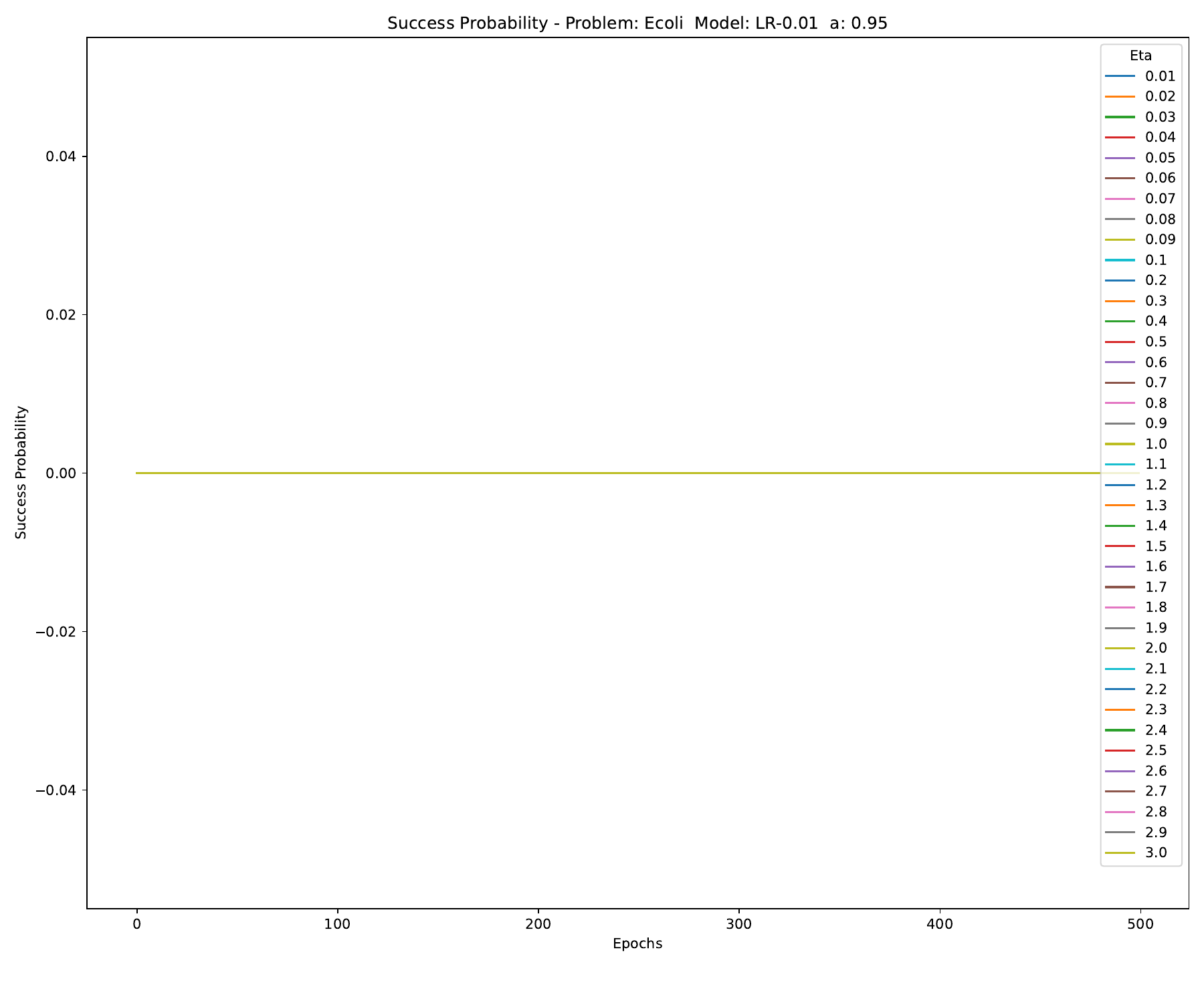} &
    \includegraphics[clip,trim=0 0 0 0,width=.52\linewidth]{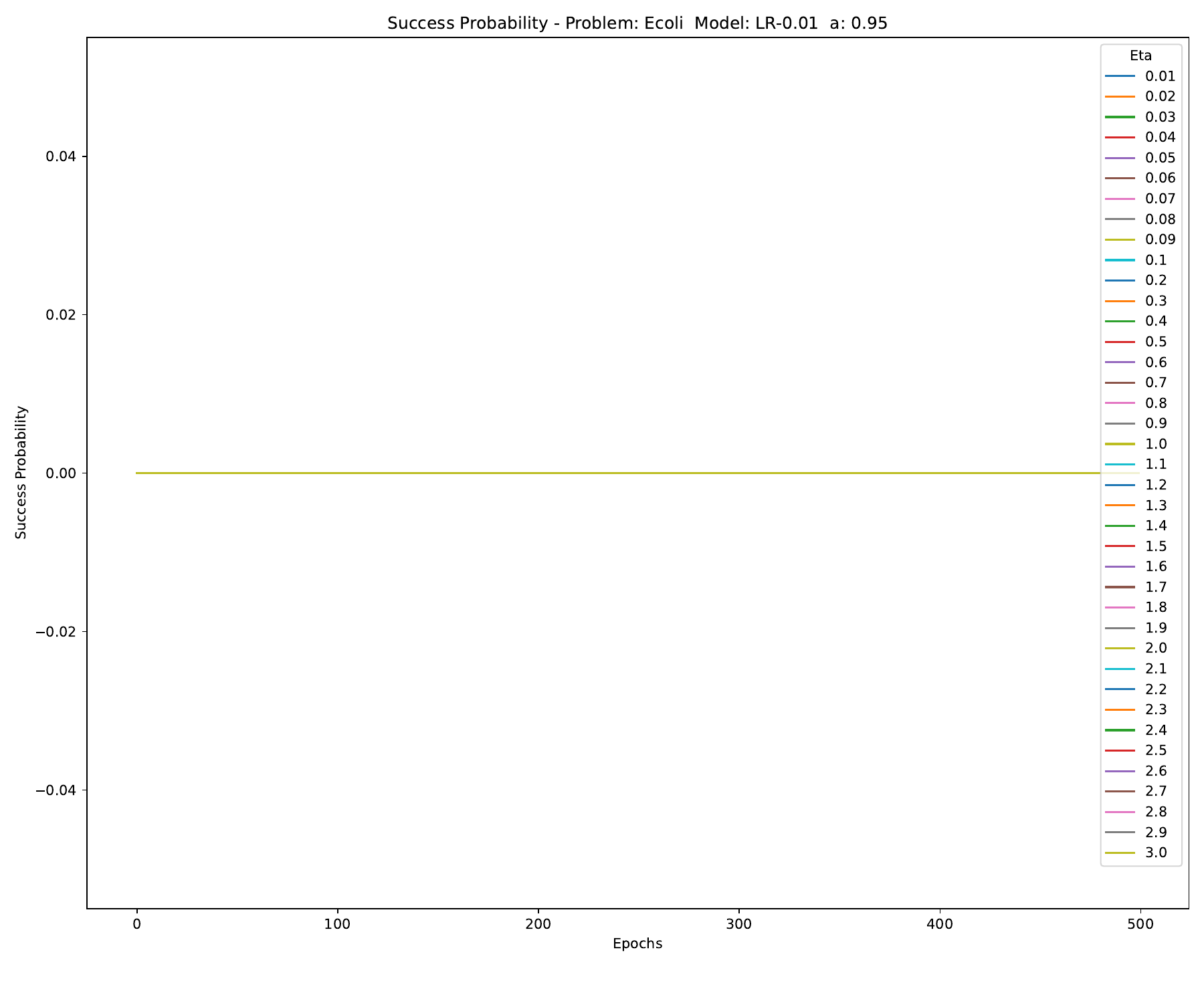} \\[-3mm]
  \end{tabular}
\caption{Success probability ($a=0.95$) for LR-0.01 model.}
  \label{fig:success_probability_LR_0.01_model_0.95}
\end{figure*}

\begin{figure*}[p]
  \centering
  \ContinuedFloat
  \begin{tabular}{c@{}c}
    Training Success Probability & Validation Success Probability \\
    \includegraphics[clip,trim=0 0 0 0,width=.52\linewidth]{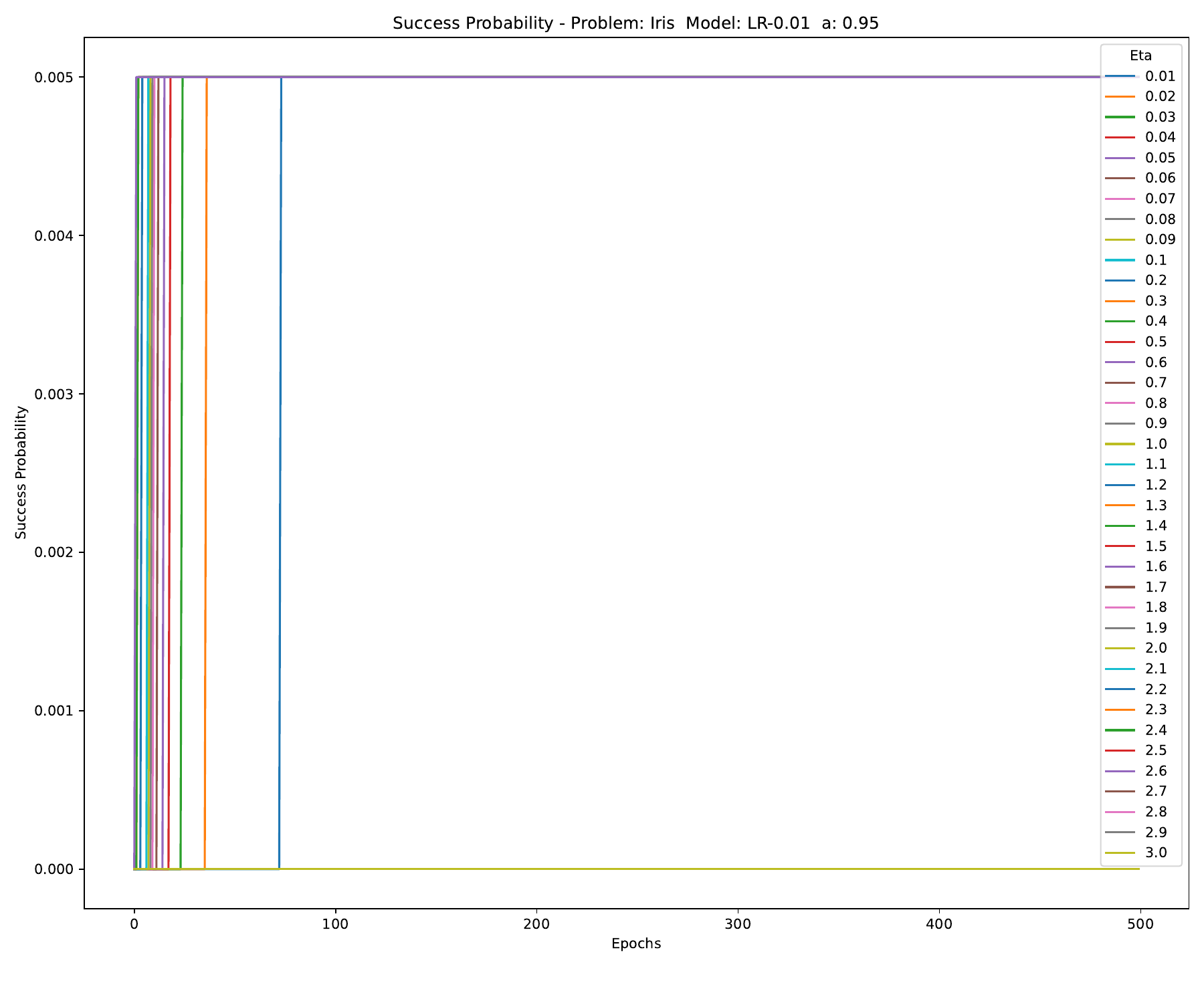} &
    \includegraphics[clip,trim=0 0 0 0,width=.52\linewidth]{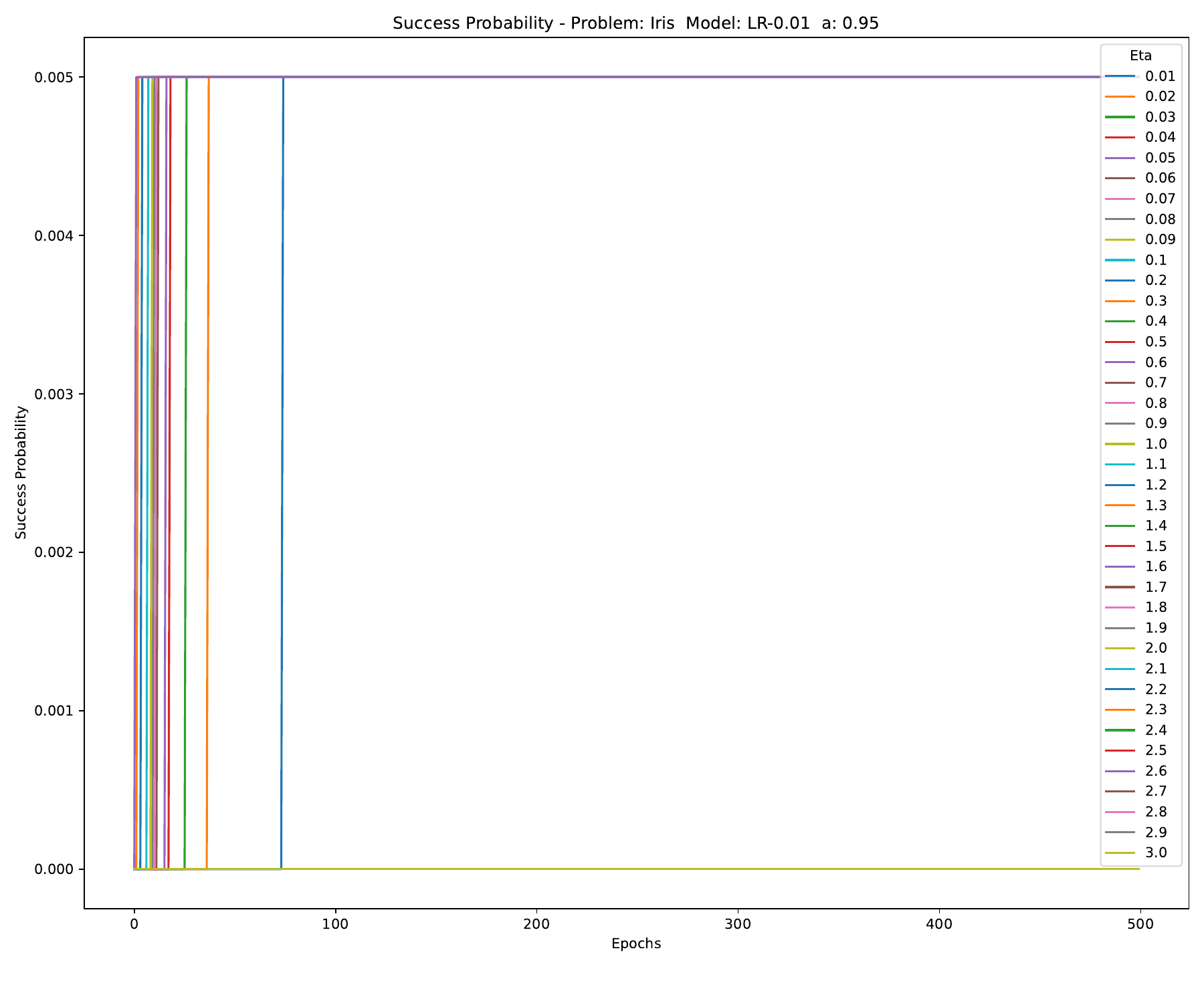} \\[-3mm]
    \includegraphics[clip,trim=0 0 0 0,width=.52\linewidth]{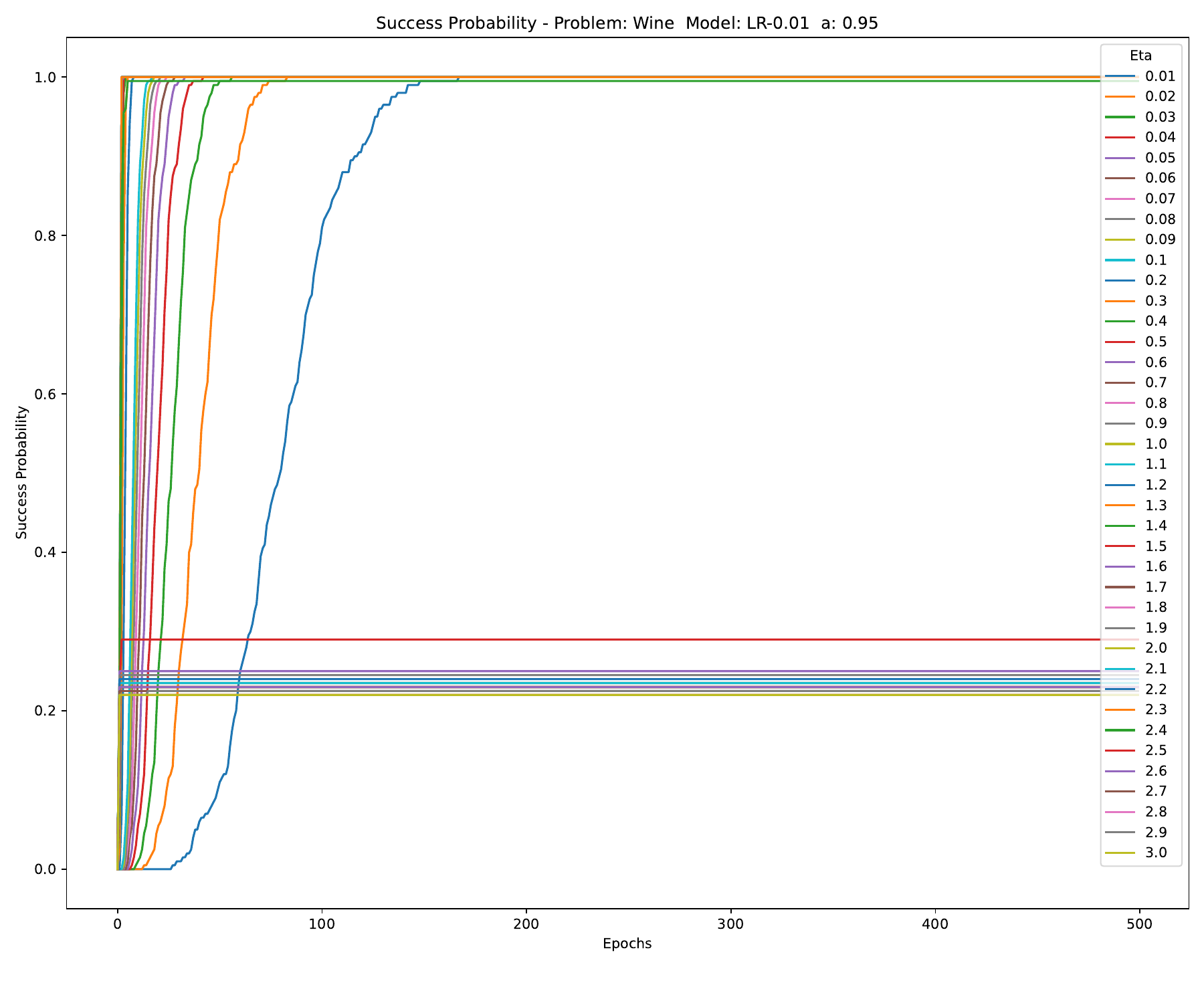} &
    \includegraphics[clip,trim=0 0 0 0,width=.52\linewidth]{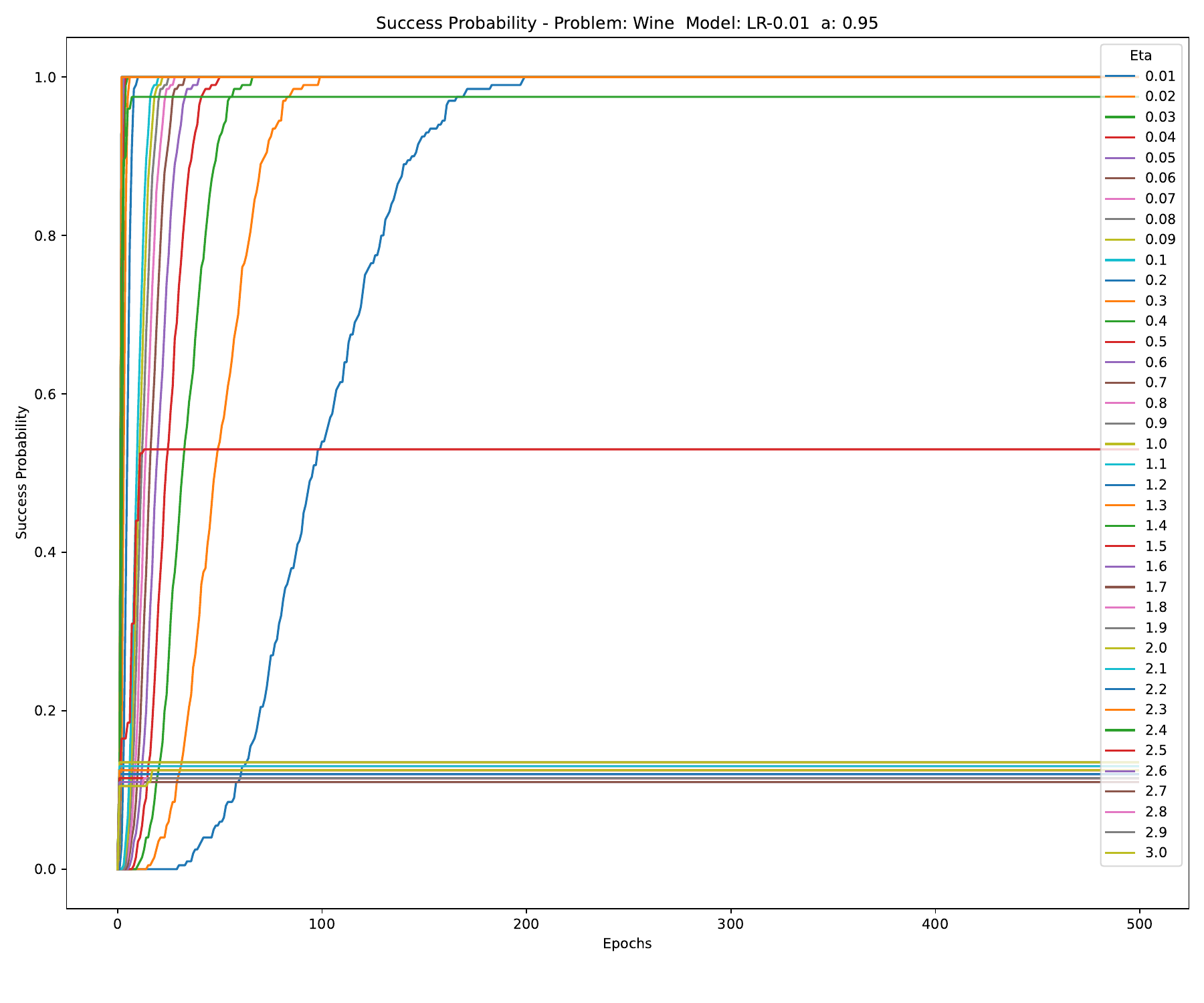} \\[-3mm]
  \end{tabular}
\caption{Success probability ($a=0.95$) for LR-0.01 model (continued).}
\end{figure*}

\begin{figure*}[p]
  \centering
  \begin{tabular}{c@{}c}
    Training Success Probability & Validation Success Probability \\
    \includegraphics[clip,trim=0 0 0 0,width=.52\linewidth]{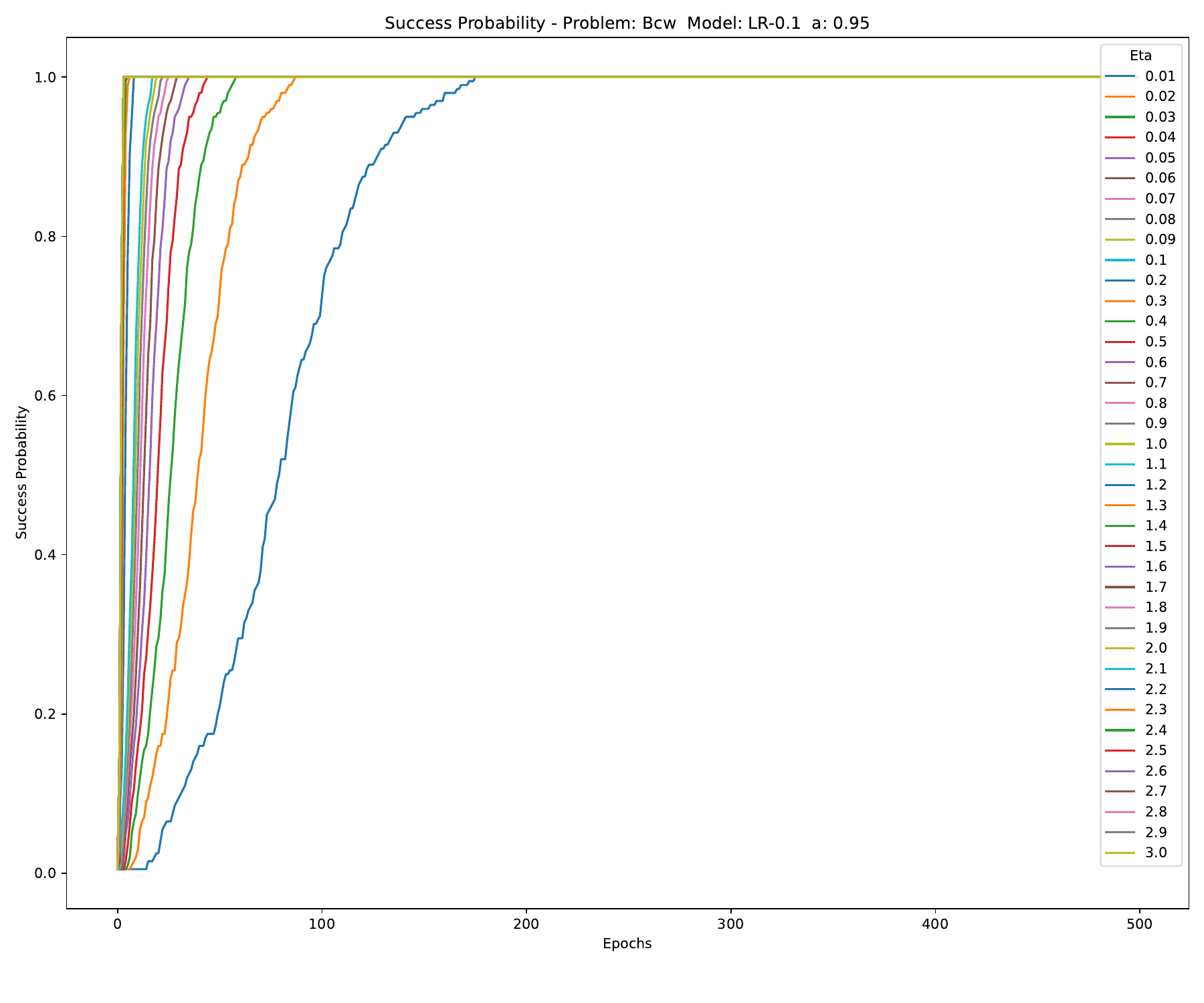} &
    \includegraphics[clip,trim=0 0 0 0,width=.52\linewidth]{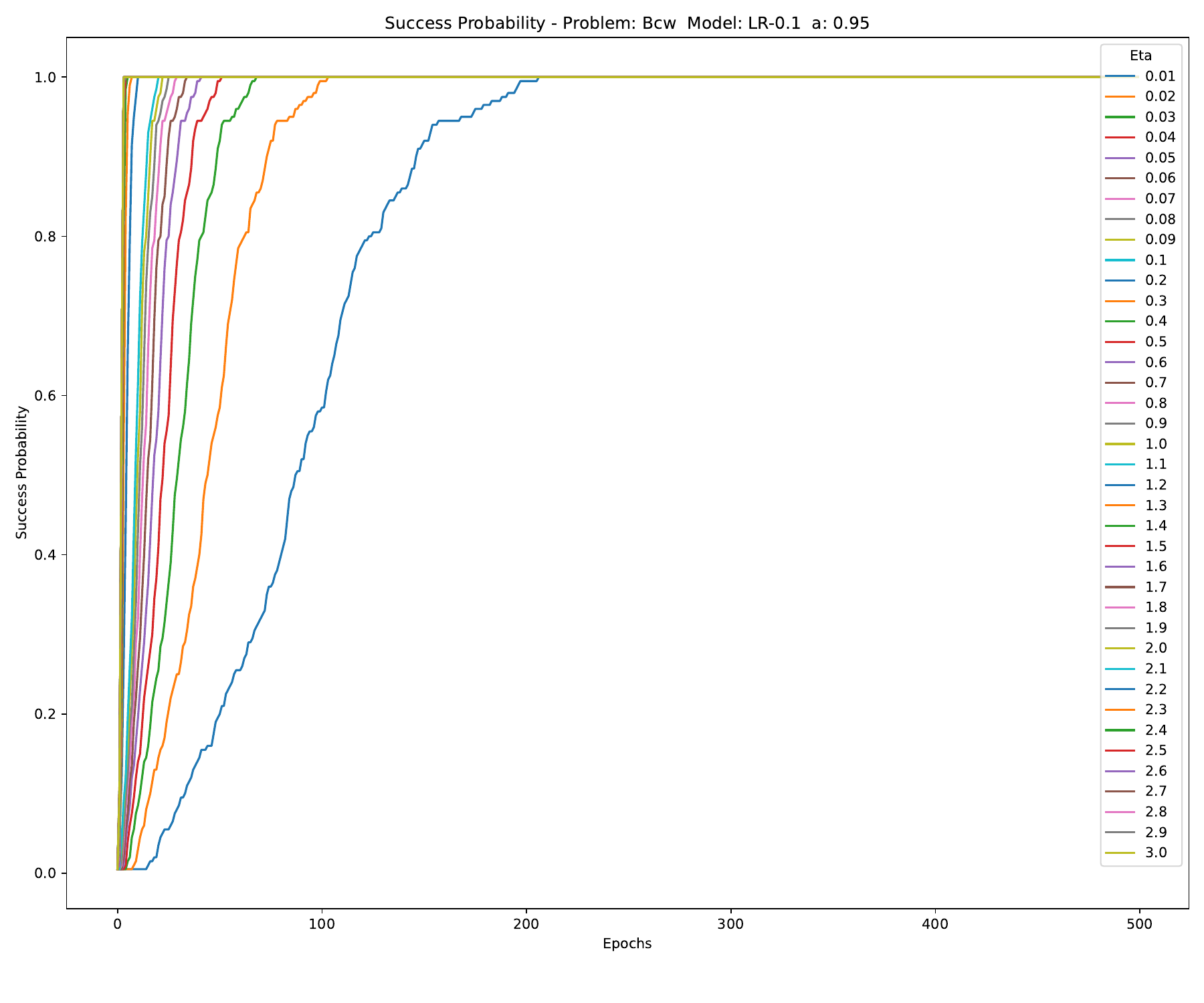} \\[-3mm]
    \includegraphics[clip,trim=0 0 0 0,width=.52\linewidth]{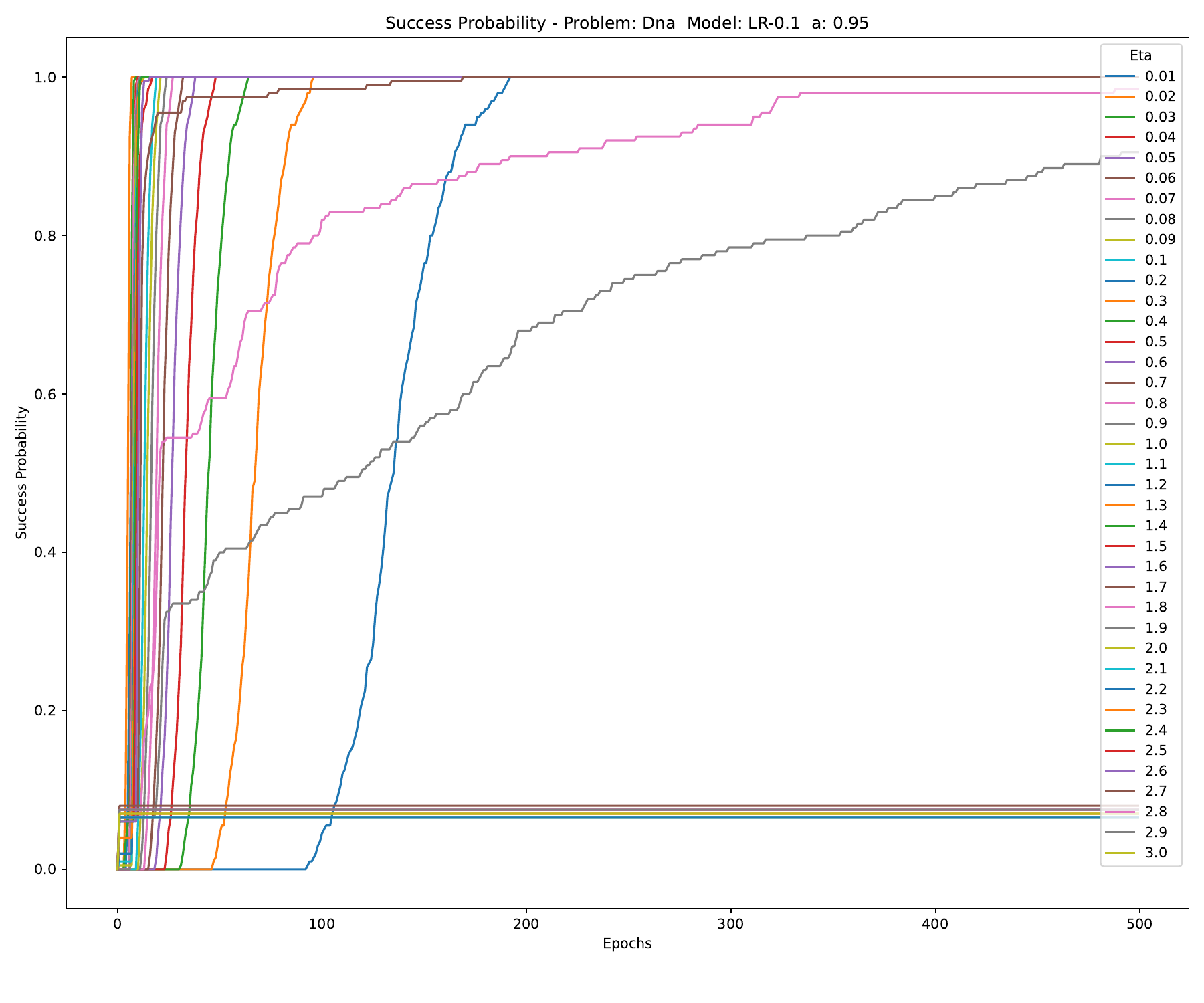} &
    \includegraphics[clip,trim=0 0 0 0,width=.52\linewidth]{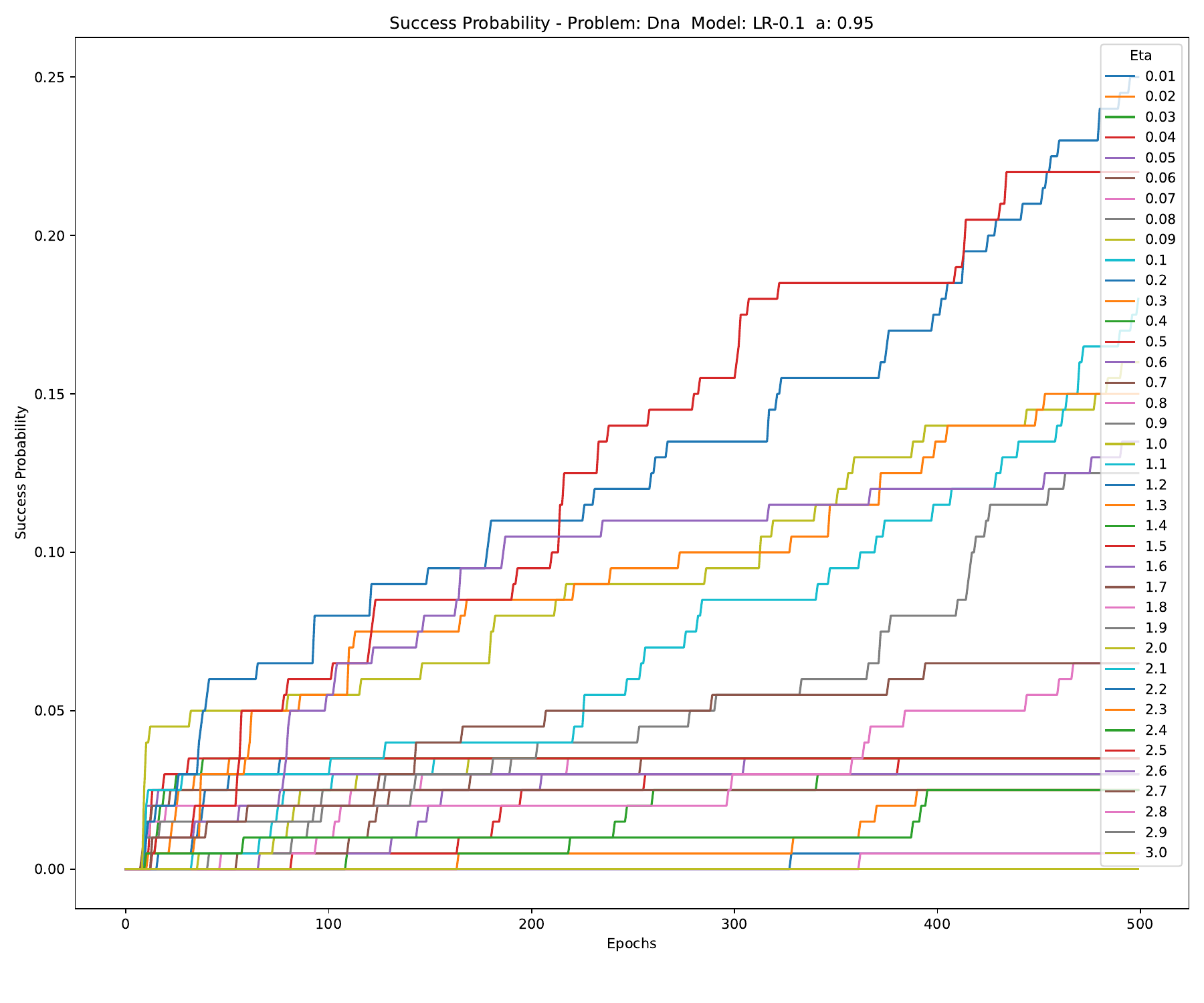} \\[-3mm]
    \includegraphics[clip,trim=0 0 0 0,width=.52\linewidth]{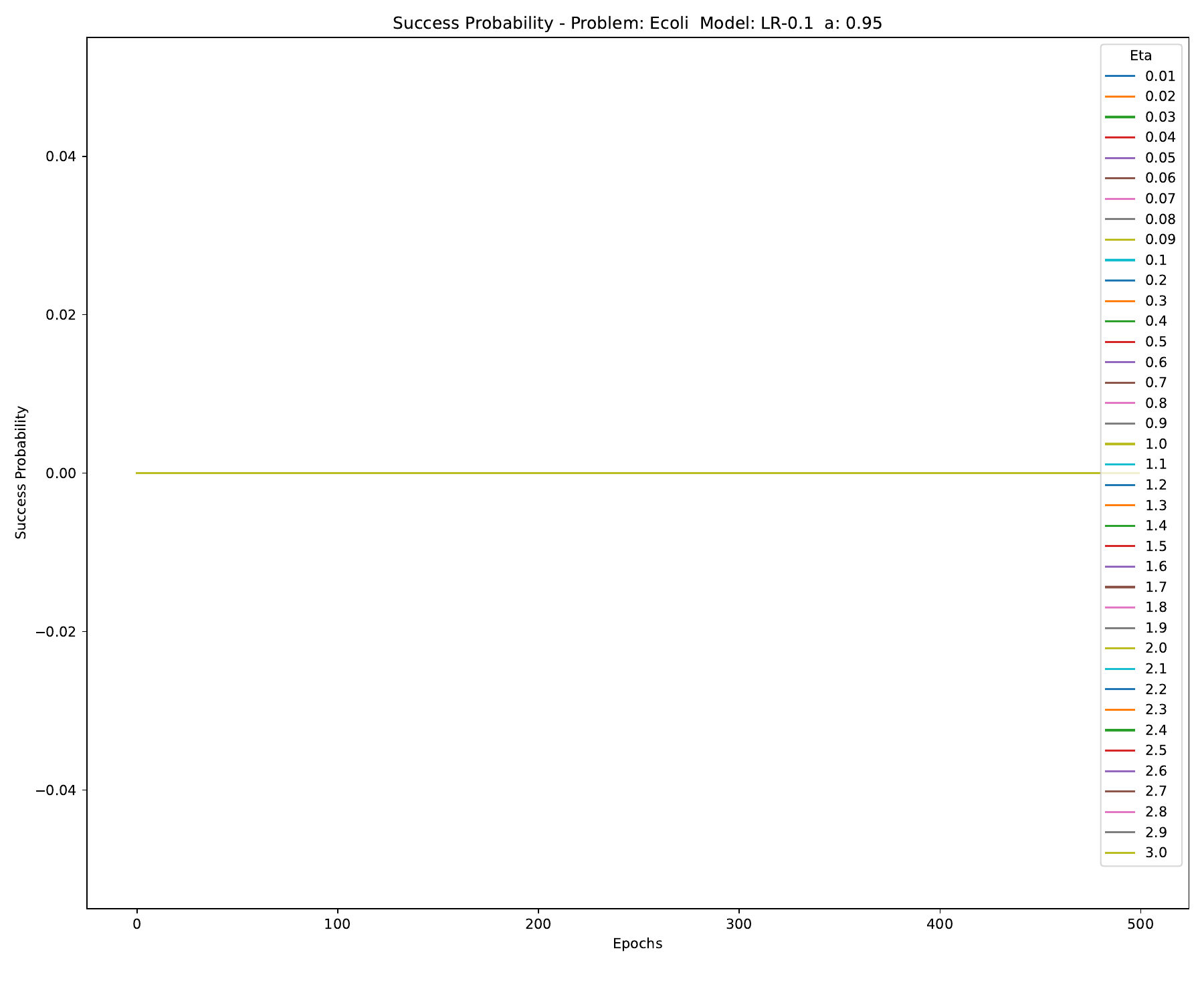} &
    \includegraphics[clip,trim=0 0 0 0,width=.52\linewidth]{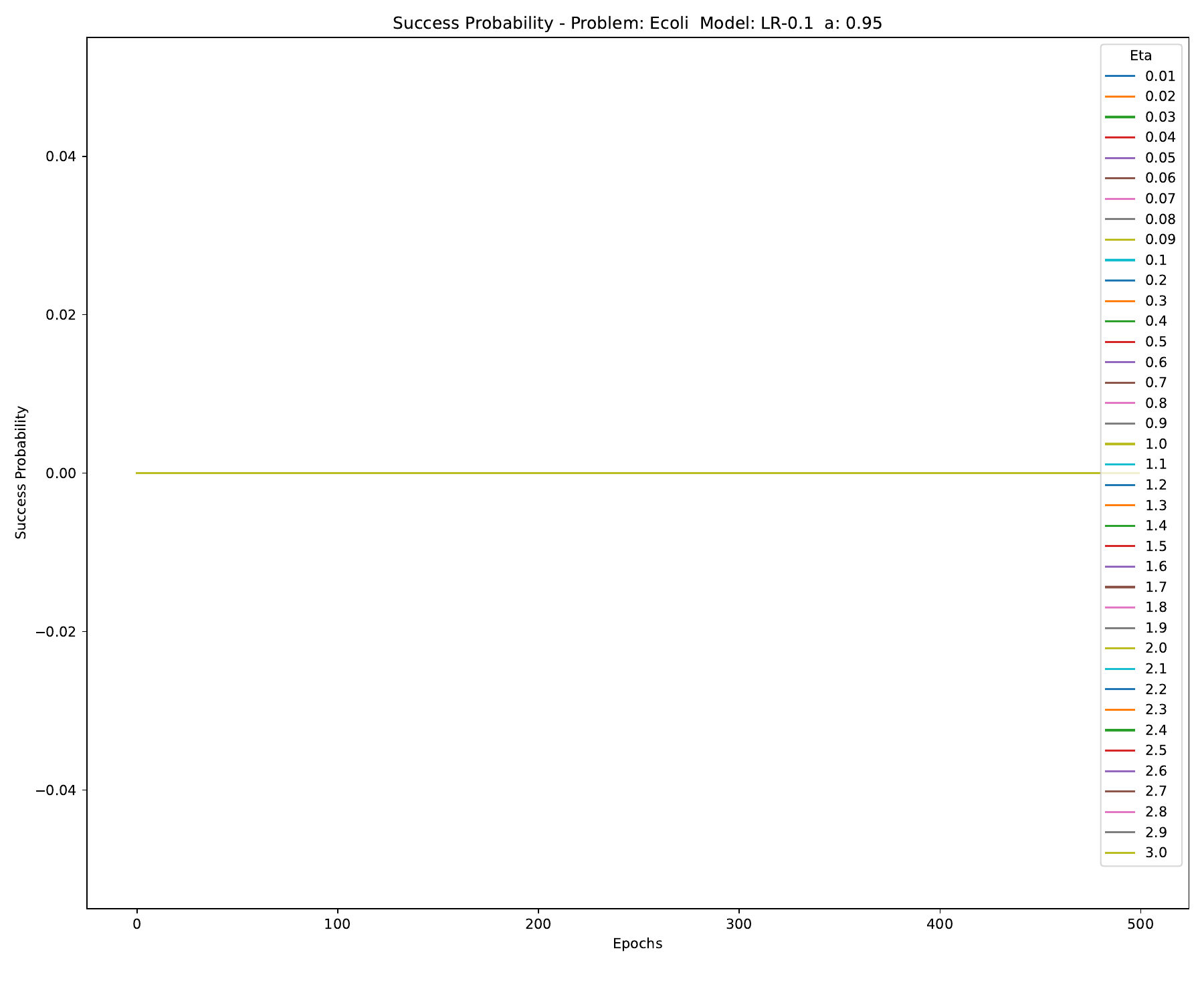} \\[-3mm]
  \end{tabular}
\caption{Success probability ($a=0.95$) for LR-0.1 model.}
  \label{fig:success_probability_LR_0.1_model_0.95}
\end{figure*}

\begin{figure*}[p]
  \centering
  \ContinuedFloat
  \begin{tabular}{c@{}c}
    Training Success Probability & Validation Success Probability \\
    \includegraphics[clip,trim=0 0 0 0,width=.52\linewidth]{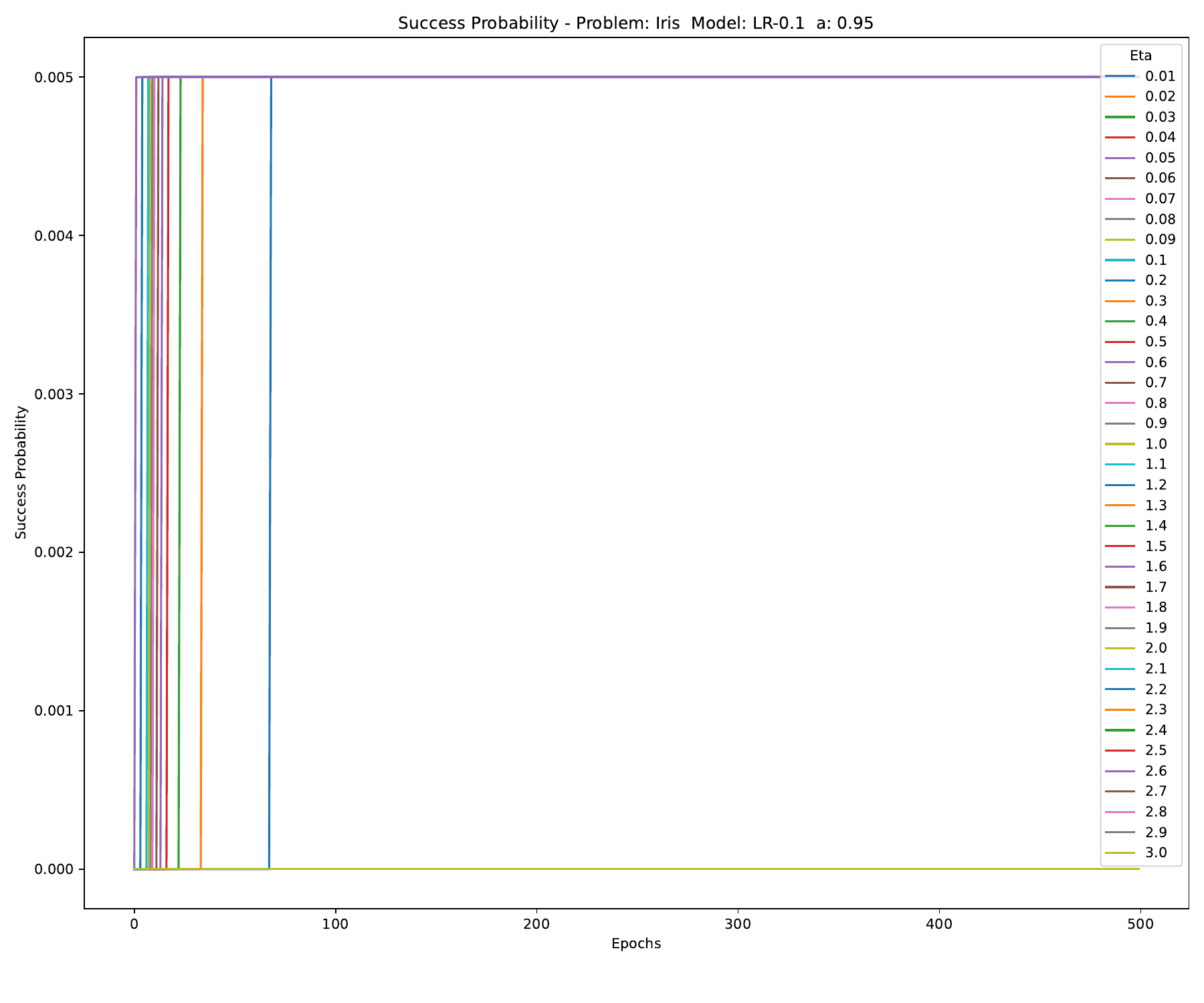} &
    \includegraphics[clip,trim=0 0 0 0,width=.52\linewidth]{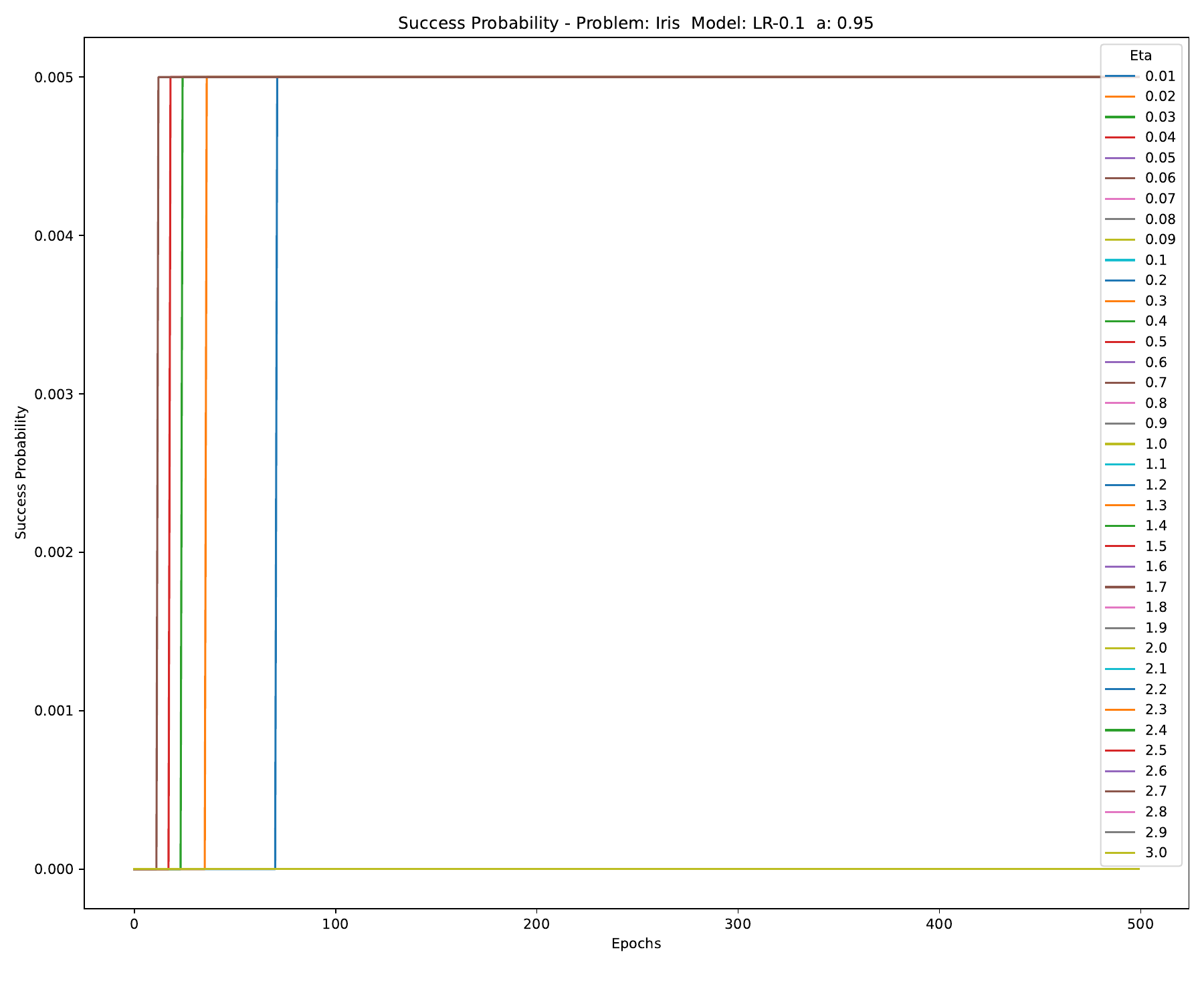} \\[-3mm]
    \includegraphics[clip,trim=0 0 0 0,width=.52\linewidth]{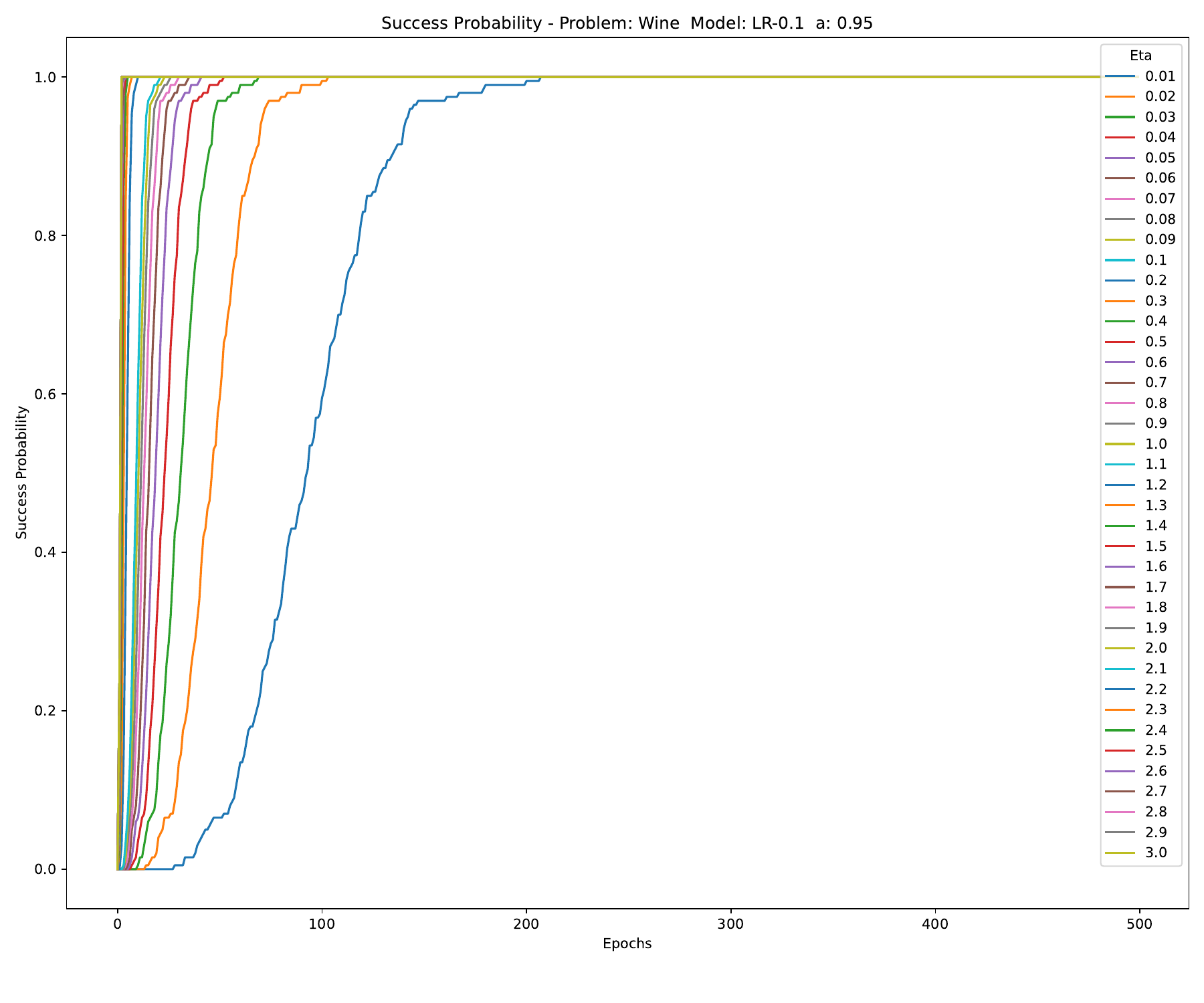} &
    \includegraphics[clip,trim=0 0 0 0,width=.52\linewidth]{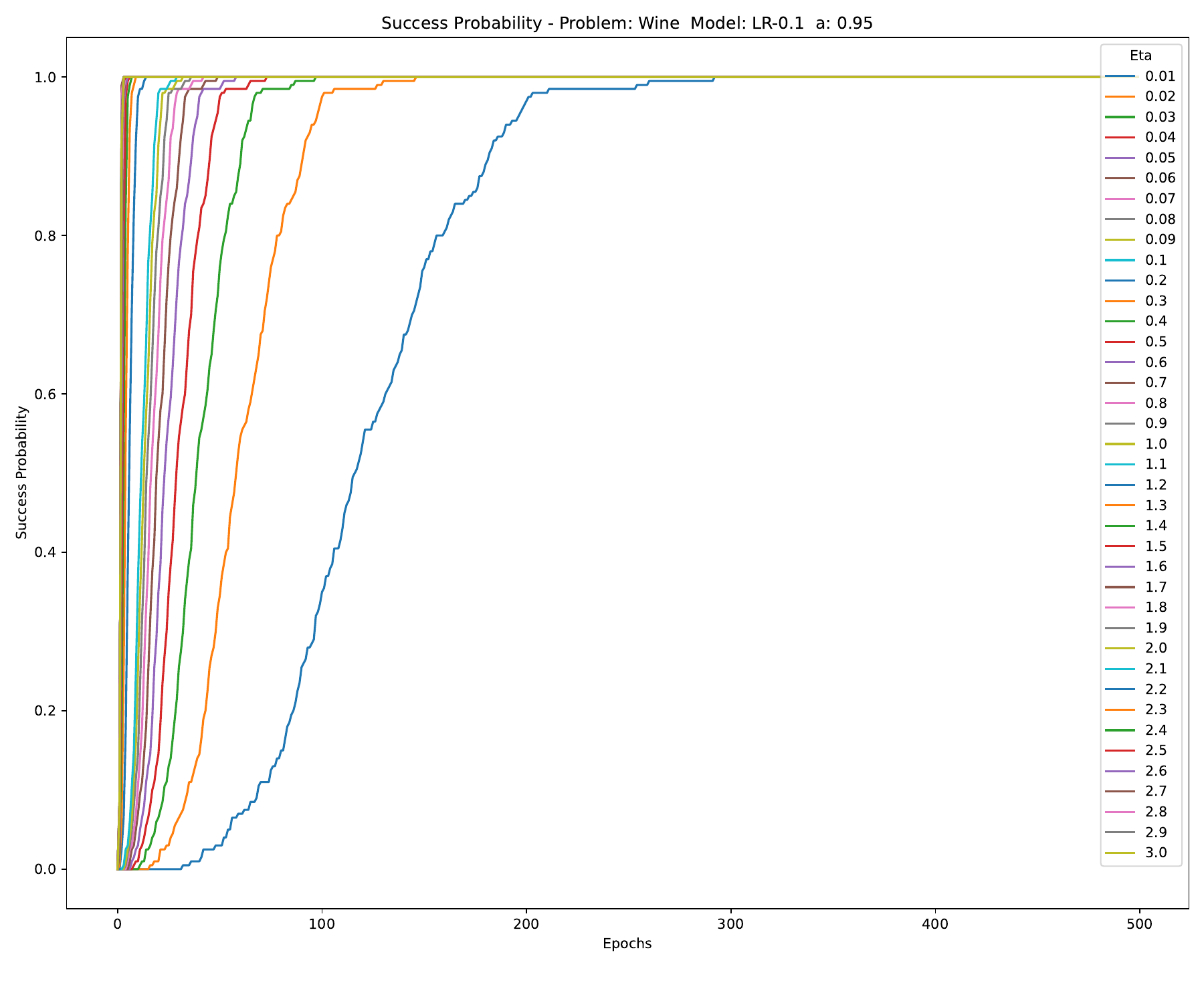} \\[-3mm]
  \end{tabular}
\caption{Success probability ($a=0.95$) for LR-0.1 model (continued).}
\end{figure*}

\begin{figure*}[p]
  \centering
  \begin{tabular}{c@{}c}
    Training Success Probability & Validation Success Probability \\
    \includegraphics[clip,trim=0 0 0 0,width=.52\linewidth]{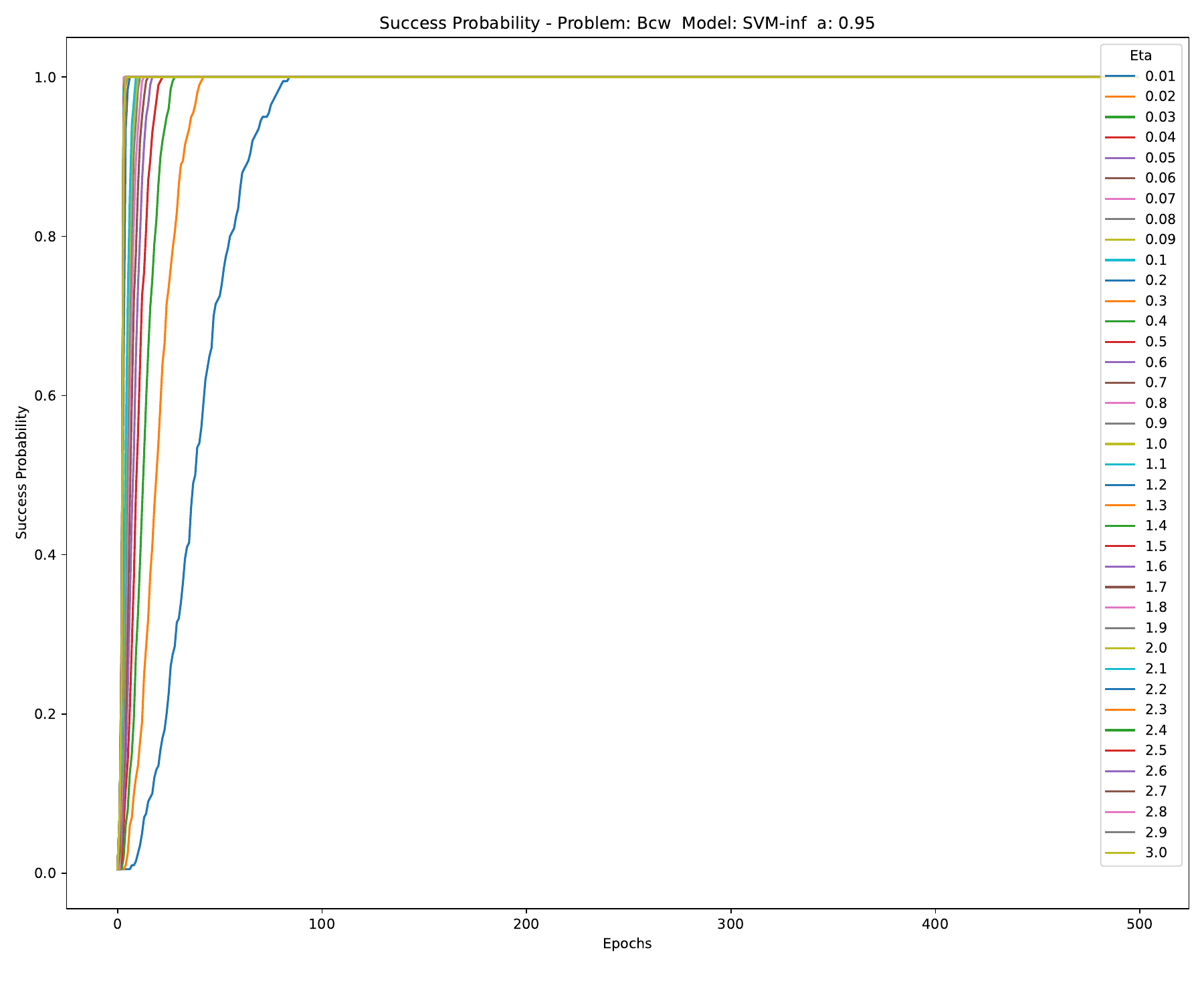} &
    \includegraphics[clip,trim=0 0 0 0,width=.52\linewidth]{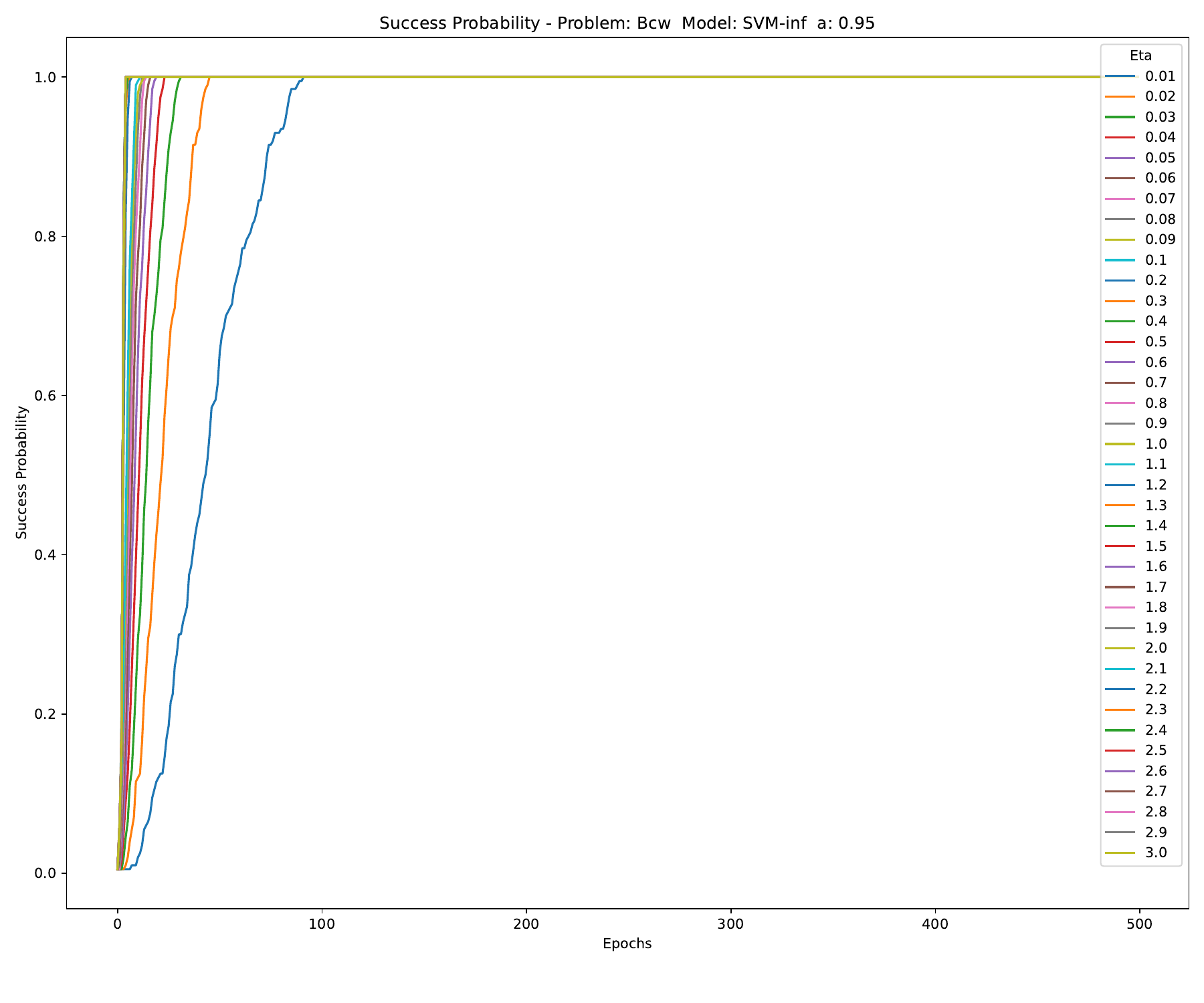} \\[-3mm]
    \includegraphics[clip,trim=0 0 0 0,width=.52\linewidth]{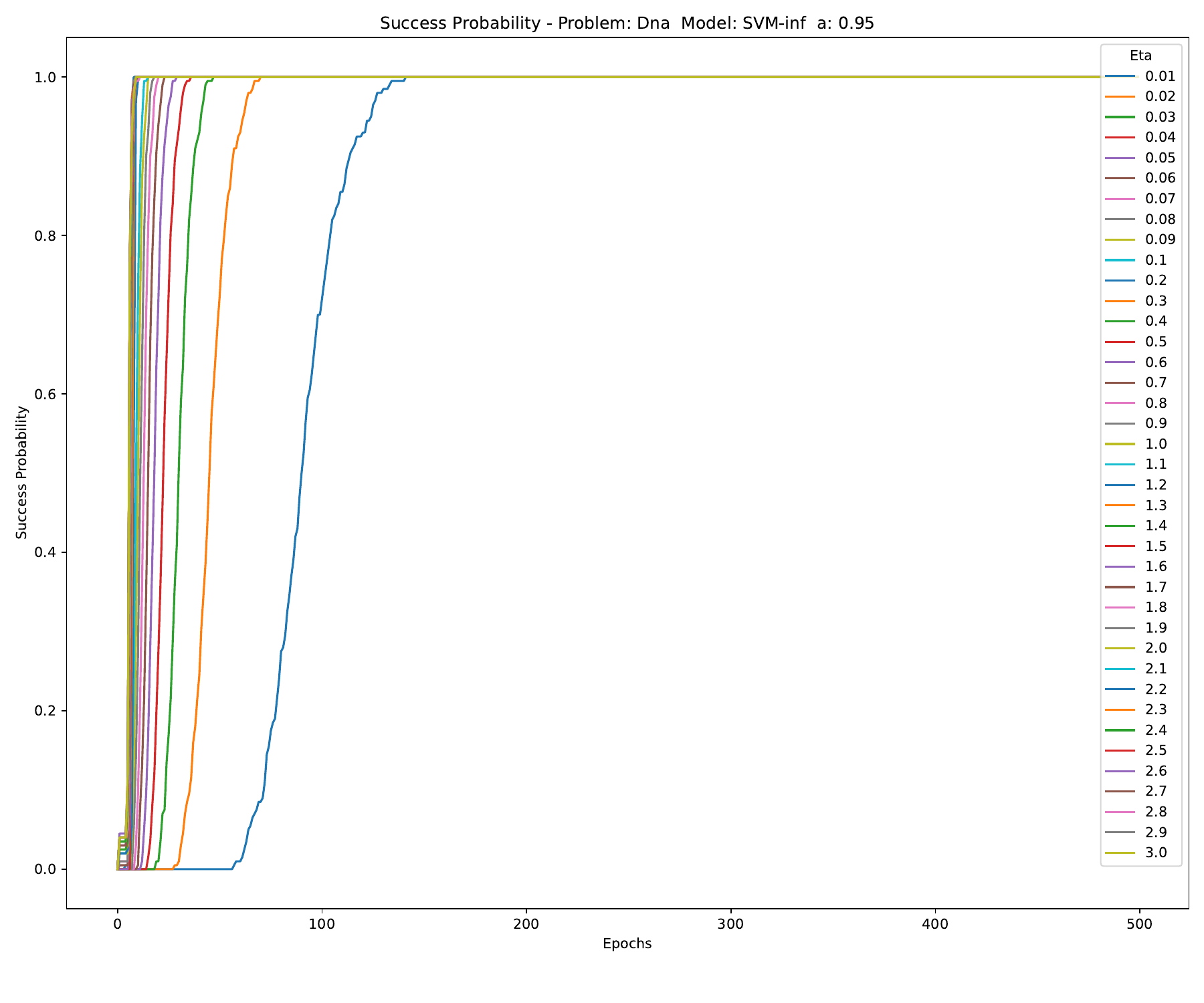} &
    \includegraphics[clip,trim=0 0 0 0,width=.52\linewidth]{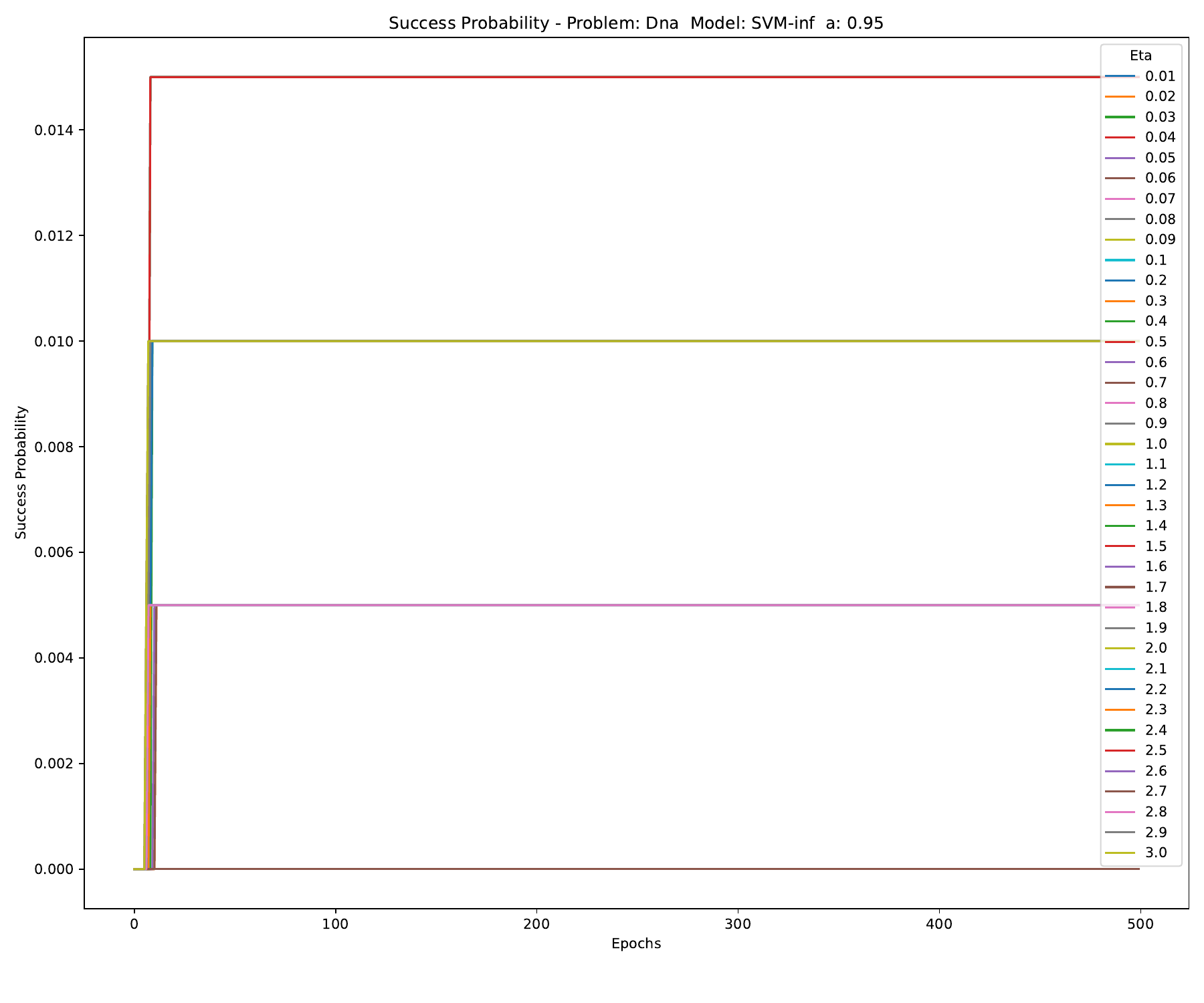} \\[-3mm]
    \includegraphics[clip,trim=0 0 0 0,width=.52\linewidth]{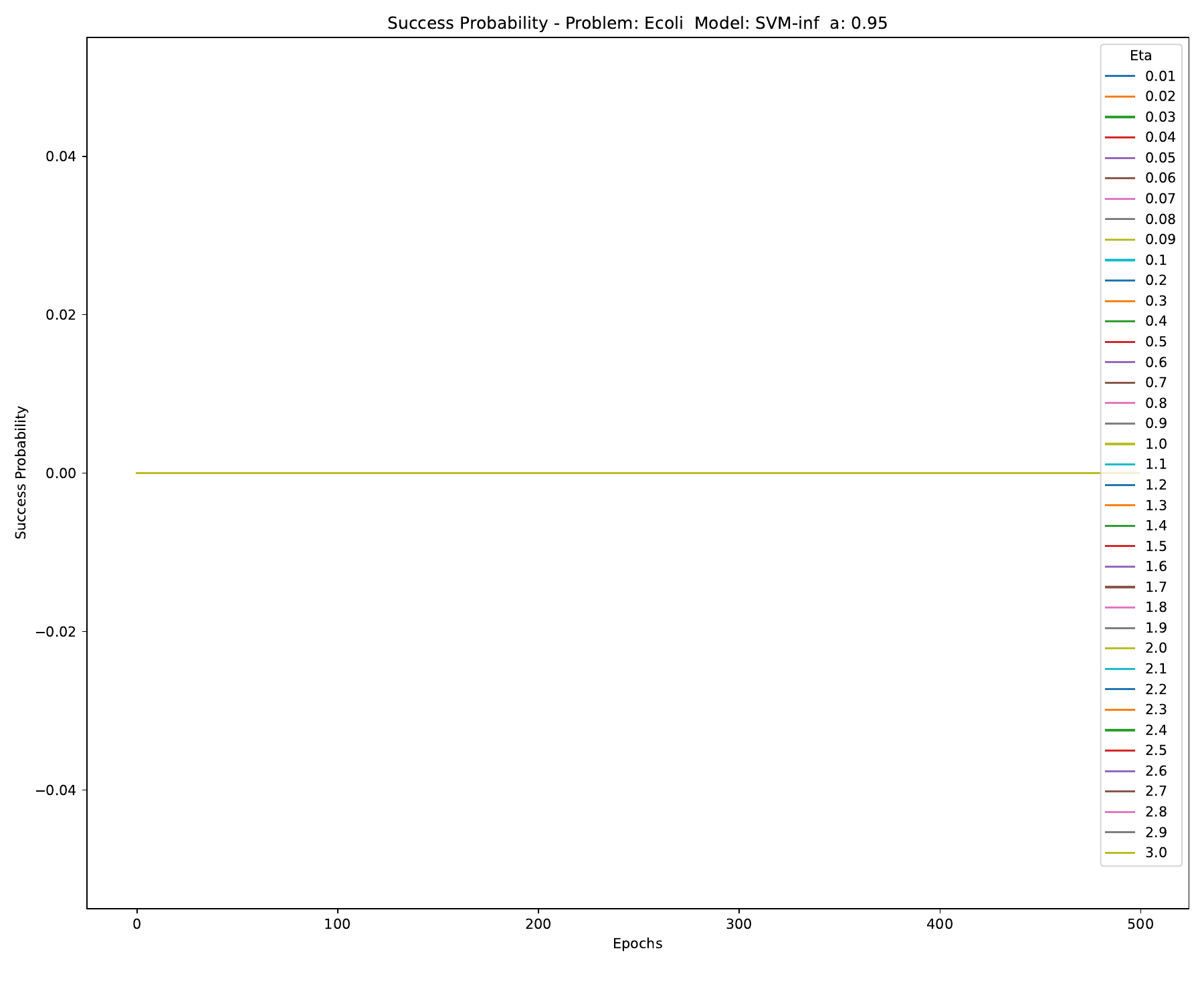} &
    \includegraphics[clip,trim=0 0 0 0,width=.52\linewidth]{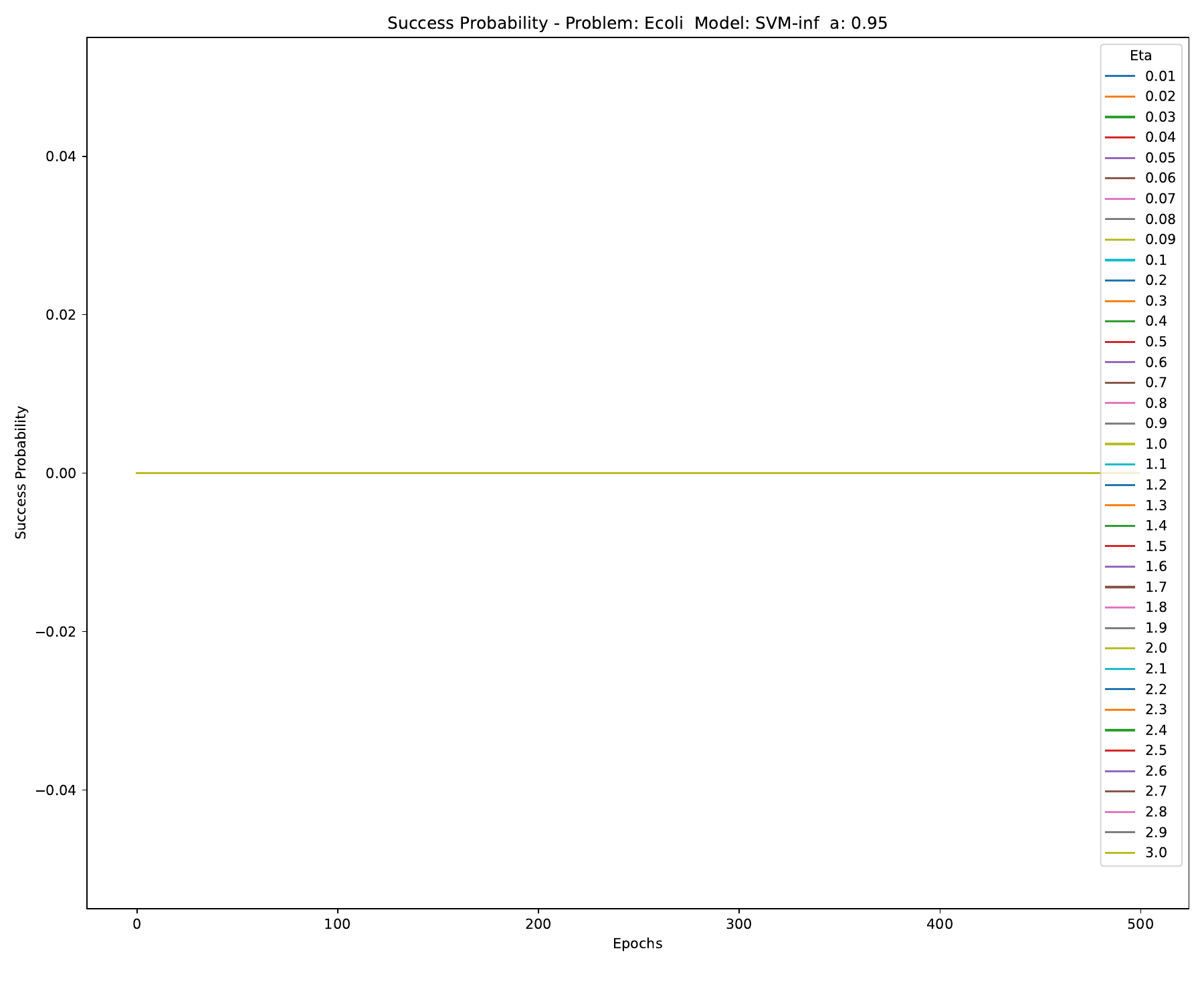} \\[-3mm]
  \end{tabular}
\caption{Success probability ($a=0.95$) for SVM-Inf model.}
  \label{fig:success_probability_SVM_Inf_model_0.95}
\end{figure*}

\begin{figure*}[p]
  \centering
  \ContinuedFloat
  \begin{tabular}{c@{}c}
    Training Success Probability & Validation Success Probability \\
    \includegraphics[clip,trim=0 0 0 0,width=.52\linewidth]{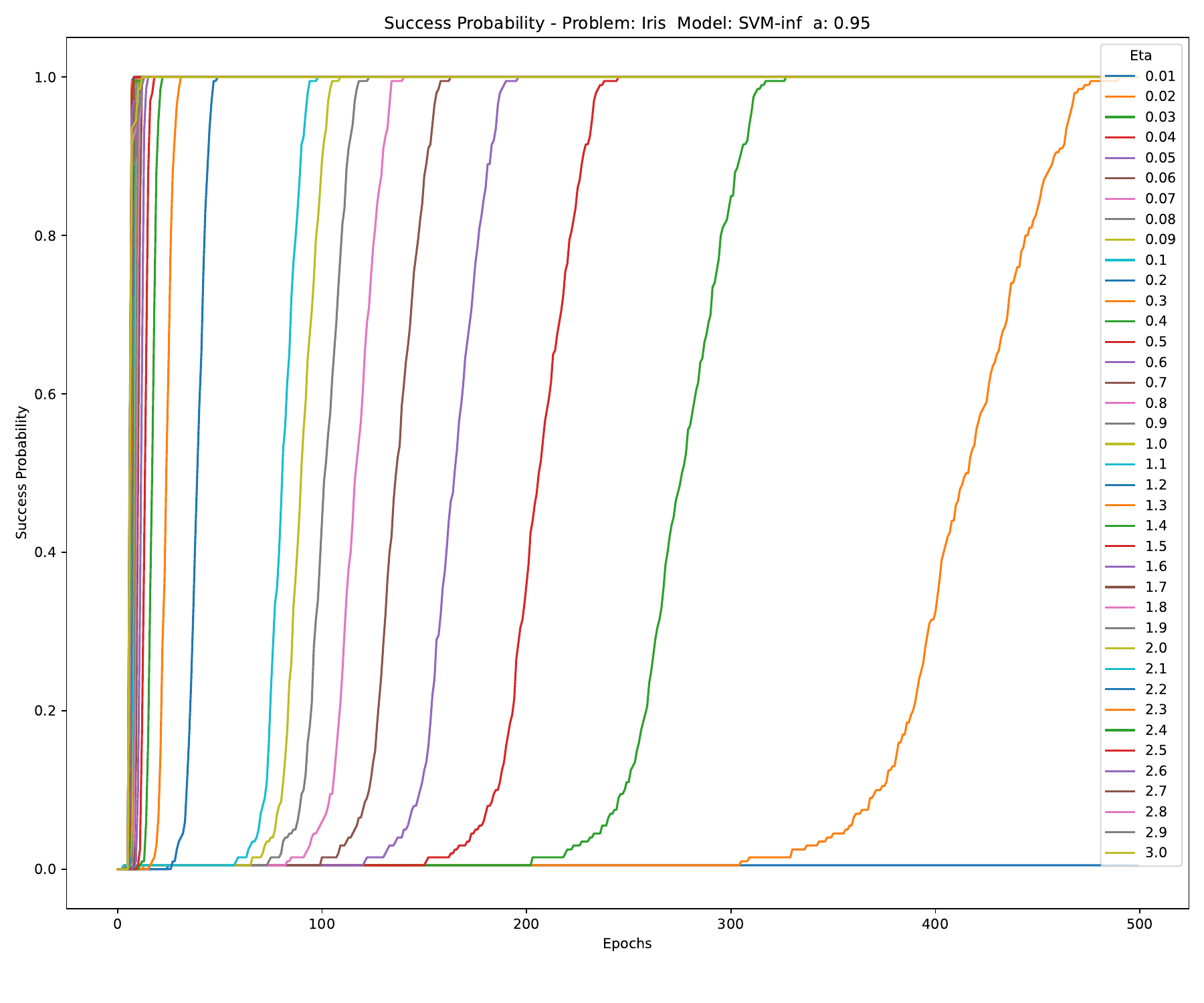} &
    \includegraphics[clip,trim=0 0 0 0,width=.52\linewidth]{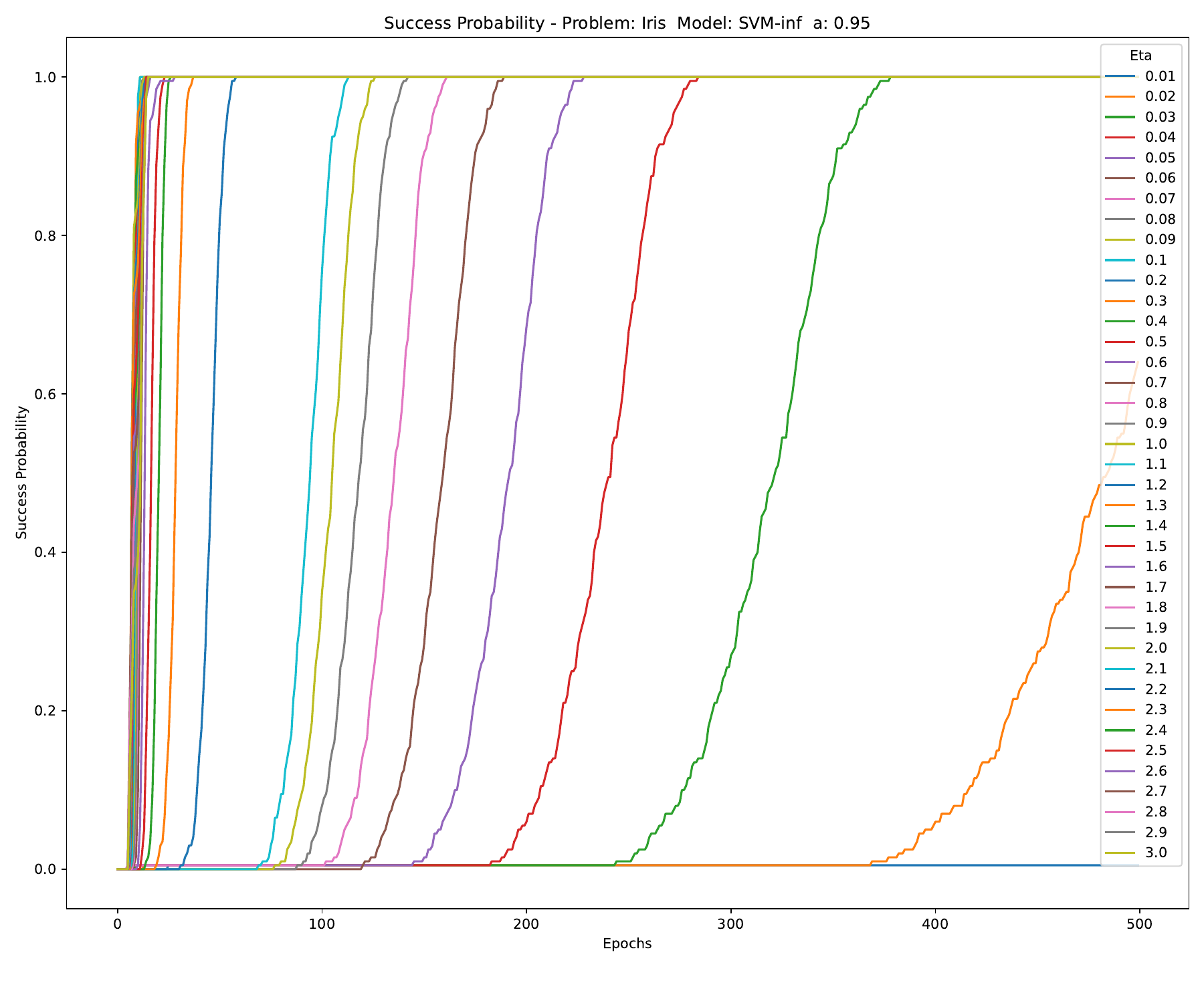} \\[-3mm]
    \includegraphics[clip,trim=0 0 0 0,width=.52\linewidth]{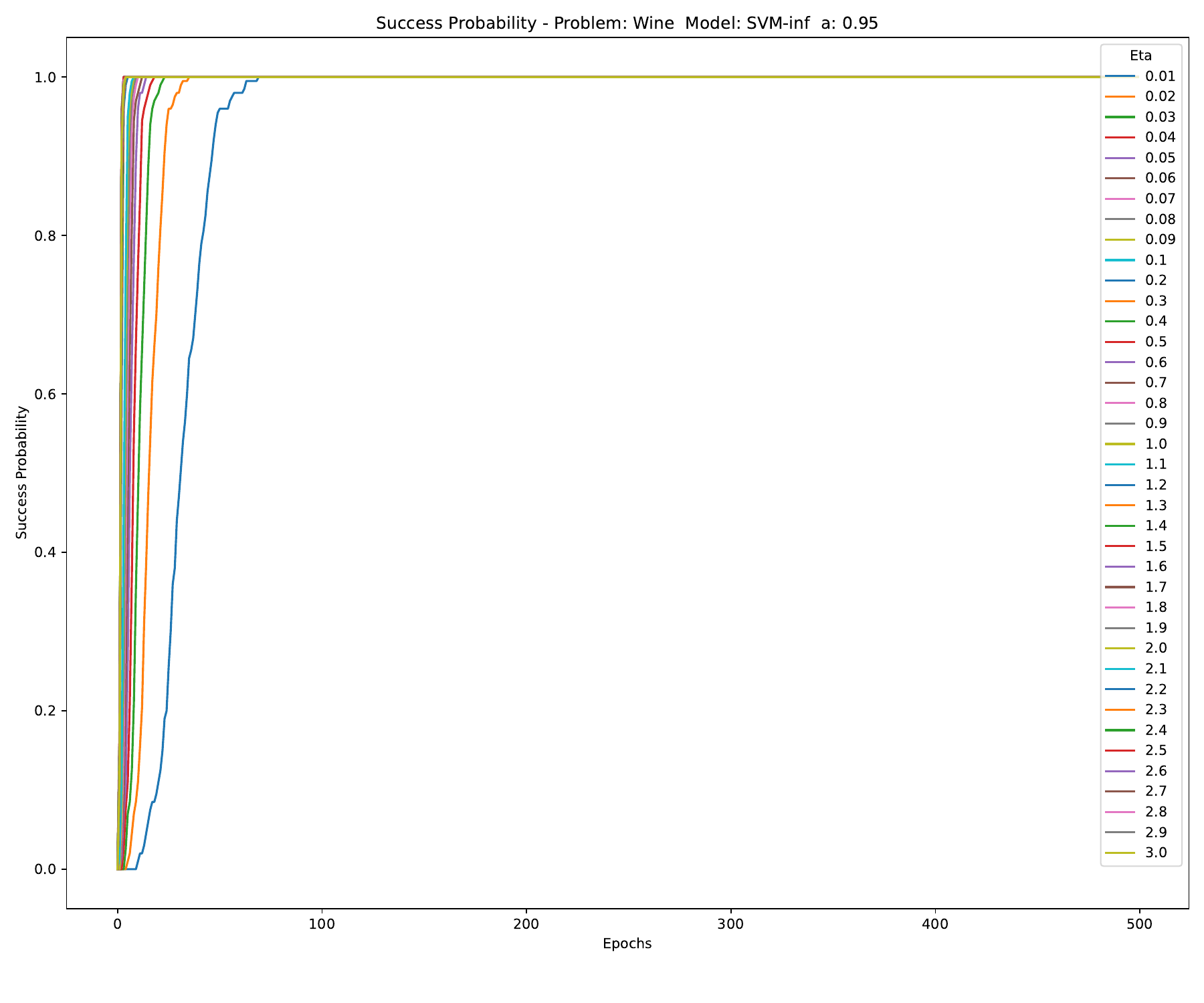} &
    \includegraphics[clip,trim=0 0 0 0,width=.52\linewidth]{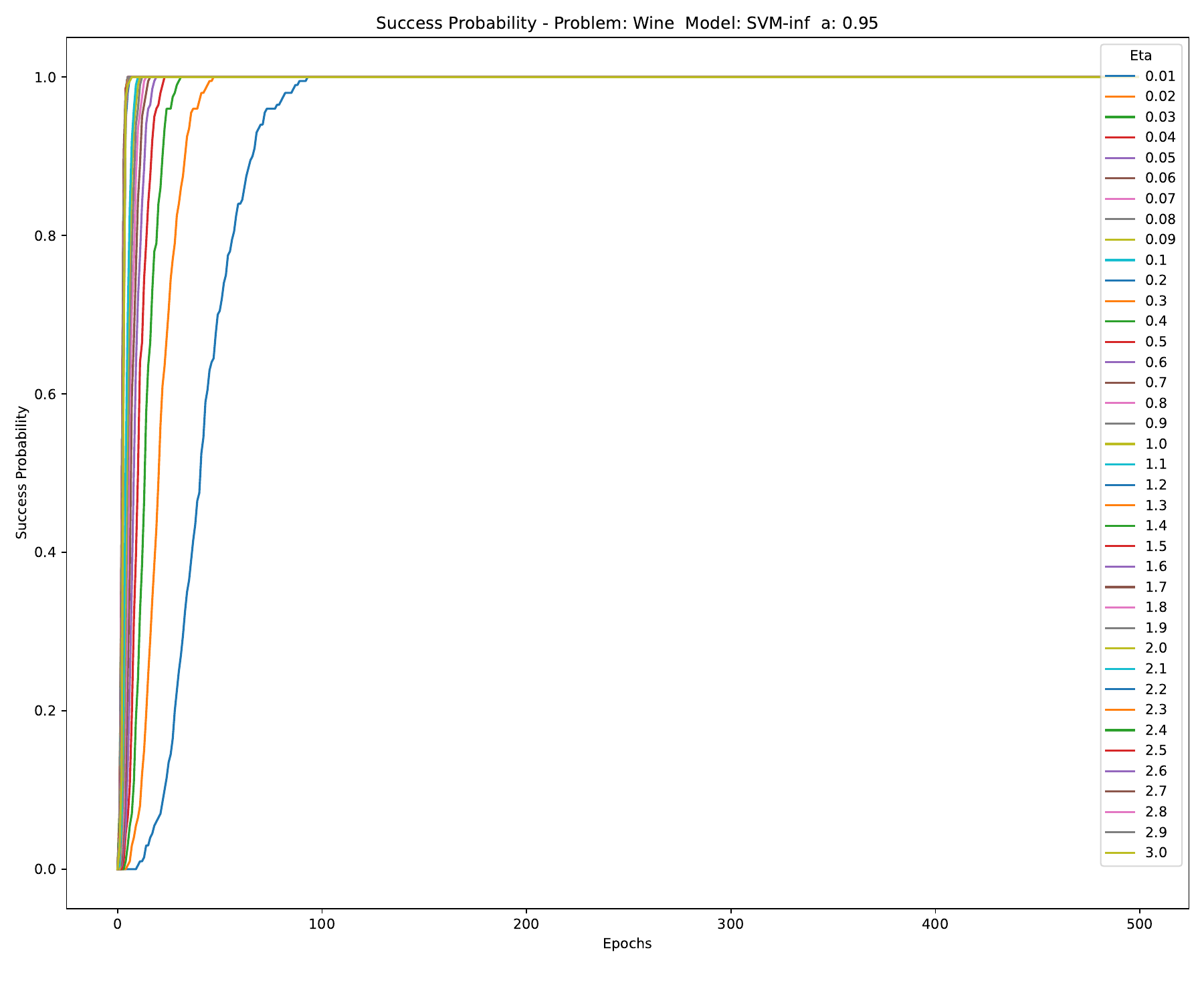} \\[-3mm]
  \end{tabular}
\caption{Success probability ($a=0.95$) for SVM-Inf model (continued).}
\end{figure*}

\begin{figure*}[p]
  \centering
  \begin{tabular}{c@{}c}
    Training Success Probability & Validation Success Probability \\
    \includegraphics[clip,trim=0 0 0 0,width=.52\linewidth]{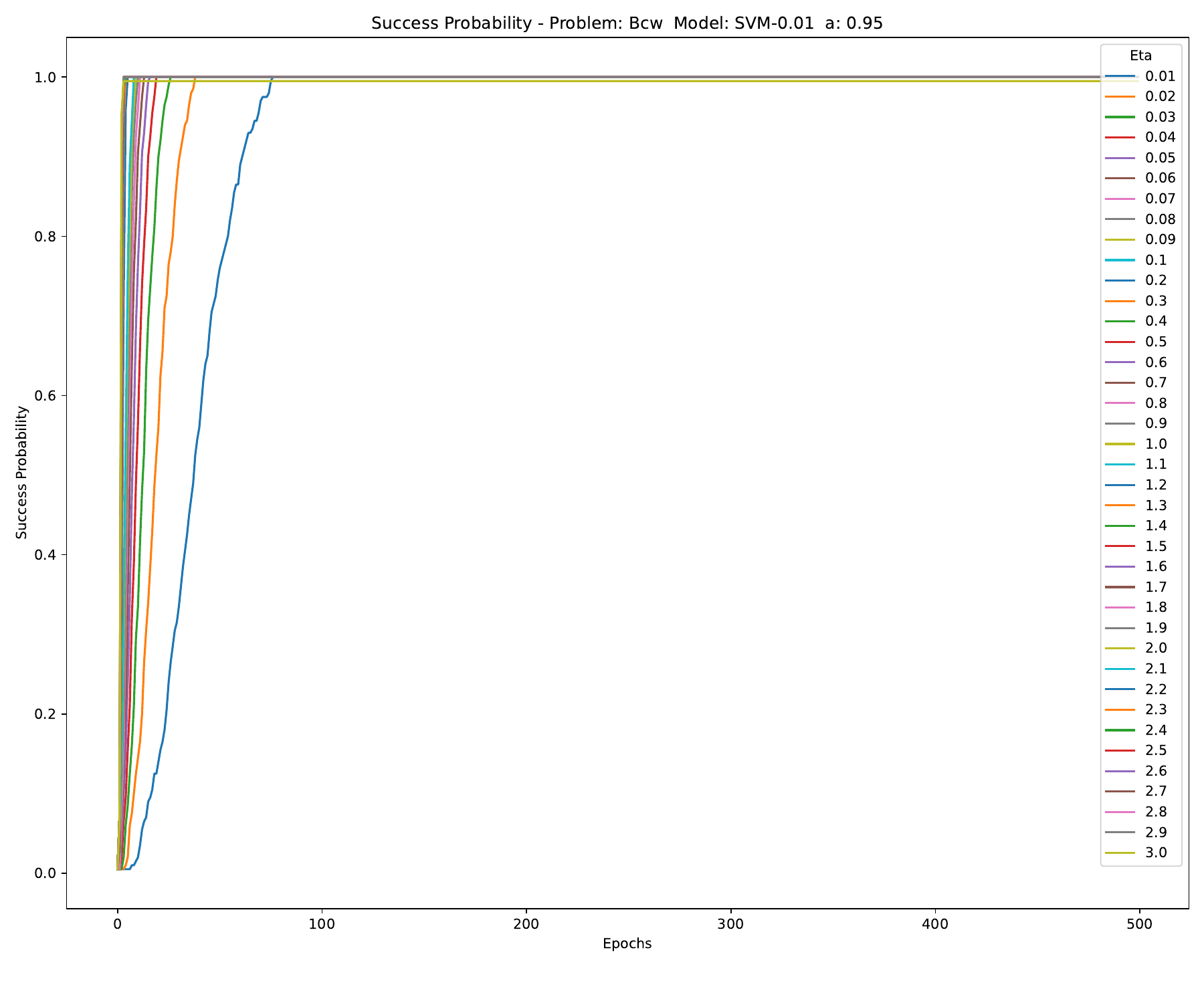} &
    \includegraphics[clip,trim=0 0 0 0,width=.52\linewidth]{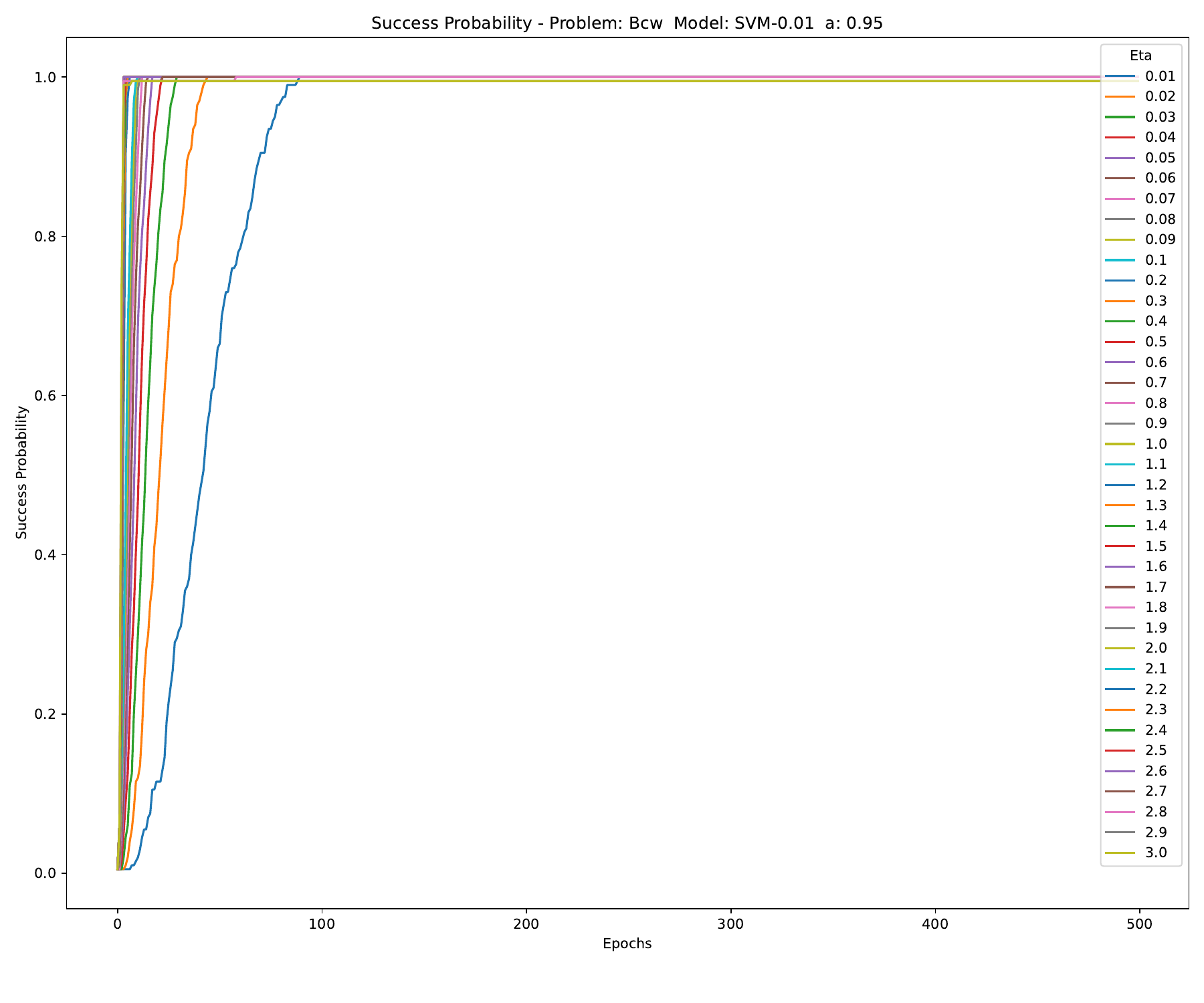} \\[-3mm]
    \includegraphics[clip,trim=0 0 0 0,width=.52\linewidth]{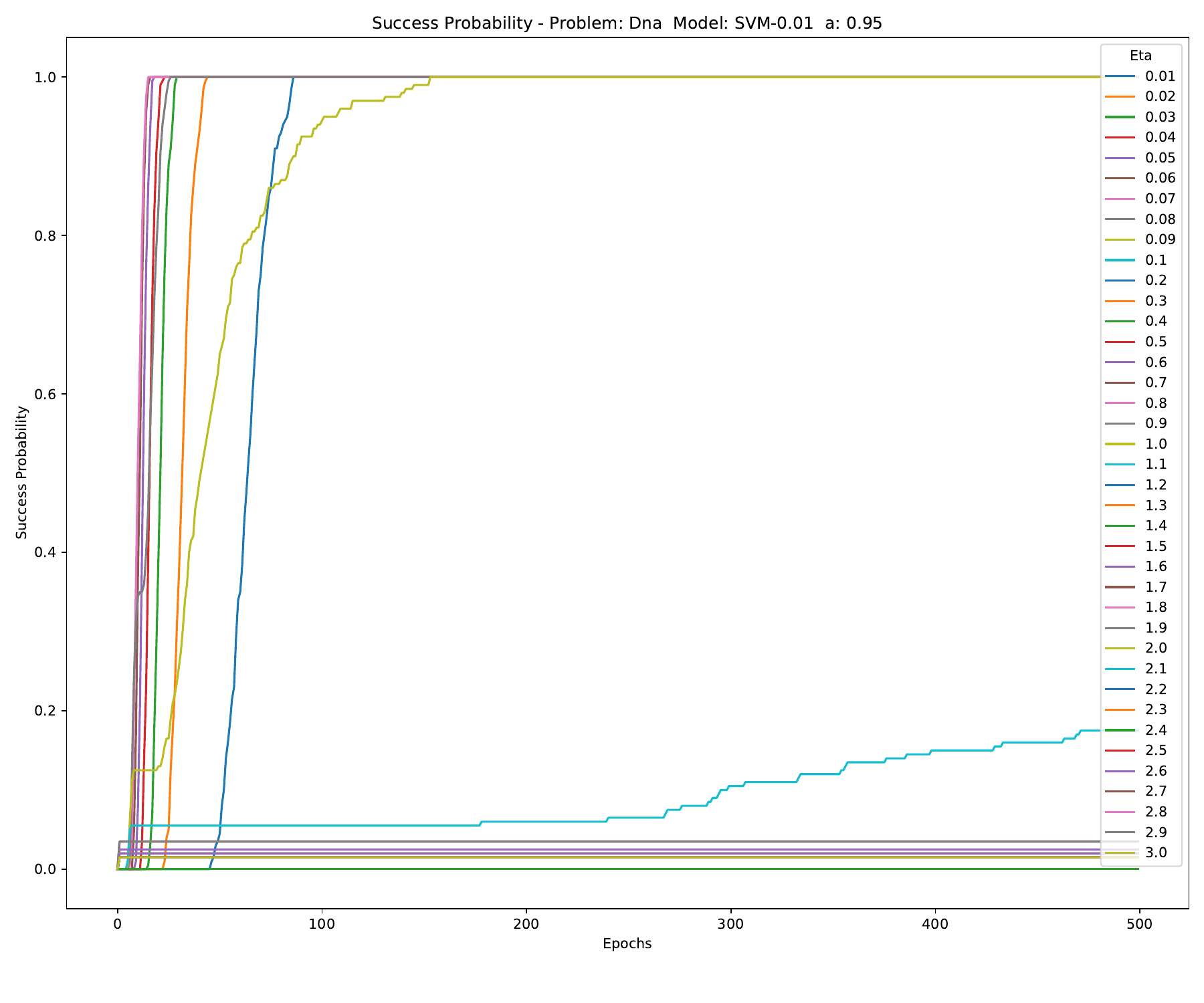} &
    \includegraphics[clip,trim=0 0 0 0,width=.52\linewidth]{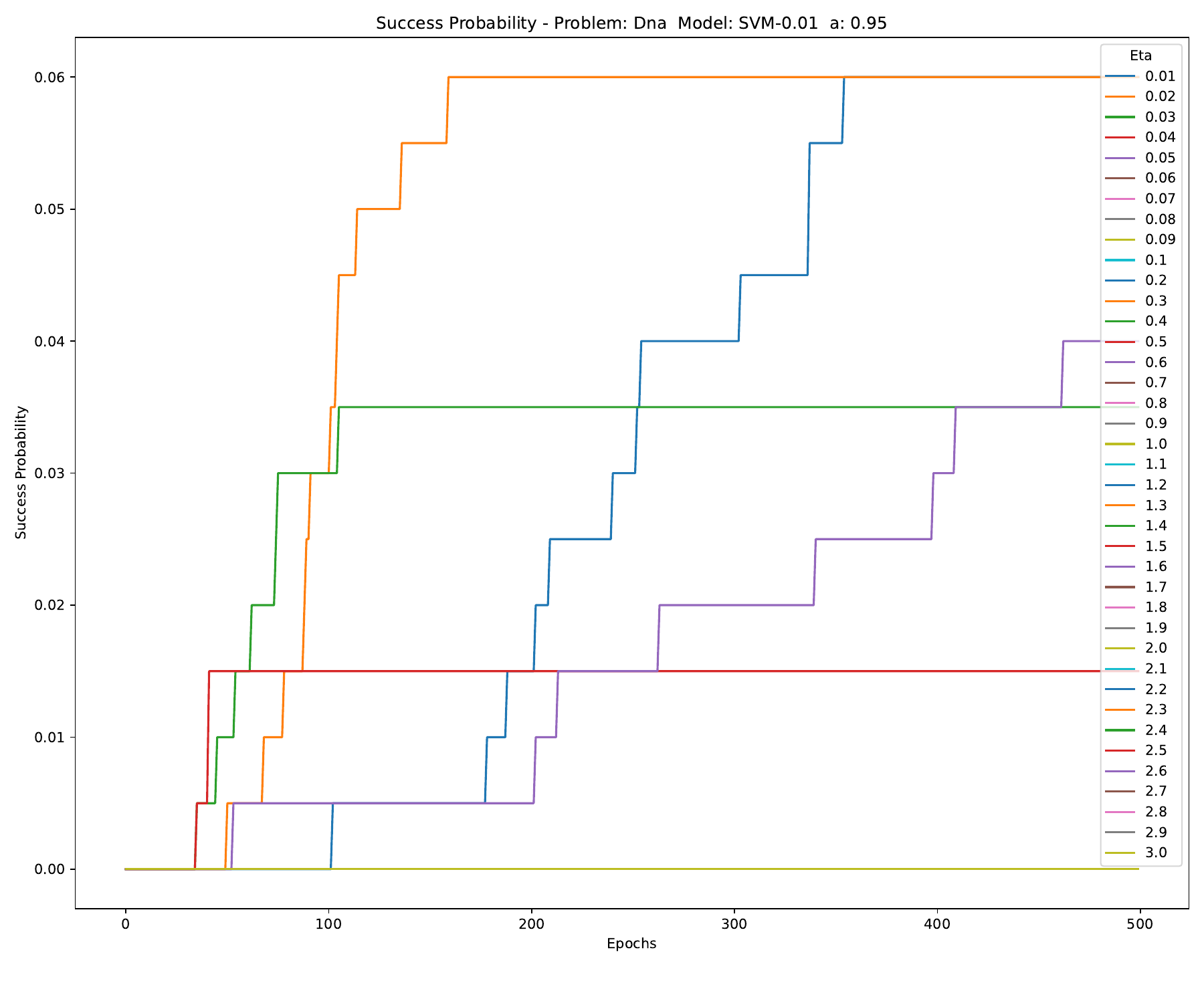} \\[-3mm]
    \includegraphics[clip,trim=0 0 0 0,width=.52\linewidth]{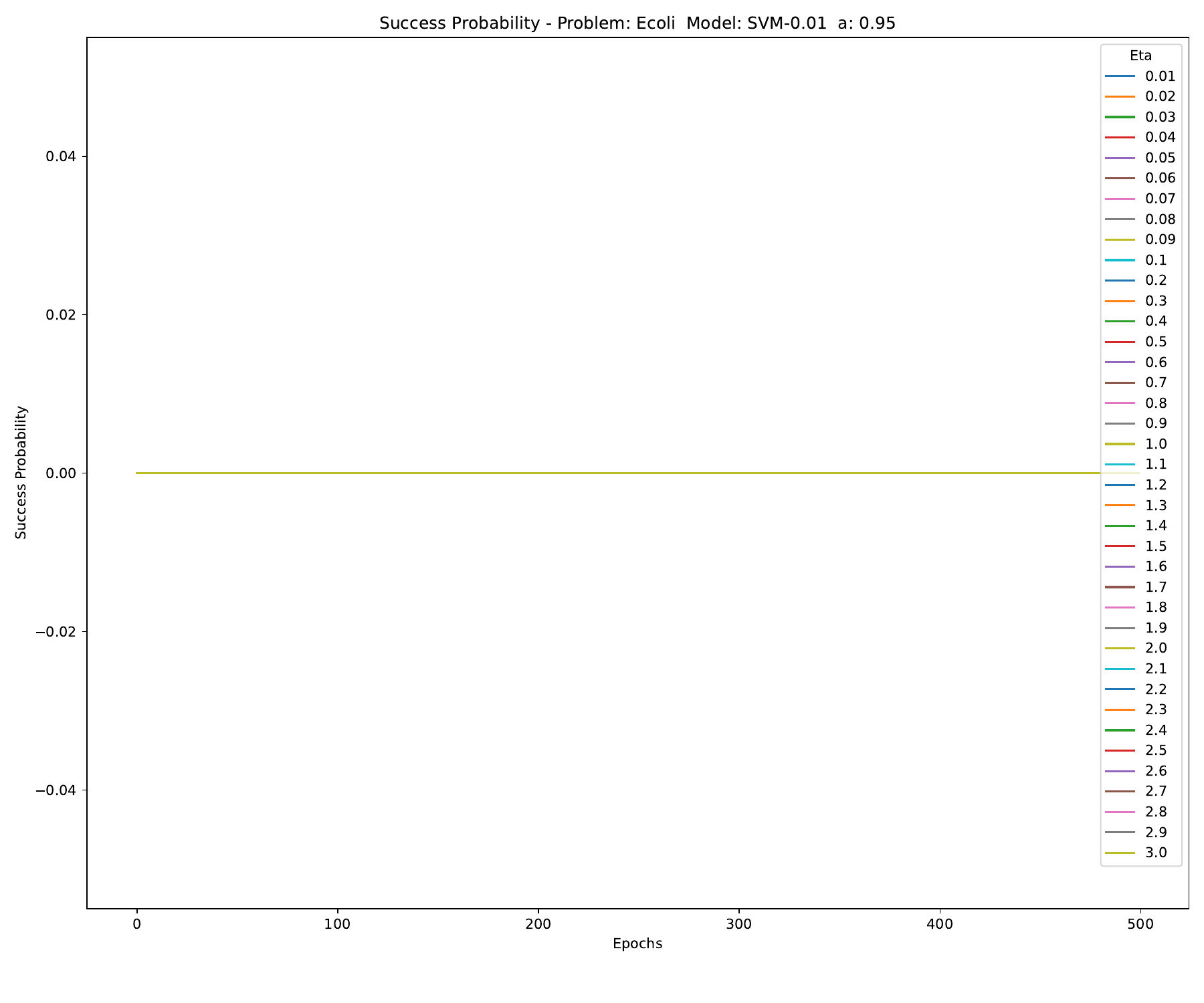} &
    \includegraphics[clip,trim=0 0 0 0,width=.52\linewidth]{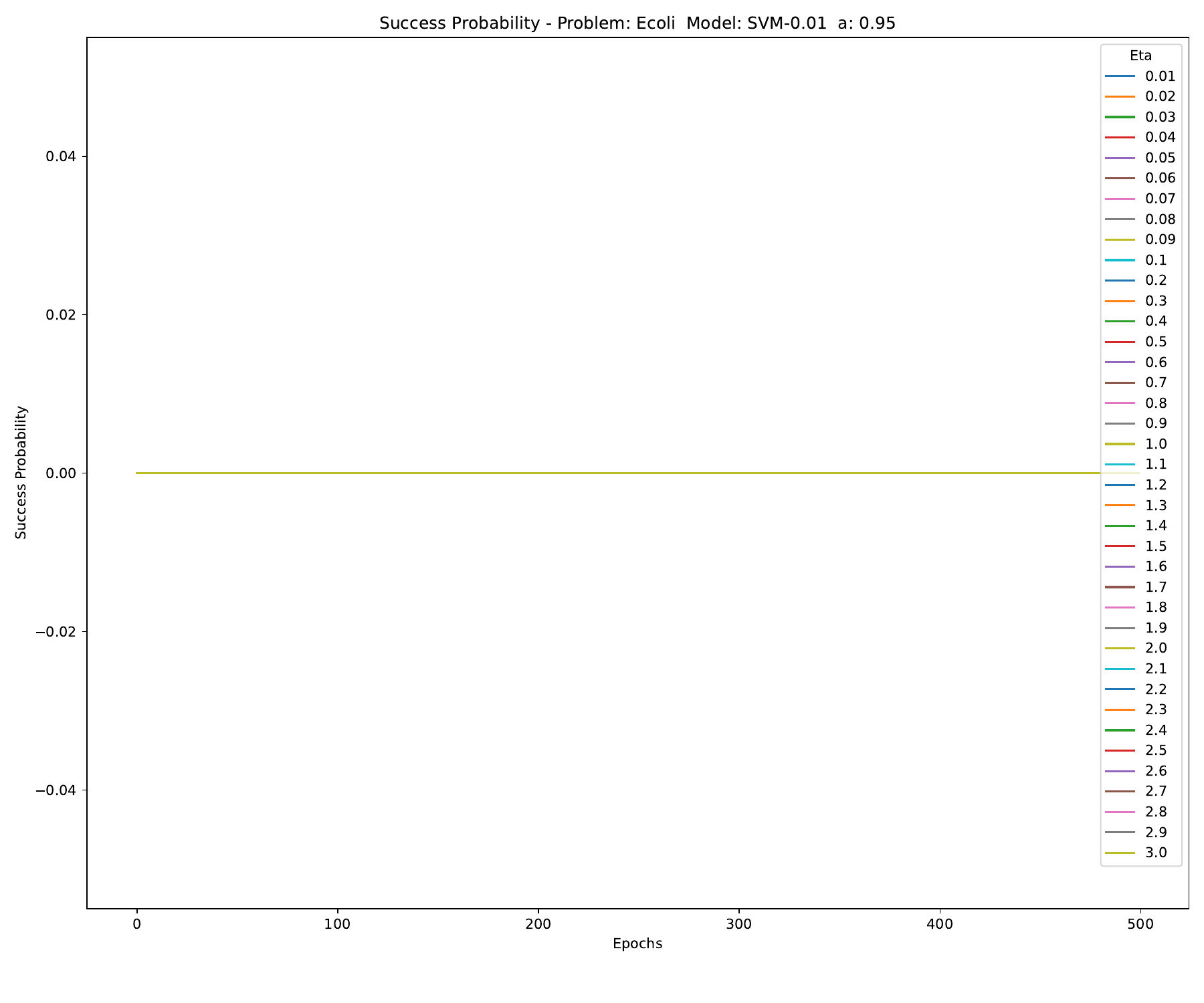} \\[-3mm]
  \end{tabular}%
\caption{Success probability ($a=0.95$) for SVM-0.01 model (continued).}
\end{figure*}

\begin{figure*}[p]
  \centering
  %
\caption{Success probability ($a=0.95$) for SVM-0.1 model (continued).}
\end{figure*}

\FloatBarrier

\pagebreak
\section{Computational Effort}
\label{app:traintest_comp}

In this section we report computational efforts as a function of
epochs numbers $i$ and learning rate $\eta$ for
$a\in\{0.85, 0.9, 0.95\}$ for all the problems and models used in this
article.  We do have the plots for all other values of $a$ between
0.75 and 1.0 (in steps of 0.005), which we would be happy to provide
upon request.

\subsection{Computational Effort for $a=0.85$}

In
Figures~\ref{fig:comp_effort_NN_ReLU_model_0.85}---\ref{fig:comp_effort_SVM_0.1_model_0.85}
we show the plots of the computational efforts as a function of epochs
numbers $i$ and learning rate $\eta$ for $a=0.85$ for all the problems
and models used in this article.

\begin{figure*}[p]
  \centering

\caption{Computational effort ($a=0.85$) for NN-ReLU model.}
  \label{fig:comp_effort_NN_ReLU_model_0.85}
\end{figure*}

\begin{figure*}[p]
  \centering
  \ContinuedFloat
  %
\caption{Computational effort ($a=0.85$) for NN-ReLU model (continued).}
\end{figure*}

\begin{figure*}[p]
  \centering
  %
\caption{Computational effort ($a=0.85$) for DNN-ReLU model.}
  \label{fig:comp_effort_DNN_ReLU_model_0.85}
\end{figure*}

\begin{figure*}[p]
  \centering
  \ContinuedFloat
  %
\caption{Computational effort ($a=0.85$) for DNN-ReLU model (continued).}
\end{figure*}

\begin{figure*}[p]
  \centering
  %
\caption{Computational effort ($a=0.85$) for LDA-Inf model.}
  \label{fig:comp_effort_LDA_Inf_model_0.85}
\end{figure*}

\begin{figure*}[p]
  \centering
  \ContinuedFloat
  %
\caption{Computational effort ($a=0.85$) for LDA-Inf model (continued).}
\end{figure*}

\begin{figure*}[p]
  \centering
  %
\caption{Computational effort ($a=0.85$) for LDA-0.01 model.}
  \label{fig:comp_effort_LDA_0.01_model_0.85}
\end{figure*}

\begin{figure*}[p]
  \centering
  \ContinuedFloat
  %
\caption{Computational effort ($a=0.85$) for LDA-0.01 model (continued).}
\end{figure*}

\begin{figure*}[p]
  \centering
  %
\caption{Computational effort ($a=0.85$) for LDA-0.1 model.}
  \label{fig:comp_effort_LDA_0.1_model_0.85}
\end{figure*}

\begin{figure*}[p]
  \centering
  \ContinuedFloat
  %
\caption{Computational effort ($a=0.85$) for LDA-0.1 model (continued).}
\end{figure*}

\begin{figure*}[p]
  \centering
  %
\caption{Computational effort ($a=0.85$) for LR-Inf model.}
  \label{fig:comp_effort_LR_Inf_model_0.85}
\end{figure*}

\begin{figure*}[p]
  \centering
  \ContinuedFloat
  %
\caption{Computational effort ($a=0.85$) for LR-Inf model (continued).}
\end{figure*}

\begin{figure*}[p]
  \centering
  %
\caption{Computational effort ($a=0.85$) for LR-0.01 model.}
  \label{fig:comp_effort_LR_0.01_model_0.85}
\end{figure*}

\begin{figure*}[p]
  \centering
  \ContinuedFloat
  %
\caption{Computational effort ($a=0.85$) for LR-0.01 model (continued).}
\end{figure*}

\begin{figure*}[p]
  \centering
  %
\caption{Computational effort ($a=0.85$) for LR-0.1 model.}
  \label{fig:comp_effort_LR_0.1_model_0.85}
\end{figure*}

\begin{figure*}[p]
  \centering
  \ContinuedFloat
  %
\caption{Computational effort ($a=0.85$) for LR-0.1 model (continued).}
\end{figure*}

\begin{figure*}[p]
  \centering
  %
\caption{Computational effort ($a=0.85$) for SVM-Inf model.}
  \label{fig:comp_effort_SVM_Inf_model_0.85}
\end{figure*}

\begin{figure*}[p]
  \centering
  \ContinuedFloat
  %
\caption{Computational effort ($a=0.85$) for SVM-Inf model (continued).}
\end{figure*}

\begin{figure*}[p]
  \centering
  %
\caption{Computational effort ($a=0.85$) for SVM-0.01 model.}
  \label{fig:comp_effort_SVM_0.01_model_0.85}
\end{figure*}

\begin{figure*}[p]
  \centering
  \ContinuedFloat
  %
\caption{Computational effort ($a=0.85$) for SVM-0.01 model (continued).}
\end{figure*}

\begin{figure*}[p]
  \centering
  %
\caption{Computational effort ($a=0.85$) for SVM-0.1 model.}
  \label{fig:comp_effort_SVM_0.1_model_0.85}
\end{figure*}

\begin{figure*}[p]
  \centering
  \ContinuedFloat
  %
\caption{Computational effort ($a=0.85$) for SVM-0.1 model (continued).}
\end{figure*}
\FloatBarrier

\subsection{Computational Effort for $a=0.9$}

In
Figures~\ref{fig:comp_effort_NN_ReLU_model_0.9}---\ref{fig:comp_effort_SVM_0.1_model_0.9}
we show the plots of the computational efforts as a function of epochs
numbers $i$ and learning rate $\eta$ for $a=0.90$ for all the problems
and models used in this article.

\begin{figure*}[p]
  \centering
  %
\caption{Computational effort ($a=0.9$) for NN-ReLU model.}
  \label{fig:comp_effort_NN_ReLU_model_0.9_duplicate}
\end{figure*}

\begin{figure*}[p]
  \centering
  \ContinuedFloat
  %
\caption{Computational effort ($a=0.9$) for NN-ReLU model (continued).}
\end{figure*}

\begin{figure*}[p]
  \centering
  %
\caption{Computational effort ($a=0.9$) for DNN-ReLU model.}
  \label{fig:comp_effort_DNN_ReLU_model_0.9}
\end{figure*}

\begin{figure*}[p]
  \centering
  \ContinuedFloat
  %
\caption{Computational effort ($a=0.9$) for DNN-ReLU model (continued).}
\end{figure*}

\begin{figure*}[p]
  \centering
  %
\caption{Computational effort ($a=0.9$) for LDA-Inf model.}
  \label{fig:comp_effort_LDA_Inf_model_0.9}
\end{figure*}

\begin{figure*}[p]
  \centering
  \ContinuedFloat
  %
\caption{Computational effort ($a=0.9$) for LDA-Inf model (continued).}
\end{figure*}

\begin{figure*}[p]
  \centering
  %
\caption{Computational effort ($a=0.9$) for LDA-0.01 model.}
  \label{fig:comp_effort_LDA_0.01_model_0.9}
\end{figure*}

\begin{figure*}[p]
  \centering
  \ContinuedFloat
  %
\caption{Computational effort ($a=0.9$) for LDA-0.01 model (continued).}
\end{figure*}

\begin{figure*}[p]
  \centering
  %
\caption{Computational effort ($a=0.9$) for LR-Inf model.}
  \label{fig:comp_effort_LR_Inf_model_0.9_duplicate}
\end{figure*}

\begin{figure*}[p]
  \centering
  \ContinuedFloat
  %
\caption{Computational effort ($a=0.9$) for LR-Inf model (continued).}
\end{figure*}

\begin{figure*}[p]
  \centering
  %
\caption{Computational effort ($a=0.9$) for LR-0.01 model.}
  \label{fig:comp_effort_LR_0.01_model_0.9}
\end{figure*}

\begin{figure*}[p]
  \centering
  \ContinuedFloat
  %
\caption{Computational effort ($a=0.9$) for LR-0.01 model (continued).}
\end{figure*}

\begin{figure*}[p]
  \centering
  %
\caption{Computational effort ($a=0.9$) for SVM-Inf model.}
  \label{fig:comp_effort_SVM_Inf_model_0.9}
\end{figure*}

\begin{figure*}[p]
  \centering
  \ContinuedFloat
  %
\caption{Computational effort ($a=0.9$) for SVM-Inf model (continued).}
\end{figure*}

\begin{figure*}[p]
  \centering
  %
\caption{Computational effort ($a=0.9$) for SVM-0.01 model.}
  \label{fig:comp_effort_SVM_0.01_model_0.9}
\end{figure*}

\begin{figure*}[p]
  \centering
  \ContinuedFloat
  %
\caption{Computational effort ($a=0.9$) for SVM-0.01 model (continued).}
\end{figure*}

\begin{figure*}[p]
  \centering
  %
\caption{Computational effort ($a=0.9$) for SVM-0.1 model (continued).}
\end{figure*}

\FloatBarrier

\subsection{Computational Effort for $a=0.95$}

In
Figures~\ref{fig:comp_effort_NN_ReLU_model_0.95}---\ref{fig:comp_effort_SVM_0.1_model_0.95}
we show the plots of the computational efforts as a function of epochs
numbers $i$ and learning rate $\eta$ for $a=0.95$ for all the problems
and models used in this article.

\begin{figure*}[p]
  \centering
  %
\caption{Computational effort ($a=0.95$) for NN-ReLU model.}
  \label{fig:comp_effort_NN_ReLU_model_0.95}
\end{figure*}

\begin{figure*}[p]
  \centering
  \ContinuedFloat
  %
\caption{Computational effort ($a=0.95$) for NN-ReLU model (continued).}
\end{figure*}

\begin{figure*}[p]
  \centering
  %
\caption{Computational effort ($a=0.95$) for DNN-ReLU model.}
  \label{fig:comp_effort_DNN_ReLU_model_0.95}
\end{figure*}

\begin{figure*}[p]
  \centering
  \ContinuedFloat
  %
\caption{Computational effort ($a=0.95$) for DNN-ReLU model (continued).}
\end{figure*}

\begin{figure*}[p]
  \centering
  %
\caption{Computational effort ($a=0.95$) for LDA-Inf model.}
  \label{fig:comp_effort_LDA_Inf_model_0.95}
\end{figure*}

\begin{figure*}[p]
  \centering
  \ContinuedFloat
  %
\caption{Computational effort ($a=0.95$) for LDA-Inf model (continued).}
\end{figure*}

\begin{figure*}[p]
  \centering
  %
\caption{Computational effort ($a=0.95$) for LR-Inf model.}
  \label{fig:comp_effort_LR_Inf_model_0.95}
\end{figure*}

\begin{figure*}[p]
  \centering
  \ContinuedFloat
  %
\caption{Computational effort ($a=0.95$) for LR-Inf model (continued).}
\end{figure*}

\begin{figure*}[p]
  \centering
  %
\caption{Computational effort ($a=0.95$) for SVM-Inf model.}
  \label{fig:comp_effort_SVM_Inf_model_0.95}
\end{figure*}

\begin{figure*}[p]
  \centering
  \ContinuedFloat
  %
\caption{Computational effort ($a=0.95$) for SVM-Inf model (continued).}
\end{figure*}

\begin{figure*}[p]
  \centering
  %
\caption{Computational effort ($a=0.95$) for SVM-0.01 model (continued).}
\end{figure*}

\begin{figure*}[p]
  \centering
  %
\caption{Computational effort ($a=0.95$) for SVM-0.1 model (continued).}
\end{figure*}

\FloatBarrier

\pagebreak
\section{Optimal Hyper-parameters}
\label{app:optim-learn-hyperp}

In this section we report minimum computational efforts and optimal
hyper-parameters as a function of the accuracy threshold  $a\in\{0.75,
0.755, 0.76, \dots, 0.99, 0.995,1\}$ for all the problems and models
used in this article.

\begin{figure*}[p]
  \centering
  %
\caption{Optimal computational effort and hyper-parameters for different values of $a$ for NN-ReLU model.}
  \label{fig:E_vs_accuracy_Pareto_NN_ReLU_model_duplicate}
\end{figure*}

\begin{figure*}[p]
  \centering
  \ContinuedFloat
  %
\caption{Optimal computational effort and hyper-parameters for different values of $a$ for NN-ReLU model (continued).}
\end{figure*}

\begin{figure*}[p]
  \centering
  %
\caption{Optimal computational effort and hyper-parameters for different values of $a$ for DNN-ReLU model.}
  \label{fig:E_vs_accuracy_Pareto_DNN_ReLU_model}
\end{figure*}

\begin{figure*}[p]
  \centering
  \ContinuedFloat
  %
\caption{Optimal computational effort and hyper-parameters for different values of $a$ for DNN-ReLU model (continued).}
\end{figure*}

\begin{figure*}[p]
  \centering
  %
\caption{Optimal computational effort and hyper-parameters for different values of $a$ for LDA-Inf model.}
  \label{fig:E_vs_accuracy_Pareto_LDA_Inf_model}
\end{figure*}

\begin{figure*}[p]
  \centering
  \ContinuedFloat
  %
\caption{Optimal computational effort and hyper-parameters for different values of $a$ for LDA-Inf model (continued).}
\end{figure*}

\begin{figure*}[p]
  \centering
  %
\caption{Optimal computational effort and hyper-parameters for different values of $a$ for LDA-0.01 model.}
  \label{fig:E_vs_accuracy_Pareto_LDA_0.01_model}
\end{figure*}

\begin{figure*}[p]
  \centering
  \ContinuedFloat
  %
\caption{Optimal computational effort and hyper-parameters for different values of $a$ for LDA-0.01 model (continued).}
\end{figure*}

\begin{figure*}[p]
  \centering
  %
\caption{Optimal computational effort and hyper-parameters for different values of $a$ for LDA-0.1 model.}
  \label{fig:E_vs_accuracy_Pareto_LDA_0.1_model}
\end{figure*}

\begin{figure*}[p]
  \centering
  \ContinuedFloat
  %
\caption{Optimal computational effort and hyper-parameters for different values of $a$ for LDA-0.1 model (continued).}
\end{figure*}

\begin{figure*}[p]
  \centering
  %
\caption{Optimal computational effort and hyper-parameters for different values of $a$ for LR-Inf model.}
  \label{fig:E_vs_accuracy_Pareto_LR_Inf_model_duplicate}
\end{figure*}

\begin{figure*}[p]
  \centering
  \ContinuedFloat
  %
\caption{Optimal computational effort and hyper-parameters for different values of $a$ for LR-Inf model (continued).}
\end{figure*}

\begin{figure*}[p]
  \centering
  %
\caption{Optimal computational effort and hyper-parameters for different values of $a$ for LR-0.01 model.}
  \label{fig:E_vs_accuracy_Pareto_LR_0.01_model}
\end{figure*}

\begin{figure*}[p]
  \centering
  \ContinuedFloat
  %
\caption{Optimal computational effort and hyper-parameters for different values of $a$ for LR-0.01 model (continued).}
\end{figure*}

\begin{figure*}[p]
  \centering
  %
\caption{Optimal computational effort and hyper-parameters for different values of $a$ for LR-0.1 model.}
  \label{fig:E_vs_accuracy_Pareto_LR_0.1_model}
\end{figure*}

\begin{figure*}[p]
  \centering
  \ContinuedFloat
  %
\caption{Optimal computational effort and hyper-parameters for different values of $a$ for LR-0.1 model (continued).}
\end{figure*}

\begin{figure*}[p]
  \centering
  %
\caption{Optimal computational effort and hyper-parameters for different values of $a$ for SVM-Inf model.}
  \label{fig:E_vs_accuracy_Pareto_SVM_Inf_model}
\end{figure*}

\begin{figure*}[p]
  \centering
  \ContinuedFloat
  %
\caption{Optimal computational effort and hyper-parameters for different values of $a$ for SVM-Inf model (continued).}
\end{figure*}

\begin{figure*}[p]
  \centering
  %
\caption{Optimal computational effort and hyper-parameters for different values of $a$ for SVM-0.01 model.}
  \label{fig:E_vs_accuracy_Pareto_SVM_0.01_model}
\end{figure*}

\begin{figure*}[p]
  \centering
  \ContinuedFloat
  %
\caption{Optimal computational effort and hyper-parameters for different values of $a$ for SVM-0.01 model (continued).}
\end{figure*}

\begin{figure*}[p]
  \centering
  %
\caption{Optimal computational effort and hyper-parameters for different values of $a$ for SVM-0.1 model.}
  \label{fig:E_vs_accuracy_Pareto_SVM_0.1_model}
\end{figure*}

\begin{figure*}[p]
  \centering
  \ContinuedFloat
  %
\caption{Optimal computational effort and hyper-parameters for different values of $a$ for SVM-0.1 model (continued).}
\end{figure*}

\FloatBarrier

\end{document}